\documentclass{amsbook}
\usepackage{amsmath,amsthm,amsfonts,amssymb,amscd}
\usepackage{stmaryrd}
\usepackage{mathrsfs,verbatim} 
\usepackage{epsfig}
\usepackage{graphics}
\usepackage{color}
\usepackage{epic,eepic}
\usepackage{xfrac}

\usepackage{natbib}

\usepackage{hyperref}

\usepackage[brazilian]{babel}
\usepackage[latin1]{inputenc}
\usepackage{times}

\hypersetup{
  colorlinks=true,       
  linkcolor=magenta,          
    filecolor=magenta,      
    urlcolor=cyan           
}

\newtheorem{theorem}{Theorem}[section]

\newtheorem{lemma}[theorem]{Lemma}
\newtheorem{teorema}[theorem]{Teorema}
\newtheorem{lema}[theorem]{Lema}
\newtheorem{corollary}[theorem]{Corollary}
\newtheorem{corolario}[theorem]{Corol\'ario}

\newtheorem{proposicao}[theorem]{Proposi\c c\~ao}
\newtheorem{conjetura}[theorem]{Conjetura}

\newtheorem{fato}[theorem]{fato}

\theoremstyle{definition}
\newtheorem{definition}[theorem]{Definition}
\newtheorem{definicao}[theorem]{Defini\c c\~ao}

\newtheorem{exemplo}[theorem]{Exemplo}

\newtheorem{exercicio}[theorem]{Exerc\'\i cio}

\theoremstyle{remark}

\newtheorem{observacao}[theorem]{Observa\c c\~ao}

\numberwithin{section}{chapter}
\numberwithin{equation}{chapter}
\numberwithin{figure}{chapter}
\numberwithin{table}{chapter}

\newcommand{\R}{{\mathbb R}}
\newcommand{\I}{{\mathbb I}}

\newcommand{\C}{{\mathbb C}}

\newcommand{\E}{{\mathbb E}}
\newcommand{\K}{{\mathbb K}}
\newcommand{\F}{{\mathcal{F}}}
\newcommand{\Z}{{\mathbb Z}}
\newcommand{\VC}{{\mbox{VC-dim}}}
\newcommand{\s}{\mathbb{S}}
\newcommand{\N}{{\mathbb N}}

\newcommand{\im}{{\mathbf i}}
\newcommand{\Q}{{\mathbb Q}}
\def\bij{{\rightarrowtail\!\!\!\!\!\to \,}}

\newcommand{\T}{{\mathbb T}}

\newcommand{\e}{{\epsilon}}
\newcommand{\conv}{{\mathrm{conv}\,}}

\newcommand{\Int}{{\mathrm{Int}\,}}
\newcommand{\var}{{\mathrm{var}\,}}

\def\ls #1#2{{[#1]^{\leq#2~0|1}}}

\def\diam{{\mathrm{diam}\,}}

\def\obs{{\mbox{obs-diam}\,}}
\newcommand{\supp}{{\mathrm{supp}\,}}
\newcommand{\ve}{{\varepsilon}}
\newcommand{\tri}{\hfill$\blacktriangle$} 
\def\norm #1{{\left\Vert\,#1\,\right\Vert}}
\newcommand{\abs}[1]{\lvert#1\rvert}
\def\fr(#1/#2){{^{\mbox{$_#1$}}\!/_{#2}}}
\font\mathf=cmex10
\def\va{\hbox{\mathf\char'76}}
\def\pipe{\hbox{${\raise.28em\va\atop\raise.50em\va}$}}

\makeatletter
\def\Ddots{\mathinner{\mkern1mu\raise\p@
\vbox{\kern7\p@\hbox{.}}\mkern2mu
\raise4\p@\hbox{.}\mkern2mu\raise7\p@\hbox{.}\mkern1mu}}
\makeatother

\makeindex

\begin{document}

\title{Elementos da teoria de aprendizagem de m\'aquina supervisionada}

\author{Vladimir Pestov}

\address{Instituto de Matem\'atica e Estat\'\i stica, Universidade Federal da Bahia, Ondina,  Salvador, BA, 40.170-115, Brasil}
\address{Departamento de Matem\'atica, Universidade Federal de Santa Catarina, Trindade, Florian\'opolis, SC, 88.040-900, Brasil}
 \address{Department of Mathematics and Statistics,
 University of Ottawa, Ottawa, Ontario, K1N 6N5, Canada}

\email{vpestov2010@gmail.com}
\thanks{\textcopyright IMPA, 2019}

\subjclass{68Q32; 62H30, 68T05, 68T10}


\maketitle
\frontmatter

\setcounter{page}{3}

\tableofcontents

%
%

\chapter*{Pref\'acio} 

No Outono do Hemisf\'erio Norte de 2007, o autor decidiu educar-se sobre o assunto de aprendizagem de m\'aquina e ofereceu-se para ministrar um curso de p\'os-gradua\c c\~ao na Universidade de Ottawa, Canad\'a, baseado inicialmente nas notas \citep*{mendelson} e no livro \citep*{AB}. O curso foi oferecido novamente em 2012. Com meus alunos de pesquisa nesta \'area, n\'os n\~ao nos limitamos ao estudo te\'orico, mas projetamos novos algoritmos e participamos numa s\'erie de competi\c c\~oes internacionais de minera\c c\~ao de dados, ganhando uma delas em 2013. O nosso modesto semin\'ario de Data Science and Machine Learning cresceu em um verdadeiro grupo de pesquisa na Universidade de Ottawa.
Depois de me mudar para o Brasil em 2016, eu ministrei o curso duas vezes na UFSC, Florian\'opolis, em 2017 e 2018, bem como iniciei um semin\'ario, Grupo de Estudos em Aprendizagem de M\'aquina (GEAM), na ilha de Santa Catarina. Essas notas de aula, ainda bastante informais, resultam destes cursos e semin\'arios.

As notas s\~ao dirigidas principalmente para os matematicamente inclinados. Ao mesmo tempo, elas s\~ao quase autosuficientes, gra\c cas a um extenso conjunto de ap\^endices. \'E claro que n\~ao \'e realista esperar por um tratamento abrangente de uma \'area t\~ao vasta; as notas cobrem uma sele\c c\~ao de assuntos altamente subjetiva. No entanto, acredito que elas permitem aprender algumas ideias importantes. Eu evitei propositadamente qualquer tentativa de descrever ``aplica\c c\~oes pr\'aticas'': nesta \'area do conhecimento n\~ao h\'a necessidade de faz\^e-lo, dada a curta dist\^ancia que separa teoria, algoritmos, e implementa\c c\~ao.

Esta \'e uma boa ocasi\~ao para reconhecer o qu\~ao importante para mim foi a intera\c c\~ao nesta \'area com os meus estudantes de pesquisa ao longo dos anos. Quero mencionar de nome Aleksandar Stojmirovi\'c, Christian Despres, Damjan Kalajdzievski, \'Emilie Id\`ene, Ga\"el Giordano, Hubert Duan, Igor Artemenko, Ilya Volnyansky, Robert Davies, Sabrina Sixta, Samuel Buteau, Stan Hatko, Sushma Kumari, Varun Singla, Yue Dong.

Sem d\'uvida, o texto est\'a infestado de erros e lacunas, tanto matem\'aticos quanto os da express\~ao portuguesa. Meus agradecimentos v\~ao para os colegas, amigos, e alunos brasileiros que me ajudaram a reduzir o n\'umero destes \'ultimos: Aishameriane Schmidt, Aldrovando Luis Azeredo Ara\'ujo, Brittany Beauchamp e Pedro Levit Kaufmann, Maria Inez Cardoso Gon\c calves, Nat\~a Machado, e Ricardo Borsoi, assim como o editor do IMPA, Paulo Ney de Souza.

Gostaria de exprimir a minha gratid\~ao pelo apoio do CNPq (bolsa Pesquisador Visitante, N\'\i vel 1, Out 2016---Set 2017, processo 310012/2016), da CAPES (bolsa Professor Visitante Estrangeiro S\^enior, ao longo do ano 2018, processo 88881.117018/2016-01), e da UFBA, onde sou  Professor Visitante Titular desde Mar\c co 2019.
\vskip 3pt

{\em Feci quod potui, faciant meliora potentes.}

\vskip 10pt
\rightline{Vladimir Pestov}
\rightline{Rio Vermelho, Salvador, Bahia}
\rightline{5 de Outubro 2019}

%
%

\chapter*{Tr\^es cita\c c\~oes orientadoras} 
 \begin{minipage}[b]{3.5cm}
    \scalebox{0.23}[0.23]{\includegraphics{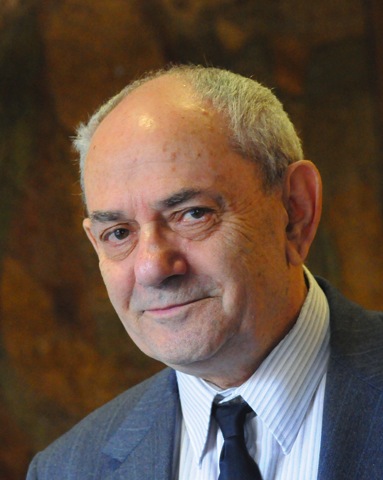}} 
\\
{\sc Vladimir Vapnik}
\end{minipage} \hskip .2cm
\begin{minipage}[t]{9cm}
  \vskip -3.8cm
  ``Statistical learning theory does not belong to any specific branch of science: It has its own goals, its own paradigm, and its own techniques.
  
Statisticians (who have their own paradigm) never considered this theory as part of statistics''. \hfill \citep*{vapnik2}
\end{minipage}

  \vskip 1cm

\begin{minipage}[t]{9cm}
  \vskip -4.8cm
  "As the power of computers approaches the theoretical limit and as we turn
to more realistic (and thus more complicated) problems, we face 'the curse
of dimension', which stands in the way of successful implementations of
numerics in science and engineering. Here one needs a much higher level of
mathematical sophistication in computer architechture as well as in
computer programming. ...Successes here may provide theoretical means for
performing computations with high power growing arrays of data. ...We
shall need ... the creation of a new breed of mathematical professionals
able to mediate between pure mathematics and applied science. The
cross-fertilization of ideas is crucial for the health of the science and
mathematics." \hfill \citep*{gromov_trends}
\end{minipage}
\hskip.8cm
\begin{minipage}[b]{3cm}
  \scalebox{0.32}[0.32]{\includegraphics{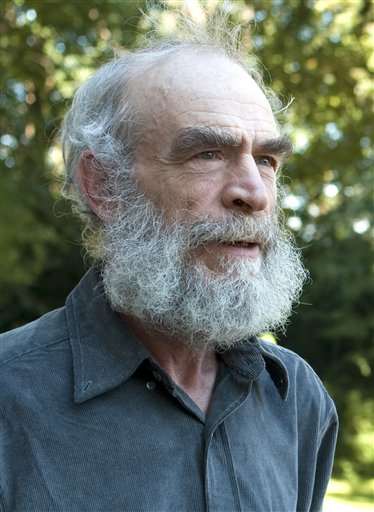}} 
\\
{\sc Mikha\"\i l Gromov}
\end{minipage}
\vskip 1cm

\begin{minipage}[b]{3.5cm}
    \scalebox{0.23}[0.23]{\includegraphics{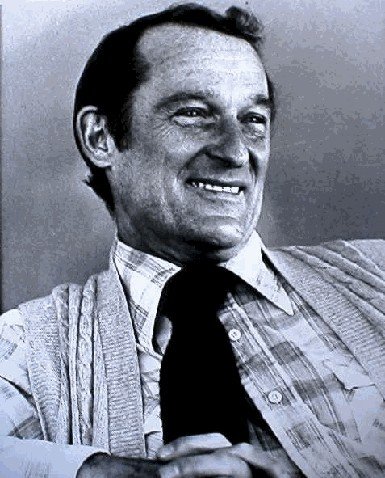}} 
\\
{\sc Seymour Cray}
\end{minipage} \hskip .2cm
\begin{minipage}[t]{9cm}
  \vskip -3.8cm
  "One of my guiding principles is don't do anything that other people are doing." 
\hfill (\footnote{Computer History Museum, Mountain View, CA, permanent exposition.})
\end{minipage}



\mainmatter
\setcounter{chapter}{0}
%
%

\chapter*{Introdu\c c\~ao}

\begin{figure}[ht]
\begin{center}
\scalebox{0.33}{\includegraphics{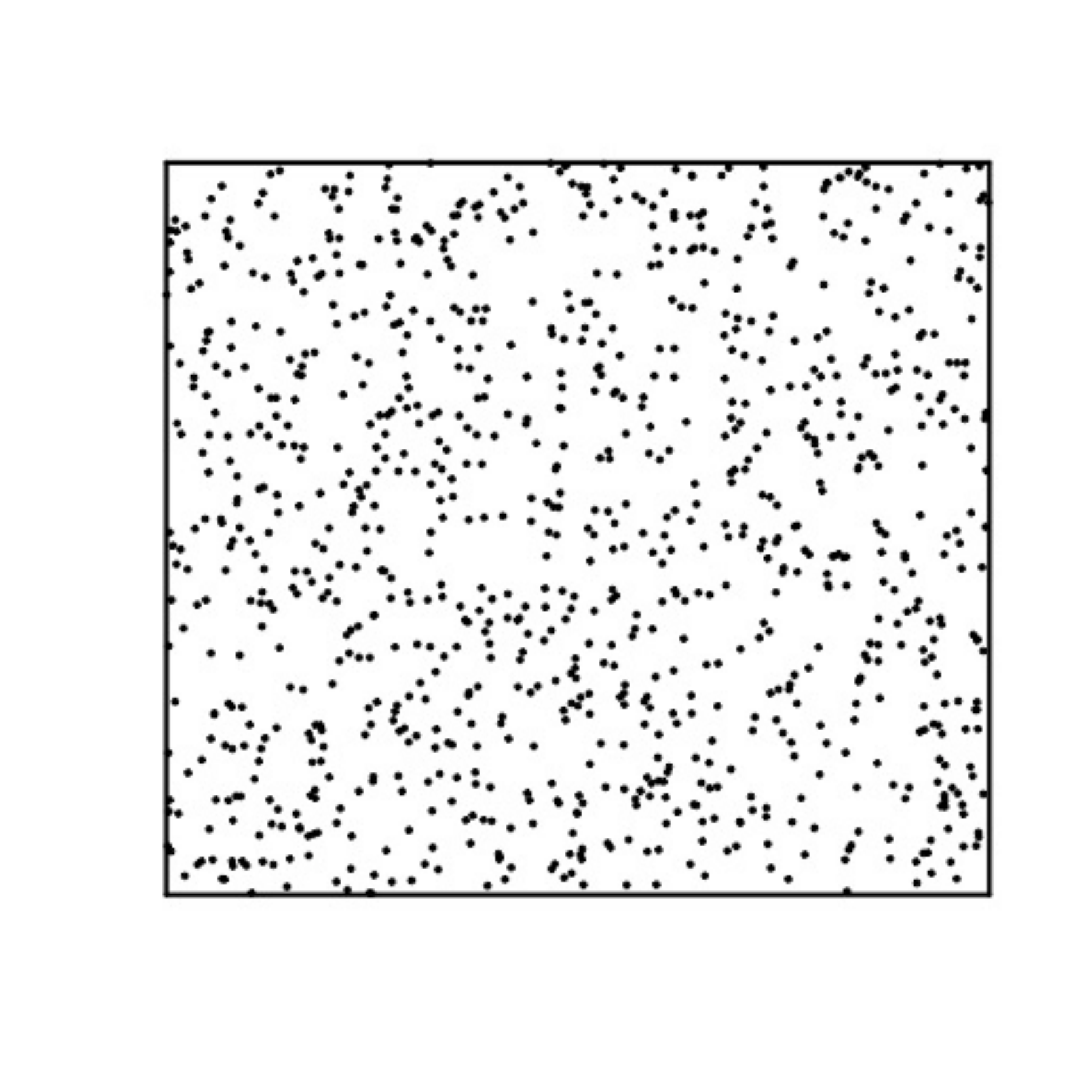}} 
\scalebox{0.33}{\includegraphics{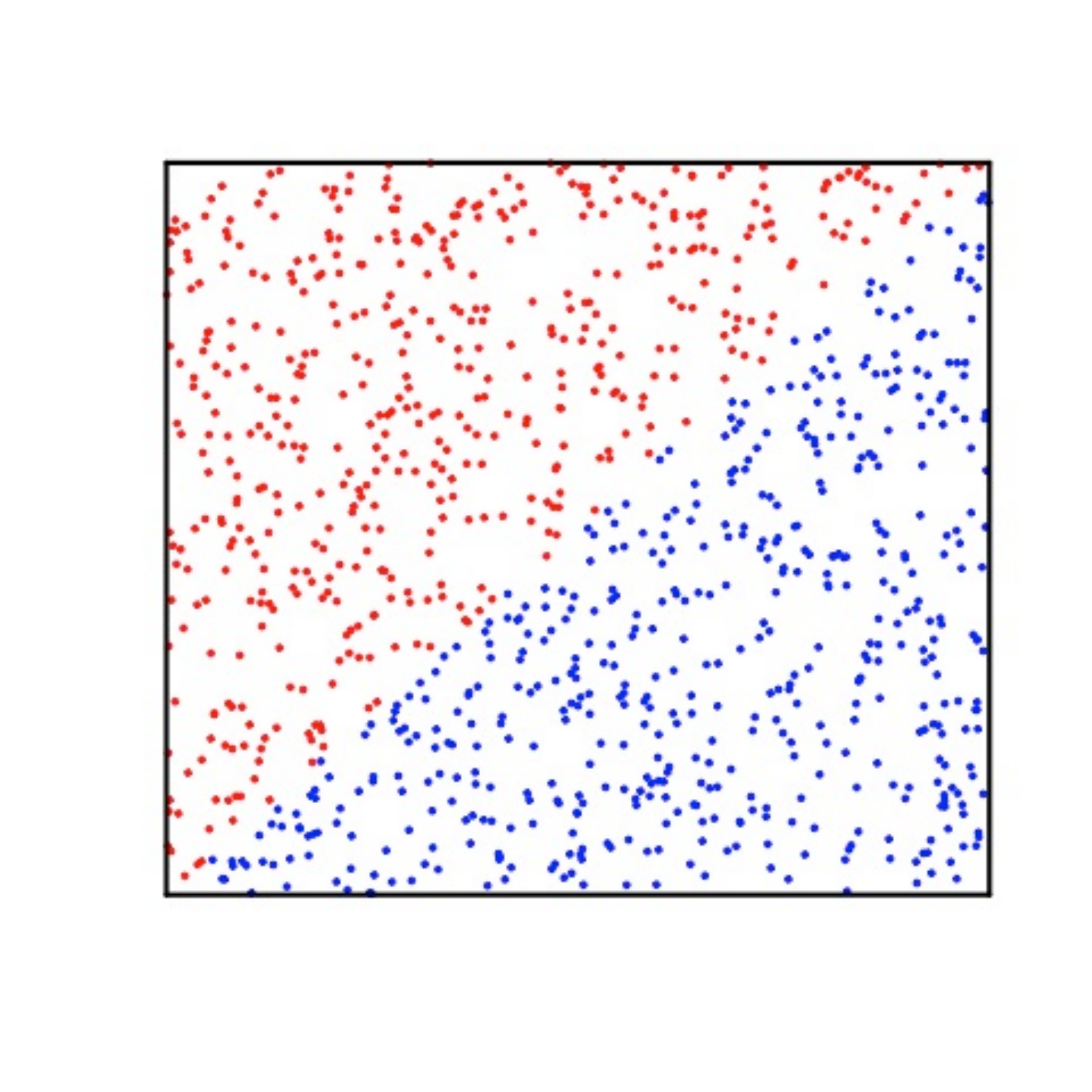}} 
\caption{Amostra aleat\'oria de $1000$ pontos tirados uniformemente do quadrado, n\~ao rotulada (esquerda) e rotulada (dir.)}
\label{fig:1000random_pts_unlabelled}
\end{center}
\end{figure}

Vamos come\c car pela no\c c\~ao b\'asica da {\em aprendizagem supervisionada}: o {\em problema de classifica\c c\~ao bin\'aria}. Para tanto, tomemos uma experi\^encia simples. Geremos $n=1000$ pontos aleat\'orios no quadrado unit\'ario $[0,1]^2$, distribu\'\i dos uniformemente e independentemente um do outro. (A {\em distribui\c c\~ao uniforme} significa que a probabilidade de que um ponto $x$ perten\c ca a um pequeno quadrado $[a,a+\e]\times [b,b+\e]$ de lado $\e>0$ \'e proporcional -- com efeito, igual -- \`a \'area do quadrado, $\e^2$.) 
O que conjunto de dados resultante pode parecer, claro, n\~ao \'e uma grade uniforme, mas em vez disso, algo assim (Figura \ref{fig:1000random_pts_unlabelled}, esquerda).

Note, em particular, a presen\c ca do que aparece como uma estrutura interna de dados significativa: os grandes buracos aqui e ali, agrupamentos de pontos... Estes s\~ao, na verdade, desvios aleat\'orios, n\~ao carregando nenhuma informa\c c\~ao \'util.

O nosso conjunto de dados, 
\[X = \{x_1,x_2,\ldots,x_{1000}\},\]
\'e uma {\em amostra}. O quadrado $[0,1]^2$ \'e o {\em dom\'\i nio.}

Agora dividimos os dados em duas classes: a classe $A$ dos pontos sobre ou acima da diagonal (marcados com vermelho) e a classe $B$ dos pontos abaixo da diagonal (marcados com azul). Obtemos o que \'e chamado uma {\em amostra rotulada} ({\em labelled sample}). Ver Figura \ref{fig:1000random_pts_unlabelled}, (dir.)

Observe um efeito visual interessante: parece que a fronteira entre as duas classes \'e uma curva ondulada, ao inv\'es de uma linha reta! No entanto, voc\^e pode usar uma r\'egua para convencer-se de que n\~ao h\'a nenhum erro e os centros de todas as bolinhas vermelhas est\~ao realmente acima da diagonal, enquanto os centros das azuis est\~ao abaixo.

Marcando os elementos de $A$ com $1$ e os elementos de $B$ com $0$, a nossa amostra rotulada pode ser escrita da seguinte maneira:
\[\sigma =(x_1,x_2,\dots,x_{1000},\ve_1,\e_2\,\ldots,\ve_{1000}),\]
onde por valor $\ve_i$ do r\'otulo do ponto $x_i$, temos $\ve_i\in\{0,1\}$, $i=1,2,\ldots,1000$. (Certo, ao inv\'es dos r\'otulos $0$ e $1$ se pode usar, por exemplo, $-1$ e $+1$, ou bem $-$ e $+$...)

Neste exemplo ``de brinquedo'' a dimens\~ao dos dados \'e $2$, e o conjunto de dados pode ser visualizado, o que ajuda muito para determinar a sua estrutura. Cada ponto \'e um elemento de $\R^2$, representado por duas coordenadas, $x_i=(x_i^{(1)},x_i^{(2)})$. A amostra rotulada $\sigma$ pode ser tratada como um subconjunto (ordenado) de $[0,1]\times [0,1]\times \{0,1\}$, e escrita como uma matriz de dimens\~ao $1,000\times 3$: cada linha $(x_i^{(1)},x_i^{(2)},\ve_i)$ representa um elemento de $X$, bem como seu r\'otulo. 
Essa representa\c c\~ao matricial dos conjuntos de dados \'e bastante comum. De uma maneira mais abstrata, podemos escrever
\[\sigma\in ([0,1]\times [0,1])^n\times \{0,1\}^n.\]

Chegamos ao seguinte {\em problema de classifica\c c\~ao bin\'aria}: a partir da amostra rotulada $\sigma$, construir uma fun\c c\~ao
\[T\colon [0,1]^2\to\{0,1\}\]
(chamada {\em classificador,} {\em preditor,} ou {\em fun\c c\~ao de transfer\^encia}),
definida sobre todo o dom\'\i nio, que seja capaz de predizer com confian\c ca um r\'otulo n\~ao s\'o para os dados existentes, mas tamb\'em para novos dados.
Pode-se dizer que esse \'e o problema central da aprendizagem autom\'atica  estat\'\i stica {\em supervisionada.}
 
Claro que sabemos a resposta para nosso ``problema de brinquedo'': ela \'e dada pelo classificador de verdade
\[T_{true}(x)=\eta(x^{(2)}-x^{(1)}),\]
onde $\eta$ \'e a {\em fun\c c\~ao de Heaviside,}
\[\eta(x)=\left\{\begin{array}{cl}1,&\mbox{ se }x\geq 0,\\
0,&\mbox{ se }x<0.\end{array}\right.\]

Mas se o problema for mostrado a alguma outra pessoa (ou m\'aquina) que n\~ao sabe como as duas classes $A$ e $B$ foram formadas, voc\^e pode obter outras respostas.
Por exemplo, o seu pr\'oprio c\'ortex visual, ao analisar a imagem na figura \ref{fig:1000random_pts_unlabelled} (dir.), sugere separar as duas classes com uma linha ondulada!
Um tal classificador poderia n\~ao ser exato, mas estar perto da verdade para ser aceit\'avel. As chances de {\em classifica\c c\~ao err\^onea} (o {\em erro de classifica\c c\~ao}) para um novo ponto de dados seriam relativamente pequenos.

Algu\'em pode sugerir a seguinte solu\c c\~ao simplista: atribuir o valor $1$ a todos os pontos de dados atuais que est\~ao acima da diagonal, e o valor $0$ a todos os outros pontos, atuais e futuros:
\[T(x)=\left\{\begin{array}{cl}1,&\mbox{ se }x\in A,\\
0,&\mbox{ sen\~ao.}\end{array}\right.\]
Este classificador d\'a uma resposta correta para todos os pontos atuais $x_i\in X$, $i=1,2,\ldots,1,000$. 
No entanto, se n\'os gerarmos aleatoriamente um novo ponto $y\in [0,1]^2$, com probabilidade $1/2$ ele ficar\'a acima da diagonal. Ao mesmo tempo, a probabilidade de escolher um ponto em $X$ \'e zero. Assim, com probabilidade de $1/2$, o classificador $T$ ir\'a retornar um valor falso para $y$. Entre todos os pontos gerados no futuro,
\[x_{1001},x_{1002},\ldots,x_{1000+n},\ldots,\]
aproximadamente metade deles ser\~ao classificados erroneamente. Para $n$ suficientemente grande, o classificador $T$ fornecer\'a uma resposta errada aproximadamente na metade dos casos --- certamente um fracasso completo. Jogando a moeda equilibrada podemos conseguir a mesma taxa de sucesso de $1/2$, sem usar qualquer classificador, simplesmente atribuindo a um ponto um valor aleat\'orio $0$ ou $1$.

Como podemos distinguir um bom classificador de um ruim? Ou seja, dado um classificador, $T$, existe uma maneira de verificar se $T$ \'e suscet\'\i vel de atribuir a {\em maioria} dos pontos de dados futuros \`a classe correta?

\`A primeira vista, o problema parece completamente intrat\'avel: como possivelmente podemos mostrar algo sobre os dados que ainda n\~ao existem? Na verdade, \'e quase incr\'\i vel que -- pelo menos dentro de um  modelo te\'orico -- tais predi\c c\~oes podem ser feitas com um grau consider\'avel de certeza.

Todavia, vamos deixar este problema para mais tarde. 
Consideremos um exemplo real: um conjunto de dados da competi\c c\~ao CDMC'2013 de minera\c c\~ao de dados para o problema de dete\c c\~ao de intrusos numa rede, coletados por um sistema real IDS (Intrusion Detection System) na Coreia. (Para mais informa\c c\~oes, consulte \citep*{STK}. Este conjunto n\~ao est\'a dispon\'\i vel publicamente, mas outros conjuntos semelhantes est\~ao, por exemplo\footnote{DARPA Intrusion Detection Data Sets, MIT Lincoln Lab, \href{https://www.ll.mit.edu/r-d/datasets}
{https://www.ll.mit.edu/r-d/datasets}}.)
  
Cada linha da matriz corresponde a uma sess\~ao, onde as $7$ coordenadas s\~ao os valores dos par\^ametros da sess\~ao.
O conjunto de dados cont\'em $n=77,959$ elementos, incluindo $71,758$ sess\~oes normais (sem intruso), rotuladas $+1$, e $6,201$ sess\~oes ataque (com intruso), rotuladas $-1$. 
A Figure \ref{fig:15linhas} mostra um extrato das $15$ linhas da matriz.
  
\begin{figure}[ht]
\begin{center}
{\small
\begin{verbatim}
                    ............
+1 1:-1.00 2:-0.03 3:-0.09 4:-0.49 5:-0.05 6:-0.15 7:-1.08
+1 1:-1.00 2:-0.03 3:-0.09 4:-0.49 5:-0.05 6:-0.15 7:-1.08
+1 1:-1.00 2:-0.03 3:-0.09 4:-0.49 5:-0.05 6:-0.15 7:-1.08
+1 1:-1.00 2:-0.03 3:-0.09 4:-0.49 5:-0.05 6:-0.15 7:-1.08
+1 1:-0.67 2:-0.03 3:0.04  4:1.95  5:-0.05 6:-0.10 7:1.11
+1 1:-1.00 2:-0.03 3:-0.09 4:-0.49 5:-0.05 6:-0.15 7:-1.08
+1 1:-1.00 2:-0.03 3:-0.09 4:-0.49 5:-0.05 6:-0.15 7:-1.08
+1 1:-0.63 2:-0.03 3:0.03  4:1.89  5:-0.05 6:-0.10 7:1.11
+1 1:-0.59 2:-0.03 3:0.03  4:1.83  5:-0.05 6:-0.09 7:1.11
-1 1:-1.00 2:-0.03 3:-0.09 4:-0.49 5:-0.05 6:-0.15 7:-1.08
+1 1:-1.00 2:-0.03 3:-0.09 4:-0.49 5:-0.05 6:-0.15 7:-1.08
+1 1:-1.00 2:-0.03 3:-0.09 4:-0.49 5:-0.05 6:-0.15 7:-1.08
+1 1:-1.00 2:-0.03 3:-0.09 4:-0.49 5:-0.05 6:-0.15 7:-1.08
-1 1:-1.00 2:-0.03 3:-0.09 4:-0.49 5:-0.05 6:-0.15 7:-1.08
+1 1:1.09  2:-0.03 3:-0.02 4:-0.49 5:-0.05 6:-0.15 7:1.11
                    ............
\end{verbatim}
}
\caption{Fragmento do conjunto de dados para dete\c c\~ao de intrusos na rede.}
\label{fig:15linhas}
\end{center}
\end{figure}

O objetivo \'e de construir um classificador capaz de alertar de um intruso em tempo real com um erro m\'\i nimo e uma confian\c ca alta.
O que seria o classificador mais natural de se usar, baseado em nossa experi\^encia cotidiana e o senso comum?

Suponha que voc\^e queira vender seu carro. Para determinar um pre\c co razo\'avel, voc\^e vai buscar algumas informa\c c\~oes sobre a venda dos carros do mesmo modelo, idade, milhagem, at\'e a cor. Em outras palavras, voc\^e busca um carro o mais semelhante ao seu, e a sua cota\c c\~ao de venda d\'a uma boa ideia do pre\c co a escolher. 

\'E exatamente como o {\em classificador de vizinhos mais pr\'oximos}, ou o {\em classificador NN} (Nearest Neighbour Classifier) funciona. Dado um ponto qualquer $y$ do dom\'\i nio, $\Omega$, buscamos o ponto $x$ do conjunto de dados atual, $X$, mais pr\'oximo a $y$. O classificador $NN$ atribui a $y$ o mesmo r\'otulo que o r\'otulo de $x$. Obviamente, a fim de determinar o vizinho mais pr\'oximo, precisamos de uma fun\c c\~ao de semelhan\c ca qualquer sobre o dom\'\i nio:
\[S\colon \Omega\times\Omega \to\R.\]
Tipicamente, $S$ \'e uma m\'etrica, por exemplo, a m\'etrica euclidiana. 

Voltando \`a venda do carro, provavelmente \'e mais razo\'avel buscar mais de um carro semelhante ao seu, e determinar o pre\c co baseado sobre uma variedade dos pre\c cos destes carros. Obtemos o {\em classificador de $k$ vizinhos mais pr\'oximos}, ou {\em classificador $k$-NN}, onde $k$ \'e um n\'umero fixo. Dada a amostra rotulada, 
\[\sigma=(x_1,x_2,\ldots,x_n,\ve_1,\ve_2,\ldots,\ve_n)\in\Omega^n\times\{0,1\}^n,\]
e o ponto da entrada $y\in\Omega$, o classificador $k$-NN escolhe $k$ vizinhos mais pr\'oximos a $y$, $x_{i_1},x_{i_2},\ldots,x_{i_k}\in X$, e determina o r\'otulo de $y$ pelo voto majorit\'ario entre os r\'otulos $\ve_{i_1},\ve_{i_2},\ldots,\ve_{i_k}$. Se a vota\c c\~ao for indecisa (o que \'e poss\'\i vel se $k$ for par), o r\'otulo de $y$ \'e escolhido aleatoriamente.

\begin{figure}
  
  \begin{center}
  \scalebox{0.4}[0.4]{\includegraphics{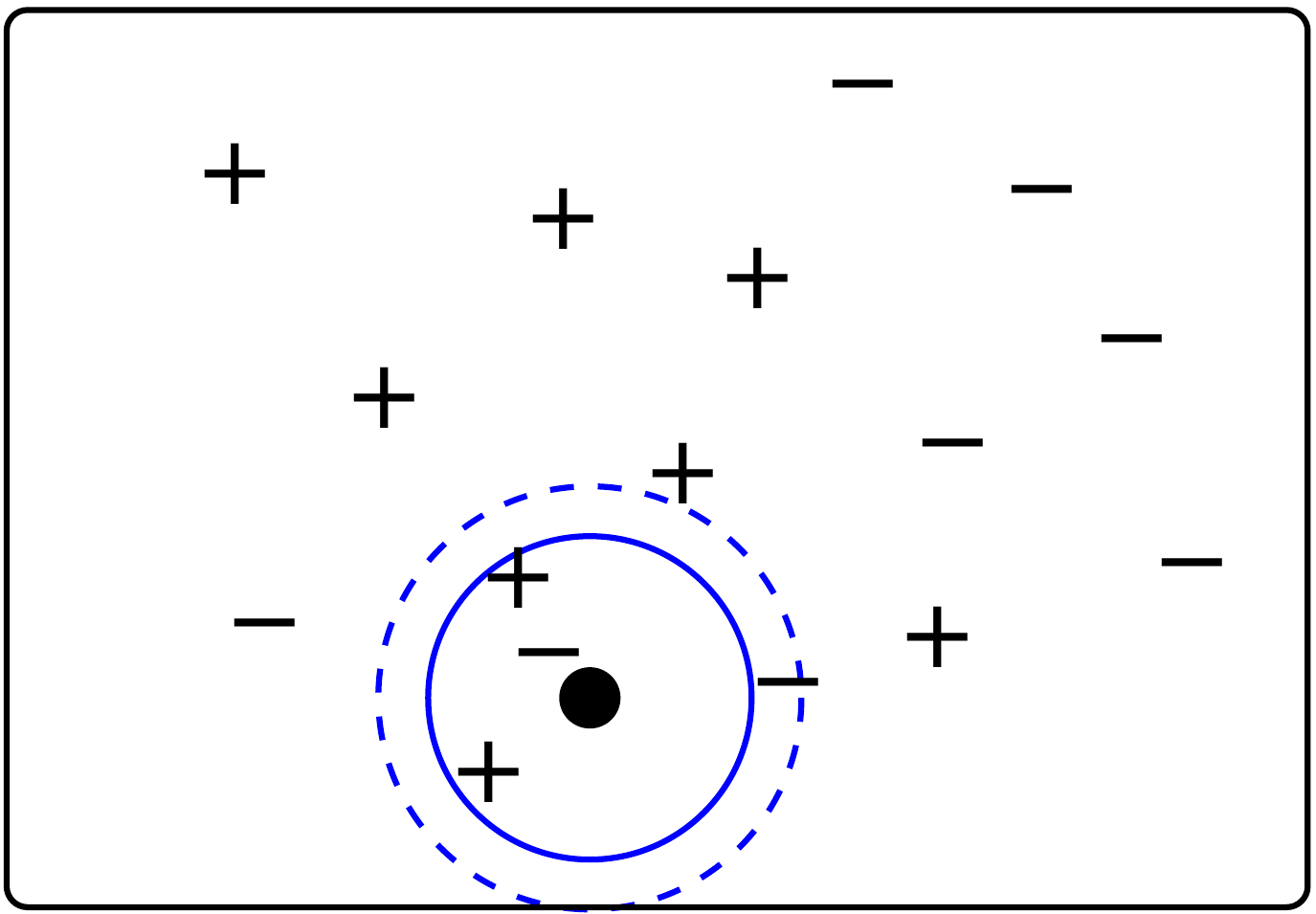}} 
  \caption{O voto majorit\'ario para $k=3$, entre $+,+,-$, retorna $+$, a para $k=4$, entre $+,+,-,-$, \'e indeciso.}
\end{center} 
\end{figure}

Como podemos garantir que as previs\~oes dadas por um classificador s\~ao confi\'aveis? Na pr\'atica, a t\'ecnica comumente usada \'e a {\em valida\c c\~ao cruzada}. O conjunto de dados \'e dividido aleatoriamente no {\em conjunto de treinamento} (tipicamente, 75 a 90 por cento dos pontos) e o conjunto de avalia\c c\~ao (o restante 10 a 25 por cento). Somente os dados de treinamento s\~ao usados pelo algoritmo, e os dados de avalia\c c\~ao s\~ao aplicados para estimar o erro de predi\c c\~ao.

Denotemos $T\colon\Omega\to\{0,1\}$ o classificador, $X_t$ o conjunto de treinamento, e $X_a$ o conjunto de avalia\c c\~ao:
\[X = X_t\cup X_a,~~X_t\cap X_a=\emptyset.\]
O valor seguinte \'e o estimador estat\'\i stico do erro de predi\c c\~ao (ou: erro de classifica\c c\~ao) de $T$:
\[\frac{\left\vert\{i\colon x_i\in X_a,~ T(x_i)\neq \ve_i\} \right\vert}{\abs{X_a}}.\]

O procedimento \'e iterado muitas vezes, e o valor m\'edio dos erros cada vez estimados serve como uma boa aproxima\c c\~ao ao valor do erro verdadeiro de classifica\c c\~ao de $T$.

Para aplicar classificadores aos conjuntos de dados concretos, \'e preciso escolher uma linguagem de programa\c c\~ao. Teoricamente, qualquer linguagem pode ser utilizada: todas s\~ao equivalentes \`a m\'aquina de Turing! Uma das linguagens utilizada mais comunmente em ci\^encia de dados 
\'e R, a linguagem de programa\c c\~ao estat\'\i stica, criada no Departamento de Estat\'\i stica da Universidade de Auckland, Nova Zel\^andia e baseado em software livre (no formato do projeto GNU)\footnote{The R project for statistical computing, \href{http://www.r-project.org/}{http://www.r-project.org/}}. A fonte mais abrangente com informa\c c\~oes sobre R, {\em The R Book,} \'e dispon\'\i vel livremente na web \citep*{C}.

Sugere-se baixar a linguagem R seguindo as instru\c c\~oes de qualquer uma das fontes mencionadas acima (\citep*{O} ou \citep*{M}), e experimentar com ela,
come\c cando com os exerc\'\i cios destes livros. 
Existem muitas implementa\c c\~oes dispon\'\i veis do classificador $k$-NN em R, por exemplo, o classificador IBk do pacote RWeka, ou o do pacote FNN (Fast Nearest Neighbor Search Algorithms and Applications).

\begin{exercicio}
Baixar algum conjunto de dados de Stanford\footnote{\href{https://web.stanford.edu/~hastie/ElemStatLearn//datasets/}
{https://web.stanford.edu/~hastie/ElemStatLearn//datasets/}}, por exemplo {\em Phoneme}, e treinar o classificador $k$-NN em R.
\end{exercicio}
\label{pagina:phoneme}

Aplicando o classificador $k$-NN ao nosso conjunto de dados para dete\c c\~ao de intrusos na rede, obtemos um classificador cujo erro de classifica\c c\~ao \'e ao torno de $0.3\%$.

Certo, \'e um bom resultado. Todavia, se voc\^e participar numa competi\c c\~ao, claro que todos os outros participantes v\~ao usar os classificadores padr\~ao. Para melhorar o resultado, \'e preciso combinar as t\'ecnicas conhecidas com as novas abordagens. E antes de melhorar o desempenho do algoritmo, precisamos compreender o que pode ser melhorado, onde h\'a um problema poss\'\i vel? 

Mas antes mesmo de examinar esta pergunta, temos uma ainda mais fundamental: por que n\'os esperamos que o classificador $k$-NN funcione, d\^e resultados confi\'aveis? 

A \'unica maneira de analisar as perguntas deste tipo \'e no formato de um modelo matem\'atico da aprendizagem supervisionada. 

Os dados s\~ao modelados pelas {\em vari\'aveis aleat\'orias} independentes com valores no nosso {\em dom\'\i nio,} $\Omega$. 
No nosso exemplo ``de brinquedo,'' o dom\'\i nio \'e o quadrado, $\Omega=[0,1]^2$. No segundo exemplo, o dom\'\i nio \'e $\Omega=\R^7$. 

A no\c c\~ao de uma vari\'avel aleat\'oria n\~ao \'e apenas o conceito mais fundamental da teoria de probabilidade, mas \'e, sem d\'uvida, uma das mais importantes no\c c\~oes em todas as ci\^encias matem\'aticas. Alguns matem\'aticos argumentam que, eventualmente, os fundamentos da matem\'atica devem ser alterados de modo que as vari\'aveis aleat\'orias sejam tratadas juntamente com conjuntos... Nos vamos tratar vari\'aveis aleat\'orias {\em ab inicio} no ap\^endice \ref{a:variaveis}.

O dom\'\i nio $\Omega$ \'e um espa\c co m\'etrico separ\'avel e completo (como, por exemplo, $\R^d$). Um ponto $(x,\ve)$ de dados rotulado, onde $x\in \Omega$, $\ve\in\{0,1\}$, \'e modelado por uma vari\'avel aleat\'oria $(X,Y)$ com valores no produto $\Omega\times \{0,1\}$. Aqui, $X\in\Omega$ representa um ponto no dom\'\i nio, e $Y\in\{0,1\}$, o r\'otulo marcando o ponto. A lei conjunta de $(X,Y)$ \'e uma medida de probabilidade, $\tilde\mu$, sobre $\Omega\times \{0,1\}$. Agora, o ponto $x\in\Omega$ \'e dito {\em inst\^ancia} da v.a. $X$, e o r\'otulo $\ve$ \'e uma inst\^ancia da v.a. $Y$.

\'E importante ressaltar que, mesmo se sempre supormos que a lei $\mu$ existe, ela \'e sempre desconhecida. Tamb\'em, \`as vezes o mesmo ponto $x\in\Omega$ pode aparecer na lista mais de uma vez, obtendo r\'otulos diferentes. 

Uma abordagem alternativa e as vezes mais c\^omoda \'e atrav\'es de uma medida de probabilidade, $\mu$, sobre $\Omega$ (a distribui\c c\~ao de pontos n\~ao rotulados), juntada com a {\em fun\c c\~ao de regress\~ao} $\eta\colon\Omega\to [0,1]$, cuja valor $\eta(x)$ \'e a probabilidade de ponto $x$ ser rotulado $1$. As duas abordagens s\~ao equivalentes.

Um {\em classificador} \'e uma fun\c c\~ao boreliana (mensur\'avel)
\[T\colon\Omega\to\{0,1\}.\]
Dado um classificador, o seu {\em erro de classifica\c c\~ao} \'e o valor real, a probabilidade de classifica\c c\~ao errada:
\begin{eqnarray*}
  {\mathrm{err}}_{\mu}(T) &=& P[T(X)\neq Y] \\
  &=& \mu\{(x,y)\in \Omega\times\{0,1\}\colon T(x) \neq y\}.\end{eqnarray*}
O {\em erro de Bayes} \'e o \'\i nfimo dos erros de classifica\c c\~ao de todos os classificadores poss\'\i veis sobre $\Omega$:
\[\ell^{\ast}=\ell^{\ast}(\mu)=\inf_{T}{\mathrm{err}}_{\mu}(T).\]
\'E simples mostrar que, com efeito, o \'\i nfimo \'e o m\'\i nimo, atingido pelo {\em classificador de Bayes:}
\[T_{bayes}(x) =\left\{\begin{array}{ll} 0,&\mbox{ se }\eta(x)<\frac 12,\\
  1,&\mbox{ se }\eta(x)\geq \frac 12,
\end{array}\right.\]
\[{\mathrm{err}}_{\mu}(T_{bayes})=\ell^{\ast}(\mu).\]
O significado do classificador de Bayes \'e puramente te\'orico, porque a fun\c c\~ao de regress\~ao, $\eta$, \'e desconhecida, assim como a lei $\mu$. 

Uma {\em regra da aprendizagem} \'e uma aplica\c c\~ao associando a cada amostra rotulada, $\sigma$, um classificador, $T$. Dado uma amostra 
\[\sigma = (x_1,x_2,\ldots,x_n,\ve_1,\ve_2,\ldots,\ve_n),\]
a regra produz um classificador, $T={\mathcal L}_n(\sigma)$, que \'e uma fun\c c\~ao boreliana de $\Omega$ dentro $\{0,1\}$. 

De maneira mais formal, podemos dizer que uma regra de aprendizagem \'e uma fam\'\i lia 
${\mathcal L}=\left({\mathcal L}_n\right)_{n=1}^{\infty}$, onde para cada $n=0,1,2,\ldots$,
\[{\mathcal L}_n\colon \Omega^n\times \{0,1\}^n \to \Omega^{\{0,1\}}.
\]
Uma maneira conveniente de escrever ${\mathcal L}_n$ \'e de avaliar a fun\c c\~ao ${\mathcal L}_n(\sigma)$ em ponto $x\in\Omega$, obtendo a aplica\c c\~ao
\[{\mathcal L}_n\colon \Omega^n\times \{0,1\}^n\times\Omega\to \{0,1\}.\]
Estas aplica\c c\~oes de avalia\c c\~ao,
\[\Omega^n\times \{0,1\}^n\times \Omega \ni (\sigma,x)\mapsto {\mathcal L}_n(\sigma)(x) \in \{0,1\},\]
devem ser tamb\'em mensur\'aveis.
Por exemplo, o classificador $k$-NN \'e uma regra de aprendizagem.

A amostra rotulada $(x_1,x_2,\ldots,x_n,\ve_1,\ldots,\ve_n)$ \'e modelada pela sequ\^encia 
\[(X_1,Y_1),(X_2,Y_2),\ldots,(X_n,Y_n)\] das vari\'aveis independentes com valores em $\Omega\times\{0,1\}$, seguindo a lei fixa por\'em desconhecida, $\mu$.
Para cada $n$, a regra de aprendizagem s\'o ``v\^e'' os $n$ primeiros pares de vari\'aveis. 

O erro de classifica\c c\~ao de uma regra de aprendizagem ao n\'\i vel $n$ \'e a esperan\c ca do erro sobre todas as amostras rotuladas de tamanho $n$:
\[{\mathrm{err}}_{\mu}({\mathcal L}_n)=\E_{\sigma\sim\mu} {\mathrm{err}}_{\mu}\left( {\mathcal L}_n(\sigma)\right).\]

Existem dois paradigmas principais da teoria de aprendizagem: {\em consist\^encia universal} e {\em aprendizagem dentro de uma classe.}

A consist\^encia universal \'e uma no\c c\~ao mais antiga de duas.
A regra de aprendizagem $\mathcal L$ \'e chamada {\em consistente} se o erro de classifica\c c\~ao converge para o erro de Bayes (o menor poss\'\i vel) em probabilidade quando $n\to\infty$:
\[\forall\ve>0,~~P\left[{\mathrm{err}}_{\mu}{\mathcal L}_n >\ell^\ast(\mu)+\ve\right]\to 0\mbox{ quando }n\to\infty.\]

Porque n\~ao conhecemos a lei subjacente, $\mu$, precisamos que a regra de aprendizagem seja consistente para todas as leis poss\'\i veis. Isto leva \`a seguinte defini\c c\~ao. A regra $\mathcal L$ \'e {\em universalmente consistente} se ela \'e consistente para cada medida de probabilidade $\mu$ sobre $\Omega\times\{0,1\}$. O classificador $k$-NN foi a primeira regra de aprendizagem cuja consist\^encia universal foi estabelecida.

\begin{teorema}[Charles Stone \citep*{stone:77}] 
  Suponha que $k=k_n\to\infty$ e $k_n/n\to 0$. Ent\~ao o classificador $k$-NN em $\R^d$ (com a dist\^ancia euclidiana) \'e universalmente consistente.
\end{teorema}

Assim, dentro do modelo atual da aprendizagem estat\'\i stica, o classificador $k$-NN, com alta confian\c ca, dar\'a uma resposta correta a longo prazo, quando o tamanho da amostra \'e bastante grande. 

O teorema de Stone falha nos espa\c cos m\'etricos mais gerais, mesmo no espa\c co de Hilbert de dimens\~ao infinita. Estudar a consist\^encia universal do classificador $k$-NN em v\'arios espa\c cos m\'etricos \'e um problema interessante, ainda n\~ao resolvido completamente. Vamos tratar-nos deste assunto no cap\'\i tulo \ref{c:kNN}.

Paradoxalmente, o classificador mais popular e mais eficaz do nosso tempo, a Floresta Aleat\'oria ({\em Random Forest classifier}, RF), n\~ao \'e universalmente consistente, pelo menos dentro de um modelo simplificado \citep*{BDL}.

Na defini\c c\~ao de um algoritmo consistente, o erro de classifica\c c\~ao converge para o erro de Bayes em probabilidade. Isso \'e equivalente ao seguinte. Dado  $\e>0$ ({\em erro}) e $\delta>0$ ({\em risco}), existe $N$ tal que, se $n\geq N$, a probabilidade ({\em confian\c ca}) que o erro de classifica\c c\~ao de ${\mathcal L}_n(\sigma)$ seja inferior a $\ell^\ast+\e$ \'e maior que $1-\delta$. Ent\~ao, podemos aprender o conceito desconhecido com um qualquer grau de confian\c ca, se temos bastante dados rotulados. 

Portanto, n\~ao \'e poss\'\i vel garantir a taxa de converg\^encia uniforme de $\e$ e $\delta$ para zero com $N$. Com efeito, existem as distribui\c c\~oes $\mu$ sobre o espa\c co euclidiano $\R^m$ cuja converg\^encia do erro e do risco para zero \'e t\~ao lento quanto se queira.

O segundo paradigma, {\em aprendizagem dentro uma classe,} se aplica \`a situa\c c\~ao onde o classificador n\~ao \'e universalmente consistente, por exemplo, uma rede neural da geometria fixa, com um n\'umero limitado de par\^ametros. Aqui, a teoria pode garantir o seguinte: supondo que a amostra pode ser separada pelo classificador bastante bem, com o pequeno {\em erro de aprendizagem,} o {\em erro de generaliza\c c\~ao} estar\'a pequeno tamb\'em. Em particular, temos os limites uniformes para converg\^encia do erro e do risco para zero.

 Relembramos que um classificador bin\'ario \'e uma fun\c c\~ao mensur\'avel $f\colon \Omega\to\{0,1\}$. Frequentemente, n\~ao distinguimos entre tal fun\c c\~ao \'e o conjunto $C$ de pontos onde $f$ toma o valor um: 
\[C =\{x\in\Omega\colon f(x)=1\}.\]
Neste caso, $f$ \'e a fun\c c\~ao indicadora de $C$:
\[f = \chi_C.\]
O conjunto $C$ neste contexto \'e chamado um {\em conceito.} O problema de classifica\c c\~ao bin\'aria dentro de uma classe \'e frequentemente analisado sob a hip\'otese simplificada do que a fun\c c\~ao de regress\~ao $\eta$ \'e determin\'\i stica, a fun\c c\~ao indicadora de um conceito. Deste modo, o problema se reduz \`a aprendizagem de um conceito desconhecido $C\subseteq\Omega$. No nosso exemplo ``de brinquedo,'' o conceito desconhecido \'e o conjunto $A$.

Uma das medidas mais importantes da complexidade de uma classe de conceitos $\mathscr C$ \'e um par\^ametro not\'avel, a {\em dimens\~ao de Vapnik--Chervonenkis} da classe $\mathscr C$.  
Para um subconjunto $A\subseteq\Omega$, dizemos que $\mathscr C$ {\em fragmenta} $A$, se cada subconjunto $B\subseteq A$ pode ser obtido como a interse\c c\~ao de $A$ com um conjunto $C\in {\mathscr C}$:
\[\{A\cap C\colon C\in{\mathscr C}\} = 2^A.\]
Em outras palavras, cada rotulagem de $A$ pode ser classificada corretamente com uma fun\c c\~ao da forma $\chi_C$, $C\in{\mathscr C}$.

Por exemplo, o conjunto $\mathscr P$ de todos os semiplanos de $\R^2$ fragmenta cada subconjunto de tr\^es pontos n\~ao em posi\c c\~ao geral. Ao mesmo tempo, \'e f\'acil mostrar que nenhum conjunto com quatro pontos \'e fragmentado pelos semiplanos.

A dimens\~ao de Vapnik--Chervonenkis (dimens\~ao $VC$) de uma classe $\mathscr C$ de subconjuntos de $\Omega$ \'e o supremo de cardinalidades de subconjuntos finitos de $\Omega$ fragmentados por $\mathcal C$.
Por exemplo, a dimens\~ao VC da classe de semiplanos em $\R^2$ \'e $3$.

Agora, um dos resultados principais da teoria de Vapnik-Chervonenkis pode ser formulado assim. Seja $\mathscr C$ uma classe (fam\'\i lia de classificadores). As condi\c c\~oes seguintes s\~ao equivalentes:
\begin{itemize}
\item Existem cotas uniformes para o erro de generaliza\c c\~ao e o risco que dependem apenas de $n$, o tamanho da amostra, e n\~ao dependem da lei subjacente $\mu$.
\item A dimens\~ao VC de $\mathscr C$ \'e finita.
\end{itemize} 

Algoritmos baseados sobre esta teoria seguem o {\em Princ\'\i pio de Minimiza\c c\~ao do Risco Emp\'\i rico}  ({\em ERM}). A regra $\mathcal L$ est\'a tentando minimizar o erro de aprendizagem, ou bem o {\em risco emp\'\i rico,}
\[\frac 1n\sharp \{i\colon \ve_i\neq T(x_i)\},\]
onde $T={\mathcal L}_n(\sigma)$.
O pre\c co a pagar para controlarmos o erro de classifica\c c\~ao \'e o que tais algoritmos n\~ao s\~ao universalmente consistentes.  

Afinal, h\'a algoritmos que combinam o melhor de dois mundos. Eles s\~ao baseados sobre o {\em Princ\'\i pio de Minimiza\c c\~ao do Risco Estrutural,} \'e s\~ao universalmente consistentes ao mesmo tempo que permitem controlar o erro de generaliza\c c\~ao.

Falemos um pouco sobre a estrutura do texto. No cap\'\i tulo \ref{ch:rotulagens}, estudamos o {\em cubo de Hamming,} $\{0,1\}^n$, um objeto  da teoria combinat\'oria, que consiste de todas as sequ\^encias bin\'arias de comprimento fixo, $n$. Para nos, o cubo tem interesse como o espa\c co de todas as rotulagens poss\'\i veis sobre uma amostra fixa com $n$ pontos. O cubo de Hamming admite uma m\'etrica e uma medida naturais, e a intera\c c\~ao entre as duas estruturas leva a uma geometria e uma vers\~ao da an\'alise interessantes. O que \'e mais importante para nos, \'e o {\em fen\^omeno de concentra\c c\~ao de medida:} as fun\c c\~oes Lipschitz cont\'\i nuas sobre o cubo, com constantes de Lipschitz controladas, s\~ao altamente concentradas em torno do seu valor mediano (ou: m\'edio), quando $n$ \'e bastante grande. Em outras palavras, tais fun\c c\~oes t\^em uma pequena vari\^ancia.

Pode-se dizer que, intuitivamente, a aprendizagem estat\'\i stica exige de uma combina\c c\~ao de alta concentra\c c\~ao com baixa complexidade. A parte da ``alta concentra\c c\~ao'' est\'a tratada no cap\'\i tulo \ref{ch:rotulagens}. Revisitamos a concentra\c c\~ao no ap\^endice \ref{a:esfera}, onde ela vai ser estabelecida para a esfera euclidiana $\s^n$ (a desigualdade cl\'assica de Paul L\'evy, que foi o primeiro matem\'atico  a isolar o fen\^omeno explicitamente).
A parte da ``baixa complexidade'' \'e estudada no cap\'\i tulo \ref{ch:VC}, consagrado \`a dimens\~ao de Vapnik--Chervonenkis. Em particular, estudamos como a dimens\~ao VC se mudar quando as unidades computacionais est\~ao combinadas em uma rede. 

Cap\'\i tulo \ref{ch:PAC} j\'a trata-se de algumas no\c c\~oes b\'asicas da teoria de aprendizagem autom\'atica estat\'\i stica, inicialmente no contexto de uma lei puramente at\^omica, quando a distribui\c c\~ao \'e suportada por uma sequ\^encia finita ou enumer\'avel de pontos de massa estritamente positiva. Esta simplifica\c c\~ao permite evitar as sutilezas da teoria de medida, e serve como mera ilustra\c c\~ao do modelo. Por\'em, j\'a neste contexto mostramos alguns resultados interessantes, tais como a an\'alise de taxa de converg\^encia do erro de um classificador universalmente consistente para zero, que pode ser t\~ao lenta quanto se queira. Depois de passarmos para medidas gerais, provamos no mesmo cap\'\i tulo \ref{ch:PAC} o teorema de Benedek--Itai, afirmando que uma classe de conceitos, $\mathscr C$, \'e provavelmente aproximadamente aprendiz\'avel sob uma lei fixa, $\mu$, se e somente se $\mathscr C$ \'e pr\'e-compacto em rela\c c\~ao \`a dist\^ancia $L^1(\mu)$.

A lei dos grandes n\'umeros diz que, se o tamanho de uma amostra aleat\'oria $\sigma$ \'e bastante grande, ent\~ao, com alta confian\c ca, a medida emp\'\i rica de um conceito dado, $C\subseteq\Omega$, ou seja, a fra\c c\~ao de pontos de $\sigma$ contidos em $C$, se aproxima do valor da medida $\mu(C)$. Uma classe de conceitos $\mathscr C$ \'e dito {\em classe de Glivenko--Cantelli} se, com alta confian\c ca, a mesma conclus\~ao vale para todos os elementos da classe simultaneamente, ou seja, com alta confian\c ca, a medida de cada elemento $C\in {\mathscr C}$ pode ser aproximada pela medida emp\'\i rica usando apenas uma amostra aleat\'oria.  No cap\'\i tulo \ref{ch:GC} estamos apresentando v\'arias carateriza\c c\~oes de classes de Glivenko--Cantelli sob uma medida fixa, e deduzimos consequ\^encias para aprendizagem. Cada classe de Glivenko--Cantelli \'e {\em consistentemente aprendiz\'avel,} ou seja, cada regra de aprendizagem que busca um elemento da classe induzindo a rotulagem original sempre que poss\'\i vel, aprende $\mathscr C$. E as classes que s\~ao Glivenko--Cantelli uniformemente em rela\c c\~ao \`a medida s\~ao exatamente as classes aprendiz\'aveis uniformemente em rela\c c\~ao \`a medida e exatamente as classes de dimens\~ao VC finita.

No cap\'\i tulo \ref{c:kNN} estudamos a no\c c\~ao de consist\^encia universal de uma regra de aprendizagem, e estabelecemos a consist\^encia universal do classificador $k$-NN numa grande classe de espa\c cos m\'etricos, os ditos espa\c cos de dimens\~ao de Nagata sigma-finita. Contudo, a descri\c c\~ao completa da classe de espa\c cos m\'etricos onde $k$-NN \'e universalmente consistente resta um problema em aberto.

O assunto do cap\'\i tulo \ref{ch:reducao} \'e o problema seguinte. Em dom\'\i nios de alta dimens\~ao $ d\gg 1$, v\'arios algoritmos da ci\^encia de dados muitas vezes levam muito tempo e tornam-se ineficientes. Este fen\^omeno, conhecido como a maldi\c c\~ao de dimensionalidade, \'e ainda pouco entendido. A redu\c c\~ao de dimensionalidade \'e qualquer fun\c c\~ao do dom\'\i nio de dimens\~ao alta para um dom\'\i nio de dimens\~ao baixa, transferindo o problema de aprendizagem para l\'a. N\'os discutimos alguns aspectos da maldi\c c\~ao de dimensionalidade e suas liga\c c\~oes poss\'\i veis com o fen\^omeno de concentra\c c\~ao de medida, assim como duas abordagens para reduzir a dimensionalidade: o lema de Johnson--Lindenstrauss (para qual ainda n\~ao existe alguma base te\'orica no cen\'ario da aprendizagem), assim como a redu\c c\~ao usando fun\c c\~oes injetoras borelianas, que resulta em regras universalmente consistentes.

No cap\'\i tulo \ref{ch:aproximacao}, estudamos a capacidade para redes neurais aproximarem uma dada fun\c c\~ao. Selecionamos dois resultados importantes. O primeiro, o teorema da superposi\c c\~ao de Kolmogorov, diz que, dado $d\in\N$, existe uma rede de arquitetura fixa, cujos par\^ametros s\~ao fun\c c\~oes reais de uma vari\'avel, que pode gerar todas as fun\c c\~oes cont\'\i nuas em $d$ vari\'aveis. No entanto, o significado pr\'atico desta rede at\'e agora parece ser limitada, pois ela n\~ao \'e facilmente implement\'avel. O segundo resultado, o teorema da aproxima\c c\~ao universal de Cybenko, diz que uma rede  cujas unidades s\~ao fun\c c\~oes afins com $d$ argumentos, e que s\'o tem uma camada escondida, pode aproximar qualquer fun\c c\~ao $L^1(\R^d,\mu)$ qualquer que seja $\mu$, quando o n\'umero de unidades cresce. 

Finalmente, no cap\'\i tulo \ref{ch:compressao} estudamos a compress\~ao amostral. Assim \'e chamado um modo de codificar cada elemento $C$ de uma classe de conceitos, $\mathscr C$, com uma amostra (rotulada ou n\~ao) de tamanho $\leq d$, onde $d$ se chama o {\em tamanho} do esquema de compress\~ao. O esquema deve satisfazer a restri\c c\~ao seguinte: dada uma amostra finita $\sigma$, os conceitos codificados com subamostras de $\sigma$ tendo $\leq d$ elementos devem produzir todas as mesmas rotulagens sobre $\sigma$ que os elementos da classe $\mathscr C$. A maior hip\'otese em aberto sugere que cada classe de dimens\~ao de Vapnik--Chervonenkis $\leq d$ admite um esquema de compress\~ao de tamanho $\leq d$. Discutimos alguns resultados nesta dire\c c\~ao, o mais importante deles sendo o teorema recente de Moran--Yehudayoff afirmando que cada classe de dimens\~ao VC $\leq d$ admite um esquema de compress\~ao de tamanho exponencial em $d$.

Tent\'amos tornar o texto razoavelmente aut\^onomo com ajuda de ap\^endices. O ap\^endice \ref{a:variaveis} cont\'em uma introdu\c c\~ao informal \`a no\c c\~ao de uma vari\'avel aleat\'oria, levando o leitor para no\c c\~oes de um espa\c co boreliano padr\~ao (ap\^endice \ref{apendice:padrao}), medida de probabilidade (ap\^endice \ref{a:medidas}), e esperan\c ca / integral (ap\^endice \ref{a:integral}). Como pr\'e-requisitos, precisamos elementos da teoria de conjuntos (ap\^endice \ref{a:conjuntos}), espa\c cos m\'etricos (ap\^endice \ref{ch:metricos}), e espa\c cos normados (ap\^endice \ref{ch:norma}), onde introduzimos algumas ferramentas b\'asicas tais como o teorema da categoria de Baire, o teorema de Hahn--Banach, o teorema de Stone--Weierstrass, o teorema de representa\c c\~ao de Riesz, etc. Em dois ap\^endices separados mostramos resultados que n\~ao se encaixam na lista acima: o lema de Kronecker (ap\^endice \ref{a:kronecker}) e o teorema Minimax ``concreto'' de von Neumann (ap\^endice \ref{a:minimax}). Como j\'a mencionamos, o ap\^endice \ref{a:esfera} cont\'em a prova da desigualdade de concentra\c c\~ao de Paul L\'evy.

Alguns coment\'arios sobre a leitura posterior. Meus textos favoritos na \'area de aprendizagem de m\'aquina te\'orica s\~ao provavelmente \citep*{shalev-shwartz_ben-david} e \citep*{vidyasagar}. Acho o livro \citep*{AB} muito leg\'\i vel tamb\'em. A monografia \citep*{DGL} oferece uma refer\^encia enciclop\'edica. O livro \citep*{vapnik2} \'e rico em ensinamentos. Tamb\'em sugiro o livro em prepara\c c\~ao interessante \citep*{BHK}. A monografia \citep*{dudley} \'e dedicada \`a teoria de classes de Glivenko--Cantelli e classes relacionadas. Para algu\'em que trabalha na an\'alise e teoria de aproxima\c c\~ao, a abordagem adotada em \citep*{cucker_smale,cucker_zhou} vai ser atraente. O livro \citep*{torgo} pode ajudar a dominar alguns aspectos pr\'aticos da ci\^encia de dados.

Estas notas de aula ainda est\~ao em uma forma \'aspera.
Quero convidar o leitor a enviar coment\'arios, criticismos, e corre\c c\~oes para {\small\tt <vpest283@uottawa.ca>}. Talvez, eventualmente, a segunda edi\c c\~ao aparecer\'a, escrita mais cuidadosamente.


%
%

\chapter{A geometria de rotulagens\label{ch:rotulagens}}

O {\em cubo de Hamming}
\index{cubo! de Hamming}
 pode ser visto como o conjunto de todas as rotulagens poss\'\i veis de uma dada amostra finita
\[x_1,x_2,\ldots,x_n,\]
ou seja, formas de associar a cada ponto de amostra $x_i$ o r\'otulo $0$ ou $1$. O cubo de Hamming, por si s\'o, \'e um interessante objeto de estudo da geometria discreta.  

Neste cap\'\i tulo, vamos explorar a an\'alise e a geometria do cubo, que nos remeter\'a \`as {\em desigualdades de concentra\c c\~ao.}

\begin{definicao}
Seja $n\in\N$. 
O {\it cubo de Hamming} de dimens\~ao $n$ is a cole\c c\~ao de todas as sequ\^encias bin\'arias de comprimento $n$. \'E denotado por
$\{0,1\}^n$ ou $\Sigma^n$.
\end{definicao}

\index{sigma@$\Sigma^n$}

Assim, um elemento $\sigma\in\Sigma^n$ \'e da forma
\[\sigma=\sigma_1\sigma_2\cdots\sigma_n,\]
onde $\sigma_i\in\Sigma=\{0,1\}$, o {\em alfabeto} de dois s\'\i mbolos.

Nosso objetivo \'e equipar $\Sigma^n$ com algumas estruturas adicionais. A primeira delas ser\'a uma dist\^ancia. 

\section{Cubo de Hamming como espa\c co m\'etrico}

\subsection{Dist\^ancia de Hamming}

\begin{definicao}
Seja $n\in\N$. A {\it dist\^ancia de Hamming} entre duas sequ\^encias de comprimento $n$,
$\sigma,\tau\in\Sigma^n$, \'e definida por
\[d(\sigma,\tau)=\sharp\{i\colon \sigma_i\neq\tau_i\}.\]
\end{definicao}

\index{dist\^ancia! de Hamming!}

Aqui $\sharp$ denota a cardinalidade (o n\'umero de elementos) de um conjunto finito. Assim, a dist\^ancia de Hamming entre duas sequ\^encias de mesmo comprimento \'e igual ao n\'umero de posi\c c\~oes nas quais elas diferem entre si.  
O seguinte \'e um exerc\'\i cio simples.

\begin{proposicao}
A dist\^ancia de Hamming satisfaz as seguintes propriedades. 
\begin{enumerate}
\item (Propriedade de separa\c c\~ao) $d(\sigma,\tau)=0$ $\Leftrightarrow$ $\sigma=\tau$.
\item (Simetria)
$d(\sigma,\tau)=d(\tau,\sigma)$.
\item (Desigualdade triangular)
$d(\sigma,\tau)\leq d(\sigma,\varsigma)+d(\varsigma,\tau)$.
\end{enumerate}
\end{proposicao}

\index{m\'etrica}

Ou seja, a proposi\c c\~ao acima garante que a dist\^ancia de Hamming \'e uma m\'etrica, e portanto o par $(\Sigma^n,d)$ \'e um {\it espa\c co m\'etrico} (subs. \ref{ss:espmet}).

Definiremos o {\em di\^ametro} de um espa\c co m\'etrico $(X,d)$ como o supremo de dist\^ancias entre todos os pares de pontos de $X$:
\[\mathrm{diam}(X)=\sup_{x,y\in X} d(x,y).\]
\index{di\^ametro! de espa\c co m\'etrico}
Temos ent\~ao que $\mathrm{diam}(\Sigma^n,d)=n$, assim o di\^ametro do cubo de dimens\~ao $n$ tende ao infinito quando $n\to\infty$. Para ter um certo controle sobre o tamanho dos cubos no caso assint\'otico $n\to\infty$, vamos introduzir a seguinte no\c c\~ao. 

\begin{definicao}
A {\it dist\^ancia de Hamming normalizada} entre duas sequ\^encias de comprimento $n$ \'e dada por 
\[\bar d(\sigma,\tau)=\frac 1 n d(\sigma,\tau).\]
\end{definicao}

\index{dist\^ancia! de Hamming! normalizada}

Por exemplo,
\[\bar d(00110, 10101)=\frac 35.\]
A vantagem da dist\^ancia de Hamming normalizada \'e o fato do di\^ametro de $(\Sigma^n,\bar d)$ ser igual a um, independentemente do valor de $n$.
A seguir apresentamos alguns conjuntos associados \`a dist\^ancia de Hamming. 

\begin{definicao}
Seja $x$ um ponto num espa\c co m\'etrico $(X,d)$ e seja $\e>0$.
A {\em bola aberta} em torno de $x$ de raio $\e$ \'e o conjunto
\[B_\e(x)=\{y\in X\mid d(x,y)<\e\},\]
\index{bola! aberta}
e a {\em bola fechada} em torno de $x$ de raio $\e$,
\[\bar B_\e(x)=\{y\in X\mid d(x,y)\leq \e\}.\]
\end{definicao}
\index{bola! fechada}

\begin{exemplo}
$B_1(\mathbf{0})=\{\mathbf{0}\}$.
\end{exemplo}

\begin{exemplo}
A bola aberta de raio $\frac 2n$ em torno de $\mathbf{0}$ no espa\c co $(\Sigma^n,\bar d)$ consiste de todas as sequ\^encias contendo no m\'aximo um $1$. 
\end{exemplo}

\begin{observacao}
A estrutura m\'etrica do cubo de Hamming tem sua origem em teoria de c\'odigos corretores de erros, onde as sequ\^encias bin\'arias s\~ao vistas como mensagens que ser\~ao transmitidas por um canal de comunica\c c\~oes, na presen\c ca de ru\'\i do, o que significa que alguns bits podem ser corrompidos durante a transmiss\~ao. Assumindo que n\~ao s\~ao muitos os bits corrompidos durante uma transmiss\~ao qualquer, digamos n\~ao mais do que um dado percentual $\e<1$, existe um eficiente conceito de {\em c\'odigo corretor de erros}. 

Um {\em c\'odigo} \'e nada mais do que um subconjunto $A\subseteq\Sigma^n$. Digamos que $A$ corrige erros de ordem at\'e $\e$ se para todos 
$a,b\in A$, as bolas abertas de raio $\e$ em torno de $a$ e $b$ s\~ao disjuntas:
\[B_\e(a)\cap B_\e(b)=\emptyset.\]
Esta condi\c c\~ao tem o seguinte corol\'ario. Suponha que uma palavra 
$\sigma\in A$ foi transmitida e que a palavra $\sigma^\prime$ foi recebida, com $\bar d(\sigma,\sigma^\prime)<\e$ (ou seja, somente uma fra\c c\~ao menos do que $\e$ dos bits foi corrompida). Ent\~ao, $\sigma^\prime\in B_\e(\sigma)$ e o fato de que $a\in A$ e $\sigma^\prime\in B_\e(a)$ implica $\sigma=a$. Assim, para corrigir o erro, \'e suficiente encontrar em $A$ o vizinho mais pr\'oximo da palavra corrompida, $\sigma^\prime$, com rela\c c\~ao \`a dist\^ancia normalizada de Hamming. 

Para uma introdu\c c\~ao na teoria de c\'odigos corretores de erros, veja \citep*{HV}.
\end{observacao}

\subsection{Homogeneidade}
A estrutura m\'etrica do cubo de Hamming \'e bastante rica.  Os resultados que seguem (Proposi\c c\~ao \ref{homo} e observa\c c\~ao
\ref{semi}) s\~ao v\'alidos para a dist\^ancia de Hamming e a sua vers\~ao normalizada. 

\begin{definicao}
Seja $(X,d)$ um espa\c co m\'etrico. 
Uma {\em isometria} de $X$ \'e uma fun\c c\~ao bijetora 
$i\colon X\to X$ que preserva as dist\^ancias, isso \'e, tem a propriedade $d(x,y)=d(i(x),i(y))$ para cada $x,y\in X$. 
\end{definicao}
\index{isometria}

Por exemplo, toda transla\c c\~ao $x\mapsto x+a$ da reta
$\R$ munida da dist\^ancia usual \'e uma isometria.  Um outro exemplo da isometria \'e uma rota\c c\~ao do c\'\i rculo unit\'ario munido da dist\^ancia euclideana.

\begin{observacao}
Se o espa\c co m\'etrico $X$ \'e finito, ent\~ao cada fun\c c\~ao que preserva as dist\^ancias \'e automaticamente bijetora. 
\end{observacao}

\begin{proposicao}
\label{homo}
O cubo de Hamming $\Sigma^n$ \'e metricamente homog\^eneo, ou seja,
para todo $\sigma,\tau\in\Sigma^n$ existe uma isometria $i$ de
$\Sigma^n$ tal que $i(\sigma)=\tau$.
\end{proposicao}
\index{espa\c co! metricamente homog\^eneo}

\begin{proof}
Primeiramente observe que toda aplica\c c\~ao de forma
\[\Sigma^n\ni\sigma\mapsto\sigma+\tau\in\Sigma^n\]
\'e uma isometria, onde a soma \'e realizada componente a componente m\'odulo $2$: 
\[(\sigma+\tau)_i=\sigma_i+\tau_i \mod 2.\]
Sejam $\sigma,\tau\in\Sigma^n$ elementos quaisquer de $\Sigma^n$, e seja
\[\varsigma=\sigma+\tau.\]
A isometria $i=\sigma\mapsto\sigma+\varsigma$
envia $\sigma$ para $\tau$.
\end{proof}

\begin{observacao}
Na demonstra\c c\~ao acima temos que $i$ permuta $\sigma$ e $\tau$. Por\'em, nem todo espa\c co metricamente homog\^eneo satisfaz esta propriedade (exerc\'\i cio!), a qual \'e uma forma mais forte de homogeneidade. 
\end{observacao}

\begin{observacao}
\label{semi}
As isometrias $i$ como as apresentadas na demonstra\c c\~ao acima s\~ao transla\c c\~oes em $\Sigma^n$ quando considerado como um grupo munido da adi\c c\~ao (componente por componente) m\'odulo $2$. Todavia, existem isometrias de outra natureza. A saber, se 
$s$ \'e uma permuta\c c\~ao do conjunto $\{1,2,\cdots,n\}$, ent\~ao podemos definir uma isometria de $\Sigma^n$, denotada pela mesma letra $s$, da seguinte maneira:
\[s(\sigma)_i=\sigma_{s(i)}.\]
Observe que estas isometrias s\~ao diferentes das anteriores, pois as permuta\c c\~oes mant\^em o zero (palavra nula) fixo, enquanto a transla\c c\~ao, n\~ao.
 A cole\c c\~ao de todas as isometrias $s$ \'e isomorfa ao grupo de permuta\c c\~oes de posto $n$, $S_n$. Pode-se mostrar agora que toda isometria de $\Sigma^n$ \'e uma composi\c c\~ao de isometrias dos dois tipos apresentados (transla\c c\~oes e permuta\c c\~oes). Mais exatamente, o grupo de isometrias de $\Sigma^n$ \'e o chamado {\em produto semidireto}
$S_n\ltimes\Sigma^n$.
\end{observacao}

\begin{definicao}
Um espa\c co m\'etrico $(X,d)$ \'e chamado $n$-{\it homog\^eneo}, onde
$n$ \'e um n\'umero natural, se, dados as subespa\c cos isom\'etricos $A$ e $B$ quaisquer com at\'e $n$ elementos em cada um e uma isometria $i\colon A\to B$, existe uma isometria $j$ de $X$ que estende $i$: $j\vert_A=i$.
\end{definicao}

Por exemplo, $1$-homogeneidade \'e a homogeneidade m\'etrica usual. 

Para uma palavra $\sigma$, denote $\mathrm{supp}\,\sigma=\{i\colon \sigma_i=1\}$.

\begin{proposicao}
O cubo de Hamming \'e $3$-homog\^eneo.
\end{proposicao}

\begin{proof}
Sejam $A=\{\sigma_1,\sigma_2,\sigma_3\}$ e $B=\{\tau_1,\tau_2,\tau_3\}$ duas triplas de palavras quaisquer, tais que $d(\sigma_i,\sigma_j)=d(\tau_i,\tau_j)$ para todo
$i,j=1,2,3$. 

1. Primeiramente assuma que $\sigma_1={\mathbf{0}}=\tau_1$ e
$\sigma_2=\tau_2$. Ent\~ao $w(\sigma_3)=d(\sigma_3,\mathbf{0})=
w(\tau_3)$, e como $d(\sigma_3,\sigma_2)=d(\tau_2,\tau_3)$, 
os conjuntos $\mathrm{supp}\,\sigma_3\setminus\mathrm{supp}\,\sigma_2$ e
$\mathrm{supp}\,\tau_3\setminus\mathrm{supp}\,\sigma_2$ (onde $\sigma_2=\tau_2$) tem o mesmo n\'umero de elementos. Logo, existe uma permuta\c c\~ao de coordenadas, $s$, tal que
$s$ tem $\mathrm{supp}\,\sigma_2$ fixo e permuta
 $\mathrm{supp}\,\sigma_3\setminus\mathrm{supp}\,\sigma_2$ e
$\mathrm{supp}\,\tau_3\setminus\mathrm{supp}\,\sigma_2$. Tal $s$ defina uma isometria do cubo, enviando $\sigma_1={\mathbf{0}}$ e $\sigma_2$ para si mesmos e permutando $\sigma_3$ com $\tau_3$.

2. Agora suponhamos que $\sigma_1={\mathbf{0}}=\tau_1$.
Neste caso,  $w(\sigma_2)=d(\sigma_2,{\mathbf{0}})=
w(\tau_2)$, ou seja, $\mathrm{supp}\,\sigma_2$ e $\mathrm{supp}\,\tau_2$ tem o mesmo n\'umero de elementos, e existe uma permuta\c c\~ao de coordenadas, digamos $t$, permutando  
$\mathrm{supp}\,\sigma_2$ e $\mathrm{supp}\,\tau_2$. O problema \'e reduzido para o caso 1, pois $t(\sigma_1)={\mathbf{0}}=t(\tau_1)$, entanto
$t(\sigma_2)=t(\tau_2)$. Podemos escolher $s$ como no caso 1, e a isometria resultante \'e da forma
\[t^{-1}\circ s \circ t\]
(verifique!)

3. Finalmente, tratamos o caso geral de duas triplas como acima. Definiremos duas isometrias, $i_1$ e $i_2$, como segue:
\[i_1(\alpha)= \alpha+\sigma_1,\]
\[i_2(\alpha)=\alpha+\tau_1.\]
Ent\~ao $i_1(\sigma_1)={\mathbf{0}}=i_2(\tau_1)$, e as triplas
\[\{i_1(\sigma_1),i_1(\sigma_2),i_1(\sigma_3)\}\mbox{ e }
\{i_2(\tau_1),i_2(\tau_2),i_2(\tau_3)\}\] 
satisfazem as hip\'oteses do caso 2. Escolhemos uma isometria, $i_3$, enviando a segunda tripla para a primeira. Logo, a isometria final, enviando 
$\{\sigma_1,\sigma_2,\sigma_3\}$ para $\{\tau_1,\tau_2,\tau_3\}$, \'e dada pela composi\c c\~ao: 
\[ i_2^{-1}\circ i_3\circ i_1.\]
\end{proof}

\begin{observacao} \'E interessante que o cubo de Hamming n\~ao \'e $4$-homog\^eneo, como vermos no exemplo abaixo. A demonstra\c c\~ao do lema \ref{mike} e do exemplo
\ref{non-four} foram feitas por Mike Doherty nas suas solu\c c\~oes de um dever em 2002. 
\end{observacao}

\begin{lema}
\label{mike}
Se uma isometria do cubo de Hamming deixa $\mathbf{0}$ fixo, ent\~ao ela \'e uma permuta\c c\~ao de coordenadas. 
\end{lema}

\begin{proof} Se $i({\mathbf{0}})={\mathbf{0}}$, ent\~ao $i$ permuta as palavras com o suporte de um elemento entre elas, e desta maneira, ela determina uma permuta\c c\~ao de coordenadas, $\pi$. Precisamos verificar que $i$ age pela permuta\c c\~ao sobre todas outras palavras. Toda palavra $\sigma$ \'e determinada unicamente pela cole\c c\~ao de dist\^ancias entre ela e as palavras  $00\ldots 0$, $10\ldots 0$,
$010\ldots 0$, $\ldots$, $000\ldots 01$. Logo, $i(\sigma)$ \'e determinada unicamente pela cole\c c\~ao de dist\^ancias entre $i(\sigma)$ e as mesmas palavras permutadas pelo $i$, logo \'e igual a 
$\pi(\sigma)=\pi(\sigma_1)\pi(\sigma_2)\ldots\pi(\sigma_n)$.
\end{proof} 

\begin{exemplo}
\label{non-four}
Considere dois qu\'adruplos de palavras (horizontais) em $\Sigma^4$:
\[
\begin{array}{cccc}
1&0&0&1\\
0&1&0&1\\
0&0&1&1 \\
0&0&0&0
\end{array} ~~\mbox{ e }~~
\begin{array}{rccccl}
0&1&1&0\\
0&1&0&1\\
0&0&1&1 \\
0&0&0&0
\end{array}
\]
Eles s\~ao isom\'etricos (congruentes). Ao mesmo tempo, n\~ao existe isometria de $\Sigma^n$ que envia as palavras correspondentes do primeiro quadro \`as palavras do segundo. Uma tal isometria, $i$, preservaria a palavra zero, ent\~ao, pelo lema \ref{mike}, seria gerada por uma permuta\c c\~ao, $\pi$, de quatro coordenadas. Examinando as palavras nas 2$\textsuperscript{\underline{a}}$ e 3$\textsuperscript{\underline{a}}$ linhas, conclu\'\i mos que a permuta\c c\~ao deixa a primeira coordenada fixa: 
$\pi(1)=1$. Isso n\~ao permite \`a primeira palavra no primeiro quadro ser enviada para a primeira palavra no segundo quadro. 
\end{exemplo}

\subsection{Fun\c c\~oes}

Nosso pr\'oximo objetivo \'e desenvolver uma teoria de fun\c c\~oes cujo dom\'\i nio \'e $\Sigma^n$. Em particular, gostar\'\i amos de saber quais fun\c c\~oes s\~ao a ser escolhidas como as ``agrad\'aveis'' de se trabalhar. Os conceitos fundamentais de continuidade e diferenciabilidade, t\~ao importantes em an\'alise, perdem seu significado no caso discreto, pois por exemplo toda fun\c c\~ao $f\colon\Sigma^n\to\R$ \'e cont\'\i nua:
\[\forall\e>0,\exists\delta>0, \forall \sigma,\tau\in\Sigma^n,
d(\sigma,\tau)<\delta \Rightarrow \abs{f(\sigma)-f(\tau)}<\tau,\]
ou seja, dado $\e>0$, basta definir $\delta=1$ 
(para a dist\^ancia de Hamming normalizada, $\delta=\frac 1n$).\footnote{Com efeito, a topologia de $\Sigma^n$ \'e discreta, isso \'e, cada aplica\c c\~ao de $\Sigma^n$ para um espa\c co topol\'ogico qualquer \'e cont\'\i nua.} 
A seguir apresentamos uma no\c c\~ao adequada de uma fun\c c\~ao ``agrad\'avel'' neste contexto.  

\begin{definicao} Uma fun\c c\~ao real $f$ sobre um espa\c co m\'etrico $(X,d)$ \'e chamada
{\it Lipschitz cont\'\i nua} se existe um n\'umero real $L$ tal que
para todos $x,y\in X$
\[\abs{f(x)-f(y)}\leq L\cdot d(x,y).\]
\index{fun\c c\~ao! Lipschitz cont\'\i nua}
Este $L$ \'e chamado uma {\it constante de Lipschitz} de $f$. 
\index{constante! de Lipschitz}

O \'\i nfimo de todas as constantes de Lipschitz para uma fun\c c\~ao Lipschitz cont\'\i nua $f$ (que \'e uma constante de Lipschitz tamb\'em, exerc\'\i cio), \'e denotado $L_f$ ou bem $\mathrm{Lip}(f)$. 
Uma fun\c c\~ao Lipschitz cont\'\i nua com $L=1$ \'e chamada uma fun\c c\~ao {\em n\~ao expansiva,} ou 
$1${\it -Lipschitz cont\'\i nua.} 
\end{definicao}

Assim, a constante de Lipschitz $L$ nos diz que uma fun\c c\~ao aumenta as dist\^ancias entre os pontos do seu dom\'\i nio por um fator de $L$ no m\'aximo.

\begin{observacao}
De fato, {\it toda} fun\c c\~ao definida no cubo de Hamming, ou em qualquer espa\c co m\'etrico finito, $X$, \'e Lipschitz cont\'\i nua: neste caso, temos 
\[\mathrm{Lip}(f)=\max_{x,y\in X,x\neq y}
\frac{\abs{f(x)-f(y)}}{d(x,y)}<+\infty,\]
pois estamos tomando o m\'aximo em um conjunto finito.

Seria ent\~ao o caso de se afirmar que o conceito de fun\c c\~ao Lipschitz cont\'\i nua n\~ao apresenta nenhum significado importante? A resposta \'e n\~ao, pois \'e justamente o fato de se poder conhecer o valor exato de $L=\mathrm{Lip}(f)$ que \'e importante. Informalmente falando, $L$ mostra qu\~ao Lipschitziana uma fun\c c\~ao Lipschitz cont\'\i nua \'e. Os exemplos a seguir v\~ao evidenciar este ponto.
\label{everylip}
\end{observacao}

\begin{exemplo}
Toda fun\c c\~ao constante \'e Lipschitz cont\'\i nua com $L=0$.
\end{exemplo}

\begin{exemplo}
\label{n-lip}
A fun\c c\~ao $\pi_1$ enviando a palavra para o primeiro bit,
\[\pi_1\colon\Sigma^n\ni\sigma\mapsto \sigma_1\in\{0,1\}\subset\R,\]
\'e Lipschitz cont\'\i nua com $L=1$ em rela\c c\~ao \`a dist\^ancia de Hamming, e com $L=n$ em rela\c c\~ao \`a dist\^ancia de Hamming normalizada. A constante n\~ao pode se melhorar: de fato, 
se $\sigma=100\cdots 0$, ent\~ao $\bar d(\mathbf{0},\sigma)=\frac 1n$, e no mesmo tempo a dist\^ancia entre as imagens \'e $1$.
\end{exemplo}

\begin{exemplo}
O n\'umero de $1$s numa palavra bin\'aria $\sigma$ \'e o {\em peso} da palavra, denotado $w(\sigma)$:

\begin{align*}
w(\sigma)\ &= \ \sharp\{i\colon \sigma(i)=1\}\\
\ &=\  \sum_{i=1}^n\sigma_i.
\end{align*}
De maneira equivalente, 
\[w(\sigma)=d(\mathbf{0},\sigma).\]
O peso, visto como a fun\c c\~ao
\[\Sigma^n\ni\sigma\mapsto w(\sigma)\in \R,\]
\'e Lipschitz cont\'\i nua com a constante $1$ em rela\c c\~ao \`a dist\^ancia de Hamming. 
\end{exemplo}

\begin{exemplo} 
O {\it peso normalizado} de uma palavra de comprimento $n$ \'e definido por
\[\bar w(\sigma) =\frac 1n w(\sigma).\]
Por exemplo,
\[\bar w(00110110)=\frac 12.\]
O peso normalizado $\bar w$ \'e uma fun\c c\~ao Lipschitz cont\'\i nua sobre $(\Sigma^n,\bar d)$
com $\mathrm{Lip}(\bar w)=1$.
\end{exemplo}

\begin{exemplo}
Visto como uma fun\c c\~ao 
\[w\colon (\Sigma^n,\bar d)\to \{0,1\},\]
o peso tem a constante de Lipschitz $n$, enquanto o peso normalizado visto como uma fun\c c\~ao 
\[\bar w\colon (\Sigma^n, d)\to \{0,1\}\]
tem a constante $L=\frac 1n$.
\end{exemplo}

A constru\c c\~ao seguinte \'e uma fonte importante de fun\c c\~oes Lipschitz cont\'\i nuas. 

\begin{definicao}
Seja $A$ um subconjunto n\~ao vazio de um espa\c co m\'etrico $(X,d)$. A {\em fun\c c\~ao dist\^ancia de $A$} \'e uma fun\c c\~ao definida para cada 
$x\in X$ por
\[d_A(x)\equiv \mathrm{dist}(x,A)=\inf_{a\in A} d(x,a).\]
Se $A$ \'e finito, o \'\i nfimo na defini\c c\~ao torna-se o m\'\i nimo.
\end{definicao}

\begin{observacao}
Claro que $A$ deve ser n\~ao vazio: a fun\c c\~ao  $d_\emptyset$ toma identicamente o valor $+\infty$...
\end{observacao}

\begin{lema}
\label{distance}
Cada fun\c c\~ao dist\^ancia, $d_A$, \'e Lipschitz cont\'\i nua com $L=1$.
\label{l:distanciadeA}
\end{lema}

\begin{proof}  
Sejam $x,y\in X$ quaisquer. Dado $\e>0$, existem elementos
$a,b\in A$ tais que $d(x,a)< d_A(x)+\e$ e $d(y,b)<d_A(y)+\e$.
Usando a defini\c c\~ao de $d_A$ e a desigualdade triangular, temos
\begin{align*}
d_A(x) & \leq d(x,b) \\
 & \leq d(x,y)+d(y,b)\\
 & <  d(x,y)+ d_A(y)+\e,
\end{align*}
e de maneira semelhante,
\[d_A(y)\leq d(x,y) + d_A(x)+\e.\]
Como as desigualdades s\~ao verdadeiras para cada $\e>0$, conclu\'\i mos que de fato
\[d_A(x)\leq d(x,y) + d_A(y)\mbox{
e }
d_A(y)\leq d(x,y) + d_A(x),\]
como desejado. 
\end{proof}

\begin{observacao}
A observa\c c\~ao \ref{everylip} pode ser invertida: um espa\c co m\'etrico $(X,d)$ \'e finito se e somente se toda fun\c c\~ao
$f\colon X\to\R$ \'e Lipschitz cont\'\i nua. Para estabelecer a implica\c c\~ao $\Leftarrow$, suponhamos que $X$ \'e infinito, ou seja, existe uma aplica\c c\~ao sobrejetora 
$\phi\colon X\to\N$. Escolhe um elemento $x_0\in X$ e define a fun\c c\~ao
\[f(x)=\phi(x)d(x,x_0).\]
Como $\phi(x)$ assume valores positivos arbitrariamente grandes,
a fun\c c\~ao $f$ n\~ao \'e Lipschitz cont\'\i nua.
\end{observacao}

\begin{definicao}
Seja $f\colon \Gamma\to\R$ uma fun\c c\~ao limitada sobre um conjunto n\~ao vazio, $\Gamma$. O valor da {\em norma}
$\ell^\infty$ de $f$ \'e dado por
\[\norm f_\infty = \sup_{x\in \Gamma}\abs{f(x)}.\]
Se $\Gamma$ for finito (no caso de nosso interesse, $\Gamma=\Sigma^n$ ser\'a o cubo de Hamming), o supremo \'e o m\'aximo.
\end{definicao}

Os fatos seguintes s\~ao muito claros.

\begin{proposicao}
A norma $\ell^\infty$ tem as seguintes propriedades.
\begin{enumerate}
\item $\norm f_\infty =0 \Leftrightarrow f\equiv 0$.
\item $\norm{\lambda f}_\infty =\abs{\lambda}\norm f_\infty$
para todas fun\c c\~oes $f$ e todos $\lambda\in\R$.
\item $\norm{f+g}_\infty\leq \norm f_\infty + \norm g_\infty.$
\item $\norm{fg}_{\infty}\leq \norm{ f}_\infty\cdot \norm{g}_{\infty}$ (Submultiplicatividade da norma $\ell^\infty$).\label{submulti}
\item $\norm{e^f}_\infty\leq e^{\norm f_\infty}$. \label{expnorm}
\end{enumerate}
(1)-(3) significam que $\norm\cdot_\infty$ \'e uma norma (se\c c\~ao \ref{s:normados}) sobre o espa\c co vetorial $\ell^{\infty}(\Gamma)$ das fun\c c\~oes limitadas sobre $\Gamma$.
\qed
\end{proposicao}

\section{Teorema isoperim\'etrico de Harper}

Nossa pr\'oxima estrutura sobre o cubo de Hamming nasce de uma observa\c c\~ao muito simples: dado um subconjunto $A\subseteq \Sigma^n$, podemos {\em contar} o n\'umero dos elementos dele. Este n\'umero, $\sharp A$, satisfaz as seguintes propriedades evidentes.

\begin{proposicao} 
$\,$
\begin{enumerate}
\item $\sharp\emptyset =0$, $\sharp\Sigma^n = 2^n$.
\item Se $A\cap B=\emptyset$, ent\~ao $\sharp(A\cup B)=\sharp A+\sharp B$.
\end{enumerate}
\end{proposicao}

Desse modo, pode-se dizer que a correspond\^encia $A\mapsto \sharp A$ \'e uma medida, chamada a {\em medida de contagem}.
\index{medida! de contagem}
Com efeito, como o conjunto $\Sigma^n$ \'e finito, toda fam\'\i lia de subconjuntos dois a dois disjuntos \'e necessariamente finita. Logo, a medida de contagem \'e mesmo sigma-aditiva: 

\begin{enumerate}
\item[(3)] {\em Seja $A_1,A_2,\ldots,A_n,\ldots$ uma sequ\^encia qualquer de subconjuntos do cubo dois a dois disjuntos. Ent\~ao,}
\[\sharp\left(\cup_{i=1}^\infty A_i\right) = \sum_{i=1}^\infty\sharp(A_i).\]
\end{enumerate}

A soma no lado direito \'e de fato finita, porque quase todos os termos se anulam.

\begin{exemplo}
A esfera $S_r(\sigma)$ de raio $r\geq 0$,
\[S_r(\sigma)=\{\tau\in\Sigma^n\colon d(\sigma,\tau)=r\},\]
\'e n\~ao-vazia se e somente se $r\in \{0,1,\ldots,n\}$. Neste caso, ela satisfaz
\[\sharp S_r(\sigma) = {n\choose r},\]
o coeficiente binomial.
De fato, a transla\c c\~ao $x\mapsto x+\sigma\mod 2$, estabelece a correspond\^encia bijetora entre a esfera $S_r(\sigma)$ e a esfera em torno de zero, $S_r(0)$, e a \'ultima pode ser identificada com a fam\'\i lia de todos os subconjuntos de $[n]=\{1,2,3,\ldots,n\}$ contendo exatamente $r$ elementos.
\end{exemplo}

\begin{exercicio}
Seja $k\in\N$. Usar a indu\c c\~ao em $k$ para mostrar que
\[{2k\choose k}\leq \frac{4^k}{\sqrt{3k+1}}.\]
Deduzir que, se $k$ \'e um n\'umero par, ent\~ao a medida normalizada da esfera de raio $k/2$ no cubo de Hamming $\{0,1\}^k$ satisfaz
\[\mu_{\sharp}(S_k)\leq \frac{1}{\sqrt{\frac 32 k+1}}.\]
\label{ex:tamanho_esfera}
\end{exercicio}

\begin{exemplo}
A bola fechada, $\bar B_d(\sigma)$, de raio inteiro $d\geq 0$, \'e a uni\~ao das esferas:
\[\bar B_d(\sigma)=\bigcup_{i=0}^d S_i(\sigma).\]
Como consequ\^encia, temos
\begin{equation}
\sharp \bar B_r(\sigma) = \sum_{i=0}^d {n \choose i}.
\label{eq:tamanho_bola}
\end{equation}
\end{exemplo}

A estimativa na Eq. (\ref{eq:tamanho_bola}) n\~ao \'e muito conveniente, e por isso vamos obter duas cotas superiores para ela, as quais ser\~ao usadas em dois contextos diferentes. 

\begin{lema}[Desigualdade de Euler]
Para todo $x>0$ e $a\in\R$, $a\neq 0$,
\begin{equation}
\left(1+\frac{a}{x}\right)^{x}< e^a.\end{equation}
\index{desigualdade! de Euler}
\end{lema}

Este resultado \'e mostrado nos cursos de C\'alculo elementar do primeiro ano, no momento quando a constante de Euler, $e$, for definida como o montante no banco que um investidor ter\'a um ano ap\'os ser investido R\$ 1 \`a taxa de juro de 100 \% aplicada continuamente -- em outras palavras, como o limite
\[e=\lim_{x\to\infty}\left(1+\frac{1}{x}\right)^{x}.\]
O limite existe pois a fun\c c\~ao sob o sinal do limite \'e estritamente crescente e limitada acima. A desigualdade de Euler segue-se. 

A seguir, apresentamos a primeira estimativa do tamanho da bola no cubo de Hamming.

\begin{lema}
Sejam $1\leq d\leq n$. Ent\~ao
\begin{equation}
\sharp \bar B_r(\sigma)<\left(\frac{en}d\right)^d.
\end{equation}
\label{l:<}
\end{lema}

\begin{proof}
Para todo $0\leq i\leq d$, temos
\[\left(\frac nd\right)^d\left(\frac d n\right)^i=\left(\frac nd\right)^{d-i}\geq 1,\]
e por conseguinte, usando a f\'ormula binomial e a desigualdade de Euler, conclu\'\i mos:
\begin{align*}
\sum_{i=0}^d{n \choose i} &\leq \left(\frac nd\right)^d\sum_{i=0}^d{n \choose i} \left(\frac d n\right)^i\\
&\leq  \left(\frac nd\right)^d\sum_{i=0}^n{n \choose i}\left(\frac d n\right)^i \\
&=  \left(\frac nd\right)^d\left(1+\frac d n\right)^n
\\
&< \left(\frac nd\right)^de^d \\
&= \left(\frac{en}d\right)^d.
\end{align*}
\end{proof}

\begin{definicao}
Dado um subconjunto $A\subseteq\Sigma^n$ e um n\'umero natural $r$, denotemos
\[\Gamma_rA = \{\sigma\in\Sigma^n\colon \exists \alpha\in A,~d(\sigma,\alpha)\leq r\}\]
a $r$-vizinhan\c ca fechada de $A$ (em rela\c c\~ao \`a dist\^ancia de Hamming n\~ao normalizada). 
\end{definicao}

\begin{exercicio}
Verificar que a $r$-vizinhan\c ca de uma bola fechada \'e uma bola fechada:
\[\Gamma_r(B_d(\sigma)) = B_{d+r}(\sigma).\]
\end{exercicio}

O resultado principal desta se\c c\~ao diz que entre todos os subconjuntos do cubo de Hamming, as bolas possuem as menores vizinhan\c cas. Precisamos da defini\c c\~ao seguinte.

\begin{definicao}
Um subconjunto $B$ de $\Sigma^n$ \'e chamado uma {\em bola de Hamming} se existem $r\in\N$ e $\sigma\in\Sigma^n$ tais que
\[B_r(\sigma)\subseteq B \subseteq B_{r+1}(\sigma).\]
\index{bola! de Hamming}
\end{definicao} 

Em outras palavras, uma bola de Hamming \'e uma bola fechada mais um peda\c co da esfera de raio mais um, ou bem uma regi\~ao compreendida entre a bola de centro $\sigma$ e raio $r$, e a bola de centro $\sigma$ e raio $r+1$.

\begin{teorema}[Desigualdade isoperim\'etrica de Harper]
Para qualquer $A\subseteq\Sigma^n$ subconjunto n\~ao vazio e $r\in\N$, com $r\leq n$, existe uma bola de Hamming $B$ satisfazendo $\sharp B = \sharp A$ e $\sharp\Gamma_rB\leq\sharp\Gamma_rA$. 
\index{desigualdade! isoperim\'etrica! de Harper}
\end{teorema}

A prova (mais ou menos seguindo \citep*{FF}) ocupa o resto da se\c c\~ao.

\begin{definicao}
A {\em dist\^ancia entre dois subconjuntos n\~ao-vazios} $A$ e $B$ de um espa\c co m\'etrico $X$ \'e a quantidade
\[d(A,B)=\inf\{d(a,b)\colon a\in A, b\in B\}.\]
(Se pelo menos um de $A,B$ \'e compacto, por exemplo finito, ent\~ao temos o m\'\i nimo).
\end{definicao}

A express\~ao ``dist\^ancia'' \'e um pouco enganosa, porque a dist\^ancia entre dois conjuntos n\~ao satisfaz necessariamente a desigualdade triangular.

Denotando
\[B = \Sigma^n\setminus\Gamma_rA,\]
temos que $d(A,B)\geq r+1$. 
Basta mostrar o resultado seguinte.

\begin{lema} Dado dois subconjuntos $A$ e $B$ de $\Sigma^n$ e um n\'umero natural $r$, com a propriedade $d(A,B)\geq r$, existem uma bola de Hamming $\tilde A$ em torno de $0$ e uma bola de Hamming $\tilde B$ em torno de $1$, tais que $\sharp A=\sharp\tilde A$, $\sharp B=\sharp\tilde B$, e $d(\tilde A,\tilde B)\geq r$.
\label{l:ab}
\end{lema}

Uma vez que o lema esteja mostrado, teremos $\Gamma_r\tilde A\cap\tilde B=\emptyset$, de onde $\sharp\Gamma_r\tilde A\leq 2^n - \sharp\tilde B$, e por conseguinte
\[\sharp\Gamma_rA =2^n- \sharp B =2^n-\sharp\tilde B \geq \sharp\Gamma_r\tilde A,\]
como exigido. 

Vamos executar uma sequ\^encia de transforma\c c\~oes simult\^aneas dos conjuntos $A$ e $B$ que n\~ao aumentem o tamanho de suas vizinhan\c cas e que os tornam cada vez mais ``redondos'' de maneira que no final obtemos uma bola de Hamming em torno de $0$ e uma bola de Hamming em torno de $1$, respetivamente. 

\begin{exercicio}
Observe que um subconjunto n\~ao vazio $A\subseteq\Sigma^n$ \'e uma bola de Hamming em torno de $0$, se e somente se, quaisquer que sejam $\alpha,\beta\in\Sigma^n$, se $\alpha\in A$ e $w(\beta)<w(\alpha)$, ent\~ao $\beta\in A$. De uma forma rigorosamente equivalente: para cada $\alpha\in A$ 
 e  $\tau\in\Sigma^n$, se $w(\alpha+\tau)< w(\alpha)$, ent\~ao $\alpha+\tau\in A$. 

De mesma maneira, um conjunto $B\neq\emptyset$ \'e uma bola de Hamming em torno de $1$ se e somente se, qualquer que sejam $\alpha\in A$ e $\tau\in\Sigma^n$, se $w(\alpha+\tau)> w(\alpha)$, ent\~ao $\alpha+\tau\in A$. 
\end{exercicio}

Dado $A$ e $B$ como no lema \ref{l:ab}, escolhamos $\tau$ do menor peso tal que ou existe $\alpha\in A$ com $w(\alpha+\tau)<w(\alpha)$ e $\alpha+\tau\notin A$, ou existe $\beta\in B$ com $w(\beta+\tau)>w(\beta)$ e $\beta+\tau\notin B$. (Se um tal $\tau$ n\~ao existe, ent\~ao trata-se de duas bolas de Hamming em torno de $0$ e $1$, e n\~ao h\'a nada para mostrar). Definamos a aplica\c c\~ao $T=T_{A,B,\tau}$ de $A\cup B$ para $\Sigma^n$ como segue:
\[\mbox{se }\alpha\in A,~T(\alpha)=\begin{cases} &\alpha+\tau,\mbox{ se }w(\alpha+\tau)<w(\alpha)\mbox{ e }\alpha+\tau\notin A,\\
& \alpha,\mbox{ de outro modo;}
\end{cases}\]
\[\mbox{se }\beta\in B,~T(\beta)=\begin{cases} &\beta+\tau,\mbox{ se }w(\beta+\tau)>w(\beta)\mbox{ e }\beta+\tau\notin B,\\
& \beta,\mbox{ de outro modo.}
\end{cases}\]
\'E claro que as restri\c c\~oes de $T$ sobre $A$ e sobre $B$ s\~ao injetoras. Pode-se mostrar facilmente que se $a\in A$, $b\in B$, e $T(a)=b$, ent\~ao $T(b)=a$. De fato, isso n\~ao \'e necess\'ario, pois vamos mostrar em breve que $d(T(A),T(B))\geq r$. 

Al\'em disso, se $\alpha\in A$, $\beta\in B$, ent\~ao $w(T(\alpha))\leq w(\alpha)$, $w(T(\beta))\geq w(\beta)$, o que significa que o operador $T$ n\~ao aumenta a quantidade
\begin{equation}
\sum_{\alpha\in A}w(\alpha)-\sum_{\beta\in B}w(\beta),
\label{eq:quantidade}
\end{equation}
e de fato, 
existe pelo menos um elemento $\gamma\in A\cup B$ que diminui esta quantidade em pelo menos $1$. O valor absoluto da express\~ao (\ref{eq:quantidade}) n\~ao excede $n2^n$.
Isso significa que o procedimento acima (a escolha de $\tau$ e da aplica\c c\~ao de $T$) s\'o pode ser repetida um n\'umero finito de vezes, ap\'os o qual chegaremos \`as duas bolas satisfazendo as conclus\~oes do lema.

A \'unica coisa restante a verificar \'e a condi\c c\~ao
\[d(T(A),T(B))\geq r.\]
Em outras palavras, quaisquer que sejam $a\in A$, $b\in B$, temos $d(T(a),T(b))\geq r$.
Temos tr\^es casos distintos.

(1) Ambos os elementos $T(a),T(b)$ s\~ao ``antigos'', isso \'e, $a\in A\cap T(A)$, $b\in B\cap T(B)$. Neste caso h\'a nada a mostrar, pois $d(a,b)\geq r$.

(2) Ambos os elementos s\~ao ``novos'': $a^\prime\in T(A)\setminus A$, $b^\prime\in T(B)\setminus B$. Neste caso, $a^\prime=a+\tau$, $b^\prime=b+\tau$, onde $a\in A$, $b\in B$, e temos
\[d(a^\prime,b^\prime)=d(a+\tau,b+\tau)=d(a,b)\geq r,\]
pois a dist\^ancia de Hamming \'e invariante pelas transla\c c\~oes.

(3) Um elemento \'e ``novo'' e outro, ``antigo''. Suponha que $a^\prime\in T(A)\setminus A$ e $b\in B\cap T(B)$. (O caso sim\'etrico $a\in A\cap T(A)$ e $b^\prime\in B\setminus T(B)$ \'e um exerc\'\i cio.) 

Se $b+\tau\in B$, ent\~ao podemos concluir:
\[d(a^\prime,b)=d(a+\tau,b)=d(a,b+\tau)\geq r,\]
pois $a\in A$ e $b+\tau\in B$. Ent\~ao, a partir desse momento, suponhamos que $b+\tau\notin B$. Como $T(b)=b$, conclu\'\i mos que 
\[w(b+\tau)\leq w(b).\]
A adi\c c\~ao de $\tau$ tem efeito de alterar todos os bits em $\supp\tau$. 
A desigualdade acima significa que
o n\'umero de zeros em $b\tau$ \'e menor ou igual ao n\'umero de uns em $b\tau$, ou seja:
\begin{equation}
w(b\tau) = d(0,b\tau)\geq d(\tau,b\tau).
\label{eq:btau}
\end{equation}

Subcaso (3a): 
$\supp b\tau \subseteq\supp a$, ou seja, para todo $i$, se $\tau_i=1$ e $b_i=1$, ent\~ao $a_i=1$. 
Escrevamos $b=b_1+b\tau$. Agora, $a^\prime$ e $b_1$ t\^em zeros em todas as posi\c c\~oes de $\supp b\tau=\supp\tau\cap\supp b$, e por isso:
\begin{align*}
d(a^\prime,b) & = d(a^\prime,b_1)+w(b\tau)\\
&\geq
 d(a^\prime,b_1)+d(\tau,b\tau)\\
&=
d(a^\prime, b_1+b\tau) + d(b+\tau,b+b\tau) \\
 &\geq d(a^\prime,b+\tau) \\
&= d(a,b)\geq r.
\end{align*}

Subcaso (3b): $\supp\tau b\not\subseteq\supp a$, ou seja, existe $i$ tal que $\tau_i=1$, $b_i=1$, e $a_i=0$. Escolhemos um tal \'\i ndice $i$.
Como $w(a+\tau)<w(a)$, o n\'umero de uns em $a\tau$ \'e estritamente maior do que o n\'umero de zeros. Em outras palavras, $\sharp(\supp a\cap \supp\tau)> (1/2)\sharp\supp\tau$.
Relembremos que o n\'umero de uns em $b\tau$ \'e maior ou igual ao n\'umero de zeros, ou seja, $\sharp(\supp b\cap\supp\tau)\geq (1/2)\sharp\supp\tau$. Por conseguinte, os subconjuntos $\supp a\cap \supp\tau$ e $\supp b\cap \supp\tau$ se encontram, e existe $j$ tal que $\tau_j=1$, $b_j=1$, e $a_j=1$. Escolhamos um tal $j$.
Elimine do suporte de $\tau$ os \'\i ndices $i$ e $j$, para obter a palavra $\tilde\tau$ com $\supp\tilde\tau=\supp\tau\setminus\{i,j\}$. \'E claro que $w(a+\tilde\tau)=w(a+\tau)<w(a)$. Como $\tau$ tem o peso minimal, conclu\'\i mos que $a+\tilde\tau\in A$, logo $d(a+\tilde\tau,b)\geq r$. A palavra $a^\prime=a+\tau$ s\'o difere de $a+\tilde\tau$ em coordenadas $i$ e $j$. Temos $(a+\tilde\tau)_i=a_i=0$ e $b_i=1$, de onde $\abs{(a+\tilde\tau)_i)-b_i}=1$. Ao mesmo tempo, $a^\prime_i=1$, de onde $\abs{a^\prime_i-b_i}=0$. De mesma maneira, $\abs{(a+\tilde\tau)_j)-b_j}=0$ e $\abs{a^\prime_j-b_j}=1$. Conclu\'\i mos: 
\[d(a^\prime,b) = d(a+\tilde\tau, b)\geq r.\]

\begin{observacao} Eis o exemplo de espa\c co m\'etrico finito natural e importante, de fato semelhante ao cubo de Hamming $\{0,1\}^n$, cujo problema isoperim\'etrico parece ainda estar em aberto. 

O grupo sim\'etrico $S_n$
\index{grupo! sim\'etrico}
 de posto $n$ consiste de todas as autobije\c c\~oes do conjunto $[n]=\{1,2,\ldots,n\}$ com $n$ elementos
\index{n@$[n]$} 
(este $n$ entre colchetes \'e uma nota\c c\~ao da teoria combinat\'oria), munido da lei de composi\c c\~ao de aplica\c c\~oes.
Definamos a {\em m\'etrica de Hamming} sobre $S_n$ como segue:
\[d(\sigma,\tau)=\sharp\{i\in [n]\colon \sigma(i)\neq\tau(i)\}.\]
Pode se verificar facilmente que $d$ \'e uma m\'etrica {\em bi-invariante,} ou seja: quaisquer que sejam $\sigma,\tau,\alpha\in S_n$,
\[d(\alpha\sigma,\alpha\tau)=d(\sigma,\tau)= d(\sigma\alpha,\tau\alpha).\]
Quais conjuntos $B$ possuem o menor tamanho da $r$-vizinhan\c ca $\Gamma_rB$ entre todos os conjuntos $A$ com $\sharp A=\sharp B$? Se sabe que as bolas fechadas em $S_n$ {\em n\~ao t\^em} esta propriedade.
\end{observacao}

\section{Cubo de Hamming como espa\c co probabil\'\i stico}
\subsection{Medida de contagem normalizada}

A cardinalidade do cubo de posto $n$ \'e igual a $2^n$, e com o intuito de conseguir trabalhar com os cubos de postos diferentes, $\Sigma^n$ e $\Sigma^m$, simultaneamente, vamos realizar a normaliza\c c\~ao.

\begin{definicao}
Seja $A\subseteq\Sigma^n$ um subconjunto qualquer. A {\it medida de contagem normalizada} de $A$ \'e dada por
\[\mu_\sharp(A)=\frac{\sharp A}{2^n}.\]
\index{medida! de contagem! normalizada}
\end{definicao}

\begin{exemplo}
A medida normalizada de um conjunto unit\'ario \'e
\[\mu_\sharp(\{\sigma\})=\frac 1{2^n}.\]
\end{exemplo}

As propriedades da medida de contagem normalizada s\~ao seguintes.

\begin{proposicao} 
$\,$
\begin{enumerate}
\item $\mu_\sharp(\emptyset) =0$, $\mu_\sharp(\Sigma^n)= 1$.
\item Se $A\subseteq B$, ent\~ao $\mu_\sharp(A)\leq\mu_\sharp(B)$.
\item Se $A\cap B=\emptyset$, ent\~ao $\mu_\sharp(A\cup B)=
\mu_\sharp (A)+\mu_\sharp (B)$.
\end{enumerate}
\end{proposicao}

Podemos substituir a condi\c c\~ao (3) por uma condi\c c\~ao formalmente mais forte, a {\em sigma-aditividade:} dada uma sequ\^encia qualquer de subconjuntos dois a dois disjuntos, $A_1,A_2,\cdots$ de $\Sigma^n$, temos
\[\mu_\sharp(\cup_i A_i)=\sum_i\mu_\sharp(A_i).\] 
\'E verdadeiro porque temos $A_i=\emptyset$ a partir de um $i$ suficientemente grande.

\begin{observacao}
A condi\c c\~ao
\[\mu_{\sharp}\left(\Sigma^n\right)=1\]
significa que $\mu_{\sharp}$ \'e uma {\em medida de probabilidade.} Do ponto de vista probabil\'\i stico, $\Sigma^n$ \'e o {\em espa\c co de eventos elementares,} ou o {\em espa\c co amostral,} que corresponde ao experimento aleat\'orio de jogar uma moeda equilibrada $n$ vezes. A sequ\^encia bin\'aria codifica o resultado do experimento, com, por exemplo, $0$ correspondente \`a cara e $1$ \`a coroa. Um {\em evento} \'e um subconjunto qualquer de $A\subseteq\Sigma^n$. Por exemplo, o subconjunto
\[A=\{\sigma\mid \sigma_1=0\}\]
corresponde ao evento ``no primeiro lan\c camento, a moeda tirou a cara''. Neste contexto, a medida de contagem normalizada de um conjunto se interpreta como a {\em probabilidade} do evento $A$:
\[\mu_\sharp(A) = P(A).\]
Para um tratamento um pouco mais sistem\'atico das medidas de probabilidade, veja ap\^endice \ref{a:medidas}.
\end{observacao}

\begin{exercicio}
Sejam $n,r\in\N$, onde $r<n/2$. Mostrar que a medida normalizada de contagem da bola fechada $\bar B_r$ no cubo $\Sigma^{n+1}$ \'e estritamente menor do que a medida $\mu_{\sharp}$ da bola fechada de mesmo raio no cubo $\Sigma^{n}$. 
\par
[ {\em Sugest\~ao:} ${n+1\choose i} = {n\choose i} + {n\choose i-1}$. ]
\label{ex:bola_menor}
\end{exercicio}

A medida de contagem (como qualquer outra medida) pode ser usada para definir uma integral. Vamos integrar fun\c c\~oes sobre o cubo de Hamming. 

\begin{definicao}
Seja $f\colon\Sigma^n\to\R$ uma fun\c c\~ao {\it qualquer}. A
{\it integral} de $f$ com rela\c c\~ao a medida de contagem normalizada \'e o n\'umero
\[\int_{\Sigma^n}f(\sigma) d\mu_\sharp(\sigma) =
\frac{\sum_{\sigma\in\Sigma^n}f(\sigma)}{2^n}.\]
Em outras palavras, \'e o {\em valor m\'edio} de $f$. 

Na tradi\c c\~ao probabil\'\i stica, a integral \'e chamada a {\em esperan\c ca} de $f$ (ou: da uma vari\'avel aleat\'oria cuja realiza\c c\~ao \'e $f$), e denotada $\E(f)$, ou $\E_{\mu_{\sharp}}(f)$, se for necess\'ario destacar a medida usada.
\end{definicao}

As seguintes propriedades da esperan\c ca s\~ao imediatas.

\begin{proposicao}
$\,$
\begin{enumerate}
\item $\E(f+g)=\E(f)+\E(g)$.
\item $\E(\lambda f)=\lambda \E(f)$.
\item Se $f(\sigma)\geq 0$ para todos $\sigma$, ent\~ao
$\E(f)\geq 0$.
\item $\E(\mathbf{1})=1$.
\end{enumerate}
\end{proposicao}

\begin{exemplo}
Mostrar que a esperan\c ca do peso normalizado, $\bar w$, \'e igual \`a
$\frac 12$. 
\end{exemplo}

Para mais sobre a integral e esperan\c ca, consulte o ap\^endice \ref{a:integral}.
 
\begin{lema}[Desigualdade de Markov]
\label{majorize}
Seja $f$ uma fun\c c\~ao real sobre um conjunto finito $X$ tal que $f$ assume somente valores n\~ao-negativos. Ent\~ao 
\[\mu_\sharp\{x\in X\mid f(x)\geq 1\}\leq
\E(f).\]
\index{desigualdade! de Markov}
\end{lema}

\begin{proof}
Denotemos
\[A=\{x\in X\mid f(x)\geq 1\}.\]
Agora,
\begin{align}
\E(f) &= \int_Xf(x)d\mu_\sharp(x) \nonumber \\ 
&=
\int_Af(x)d\mu_\sharp(x) +\int_{X\setminus A}f(x)d\mu_\sharp(x) 
\nonumber \\
&\geq  \mu_\sharp(A).
\end{align}
\end{proof}

Al\'em da esperan\c ca, pode-se definir o valor mediano. 

\begin{definicao}
Seja $f$ uma fun\c c\~ao real sobre um conjunto finito, $X$. Um n\'umero $M\in\R$ \'e chamado um {\em valor mediano} de $f$ se ele satisfaz 
as seguintes condi\c c\~oes:
\[\mu_\sharp\{x\in X\colon f(x)\geq M\}\geq\frac 12
\mbox{ e } 
\mu_\sharp\{x\in X\colon f(x)\leq M\}\geq\frac 12.\]
\label{medianvalue}
\index{valor! mediano}
\end{definicao}

\begin{proposicao}
Cada fun\c c\~ao sobre um conjunto finito tem um valor mediano com rela\c c\~ao \`a medida de contagem normalizada.
\end{proposicao}

\begin{proof}
\'E f\'acil ordenar $X$,
\[X=\{x_1,x_2,\dots,x_n\},\]
de tal maneira que $f(x_i)\leq f(x_j)$ quando $i\leq j$.
Agora $M=f(x_{\lceil n/2\rceil})$ \'e um valor mediano de $f$.
\end{proof}

Aqui, $\lceil x\rceil$ denota o {\em teto} de um n\'umero real, $x$: 
\[\lceil x\rceil = \min\{n\in\Z\colon x\leq n\}.\]
\index{teto}
\index{teto@$\lceil x\rceil$}

\begin{exemplo}
O valor $\frac 12$ \'e um valor mediano do peso normalizado $\bar w$.
\end{exemplo}

\begin{observacao}
Intuitivamente, o valor mediano tem a propriedade de que os eventos $f(x)\leq M$ e $f(x)\geq M$ ocorrem com a probabilidade $1/2$ cada um. Portanto, os exemplos f\'aceis mostram que a igualdade n\~ao \'e sempre poss\'\i vel. 
\end{observacao}

\begin{observacao}
O valor mediano n\~ao \'e, em geral, \'unico. Mais propriamente, os valores medianos de uma fun\c c\~ao enchem um intervalo fechado. Por exemplo, cada n\'umero real estritamente entre $0$ e $1$ \'e um valor mediano da fun\c c\~ao $\pi_1$ a qual envia uma palavra para seu primeiro bit. 
\end{observacao}

\begin{observacao}
Geralmente, a esperan\c ca de uma fun\c c\~ao n\~ao tem de ser um valor mediano.
Seja $x_0\in X$, e seja $f$ a fun\c c\~ao delta localizada em $x_0$:
\[f(x)=\begin{cases} 0, & \mbox{if } x\neq x_0, \\
1, & \mbox{if } x=x_0.
 \end{cases}
 \]
Ent\~ao, $\E(f)=\frac 1n$, mais neste caso o \'unico valor mediano  \'e $M=0$. 
\end{observacao}

Como uma das aplica\c c\~oes da pequena teoria a ser desenvolvida, vamos mostrar que se $f$ \'e uma fun\c c\~ao real Lipschitz cont\'\i nua com $L=1$ sobre o cubo de Hamming $\Sigma^n$, ent\~ao a esperan\c ca e qualquer valor mediano de $f$ s\~ao muito pr\'oximos, $\abs{\E f-M_f}\leq 1/n$.

\subsection{Parti\c c\~oes e esperan\c ca condicional}

\begin{definicao}
Seja $X$ um conjunto. Uma {\it cobertura} de $X$ \'e uma fam\'\i lia $\gamma$ de subconjuntos n\~ao vazios de $X$ cuja uni\~ao \'e $X$:
\[\cup\gamma = X.\]
Em outras palavras, todo elemento de $X$ \'e contido em pelo menos um elemento de $\gamma$:
\[\forall x\in X,~\exists A\in\gamma,\mbox{ tal que }x\in A.\]
\index{cobertura}
\end{definicao}

\begin{definicao}
Uma {\it parti\c c\~ao}, $\Omega$, de um conjunto $X$ \'e uma cobertura de $X$,
\[\cup\Omega = X,\]
cujos elementos s\~ao dois a dois disjuntos:
\[A\cap B=\emptyset \mbox{ se $A,B\in\Omega$, $A\neq B$.}\]
Nesse caso, todo elemento de $X$ \'e contido em exatamente um elemento de $\Omega$:
\[\forall x\in X,~\exists ! A\in\Omega,\mbox{ tal que }x\in A.\]
\index{parti\c c\~ao}
\end{definicao}

\begin{exemplo}
A parti\c c\~ao do intervalo unit\'ario $[0,1)$ em subintervalos,
\[\Omega=\left\{\left[\frac in,\frac{i+1}n\right) 
\mid i=0,1,\dots,n-1\right\},\]
\'e importante na teoria de integra\c c\~ao de Riemann.
\end{exemplo}

\begin{exemplo}
A {\it parti\c c\~ao trivial} de $X$:
\[\Omega=\{X\}.\]
\end{exemplo}

\begin{exemplo}
A {\it parti\c c\~ao mais fina} de $X$:
\[\Omega_f(X)=\{\{x\}\mid x\in X.\}\]
\end{exemplo}

As parti\c c\~oes que nos interessam s\~ao as do cubo de Hamming. 

\begin{definicao}
Sejam $n\in\N_+$ e $k\leq n$ n\'umeros naturais.
A aplica\c c\~ao
\[\pi^n_k\colon \Sigma^n\to\Sigma^k\]
\'e definida por
\[\pi^n_k(\sigma_1\sigma_2\dots\sigma_k\sigma_{k+1}\dots\sigma_n)
=\sigma_1\sigma_2\dots,\sigma_k.\]
Dada uma palavra $\sigma$ de comprimento $n$, a aplica\c c\~ao $\pi^n_k$ acima trunca esta palavra, deixando somente o prefixo dos $k$ primeiros bits. 
\end{definicao}

\begin{exemplo}
$\pi^6_3(001011) = 001$.
\end{exemplo}

\begin{definicao}
\label{d:standard}
Uma {\em parti\c c\~ao padr\~ao} $\Omega_k$ (ou, mais exatamente, 
$\Omega_k(\Sigma^n)$) do cubo de Hamming $\Sigma^n$ consiste de subconjuntos
\[A_\tau=\{\sigma\in\Sigma^n \mid \pi^n_k(\sigma)=\tau\},\]
onde $\tau\in\Sigma^k$. 
\end{definicao}

Os membros da parti\c c\~ao $\Omega_k$ s\~ao definidos pelos primeiros $k$ bits. 

\begin{exemplo}
$\Omega_1(\Sigma^n)=\{\{\sigma\mid \sigma_1=0\},
\{\sigma\mid \sigma_1=1\}\}$.
\end{exemplo}

\begin{exemplo}
$\Omega_0(\Sigma^n)$ \'e a parti\c c\~ao trivial.
\end{exemplo}

\begin{exemplo}
$\Omega_n(\Sigma^n)$ \'e a parti\c c\~ao mais fina de $\Sigma^n$.
\end{exemplo}

\begin{definicao}
Sejam $\Omega$ e $\Psi$ duas parti\c c\~oes de um conjunto $X$.
Dizemos que $\Omega$ {\it refina} $\Psi$ (ou que 
$\Omega$ \'e {\it mais fina} do que $\Psi$, ou ainda que $\Psi$ \'e {\em mais grossa} do que $\Omega$), se todo elemento $A\in\Omega$ est\'a contido num elemento
$B\in\Psi$:
\[A\subseteq B.\]
Nota\c c\~ao:
\[\Omega \prec \Psi.\]
\end{definicao}

Informalmente, \'e o caso onde todos elementos da parti\c c\~ao 
$\Omega$ s\~ao obtidos quebrando os elementos de $\Psi$ em peda\c cos menores. 

\begin{exemplo}
Toda parti\c c\~ao refina-se a si mesma: $\Omega\prec\Omega$.
\end{exemplo}

\begin{exemplo}
Qualquer parti\c c\~ao \'e mais fina que a parti\c c\~ao trivial $\{X\}$:
\[\Omega\prec \{X\}.\]
\end{exemplo}

\begin{exemplo}
A parti\c c\~ao mais fina, $\Omega_f(X)$, justifica plenamente o seu nome refinando cada outra parti\c c\~ao $\Omega$ de $X$:
\[\Omega_f(X)\prec\Omega.\]
\end{exemplo}

\begin{definicao} Seja $\Omega$ uma parti\c c\~ao de um conjunto finito $X$ qualquer, e seja $f\colon X\to \R$ uma fun\c c\~ao. A {\it esperan\c ca condicional} de $f$ dada $\Omega$ \'e uma fun\c c\~ao
de $X$ para $\R$, denotada $\E(f\mid\Omega)$, definida da seguinte maneira.
Para cada $A\in\Omega$, a fun\c c\~ao $\E(f\mid\Omega)$ assume um valor 
constante sobre $A$, igual ao valor m\'edio de $f$ sobre $A$:
\begin{align*}
\E(f\mid\Omega)(a) &= 
\frac{\sum_{x\in A}f(x)}{\sharp(A)},
\end{align*}
qualquer que seja $a\in A$.
\index{esperan\c ca! condicional}
\label{d:esperancacondicional}
\end{definicao}

\begin{exemplo} A esperan\c ca condicional de $f$ dada a parti\c c\~ao trivial de $X$ \'e uma fun\c c\~ao constante assumindo o valor $\E(f)$. Podemos escrever:
\[\E(f\mid\{X\})=\E(f).\]
\end{exemplo}

\begin{exemplo}
A esperan\c ca condicional de uma fun\c c\~ao $f$ dada a parti\c c\~ao mais fina \'e a pr\'opria fun\c c\~ao $f$:
\[\E(f\mid\Omega_f(X))=f.\]
\end{exemplo}

As seguintes propriedades da esperan\c ca condicional s\~ao de simples verifica\c c\~ao.

\begin{proposicao} Sejam $f$ e $g$ duas fun\c c\~oes reais sobre um conjunto finito $X$ qualquer,
e seja $\Omega$ uma parti\c c\~ao de $X$.
\begin{enumerate}
\item \label{mono}
Se $f\leq g$, ent\~ao $\E(f\mid\Omega)\leq \E(g\mid \Omega)$.
\item Se $g=\E(g\mid\Omega)$ (ou seja, se $g$ \'e constante
sobre os elementos de $\Omega$), ent\~ao \\
$\E(gf\mid \Omega)=g\E(f\mid\Omega)$.
\label{factorout}
\item Em particular,
$\E(\lambda f\mid\Omega)=\lambda \E(f\mid\Omega)$.
\item $\E(f+g\mid\Omega)= \E(f\mid\Omega)+\E(g\mid\Omega)$.
\end{enumerate}
\label{properties}
\qed
\end{proposicao} 

\begin{lema}
\label{normest}
$\norm{\E(f\mid\Omega)}_{\infty}\leq\norm f_\infty$. \qed
\end{lema}

\begin{proposicao}
\label{conseq}
Sejam $\Psi\prec\Omega$ duas parti\c c\~oes de um conjunto finito $X$
e seja $f\colon X\to\R$. Ent\~ao,
\[\E(f\mid\Omega) = \E(\E(f\mid\Psi)\mid\Omega).\]
Em particular,
\[\E(f) = \E(\E(f\mid \Psi)).\]
\end{proposicao}

\begin{proof}
Dado $x\in X$, precisamos mostrar que 
\[\E(f\mid\Omega)(x) = \E(\E(f\mid\Psi)\mid\Omega)(x)\]
Seja $A\in\Psi$ um elemento contendo $x$. Denotemos 
\[\Xi=\{B\in\Omega\colon B\subseteq A\}.\]
Este $\Xi$ \'e uma parti\c c\~ao de $A$. Temos:
\begin{align*}
\E(\E(f\mid\Psi)\mid\Omega)(x) 
&= \abs{A}^{-1}\sum_{a\in A}\E(f\mid\Psi)(a) \\
&=\abs{A}^{-1}\sum_{B\in\Xi}\sum_{b\in B} \E(f\mid\Psi)(b)
\\ &=
\abs{A}^{-1}\sum_{B\in\Xi}\abs{B} \E(f\mid\Psi)(b) \\
&= \abs{A}^{-1}\sum_{B\in\Xi}\abs{B}\left(\abs{B}^{-1}
\sum_{b\in B}f(b)\right) \\
&= \abs{A}^{-1}\sum_{B\in\Xi}\sum_{b\in B}f(b) \\
&= \abs{A}^{-1}\sum_{a\in A} f(a) \\
&= \E(f\mid\Omega)(x).
\end{align*}
\end{proof}

\subsection{Martingales\label{ss:martingales}}

\begin{definicao} Uma {\it sequ\^encia refinadora de parti\c c\~oes} de um conjunto finito $X$ \'e uma sequ\^encia $(\Omega_k)_{k=0}^n$ onde
$\Omega_0$ \'e uma parti\c c\~ao trivial, $\Omega_n$ \'e a 
parti\c c\~ao mais fina, e cada parti\c c\~ao est\'a refinando a parti\c c\~ao precedente:
\[\Omega_n\prec\dots\prec \Omega_{k+1}\prec\Omega_k\prec
\Omega_{k-1}\prec\dots\prec\Omega_0.\]
\end{definicao}

\begin{exemplo} No cubo de Hamming, temos a sequ\^encia refinadora can\^onica de parti\c c\~oes:
\[\{\Sigma^n\}=\Omega_0(\Sigma^n)\succ \Omega_1(\Sigma^n)\succ \dots \succ
\Omega_k(\Sigma^n)\succ\dots \succ \Omega^n(\Sigma^n)=\Omega_f(\Sigma^n).\]
\label{refiningcube}
\end{exemplo}

\begin{definicao}
Um {\it martingale}
 (formado em rela\c c\~ao a uma sequ\^encia refinadora de parti\c c\~oes de um conjunto finito $X$) \'e uma cole\c c\~ao de fun\c c\~oes reais $(f_0,f_1,\dots,f_n)$
sobre $X$ tais que para todo $i=1,2,\dots,n$,
\[E(f_i\mid\Omega_{i-1})=f_{i-1}.\]
\index{martingale}
\end{definicao}

\begin{observacao}
A palavra {\em martingale} (masculino; vers\~ao mais rara: {\em martingala,} feminino) chegou no portugu\^es do franc\^es ({\em martingale,} feminino), e no franc\^es, em sua vez, do proven\c cal ({\em martegalo,} masculino). O significado original da palavra foi uma estrat\'egia de jogo fora de usual, pouco razo\'avel. Veja \citep*{mansuy}.
\end{observacao}

\begin{lema}
Seja $(f_0,f_1,\dots,f_n)$ um martingale sobre $X$ formado em rela\c c\~ao a uma sequ\^encia $(\Omega_k)_{i=1}^n$ refinadora de parti\c c\~oes. Para quaisquer que sejam $i\leq j$, temos
\[\E(f_j\mid\Omega_i)=f_i.\]
\end{lema}

\begin{proof}
Segue-se pela indu\c c\~ao matem\'atica da Proposi\c c\~ao \ref{conseq}.
\end{proof}

O resultado seguinte \'e imediato, segue por indu\c c\~ao. 

\begin{teorema}
Se $f$ \'e uma fun\c c\~ao sobre $X$, ent\~ao a regra
\[f_i=\E(f\mid\Omega_i),~~ i=0,1,\dots,n\] 
define um martingale sobre $X$. De fato, todo martingale \'e obtido dessa maneira da fun\c c\~ao $f=f_n$.
\label{marti}\qed
\end{teorema}

\begin{exemplo}
Se $\Omega_i$ s\~ao as parti\c c\~oes padr\~ao do cubo de Hamming 
$\Sigma^n$ (defini\c c\~ao \ref{d:standard}), ent\~ao um c\'alculo f\'acil mostra que para cada $k=0,1,\dots,n$ 
\[E(w\mid\Omega_k)(\sigma) = w_k(\pi^n_k(\sigma))+
\frac{n-k}{2}.\]
Aqui $w_k$ \'e o peso de Hamming sobre o cubo $\Sigma^k$.
\end{exemplo}

\begin{definicao}
Seja $(f_0,f_1,\dots,f_n)$ um martingale sobre um conjunto finito $X$, formado em rela\c c\~ao a uma sequ\^encia de parti\c c\~oes refinadoras. As {\em diferen\c cas de martingale} s\~ao as fun\c c\~oes
\[d_i=f_i-f_{i-1}.\]
\index{diferen\c cas de martingale}
\label{d:diferencasdemartingale}
\end{definicao}

\begin{observacao}\label{notice}
Temos a soma telesc\'opica:
\[f=\E(f)+d_1+d_2+\ldots+d_{n-1}+d_n.\]
\end{observacao}

\begin{proposicao}
\label{p:diffprop}
\[\E(d_i\mid\Omega_{i-1})=0.\]
\end{proposicao}

\begin{proof}
$\E(d_i\mid\Omega_{i-1})=\E(f_i\mid\Omega_{i-1})-\E(f_{i-1}\mid\Omega_{i-1})
= f_{i-1}-f_{i-1}$.
\end{proof}

\begin{lema}
Seja $f\colon \Sigma^n\to\R$ uma fun\c c\~ao Lipschitz cont\'\i nua de constante $L=1$ em rela\c c\~ao \`a dist\^ancia de Hamming. Seja $(f_0,f_1,\ldots,f_n)$ o martingale correspondente formado em rela\c c\~ao \`a sequ\^encia can\^onica de parti\c c\~oes $\Omega_0,\Omega_1,\dots,\Omega_n$ de $\Sigma^n$. Ent\~ao, para todo $i=1,2,\ldots,n$, 
as diferen\c cas de martingale, $d_i=f_i-f_{i-1}$, satisfazem
\[\norm{d_i}_\infty\leq\frac 12.\]
\label{l:differencasdemartingale}
\end{lema}

\begin{proof}
Seja $0\leq i\leq n$. 
O valor $f_i(\sigma)$ \'e o valor m\'edio de $f$ sobre o conjunto 
de todas as palavras tendo o prefixo $\sigma_1\sigma_2\cdots\sigma_i$:
\begin{align*}
f_i(\sigma)&=
2^{-n+i} \sum_{\delta\in\Sigma^{n-i}}f(\sigma_1\sigma_2\cdots\sigma_i\delta).
\end{align*}
Temos ent\~ao, de mesma forma,
\begin{align*}
f_{i-1}(\sigma)&=
2^{-n+i-1}\sum_{\delta\in\Sigma^{n-i+1}}f(\sigma_1\sigma_2\cdots\sigma_{i-1}\delta) \\
&= 2^{-n+i}\sum_{\delta\in\Sigma^{n-i}}\frac 12\left[f(\sigma_1\sigma_2\cdots\sigma_{i-1}0\delta)+
f(\sigma_1\sigma_2\cdots\sigma_{i-1}1\delta)\right].
\end{align*}
Por conseguinte,
\begin{align*}
\abs{d_i(\sigma)} &= \abs{f_i(\sigma)-f_{i-1}(\sigma)} \\
&=  2^{-n+i} \sum_{\delta\in\Sigma^{n-i}} \left\vert 
f(\sigma_1\sigma_2\cdots\sigma_{i-1}\sigma_{i}\delta) - 
\frac 12\left[f(\sigma_1\sigma_2\cdots\sigma_{i-1}0\delta)+
f(\sigma_1\sigma_2\cdots\sigma_{i-1}1\delta)\right]
\right\vert \\
&= 2^{-n+i} \sum_{\delta\in\Sigma^{n-i}} \left\vert 
\frac 12\left[f(\sigma_1\sigma_2\cdots\sigma_{i-1}\sigma_{i}\delta)-
f(\sigma_1\sigma_2\cdots\sigma_{i-1}\bar\sigma_{i}\delta)\right]
\right\vert\\
&\leq  2^{-n+i}\times 2^{n-i}\times \frac 12 = \frac 12.
\end{align*}
\end{proof}

\section{Concentra\c c\~ao de medida}

\subsection{Desigualdade de Azuma}

\begin{exercicio} 
O {\em cosseno hiperb\'olico}, $\cosh x$, \'e definido por
\[\cosh x = \frac{e^x+e^{-x}}2.\]
Mostrar que para todo $x\in\R$,
\[\cosh x\leq e^{x^2/2}.\]
\label{l:cosh}
\end{exercicio}

\begin{observacao}
Uma fun\c c\~ao $f\colon\R\to\R$ \'e dita {\em convexa} se para todos $x,y\in\R$ e $t\in [0,1]$,
\[f(tx+(1-t)y)\leq tf(x)+(1-t)f(y).\]
Em particular, se $f$ \'e $C^2$ e $f^{\prime\prime}(x)\geq 0$ para todo $x$, ent\~ao $f$ \'e convexa. 
Por exemplo, $f(x)=e^x$ \'e uma fun\c c\~ao convexa.
\index{fun\c c\~ao! convexa}
\end{observacao}

O resultado seguinte est\'a entre os principais resultados t\'ecnicos do primeiro cap\'\i tulo. Ele avalia a probabilidade de que o valor de uma fun\c c\~ao, $f$, desvie da sua esperan\c ca, $\E f$, por mais de um valor dado, $c>0$.

\begin{teorema}[Desigualdade de Azuma]
Seja $f\colon X\to\R$ uma fun\c c\~ao sobre um conjunto finito, $X$, seja
$(\Omega_0,\Omega_{1},\dots,$ $\Omega_{n-1},\Omega_n)$ uma sequ\^encia 
de parti\c c\~oes refinadoras de $X$, e seja $(f_0,f_1,\dots,f_n)$ o martingale correspondente com as diferen\c cas $(d_1,d_2,\dots,d_n)$.
Ent\~ao para todo $c>0$
\[\mu_\sharp(\{x\in X\mid f(x)-\E(f)\geq c\} \leq
\exp\left(-\frac{c^2}{2\sum_{i=1}^n\norm{d_i}_\infty^2}\right),\]
em particular,
\[\mu_\sharp(\{x\in X\mid \abs{f(x)-\E(f)}\geq c\} \leq
2\exp\left(-\frac{c^2}{2\sum_{i=1}^n\norm{d_i}_\infty^2}\right).\]
\label{azema}
\index{desigualdade! de Azuma}
\end{teorema}

\begin{proof}
Denotemos 
\[c_i=\norm{d_i}_\infty\equiv \max_{x\in X}\abs{d_i(x)}.\]
Fixe o valor $i=1,2,\ldots,n$, e denote
\[t=\frac{1-\frac{d_i}{c_i}}{2}.\]
Esse $t$ \'e uma fun\c c\~ao sobre $\Sigma^n$, tomando seus valores no intervalo $[0,1]$. Temos 
\[1-t = \frac{1+\frac{d_i}{c_i}}{2},\]
e as diferen\c cas de martingale, $d_i$, tornam-se umas combina\c c\~oes convexas de $\pm c_i$ com coeficientes vari\'aveis:
\begin{equation}
\label{eq:01}
d_i=t(-c_i)+(1-t)c_i.\end{equation}
Seja $\lambda\in\R$ qualquer (o valor exato ser\'a escolhido no final da prova). Multiplique a Eq. (\ref{eq:01}) por $\lambda$:
\begin{equation}
\label{eq:col}
\lambda d_i=t(-\lambda c_i)+(1-t)\lambda c_i.\end{equation}
Usando a convexidade de $e^x$, conclu\'\i mos:
\begin{align}
\label{eq:coln}
e^{\lambda d_i}&\leq te^{-\lambda c_i}+(1-t)e^{\lambda c_i} \nonumber \\
&= e^{-\lambda c_i}\frac{1-\frac{d_i}{c_i}}{2} + e^{\lambda c_i}\frac{1+\frac{d_i}{c_i}}{2} \nonumber \\
&= \frac{e^{\lambda c_i}+e^{-\lambda c_i}}{2} + \frac{d_i}{2c_i}\left(e^{\lambda c_i}-e^{-\lambda c_i}\right).
\end{align}
Estimamos a esperan\c ca condicional da fun\c c\~ao $e^{\lambda d_i}$ dado $\Omega_{i-1}$:
\begin{align}
\label{eq:estim}
\E\left( e^{\lambda d_i}\mid\Omega_{i-1}\right) &\leq 
\E\left( \frac{e^{\lambda c_i}+e^{-\lambda c_i}}{2}\mid\Omega_{i-1}\right)+
\E\left(\frac{d_i}{2c_i}\left(e^{\lambda c_i}-e^{-\lambda c_i}\right) \mid\Omega_{i-1}\right) \nonumber \\
&= \frac{e^{\lambda c_i}+e^{-\lambda c_i}}{2} + \frac{1}{2c_i}\left(e^{\lambda c_i}-e^{-\lambda c_i}\right)\E\left(d_i\mid\Omega_{i-1}\right) \nonumber\\
\mbox{\small (proposi\c c\~ao \ref{p:diffprop})}
&= \frac{e^{\lambda c_i}+e^{-\lambda c_i}}{2} \nonumber\\
\mbox{\small (ex. \ref{l:cosh})} 
&\leq e^{(\lambda c_i)^2/2}.
\end{align}
(Os valores $c_i$ e fun\c c\~oes deles s\~ao {\em constantes}).

Toda diferen\c ca de martingale, $d_j$, $j<i$ \'e constante sobre os elementos da parti\c c\~ao $\Omega_{i-1}$. Segue-se que:
\begin{align}
\E\left(e^{\lambda\sum_{j=1}^i d_j}\right) &= 
\E\left( \E\left(e^{\lambda\sum_{j=1}^i d_j}\mid\Omega_{i-1}\right)\right)
\nonumber \\
\mbox{\small (pela proposi\c c\~ao \ref{properties}(\ref{factorout})}
&= \E\left(\left(e^{\lambda\sum_{j=1}^{i-1} d_j}\right) 
\E\left(e^{\lambda d_i}\mid\Omega_{i-1} \right)\right) \nonumber \\
\mbox{\small (pela desigualdade (\ref{eq:estim}))} &\leq 
\E\left(e^{\lambda\sum_{j=1}^{i-1} d_j}\right)
e^{\lambda^2c_i^2/2}.
\label{ineqq}
\end{align}
Agora suponha que $\lambda>0$. 
Levando em conta a observa\c c\~ao \ref{notice}, conclu\'\i mos que
\begin{align}
\mu_\sharp\{\sigma\mid f(\sigma)-\E(f)\geq c\} 
& =
\mu_\sharp\left\{\sigma\mid \sum_{i=1}^n d_i(\sigma) \geq c\right\}
\nonumber \\
& =
\mu_\sharp\left\{\sigma\mid \lambda\sum_{i=1}^n d_i(\sigma)-
\lambda c \geq 0\right\}
\nonumber \\
& =
\mu_\sharp\left\{\sigma\mid e^{ \lambda\sum_{i=1}^n d_i(\sigma)-
\lambda c} \geq 1\right\} \nonumber \\
\mbox{(pela desigualdade de Markov)} 
&\leq 
\E\left(e^{ \lambda\sum_{i=1}^n d_i-\lambda c} \right)
\nonumber \\
&= \E\left(e^{\lambda d_1}e^{\lambda d_2}\dots e^{\lambda d_{n-1}}
e^{\lambda d_n} \right)e^{-\lambda c} \nonumber \\
\mbox{(pela desigualdade (\ref{ineqq}))} 
&\leq 
\E\left(e^{\lambda d_1}e^{\lambda d_2}\dots e^{\lambda d_{n-1}}
 \right)e^{\lambda^2c_n^2/2}e^{-\lambda c}
\nonumber \\
\mbox{(usando (\ref{ineqq}))
recursivamente, $n$ vezes)} 
&\leq  \dots \nonumber \\
&\leq e^{\lambda^2c_1^2/2}e^{\lambda^2c_2^2/2}
\cdots e^{\lambda^2c_n/2^2}e^{-\lambda c}
\nonumber \\
&= e^{\lambda^2\sum_{j=1}^nc_j^2/2 -\lambda c}.
\label{almostover}
\end{align}
Substituamos em (\ref{almostover}) o valor
\[\lambda=\frac{c}{\sum_{j=1}^nc_j^2}\]
para concluir
\begin{equation}
\mu_\sharp\{\sigma\colon f(\sigma)-\E(f)\geq c\}\leq
e^{-\frac{c^2}{2\sum_{j=1}^n\norm{d_j}_\infty^2}}.
\label{odno}
\end{equation}
A desigualdade g\^emea \'e obtida usando $-f$ em lugar de $f$:
\begin{equation}
\mu_\sharp\{\sigma\colon \E(f)-f(\sigma)\geq c\}\leq
e^{-\frac{c^2}{2\sum_{j=1}^n\norm{d_j}_\infty^2}}.
\label{vtoroe}
\end{equation}
Em conjunto, as duas desigualdades (\ref{odno}) e
(\ref{vtoroe}) implicam
\begin{equation}
\mu_\sharp\{\sigma\colon \abs{\E(f)-f(\sigma)}\geq c\}\leq
2e^{-\frac{c^2}{2\sum_{j=1}^n\norm{d_j}_\infty^2}},
\end{equation}
como desejado. (Aqui usamos a {\em desigualdade de Boole}, conhecido como o {\em union bound} em ingl\^es:
$\mu(A\cup B)\leq\mu(A)+\mu(B)$.)
\end{proof}

\subsection{Cotas de Chernoff}

\begin{teorema}
Seja $f\colon \Sigma^n\to\R$ uma fun\c c\~ao Lipschitz cont\'\i nua, de constante $L=1$ (em rela\c c\~ao \`a dist\^ancia de Hamming sobre $\Sigma^n$). Ent\~ao, para todo $c>0$ temos
\[\mu_\sharp\{x \colon \abs{f(x)- \E(f)}\geq c\}\leq
2e^{-\frac{2c^2}{n}}.\]
\label{t:interplay}
\end{teorema}

\begin{proof}
Lema \ref{l:differencasdemartingale} junto com a desigualdade de Azuma. 
\end{proof}

\subsection{Lei geom\'etrica dos grandes n\'umeros}
Eis uma vers\~ao para a dist\^ancia de Hamming normalizada. 

\begin{corolario}
Seja $f\colon \Sigma^n\to\R$ uma fun\c c\~ao 1-Lipschitz cont\'\i nua em rela\c c\~ao \`a dist\^ancia de Hamming normalizada. Ent\~ao para todo $\e>0$
\[\mu_\sharp\{x \colon \abs{f(x)- \E(f)}\geq \e\}\leq
2e^{-2\e^2n}.\]
\label{c:interplay2}
\end{corolario}

\begin{proof}
A fun\c c\~ao $nf$ \'e 1-Lipschitz cont\'\i nua em rela\c c\~ao com a dist\^ancia de Hamming n\~ao normalizada: 
\begin{align*}
\abs{nf(\sigma)-nf(\tau)} 
&= n\abs{f(\sigma)-f(\tau)} \\
&\leq  n \bar d(\sigma,\tau) \\
&= d(\sigma,\tau).
\end{align*}

O teorema \ref{t:interplay} aplicada \`a fun\c c\~ao $nf$ com $c=n\e$ implica:
\[\mu_\sharp\{x \colon \abs{nf(x)- \E(nf)}\geq n\e\}\leq
2e^{-\frac{\e^2n^2}{n}},\]
\'e como $\E(nf)=n\E(f)$, terminamos.
\end{proof}

Aplicadas ao peso normalizado $\bar w$, as cotas de Chernoff (corol\'ario \ref{c:interplay2}) levam a concluir:
\[\forall \ve>0,~~\mu_{\sharp}\left\{x\in\Sigma^n\colon \left\vert\bar w -\frac 12\right\vert >\ve\right\} \leq 2e^{-2\ve^2n}.\]
Esta desigualdade, conhecida na teoria de probabilidade como a {\em (fraca) Lei dos Grandes N\'umeros,} 
\index{lei dos grandes n\'umeros! fraca}
diz que a probabilidade que o n\'umero m\'edio de sucessos (caras) obtidos ap\'os $n$ lan\c camentos de uma moeda equilibrada diferir de $1/2$ se torna exponencialmente pequena quando $n\to\infty$.

\'E por isso que a desigualdade mais geral no corol\'ario \ref{c:interplay2}, v\'alida para todas fun\c c\~oes Lipschitz cont\'\i nuas de constante $L=1$, \'e as vezes chamada {\em a lei geom\'etrica dos grandes n\'umeros}.
\index{lei dos grandes n\'umeros! geom\'etrica}

\subsection{Volumes das bolas: a segunda estimativa\label{ss:volume}}

Como $\bar w(x+1)+\bar w(x)=1$, temos 
$\bar d(x,1)=\bar d(x+1,0) =1-\bar w(x)$, e por conseguinte
\[\bar B_\e(1) = \{x\in\Sigma^n\colon \bar w(x)-\frac 12\geq \frac 12-\e\}.\]
Desta maneira, obtemos a segunda estimativa da medida da bola:
\begin{equation}
\mu_{\sharp}(\bar B_\e) \leq e^{-2\left(\frac 12-\e\right)^2n}.
\label{eq:vol2}
\end{equation}
Esta estimativa \'e melhor do que a dada pelo lema \ref{l:<} no regi\~ao onde $\e>0$ \'e constante, isso \'e, o raio n\~ao normalizado da bola \'e proporcional \`a $n$:
\[d\sim \e n.\]
A primeira estimativa, a do lema \ref{l:<},
\begin{equation}
\mu_{\sharp}(\bar B_\e) \leq 2^{-n}\left(\frac{en}{\lfloor \e n\rfloor}\right)^{\lfloor\e n\rfloor},
\end{equation}
\'e melhor na regi\~ao onde $d=\lfloor \e n \rfloor$ \'e constante, ou seja, $\e=O(n^{-1})$. Vamos utilizar ambas simultaneamente na an\'alise de aprendizabilidade.

\subsection{Fun\c c\~ao de concentra\c c\~ao}
As desigualdades acima -- as de Azuma e de Chernoff -- s\~ao manifesta\c c\~oes do {\em fen\^omeno de concentra\c c\~ao de medida sobre as estruturas de alta dimens\~ao}, que pode ser expresso informalmente da seguinte maneira:

\begin{quote}
{\em Tipicamente,
sobre uma estrutura $\Omega$ de alta dimens\~ao, cada fun\c c\~ao Lipschitz cont\'\i nua concentra-se em torno de um valor, ou seja, \'e quase constante em toda parte exceto sobre um conjunto da medida demasiadamente pequena.
}
\end{quote}

Como o valor t\'\i pico da fun\c c\~ao, pode ser escolhido ou a esperan\c ca $\E f$, ou um valor mediano, $M_f$. Neste contexto, o fen\^omeno pode ser reformulado mais uma vez:

\begin{quote}
{\em
Tipicamente,
numa estrutura $\Omega$ de alta dimens\~ao, para cada subconjunto $A\subseteq \Omega$ que cont\'em pelo menos a metade dos pontos, a maior parte dos pontos de $\Omega$ est\~ao pr\'oximos ao $A$.}
\end{quote}

\index{fen\^omeno de concentra\c c\~ao de medida}

As no\c c\~oes ``dimens\~ao alta'' e ``pontos pr\'oximos'' sendo relativas, e a no\c c\~ao ser\'a formalizada no contexto assint\'otico. 
Um instrumento conveniente para quantificar o fen\^omeno da concentra\c c\~ao de medida \'e a {\em fun\c c\~ao de concentra\c c\~ao.} 
Recordemos que
\[A_\e=\{x\in\Omega\colon \exists a\in A~~\rho(x,a)<\e\}\]
\'e a $\e$-vizinhan\c ca do subconjunto $A$ de $\Omega$.

\begin{definicao}
Seja $(\Omega,d,\mu)$ um espa\c co m\'etrico munido de uma medida de probabilidade. A {\em fun\c c\~ao de concentra\c c\~ao} de $\Omega$, denotada por $\alpha_\Omega(\e)$ ou $\alpha(\Omega,\e)$ ou simplesmente $\alpha(\e)$, \'e definida pelas condi\c c\~oes seguintes:

\[\alpha(\e)=\left\{
  \begin{array}{ll} \frac 12, & \mbox{se $\e=0$,} \\
1-\inf\left\{\mu_\sharp\left(A_\e\right) \colon
A\subseteq\Sigma^n, ~~ \mu_\sharp(A)\geq\frac 12\right\}, &
\mbox{se $\e>0$.}
\end{array}\right.
\]
\label{def:concfn}
\index{fun\c c\~ao! de concentra\c c\~ao}
\end{definicao}

\begin{figure}[htp]
  \centerline{\includegraphics[width=6cm]{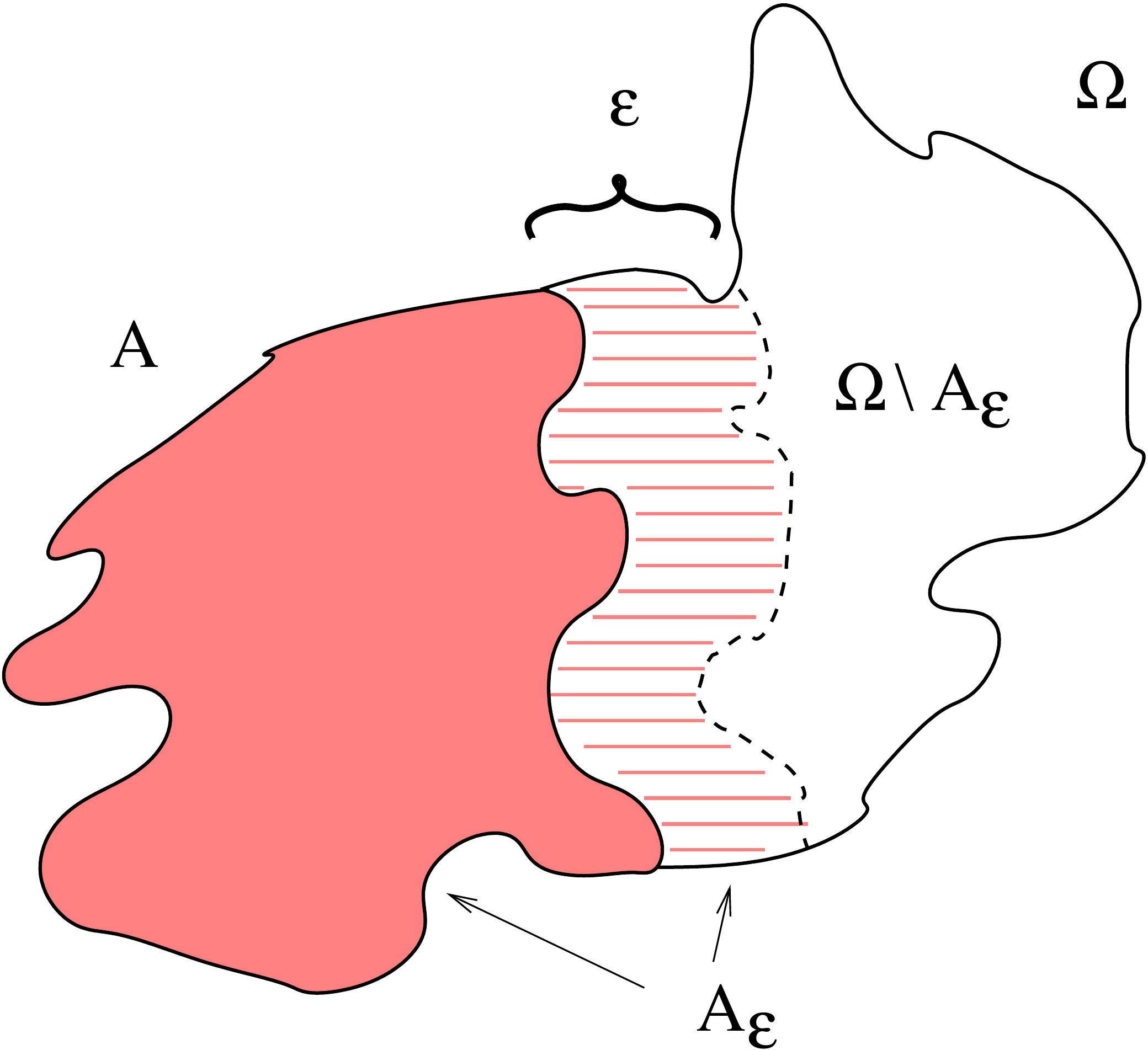}}
  \caption{A fun\c c\~ao de concentra\c c\~ao $\alpha(\Omega,\e)$.}
\label{fig:illconc}
\end{figure}

\begin{exercicio} Mostrar que
  \[\alpha(\Omega,\e)\to 0\mbox{ quando }\e\to\infty.\]
\end{exercicio}

\begin{definicao}
  Uma fam\'\i lia $(\Omega_n,\rho_n,\mu_n)$ de espa\c cos m\'etricos com medida de probabilidade \'e dita uma {\em fam\'\i lia de L\'evy} se as fun\c c\~oes de concentra\c c\~ao tendem a zero pontualmente,
  \[\alpha(\Omega_n,\e)\to 0\mbox{ para cada }\e>0.\]
Ela \'e uma {\em fam\'\i lia de L\'evy normal} se existem $C_1,C_2>0$ tais que
  \[\alpha(\Omega_n,\e)\leq C_1e^{-C_2\e^2n}.\]
\index{fam\'\i lia! de L\'evy}
\index{fam\'\i lia! de L\'evy! normal}
\end{definicao}

\begin{figure}[htp]
\centerline{\scalebox{0.72}[1]{\includegraphics[width=8cm]
{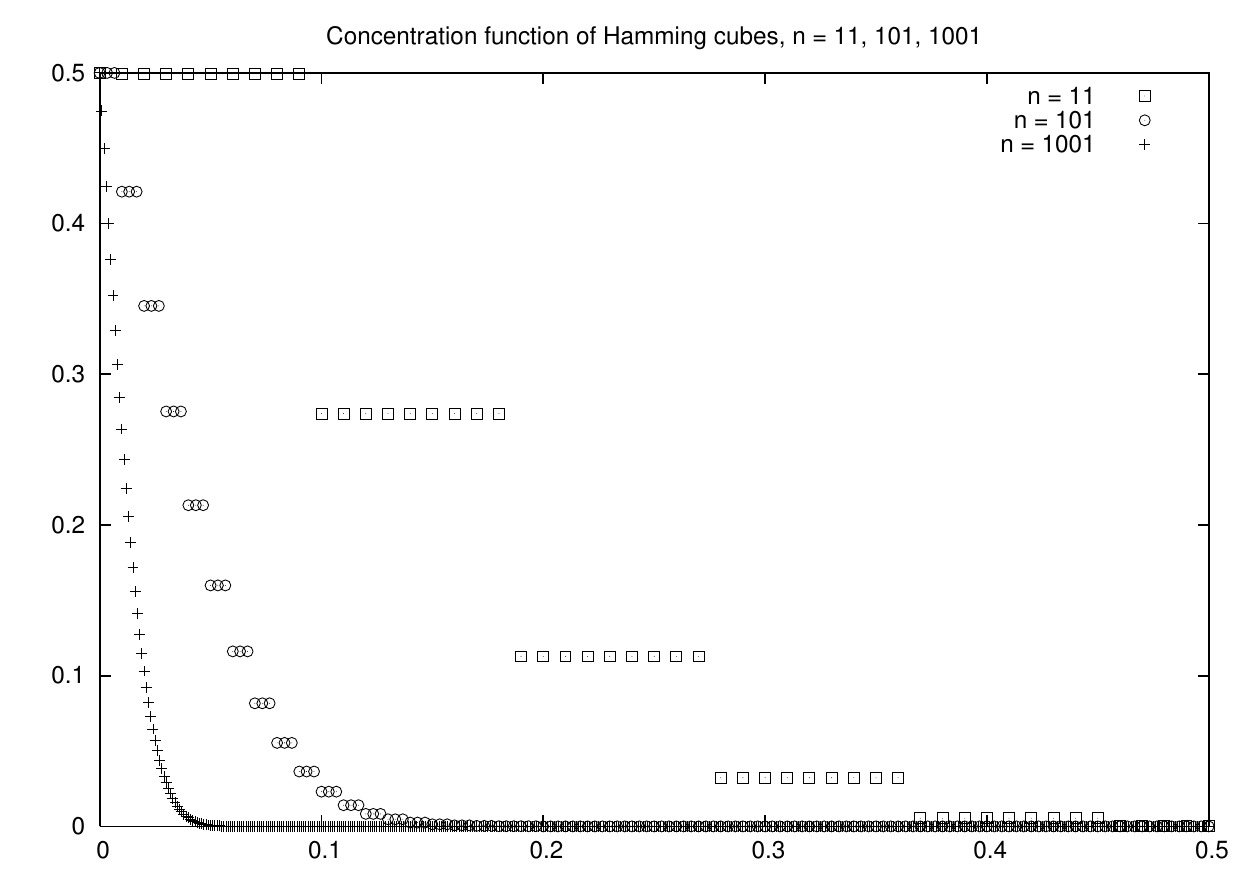}}}
\caption{As fun\c c\~oes de concentra\c c\~ao dos cubos de Hamming 
para $d=11,101,1001$.}
\end{figure}

\begin{exercicio}
Seja $r\geq 0$ inteiro e $\e>0$ qualquer. Ent\~ao, a $\e$-vizinhan\c ca aberta de uma bola fechada de raio $r$ no cubo de Hamming \'e uma bola aberta de mesmo centro dada por
\[\left(\bar B_r\right)_{\e}(\sigma) = B_{r+\e}(\sigma).\]
(Mostra que para $r,\e>0$ reais quaisquer isso n\~ao \'e mais verdadeiro).
\end{exercicio}

\begin{exercicio}
O complemento de uma bola aberta no cubo de Hamming \'e dado por
\[\Sigma^n\setminus B_\delta(\sigma) = \bar B_{1-\delta}(\sigma+\bar 1).\]
\end{exercicio}

\begin{exercicio}
Seja $n=2m+1$ um n\'umero \'\i mpar. Mostrar que a bola $\bar B_{m}(\sigma)$ ($\bar B_{m/n}(\sigma)$, com rela\c c\~ao \`a dist\^ancia de Hamming normalizada) tem a medida de contagem normalizada $1/2$. 
\par
[ {\em Sugest\~ao:} estudar o seu complemento. ]
\end{exercicio}

\begin{exercicio}
Deduzir que no caso de $n=2m+1$ \'\i mpar, as bolas $\bar B_{m}(\sigma)$ s\~ao as \'unicas bolas de Hamming com $2^{n-1}$ elementos.
\end{exercicio}

\begin{teorema}
Os cubos de Hamming $\Sigma^n$ munidos da dist\^ancia de Hamming e a medida de contagem normalizadas formam uma fam\'\i lia de L\'evy normal:
\[\alpha(\Sigma^n,\e) \leq e^{-\e^2n}.
\]
\end{teorema}

\begin{proof}
Seja $A$ um subconjunto do cubo que contem pelo menos a metade dos elementos: $\mu_{\sharp}(A)\geq 1/2$, e seja $\e>0$ qualquer. Precisamos estimar por cima $\mu_{\sharp}(\Sigma^n\setminus A_\e)$. \'E claro que pode-se supor sem perda de generalidade que $\mu_{\sharp}(A)=1/2$ (selecionando um subconjunto de $A$ com exatamente $2^{n-1}$ elementos: o valor de $\mu_{\sharp}(\Sigma^n\setminus A_\e)$ s\'o vai aumentar). Seja $r\geq 0$ o maior valor inteiro tal que $r/n<\e$. Temos $\bar B_r(\sigma) = B_\e(\sigma)$, qual quer seja $\sigma$.

Come\c camos com $n$ \'\i mpar, $n=2m+1$. 
Segundo o teorema de Harper, assim como os exerc\'\i cios acima, a bola $\bar B_r(\bar 0)$ tem a propriedade $\mu_{\sharp}(\Gamma_R\bar B_m(\bar 0))\leq\mu_{\sharp}(A_\e)$, isto \'e, $\mu_{\sharp}(B_{m/n+\e})\leq \mu_{\sharp}(A_\e)$, \'e por conseguinte, $\mu_{\sharp}(\bar B_{1-m/n - \e}) \geq \mu_{\sharp}(\Sigma^n\setminus A_\e)$. Temos
\[1-\frac m n -\e = \frac 12 + \frac 1{2n} - \e,
\]
e segundo a estimativa de tamanho das bolas,
\[\mu_{\sharp}(\bar B_{1-m/n - \e})\leq e^{-\left(\e - \frac 1{2n}\right)^2n}.\]
Como
\[-2\left(\e-\frac 1{2n}\right)^2n = -2\e^2n+{2\e}-\frac 1{2n} \leq -\frac 32\e^2n\]
(pois $\frac 12\e^2n^2-2\e n+2\geq 0$), deduzimos do corol\'ario \ref{c:interplay2}:
\begin{equation}
\alpha(\Sigma^n,\e)\leq e^{-\frac 32\e^2 n}.
\label{eq:par}
\end{equation}

Para $n$ par, $n=2m$, o argumento \'e semelhante, mas nesse caso o cubo de Hamming consiste da bola fechada de raio $m-1$ ($(m-1)/n$, caso normalizado), mais uma metade de elementos da esfera de raio $m$. Em particular, o complemento de conjunto $A_\e$ tem uma medida menor ou igual da medida da bola de raio $\frac 12 + \frac 1{n} - \e$, cuja medida de contagem normalizada tem estimativa acima 
\[e^{-\left(\e - \frac 1{n}\right)^2n}.\]
Como 
\[-2\left(\e-\frac 1{n}\right)^2n = -2\e^2n+{4\e}-\frac 2{n} \leq -  \e^2n\]
(pois $\e^2n^2 - 4\e n +2\geq 0$ para qualquer que seja $n$), conclu\'\i mos:

\[\alpha(\Sigma^n,\e)<e^{-\e^2 n}.\]
\end{proof}

\begin{observacao}
Sem d\'uvida, as constantes podem ser melhoradas. Em particular, eu n\~ao sei a resposta para o seguinte. Dado $\e>0$ e $n$ quaisquer, \'e verdadeiro que
\[\alpha_{\Sigma^{n+1}}(\e)\leq \alpha_{\Sigma^n}(\e)?\]
Em outras palavras, \'e verdade que as fun\c c\~oes de concentra\c c\~ao de cubos de Hamming convergem para zero monotonicamente quando $n\to\infty$? Aqui, s\'o a transi\c c\~ao de um $n$ \'\i mpar para $n+1$ deve ser estudada.
\end{observacao}

\begin{figure}[htp]
\centering
\scalebox{0.6}[0.6]{\includegraphics{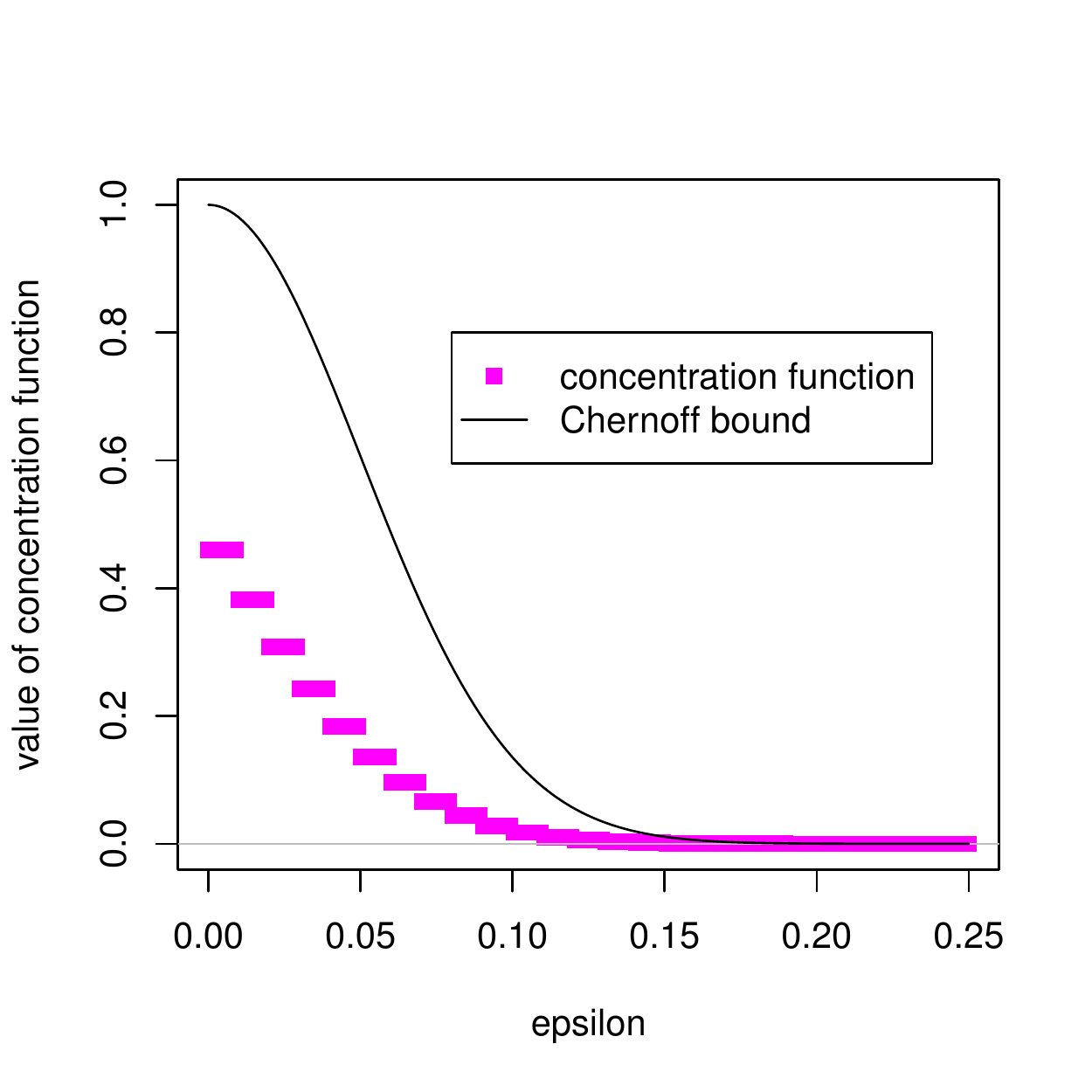}}
\caption{Fun\c c\~ao de concentra\c c\~ao do cubo de Hamming $\Sigma^{100}$ e a cota superior gaussiana de Chernoff}
\label{fig:bound101}
\end{figure}

\begin{exercicio}
  Seja $f$ uma fun\c c\~ao Lipschitz cont\'\i nua com a constante de Lipschitz $L\geq 0$ sobre um espa\c co m\'etrico com medida de probabilidade, $(\Omega,\rho,\mu)$. Seja $M=M_f$ um valor mediano de $f$.
Provar que
  \[\mu\{\abs{f(x)-M}>\e\}\leq 2\alpha \left(\Omega,\frac{\e}{L}\right).\]
  Mais geralmente, se $f$ \'e uniformemente cont\'\i nua de tal modo que
\[\forall x,y\in X,~~d(x,y)<\delta\Rightarrow \abs{fx-fy}<\e,\]
ent\~ao
\[\mu\{\abs{f(x)-M}>\e\}\leq 2\alpha(\Omega,\delta).\]
\label{ex:M}
\end{exercicio}

O fen\^omeno de concentra\c c\~ao da medida \'e o assunto de estudo de uma disciplina matem\'atica relativamente nova: a {\em an\'alise geom\'etrica assint\'otica.}
Esta introspe\c c\~ao na geometria dos objetos de alta dimens\~ao \'e da mais alta import\^ancia, e tem muitas aplica\c c\~oes e consequ\^encias amplas em ci\^encias matem\'aticas. Veja a monografia \citep*{AAGM}.

A apresenta\c c\~ao neste cap\'\i tulo geralmente seguiu a no livro cl\'assico \citep*{MS}, mas com constantes melhoradas como em \citep*{alon_spencer}. O livro \citep*{L} \'e uma fonte excelente para a concentra\c c\~ao de medida. Veja tamb\'em o panor\^amico de \citep*{schechtman}, assim como o livro de \citep*{matousekBook}.

%
%

\chapter{Dimens\~ao de Vapnik--Chervonenkis\label{ch:VC}}

\section{Defini\c c\~ao e exemplos iniciais}
Vamos ampliar a nossa perspetiva e estudar as rotulagens sobre os subconjuntos finitos de um dom\'\i nio, $\Omega$, finito ou infinito. Um classificador, $T\colon\Omega\to\{0,1\}$, induz sobre uma amostra n\~ao rotulada qualquer, $\sigma=(x_1,\ldots,x_n)$, uma rotulagem $(\e_1,\ldots,\e_n)$ bastante naturalmente:
\[\e_i= T(x_i),~~i=1,2,\ldots,n.\]
Um classificador, ou uma fun\c c\~ao bin\'aria, \'e exatamente a fun\c c\~ao indicadora de um subconjunto, $C\subseteq\Omega$:
\[T=\chi_C,\mbox{ onde }\chi_C(x)=\begin{cases}1,&\mbox{ se }x\in C,\\
0,&\mbox{ se }x\notin C.
\end{cases}\]
Em vez de fun\c c\~oes bin\'arias, podemos trabalhar com subconjuntos do dom\'\i nio, $C\in 2^{\Omega}$. Tais subconjuntos $C\subseteq\Omega$ s\~ao chamados de {\em conceitos.} Dado um conceito $C$, um ponto $x\in\Omega$ \'e ent\~ao rotulado $1$ se e somente se $x\in C$. 

Agora suponha que trabalhamos com uma fam\'\i lia qualquer $\mathscr C\subseteq 2^{\Omega}$ de conceitos (uma {\em classe de conceitos,} {\em concept class}). 
Mediante a motiva\c c\~ao, $\mathscr C$ \'e uma fam\'\i lia de todos os classificadores produzidos por uma regra de aprendizagem particular, como por exemplo uma rede neural $\mathscr N$ de uma arquitetura fixa, que depende de uma fam\'\i lia de par\^ametros reais. Os diferentes valores de par\^ametros resultam em conceitos diferentes. O algoritmo de aprendizagem procura um conceito $C$ dentro da classe $\mathscr C$ que melhor se ajusta a uma dada amostra rotulada, $\sigma$, a fim de minimizar o erro emp\'\i rico (o erro de aprendizagem). Queremos evitar o fen\^omeno de ``overfitting'' (sobreajuste), esperando que o {\em erro de generaliza\c c\~ao} seja baixo, ou seja, que o conceito escolhido se mostra eficaz para predizer novos r\'otulos. Isso \'e poss\'\i vel se a complexidade da classe $\mathscr C$ for bastante baixa. Falando de uma maneira informal, a complexidade de $\mathscr C$ d\'a uma medida de quantas rotulagens diferentes $\mathscr C$ pode produzir em amostras finitas. Uma medida te\'orica importante da complexidade de uma classe de conceitos \'e a famosa {\em dimens\~ao de Vapnik--Chervonenkis}.\footnote{Pronunciar: V\'apnik--Tcherv\'onenkis.}

\begin{definicao}
Seja $\mathscr C$ uma classe de conceitos no dom\'\i nio $\Omega$. Para um subconjunto $B\subseteq\Omega$, denotemos por
\[{\mathscr C}\vert_B = \{\chi_C\vert_B\colon C\in {\mathscr C}\}\]
a fam\'\i lia de todas as rotulagens sobre $B$ induzidas pelos membros da classe $\mathscr C$. Digamos que um subconjunto finito $B\subseteq\Omega$ \'e {\em fragmentado} ({\em shattered}) por $\mathscr C$ se cada rotulagem poss\'\i vel de $B$ \'e atingida: 
\[{\mathscr C}\vert_B = 2^B =\{0,1\}^{\sharp B}.\]
Em outras palavras, $B$ \'e fragmentado por $\mathscr C$ se cada fun\c c\~ao bin\'aria sobre $B$ pode se estender a uma fun\c c\~ao da forma $\chi_C$, $C\in\mathscr C$. 
\index{subconjunto! fragmentado}
\end{definicao}

Aqui \'e uma outra maneira de reformular a defini\c c\~ao.

\begin{definicao}
Seja $\mathscr C$ uma fam\'\i lia de subconjuntos de $\Omega$. Um conjunto finito $A\subseteq\Omega$ \'e {\em fragmentado} por $\mathscr C$ se todo subconjunto $B\subseteq A$ pode ser ``retirado'' de $A$ com a ajuda de um elemento apropriado $C$ de $\mathscr C$:
\[B = A\cap C.\]
\end{definicao}

\begin{figure}[ht]
\begin{center}
\scalebox{0.25}[0.25]{\includegraphics{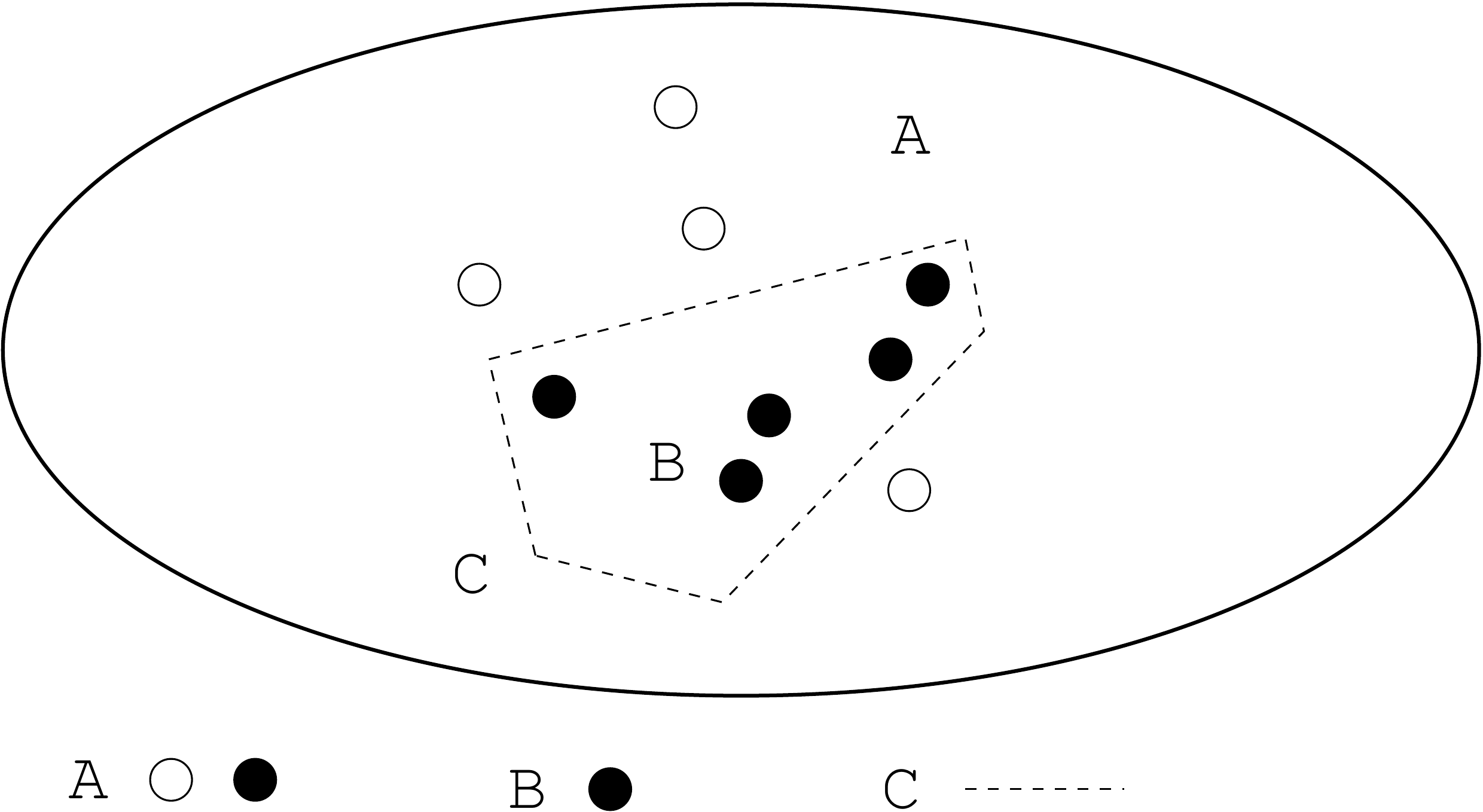}} 
\caption{Um subconjunto $A$ fragmentado pela classe $\mathscr C$.}
\end{center}
\end{figure}  

\begin{definicao}
A {\em dimens\~ao de Vapnik--Chervonenkis} ({\em dimens\~ao VC}) 
de uma classe de conceitos $\mathscr C$ \'e o supremo de cardinalidades de subconjuntos finitos $A\subseteq\Omega$ fragmentados por $\mathscr C$. Nota\c c\~ao: $\VC({\mathscr C})$.  

Em outras palavras, a dimens\~ao VC de $\mathscr C$ \'e a maior cardinalidade de um conjunto fragmentado por $\mathscr C$, se existir, e $+\infty$ se n\~ao existir.
\index{VCdimF@$\VC({\mathscr C})$}
\index{dimens\~ao! de Vapnik--Chervonenkis}
\end{definicao}

\begin{observacao} 
Para estabelecer a desigualdade $\VC({\mathscr C})\geq d$, basta exibir um conjunto $A\subseteq\Omega$ com $d$ elementos, fragmentado por $\mathscr C$. Por outro lado, para estabelecer a desigualdade $\VC({\mathscr C})\leq d$, \'e preciso mostrar que nenhum subconjunto com $d+1$ elementos \'e fragmentado. Isso \'e normalmente a parte mais dif\'\i cil.  
\end{observacao}

\begin{exemplo}
\label{ex:so}
Uma classe que consiste de um conceito s\'o, $\mathscr C=\{C\}$, possui dimens\~ao VC igual a zero. A classe $\mathscr C$ n\~ao fragmenta nenhum conjunto unit\'ario, $\{x\}$, pois ele sempre induza o mesmo r\'otulo ($1$, se $x\in C$, e $0$, se $x\notin C$), nunca ambos. Por isso, $\VC(\{C\})<1$. Por outro lado, o conjunto vazio, $\emptyset$, \'e fragmentado por $\mathscr C$: a restri\c c\~ao da fun\c c\~ao $\chi_C$ sobre o conjunto vazio \'e a fun\c c\~ao vazia, a \'unica rotulagem poss\'\i vel de $\emptyset$. Conclu\'\i mos: $\VC(\mathscr C)=0$.
\end{exemplo}

\begin{exemplo}
Seja $\Omega$ um dom\'\i nio n\~ao vazio qualquer, e seja $\mathscr C$ a classe que consiste de dois conjuntos: $\emptyset$ e $\Omega$. Nesse caso, a dimens\~ao VC da classe $\mathscr C$ \'e $1$. Todo conjunto unit\'ario $\{x\}$, $x\in\Omega$, \'e fragmentado: $\emptyset = \emptyset\cap \{x\}$ e $\{x\}=\Omega\cap \{x\}$.
Nenhum conjunto com dois pontos diferentes, $\{x,y\}$, \'e fragmentado por $\mathscr C$, pois um tal conjunto cont\'em $4$ subconjuntos diferentes, $\emptyset$, $\{x\}$, $\{y\}$, e $\{x,y\}$, e s\'o $\emptyset$ e $\{x,y\}$ podem ser obtidos como interse\c c\~oes de elementos de $\mathscr C$ com $\{x,y\}$.
\end{exemplo}

\begin{observacao} 
Mais geralmente, a dimens\~ao VC de uma classe $\mathscr C$ com $n$ conceitos n\~ao excede $\log_2n$. Se um conjunto com $d$ pontos for fragmentado por $\mathscr C$, a classe tem que conter pelo menos $2^d$ elementos dois a dois distintos.
\end{observacao}

\begin{exemplo}
Seja $\Omega=\R$, e seja $\mathscr C$ a classe de todos os intervalos semiinfinitos crescentes:
\[{\mathscr C}=\{[a,+\infty)\colon a\in\R\}.\]
Ent\~ao,
\[\VC({\mathscr C})=1.\]
\'E claro que todo conjunto unit\'ario, $\{x\}$, \'e fragmentado por $\mathscr C$: s\'o h\'a dois subconjuntos, o pr\'oprio $\{x\}$ e o conjunto vazio, e ambos s\~ao interse\c c\~oes de $\{x\}$ com um intervalo apropriado. Por outro lado, nenhum conjunto com dois pontos, $A=\{x,y\}$, $x<y$, \'e fragmentado por $\mathscr C$: qualquer que seja $a\in\R$, temos $\{x,y\}\cap [a,\infty)\neq\{x\}$. 
\end{exemplo}

\begin{exemplo} Se agora tomamos todos os intervalos semiinfinitos crescentes bem como decrescentes,
\[{\mathscr C} = \{[a,+\infty)\colon a\in\R\}\cup \{(-\infty,b]\colon b\in\R \},\]
o valor da dimens\~ao de Vapnik--Chervonenkis aumenta:
\[\VC({\mathscr C})=2.\]
Todo subconjunto com dois pontos, $A=\{x,y\}$ \'e fragmentado por $\mathscr C$:
\begin{eqnarray*}
\{x,y\}=\{x,y\}\cap [x,\infty),&~~~&\{x\} = \{x,y\}\cap (-\infty,x],\\
\{y\} =\{x,y\}\cap [y,\infty),&~~&\emptyset = \{x,y\}\cap [y+1,\infty).\end{eqnarray*}
Ao mesmo tempo, nenhum subconjunto com tr\^es pontos distintos \'e fragmentado por $\mathscr C$. Sejam $x<y<z$, $A=\{x,y,z\}$, ent\~ao o conjunto $B=\{y\}$ nunca pode ser obtido como a interse\c c\~ao de $B$ com um intervalo semiinfinito. 
\end{exemplo}

\begin{exemplo}
Agora seja $\mathscr C$ a classe de todos os intervalos fechados finitos,  $[a,b]$, $a,b\in\R$. Ent\~ao, mais uma vez,
\[\VC({\mathscr C})=2.\]
\index{dimens\~ao! de Vapnik--Chervonenkis! de intervalos}
Todo conjunto com dois pontos \'e fragmentado, mais nenhum conjunto com tr\^es pontos, $A=\{x,y,z\}$, \'e: o conjunto $B=\{x,z\}$ de pontos extremos n\~ao pode ser obtido como a interse\c c\~ao de $B$ com um intervalo. 
\end{exemplo}

\begin{exemplo}
\label{ex:half}
Agora seja $\Omega=\R^2$, o plano.
A classe $\mathscr C$ consiste de todos os semiplanos (abertos, ou fechados, ou bem abertos e fechados, o resultado \'e o mesmo). Recordemos pois que trata-se dos semiplanos fechados:
\[H\equiv H_{\vec v,b}=\{\vec x\in\R^2\colon \langle\vec x,\vec v\rangle\geq b\},
\]
onde $\vec v\in\R^2$ e $b\in\R$ s\~ao quaisquer:
\[{\mathscr C}=\{H_{\vec v,b}\colon \vec v\in\R^2,~b\in\R\}.\]

Ent\~ao, 
\[\VC({\mathscr C})=3.\]
\index{dimens\~ao! de Vapnik--Chervonenkis! de semiplanos}

Todo conjunto de tr\^es pontos n\~ao colineares, $a,b,c$, \'e fragmentado pelos semiplanos, o que \'e mais ou menos evidente geom\'etricamente, veja Figura \ref{fig:2}. 

\begin{figure}[ht]
\begin{center}
\scalebox{0.2}[0.2]{\includegraphics{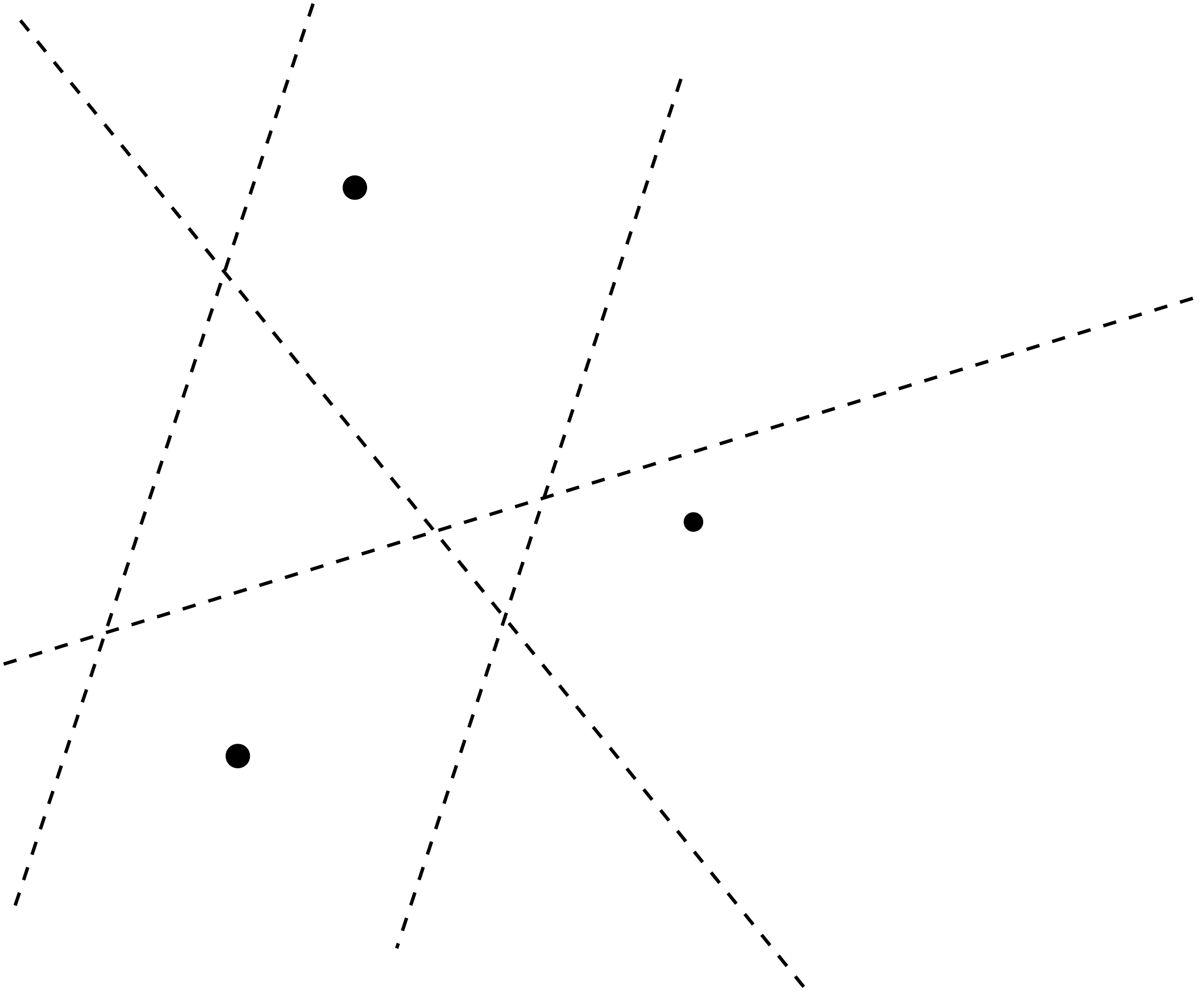}} 
\caption{Todo conjunto de tr\^es pontos em posi\c c\~ao geral \'e fragmentado pelos semiplanos.}
\label{fig:2}
\end{center}
\end{figure}
Ao mesmo tempo, nenhum subconjunto com quatro pontos distintos \'e fragmentado pelos semiplanos. Sejam $a,b,c,d$ quaisquer, dois a dois distintos. Existem dois casos a considerar. 

Se pelo menos um dos pontos $a,b,c,d$ for contido dentro do tri\^angulo formado pelos outros tr\^es pontos, digamos o ponto $d$, ent\~ao cada semiplano que cont\'em o conjunto $\{a,b,c\}$, sendo um conjunto convexo, tem que conter $d$ tamb\'em, veja a Figura  \ref{fig:3}.

\begin{figure}[ht]
\begin{center}
\scalebox{0.25}[0.25]{\includegraphics{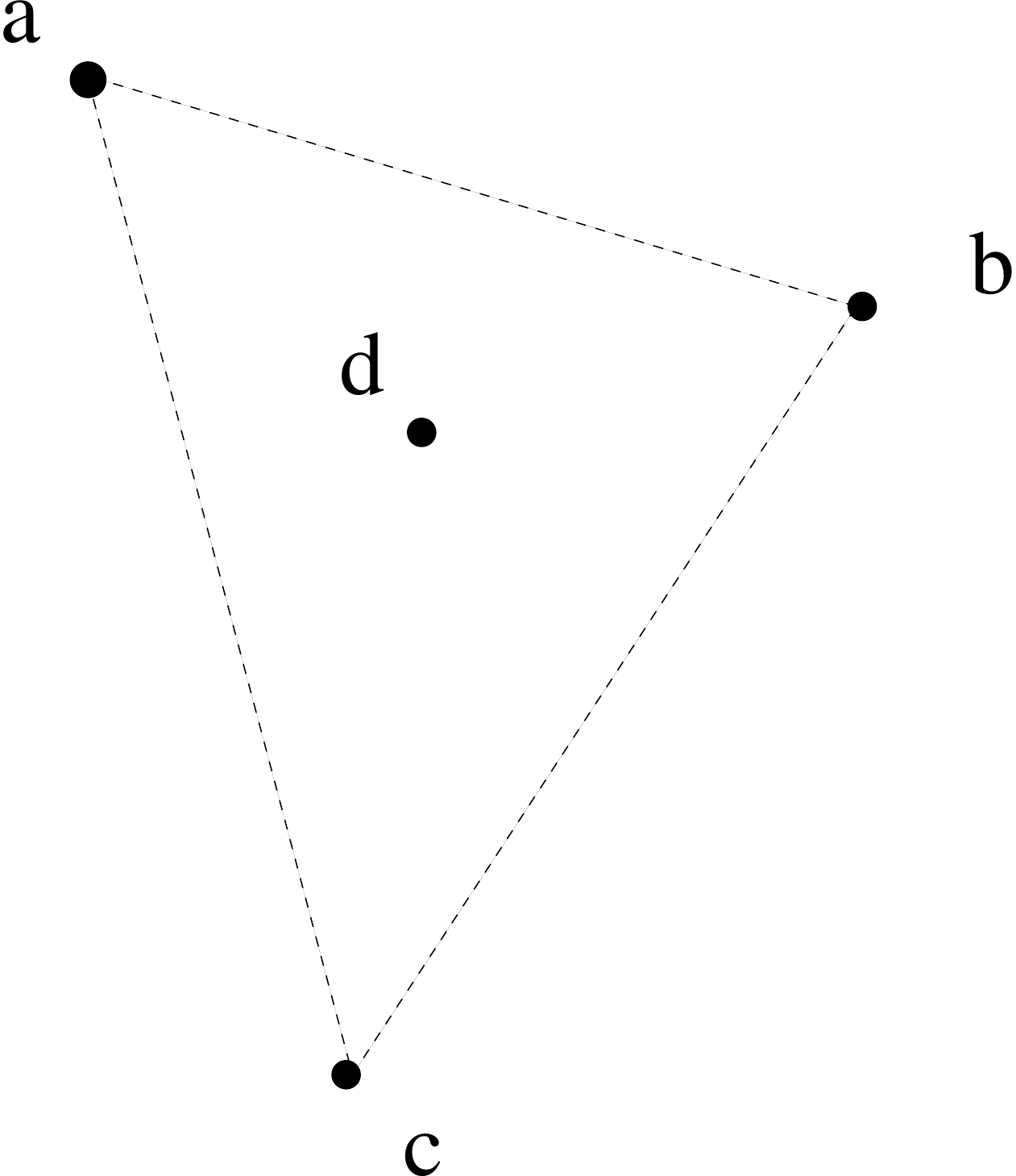}} 
\caption{Todo semiplano contendo $C=\{a,b,c\}$ vai conter $d$.}
\label{fig:3}
\end{center}
\end{figure}

Se nenhum dos pontos pertence \`a envolt\'oria convexa dos tr\^es pontos restantes, o quadril\'atero $a,b,c,d$ \'e convexo. Neste caso, um subconjunto contendo o par de v\'ertices opostos n\~ao pode ser a interse\c c\~ao de um semiplano com  $A=\{a,b,c,d\}$, veja a Fig. \ref{fig:4}.

\begin{figure}[ht]
\begin{center}
\scalebox{0.25}[0.25]{\includegraphics{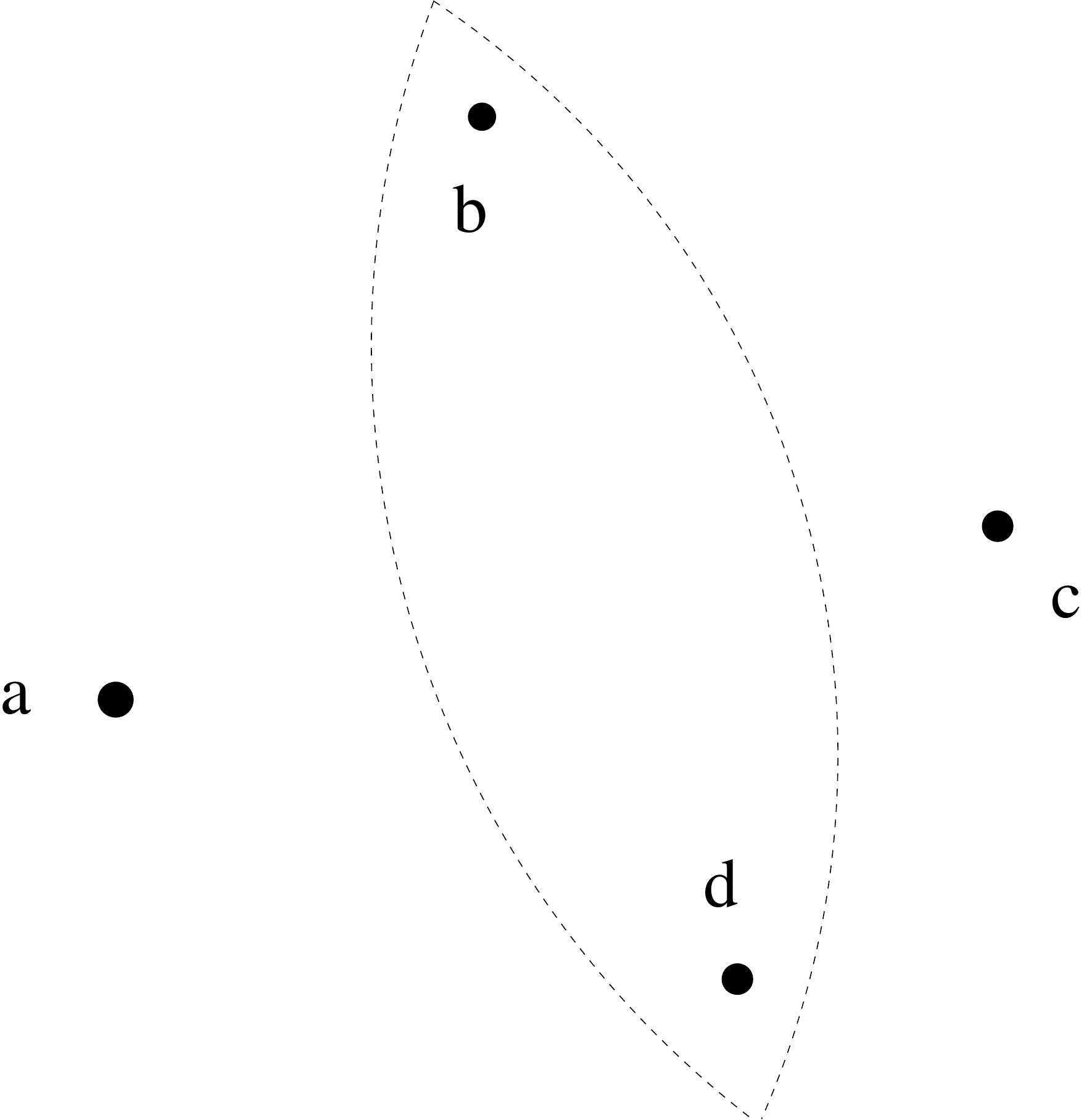}} 
\caption{Todo semiplano que cont\'em $C=\{b,d\}$ vai conter $a$ ou $c$.}
\label{fig:4}
\end{center}
\end{figure}
\end{exemplo}

\begin{exemplo}
\label{ex:rect}
Seja $\mathscr C$ a fam\'\i lia de todos os ret\^angulos em $\R^2$ cujos lados s\~ao paralelos aos eixos coordenados, ou seja,
\[{\mathscr C}=\{[a,b]\times [c,d]\colon a,b,c,d\in\R,~a<b,~c<d\}.\]
Ent\~ao,
\[\VC({\mathscr C})=4.\]
Para come\c car, o conjunto
\[A=\{(1,0),(-1,0),(0,1),(0,-1)\}\]
\'e fragmentado pela classe $\mathscr C$, o que pode ser verificado diretamente, veja Fig. \ref{fig:5}.
\index{dimens\~ao! de Vapnik--Chervonenkis! de ret\^angulos}

\begin{figure}[ht]
\begin{center}
\scalebox{0.25}[0.25]{\includegraphics{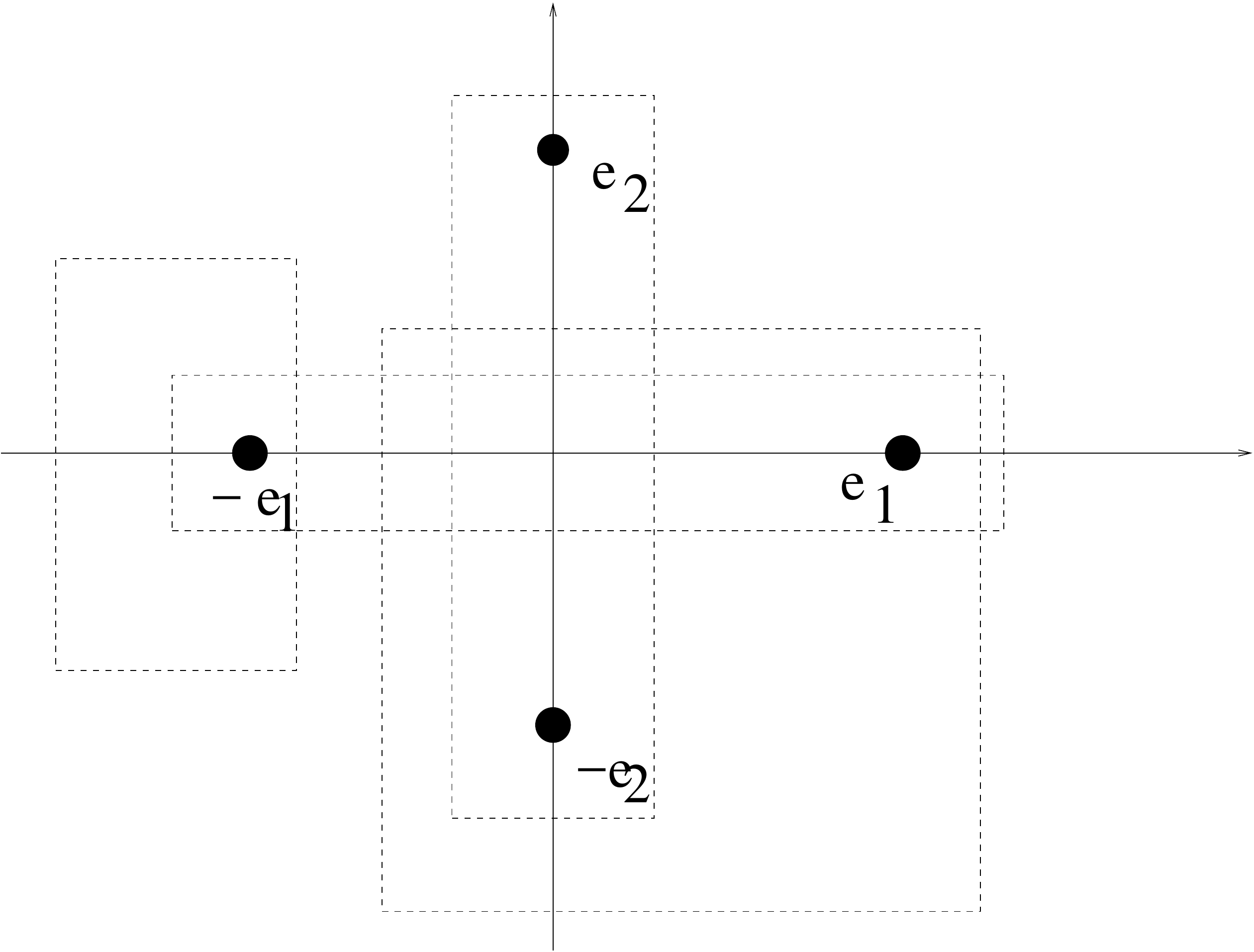}} 
\caption{O conjunto $A=\{\pm e_i\}_{i=1}^2$ \'e fragmentado pelos ret\^angulos.}
\label{fig:5}
\end{center}
\end{figure}

Ao mesmo tempo, vamos mostrar que nenhum conjunto com cinco pontos diferentes pode ser fragmentado pelos ret\^angulos. Seja $A$ um tal conjunto. Escolhemos o ponto ``mais \`a esquerda'', $a$, ou seja, tal que a primeira coordenada $a_1\leq x_1$ para todos $x=(x_1,x_2)\in A$. De mesmo modo, escolha o ponto ``mais \`a direita'',  $b$, o ponto ``mais alto'' $d$ (cuja segunda coordenada \'e maior), assim como o ponto ``mais baixo'', $c$. Veja Fig. \ref{fig:6}. (Pode acontecer que n\~ao todos os pontos $a,b,c,d$ sejam dois a dois distintos).

\begin{figure}[ht]
\begin{center}
\scalebox{0.25}[0.25]{\includegraphics{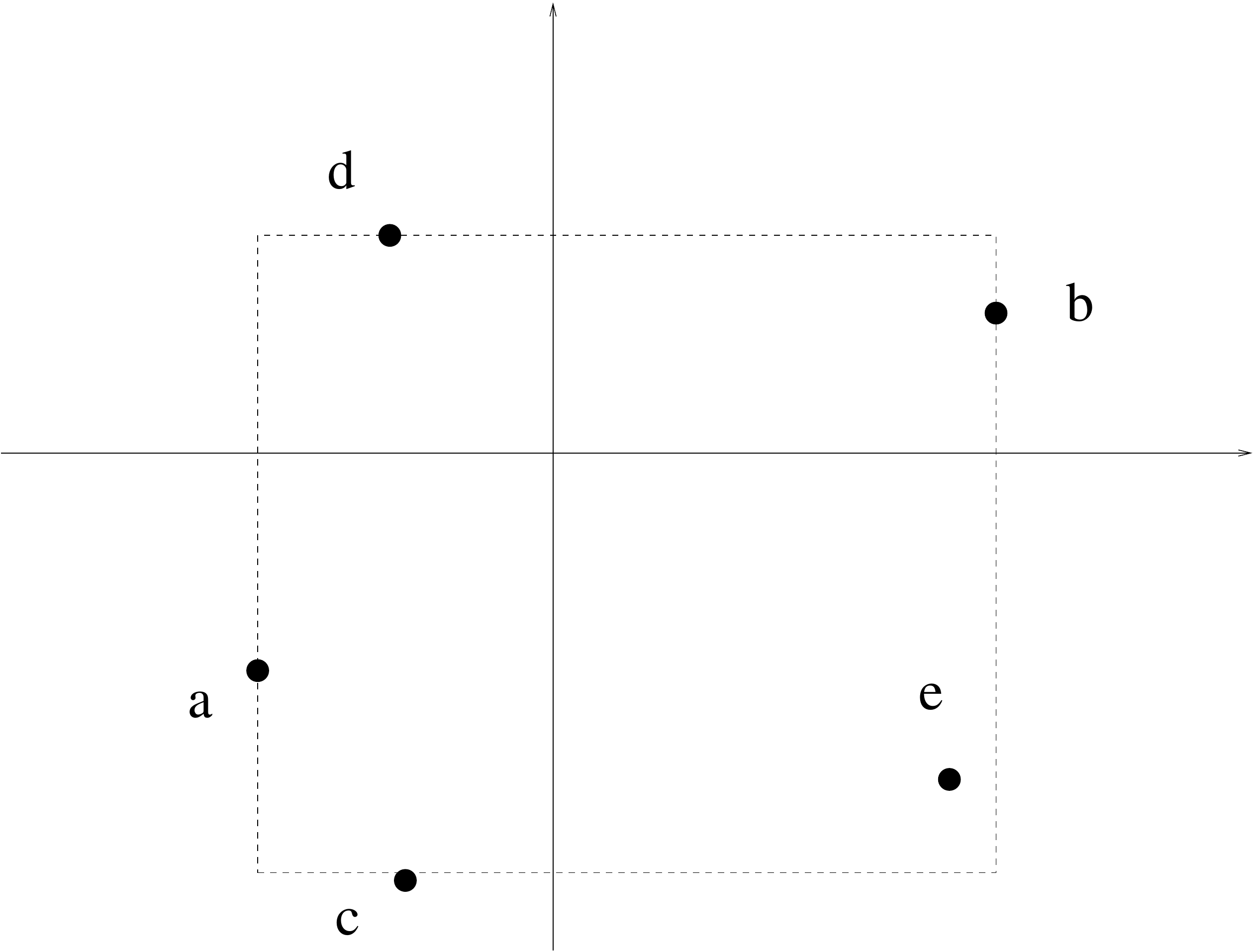}} 
\caption{Nenhum conjunto $A=\{a,b,c,d,e\}$ com cinco pontos \'e fragmentado pelos ret\^angulos.}
\label{fig:6}
\end{center}
\end{figure}

Defina $B=\{a,b,c,d\}$. Como $\abs A\geq 5$, existe pelo menos um ponto $e$ em $A\setminus B$ que n\~ao foi escolhido. Agora, todo ret\^angulo $[x,y]\times [z,w]$ que cont\'em $B$ tem que satisfazer
\[x\leq a_1,~~b_1\leq y,~~z\leq c_2,~~d_2\leq w,\]
ou seja,
\[x\leq a_1\leq e_1\leq b_1\leq y,\]
assim como
\[z\leq c_2\leq e_2\leq d_2\leq w.\]
Por conseguinte, $e\in [x,y]\times [z,w]$ e $A\cap [x,y]\times [z,w]=A$. O subconjunto $B\subsetneqq A$ n\~ao pode ser obtido de $A$ como a interse\c c\~ao com um ret\^angulo.  
\end{exemplo}

\begin{exemplo}
Seja $\mathscr C$ a fam\'\i lia de todos os paralelep\'\i pedos ret\^angulos no espa\c co $\R^d$ com os lados paralelos aos eixos coordenados, ou seja, os subconjuntos da forma 
\[\Pi = [a_1,b_1]\times [a_2,b_2]\times\ldots\times [a_d,b_d].\]
\index{dimens\~ao! de Vapnik--Chervonenkis! de paralelep\'\i pedos}
Ent\~ao,
\[\VC({\mathscr C})=2d.\]

Dado um subconjunto finito $X$ de $\R^d$, denotemos por $\Pi(X)$ o menor  paralelep\'\i pedo (fechado) contendo $X$, ou seja,
\[\Pi(X) =\bigcap_{i=1}^d \pi_i^{-1}\left(\conv(\pi_i(X)\right).\]
Aqui $\pi_i$ \'e a $i$-\'esima proje\c c\~ao coordenada de $\R^d$ para $\R$, e a envolt\'oria convexa em $\R$ \'e simplesmente o menor intervalo que cont\'em $\pi_i(X)$. 

A prova da desigualdade $\leq$ \'e id\^entica \`a prova no exemplo \ref{ex:rect}. Para mostrar a desigualdade $\geq$, observemos que
o conjunto
\[A=\{\pm e_i\}_{i=1}^d\]
com $2d$ pontos \'e fragmentado pelos paralelep\'\i pedos. Seja $B\subseteq A$. Vamos mostrar que $\Pi(B)\cap A = B$, terminando a demonstra\c c\~ao. A inclus\~ao $\supseteq$ \'e \'obvia. Suponha que $x\in A\setminus B$. Existe $i$ tal que $\pi_ix$ \'e da forma $\pm e_i$. Suponhamos $b=e_i$ (o argumento no segundo caso $b=-e_i$ \'e quase id\^entico). O ponto $e_i$ \'e o \'unico ponto no conjunto $A$ cuja $i$-\'esima coordenada \'e um. Isso implica 
\[p_i(B)\subseteq \{-1,0\},\] 
e pela defini\c c\~ao de $\Pi(A)$,
\[b=e_i\notin \Pi(A).\]
\end{exemplo}

O exemplo \ref{ex:half} estende-se aos espa\c cos vetoriais de dimens\~ao qualquer. Precisamos um resultado t\'ecnico seguinte.

\begin{teorema} Seja $V$ um subespa\c co vetorial do espa\c co $\R^\Omega$ de todas as fun\c c\~oes reais sobre o dom\'\i nio $\Omega$. A dimens\~ao VC da classe de conjuntos
\[C_f = \{x\in\Omega\colon f(x)\geq 0\},~~f\in V,\]
\'e menor ou igual a $\dim V$.
\label{t:cf}
\end{teorema}

\begin{proof}
Seja $\dim V=d$. 
Com todo ponto $x\in\Omega$ pode ser associado um funcional linear $\hat x$ de avalia\c c\~ao no ponto $x$, da forma $f\mapsto f(x)$, onde $f\in V$. Este $\hat x$ pertence ao espa\c co vetorial dual, $V^\ast$.
Seja agora $A\subseteq\Omega$ um conjunto com $d+1$ pontos distintos. Como a dimens\~ao do espa\c co dual de $V$ \'e $d$, um dos funcionais lineares de avalia\c c\~ao aos pontos, que denotemos por $x_{d+1}\in A$, pode ser expresso como a combina\c c\~ao linear dos pontos restantes,
\[\hat x_{d+1}=\sum_{i=1}^d\lambda_i\hat x_i,~~x_i\in A.\]
Definamos
\[B= \{x_i\colon 0\leq i\leq d,~\lambda_i\geq 0\}\subseteq A.\]
Seja $f\in V$ tal que $C_f\cap \{x_1,x_2,\ldots,x_d\}=B$. Neste caso, 
$f(x_i)\geq 0 \iff x_i\in B$, e conclu\'\i mos
\[f(x_{d+1}) = \hat x_{d+1}(f) = \sum_{i=1}^d\lambda_i\hat x_i(f) =
\sum_{i=1}^d\lambda_if(x_i) \geq 0,
\]
de onde $x_{d+1}\in C_f$. O conjunto $B$ n\~ao \'e da forma $A\cap C_f$ para $f\in V$ qualquer.
\end{proof}

\begin{exercicio}
Investigar a pergunta seguinte: \'e sempre verdadeiro que a dimens\~ao VC da classe de conceitos $C_f$, $f\in V$, \'e {\em igual} a $\dim V$, ou pode ser estritamente menor?
\end{exercicio}

\begin{exemplo}
\label{ex:spaces}
Seja $\mathscr C$ a cole\c c\~ao de todos os semiespa\c cos (fechados) no espa\c co vetorial $\R^d$:
\[{\mathscr C}=\left\{H_{\vec v,b} \colon\vec v\in\R^d,~b\in\R\right\},\]
onde, como no exemplo \ref{ex:half},
\[H_{\vec v,b}=\{\vec x\in\R^d\colon \langle\vec x,\vec v\rangle\geq b\}.
\]
\index{dimens\~ao! de Vapnik--Chervonenkis! de semiespa\c cos}
Ent\~ao,
\[\VC({\mathscr C})=d+1.\]
A desigualdade $\geq$ \'e bastante f\'acil a verificar: por exemplo, os $d+1$ pontos
\[0,e_1,e_2,\ldots,e_d\]
s\~ao fragmentados por $\mathscr C$, onde $e_i$ denota o $i$-\'esimo vetor de base padr\~ao. Seja $B\subseteq \{ 0,e_1,e_2,\ldots,e_d\}$ qualquer.
Defina o vetor $\vec v\in\R^n$ por
\[\vec v_i =\begin{cases} -1,&\mbox{ if } e_i\notin B,\\
1,&\mbox{ if }e_i\in B.
\end{cases}\]
Se $x\neq 0$ e $x\in B$, temos $\langle x,\vec v\rangle =1$, se $x\neq 0$ e $x\notin B$, temos
$\langle x,\vec v\rangle =-1$, e finalmente $\langle 0,\vec v\rangle =0$. Se $0\in B$, definamos $b=-1/2$, e se $0\notin B$, definamos $b=1/2$. Em ambos os casos, o semiespa\c co $H_{\vec v,b}$ separa $B$ do resto do conjunto $\{ 0,e_1,e_2,\ldots,e_d\}$. 

A cota superior, 
\[\VC({\mathscr C})\leq d+1,\]
segue do teorema \ref{t:cf}, notando que a dimens\~ao do espa\c co $V$ de todas as fun\c c\~oes afins sobre $\R^d$, ou seja, as fun\c c\~oes da forma
\[\vec x\mapsto \langle\vec x,\vec v\rangle - b,\]
\'e igual a $d+1$. 
\end{exemplo}

\begin{exemplo}
\label{ex:balls}
A dimens\~ao VC da fam\'\i lia de todas bolas abertas euclideanas em $\R^d$,
\[B_r(x),~~r>0,~~x\in\R^d,\]
\'e igual a $d+1$. (O mesmo vale para as bolas fechadas, ou bem a mistura de ambos tipos).

O resultado segue da observa\c c\~ao seguinte: um conjunto $A\subseteq\R^d$ \'e fragmentado pelas bolas se e somente se ele \'e fragmentado pelos semiespa\c cos. 

Em primeiro lugar, todo conjunto $A$ fragmentado pelos semiespa\c cos pode ser fragmentado pelas bolas de um raio bastante grande. De fato, as bolas podem aproximar um semiplano com alguma precis\~ao. Em particular, n\~ao \'e dif\'\i cil escolher explicitamente o centro e o raio de uma bola para todo subconjunto de  $X=\{0,e_1,e_2,\ldots,e_n\}$. 

Agora, seja $A$ um conjunto fragmentado pelas bolas. Seja $B\subseteq A$. Existem as bolas $B_1$ e $B_2$ com $B_1\cap A=B$, $B_2\cap A=A\setminus B$. Se $B_1\cap B_2=\emptyset$, ent\~ao pelo teorema de Hahn-Banach as bolas s\~ao separadas por um hiperplano, que separa tamb\'em $B$ e $A\setminus B$. Caso contr\'ario, a interse\c c\~ao de esferas correspondentes \`as bolas $B_1$, $B_2$ \'e uma esfera num hiperplano, e como $B_1\cap B_2\cap A=\emptyset$, os subconjuntos $B$ e $A\setminus B$ s\~ao separados pelos semiespa\c cos determinados pelo hiperplano. 
\end{exemplo}

\begin{exercicio} Deduzir o {\em teorema de Helly:} sejam $x_1,x_2,\ldots,x_{d+2}$ pontos em $\R^d$. Ent\~ao o conjunto de \'\i ndices $\{1,2,\ldots,d+2\}$ pode ser partilhado em dois subconjuntos disjuntos, $I$ e $J$, de tal maneira que as envolt\'orias convexas de conjuntos 
\[\{x_i\colon i\in I\}\mbox{ e }\{x_j\colon j\in J\}\]
t\^em um ponto comum. 
\index{teorema! de Helly}
\end{exercicio}

\begin{observacao}
A situa\c c\~ao com a dimens\~ao VC da fam\'\i lia de todas as bolas num espa\c co normado $E$ de dimens\~ao finita \'e mais complicada. Para o espa\c co $\ell^{\infty}(d)$, ela \'e igual a $\lfloor(3d+1)/2\rfloor$, calculada em \citep*{despres}. O mesmo artigo cont\'em um exemplo de espa\c co normado de dimens\~ao $3$ cuja fam\'\i lia de bolas tem dimens\~ao VC infinita.
\end{observacao}

A nossa apresenta\c c\~ao desta se\c c\~ao segue em grande parte (n\~ao totalmente) \citep*{AB}.

\section{Teorema de Pajor e lema de Sauer--Shelah}

Sejam $\Omega$ um dom\'\i nio (conjunto qualquer) e $\mathscr C$ uma classe de conceitos, ou seja, uma fam\'\i lia n\~ao vazia de partes de $\Omega$. Suponhamos que a dimens\~ao VC de $\mathscr C$ \'e finita, $\VC(\mathscr C)=d$. Agora suponha que $\sigma=\{x_1,\ldots,x_n\}\subseteq\Omega$ \'e uma amostra, tamb\'em finita, mas intuitivamente de uma cardinalidade $n\gg d$. A classe $\mathscr C$ induz v\'arias rotulagens sobre $\sigma$: se $C\in {\mathscr C}$, ent\~ao
\[\chi_C\vert_\sigma\colon x_i\mapsto \chi_C(x_i)\in \{0,1\}\]
\'e uma rotulagem. Denotemos
\[{\mathscr C}\vert_\sigma = \{C\cap\sigma\colon C\in {\mathscr C}\}.\]
Se $n>d$, nem todas as rotulagens sobre $\sigma$ s\~ao desta forma.
Qual \'e o tamanho da classe ${\mathscr C}\vert_\sigma$ de todas as rotulagens induzidas pela classe $\mathscr C$ sobre $\sigma$? A resposta \'e dada pelo importante resultado, mostrado independentemente e aproximadamente ao mesmo tempo por \citep*{sauer}, \citep*{shelah}, e \citep*{VC:71}, e chamado lema de Sauer ou lema de Sauer--Shelah.

\begin{teorema}[Lema de Sauer--Shelah]
Seja $\mathscr C$ uma classe de conceitos satisfazendo 
$\VC({\mathscr C})\leq d$.
 Se $\sigma\subseteq\Omega$, $\sharp\sigma=n\geq d$, ent\~ao
\[\sharp{\mathscr C}\vert_\sigma\leq \sharp B_d.\]
Em particular,
\[\sharp{\mathscr C}\vert_\sigma < \left(\frac{en}{d} \right)^d.\]
\label{l:sauer-shelah}
\index{lema! de Sauer--Shelah}
\end{teorema}

Aqui $B_d$ \'e uma bola fechada de raio $d$ no cubo de Hamming de dimens\~ao $n$ n\~ao normalizado. 

\begin{definicao}
Dada uma classe de conceitos, $\mathscr C$, definamos o {\em $n$-\'esimo coeficiente de fragmenta\c c\~ao} ({\em $n$-th shattering coefficient}) de $\mathscr C$ como o maior n\'umero de rotulagens induzidas por $\mathscr C$ sobre as amostras com $\leq n$ pontos:
\[s(n, {\mathscr C}) = \sup\{\sharp{\mathscr C}\vert_\sigma\colon \sigma\subseteq\Omega,~\sharp\sigma\leq n\}.\]
\index{coeficiente! de fragmenta\c c\~ao}
\end{definicao}

Por exemplo, a dimens\~ao VC de $\mathscr C$ \'e o supremo de cardinalidades $n$ tais que
\[s(n,{\mathscr C})=2^n.\]

Ent\~ao, o lema de Sauer--Shelah pode ser reformulado assim:
\[s(n, {\mathscr C})  < \left(\frac{en}{\VC({\mathscr C})} \right)^{\VC({\mathscr C})}.\] 
A mensagem mais importante deste resultado \'e que o n\'umero de rotulagens induzidas sobre uma amostra de tamanho $n\to\infty$ cresce polinomialmente em $n$, isto \'e, relativamente devagar:
\[s(n,{\mathscr C})={\mathrm{poly}}(n)\mbox{ se }\VC({\mathscr C})<\infty.\]

Vamos deduzir o lema de Sauer--Shelah do teorema seguinte.

\begin{teorema}[Teorema de Pajor \citep*{pajor}]
Seja $\mathscr C$ uma classe de conceitos com $m$ elementos, $m\geq 1$. Ent\~ao $\mathscr C$ fragmenta pelo menos $m$ subconjuntos de $\Omega$ dois a dois diferentes.
\index{teorema! de Pajor}
\label{t:pajor}
\end{teorema}

\begin{proof}
N\'os j\'a vimos o caso particular de $m=1$ no exemplo \ref{ex:so}: uma classe que consiste de \'unico conceito fragmenta pelo menos um conjunto, a saber, o conjunto vazio. Isso \'e a base da prova.

Suponhamos agora que que a afirma\c c\~ao do teorema \'e verdadeira para todos $i$, $1\leq i\leq m$, onde $m\geq 1$. Seja $\mathscr C$ uma fam\'\i lia qualquer de subconjuntos de $\Omega$ com $m+1$ elementos. Segue-se que $\cup{\mathscr C}\supsetneqq \cap{\mathscr C}$, e podemos escolher um elemento $x_0$ que pertence a alguns membros de $\mathscr C$, mas n\~ao a todos deles. Dividamos $\mathscr C$ em duas subclasses: $\mathscr C_0$ consiste de todos $A\in {\mathscr C}$ que cont\'em $x_0$, e $\mathscr C_1$ consiste de todos $B\in {\mathscr C}$ que n\~ao cont\'em $x_0$. Ambas s\~ao n\~ao vazias, e por conseguinte os n\'umeros
\[k=\sharp {\mathscr C}_0,~l=\sharp {\mathscr C}_1,\]
satisfazem $1\leq k,l\leq m$, $k+l=m+1$. Segundo a hip\'otese indutiva, existem conjuntos $A_1,\ldots,A_k$ fragmentados por $\mathscr C_0$, bem como conjuntos $B_1,\ldots,B_l$ fragmentados por $\mathscr C_1$.
Se todos os conjuntos na lista
\[A_1,A_2,\ldots,A_k,B_1,B_2,\ldots,B_l\]
s\~ao dois a dois distintos, ent\~ao temos uma fam\'\i lia de $m+1$ conjuntos fragmentados pela classe $\mathscr C={\mathscr C}_0\cup {\mathscr C}_1$. Por\'em, pode ocorrer que alguns conjuntos nas duas listas $A_i$ e $B_j$ s\~ao iguais. 

Se $A_i=B_j$, ent\~ao este conjunto \'e fragmentado por $\mathscr C_0$, assim como por $\mathscr C_1$. Neste caso, vamos observar que o conjunto $A_i\cup\{x_0\}$ \'e fragmentado por $\mathscr C$. Se $B\subseteq A_i\cup\{x_0\}$ e $B\ni x_0$, ent\~ao existe um conjunto $C\in {\mathscr C}_0$ com $C\cap A= B\setminus \{x_0\}$, e neste caso, $C\cap A_i\cup\{x_0\}= B$. Se $B\subseteq A_i\cup\{x_0\}$ e $B\not\ni x_0$, existe um conjunto $C\in {\mathscr C}_1$ com $C\cap A= B$, o que implica $C\cap A_i\cup\{x_0\}= B$. Ao mesmo tempo, $A_i\cup\{x_0\}$ n\~ao pode ser fragmentado por $\mathscr C_0$ (cujos elements cont\'em $x_0$) nem por $\mathscr C_1$ (cujos elementos n\~ao cont\'em $x_0$). Logo, $A_i\cup\{x_0\}$ \'e diferente de todos os outros conjuntos na lista. 

Conclu\'\i mos: os conjuntos na primeira lista, $A_1,\ldots,A_k$, bem como os conjuntos na segunda lista $B_1,\ldots,B_l$ que n\~ao se encontram na primeira lista, e os conjuntos $A_i\cup\{x_0\}$ para todos os conjuntos que se encontram na interse\c c\~ao de duas listas, s\~ao todos fragmentados pela classe $\mathscr C$ e s\~ao dois a dois diferentes. O n\'umero de tais conjuntos \'e $m+1$. 
\end{proof}

Para deduzir o lema de Sauer--Shelah do teorema de Pajor, seja $\mathscr C$ uma classe de conceitos com a dimens\~ao de Vapnik--Chervonenkis $\leq d$. Ent\~ao, a classe $\mathscr C$, e por conseguinte a classe ${\mathscr C}\vert\sigma$, n\~ao fragmentam nenhum conjunto com mais de $d$ pontos. 
Todos os subconjuntos de $\sigma$ fragmentados pelo ${\mathscr C}\vert\sigma$ tem a cardinalidade $\leq d$. Segundo o teorema de Pajor, a classe ${\mathscr C}\vert\sigma$ fragmenta pelo menos $s({\mathscr C},n)$ subconjuntos dois a dois diferentes de $\sigma$.
Basta notar que a fam\'\i lia de subconjuntos de $\sigma$ com $\leq d$ pontos \'e equipotente \`a bola $B_d(0)$. Conclu\'\i mos: $s({\mathscr C},n)$ n\~ao pode ser maior do que $\sharp B_d(0)$. \qed

Revisitaremos o lema de Sauer--Shelah no subse\c c\~ao \ref{ss:prabaixo}.

\section{Redes de unidades computacionais}

\subsection{Unidade de computa\c c\~ao}

Como nos j\'a mencionamos, a origem de classes de conceitos \'e algor\'\i tmico: cada tal classe, $\mathscr C$, \'e uma classe de conceitos (fun\c c\~oes bin\'arias) gerados por um algoritmo, para todos os valores de par\^ametros poss\'\i veis. Um tal algoritmo \'e conhecido como uma {\em unidade de computa\c c\~ao}. Eis um exemplo simples mas importante.

\begin{exemplo}[Perceptron] Uma unidade de computa\c c\~ao que gera o espa\c co de todos os semiespa\c cos do espa\c co euclidiano $\R^d$ (como no exemplo \ref{ex:spaces}) \'e conhecida na aprendizagem de m\'aquina como {\em perceptron}. \'E uma fun\c c\~ao 
\[\eta\left(\sum_{i=1}^d w_ix_i -\theta \right),\]
onde $w_1,w_2,\ldots,w_n,\theta\in\R$ s\~ao os valores de par\^ametros, $x_1,\ldots,x_d$ s\~ao os valores de entrada (inputs), e $\eta$ \'e a fun\c c\~ao de Heaviside:
\[\eta(x) =\begin{cases}
1,&\mbox{ se }x\geq 0,\\
0,&\mbox{ se }x<0.
\end{cases}
\]
O {\em estado} de um perceptron \'e o vetor de valores de par\^ametros,
$\omega=(w_1,w_2,\ldots,w_d,\theta)\in\R^{d+1}$, onde os $w_i$ s\~ao os {\em pesos,} e $\theta$, o {\em valor limiar}. Cada estado determina uma fun\c c\~ao bin\^aria (isto \'e, um classificador) sobre $\R^d$, e um correspondente semiespa\c co. Deste modo, o perceptron pode ser visto como uma aplica\c c\~ao
\[\R^{d+1}\times\R^d\to\{0,1\},\]
onde $\R^{d+1}$ \'e o espa\c co de estados dele, e $\R^d$, o dom\'\i nio.
\index{estado! de perceptron}
\index{valor! limiar de perceptron}
\index{perceptron}
\end{exemplo}

Com a base no exemplo acima, vamos isolar a no\c c\~ao formal seguinte, mais geral e mais b\'asico do que a no\c c\~ao de uma regra de aprendizagem (a seguir mais tarde).

\begin{definicao}
Uma {\em unidade computacional (bin\'aria)} \'e uma aplica\c c\~ao 
\[u\colon {\mathcal W}\times \Omega \to \{0,1\},\]
onde $\mathcal W$ \'e o {\em espa\c co de estados} da unidade e $\Omega$ \'e o dom\'\i nio.
\index{unidade computacional}
\index{espa\c co! de estados! da unidade computacional}
\end{definicao}

Desse modo, dado um estado $\omega\in {\mathcal W}$ e qualquer ponto $x\in \Omega$, a unidade calcula o valor $u^\omega(x)$, seja $0$, seja $1$. Toda aplica\c c\~ao $u^\omega\colon x\mapsto u^\omega(x)$ \'e um classificador sobre $\Omega$, e desse modo, a fun\c c\~ao indicadora de um subconjunto $C^\omega\subseteq\Omega$ (um conceito). 

\begin{observacao}
A unidade de computa\c c\~ao $u$ pode ser assim vista como uma fam\'\i lia de conceitos $C^\omega$, $\omega\in {\mathcal W}$, indexada com um conjunto de estados. Como o conjunto de estados, pode ser usada a pr\'opria classe de conceitos $\mathscr C$. Assim, n\~ao h\'a diferen\c ca principal entre uma unidade computacional e uma classe de conceitos.
\end{observacao}

\begin{observacao}
Eis mais um ponto de vista sobre a no\c c\~ao de unidade de com\-pu\-ta\-\c c\~ao. As informa\c c\~oes sobre uma tal unidade, $u$, podem ser completamente codificadas no subconjunto seguinte do produto cartesiano $\Omega\times {\mathcal W}$:
\[{\mathcal C}_u =\{(x,\omega)\colon x\in\Omega,~ u\in {\mathcal W},~ u^\omega(x)=1\} \equiv \{(x,\omega)\colon x\in C^\omega\}.\]
Ent\~ao, pode-se dizer que uma unidade de computa\c c\~ao (de ponto de vista puramente conjunt\'\i stico) \'e simplesmente um subconjunto qualquer,
\[{\mathcal C}_u\subseteq {\mathcal W}\times \Omega.\]
As ``se\c c\~oes horizontais'' do conjunto ${\mathcal C}_u$ (ao n\'\i vel $\omega$) s\~ao os conceitos $C^\omega$.
\end{observacao}

\begin{observacao}
Vamos eventualmente impor sobre a defini\c c\~ao de uma unidade de computa\c c\~ao o requirimento adicional que a fun\c c\~ao $u\colon {\mathcal W}\times\Omega\to\{0,1\}$ (ou, de modo equivalente, o conjunto ${\mathcal C}_u$) seja {\em mensur\'avel no sentido de Borel} relativo a uma {\em estrutura boreliana padr\~ao} sobre $\mathcal W$ e $\Omega$. Estas no\c c\~oes est\~ao formalizadas nos ap\^endices \ref{a:variaveis} e \ref{apendice:padrao}. Tais restri\c c\~oes n\~ao s\~ao restringentes, e unicamente servem para eliminar algumas patologias da natureza conjunt\'\i stica.
\end{observacao}

\begin{definicao} Pode acontecer que o dom\'\i nio de uma unidade de computa\c c\~ao, $\Omega$, \'e um produto cartesiano:
\[\Omega = \Omega_1\times\Omega_2\times \ldots \times \Omega_k.\]
Nesse caso, digamos que $u$ \'e uma unidade com $k$ entradas (inputs):
\[u^\omega(x) = u^\omega(x_1,x_2,\ldots,x_d).\]
\end{definicao}

\begin{exemplo}
O {\em perceptron} \'e uma unidade com $d$ entradas.
\end{exemplo} 

\begin{definicao}
A dimens\~ao de Vapnik--Chervonenkis de uma unidade de computa\c c\~ao, $u$, \'e definida como a dimens\~ao VC da classe de conceitos correspondente,
\[{\mathscr C}_u=\{C^\omega\colon \omega\in {\mathcal W}\},\]
gerada por $u$.
\end{definicao}

\begin{exemplo}
A dimens\~ao VC de um perceptron com $d$ entradas \'e igual \`a $d+1$.
\end{exemplo}

\begin{observacao}
Na {\em teoria algor\'\i tmica} (ou: {\em computacional}) {\em de aprendizagem de m\'aquina} o estudo da natureza e do funcionamento das unidades de computa\c c\~ao \'e particularmente importante, com enfase sobre a complexidade computacional delas. N\~ao tocamos neste assunto aqui. Veja, por exemplo, \citep*{kearns_vazirani}.
\end{observacao}

\begin{observacao}
Nessa perspetiva, uma {\em regra de aprendizagem} \'e uma unidade computacional cujos estados s\~ao todas as amostras rotuladas. 
\end{observacao}

Alarguemos um pouco a nossa cole\c c\~ao de exemplos das unidades computacionais.

\begin{exemplo}[Regra de histograma] 
Seja $\Omega$ um dom\'\i nio qualquer.
Uma unidade computacional para $\Omega$ chamada {\em regra de histograma} \'e determinada pelos valores de dois par\^ametros: $\gamma$, uma parti\c c\~ao finita de $\Omega$, e $\sigma=(x_1,x_2,\ldots,x_n,\e_1,\e_2,\ldots,\e_n)$, uma amostra rotulada, onde $x_i\in\Omega$ e $\e_i\in\{0,1\}$. Desse modo, o espa\c co de estados do histograma pode ser identificado com o produto cartesiano de fam\'\i lia de todas as parti\c c\~oes finitas de $\Omega$ com o conjunto $\cup_{i=1}^{\infty}\Omega^n\times \{0,1\}^n$ de todas as amostras rotuladas em $\Omega$.

Dado um ponto $x\in\Omega$ qualquer, assim como uma parti\c c\~ao $\gamma$ e uma amostra rotulada $\sigma$, existe um e apenas um elemento $A\in\gamma$ que cont\'em $x$ como elemento. Denotemos por $\sigma\restriction_A$ uma amostra rotulada que consiste de todos os pontos $x_i$, $i=1,2,\ldots,n$ em $A$, junto com os seus r\'otulos originais. A regra de histograma assina a $x$ o r\'otulo majorit\'ario da amostra $\sigma\restriction_A$. No caso de empates (o mesmo n\'umero de $0$ que de $1$ entre os r\'otulos) a gente pode, por exemplo, sempre escolher $1$, de mesmo que no caso onde $\sigma\restriction_A$ \'e vazia:
\[u^{\gamma,\sigma}(x) = \eta\left(\sum_{x_i\in A}\left(\e_i-\frac 12\right)\right).\]
Em particular, a f\'ormula vai assinar o r\'otulo $1$ a $x$ no case se $\sigma$ \'e disjunto de $A$ (pois a soma de uma fam\'\i lia vazia de reais \'e igual a $0$).

A origem do termo ``histograma'' \'e que o valor da nossa fun\c c\~ao em todo ponto do dom\'\i nio \'e completamente determinado por um ``histograma'' (ou, melhor, dois histogramas) contando os n\'umeros de $0$s e de $1$s em cada elemento da parti\c c\~ao $\gamma$. O r\'otulo de $x$ depende de qual de dois histogramas seja mais alto em $x$.
\index{regra! de histograma}
\end{exemplo}

\begin{exemplo}[\'Arvore de decis\~ao] 
Uma \'arvore de decis\~ao \'e uma unidade de computa\c c\~ao de tipo regra de histograma, cuja parti\c c\~ao do dom\'\i nio $\Omega=\R^m$ consiste de ret\^angulos, ou seja, produtos cartesianos de intervalos (da forma $[a,b)$, $(-\infty,a)$, $[b,+\infty)$, e $(-\infty,+\infty)$).

Os ret\^angulos s\~ao tipicamente constru\'\i das recursivamente, a partir de uma amostra rotulada, $\sigma$. Ao primeiro passo, escolhemos uma coordenada $i\in [m]$ e um n\'umero real $a_1$, dividindo o espa\c co $\R^m$ em dois ret\^angulos pelo hiperespa\c co $x_i=a$. A sele\c c\~ao de $a$ \'e feita de modo para maximizar uma medida de separa\c c\~ao de uma amostra rotulada em duas classes com r\'otulas mais diferentes poss\'\i veis. Cada um de dois ret\^angulos resultantes \'e dividido em dois subret\^angulos ao longo de uma outra coordenada $j\in [m]$, escolhendo os valores $a_{21}$ e $a_{22}$, e o procedimento continua at\'e o momento quando todo ret\^angulo conter menos de $k$ pontos de $\sigma$ (o valor $k$ \'e predefinido). 
\index{\'arvore de decis\~ao}
\end{exemplo}

\subsection{Agrupando unidades de computa\c c\~ao em redes}
As unidades de com\-pu\-ta\-\c c\~ao podem ser agrupadas em redes, produzindo as unidades cada vez mais complexas. 

\begin{exemplo}[Aprendizagem ensemble (ensemble learning)] Uma estrat\'egia importante para criar novos algoritmos de aprendizagem \'e de tomar o voto majorit\'ario entre muitas vers\~oes de regras de aprendizagem (tipicamente, do mesmo tipo mas com os par\^ametros vari\'aveis). 

Por exemplo, o algoritmo provavelmente mais popular na pr\'atica atual de cientistas de dados, Floresta Aleat\'oria (Random Forest, RF), usa o voto majorit\'ario entre um alto n\'umero (por exemplo, $500$) de \'arvores de decis\~ao constru\'\i dos usando par\^ametros escolhidos aleatoriamente (tais como as coordenadas $i$ de divis\~ao).

Dado um {\em ensemble} de classificadores, $u_1,u_2,\ldots,u_k$, e um ponto $x$, os valores de predi\c c\~oes $u_1(x),u_2(x),\ldots,u_k(x)$, s\~ao alimentados a uma unidade computacional, que conta os votos e sai o valor, $0$ ou $1$, recebendo mais votos. Uma tal unidade pode ser um perceptron:
\[\eta\left(\sum_i \left(x_i-\frac 12\right)\right).\]
Graficamente, a rede combinada das unidades de computa\c c\~ao pode ser representada assim:
\begin{figure}[ht]
\begin{center}
\scalebox{0.3}[0.3]{\includegraphics{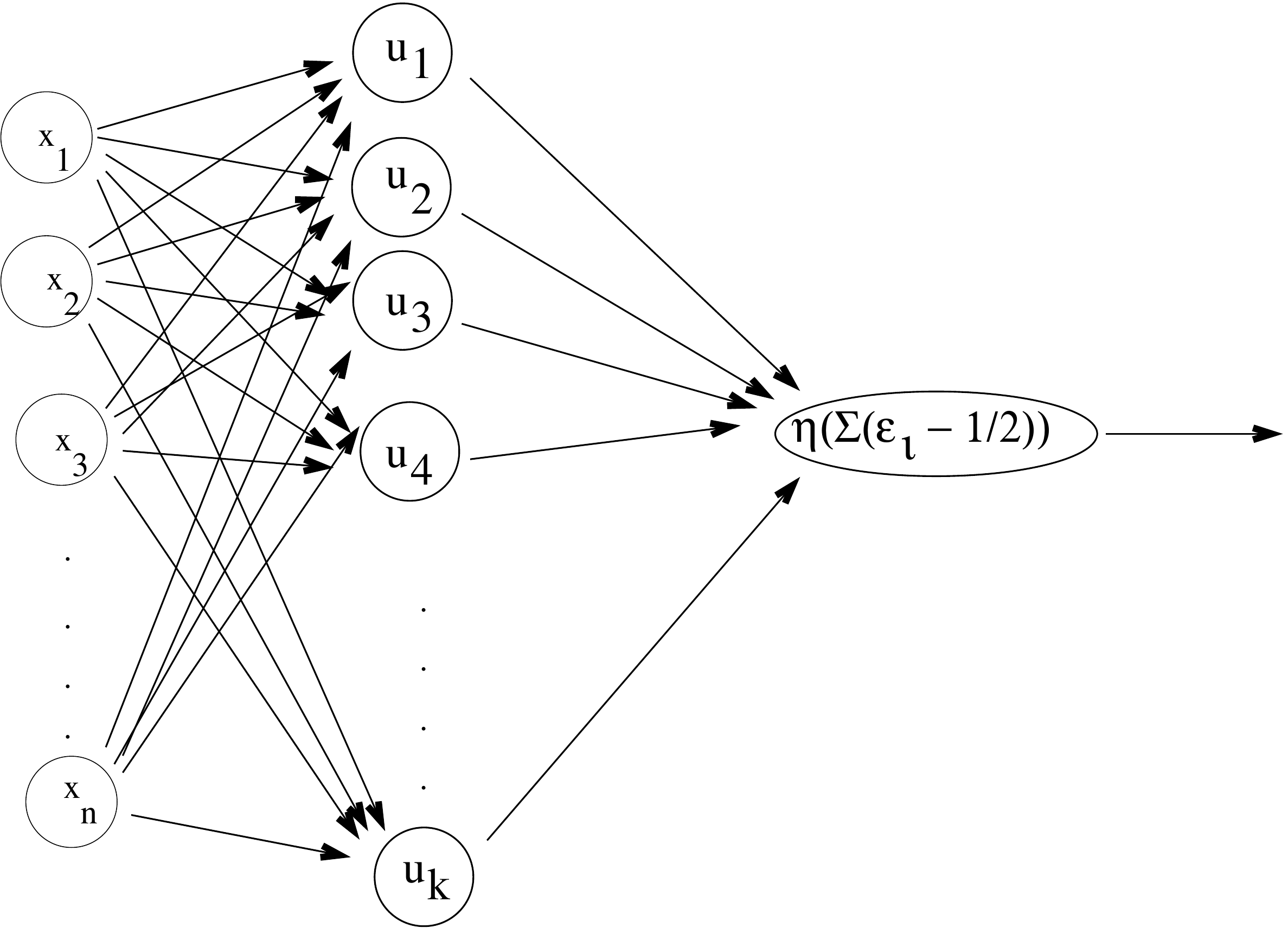}} 
\caption{Uma rede de aprendizagem ensemble.}
\label{fig:ensemble}
\end{center}
\end{figure}
Os $x_i$, $i=1,2,\ldots,n$, representam os valores da entrada (inputs), todos alimentados \`as todas unidades de computa\c c\~ao $u_j$, $j=1,2,\ldots,k$. As sa\'\i das de unidades $u_j$ s\~ao alimentadas ao perceptron determinando o voto majorit\'ario. A sa\'\i da do perceptron \'e a sa\'\i da da rede. Duas unidades s\~ao ligadas por uma aresta se a sa\'\i da da primeira unidade serve de uma entrada da segunda. As unidades s\~ao agrupadas em tr\^es camadas: a camada $0$ de valores de entrada, a camada $1$ de unidades $u_j$, e a camada $2$ que s\'o cont\'em a unidade de sa\'\i da. 
\index{aprendizagem ensemble}
\end{exemplo}

M\'etodos de aprendizagem ensemble s\~ao uma ferramenta entre as mais importantes para construir novos algoritmos. Veja por exemplo o livro \citep*{zhou_ensemble}.

Com base no \'ultimo exemplo, vamos formalizar a no\c c\~ao de uma rede de unidades de computa\c c\~ao bin\'arias.

\begin{definicao}
Um grafo dirigido (simples) \'e um par, $(V,E)$, que consiste de um conjunto $V$ de {\em v\'ertices,} ou (mais tipicamente nesse contexto) {\em n\'os,} e um conjunto $E\subseteq V^2$ de pares dirigidos $(x,y)$, chamados {\em arestas} ou {\em arcos}. Suponhamos que $x\neq y$ (n\~ao tem la\c cos). O {\em grau de entrada} do n\'o $x$, $\mathrm{indeg}(x)$, \'e o n\'umero $\sharp\{y\in V\colon (y,x)\in E\}$, e o {\em grau de sa\'\i da} de $x$, $\mathrm{outdeg}(x)$, \'e definido similarmente. 

\begin{figure}[ht]
\begin{center}
\scalebox{0.3}[0.3]{\includegraphics{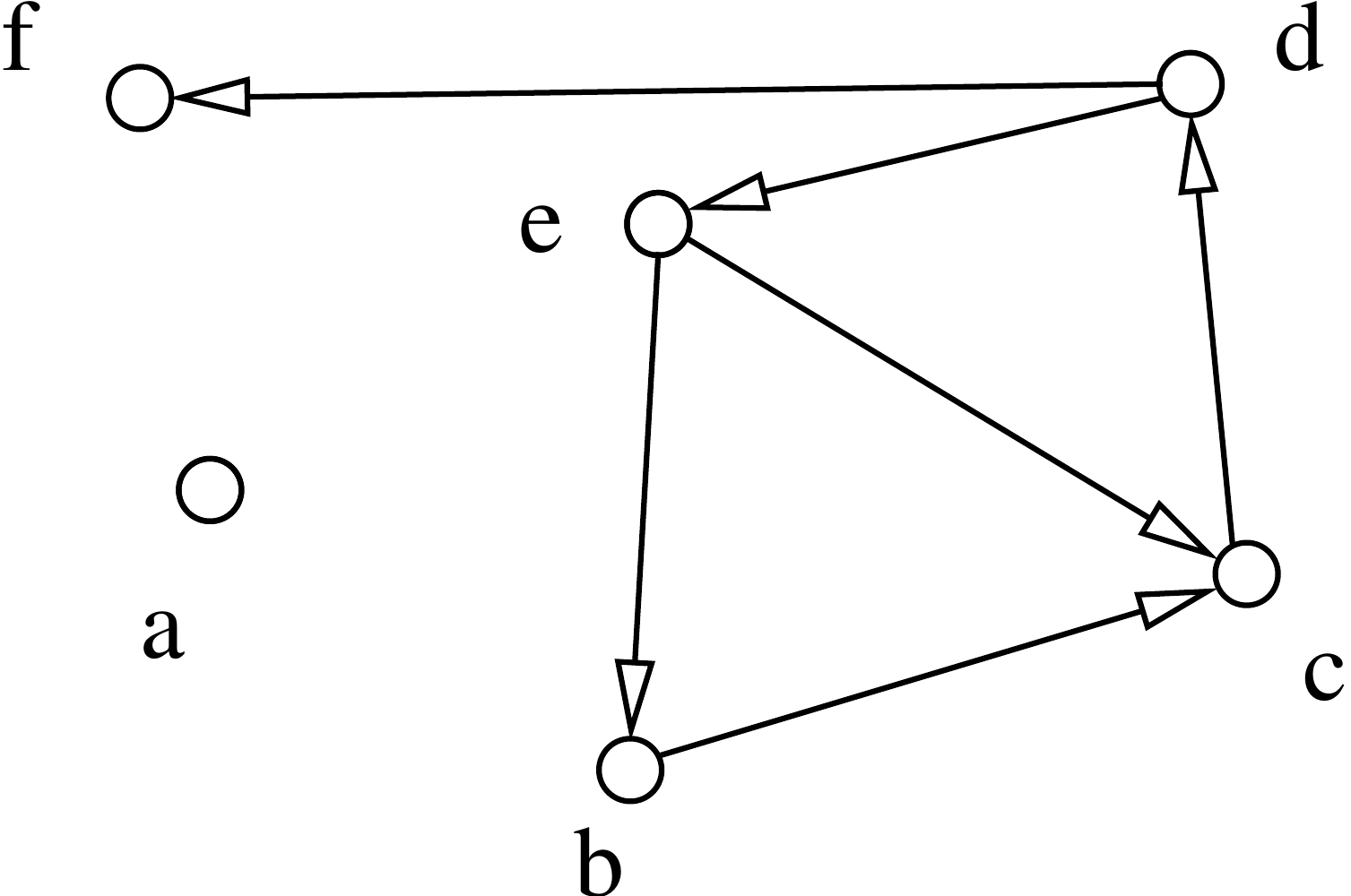}} 
\caption{Um grafo dirigido com $6$ v\'ertices.}
\label{fig:graphs}
\end{center}
\end{figure}
\index{grafo! dirigido}
\index{grau! de entrada}
\index{grau! de sa\'\i da}
\end{definicao}

Sup\~oe-se que existe pelo menos um n\'o $x$ com $\mathrm{indeg}(x)=0$. Tais n\'os, chamados {\em n\'os de entrada,} aceitam os valores de input, tipicamente os valores reais. Por exemplo, se queremos tratar um vetor $x\in\R^m$, precisamos $m$ n\'os de input, um para cada coordenada $x_i$. 
\index{n\'o! de entrada}

A outra hip\'otese \'e que existe pelo menos um n\'o $x$ com $\mathrm{outdeg}(x)=0$. S\~ao os {\em n\'os de sa\'\i da,} produzindo os resultados de computa\c c\~ao. Para simplificar as coisas, suponhamos que existe o \'unico n\'o de sa\'\i da. 
\index{n\'o! de sa\'\i da}

\begin{definicao}
Uma {\em rede de unidades de computa\c c\~ao}, $\mathscr N$, \'e um grafo dirigido, que possui pelo menos um n\'o de entrada e cujos n\'os que n\~ao s\~ao os de entrada s\~ao unidades de computa\c c\~ao bin\'arias, com $\mathrm{indeg}(u)$ valores de entrada cada uma. 
\index{rede! de unidades de computa\c c\~ao}
\end{definicao}

Ent\~ao, para cada estado $\omega$ de $u$, a fun\c c\~ao bin\'aria $u^\omega$ 
aceita $\mathrm{indeg}(u)$ argumentos. 
O dom\'\i nio de $u^\omega$ \'e o produto $\R^p\times\{0,1\}^q$, onde $p+q=\mathrm{indeg}(u)$. Aqui, $p$ \'e o n\'umero de $x$ tais que $(x,u)\in E$ e $x$ \'e um n\'o de entrada, e $q$ \'e o n\'umero de $v$ tais que $(v,u)\in E$ e $v$ \'e uma unidade computacional.

\begin{definicao}
O {\em espa\c co de estados} da rede \'e um subconjunto, $\mathcal W$, do produto $\prod_{u}{\mathcal W}_u$. Desse modo, um estado $\omega =(\omega_u)_{u}$ da rede determina um estado, $\omega_u$, de cada unidade computacional, $u$. 
\index{espa\c co! de estados! de uma rede}
\end{definicao}

Na pr\'atica, ao inv\'es desse subconjunto, pode-se tratar de uma {\em parametriza\c c\~ao} de $\mathcal W$. O {\em treinamento} da rede \'e um procedimento algor\'\i tmico para escolher um estado, $\omega$. 

\begin{definicao}
No caso se uma rede de unidades computacionais, $\mathscr N$, n\~ao tem ciclos (como um grafo dirigido), ela \'e dita uma rede {\em sem realimenta\c c\~ao} ({\em feed-forward}). 
\index{rede! sem realimenta\c c\~ao}
\end{definicao}

A partir desse momento, estudemos somente as redes sem realimenta\c c\~ao.
Dado um estado, $\omega$, de uma tal rede, podem-se definir recursivamente as fun\c c\~oes $f^\omega_u$ em toda unidade de computa\c c\~ao, como composi\c c\~oes das fun\c c\~oes  $u^\omega$, na ordem determinada pelo grafo. Os argumentos da fun\c c\~ao $u^\omega$ s\~ao ou os valores das fun\c c\~oes $u^\omega$ onde a unidade $v$ \'e ligada com $u$ ($(v,u)\in E$), ou os valores de entrada atribu\'\i dos aos n\'os de entrada $x$ ligados a $u$ ($(x,u)\in E$). A fun\c c\~ao bin\'aria $f^{\omega}_{u_{out}}$, onde $u_{out}$ \'e o n\'o de sa\'\i da, aceita os valores de input e produz $0$ ou $1$. 

Mais especificamente, enumeremos os n\'os da rede, $x_1,x_2,\ldots,x_k$ da maneira que se $(x_i,x_j)\in E$, ent\~ao $i<j$. Agora, suponha que um valor seja atribu\'\i do a cada um dos $m$ n\'os de entrada, $\sigma_1,\ldots,\sigma_m$. Para todo $i=1,2,\ldots,k$, calculamos recursivamente o valor $f^{\omega}_i(\sigma)=f^{\omega}_{x_i}(\sigma)$ como segue:
\begin{itemize}
\item
$f^{\omega}_i(\sigma)=\sigma$ se $x_i$ \'e um n\'o de entrada;
\item $f^{\omega}_i(\sigma)=x_i^{\omega}(f^\omega_{i_1}(\sigma),\ldots,f^\omega_{i_s}(\sigma))$, 
onde $x_{i_1},\ldots,x_{i_{\mathrm{indeg}(x)}}$ s\~ao os n\'os ligados \`a unidade computacional $x_i$. 
\end{itemize}

O que faz o procedimento bem-definido, \'e o fato de que quando $j$ for conectado com $i$, temos $j<i$ e logo o valor de $f^{\omega}_j(\sigma)$ j\'a foi calculado quando chegar a hora de calcular o valor $f^{\omega}_i(\sigma)$.

Denotemos por ${\mathscr C}_{\mathcal N}$ a fam\'\i lia de todas as fun\c c\~oes $f^{\omega}_{u_{out}}$, onde $\omega\in {\mathcal W}$. \'E a {\em classe de conceitos gerada} pela rede $\mathcal N$. Desse modo, a rede $\mathcal N$ torna-se em uma unidade de computa\c c\~ao. Em particular, pode-se falar da sua dimens\~ao de Vapnik--Chervonenkis.

Uma classe importante de redes de unidades de computa\c c\~ao consiste de {\em redes neurais artificiais} ({\em artificial neural networks,} {\em ANN}).

\subsection{Redes neurais sem realimenta\c c\~ao com valor limiar linear}
Estas redes ({\em feed-forward linear threshold networks}) t\^em uma arquitetura particular, dita {\em multicamada}. Tal rede \'e caraterizada pelas propriedades seguintes.

\begin{itemize}
\item Cada unidade computacional \'e um perceptron, e
\item os n\'os do grafo s\~ao agrupados em {\em camadas,} enumeradas com os n\'umeros naturais, 
\[\ell(x_i)=0,1,\ldots,l,\]
de maneira que
\begin{itemize}
\item os n\'os de entrada pertencem a camada zero, e 
\item se $x_i$ \'e conectado a $x_j$, ent\~ao $\ell(x_i)<\ell(x_j)$.
\end{itemize}
\index{rede! sem realimenta\c c\~ao! com valor limiar linear}
\end{itemize}

Ent\~ao, os n\'os dentro uma camada nunca s\~ao conectados, e a \'ultima ($\ell$-\'esima) camada tem somente um n\'o, o n\'o de sa\'\i da.

\begin{exemplo}[Perceptron como uma rede neural]
O perceptron visto no contexto das redes neurais \'e o exemplo mais simples e ao mesmo tempo mais antigo historicamente.

\begin{figure}[ht]
\begin{center}
\scalebox{0.5}[0.5]{\includegraphics{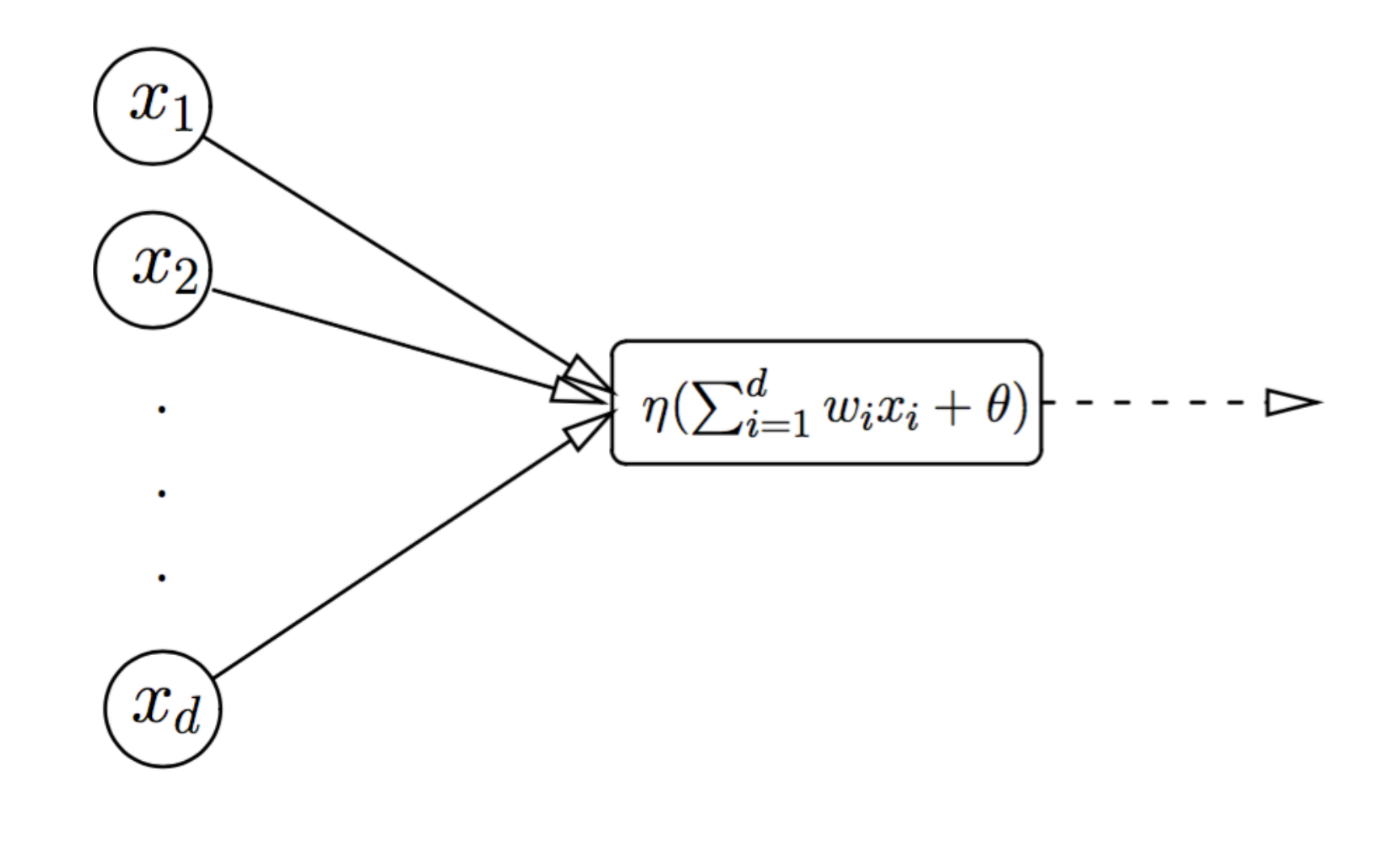}} 
\caption{Perceptron.}
\label{fig:perceptron}
\end{center}  
\end{figure}

Um {\em perceptron} 
\index{perceptron} 
\'e uma rede neural artificial com $d\geq 1$ n\'os de entrada e um n\'o de sa\'\i da, munido de uma fun\c c\~ao da forma
\[g=\eta\circ z,\]
onde 
\[z\colon\R^d\to \R\]
\'e uma fun\c c\~ao afim, 
\[z(x_1,x_2,\ldots,x_d)=\sum_{i=1}^d w_ix_i-\theta,\]
chamada de {\em fun\c c\~ao modeladora} ({\em shaping function}),
cujos coeficientes $w_i\in\R$, $i=1,2,\ldots,d$ s\~ao ditos {\em pesos},
\index{pesos! de perceptron}
e $\theta\in\R$ \'e o {\em valor limiar}.  
Enquanto
\[\eta\colon\R\to\{0,1\}\]
e a fun\c c\~ao de Heaviside:
\[\eta(x)=\begin{cases}
1,&\mbox{ if }x\geq 0,\\
0,&\mbox{ if }x<0.
\end{cases}\]
A fun\c c\~ao $\eta$ \'e dita {\em fun\c c\~ao de ativa\c c\~ao} ({\em activation function}).

A fun\c c\~ao de ativa\c c\~ao dispara um sinal $1$ se a intensidade do valor $\sum_{i=1}^d w_ix_i$ \'e maior do que o valor limiar $\theta$. 
\end{exemplo}

\subsection{Dimens\~ao VC de redes das unidades de computa\c c\~ao\label{ss:redes}}

O restante desta subse\c c\~ao \'e dedicada \`a prova do seguinte teorema.

\begin{teorema}
\label{th:vc}
Seja $\mathscr N$ uma rede sem realimenta\c c\~ao, tendo $k$ unidades computacionais bin\'arias, incluindo um n\'o de sa\'\i da. Seja $W$ a soma total de dimens\~oes VC de todas as unidades de $\mathcal N$. Ent\~ao, para todo $n$,
\begin{equation}
\label{eq:s}
s({\mathscr N},n)\leq \left(\frac{enk}{W}\right)^W,\end{equation}
e
\begin{equation}
\VC({\mathscr N})\leq 2W\log_2\left(\frac{2k}{\log 2}\right).
\label{eq:vcdim}
\end{equation}
\end{teorema}

Aqui, $\VC({\mathscr N})$ \'e a dimens\~ao VC da classe de conceitos ${\mathscr C}_{\mathcal N}$ gerada pela rede $\mathcal N$.

\begin{observacao}
Por exemplo, o teorema se aplica a cada rede neural sem re\-a\-li\-men\-ta\-\c c\~ao com valor limiar linear. Neste caso, $W$ \'e simplesmente o n\'umero total de par\^ametros ajust\'aveis da rede que s\~ao os pesos, $w_{ij}$, e os valores limiares, $\theta_i$, dos perceptrons.
\end{observacao}

A nossa prova segue, nas suas grandes linhas, \citep*{AB}.

\begin{lema}
Seja $f\colon X\to Y$ uma aplica\c c\~ao qualquer, e seja $\mathscr C$ uma classe de conceitos sobre $Y$, ${\mathscr C}\subseteq 2^Y$. Ent\~ao a dimens\~ao VC da fam\'\i lia de fun\c c\~oes bin\'arias do tipo
\[\chi_C\circ f,~~C\in {\mathscr C},\]
\'e menor ou igual de $\VC({\mathscr C})$.
\label{l:composicao}
\end{lema}

\begin{proof} Exerc\'\i cio.
\end{proof}

\begin{lema}
Para todos $\alpha,x>0$,
\[\log x\leq\alpha x-\log\alpha-1,\]
e a igualdade \'e atingida se e apenas se $\alpha x=1$.
\label{l:trivial}
\end{lema}

\begin{proof}
Levar $\log\alpha$ no lado esquerdo, exponenciar, e usar a expans\~ao de Taylor.
\end{proof}

\begin{definicao}
Um {\em espa\c co probabil\'\i stico finito} \'e um conjunto finito, $X=\{x_1,x_2,\linebreak \ldots,x_n\}$, onde cada ponto $x_i$ tem um peso associado, $p_i\geq 0$, tendo a propriedade
\[\sum_{i=1}^n p_i=1.\]
A medida de probabilidade, $\mu$, \'e definida pelos pesos como segue: se $A\subseteq X$,
\[\mu(A) = \sum_{x_i\in A}p_i.\]
\index{espa\c co! probabil\'\i stico! finito}
\end{definicao}

\'E f\'acil de verificar que $\mu$ \'e uma medida de probabilidade. Por exemplo, a medida de contagem normalizada sobre o cubo de Hamming \'e uma {\em medida uniforme} no sentido que os pesos correspondentes s\~ao todos iguais:
\[\forall \sigma\in\{0,1\}^n,~~p_{\sigma}=\frac{1}{2^n}.\]

\begin{definicao} 
Seja $X$ um espa\c co probabil\'\i stico finito, com pontos  $x_1,x_2,\ldots,x_n$ e probabilidades $p_i$. A {\em entropia} de $X$ \'e a quantidade
\[H(X)=\sum_{i=1}^n -p_i\log p_i.\]
\index{entropia}
\end{definicao}

\begin{lema}
O valor m\'aximo da entropia de um espa\c co probabil\'\i stico finito, $X$, como acima \'e igual $\log n$, atingido sobre a distribui\c c\~ao uniforme:
\[p_i=\frac 1n,~i=1,2,\ldots,n.\]
\label{l:entropy}
\end{lema}

\begin{proof}
O logaritmo \'e uma fun\c c\~ao c\^oncava no seu dom\'\i nio de defini\c c\~ao, ou seja, para quaisquer que sejam $\lambda_1,\lambda_2>0$ e $p_1,p_2\in [0,1]$, $p_1+p_2=1$, temos
\[\log(p_1\lambda_1+p_2\lambda_2)\geq p_1\log \lambda_1 +p_2\log \lambda_2.\]
Pela indu\c c\~ao finita, conclu\'\i mos que, para qualquer que seja a cole\c c\~ao $\lambda_i> 0$, $i=1,2,\ldots,n$,
\[\log\left(\sum_{i=1}^n p_i\lambda_i\right)\geq \sum_{i=1}^n p_i\log(\lambda_i).\]
No caso $\lambda_i=1/p_i$, deduzimos
\[\log n \geq \sum_{i=1}^n p_i\log\left(\frac 1{p_i}\right) = H(X),\]
e al\'em disso, se $p_i=1/n$, $i=1,2,\ldots,n$,
\[H(X)=\sum_{i=1}^n -p_i\log p_i=\sum_{i=1}^n \frac{\log n}n =\log n.\]
\end{proof}

\begin{observacao} A no\c c\~ao de entropia foi definida por Claude Shannon no contexto da sua teoria de informa\c c\~ao \citep*{shannon}. A entropia \'e uma medida de incerteza, no sentido seguinte. Suponha que $x_1,x_2,\ldots,x_n$ s\~ao as mensagens, e que $p_1,p_2,\ldots,p_n$ s\~ao as probabilidades de receber cada uma delas. Quando as probabilidades s\~ao todas iguais a $1/n$, n\~ao temos nenhuma informa\c c\~ao sobre qual mensagem \'e mais prov\'avel. Nesse caso, a incerteza \'e completa, e a entropia atinge o valor m\'aximo, $\log n$. Ao contr\'ario, suponha que, por exemplo, $p_1=1$ e $p_j=0$, $i\neq j$, ou seja, a mensagem $x_1$ vai chegar com probabilidade um. Nesse caso, a certeza \'e total, e a entropia atinge o seu m\'\i nimo, $0$. (A express\~ao $- 0\cdot\log 0$ surgindo na defini\c c\~ao da entropia nesse caso, est\'a tratada como o limite $\lim_{\e\to 0} - \e \log\e$, que existe).
\end{observacao}

Agora nos temos todas as ferramentas para mostrar o teorema \ref{th:vc}.
Escolhemos uma ordem total no conjunto de todas as unidades computacionais, \[u_1,u_2,\ldots,u_k,\]
de maneira que se existe uma conex\~ao de $u_i$ para $u_j$, ent\~ao $i<j$. (\'E sempre poss\'\i vel de estender uma ordem parcial a uma ordem total). Denotemos por $\mathcal W$ o espa\c co de estados da rede $\mathcal N$.
Temos
\begin{equation}
\label{eq:prob}
\sum_{i=1}^kd_i= W.\end{equation}
Dado um estado $\omega$ qualquer, denotemos por $f^\omega_i$ a fun\c c\~ao de sa\'\i da da unidade $i$, que \'e uma composi\c c\~ao recursiva de fun\c c\~oes $u^\omega_j$, $j\leq i$, numa ordem determinada pelas conex\~oes. Em particular, $f^{\omega}_k$ \'e a fun\c c\~ao bin\'aria de sa\'\i da da nossa rede.

Sejam $\Omega$ o dom\'\i nio (por exemplo, pode ser $\Omega=\R^m$, onde $m$ \'e o n\'umero de n\'os de entrada da rede, mas $m$ n\~ao desempenha papel nessa demonstra\c c\~ao), e $\sigma\subseteq\Omega^n$, uma amostra fixa. Vamos estimar superiormente o n\'umero de fun\c c\~oes bin\'arias distintas da forma $f^\omega_k\vert_{\sigma}$, $\omega\in {\mathcal W}$.

Para todo $i=1,2,\ldots,k$, digamos que dois estados, $\omega$ e $\omega^\prime$, s\~ao {\em indistingu\'\i veis at\'e a unidade $i$} se para todos $j=1,2,\ldots,i$ e todos valores de entrada $x\in\sigma$ temos
\[f^\omega_j(x)=f^{\omega^\prime}_j(x).\]
Denotemos essa rela\c c\~ao por $\omega\overset{i}{\sim}\omega^\prime$. Ser indistingu\'\i veis at\'e a unidade $i$ significa que um observador n\~ao pode distinguir o estado $\omega$ do estado $\omega^\prime$ somente observando os resultados de c\'alculo de valores de todas as fun\c c\~oes bin\'arias $f^{\omega}_j$ aparecendo at\'e e incluindo o n\'o $i$ sobre o conjunto $\sigma$. 

\'E f\'acil a ver que $\overset{i}{\sim}$ \'e uma rela\c c\~ao de equival\^encia sobre o espa\c co de estados $\mathcal W$ para todo $i=1,2,\ldots,k$, e por conseguinte ela define uma parti\c c\~ao finita $\Upsilon_i$ de $\mathcal W$ em  subconjuntos dois a dois disjuntos. Al\'em disso, se $i<j$, ent\~ao a parti\c c\~ao $\Upsilon_j$ refina a parti\c c\~ao $\Upsilon_i$:
\[\Upsilon_j\prec\Upsilon_i.\]

Se os valores de duas fun\c c\~oes de sa\'\i da, $f^\omega_k$ e $f^{\omega^\prime}_k$, diferem sobre um $x\in\sigma$, ent\~ao obviamente $\omega\not\overset{k}{\sim}\omega^\prime$. Por conseguinte, o n\'umero m\'aximo de fun\c c\~oes duas a duas distintas da forma $f^\omega_k\vert_{\sigma}$, $\omega\in{\mathcal W}$, \'e limitado de acima pelo tamanho da parti\c c\~ao final, $\Upsilon_k$. Vamos estimar $\sharp \Upsilon_i$ recursivamente em $i$, come\c cando com $i=1$. 

Dois estados $\omega$ e $\omega^\prime$ quaisquer podem ser distinguidos na unidade $u_1$ se e apenas se existe $x\in\sigma$ tal que $f^\omega_1(x)\neq f^{\omega^\prime}_1(x)$. De modo equivalente: $u^\omega_1(x)\neq u^{\omega^\prime}_1(x)$.
Neste caso particular, temos 
\[\sharp{\Upsilon_1} = \sharp{\{u_1^\omega\upharpoonright\sigma\colon\omega\in\R^W\}}\leq \left(\frac{en}{d_1}\right)^{d_1},\]
gra\c cas ao lema de Sauer-Shelah e a hip\'otese $\VC(u_1)=d_1$.  

Agora suponha que o tamanho da parti\c c\~ao $\Upsilon_{i-1}$ j\'a foi estimado. Seja $A\in \Upsilon_{i-1}$ uma classe de equival\^encia qualquer dessa parti\c c\~ao. Em quantos subconjuntos a rela\c c\~ao $\overset{i}{\sim}$ vai dividir $A$? Se $\omega,\omega^\prime\in A$, ent\~ao para todos $x\in\sigma$ e todos $j\leq i-1$ temos $f^\omega_j(x)=f^{\omega^\prime}_j(x)$. 
Em particular, as fun\c c\~oes $u^\omega_i$ e $u^{\omega^\prime}_i$ tem os mesmos valores de todos os seus argumentos. Em outras palavras, qual quer seja $\omega\in A$, a fun\c c\~ao $f^{\omega}_i$ pode ser escrita como $u^\omega_i\circ g$, onde a fun\c c\~ao $g$ n\~ao depende de $\omega$. Segundo o lema \ref{l:composicao}, a dimens\~ao VC do conjunto de fun\c c\~oes $f^{\omega}_i$, $\omega\in A$, \'e menor ou igual a $d_i$.
O lema de Sauer-Shelah implica
\[\sharp\{f^\omega_i\colon \omega\in A\}\leq \left(\frac{en}{d_i}\right)^{d_i}.\]
Conclu\'\i mos: o conjunto $A$ \'e dividido pela rela\c c\~ao $\overset{i}{\sim}$ em
\[\leq \left(\frac{en}{d_i}\right)^{d_i}\]
subconjuntos. Cada tal subconjunto corresponde a uma restri\c c\~ao diferente do tipo $f^\omega_i\vert_{\sigma}$, $\omega\in C$. 
Por conseguinte,
\[\sharp{\Upsilon_i}\leq \left(\frac{en}{d_i}\right)^{d_i}\sharp{\Upsilon_{i-1}},\]
ent\~ao
\[
\left\vert {\mathscr C}_{\mathscr N}\upharpoonright\sigma\right\vert\leq
\abs{\Upsilon_k}\leq\prod_{i=1}^k\left(\frac{en}{d_i}\right)^{d_i}.\]
Conclu\'\i mos:
\[\log s({\mathcal N},n) \leq\sum_{i=1}^k d_i\log \left(\frac{en}{d_i}\right).\]
A \'ultima express\~ao \'e reminiscente da entropia de um espa\c co probabil\'\i stico finito, e agora vamos manipul\'a-la. De fato, os valores $d_i/W$ s\~ao as probabilidades, em vista da Eq. (\ref{eq:prob}). Denotemos por $X$ um espa\c co finito com pesos de pontos iguais a $d_i/W$, $i=1,2,\ldots,k$.
Temos:
\begin{eqnarray*}
\log s({\mathscr N},n)  & \leq &
\sum_{i=1}^k d_i\log \left(\frac{en}{d_i}\right)\\ &=& W\sum_{i=1}^k \frac{d_i}{W}\left[\log\frac{W}{d_i} + \log(en) - \log W\right] \\
&=& W\cdot H(X) + W\log\frac{en}W \\
&\leq & W\log k + W\log\frac{en}W \\
&=& W\log\frac {enk}W.
\end{eqnarray*}
Exponenciando, obtemos a Eq. (\ref{eq:s}).

Temos $\VC({\mathscr N})< n$ se e apenas se 
\[s({\mathscr N},n) < 2^n,\]
\label{p:trick}
em particular, for o caso quando 
\[\left(\frac{enk}W\right)^W\leq 2^n,\]
ou seja, 
\begin{equation}
\label{eq:ass}
n\geq W\log_2\left(\frac{enk}W\right).
\end{equation}
Quando esta desigualdade, (\ref{eq:ass}), tem lugar? Aplicamos lema \ref{l:trivial} com
\[x=\frac{enk}W\mbox{ e }\alpha=\frac{\log 2}{2ek}\]
para deduzir
\[\log\left(\frac{enk}W\right)\leq \frac{n\log 2}{2W}-\log\left(\frac{\log 2}{2ek}\right)-1,\]
ou seja,
\[\log\left(\frac{enk}W\right)\leq \frac{n\log 2}{2W}+\log\left(\frac{2k}{\log 2}\right),\]
e como $\log x / \log 2 = \log_2x$,
\[W\log_2\left(\frac{enk}W\right)\leq \frac n2 +W\log_2\left(\frac{2k}{\log 2}\right). \]
Agora (\ref{eq:ass}) \'e v\'alido quando
\[W\log_2\left(\frac{2k}{\log 2}\right)\leq\frac n2,\]
ou seja,
\[n\geq 2W\log_2\left(\frac{2k}{\log 2}\right).\]
de onde obtemos (\ref{eq:vcdim}).

Ent\~ao, a dimens\~ao VC da rede satisfaz
\[d=O(W\log k),\]
onde $W$ \'e a soma de dimens\~oes VC de todas as $k$ unidades computacionais. Pode-se mostrar que a ordem de grandeza \'e \'otima.

\subsection{A rede sigmoide de Sontag} 
A fun\c c\~ao de Heaviside, $\eta$, \'e um exemplo particular de uma fun\c c\~ao {\em sigmoide}, $s\colon\R\to [0,1]$, isto \'e, uma fun\c c\~ao mon\'otona (n\~ao necessariamente cont\'\i nua) com $\lim_{t\to-\infty}s(t)=0$ e $\lim_{t\to+\infty}s(t)=1$. Por exemplo, a fun\c c\~ao
\[f(t) =\frac 1{\pi}\arctan t +\frac 12\]
\'e sigmoide, entre muitas outras.
Tais fun\c c\~oes s\~ao frequentemente usadas na teoria de redes neurais como fun\c c\~oes de ativa\c c\~ao. O exemplo de Sontag \citep*{sontag} mostra uma rede neural simples, com s\'o $4$ n\'os, cujas fun\c c\~oes de modelagem s\~ao lineares e as fun\c c\~oes de ativa\c c\~ao s\~ao sigmoide, e que possui um grande poder de fragmenta\c c\~ao: n\~ao somente a rede tem a dimens\~ao VC infinita, mais ela vai fragmentar ``quase todos'' subconjuntos finitos de $\R$. 

\subsubsection{Lema de Kronecker}
Denotemos por $\T$ o grupo de rota\c c\~oes do c\'\i rculo, munido da multiplica\c c\~ao usual de n\'umeros complexos:
\[\T=\{z\in\C\colon \abs z=1\}.\]
Este grupo pode ser tamb\'em identificado com o grupo aditivo de n\'umeros reais modulo $1$,
\[\T = \R/\Z,\]
com a ajuda da aplica\c c\~ao exponencial, $\R\ni t\mapsto \exp(2\pi\mathbf{i} t)\in\T$. 
O produto $\T^n$ de $n$ c\'opias de $\T$ \'e chamado o {\em toro} de posto $n$. Temos a identifica\c c\~ao
\[\T^n=\R^n/\Z^n.\]

A reta $\R$ pode ser vista como um espa\c co vetorial sobre o corpo $\Q$ dos n\'umeros racionais. Como tal, $\R$ tem dimens\~ao infinita (com efeito, a dimens\~ao de cont\'\i nuo, $\mathfrak c$). Uma $n$-upla dos reais, $x_1,x_2,\ldots,x_n$, \'e dita {\em racionalmente independente} se ela forma uma fam\'\i lia linearmente independente em $\R$ visto como um espa\c co vetorial sobre $\Q$. Ou seja, nenhuma combina\c c\~ao linear n\~ao trivial com os coeficientes racionais, $\sum_{i=1}^n q_ix_i$, se anula. Ao inv\'es de racionais, bastaria considerar os coeficientes inteiros.
\index{reais racionalmente independentes}

Por uma quest\~ao de simplifica\c c\~ao cometamos um abuso de linguagem e digamos que uma $n$-upla $x=(x_1,x_2,\ldots,x_n)\in\T^n$ dos elementos de $\T$ \'e racionalmente independente se qualquer $n$-upla de reais, $x^\prime_1,x^\prime_2,\ldots,x^\prime_n$, enviada sobre a nossa tupla pela fun\c c\~ao exponencial, $x_i=\exp(2\pi\mathbf{i} x^\prime_i)$, \'e racionalmente independente. (Neste caso, pode-se ver que {\em cada} tupla de reais enviada para a nossa tupla \'e racionalmente independente). 

Finalmente, recordemos que um subconjunto $A$ de um espa\c co m\'etrico \'e {\em denso} se a ader\^encia de $A$ em $X$ \'e igual a $X$. 

\begin{teorema}[Lema de Kronecker] 
Seja $x=(x_1,x_2,\ldots,x_n)\in\T^n$. As condi\c c\~oes seguintes s\~ao equivalentes:
\begin{enumerate}
\item A \'orbita de $x$ no grupo $\T^n$ (o conjunto de pot\^encias $x^k$, $k\in\Z$) \'e denso em $\T^n$.
\item A $n+1$-upla $x_1,x_2,\ldots,x_n,1$ \'e racionalmente independente.
\end{enumerate}
\index{lema! de Kronecker}
\end{teorema}

Uma prova elementar \'e apresentada no Ap\^endice \ref{a:kronecker}.
Outra prova, talvez mais transparente (embora n\~ao trivial), \'e baseada sobre a teoria de dualidade de grupos abelianos localmente compactos desenvolvida por Pontryagin e van Kampen (veja, por exemplo, Ch. 7 em \citep*{hofmann_morris}). Um {\em car\'ater} de um tal grupo, $G$, \'e um homomorfismo cont\'\i nuo $\chi\colon G\to \T$. Por exemplo, se $f\colon\R^n\to\R$ \'e um funcional linear tal que o valor de $f$ sobre cada elemento do reticulado $Z^n$ \'e inteiro, ent\~ao a f\'ormula $\chi(x\mod \Z)=\exp(2\pi{\mathbf{i}}f(x))$ define um car\'ater de $\T^n$, e pode-se mostrar, usando os resultados b\'asicos da topologia homot\'opica, que cada car\'ater de $\T^n$ \'e desta forma. O funcional linear $f$ \'e necessariamente da forma $f(x)= \langle x,v\rangle$ para um vetor $z\in\Z^n$.  

Um dos resultados de base da teoria de Pontryagin e van Kampen diz que se $H$ \'e um subgrupo fechado de um grupo abeliano localmente compacto $G$ e $x\in G\setminus H$, ent\~ao existe um car\'ater $\chi$ de $G$ tal que $\chi\vert_H\equiv e$, e $\chi(x)\neq e$. A condi\c c\~ao (1) equivale \`a propriedade de que o menor subgrupo fechado que cont\'em $x$ \'e igual \`a $\T^n$, ou seja, de que cada car\'ater $\chi$ tal que $\chi(x)=e$, \'e trivial. De maneira equivalente, se $z\in\Z^n$ e $\langle z,x\rangle=0$, ent\~ao $z=0$. Mas isso \'e a condi\c c\~ao (2).

\subsubsection{A rede auxiliar} Como um passo intermedi\'ario, consideremos uma rede $\mathcal N$ com uma s\'o unidade de computa\c c\~ao, cuja fun\c c\~ao de modelagem \'e da forma $f^w(x)=\cos(wx)$, e a fun\c c\~ao de ativa\c c\~ao \'e a fun\c c\~ao de Heaviside usual, $\eta$. A rede $\mathcal N$ \'e bin\'aria, com valores de entrada reais e um par\^ametro, $w$, que toma os valores reais (de fato, inteiros). 

\begin{lema}
A rede $\mathcal N$ fragmenta cada $n$-upla de reais $(x_1,\ldots,x_n)$ tal que \[(x_1,\ldots,x_n,1)\]
 s\~ao racionalmente independentes.
\end{lema}

\begin{proof}
Denotemos $y_i=\exp(2\pi {\mathbf{i}}x_i)\in\T$. Seja $I\subseteq [n]$. O lema de Kronecker implica que existe um valor do par\^ametro $w\in\Z$ tal que para todo $i\in I$ a parte real de $f^w(y_i)$ \'e positiva, e para todo $i\in [n]\setminus I$ a parte real de $f^w(y_i)$ \'e estritamente negativa. Conclu\'\i mos: o valor de sa\'\i da da nossa rede \'e igual a $1$ para $x_i,i\in I$, e \'e igual a $0$, para $x_i,i\notin I$. 
\end{proof}

\'E claro que essa rede n\~ao \'e sigmoide, mas isto \'e f\'acil a contornar.

\subsubsection{Um jeitinho} A ideia de Sontag foi de esconder a fun\c c\~ao $\cos x$ dentro uma fun\c c\~ao sigmoide, adicionando um fator de cosseno, $a\cos x$, com o coeficiente $a$ vari\'avel t\~ao pequeno que a adi\c c\~ao visivelmente tem efeito nenhum sobre a forma de uma fun\c c\~ao padr\~ao, como $\frac 1{\pi}\arctan x+\frac 12$. Depois, a gente pode se livrar de $\arctan $ facilmente. 

\begin{figure}
{\includegraphics[width=1.5in]{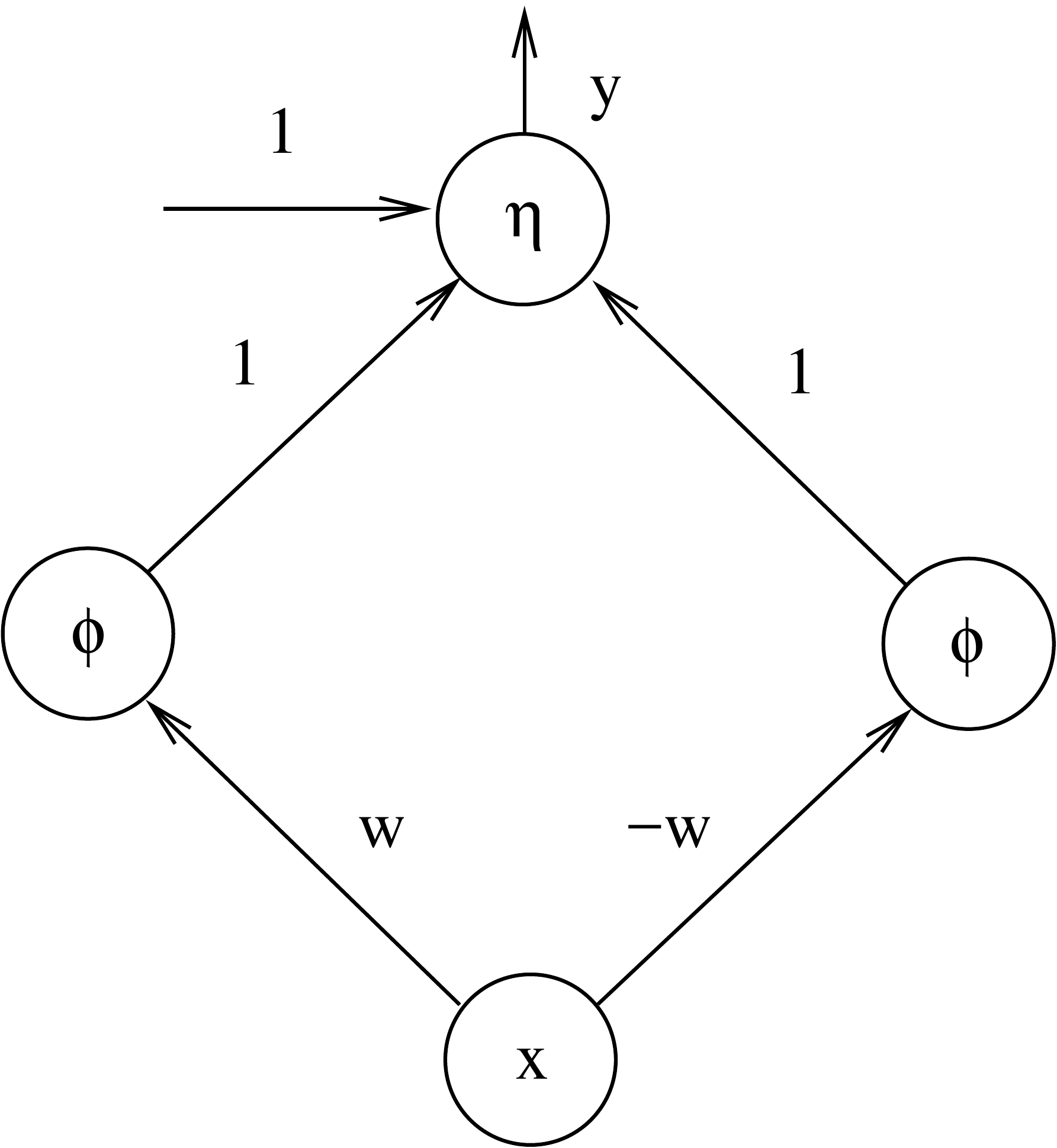}}
\label{fig_sontag}
\caption{Arquitetura da rede de Sontag}
\end{figure}

\begin{figure}
{\includegraphics[width=2in]{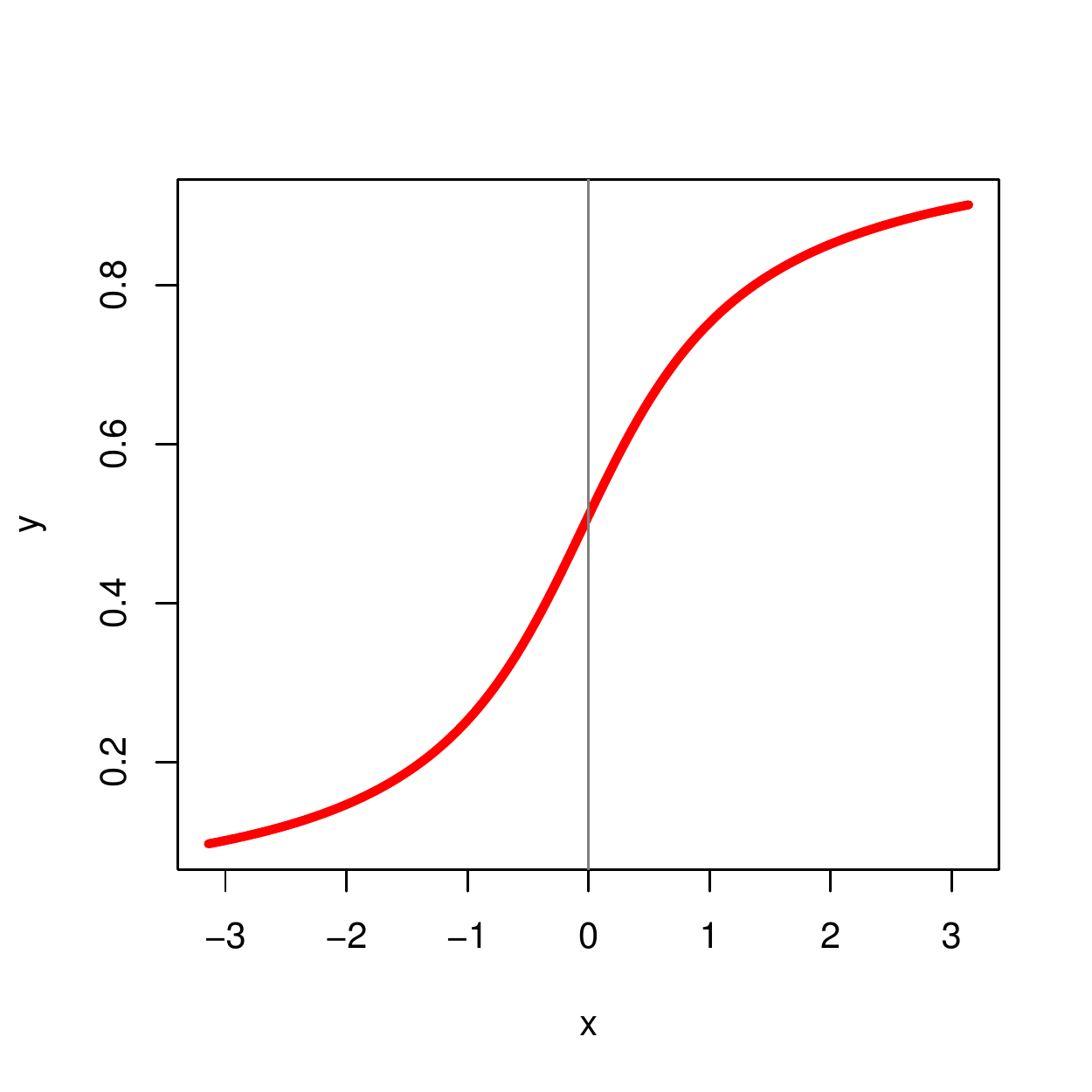}}
\label{fig:sigmoid}
\caption{O sigmoide de ativa\c c\~ao $\phi$ com $\alpha=100$}
\end{figure}

A fun\c c\~ao sigmoide de ativa\c c\~ao \'e da forma
\[\phi(x) =\frac 1\pi\arctan x+\frac{\cos x}{\alpha(1+x^2)}+\frac 12,\]
onde $\alpha\gg 2\pi$ \'e fixo, por exemplo, $\alpha=100$. O perceptron da camada de sa\'\i da tem dois pesos de entrada iguais a um, e o valor limiar, um. 
A rede gera a fun\c c\~ao bin\'aria 
\[y = \eta[\rho(x)],\]
onde
\[\rho(x) = \frac{2\cos wx}{\alpha(1+w^2x^2)}.\]
Os valores de entrada s\~ao reais. 

Como o sinal da fun\c c\~ao $\rho(x)$ \'e igual ao sinal de $\cos wx$, as duas redes geram a mesma fun\c c\~ao bin\'aria e tem as mesmas propriedades de fragmenta\c c\~ao.
\index{rede! de Sontag}

\begin{figure}[!t]
\begin{center}
\scalebox{0.35}[0.35]{\includegraphics{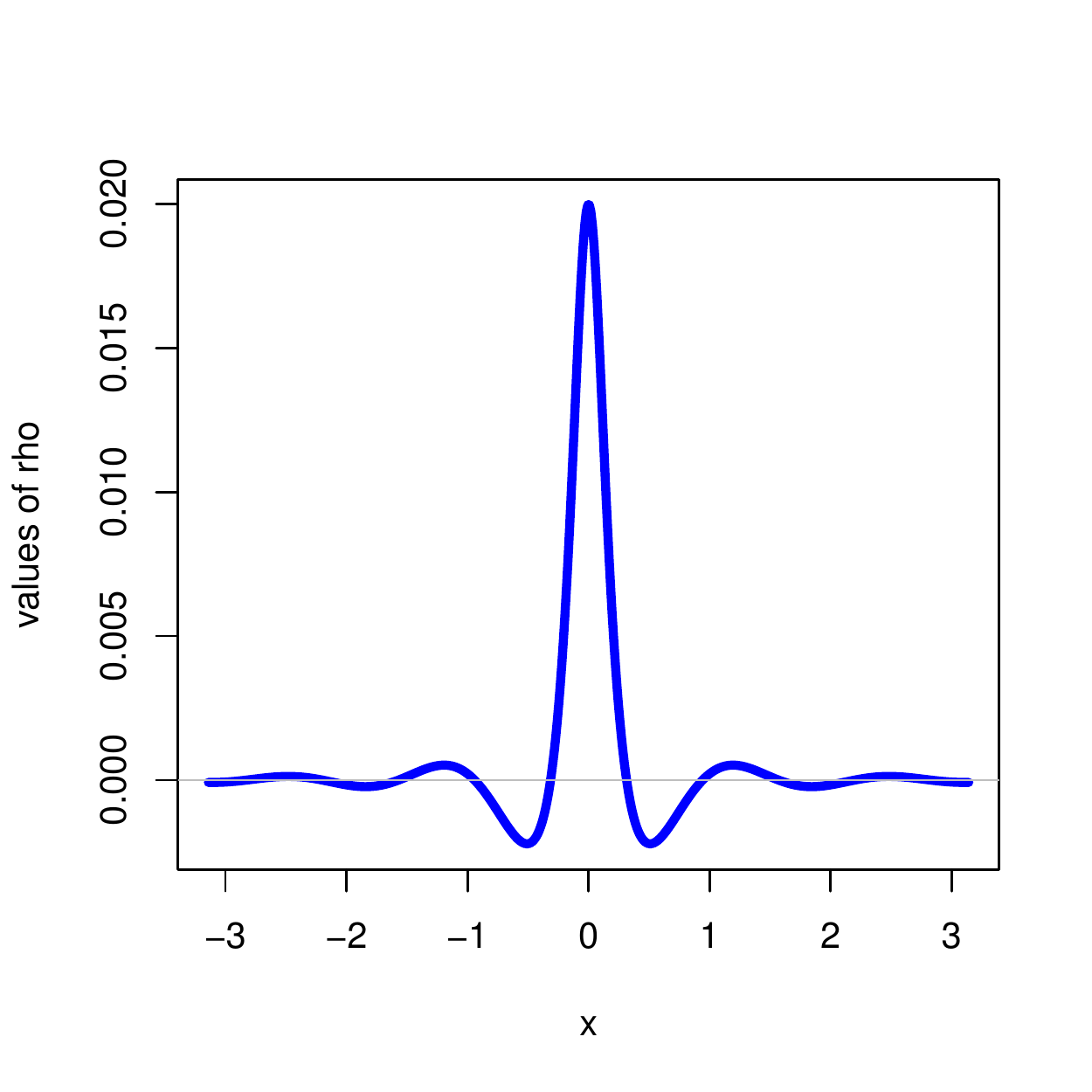}}
\scalebox{0.35}[0.35]{\includegraphics{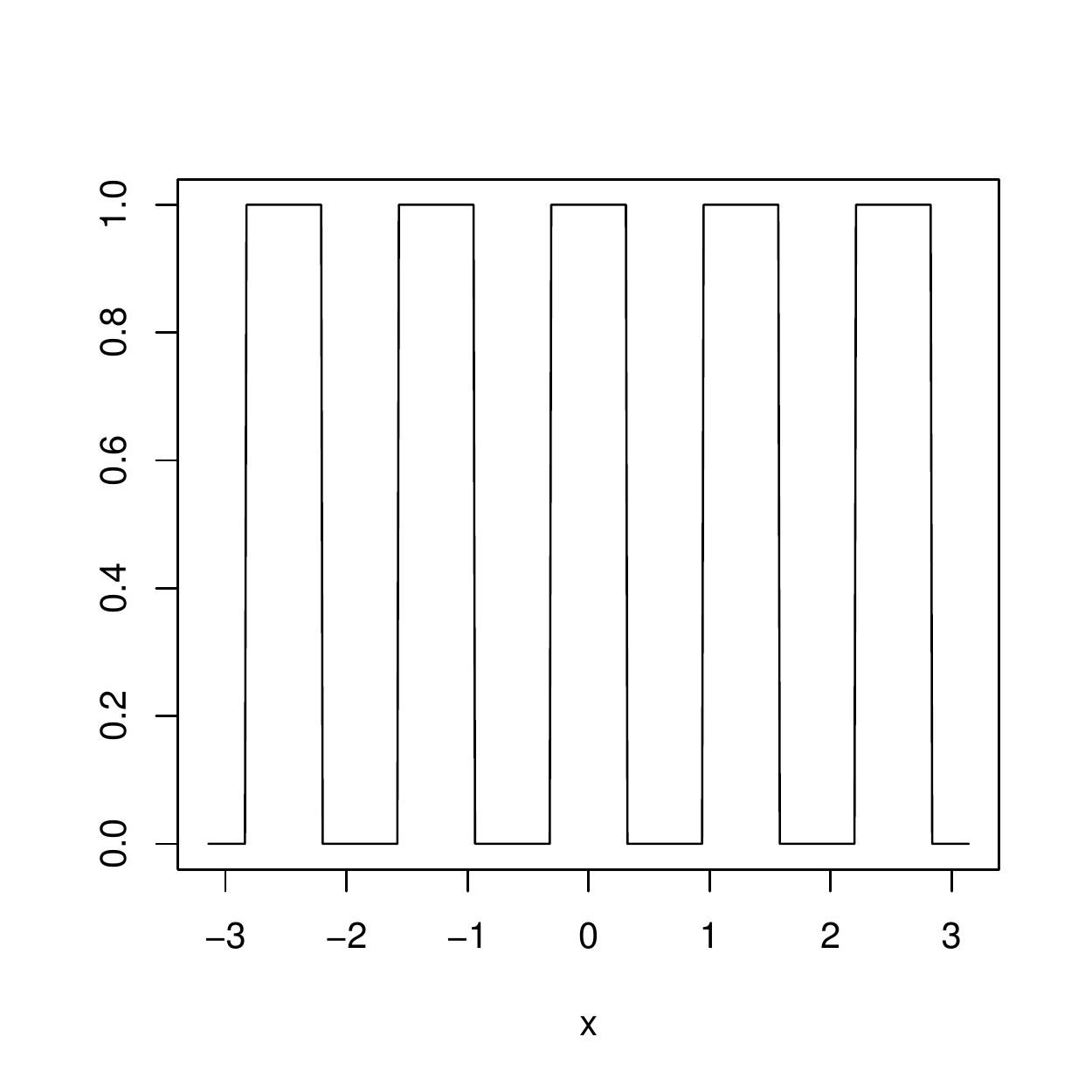}}
\end{center}
\caption{A fun\c c\~ao $\rho$ com $\alpha=100$ e $w=5$ (a esquerda) e a fun\c c\~ao bin\'aria de sa\'\i da correspondente (a dir.)}
\label{fig:rho}
\end{figure}

Por conseguinte, a rede de Sontag tem dimens\~ao VC infinita. De fato, as suas propriedades de fragmenta\c c\~ao s\~ao muito mais fortes. Como a dimens\~ao do $\R$ como um espa\c co vetorial racional \'e a igual \`a cardinalidade de cont\'\i nuo, $\dim_{\Q}\R={\mathfrak c}$, existe um conjunto $X\subseteq\R$ racionalmente independente da mesma cardinalidade que $\R$. A rede de Sontag fragmenta todo subconjunto finito de $X$. 

(Mesmo melhor, pode-se provar que para todo $n$, o conjunto de $n$-uplas que s\~ao racionalmente dependentes tem a medida de Lebesgue nula, ent\~ao a rede de Sontag fragmenta quase todos conjuntos finitos de reais....)

\section{Dimens\~ao sens\'\i vel \`a escala (fat-shattering dimension)}

Seja $\mathcal F$ uma {\em classe de fun\c c\~oes,} ou seja, uma fam\'\i lia de fun\c c\~oes sobre $\Omega$ tomando valores reais (tipicamente, num intervalo da reta, por exemplo, $[0,1]$). Qual \'e a vers\~ao correta da dimens\~ao VC para as fun\c c\~oes que n\~ao s\~ao necessariamente bin\'arias? Tem v\'arias, por exemplo, a seguinte.

\begin{definicao}
Seja $\mathscr F$ uma classe de fun\c c\~oes reais sobre um conjunto $\Omega$. Um subconjunto $\sigma\subseteq\Omega$ \'e {\em pseudofragmentado} ({\em pseudoshattered})
\index{pseudofragmenta\c c\~ao}
por $\mathscr F$ se existe uma fun\c c\~ao $g\colon\sigma\to\R$ tal que, qualquer seja $I\subseteq\sigma$, existe uma fun\c c\~ao $f_I\in {\mathscr F}$ satisfazendo as condi\c c\~oes
\begin{eqnarray}
\forall x\in I,~~f_I(x)\geq g(x), \nonumber \\
\forall x\notin I,~~f_I(x)< g(x).
\end{eqnarray}
A {\em pseudodimens\~ao,}
\index{pseudodimens\~ao}
ou {\em P-dimens\~ao,} 
\index{P-dimens\~ao}
de $\mathscr F$, denotada $\mbox{P-dim}({\mathscr F})$, \'e o supremo de cardinalidades de subconjuntos finitos $\sigma$ de $\Omega$ pseudofragmentados por $\mathscr F$.
\end{definicao}

\begin{exercicio} 
Mostrar que a pseudodimens\~ao de $\mathcal F$ \'e igual a dimens\~ao VC da fam\'\i lia de todos os {\em subgr\'aficos} de fun\c c\~oes  $f\in{\mathscr F}$, ou seja, os conjuntos
\[S_f=\{(x,y)\in\Omega\times \R\colon f(x)\leq y\}.\]
\index{subgr\'afico}
\end{exercicio}

\begin{exercicio} 
Mostrar que se $\mathcal F$ consiste de fun\c c\~oes bin\'arias, ent\~ao $\mbox{P-dim}({\mathscr F})=\VC({\mathscr F})$.
\end{exercicio}

\begin{exercicio} Seja $\Omega=\R^d$, e seja $\mathscr F$ a classe de todas as fun\c c\~oes dist\^ancia de pontos $x_0\in\R^n$,
\[{\mathscr F}=\left\{d_{x_0}\colon x_0\in\R^d\right\},\]
onde
\[d_{x_0}(x) =\norm{x-x_0}_2. \]
Mostrar que
\[\mbox{P-dim}({\mathscr F})=d+1.\]
 
\end{exercicio}

De fato, a no\c c\~ao de $P$-dimens\~ao n\~ao \'e completamente satisfat\'oria. A no\c c\~ao seguinte \'e mais fina. A {\em fat-shattering dimension,} $\mathrm{fat}_\e$, na verdade n\~ao toma um valor \'unico, mas melhor uma cole\c c\~ao de valores \`as escalas $\e>0$ diferentes.

\begin{figure}[ht]
\begin{center}
\scalebox{0.35}[0.35]{\includegraphics{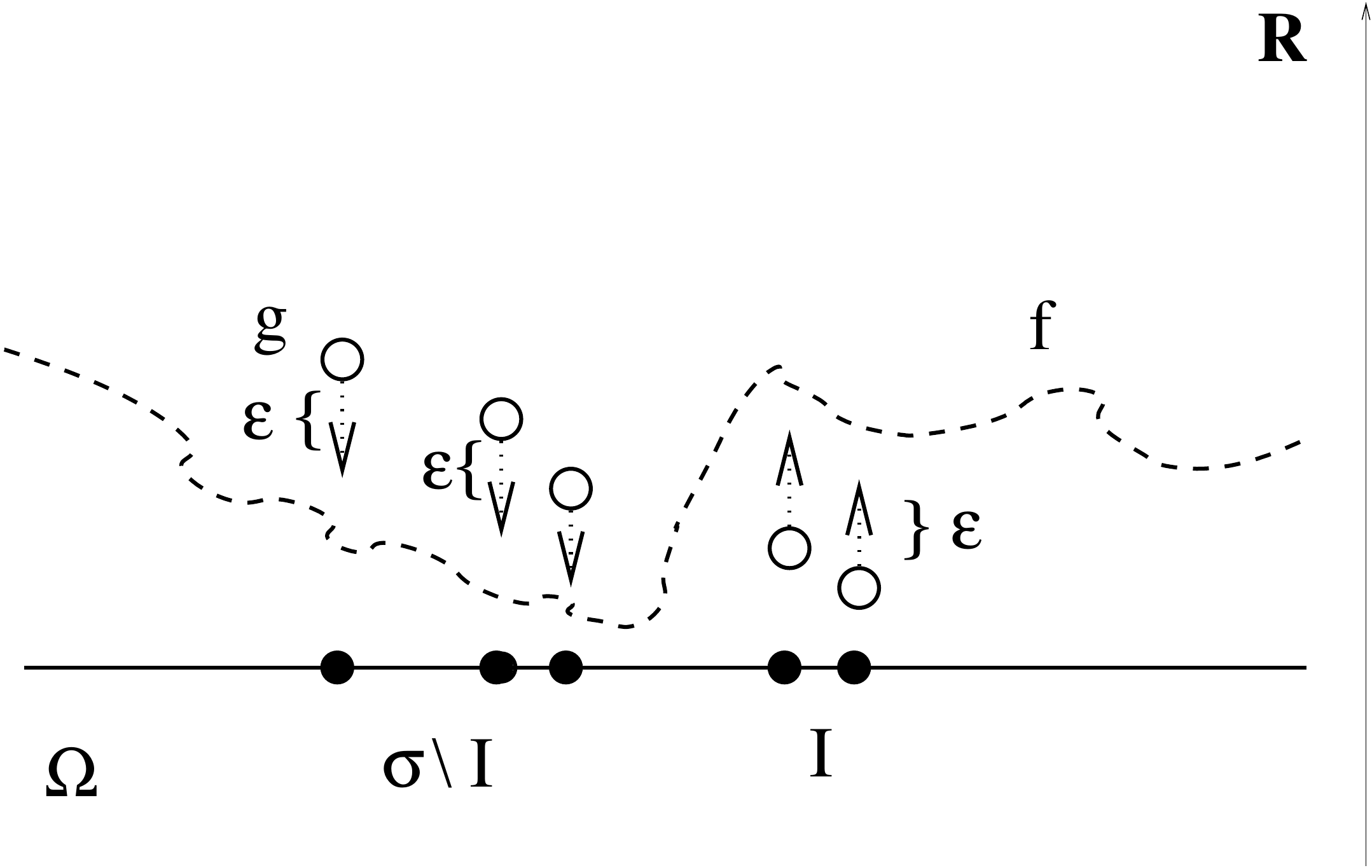}} 
\caption{$\e$-fragmenta\c c\~ao.}
\label{fig:fatshat}
\end{center}
\end{figure}

\begin{definicao}
Seja $\mathscr F$ uma classe de fun\c c\~oes reais sobre um conjunto $\Omega$. Dado um $\e>0$, digamos que o subconjunto $\sigma\subseteq\Omega$ \'e {\em $\e$-fragmentado} por $\mathscr F$ se existe uma fun\c c\~ao $g\colon\sigma\to\R$ ({\em testemunha de fragmenta\c c\~ao})
\index{testemunha! de fragmenta\c c\~ao}
tal que, qualquer que seja $I\subseteq\sigma$, existe uma fun\c c\~ao $f_I\in {\mathscr F}$ satisfazendo as condi\c c\~oes
\begin{eqnarray}
\forall x\in I,~~f_I(x)> g(x)+\e, \nonumber \\
\forall x\notin I,~~f_I(x)< g(x)-\e.
\end{eqnarray}
A {\em dimens\~ao $\e$-fragmentando} de $\mathscr F$ ({\em $\e$-fat shattering dimension}, em ingl\^es), 
\index{fat shattering dimension}
denotada $\mbox{fat}_{\e}({\mathscr F})$, \'e o supremo de cardinalidades de todos os subconjuntos finitos de $\Omega$ que s\~ao $\e$-fragmentados por 
\index{fatepsilonf@$\mbox{fat}_{\e}({\mathscr F})$}
$\mathscr F$.
\end{definicao}

Mais uma vez, n\~ao \'e dif\'\i cil de verificar que se $\mathscr F$ consiste de fun\c c\~oes bin\'arias e $\e>0$ \'e bastante pequeno (por exemplo, $\e\leq \frac 12$), ent\~ao $\mbox{fat}_{\e}({\mathscr F})$ \'e a dimens\~ao VC de $\mathscr F$. 

\begin{exercicio}
Para todo $\e>0$,
\[\mbox{fat}_{\e}({\mathscr F})\leq \mbox{P-dim}({\mathscr F}).\]
Se $0<\e<\gamma$, ent\~ao
\[\mbox{fat}_{\e}({\mathscr F})\leq \mbox{fat}_{\gamma}({\mathscr F}).\]
\end{exercicio}

\begin{exercicio}
\[\mbox{P-dim}({\mathscr F})=\sup_{\e>0}\mbox{fat}_{\e}({\mathscr F}).\]
\end{exercicio}

N\~ao teremos muita chance de trabalhar com essa no\c c\~ao em detalhes.
Para mais sobre a dimens\~ao {\em fat-shattering} pedimos-lhe que consulte \citep*{ABDCBH}.

%
%

\chapter{Aprendizagem PAC\label{ch:PAC}}

Neste cap\'\i tulo introduziremos algumas no\c c\~oes b\'asicas tais que uma regra de aprendizagem, a aprendizagem PAC (Provavelmente Aproximadamente Correta), e uma classe aprendiz\'avel, ilustrando-as no quadro de aprendizagem dentro uma classe sob a distribui\c c\~ao fixa. 

O cen\'ario geral da aprendizagem supervisionada \'e o seguinte. H\'a um dom\'\i nio $\Omega$, e um conceito desconhecido, $C\subseteq\Omega$ (ou: um classificador bin\'ario desconhecido, $\chi_C$). Este $C$ \'e a ser aprendido numa sequ\^encia de experi\^encias. Sobre cada amostra finita $x_1,x_2,\ldots,x_n$ de pontos de $\Omega$, o conceito $C$ induz uma rotulagem, $\e_1=\chi_C(x_i)$, $\e_2=\chi_C(x_2)$, $\ldots$, $\e_n=\chi_C(x_n)$. Somos mostrados a amostra rotulada \[(x_1,x_2,\ldots,x_n,\e_1, \e_2, \ldots,\e_n)\]
 e, com base nesta amostra, temos que adivinhar o conceito $C$, sugerindo uma {\em hip\'otese}, $H\subseteq\Omega$.

A bondade da nossa hip\'otese $H$ \'e avaliada atrav\'es do {\em erro de generaliza\c c\~ao}, ou seja, a probabilidade do evento que o r\'otulo de um ponto qualquer, $x$, do dom\'\i nio seja adivinhado incorretamente:
\[\chi_H(x)\neq\chi_C(x).\]
Isso torna inevit\'avel um contexto probabil\'\i stico. Para dar o significado a uma express\~ao do tipo
\[P[\chi_H(x)\neq\chi_C(x)],\]
o valor aleat\'orio, $x$, deve ser modelado por uma {\em vari\'avel aleat\'oria} com valores em $\Omega$, de costume denotada por uma letra mai\'uscula, $X$. O significado disso \'e que n\~ao somente $X_i$ \'e uma vari\'avel (elemento desconhecido) do dom\'\i nio, mas, al\'em disso, dado um subconjunto $A$ do dom\'\i nio, sabe-se a probabilidade do valor de $X$ pertencer a $A$:
\[P[X\in A]=\mu(A),\]
onde $\mu$ \'e um medida de probabilidade sobre o dom\'\i nio, chamada a {\em lei} de $X$.
Assim, uma vari\'avel aleat\'oria \'e uma vers\~ao ``melhorada'' de uma vari\'avel usual.

Vamos introduzir este contexto come\c cando com duas situa\c c\~oes particularmente transparentes, de fato quase triviais, onde tudo corre bem e pode ser perfeitamente entendido: aprendizagem sob uma medida at\^omica, e aprendizagem de uma classe de conceitos finita.

\section{Aprendizagem sob a distribui\c c\~ao puramente at\^omica}
\label{s:atomica}
\subsection{Aprendizagem num dom\'\i nio finito}

Para come\c car, suponha que $\Omega=(\Omega,\mu)$ seja um espa\c co probabil\'\i stico finito:
\[\Omega=\{a_1,a_2,\ldots,a_k\},\]
onde a medida de probabilidade $\mu$ \'e determinada pelos pesos, $p_1,p_2,\ldots,p_k$, de todos os pontos, de modo que, qualquer que seja $A\subseteq\Omega$,
\[\mu(A)=\sum_{a_i\in A}p_i.\]
Particularmente,
\[\mu\{a_i\}=p_i.\]
Desse modo, a probabilidade de que a vari\'avel aleat\'oria $X_i$ toma valor $a_j$, \'e dada por
\[P[X_i=a_j]=\mu\{a_j\}=p_j.\]

Segundo o paradigma de aprendizagem,
os dados (amostras n\~ao-ro\-tu\-la\-das) s\~ao modelados por uma sequ\^encia infinita $X_1,X_2,\ldots,X_n,\ldots$ de elementos ale\-a\-t\'o\-ri\-os de $\Omega$, independentes e identicamente distribu\'\i dos, com a lei comum $\mu$. Ent\~ao, dado um conjunto $A\subseteq\Omega$, temos para todo $i$,
\[P[X_i\in A]=\mu(A),\]
enquanto a independ\^encia significa que, dado pontos quaisquer $a_{i_1},a_{i_2},\ldots,a_{i_m}$, temos
\begin{align*}
P[X_1=a_{i_1},X_2=a_{i_2},\ldots,X_m =
a_{i_m}]
&=
P[X_1=a_{i_1}] P[X_2=a_{i_2}] \ldots P[X_m=a_{i_m}]
\\
&= p_{i_q}p_{i_2}\ldots p_{i_m}.
\end{align*}
De modo equivalente (exerc\'\i cio), mesmo se formalmente mais geral, quaisquer que sejam conjuntos $A_{i_1},A_{i_2},\ldots,A_{i_m}\subseteq\Omega$, 
\begin{align*}
P[X_1\in A_{i_1},~X_2\in A_{i_2},\ldots,&X_m\in A_{i_m}] =
\\
&
P[X_1\in A_{i_1}]\cdot P[X_2\in A_{i_2}]\cdot\ldots\cdot P[X_m\in A_{i_m}]. 
\end{align*}
(Para uma discuss\~ao da no\c c\~ao de uma vari\'avel aleat\'oria num contexto mais geral, consulte ap\^endice \ref{a:variaveis}, seguido pelos Ap\^endices \ref{apendice:padrao}, \ref{a:medidas}, e \ref{a:integral}.)

Seja $C\subseteq\Omega$ um conceito desconhecido. Dada uma amostra aleat\'oria $(X_1, X_2, \ldots, X_n)$ (ou seja, uma cole\c c\~ao finita de vari\'aveis i.i.d. seguindo a lei $\mu$), o conceito $C$ induz sobre a amostra uma rotulagem, isto \'e, a sequ\^encia de r\'otulos $Y_1,Y_2,\ldots,Y_n$:
\[Y_i = \chi_C(X_i),~i=1,2,\ldots,n.\]
Cada r\'otulo \'e uma composi\c c\~ao de uma vari\'avel aleat\'oria, $X_i$, com uma fun\c c\~ao determin\'\i stica, $\chi_C$, e desse modo, $Y_i$ \'e uma vari\'avel aleat\'oria tamb\'em. A lei de $Y_i$ \'e a imagem direita, $\chi_{C\ast}(\mu)$, da lei $\mu$ de $X_i$ sob a aplica\c c\~ao $\chi_C$:
\begin{align*}
P[Y_i=1] &= P[X_i\in C] \\
&= \mu(C) \\
&=
\chi_{C\ast}(\mu)(\{1\}), \\
P[Y_i=0] 
&= 1 - \mu(C).
\end{align*}

\begin{exercicio}
Verificar que as vari\'aveis $Y_1,Y_2,\ldots,Y_n$ s\~ao i.i.d.
\end{exercicio}

A observa\c c\~ao de base que permite advinhar $C$ \'e a seguinte: se o tamanho $n$ de amostra \'e bastante grande, ent\~ao, com alta probabilidade, cada ponto de nosso dom\'\i nio finito vai aparecer pelo menos uma vez entre os pontos da amostra, assim determinando $C$ completamente.

\begin{fato} 
Seja $a\in\Omega$, com $P[X=a]=\eta$. Ent\~ao, a probabilidade que nenhuma de v.a. i.i.d. $X_1,X_2,\ldots,X_n$ toma o valor $a$ \'e igual a $(1-\eta)^n$.
\label{f:1}
\end{fato}

\begin{proof}
Para cada $i=1,2,\ldots,n$, 
\[P[X_i\neq a] = \mu(\Omega\setminus\{a\})= 1-\eta,\]
e como as vari\'aveis $X_i$ s\~ao independentes,
\[P[X_1\neq a, X_2\neq a,\ldots, X_n\neq a] = (1-\eta)^n.\]
\end{proof}

\begin{fato} Seja
\[\eta=\min_{i=1}^k p_i.\]
A probabilidade que existe $a\in\Omega$ tal que $X_i\neq a$ para todos $i$ \'e limitada acima pelo valor $k(1-\eta)^n$.
\label{f:2}
\end{fato}

\begin{proof}
O evento ``existe $a\in\Omega$ tal que $X_i\neq a$ para todos $i$'' corresponde, ao n\'\i vel conjunt\'\i stico, ao subconjunto de todas as sequ\^encias $(x_1,x_2,\ldots,x_n)\in \Omega^n$ quem n\~ao cont\'em um dos elementos de $\Omega$. Isso pode se escrever assim:
\[\bigcup_{j=1}^k \{x\in \Omega^n\colon \forall i,~x_i\neq a_j\}.\]
A probabilidade neste contexto corresponde \`a medida de produto, $\mu^{\otimes n}$, determinada por
\[\mu^{\otimes n}\{x_1,x_2,\ldots,x_k\} = \mu\{x_1\}\mu\{x_2\}\ldots\mu\{x_k\}.\]
A medida de uni\~ao de $k$ conjuntos n\~ao excede $k$ vezes a medida do maior entre eles, como estimado no fato \ref{f:1}.
\end{proof}

Agora denotemos 
\begin{align*}
\supp\mu &= \{a\in\Omega\colon \mu\{a\}>0\}\\
&=\{a_j\colon p_j>0,~j=1,2,\ldots,k\}
\end{align*}
o {\em suporte} da medida $\mu$. O suporte, neste caso, consiste de todos os \'atomos da medida, ou seja, pontos $a$ satisfazendo $\mu\{a\}>0$. \'E claro que a medida do suporte \'e igual a $1$. 

\begin{fato}
Seja $0<\delta<1$ qualquer (o {\em risco}). Denotemos
\[\eta=\min\{p_i\colon p_i>0\}\]
o valor do menor \'atomo.
Se
\begin{equation}
 n\geq \frac 1\eta\ln\frac{k}{\delta},
\label{eq:riscon}\end{equation}
ent\~ao, com a probabilidade (a {\em confian\c ca}) $\geq 1-\delta$, cada ponto do suporte de $\mu$ vai aparecer pelo menos uma vez entre os valores tomados pelas vari\'aveis aleat\'orias i.i.d. $X_1,X_2,\ldots,X_n\sim \mu$.
\label{f:3}
\index{risco}
\index{confian\c ca}
\end{fato}

\begin{proof}
Segundo fato \ref{f:2}, bastaria que o valor de $n$ satisfa\c ca
\[ k(1-\eta)^n\leq\delta.\]
Tomando o logaritmo, obtemos
\[\log k + n\log (1-\eta)\leq\log\delta,\]
ou seja (como $\log(1-\eta)<0$),
\[n\geq \log\frac{\delta}k\cdot \frac 1{\log{(1-\eta)}}=\log\frac k{\delta}\cdot\frac{ 1}{-\log(1-\eta)}.\]
De acordo com lema \ref{l:trivial}, substituindo os valores $\alpha=1$ e $x=1-\eta$, temos
\[\log(1-\eta)\leq -\eta,\]
de onde
\[\frac{ 1}{-\log(1-\eta)}\leq \frac 1\eta,\]
concluindo o argumento.
\end{proof}

Conclu\'\i mos: dado um valor $\delta>0$ de risco qualquer, desde que $n$ seja bastante grande (como dado pela Eq. \ref{eq:riscon}), com confian\c ca $\geq 1-\delta$, temos o conhecimento completo de r\'otulos de todos os pontos de $\Omega$. A hip\'otese $H$ pode consistir de todos os pontos do dom\'\i nio que aparecem na amostra $x_1,x_2,\ldots,x_n$ e t\^em o r\'otulo $1$:
\begin{align*}
H &= \{a\in\Omega\colon \exists i, x_i=a\mbox{ e }\e_i=1\} \\
&= \{x_i\colon \e_i=1,~i=1,2,\ldots,n\}.
\end{align*}
Note que nesta f\'ormula ({\em regra de aprendizagem}) as vari\'aveis s\~ao notadas pelas letras min\'usculos, pois elas s\~ao elementos do dom\'\i nio, n\~ao importa a lei de distribui\c c\~ao. A regra de distribui\c c\~ao n\~ao pode depender de uma lei subjacente (que \'e sempre desconhecida).

Com a confian\c ca $\geq 1-\delta$, temos 
\[H\cap \supp\mu = C\cap\supp\mu.\]
Por conseguinte, a diferen\c ca sim\'etrica $H\Delta C$ \'e um subconjunto de $\Omega\setminus\supp\mu$, que tem medida nula. Nesse caso, 
o erro de generaliza\c c\~ao da nossa hip\'otese \'e zero:
\[P[\chi_C(X)\neq \chi_H(X)] = \mu (H\Delta C) =0.\]
A estimativa n\~ao depende de $C$. Podemos concluir:

\begin{proposicao}
Seja $\Omega$ um espa\c co probabil\'\i stico finito, e seja $\eta>0$ a medida do menor \'atomo de $\Omega$. Dado um risco $\delta>0$, se 
\[n\geq \frac 1\eta\ln\frac{k}{\delta},\]
ent\~ao, com confian\c ca $\geq 1-\delta$, podemos advinhar qualquer conceito $C$ exatamente.
\end{proposicao}

Nesse caso, trata-se da aprendizagem {\em provavelmente (exatamente) correta}. A palavra ``provavelmente'' se aplica \`a alta confian\c ca para adivinharmos o conceito $C$ corretamente.

Agora passemos a um caso um pouco mais interessante.

\subsection{Aprendizagem num dom\'\i nio enumer\'avel\label{ss:enumeravel}}

Nesta subse\c c\~ao um {\em dom\'\i nio,} $\Omega$, pode significar um conjunto qualquer.
Seja $\mu$ uma medida de probabilidade discreta, ou puramente at\^omica, sobre $\Omega$. Isso significa que existe uma sequ\^encia (finita ou infinita) de pontos dois-a-dois distintos $a_j\in\Omega$, $j=1,2,3,\ldots$ ({\em \'atomos} de medida $\mu$), com $\mu\{a_j\}=p_j>0$, tal que
\[\sum_{j=1}^\infty p_j=1.\]
Dado $A\subseteq\Omega$, definamos o valor da medida de $A$ por
\[\mu(A)=\sum\{p_j\colon a_j\in A\}=\sum_{j=1}^\infty \chi_A(a_j)p_j=\int_\Omega\chi_Ad\mu.\]
Segue-se, em particular, que
\[\mu\left(\Omega\setminus\{a_1,a_2,\ldots\}\right)=0.\]
Nesse caso, o suporte de $\mu$ \'e tamb\'em o conjunto de todos os \'atomos. Essencialmente, podemos ignorar o resto de $\Omega$ e assumir que $\Omega$ seja enumer\'avel, porque a probabilidade para um ponto aleat\'orio a pertencer ao resto \'e zero:
\[P[X\notin \{a_1,a_2,\ldots\}]=0.\]

Como antes, os dados s\~ao modelados por uma sequ\^encia infinita $X_1,X_2,\ldots,X_n,\ldots$ de elementos ale\-a\-t\'o\-ri\-os de $\Omega$, independentes e identicamente distribu\'\i dos, com a lei comum $\mu$. 
A independ\^encia significa a mesma coisa que no caso de $\Omega$ finito: dado pontos quaisquer $a_{i_1},a_{i_2},\ldots,a_{i_k}$, temos
\begin{eqnarray*}
P[X_1=a_{i_1},X_2=a_{i_2},\ldots,X_k =
a_{i_k}]
&=&
P[X_1=a_{i_1}] P[X_2=a_{i_2}] \ldots P[X_k=a_{i_k}]
\\
&=& p_{i_q}p_{i_2}\ldots p_{i_k}.
\end{eqnarray*}

Esta vez, a observa\c c\~ao de base \'e a seguinte. Seja $k\in\N_+$ qualquer. Se esperarmos bastante tempo, ent\~ao, com alta confian\c ca, todos os pontos $a_1,a_2,\ldots,a_k$ aparecer\~ao entre os pontos da amostra $x_1,x_2,\ldots,x_n$, desde que $n$ \'e bastante grande. 

Formalizemos esta ideia.
Podemos supor sem perda de generalidade que os \'atomos $a_j$ sejam ordenados da tal maneira que
\[p_1=\mu\{a_1\}\geq p_2=\mu\{a_2\}\geq\ldots\geq p_k=\mu\{a_k\}\geq\ldots.\]
Usando o mesmo argumento que na prova do fato \ref{f:3}, conclu\'\i mos que, dado $k\geq 1$ e $\delta>0$, se $n$ \'e maior ou igual ao valor  
\[\frac 1{p_k}\ln\frac{k}{\delta},\] 
ent\~ao, com confian\c ca $1-\delta$, cada ponto $a_1,\ldots,a_k$ ocorre pelo menos uma vez na sequ\^encia gerada aleatoriamente, $X_1,X_2,\ldots,X_n$. 

Se $\e>0$, ent\~ao existe $k$ tal que a soma das medidas de $k$ primeiros \'atomos \'e pelo menos $1-\e$, ou seja,
\[\sum_{j=k+1}^{\infty} p_j < \e.\]
Denotemos o menor $k$ com esta propriedade $k(\e)$. Conclu\'\i mos: se
\[n\geq \frac 1{p_{k(\e)}}\ln\frac{k}{\delta},\] 
ent\~ao
\[\mu^{\otimes n}\{\sigma=(x_1,x_2,\ldots,x_n)\in\Omega^n\colon \{a_1,a_2,\ldots,a_k\}\subseteq\{x_1,x_2,\ldots,x_n\}\}\geq 1-\delta.\]

Agora mais uma vez passemos a aprendizagem {\em supervisionada,} que se trata das amostras {\em rotuladas}. Esta vez, discutamos o problema em mais detalhes. Queremos aprender um {\em conceito desconhecido}, $C\subseteq\Omega$, escolhido pelo {\em professor} ({\em teacher}). 
\index{professor}
Uma amostra aleat\'oria n\~ao rotulada, $\sigma=(x_1,x_2,\ldots,x_n)$, \'e gerada pelo gerador de dados aleat\'orios, e o professor induz a rotulagem $C\upharpoonright\sigma=(\e_1,\e_2,\ldots,\e_n)$ usando o seu conceito segredo, $C$:
\[\e_i=\begin{cases} 1,&\mbox{ se }x_i\in C,\\
0,&\mbox {se n\~ao.}\end{cases}\]
O professor depois mostra a amostra rotulada
\[(\sigma,C\upharpoonright \sigma)=(x_1,x_2,\ldots,x_n,\e_1,\e_2,\ldots,\e_n)\]
ao {\em aprendiz} ({\em learner}), \index{aprendiz}
que far\'a o melhor que puder para adivinhar o conceito $C$. A sua {\em hip\'otese,} $H={\mathcal L}(\sigma,C\upharpoonright \sigma)$, ser\'a enviada ao professor, que estima o {\em erro} (o {\em erro de generaliza\c c\~ao}), ou seja, a probabilidade que o r\'otulo de verdade de um ponto aleat\'orio $X$ n\~ao seja igual ao r\'otulo predito pelo aprendiz:
\begin{align*}
\mbox{erro}_{\mu,C}(H) &=P[Y\neq H] \\
&=
P[\chi_C(X)\neq \chi_H(X)] \\
&= \mu (C\Delta H).
\end{align*}
\index{erro! de generaliza\c c\~ao}
Aqui,
\[C\Delta H = (C\setminus H)\cup (H\setminus C)\]
\'e a {\em diferen\c ca sim\'etrica} de $C$ e $H$. 
\index{diferen\c ca sim\'etrica}

A correspond\^encia
\[(\sigma,C\upharpoonright \sigma)\mapsto H={\mathcal L}(\sigma,C\upharpoonright \sigma),\]
ou seja, o algoritmo usado pelo aprendiz para escolher suas hip\'oteses,
\'e a {\em regra de aprendizagem}.
\index{regra! de aprendizagem} 
A regra {\em aprende} o conceito $C$ ({\em sob a distribui\c c\~ao $\mu$}) se, quando o tamanho de dados crescer, $n\to\infty$, o erro converge para zero. 

Sem se preocupar com a formaliza\c c\~ao mais exata destas no\c c\~oes, tentemos definir uma regra que aprende o conceito desconhecido, $C$, sob a nossa distribui\c c\~ao discreta, $\mu$. 

Isso \'e muito simples! Para cada $\e>0$ e $\delta>0$, se o tamanho da amostra, $n$, for bastante grande, ent\~ao, com confian\c ca $\geq 1-\delta$, temos entre os pontos de amostra tantos \'atomos, $a_1,a_2,\ldots,a_{k(\e)}$, que a soma de suas medidas \'e $>1-\e$. Desta maneira, sabemos as rotulagens da maioria de pontos do dom\'\i nio (a maioria no sentido da lei, $\mu$). Podemos determinar com alta confian\c ca o conjunto 
\[C\cap \{a_1,a_2,\ldots,a_{k(\e)}\}.\]
A nossa predi\c c\~ao pode consistir, por exemplo, de todos os pontos da amostra $\sigma$ tendo o r\'otulo $1$:
\[H={\mathcal L}_n(C\upharpoonright\sigma) = \{x_i\colon \e_i=1,~i=1,2,\ldots,n\}.\]

Com confian\c ca $\geq 1-\delta$, 
\[\{a_1,a_2,\ldots,a_{k(\e)}\}\subseteq \{x_1,\ldots,x_n\},\]
o que implica
\[\mu(C\setminus H)<\e\]
e, como $H\subseteq C$, conclu\'\i mos:
\[\mbox{erro}_{\mu,C}(H)<\e,\]
com a probabilidade $\geq 1-\delta$. A {\em complexidade amostral}
\index{complexidade! amostral} 
de uma regra de aprendizagem \'e o m\'\i nimo tamanho da amostra necess\'ario para atingir a confian\c ca $1-\delta$ e a precis\~ao $\e>0$ exigidos. Ent\~ao, a regra definida acima aprende $C$ com a complexidade amostral
\[s(\e,\delta)\leq \frac 1{p_{k(\e)}}\ln\frac{k}{\delta}.
\]
A regra aprende $C$ {\em provavelmente aproximadamente corretamente,} onde {\em provavelmente} refere-se a $\delta$, e {\em aproximadamente}, a $\e$. 

Vamos fazer duas observa\c c\~oes.
Note-se que a nossa regra possui uma propriedade mais forte: ela pode aprender todos os conceitos $C\in2^{\Omega}$ {\em simultaneamente,} ou seja:
\[\forall\e>0~\forall\delta>0\mbox{ se }n\geq s(\e,\delta),\mbox{ ent\~ao }
P\left[\forall C\in 2^{\Omega}~\mbox{erro}_{\mu,C}({\mathcal L}(C\upharpoonright \sigma)\leq\e \right]>1-\delta.
\]
Al\'em disso, em vez dessa regra particular, pode-se usar uma regra qualquer que apenas tem a propriedade que a hip\'otese $H$ induz sobre a amostra $\sigma$ a mesma rotulagem que o conceito $C$:
\[{\mathcal L}(C\upharpoonright \sigma)\upharpoonright \sigma = C\upharpoonright \sigma.\]
Uma tal regra chama-se {\em consistente} (com a classe, no nosso caso $2^\Omega$). Pode-se dizer que a classe $2^{\Omega}$ \'e {\em consistentemente aprendiz\'avel} sob qualquer medida discreta. 
\index{regra! consistente com uma classe}

\section{Aprendizagem de classes finitas}

Vamos passar agora para o caso mais geral, o de uma distribui\c c\~ao fixa $\mu$ qualquer, n\~ao necessariamente discreta, a fim de investigar a pergunta seguinte: \'e verdade que existe uma regra de aprendizagem, $\mathcal L$, que aprende a classe de todos os conceitos (borelianos), ${\mathscr B}_\Omega$, sob a distribui\c c\~ao $\mu$, de mesmo modo que para $\mu$ puramente at\^omica? A resposta \'e negativa, o que nos motiva a procurar por classes que podem ser aprendidas. Os primeiros exemplos s\~ao as classes finitas.

Comecemos com algumas formaliza\c c\~oes.

\subsection{Aprendizagem sob uma medida fixa: formaliza\c c\~oes}
Seja $\Omega$ um dom\'\i nio, ou seja, um espa\c co boreliano padr\~ao (veja o Ap\^endice \ref{apendice:padrao}). Uma {\em regra de aprendizagem} sobre $\Omega$ \'e uma aplica\c c\~ao que aceita uma amostra rotulada e gera um subconjunto boreliano de $\Omega$, normalmente chamado uma {\em hip\'otese,}
\[{\mathcal L}\colon\bigcup_{n=1}^{\infty}\Omega^n\times\{0,1\}^n\to {\mathcal B}_{\Omega}.\]
\index{hip\'otese}
\'E mais conveniente tratar $\mathcal L$ como uma fam\'\i lia de aplica\c c\~oes, ${\mathcal L}=({\mathcal L}_n)_{n=1}^{\infty}$, uma para cada n\'\i vel $n=1,2,3,\ldots$:
\[{\mathcal L}_n\colon\Omega^n\times\{0,1\}^n\to {\mathcal B}_{\Omega}.\]
A maneira alternativa de representar a regra de aprendizagem \'e como uma  fun\c c\~ao que aceita uma amostra rotulada $\sigma=(x_1,x_2,\ldots,x_n,\e_1,\e_2\ldots,\e_n)\in \Omega^n\times\{0,1\}^n$, bem como um ponto $x\in\Omega$, e produz um r\'otulo, $0$ ou $1$, para $x$ baseado sobre as informa\c c\~oes dadas por $\sigma$:
\[{\mathcal L}_n\colon\Omega^n\times\{0,1\}^n\times\Omega\to \{0,1\}.\]
Nesta interpreta\c c\~ao, para dar sentido \`as express\~oes tipo $P[{\mathcal L}_n=\chi_C]$, as fun\c c\~oes ${\mathcal L}_n$ devem ser borelianas. 
\index{regra! de aprendizagem}

Seja $C\subseteq\Omega$ um conceito, ou seja, um subconjunto boreliano de $\Omega$.
\index{conceito}
Dada uma amostra n\~ao rotulada, $\sigma\in\Omega^n$, o conceito $C$ induz uma rotulagem, $C\upharpoonright \sigma$. A regra $\mathcal L$ associa com a amostra rotulada $(\sigma,C\upharpoonright \sigma)$ uma hip\'otese, $H={\mathcal L}_n(\sigma,C\upharpoonright \sigma)$. Para facilitar, vamos escrever $H={\mathcal L}_n(C\upharpoonright \sigma)$. 

Agora suponha que o dom\'\i nio seja munido de uma distribui\c c\~ao, ou seja, uma medida de probabilidade boreliana qualquer, $\mu$ (Ap\^endice \ref{a:medidas}). Para cada hip\'otese $H\in{\mathscr B}_\Omega$, pode-se calcular o {\em erro de generaliza\c c\~ao} (com rela\c c\~ao ao conceito $C$ e a medida $\mu$):
\index{erro de generaliza\c c\~ao}
\[\mbox{erro}_{\mu,C}(H)=\mu(H\Delta C)=P[\chi_H(X)\neq\chi_C(X)]
=\E(\abs{\chi_H-\chi_C})=\norm{\chi_H-\chi_C}_{L^1(\mu)}.\]
Em particular, chegamos ao {\em erro de generaliza\c c\~ao} da regra $\mathcal L$ (ao n\'ivel $n$):
\[\mbox{erro}_{\mu,C}{\mathcal L}_n(C\upharpoonright \sigma).\]
Este erro \'e em si uma vari\'avel aleat\'oria, uma composi\c c\~ao da amostra aleat\'oria\footnote{Uma advert\^encia: o s\'\i mbolo $\sigma$, nesta \'area de conhecimento, pode denotar uma amostra aleat\'oria, $\sigma=(X_1,X_2,\ldots,X_n)$, bem como uma amostra determin\'\i stica, $\sigma=(x_1,x_2,\ldots,x_n)$, ou seja, um elemento do conjunto $\Omega^n$. De mesmo, para as amostras rotuladas. As vezes, vamos usar $W$ quando for necess\'ario de distinguir entre as duas.} $\sigma=(X_1,X_2,\ldots,X_n)$ com tr\^es aplica\c c\~oes determin\'\i sticas: a de rotulagem,
\[\Omega^n\ni\sigma\mapsto (\sigma,C\upharpoonright\sigma)= (x_1,x_2,\ldots,x_n,\chi_C(x_1),\chi_C(x_2),\ldots,\chi_C(x_n))\in \Omega^n\times\{0,1\}^n,\]
a da regra de aprendizagem,
\[\Omega^n\times\{0,1\}^n\ni(\sigma,C\upharpoonright\sigma)\mapsto H = {\mathcal L}(\sigma,C\upharpoonright \sigma)\in {\mathscr B}_{\Omega},\]
e afinal, a de erro de generaliza\c c\~ao:
\[{\mathscr B}_{\Omega}\ni H\mapsto\mbox{erro}_{\mu,C}(H)\in\R.\]
Pode-se escrever assim:
\[\mbox{erro}_{\mu,C}({\mathcal L}((X_1,X_2,\ldots,X_n),C\upharpoonright (X_1,X_2,\ldots,X_n))).\]
A esperan\c ca desta v.a. \'e o erro m\'edio de generaliza\c c\~ao da regra ${\mathcal L}_n$:
\begin{align*}
\mbox{erro}_{\mu,C}({\mathcal L}_n)&=\E\left(\mbox{erro}_{\mu,C}({\mathcal L}((X_1,X_2,\ldots,X_n),C\upharpoonright (X_1,X_2,\ldots,X_n))) \right)\\
&= \E\left(\mu\left(C\Delta {\mathcal L}((X_1,X_2,\ldots,X_n,\chi_C(X_1),\chi_C(X_2),\ldots,\chi_C(X_n)))\right) \right)\\
&= \E\left(P\left[\chi_C(X)\neq {\mathcal L}((X_1,X_2,\ldots,X_n),\chi_C(X_1),\chi_C(X_2),\ldots,\chi_C(X_n))(X)\right] \right).
\end{align*}
Usualmente, escrevemos a \'ultima express\~ao assim:
\[\E_{\sigma\sim\mu^{\otimes n}}\left(P_{X\sim\mu}\left[\chi_C(X)\neq {\mathcal L}(C\upharpoonright \sigma)(X)\right] \right),\]
ou, por um ligeiro abuso de nota\c c\~ao,
\[\E_{\sigma\sim\mu}\left(P_{X\sim\mu}\left[\chi_C(X)\neq {\mathcal L}(C\upharpoonright \sigma)(X)\right] \right).\]

Agora podemos formular algumas defini\c c\~oes exatas.

\begin{definicao}
A regra $\mathcal L$ aprende o conceito $C$ sob a distribui\c c\~ao $\mu$ {\em provavelmente aproximadamente corretamente} ({\em probably approximately correctly, PAC})  se
\[\mbox{erro}_{\mu,C}({\mathcal L}_n)\to 0\mbox{ quando }n\to\infty.\]
Da maneira equivalente: a sequ\^encia de v.a. reais $\mbox{erro}_{\mu,C}({\mathcal L}_n(C\upharpoonright \sigma))$, $n=1,2,3,\ldots$, converge para zero em probabilidade:
\begin{eqnarray*}
\forall\e>0,~\forall\delta>0,~\exists s=s(\e,\delta),~~\forall n\geq s,~~
P\left[\mbox{erro}_{\mu,C}({\mathcal L}_n(C\upharpoonright \sigma))>\ve \right]\leq\delta.
\end{eqnarray*}
[ O que \'e denotado
\[\mbox{erro}_{\mu,C}({\mathcal L}_n(C\upharpoonright \sigma))\overset p\to 0~].\]
\end{definicao}

\begin{exercicio}
Mostrar a equival\^encia de duas no\c c\~oes acima. 
\par
[ {\em Sugest\~ao:} ela \'e verdadeira para uma sequ\^encia qualquer de v.a. reais que tomam seus valores num intervalo, neste caso, o intervalo $[0,1]$. ]
\end{exercicio}

\begin{definicao}
Seja $\mathscr C$ uma classe de conceitos, ou seja, uma fam\'\i lia de conjuntos borelianos $\mathscr C\subseteq {\mathscr B}_{\Omega}$. A regra $\mathcal L$ aprende a classe $\mathscr C$ sob a distribui\c c\~ao $\mu$ {\em provavelmente aproximadamente corretamente} se 
\[\sup_{C\in{\mathscr C}}\mbox{erro}_{\mu,C}({\mathcal L}_n)\to 0\mbox{ quando }n\to\infty.\]
De maneira equivalente: dado $\ve>0$ ({\em precis\~ao}) e $\delta>0$ ({\em risco}), existe o valor $s=s(\e,\delta)$ (a {\em complexidade amostral}) tal que, se $n\geq s(\e,\delta)$, ent\~ao, qualquer que seja $C\in {\mathscr C}$,
\[\mu^{\otimes n}\{\sigma\in\Omega^n\colon\mbox{erro}_{\mu,C}{\mathcal L}_n(C\upharpoonright\sigma)>\ve\}\leq\delta.\]
\index{regra! provavelmente aproximadamente correta (PAC)}
\end{definicao}

\begin{definicao}
Diz-se que uma classe de conceitos $\mathscr C$ \'e {\em aprendiz\'avel} (ou: {\em PAC aprendiz\'avel}) {\em sob a medida} $\mu$ se existe uma regra de aprendizagem, $\mathscr C$, que aprende a classe $\mathscr C$ sob a distribui\c c\~ao $\mu$.
\index{classe! PAC aprendiz\'avel}
\end{definicao}

\begin{observacao}
O modelo de aprendizagem provavelmente aproximadamente correta foi sugerido em \citep*{valiant84learnable}. A no\c c\~ao \'e particularmente importante no contexto da aprendizagem independente da medida (se\c c\~ao \ref{s:pacindepmedida}).
\end{observacao}

\subsection{Propriedade geom\'etrica de uma classe $\mathscr C$ aprendiz\'avel\label{s:propgeom}}

Vamos examinar a pergunta seguinte, que fica no cora\c c\~ao do assunto.  
Seja $\mathscr C$ \'e uma classe de conceitos num dom\'\i nio $\Omega$ munido de uma lei, $\mu$. Suponhamos que existe uma regra $\mathcal L$ que aprende $\mathscr C$ provavelmente aproximadamente corretamente. Vamos notar uma propriedade geom\'etrica de $\mathscr C$ que segue desta suposi\c c\~ao.

\begin{exercicio} Verificar que a fun\c c\~ao de dois argumentos, $\mu(C\Delta C^\prime)$, \'e uma {\em pseudom\'etrica}, ou seja, satisfaz $\mu(C\Delta C)=0$, \'e sim\'etrica, e satisfaz a desigualdade triangular. De fato, este valor \'e igual \`a dist\^ancia $L^1$ entre as fun\c c\~oes indicadoras de conjuntos $C$ e $C^\prime$:
\[\lambda(C,C^\prime) = \int_0^1\abs{\chi_C-\chi_{C^\prime}}d\lambda(x) =
\norm{\chi_C-\chi_{C^\prime}}_1.\]
\index{pseudom\'etrica}
\end{exercicio}

Sejam $0<\ve,\delta<1$ quaisquer. Seja $C_1,C_2,\ldots,C_k$ um subconjunto finito de $\mathscr C$ de elementos \`a dist\^ancia dois a dois $\geq 2\ve$. 
Seja $n\geq s(\e,\delta)$. Para cada $i=1,2,\ldots, k$, denotemos
\[A_i=\{\sigma\in\Omega^n\colon \mu(C_i,{\mathcal L}(C_i\upharpoonright\sigma))>\ve \}.\]
Definamos a fun\c c\~ao $f\colon\Omega^n\to \N$ por $f=\sum_{i=1}^k \chi_{A_i}$. Como $\mu^{\otimes n}(A_i)\leq\delta$, temos $\int_{\Omega}f\,d \mu^{\otimes n}\leq k\delta$, logo existe pelo menos uma amostra $\sigma$ tal que 
\[\sharp J=\sharp\{i\colon \sigma\in A_i\}\leq k\delta.\]
 Para o subconjunto complimentar $I=[k]\setminus J$ com $> (1-\delta)k$ \'\i ndices, temos
\[i\in I\Rightarrow\mu(C_i,{\mathcal L}(C_i\upharpoonright\sigma))\leq\ve. \]
Gra\c cas a nossa escolha de conceitos $C_i$, conclu\'\i mos que, se $i,j\in I$ e $i\neq j$, ent\~ao
\[\mu({\mathcal L}(C_i\upharpoonright\sigma)\Delta {\mathcal L}(C_j\upharpoonright\sigma))>0,\]
em particular, 
\[{\mathcal L}(C_i\upharpoonright\sigma)\neq{\mathcal L}(C_j\upharpoonright\sigma).\]
Mas isso significa que os inputs da regra $\mathcal L$, $C_i\upharpoonright\sigma$ e $C_j\upharpoonright\sigma$, s\~ao diferentes. Em outras palavras, o cubo de Hamming $\{0,1\}^n$ de todas as rotulagens da amostra $\sigma$ com $n$ elementos admite uma inje\c c\~ao $[\lceil k(1-\delta)\rceil]\to 2^n$. Conclu\'\i mos: $k(1-\delta)\leq 2^n$, ou seja, $\log_2k+\log_2(1-\delta)\leq n$. Em particular, quando $\delta\leq 1/2$, $\log_2(1-\delta)\geq -1$, e quando $\delta<1/k$, pode-se ver que $n\geq \log_2k$.

Resumamos as observa\c c\~oes acima.

\begin{lema}
Seja $\mathscr C$ uma classe de conceitos PAC aprendiz\'avel sob uma medida $\mu$. Ent\~ao, para cada $\e>0$, as cardinalidades de subfam\'\i lias de $\mathscr C$ de conjuntos $2\e$-discretas (isto \'e, dois a dois \`a dist\^ancia $L^1> 2\e$) s\~ao limitadas por acima por $2^{s(\e,\delta)+1}$, quando $\delta\leq 1/2$.
\label{l:2ediscrete}
\end{lema}

Isto significa, em particular, que $\mathscr C$ \'e {\em pr\'e-compacto} (ou: {\em totalmente limitado}) com rela\c c\~ao \`a dist\^ancia $L^1$ (veja a subse\c c\~ao \ref{ss:espacospre-compactos}). A observa\c c\~ao seguinte \'e imediata.
\index{espa\c co! m\'etrico! pr\'e-compacto}
\index{espa\c co! m\'etrico! totalmente limitado}

\begin{fato}
Suponha que a classe $\mathscr C$ cont\'em uma sequ\^encia infinita de conceitos $C_1,C_2,\ldots,C_k,\ldots$, dois a dois a dist\^ancia $>\e>0$ com rela\c c\~ao \`a m\'etrica $L^1(\mu)$. Ent\~ao $\mathscr C$ n\~ao \'e PAC aprendiz\'avel sobre $\mu$. 
\end{fato}

\begin{exemplo}
Existe uma fam\'\i lia de subconjuntos borelianos do intervalo que \'e infinita \'e $1/2$-discreta. \'E a fam\'\i lia de conjuntos de {\em Rademacher,} constru\'\i dos recursivamente assim:
\begin{eqnarray*}
R_1 &=& [0,1],\\
R_2 &=& [0,1/2],\\
R_3&=& [0,1/4]\cup [1/2,3/4],\\
R_4&=& [0,1/8]\cup [1/4,3/8]\cup [1/2,5/8]\cup [3/4,7/8], \\
\ldots && \ldots \\
R_n &=& \bigcup_{i=0}^{2^{n-1}-1} [2^{-n+1}i,2^{-n+1}i+2^{-n}],\\
\ldots && \ldots \\
\end{eqnarray*}
\index{fam\'\i lia! de Rademacher}
\'E f\'acil a verificar a propriedade desejada: se $n\neq m$, ent\~ao $\lambda(R_n\Delta R_m)=1/2$. Conclu\'\i mos: a classe de todos os conceitos (borelianos), ${\mathscr B}_\Omega$, n\~ao \'e aprendiz\'avel sob a medida uniforme no intervalo $[0,1]$.
\end{exemplo}

\begin{exercicio}
Mostre que a mesma conclus\~ao vale para qualquer medida $\mu$ sobre um espa\c co boreliano padr\~ao que tem uma parte n\~ao at\^omica. \par
[ {\em Sugest\~ao:} use o teorema \ref{t:parametrizacao} para reduzir este caso ao caso acima. ]
\end{exercicio}

Agora, duas perguntas merecem a ser investigadas: quando uma dada classe de conceitos, $\mathscr C$, for PAC aprendiz\'avel sob uma medida qualquer $\mu$, e quando $\mathscr C$ for consistentemente aprendiz\'avel sob $\mu$. 

Em breve, vamos apresentar uma solu\c c\~ao completa do primeiro problema
dada pelo teorema de Benedek e Itai.
Por enquanto, vamos provar que todas as classes finitas s\~ao PAC aprendiz\'aveis.

\subsection{Uma forma mais geral da lei geom\'etrica dos grandes n\'umeros}
No primeiro cap\'\i tulo mostramos o resultado seguinte (corol\'ario \ref{c:interplay2}). 
Seja $f\colon \Sigma^n\to\R$ uma fun\c c\~ao 1-Lipschitz cont\'\i nua em rela\c c\~ao a dist\^ancia de Hamming normalizada. Ent\~ao para todo $\e>0$
\[\mu_\sharp\{x \colon \abs{f(x)- \E(f)}\geq \e\}\leq
2e^{-2\e^2n}.\] 

O cubo de Hamming $\Sigma^n$, como espa\c co probabil\'\i stico, \'e o produto de $n$ c\'opias do espa\c co probabil\'\i stico simpl\'\i ssimo: $\Omega=\{0,1\}$, munido da medida de probabilidade 
\[\mu\{0\}=\mu\{1\}=1/2.\]
Este espa\c co, o espa\c co de Bernoulli, modeliza o \'unico lan\c camento de uma moeda equilibrada. 
No entanto, n\~ao h\'a nenhuma raz\~ao matem\'atica para que espa\c co de Bernoulli n\~ao possa ser substitu\'\i do por um espa\c co probabil\'\i stico padr\~ao qualquer, $(\Omega,\mu)$. A demonstra\c c\~ao permanece a mesma, com algumas modifica\c c\~oes \'obvias.

Seja $(\Omega,\mu)$ um espa\c co boreliano padr\~ao qualquer, munido de uma medida de probabilidade, e seja $n\in\N$. Sobre o produto cartesiano $\Omega^n$, definamos a m\'etrica (a dist\^ancia de Hamming normalizada): 
\[\bar d(x,y) = \frac 1n \sharp\{i\colon x_i\neq y_i\}.\]
Uma advert\^encia: esta m\'etrica, em geral, n\~ao gera a estrutura boreliana do produto, e de fato ela n\~ao \'e necessariamente mensur\'avel. Contudo, ela se comporta bem.

\begin{exercicio}
Seja $B$ um subconjunto mensur\'avel de $\Omega^n$ com rela\c c\~ao \`a estrutura boreliano produto. Mostrar que para todo $\ve>0$ a $\ve$-vizinhan\c ca de $B$ formada usando a m\'etrica $\bar d$ \'e um conjunto mensur\'avel.
\end{exercicio}

\begin{teorema}[Lei geom\'etrica dos grandes n\'umeros]
Seja $f\colon (\Omega^n,\bar d)\to\R$ uma fun\c c\~ao $1$-Lipschitz cont\'\i nua mensur\'avel com rela\c c\~ao \`a estrutura boreliana do produto sobre $\Omega^n$. Ent\~ao para todo $\ve>0$ temos
\[\mu^{\otimes n}\{x\in\Omega^n\colon \left\vert f(x)-\E f\right\vert\geq\ve\}\leq 2e^{-2\ve^2n}.\]
\qed\label{t:leigeneralizada}
\index{lei dos grandes n\'umeros! geom\'etrica}
\end{teorema}

\begin{exercicio}
Refazer a sequ\^encia de resultados no primeiro cap\'\i tulo resultando em corol\'ario \ref{c:interplay2} para obter formalmente o resultado mais geral acima.
\end{exercicio}

O teorema admite a forma equivalente seguinte.

\begin{teorema}
Sejam $X_i$, $i=1,2,3,\ldots$, elementos aleat\'orios i.i.d. de um espa\c co boreliano padr\~ao, $\Omega$. Seja $f\colon (\Omega^n,\bar d)\to\R$ uma fun\c c\~ao $1$-Lipschitz cont\'\i nua e mensur\'avel com rela\c c\~ao a estrutura boreliana de produto sobre $\Omega^n$. Ent\~ao para todo $\ve>0$ temos
\[P\left[\left\vert f(X_1,X_2,\ldots,X_n) - \E f\right\vert\geq \e\right] \leq 2e^{-2\ve^2n}.\]
\label{t:leigeneralizada_va}
\qed
\end{teorema}

\begin{exercicio}
Estabelecer a equival\^encia dos teoremas \ref{t:leigeneralizada} e \ref{t:leigeneralizada_va}.
\end{exercicio}

\subsection{Medida emp\'\i rica contra a medida subjacente\label{ss:casoimportante}}
Seja $C$ um conceito, isto \'e, um subconjunto boreliano de $\Omega$. Queremos estimar o valor da medida de $C$, $\mu(C)$, usando a amostragem. Dado uma amostra com $n$ pontos, $x_1,x_2,\ldots,x_n$, pode-se calcular a fra\c c\~ao de pontos dela que pertencem a $C$:
\begin{equation}
\mu_n(C) = \frac 1n \sharp\{i\colon x_i\in C\} = \frac 1n\sum_{i=1}^n \chi_C(x_i).
\end{equation}
A quantidade $\mu_n$ \'e a {\em medida emp\'\i rica} suportada em  $\sigma=\{x_1,x_2,\ldots,x_n\}$. 
\index{medida! emp\'\i rica}
Vamos denot\'a-la $\mu_\sigma$ ou \`as vezes $\mu_n$. A medida emp\'\i rica \'e uma medida de probabilidade em $\Omega$, um substituto de $\mu$ dispon\'\i vel a n\'os.

A pergunta \'e, qual \'e a probabilidade de que $\mu_n(C)$ estima $\mu(C)$ razoavelmente bem? Ou seja, dado um erro $\ve>0$, qual \'e o {\em risco} de um desvio grande,
\[\mu^{\otimes n}\left\{\sigma\in\Omega^n\colon \abs{\mu_\sigma(C)-\mu(C)}>\ve\right\}?\]

A fun\c c\~ao
\[\Omega^n\ni \sigma \mapsto \mu_{\sigma}(C) =\frac 1n\sum_{i=1}^n \chi_C(x_i) \in\R\] 
possui as propriedades seguintes: 

\begin{enumerate}
\item $\sigma\mapsto\mu_{\sigma}(C)$ \'e $1$-Lipschitz cont\'\i nua com rela\c c\~ao \`a m\'etrica de Hamming normalizada $\bar d$ sob $\Omega^n$. De fato, se $\bar d(\sigma,\tau)=a$, ent\~ao $an$ coordenadas de $\sigma$ e $\tau$ s\~ao diferentes, e as duas somas diferem por $an$ no m\'aximo; depois da normaliza\c c\~ao, conclu\'\i mos que $\mu_\sigma(C)$ e $\mu_\tau(C)$ diferem por $\leq a$.
\vskip .2cm
\item A fun\c c\~ao \'e Borel mensur\'avel com rela\c c\~ao \`a estrutura boreliana de produto (exer\-c\'\i \-ci\-o).
\vskip .2cm
\item A fun\c c\~ao satisfaz
\[\E_{\sigma\sim\mu}[\mu_\sigma(C)] = \mu(C).\]
De fato,
\begin{align*}
\E_{\sigma\sim\mu}[\mu_\sigma(C)] &= \int_{\Omega^n}\frac 1n\sum_{i=1}^n \chi_C(x_i)\,d\mu^{\otimes n}(x) \\
&=  \frac 1n \sum_{i=1}^n \int_{\Omega}\chi_{C}(x_i)\,d\mu(x_i) 
\\
&= \frac 1n\times n\mu(C) \\
&=\mu(C).
\end{align*}
\end{enumerate}

A cota de Chernoff (teorema \ref{t:leigeneralizada}) implica
\begin{equation}
\label{eq:meas}
\mu^{\otimes n}\left\{x\in\Omega^n\colon \abs{\mu_n(C)-\mu(C)}\geq \ve\right\}\leq 2e^{-2\ve^2n}.\end{equation}
De mesmo, usando nota\c c\~ao probabil\'\i stica, l\^e-se
\begin{equation}
\label{eq:prob1}
P\left[\abs{\mu_n(C)-\mu(C)}\geq \ve
\right]\leq 2e^{-2\ve^2n}.\end{equation}
Aqui, a medida emp\'\i rica $\mu_n$ em si \'e uma vari\'avel aleat\'oria com valores no espa\c co de medidas de probabilidade sobre $\Omega$, isso \'e, uma medida aleat\'oria. Segue-se a f\'ormula escrita mais cuidadosamente:
\[P\left[\left\vert \frac 1n\sum_{i=1}^n \chi_C(X_i) -\mu(C)\right\vert >\ve
\right]<2e^{-2\ve^2n}.\]
O sinal de probabilidade refere-se \`a medida produto $\mu^{\otimes n}$.

Agora suponha que ao inv\'es de um conceito, temos uma fam\'\i lia finita
\[{\mathscr C} = \{C_1,C_2,\ldots,C_k\}.\]
Queremos estimar suas medidas simultaneamente, usando a \'unica amostra aleat\'oria dispon\'ivel,
\[\sigma=\{x_1,x_2,\ldots,x_n\}.\]

Isso pode ser feito, embora com uma confian\c ca um pouco pior, usando a cota de uni\~ao:
\begin{align*}
\mu^{\otimes n}\left\{x\in\Omega^n\right.&\colon\left. \max_{j=1}^k\left\vert \frac 1n\sum_{i=1}^n \chi_{C_j}(X_i) -\mu(C_j)\right\vert \geq\ve\right\} \\ &\leq
k\cdot \max_{j=1}^k \mu^{\otimes n}\left\{x\in\Omega^n\colon \abs{\mu_n(C_j)-\mu(C_j)}\leq \ve\right\} \\
&\leq 2ke^{-2\ve^2n}.
\end{align*}

\begin{corolario}
Dado $k$ conceitos $C_1,C_2,\ldots,C_k$ e um $\e>0$, 
com confian\c ca $\geq 1- 2ke^{-2\ve^2n}$, a medida emp\'\i rica de cada um de $k$ conceitos difere da sua medida por menos de $\e>0$:
\[\mu^{\otimes n}\left\{\exists i=1,2,\ldots,k~\left\vert \mu_{\sigma}(C_i)-\mu(C)\right\vert\geq \e
\right\}\leq 2ke^{-2\ve^2n}.
\]
\label{c:medidas}
\end{corolario}

Usando a nota\c c\~ao probabil\'\i stica:

\begin{equation}
\label{eq:probk}
P\left[\max_{j=1}^k\abs{\mu_n(C_j)-\mu(C_j)}>\ve
\right]\leq 2ke^{-2\ve^2n}.\end{equation}

De fato, a mesma prova se aplica no contexto mais geral quando as fun\c c\~oes indicadores s\~ao substitu\'\i das por fun\c c\~oes quaisquer com valores no intervalo $[0,1]$. A papel da medida emp\'\i rica $\mu_\sigma$ ser\'a exercida pela {\em esperan\c ca emp\'\i rica} de uma fun\c c\~ao,
\[\E_{\mu_{\sigma}}(f) = \frac 1 n\sum_{i=1}^n f(x_i).\]
\index{esperan\c ca! emp\'\i rica}

\begin{teorema}
Sejam $f_1,f_2,\ldots,f_k$ fun\c c\~oes borelianas sobre $\Omega$ com valores em $[0,1]$. Com a confian\c ca $\geq 1-2k e^{-2\ve^2n}$, a esperan\c ca emp\'\i rica $\E_{\mu_{\sigma}}$ simultaneamente aproxima a esperan\c ca $\E_{\mu}$ para todas as fun\c c\~oes $f_i$, $i=1,2,\ldots,k$ com precis\~ao $\leq\ve$:
\[\mu^{\otimes n}\left\{\sigma\in\Omega^n\colon \exists i=1,2,\ldots k~\left\vert \E_{\mu}(f_i)-\E_{\mu_{\sigma}}(f_i)\right\vert>\ve \right\}\leq 2k e^{-2\ve^2n}.\]
\label{t:aproximacao}
\end{teorema}

\begin{exercicio}
Mostrar o teorema \ref{t:aproximacao}.
\end{exercicio}

Uma classe de fun\c c\~oes cujas esperan\c cas podem ser aproximadas simultaneamente pelas esperan\c cas emp\'\i ricas \'e dita uma {\em classe de Glivenko--Cantelli}. Vamos formalizar esta no\c c\~ao em breve.
\index{classe! de Glivenko--Cantelli}

\begin{corolario}
Cada classe finita de conceitos, ${\mathscr C}= \{C_1,C_2,\ldots, C_k\}$, \'e  PAC aprendiz\'avel sob qualquer medida fixa, $\mu$, com a complexidade amostral 
$s(\e,\delta)\leq\frac  1{2\e^2}\log\frac{2k}{\delta}$.
\end{corolario}

\begin{proof}
Dado um conceito desconhecido, $C=C_i\in {\mathscr C}$, e uma amostra $\sigma$, assim como a rotulagem $C\upharpoonright \sigma$ induzida por $C$, vamos buscar o elemento $C_m$ mais pr\'oximo a $C$ no sentido da dist\^ancia emp\'\i rica $L^1(\mu_\sigma)$:
\begin{align*}
m&=\arg\min_{j=1}^k \mu_{\sigma}(C\Delta C_j)\\
&= \arg\min_{j=1}^k \frac 1n\sum_{i=1}^n \abs{\chi_C(x_i)-\chi_{C_j}(x_i)}.
\end{align*}
A hip\'otese gerada pela nossa regra de aprendizagem ser\'a $H=C_m$. 

Como $\mu_{\sigma}(C\Delta C_i)=0$, segue-se que $\mu_{\sigma}(C\Delta C_m)=0$ tamb\'em. 
Seja $\e>0$.
Aplicando corol\'ario \ref{c:medidas} aos conceitos
\[C\Delta C_1, C\Delta C_2,\ldots, C\Delta C_k,\]
conclu\'\i mos que, com confian\c ca $\geq 1- 2ke^{-2\ve^2n}$, 
\begin{align*}
\mbox{erro}_H(C) &= \mu(C\Delta C_m) \\
& \overset\e\approx  \mu_{\sigma}(C\Delta C_m) =0.
\end{align*}
Dado $\delta>0$ qualquer, conclu\'\i mos: se $n\geq \frac 1{2\e^2}\log\frac{2k}{\delta}$, logo, com confian\c ca $\geq 1-\delta$,
$\mbox{erro}_H(C) <\e$.
\end{proof}

\begin{observacao}
A express\~ao $\mu_{\sigma}(C\Delta H)$ \'e conhecida como o {\em erro} (as vezes: {\em risco}) {\em emp\'\i rico.} Por isso, a nossa estrat\'egia de escolha da hip\'otese se chama {\em minimiza\c c\~ao do erro} (ou: de {\em risco}) {\em emp\'\i rico.}
\index{erro! emp\'\i rico}
\end{observacao}

\begin{observacao} 
Torna-se explicito que a complexidade amostral da aprendizagem de uma classe finita n\~ao depende da medida $\mu$. Diz-se que uma tal classe \'e {\em uniformemente PAC aprendiz\'avel} (tendo em mente: uniformemente sobre todas as medidas de probabilidade). 
\end{observacao}

\subsection{Redu\c c\~ao de dimensionalidade no cubo de Hamming\label{ss:redcubo}}

O teorema \ref{t:aproximacao} pode ser visto como uma t\'ecnica de redu\c c\~ao de dimensionalidade aleatorizada. Suponha que os dados s\~ao realizados como fun\c c\~oes borelianas sobre um espa\c co probabil\'\i stico $(\Omega,\mu)$, e munidas da dist\^ancia $L^1(\mu)$, dada pela norma
\[\norm{f}_{L^1(\mu)}=\int_{\Omega}\abs{f(x)}d\mu(x)=\E_{\mu}(\abs f).\]
No caso particular onde $\Omega$ \'e finito, ${\Omega}=\{x_1,x_2,\ldots,x_m\}$, e a medida $\mu$ \'e uma medida uniforme sobre $[m]$, cada fun\c c\~ao $f\colon\Omega\to\R$ \'e um $m$-vetor, e a norma $L^1(\mu)$ \'e a norma $\ell^1(m)$ normalizada (!) sobre o espa\c co vetorial $\R^m$ de dimens\~ao $m$:
\[\norm{(f(x_1),f(x_2),\ldots,f(x_m))}_1 = \frac 1m\sum_{i=1}^m\abs{f(x_i)}.\]
A norma acima \'e a norma no espa\c co $L^1(\mu_m)$, em rela\c c\~ao a medida emp\'\i rica.

Voltemos no caso de $\Omega$ geral.
Dado uma amostra $\sigma=(x_1,x_2,\ldots,x_m)\in\Omega^m$, a aplica\c c\~ao
\[f\mapsto (f(x_1),f(x_2),\ldots,f(x_m)),\]
pode ser vista como uma proje\c c\~ao do espa\c co $L^1(\Omega,\mu)$ sobre o espa\c co $\frac 1m\ell^1(m)=L^1(\mu_m)$. 

\begin{teorema}
Sejam $f_1,f_2,\ldots,f_n$ fun\c c\~oes borelianas quaisquer sobre um es\-pa\-\c co probabil\'\i stico padr\~ao $\Omega$ com valores em $[0,1]$. Seja $\mu$ uma medida de probabilidade boreliana sobre $\Omega$.
Dado $\ve>0$ e $\delta>0$, se
\begin{equation}
m\geq \frac{1}{\ve^2}\ln\frac{n}{\delta},
\label{eq:scompl}
\end{equation}
ent\~ao com confian\c ca $\geq 1-\delta$, uma proje\c c\~ao aleat\'oria
\[L^1(\Omega,\mu)\ni f\mapsto f\upharpoonright \sigma_m\in L(\mu_m),\]
 conserva as dist\^ancias a menos $\e$:
\[\forall i,j=1,2,\ldots,n,~~\norm{f_i-f_j}_{L^1(\mu)}\overset{\ve}\approx
\norm{f_i\upharpoonright \sigma_m-f_j\upharpoonright \sigma_m}_{L(\mu_m)}.\]
\label{t:projaleator}
\end{teorema}

Usamos a nota\c c\~ao conveniente:
\[a\overset{\ve}\approx b \iff \abs{a-b}\leq\ve.\]

\begin{proof}
Aplicando o teorema \ref{t:aproximacao} para $n(n-1)/2$ fun\c c\~oes $\abs{f_i-f_j}$, $i\neq j$, temos a conclus\~ao desejada com confian\c ca $\geq 1-n(n-1) e^{-2\ve^2 m}$. A desigualdade
\[n(n-1) e^{-2\ve^2 m}\leq \delta\]
\'e equivalente a
\[\ln (n(n-1))-2\ve^2 m \leq\ln\delta,\]
o que \'e satisfeito sob a hip\'otese (\ref{eq:scompl}).
\end{proof}

Deste modo, podemos realizar o conjunto de $n$ dados {\em provavelmente aproximadamente corretamente} dentro de um espa\c co vetorial de dimens\~ao $O(\log n)$. 

\begin{exercicio}
Obter uma modifica\c c\~ao do teorema \ref{t:projaleator} v\'alida para todas as dist\^ancias $L^p(\mu)$, $1\leq p<+\infty$:
\[\norm{f}_{L^p(\mu)}=\left(\int_{\Omega} \abs{f}^pd\mu\right)^{1/p}.\]
\end{exercicio}

Segue-se um caso particular importante.

\begin{exercicio}
Seja $\Omega=[d]=\{1,2,3,\ldots,d\}$ um conjunto finito munido da medida de contagem normalizada, $\mu_{\sharp}$. Verificar que o espa\c co $L^1(\mu_{\sharp})$ \'e o cubo de Hamming $\Sigma^d$, e a dist\^ancia $L^1(\mu_{\sharp})$ \'e a dist\^ancia de Hamming normalizada.
\end{exercicio}

\begin{corolario}
Seja $X\subseteq\Sigma^n$ um conjunto de dados realizado no cubo de Hamming, munido da dist\^ancia de Hamming normalizada. Denotemos $n=\abs X$. Sejam $\ve,\delta>0$. Tiremos aleatoriamente de $[n]$ um subconjunto $I$ com 
\[\abs I\geq \frac{1}{\ve^2}\ln\frac{n}{\delta}=O(\log n)\]
elementos. 
Ent\~ao, com confian\c ca $\geq 1-\delta$, a proje\c c\~ao
\[x\mapsto x\upharpoonright I\]
de $X$ sobre um subcubo aleat\'orio $\Sigma^{I}$
conserva as dist\^ancias entre os elementos de $X$ a menos $\ve$:
\[\forall i,j,~\bar d^{\,\Sigma^d}(x,y) \overset\ve\approx \bar d^{\,\Sigma^I}(x\upharpoonright I,y\upharpoonright I).\]
\end{corolario}

\begin{observacao}
Este g\^enero de resultados foi usado no artigo \citep*{KOR} para construir uma esquema de indexa\c c\~ao eficaz para busca de vizinhos pr\'oximos {\em aproximadamente correta} no cubo de Hamming. No mesmo tempo, o problema da exist\^encia das esquemas da busca {\em exata} (subse\c c\~ao \ref{ss:mdck}) n\~ao est\'a resolvida. A validade da seguinte conjetura ainda est\'a um problema em aberto.
\vskip .3cm

\noindent
{\bf Conjetura de maldi\c c\~ao de dimensionalidade} \citep*{indyk}
{\em Seja $X$ um conjunto de dados com $n$ pontos no cubo de Hamming $\{0,1\}^d$.
Suponha que $d=n^{o(1)}$ e $d=\omega(\log n)$. Ent\~ao cada estrutura de dados para a pesquisa exata de semelhan\c ca em $X$, com o tempo da pesquisa $d^{O(1)}$, deve usar o espa\c co $n^{\omega(1)}$.}
\vskip .3cm

As estruturas de dados e os algoritmos s\~ao entendidos no sentido do {\em modelo de sonda de c\'elula} da computa\c c\~ao ({\em cell probe model})
\citep*{miltersen}.
Os melhores limites conhecidos \citep*{BR,PT} est\~ao longe de resolver o problema. A conjetura foi somente mostrada para algumas estruturas concretas \citep*{pestov2012,pestov:13}. 
\end{observacao}

\section{Teorema de Benedek--Itai}

Como foi mostrado na se\c c\~ao \ref{s:propgeom}, uma classe aprendiz\'avel sob uma medida fixa n\~ao cont\'em fam\'\i lias infinitas uniformemente discretas. De fato, esta propriedade carateriza as classes aprendiz\'aveis. 

Mediante a motiva\c c\~ao, esta propriedade \'e exatamente o que torna a classe $2^{\Omega}$ de todos os subconjuntos de um dom\'\i nio $\Omega$ aprendiz\'avel sob uma medida discreta qualquer.
Dado $\e>0$, existe $k=k(\e)$ tal que a medida total de \'atomos $a_i$, $i>k$, \'e menor ou igual a $\e$. Por conseguinte, o conjunto finito de $2^k$ conceitos que s\~ao subconjuntos de $\{a_1,a_2,\ldots,a_k\}$ forma uma {\em $\e$-rede} em $2^{\Omega}$, com rela\c c\~ao \`a dist\^ancia $d(C,D)=\mu(C\Delta D)$ medidora do erro de generaliza\c c\~ao: qualquer que seja $D\in 2^{\Omega}$, existe $C\subseteq \{a_1,a_2,\ldots,a_k\}$, tal que $\mu(C\Delta D)< \e$. (Basta definir $D=C\cap \{a_1,a_2,\ldots,a_k\}$). Esta propriedade significa que o conjunto $2^{\Omega}$ munido da dist\^ancia $d$ \'e {\em totalmente limitado,} \'e no final das contas, essa propriedade \'e  respons\'avel pela aprendizabilidade da classe $2^{\Omega}$.

Para medidas mais gerais, n\'os j\'a sabemos que as classes finitas s\~ao PAC aprendiz\'aveis. Esta observa\c c\~ao est\'a ao cora\c c\~ao do resultado de Benedek e Itai que d\'a umas condi\c c\~oes necess\'arias e suficientes da aprendizabilidade sob uma medida fixa. Vamos mostrar o resultado nessa se\c c\~ao.

\subsection{N\'umeros de cobertura e de empacotamento\label{s:cobertura}}

Relembramos que uma {\em pseudom\'etrica} sobre um conjunto \'e uma fun\c c\~ao de duas vari\'aveis, $d(x,y)$, que satisfaz todos os axiomas de uma m\'etrica exceto a primeira: pode ser que $d(x,y)=0$ para $x\neq y$.
\index{pseudom\'etrica}
 Um conjunto munido de uma pseudom\'etrica \'e chamado um {\em espa\c co pseudom\'etrico.} 
\index{espa\c co! pseudom\'etrico}
Por exemplo, a rigor, a dist\^ancia $L^1(\mu)$ \'e uma m\'etrica se e somente se a medida $\mu$ \'e puramente at\^omica (exerc\'\i cio).

\begin{exercicio}
Dado um espa\c co pseudom\'etrico, $(X,d)$, mostrar que a rela\c c\~ao
\[x\overset d\sim y\iff d(x,y)=0\]
\'e uma rela\c c\~ao de equival\^encia, e que o conjunto quociente $\tilde X=X/\overset d\sim$ de classes de equival\^encia $[x]$, $x\in X$, admite uma m\'etrica, definida corretamente pela regra
\[\tilde d([x],[y]) = d(x,y).\]
Al\'em disso, mostrar que a m\'etrica $\tilde d$ sobre $\tilde X$ \'e a maior m\'etrica tal que a aplica\c c\~ao quociente,
\[X\ni x\mapsto [x]\in \tilde X,\]
\'e Lipschitz cont\'\i nua com a constante $L=1$.

O espa\c co m\'etrico $(\tilde X,\tilde d)$ \'e chamado {\em o espa\c co m\'etrico associado ao espa\c co pseudom\'etrico} $(X,d)$, ou simplesmente o {\em quociente m\'etrico} de $(X,d)$. 
\label{ex:espacometricoquociente}
\end{exercicio}

Dado um espa\c co probabil\'\i stico padr\~ao, $(\Omega,\mu)$, o s\'\i mbolo $L^1(\Omega,\mu)$ tipicamente significa o quociente m\'etrico do espa\c co de todas as fun\c c\~oes borelianas sobre $\Omega$ munidas da pseudom\'etrica $L^1(\mu)$ tais que $L^1(\abs f)<\infty$. (Veja tamb\'em a subse\c c\~ao \ref{ss:l1mu}.) Vamos ser amb\'\i guos, e na maioria de casos no contexto de aprendizagem vamos trabalhar direitamente com o espa\c co pseudom\'etrico original, sem passar ao espa\c co m\'etrico quociente, embora usando o mesmo s\'\i mbolo.

\begin{definicao}
Sejam $(X,d)$ um espa\c co pseudom\'etrico e $Y\subseteq X$ um subconjunto. Ent\~ao $Y$ \'e dito {\em totalmente limitado} em $X$, ou: {\em pr\'e-compacto} em $X$, se para todo $\ve>0$ existe uma cobertura finita de $Y$ com bolas abertas em $X$ (definidas de mesmo modo que num espa\c co m\'etrico):
\[\exists x_1,x_2,\ldots,x_k\in X,~~Y\subseteq \bigcup_{i=1}^kB_\e(x_i).\]
\index{subconjunto! totalmente limitado}
\index{subconjunto! pr\'e-compacto}
\end{definicao}

(Todas as no\c c\~oes necess\'arias podem ser achadas no ap\^endice \ref{ch:metricos}.)

Esta no\c c\~ao n\~ao deve ser confundida com a no\c c\~ao de um subconjunto {\em relativamente compacto:} $Y$ \'e relativamente compacto em $X$ se e apenas se a ader\^encia de $Y$ em $X$ \'e compacta. 
\index{subconjunto! relativamente compacto}
Cada subconjunto relativamente compacto \'e pr\'e-compacto, mas n\~ao o contr\'ario. As no\c c\~oes de um conjunto totalmente limitado e de um conjunto pr\'e-compacto s\~ao rigorosamente sin\^onimas.

\begin{exercicio}
Mostrar que $Y$ \'e totalmente limitado em $X$ se e somente se $Y$ \'e totalmente limitado em si mesmo.
\end{exercicio}

\begin{definicao} Seja $\ve>0$. Um {\em n\'umero de cobertura} ({\em covering number}), $N_X(\e,Y)$, 
de um subconjunto $Y$ em um espa\c co pseudom\'etrico $X$ \'e o menor tamanho de uma cobertura de $Y$ pelas bolas abertas em $X$:
\[N_X(\ve,Y)=\min\left\{ k\in\N\colon \exists x_1,x_2,\ldots,x_k\in X,~~Y\subseteq \bigcup_{i=1}^kB_\ve(x_i)
\right\}.\]
\index{n\'umero! de cobertura}
\end{definicao}

\'E claro que o n\'umero $B_X(\e,Y)$ \'e finito para todo $\ve>0$ se e somente se $Y$ \'e pr\'e-compacto. No mesmo tempo, os n\'umeros de cobertura s\~ao as carater\'\i sticas relativas e dependem do espa\c co ambiente, $X$.

\begin{exercicio}
Construir um espa\c co (pseudo)m\'etrico $X$ e um subespa\c co $Y$ tais que existe $\e>0$ com
\[N_X(\e,Y)\lneqq N_Y(\e,Y).\]
\end{exercicio}

A pr\'e-compacidade de um espa\c co pseudom\'etrico $Y$ pode ser tamb\'em exprimida em termos de {\em n\'umeros de empacotamento,} que s\~ao absolutas (n\~ao dependem do espa\c co ambiente).

\begin{definicao} Sejam $X$ um espa\c co pseudom\'etrico, $\ve>0$. O {\em n\'umero de empacotamento} ({\em packing number}), $D(\e,X)$, \'e o maior n\'umero de pontos \`a dist\^ancia $\geq\e$ dois a dois:
\[D(\e,X) = \sup\{n\in\N\colon \exists x_1,x_2,\ldots,x_k\in X,~(i\neq j) d(x_i,x_j)\geq \e\}.
\]
\index{n\'umero! de empacotamento}
\end{definicao}

\begin{lema}
Sejam $X$ um espa\c co pseudom\'etrico, $Y\subseteq X$, e $\e>0$. Ent\~ao,
\[N_X(\e,Y)\leq N_Y(\e,Y)\leq D(\e,Y)\leq N_X(\e/2,Y)\leq N_Y(\e/2,Y).\]
\end{lema}

\begin{proof}
Se temos um $\e$-empacotamento m\'aximo de $Y$, $x_1,x_2,\ldots,x_k$, ent\~ao n\~ao podemos adicionar mais um ponto $y$ de modo que as dist\^ancias $d(y,x_i)$ satisfa\c cam $\geq\e$. Isso implica que as bolas abertas de raio $\e$ em torno de pontos $x_i$ formam uma cobertura de $X$.

Agora, seja $B_{\e/2}(x_i)$, $i=1,2,\ldots,k$ uma cobertura de $Y$ pelas bolas abertas em $X$. Seja $y_1,\ldots,y_m$ um subconjunto finito de $Y$ de pontos dois a dois \`a dist\^ancia $\geq \e$. Cada bola $B_{\e/2}(x_i)$ pode conter no m\'aximo um ponto $y_j$, o que significa $m\leq k$.
\end{proof}

\begin{proposicao}
O n\'umero de empacotamento do cubo de Hamming satisfaz
\[D(\e,\Sigma^n)\geq e^{2\left(\frac 12 -\e\right)^2n},\]
quando $\e<1/2$.
\label{p:empacotamento}
\end{proposicao}

\begin{proof} Escolhemos um $\e$-empacotamento m\'aximo, $X$, de $\Sigma^n$. 
A maximalidade implica que as bolas $B_\e(x)$, $x\in X$ cobrem $\Sigma^n$, e por conseguinte
\[1=\mu_{\sharp}(\Sigma^n)\leq\sum_{x\in X}\mu_{\sharp}(B_\e(x))=\sharp(X)\mu_{\sharp}(B_\e(0))\leq \sharp(X)e^{-2\left(\frac 12 -\e\right)^2n},\]
usando a estimativa do volume da bola da subse\c c\~ao \ref{ss:volume}, Eq. (\ref{eq:vol2}).
\end{proof}

\subsection{Teorema de Benedek--Itai: aprendizagem sob uma distribui\c c\~ao fixa}

\begin{teorema}[\citet*{BI}]
Seja $(\Omega,\mu)$ um espa\c co probabil\'\i stico padr\~ao.
Para uma classe de conceitos ${\mathscr C}\subseteq {\mathscr B}_{\Omega}$, as condi\c c\~oes seguintes s\~ao equivalentes.
\begin{itemize}
\item $\mathscr C$ \'e PAC aprendiz\'avel.
\item $\mathscr C$ \'e pr\'e-compacto em rela\c c\~ao \`a pseudom\'etrica $L^1(\mu)$.
\end{itemize}
A complexidade amostral de uma classe $\mathscr C$ que satisfaz uma destas condi\c c\~oes equivalentes verifica
\[\log_2 D(2\e,{\mathscr C})-1\leq s(\e,\delta)\leq \frac{9}{2\ve^2} \ln\frac{2N(\ve/3,{\mathscr C},L^1(\mu))}{\delta},
\]
quando $\delta\leq 1/2$.
\label{t:benedek-itai}
\index{teorema! de Benedek--Itai}
\end{teorema}

\subsubsection{A regra de aprendizagem}
Suponha primeiramente que $\mathscr C$ seja $L^1(\mu)$-pre\-compacto. Dado $\e>0$, escolha uma $\e/3$-rede finita, ou seja, conjuntos borelianos
\[H_1,H_2,\ldots,H_k\]
tais que, qualquer seja $C\in {\mathscr C}$, existe $i=1,2,\ldots,k$ com $\mu(C\Delta H_i)<\e/3$. Os conjuntos $H_i$ (hip\'oteses) n\~ao t\^em que necessariamente pertencer \`a classe $\mathscr C$. Elas formam uma {\em classe de hip\'oteses,} $\mathcal H$.

Seja $C\in {\mathscr C}$ um conceito desconhecido qualquer.
Dada uma amostra rotulada, 
\[\sigma=(x_1,x_2,\ldots,x_n,\e_1,\e_2,\ldots,\e_n),\]
temos a medida emp\'\i rica, $\mu_\sigma$,
\[\mu_\sigma(A)=\frac 1n\sharp\{i\colon x_i\in A\}.\]
O que se sabe, \'e o tra\c co de $C$ sobre $\sigma$:
\[C\cap\supp\sigma= \{x_i\colon \e_i=1\}.\]
Dada uma hip\'otese $H$ qualquer, pode-se calcular o {\em erro emp\'\i rico} de $H$:
\[\mbox{erro}_{C}(H,\mu_{\sigma}) = \mu_{\sigma}(H\Delta C\cap\supp \sigma)=\mu_{\sigma}(C\Delta H)=\frac 1n\sum_{i=1}^n \abs{\chi_C(x_i)-\chi_H(x_i)},\]
pois sabemos calcular os valores $\chi_C(x_i)$ e $\chi_H(x_i)$ para todos elementos $x_i\in\sigma$ da amostra. (Os valores $\chi_C(x_i)$ s\~ao determinados pela dada rotulagem da $\sigma$, e a hip\'otese $H$ \'e conhecida).

A regra de aprendizagem \'e a de {\em minimiza\c c\~ao de erro emp\'\i rico} na classe de hip\'oteses $\mathcal H$. Em outras palavras, buscamos a hip\'otese $H_j$ que minimiza o erro emp\'\i rico com rela\c c\~ao ao conceito $C$:
\[j=\arg\min_{i} \mu_{\sigma}(C\Delta H_i).\]
\index{regra! de minimiza\c c\~ao de erro emp\'\i rico}

Apliquemos o corol\'ario \ref{c:medidas} \`a classe de conjuntos da forma $C\Delta H_i$, $i=1,2,\ldots, k$, com precis\~ao $\ve/3$. Conclu\'\i mos que, com confian\c ca $\geq 1-2ke^{-2(\ve/3)^2n}$, a medida emp\'\i rica de cada um de $k$ conceitos $C\Delta H_i$ difere da sua medida verdadeira por menos de $\e/3$:
\[\mu^{\otimes n}\left\{\exists i=1,2,\ldots,k~\left\vert \mu_{\sigma}(C\Delta H_i)-\mu(C\Delta H_i)\right\vert\geq \frac\e 3
\right\}\leq 2ke^{-2(\ve/3)^2n},
\]
ou seja,
\[\mu^{\otimes n}\left\{\forall i=1,2,\ldots,k~\left\vert \mbox{erro}_{\mu,C}(H_i,\mu_{\sigma})-\mbox{erro}_{\mu,C}(H_i,\mu)\right\vert\leq \frac\e 3
\right\}> 1- 2ke^{-2(\ve/3)^2n}.
\]

O argumento seguinte se aplica com confian\c ca $\geq 1-2ke^{-2(\ve/3)^2n}$ (isto \'e: a probabilidade de que o argumento inteiro seja correto) \'e pelo menos o valor acima.
 
Sabemos que existe $i$ tal que $\mbox{erro}_{\mu,C}(H_i,\mu)<\ve/3$. Isso implica que
\[\mbox{erro}_{\mu,C}(H_i,\mu_{\sigma})<\frac{\ve}3+\frac{\ve}3=\frac{2\ve}3.\]
 A hip\'otese $H=H_j$ escolhida pela nossa regra de minimiza\c c\~ao do erro emp\'\i rico pode ser diferente de $H_i$, mas ela satisfaz tudo mesmo 
\[\mbox{erro}_{\mu,C}(H_j,\mu_{\sigma})\leq \mbox{erro}_{\mu,C}(H_i,\mu_{\sigma})=\frac{2\ve}3,\]
e por conseguinte
\[\mbox{erro}_{\mu,C}(H_j,\mu)<\frac{2\ve}3+\frac{\ve}3=\ve.\]
Ent\~ao, com confian\c ca $\geq 1-2ke^{-2(\ve/3)^2n}$, a nossa regra aprende o conceito $C$ com precis\~ao $\e$. 

Resta estimar a complexidade amostral: a desigualdade desejada
\[\delta\leq 2ke^{-2(\ve/3)^2n}\]
transforma-se em 
\[\ln\delta \leq \ln(2k) - \frac{2\ve^2}{9}n,\]
ou seja,
\[n\geq \frac{9}{2\ve^2} \ln\frac{2k}{\delta}.
\]

\subsubsection{Necessidade de pr\'e-compacidade da classe} Agora suponhamos que a classe $\mathscr C$ seja aprendiz\'avel. Denotemos $\mathcal L$ uma regra de aprendizagem que aprende $\mathscr C$. 
Lema \ref{l:2ediscrete} significa que todo $2\ve$-empacotamento de $\mathscr C$ \'e finito, em particular $\mathscr C$ \'e pr\'e-compacto, com
\[D(2\e,{\mathscr C})\leq 2^{n+1},\]
e a complexidade amostral de aprendizagem de $\mathscr C$ satisfaz
\[s(\e,\delta)\geq \log_2 D(2\e,{\mathscr C})-1,\]
quando $\delta\leq 1/2$.

\begin{observacao} 
Note como pouco sens\'\i vel \'e a depend\^encia da complexidade amostral sobre $\delta$. Diz-se que ``confian\c ca \'e barata'' (``confidence is cheap'').
\end{observacao}

\subsection{Regras consistentes}

\begin{definicao}
Uma regra de aprendizagem $\mathcal L$ \'e dita {\em consistente com uma classe de conceitos}, $\mathscr C$, se
\begin{itemize}
\item todas as hip\'oteses produzidas por $\mathcal L$ pertencem a $\mathscr C$, 
\[\forall n,~\forall \sigma\in\Omega^n\times\{0,1\}^n,~{\mathcal L}(\sigma)\in{\mathscr C},\]
\item a hip\'otese induz sobre a amostra a rotulagem original sempre que poss\'\i vel:
\[\forall C\in {\mathscr C},~\forall n,~\forall \sigma\in\Omega^n,~
{\mathcal L}(C\upharpoonright \sigma)\upharpoonright \sigma = C\upharpoonright \sigma.\]
\end{itemize}
\index{regra! consistente com uma classe}
\end{definicao}

\begin{definicao}
Uma classe de conceitos $\mathscr C$ \'e {\em consistentemente aprendiz\'avel} se cada regra $\mathcal L$ consistente com $\mathscr C$ PAC aprende $\mathscr C$.
\index{classe! consistentemente aprendiz\'avel}
\end{definicao}

\begin{observacao}
Se uma classe $\mathscr C$ \'e consistentemente aprendiz\'avel, ent\~ao ela \'e aprendiz\'avel. Basta notar que cada classe de conceitos, $\mathscr C$, admite pelo menos uma regra de aprendizagem consistente com ela. Intuitivamente, isso \'e claro: dada uma amostra rotulada cuja rotulagem \'e produzida por um elemento de $\mathscr C$, escolha a hip\'otese entre todos os conceitos que produz a mesma rotulagem. O que n\~ao \'e totalmente \'obvio, \'e a possibilidade de fazer a escolha de hip\'oteses de maneira que a aplica\c c\~ao resultante,
\[{\mathcal L}_n\colon\Omega^n\times\{0,1\}^n\times\Omega\to \{0,1\},\]
seja boreliana. Vamos adiar a discuss\~ao para mais tarde (se\c c\~ao \ref{s:cuidado}), para n\~ao nos perdermos em tecnicalidades prematuramente.
\end{observacao}

\begin{observacao} 
Como vimos na se\c c\~ao \ref{s:atomica}, cada classe de conceitos \'e consistentemente aprendiz\'avel sob uma medida discreta.
\end{observacao}

No mesmo tempo, nem toda classe PAC aprendiz\'avel \'e consistentemente aprendiz\'avel. Aqui um exemplo cl\'assico. 

\begin{exemplo}
Seja $\Omega=[0,1]$, o intervalo fechado, munido da medida de Lebesgue, $\lambda$, ou seja, a distribui\c c\~ao uniforme. A classe de conceitos, $\mathscr C$, consiste de todos os subconjuntos finitos e cofinitos do intervalo. (Um subconjunto $A\subseteq\Omega$ \'e {\em cofinito} se $\Omega\setminus A$ \'e finito).
O espa\c co m\'etrico associado ao espa\c co pseudom\'etrico $\mathscr C$, munido da dist\^ancia $\lambda(C\Delta D)$, consiste de dois pontos: a classe de equival\^encia do conceito vazio e a do intervalo, a dist\^ancia $1$ um de outro. Em outras palavras, \'e o espa\c co m\'etrico $\{0,1\}$ com a dist\^ancia usual. Segundo o teorema de Benedek-Itai, esta classe \'e PAC aprendiz\'avel.

No mesmo tempo, existe uma regra de aprendizagem consistente cujas todas hip\'oteses s\~ao conceitos finitos. Por exemplo, dada uma amostra rotulada \[\sigma=(x_1,x_2,\ldots,x_n,\e_1,\e_2,\ldots,\e_n),\] pode-se definir a hip\'otese
\[H={\mathcal L}_n(\sigma) = \{x_i\colon \e_i=1,~i=1,2,\ldots,n\}.\]
Esta hip\'otese, $H$, pertence \`a classe $\mathscr C$ e induz a rotulagem original sobre $\{x_1,x_2,\ldots,x_n\}$. No mesmo tempo, a regra $\mathcal L$ nunca vai aprender o conceito $[0,1]$: qualquer que seja a hip\'otese $H$ gerada pela regra, o erro de aprendizagem \'e sempre p\'essimo:
\[\mbox{erro}_{[0,1]}(H)=\lambda(H\Delta [0,1])=\lambda([0,1]\setminus H)=1.\]
Conclu\'\i mos: $\mathscr C$ n\~ao \'e consistentemente aprendiz\'avel.
\label{ex:precnaocons}
\end{exemplo}

Quando uma classe \'e consistentemente aprendiz\'avel sob uma medida fixa? Mesmo que n\~ao haja crit\'erio conhecido para esta condi\c c\~ao, tem a condi\c c\~ao suficiente. Uma classe de Glivenko--Cantelli, $\mathscr C$, \'e caraterizado pela condi\c c\~ao seguinte: com alta confian\c ca, a medida emp\' \i rica de todo elemento de $\mathscr C$ aproxima bem a medida de verdade. Verifica-se que as classes de Glivenko--Cantelli s\~ao consistentemente aprendiz\'aveis, e a condi\c c\~ao de ser Glivenko-Cantelli exprima-se na linguagem de fragmenta\c c\~ao. As classes de Glivenko--Cantelli formam o assunto do pr\'oximo  cap\'\i tulo. 

\section{Taxa de aprendizagem}

A taxa de aprendizagem de uma classe $\mathscr C$ sobre uma medida $\mu$ \'e o m\'\i nimo tamanho de dados, $n$, necess\'ario para aprender $\mathscr C$ a uma precis\~ao e com uma confian\c ca desejadas. Em outras palavras, \'e a taxa de crescimento da fun\c c\~ao $s(\e,\delta)$ da complexidade amostral. Como a depend\^encia de $\delta$ n\~ao \'e muito sens\'\i vel, usualmente a depend\^encia de $\e$ \'e o que \'e de interesse. Chamamos a {\em taxa de aprendizagem} a fun\c c\~ao $\e\mapsto s(\e,\delta_0)$ com o valor do risco $\delta_0>0$ fixo, imaginando alguma coisa como $\delta_0=0,05$ ou $\delta_0=0,01$.
\index{taxa! de aprendizagem}

Mesmo se cada classe de conceitos \'e PAC aprendiz\'avel sobre cada medida discreta, a taxa de aprendizagem depende da medida e pode ser qualquer, de fato t\~ao lenta quanto desejado.

\subsection{O resultado}

\begin{teorema}[\cite{pestov2010sbrn}]
Seja $\mathscr C$ uma classe de conceitos borelianos num dom\'\i nio boreliano padr\~ao $\Omega$, com a propriedade seguinte: existe um conjunto infinito $A$ tal que todos subconjuntos finitos de $A$ s\~ao fragmentados por $\mathscr C$.
Sejam $(\ve_k)$ e $(f_k)$ duas sequ\^encias de reais positivos tais que $f_i\uparrow+\infty$ e $\ve_i\downarrow 0$, de modo que
\[\frac 17>\ve_1>\ve_2>\ldots >\ve_i > \ldots.\]
Seja $0<\delta_0< 1$ qualquer fixo. Ent\~ao
existe uma medida de probabilidade discreta, $\mu$, sobre $\Omega$, tal que cada regra de aprendizagem consistente com a classe ${\mathscr C}$ exige uma amostra aleat\'oria de tamanho $n\geq f_k$ para aprender ${\mathscr C}$ com precis\~ao $\ve_k$ e confian\c ca constante $1-\delta_0$, qualquer que seja $i$:
\[s_{\mathscr C}(\ve_k,\delta_0,\mu)\geq f_k.\]
\label{t:taxa}
\end{teorema}

Isso significa que a complexidade amostral de aprendizagem pode crescer exponencialmente, ou como
\[2,2^2, 2^{2^2},\ldots, 2^{2^{2^{2^{\Ddots}}}},\ldots,\]
ou mesmo mais r\'apido do que isso, por exemplo, como uma fun\c c\~ao n\~ao recursivamente comput\'avel. O fato de ser aprendiz\'avel com uma taxa de crescimento igual n\~ao \'e particularmente informativo. Usualmente na ci\^encia de computa\c c\~ao s\'o as taxas de crescimento polinomiais s\~ao consideradas desejadas. 

\begin{observacao}
Seja $\mu$ uma medida uniforme sobre um conjunto finito, $X$, com $n$ elementos, que tem a massa total $\gamma>0$. Isso significa $\mu\{x\}=\gamma/n$ para todo $x\in X$. Ent\~ao a dist\^ancia $L^1(\mu)$ sobre o espa\c co $2^X$ de todos os subconjuntos de $X$ \'e igual \`a dist\^ancia de Hamming normalizada a menos um fator de $\gamma$:
\[d(C,D) = \norm{\chi_C-\chi_D}_{L^1(\mu)} = \int_X \abs{\chi_C-\chi_D}d\mu 
= \mu(C\Delta D) = \gamma \bar d(\chi_C,\chi_D).\]
\end{observacao}

\begin{proof}[Prova do teorema \ref{t:taxa}]
Gra\c cas ao teorema \ref{t:benedek-itai} de Benedek-Itai, basta construir a medida discreta $\mu$ de modo que para cada $k$ existe uma fam\'\i lia de $\geq 2^{f_k}$ conceitos dois a dois \`a dist\^ancia $\geq 2\ve_k$ um de outro. 
Substituindo $f_k$ por $\max\{f_1,f_2,\ldots,f_k\}$ se for necess\'ario, podemos supor que $(f_k)$ cresce monotonicamente.
Denotemos $\gamma_0=1-7\ve_1>0$ e para todo $k\geq 1$,
\[\gamma_k = 7(\ve_{k}-\ve_{k+1}).\]
Ent\~ao, $\gamma_k>0$, $k\in\N$, e $\sum_{k=0}^{\infty}\gamma_k=1$. 

Tendo em vista a proposi\c c\^ao \ref{p:empacotamento}, escolhemos
para cada $k\geq 1$ um n\'umero natural $m_k$ de mode que 
\[e^{2\left(\frac 12-\frac{1}3\right)^2m_k}\geq 2^{f_k},\]
ou seja, $m_k\geq (18\ln 2) f_k$. Posemos $m_0=1$. Agora, para todo $k\in\N$, escolhemos uma fam\'\i lia de elementos $\sigma^k_i\in \Sigma^{m_k}$, $i=1,2,\ldots,f_k$, com $\bar d(\sigma_i,\sigma_j)\geq 1/3$ se $i\neq j$.

Escolhemos um subconjunto infinito enumer\'avel, $A$, de $\Omega$, cujos subconjuntos finitos s\~ao todos fragmentados pela classe $\mathscr C$. Dividimos $A$ em subconjuntos $A_k$, $k\in\N$, dois a dois disjuntos e finitos, $\sharp A_k=m_k$. Finalmente, definamos $\mu$ pela condi\c c\~ao: se $a\in A_k$, ent\~ao
\[\mu\{a\}=\frac{\gamma_k}{m_k}.\]
\'E claro que $\mu(A_k)=\gamma_k$, $k\in\N$, e $\mu$ \'e uma medida de probabilidade discreta sobre $\Omega$. 

Seja $k\geq 1$ qualquer. Como $\sum_{i=k}^{\infty}\gamma_i =7\ve_k$,
existe $N>k$ tal que $\sum_{i=k}^N\gamma_i \geq 6\ve_k$. Segundo a escolha de $A$, existem os conceitos $C_1,C_2,\ldots,C_{f_k}\in {\mathscr C}$ tais que, quaisquer que sejam $i=k,k+1,\ldots,N$ e $j=1,2,\ldots,f_k$, temos
\[C_j\upharpoonright A_i = \sigma^i_j.\]
Se $j,j^{\prime}=1,2,\ldots,f_k$, $j\neq j^\prime$, ent\~ao
\begin{eqnarray*}
\mu(C_j\Delta C_{j^\prime}) &=& \sum_{i=k}^N \mu((C_j\cap A_i)\Delta (C_{j^\prime}\cap A_{j^\prime})) \\
&=& \sum_{i=k}^N \gamma_i \bar d(\sigma^i_j,\sigma^i_{j^\prime}) \\
&\geq & \sum_{i=k}^N \frac 13 \gamma_i \\
&>& \frac 13\cdot 6\ve_k  \\
&=& 2\ve_k.
\end{eqnarray*}
\end{proof}

\subsection{Consequ\^encias para medidas n\~ao at\^omicas\label{ss:difusas}}
Modifiquemos o resultado acima a fim de construir um exemplo de uma classe de conceitos, $\mathscr C$, que s\~ao subconjuntos de $\Omega=\R$, com as propriedades seguintes:

\begin{itemize}
\item Qualquer que seja uma medida de probabilidade boreliana sobre $\Omega$, 
 a classe n\~ao \'e trivial em rela\c c\~ao \`a dist\^ancia $L^1(\mu)$;
\item $\mathscr C$ \'e PAC aprendiz\'avel sob toda medida de probabilidade boreliana sobre $\Omega$, e
\item a taxa de aprendizagem pode ser t\~ao lenta quanto se queira, incluindo sob medidas difusas.
\end{itemize}

A classe $\mathscr C$ consiste de todas as uni\~oes de intervalos semiabertos da forma $[n,n+1)$, $n\in\Z$. Dado $I\subseteq\Z$, denotemos
\[C_I=\bigcup_{n\in I}[n,n+1).\]
Desse modo, existe uma bije\c c\~ao natural entre os elementos de $\mathscr C$ e os de $2^{\Z}$, dada por
\[\phi(C) = C\cap\Z.\]
Se $\mu$ \'e uma medida de probabilidade sobre $\Omega$, ent\~ao $\phi_{\ast}$ \'e a sua imagem direta, uma medida de probabilidade sobre $\Z$ dada por
\[\phi_{\ast}(\mu)\{k\}=\mu([k,k+1)).\]
Dada uma regra de aprendizagem, $\mathscr L$, no dom\'\i nio $\Z$, pode se definir a regra para $\Omega$, que denotaremos por ${\mathscr L}^{\phi}$. Ela \'e dada por
\[{\mathscr L}^{\phi}_n(\sigma)=\phi^{-1}({\mathscr L}_n(\phi(\sigma))).\]

\begin{exercicio}
Se a regra $\mathscr L$ PAC aprende a classe $2^{\Z}$ sob a medida $\phi_{\ast}(\mu)$, ent\~ao ${\mathscr L}^{\phi}$ aprende $\mathscr C$ sob a medida $\mu$, com a mesma complexidade amostral. 
\end{exercicio}

Dada uma regra de aprendizagem $\mathscr L$ sobre o dom\'\i nio $\Omega$, pode se definir uma regra para $\Z$, usando a imers\~ao can\^onica, $i$, de $\Z$ dentro $\R$:
\[{\mathscr L}^i(\sigma) = {\mathscr L}(\sigma)\cap\Z.\]

\begin{exercicio}
Se a regra $\mathscr L$ PAC aprende a classe $\mathscr C$ sob a medida $\mu$, ent\~ao ${\mathscr L}^{i}$ aprende $2^{\Z}$ sob a medida $\phi_{\ast}(\mu)$, com a  complexidade amostral menor ou igual \`a complexidade de $\mathscr L$.
\end{exercicio}

Juntos, os dois exerc\'\i cios estabelecem a validade da afirma\c c\~ao desejada.

\subsection{Consequ\^encias para o classificador $1$-NN\label{ss:1-nn}}
O classificador de vizinho mais pr\'oximo ($1$-NN) pode ser definido em todo o dom\'\i nio munido da uma ``medida de se\-me\-lhan\-\c ca'' qualquer (n\~ao necessariamente uma m\'etrica). Mais geralmente, $\Omega$ deve admitir uma fam\'\i lia de pr\'e-ordens, $\prec_x$, uma para cada ponto $x\in\Omega$, tais que $x$ \'e o ponto m\'\i nimo desta pr\'e-ordem e a condi\c c\~ao $y\prec_x z$ \'e interpretado como significando que $y$ \'e mais perto de $x$ que $z$. As pr\'e-ordens $\prec_x$ devem satisfazer certas condi\c c\~oes de mensurabilidade. 
No caso de mais de um vizinho mais pr\'oximo de $x$, vamos desempatar escolhendo o vizinho $x_i$, seja aleatoriamente, seja com o \'\i ndice $i$ menor.
Pelo momento, uma teoria falta nessa generalidade, e apenas existe para espa\c cos m\'etricos.

Vamos tratar o caso que parece at\'e ser trivial, o da m\'etrica zero-um, que so toma dois valores:
\[d(x,y)=\begin{cases} 0,&\mbox{ se }x=y, \\
1,&\mbox{ se }x\neq y.
\end{cases}\]
A estrutura boreliana induzida pela tal m\'etrica contem todos os subconjuntos do dom\'inio, que por conseguinte deve ser enumer\'avel. Conclu\'\i mos que toda medida de probabilidade sobre $\Omega$ \'e discreta. Dado um conjunto finito dos \'atomos, $a_1,a_2,\ldots,a_k$, se $n$ for bastante grande, com alta confian\c ca todos $a_j$ v\~ao aparecer entre os elementos da amostra aleat\'oria, $x_1,x_2,\ldots,x_n$. Se $x_i=a_j$, o seu vizinho mais pr\'oximo \'e $a_j$ mesmo, e o r\'otulo de $a_j$ vai ser escolhido para $x_i$ pela regra $1$-NN. Esse argumento estabelece que a regra $1$-NN neste contexto 
PAC aprende a classe $2^{\Omega}$. No mesmo tempo, o teorema \ref{t:taxa} implica que a taxa de aprendizagem pode ser t\~ao lenta quanto se queira, para uma medida $\mu$ apropriada. Esta conclus\~ao vale para caso geral de classificadores $k$-NN nos dom\'\i nios quaisquer.

Mesmo se nessa situa\c c\~ao o classificador $1$-NN pode aparecer como uma regra de aprendizagem dentro uma classe, na verdade, ele \'e de uma natureza  diferente. Vamos estud\'a-lo, dentro de um modelo pr\'oprio de aprendizagem, no cap\'\i tulo \ref{c:kNN}.

%
%

\chapter{Classes de Glivenko-Cantelli\label{ch:GC}}

A no\c c\~ao seguinte est\'a uma das mais importantes na teoria de aprendizagem estat\'\i stica supervisionada.

\begin{definicao}
Uma classe de conceitos, $\mathscr C$, sobre um dom\'\i nio boreliano padr\~ao $\Omega$ \'e uma {\em classe de Glivenko--Cantelli} sob a medida de probabilidade $\mu$, ou tem a propriedade de {\em converg\^encia uniforme de medidas emp\'\i ricas} ({\em uniform convergence of empirical means}, {\em UCEM} property) sob $\mu$, se $\mathscr C$ possui a propriedade seguinte.
Para quaisquer que sejam $\e>0$ e $\delta>0$, existe o valor $s=s(\e,\delta)$ (a complexidade amostral) tal que, se $n\geq s(\e,\delta)$, ent\~ao, com confian\c ca $>1-\delta$ a medida emp\'\i rica $\e$-aproxima a medida verdadeira para todo conceito $C\in {\mathscr C}$: 
\begin{equation}
\label{eq:glivenko}
\mu^{\otimes n}\{\sigma\in\Omega^n\colon \forall C\in{\mathscr C},~\mu_{\sigma}(C)\overset\e\approx \mu(C)\}>1-\delta.
\end{equation}
De maneira equivalente,
\begin{equation}
\label{eq:glivenko2}
\E_{\sigma\sim\mu}\sup_{C\in {\mathscr C}}\abs{\mu_\sigma(C)-\mu(C)}\to 0\mbox{ quando }n\to\infty.
\end{equation}
\index{classe! de Glivenko--Cantelli}
\index{converg\^encia! uniforme de medidas emp\'\i ricas}
\end{definicao}

Por exemplo, segundo teorema \ref{t:aproximacao}, toda classe finita \'e uma classe de Glivenko--Cantelli sob qualquer medida. Segue-se facilmente dos resultados da Subse\c c\~ao \ref{ss:enumeravel} que a classe $2^{\Omega}$ de todos os conceitos \'e uma classe de Glivenko--Cantelli sob uma medida discreta.

De fato, esta no\c c\~ao est\'a bem definida para classes de {\em fun\c c\~oes,} e a maioria dos resultados tem an\'alogos. Mas n\'os vamos nos preocupar principalmente com as classes de conceitos. Neste cap\'\i tulo vamos estudar as propriedades de classes de Glivenko--Cantelli, as suas v\'arias carateriza\c c\~oes, e o seu papel na aprendizagem de m\'aquina.

\section{Complexidades de Rademacher}

Vamos brincar com a defini\c c\~ao de uma classe G--C, reformulando-a em uma forma equivalente que pode ser interpretada na linguagem da aprendizagem estat\'\i stica.

\subsection{Simetriza\c c\~ao com sinais}
Seja $\mathscr C$ uma classe de Glivenko--Cantelli, onde vamos adotar a defini\c c\~ao na eq. (\ref{eq:glivenko2}). Seja $\ve>0$, e suponha que $n$ \'e t\~ao grande que 
\[\E_{\sigma\sim\mu}\sup_{C\in {\mathscr C}}\abs{\mu_\sigma(C)-\mu(C)}<\ve.\]
Seja $\sigma^\prime=(X^\prime_1,X^\prime_2,\ldots,X^\prime_n)$ uma amostra independente de $\sigma=(X_1,X_2,\ldots,X_n)$, o que significa que a lei do par $(\sigma,\sigma^\prime)$ \'e a medida de produto $\mu^{\otimes n}\otimes \mu^{\otimes n}$ sobre $\Omega^n\times\Omega^n$. Usando a desigualdade triangular junto com as propriedades b\'asicas do supremo e esperan\c ca, temos
\begin{align*}
\E_{\sigma,\sigma^\prime\sim\mu}\sup_{C\in {\mathscr C}}\abs{\mu_\sigma(C)&-\mu_{\sigma^\prime}(C)} \\
& = \E_{\sigma,\sigma^\prime\sim\mu}\sup_{C\in {\mathscr C}}\abs{\mu_\sigma(C)-\mu(C)+\mu(C)-\mu_{\sigma^\prime}(C)}
\\
&\leq\E_{\sigma,\sigma^\prime\sim\mu}\sup_{C\in {\mathscr C}}\abs{\mu_\sigma(C)-\mu(C)}+
\E_{\sigma,\sigma^\prime\sim\mu}\sup_{C\in {\mathscr C}}\abs{\mu_{\sigma^\prime}(C)-\mu(C)} \\
&= \E_{\sigma\sim\mu}\sup_{C\in {\mathscr C}}\abs{\mu_\sigma(C)-\mu(C)}+
\E_{\sigma^\prime\sim\mu}\sup_{C\in {\mathscr C}}\abs{\mu_{\sigma^\prime}(C)-\mu(C)} \\
&< 2\ve.
\end{align*}

Cada permuta\c c\~ao $\tau$ de coordenadas de $\Omega^{2n}$, ou seja, um elemento $\tau\in S_{2n}$ do grupo sim\'etrico de posto $2n$, define um automorfismo boreliano de $\Omega^{2n}$,
\[(x_1,x_2,\ldots,x_{2n})\mapsto (x_{\tau(1)},x_{\tau(2)},\ldots,x_{\tau(2n)}),\]
que conserva a medida de produto $\mu^{\otimes 2n}$. Em outras palavras, qualquer que seja um subconjunto boreliano $B\subseteq\Omega^{2n}$, temos
\[\mu(\tau(B))=\mu(B).\]
(Para mostrar este fato, notemos que a medida de conjuntos retangulares \'e obviamente conservada pelas permuta\c c\~oes de coordenadas,
\[\mu(B_1\times\ldots B_{2n}) = \prod_{i}\mu(B_i) = \mu(B_{\tau(1)}\times\ldots B_{\tau(2n)}),\]
e segundo o teorema de Carath\'eodory \ref{t:extensao} sobre a extens\~ao de medidas, os valores de medida de conjuntos retangulares definem unicamente a medida de produto).

Em particular, a conclus\~ao acima se aplica \`a transposi\c c\~ao, $\tau_i$, de $i$-\'esimas coordenadas de amostras $\sigma=(x_1,x_2,\ldots,x_n)$ e $\sigma^\prime=(x_1^\prime,x_2^\prime,\ldots,x_n^\prime)$:
\[\tau_i:  i\leftrightarrow n+i,~~1\leq i\leq n.\]
Conclu\'\i mos que a esperan\c ca da vari\'avel aleat\'oria
\[f(\sigma,\sigma^\prime)=\sup_{C\in {\mathscr C}}\abs{\mu_\sigma(C)-\mu_{\sigma^\prime}(C)}\]
\'e igual \`a esperan\c ca da composi\c c\~ao $f\circ\tau_i$. 
Escrevemos $f$ assim:
\begin{align*}
f(\sigma,\sigma^\prime)&=\sup_{C\in {\mathscr C}}\left\vert
\frac 1n\sum_{i=1}^n \chi_C(x_i)-\frac 1n\sum_{i=1}^n \chi_C(x^\prime_i)\right\vert
\\
&=\sup_{C\in {\mathscr C}}\left\vert
\frac 1n\sum_{i=1}^n \left(\chi_C(x_i)- \chi_C(x^\prime_i)\right)\right\vert.
\end{align*}
O efeito de aplicar a transposi\c c\~ao $\tau_i$ \'e de substituir o termo  $\chi_C(x_i)-\chi_C(x^\prime_i)$ pelo termo $\chi_C(x^\prime_i)-\chi_C(x_i)$, ou seja, multiplic\'a-lo por $-1$. A composi\c c\~ao $\tau$ de um n\'umero finito de transposi\c c\~oes desse tipo resulta em express\~ao seguinte:
\[(f\circ\tau)(\sigma,\sigma^\prime)=\sup_{C\in {\mathscr C}}\left\vert
\frac 1n\sum_{i=1}^n \eta_i\left(\chi_C(x_i)- \chi_C(x^\prime_i)\right)\right\vert,\]
onde $\eta = (\eta_1,\eta_2,\ldots,\eta_n)\in\{-1,1\}^n$ \'e uma sequ\^encia de sinais $\pm 1$. O conjunto $\{\pm 1\}^n$ chama-se o {\em cubo de Rademacher}, um parente pr\'oximo do cubo de Hamming. 
\index{cubo! de Rademacher}

Resumimos: qualquer que seja $\eta\in \{\pm 1\}^n$, 
\begin{align}
\E_{\sigma,\sigma^\prime\sim\mu}\sup_{C\in {\mathscr C}}\left\vert
\frac 1n\sum_{i=1}^n \eta_i \chi_C(x_i)- \frac 1n\sum_{i=1}^n \eta_i 
\chi_C(x^\prime_i)\right\vert &= 
\E_{\sigma,\sigma^\prime\sim\mu}(f\circ\tau)(\sigma,\sigma^\prime)\nonumber \\
&= \E_{\sigma,\sigma^\prime\sim\mu}f(\sigma,\sigma^\prime) \nonumber\\
&=
\E_{\sigma,\sigma^\prime\sim\mu}\sup_{C\in {\mathscr C}}\abs{\mu_\sigma(C)-\mu_{\sigma^\prime}(C)}\nonumber \\
&< 2\e. \label{eq:symmm}
\end{align}

\subsection{Cubo de Rademacher}
Se pensarmos em $\{\pm 1\}^n$ como formado pelas fun\c c\~oes de $[n]$ para $\{1,-1\}\subseteq\R$, ent\~ao a dist\^ancia (normalizada) no cubo de Rademacher seria a dist\^ancia $L^1(\mu_{\sharp})$ relativa \`a medida de contagem normalizada sobre $[n]$:
\[d(\eta,\eta^\prime)=\frac 1n\sum_{i=1}^n\abs{\eta_i-\eta^\prime_i}=\frac 2n\sharp\{i\colon \eta_i\neq \eta^\prime_i\}.\]
A medida sobre $\{\pm 1\}^n$ \'e a medida normalizada de contagem.
A aplica\c c\~ao $i\colon \eta\mapsto \frac 12(\eta+1)$ de $\{-1,+1\}^n$ para $\{0,1\}^n$ conserva a medida de contagem normalizada, e multiplica as dist\^ancias pelo fator constante $1/2$. Em particular, a imagem de uma esfera (bola) de raio $r$ \'e uma esfera (bola) de raio $r/2$ (os valores da dist\^ancia normalizada no cubo de Rademacher s\~ao os m\'ultiplos de $2/n$). Todas as no\c c\~oes e constru\c c\~oes v\'alidas para o cubo de Hamming tem sentido para o cubo de Rademacher, tais que, entre muitas outras, o peso normalizado:
\[\bar w(\eta) =\bar w_{\{\pm 1\}^n}(\eta) =\frac 1n\sum_{i=1}^n\eta_i.\]

\begin{lema} No cubo de Rademacher,
\[\E_\eta\abs{\bar w_{\{\pm 1\}^n}(\eta)}\leq \frac {\sqrt{2\pi}}{\sqrt n}.\]
\label{l:cota}
\end{lema}

\begin{proof} 
Passemos ao cubo de Hamming:

\begin{align*}
\E_{\eta}\left\vert\bar w_{\{\pm 1\}^n}\right\vert 
&= \int_{\{\pm 1\}^n}\abs{\bar w_{\{\pm 1\}^n}(\eta)}d\mu_{\sharp}(\eta) \\
&=
2\int_{\{0,1\}^n}\abs{w_{\{0,1\}^n(\sigma)}-1/2}d\mu_{\sharp}(\sigma) \\
&= 2\sum_{\sigma\in\{0,1\}^n}\left\vert\bar w(\sigma)-1/2\right\vert\cdot\frac 1{2^n}.
\end{align*}
Agrupamos os termos da soma, usando o fato que $\bar w(\sigma) = \bar d(0,\sigma)$ e por conseguinte o peso \'e constante em cada esfera de Hamming,
\begin{align*}
\sum_{\sigma\in\{0,1\}^n}\left\vert\bar w(\sigma)-1/2\right\vert\cdot\frac 1{2^n} &=
\sum_{i=0}^n \left\vert \frac in-\frac 12\right\vert \mu_{\sharp}S_{i/n}(\bar 0)
\\
&= 2\sum_{i=0}^m \left( \frac 12-\frac in\right) \mu_{\sharp}S_{i/n}(\bar 0)\\
&= \frac 2n \sum_{i=0}^m \left( \frac n2-i\right) \mu_{\sharp}S_{i/n}(\bar 0)
\end{align*}
onde $m=\lceil n/2\rceil-1$. (Usamos a simetria de termos sob a transposi\c c\~ao $i\leftrightarrow n-i$). 

Para $n=2m$ par, 
\begin{align*}
\frac 1n \sum_{i=0}^m \left( \frac n2-i\right) \mu_{\sharp}S_{i/n}(\bar 0)
&= \frac 1n \sum_{i=0}^{m-1} \mu_{\sharp}B_{i/n}(\bar 0),
\end{align*}
e agora usamos a estimativa do volume da bola da subse\c c\~ao \ref{ss:volume} e depois fazemos a substitui\c c\~ao $j=m-i$:
\begin{align*}
\E_{\eta}\left\vert\bar w_{\{\pm 1\}^n}\right\vert 
&< \frac 4n \sum_{i=0}^{m-1}  e^{-2(1/2-i/n)^2n}\\
&= \frac 4n \sum_{j=1}^m e^{-2(j/n)^2n}\\
&< 4 \int_{0}^{1/2} e^{-2x^2n}dx\\
&\leq  \frac {4}{\sqrt{2n}}\int_0^{+\infty} e^{-(\sqrt 2 x\sqrt{n})^2} d(\sqrt 2 x\sqrt n) \\
&= \frac {\sqrt{2\pi}}{\sqrt{n}}.
\end{align*}

Para $n=2m+1$ \'\i mpar, $n/2=m+1/2$, e temos
\begin{eqnarray*}
\frac 1n \sum_{i=0}^m \left( \frac n2-i\right) \mu_{\sharp}S_{i/n}(\bar 0)
&=& \frac 1n \sum_{i=0}^m \left( m-i+\frac 12\right) \mu_{\sharp}S_{i/n}(\bar 0) \\
&=& \frac 1n \sum_{i=0}^{m-1} \mu_{\sharp}B_{i/n}(\bar 0)+\frac 1{2n} \mu_{\sharp}(B_{m/n}(0))\\
&<& \frac 1n \sum_{j=1}^m e^{-2(1/2n+j/n)^2n} + \frac 1{2n} e^{-2(1/2n)^2n}
\\
&<& \int_{0}^{\infty} e^{-2x^2n}dx,
\end{eqnarray*}
etc.
\end{proof}

O argumento acima pode ser generalizado como segue.

\begin{exercicio} Seja $X=(X,d,\mu)$ um espa\c co m\'etrico com medida de probabilidade boreliana, tendo a fun\c c\~ao de concentra\c c\~ao $\alpha_X$. Mostrar que, qualquer que seja uma fun\c c\~ao Lipschitz cont\'\i nua, $f$, com a constante de Lipschitz um, temos
\[\abs{M(f)-\E(f)} \leq \int_0^\infty \alpha_X(\e)\,d\e.\]
\end{exercicio} 

(Para uma prova, veja lema \ref{l:mm}). O lema \ref{l:cota}, com constantes um pouco piores, segue-se em aplicando o exerc\'\i cio \`as fun\c c\~oes $\bar w_+ =\max\{\bar w,0\}$ e $\bar w_-=\max\{-\bar w,0\}$ separadamente (cada uma delas tem $0$ com valor mediano).

\begin{exercicio} Deduzir que, se $X_n=(X_n,d_n,\mu_n)$, $n\in\N_+$ \'e uma fam\'\i lia de L\'evy de espa\c cos m\'etricos com medida de probabilidade, e se $f_n\colon X_n\to\R$ s\~ao fun\c c\~oes Lipschitz cont\'\i nuas com a constante de Lipschitz um, ent\~ao
\[\abs{M(f_n)-\E(f_n)}\to 0\mbox{ quando }n\to\infty.\]
(Para uma solu\c c\~ao, consulte a prova do lema \ref{l:mm}.)
\label{e:medianomedio}
\end{exercicio}

\begin{exercicio}
Seja ${\mathscr X}=(X_n,d_n,\mu_n)_{n=1}^\infty$ uma fam\'\i lia de L\'evy normal. Mostrar que existe uma constante $C_{\mathscr X}>0$ tal que, para todo $n$ e toda fun\c c\~ao Lipschitz cont\'\i nua com a constante um, $f_n\colon X_n\to\R$,
\[\abs{M(f_n)-\E(f_n)}\leq \frac{C_{\mathscr X}}{\sqrt n}.\]
(Para uma solu\c c\~ao, veja a prova do corol\'ario \ref{c:paracotagaussiana}.)
\label{ex:m-e}
\end{exercicio}

\begin{exercicio}
Mostrar que a constante da fam\'\i lia de cubos de Hamming satisfaz $C_{(\Sigma^n)}< 1$.
\end{exercicio}

\subsection{M\'edias de Rademacher}
Concentremos a nossa aten\c c\~ao na express\~ao seguinte da Eq. (\ref{eq:symmm}):

\begin{equation}
\frac 1n \sum_{i=1}^n \eta_i\chi_C(x_i).
\label{eq:rademacher}
\end{equation}

A escolha de sinais, $\eta$, pode ser vista como uma rotulagem de $\sigma$, usando os r\'otulos $-1$ e $+1$. Por causa da simetria de sinais, pode-se mostrar o seguinte.

\begin{exercicio}
\[\E_{\eta}\frac 1n \sum_{i=1}^n \eta_i\chi_C(x_i) =0.\]
\label{e:esperanceeta}
\end{exercicio}

Denotemos 
\[\eta_+ = \{i=1,2,\ldots,n\colon \eta_i=+1\},~~\eta_-= \{i=1,2,\ldots,n\colon \eta_i=-1\}.\]
O peso normalizado de $\eta$,
\[\bar w(\eta)=\frac 1n\sum_{i}\eta_i,\]
que toma os valores no intervalo $[-1,1]$, satisfaz as rela\c c\~oes seguintes:
\[\bar w(\eta)=\frac 1n(\sharp\eta_+-\sharp\eta_-),~\sharp\eta_+=\frac n2(1+\bar w(\eta)),~\sharp\eta_-=\frac n2(1-\bar w(\eta)).\]
Dada uma amostra $\sigma\in\Omega^n$, denotemos por $\sigma_{\eta_+}$, ou melhor, simplesmente $\sigma_+$, uma subamostra de $\sigma$ do tipo
\[\sigma_+ = (x_{i_1},x_{i_2},\ldots,x_{i_k}),~(i_1,i_2,\ldots,i_k)=\eta_+,\]
e da mesma forma para $\sigma_-$. Estas $\sigma_{\pm}$ s\~ao amostras i.i.d. de comprimento $\sharp\eta_{\pm}$ respectivamente, seguindo a distribui\c c\~ao $\mu$, e $\sigma_+$ e $\sigma_-$ s\~ao independentes. 

Para interpretar a express\~ao na eq.
(\ref{eq:rademacher}), calculemos o erro de aprendizagem da amostra $\sigma$ rotulada com sinais com o conceito $C$. Ao inv\'es da fun\c c\~ao bin\'aria $\chi_C$, temos que usar a fun\c c\~ao $2\chi_C-1$ tomando valores em $\{\pm 1\}$:
\[(2\chi_C-1)(x) =\begin{cases} 1,&\mbox{ se }x\in C,\\
-1,&\mbox{ se }x\notin C.
\end{cases}
\]

Como 
\[\eta_i(2\chi_C(x_i)-1) =\begin{cases} +1,&\mbox{ se } 
x_i\in\sigma_+\mbox{ e }x_i\in C, \\
&\mbox{ ou } x_i\notin\sigma_+\mbox{ e }x_i\notin C, \\
-1,&\mbox{ se } x_i\in\sigma_+\mbox{ e }x_i\notin C, \\
&\mbox{ ou } x_i\notin\sigma_+\mbox{ e }x_i\in C, \\
\end{cases}
\]
temos 
\begin{align*}
\frac 1n \sum_{i=1}^n \eta_i(2\chi_C(x_i)-1) &= \frac{\sharp(+1)}{n}-\frac{\sharp(-1)}{n} \\
&= 1-2\mu_{\sigma}(C\Delta\sigma_+) \\
&= 1-2\mbox{erro}_{\mu_\sigma,\sigma_+}(C).
\end{align*}
O valor pode ser visto como a {\em bondade de ajuste} ({\em goodness of fit}) {\em emp\'\i rica} de $\sigma_+$ com $C$. O valor m\'aximo de $1$ significa o ajuste perfeito.
\index{bondade de ajuste emp\'\i rica}

Daqui, deduzimos 
\begin{equation}
\frac 1n \sum_{i=1}^n \eta_i \chi_C(x_i) = \frac 12 + \frac 12 \bar w(\eta) - \mbox{erro}_{\mu_{\sigma},\sigma_+}(C).
\label{eq:interpretacao}
\end{equation}
A interpreta\c c\~ao desta express\~ao \'e menos transparente. Por\'em, se $\eta$ \'e uma amostra aleat\'oria, ent\~ao, quando $n\gg 1$, com alta confian\c ca, $\bar w(\eta)\approx 0$, ent\~ao, os valores da express\~ao acima v\~ao se aproximar de $1/2-\mbox{erro}_{\mu_{\sigma},\sigma_+}(C)$. Deste modo, vista como uma vari\'avel aleat\'oria, a express\~ao acima \'e uma medida de bondade de ajuste emp\'\i rica tamb\'em.

Revisitamos o exerc\'\i cio \ref{e:esperanceeta}, calculando a esperan\c ca da nossa soma, dessa vez em $\sigma$:
\begin{align*}
\E_{\sigma\sim\mu}\left(\frac 1n \sum_{i=1}^n \eta_i\chi_C(x_i)\right) 
&= \E_{\sigma\sim\mu}\left(\frac {\sharp\sigma_+}n\cdot \frac 1{\sharp\sigma_+}\sum_{i\in\sigma_+} \chi_C(x_i)-\frac {\sharp\sigma_-}n\cdot \frac 1{\sharp\sigma_-}\sum_{i\in\sigma_-} \chi_C(x_i)\right) \\
&= \frac {\sharp\sigma_+}n\E_{\sigma_+\sim\mu}\mu_{\sigma_+}(C) - \frac {\sharp\sigma_-}n\E_{\sigma_-\sim\mu}\mu_{\sigma_-}(C)\\
&= 
\frac 12(1+\bar w(\eta))\mu(C) - \frac 12(1-\bar w(\eta))\mu(C) \\
&= \bar w(\eta)\mu(C).
\end{align*}
Aqui, usamos a observa\c c\~ao que $\sigma_+$ e $\sigma_-$ s\~ao amostras independentes, distribu\'\i das segundo a medida $\mu$, junto com a propriedade $\E_{\sigma\sim\mu}\E_{\mu_\sigma}f=\E_{\mu}f$, estabelecida na subse\c c\~ao \ref{ss:casoimportante}.

Deduzimos:
\begin{align*}
2\e &> \E_{\eta}\E_{\sigma}
\E_{\sigma^\prime}\sup_{C\in {\mathscr C}}\left\vert
\frac 1n\sum_{i=1}^n \eta_i \chi_C(x_i)- \frac 1n\sum_{i=1}^n \eta_i 
\chi_C(x^\prime_i)\right\vert \\
&\geq  \E_{\eta}\E_{\sigma}\sup_{C\in {\mathscr C}}\left\vert
\E_{\sigma^\prime}\left(\frac 1n\sum_{i=1}^n \eta_i \chi_C(x_i)- \frac 1n\sum_{i=1}^n \eta_i 
\chi_C(x^\prime_i) \vert\sigma\right) \right\vert \\
&= \E_{\sigma}\E_{\eta}\sup_{C\in {\mathscr C}}\left\vert\frac 1n\sum_{i=1}^n \eta_i \chi_C(x_i)-\bar w(\eta)\mu(C)\right\vert
\\
&\geq \E_{\sigma}\E_{\eta}\sup_{C\in {\mathscr C}}\left\vert\frac 1n\sum_{i=1}^n \eta_i \chi_C(x_i)\right\vert - \E_{\eta}\left\vert \bar w(\eta)\right\vert\mu(C) \\
&> \E_{\sigma}\E_{\eta}\sup_{C\in {\mathscr C}}\left\vert\frac 1n\sum_{i=1}^n \eta_i \chi_C(x_i)\right\vert -\sqrt{\frac{2\pi}{n}}.
\end{align*}

Ou seja,
\[\E_{\sigma}\E_{\eta}\sup_{C\in {\mathscr C}}\left\vert\frac 1n\sum_{i=1}^n \eta_i \chi_C(x_i)\right\vert<2\ve+\sqrt{\frac{2\pi}{n}}.\]
Quando $\mathscr C$ \'e uma classe de Glivenko--Cantelli, a express\~ao
\begin{equation}
\E_{\eta}\sup_{C\in {\mathscr C}}\left\vert\frac 1n\sum_{i=1}^n \eta_i \chi_C(x_i)\right\vert,\end{equation}
 converge para zero em esperan\c ca (de maneira equivalente, em probabilidade). Por conseguinte, o mesmo se aplica \`a express\~ao
 \begin{equation}
 \label{eq:commodulo}
 \hat R_n({\mathscr C}) =\E_{\eta}\sup_{C\in {\mathscr C}}\frac 1n\sum_{i=1}^n \eta_i \chi_C(x_i).
\end{equation}

A express\~ao acima, (\ref{eq:commodulo}),
\'e chamada a {\em complexidade de Rademacher emp\'\i rica} da classe $\mathscr C$. A complexidade de Rademacher emp\'\i rica, $\hat R_n({\mathscr C})$, \'e uma {\em fun\c c\~ao determin\'\i stica,} de $\Omega^n$ para $[0,1]$. Na presen\c ca de uma medida $\mu$ sobre $\Omega$, ela torna-se uma vari\'avel aleat\'oria. A sua esperan\c ca,
\[R_n=\E_{\sigma\sim\mu}\hat R_n({\mathscr C}),\]
 chama-se a {\em complexidade de Rademacher} de $\mathscr C$.
\index{complexidade! de Rademacher! emp\'\i rica}
\index{complexidade! de Rademacher}

\begin{observacao}
\`A luz de eq. (\ref{eq:interpretacao}), 
\[\hat R_n({\mathscr C})(\sigma) = \frac 12 - \E_\eta\inf_{C\in{\mathscr C}}\mbox{erro}_{\mu_{\sigma},\sigma_+}(C).\]
Ent\~ao, o valor de $\hat R_n({\mathscr C})$ em $\sigma$ \'e a esperan\c ca da melhor bondade de ajuste com um conceito da classe $\mathscr C$, da amostra $\sigma$ rotulada aleatoriamente. Deste modo, o valor de $R_n({\mathscr C})$,
\[R_n=\frac 12 - \E_{\sigma,\eta}\inf_{C\in{\mathscr C}}\mbox{erro}_{\mu_{\sigma},\sigma_+}(C),\]
 \'e a esperan\c ca da melhor bondade de ajuste com $\mathscr C$ de uma $n$-amostra rotulada aleat\'oria.
\label{o:bondade}
\end{observacao}

\begin{exercicio}
Mostrar que $\hat R_n({\mathscr C})(\sigma)\geq 0$ para cada amostra $\sigma$.
\par
[ {\em Sugest\~ao:} se $\sup_{C\in {\mathscr C}}\frac 1n\sum_{i=1}^n \eta_i \chi_C(x_i)<0$, ent\~ao 
\[\sup_{C\in {\mathscr C}}\frac 1n\sum_{i=1}^n (-\eta)_i \chi_C(x_i)\geq \left\vert \sup_{C\in {\mathscr C}}\frac 1n\sum_{i=1}^n \eta_i \chi_C(x_i)\right\vert.~~]\]
\label{e:rnpositiva}
\end{exercicio}

Para resumir o que n\'os mostramos at\'e agora: se $\mathscr C$ \'e uma classe de Glivenko--Cantelli, ent\~ao
\[R_n({\mathscr C})\to 0\mbox{ quando }n\to\infty.\]

\subsection{O crit\'erio} Vamos deduzir a implica\c c\~ao no sentido contr\'ario: se $R_n({\mathscr C})\to 0$ (ou, da maneira equivalente, $\hat R_n({\mathscr C})\to 0$ em probabilidade), ent\~ao $\mathscr C$ \'e uma classe de Glivenko--Cantelli. O argumento \'e baseado sobre as mesmas ideias de simetriza\c c\~ao com sinais, e por isso precisamos usar uma c\'opia, $\sigma^\prime=(X^\prime_1,X^\prime_2,\ldots,X^\prime_n)$, independente e identicamente distribu\'\i da, da amostra aleat\'oria, $\sigma$. 
Temos:

\begin{align*}
\E_{\sigma\sim\mu}\sup_{C\in {\mathscr C}}\left(\mu_\sigma(C)-\mu(C)\right) 
&=
\E_{\sigma} \sup_{C\in{\mathscr C}}\left[\frac 1n\sum_{i=1}^n \chi_C(X_i) - \mu(C)\right] \\
&=
\E_{\sigma} \sup_{C\in{\mathscr C}}
\left[
\frac 1n\sum_{i=1}^n 
\left(\chi_C(X_i) - \E_{\sigma^\prime}\chi_C(X^\prime_i)\right)\right]
\\
&=
\E_{\sigma} \sup_{C\in{\mathscr C}}
\left[
\E_{\sigma^\prime}\left(
\frac 1n\sum_{i=1}^n 
(
\chi_C(X_i) - \chi_C(X^\prime_i)
)
\bigg|
\sigma 
\right)
\right]
\end{align*}

Na \'ultima express\~ao trata-se de uma esperan\c ca condicional, que corresponde \`a in\-te\-gra\-\c c\~ao ao longo das vari\'aveis $\sigma^\prime$ quando as vari\'aveis $\sigma$ s\~ao fixas. Agora vamos usar o fato que o supremo de integrais de uma fam\'\i lia de fun\c c\~oes \'e limitado superiormente pela integral do supremo da fam\'ilia, e depois aplicar o teorema de Fubini:
\begin{eqnarray*}
\E_{\sigma\sim\mu}\sup_{C\in {\mathscr C}}\left(\mu_\sigma(C)-\mu(C)\right) &\leq & \E_{(\sigma,\sigma^\prime)} \sup_{C\in{\mathscr C}}
\left[
\frac 1n\sum_{i=1}^n 
(
\chi_C(X_i) - \chi_C(X^\prime_i)
)
\right]
\end{eqnarray*}

Como as transposi\c c\~oes $i\leftrightarrow n+i$ conservam a medida $\mu^{\otimes 2n}$, conclu\'\i mos que para cada $\eta\in \{\pm 1\}^n$, 
\begin{eqnarray*}
\E_{\sigma\sim\mu}\sup_{C\in {\mathscr C}}\left(\mu_\sigma(C)-\mu(C)\right)
&\leq&
\E_{(\sigma,\sigma^\prime)} \sup_{C\in{\mathscr C}}
\left[
\frac 1n\sum_{i=1}^n 
\eta_i(
\chi_C(X_i) - \chi_C(X^\prime_i)
)
\right]. \\
\end{eqnarray*}
Por conseguinte,
\begin{eqnarray*}
\E_{\sigma\sim\mu}\sup_{C\in {\mathscr C}}\left(\mu_\sigma(C)-\mu(C)\right) 
&\leq&
\E_{\eta}\E_{(\sigma,\sigma^\prime)} \sup_{C\in{\mathscr C}}
\left[
\frac 1n\sum_{i=1}^n 
\eta_i(
\chi_C(X_i) - \chi_C(X^\prime_i)
)
\right] \\
&\leq& \E_{\eta,\sigma,\sigma^\prime}\sup_{C\in{\mathscr C}}
\left[
\frac 1n\sum_{i=1}^n 
\eta_i
\chi_C(X_i) + \frac 1n\sum_{i=1}^n (-\eta_i)\chi_C(X^\prime_i)
\right] \\
&\leq& \E_{\eta,\sigma}\sup_{C\in{\mathscr C}}
\frac 1n\sum_{i=1}^n 
\eta_i
\chi_C(X_i)
+\E_{\eta,\sigma^\prime}\sup_{C\in{\mathscr C}}
\frac 1n\sum_{i=1}^n (-\eta_i)\chi_C(X^\prime_i)
\\
&=& 2R_n({\mathscr C}).
\end{eqnarray*}

Para deduzir a propriedade de Glivenko--Cantelli, vamos aplicar o fen\^omeno de concentra\c c\~ao de medida, que tamb\'em permite de quantificar o resultado.

\begin{exercicio}
Verificar que a fun\c c\~ao real
\[\sigma\mapsto \sup_{C\in {\mathscr C}}(\mu_\sigma(C)-\mu(C))\]
\'e Lipschitz cont\'\i nua com rela\c c\~ao \`a dist\^ancia de Hamming normalizada sobre $\Omega^n$, de constante $L=1$.
\end{exercicio}

Por conseguinte, qualquer que seja $\ve>0$,
\begin{eqnarray*}
&&\mu^{\otimes n}\{\sigma\colon \sup_{C\in {\mathscr C}}(\mu_\sigma(C)-\mu(C))-2R_n({\mathscr C})>\ve\}  \\ &&
\leq
\mu^{\otimes n}\{\sigma\colon \sup_{C\in {\mathscr C}}(\mu_\sigma(C)-\mu(C))-\E\sup_{C\in {\mathscr C}}(\mu_\sigma(C)-\mu(C))>\ve\} \\ &&
\leq e^{-2\ve^2n}.
\end{eqnarray*}
O argumento id\^entico mostra que tamb\'em
\[\mu^{\otimes n}\{\sigma\colon \sup_{C\in {\mathscr C}}(\mu(C)-\mu_\sigma(C))-2R_n({\mathscr C})>\ve\} \leq e^{-2\ve^2n}.\]
Deduzimos: com confian\c ca $\geq 1-2e^{-2\ve^2n}$, qualquer que seja $C\in{\mathscr C}$,
\[\abs{\mu_\sigma(C)-\mu(C)}\leq 2R_n({\mathscr C})+\ve.\]
Nota-se que
  \[2e^{-2\ve^2n}\leq\delta\]
quando 
 \[\e\geq \sqrt{\frac{\ln (2/\delta)}{2n}}.\]

Podemos formular o resultado.

\begin{teorema}
Dada uma classe de conceitos $\mathscr C$, para cada $n$, temos com confian\c ca $1-\delta$, 
\[\sup_{C\in{\mathscr C}}\left\vert \mu_\sigma(C)-\mu(C)\right\vert\leq 2R_n({\mathscr C}) + \sqrt{\frac{\ln (2/\delta)}{2n}}.\]
A classe $\mathscr C$ \'e uma classe de Glivenko--Cantelli se e somente se as complexidades de Rademacher de $\mathscr C$ convergem para zero:
\[R_n({\mathscr C})\to 0\mbox{ quando }n\to\infty.\]
\label{t:rademacher}
\end{teorema}

\begin{observacao}
Se uma classe $\mathscr C$ ajusta bem as amostras rotuladas aleat\'orias, ent\~ao ela sofre de {\em sobreajuste} ({\em overfitting}). Desse modo, teorema \ref{t:rademacher} diz que uma classe \'e a de Glivenko--Cantelli se e somente se ele n\~ao sofre de sobreajuste.
\label{sobreajustevserro}
\index{sobreajuste}
\end{observacao}

\subsection{Propriedades de complexidades de Rademacher}
A no\c c\~ao de complexidades de Rademacher tem sentido para qualquer classe $\mathscr F$ de fun\c c\~oes reais sobre o dom\'\i nio. 

\begin{definicao}
Seja $\mathscr F$ uma classe de fun\c c\~oes reais.
A express\~ao
\[\hat R_n({\mathscr F})(\sigma) = \E_{\eta}\sup_{f\in{\mathscr F}}
\frac 1n \sum_{i=1}^n \eta_if(x_i)\]
chama-se a {\em complexidade de Rademacher emp\'\i rica} de classe $\mathscr F$. 
\end{definicao}

Mais uma vez, $\hat R_n({\mathscr F})$ \'e uma fun\c c\~ao real determin\'\i stica com valores positivos sobre o dom\'\i nio, que n\~ao depende de uma medida. Na presen\c ca de uma medida $\mu\in P(\Omega)$, torna-se uma vari\'avel aleat\'oria. Estamos principalmente interessados do caso onde ${\mathscr F}=\{\chi_C\colon C\in{\mathscr C}\}$ \'e uma classe de fun\c c\~oes bin\'arias. Nesse caso, vamos escrever $\hat R_n({\mathscr C})$, etc.

\begin{definicao}
Sejam $\mathscr F$ uma classe de fun\c c\~oes borelianas e $\mu$ uma medida de probabilidade boreliana sobre o dom\'\i nio.
A esperan\c ca da complexidade de Rademacher emp\'\i rica chama-se a {\em complexidade de Rademacher} da classe $\mathscr F$:
\[R_n({\mathscr F},\mu)= \E_{\sigma\sim\mu}\hat R_n({\mathscr F}),\]
ou: a {\em m\'edia de Rademacher} ({\em Rademacher average}), no contexto original da an\'alise funcional.
\end{definicao}

Por exemplo, $R_n({\mathscr F},\mu)$ \'e bem definida quando a classe $\mathscr F$ consiste de fun\c c\~oes com valores num intervalo.

\begin{proposicao}
Sejam $\mathscr F$ e $\mathscr G$ duas classes de fun\c c\~oes com valores num intervalo, $g$ uma fun\c c\~ao limitada qualquer, e $\lambda\in\R$. Ent\~ao
\begin{enumerate}
\item $\hat R_n({\mathscr F}+{\mathscr G})= \hat R_n({\mathscr F}) + \hat R_n({\mathscr G})$,
\item $\hat R_n({\mathscr F}+g)=\hat R_n({\mathscr F})$,
\item $\hat R_n(\lambda{\mathscr F})=\abs\lambda\hat R_n({\mathscr F})$.
\end{enumerate}
\label{ex:maisf}
\end{proposicao}

\begin{proof}
\begin{align*}
(1):~~\hat R_n({\mathscr F}+{\mathscr G}) 
&= \E_\eta\sup_{f\in{\mathscr F},g\in{\mathscr G}}\frac 1n\sum_{i=1}^n \eta_i (f(x_i)+g(x_i)) 
\\
&= 
\E_\eta\sup_{f\in{\mathscr F}}\frac 1n\sum_{i=1}^n \eta_i f(x_i) +
\E_\eta\sup_{g\in{\mathscr G}}\frac 1n\sum_{i=1}^n \eta_i g(x_i)
\\ 
&=
 \hat R_n({\mathscr F}) + \hat R_n({\mathscr G}).
\end{align*}

\begin{align*}
(2):~~\hat R_n({\mathscr F}+f)
&=\E_\eta\sup_{f\in{\mathscr F}}\frac 1n\sum_{i=1}^n \eta_i(f(x_i)+g(x_i)) 
\\
&= \E_\eta\sup_{f\in{\mathscr F}}\frac 1n\sum_{i=1}^n \eta_i f(x_i) + \E_\eta\frac 1n\sum_{i=1}^n \eta_i g(x_i)
\\
&=
\hat R_n({\mathscr F}).
\end{align*}

\begin{align*}
(3):~~\hat R_n(\lambda{\mathscr F})
&=\E_\eta\sup_{f\in{\mathscr F}}\frac 1n\sum_{i=1}^n \lambda \eta_i f(x_i) 
\\
&=\E_\eta\sup_{f\in{\mathscr F}}\frac 1n\sum_{i=1}^n \abs{\lambda} \eta_i f(x_i) 
\\
&=
\abs\lambda\hat R_n({\mathscr F}).
\end{align*} 
\end{proof}

\begin{corolario}
Sejam $\mathscr F$ e $\mathscr G$ duas classes de fun\c c\~oes tais que $0\in{\mathscr F}$, $0\in {\mathscr G}$. Ent\~ao,
\[\hat R_n({\mathscr F}\cup{\mathscr G})\leq \hat R_n({\mathscr F}) + \hat R_n({\mathscr G}).
\]
\label{c:rndauniao}
\end{corolario}

\begin{proof}
Como $0\in{\mathscr F}$, conclu\'\i mos que ${\mathscr G}\subseteq {\mathscr F}+{\mathscr G}$, e igualmente para $\mathscr F$. Agora o resultado segue-se de proposi\c c\~ao \ref{ex:maisf},(1).
\end{proof}

Sem a hip\'otese $0\in{\mathscr F}$, $0\in {\mathscr G}$, o resultado \'e falso.

\begin{exemplo}
Seja ${\mathscr C}=\{f\}$ uma classe que consiste de uma fun\c c\~ao s\'o. Como no exerc\'\i cio \ref{e:esperanceeta},
\[R_n(\{f\}) = 0.\]
Agora seja $\Omega=\{x_1,x_2,\ldots,x_n\}$ um dom\'\i nio finito qualquer com $n$ pontos, e ${\mathscr C}=2^\Omega$, a classe de todos subconjuntos de $\Omega$. Denotemos $\sigma=(x_1,x_2,\ldots,x_n)$. Se a conclus\~ao do corol\'ario \ref{c:rndauniao} fosse verdadeira nessa situa\c c\~ao, ela implicaria indutivamente que $\hat R_n(2^\Omega)(\sigma)=0$. Na realidade,
\[\hat R_n(2^\Omega)(\sigma) = \frac 12,\]
atingindo o valor m\'aximo, porque toda rotulagem de $\sigma$ pode ser ajustada perfeitamente com um conceito $C\in 2^\Omega$.
\label{ex:umconceitoso}
\end{exemplo}

Ao mesmo tempo, vamos ver em breve que o corol\'ario \ref{c:rndauniao} \'e ``quase'' verdadeiro, a menos de um pequeno termo de ajuste. 

\begin{proposicao}[Desigualdade de Jensen] Seja $\phi$ uma fun\c c\~ao real c\^oncava, e seja $X$ uma vari\'avel aleat\'oria real. Ent\~ao,
\[\E(\phi(X))\leq \phi(\E(X)).\]
\index{desigualdade! de Jensen}
\end{proposicao}

A desigualdade \'e melhor ilustrada usando uma vari\'avel aleat\'oria $X$ tomando dois valores quaisquer, $x$ e $y$, com as probabilidades $P[X=x]=t$, $P[X=y]=1-t$. Neste caso,
\[\E(\phi(X)) =t\phi(x)+(1-t)\phi(y)\leq \phi\left(tx+(1-t)y\right)=\phi(\E(X)),\]
porque a linha que junta os pontos $(x,\phi(x))$ e $(y,\phi(y))$, no gr\'afico de $\phi$, fica abaixo do gr\'afico, o que \'e exatamente a defini\c c\~ao de uma fun\c c\~ao c\^oncava. Este argumento se generaliza por indu\c c\~ao sobre todas as combina\c c\~oes convexas finitas, e posteriormente o valor $\E(\phi(X))=\int f(x)\,d\mu(x)$ \'e aproximado pelos valores de $\phi$ em combina\c c\~oes convexas finitas.  

\begin{lema}[Lema de Massart] 
Sejam $\mathscr F$ uma classe de fun\c c\~oes com valores no intervalo $[-1,1]$, e $\sigma$ uma amostra com $n$ pontos.
Suponha que $k=\sharp({\mathscr F}\upharpoonright\sigma)<\infty$. Ent\~ao,
\[\hat R_n({\mathscr F})(\sigma)\leq \sqrt{\frac{2\log k}{n}}.\]
\label{l:massart}
\end{lema}

\begin{proof}
Denotemos a express\~ao \`a direita por $\lambda$. Identifiquemos $\mathscr F\upharpoonright\sigma$ com o conjunto de todos os vetores $v\in [-1,1]^n$ dados por $v_i=f(x_i)$.
Temos
\begin{align*}
\lambda n \hat R_n({\mathscr F})(\sigma) 
&=
\E_{\eta}\lambda \sup_{f\in{\mathscr F}} \sum_{i=1}^n\eta_i f(x_i) \\
&=
\E_{\eta}\max_{v\in\mathscr F\upharpoonright\sigma} \lambda\langle \eta,v\rangle \\
&= \E_{\eta}\log\max_{v\in {\mathscr F}\upharpoonright\sigma} e^{\lambda\langle \eta,v\rangle}\\
\mbox{\small (Jensen) }
&\leq \log \E_{\eta} \max_{v\in {\mathscr F}\upharpoonright\sigma} e^{\lambda\langle \eta,v\rangle}\\ 
&\leq 
\log \E_{\eta} \sum_{v\in {\mathscr F}\upharpoonright\sigma} e^{\lambda\langle \eta,v\rangle}\\ 
& = 
\log  \sum_{v\in {\mathscr F}\upharpoonright\sigma} \E_{\eta}
\prod_{i=1}^ne^{\lambda\eta_iv_i}\\ 
\end{align*}
A esperan\c ca do produto das vari\'aveis aleat\'orias independentes \'e igual ao produto das esperan\c cas (teorema de Fubini). Continuamos:

\begin{align*}
\lambda n \hat R_n({\mathscr F})(\sigma)
&\leq  \log  \sum_{v\in{\mathscr F}\upharpoonright\sigma} 
\prod_{i=1}^n\E_{\eta}e^{\lambda\eta_iv_i}\\
&=
\log  \sum_{v\in{\mathscr F}\upharpoonright\sigma} 
\prod_{i=1}^n\frac 12\left(e^{\lambda v_i}+e^{-\lambda v_i} \right) \\
\mbox{(exerc\'\i cio \ref{l:cosh})}
&\leq  \log  \sum_{v\in{\mathscr F}\upharpoonright\sigma}  
\prod_{i=1}^ne^{\lambda^2 v_i^2/2}\\
&\leq \log\left(k e^{n\lambda^2/2}\right)\\
&= \log k + \frac{n}2\lambda^2.
\end{align*}
Substituindo o valor original de $\lambda$, obtemos o resultado.
\end{proof}

\begin{lema}
Seja $\mathscr F$ uma classe de fun\c c\~oes com valores no intervalo $[-1,1]$. Ent\~ao,
\[\hat R_n({\mathscr F}\cup \{0\})\leq \hat R_n({\mathscr F}) + \sqrt{\frac{2\log 2}{n}}.\]
\end{lema}

\begin{proof}
Seja $f_0\in {\mathscr F}$ um elemento qualquer. Nota-se que a classe ${\mathscr F}-f_0$ cont\'em zero. Usando corol\'ario \ref{c:rndauniao} e lema de Massart \ref{l:massart}, obtemos:
\begin{eqnarray*}
\hat R_n({\mathscr F}\cup \{0\}) &=&
\hat R_n(({\mathscr F}\cup \{0\})-f_0) \\
&=& \hat R_n(({\mathscr F}-f_0)\cup \{-f_0\}) \\
&\leq &\hat R_n(({\mathscr F}-f_0)\cup \{0,-f_0\}) \\
&\leq & \hat R_n({\mathscr F}-f_0) + \hat R_n\{0,-f_0\} 
\\
&\leq  &
\hat R_n({\mathscr F}) + \sqrt{\frac{2\log 2}{n}}.
\end{eqnarray*}
\end{proof}

\begin{proposicao}
Sejam $\mathscr F$ e $\mathscr G$ duas classes de fun\c c\~oes com valores no intervalo $[-1,1]$. Ent\~ao,
\[\hat R_n({\mathscr F}\cup{\mathscr G})\leq \hat R_n({\mathscr F}) + \hat R_n({\mathscr G})+2\sqrt{\frac{2\log 2}{n}}.
\]
\label{p:rndauniao}
\end{proposicao}

\begin{proof}
\begin{eqnarray*}
\hat R_n({\mathscr F}\cup{\mathscr G}) &\leq & 
\hat R_n(({\mathscr F}\cup\{0\})\cup ({\mathscr G}\cup\{0\})) \\
&\leq & \hat R_n({\mathscr F}\cup\{0\})+\hat R_n({\mathscr G}\cup\{0\}) 
\\
&\leq & \hat R_n({\mathscr F}) + \hat R_n({\mathscr G}) + 2\sqrt{\frac{2\log 2}{n}}.
\end{eqnarray*}
\end{proof}

\begin{exercicio}
Para uma classe de fun\c c\~oes sobre um dom\'\i nio $\Omega$, denotemos por $\mbox{conv}\,(\mathscr F)$ a {\em envolt\'oria convexa} de $\mathscr F$, ou seja, o menor conjunto convexo do espa\c co linear $\R^{\Omega}$ que cont\'em $\mathscr F$:
\[\mbox{conv}\,(\mathscr F)=\left\{\sum_{i=1}^n\lambda_if_i\colon n\in\N,~\lambda_i\geq 0,~\sum\lambda_i=1,~f_i\in {\mathscr F}\right\}.\]
Nota-se que se $\mathscr F$ consiste de fun\c c\~oes borelianas com valores num intervalo $[a,b]$, ent\~ao o mesmo vale para $\mbox{conv}\,(\mathscr F)$.
Mostrar que $\hat R_n(\mbox{conv}\,(\mathscr F))=\hat R_n(\mathscr F)$.
\index{envolt\'oria convexa}
\end{exercicio}

\begin{exercicio}
Deduzir que, dada uma classe $\mathscr F$ de fun\c c\~oes com valores no intervalo $[-1,1]$, a sua {\em envolt\'oria convexa sim\'etrica} satisfaz
\[\hat R_n(\mbox{conv}\,(\mathscr F\cup -\mathscr F))\leq 2\hat R_n(\mathscr F) +2\sqrt{\frac{2\log 2}{n}}. 
\]
\end{exercicio}

J\'a o seguinte resultado n\~ao \'e completamente trivial.

\begin{lema} Dada uma classe $\mathscr F$, denotemos $\abs{\mathscr F}=\{\abs f\colon f\in{\mathscr F}\}$. Ent\~ao,
\[\hat R_n(\abs{\mathscr F})\leq \hat R_n(\mathscr F).\]
\end{lema}

\begin{proof}
Omitindo o fator $1/n$, temos a desigualdade a mostrar para uma amostra qualquer, $\sigma=(x_1,x_2,\ldots,x_n)$:
\begin{equation}
\E_\eta \sup_{f\in{\mathscr F}}\sum_{i=1}^n \eta_i \abs{f(x_i)}\leq \E_\eta \sup_{f\in{\mathscr F}}\sum_{i=1}^n \eta_i f(x_i).
\label{eq:contraction}
\end{equation}
A chave para a prova \'e dada pelo argumento no caso mais simples de $n=1$. Temos:
\begin{align*}
\frac 12[\sup_{f\in{\mathscr F}}\abs{f(x_1)}+\sup_{f\in{\mathscr F}}(-1)\abs{f(x_1)}] &=  
\frac 12 \sup_{f,g\in{\mathscr F}} 
\left(\abs{f(x_1)}-\abs{g(x_1)}\right) \\
\mbox{\small [ porque $\abs{a}-\abs b\leq \abs{a-b}$ ]}
&\leq \frac 12 \sup_{f,g\in{\mathscr F}} 
\abs{f(x_1)-g(x_1)}  \\
\mbox{\small [ \'e claro que o m\'odulo pode ser suprimido ] }
&= \frac 12 \sup_{f,g\in{\mathscr F}} 
(f(x_1)-g(x_1))  \\
&=  \frac 12[\sup_{f\in{\mathscr F}}f(x_1)+\sup_{f\in{\mathscr F}}(-1)f(x_1)].
\end{align*}
No caso de $n$ qualquer, n\'os simplesmente aplicamos o mesmo jeito a cada coordenada, uma ap\'os a outra, come\c cando com $i=1$:
\begin{align*}
\E_\eta \sup_{f\in{\mathscr F}}\sum_{i=1}^n \eta_i \abs{f(x_i)} 
&=
\frac 12\E_{\eta_2,\ldots,\eta_n}\left[\sup_{f\in{\mathscr F}}\left(\abs{f(x_1)}+ \sum_{i=2}^n \eta_i \abs{f(x_i)}\right)\right.+
\\ & 
\phantom{xxxxxxxxxxx}
\left.\sup_{f\in{\mathscr F}}\left(-\abs{f(x_1)}+ \sum_{i=2}^n \eta_i \abs{f(x_i)}\right)\right]\\
&= \frac 12\E_{\eta_2,\ldots,\eta_n}\sup_{f,g\in{\mathscr F}}
\left[\abs{f(x_1)}-\abs{g(x_1)} + \sum_{i=2}^n \eta_i \abs{f(x_i)}+\sum_{i=2}^n \eta_i \abs{g(x_i)}
 \right]
\\
&\leq  \frac 12\E_{\eta_2,\ldots,\eta_n}\sup_{f,g\in{\mathscr F}}
\left[f(x_1)-g(x_1) + \sum_{i=2}^n \eta_i \abs{f(x_i)}+\sum_{i=2}^n \eta_i \abs{g(x_i)}
 \right]
\\
&= \E_\eta \sup_{f\in{\mathscr F}}\left[ \eta_1 f(x_1)+\sum_{i=2}^n \eta_i \abs{f(x_i)} \right] \\
&\leq  \ldots \\
&\leq  \E_\eta \sup_{f\in{\mathscr F}}\sum_{i=1}^n \eta_i f(x_i).
\end{align*}
\end{proof}

\begin{exercicio} Generalizar o argumento acima para mostrar a {\em desigualdade de contrata\c c\~ao de Ledoux--Talagrand}.
Sejam $\phi\colon\R\to\R$ uma fun\c c\~ao Lipschitz cont\'\i nua de constante $L\geq 0$, e $\mathscr F$ uma classe de fun\c c\~oes reais no dom\'\i nio $\Omega$. Denotemos
\[\phi\circ{\mathscr F}=\{\phi\circ f\colon f\in {\mathscr F}\}.\]
 Ent\~ao,
\[\hat R_n(\phi\circ {\mathscr F}) \leq L\cdot \hat R_n({\mathscr F}).\]
\index{desigualdade! de contrata\c c\~ao de Ledoux--Talagrand}
\end{exercicio}

\begin{exercicio}
Deduzir o seguinte. 
Sejam $\mathscr C$ e $\mathscr D$ duas classes de Glivenko--Cantelli. Ent\~ao, a classe
\[{\mathscr C}\Delta {\mathscr D}=\{C\Delta D\colon C\in {\mathscr C},~D\in {\mathscr D}\}\]
\'e uma classe de Glivenko--Cantelli, que satisfaz
\[R_n({\mathscr C}\Delta {\mathscr D})\leq R_n({\mathscr C})+ R_n({\mathscr D})+2\sqrt{\frac{2\log 2}{n}}.\]
[ {\em Sugest\~ao:} $\chi_{C\Delta D}=\abs{\chi_C-\chi_D}$. ]
\label{ex:delta}
\end{exercicio}

\begin{exercicio}
Deduzir que se $\mathscr C$ \'e uma classe de Glivenko--Cantelli e $D\subseteq\Omega$ um conceito qualquer, ent\~ao a classe ${\mathscr C}\Delta D=\{C\Delta D\colon C\in {\mathscr C}\}$ \'e a de Glivenko--Cantelli, tendo a mesma complexidade de Rademacher que $\mathscr C$:
\[\hat R_n({\mathscr C}\Delta D)=\hat R_n({\mathscr C}).\]
\label{ex:cdeltad}
\end{exercicio}

\begin{observacao}
A defini\c c\~ao da complexidade de Rademacher emp\'\i rica na eq. (\ref{eq:commodulo}) n\~ao \'e a \'unica usada na literatura. Frequentemente, a complexidade de Rademacher \'e definida como a esperan\c ca da express\~ao 
\[\sup_{C\in {\mathscr C}}\left\vert\frac 1n\sum_{i=1}^n \eta_i \chi_C(x_i)\right\vert.\]
\end{observacao}

Obviamente, 
\[\hat R_n({\mathscr F})(\sigma)\leq \E_\eta \sup_{C\in {\mathscr C}}\left\vert\frac 1n\sum_{i=1}^n \eta_i \chi_C(x_i)\right\vert.\]
Ao mesmo tempo, as duas no\c c\~oes diferem j\'a nos casos mais simples.

\begin{exemplo}
Como vimos no exerc\'\i cio \ref{ex:umconceitoso}, se ${\mathscr C}=\{C\}$ \'e uma classe que consiste de um conceito s\'o, temos
\[R_n({\mathscr C}) = 0.\]
No entanto, 
\begin{eqnarray*}
\E_\sigma\E_\eta \left\vert \frac 1n \sum_{i=1}^n \eta_i\chi_C(x_i)\right\vert 
&=&   \E_\eta \abs{\bar w}\mu(C),
\end{eqnarray*}
um valor positivo de ordem de grandeza $O(1/\sqrt n)$.
\label{ex:umconceitoso2}
\end{exemplo}

Portanto, as duas no\c c\~oes n\~ao s\~ao t\~ao diferentes. 

\begin{lema}
Se $\mathscr F$ \'e uma classe que cont\'em a fun\c c\~ao nula, ent\~ao
\[\E_\eta \sup_{f\in {\mathscr F}}\left\vert\frac 1n\sum_{i=1}^n \eta_i f(x_i)\right\vert\leq 2 \hat R_n({\mathscr F})(\sigma).\]
\end{lema}
 
\begin{proof}
Seja $\eta$ qualquer. Suponha que $f\in {\mathscr F}$ \'e tal que 
\[\left\vert\frac 1n \sum_{i=1}^n \eta_if(x_i)\right\vert>0.\]
Se a express\~ao sob o valor de m\'odulo \'e positiva, este valor \'e atingido por $\frac 1n \sum_{i=1}^n \eta_if(x_i)$. Sen\~ao, ele \'e atingido por $\frac 1n \sum_{i=1}^n (-\eta_i)f(x_i)$. Conclu\'\i mos: existe um subconjunto $A\subseteq\{\pm 1\}^n$ tal que $\pm A=\{-1,1\}^n$ (em particular, $\mu_{\sharp}(A)\geq 1/2$), e para $\eta\in A$, os valores das express\~oes 
\[\sup_{f\in{\mathscr F}}
\left\vert\frac 1n \sum_{i=1}^n \eta_if(x_i)\right\vert
\mbox{ e }\sup_{f\in{\mathscr F}}
\frac 1n \sum_{i=1}^n \eta_if(x_i)\]
s\~ao id\^enticos. Como $0\in {\mathscr F}$, a \'ultima express\~ao n\~ao \'e negativa.
\end{proof}

\begin{exercicio}
Deduzir que, qualquer que seja uma classe de fun\c c\~oes, $\mathscr F$, com valores em $[-1,1]$,
\[\E_\eta \sup_{C\in {\mathscr C}}\left\vert\frac 1n\sum_{i=1}^n \eta_i \chi_C(x_i)\right\vert\leq 2\hat R_n({\mathscr F}) + \sqrt{\frac{2\log 2}{n}}.\]
\end{exercicio}

\begin{observacao}
A nossa defini\c c\~ao de complexidades de Rademacher (sem valor absoluto) \'e mais natural por pelo menos duas raz\~oes. Ela tem um significado muito claro (observa\c c\~ao \ref{o:bondade}). Al\'em disso, o lema de Massart torna-se mais elegante.
\end{observacao}

\begin{observacao}
Recordemos (obs. \ref{sobreajustevserro}) que a complexidade de Rademacher,  $R_n({\mathscr C},\mu)$, de uma classe de conceitos $\mathscr C$ interpreta-se como a bondade de ajuste emp\'\i rica esperada por $\mathscr C$ de uma amostra rotulada aleat\'oria. Segundo o nosso resultado principal (teorema \ref{t:rademacher}), uma classe $\mathscr C$ \'e de Glivenko--Cantelli se e somente se sua bondade de ajuste \'e assintoticamente nula, ou seja, ruim.

Esse resultado mostra que aprendizagem dentro uma classe de Glivenko--Cantelli \'e imposs\'\i vel? N\~ao, porque na realidade nunca tentamos aprender nos baseando sobre uma rotulagem totalmente aleat\'oria para cada amostra. Dentro do nosso modelo adotado, assumimos que a rotulagem \'e induzida por um conceito desconhecido, mas sempre o mesmo (ou mudando, mas devagar, como na no\c c\~ao recente do {\em concept drift}\footnote{Que merece uma boa investiga\c c\~ao matem\'atica, baseada sobre a teoria de sistemas din\^amicos. O {\em concept drift} significa a mudan\c ca temporal do conceito a ser aprendido, ou seja, uma a\c c\~ao de $\R_+$ sobre o espa\c co de conceitos.}). O fato de m\'aquina de aprendizagem funcionar mal que alimentada com rotulagens aleat\'orias \'e um bom ind\'\i cio, mostra-nos a sua estabilidade, e significa que a m\'aquina n\~ao sofre o sobreajuste e vai generalizar bem.
\end{observacao}

\section{Minimiza\c c\~ao da perda emp\'\i rica}

\subsection{}
Vamos deduzir os primeiros corol\'arios do teorema \ref{t:rademacher}.

\begin{corolario}
Seja $\mathscr C$ uma classe de conceitos de Glivenko--Cantelli. Ent\~ao $\mathscr C$ \'e consistentemente PAC aprendiz\'avel. Se $\mathcal L$ \'e uma regra de aprendizagem consistente com a classe $\mathscr C$, ent\~ao para todo $n$ e todo $\delta>0$, dado um conceito $C\in {\mathscr C}$, temos, com confian\c ca $1-\delta$,
\[\mbox{erro}_{\mu,C}({\mathcal L}_n(C\upharpoonright\sigma)) \leq 2R_n({\mathscr C}) + \frac{\sqrt{\ln (2/\delta)}}{\sqrt{2n}}.\]
\label{c:consistente}
\end{corolario}

\begin{proof} Seja $C\in {\mathscr C}$ qualquer. A classe ${\mathscr C}\Delta C$ \'e Glivenko--Cantelli, com $\hat R_n({\mathscr C}\Delta C)= \hat R_n({\mathscr C})$ (exerc\'\i cio \ref{ex:cdeltad}).
Como $\mathcal L$ \'e consistente, para cada $\sigma\in\Omega^n$ o erro emp\'\i rico de aprendizagem \'e zero, e temos
\begin{align*}
\mbox{erro}_{\mu,C}({\mathcal L}_n(C\upharpoonright\sigma)) 
&=
\mu\left({\mathcal L}_n(C\upharpoonright\sigma)\Delta C\right) \\
&=
\left\vert 
\mu\left({\mathcal L}_n(C\upharpoonright\sigma)\Delta C\right)-
\mu_{\sigma}\left({\mathcal L}_n(C\upharpoonright\sigma)\Delta C\right)\right\vert \\
\mbox{\small [ com confian\c ca $1-\delta$, por teor. \ref{t:rademacher} ]}
&\leq 
2R_n({\mathscr C}\Delta C) + \sqrt{\frac{\ln (2/\delta)}{2n}}\\
&= 2R_n({\mathscr C}) + \frac{\sqrt{\ln (2/\delta)}}{\sqrt{2n}}.
\end{align*}
\end{proof}

\begin{observacao}
No entanto, nem toda classe consistentemente aprendiz\'avel \'e uma classe de Glivenko--Cantelli. O contraexemplo mais simples \'e a classe que consiste de todos os subconjuntos finitos do intervalo $[0,1]$ munido da medida de Lebesgue. 

Seja $\mathcal L$ uma regra consistente com a classe (ou mesmo uma regra qualquer tomando valores em $\mathscr C$). Qualquer seja $C\in {\mathscr C}$,
\[\mbox{erro}_{\mu,C}{\mathcal L}_n(C\upharpoonright\sigma) = \mu\left({\mathcal L}_n(C\upharpoonright\sigma)\Delta C\right) =0,\]
porque ${\mathcal L}_n(C\upharpoonright\sigma)\Delta C$ \'e finito. 

Ao mesmo tempo, esta classe n\~ao \'e de Glivenko--Cantelli. Nenhuma amostra $\sigma$ vai estimar corretamente, por exemplo, a sua pr\'opria medida (mais exatamente, a do conceito $\{x_i\colon \e_i=1\}$):
\[\sup_{C\in{\mathscr C}}\left\vert \mu_\sigma(C)-\mu(C)\right\vert =
\mu_\sigma(\sigma) - \mu(\sigma) = 1-0=1.\]

Mesmo assim, a intui\c c\~ao \'e que as classes consistentemente aprendiz\'aveis n\~ao podem ser muito diferentes das classes de Glivenko--Cantelli. Talvez, num sentido, o exemplo acima exprima a \'unica patologia poss\'\i vel.
Seria interessante formular um crit\'erio necess\'ario e suficiente. Vamos conjeturar um tal crit\'erio mais tarde (conjetura \ref{conj:pacaprend}).
\label{o:consnaogc}
\end{observacao}

\subsection{}
Com alta confian\c ca, a dist\^ancia $L^1$ emp\'\i rica aproxima bem a dist\^ancia $L^1$ entre todos os pares de elementos de uma classe de Glivenko-Cantelli. 

\begin{corolario}
Seja $\mathscr C$ uma classe de Glivenko-Cantelli. Dado $\delta>0$ e $n$ quaisquer, com confian\c ca $1-\delta$ temos
\[\forall C,D\in {\mathscr C},~\mu(C\Delta D)\overset\ve\approx \mu_{\sigma}(C\Delta D),\]
onde
\[\e = 2R_n({\mathscr C}) + 2\sqrt{\frac{2\log 2}{n}} +
\sqrt{\frac{\ln (2/\delta)}{2n}}.\]
\end{corolario}

\begin{proof}
Usemos o exerc\'\i cio \ref{ex:delta}.
\end{proof}

\begin{exercicio} 
Formule e mostre o resultado mais geral para classes de fun\c c\~oes.
\end{exercicio}

Como toda classe PAC aprendiz\'avel, uma classe de Glivenko--Cantelli \'e pr\'e-compacta. Isso pode ser estabelecido diretamente, sem passar pela aprendizabilidade.

\begin{corolario}
Toda classe de Glivenko--Cantelli \'e pr\'e-compacta na pseudom\'etrica $L^1(\mu)$.
\end{corolario}

\begin{proof} 
O argumento \'e quase mesmo que na prova do teorema \ref{t:benedek-itai} de Benedek-Itai.
Seja $\ve>0$ qualquer, e $\delta<1$. Suponha que os conceitos $C_1,C_2,\ldots,C_k$ formam um $2\ve$-empacotamento na dist\^ancia $L^1(\mu)$.
Quando $n\geq s(\e,\delta, {\mathscr C}\Delta {\mathscr C})$, existe pelo menos uma amostra $\sigma\in\Omega^n$ estimando as dist\^ancias dois a dois entre $C_i$ com a precis\~ao $\ve$. Segue-se que as imagens $C_i\upharpoonright\sigma$ s\~ao dois a dois distintas, e por conseguinte $k\leq 2^n$. Conclu\'\i mos:
\[D(2\e,{\mathscr C},L^1(\mu))\leq 2^{s(\e,\delta, {\mathscr C}\Delta {\mathscr C})}.\]
\end{proof}

J\'a vimos um exemplo de uma classe pr\'e-compacta que n\~ao \'e consistentemente aprendiz\'avel (exemplo \ref{ex:precnaocons}). \'E f\'acil ver diretamente que esta classe n\~ao \'e de Glivenko--Cantelli, pela mesma raz\~ao que na observa\c c\~ao \ref{o:consnaogc}.

\subsection{}
\'E claro que o significado do corol\'ario \ref{c:consistente} \'e puramente te\'orico: n\~ao tem raz\~ao nenhuma para um conceito desconhecido pertencer \`a classe $\mathscr C$. Ainda assim, as no\c c\~oes, resultados, e provas podem ser adaptados simplesmente \`a situa\c c\~ao seguinte mais realista. 

\begin{definicao}
Sejam $\mathscr C$ uma classe de conceitos, $D$ um conceito qualquer (ou seja, um subconjunto boreliano do dom\'\i nio $\Omega$), e $\mu$ uma distribui\c c\~ao  sobre $\Omega$. O erro de aprendizagem do conceito $D$ pela classe $\mathscr C$ com rela\c c\~ao \`a medida $\mu$
\'e a \'\i nfima discrep\^ancia entre $D$ e os ajustes de $D$ em $\mathscr C$:
\[\mbox{erro}_{\mu,D}({\mathscr C})
 =\inf_{C\in {\mathscr C}} \mu(C\,\Delta\,D).\]
\end{definicao}

\'E uma fun\c c\~ao determin\'\i stica sobre a fam\'\i lia ${\mathscr B}(\Omega)$ de todos os subconjuntos borelianos de $\Omega$, tomando os valores reais n\~ao negativos. Por exemplo, quando $D\in {\mathscr C}$, o valor
$\mbox{erro}_{\mu,D}({\mathscr C})$ \'e nulo qualquer que seja a medida $\mu$. Ela mede o \'\i nfimo de todos os erros de classifica\c c\~ao atingidos quando aprendemos o conceito $D$ usando a classe $\mathscr C$, com rela\c c\~ao \`a medida subjacente $\mu$.

Visto que, na pr\'atica, n\~ao h\'a como calcular os valores da fun\c c\~ao acima, conforma-se com o 
{\em erro de aprendizagem emp\'\i rico,} onde a medida $\mu$ \'e substitu\'\i da com a medida emp\'\i rica $\mu_n$ suportada por uma amostra finita $\sigma$:
\[\mbox{erro}_{\mu_{\sigma},D}({\mathscr C})
= \inf_{C\in {\mathscr C}}\mu_{\sigma}(C\,\Delta\,D).\]
Neste caso, o \'\i nfimo \'e sempre atingido, logo \'e o m\'\i nimo. A fun\c c\~ao de erro emp\'\i rico  \'e uma vari\'avel aleat\'oria, porque ela depende da amostra aleat\'oria $\sigma$. 

\begin{definicao}
Seja $\mathscr C$ uma classe de conceitos. Diz-se que uma regra de aprendizagem, $\mathcal L$, segue o princ\'\i pio de {\em minimiza\c c\~ao de erro emp\'\i rico} ({\em empirical risk minimization,} {\em ERM}) {\em dentro a classe} $\mathscr C$, se existe uma sequ\^encia de constantes
\[(\ve_n),~~\ve_n\geq 0,~~\ve_n\to 0\mbox{ quando }n\to\infty,\]
tal que, para todo $D\in {\mathscr B}(\Omega)$ e toda $\sigma\in\Omega^n$,
\[\mu_\sigma({\mathcal L}_n(\sigma,D\upharpoonright \sigma)\,\Delta\,D)\leq \mbox{erro}_{\mu_{\sigma},D}({\mathscr C}) + \ve_n.\]
\label{d:erm}
\index{princ\'\i pio! de minimiza\c c\~ao! de erro emp\'\i rico}
\end{definicao}

Desta maneira, a regra acha um ajuste empiricamente quase \'otimo para $D$, com uma margem de erro permitido $\ve_n$. 

A prova do seguinte resultado mimetiza a do corol\'ario \ref{c:consistente}.

\begin{teorema}
Seja $\mathscr C$ uma classe de conceitos, e seja $\mathcal L$ uma regra de aprendizagem seguindo o princ\'\i pio de minimiza\c c\~ao da perda emp\'\i rica dentro $\mathscr C$ com as constantes de erro $(\ve_n)$. Quaisquer sejam $n$ e $\delta>0$, dado um subconjunto boreliano $D\subseteq\Omega$ qualquer,
uma amostra aleat\'oria $\sigma$ satisfaz, com confian\c ca $>1-\delta$:
\[\mbox{erro}_{\mu,D}({\mathcal L}_n(D\upharpoonright\sigma)) \leq 
\mbox{erro}_{\mu_{\sigma},D}({\mathscr C}) +\ve_n +
2R_n({\mathscr C}) + \frac{\sqrt{\ln (2/\delta)}}{\sqrt{2n}}.\]
\qed
\end{teorema}

Desta maneira, assintoticamente, a regra $\mathcal L$ provavelmente aproximadamente corretamente aprende, dentro a classe $\mathscr C$, um conceito boreliano, $D$, {\em qualquer,} atingindo no limite $n\to\infty$ o menor erro teoricamente poss\'\i vel.

\begin{observacao}
Eu n\~ao sei responder a pergunta seguinte. Seja $\mathscr C$ uma classe consistentemente aprendiz\'avel sob a medida $\mu$. Seja $\mathcal L$ uma regra satisfazendo o princ\'\i pio de minimiza\c c\~ao do erro emp\'\i rico como na defini\c c\~ao \ref{d:erm}. Segue-se que para cada $\e,\delta>0$ existe $s=s(\e,\delta)$ tal que, se $n\geq s(\e,\delta)$, ent\~ao, dado um subconjunto boreliano $D\subseteq\Omega$ qualquer, temos
\[\mbox{erro}_{\mu,D}{\mathcal L}_n(D\upharpoonright\sigma) \leq \e\]
com confian\c ca $1-\delta$?
\end{observacao}

\begin{observacao}
Enquanto as complexidades de Rademacher formam muito tempo uma ferramenta padr\~ao na an\'alise funcional e geometria de espa\c cos de Banach, na teoria de aprendizagem estat\'\i stica elas surgiram da forma explicita s\'o recentemente  \citep*{koltchinskii_panchenko,bartlett_mendelson,mendelson}.
\end{observacao}

\section{Entropia m\'etrica e testemunha de irregularidade}

\subsection{}
Nessa se\c c\~ao, vamos discutir e mostrar o teorema seguinte, cujas condi\c c\~oes (1)-(4) formam o resultado cl\'assico de Vapnik--Chervonenkis \citep*{VC:71}, enquanto a condi\c c\~ao (5) foi adicionada por Talagrand mais tarde \citep*{talagrand84,talagrand96}.

\begin{teorema}
Seja $\mathscr C$ uma classe de conceitos. As condi\c c\~oes seguintes s\~ao equivalentes.
\begin{enumerate}
\item $\mathscr C$ \'e uma classe de Glivenko--Cantelli sob a medida $\mu$.
\item Para cada $\ve>0$, $\frac 1n
\E_{\sigma}\log N(\e,{\mathscr C},L^1(\mu_\sigma))\to 0$ quando $n\to\infty$.
\item Para cada $\ve>0$, $\frac 1n\E_{\sigma}\log \sharp({\mathscr C}\upharpoonright\sigma)\to 0$ quando $n\to\infty$.
\end{enumerate}
As condi\c c\~oes seguintes tamb\'em s\~ao dois a dois equivalentes:
\begin{enumerate}
\item[($\neg$ 1)] $\mathscr C$ n\~ao \'e uma classe de Glivenko--Cantelli  sob a medida $\mu$.
\item[(4)] Existe $\alpha>0$ com a propriedade que, com alta confian\c ca, uma amostra aleat\'oria $\sigma$ tem uma fra\c c\~ao de $\geq\alpha n$ pontos fragmentados por $\mathscr C$.
\item[(5)] Existe um subconjunto $A$ de medida estritamente positiva $\alpha=\mu(A)>0$ tal que $\mu$-quase toda $n$-upla de elementos de $A$ \'e fragmentada por $\mathscr C$, qualquer que seja $n$.
\end{enumerate}
\label{t:vct}
\end{teorema}

O logaritmo aqui \'e a base $2$.

\subsection{Entropia m\'etrica. (2) $\Rightarrow$ (1)}
Mesmo se a pr\'e-compacidade da classe $\mathscr C$ ``do ponto de vista'' da medida $\mu$ n\~ao \'e suficiente para ser Glivenko--Cantelli, a pr\'e-compacidade de $\mathscr C$ estimada ``do ponto de vista coletivo'' das medidas emp\'\i ricas torna-se uma condi\c c\~ao necess\'aria e suficiente. \'E claro que para cada amostra $\sigma$, o n\'umero de cobertura $N(\e,{\mathscr C},L^1(\mu_\sigma))$ \'e limitado por $2^n$, e o logaritmo dele, por $n$. (Aqui, assumamos, pela conveni\^encia, que o n\'umero de cobertura de $\mathscr C$ seja estimado no espa\c co pseudom\'etrico maior, $(2^{\Omega},L^1(\mu_\sigma))$. De modo parecido calculamos o n\'umero de cobertura do conjunto ${\mathscr C }\upharpoonright\sigma$ de todas as rotulagens induzidas por elementos de $\mathscr C$, no cubo de Hamming correspondente). A condi\c c\~ao (2) diz que os logaritmos de n\'umeros de cobertura divididos por $n$ convergem para zero em probabilidade. Em outras palavras, a taxa m\'edia de crescimento de n\'umeros de cobertura em rela\c c\~ao com a medida emp\'irica \'e subexponencial em $n$.

Com efeito, o limite na condi\c c\~ao (2) existe para classes $\mathscr C$ quaisquer.

\begin{lema}[Lema de Fekete (1923)]
Seja $(a_n)$ uma sequ\^encia subaditiva, ou seja, $a_{m+n}\leq a_m+a_n$ para todos $m,n\in\N_+$. Ent\~ao a sequ\^encia $(a_n/n)$ \'e convergente, com
\[\lim_{n\to\infty}\frac{a_n}{n}=\inf_{n\in\N_+}\frac{a_n}{n}.\]
Na forma rigorosamente equivalente, seja $(a_n)$ uma sequ\^encia submultiplicativa de reais positivos, ou seja, $a_{m+n}\leq a_ma_n$. Ent\~ao a sequ\^encia $(a_n^{1/n})$ \'e convergente, com
\[\lim_{n\to\infty}(a_n^{1/n})=\inf_{n\in\N_+}(a_n^{1/n}).\]
\label{l:fekete}
\index{sequ\^encia! subaditiva}
\index{sequ\^encia! submultiplicativa}
\index{lema! de Fekete}
\end{lema}

\begin{proof} A vers\~ao submultiplicativa \'e um pouco mais simples a manejar, porque s\'o h\'a um caso a ser considerado. Denotemos $\alpha=\inf_{n\in\N_+}a_n^{1/n}\geq 0$. Seja $\ve>0$ qualquer. Existe $k$ tal que $\abs{a_k^{1/k}-\alpha}<\ve/2$. Seja $N$ t\~ao grande que para todo $i=1,2,\ldots,k$ $(a_i^{1/N}-1)a_k^{1/k}<\ve/2$. Agora, seja $n\geq N$ qualquer. Pode-se escrever $n=mk+i$, onde $0\leq i <k$. Temos:
\begin{eqnarray*}
\alpha\leq 
a_n^{1/n}\leq (a_k^ma_i)^{1/n}\leq {a_k}^{1/k}{a_i}^{1/N}<a_k^{1/k}+\ve/2<\alpha+\ve.
\end{eqnarray*}
A forma aditiva se infere exponenciando a sequ\^encia subaditiva, e depois tomando o logaritmo. O caso onde $\alpha=0$ corresponde, para uma sequ\^encia aditiva, ao limite $-\infty$. 
\end{proof}

\begin{exercicio} Seja $\sigma=(x_1,x_2,\ldots,x_m,x_{m+1},\ldots,x_{m+n})$ uma amostra de comprimento $m+n$. Denote por $\sigma_m$ a amostra que consiste dos primeiros $m$ elementos de $\sigma$, e por $\sigma_n$, a amostra dos elementos $x_{m+1},\ldots, x_{m+n}$. Mostre que
\[N(\e,{\mathscr C},L^1(\mu_\sigma))\leq N(\e,{\mathscr C},L^1(\mu_{\sigma_m}))\times N(\e,{\mathscr C},L^1(\mu_{\sigma_n})).\]
\end{exercicio}

\begin{exercicio}
Deduza que a sequ\^encia de valores 
\[
\E_{\sigma}\log N(\e,{\mathscr C},L^1(\mu_\sigma))\]
\'e subaditiva.
\end{exercicio}

\begin{definicao}
O valor
\begin{align*}
{\mathcal H}(\e,{\mathscr C},\mu)
&=\lim_{n\to\infty}\frac 1n
\E_{\sigma}\log N(\e,{\mathscr C},L^1(\mu_\sigma)) \\
&=\inf_{n\in\N_+}\frac 1n
\E_{\sigma}\log N(\e,{\mathscr C},L^1(\mu_\sigma))
\end{align*}
chama-se a {\em entropia m\'etrica} da classe $\mathscr C$ (em escala $\ve>0$, sob a medida $\mu$).
\index{entropia! m\'etrica}
\end{definicao}

Ent\~ao, a condi\c c\~ao (2) diz que a entropia m\'etrica de $\mathscr C$ sob $\mu$ \'e zero em cada escala. 

\begin{exercicio}
Mostre que a fun\c c\~ao 
\[
\hat {\mathcal H}(\e,{\mathscr C},\mu)(\sigma) =
\frac 1n\log N(\e,{\mathscr C},L^1(\mu_\sigma))
\]
({\em entropia m\'etrica emp\'\i rica}) \'e
Lipschitz cont\'\i nua com a constante $L\leq 1$ sobre $\Omega^n$ com rela\c c\~ao \`a dist\^ancia de Hamming normalizada. Conclua que o valor da entropia emp\'\i rica d\'a uma estimativa provavelmente aproximadamente correta (correndo o risco subgaussiano) para o valor da entropia m\'etrica.
\index{entropia! m\'etrica! emp\'\i rica}
\end{exercicio}

Note que o valor de complexidade de Rademacher emp\'\i rica em uma amostra $\sigma$ \'e completamente determinado pela restri\c c\~ao da classe $\mathscr C$ sobre $\sigma$. 

\begin{proof}[Prova da implica\c c\~ao (2) $\Rightarrow$ (1)]
\label{p:impl1implies2}
Fixemos uma amostra $\sigma$ qualquer e estimemos o valor da complexidade emp\'\i rica de Rademacher de $\mathscr C$ em $\sigma$. Seja $\ve>0$ qualquer. 
Escolhemos uma $\ve$-rede, ${\mathscr C}^\prime\subseteq \{0,1\}^n$, com $N(\e,{\mathscr C},L^1(\mu_\sigma))$ elementos, para o espa\c co m\'etrico ${\mathscr C}\upharpoonright\sigma$ munido da dist\^ancia de Hamming normalizada (ou seja, a dist\^ancia $L^1(\mu_\sigma)$). Segue-se que 
\[{\mathscr C}\upharpoonright\sigma\subseteq {\mathscr C}^\prime+B_\ve(0),\]
onde a bola aberta \'e formada no cubo de Hamming. Por conseguinte,
\begin{align*}
\hat R_n({\mathscr C})(\sigma) 
&\leq  \hat R_n({\mathscr C}^\prime)(\sigma) + \hat R_n(B_\ve(0))(\sigma) \\
\mbox{\small (lema de Massart)}
&\leq \sqrt{\frac{2\log N(\e,{\mathscr C},L^1(\mu_\sigma))}{n}}+\ve.
\end{align*}
A raiz quadrada \'e uma fun\c c\~ao c\^oncava, logo a desigualdade de Jensen se aplica:
\begin{align*}
R_n({\mathscr C}) &= \E_{\sigma} \hat R_n({\mathscr C})(\sigma) \\
&\leq \E_{\sigma}\sqrt{\frac{2\log N(\e,{\mathscr C},L^1(\mu_\sigma))}{n}}+\e \\
&\leq \sqrt{2\E_{\sigma} \frac{\log N(\e,{\mathscr C},L^1(\mu_\sigma))}{n}}+\ve
\\
&< 2\ve
\end{align*}
quando $n$ \'e bastante grande gra\c cas \`a hip\'otese (2).
\end{proof}

\subsection{Coeficiente de fragmenta\c c\~ao m\'edio. (3) $\Rightarrow$ (2)
\label{ss:cond3}}
A express\~ao em (3),
\begin{equation}
\frac 1n\E_{\sigma}\log \sharp({\mathscr C}\upharpoonright\sigma),
\label{eq:expressaoem3}
\end{equation}
pode ser vista como uma variante m\'edia do coeficiente de fragmenta\c c\~ao de $\mathscr C$:
\[s(n,{\mathscr C}) =\sup_{\sigma\in\Omega^n}\sharp({\mathscr C}\upharpoonright\sigma).\]
\index{coeficiente! de fragmenta\c c\~ao! m\'edio}
\'E claro que
\[\frac 1n\E_{\sigma}\log \sharp({\mathscr C}\upharpoonright\sigma)\leq \frac 1n\log s(n,{\mathscr C}).
\]
Observemos que o limite na condi\c c\~ao (3) existe pelas mesmas raz\~oes que na condi\c c\~ao (2). 

A condi\c c\~ao (3) \'e formalmente mais forte do que a  condi\c c\~ao (2), porque a express\~ao em (3) domina a express\~ao na condi\c c\~ao (2) para todo $\ve>0$. Por isso, (3) $\Rightarrow$ (2) trivialmente. 

\subsection{Montante m\'edio de fragmenta\c c\~ao. ($\neg$ 4) $\Rightarrow$ (3) e (1) $\Rightarrow$ ($\neg$4)}
A quantidade $\alpha\geq 0$ tal que uma amostra aleat\'oria $\sigma$ tenha uma parte com $\sim\alpha n$ pontos fragmentada por $\mathscr C$ foi introduzida por Vapnik e Chervonenkis. A propriedade reflete o princ\'\i pio geral segundo o qual mais fragmenta\c c\~ao significa estar mais longe de uma classe de Glivenko--Cantelli. (As m\'edias de Rademacher s\~ao uma outra realiza\c c\~ao da mesma ideia). Por exemplo, se a classe $\mathscr C$ tiver a dimens\~ao VC finita, ent\~ao $\alpha=0$, qualquer que seja a medida subjacente $\mu$. 

Primeiramente, vamos mostrar a exist\^encia da constante $\alpha\geq 0$ para cada classe $\mathscr C$.

\begin{exercicio}
Mostre que a sequ\^encia de reais
\begin{eqnarray*}
\E_{\sigma_n} \VC({\mathscr C}\upharpoonright\sigma) =
\E_{\sigma_n}\max\{d\colon \exists I\subseteq [n],~\sharp I=d,~
{\mathscr C}\mbox{ fragmenta }\{x_i\colon i\in I\}\}
\end{eqnarray*}
\'e subaditiva. Deduza que a sequ\^encia
\[\E_{\sigma_n}\frac 1n\VC({\mathscr C}\upharpoonright\sigma)
\]
converge para um valor $\alpha=\alpha({\mathscr C},\mu)\in [0,1]$ quando $n\to\infty$.
\end{exercicio}

\begin{exercicio}
Mostre que a fun\c c\~ao
\[\sigma\mapsto \frac 1n\VC({\mathscr C}\upharpoonright\sigma)\]
\'e Lipschitz cont\'\i nua sobre $\Omega^n$ munido da dist\^ancia de Hamming normalizada, com $L=1$. Deduza que para qualquer que seja $\ve>0$, se $n$ for bastante grande, com alta confian\c ca uma amostra aleat\'oria tem um subconjunto de $\geq (1-\ve)\alpha$ pontos fragmentado por $\mathscr C$.
\end{exercicio}

\begin{proof}[Prova da implica\c c\~ao ($\neg$4) $\Rightarrow$ (3)] 
Suponha que $\frac {d_\sigma}n $ converge para zero em probabilidade (ou: em esperan\c ca), onde ${d_\sigma}=\VC({\mathscr C}\upharpoonright\sigma)$ \'e a ``dimens\~ao VC emp\'\i rica'' da classe $\mathscr C$, uma vari\'avel aleat\'oria. De acordo com o lema de Sauer-Shelah, segue-se que a vari\'avel aleat\'oria
\begin{eqnarray}
\frac 1n\log \sharp \left({\mathscr C}\upharpoonright\sigma\right)&\leq& \frac 1n\log
\left(\frac{en}{d_\sigma}\right)^{d_\sigma} \nonumber \\
&=& \frac {{d_\sigma}}n\left(1-\log \frac {{d_\sigma}}n\right) \label{eq:sauerbound}
\end{eqnarray}
converge para zero em probabilidade tamb\'em.
\end{proof}

O caso importante \'e o onde a classe $\mathscr C$ tem a dimens\~ao VC finita. Neste caso, ${d_\sigma}\leq d<\infty$, e em particular, a entropia m\'etrica converge para zero com uma taxa independente da medida $\mu$. \'E uma observa\c c\~ao b\'asica para {\em distribution-free learning}.

\begin{proof}[Prova da implica\c c\~ao (1) $\Rightarrow$ ($\neg$4)]
Mostramos a contrapositiva: (4) $\Rightarrow$ ($\neg$1). Suponha que, com alta confian\c ca, uma amostra aleat\'oria $\sigma$ tem uma subamostra, $\sigma_I$, $I\subseteq [n]$,  fragmentada por $\mathscr C$, com $\abs I\geq\alpha n$ pontos. O valor da complexidade de Rademacher emp\'\i rica, $\hat R_{\lceil \alpha n\rceil}(\mathscr C)(\sigma_I)$, em $\sigma_I$ atinge o valor m\'aximo, $1/2$. \'E f\'acil de concluir que $\hat R_n(\mathscr C)(\sigma)\geq \alpha/2>0$, com alta confian\c ca. Por conseguinte, n\~ao ocorre a converg\^encia em probabilidade de $\hat R_n(\mathscr C)$ para zero, e a classe $\mathscr C$ n\~ao \'e G-C.
\end{proof}

\subsection{Testemunha de irregularidade. (5) $\Rightarrow$ (4)
\label{ss:testemunha}}
Assumindo a condi\c c\~ao (5), segundo a lei de grandes n\'umeros, se $n$ for bastante grande, ent\~ao, com alta confian\c ca, uma amostra aleat\'oria cont\'em uma fra\c c\~ao de $\sim\alpha n$ pontos que pertencem a $A$, logo, s\~ao fragmentados.  Por conseguinte, (5) implica (4). Portanto, formalmente, a condi\c c\~ao (5) \'e mais forte do que (4). N\~ao somente uma propor\c c\~ao fixa de elementos de cada amostra \'e fragmentada, mas as amostras fragmentadas s\~ao bem situadas em $\Omega$. 
Deste modo, a condi\c c\~ao (5) de Talagrand \'e bastante surpreendente. 

\begin{definicao}
Um subconjunto boreliano $A$ de $\Omega$ \'e dito {\em testemunha de irregularidade} se $\mu(A)>0$ e quase toda amostra finita de pontos de $A$ \'e fragmentada, ou seja,
\[\forall n\in\N,~\mu^{\otimes n}\{\sigma=(x_1,\ldots,x_n)\in A^n,~\sigma\mbox{ n\~ao \'e fragmentada por }\mathscr C\}=0.\]
\index{testemunha! de irregularidade}
\end{definicao}

A fragmenta\c c\~ao entende-se como a de uma amostra {\em ordenada:}
\[\forall I\subseteq [n],~\exists C\in {\mathscr C},~\forall i\in [n],~i\in I\iff \chi_C(x_i)=1.\]
Em particular, se $\sigma$ \'e fragmentada por $\mathscr C$, ent\~ao os elementos de $\sigma$ s\~ao dois a dois distintos. 

\begin{exercicio}
Deduzir da nossa defini\c c\~ao que uma testemunha de irregularidade nunca cont\'em \'atomos.
\end{exercicio}

\subsection{(4) $\Rightarrow$ (4$\sfrac 12$), uma condi\c c\~ao t\'ecnica intermedi\'aria}
Neste momento, temos a equival\^encia das condi\c c\~oes (1), (2), (3), e ($\neg$ 4), assim como a implica\c c\~ao (5) $\Rightarrow$ (4). 
A \'ultima implica\c c\~ao que nos resta, (4) $\Rightarrow$ (5), vai ser quebrada em duas.
Definamos uma condi\c c\~ao t\'ecnica intermedi\'aria.

\begin{enumerate}
\item[(4$\sfrac 12$)] {\em Seja $A_n$ o subconjunto de $\Omega^n$ de todas as amostras fragmentadas por $\mathscr C$. Ent\~ao existe uma constante $\alpha>0$ tal que}
\[\forall n,~\mu^{\otimes n}(A_n)\geq \alpha^n.\]
\end{enumerate}

\begin{exercicio}
Mostre que a sequ\^encia $(\mu^{\otimes n}(A_n))$ \'e submultiplicativa, e deduza do lema de Fekete \ref{l:fekete} que a sequ\^encia 
\[\left(\mu^{\otimes n}(A_n)\right)^{1/n}\]
converge para o seu \'\i nfimo, o valor m\'aximo de $\alpha$ na propriedade acima. Denotemos
\[ r\left((A_n)_{n=1}^{\infty}\right)=\lim_{n\to\infty}\left(\mu^{\otimes n}(A_n)\right)^{1/n}.\]
As condi\c c\~oes seguintes s\~ao equivalentes:
\begin{itemize}
\item para todos $n$, $\mu^{\otimes n}(A_n)\geq\alpha^n$, e
\item $r\left((A_n)_{n=1}^{\infty}\right)\geq\alpha$.
\end{itemize}
\label{ex:alpha}
\end{exercicio}

\begin{proof}[Prova da implica\c c\~ao (4) $\Rightarrow$ (4$\sfrac 12$)]
Seja $n\in\N$ fixo qualquer. Para uma amostra aleat\'oria $W=(X_1,X_2,\ldots,X_n)$, onde $X_i$ s\~ao elementos aleat\'orios i.i.d. do dom\'\i nio $\Omega$, seguindo a lei $\mu$, vamos estimar por baixo a probabilidade do evento $W\in A_n$. A amostra $W$ pode ser vista como o fragmento inicial de uma amostra aleat\'oria com $N$ elementos, onde $N\geq n$:
\begin{align*}
P[W\in A_n] =
\mu^{\otimes N}\{\sigma=&(x_1,x_2,\ldots,x_n,\ldots,x_N)\in\Omega^N\colon  
\\ & 
(x_1,x_2,\ldots,x_n)\mbox{ \'e fragmentado por }\mathscr C\}.
\end{align*}
Cada permuta\c c\~ao $\tau\in S_N$ de $N$ coordenadas conserva a medida de produto $\mu^{\otimes N}$, e por conseguinte, qualquer que seja $\tau$,
\[P[W\in A_n] = \mu^{\otimes N}\{\sigma=(x_1,\ldots,x_N)\in\Omega^N\colon  
(x_{\tau(1)},x_{\tau(2)},\ldots,x_{\tau(n)})\mbox{ \'e fragmentado}\}.\]
Podemos igualmente, sem mudar o valor, tomar a esperan\c ca relativa a medida uniforme sobre o grupo $S_N$ de permuta\c c\~oes:
\begin{align*}
P[W\in A_n]&=
\E_{\tau}\mu^{\otimes N}\{\sigma\in\Omega^N\colon  
(x_{\tau(1)},x_{\tau(2)},\ldots,x_{\tau(n)})\mbox{ \'e fragmentado}\}\\
&= \E_{\tau}\E_{\sigma\sim\mu^{\otimes N}}\chi_{\{\sigma\colon 
 (x_{\tau(1)},x_{\tau(2)},\ldots,x_{\tau(n)})\mbox{\footnotesize ~\'e fragmentado}\}} \\
&= \E_{\sigma\sim\mu^{\otimes N}}\E_{\tau}\chi_{\{\sigma\colon 
 (x_{\tau(1)},x_{\tau(2)},\ldots,x_{\tau(n)})\mbox{\footnotesize ~\'e fragmentado}\}} \\
&= \E_{\sigma\sim\mu^{\otimes N}} P_{\tau}[(x_{\tau(1)},x_{\tau(2)},\ldots,x_{\tau(n)})\mbox{ \'e fragmentado}~~\vert~~ \sigma].
\end{align*}
Usamos o ``truque de condicionamento'' (``conditioning trick'', veja observa\c c\~ao \ref{o:truquedecond}): na express\~ao interior os valores de $\sigma$ s\~ao fixos, $X_1=x_2,X_2=x_2,\ldots,X_n=x_n$, e a probabilidade refere-se ao espa\c co probabil\'\i stico finito $S_N$ munido da medida uniforme, $\mu_{\sharp}$. Dada a amostra determin\'\i stica, $\sigma$, escolhemos um subconjunto $D\subseteq [n]$ da cardinalidade $d=d(\sigma)=\VC(\mathscr C\upharpoonright\sigma)$, fragmentado por $\mathscr C$. Agora, temos
\[P_{\tau}[(x_{\tau(1)},x_{\tau(2)},\ldots,x_{\tau(n)})\mbox{ \'e fragmentado}]\geq P[x_{\tau(1)},x_{\tau(2)},\ldots,x_{\tau(n)}\in D].\]
Trata-se de um problema da primeira aula de um curso de probabilidade: uma urna cont\'em $N$ bolas, incluindo exatamente $d$ vermelhas, e uma pessoa retira $n$ bolas aleatoriamente. Qual \'e a probabilidade do que todas as bolas retiradas sejam vermelhas? A probabilidade \'e dada pela express\~ao
\[\frac dN\cdot\frac{d-1}{N-1}\cdot\frac{d-2}{N-2}\cdot\ldots\cdot \frac{d-n+1}{N-n+1},\]
de onde
\begin{eqnarray*}
P[W\in A_n] &\geq& \E_{\sigma\sim\mu^{\otimes N}}\frac {d(\sigma)}N\cdot\frac{{d(\sigma)}-1}{N-1}\cdot\frac{{d(\sigma)}-2}{N-2}\cdot\ldots\cdot \frac{{d(\sigma)}-n+1}{N-n+1}\\
&\to & \alpha^n\mbox{ quando }N\to\infty.
\end{eqnarray*}

\end{proof}

\subsection{Lema de Fremlin. (4$\sfrac 12$) $\Rightarrow$ (5)}

\begin{lema}
Sejam $\Omega$ um conjunto e $(A_n)$ uma sequ\^encia de subconjuntos, $A_n\subseteq\Omega^n$, $n\in\N_+$. As condi\c c\~oes seguintes s\~ao equivalentes.
\begin{enumerate}
\item Existe uma classe $\mathscr C$ tal que para todo $n$, $A_n$ consiste de todas as amostras de comprimento $n$ fragmentadas por $\mathscr C$.
\item Para todos $m,n$, $m\leq n$, os conjuntos $A_n$ s\~ao invariantes pelas permuta\c c\~oes de coordenadas, 
\[\tau(A_n)\subseteq A_n,~~\tau\in S_n,\]
assim como pelas proje\c c\~oes:
\[\pi^n_m(A_n)\subseteq V_m,~\mbox{ onde }~\pi^n_m(x_1,x_2,\ldots,x_m,\ldots,x_n)= (x_1,x_2,\ldots,x_m).\]
\end{enumerate}
\end{lema}

\begin{proof}
A implica\c c\~ao (1) $\Rightarrow$ (2) \'e evidente, e para mostrar (2)  $\Rightarrow$ (1), basta criar a classe $\mathscr C$ que consiste de todos os conjuntos finitos $\{x_1,x_2,\ldots,x_n\}$ tendo a propriedade $(x_1,x_2,,\ldots,x_n)\in A_n$, $n\in\N$.
\end{proof}

Vamos chamar provisoriamente (s\'o dentro desta subse\c c\~ao) uma sequ\^encia $(A_n)$ de conjuntos borelianos que possui uma das propriedades (1) ou (2) uma sequ\^encia {\em fragment\'avel}.

\begin{exercicio}
Mostrar que cada medida boreliana sigma-aditiva, $\mu$, satisfaz a propriedade: se $(B_n)$ \'e uma sequ\^encia decrescente de conjuntos borelianos encaixados, ent\~ao
\[\mu\left(\cap_nB_n\right) = \lim\mu(B_n).\]
Uma propriedade an\'aloga tamb\'em tem lugar para as sequ\^encias crescentes.
\label{ex:crescentes}
\end{exercicio}
 
\begin{lema}
Seja $(A_n)$ uma sequ\^encia fragment\'avel tal que 
$\mu^{\otimes n}(A_n)\geq\alpha^n$, ou seja, $r(A_n)\geq\alpha$, onde $\alpha\in\R_+$.
Ent\~ao existe uma sequ\^encia fragment\'avel $(A^\prime_n)$, tal que $A^\prime_n\subseteq A_n$, $r(A^\prime_n)\geq\alpha$, e $(A^\prime_n)$ \'e minimal com estas propriedades, ou seja, dado uma sequ\^encia fragment\'avel $(W_n)$ tal que $W_n\subseteq A^\prime_n$ e $r(W_n)\geq\alpha$, temos para todo $n$
\[\mu^{\otimes n}(A^\prime_n\setminus W_n)=0.\]
\end{lema}

\begin{proof} Vamos definir $A^\prime_n$ recursivamente. Para $n=1$, definamos
\[s=\sup_{(W_n)_n} \mu(A_1\setminus W_1),\]
onde o supremo \'e formado sobre todas as sequ\^encias fragment\'aveis $(W_n)_{n=1}^{\infty}$ tendo as propriedades $W_n\subseteq A_n$ e $r(W_n)\geq\alpha$. Se $s=0$, ent\~ao colocamos $A^\prime_1=A_1$. 
Se $s>0$, ent\~ao para cada $k\in\N_+$ existe uma sequ\^encia fragment\'avel $(W^k_n)$ tal que $W^k_n\subseteq A_n$, $\mu^{\otimes n}(W^k_n)\geq\alpha^n$, e $\mu(A_1\setminus W^k_1)>s-1/k$. Verifica-se facilmente que a sequ\^encia de conjuntos
\[W_n=\bigcap_{k=1}^{\infty}W^k_n\]
\'e fragment\'avel, tenha a propriedade $\mu^{\otimes n}(W_n)\geq\alpha^n$, e qualquer que seja uma sequ\^encia fragment\'avel $(Z_n)$ com $\mu^{\otimes n}(Z_n)\geq\alpha^n$ e $Z_n\subseteq A_n$, tenhamos necessariamente $\mu(W_1\setminus Z_1)=0$. Escolhemos $A^\prime_1=W_1$ e continuamos recursivamente. 
\end{proof}

\begin{teorema}[Lema de Fremlin]
Seja $\Omega$ um espa\c co probabil\'\i stico padr\~ao, e seja $(A_n)$ uma sequ\^encia fragment\'avel tal que
\[\mu^{\otimes n}(A_n)\geq\alpha^n\mbox{ para todo }n.\]
Ent\~ao existe um conjunto boreliano, $A\subseteq\Omega$, tal que $\mu(A)=\alpha$ e para todo $n$, $\mu^{\otimes n}(A_n\setminus A^n)=0$, ou seja,  a menos de um conjunto negligenci\'avel, $A^n$ \'e contido em $A_n$.
\index{lema! de Fremlin}
\end{teorema}

Vamos supor sem perda de generalidade que $(A_n)$ \'e uma sequ\^encia fragment\'avel minimal com a propriedade $r(A_n)\geq\alpha$. Para cada $x\in\Omega$ e $n\in\N_+$, definamos um conjunto boreliano $A_{n,x}$, como a fatia vertical de $V_{n+1}$ sobre $x$:
\[A_{n,x}=\{\sigma\in\Omega^n\colon x\sigma\in A_{n+1}\},\]
onde $x\sigma$ \'e a concatena\c c\~ao de $x$ com $\sigma$. 
Verifica-se facilmente que a sequ\^encia $(A_{n,x})$ \'e fragment\'avel e tem a propriedade $A_{n,x}\subseteq A_n$. Logo, o valor de $r((A_{n,x})_n)$ \'e bem definido, e n\~ao pode ultrapassar $\alpha$:
\[r((A_{n,x})_n)=\lim_{n\to\infty}\left(\mu^{\otimes n}A_{n,x}\right)^{1/n}\leq\alpha.\]
O nosso conjunto desejado \'e agora definido como segue:
\[A=\{x\in\Omega\colon r((A_{n,x})_n)=\alpha\}.\]

Primeiramente, vamos verificar que $\mu(A)\geq\alpha$. Para obter uma contradi\c c\~ao, suponha que $\mu(A)=\alpha^\prime<\alpha$.
Usando o exerc\'\i cio \ref{ex:crescentes}, podemos escolher um valor $\alpha_1$ tal que $\alpha^\prime<\alpha_1<\alpha$ e
\[\mu\{x\in\Omega\colon r((A_{n,x})_n)\geq\alpha_1\}<\alpha_1,\]
ou seja
\[\mu\{x\in\Omega\colon r((A_{n,x})_n)<\alpha_1\}\geq 1-\alpha_1.\]
Definamos
\[B_n=\{x\in\Omega\colon \forall N\geq n,~\left(\mu^{\otimes N}A_{N,x}\right)^{1/N}<\alpha_1\}.\]
Usando exerc\'\i cio \ref{ex:crescentes}, conclu\'\i mos: $\lim_{n\to\infty}\mu(B_n)\geq 1-\alpha_1$. Por conseguinte,
se $n$ \'e bastante grande,
\[\mu(\Omega\setminus B_n)<\frac12(\alpha+\alpha_1).
\]
Temos
\begin{align*}
\mu^{\otimes (n+1)}(A_{n+1}) 
&= \int_{\Omega}\mu^{\otimes n}(A_{n,x})\,d\mu(x) \\
&= \int_{B_n} + \int_{\Omega\setminus B_n} \\
&< \alpha_1^n + \frac 12(\alpha+\alpha_1)\mu^{\otimes n}(A_{n}) \\
&= \alpha^n\left(\frac{\alpha_1^n}{\alpha^n}+ \frac 12(\alpha+\alpha_1)\frac{\mu^{\otimes n}(A_{n})}{\alpha^n} \right) \\
&\leq  \left(\frac 34\alpha +\frac 14 \alpha_1\right)\mu^{\otimes n}(A_{n})
\end{align*}
quando $n$ \'e bastante grande, porque $\alpha_1^n/\alpha_n\to 0$, enquanto $\mu^{\otimes n}(V_{n})\geq\alpha^n$. Conclu\'\i mos indutivamente: para $n$ bastante grande e fixo, e $N > n$ qualquer, 
\[\mu^{\otimes (N)}(V_{N})\leq \left(\frac 34\alpha +\frac 14 \alpha_1\right)^{N-n}\mu^{\otimes n}(A_{n}),\]
 e por conseguinte,
\[r((A_N)_{N})=\lim\left(\mu^{\otimes N}(A_{N})\right)^{1/N}\leq \frac 34\alpha +\frac 14 \alpha_1<\alpha,\]
 uma contradi\c c\~ao.

Ent\~ao, $\mu(A)\geq\alpha$. Segundo a defini\c c\~ao de $A$, para todo $x\in A$, a sequ\^encia fragment\'avel $(A_{n,x})_{n=1}^{\infty}$ satisfaz $r((A_{n,x})_n)=\alpha$ e $A_{n,x}\subseteq A_n$. Como $(A_n)$ \'e minimal, temos $\mu^{\otimes n}(A_n\setminus A_{n,x})=0$. 
Segue-se, indutivamente em $k$, que qualquer que seja $\varsigma\in A^k$, $\mu$-quase certamente o conjunto
\[V_{n,\varsigma}=\{\sigma\in\Omega^n \colon \varsigma\sigma\in V_{n+k}\}\]
satisfaz a propriedade $\mu^{\otimes n}(A_n\setminus V_{n,\varsigma})=0$. Mas isso implica que para todo $k$, $\mu$-quase todas amostras $\varsigma\in A^k$ pertencem a $A_n$. O lema de Fremlin \'e mostrado.

A implica\c c\~ao procurada (4$\sfrac 12$) $\Rightarrow$ (5) segue-se imediatamente no caso onde $A_n$ consistem de todas as amostras fragmentadas por $\mathscr C$. Isso termina a prova do teorema \ref{t:vct}.

Agora pode-se formular a conjetura seguinte.
 
\begin{conjetura}
Uma classe de conceitos $\mathscr C$ \'e consistentemente PAC aprendiz\'avel sob uma medida de probabilidade $\mu$ se e somente se, para cada testemunha de irregularidade $A$, o espa\c co m\'etrico quociente da restri\c c\~ao ${\mathscr C}\upharpoonright A$ na dist\^ancia $L^1(\mu\vert A)$ \'e um conjunto unit\'ario:
\[\forall C,D\in {\mathscr C},~\mu((C\Delta D)\cap A)=0.\]
\label{conj:pacaprend}
\end{conjetura}

\begin{exercicio}
Mostre que as testemunhas de irregularidade podem ser muitas distintas no mesmo dom\'\i nio, disjuntas ou parcialmente sobrepondo-se. Construir um exemplo de um espa\c co probabil\'\i stico padr\~ao, $(\Omega,\mu)$, bem como uma classe de conceitos $\mathscr C$, tais que $\Omega$ cont\'em v\'arias testemunhas de irregularidade.
\end{exercicio}

Pode ser que, dado uma classe $\mathscr C$ no dom\'\i nio $(\Omega,\mu)$, existe uma aplica\c c\~ao quociente can\^onica $\phi$ de $\Omega$ para um outro espa\c co probabil\'\i stico, $(\Upsilon,\nu)$, munido de uma classe $\mathscr D$, de modo que $\nu=\phi_{\ast}(\mu)$, $\mathscr C=\phi^{-1}(\mathscr D)$, e $\mathscr C$ \'e consistentemente PAC aprendiz\'avel se e somente se $\mathscr D$ \'e Glivenko--Cantelli? Informalmente, uma tal aplica\c c\~ao hipot\'etica, $\phi$, colaria cada testemunha de irregularidade em um ponto de $\Upsilon$. 

\section{Aprendizagem independente de distribui\c c\~ao\label{s:pacindepmedida}}

A parte mais importante da teoria da aprendizagem dentro de uma classe \'e a aprendizagem independente da distribui\c c\~ao subjacente (``distribution-free learning''). O conceito significa que a complexidade amostral n\~ao depende da medida $\mu$, e \'e uniforme sobre o conjunto $P(\Omega)$ de todas as medidas de probabilidade borelianas. Esta abordagem justifica-se plenamente, dado que a medida $\mu$ \'e geralmente desconhecida. Todas as no\c c\~oes e resultados principais tem vers\~oes uniformes, independentes de uma medida. Vamos come\c car as formula\c c\~oes.

\begin{definicao}
Uma regra de aprendizagem $\mathcal L$ \'e provavelmente aproximadamente correta (PAC) em rela\c c\~ao \`a classe de conceitos $\mathscr C$ (independentemente de medida), se para todo $\e>0$, todo $C\in {\mathscr C}$ e toda medida de probabilidade boreliana $\mu$ sobre $\Omega$,
\begin{equation}
P\left[\mu\left({\mathcal L(\sigma,C\upharpoonright\sigma)}\,\Delta\, C\right)>\e\right] \to 0\mbox{ uniformemente em $\mu$ e $C$.}
\end{equation}
Em outras palavras,
\begin{equation}
\label{eq:pac}
\sup_{\mu\in P(\Omega)}\sup_{C\in {\mathscr C}}
\mu^{\otimes n}\left\{\sigma\in\Omega^n\colon 
\mu\left({\mathcal L(\sigma,C\upharpoonright\sigma)}\,\Delta\, C\right)>\e \right\} \to 0\mbox{ quando }n\to\infty,\end{equation}
ou, usando uma nota\c c\~ao mais econ\^omica,
\[\sup_{\mu\in P(\Omega)}
\E_{\sigma\sim\mu}
\sup_{C\in {\mathscr C}}
\mbox{erro}_{\mu,C}{\mathcal L}(C\upharpoonright\sigma)\to 0\mbox{ quando }n\to\infty.
\]

Da maneira equivalente, existe uma fun\c c\~ao $s(\e,\delta)$ ({\em complexidade amostral} de $\mathcal L$, independente da medida) tal que para todo $C\in{\mathscr C}$ e toda $\mu\in P(\Omega)$, uma amostra i.i.d. $\sigma\sim\mu$ de tamanho $n\geq s(\e,\delta)$ satisfaz $\mu(C\,\Delta\, {\mathcal L}(\sigma,C\upharpoonright\sigma))<\e$ com confian\c ca $\geq 1-\delta$.
\end{definicao}

\begin{definicao}
Uma classe $\mathscr C$ \'e {\em PAC aprendiz\'avel} se existe uma regra de aprendizagem PAC para $\mathscr C$, independente de medida.
\end{definicao}

\begin{observacao}
A classe $\mathscr C$ constru\'\i da na subse\c c\~ao \ref{ss:difusas}, que \'e PAC aprendiz\'avel sob toda medida de probabilidade $\mu$ no dom\'\i nio, n\~ao satisfaz a defini\c c\~ao acima, pois cada vez a regra de aprendizagem \'e diferente, ou seja, ela depende da medida. De fato, a classe n\~ao \'e PAC aprendiz\'avel independentemente da medida, pois a taxa de aprendizagem, como foi mostrado, pode ser arbitrariamente lenta.
\end{observacao}

\begin{definicao}
Uma classe $\mathscr C$ \'e {\em consistentemente PAC aprendiz\'avel} se toda regra de aprendizagem consistente com $\mathscr C$ \'e provavelmente aproximadamente correta, independentemente de medida.
\end{definicao}

Precisamos tamb\'em formular uma vers\~ao uniforme (independente da medida) de uma classe de Glivenko--Cantelli.

\begin{definicao}
Uma classe de conceitos, $\mathscr C$, sobre um dom\'\i nio $\Omega$ \'e uma c {\em classe de Glivenko-Cantelli uniforme} se, dado $\ve>0$ (precis\~ao) e $\delta>0$ (risco), existe um valor $s=s(\e,\delta,{\mathscr C})$ tal que, qualquer que seja $n\geq s$, temos para toda medida de probabilidade $\mu$ sobre $\Omega$:
\begin{equation}
\label{eq:gcc}
\mu^{\otimes n}\left[\sup_{C\in{\mathscr C}} \left\vert \frac 1n \sum_{i=1}^n \chi_C(X_i)-\mu(C)\right\vert>\ve
\right]\leq \delta,
\end{equation}
onde $X_1,X_2,\ldots,X_n$ s\~ao elementos aleat\'orias de $\Omega$ i.i.d. distribu\'\i dos segundo a lei $\mu$. 

De maneira equivalente e mais econ\^omica,
\[\sup_{\mu\in P(\Omega)}\E_{\mu}\sup_{C\in {\mathscr C}}\left\vert \mu_{\sigma}(C)-\mu(C) \right\vert\to 0\mbox{ quando }n\to\infty.
\]
\end{definicao}

O resultado seguinte \'e a pedra angular da teoria de aprendizagem dentro de uma classe \citep*{VC:71,BEHW}.
  
\begin{teorema}
As seguintes proposi\c c\~oes s\~ao equivalentes:
\begin{enumerate}
\item\label{vc:1} 
$\mathscr C$ \'e PAC aprendiz\'avel.
\item\label{vc:2} $\mathscr C$ \'e consistentemente PAC aprendiz\'avel.
\item\label{vc:3} $\mathscr C$ \'e uma classe de Glivenko--Cantelli uniforme.
\item \label{vc:3a} $\sup_{\mu\in P(\Omega)}R_n({\mathscr C},\mu)\to 0$ quando $n\to\infty$.
\item\label{vc:4} $\VC(\mathscr C)<\infty$.
\end{enumerate}
\label{t:maintVC}
\end{teorema}

Vamos apresentar uma prova circular, no sentido anti-hor\'ario.

(\ref{vc:1}) $\Rightarrow$ (\ref{vc:4}). Seja $\mathscr C$ uma classe PAC aprendiz\'avel independentemente da medida, com a complexidade amostral $s(\e,\delta)$. Qualquer seja a medida de probabilidade $\mu$ sobre $\Omega$,
segundo o teorema \ref{t:benedek-itai} de Benedek--Itai, temos
\[\log_2 D(2\e,{\mathscr C},L^1(\mu)))-1\leq s(\e,\delta),
\]
quando $\delta\leq 1/2$. Seja $\sigma$ uma amostra com $n$ pontos, fragmentada pela classe $\mathscr C$. Aplicando a desigualdade acima \`a medida emp\'\i rica $\mu=\mu_\sigma$, ou seja, a medida uniforme suportada sobre $\sigma$, e usando a proposi\c c\~ao \ref{p:empacotamento} com $\e=1/8$ e $\delta=1/2$, deduzimos
\[\frac n 8\leq \log 2\left[s(1/8,1/2)\right] + 1,\]
de onde
\[\VC(\mathscr C)\leq 8 \log 2\left[s(1/8,1/2)\right] + 1.
\]

(\ref{vc:4}) $\Rightarrow$ (\ref{vc:3a}).
Veja a subse\c c\~ao \ref{ss:cond3}, eq. (\ref{eq:sauerbound}).

(\ref{vc:3a}) $\Rightarrow$ (\ref{vc:3}). Segue-se do teorema \ref{t:rademacher}.

(\ref{vc:3}) $\Rightarrow$ (\ref{vc:2}). Usamos o corol\'ario \ref{c:consistente}. Note que a taxa de aprendizagem \'e uniforme para todas as regras de aprendizagem consistentes. 

(\ref{vc:2}) $\Rightarrow$ (\ref{vc:1}). Segue da observa\c c\~ao seguinte.

\begin{proposicao}
Cada classe $\mathscr C$ admite pelo menos uma regra de aprendizagem consistente.
\end{proposicao}

\begin{proof}
Uma forma equivalente do Axioma de Escolha (AC) diz que todo conjunto $X$ admite uma boa ordem, ou seja, uma ordem total, $\prec$, tal que cada subconjunto n\~ao vazio de $X$ tem o elemento m\'\i nimo com rela\c c\~ao \`a ordem $\prec$:
\[\forall Y,~\emptyset \neq Y\subseteq X \Rightarrow \exists y\in Y~
\forall z\in Y~y\prec z.\]
Fixemos uma tal ordem sobre $\mathscr C$. Dada uma amostra $\sigma$ rotulada por um conceito que pertence a $\mathscr C$, a fam\'\i lia ${\mathscr C}_\sigma$ de todos os conceitos $C$ induzindo a mesma rotulagem sobre $\sigma$ n\~ao \'e vazia, logo possui o elemento minimal, que escolhemos como ${\mathscr L}(\sigma)$. A regra $\mathscr L$ \'e consistente.
\end{proof}

(Vamos reexaminar este resultado na pr\'oxima se\c c\~ao.)

\begin{exercicio}
N\'os j\'a caracterizamos as classes $\mathscr C$ que possuem a propriedade de Glivenko--Cantelli (uniforme) sob a classe de todas as medidas, assim como sob a medida fixa. As primeiras s\~ao as de dimens\~ao VC finita, as \'ultimas s\~ao as classes que n\~ao possuem a testemunha de irregularidade de Talagrand. Ambas propriedades est\~ao limitando a quantidade de fragmenta\c c\~ao pela classe. Agora estude o problema seguinte: quais condi\c c\~oes sobre uma classe $\mathscr C$, tamb\'em exprimidas na linguagem de fragmenta\c c\~ao, s\~ao necess\'arias e suficientes para $\mathscr C$ possuir a propriedade de Glivenko--Cantelli uniforme sob a classe de todas as medidas {\em n\~ao at\^omicas}?

O problema foi sugerido em \citep*{vidyasagar}, problema 12.8. Depois de refletir um pouco, voc\^e pode comparar a sua solu\c c\~ao com a sugerida em \citep*{pestov:13tcs}.
\end{exercicio}
  
\section{Tomar cuidado com a mensurabilidade\label{s:cuidado}}

Nesta se\c c\~ao, vamos fazer algo de paradoxal: construiremos contraexemplos aos teoremas principais deste cap\'\i tulo.

\subsection{Dois contraexemplos\label{ss:dois}}

Lembramos que uma ordem total, $\preceq$, \'e chamada uma {\em boa ordem} se todo subconjunto n\~ao vazio de $\Omega$ tem m\'\i nimo. Uma boa ordem sobre $X$ \'e {\em minimal} se todo segmento aberto inicial, $\{x\in X\colon x\prec y\}$, tem a cardinalidade estritamente menor do que $X$. Para um conjunto $X$ infinito, a condi\c c\~ao equivalente \'e que todo segmento fechado inicial,
\[I_x=\{y\in X\colon y\preceq x\},\]
tem a cardinalidade estritamente menor do que a de $X$. 

Segundo uma forma equivalente do Axioma de Escolha (AC), o {\em Princ\'\i pio de Boa Ordem} ({\em Well-Ordering Principle}), todo conjunto admite uma boa ordem minimal (se\c c\~ao \ref{ss:boaordem}).

\begin{exemplo}[\citet*{DD},Proposition 2.2]
\label{ex:dd}
Assumamos a Hip\'otese do Cont\'\i nuo (CH). Ou seja, n\~ao existem cardinalidades estritamente intermedi\'arias entre a cardinalidade de $\N$ e a de $\R$ (se\c c\~ao \ref{s:CH}).

Seja $\Omega$ um espa\c co boreliano padr\~ao n\~ao enumer\'avel, isto \'e, um espa\c co boreliano isomorfo ao intervalo $[0,1]$. 
Escolhemos uma boa ordem minimal, $\preceq$, sobre $\Omega$. \'E \'util a observar que a ordem $\preceq$ n\~ao tem nada a ver com as ordens usuais bem conhecidas, tais como a ordem usual, $\leq$, sobre $\R$. Sob a Hip\'otese do Cont\'\i nuo, todo segmento fechado inicial, $I_y$, tendo uma cardinalidade menor que $\mathfrak c = 2^{\aleph_0}$, \'e enumer\'avel. 
Formamos uma classe de conceitos $\mathscr C$ que consiste de todos os segmentos iniciais, $I_y$, $y\in \Omega$. \'E claro que a dimens\~ao VC da classe $\mathscr C$ \'e um.

Agora, seja $\mu$ uma medida de probabilidade boreliana n\~ao at\^omica sobre $\Omega$ (por exemplo, a medida de Lebesgue sobre o intervalo $[0,1]$). Sob CH, todo elemento de $\mathscr C$ \'e um conjunto enumer\'avel, logo boreliano, de medida zero (um conjunto nulo).

Ao mesmo tempo, para cada amostra finita $\sigma$, existe um segmento inicial $I_y\in {\mathscr C}$ que cont\'em todos os elementos de $\sigma$: basta definir $y=\max_{\prec}\{x_1,x_2,\ldots,x_n\}$. A medida emp\'\i rica de $C$ com rela\c c\~ao a $\sigma$ \'e um. Conclu\'\i mos: nenhuma amostra finita pode advinhar a medida de todos os elementos de $\mathscr C$ com uma precis\~ao $\e<1$ qualquer e uma confian\c ca n\~ao nula. A classe $\mathscr C$ n\~ao \'e uma classe de Glivenko--Cantelli, apesar de ter a dimens\~ao VC um. 
\label{ex:DD}
\end{exemplo}

\begin{observacao}
Com efeito, no exemplo acima, CH n\~ao \'e necess\'aria, o mesmo argumento \'e v\'alido sobre uma hip\'otese bem mais fraca, o Axioma de Martin (MA).
Eu n\~ao sei se um exemplo com as mesmas propriedades pode ser constru\'\i do dentro o sistema ZFC, sem hip\'oteses conjunt\'\i sticas adicionais.
\end{observacao}

O exemplo pode ser modificado, para obter um exemplo de uma classe de conceitos de dimens\~ao VC finita (um) que n\~ao \'e consistentemente PAC aprendiz\'avel.

\begin{exemplo}[\citet*{BEHW}, p. 953]
\label{ex:behw}
Mais uma vez, sobre CH, adicione \`a classe de conceitos $\mathscr C$ do exemplo \ref{ex:DD} o conjunto $\Omega$ como um elemento. Denotemos a nova classe ${\mathscr C}^\prime={\mathscr C}\cup\{\Omega\}$. Sempre temos $\VC(\mathscr C^\prime)=1$. A toda amostra finita rotulada, $(\sigma,\tau)$, associamos a hip\'otese
\begin{equation}
\label{eq:rule}
{\mathcal L}(\sigma,\tau) = I_z,~~z=  \max_{\prec}\tau.
\end{equation}
\'E f\'acil a ver que a regra de aprendizagem $\mathcal L$ \'e consistente com a classe ${\mathscr C}^\prime$. 

Ao mesmo tempo, $\mathcal L$ n\~ao \'e provavelmente aproximadamente correta.
Para o conceito $C=\Omega$ a regra ${\mathcal L}(\sigma,\Omega\upharpoonright\sigma)={\mathcal L}(\sigma,\sigma)$ vai sempre gerar um conceito enumer\'avel da forma $I_y$, $y=\max\sigma$. Por conseguinte, quando a medida $\mu$ \'e n\~ao at\^omica, $\mu(\Omega\,\Delta\, I_y)=\mu(\Omega\setminus I_y)=1$. O conceito $C=\Omega$ n\~ao pode ser aprendido com a precis\~ao $\e<1$ e uma confian\c ca n\~ao nula. Ao mesmo tempo, a dimens\~ao VC da classe ${\mathscr C}^\prime$ \'e um. 
\end{exemplo}

Como \'e poss\'\i vel que os mesmos resultados possam ser mostrados e refutados ao mesmo tempo? Onde reside o problema?
A fim de entender a situa\c c\~ao, vamos discutir um paradoxo importante, descoberto na matem\'atica bastante recentemente.

\subsection{Paradoxo de Chris Freiling: uma ``refuta\c c\~ao'' da Hip\'otese do Cont\'\i nuo}

Vamos discutir um argumento not\'avel
contra a validade da Hip\'otese do Cont\'\i nuo, sugerido por Chris Freiling \citep*{freiling}.

Considere uma experi\^encia mental de {\em jogar dardo,} que est\'a entre os passatempos favoritos de probabilistas (em terceiro lugar somente depois de lan\c car a moeda e lan\c car um dado). Voc\^e joga um dardo em um alvo, por exemplo, o quadrado de
lado unit\'ario, $[0,1]^2$. O resultado de uma experi\^encia \'e um ponto, 
$x\in [0,1]^2$, atingido pelo dardo. Suponha que a probabilidade de atingir uma regi\~ao, $A$, \'e proporcional \`a \'area (a medida uniforme) dela:
\[ P[ X\in A ] =\mbox{area}(A)=\lambda^{\otimes 2}(A),\]
onde $\lambda^{\otimes 2}$ \'e a medida de Lebesgue.
Desse modo, a distribui\c c\~ao de pontos atingidos pelo dardo segue a lei uniforme no quadrado. Em particular, a medida uniforme (de produto) \'e n\~ao at\^omica: a probabilidade de atingir um ponto particular \'e zero.

Suponha que a Hip\'otese do Cont\'\i nuo seja v\'alida. Como uma consequ\^encia, existe uma boa ordem $\prec$ sobre o quadrado, $[0,1]^2$, com a propriedade do que, qualquer que seja $x\in [0,1]^2$, o conjunto de pontos que precedem $x$,
\[I_x = \{y\in [0,1]^2\colon y\preceq x\},\]
\'e enumer\'avel:
\[\forall x\in [0,1]^2,~~ \abs{I_x}\leq\aleph_0.\]
Lembremos, mais uma vez, que esta ordem n\~ao \'e construtiva, n\~ao h\'a algum algoritmo para constru\'\i -la. \'E uma ordem cuja exist\^encia segue do Axioma de Escolha (p\'agina \pageref{page:axiomadaescolha}) (o Princ\'\i pio de Boa Ordem de Zermelo \ref{t:well-ordering}).
 
Suponha agora dois jogadores de dardo, A e B, ou (no esp\'\i rito da
criptografia de chave p\'ublica), Alice e Bob, est\~ao jogando dardos,
independentemente um do outro. Em outras palavras, h\'a
dois independentes vari\'aveis aleat\'orias, $X$ e $Y$, distribu\'\i das uniformemente no quadrado $[0,1]^2$. 

Lembra-se que a independ\^encia de vari\'aveis aleat\'orias $X$ e $Y$ significa que para cada par de subconjuntos borelianos do quadrado, digamos $\Pi_1$ e $\Pi_2$, os eventos $[X\in \Pi_1]$ e $[Y\in \Pi_2]$ s\~ao independentes, ou seja,
\[P[X\in \Pi_1\wedge Y\in \Pi_2] = P[X\in \Pi_1]\cdot P[Y\in \Pi_2].\]
Por conseguinte, n\~ao importa se Alice lan\c ca seu dardo primeiro, ou Bob faz:
de nenhum jeito isto afetar\'a o resultado da experi\^encia. Por
exemplo, pode ser que Alice tem jogado o seu dardo h\'a dez anos, o
resultado (um elemento $x\in [0,1]^2$) foi gravado e mantido em segredo, e agora Bob lan\c ca seu dardo, e depois os resultados nos s\~ao encaminhados simultaneamente, em envelopes selados, por pessoas diferentes. \'E poss\'\i vel tamb\'em que Alice e Bob n\~ao saibam da exist\^encia um do outro, e mesmo que viverem -- e jogarem dardos -- em duas extremidades diferentes da terra. 

Em qualquer caso, depois de eles jogarem dardos, recebemos dois n\'umeros: $a\in [0,1]^2$ para Alice, e $b\in [0,1]^2$ para Bob. H\'a duas possibilidades: $a\preceq b$ ou $b\preceq a$. Consideraremos ambos os casos.

Suponha primeiro que $a\preceq b$. Como os dois eventos eram independentes, podemos assumir sem perda de generalidade que Alice jogou seu dardo primeira.
Qual \'e a probabilidade de que, quando $a$ for fixado, o resultado do Bob
preceder $a$? Como o conjunto de predecessores de $a$, que n\'os
denotamos $S_a$, \`a enumer\'avel, a probabilidade do evento 
\'e zero:
\[P[Y\in I_a]=0.\]
\'E um evento improv\'avel. 

Agora suponha que $b\preceq a$. Dessa vez, podemos assumir igualmente bem que Bob tinha jogado seu dardo primeiro e n\'os tenhamos o seu resultado, $b$. Qual \'e a probabilidade de que o resultado de Alice seja menor do que o do Bob, $a\preceq b$? O conjunto $I_b$ \'e tamb\'em enumer\'avel, logo
\[P[X\in I_b]=0.\] 
Deste modo, conclu\'\i mos que ambos os resultados, $a\prec b$ e $b\prec a$, s\~ao improv\'aveis. Por conseguinte, o evento 
\[(a\preceq b) \vee (b\preceq a)\]
\'e improv\'avel. Mas claramente {\em ou} $a\preceq b $ {\em ou}
$b\prec a$ sempre acontece, ent\~ao a probabilidade do evento acima deve ser um! Chegamos a uma contradi\c c\~ao. Ela resulta da Hip\'otese do Cont\'\i nuo. Conclu\'\i mos: a CH \'e falsa.
\qed
\index{paradoxo! de Freiling}

O argumento de Freiling {\em refuta} a CH? N\~ao realmente: enganamos um
pouquinho. Se voc\^e olhar para a ``impress\~ao fina'' do argumento, torna-se
claro que ele n\~ao \'e uma boa prova rigorosa.

Queremos falar da probabilidade do evento que o resultado da primeira experi\^encia precede, com rela\c c\~ao \`a ordem $\prec$, o resultado
da segunda experi\^encia: $a\preceq b $. Ou seja, estamos interessados no
probabilidade do evento
\[ [X\preceq Y]. \]
A probabilidade \'e igual \`a medida uniforme do conjunto de todos os pares $(a,b)\in [0,1]^2\times [0,1]^2$ tendo a propriedade $a\preceq b$:
\[P[X\preceq Y] = \lambda^{\otimes 2}\{(a,b)\colon a\preceq b\}\subseteq [0,1]^2\times [0,1]^2.
\]
\'E neste momento particular que o nosso argumento intuitivo perde o rigor matem\'atico: o conjunto acima, ou seja, o gr\'afico da rela\c c\~ao $\preceq$, n\~ao \'e mensur\'avel, e n\~ao pode ser atribu\'\i do nenhum valor da medida! A probabilidade do evento $[X\preceq Y]$ \'e indefinida. O argumento de Freiling n\~ao pode ser traduzido em uma prova.

Concretamente, denotemos
\[A = \{(a,b)\colon a\preceq b\}.\]
Para todo $b\in [0,1]^2$, a ``fatia horizontal'', $A_b=\{a\in [0,1]^2\colon (a,b)\in A\}$, \'e enumer\'avel, logo tem medida nula. Se $A$ for mensur\'avel, ter\'\i amos
\[\lambda^{\otimes 2}(A) = \int_{[0,1]^2}\lambda (A_b)\, d\lambda(b) = 0,\]
e de modo semelhante, a medida do conjunto complementar seria nula tamb\'em, uma contradi\c c\~ao. O que o argumento de Freiling mostra, \'e s\'o que $A$ n\~ao \'e mensur\'avel.

No entanto, o significado do argumento acima \'e muito maior do que um truque divertido. Estamos apenas nos atendo a um tecnicismo puro. E se um dia os matem\'aticos conseguirem construir uma nova base axiom\'atica que incorporaria tanto os conjuntos quanto as vari\'aveis aleat\'orias em um fundamento, o argumento de Freiling se tornaria um {\em teorema}, mostrando rigorosamente que a Hip\'otese do Cont\'\i nuo seja {\em falsa}. 

Freiling ele mesmo prop\^os incorporar nos fundamentos da matem\'atica um axioma adicional, o chamado {\em axioma da simetria} ({\em AX}). 

Deixe-me concluir a subse\c c\~ao com uma opini\~ao sobre o argumento de Freiling pertencente \`a David Mumford \citep*{mumford}.

\begin{quote} {\it ``Why is not this result as widely known as G\"odel's
and Cohen's, I don't know. (...) The Continuum Hypothesis is surely
similar to the scholastic issue on how many angels can stand on the head
of a pin: an issue which disappears if you change your point of view.''}
\end{quote}

\subsection{Classes universalmente separ\'aveis\label{ss:univsep}}
O paradoxo de Freiling nos d\'a uma chave para entender os exemplos aparentemente contradit\'orios: a Hip\'otese do Cont\'\i nuo n\~ao combina bem com mensurabilidade.
Examinando as provas de nossos teoremas desta se\c c\~ao, conclu\'\i mos que o argumento de simetriza\c c\~ao n\~ao funciona para a classe $\mathscr C$ constru\'\i da na subse\c c\~ao \ref{ss:dois}. Pode-se mostrar (usando o mesmo argumento simples que acima) que para nossa classe, a express\~ao
\[\E_{\sigma,\sigma^\prime\sim\mu}\sup_{C\in {\mathscr C}}\abs{\mu_\sigma(C)-\mu_{\sigma^\prime}(C)} \]
\'e mal definida, pois a fun\c c\~ao 
\[\sup_{C\in {\mathscr C}}\abs{\mu_\sigma(C)-\mu_{\sigma^\prime}(C)} \]
sobre $\Omega^n\times \Omega^n$ n\~ao \'e mensur\'avel. 
Isso mostra que algumas hip\'oteses de mensurabilidade adicionais sobre a classe de conceitos, $\mathscr C$, s\~ao inevit\'aveis.

H\'a muitas formas de contornar esta restri\c c\~ao. Escolhemos a mais simples delas, a no\c c\~ao de uma classe universalmente separ\'avel.

\begin{definicao}
Uma fam\'\i lia de conjuntos borelianos, $\mathscr C$, de um espa\c co boreliano padr\~ao, $\Omega$, \'e {\em universalmente separ\'avel} se existe uma subfam\'\i lia enumer\'avel, ${\mathscr C}^\prime$, tal que todo $C\in {\mathscr C}$ \'e o limite de uma sequ\^encia $(C_n)$ de elementos de ${\mathscr C}^\prime$ na topologia de converg\^encia pontual:
\[\forall x\in\Omega,~\exists N~\forall n\geq N,~\chi_{C_n}(x)=\chi_C(x).\]
(Diz-se que ${\mathscr C}^\prime$ \'e {\em universalmente denso} em $\mathscr C$).
\index{classe! universalmente! separ\'avel}
\index{classe! universalmente! densa}
\end{definicao}

Vamos deixar como um exerc\'\i cio a verifica\c c\~ao que para uma classe universalmente separ\'avel todos os conjuntos que aparecem nas provas deste cap\'\i tulo s\~ao mensur\'aveis. De maneira ilustrativa, s\'o vamos mostrar uma propriedade t\'\i pica de mensurabilidade de tais classes.

\begin{proposicao}
\label{p:usep}
Cada classe universalmente separ\'avel admite uma regra de aprendizagem consistente que satisfaz a condi\c c\~ao  de mensurabilidade seguinte. Para todo conjunto boreliano $B\in{\mathscr B}$, toda medida de probabilidade $\mu\in P(\Omega)$, e todo $n\in\N_+$, a fun\c c\~ao
\begin{equation}
\label{eq:measurabilityl}
\Omega^n\ni\sigma\mapsto \mu\left({\mathcal L(\sigma,C\upharpoonright\sigma)}\,\Delta\, B\right) \in \R 
\end{equation}
\'e mensur\'avel.
\end{proposicao}

\begin{proof}
Seja $\mathscr C$ uma classe universalmente separ\'avel. Selecionamos uma subclasse enumer\'avel e universalmente densa de $\mathscr C$,
\[{\mathscr C}^\prime = \{C_1,C_2,\ldots,C_n,\ldots\}.\]
Definamos uma regra de aprendizagem, $\mathcal L$, como segue: dado uma amostra rotulada de tamanho $n$, $(\sigma,\tau)$, definamos
\[{\mathcal L}_n(\sigma,\tau) = \begin{cases}\emptyset,&\mbox{ se }\tau\notin {\mathscr C}\upharpoonright\sigma,\\
C_{\min\{k\colon C_k\upharpoonright\sigma = \tau \}},&\mbox{ caso contr\'ario.}
\end{cases}
\]
Informalmente, o algoritmo examina todos os conceitos $C_1$, $C_2$, $\ldots$ nesta ordem, e para quando o conceito induzindo a rotulagem desejada for achado. 

\'E f\'acil a ver que a classe ${\mathscr C}^\prime$ induz todas as mesmas rotulagens que $\mathscr C$:
\[{\mathscr C}\upharpoonright\sigma = {\mathscr C}^\prime\upharpoonright\sigma.\]
Logo, a aplica\c c\~ao $\mathcal L$ \'e bem definida. \'E claro que a regra $\mathcal L$ \'e consistente com a classe $\mathscr C$. Para verificar que $\mathcal L$ satisfaz a condi\c c\~ao de mensurabilidade, definamos para todo $C\in {\mathscr C}$ e todo $n$ o conjunto
\[A_C = \{(\sigma,\tau)\colon \sigma\in\Omega^n,~\tau = C\upharpoonright\sigma\}\subseteq\Omega^n\times\{0,1\}^n.\]
Dado um $\tau\in \{0,1\}^n$, a ``fatia vertical'' correspondente de $A_C$,
\[A_C\cap \left(\Omega^n\times\{\tau\}\right),\]
pode ser escrita como
\[C_1\times C_2\times\ldots\times C_n\times \{\tau\},\]
onde $C_i$ \'e igual a $C$ se $\tau_i=1$ e $C_i=\Omega\setminus C$ se $\tau_i=0$. Por isso, $A_C$ \'e um conjunto boreliano. 

Definamos para todos $k$
\[B_k = \{(\sigma,\tau)\colon {\mathcal L}_n(\sigma,\tau) = C_k\}\]
e 
\[B_0 = \Omega\setminus\bigcup_{k=1}^\infty B_k.\]
Temos
\[B_1 = A_{C_1},~~B_2 = A_{C_2}\setminus A_{C_1},\ldots,\]
\[B_k = A_{C_k}\setminus\bigcup_{i=1}^{k-1}A_{C_i},\ldots,\]
logo todos os conjuntos $B_k$ s\~ao borelianos, incluindo $B_0$. Eles formam uma parti\c c\~ao boreliana enumer\'avel de $\Omega$, e agora basta mostrar que a nossa fun\c c\~ao
\[\Omega\ni\sigma\mapsto \mu\left({\mathcal L(\sigma,C\upharpoonright\sigma)}\,\Delta\, B\right) \in \R ,\]
\'e mensur\'avel quando restrita sobre cado peda\c co boreliano, $B_k$, desta parti\c c\~ao. Mas sobre cada peda\c co, a fun\c c\~ao \'e constante: sobre $B_k$, $k\geq 1$ ela toma o valor
\[\mu(C_k\,\Delta\,B), \]
enquanto sobre $B_0$ ela toma o valor $\mu(B)$.
\end{proof}

O seguinte pode ser mostrado com algumas modifica\c c\~oes cosm\'eticas.

\begin{proposicao}
Toda classe $\mathscr C$ universalmente separ\'avel admite uma regra de aprendizagem $\mathcal L$ mensur\'avel que segue o princ\'\i pio de minimiza\c c\~ao de perda emp\'\i rica.  \qed
\end{proposicao}

Desse modo, segue-se que todos os resultados de nosso cap\'\i tulo restam v\'alidos sob a hip\'otese de que a classe $\mathscr C$ seja universalmente separ\'avel.

\begin{observacao}
\'E interessante que, sob CH (ou MA), pode-se mostrar \citep*{pestov2010} que a classe ${\mathscr C}^\prime$ do exemplo \ref{ex:behw} \'e PAC aprendiz\'avel independentemente da medida.

Redefinamos a boa ordem sobre $\mathscr C=\{I_x\colon x\in\R\}\cup\{\Omega\}$ fazendo $\Omega$ o elemento minimal (em vez do maximal), e preservando a rela\c c\~ao de ordem entre outros elementos. Denotemos a nova ordem $\prec_1$. Definamos a regra de aprendizagem $\mathcal L_1$ de mesmo modo que na eq. (\ref{eq:rule}), mas dessa vez o m\'\i nimo \'e entendido no sentido da ordem $\prec_1$:
\begin{equation}
\label{eq:rule1}
{\mathcal L}_1(\sigma,\tau) = \min_{(\prec_1)}\left\{C\in{\mathscr C}\colon C\upharpoonright\sigma = \bigcap_{\tau\subseteq D}D\right\}.\end{equation} 

Para ver o que faz a diferen\c ca, seja $\mu$ uma medida n\~ao at\^omica sobre $\Omega$. Se $C=\Omega$, ent\~ao para toda amostra $\sigma$ consistentemente rotulada com $\Omega$ a regra $\mathcal L_1$ vai produzir $C$, porque \'e o conceito m\'\i nimo induzindo a rotulagem consistente. Se $C\neq\Omega$, ent\~ao para $\mu$-quase todas amostras $\sigma$ a rotulagem sobre $\sigma$ produzida por $C$ \'e vazia, \'e o conceito ${\mathcal L}_1(\sigma,\emptyset)$ gerado pela regra $\mathcal L_1$, enquanto possivelmente diferente de $C$, ser\'a um conceito enumer\'avel, significando que $\mu(C\,\Delta\,{\mathcal L}(\sigma,\emptyset))=0$. 

Para dar uma prova formal que $\mathcal L_1$ \'e PAC, note que, quaisquer que sejam $C\in {\mathscr C}^\prime$ e $n\in\N$, a cole\c c\~ao de conceitos dois a dois distintos ${\mathcal L}_1(\sigma\upharpoonright C)$, $\sigma\in\R^n$ \'e enumer\'avel (sob CH), pois elas s\~ao todos contidos no segmento inicial da ordem $\prec_1$ do conjunto $\mathscr C$ de cardinalidade cont\'\i nuo. Por conseguinte, a classe de conceitos
\begin{equation}
\label{eq:lc}
{\mathcal L}_1^{C}=\{{\mathcal L}_1(\sigma\cap C)\colon\sigma\in\R^n,n\in\N\}\subseteq {\mathscr C}^\prime\end{equation}
\'e enumer\'avel (assumindo CH). A dimens\~ao VC da classe ${\mathcal L}_1^{C}\cup\{C\}$ \'e $\leq 1$, e por ser enumer\'avel, logo universalmente separ\'avel, ela \'e uniforme Glivenko--Cantelli com a complexidade amostral usual $s(\e,\delta,1)$. Por conseguinte, dado $\e,\delta>0$, se $n$ \'e bastante grande, temos para cada medida de probabilidade $\mu$ sobre $\Omega$ e para cada $\sigma\in\Omega^n$
\[\mu(C\,\Delta\,{\mathcal L}(\sigma,C\cap\sigma))<\e\]
quando $n\geq N(\e,\delta,1)$.
\end{observacao}

%
%

\chapter{Consist\^encia universal\label{c:kNN}}

O assunto deste cap\'\i tulo \'e o classificador $k$-NN de $k$ vizinhos mais pr\'oximos em um espa\c co m\'etrico separ\'avel.

\section{Classificador $1$-NN de vizinhos mais pr\'oximos}

Seja $\Omega$ um espa\c co m\'etrico.
Dado um ponto $x\in\Omega$ \'e uma amostra n\~ao rotulada $\sigma$, denotemos por $\mbox{NN}_\sigma(x)$ o vizinho mais pr\'oximo de $x$ em $\sigma$. No caso de empates, ou seja, alguns pontos $x_i$ que todos atingem o valor da dist\^ancia $d(x,\sigma)$, vamos desempatar, por exemplo, escolhendo $x_i$ com o menor \'\i ndice $i$, ou escolhendo um dos vizinhos aleatoriamente e uniformemente. Pelo momento a estrat\'egia de desempate n\~ao \'e importante. Quando $\Omega$ \'e munido de uma medida de probabilidade boreliana, $\mu$, vamos denotar por $W_n$ uma amostra aleat\'oria com $n$ elementos i.i.d., seguindo a lei $\mu$.

\begin{teorema}[Lema de Cover--Hart \citep*{cover_hart}]
Seja $\mu$ uma medida de probabilidade boreliana sobre um espa\c co m\'etrico separ\'avel $(\Omega,d)$. Ent\~ao, quase certamente, 
\begin{equation}
\mbox{NN}_{W_n}(X)\to X\mbox{ quando }n\to\infty.
\label{eq:caminho}
\end{equation}
\label{t:cover-hart}
\index{lema! de Cover--Hart}
\end{teorema}

Vamos esclarecer a proposi\c c\~ao, reformulando-a na linguagem conjunt\'\i stica e da teoria de medida. Um {\em caminho amostral} \'e um elemento qualquer,
\[\varsigma=(x_1,x_2,\ldots,x_n,\ldots),\]
do produto cartesiano infinito $\Omega^{\infty}=\Omega^{\N_+}$ de $\N_+$ c\'opias do dom\'\i nio, $\Omega$. Ou seja, $\varsigma$ \'e simplesmente uma sequ\^encia de elementos do dom\'\i nio. Todo segmento inicial de um caminho amostral, 
\[\varsigma_n=(x_1,x_2,\ldots,x_n)\in\Omega^n,\]
\'e uma amostra.
\index{caminho! amostral}
Na presen\c ca de uma medida de probabilidade $\mu$ sobre $\Omega$,
o caminho amostral torna-se uma vari\'avel aleat\'oria seguindo a lei $\mu^\infty=\mu^{\otimes \N_+}$:
\[W=(X_1,X_2,\ldots,X_n,\ldots)\in \Omega^{\infty}.\] A express\~ao na eq. (\ref{eq:caminho}) depende da vari\'avel aleat\'oria $(X,X_1,X_2,\ldots,X_n,\ldots)=(W,X)$, combinando o caminho amostral $W$ com um elemento aleat\'orio $X$ do dom\'\i nio, que \'e independente do caminho. O significado probabil\'\i stico da eq. (\ref{eq:caminho}) \'e que
\[P[\lim_{n\to\infty}d(X,\mbox{NN}_{W_n}(X))=0]=1.\]
(Aqui, $W_n$ \'e o segmento inicial de $W$.)
A reformula\c c\~ao conjunt\'\i stica:
\begin{equation}
(\mu\otimes \mu^{\infty})\left\{(x,\varsigma)\in\Omega\times\Omega^{\infty}\colon
\lim_{n\to\infty}d(x,\mbox{NN}_{\varsigma_n}(x))=0 \right\}=1,
\end{equation}
onde
\[d(x,\mbox{NN}_{\varsigma_n}(x))=d(x,\varsigma_n)=\min_{i=1}^n d(x,x_i).\]

\begin{exercicio}
Mostre que o conjunto
\[\left\{(x,\varsigma)\in\Omega\times\Omega^{\infty}\colon
\lim_{n\to\infty}d(x,\varsigma_n)=0 \right\}\]
\'e boreliano.
\end{exercicio}

\begin{definicao}
Seja $\mu$ uma medida boreliana sobre um espa\c co m\'etrico $\Omega$. Um ponto $x\in\Omega$ pertence ao {\em suporte de} $\mu$ se o valor da medida de toda bola aberta em torno de $x$ \'e estritamente positivo:
\[x\in\mbox{supp}\,\mu\iff \forall \ve>0~\mu(B_{\ve}(x))>0.\]
\index{suporte! da medida}
\end{definicao}

\begin{exercicio}
Mostre que o suporte de uma medida \'e um conjunto fechado.
\end{exercicio}

\begin{exercicio}
Mostre que o suporte de uma medida boreliana sobre um espa\c co m\'etrico qualquer \'e separ\'avel.
\par
[ {\em Sugest\~ao:} observe que dado $\ve>0$, cada fam\'\i lia $\ve$-discreta de pontos do suporte da medida \'e necessariamente enumer\'avel... ]
\end{exercicio}

\begin{exercicio} Seja $\mu$ uma medida de probabilidade sobre um espa\c co m\'etrico separ\'avel. Mostre que
\begin{equation}
\mu\left(\mbox{supp}\,\mu\right)=1.
\label{eq:suportemedida1}
\end{equation}
\label{ex:suppmu}
\end{exercicio}

\begin{observacao} 
Este \'e o resultado que exige a separabilidade do espa\c co $\Omega$. Mais exatamente, ele vale para qualquer espa\c co m\'etrico cuja densidade n\~ao \'e um cardinal {\em mensur\'avel.} Uma das condi\c c\~oes equivalentes \`a defini\c c\~ao de cardinal mensur\'avel, $\tau$, \'e a seguinte: um espa\c co m\'etrico $\Omega$ de cardinalidade $\tau$, munido da m\'etrica zero-um, admite uma medida de probabilidade boreliana n\~ao at\^omica (uma medida de probabilidade definida sobre todos os conjuntos de $\Omega$, tendo a propriedade que a medida de cada conjunto unit\'ario \'e zero). O suporte de uma tal medida \'e obviamente vazio!
\index{cardinal! mensur\'avel}

Portanto, em ZFC, n\~ao somente a exist\^encia de cardinais mensur\'aveis n\~ao pode ser mostrada, mas at\'e mesmo a consist\^encia do sistema ``ZFC $+$ existem cardinais mensur\'aveis''. 
\label{o:cardmensuravel}
\end{observacao}

\begin{exercicio}[$\ast$] 
Seja $\mu$ uma medida de probabilidade sobre espa\c co m\'etrico cuja densidade n\~ao \'e um cardinal mensur\'avel. Mostre que vale a conclus\~ao na eq. (\ref{eq:suportemedida1}). 
\par
[ {\em Sugest\~ao:}  todo espa\c co m\'etrico admite uma base topol\'ogica que \'e a uni\~ao de uma sequ\^encia $\gamma_n$ de fam\'\i lias discretas (para todo $x$ existe $\ve>0$ tal que $B_{\ve}(x)$ encontra um elemento de $\gamma_n$ ao m\'aximo). (\'E uma metade do teorema de metriza\c c\~ao de Bing.) Tente mostrar antes de consultar, por exemplo, \citep*{engelking}, sect. 5.4. Supondo que a medida de suporte de $\mu$ \'e menor que $1$, e aplicando o resultado acima ao espa\c co $X\setminus\mbox{supp}\,\mu$, mostre que existe uma fam\'\i lia discreta (em particular, disjunta) de abertos negligenci\'aveis cuja uni\~ao tem a medida estritamente positiva. Deduza que o cardinal $\abs\gamma$ \'e mensur\'avel. ]
\end{exercicio}

Deste modo, tudo se reduz ao caso separ\'avel. Isso significa que a nossa hip\'otese de $\Omega$ ser separ\'avel n\~ao \'e restritiva. 

\begin{proof}[Prova do lema de Cover--Hart \ref{t:cover-hart}] Seja $x\in \mbox{supp}\,\mu$ fixo qualquer, e seja $\ve>0$. Para um elemento aleat\'orio $X_n\sim\mu$, a probabilidade do evento $X_n\notin B_{\ve}(x)$ \'e igual a $1-\mu(B_\ve(x))<1$. Logo, para um caminho amostral aleat\'orio $W=(X_1,X_2,\ldots,X_n,\ldots)$, qualquer seja $n$,
\begin{align*}
P[\forall n~X_n\notin B_\ve(x)]
&\leq
 P[X_1\notin B_\ve(x),\ldots,X_n\notin B_\ve(x)]\\
&=
(1-\mu(B_\ve(x)))^n\to 0\mbox{ quando }n\to\infty.
\end{align*}
A sequ\^encia de reais $d(\mbox{NN}_{\varsigma_n},x)$ \'e mon\'otona. Logo,
com a probabilidade um, $\mbox{NN}_{\varsigma_n}\in B_\ve(x)$ se $n\geq N$, onde $N$ \'e bastante grande. Aplicando esta observa\c c\~ao para a sequ\^encia de valores $\ve=1/k$ e usando o fato que a interse\c c\~ao de uma sequ\^encia de conjuntos de medida um tem medida um, conclu\'\i mos que, com probabilidade um, $\mbox{NN}_{\varsigma_n}\to x$.
(Aqui, a probabilidade refere-se \`a medida $\mu^\infty$ sobre o espa\c co de caminhos amostrais). 

Segundo o teorema de Fubini \ref{t:fubini}, e usando o exerc\'\i cio \ref{ex:suppmu},
\begin{align*}
(\mu\otimes \mu^{\infty})\left\{(x,\varsigma)\colon
\lim_{n\to\infty}d(x,\varsigma_n)=0 \right\} 
&=
\int_{\Omega}\mu^{\infty}\left\{(x,\varsigma)\colon
\lim_{n\to\infty}d(x,\varsigma_n)=0 \right\}\,d\mu(x) \\
&=
 \int_{\mbox{supp}\,\mu}+\int_{\Omega\setminus \mbox{supp}\,\mu}
\\
&= 1+0=1.
\end{align*}
\end{proof}

\begin{observacao}
Seria interessante saber se aprendizagem \'e poss\'\i vel nos espa\c cos mensur\'aveis com medida n\~ao separ\'aveis (por exemplo, nos espa\c cos probabil\'\i sticos de Loeb \citep*{loeb}), talvez, sob um modelo de aprendizagem diferente.
\end{observacao}

\begin{exercicio}
Analise onde os axiomas de uma m\'etrica foram usados na demonstra\-\c c\~ao do lema de Cover--Hart. Tente formular e mostrar vers\~oes do resultado para ``medidas de semelhan\c ca'' $d(x,y)$ que n\~ao satisfazem todos os axiomas da m\'etrica.
\index{medida! de semelhan\c ca}
\end{exercicio}

Aqui, uma consequ\^encia imediata do lema de Cover--Hart.

\begin{proposicao}
Seja $\mu$ uma medida de probabilidade boreliana sobre um espa\c co m\'etrico separ\'avel qualquer, $\Omega$. Se $C$ \'e um conceito boreliano em $\Omega$ tal que a fronteira de $C$,
\[\mbox{fr}(C)=\bar C\setminus \overset\circ C,\]
\'e negligenci\'avel,
\[\mu(\mbox{fr}(C))=0,\]
ent\~ao o classificador de vizinhos mais pr\'oximos aprende $C$, e a converg\^encia de erro para zero \'e quase certa:
\[\mu({\mathcal L}^{\mbox{\tiny NN}}_n(C\upharpoonright \varsigma_n)\Delta C)\overset{q.c.}\to 0\mbox{ quando }n\to\infty.\]
\index{fronteira de um conjunto}
\label{p:fronteiranula}
\end{proposicao}

\begin{observacao}
A converg\^encia quase certa \'e uma propriedade mais forte do que a converg\^encia em probabilidade. Ela significa que o erro converge para zero ao longo de quase todo caminho amostral. De fato, temos, para quase todos $x\in\Omega$ e quase todos os caminhos amostrais $\varsigma$,
\begin{equation}
{\mathcal L}^{\mbox{\tiny NN}}_n(C\upharpoonright \varsigma_n)(x) \to \chi_C(x).
\label{eq:elltochiCx}
\end{equation}
Segundo o teorema de Lebesgue sobre converg\^encia dominada (teorema \ref{t:dominada}), a converg\^encia quase certa implica a converg\^encia em probabilidade.
\index{converg\^encia! quase certa}
\end{observacao}

\begin{proof}[Demonstra\c c\~ao da proposi\c c\~ao \ref{p:fronteiranula}]
O conjunto aberto $\overset\circ C$ \'e rotulado identicamente com o r\'otulo $1$, e o conjunto aberto $\Omega\setminus\bar C$ \'e rotulado com $0$. A uni\~ao de dois conjuntos tem medida plena (um). Seja $x$ um elemento qualquer desta uni\~ao. Suponha que $x$ perten\c ca ao interior de $C$. Segundo o lema de Cover--Hart, para quase todos os caminhos amostrais, existe um $n$ tal que $X_n$ toma valor dentro do interior de $C$, e desse modo, o vizinho mais pr\'oximo de $x$ vai receber o r\'otulo $1$. Por conseguinte, a regra $1$-NN vai associar a $x$ o r\'otulo correto, $1$, a partir deste $n$. Igualmente, para cada elemento $x$ de $\Omega\setminus\bar C$ e quase todos os caminhos amostrais $\varsigma$, a regra ${\mathcal L}^{\mbox{\tiny NN}}_n$ vai associar a $x$ o r\'otulo correto, $0$, a partir de um $n$. O teorema de Fubini \ref{t:fubini} permite a concluir que o conjunto de pares $(x,\varsigma)$ com a propriedade de que a partir de um $n$ o r\'otulo assinado a $x$ \'e correto, tem a medida um.
\end{proof}

O resultado se aplica a cada espa\c co m\'etrico separ\'avel, sem restri\c c\~ao nenhuma. Infelizmente, ele n\~ao \'e muito \'util, pois num sentido ``a maioria'' dos conjuntos borelianos tem uma fronteira de medida n\~ao nula.

\begin{exemplo}
O {\em conjunto de Cantor,} $C$,
\index{conjunto! de Cantor}
\'e obtido do intervalo fechado $[0,1]$ removendo o ter\c co m\'edio, $(1/3,2/3)$, e em seguida, removendo os ter\c cos m\'edios $(1/9,2/9)$ e $(7/9,8/9)$, de dois intervalos restantes, e cetera. Mais precisamente, a constru\c c\~ao \'e recursiva. Definamos $C_0=[0,1]$. Para todo $n\in\N$, o conjunto $C_n$ \'e a uni\~ao de $2^n$ intervalos fechados de comprimento $3^{-n}$ cada um. Denotemos $C_{n+1}$ o conjunto obtido de $C_n$ eliminando os ter\c cos m\'edio de todos os intervalos que formam $C_n$. Finalmente, definamos
\[C =\bigcap_{n=0}^\infty C_n.\] 

\begin{figure}
\centering
\scalebox{0.25}{\includegraphics{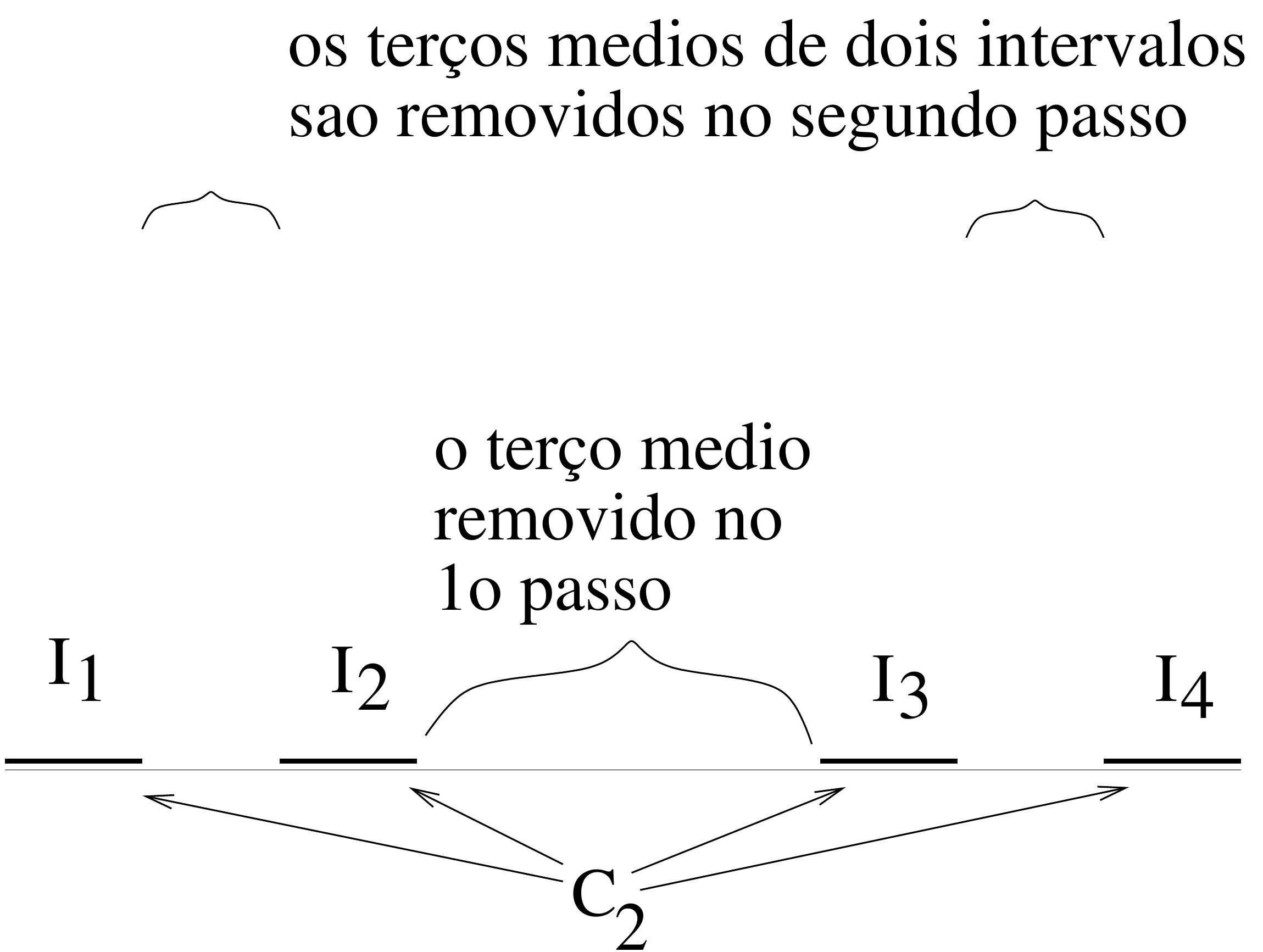}}
\caption{A constru\c c\~ao do conjunto de Cantor, depois $n=2$ passos.}
\label{fig:cantor}
\label{ex:conjuntodecantor}
\end{figure}

O comprimento total de $C_n$ \'e igual a $(2/3)^n$, logo a medida de Lebesgue de $C$ est nula. No entanto, \'e um objeto muito interessante.

Os pontos extremos de todos os intervalos abertos removidos pertencem a $C$; eles se chamam os {\em pontos de primeira esp\'ecie}. Eles formam um conjunto enumer\'avel. O restante de elementos de $C$ chamam-se os {\em pontos de segunda esp\'ecie.} Para ver que eles existem, note que os pontos removidos podem se caraterizar como todos os pontos cuja expans\~ao tern\'aria,
\[x = \sum_{n=1}^\infty 3^{-n}d_n,~~d_n\in\{0,1,2\}, \]
cont\'em pelo menos um d\'\i gito $1$. Por conseguinte, $C$ consiste de todos os pontos $x\in [0,1]$ que admitem uma expans\~ao tern\'aria com os d\'\i gitos $0$ e $2$ s\'o. Em particular, os pontos de primeira esp\'ecie s\~ao os que admitem uma tal expans\~ao tern\'aria finita. Eis um exemplo de ponto de segunda esp\'ecie:
\[0.202020202\ldots\]
\end{exemplo}

\begin{exercicio} 
Observe que o conjunto de Cantor tem o interior vazio.
\end{exercicio}

\begin{exercicio}
Construa uma vers\~ao do conjunto de Cantor do tipo ``ter\c co m\'edio'' que tem a medida de Lebesgue t\~ao pr\'oxima de $1$ quanto desejado. 
\par
[ {\em Sugest\~ao:} eliminar em vez de uma ter\c ca parte, uma parte menor, e calcular a soma de uma s\'erie geom\'etrica.... ] Conclua que, dado $\delta>0$, existe um subconjunto aberto, $U$, de $[0,1]$, cuja fronteira tem medida de Lebesgue $>1-\delta$.
\end{exercicio}

\begin{exercicio}
Mostre que no caso geral, quase certamente, todo \'atomo de $\mu$ aparece pelo menos uma vez (de fato, um n\'umero infinito de vezes) dentro do caminho amostral. Deduza que no caso puramente at\^omico, o erro de aprendizagem converge para zero quase certamente. De modo mais geral, quase certamente, cada \'atomo $x\in\Omega$ satisfaz Eq. (\ref{eq:elltochiCx}).
\label{ex:learningatoms}
\end{exercicio}

Vamos ver dois exemplos de subconjuntos fechados nos espa\c cos m\'etricos separ\'aveis que n\~ao podem ser aprendidos pelo classificador de vizinhos mais pr\'oximos. Nossa pr\'oxima tarefa \'e entender a geometria de um espa\c co m\'etrico $\Omega$ que permite a aprendizagem de conceitos pelo classificador $1$-NN.

\begin{observacao}
Na prova da proposi\c c\~ao \ref{p:fronteiranula}, n\'os estabelecemos um resultado mais forte do que a mera converg\^encia do erro para zero em probabilidade: mostramos que o erro converge para zero quase certamente. (Veja se\c c\~ao \ref{s:quasecerta}).
Na parte da aprendizagem autom\'atica motivada pela estat\'\i stica, refor\c car os resultados sobre a converg\^encia de erro em probabilidade para converg\^encia quase certa \'e considerado importante. No entanto, continuaremos com a converg\^encia do erro para zero em probabilidade (ou: esperan\c ca).
\end{observacao}

Eis o resultado central desta subse\c c\~ao.

\begin{teorema}
Seja $C$ um conjunto boreliano qualquer do espa\c co euclidiano $\R^d$. Ent\~ao, o classificador de vizinho mais pr\'oximo aprende $C$:
\[\E_{\sigma\sim\mu^n}\mbox{erro}_{\mu,C}{\mathcal L}^{\mbox{\tiny NN}}_n(C\upharpoonright\sigma)\to 0\mbox{ quando }n\to\infty.\]
\label{t:1-NN}
\end{teorema}

O resultado \'e um pouco surpreendente, pois a fronteira de $C$ pode ser grande, inclusive ser igual ao espa\c co $\Omega$ inteiro, e cada vizinhan\c ca de um ponto da fronteira vai conter elementos de $C$ bem como elementos de $C^c$. Vamos mostrar que os elementos de $C$ s\~ao mais comuns entre os vizinhos de $x$ quando $x\in C$, e menos comuns quando $x\notin C$.
Observemos que o caso onde $x$ \'e um \'atomo foi j\'a tratado (exerc\'\i cio \ref{ex:learningatoms}).
A demonstra\c c\~ao ocupa o resto da subse\c c\~ao.

Seja $\ve>0$ qualquer.
Usando o teorema \ref{t:regularidadeKU}, escolhamos dois conjuntos compactos $K_i$, $i=1,2$, com as propriedades $K_1\subseteq C$, $K_2\subseteq \Omega\setminus C$, $\mu(C\setminus K_1)<\ve/2$, e $\mu((\Omega\setminus C)\setminus K_2)<\ve/2$. Denotemos tamb\'em $U=\Omega\setminus(K_1\cup K_2)$. Temos: $C\cap K_2=\emptyset$ e $\mu(U)<\ve$. O nosso conceito $C$ fica espremido entre dois compactos, $K_1$ e $K_2$. 

\begin{exercicio}
Mostrar que a dist\^ancia,
\[d(K_1,K_2)=\inf \{d(x,y)\colon x\in K_1,y\in K_2\},\]
entre dois compactos n\~ao vazios e disjuntos, $K_1$ e $K_2$, num espa\c co m\'etrico qualquer \'e estritamente positiva.
\end{exercicio}

Dado um caminho amostral $\varsigma$ e um n\'umero natural positivo $n$, definamos a fun\c c\~ao $r^{\varsigma_n}_{\mbox{\tiny NN}}(x)$ como a dist\^ancia entre $x$ e os elementos do segmento inicial $\varsigma_n$:
\[r^{\varsigma_n}_{\mbox{\tiny NN}}(x)=\min_{i=1,2,\ldots,n}d(x,x_i).\]
Em outras palavras, \'e o valor da dist\^ancia entre $x$ e o vizinho mais pr\'oximo dele em $\varsigma_n$. 
Note que a fun\c c\~ao dist\^ancia de um conjunto, $r^{\varsigma}_n(x)$ \'e Lipschitz cont\'\i nua de constante $L=1$. 
Note que a fun\c c\~ao $r^{\varsigma_n}_{\mbox{\tiny NN}}$ depende do caminho amostral, e como tal, pode ser vista como uma fun\c c\~ao aleat\'oria, ou seja, um elemento aleat\'orio do espa\c co de fun\c c\~oes Lipschitz cont\'\i nuas.

\begin{definicao}
Uma fam\'\i lia $\mathscr F$ de fun\c c\~oes entre dois espa\c cos m\'etricos, $X$ e $Y$, \'e dita {\em uniformemente equicont\'\i nua} se para cada $\ve>0$ existe $\delta>0$ tal que, quaisquer que sejam $x,y\in X$, 
\[d_X(x,y)>\delta \Rightarrow d_Y(f(x),f(y))<\ve.\]
\index{fam\'\i lia! uniformemente equicont\'\i nua}
\end{definicao}

\begin{exercicio}
Seja $(f_n)$ uma sequ\^encia uniformemente equicont\'\i nua de fun\c c\~oes (por exemplo, uma sequ\^encia de fun\c c\~oes $1$-Lipschitz cont\'\i nuas) sobre um conjunto pr\'e-compacto, $K$, que converge para zero simplesmente sobre $K$ (ou seja, em todo ponto de $K$). Mostrar que $f_n$ converge para zero uniformemente sobre $K$.
\end{exercicio}

\begin{lema} 
A converg\^encia no lema de Cover--Hart \'e uniforme sobre todo subconjunto pr\'e-compacto do suporte da medida. Ou seja, quase certamente, a fun\c c\~ao $r^{\varsigma_n}_{\mbox{\tiny NN}}$ converge para zero uniformemente sobre todo conjunto pr\'e-compacto $K\subseteq\mbox{supp}\,\mu$.
\end{lema}

\begin{proof}
Seja $A$ um subconjunto enumer\'avel e denso de $\mbox{supp}\,\mu$. Seja $\ve>0$ qualquer. Um argumento padr\~ao mostra que, quase certamente, toda bola $B_{\ve}(x)$, $x\in A$, cont\'em um n\'umero infinito de elementos do caminho amostral (exerc\'\i cio; aqui usamos o fato que a medida da bola n\~ao \'e nula). Aplicando esta observa\c c\~ao para uma sequ\^encia de valores $\ve=1/m$, conclu\'\i mos: quase certamente, para todos $\ve>0$ e $x\in A$, a bola $B_{\ve}(x)$, $x\in A$, cont\'em um n\'umero infinito de elementos do caminho amostral. Seja $\varsigma$ um caminho amostral qualquer tendo esta propriedade. 

Agora seja $K$ um subconjunto pr\'e-compacto qualquer do suporte da medida, e $\ve>0$. Cubramos $K$ com uma fam\'\i lia finita de bolas abertas de raio $\ve$ com centros em $A$: 
\[K\supseteq \cup_{i=1}^k B_{\ve}(x_i),~x_i\in K.\]
Toda bola $B_{\ve}(x_i)$, $i=1,2,\ldots,k$ cont\'em pelo menos $k$ elementos de $\varsigma$. Isso implica, com ajuda da desigualdade triangular, que a partir de um valor de $n$ bastante grande, temos para todo $x\in K$
\[r^{\varsigma_n}_{k\mbox{\tiny -NN}}(x)<2\ve.\]
Conclu\'\i mos que
\[r^{\varsigma_n}_{k\mbox{\tiny -NN}}\overset{K}\rightrightarrows 0\mbox{ quando }n\to\infty.\]
\end{proof}

Eis uma consequ\^encia imediata.

\begin{fato}
Quase certamente, se $n$ \'e bastante grande, ent\~ao o vizinho mais pr\'oximo de todo elemento de $K_1$ em $\varsigma_n$ n\~ao pertence a $K_2$, e reciprocamente. 
\label{f:k1k2}
\end{fato}

\begin{figure}[!htbp] 
   \begin{center}
    \scalebox{0.2}{\includegraphics{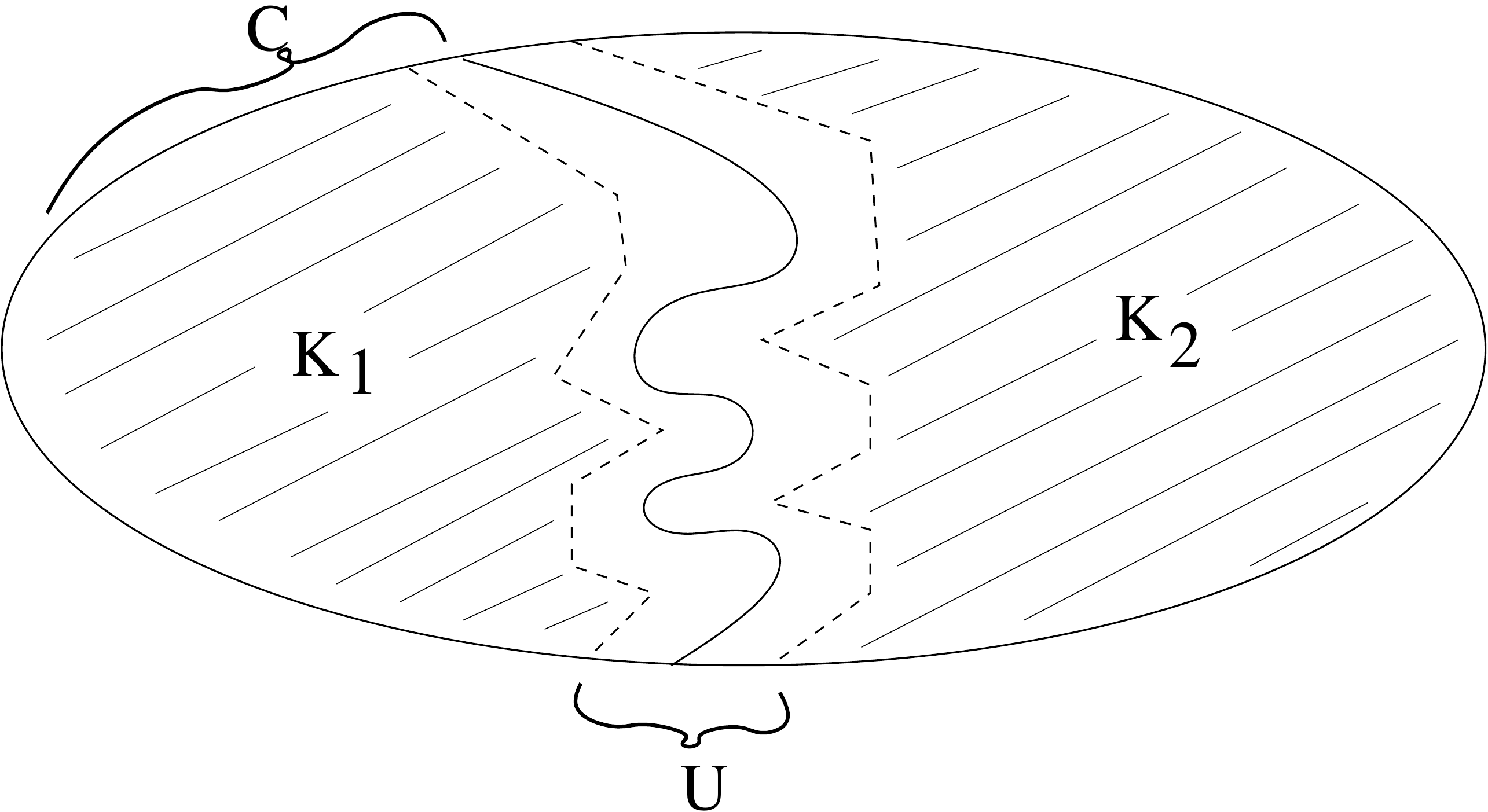}} 
  \end{center}
\caption{O conceito $C$ est\'a ``preso'' entre os compactos $K_1$ e $K_2$.}
\end{figure}

Relembremos que todos os pontos de $K_1$ s\~ao rotulados $1$ (porque $K_1\subseteq C$), enquanto os pontos de $K_2$ s\~ao rotulados $0$. 
O classificador $\mbox{NN}$ prediz o r\'otulo de um ponto $x\in K_i$ baseando a predi\c c\~ao sobre o r\'otulo do ponto $\mbox{NN}_{\varsigma_n}(x)$ que pertence ao conjunto $K_i\cup U$. Se pudermos limitar a ``influ\^encia'' sobre $x$ de pontos da amostra $\varsigma_n$ que pertencem a $U$ (por exemplo, mostrando que o valor da probabilidade $P[NN_{\varsigma_n}(X)\in U]$ \'e proporcional \`a medida de $U$), ter\'\i amos controle sobre o erro de classifica\c c\~ao. 
No espa\c co euclidiano um tal controle \'e devido ao lema importante de Stone (aqui, na vers\~ao $k=1$).

\begin{lema}[Lema geom\'etrico de Stone, caso $k=1$]
Para todo $d$ natural, existe uma constante absoluta $C=C(d)$ com a propriedade seguinte.
Seja 
\[\sigma=(x_1,x_2,\ldots,x_n),~x_i\in\R^d,~i=1,2,\ldots,n,\]
uma amostra finita qualquer em $\R^d$ (possivelmente com repeti\c c\~oes), e seja $x\in\R^d$ qualquer. O n\'umero de $i$ tais que $x\neq x_i$ e $x$ \'e o vizinho mais pr\'oximo de $x_i$ dentro da amostra
\[x,x_1,x_2,\ldots, x_{i-1},x_i,x_{i+1},\dots,x_n\]
\'e majorado por $C$. 
\label{l:stonek=1}
\index{lema! geom\'etrico de Stone}
\end{lema}

\begin{observacao}
A estrat\'egia de desempate n\~ao importa para a conclus\~ao do lema.
\end{observacao}

\begin{observacao}
O caso onde $x_i=x$ e por conseguinte tem $x$ como o vizinho mais pr\'oximo n\~ao nos preocupa, pois isso s\'o \'e poss\'\i vel com uma probabilidade positiva se $x$ for um \'atomo sob a distribui\c c\~ao $\mu$, neste caso o classificador $NN$ j\'a aprende o r\'otulo de $x$ perfeitamente bem. 
\end{observacao}

Relembremos que a {\em dist\^ancia geod\'esica}
\index{dist\^ancia! geod\'esica}
na esfera euclidiana com o centro na origem \'e o \^angulo entre dois vetores, $\vec x$ e $\vec y$. A dist\^ancia geod\'esica \'e equivalente \`a dist\^ancia euclidiana, por exemplo, na esfera unit\'aria temos
\[\norm{x-y}\leq \angle (x,y)\leq \frac{\pi}{2} \norm{x-y}.\]
Mais ainda, temos que cada uma delas \'e um chamado {\em transformado m\'etrico} da outra, por exemplo, a dist\^ancia euclidiana pode ser expressa assim:
\[\forall x,y\in\s,~\norm{x-y}=2\sin\left(\frac{\angle(x,y)}2\right).
\]
Isso implica que as bolas abertas com rela\c c\~ao  a uma dist\^ancia s\~ao as bolas com rela\c c\~ao  \`a outra.

\begin{proof}[Demonstra\c c\~ao do lema \ref{l:stonek=1}]
Como a dist\^ancia euclidiana \'e invariante pelas transla\c c\~oes, podemos supor sem perda de generalidade que $x=0$. 
Um {\em cone} com o v\'ertice na origem \'e definido assim:
\[C(\vec v,\alpha)=\{\vec y\in\R^d\colon \angle(\vec y,\vec v)<\alpha\},\]
onde $\vec v\in\R^d$ e $\alpha>0$ s\~ao fixos. A interse\c c\~ao de um cone com a esfera unit\'aria em torno de origem \'e uma bola aberta na esfera de raio $\alpha$ formada com ajuda da dist\^ancia geod\'esica, e por conseguinte, uma bola na esfera relativo \`a dist\^ancia euclidiana. Como a esfera \'e um conjunto compacto, qualquer seja $\alpha>0$, o espa\c co $\R^d$ pode ser coberto por uma fam\'\i lia finita de cones da forma $C(\vec v,\alpha)$.
Seja $\alpha=\pi/6$. Escolhemos os cones $C_1,C_2,\ldots,C_m$ cobrindo o espa\c co. (O n\'umero $m$ s\'o depende de $d$, \'e a nossa constante $C$). 

\begin{figure}
\centering
\scalebox{0.3}{\includegraphics{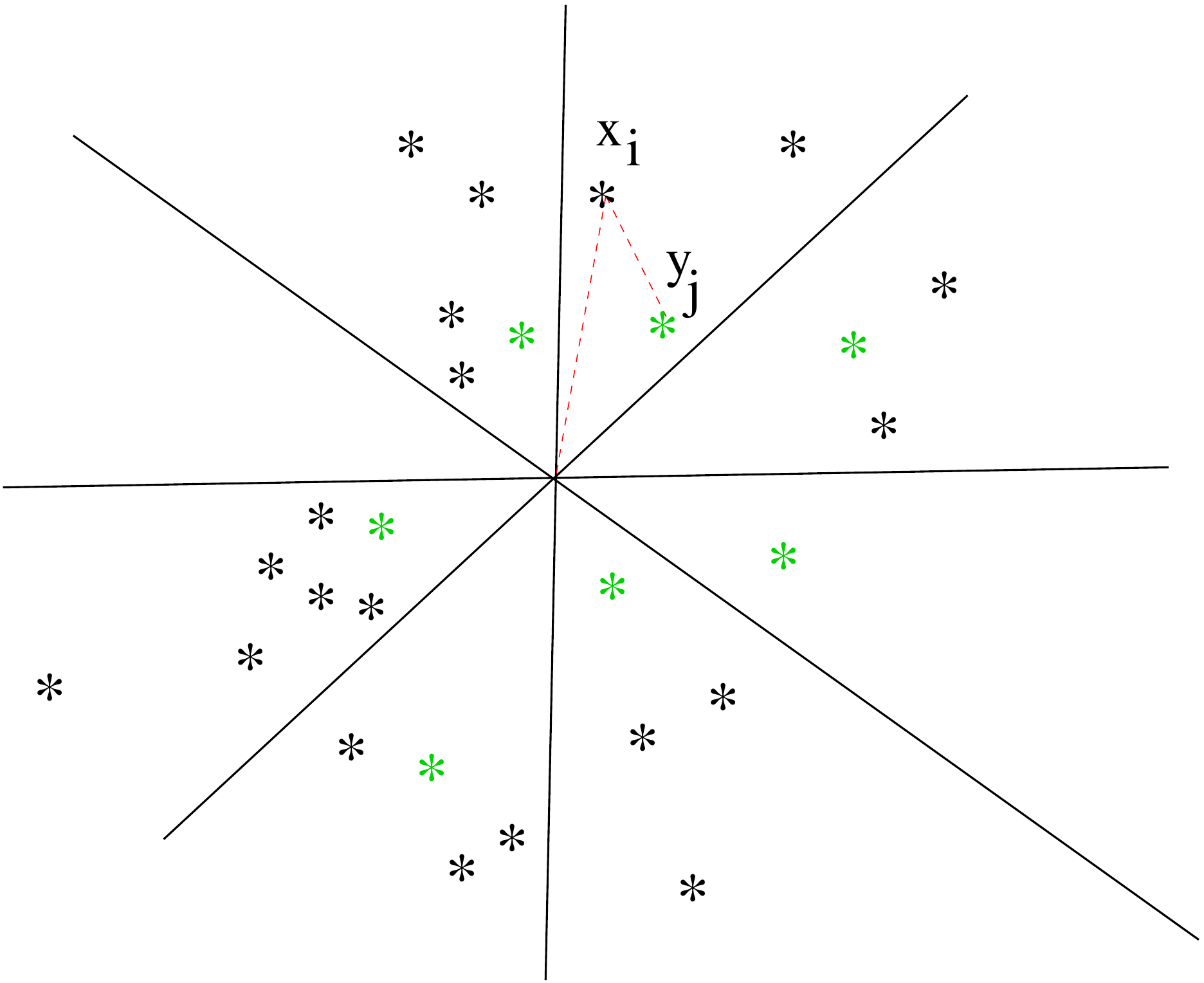}}
\caption{Prova do lema geom\'etrico de Stone (caso $k=1$) em $\R^d$.}
\label{fig:stone_1nn}
\end{figure}
Para cada cone, se existirem pontos da amostra $x_i$ que pertencem ao interior do cone, escolhamos entre eles o vizinho mais pr\'oximo de zero, e no caso de empates, qualquer um deles. Denotemos $Y$ o conjunto de todos os pontos escolhidos, $m$ deles no m\'aximo. Seja $x_i\notin Y$, $x_i\neq 0$. Existe $j$ tal que $x_i\in C_j$. Neste caso, existe $y=y_j\in Y\cap C_j$. Temos $\norm{x_i}\geq\norm{y_j}$.
Agora \'e um exerc\'\i cio simples da planimetria b\'asica, mostrar que  $\norm{x_i-y_j}<\norm{x_i}$. Por conseguinte, o vizinho mais pr\'oximo de $x_i$ no conjunto $x_1,x_2,\ldots, x_{i-1},0,x_{i+1},\dots,x_n$ nunca \'e igual a $0$. 
\end{proof}

\begin{lema}
Sejam $U$ um subconjunto boreliano qualquer de $\R^d$, e $\mu$ uma medida de probabilidade boreliana sobre $\R^d$. Ent\~ao, para todo $n$,
\[
\limsup_{n\to\infty}P[X\notin U\mbox{ e }\mbox{NN}_{\sigma_n}(X)\in U]\leq C(d)\mu(U),\]
ou, um pouco mais exatamente e na nota\c c\~ao conjunt\'\i stica,
\[(\mu\otimes\mu^{\otimes n})\{(x,\sigma_n)\in\R^d\times (\R^d)^{n}\colon x\notin U\mbox{ e }\mbox{NN}_{\sigma_n}(x)\in U\} \leq C(d)\mu(U)+\frac Cn.\]
\label{l:limsup}
\end{lema}

\begin{proof}
Apliquemos uma simetriza\c c\~ao pelas transposi\c c\~oes
\index{simetriza\c c\~ao! por transposi\c c\~oes}
$\tau_i$ que trocam $X\leftrightarrow X_i$. Toda permuta\c c\~ao de coordenadas, incluindo $\tau_i$, conserva a medida produto $\mu\otimes\mu^{\otimes n}$, e por conseguinte
\begin{align*}
P[X\notin U\mbox{ e }\mbox{NN}_{\sigma_n}(X)\in U] 
&= P[X_i\notin U\mbox{ e }\mbox{NN}_{(X_1,X_2,\ldots, X_{i-1},X,X_{i+1},\dots,X_n)}(X_i)\in U]
\\
&= \frac 1n \sum_{i=1}^n P[X_i\notin U\mbox{ e }\mbox{NN}_{(X_1,\ldots, X_{i-1},X,X_{i+1},\dots,X_n)}(X_i)\in U]
\\
&= \E\frac 1n \sum_{i=1}^n \chi_{\{(x,\sigma_n)\colon
{x_i\notin U\mbox{\tiny e }\mbox{\tiny NN}_{(x_1,\ldots, x_{i-1},x,x_{i+1},\dots,x_n)}(x_i)\in U}\}}.
\end{align*}
A soma \'e um ``contador'' que registra o n\'umero de $i$ tais que $x_i\notin U$, e ao mesmo tempo o seu vizinho mais pr\'oximo no resto da amostra $(x,\sigma)$ pertence a $U$. 
Entre os $n+1$ pontos $x,x_1,\ldots, x_{i-1},x_i,x_{i+1},\dots,x_n$, ao m\'aximo $n\mu_{\sigma_n}(U)+1$ pertencem a $U$, onde $1$ \'e adicionado por causa de $x$. Cada um destes pontos \'e, no m\'aximo, o vizinho mais pr\'oximo de $C$ pontos distintos dele, em particular, de pontos fora de $U$.
Por conseguinte, o n\'umero dos \'\i ndices $i$ tais que a fun\c c\~ao indicadora n\~ao se anula em $(x,\sigma_n)$ \'e limitado por $C( n \mu_{\sigma_n}(U)+1)$. Continuamos a avalia\c c\~ao:
\begin{align*}
P[X\notin U\mbox{ e }\mbox{NN}_{\sigma_n}(X)\in U] 
&\leq  \E\frac 1n C( n \mu_{\sigma_n}(U)+1) \\
&= C \E \mu_{\sigma_n}(U)+\frac Cn \\
&=  C\mu(U)+\frac Cn.
\end{align*}
\end{proof}

Agora podemos terminar a prova do teorema \ref{t:1-NN}. Suponha que $K_1,K_2$ s\~ao escolhidos como acima. O fato \ref{f:k1k2} implica que, para $n$ bastante grande, com a confian\c ca $\geq 1-\delta$, onde $\delta>0$ \'e o risco desejado, o vizinho mais pr\'oximo de todo ponto de $K_1$ n\~ao pertence a $K_2$, e reciprocamente. A predi\c c\~ao errado do r\'otulo no ponto $x$ s\'o pode ocorrer em dois casos: seja $x\in U$, seja $x\in K_i$, $i=1,2$ e $\mbox{NN}_{\sigma_n}(x)\in U$. Ent\~ao, a probabilidade de uma predi\c c\~ao errada \'e limitada assim:
\begin{eqnarray*}
\E_{\sigma\sim\mu^n}\mbox{erro}_C{\mathcal L}^{\mbox{\tiny NN}}_n(C\upharpoonright\sigma_n) &=& 
P[{\mathcal L}^{\mbox{\tiny NN}}_n(C\upharpoonright\sigma_n)(X)\neq \chi_C(X)]
\\
&\leq & \mu(U) + P[X\notin U\mbox{ e }\mbox{NN}_{\sigma_n}(X)\in U] 
\\
&<& \ve + C(d)\mu(U) + \frac Cn
\\
&<& (C(d)+1)\ve + \frac Cn.
\end{eqnarray*}

O \'ultimo valor pode ser t\~ao pequeno quanto quiser.

\section{Erro de Bayes}
Por que precisamos, ent\~ao, do classificador de $k$ vizinhos mais pr\'oximos, se j\'a o classificador $1$-NN aprende os conceitos? Esta parte da teoria de aprendizagem tem origem na estat\'\i stica, e o modelo de aprendizagem \'e um pouco mais sofisticado do que o usado na inform\'atica te\'orica. O modelo previamente estudado \'e {\em determin\'\i stico} no sentido que um ponto $x\in\Omega$ sempre recebe o mesmo r\'otulo, $\chi_C(x)$. O modelo estat\'\i stico leva em considera\c c\~ao a possibilidade de ru\'\i do aleat\'orio que pode corromper as observa\c c\~oes. Deste modo, o modelo \'e mais real\'\i stico, e a aprendizagem na presen\c ca de ru\'\i do, mais dif\'\i cil. Vamos ver em breve que o classificador $1$-NN falha na presen\c ca de ru\'\i do.

Dado um ponto $x$ do dom\'\i nio, o modelo atribui um r\'otulo aleat\'orio, segundo o valor da probabilidade de obter o r\'otulo $1$, denotado $\eta(x)\in [0,1]$. A fun\c c\~ao $\eta\colon\Omega\to [0,1]$ \'e chamada na estat\'\i stica de {\em fun\c c\~ao de regress\~ao}. 
\index{fun\c c\~ao! de regress\~ao}
No esp\'\i rito da {\em l\'ogica cont\'\i nua,} pode-se dizer que $\eta(x)$ \'e o valor de verdade (``grau de verdade'') da proposi\c c\~ao {\em ``o r\'otulo de $x$ \'e igual a $1$''}, cujos valores poss\'\i veis preenchem o intervalo $[0,1]$. 

No caso determin\'\i stico, a fun\c c\~ao de regress\~ao \'e bin\'aria, sendo simplesmente a fun\c c\~ao indicadora do conceito a ser aprendido:
\[\eta(x) = \chi_C(x).\]

Vamos agora descrever o modelo correspondente, com todas as mo\-di\-fi\-ca\-\c c\~oes necess\'arias.
O dom\'\i nio $\Omega$, como sempre, \'e um espa\c co boreliano padr\~ao. O ponto rotulado $(x,\e)$, onde $x\in \Omega$, $\e\in\{0,1\}$, \'e modelado por uma vari\'avel aleat\'oria $(X,Y)$ com valores no produto $\Omega\times \{0,1\}$. Aqui, $X\in\Omega$ representa um ponto no dom\'\i nio, e $Y\in\{0,1\}$, o r\'otulo marcando o ponto. A lei conjunta de $(X,Y)$ \'e uma medida de probabilidade, $\tilde\mu$, sobre $\Omega\times \{0,1\}$. Agora, o ponto $x\in\Omega$ \'e uma {\em inst\^ancia} da v.a. $X$, e o r\'otulo $\e$ \'e uma inst\^ancia da v.a. $Y$.

Denotemos $\mu_1$ a restri\c c\~ao de $\tilde\mu$ sobre $\Omega\times\{1\}$, ou, mais exatamente, uma medida sobre $\Omega$ definida por
\[\mu(A)=\tilde\mu(A\times\{1\}).\]
Esta n\~ao \'e mais necessariamente uma medida de probabilidade: o volume total de $\mu_1$ \'e igual a probabilidade de um ponto ser rotulado $1$. Semelhantemente, define a medida $\mu_0$ sobre $\Omega$. 

\begin{figure}
   \begin{center}
    \scalebox{0.25}{\includegraphics{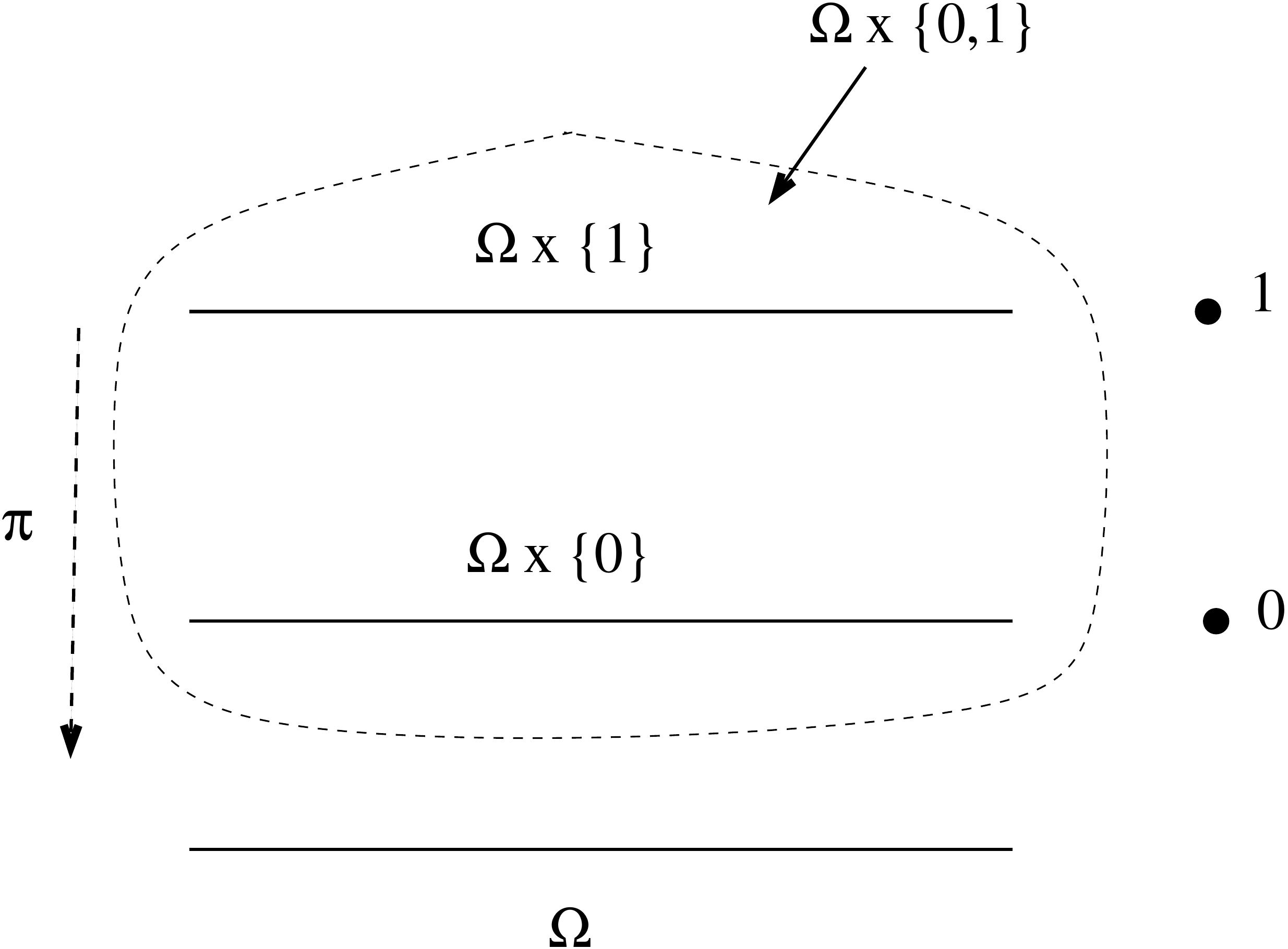}} 
  \end{center}
\caption{O dom\'\i nio rotulado e a proje\c c\~ao $\pi\colon\Omega\times\{0,1\}\to\Omega$.}
\end{figure}

Seja $\pi\colon\Omega\times\{0,1\}\to\Omega$ a proje\c c\~ao can\^onica de $\Omega\times\{0,1\}$ sobre $\Omega$, $(x,\e)\mapsto x$. A imagem direita de $\tilde\mu$ sobre $\pi$ \'e uma medida de probabilidade sobre $\Omega$,
\[\mu=\pi_{\ast}(\tilde\mu),\]
definida por
\[\mu(A)=\mu_0(A)+\mu_1(A).\]
A medida $\mu$ define a distribui\c c\~ao de pontos n\~ao rotulados. 

\'E claro que a medida $\mu_1$ \'e absolutamente cont\'\i nua com rela\c c\~ao  \`a medida $\mu$, ou seja, se $N\subseteq\Omega$ e $\mu(N)=0$, ent\~ao $\mu_1(N)=0$. De fato, ela satisfaz a condi\c c\~ao mais forte: para cada $A$, $\mu_1(A)\leq \mu(A)$. Segundo o caso particular do teorema de Radon-Nikod\'ym \ref{t:RNcasoespecial}, 
\label{page:usingRN}
existe uma derivada de Radon--Nikod\'ym de $\mu_1$ com rela\c c\~ao  a $\mu$, ou seja, uma fun\c c\~ao mensur\'avel
\[\eta(x) = \frac{d\mu_1(x)}{d\mu},~~\eta\colon\Omega\to\R,\]
tal que para todo $A\subseteq\Omega$, 
\[\mu_1(A) = \int_{A} \eta(x)\,d\mu(x).\] 
Segue-se que
\[\mu_0(A) = \int_{A} (1-\eta(x))\,d\mu(x).\]
Na linguagem probabil\'\i stica, $\eta$ \'e a probabilidade condicional para $X=x$ estar rotulado $1$:
\[\eta(x) = P[Y=1 \mid X=x ].\]
O par $(\mu,\eta)$ descreve completamente a distribui\c c\~ao $\tilde\mu$ de pontos rotulados, como segue: $X\in\Omega$ \'e uma vari\'avel aleat\'oria com a lei $\mu$, e uma vez que a inst\^ancia $x$ de $X$ \'e escolhida, o valor de $Y$ \'e escolhido lan\c cando a moeda com a probabilidade $\eta(x)$ de dar ``coroa''. 

Dado um classificador (fun\c c\~ao boreliana), $T\colon\Omega\to \{0,1\}$, o seu {\em erro de classifica\c c\~ao}
\index{erro! de classifica\c c\~ao} 
\'e definido por
\begin{eqnarray*}
  {\mathrm{erro}}_{\tilde\mu}(T) &=& {\mathrm{erro}}_{\eta,\mu}(T)
\\ &=& P[T(X)\neq Y] \\
  &=& \tilde\mu\{(x,y)\in \Omega\times\{0,1\}\colon T(x) \neq y\}.\end{eqnarray*}

\begin{proposicao}
${\mathrm{erro}}_{\eta,\mu}(T) = \norm{T-\eta}_{L^1(\mu)}= \int_{\Omega}\abs{T-\eta}\,d\mu$.
\label{p:errocomoL1}
\end{proposicao}

\begin{proof}
\begin{align*}
{\mathrm{erro}}_{\eta,\mu}(T) & = P[T(X)\neq Y] \\
&= \int_{\{T=1\}}P[Y=0\vert X=x]\,d\mu(x) + \int_{\{T=0\}}P[Y=1\vert X=x]\,d\mu(x) \\
&= \int_{\{T=1\}} (1-\eta(x))\,d\mu(x) + \int_{\{T=0\}}\eta(x)\,d\mu(x) \\
&= \int_{\{T=1\}} \abs{T-\eta}\,d\mu + \int_{\{T=0\}}\abs{T-\eta}\,d\mu\\
&= \int_{\Omega}\abs{T-\eta}\,d\mu.
\end{align*}
\end{proof}

\begin{observacao}
Se $\eta=\chi_C$ \'e uma fun\c c\~ao de regress\~ao determin\'\i stica, ent\~ao obtemos a express\~ao bem familiar:
\[{\mathrm{erro}}_{\eta,\mu}(T) = \norm{T-\chi_C}_{L^1(\mu)}=\mu(C\Delta \{T=1\}).\]
\end{observacao}

\begin{corolario}
Denote $Z_\eta=Z(\eta-1/2)=\{x\in\Omega\colon \eta(x)=1/2\}$. Ent\~ao, 
\[{\mathrm{erro}}_{\eta,\mu}(T) = \int_{\Omega\setminus Z_\eta}\abs{T-\eta}\,d\mu.\]
\label{c:erroetaZ}
\end{corolario}

\begin{proof}
Basta notar que, qualquer seja o classificador $T$, a fun\c c\~ao $\abs{T-\eta}$ toma o mesmo valor constante sobre $Z_\eta$ :
\[\abs{T-\eta}\vert_{Z_\eta}\equiv \frac 12.\]
\end{proof}

\begin{corolario}
Sejam $T,T^\prime$ dois classificadores satisfazendo
\[\forall x\in\Omega,~\eta(x)\neq \frac 12 \Rightarrow T(x) = T^\prime(x).\]
Ent\~ao,
\[{\mathrm{erro}}_{\tilde\mu}(T)={\mathrm{erro}}_{\tilde\mu}(T^\prime).\]
\label{ex:conjuntoZ}
\end{corolario}

O {\em erro de Bayes}
\index{erro! de Bayes} 
\'e o \'\i nfimo de erros de classifica\c c\~ao de todos os classificadores poss\'\i veis sobre $\Omega$:
\[\ell^{\ast}=\ell^{\ast}(\tilde\mu)=\ell^{\ast}(\mu,\eta)=
\inf_{T}{\mathrm{erro}}_{\tilde\mu}(T).\]
Com efeito, o \'\i nfimo \'e o m\'\i nimo, atingido pelo {\em classificador de Bayes:}
\index{classificador! de Bayes}
\[T_{bayes}(x) =\left\{\begin{array}{ll} 0,&\mbox{ se }\eta(x)<\frac 12,\\
  1,&\mbox{ se }\eta(x)\geq \frac 12.
\end{array}\right.\]

\begin{figure}
 \begin{center}
    \scalebox{0.2}{\includegraphics{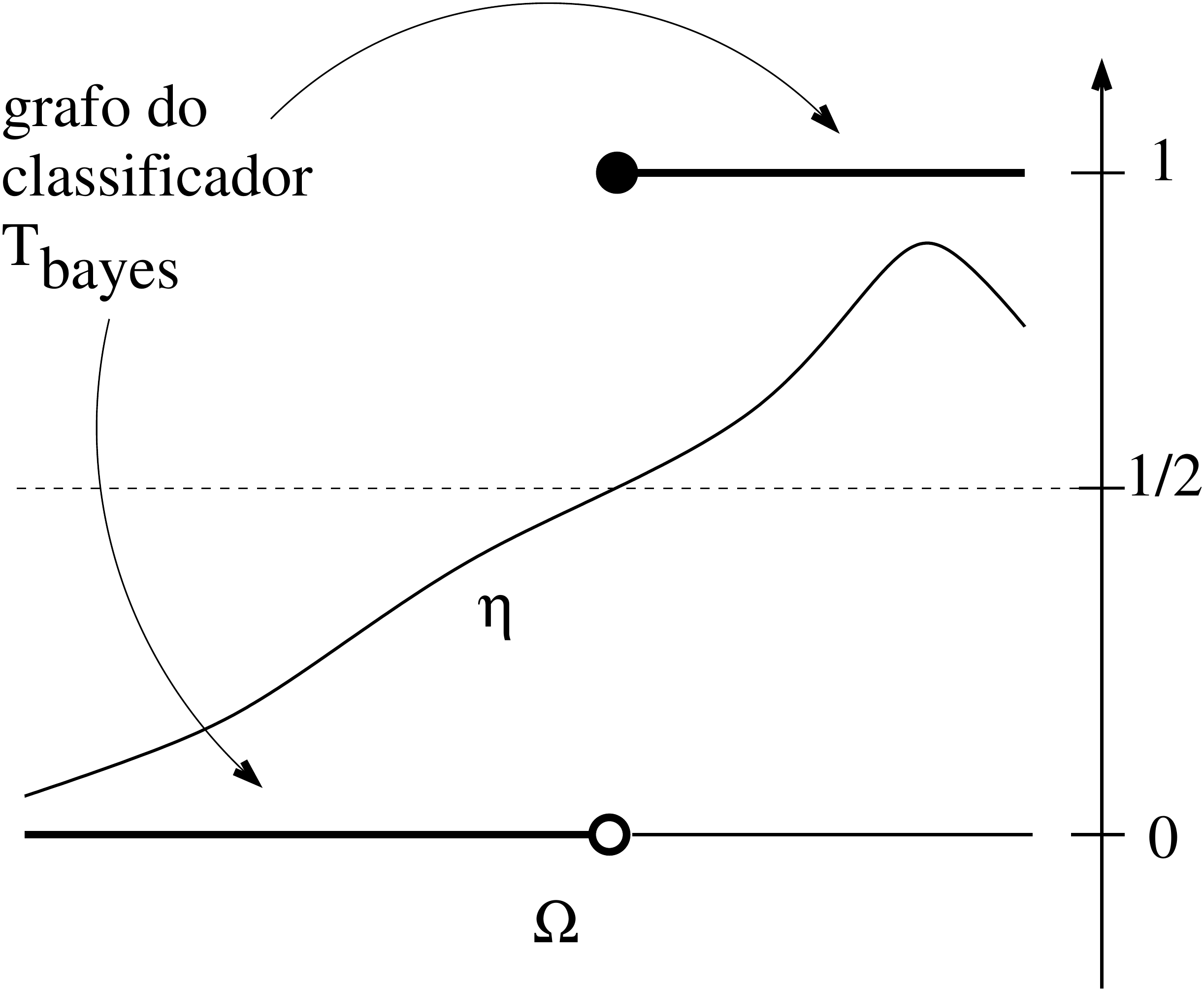}} 
  \end{center}
\caption{O classificador de Bayes}
\end{figure}

\begin{exercicio}
Mostre que
\[{\mathrm{erro}}_{\tilde\mu}(T_{bayes})=\ell^{\ast}(\tilde\mu).\]
[ {\em Sugest\~ao:} suponha que $T$ e $T_{bayes}$ sejam essencialmente diferentes fora do conjunto $Z_{\eta}$ do exerc\'\i cio \ref{ex:conjuntoZ}. Ent\~ao, existe $A$, $\mu(A)> 0$, $A\cap Z=\emptyset$, tal que $T\vert_A\neq T_{bayes}\vert_A$. Por exemplo, suponha que $T\vert_A\equiv 1$ e $T_{bayes}\vert_A\equiv 0$. Mostre direitamente que ${\mathrm{erro}}_{\tilde\mu}(T)>{\mathrm{erro}}_{\tilde\mu}(T_{bayes})$. ]
\end{exercicio}

\begin{observacao}
Segue-se do exerc\'\i cio \ref{ex:conjuntoZ} que
em pontos $x$ onde $\eta(x)=1/2$ o classificador de Bayes pode tomar qualquer valor, $0$ ou $1$. Por isso, talvez seria mais correto de falar de {\em um classificador de Bayes.}
\label{o:bayes}
\end{observacao}

\begin{exercicio}
Mostre que o erro de Bayes \'e igual a zero, $\ell^{\ast}(\tilde\mu)=0$, se e somente se a fun\c c\~ao de regress\~ao $\eta$ \'e determin\'\i stica (s\'o toma os valores $0$ e $1$ fora de um conjunto $\mu$-negligenci\'avel).
\end{exercicio}

O significado do classificador de Bayes \'e puramente te\'orico, porque a fun\c c\~ao de regress\~ao, $\eta$, \'e desconhecida, assim como a lei $\tilde\mu$. 

A amostra rotulada $(x_1,x_2,\ldots,x_n,\e_1,\ldots,\e_n)$ \'e modelada pela sequ\^encia 
\[(X_1,Y_1),(X_2,Y_2),\ldots,(X_n,Y_n)\]
das vari\'aveis independentes com valores em $\Omega\times\{0,1\}$, seguindo a lei fixa, por\'em desconhecida, $\tilde\mu$.

A regra de aprendizagem $\mathcal L$ no dom\'\i nio $\Omega$ \'e chamada {\em consistente}
\index{regra! consistente}
 se o erro de clas\-si\-fi\-ca\-\c c\~ao converge para o erro de Bayes (o menor poss\'\i vel) em probabilidade quando $n\to\infty$:
\[\forall\ve>0,~~P\left[{\mathrm{erro}}_{\mu}{\mathcal L}_n >\ell^\ast(\tilde\mu)+\ve\right]\to 0\mbox{ quando }n\to\infty.\]

Como n\~ao conhecemos a lei subjacente, $\tilde\mu$, precisamos que a regra de aprendizagem seja consistente para todas as leis poss\'\i veis. Isto leva \`a seguinte defini\c c\~ao. 

\begin{definicao}
A regra $\mathcal L$ \'e {\em universalmente consistente} se ela \'e consistente para cada medida de probabilidade $\tilde\mu$ sobre $\Omega\times\{0,1\}$. 
\index{regra! universalmente consistente}
\end{definicao}

Em particular, uma regra universalmente consistente PAC aprende cada conceito $C$ sob cada lei $\mu$ sobre o dom\'\i nio. Por\'em, a afirma\c c\~ao rec\'\i proca n\~ao \'e verdadeira.
Ocorre que o classificador de um vizinho mais pr\'oximo n\~ao \'e consistente j\'a numa situa\c c\~ao simpl\'\i ssima.

\begin{exemplo}
O dom\'\i nio \'e um conjunto unit\'ario, $\Omega=\{\ast\}$, munido da medida de probabilidade $\mu$ evidente. O \'unico caminho amostral \'e constante, $\ast,\ast,\ldots$. 
A fun\c c\~ao de regress\~ao toma o valor $2/3$ sobre o \'unico ponto do dom\'\i nio. Isso significa que com a probabilidade $2/3$, o ponto $\ast$ obteve o r\'otulo $1$, e com a probabilidade $1/3$, o r\'otulo ser\'a $0$. Como $\eta(\ast)=2/3$, o classificador de Bayes toma o valor $1$, e o
erro de Bayes \'e igual a $1/3$.

Nesta situa\c c\~ao, precisamos uma regra de desempate, e ela pode ser qualquer, uma vez ela n\~ao depende dos r\'otulos. 
Qualquer seja a regra de desempate, a probabilidade de que $NN(\ast)$ tenha o r\'otulo $\e_i=1$ \'e sempre igual a $2/3$. \'E f\'acil fazer o c\'alculo seguinte:
\begin{center}
\begin{tabular}{c|c|c}
R\'otulo de $NN(\ast)$ & Probabilidade dele & Erro \\[1mm]
\hline
$1$ & $\frac 23$ & $\frac 13$ \\[1mm]
$0$ & $\frac 13$ & $\frac 23$ \\[1mm]
\hline
\end{tabular}
\end{center}
Conclu\'\i mos, usando a probabilidade condicional:
\[\mbox{erro}_{\mu,\eta}({\mathcal L}^{\mbox{\tiny NN}})=\frac 23\cdot\frac 13+\frac 13\cdot\frac 23 = \frac 49 > \frac 13.\]
Se voc\^e olha este exemplo com mal\'\i cia, pode modific\'a-lo e substituir $\Omega$ por um espa\c co probabil\'\i stico n\~ao at\^omico, onde quase certamente todo ponto ter\'a um \'unico vizinho mais pr\'oximo, munido da fun\c c\~ao de regress\~ao $\eta\equiv 2/3$. O mesmo argumento se aplica para concluir que o classificador $1$-NN n\~ao \'e consistente.
\label{ex:simplissimo}
\end{exemplo}

Agora imaginemos que ao inv\'es de escolher o r\'otulo de $x$ nos baseando sobre o \'unico vizinho mais pr\'oximo de $x$, levamos em considera\c c\~ao os r\'otulos de $k$ vizinhos, onde $k\to\infty$. Segundo a lei dos grandes n\'umeros, quando $k$ for bastante grande, com alta confian\c ca aproximadamente $(2/3\pm\e)k$ pontos teriam o r\'otulo $1$, e o voto majorit\'ario entre $k$ vizinhos mais pr\'oximos produziria, com alta confian\c ca, o r\'otulo $1$ para $x$. No limite assint\'otico $n,k\to\infty$, o erro de aprendizagem vai convergir para o erro de Bayes, $1/3$.

\section{Classificador $k\mbox{-NN}$}

Dado $k$ e $n\geq k$, definamos o valor $r^{\varsigma_n}_{k\mbox{\tiny -NN}}(x)$  como o menor raio de uma bola fechada em torno de $x$ que cont\'em $k$ vizinhos mais pr\'oximos de $x$ em $\varsigma_n$:
\[r^{\varsigma_n}_{k\mbox{\tiny -NN}}(x)=\min\{r\geq 0\colon \sharp\{i=1,2,\ldots,n\colon x_i\in \bar B_r(x)\}\geq k\}.\]
\index{rsigmakNN@$r^{\varsigma_n}_{k\mbox{\tiny -NN}}$}

\begin{exercicio}
Verificar que a fun\c c\~ao real $r^{\varsigma_n}_{k\mbox{\tiny -NN}}\colon\Omega\to\R$ \'e Lipschitz cont\'\i nua de constante $L=1$. 
\end{exercicio}

Eis, o lema de Cover-Hart na sua vers\~ao completa.

\begin{teorema}[Lema de Cover-Hart \citep*{cover_hart}]
Seja $\Omega$ um espa\c co m\'etrico separ\'avel qualquer, e seja $\mu$ uma medida de probabilidade boreliana sobre $\Omega$. Quase certamente, a fun\c c\~ao $r^{\varsigma_n}_{k\mbox{\tiny -NN}}$ converge para zero uniformemente sobre conjuntos pr\'e-compactos $K\subseteq\mbox{supp}\,\mu$.
\index{lema! de Cover--Hart}
\end{teorema}

\begin{proof}
Seja $A$ um subconjunto enumer\'avel e denso de $\mbox{supp}\,\mu$. Seja $\ve>0$ qualquer. Um argumento j\'a conhecido mostra que, dado $x\in A$ e $\e>0$, quase certamente, a bola $B_{\ve}(x)$ cont\'em um n\'umero infinito de elementos do caminho amostral (exerc\'\i cio; usemos o fato que a medida da bola n\~ao \'e nula). Aplicando esta observa\c c\~ao \`a sequ\^encia de valores $\ve=1/m$, conclu\'\i mos: para quase todos caminhos amostrais, $\varsigma$, para todos $\ve>0$ e $x\in A$, a bola $B_{\ve}(x)$, $x\in A$, cont\'em um n\'umero infinito de elementos de $\varsigma$. 

Sejam $\varsigma$ um caminho amostral qualquer tendo a propriedade acima, $K$ um subconjunto pr\'e-compacto qualquer do suporte da medida, e $\ve>0$. Cubramos $K$ com uma fam\'\i lia finita de bolas abertas de raio $\ve$ com centros em $A$: 
\[K\supseteq \cup_{i=1}^k B_{\ve}(x_i),~x_i\in K.\]
Existe $N$ tal que, se $n\geq N$, cada bola $B_{\ve}(x_i)$, $i=1,2,\ldots,k$ cont\'em pelo menos $k$ elementos de $\varsigma$. A desigualdade triangular implica que a partir de $N$, temos para todo $x\in K$
\[r^{\varsigma_n}_{k\mbox{\tiny -NN}}(x)<2\ve.\]
Conclu\'\i mos:
\[r^{\varsigma_n}_{k\mbox{\tiny -NN}}\overset{K}\rightrightarrows 0\mbox{ quando }n\to\infty.\]
\end{proof}

Dado um ponto $x$ e uma amostra (n\~ao rotulada) $\sigma$, pode acontecer que existam v\'arios pontos de $\sigma$ \`a mesma dist\^ancia de $x$, ou seja, que existam estritamente mais do que $k$ pontos na bola fechada $\bar B_{r_{k\mbox{\tiny -NN}^{\sigma}}(x)}(x)$. Com isto em mente, precisamos de uma estrat\'egia de desempate.
\index{desempate}
Para defini-la, vamos usar a nota\c c\~ao seguinte: se $\sigma\in\Omega^n$ e $\sigma^\prime\in\Omega^k$, $k\leq n$,  o s\'\i mbolo
\[\sigma^\prime\sqsubset\sigma\]
significa que existe uma inje\c c\~ao $f\colon [k]\to [n]$ tal que
\[\forall i=1,2,\ldots,k,~\sigma^\prime_i=\sigma_{f(i)}.\] 
Uma aplica\c c\~ao de desempate \'e uma aplica\c c\~ao
\[k\mbox{-NN}^{\sigma}\colon\Omega^n\times\Omega\to\Omega^k\]
que possui as propriedades 
\[k\mbox{-NN}^{\sigma}(x) \sqsubset \sigma
\]
e
\[\sigma\cap B_{r_{k\mbox{\tiny -NN}^{\sigma}}(x)}(x)\subseteq k\mbox{-NN}^{\sigma}(x)\subseteq \bar B_{r_{k\mbox{\tiny -NN}^{\sigma}}(x)}(x).\]
A aplica\c c\~ao $k\mbox{-NN}^{\sigma}$ pode ser determin\'\i stica ou estoc\'astica, dependendo de uma var\'avel aleat\'oria $Z$, independente da lei $\mu$ de dados.

\begin{exemplo}
Uma escolha determin\'\i stica poss\'\i vel \'e usar a ordem na amostra, $x_1<x_2<\ldots<x_n$. Neste caso, $k\mbox{-NN}^{\sigma}(x)$ consiste de todos os pontos de $\sigma$ na bola aberta, $B_{r_{k\mbox{\tiny -NN}^{\sigma}}(x)}(x)$, mais o n\'umero necess\'ario de pontos de $\sigma\cap S_{r_{k\mbox{\tiny -NN}^{\sigma}}(x)}(x)$ que t\^em os menores \'\i ndices.
\end{exemplo}

\begin{exemplo}
A escolha de uma ordem de desempate pode ser tamb\'em estoc\'astica. Neste caso a fonte de aleatoriedade, $Z$, pode ser convertida numa vari\'avel aleat\'oria $Z^\prime$ com valores no grupo $S_n$ de permuta\c c\~oes, seguindo a lei uniforme. 
\end{exemplo}

Agora a regra de classifica\c c\~ao $k$-NN pode ser formalizada como segue:
\begin{eqnarray*}
{\mathcal L}^{k\mbox{\tiny -NN}}_n(\sigma,\e)(x) &=&
\theta\left[\frac 1k\sum_{x_i\in k\mbox{\tiny -NN}^{\sigma}(x)}\e_i-\frac 12 \right]
\\[1mm]
&=& 
\theta\left[\E_{\mu_{k\mbox{\tiny -NN}^{\sigma}(x)}}\e -\frac 12\right].
\end{eqnarray*}
\index{classificador! k-NN@$k$-NN}
Aqui, $\theta$ \'e a fun\c c\~ao de Heaviside, o sinal do argumento:
\[\theta(t) =\begin{cases} 1,&\mbox{ se }t\geq 0,\\
0,&\mbox{ se }t<0.\end{cases}\]
A medida emp\'\i rica $\mu_{k\mbox{\tiny -NN}^{\sigma}(x)}$ \'e suportada sobre o conjunto de $k$ vizinhos mais pr\'oximos de $x$ na amostra $\sigma$, e $\e$ \'e a rotulagem, uma fun\c c\~ao $\e\colon \{x_1,x_2,\ldots,x_n\}\to\{0,1\}$. Deste modo, a esperan\c ca tem sentido.

A aplica\c c\~ao ${\mathcal L}^{k\mbox{\tiny -NN}}_n$ pode ser determin\'\i stica ou estoc\'astica, dependendo da natureza do desempate, por\'em ela n\~ao depende da medida $\tilde\mu$ sobre $\Omega\times\{0,1\}$. 

Come\c camos a an\'alise com um resultado preliminar estabelecendo a consist\^encia  do classificador $k$-NN em espa\c co m\'etrico qualquer, sob uma hip\'otese adicional, a de continuidade da fun\c c\~ao de regress\~ao, $\eta(x)=d\mu_1(x)/d\mu(x)$. 

\begin{teorema}
Seja $\Omega$ um espa\c co m\'etrico separ\'avel munido de uma medida de probabilidade $\mu$.
Suponha que a fun\c c\~ao de regress\~ao $\eta$ \'e cont\'\i nua. Ent\~ao, quando $n,k\to\infty$ e $k/n\to 0$, o erro de classificador $k$-NN converge quase certamente para o erro de Bayes:
\[\mbox{\em erro}_{\mu,\eta}({\mathcal L}_n^{k\mbox{\tiny -NN}})\overset{q.c.}\to \ell^{\ast}(\mu,\eta).\]
\label{t:irrealista}
\end{teorema}

No caso de desempate determin\'\i stico (ou, por exemplo, se a probabilidade de empates \'e nula), quase certamente significa em rela\c c\~ao aos pares $(\varsigma,\e)\in \Omega^{\infty}\times\{0,1\}^\infty=(\Omega\times\{0,1\})^{\infty}$, seguindo a lei $\tilde\mu^\infty$. 

No caso de desempate aleat\'orio, temos que adicional uma vari\'avel aleat\'oria auxiliar, $Z$, tomando valores num espa\c co boreliano padr\~ao, $\Upsilon$, com a lei $\nu$. Por exemplo, se usamos uma ordem aleat\'oria sobre a amostra $\sigma$ para desempatar, $Z$ pode tomar valores no produto $\prod_{n=1}^{\infty} S_n$ de grupos sim\'etricos, com a lei sendo a medida produto das medidas uniformes correspondentes. Neste caso, ``quase certamente'' se aplica \`as triplas $(\varsigma,\e,z)$, onde $z\in \Upsilon^{\infty}$, um elemento aleat\'orio de $\Upsilon$ para desempatar cada amostra $\varsigma_n$ (caso for necess\'ario). O espa\c co boreliano
\[(\Omega\times\{0,1\})^{\infty}\times \Upsilon^{\infty}\]
\'e munido da lei $\tilde\mu^\infty\otimes \nu^{\infty}$.

\begin{observacao} 
Gra\c cas ao teorema de converg\^encia dominada, o teorema \ref{t:irrealista} implica, em particular, que o classificador $k$-NN \'e consistente sobre $\Omega$ se a 
fun\c c\~ao de regress\~ao $\eta$ \'e cont\'\i nua. (Veja a observa\c c\~ao \ref{o:bayes}).
\end{observacao}

Como adicionar a vari\'avel $Z$ n\~ao causa dificuldades, frequentemente vamos a suprimir e trabalhar principalmente com o caso de desempate determin\'\i stico (mesmo se o desempate aleat\'orio \'e o mais comum, pelo menos na teoria).

As vezes \'e c\^omodo expressar a lei $\tilde\mu^\infty\otimes\mu$ de modo diferente. Seja $\mu$ a proje\c c\~ao da medida $\tilde\mu$ de $\Omega\times\{0,1\}$ sobre $\Omega$. O caminho aleat\'orio n\~ao rotulado, $\varsigma$, segue a lei $\mu^{\infty}$. Agora seja $\varsigma$ (para qual usamos a mesma letra) uma realiza\c c\~ao fixa do caminho aleat\'orio, 
\[\varsigma = (x_1,x_2,\ldots,x_n,\ldots)\in\Omega^{\infty}.\]
A sequ\^encia de r\'otulos $Y=(Y_1,Y_2,\ldots,Y_n,\ldots)$ para os pontos do caminho $\varsigma$ \'e uma vari\'avel aleat\'oria com valores no produto infinito $\{0,1\}^\infty$ (um espa\c co de Cantor) munido da medida produto
\[\tilde\mu^{\varsigma} = \{\eta(x_1),1-\eta(x_1)\}\otimes \{\eta(x_2),1-\eta(x_2)\}\otimes \ldots \otimes \{\eta(x_n),1-\eta(x_n)\}\otimes\ldots\]
Aqui, usamos a nota\c c\~ao bastante comum, $\{p,q\}$, significando um espa\c co probabil\'\i stico padr\~ao com apenas dois pontos, $0$ e $1$, tendo as probabilidades $p$ e $q$ respetivamente, $p+q=1$.  

A medida $\tilde\mu^{\varsigma}$ da ``fibra'' $\{0,1\}^\infty$ sobre o caminho $\varsigma$ (que consiste de todas as rotulagens poss\'\i veis $\e$ de $\varsigma$) \'e diferente para caminhos diferentes. 
Na linguagem probabil\'\i stica, $\tilde\mu^{\varsigma}$ \'e a distribui\c c\~ao condicional dado $\varsigma$. 
A fam\'\i lia de medidas $\tilde\mu^{\varsigma}$, $\varsigma\in (\Omega^{\infty},\mu^{\infty})$ \'e um exemplo de {\em desintegra\c c\~ao} da medida $\tilde\mu^{\infty}$ (veja se\c c\~ao \ref{s:rokhlin}). A medida $\tilde\mu^{\varsigma}$ pode ser vista tamb\'em como uma medida de probabilidade sobre o espa\c co $\Omega^\infty\times\{0,1\}^\infty$, suportada sobre a fibra $\{\varsigma\}\times\{0,1\}^{\infty}$.

\begin{exercicio}
Mostrar que, qualquer seja o conjunto boreliano $B\subseteq \Omega^\infty\times\{0,1\}^\infty$, temos
\[\tilde\mu^{\infty}(B) = \int_{\Omega^{\infty}} \tilde\mu^{\varsigma}(B)\,d\mu^{\infty}(\varsigma).\]
[ {\em Sugest\~ao:} basta mostrar o resultado no caso onde $B$ \'e um conjunto cil\'\i ndrico, e o teorema de Fubini permite reduzir tudo ao caso $n=1$, onde a defini\c c\~ao da fun\c c\~ao de regress\~ao deve ser usada... ]
\end{exercicio}

Precisamos tamb\'em reexaminar alguns resultados sobre a concentra\c c\~ao de medida no cubo de Hamming. 

\begin{exercicio}
Deduzir dos teoremas \ref{t:leigeneralizada} e \ref{t:parametrizacao} a forma seguinte, formalmente mais geral, da lei geom\'etrica de grandes n\'umeros. Sejam $(\Omega_i,\mu_i)$ espa\c cos probabil\'\i sticos padr\~ao, e
seja $f\colon (\prod_{i=1}^n \Omega_i,\bar d)\to\R$ uma fun\c c\~ao $1$-Lipschitz cont\'\i nua com rela\c c\~ao  a dist\^ancia de Hamming normalizada, e mensur\'avel com rela\c c\~ao  a estrutura boreliana do produto. Ent\~ao para todo $\ve>0$ temos
\[\left(\otimes_{i=1}^n\mu_i\right)
\left\{x\in\prod_{i=1}^n \Omega_i\colon \left\vert f(x)-\E f\right\vert\geq\ve\right\}\leq 2e^{-2\ve^2n}.\]
\label{e:leigeneralizada}
\end{exercicio}

\begin{exercicio}
Sejam $(p_i,q_i)$, $i=1,2,\ldots,n$ espa\c cos de Bernoulli, $p_i,q_i\geq 0$, $p_i+q_i=1$. Coloquemos sobre o cubo de Hamming $\{0,1\}^n$ a medida produto
\[\mu=\otimes_{i=1}^n \{p_i,q_i\}.\]
Ent\~ao, a esperan\c ca do peso normalizado $\bar w$ sobre o cubo
satisfaz
\[\E_{\mu}\bar w = \frac{p_1+p_2+\ldots+p_n}n.\]
[ {\em Sugest\~ao:} usar a defini\c c\~ao de $\bar w$ junto com a da medida produto... ]
\label{e:piqi}
\end{exercicio}

O teorema \ref{t:irrealista} vai seguir da afirma\c c\~ao mais forte.

\begin{lema} Sob a hip\'otese do teorema \ref{t:irrealista} ($(\Omega,\mu)$ \'e um espa\c co m\'etrico munido de uma medida de probabilidade, $\eta$ \'e uma fun\c c\~ao de regress\~ao cont\'\i nua), h\'a converg\^encia quase certa:
\[\frac 1k\sum_{x_i\in k\mbox{\tiny -NN}^{\sigma}(x)}\e_i \to \eta(x).\]
\label{l:etaempirica}
\end{lema}

\begin{observacao}
A express\~ao a esquerda \'e a {\em fun\c c\~ao de regress\~ao emp\'\i rica}.
\end{observacao}

\begin{observacao}
``Quase certamente'' neste contexto significa com rela\c c\~ao \`as triplas $(\varsigma,\e,x)\in \Omega^{\infty}\times\{0,1\}^\infty\times\Omega=
(\Omega\times\{0,1\})^{\infty}\times\Omega$, seguindo a lei $\tilde\mu^\infty\otimes\mu$. O resultado n\~ao depende da estrat\'egia do desempate. 

No caso de desempate aleat\'orio, automaticamente segue-se a converg\^encia certa com rela\c c\~ao com os qu\'adruplos $(\varsigma,\e,x,z)\in\Omega\times\{0,1\})^{\infty}\times \Omega\times \Upsilon^{\infty}$, seguindo a lei $\tilde\mu^\infty\otimes\mu\otimes \nu^{\infty}$, onde $\nu$ \'e a lei da v.a. $Z$ usado para desempatar quando for necess\'ario.
\end{observacao}

Dada a afirma\c c\~ao do lema acima, segue-se que, quase certamente,
\[\theta\left[\frac 1k\sum_{x_i\in k\mbox{\tiny -NN}^{\sigma}(x)}\e_i -\frac 12\right] \to \theta\left[\eta(x) -\frac 12\right] = T_{bayes}(x)
\]
quando $\eta\in \Omega\setminus Z_\eta$ (ou seja, $\eta(x)\neq 1/2$), e agora o teorema de Lebesgue de converg\^encia dominada, junto com o corol\'ario \ref{c:erroetaZ} e a defini\c c\~ao do classificador $k$-NN, implicar\~ao que, quase certamente (com rela\c c\~ao  s\'o ao caminho amostral, esta vez),
\begin{align*}
\mbox{erro}_{\mu,\eta}{\mathcal L}^{k\mbox{\tiny -NN}}_n(\sigma,\e)
&= \int_{\Omega\setminus Z_\eta}\left\vert
\theta\left[\frac 1k\sum_{x_i\in k\mbox{\tiny -NN}^{\sigma}(x)}\e_i -\frac 12\right]-\eta(x)\right\vert d\mu(x) \\
&\to \int_{\Omega\setminus Z_\eta}\abs{T_{bayes}(x)-\eta(x)}\,d\mu(x) \\
&=\mbox{erro}_{\mu,\eta}T_{bayes} \\
&=\ell^{\ast}(\mu,\eta).
\end{align*}

\begin{proof}[Prova do lema \ref{l:etaempirica}] 
Segue-se do lema de Cover--Hart e da continuidade de $\eta$ que, quase certamente em $(\varsigma,x)$, o conjunto de valores $\eta(k\mbox{-NN}(x))$ converge para $\eta(x)$ no sentido \'obvio: dado $\ve>0$, existe $N$ tal que quando $n\geq N$,
\[\eta(k\mbox{-NN}(x))\subseteq (\eta(x)-\ve,\eta(x)+\ve).\]
Seja $\ve>0$ qualquer fixo, e seja $(\varsigma,x)$ um par qualquer tendo a propriedade acima.
Segundo os exerc\'\i cios \ref{e:leigeneralizada} e \ref{e:piqi}, a probabilidade do evento
\[\frac{\sum_{i=1}^k\e_i}{k}\notin (\eta(x)-2\ve,\eta(x)+2\ve)\]
\'e exponencialmente pequena, limitada por $2\exp(-4\ve_0^2 k)$. Como  $\sum_{k=1}^\infty 2\exp(-4\ve_0^2 k)<\infty$, 
o 1$^o$ lema de Borel--Cantelli (exerc\'\i cio \ref{ex:1lemaborel-cantelli})
implica (no mesmo esp\'\i rito que na prova da lei geom\'etrica forte de grandes n\'umeros, lema \ref{l:munmu}) que, $\otimes_{n=1}^{\infty} \{\eta(x_i),1-\eta(x_i)\}$-quase certamente em $\e$, o evento acima s\'o vai ocorrer para um n\'umero finito de $k$. Aplicando-o \`a sequ\^encia $\ve=1/m$, conclu\'\i mos: $\otimes_{n=1}^{\infty} \{\eta(x_i),1-\eta(x_i)\}$-quase certamente, 
\[\frac{\sum_{i=1}^k\e_i}{k}\to \eta(x).\]
\end{proof}

A fim de se livrar da hip\'otese de continuidade de $\eta$, precisamos do lema geom\'etrico de Stone para $k$ qualquer.

\begin{lema}[Lema geom\'etrico de Stone para $\R^d$]
Para todo $d$ natural, existe uma constante absoluta $C=C(d)$ com a propriedade seguinte.
Seja 
\[\sigma=(x_1,x_2,\ldots,x_n),~x_i\in\R^d,~i=1,2,\ldots,n,\]
uma amostra finita qualquer em $\R^d$ (possivelmente com repeti\c c\~oes), e seja $x\in\R^d$ qualquer. Dado $k\in\N_+$, o n\'umero de $i$ tais que $x\neq x_i$ e $x$ fica entre os $k$ vizinhos mais pr\'oximos de $x_i$ dentro da amostra
\[x,x_1,x_2,\ldots, x_{i-1},x_i,x_{i+1},\dots,x_n\]
\'e majorado por $Ck$. 
\label{l:stonek=k}
\index{lema! geom\'etrico de Stone}
\end{lema}

\begin{observacao}
A desigualdade $x\neq x_i$ \'e entendida no sentido conjunt\'\i stico. Teoricamente, no caso dos empates, pode ocorrer que a amostra inteira consiste do mesmo ponto, $x_i=0$, repetido $n$ vezes, e $x=0$ tamb\'em. Por\'em, a conclus\~ao resta verdadeira, de modo trivial.
\end{observacao}

\begin{proof}
O argumento \'e quase id\^entico ao no caso $k=1$ (lema \ref{l:stonek=1}). Podemos supor que $x=0$. 
Cubramos $\R^d$ com $C$ cones, e dentro de todo cone escolhamos o m\'aximo n\'umero poss\'\i vel de vizinhos mais pr\'oximos de $0$ diferentes de $0$ pr\'oprio, at\'e $k$. (Alguns deles podem ser repetidos, se tiver repeti\c c\~oes dentro da amostra, mas nenhum \'e igual a $0$). No caso dos empates, fazemos a escolha de modo qualquer, \'e n\~ao importa. Desta maneira, $\leq Ck$ pontos est\~ao marcados. Seja agora $i$ qualquer. Se $x_i$ \'e diferente de $0$ e n\~ao foi marcado, ent\~ao o cone que cont\'em $x_i$ j\'a tem exatamente $k$ pontos marcados, e para cada um deles, $x_j$, temos $\norm{x_i-x_j}<\norm{x_i}$. Isso implica que $0\notin k\mbox{-NN}(x_i)$, qualquer seja a estrat\'egia de desempate.
\end{proof}

Finalmente, o resultado cl\'assico, cuja prova ocupa o resto da se\c c\~ao.

\begin{teorema}[Charles J. Stone]
O classificador $k$-NN \'e universalmente consistente no espa\c co euclidiano $\R^d$, sob a estrat\'egia de desempate qualquer.
\label{t:stone}
\index{teorema! de Charles Stone}
\end{teorema}

Vamos escrever $\mu_{k\mbox{\tiny -NN}^{\varsigma_n}(x)}$ para a medida uniforme suportada pelos $k$ vizinhos mais pr\'oximos de $x$ na amostra $\varsigma_n$.

\begin{lema}
Em $\R^d$,
\[\E\left\vert \E_{\mu_{k\mbox{\tiny -NN}^{\varsigma_n}(x)}}\eta - \eta
\right\vert \to 0\mbox{ quando }n\to\infty.\]
\label{l:eta}
\end{lema}

Seja $\ve>0$. Segundo o teorema de Luzin \ref{t:luzin}, existe um compacto $K\subseteq\Omega$ com $\mu(K)>1-\ve$ tal que a restri\c c\~ao $\eta\vert_K$ \'e cont\'\i nua. Os dois exerc\'\i cios a seguir permitem estender $f\vert_K$ at\'e uma fun\c c\~ao (uniformemente) cont\'\i nua, $g$, sobre $\Omega$, com valores em $[0,1]$. 
\label{extensaodefuncoescontinuas}

\begin{exercicio}
Seja $f$ uma fun\c c\~ao $1$-Lipschitz cont\'\i nua com valores em $[0,1]$ sobre um subespa\c co m\'etrico $Y$ de um espa\c co m\'etrico $X$. Mostrar que a fun\c c\~ao $\tilde f$ sobre $X$, dada por
\[\tilde f(x) =\max\{1,\inf\{f(y)-d(x,y)\colon y\in Y\}\]
\'e uma fun\c c\~ao $1$-Lipschitz cont\'\i nua bem definida, com valores em $[0,1]$, que estende $f$. 
\par
[ Para uma solu\c c\~ao, veja a prova do lema \ref{l:extensaoL}. ]
\end{exercicio}

\begin{exercicio}
Seja $f$ uma fun\c c\~ao uniformemente cont\'\i nua sobre um subespa\c co m\'etrico $Y$ de um espa\c co m\'etrico $(X,d)$. Mostrar que existe uma m\'etrica $\rho$ sobre $X$, uniformemente equivalente a $d$, tal que $f$ \'e $1$-Lipschitz cont\'\i nua com rela\c c\~ao  a $\rho\vert_Y$.
\par
[ {\em Sugest\~ao:} a m\'etrica $\rho$ pode ser obtida como uma {\em transforma\c c\~ao m\'etrica} de $d$, ou seja, sob a forma $\rho(x,y)= h(d(x,y))$, onde $h\colon\R_+\to\R_+$ \'e uma fun\c c\~ao estritamente mon\'otona e c\^oncava, com $h(0)=0$... ]
\end{exercicio}

Segue-se que $\norm{\eta-g}_{L^1(\mu)}<\ve$. 
Podemos escrever:
\begin{align*}
\E \left\vert \E_{\mu_{k\mbox{\tiny -NN}^{\varsigma_n}(x)}}\eta - \eta
\right\vert &\leq 
\underbrace{\E \left\vert \E_{\mu_{k\mbox{\tiny -NN}^{\varsigma_n}(x)}}(\eta -  g) \right\vert}_{I} + 
\underbrace{\E \left\vert \E_{\mu_{k\mbox{\tiny -NN}^{\varsigma_n}(x)}} g  - g \right\vert}_{II} + 
\underbrace{\E \left\vert g - \eta\right\vert}_{III}. \\
\end{align*}

O termo $(III)$ \'e limitado por $\ve$ gra\c cas \`a escolha de $g$. O termo $(II)$ \'e uma sequ\^encia convergente para zero quando $n\to\infty$ gra\c cas ao teorema de converg\^encia dominada de Lebesgue, porque 
\[\E_{\mu_{k\mbox{\tiny -NN}^{\varsigma_n}(x)}} g  \to g\]
quase certamente (lema \ref{l:etaempirica}). Note que at\'e agora, tudo funciona para um espa\c co m\'etrico separ\'avel e completo qualquer, n\~ao s\'o em $\R^d$. 

Falta somente estimar o termo $(I)$. Para isso, precisamos do lema geom\'etrico de Stone, que garante que o ``conjunto ruim'', $U=\Omega\setminus K$, n\~ao exerce uma influ\^encia indevida sobre os valores preditos pelo classificador $k$-NN. Temos:

\begin{align*}
(I) &\leq  \E  \E_{\mu_{k\mbox{\tiny -NN}^{\varsigma_n}(X)}}\left\vert\eta -  g \right\vert 
\\
&=
\E \frac 1 k \sum_{X_i\in k\mbox{\tiny -NN}(X)}\left\vert\eta(X_i)-g(X_i)\right\vert 
\\
&= \E \frac 1 k \sum \left\{\left\vert\eta(X_i)-g(X_i)\right\vert \colon
X_i\in k\mbox{-NN}(X)
\right\}
\\
&= \underbrace{\E \frac 1 k \left(\ldots \mid X\in U\right)}_{A}+ \underbrace{\E \frac 1 k \left(\ldots \mid X\in K\right)}_{B}.
\end{align*}
O termo $(A)$ \'e limitado por $\ve$, pois $\mu(U)<\ve$. Para avaliar o termo $(B)$, note que se $X_i\in K$, ent\~ao o valor sob esperan\c ca se anula. Por conseguinte,
\begin{align*}
(B) & =  \E \frac 1 k \sum \left\{\left\vert\eta(X_i)-g(X_i)\right\vert \colon
X_i\in k\mbox{-NN}(X),~X\in K,~X_i\notin K
\right\} \\
&\leq  \E \frac 1 k \sum \left\{\left\vert\eta(X_i)-g(X_i)\right\vert \colon
X_i\in k\mbox{-NN}(X),~X_i\neq X
\right\}.
\end{align*}

A transposi\c c\~ao de coordenadas $\tau_i\colon X\leftrightarrow X_i$ conserva a medida $\mu$, logo:

\begin{align*}
(B) &\leq  \E \frac 1 k \sum \left\{\left\vert\eta(X)-g(X)\right\vert \colon
X\in k\mbox{-NN}(X_i),~X_i\neq X
\right\} 
\\
&=  \E \frac 1 k \left[\left\vert\eta(X)-g(X)\right\vert \sharp \{i=1,2,\ldots,n\colon X\in k\mbox{-NN}(X_i),~X_i\neq X\}\right]
\\
&\leq 
\E \left\vert\eta(X)-g(X)\right\vert \frac 1 k C(d)k 
\\
&< \ve C(d),
\end{align*}
usando o lema geom\'etrica de Stone. Isso termina a prova do lema \ref{l:eta}.

Agora a prova do teorema de Stone \ref{t:stone} \'e finalizada usando um argumento parecido ao da prova do teorema \ref{t:irrealista}, com as modifica\c c\~oes necess\'arias usando a converg\^encia em probabilidade ao inv\'es da converg\^encia quase certa.

O teorema de Stone permane\c ca v\'alido para qualquer norma sobre $\R^d$.

\begin{exercicio}
Modificar a prova do lema geom\'etrico de Stone para mostrar que o classificador $k$-NN \'e universalmente consistente em todo espa\c co normado de dimens\~ao finita. 
(Uma prova pode ser achada em \citep*{duan}.)
\end{exercicio}

\begin{observacao}
Intuitivamente, quando mais dados temos, podemos escolher um classificador melhor. Isso \'e, espera-se que o erro de classifica\c c\~ao esperado ao n\'\i vel $n+1$ seja menor do que ao n\'\i vel $n$. Mais formalmente, uma regra de classifica\c c\~ao ${\mathcal L}=({\mathcal L}_n)$,
\[{\mathcal L}_n\colon \Omega\times (\Omega\times \{0,1\})^n \to \{0,1\},
\]
\'e dita {\em inteligente} ({\sl smart} \citep*{DGL}) se para todas as medidas de probabilidade $\mu$ sobre $\Omega\times \{0,1\}$ a sequ\^encia de valores de erros de classifica\c c\~ao
\[\E_{\mu}\{L({\mathcal L}_n)\},~~n=1,2,3,\ldots\]
\'e n\~ao crescente, onde 
$L({\mathcal L}_n) = P\{{\mathcal L}_n(X,\sigma_n)\neq Y\mid \sigma_n\}$.

Surpreendentemente, as regras consistentes mais comuns (incluindo o classificador $k$-NN) n\~ao s\~ao inteligentes.  

{\em Conjetura em aberto:} nenhuma regra de aprendizagem universalmente consistente \'e inteligente (\citep*{DGL}, Problem 6.16, p. 109).
\index{regra! de aprendizagem! inteligente}
\end{observacao}

\begin{exercicio}
Mostre que o classificador $k$-NN em geral n\~ao \'e ``inteligente''. 
\par
[ {\em Sugest\~ao:} fa\c ca o c\'alculo do erro de classifica\c c\~ao no dom\'\i nio $\Omega=[-1,1]$ munido da lei seguinte: $\mu_0$ \'e a medida uniforme sobre o intervalo, com a massa total de $p$, e $\mu_1$ \'e puramente at\^omica, com $\mu_1\{0\}=q$, $p+q=1$. Para valores apropriadas de $p$ e $q$, o erro de classificador $1$-NN \'e menor do que o do classificador $2$-NN (em rela\c c\~ao \`a dist\^ancia euclidiana). ] O exemplo foi emprestado de \citep*{DGL}.
\end{exercicio}

\section{Dimens\~ao de Nagata}

Comecemos a se\c c\~ao com dois exemplos de espa\c cos normados de dimens\~ao infinita, onde o classificador $k$-NN n\~ao \'e universalmente consistente. O primeiro exemplo pertence a Roy O. Davis, e o segundo, onde o espa\c co \'e um espa\c co de Hilbert $\ell^2$, a David Preiss. Ambos exemplos s\~ao elegantes, bastante simples, e muito pouco conhecidos.

Os dois exemplos, juntamente com o teorema de Stone, sugerem que um espa\c co m\'etrico no qual o classificador $k$-NN \'e universalmente consistente, deve ter dimens\~ao finita em algum sentido intuitivo. A no\c c\~ao relevante de dimens\~ao finita para espa\c cos m\'etricos ser\'a considerada a seguir, \'e a chamada dimens\~ao de Nagata.

\subsection{Um espa\c co de Banach onde o classificador $k$-NN n\~ao \'e consistente}

Nesta subse\c c\~ao, vamos mostrar um exemplo fino, devido a Roy O. Davis \citep*{davis}. 
\index{classificador! k-NN@$k$-NN! exemplo de Davis}

\subsubsection{Panor\^amica da situa\c c\~ao}
O exemplo original de \citep*{davis} \'e um espa\c co m\'etrico compacto $\Omega$ de di\^ametro um e duas medidas borelianas $\mu_1$, $\mu_2$ sobre $\Omega$ tais que $\mu_0(\Omega)=2/3$, $\mu_1(\Omega)=1/3$, e ao mesmo tempo para todo $0<r<1$ temos $\mu_0(\bar B_r(x))=\mu_1(\bar B_r(x))$ qualquer seja $x\in \Omega$. (Exerc\'\i cio: deduzir a mesma propriedade para bolas abertas correspondentes, assim como para esferas).

Vamos apresentar o exemplo na subse\c c\~ao \ref{ss:davis}, bem como observar que esse espa\c co pode ser imerso isometricamente no espa\c co de Banach $c_0$ de sequ\^encias convergentes para zero. 

\begin{exercicio}
Seja $\mu=\mu_0+\mu_1$. Interpretemos $\mu_i$ como a distribui\c c\~ao de pontos de $\Omega$ tendo r\'otulo $i$, $i=0,1$. Segue-se que $\int_0^1\eta(x)\,d\mu(x)=1/3$, e $\ell^\ast(\mu,\eta)\leq 1/3$ (testemunhando pelo classificador $T\equiv 0$).
Mostrar que no limite $n\to\infty$ o classificador $k$-NN (com desempate digamos mais comum, aleat\'orio e uniforme) vai assinar o r\'otulo $1$ a metade de pontos:
\begin{equation}
\label{eq:metade01}
P[{\mathcal L}_n^{\mbox{\tiny k-$NN$}}(X)=1]\to \frac 12\mbox{ quando }n\to\infty.\end{equation}
Concluir que 
\[\mbox{erro}_{\mu}({\mathcal L}_n^{\mbox{\tiny k-$NN$}})\to \frac 12 > \frac 13 \geq \ell^{\ast}.\]
[ {\em Sugest\~ao:} reescrever a probabilidade na eq. (\ref{eq:metade01}) com ajuda do jeito de condicionamento (veja subse\c c\~ao \ref{ss:rokhlin}),
\[\E{\mathcal L}_n^{\mbox{\tiny k-$NN$}}(X)=
\E_r \left[\E{\mathcal L}_n^{\mbox{\tiny k-$NN$}}(X)\mid r_{\mbox{\tiny k-$NN$}}(X) =r\right],\]
e usar o lema de Cover--Hart. ]
\end{exercicio}

\subsubsection{Exemplo de Roy O. Davis\label{ss:davis}}
Seja $(N_n)$ uma sequ\^encia de inteiros positivos, a ser escolhida recursivamente mais tarde. Denotemos $E_n$ o conjunto dos elementos $(i,j)$, $1\leq i\leq N_n$, $0\leq j\leq N_n$; os elementos $(i,0)$ sendo chamados {\em centrais,} os elementos $(i,j)$, $j\geq 1$, {\em perif\'ericos}.  

Definamos uma estrutura de grafo, $G_n$, sobre $E_n$, onde todos os elementos centrais s\~ao adjacentes (ou seja, formam um grafo completo com $N_n$ vertices), e todo elemento perif\'erico $(i,j)$ \'e adjacente ao elemento central correspondente $(i,0)$. Definamos uma m\'etrica, $d_n$, sobre $E_n$ como segue: para $x,y\in E_n$, 
\[d_n(x,y) =\begin{cases} \frac{1}{2^n},&\mbox{ se $x,y$ s\~ao adjacentes,} \\
\frac{1}{2^{n-1}},&\mbox{ sen\~ao.}\end{cases}
\]

\begin{exercicio}
Dado um conjunto $\Gamma$, o espa\c co normado $\ell^{\infty}(\Gamma)$ consiste de todas fun\c c\~oes limitadas $f\colon\Gamma\to\R$ munidas da estrutura can\^onica de um espa\c co vetorial e da norma
\[\norm{f}_{\infty}=\sup_{\gamma\in\Gamma}\abs{f(\gamma)}.\]
Verificar que $\ell^{\infty}(\Gamma)$ \'e um espa\c co de Banach, e que $\ell^{\infty}(\Gamma)$ \'e isometricamente isomorfo as espa\c co $\ell^{\infty}(n)$ no caso $\sharp\Gamma=n$.
\end{exercicio}

\begin{exercicio}
Seja $X=\{x_1,x_2,\ldots,x_n\}$ um espa\c co m\'etrico finito. Mostrar que a aplica\c c\~ao 
\[X\ni x\mapsto [f\colon y\mapsto d(x,y)]\in \ell^{\infty}(n)=\ell^\infty(\{x_1,x_2,\ldots,x_n\}),\]
\'e uma imers\~ao isom\'etrica (a {\em imers\~ao de Kuratowski}, veja subse\c c\~ao \ref{ss:imersaokuratowski}).
\end{exercicio}

Seja $\Omega=\prod_{n=1}^{\infty} E_n$. 
\index{espa\c co! de Cantor}
A m\'etrica sobre $\Omega$ \'e definida pela regra: dado $x=(x_i),y=(y_i)\in\Omega$, $x\neq y$,
\[d(x,y) = d_n(x_n,y_n),\]
onde $n$ \'e o menor \'\i ndice tal que $x_n\neq y_n$.
Como o di\^ametro de $E_n$ \'e igual a $2^{n-1}$, e assim menor ou igual \`a dist\^ancia entre dois elementos distintos quaisquer em $E_{n-1}$, a defini\c c\~ao da dist\^ancia $d$ pode se escrever assim:
\begin{equation}
d(x,y) = \max_{n}d_n(x_n,y_n).
\end{equation}

\begin{exercicio}
Verificar que o espa\c co m\'etrico $\Omega$ \'e compacto, e que a m\'etrica $d$ induz a topologia produto
(veja subse\c c\~ao \ref{ss:topologiadeproduto}).
\end{exercicio} 

\begin{exercicio}
O espa\c co de Banach $c_0$
\index{espa\c co! c0@$c_0$}
consiste de todas as sequ\^encias de n\'umeros reais convergentes para zero, $x=(x_n)$, $x_n\to 0$, munidos da norma uniforme (induzida de $\ell^{\infty}$):
\[\norm{x} = \sup_{n}\abs{x_n}\]
(de fato, sobre $c_0$ \'e o m\'aximo).
Mostrar que $\Omega$ admite uma imers\~ao isom\'etrica no espa\c co de Banach $c_0$.
\end{exercicio}

Seja $B$ uma bola fechada em $\Omega$. Pode-se supor que $B=\bar B_{1/2^n}(x)$, onde $n\geq 1$ e $x=(x_i)$. Ent\~ao $B$ consiste de todos $x^\prime$ tais que $x_i^\prime=x_i$ para todos $1\leq i<n$, e $x_n^\prime$ \'e adjacente a $x_n$ em $G_n$, ou $x_n^\prime=x_n$. Se $x_n$ \'e de g\^enero $(i,0)$, ent\~ao todos os elementos poss\'\i veis $x_n^\prime$ s\~ao
\[(i,1),\ldots,(i,N_n),~~(1,0),\ldots,(N_n,0),\]
enquanto se $x_n$ \'e de forma $(i,j)$, $1\leq j\leq N_n$, ent\~ao todos os elementos poss\'\i veis $x_n^\prime$ s\~ao $(i,j)$ mesmo e $(i,0)$. Isso implica que {\em a proje\c c\~ao de $B$ sobre $E_n$ tem o n\'umero igual dos elementos centrais e perif\'ericos.}

Formamos o conjunto cil\'\i ndrico:
\[\Omega(x_1,\ldots,x_n)=\{x_1\}\times\{x_2\}\times\ldots\times \{x_n\}\times E_{n+1}\times E_{n+2}\times \ldots\]
(onde $\Omega(\emptyset)=\Omega$). Vamos definir duas sequ\^encias de n\'umeros $\alpha_n,\beta_n$, $n=0,1,2,\ldots$, com $\alpha_n>\beta_n>0$, e $\alpha_0=\frac 23$, $\beta_0=\frac 23$, bem como duas medidas borelianas $\mu_0,\mu_1$ sobre $\Omega$, tendo a propriedade: quando tiver o n\'umero dos inteiros $1\leq i\leq n$ tais que $x_i$ \'e central em $E_i$, ent\~ao
\begin{equation}
\label{eq:casoa}
\mu_0(\Omega(x_1,\ldots,x_n))=\alpha_n,~~\mu_1(\Omega(x_1,\ldots,x_n))=\beta_n,
\end{equation}
embora no caso contr\'ario
\begin{equation}
\label{eq:casob}
\mu_0(\Omega(x_1,\ldots,x_n))=\beta_n,~~\mu_1(\Omega(x_1,\ldots,x_n))=\alpha_n.
\end{equation}

As medidas $\mu_0,\mu_1$ s\~ao n\~ao at\^omicas e satisfazem $\mu_0(\Omega)=\frac 23$, $\mu_1(\Omega)=\frac 13$.
Em vista de texto em it\'alico acima, toda bola fechada $B$ de raio $r<1$ \'e igual \`a uni\~ao de um n\'umero igual de conjuntos cil\'\i ndricos de ambos tipos, e por conseguinte, satisfaz $\mu_0(B)=\mu_1(B)$.

Segundo o teorema de extens\~ao de Carath\'eodory \ref{t:extensao}, basta verificar a condi\c c\~ao seguinte:
\begin{equation}
\mu_k(\Omega(x_1,x_2,\ldots,x_{n-1})=\sum_{x_n\in E_n} \mu_k(\Omega(x_1,x_2,\ldots,x_{n-1},x_n)),~~k=0,1.
\label{eq:condicoes_cil}
\end{equation}
Vamos definir $\alpha_n,\beta_n,N_n$ recursivamente, come\c cando com $\alpha_0=\frac 23,\eta_0=\frac 13$, como segue. Supondo que $\alpha_{n-1}>\beta_{n-1}>0$, escolhamos $N_n>1$ t\~ao grande que o sistema das equa\c c\~oes lineares
\[\begin{cases}
N_n^2a+N_nb &=\alpha_{n-1}, \\
N_n^2 b +N_na &=\beta_{n-1}
\end{cases}
\]
tem uma solu\c c\~ao em n\'umeros positivos $a,b$.

\begin{exercicio}
Verificar que uma tal solu\c c\~ao existe se $N_n>\alpha_{n-1}/\beta_{n-1}$.
\end{exercicio}

\begin{exercicio}
Verificar que $a>b$ e $4a<\alpha_{n-1}$.
\end{exercicio}

Definamos $\alpha_n=a$, $\beta_n=b$. As sequ\^encias $(\alpha_n)$, $(\beta_n)$ tem as propriedades desejadas. Para verificar eq. (\ref{eq:condicoes_cil}), suponha que o n\'umero dos inteiros $i$, $1\leq i\leq n-1$, tais que $x_i$ \'e um elemento central, \'e par. Ent\~ao, (\ref{eq:casoa}) tem lugar para os elementos perif\'ericos $x_n\in E_n$ e (\ref{eq:casob}) para os elementos centrais. Para $k=1$, o valor de lado direito da eq. (\ref{eq:condicoes_cil}) \'e igual a $N_n^2\alpha_n+N_n\beta_n=\alpha_{n-1}$, igual ao valor de lado esquerda. Os tr\^es casos restantes s\~ao parecidos.

\subsection{Classificador $k$-NN n\~ao \'e consistente no espa\c co de Hilbert $\ell^2$\label{ss:preiss}} Aqui, vamos reproduzir o exemplo de David Preiss \citep*{preiss1}.
\index{classificador! k-NN@$k$-NN! exemplo de Preiss}

Seja $Q=\prod_{k=1}^{\infty} [N_k]$ um espa\c co de Cantor
\index{espa\c co! de Cantor}
com a topologia produto (subs. \ref{ss:topologiadeproduto}), onde adotemos uma nota\c c\~ao combinat\'oria para $[n]=\{1,2,\ldots n\}$, e $(N_k)$ \'e uma sequ\^encia de n\'umeros naturais $\geq 2$. Denotemos $\pi_k$ a proje\c c\~ao can\^onica de $Q$ sobre o cubo de dimens\~ao $k$,  $Q_k=\prod_{i=1}^{k} [N_i]$. Seja ${\mathcal H}=\ell^2(\cup_{k=1}^{\infty} Q_k)$ um espa\c co de Hilbert cuja base ortonormal \'e indexada com elementos da uni\~ao dos cubos $Q_k$ (veja exemplo \ref{ex:ell2}).

Para todo $\bar n=(n_1,\ldots,n_k)$ em cada cubo $Q_k$ define 
\[f(\bar n)=\sum_{i=1}^k 2^{-i}e_{(n_1,\ldots,n_i)}\in {\mathcal H}.\]

\begin{exercicio} 
Mostrar que a aplica\c c\~ao $f$ se estende por continuidade sobre a uni\~ao $Q^\ast$ de cubos sobre o espa\c co de Cantor $Q$. 
\end{exercicio}

Denote $\mu_1=f_{\ast}(\nu)$ a imagem direta da medida de Haar sobre $Q$ (o produto de medidas uniformes sobre cada conjunto finito $[N_k]$). 

\begin{exercicio}
Verificar que para todo $r>0$ satisfazendo $2^{-k}\leq r^2<2^k$, e para todo $\bar n =(n_1,n_2,\ldots)\in Q$, 
\[\mu_1(B_r(\bar n)) = (N_1N_2\ldots N_{k+1})^{-1}.\]
\end{exercicio}

Agora define uma medida puramente at\^omica $\mu_0$, suportada na imagem de $Q^\ast$:
\[\mu_0=\sum_k\sum_{\bar n\in Q_k} a_k\delta_{\bar n},\]
onde $a_k>0$ est\~ao escolhidos de modo que a medida seja finita:
\begin{equation}
\sum_{k=1}^{\infty} a_k N_1N_2\ldots N_k<\infty.
\label{eq:finite}
\end{equation}
Para $r$ e $\bar n\in Q$ como acima a bola $B_r(\bar n)$ cont\'em $f(n_1n_2,\ldots,n_k)$, logo
\[\mu_1(B_r(\bar n)) \geq a_k.\]
Supondo, al\'em disso, que 
\begin{equation}
a_kN_1N_2\ldots N_kN_{k+1} \to \infty\mbox{ as }k\to\infty,
\label{eq:infinity}
\end{equation}
conclu\'\i mos:
\[\frac{\mu_1(B_r(\bar n))}{\mu_0(B_r(\bar n))}\to 0\mbox{ when } r\downarrow 0.\]

\begin{exercicio}
Mostrar que as condi\c c\~oes (\ref{eq:finite}) e (\ref{eq:infinity}) podem ser satisfeitos pela escolha recursiva de $(N_k)$ e $(a_k)$. 
\end{exercicio}

Now renormalizemos as medidas $\mu_0$ e $\mu_1$ de modo que $\mu=\mu_0+\mu_1$ \'e uma medida de probabilidade, e interpretemos $\mu_i$ como a distribui\c c\~ao de pontos rotulados $i=0,1$. Deste modo, a fun\c c\~ao de regress\~ao, $\eta$, \'e determin\'\i stica, e estamos aprendendo o conceito  $C=f(Q)={\mathrm{supp}}\,\mu_1$, $\mu_1(C)>0$. 

Para um elemento aleat\'orio $X\in {\mathcal H}$, $X\sim\mu$, a dist\^ancia $r_{\mbox{\tiny k-$NN$}}(X)$ converge para zero quase certamente pelo lema de Cover e Hart. 

\begin{exercicio}
Deduzir das observa\c c\~oes acima que a probabilidade condicional para qualquer dos $k$ vizinhos mais pr\'oximos de $X=x$ a ser rotulado $1$ converge para zero quase certamente. 
\end{exercicio}

A regra de $k$ vizinhos mais pr\'oximos vai quase certamente predizer um classificador identicamente zero, logo n\~ao \'e consistente. 

Um exemplo mesmo mais natural no espa\c co $\ell^2$, por\'em n\~ao trivial (com $\mu$ sendo uma medida de Wiener e o conceito $C$ n\~ao aprendiz\'avel, um elipsoide), foi apresentado em \citep*{preiss}.

\subsection{Dimens\~ao de Nagata}

Relembremos que uma fam\'\i lia $\gamma$ de subconjuntos de um conjunto $\Omega$ tem {\em multiplicidade} $\leq\delta$ se a interse\c c\~ao de mais de $k$ elementos distintos de $\gamma$ \'e sempre vazia. Em outras palavras,
\[\forall x\in\Omega,~~\sum_{V\in\gamma}\chi_V(x)\leq\delta.\]

\begin{definicao}
Sejam $\delta\in\N$, $s\in (0,+\infty]$.
Digamos que um espa\c co m\'etrico $(\Omega,d)$ tem {\em dimens\~ao de Nagata} $\leq\delta$ {\em na escala $s>0$}, se toda fam\'\i lia {\em finita} de bolas fechadas $\gamma$ de raios $< s$ tem uma subfam\'\i lia $\gamma^\prime$ de multiplicidade $\leq\delta+1$ que cobre todos os centros de bolas de $\gamma$.

Um espa\c co $\Omega$ tem dimens\~ao de Nagata $\delta$ se ele tem dimens\~ao de Nagata $\delta$ em uma escala apropriada $s\in (0,+\infty]$. Vamos denotar: $\dim^s_{Nag}(\Omega)=\delta$, ou simplesmente $\dim_{Nag}(\Omega)=\delta$.
\index{dimens\~ao! de Nagata}
\index{dimsnag@$\dim^s_{Nag}$}
\end{definicao}

\begin{exercicio}
Mostre que a no\c c\~ao acima \'e equivalente \`a propriedade seguinte de $\Omega$. Dado uma sequ\^encia $x_1,\ldots,x_{\delta+2}\in \bar B_r(x)$, $r<s$, existem $i,j$ tais que $d(x_i,x_j)\leq \max\{d(x,x_i),d(x,x_j)\}$.
\par
[ {\em Sugest\~ao:} usar contraposi\c c\~ao para necessidade, e um argumento recursivo para sufici\^encia. ]
\label{ex:nagata2}
\end{exercicio}

\begin{exercicio}
Deduzir que todo espa\c co m\'etrico n\~ao arquimediano (isto \'e, satisfazendo a desigualdade triangular forte $d(x,z)\leq \max\{d(x,y),d(y,z)\}$) tem dimens\~ao de Nagata $0$ na escala $+\infty$.
\index{m\'etrica! n\~ao arquimediana}
\end{exercicio}

\begin{exercicio} 
Mostrar (tamb\'em usando o exerc\'\i cio \ref{ex:nagata2}) que $\dim_{Nag}(\R)=1$, e $\dim_{Nag}(\R^2)=2$.
\end{exercicio}

\begin{exercicio} 
Mostrar que todo espa\c co normado de dimens\~ao finita $n$ tem a dimens\~ao de Nagata finita, mais n\~ao necessariamente $n$. (Estudar o exemplo de $\ell^{\infty}(n))$, bem como os espa\c cos euclidianos de dimens\~ao $n>2$).
\end{exercicio}

\begin{exercicio} Construir uma m\'etrica equivalente na reta $\R$ cuja dimens\~ao de Nagata \'e infinita, em qualquer escala $s>0$. 
\par
[ {\em Sugest\~ao:} realizar $\R$ como uma curva cont\'\i nua e sem autointerse\c c\~oes no espa\c co de Hilbert $\ell^2$, passando pelo zero e todos vetores de uma base ortogonal, e induzir a dist\^ancia de l\'a. ] 

Modificar o exemplo para construir uma m\'etrica equivalente sobre uma sequ\^encia convergente com limite, que tem a dimens\~ao de Nagata infinita em qualquer escala.
\end{exercicio}

\begin{observacao}
Pode se mostrar que um espa\c co topol\'ogico metriz\'avel, $X$, tem a dimens\~ao de Lebesgue $n$ se e somente se a topologia de $X$ pode ser gerada por uma m\'etrica de dimens\~ao de Nagata $n$. (Teorema de Nagata--Ostrand \citep*{ostrand}).
\end{observacao}

Espa\c cos m\'etricos de dimens\~ao de Nagata finita admitem uma vers\~ao do lema geom\'etrico de Stone, pelo menos no caso onde a amostra n\~ao tem empates, ou seja, os valores das dist\^ancias $d(x_i,x_j)$, $i\neq j$, s\~ao todas dois a dois diferentes.

\begin{lema}[Lema geom\'etrico de Stone, dimens\~ao de Nagata finita, sem empates]
Seja $X$ um espa\c co m\'etrico da dimens\~ao de Nagata $\delta<\infty$.
Seja 
\[\sigma=(x_1,x_2,\ldots,x_n),~x_i\in\R^d,~i=1,2,\ldots,n,\]
uma amostra finita em $X$, e seja $x\in X$ qualquer. Suponha que n\~ao tem empates na amostra 
\[x,x_1,x_2,\ldots, x_{i-1},x_i,x_{i+1},\dots,x_n.\]
Dado $k$, o n\'umero de $i$ tais que $x\neq x_i$ e $x$ fica entre os $k$ vizinhos mais pr\'oximos de $x_i$ dentro da amostra acima
\'e majorado por $(k+1)(\delta+1)$. 
\label{l:stone_sem_empates}
\index{lema! geom\'etrico de Stone}
\end{lema}

\begin{proof}
Suponha que $x_i$, $i=1,2,\ldots,m$ tem $x$ entre os $k$ vizinhos mais pr\'oximos.
A fam\'\i lia $\gamma$ das bolas fechadas $B_{r_{k\mbox{\tiny -NN}}(x_i)}(x_i)$, $i\leq m$, admite uma subfam\'\i lia $\gamma^\prime$ de multiplicidade $\leq\delta+1$ que cobre todos pontos $x_i$, $i\leq m$. Porque n\~ao h\'a empates, cada bola em $\gamma$ contem $\leq k+1$ pontos. Segue-se que $\sharp\gamma^\prime\geq m/(k+1)$. Todas bolas em $\gamma^\prime$ contem $x$, e como a multiplicidade de $\gamma^\prime$ \'e limitada por $\delta+1$, conclu\'\i mos: $\sharp\gamma^\prime\leq\delta+1$. O resultado se segue.
\end{proof}

Agora o mesmo argumento que na p´rova de Stone mostra que o classificador $k$-NN \'e consistente sob cada distribui\c c\~ao $\mu$ sobre $\Omega\times\{0,1\}$ tal que a probabilidade de empates \'e nula. (Em outras palavras, se cada esfera de raio estritamente positivo \'e $\mu$-negligenci\'avel).

Ao contr\'ario do caso euclidiano, a afirma\c c\~ao do lema \ref{l:stone_sem_empates} \'e falsa na presen\c ca dos empates.

\begin{exemplo}
A conclus\~ao do lema geom\'etrico de Stone \'e falsa se a amostra $\sigma$ com $n>k$ pontos dois a dois distintos est\'a munida de uma m\'etrica zero-um. Por exemplo, o desempate baseado sobre a ordem na amostra escolher\'a $x_1,x_2,\ldots,x_k$ como vizinhos mais pr\'oximos de todo outro ponto.

Ao mesmo tempo, a dimens\~ao de Nagata de um espa\c co $X$ munido da m\'etrica zero-um \'e igual a $\delta=0$. Se a fam\'\i lia $\gamma$ de bolas fechadas cont\'em uma bola de raio $\geq 1$, esta bola j\'a cobre $X$. Se n\~ao, escolha de $\gamma$ uma bola de raio $<1$ para cada centro de $\gamma$. As bolas escolhidas tem a mesma uni\~ao que $\gamma$ e s\~ao duas a duas disjuntas, ou seja, da multiplicidade $1$. 

Vale a pena observar que no caso euclidiano um tal exemplo \'e imposs\'\i vel: o tamanho de uma amostra cuja m\'etrica induzida \'e a m\'etrica zero-um \'e limitada pela dimens\~ao do espa\c co!
\end{exemplo}

Poderia se esperar que a conclus\~ao do lema geom\'etrico de Stone seja v\'alida no caso de dimens\~ao de Nagata finita no sentido probabil\'\i stico, se o desempate for feito com ajuda de uma ordem aleat\'oria e uniforme sobre a amostra. Mesmo assim, o resultado \'e falso. 

\begin{exemplo} Dado $N\in\N$, existe um espa\c co m\'etrico finito $\sigma=\{x_1,\ldots,x_n\}$ de dimens\~ao de Nagata $0$, tal que a esperan\c ca de n\'umero de pontos $x_i$ tais que o ponto $x_1$ \'e o vizinho mais pr\'oximo deles sob um desempate aleat\'orio uniforme \'e pelo menos $N$.

Vamos construir $\sigma$ por recorr\^encia. Come\c camos com $x_1$ qualquer, e seja $\sigma_1=\{x_1\}$. Adicionemos $x_2$ \`a dist\^ancia $1$ de $x_1$, de modo que $\sigma_2=\{x_1,x_2\}$. Se $\sigma_n$ foi definido, adicionemos $x_{n+1}$ \`a dist\^ancia $2^n$ de todos os pontos $x_i$, $i\leq n$, e definamos $\sigma_{n+1}=\sigma_n\cup\{x_{n+1}\}$. Verifiquemos indutivamente que $\dim_{Nag}(\sigma_n)=0$. Para $n=1$ \'e trivial. Suponhamos que a conclus\~ao seja v\'alida para $\sigma_n$, e seja $\mathcal F$ uma fam\'\i lia qualquer de bolas fechadas em $\sigma_{n+1}$. Se existe uma bola que cont\'em todos os pontos, h\'a nada a mostrar. Supondo que n\~ao, ou seja, todas as bolas t\^em raios menos de $2^{n+1}$, escolhemos uma fam\'\i lia de multiplicidade $1$ da subfam\'\i lia de bolas centradas em elementos de $\sigma_n$, e adicionemos a bola centrada em $x_{n+1}$ (um ponto isolado). 

Finalmente, mostremos que se $n$ \'e bastante grande, ent\~ao o n\'umero esperado de $i$ tais que $x_1$ \'e um vizinho mais pr\'oximo de $x_i$ sob um desempate uniforme \'e t\~ao grande quanto desejado. Com este prop\'osito, para todo $i\geq 2$ calculemos a esperan\c ca do evento $x_1\in NN(x_i)$.
Para $x_2$, o \'unico vizinho mais pr\'oximo \'e $x_1$, ent\~ao $\E[x_1\in NN(x_2)]=1$. Para $x_3$, temos dois pontos \`a dist\^ancia $2$, que podem ser escolhidos com a probabilidade $1/2$ cada um, $x_1$ e $x_2$, ent\~ao $\E[x_1\in NN(x_3)]=1/2$. Para $i$ qualquer, temos $\E[x_1\in NN(x_i)]=1/i$. Conclu\'\i mos:
\[\E[\sharp\{i=1,\ldots,n\colon x_1\in NN(x_i)\}=\sum_{i=1}^n\frac 1i,\]
que converge para $+\infty$ quando $n\to\infty$.
\label{ex:exemplonagata}
\index{empates}
\end{exemplo}

Poder\'\i amos considerar que empates n\~ao ocorrem na pr\'atica? Ou seja, pode ser que pelo menos para espa\c cos m\'etricos ``n\~ao degenerados'' de dimens\~ao m\'etrica finita a probabilidade de um empate \'e assintoticamente zero, quando $n\to\infty$? Mesmo essa esperan\c ca \'e infundada.

\begin{exemplo}
Dado um valor $\delta>0$ e uma sequ\^encia $n^\prime_k\uparrow+\infty$, existe um espa\c co m\'etrico compacto de dimens\~ao de Nagata zero (o conjunto de Cantor munido de uma m\'etrica compat\'\i vel), munido de uma medida de probabilidade, n\~ao at\^omica, e uma sequ\^encia $n_k\uparrow\infty$, $n_k\geq n^\prime_k$, $k/n_k\to 0$, tendo a propriedade seguinte. Seja $X$ um elemento aleat\'orio de $\Omega$, e $W$ um caminho amostral. Com probabilidade $>1-\delta$, para todos os valores $k\in\N_+$, $X$ tem $\geq n_k$ empates entre seus $k$-vizinhos mais pr\'oximos dentro da amostra $W_{n_{k+1}}$.

O espa\c co $\Omega$ \'e o produto direto $\prod_{k=1}^{\infty} [N_k]$ de espa\c cos topol\'ogicos finitos discretos, onde $N_k$ s\~ao escolhidos recursivamente e $[N_k]=\{1,2,\ldots,N_k\}$. A dist\^ancia \'e n\~ao arquimediana:
\[d(\sigma,\tau)=\begin{cases} 0,&\mbox{ se }\sigma=\tau,\\
2^{-\min\{i\colon \sigma_i\neq\tau_i\}},&\mbox{caso contr\'ario.}
\end{cases}\]
Ela induz a topologia compacta de produto sobre $\Omega$. A medida $\mu$ sobre $\Omega$ \'e a medida produto de medidas uniformes $\mu_{N_k}$ sobre os $[N_k]$.
Essa medida \'e n\~ao at\^omica, e em particular, quase todos os empatem ocorrem \`a dist\^ancia estritamente positiva de um ponto aleat\'orio $X$.

Seja $(\delta_i)$ uma sequ\^encia, tal que $\delta_i>0$, $2\sum_i\delta_i=\delta$. Escolha $N_1$ t\~ao grande que, com a probabilidade $>1-\delta_1$, $n_1=n^\prime_1$ elementos escolhidos aleatoriamente e independentemente entre $[N_1]$ (segundo a medida uniforme) s\~ao todos dois a dois distintos. Agora, escolha $n_2\geq n^\prime_2$ t\~ao grande que, se $n_2$ elementos est\~ao escolhidos aleatoriamente em $[N_1]$, ent\~ao com a probabilidade $>1-\delta_1$ cado elemento de $[N_1]$ \'e escolhido pelo menos $n_1$ vezes. Suponha que $n_1,N_1,n_2,N_2,\ldots,n_k$ foram escolhidas. Seja $N_{k}$ t\~ao grande que, com a probabilidade $>1-\delta_{k}$, $n_k$ elementos escolhidos aleatoriamente dentro $[N_k]$ s\~ao dois a dois distintos. Escolha $n_{k+1}$ t\~ao grande que, com a probabilidade $>1-\delta_{k}$, se $n_{k+1}$ elementos s\~ao escolhidos uniformemente dentro $\prod_{i=1}^k [N_i]$, ent\~ao todo elemento de $\prod_{i=1}^k [N_i]$ ser\'a escolhido pelo menos $n_k$ vezes.

Agora seja $(X,W)$ um elemento aleat\'orio de $\Omega\times\Omega^{\infty}$.
Com a probabilidade $>1-\delta$, o seguinte ocorre:
\begin{enumerate}
\item Para todo $k$, as ${k+1}$-\'esimas coordenadas de todos elementos de $W_{n_{k+1}}$ s\~ao duas a duas distintas, logo as dist\^ancias entre $X$ e o resto da amostra $W_{n_{k+1}}$ s\~ao $\geq 2^{-k-1}$;
\item
Para todo $k$,
existem $n_k$ elementos na amostra $W_{n_{k+1}}$ que t\^em as mesmas $i$-\'esimas coordenadas que $X$, para todos $i=1,2,3,\ldots,k$. Conclu\'\i mos: eles s\~ao \`a dist\^ancia exatamente $2^{-k-1}$ de $X$ e s\~ao os vizinhos mais pr\'oximos de $X$ dentro da amostra.
\end{enumerate}
Temos $n_k$ empates entre $k$ vizinhos mais pr\'oximos de $X$ dentro $W_{n_{k+1}}$, qualquer que seja $k$.
\end{exemplo}

Os exemplos acima mostram que, na presen\c ca de empates, a prova cl\'assica n\~ao se generaliza sobre os espa\c cos m\'etricos de dimens\~ao de Nagata finita. A fim de tratar dos empates, precisa-se uma abordagem t\'ecnica diferente. 

Para come\c car, vamos estender o nosso cen\'ario at\'e o mais geral poss\'\i vel.

\begin{definicao} Um subespa\c co m\'etrico $X$ de um espa\c co m\'etrico $\Omega$ tem {\em dimens\~ao de Nagata $\leq\delta\in\N$ na escala $s>0$ dentro de} $\Omega$ se toda fam\'\i lia finita de bolas fechadas em $\Omega$ com centros em $X$ admite uma subfam\'\i lia tendo multiplicidade $\leq\delta+1$ em $\Omega$ que cobre os centros de bolas originais. Diz-se que $X$ tem dimens\~ao de Nagata finita em $\Omega$ se $X$ tem dimens\~ao finita em $\Omega$ em uma escala $s>0$. Nota\c c\~ao: $\dim^s_{Nag}(X,\Omega)$ ou $\dim_{Nag}(X,\Omega)$.
\label{d:dimnagata}
\index{dimens\~ao! de Nagata}
\end{definicao}

\begin{exercicio}
Digamos que uma fam\'\i lia de bolas \'e {\em desconexa} (segundo Preiss), se o centro de toda bola da fam\'\i lia n\~ao pertence \`as outras bolas. Mostrar que
\[\dim^s_{Nag}(X,\Omega) \leq\beta\] 
se e somente se toda fam\'\i lia desconexa de bolas fechadas em $\Omega$ de raios $<s$ com centros em $X$ tem multiplicidade $\leq\beta+1$.
\label{ex:famdesconexa}
\end{exercicio}

\begin{exercicio}
Mostrar que na defini\c c\~ao \ref{d:dimnagata}, bem como no exerc\'\i cio \ref{ex:famdesconexa}, as bolas fechadas podem ser  substitu\'\i das pelas bolas abertas.
De fato, as afirma\c c\~oes restam v\'alidas para fam\'\i lias de bolas algumas das quais s\~ao abertas, algumas fechadas.
\end{exercicio}

\begin{exercicio}
Mostrar que se $\dim^s_{Nag}(X,\Omega)\leq \delta$, ent\~ao $\dim^s_{Nag}(\bar X,\Omega)\leq \delta$, onde $\bar X$ \'e a ader\^encia de $X$ em $\Omega$.
\label{ex:closurenagata}
\end{exercicio}

\begin{exercicio}
Mostrar que se $X$ e $Y$ s\~ao dois subespa\c cos de um espa\c co m\'etrico $\Omega$ que tem dimens\~ao finita em $\Omega$, ent\~ao $X\cup Y$ tem dimens\~ao finita em $\Omega$, com $\dim_{Nag}(X\cup Y,\Omega)\leq \dim_{Nag}(X,\Omega)+\dim_{Nag}(Y,\Omega)$. 
\label{ex:uniaonagata}
\end{exercicio}

\begin{definicao}
Um espa\c co $\Omega$ tem {\em dimens\~ao de Nagata sigma-finita} se $\Omega=\cup_{i=1}^{\infty}X_n$, onde cada subespa\c co m\'etrico $X_n$ tem dimens\~ao de Nagata finita em $\Omega$ (com escalas $s_n>0$ possivelmente todas diferentes).
\index{dimens\~ao! de Nagata! sigma-finita}
\end{definicao}

\begin{observacao}
Gra\c cas ao exerc\'\i cio \ref{ex:closurenagata}, na defini\c c\~ao acima pode-se supor que $X_n$ s\~ao fechados, em particular borelianos.
\end{observacao}

Eis a ferramenta t\'ecnica que permite manejar os empates nos espa\c cos m\'etricos mais gerais.

\begin{lema}
Seja $\sigma=\{x_1,x_2,\ldots,x_n\}$ uma amostra finita num espa\c co m\'etrico $\Omega$, e seja $X$ um subespa\c co de dimens\~ao de Nagata finita $\delta$ em $\Omega$ na escala $s>0$. Seja $\alpha\in (0,1]$ qualquer. Seja $\sigma^\prime\sqsubseteq\sigma $ uma subamostra com $m$ pontos.
Assinemos a todo ponto $x_i$ da amostra uma bola (fechada ou aberta), $B_i$, em torno de $x_i$, de raio $<s$.
Ent\~ao no m\'aximo $\alpha^{-1}(\delta+1)m$ pontos $x_i$ pertencendo a $X$ tem a fra\c c\~ao de $\geq\alpha$ pontos de $\sigma^\prime$ nas bolas $B_i$:
\[\sharp\{i=1,2,\ldots,n\colon x_i\in X,~~ \sharp(B_i\cap \sigma^\prime)\geq\alpha\sharp B_i\}\leq \alpha^{-1}(\delta+1)m.
\]
\label{l:alphadelta}
\end{lema}

\begin{proof}
A fam\'\i lia de todas as bolas $B_i$ tendo as propriedades $x_i\in X$ e 
\[\sharp(B_i\cap \sigma^\prime)\geq\alpha\sharp B_i\]
admite uma subfam\'\i lia de multiplicidade $\leq\delta+1$ que contem todos os centros. Cada ponto de $\sigma^\prime$ pertence, no m\'aximo, aos $\delta+1$ bolas da subfam\'\i lia.
A soma de cardinalidades de bolas dessa subfam\'\i lia, vez $\alpha$, n\~ao excede a cardinalidade de $\sigma^\prime$ vez $\delta+1$ (pois
cada ponto de $\sigma^\prime$ est\'a contado, no m\'aximo, $\delta+1$ vezes), de onde conclu\'\i mos.
\end{proof}

\begin{observacao}
Nas aplica\c c\~oes do lema, $B_i=B_{\e_{\mbox{\tiny{k-NN}}}}(x)$ (as vezes abertas, as vezes fechadas).
\end{observacao}

\begin{lema}
Sejam $\alpha,\alpha_1,\alpha_2\geq 0$, $t_1,t_2\in [0,1]$, $t_2\leq 1-t_1$. Suponha que $\alpha_1\leq\alpha$ e
\[t_1\alpha_1+(1-t_1)\alpha_2\leq\alpha.\]
Ent\~ao,
\[\frac{t_1\alpha_1+t_2\alpha_2}{t_1+t_2}\leq\alpha.\]
\label{l:fracoes}
\end{lema}

\begin{proof} Se $\alpha_2\leq\alpha$, a conclus\~ao \'e imediata. Caso contr\'ario, $\alpha_2\geq\alpha$, segue-se da hip\'otese que 
\[t_1\alpha_1+t_2\alpha_2\leq \alpha - (1-t_1-t_2)\alpha_2\leq (t_2+t_2)\alpha. \]
\end{proof}

\begin{lema}
Seja $x,x_1,x_2,\ldots,x_n$ uma amostra finita (talvez com repeti\c c\~oes), e $\sigma^\prime\sqsubset \sigma$ uma subamostra. Seja $\alpha\geq 0$, e seja $B$ uma bola fechada em torno de $x$ de raio $r_{k\mbox{\tiny -NN}}(x)$ que contem $K$ elementos da amostra,
\[\sharp\{i=1,2,\ldots,n\colon x_i\in B\}=K.\]
Suponha que a fra\c c\~ao de pontos de $\sigma^\prime$ contidos em $B$ seja menor ou igual a $\alpha$, 
\[\sharp\{i\colon x_i\in\sigma^\prime,~~x_i\in B\}\leq \alpha K,\]
e que da mesma forma, para a bola aberta, $B^\circ$,
\[\sharp\{i\colon x_i\in\sigma^\prime,~~x_i\in B^\circ\}\leq \alpha \sharp\{i\colon x_i\in B^\circ\}.\]
 Sob um desempate uniformemente aleat\'orio de $k$ vizinhos mais pr\'oximos, a fra\c c\~ao esperada de pontos de $\sigma^\prime$ entre os $k$ vizinhos mais pr\'oximos de $x$ \'e menor ou igual a $\alpha$.
\label{l:alphak}
\end{lema}

\begin{proof}
Aplica-se o lema \ref{l:fracoes} com $\alpha_1$ e $\alpha_2$ as fra\c c\~oes de pontos de $\sigma^\prime$ na bola $B$ e na esfera $S=B\setminus B^\circ$ respectivamente, $t_1=\sharp B^\circ /\sharp B$, e $t_2$ a fra\c c\~ao de pontos da esfera $S=B\setminus B^\circ$ que temos de escolher aleatoriamente como vizinhos mais pr\'oximos de $x$ faltados na bola aberta. Basta observar que a fra\c c\~ao esperada de pontos de $\sigma^\prime$ entre os vizinhos mais pr\'oximos que pertencem \`a esfera \'e tamb\'em igual a $\alpha_2$ pois s\~ao escolhidos aleatoriamente seguindo uma distribui\c c\~ao uniforme.
\end{proof}

Agora chegamos ao resultado principal desta se\c c\~ao.

\begin{teorema}
O classificador $k$-NN sob um desempate uniformemente aleat\'orio de $k$ vizinhos mais pr\'oximos \'e universalmente consistente em qualquer espa\c co m\'etrico de dimens\~ao de Nagata sigma-finita.
\label{t:nagatasigmafinitaconsistente}
\end{teorema}

\begin{proof}
Seja $\Omega=\cup_{l=1}^\infty Y_n$, onde $Y_n$ tem dimens\~ao de Nagata finita em $\Omega$. Segundo o exerc\'\i cio \ref{ex:uniaonagata}, pode-se supor que as $Y_n$ formam uma cadeia crescente, e o exerc\'\i cio \ref{ex:closurenagata} permite supor que $Y_n$ s\~ao borelianos. Sejam $\mu$ e $\eta$ quaisquer.
Dado $\e>0$, existe $l$ tal que $\mu(Y_l)\geq 1-\e/2$, e existe $K$ compacto, $K\subseteq Y_l$, tal que $\eta\vert_K$ \'e cont\'\i nua e $\mu(K)\geq 1-\e$. A fun\c c\~ao $\eta\vert_K$ se estende a uma fun\c c\~ao uniformemente cont\'\i nua $g$ sobre $\Omega$. 

No esp\'\i rito da prova do teorema de Stone \ref{t:stone}, basta estimar por cima o termo
\begin{align*}
(B)  &=  \E \frac 1 k \sum \left\{\left\vert\eta(X_i)-g(X_i)\right\vert \colon
X_i\in k\mbox{-NN}(X),~X\in K,~X_i\notin K
\right\} \\
&= \E  \E_{j\sim \mu_{\sharp}} \frac 1 k \sum \left\{\left\vert\eta(X_i)-g(X_i)\right\vert \colon
X_{i}\in k\mbox{-NN}(X_j),~X_{j}\in K,~X_{i}\notin K,~\right. \\
& \left.i\in \{0,1,\ldots,n\}\setminus\{j\}
\right\},
\end{align*}
onde $\mu_{\sharp}$ \'e a medida uniforme sobre o conjunto $\{0,1,2,\ldots,n\}$, e $X_0=X$.
Vamos tratar do termo $(B)$ como a soma de duas esperan\c cas condicionais, $(B_1)$ e $(B_2)$, segundo $k$ vizinhos mais pr\'oximos de $X_j$ dentro da amostra $\{X_0,X_1,\ldots, X_n\}$ cont\'em mais ou menos de $\sqrt\e k$ elementos de $U=\Omega\setminus K$. 

Usando o lema \ref{l:alphadelta}, aplicado \`as bolas fechadas de raio $k\mbox{-NN}(X_j)$ bem como \`as bolas abertas correspondentes, junto com lema \ref{l:alphak}, no primeiro caso temos
\begin{align*}
(B_1) &=
 \E  \E_{j\sim \mu_{\sharp}} \left[\frac 1 k \sum \left\{\left\vert\eta(X_i)-g(X_i)\right\vert \colon
X_{i}\in k\mbox{-NN}(X_j),~X_{j}\in K,~X_{i}\notin K,~\right.\right. \\
& \left.\left.i\in \{0,1,\ldots,n\}\setminus\{j\}
\right\} \mid ~~\sharp\{i\colon X_i\in k\mbox{-NN}(X_j),~~X_i\notin K\}\geq k\sqrt \e\right] \\
&\leq 
\E  \frac 1 k k 2\e^{-1/2} (\delta+1) \frac 1n\sharp\{i=0,1,\ldots,n\colon X_i\notin K\}\\
&\leq  2\e^{-1/2} (\delta+1) \e = 2\sqrt{\e}(\delta+1),
\end{align*}
onde usamos o fato e que a soma n\~ao excede $k$, e depois
a Lei dos Grandes N\'umeros.
No segundo caso,
\begin{align*}
(B_2) &= \E  \E_{j\sim \mu_{\sharp}} \left[\frac 1 k \sum \left\{\left\vert\eta(X_i)-g(X_i)\right\vert \colon
X_{i}\in k\mbox{-NN}(X_j),~X_{j}\in K,~X_{i}\notin K,~\right.\right. \\
& \left.\left.i\in \{0,1,\ldots,n\}\setminus\{j\}
\right\} \mid ~~\sharp\{i\colon X_i\in k\mbox{-NN}(X_j),~~X_i\notin K\}\leq k\sqrt \e\right] \\
&\leq \frac 1kk\sqrt \e = \sqrt \e.
\end{align*}
\end{proof}

\begin{observacao}
Uma boa refer\^encia para uma grande variedade de dimens\~oes m\'etricas, inclusive a de Nagata, \'e o artigo \citep*{assouad_gromard}.
\end{observacao}

\begin{observacao}
Teorema \ref{t:nagatasigmafinitaconsistente} inicialmente resultou de uma combina\c c\~ao dos resultados de \citep*{preiss83} e \citep*{CG}.

Um espa\c co m\'etrico munido de uma medida de probabilidade, $(\Omega,d,\mu)$, satisfaz a {\em propriedade de Lebesgue--Besicovitch} ({\em forte}) se para toda $f\in L^1(\mu)$
\[\frac{1}{\mu(B_\ve(x)}\int_{B_\ve(x)} f(x)\,d\mu(x) \to f(x)\]
em probabilidade (respectivamente, em quase toda parte). 

Foi mostrado em \citep*{CG} que a propriedade de Lebesgue--Besico\-vitch implica que o classificador $k$-NN \'e consistente. (A prova estende a de \citep*{devroye_1981}, onde a ideia foi elaborada para $\R^d$.)

E segundo \citep*{preiss83}, um espa\c co m\'etrico $\Omega$ satisfaz a propriedade de Lebesgue--Besicovitch forte para cada medida de probabilidade $\mu$ se e somente se $\Omega$ tem a dimens\~ao de Nagata sigma-finita. Isto implica o teorema \ref{t:nagatasigmafinitaconsistente}.

(Mencionemos que a prova de Preiss foi muito curta, apenas um esbo\c co. A sufici\^encia foi elaborada de modo detalhado em \citep*{assouad_gromard}, e a necessidade, em \citep*{kumari}.)

A prova atual do teorema \ref{t:nagatasigmafinitaconsistente}, assim como diversos exemplos do cap\'\i tulo, aparecem na tese \citep*{kumari}.
\end{observacao}

\begin{observacao}
Seja $X$ um espa\c co m\'etrico (separ\'avel e completo) tal que o classificador $k$-NN \'e universalmente consistente em $X$. Segue-se que $X$ tinha dimens\~ao de Nagata sigma finita?

A resposta positiva, modulo o resultado de \citep*{CG} mencionado acima, implicaria uma resposta positiva a uma pergunta da an\'alise real, j\'a bastante antiga \citep*{preiss83}:
n\~ao se sabe se $\Omega$ tem a propriedade de Lebesgue--Besicovitch para cada medida de probabilidade se e somente se $\Omega$ tem dimens\~ao de Nagata sigma-finita.
\end{observacao}

\begin{observacao}
N\~ao enfrentamos aqui o problema de consist\^encia universal forte do classificador $k$-NN. A consist\^encia universal forte no espa\c co euclidiano foi estabelecida em \citep*{devroye_gyorfi} e \citep*{zhao:87}. O argumento \'e o seguinte. Supondo que a medida $\mu$ sobre $\R^d$ (ou um espa\c co m\'etrico qualquer) n\~ao admite empates (ou seja, cada esfera \'e $\mu$-negligenci\'avel), para cada $\alpha>0$ e cada $x$ existe o m\'\i nimo $r=r(\alpha)(x)>0$ tal que $\mu(B_{r}(x))=\alpha$. Definamos a regra de aprendizagem ${\mathcal L}_n$ em $x$ pelo voto majorit\'ario entre os r\'otulos de todos os pontos $\sigma\cap B_{r(k/n)(x)}(x)$. N\~ao \'e dif\'\i cil a verificar que a regra ${\mathcal L}_n$ e a regra $k$-NN s\~ao {\em mutualmente fortemente consistentes,} ou seja, 
\[\mbox{erro}_{\mu,\eta}k\mbox{-NN} - \mbox{erro}_{\mu,\eta}{\mathcal L}\to 0\]
ao longo de quase todo caminho amostral. Segue-se em particular que $\mathcal L$ \'e universalmente consistente.
Usando o lema geom\'etrico de Stone mais alguns argumentos topol\'ogicos, pode-se mostrar que o erro de classifica\c c\~ao da regra ${\mathcal L}_n$ \'e uma fun\c c\~ao Lipschitz cont\'\i nua sobre $\Omega^n\times\{0,1\}^n$, com a constante de Lipschitz uniformemente limitada. Usando o lema \ref{l:prob+conc->qs}, conclu\'\i mos que $\mathcal L$ \'e fortemente consistente, logo o mesmo vale para a regra $k$-NN. 

O mesmo argumento se generaliza para espa\c cos m\'etricos de dimens\~ao de Nagata sigma-finita na aus\^encia de empates \citep*{kumari}.

A condi\c c\~ao de aus\^encia de empates foi removida para espa\c co euclidiano em \citep*{DevroyeGK}. 
\end{observacao}

\begin{observacao}
Seja $\mathscr M$ uma fam\'ilia de m\'etricas sobre um espa\c co boreliano padr\~ao, todas gerando a sua estrutura boreliana. Suponha que o classificador $k$-NN seja universalmente consistente em cada espa\c co m\'etrico $(\Omega,\rho)$, $\rho\in {\mathscr M}$. Formemos a regra de aprendizagem seguinte: dado uma amostra rotulada $\sigma$, seja ${\mathcal L}_n(\sigma)$ o valor do classificador $k$-NN sobre $\sigma$, em rela\c c\~ao \`a m\'etrica $\rho\in {\mathscr M}$ minimizando o erro emp\'\i rico de classificador $k$-NN. Sobre quais condi\c c\~oes sobre a fam\'\i lia $\mathcal M$ a tal regra ser\'a universalmente consistente? Intuitivamente, \'e o caso quando $\mathcal M$ tiver uma ``complexidade baixa'', ou, talvez, num sentido apropriado, estiver ``uniformemente de dimens\~ao de Nagata sigma-finita''. 
Seria interessante tentar e desenvolver uma pequena teoria, justificando a otimiza\c c\~ao do classificador $k$-NN sobre uma fam\'\i lia de m\'etricas (ou outras dist\^ancias).

Algum trabalho nesta dire\c c\~ao foi feito em \citep*{hatko:thesis}.

O problema \'e importante porque j\'a existem algoritmos deste tipo, tais como o classificador {\em LMNN} ({\em Large Margin Nearest Neighbour classifier}) nos espa\c cos euclidianos \citep*{weinberger_saul:09}. A pergunta que parece ficar em aberto \'e a seguinte: o classificador LMNN \'e universalmente consistente?
\end{observacao}

\begin{observacao}
Tradicionalmente, o classificador de vizinhos mais pr\'oximos \'e estudado e usado no contexto dos espa\c cos m\'etricos (como neste cap\'\i tulo), e sobretudo num espa\c co euclidiano $\ell^2(d)$ de dimens\~ao finita. Todavia, nenhum axioma de m\'etrica \'e necess\'aria para definir o classificador. Por outro lado, as ``medidas de semelhan\c ca'' $d(x,y)$ intr\'\i nsecos a alguns dom\'\i nios (por exemplo, biologia molecular) n\~ao s\~ao necessariamente sim\'etricas, nem satisfazem a desigualdade triangular \citep*{PeSt06}. A raz\~ao principal para a teoria ser quase exclusivamente desenvolvida para espa\c cos m\'etricos, \'e que ela \'e baseada sobre a teoria bem estabelecida das medidas de probabilidade em tais espa\c cos, praticamente inexistente para conjuntos munidos de dist\^ancias mais gerais (quasem\'etricas, etc.) Como j\'a mencionamos, o \^ambito natural para estudar e analisar os classificadores de vizinhos mais pr\'oximos  seria um espa\c co boreliano padr\~ao $\Omega$ munido de uma fam\'\i lia de pr\'e-ordens $\prec_x$, uma para cada ponto $x\in\Omega$, compat\'\i veis com a estrutura boreliana padr\~ao separ\'avel no sentido que os intervalos geram a estrutura, onde a interpreta\c c\~ao de $y\prec_xz$ \'e que $y$ fica mais perto de $x$ que $z$. A \'unica propriedade desej\'avel seria $y\prec_x x\Rightarrow y=x$.
Na presen\c ca de uma m\'etrica (ou outra medida de semelhan\c ca), $d$, a pr\'e-ordem \'e definida por
\begin{equation}
y\prec_x z\iff d(x,y)\leq d(x,z).
\label{eq:preordens}
\end{equation}
Pode-se argumentar que a teoria de classificadores de vizinhos mais pr\'oximos desta generalidade merece, seguramente, aten\c c\~ao. 

\'E f\'acil verificar que, dado uma fam\'\i lia de pr\'e-ordens $\prec_x$, $x\in\Omega$ com a propriedade $y\prec_x x\Rightarrow y=x$, existe uma m\'etrica $d$ sobre $\Omega$ que satisfaz Eq. (\ref{eq:preordens}). Deixemos isso como um exerc\'\i cio. Por\'em, tipicamente uma tal m\'etrica vai gerar a topologia discreta, a uma estrutura boreliana n\~ao separ\'avel, e por isso vai ser in\'util. N\~ao \'e claro se a pergunta seguinte j\'a foi tratada: sob quais condi\c c\~oes existe uma tal m\'etrica $d$ que gera a estrutura boreliana original de $\Omega$?
\end{observacao}

%
%

\chapter{Redu\c c\~ao de dimensionalidade\label{ch:reducao}}

Nos dom\'\i nios de alta dimens\~ao $ d\gg 1$, v\'arios algoritmos da ci\^encia de dados muitas vezes levam muito tempo e tornam-se ineficientes. Por exemplo, nenhum classificador padr\~ao vai funcionar se aplicado diretamente ao conjunto de dados seguinte.\footnote{O do Instituto de Cardiologia da Universidade de Ottawa.} Os dados s\~ao sequ\^encias gen\^omicas,
\[X\subseteq \{A,T,G,C\}^d,\]
onde a ``dimens\~ao'' (n\'umero de carater\'\i sticas) $d\sim 870,000$, enquanto a tamanho do conjunto n\~ao \'e muito grande ($n\sim 6,000$, os dados correspondem aos pacientes individuais).
No entanto, j\'a em dimens\~oes baixas a m\'edias (tais como $7$) alguns algoritmos tornam-se visivelmente menos eficientes que em dimens\~oes $1$ ou $2$. Este fen\^omeno \'e conhecido como a {\em maldi\c c\~ao de dimensionalidade}.
\index{maldi\c c\~ao de dimensionalidade}

Para remediar a situa\c c\~ao, s\~ao usados v\'arios algoritmos sob o nome comum de {\em redu\c c\~ao de dimensionalidade}. 
\index{redu\c c\~ao! de dimensionalidade}
Isto significa escolher uma fun\c c\~ao $f\colon \Omega\to \Upsilon$ do dom\'\i nio de dimens\~ao alta, $\Omega$, para um dom\'\i nio $\Upsilon$ de dimens\~ao mais baixa, transferindo o problema para l\'a.
No contexto da aprendizagem supervisionada, a fim de classificar um ponto de dados $x\in\Omega$ baseado sobre amostra rotulada $\sigma$, aplicamos um algoritmo de classifica\c c\~ao no espa\c co $\Upsilon$ ao ponto $f(x)$ e \`a amostra $f(\sigma)$, esperando que o desempenho do algoritmo em $\Upsilon$ seja mais eficaz de que em $\Omega$ por causa da baixa dimens\~ao, e que a fun\c c\~ao $f$ conserve os padr\~oes presentes no dom\'\i nio $\Omega$.

\begin{figure}[ht]
\begin{center}
    \scalebox{0.2}{\includegraphics{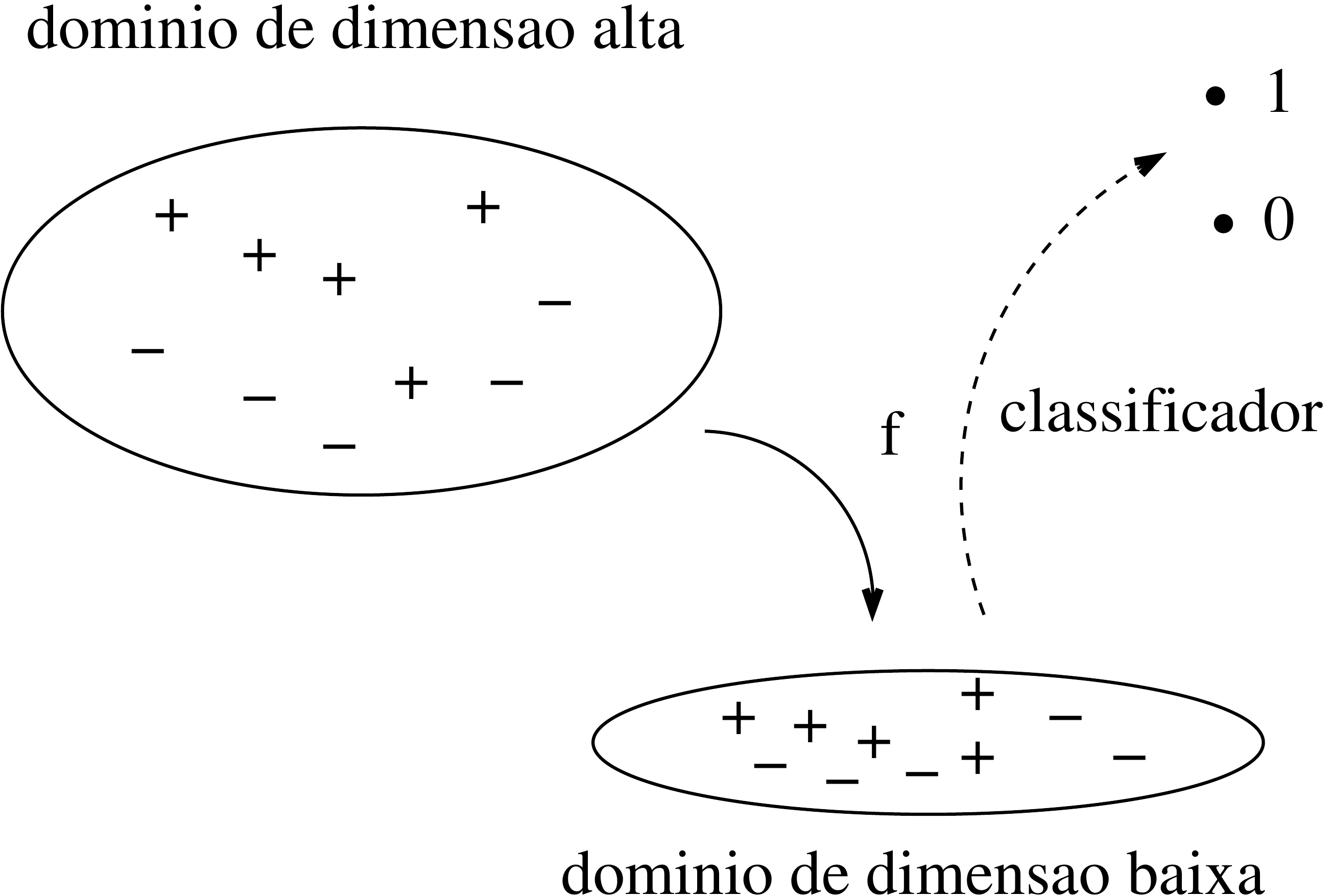}} 
    \caption{Redu\c c\~ao de dimensionalidade}
    \label{fig:reducao}
\end{center}
\end{figure}

Existem v\'arios m\'etodos de redu\c c\~ao de dimensionalidade. Para alguns aspectos da teoria, a chamada extra\c c\~ao de carater\'\i sticas, veja a cole\c c\~ao de artigos \citep{guyon}. Ao mesmo tempo, parece que ainda n\~ao existe muita base te\'orica deste procedimento no contexto de aprendizagem de m\'aquina. Isto \'e, em parte, porque o fen\^omeno de maldi\c c\~ao de dimensionalidade em si mesmo \'e ainda mal entendido. (Considere, por exemplo, a ``conjetura de maldi\c c\~ao de dimensionalidade'' em aberto, mencionada na subse\c c\~ao \ref{ss:redcubo}).
At\'e mesmo n\~ao \'e ainda claro o que significa ``dimens\~ao alta'' de um conjunto de dados: a teoria de dimens\~ao correspondente n\~ao foi ainda elaborada.
Por esta raz\~ao, qualquer apresenta\c c\~ao do assunto ser\'a necessariamente inconclusiva, disjunta, e teoricamente insatisfat\'oria. Por\'em, dado que o desafio \'e muito interessante do ponto de vista te\'orico bem como pr\'atico, arrisquemos incluir este cap\'\i tulo.

Discutimos alguns aspectos da maldi\c c\~ao de dimensionalidade, em particular o que pode significar uma ``dimens\~ao intr\'\i nseca alta'' de dados, e as liga\c c\~oes com a concentra\c c\~ao de medida. Depois, descrevemos um dos importantes algoritmos de redu\c c\~ao de dimensionalidade, o m\'etodo de proje\c c\~oes aleat\'orias baseado sobre o lema de Johnson--Lindenstrauss. Vamos discutir poss\'\i veis liga\c c\~oes do lema com aprendizagem supervisionada.
Conclu\'\i mos com a descri\c c\~ao de um novo algoritmo para redu\c c\~ao de dimensionalidade, usando as fun\c c\~oes injetoras borelianas, descont\'\i nuas em cada ponto -- um resultado simples, que no entanto pode tornar-se uma ferramenta importante. 

\section{Maldi\c c\~ao de dimensionalidade}

Como n\'os j\'a mencionamos, a natureza geom\'etrica e algor\'\i tmica da maldi\c c\~ao de dimensionalidade ainda resta um desafio te\'orico interessante. Nesta se\c c\~ao tentaremos entender um pouco, pelo menos ao n\'\i vel intuitivo, a origem de alguns problemas que afeitam o classificador $k$-NN. Por isso, o dom\'\i nio vai ser um espa\c co m\'etrico com medida de probabilidade, $(\Omega,d,\mu)$. 

\subsection{Geometria de busca por semelhan\c ca\label{ss:mdck}}
O classificador $k$-NN precisa de um algoritmo eficaz para buscar os $k$ vizinhos mais pr\'oximos do qualquer ponto $x$ do dom\'\i nio dentro da amostra, $\sigma$. Isto \'e onde os problemas come\c cam nas altas dimens\~oes, e intuitivamente, a origem de dificuldades \'e bastante clara. 

Seja $(\Omega,d,X)$ um dom\'\i nio munido de uma m\'etrica, $d$, e de uma amostra finita $X=\{x_1,x_2,\ldots,x_n\}$ (conjunto, ou {\em inst\^ancia}, de dados). O triplo $(\Omega,d,\sigma)$ \'e conhecido na teoria de busca por semelhan\c ca como uma {\em carga de trabalho de semelhan\c ca} ({\em similarity workload}).
\index{carga de trabalho de semelhan\c ca}
O que nos interessa, \'e a pesquisa de $k$ vizinhos mais pr\'oximos (consulta $k$-NN) dentro de $X$, dado um ponto de consulta ({\em query point}),
\index{ponto de consulta}
$q$. (A nossa nota\c c\~ao aqui vai ser um pouco diferente da usada na aprendizagem estat\'\i stica, porque nos passamos por uma parte diferente da inform\'atica).

\begin{figure}[ht]
\begin{center}
\scalebox{0.2}[0.2]{\includegraphics{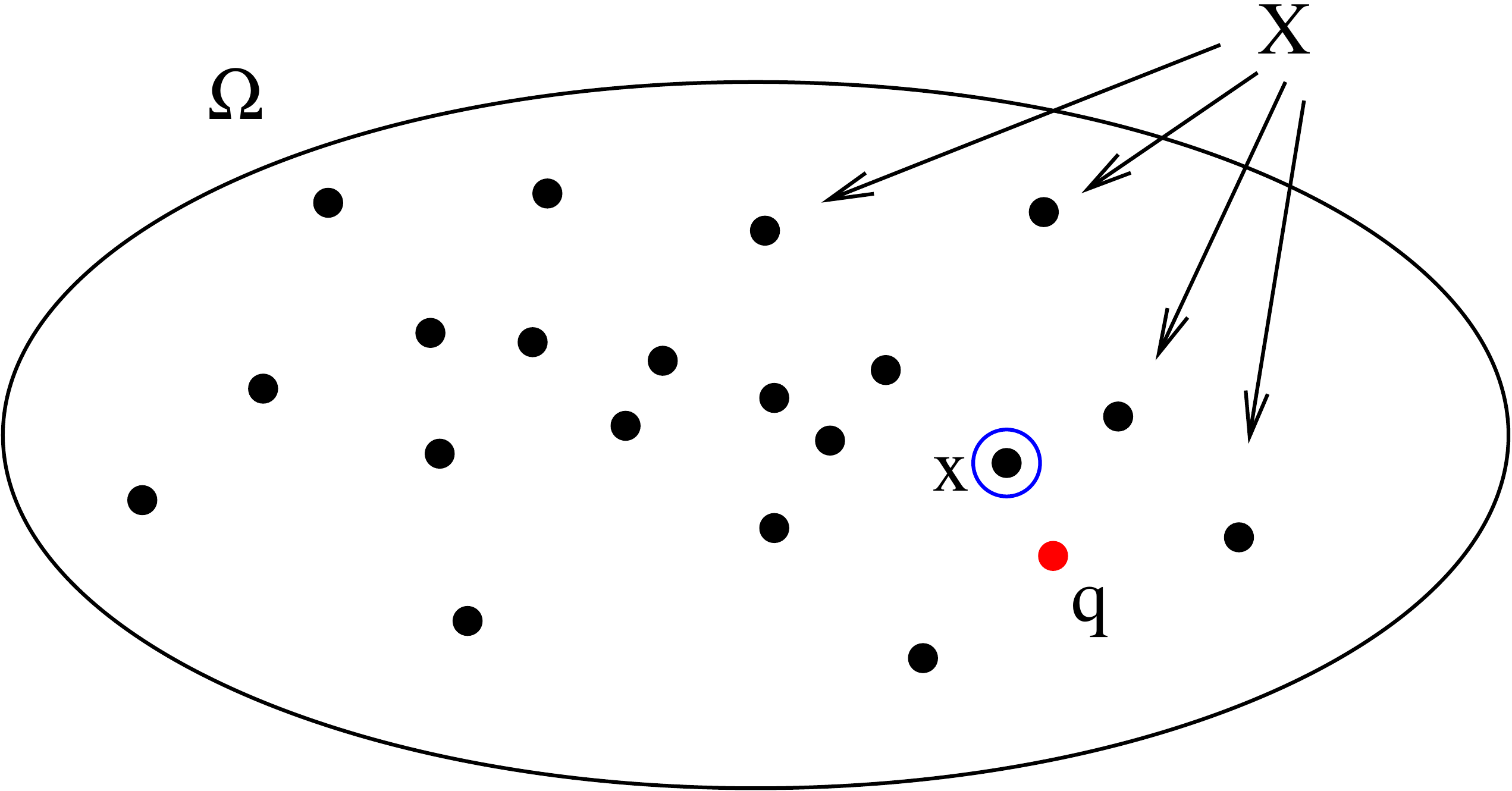}}
\hskip 1cm
\scalebox{0.2}[0.2]{\includegraphics{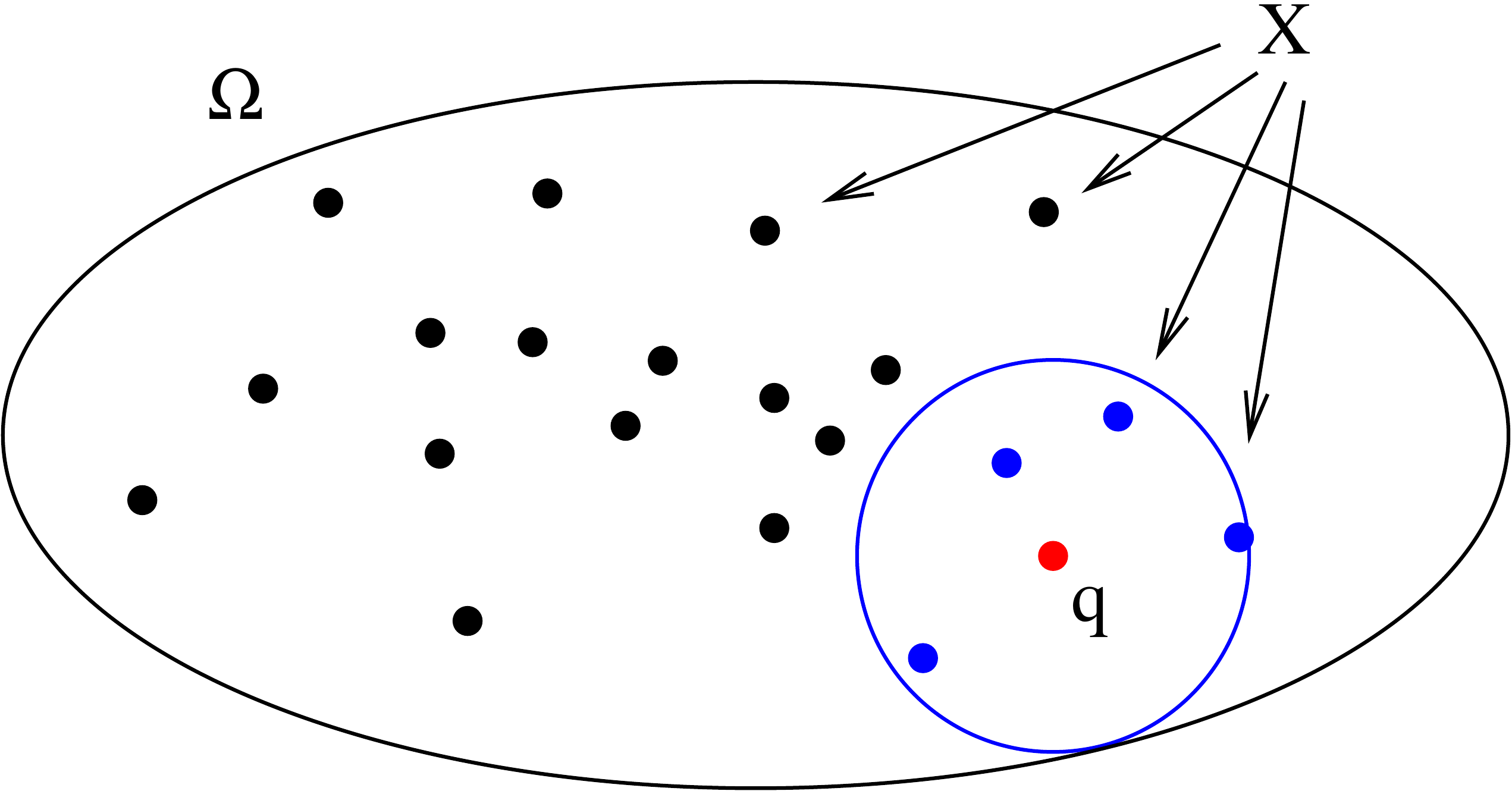}} 
\caption{Consulta do vizinho mais pr\'oximo (esquerda) e consulta de $\ve$-intervalo (dir.)}
\label{fig:pesquisa1}
\end{center}
\end{figure}

Uma possibilidade \'e a busca completa sequencial de $X$, calculando todos os valores $d(x_i,q)$, $i=1,2,\ldots,n$, ordenando-os, e escolhendo $k$ menores. Por exemplo, no caso de $k=1$, o algoritmo pode ser definido como seque:
\[i=\arg\min_i d(x_i,q).\]
A complexidade temporal deste algoritmo \'e de $O(n)$, o n\'umero de passos a executar sendo proporcional a $n$. Frequentemente, \'e o mais r\'apido algoritmo conhecido. No entanto, ele se torna devagar demais se o n\'umero de pontos, $n$, \'e grande, ou a m\'etrica \'e dif\'\i cil a calcular. \'E desej\'avel construir um {\em esquema de indexa\c c\~ao}, ou seja, uma estrutura sobre $(\Omega,d,X)$ capaz de executar uma busca mais r\'apida: de maneira ideal, no tempo $\mbox{poly}\,\log n$, ou $\mbox{poly}\, d$, onde $d$ \'e a ``dimens\~ao''. 
\index{esquema! de indexa\c c\~ao}

Um esquema de indexa\c c\~ao permite executar {\em consultas de intervalo} ({\em range queries}): 
\index{consulta de intervalo}
dado $q\in \Omega$ e $\ve>0$, retornar todos $x\in \bar B_{\ve}(q)\cap X$. (Depois, a busca bin\'aria permite reduzir a consulta $k$-NN a uma s\'erie de consultas de intervalo da maneira mais ou menos evidente e eficaz.)
Um esquema de indexa\c c\~ao tradicional, cingido ao essencial, consiste de uma fam\'\i lia de fun\c c\~oes reais $f_i$, $i\in I$ sobre $\Omega$, definidas totalmente ou parcialmente, que satisfazem a propriedade de Lipschitz com constante $L=1$:
\begin{equation}
\abs{f_i(x)-f_i(y)}\leq d(x,y).
\end{equation}
(Por exemplo, os chamados esquemas baseados sobre piv\^os usam como as fun\c c\~oes $f_i$ as da dist\^ancia $d(p_i,-)$ at\'e os {\em piv\^os} escolhidos, ou seja, pontos do dom\'\i nio $p_i\in\Omega$.)

\begin{figure}[ht]
\begin{center}
\scalebox{0.2}[0.2]{\includegraphics{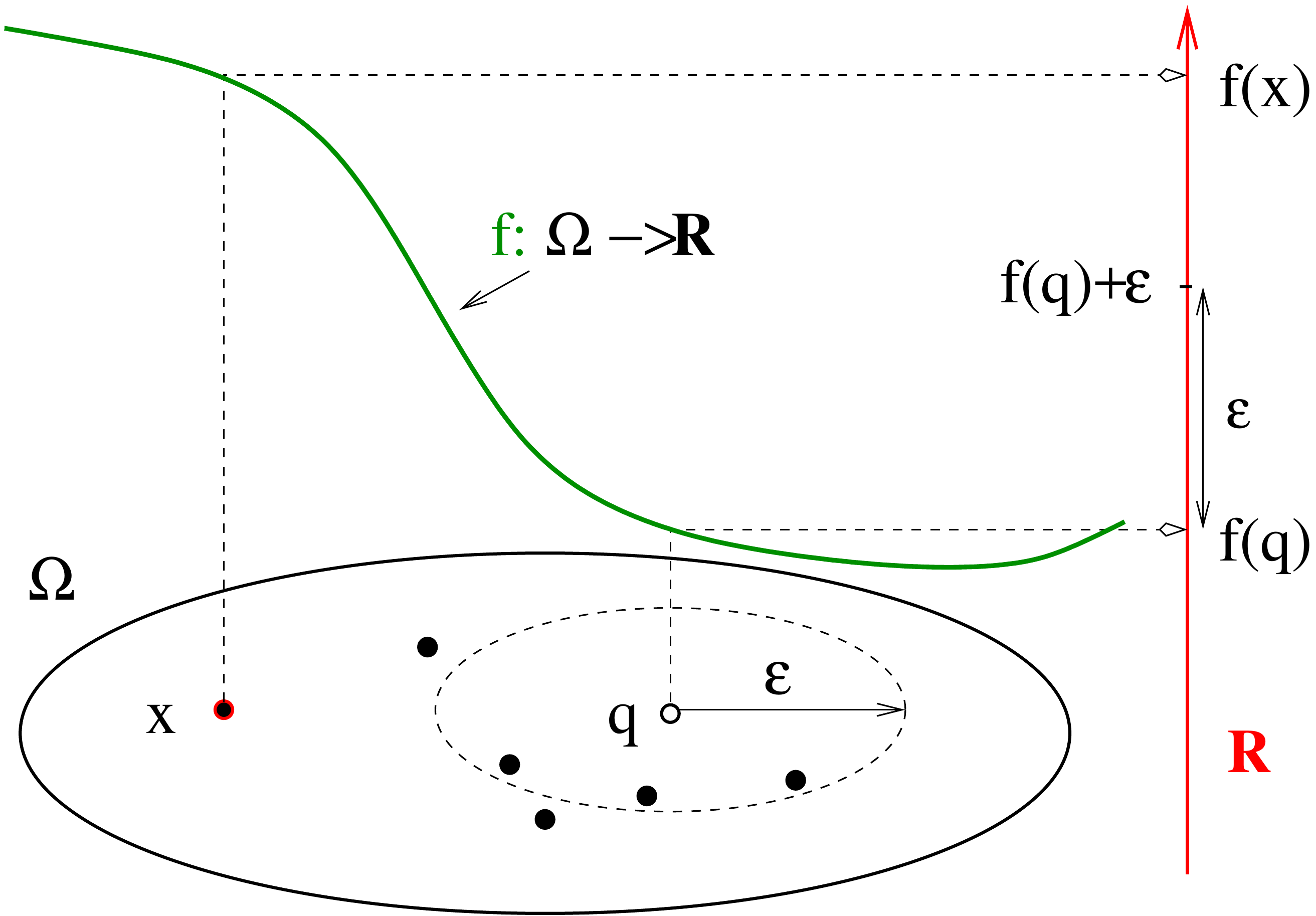}} 
\caption{O ponto $x$ pode ser descartado.
\label{fig:discarding}}
\end{center}
\end{figure}  
Dado uma consulta $(q,\ve)$, onde $q\in\Omega$ e $\ve>0$, o algoritmo escolha recursivamente uma s\'erie de \'\i ndices $i_n$, onde cada $i_{n+1}$ est\'a determinado pelos valores $f_{i_m}(q)$, $m\leq n$. As fun\c c\~oes $f_i$ servem para descartar os pontos $x_i$ que n\~ao podem responder \`a consulta. A saber, se $\abs{f_i(q)-f_i(x)}\geq\e$, ent\~ao, segundo a propriedade $1$-lipschitziana de $f_i$, temos $d(q,x)\geq \e$. Assim, o ponto $x$ \'e irrelevante e n\~ao precisa ser considerado (Figura \ref{fig:discarding}).

Ap\'os o c\'alculo terminar, o algoritmo retorna todos os pontos que n\~ao podem ser descartados, e verifica cada um deles contra a condi\c c\~ao $d(x,q)<\e$.

\begin{figure}[ht]
\centering
  \scalebox{0.2}{\includegraphics{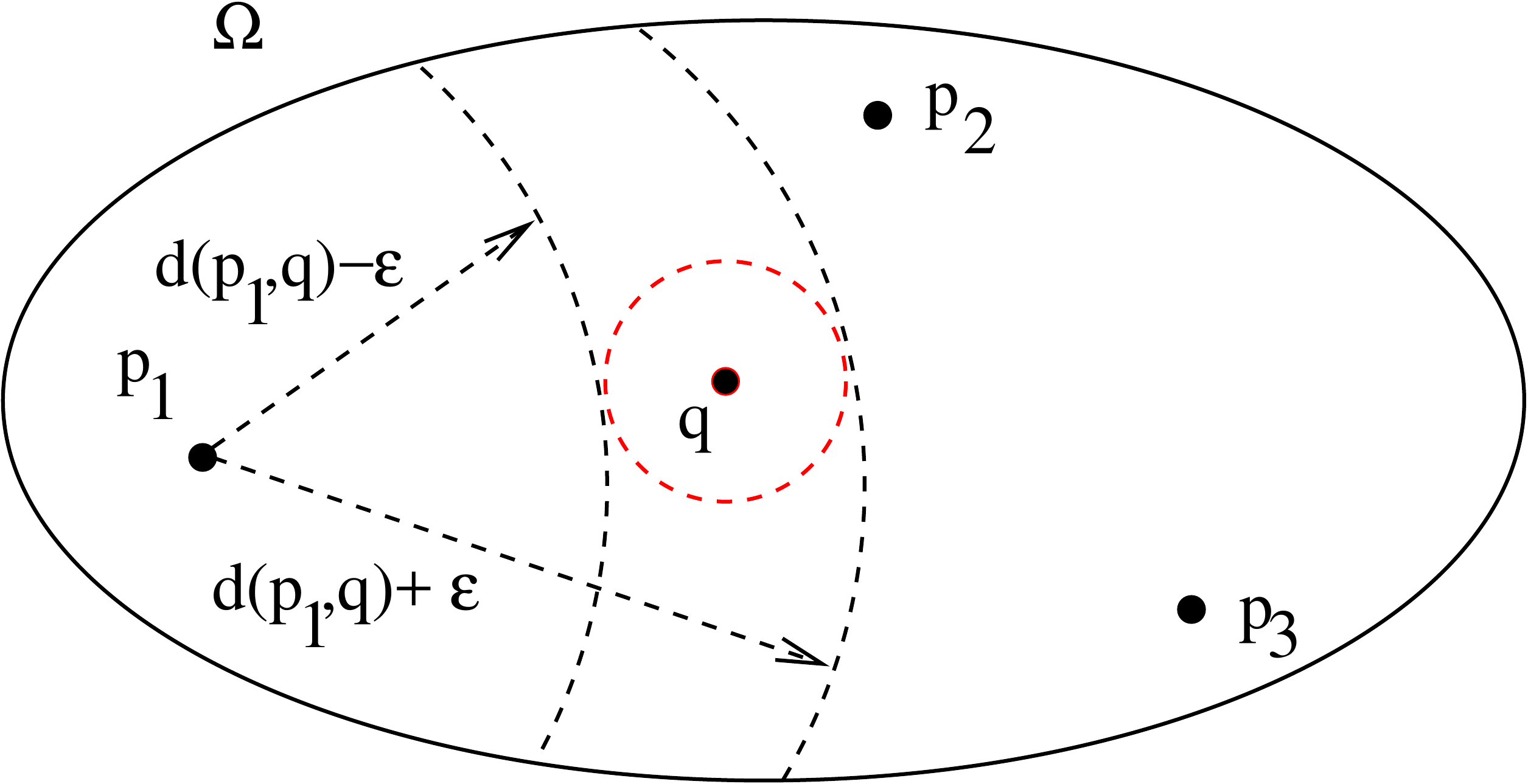}} 
\hskip 1cm
  \scalebox{0.2}{\includegraphics{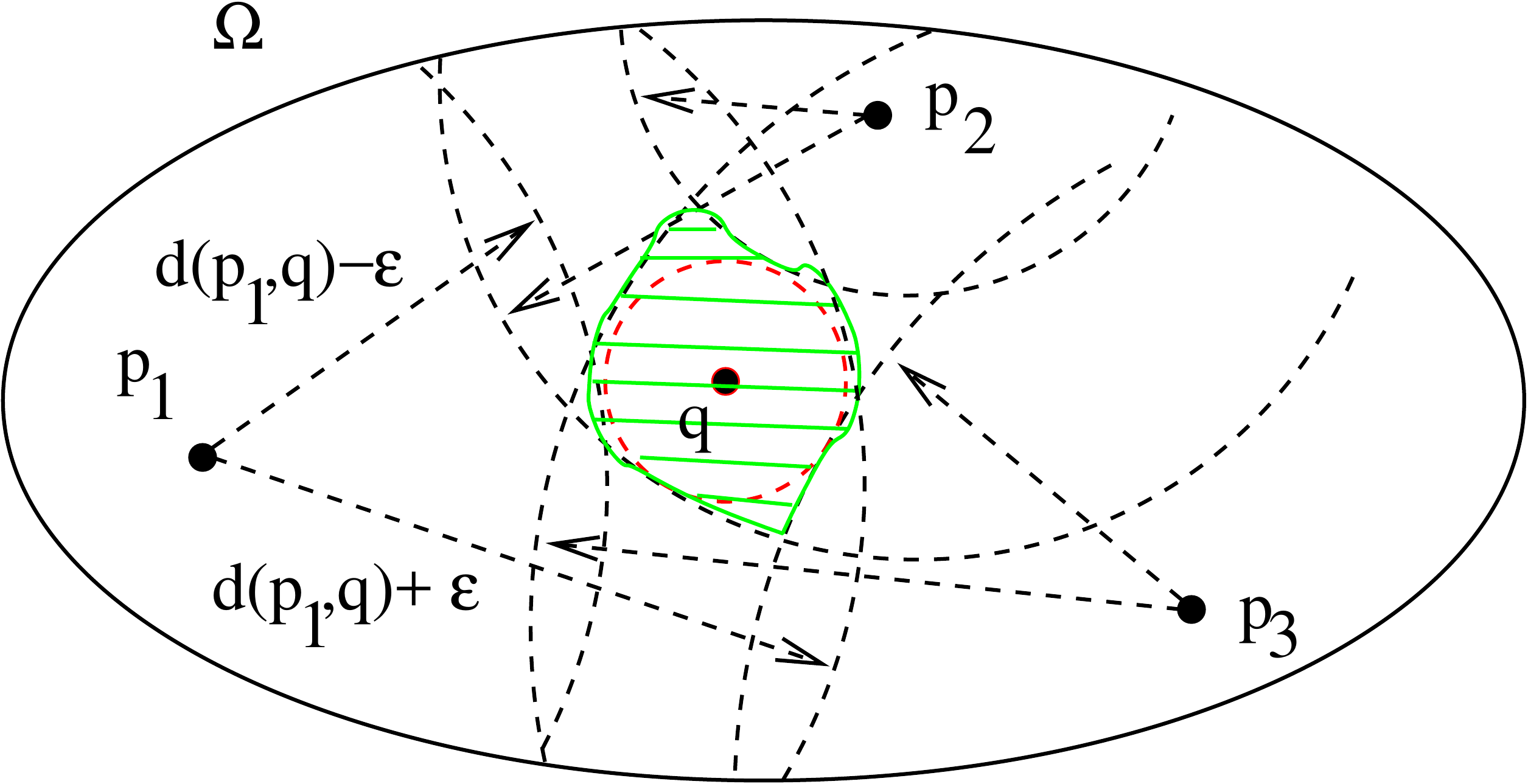}}
  \caption{Busca do vizinho mais pr\'oximo com piv\^os.}
  \label{fig:pivos}
\end{figure}

Figura \ref{fig:pivos} ilustra o funcionamento do esquema onde as fun\c c\~oes $f_i$ s\~ao dist\^ancias at\'e piv\^os. Apenas os pontos na regi\~ao verde ter\~ao que ser examinadas.

Uma observa\c c\~ao importante aqui \'e que as fun\c c\~oes $f_i,i\in I$ que formam um esquema de indexa\c c\~ao geralmente exigem uma certa quantidade de pr\'e-computa\c c\~ao e armazenamento, principalmente os ponteiros para dados a ser descartados. Por isso, o tamanho do esquema, $\sharp I$, n\~ao pode ser ilimitado.

\begin{figure}[ht]
\centering
\scalebox{0.55}{\includegraphics{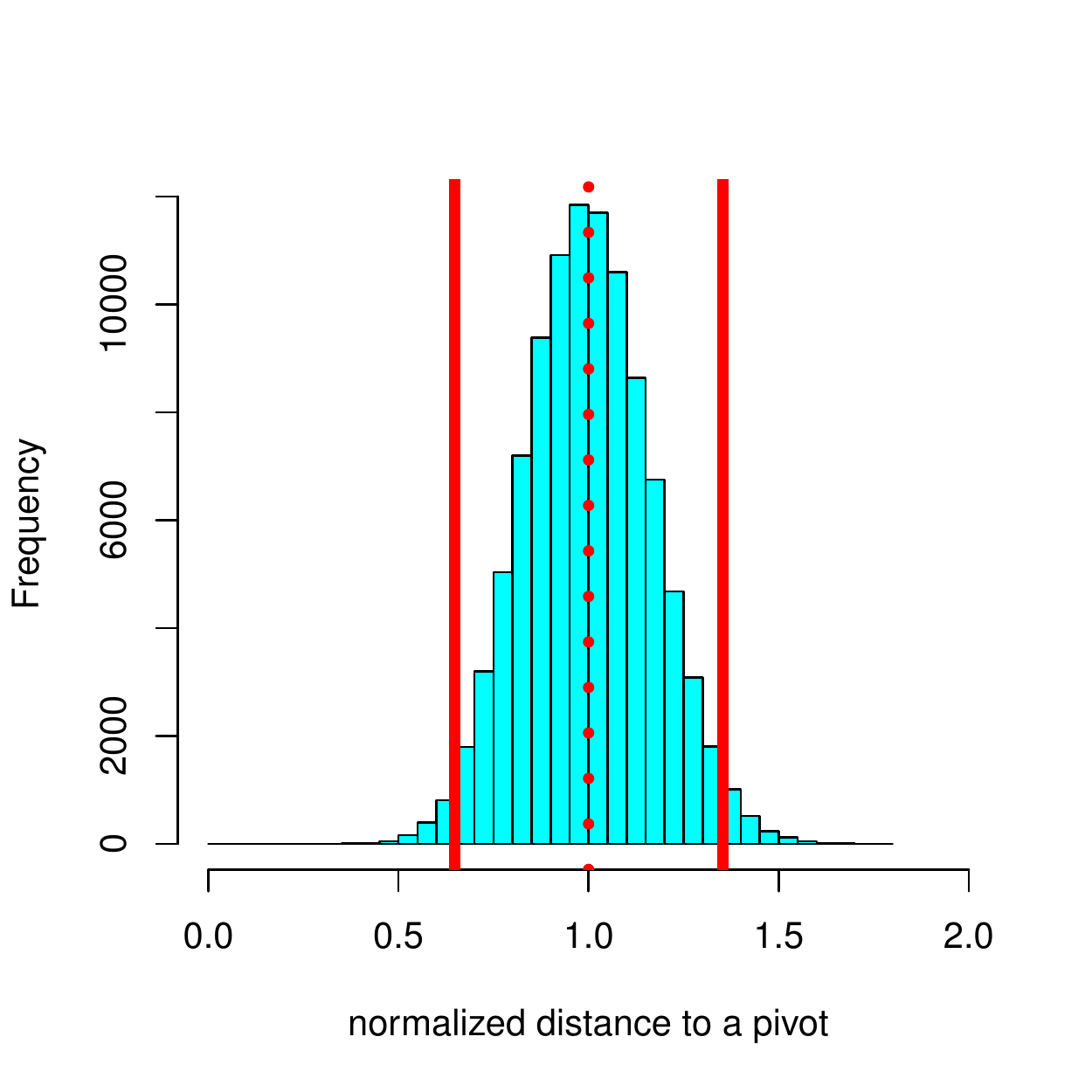}}
\caption{\label{fig:hist_14gauss}
  Histograma das dist\^ancias a um ponto escolhido aleatoriamente em um conjunto de dados $X$ com $n = 10^5$ points, tirados de uma distribui\c c\~ao gaussiana em $\R^{14}$. 
}
\end{figure}

Por causa de concentra\c c\~ao de medida, sobre as estruturas de alta dimens\~ao, as fun\c c\~oes lipschitzianas s\~ao concentradas em torno de seus valores m\'edios (ou medianas).  Esse efeito j\'a \'e pronunciado em dimens\~oes moderadas, tais como $d=14$ na Figura \ref{fig:hist_14gauss}. Aqui a fun\c c\~ao \'e a dist\^ancia de um piv\^o $p$ escolhido aleatoriamente. As linhas verticais marcam a dist\^ancia m\'edia normalizada $1\pm \e_{NN}$, onde $\e_{NN}$ \'e a dist\^ancia m\'edia do vizinho mais pr\'oximo. Supondo que o ponto de consulta $q$ esteja a uma dist\^ancia $\approx 1$ de $p$, apenas os pontos fora da regi\~ao marcada por barras verticais podem ser descartados.

\begin{figure}[ht]
 \centering
 \scalebox{0.5}{\includegraphics{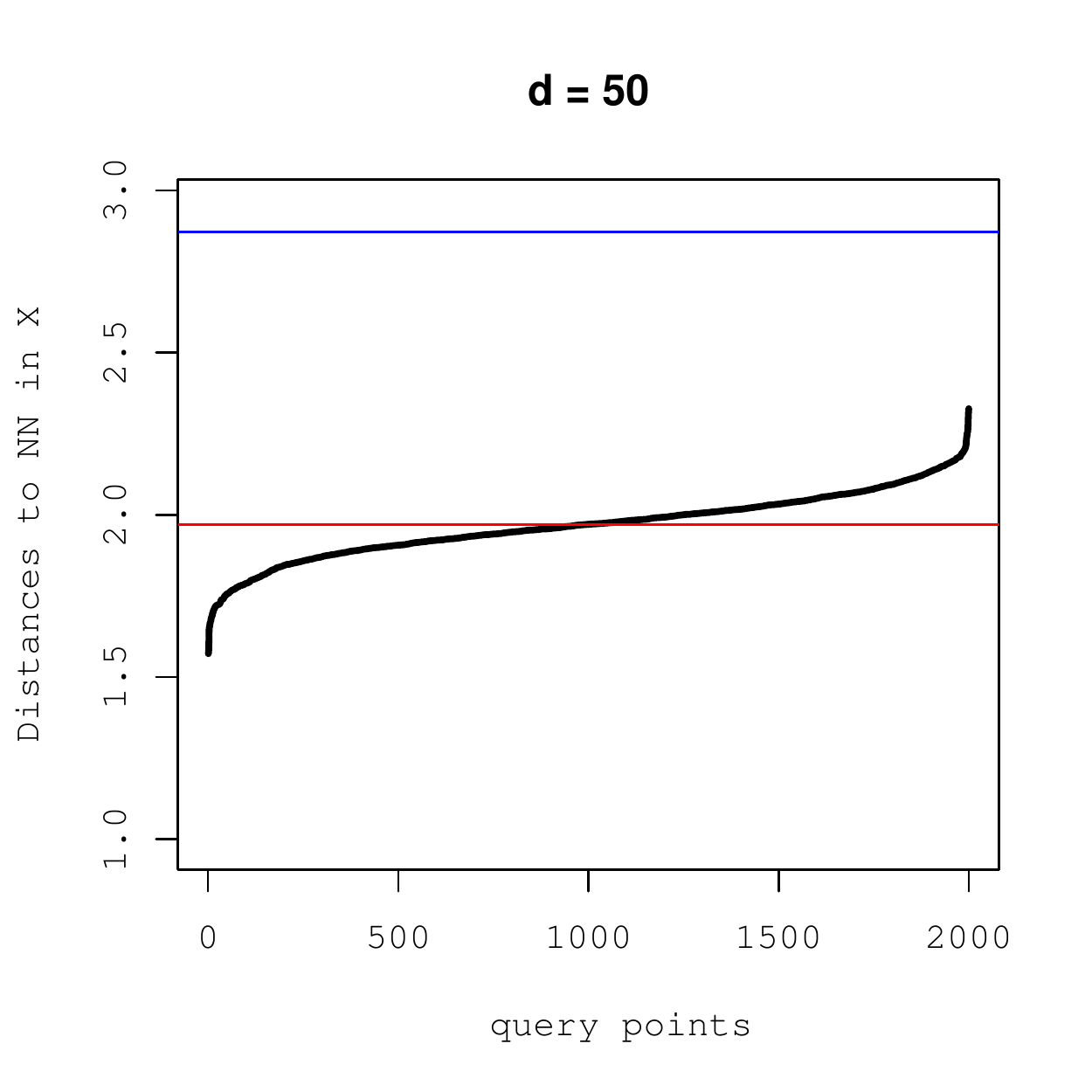}}
\caption{Dist\^ancias de $2,000$ pontos de busca aleat\'orios at\'e seus vizinhos mais pr\'oximos dentro de um conjunto de $10,000$ pontos aleat\'orios no cubo euclidiano $[0,1]^{50}$. A linha horizontal inferior marca o valor mediano $(r_{\mbox{\tiny NN}})_M =  1.9701$, a linha vertical, ${\mathbb{E}}d(x,y) =  2.872$.}
 \label{fig:nn_distances50}
 \end{figure}

Isto conduz em particular ao efeito conhecido na ci\^encia de dados como o ``paradoxo do espa\c co vazio'',
\index{paradoxo! do espa\c co vazio}
que afirma que a dist\^ancia m\'edia  $\E(\e_{NN})$ para o vizinho mais pr\'oximo aproxima-se \`a dist\^ancia m\'edia $\E(d)$ entre dois pontos de dados, quando a dimens\~ao $d$ cresce, desde que o n\'umero de dados, $n$, cresce apenas subexponencialmente em $d$. 
Figura \ref{fig:nn_distances50} \'e autoexplicativa.

\begin{figure}[ht]
\centering
\scalebox{0.55}{\includegraphics{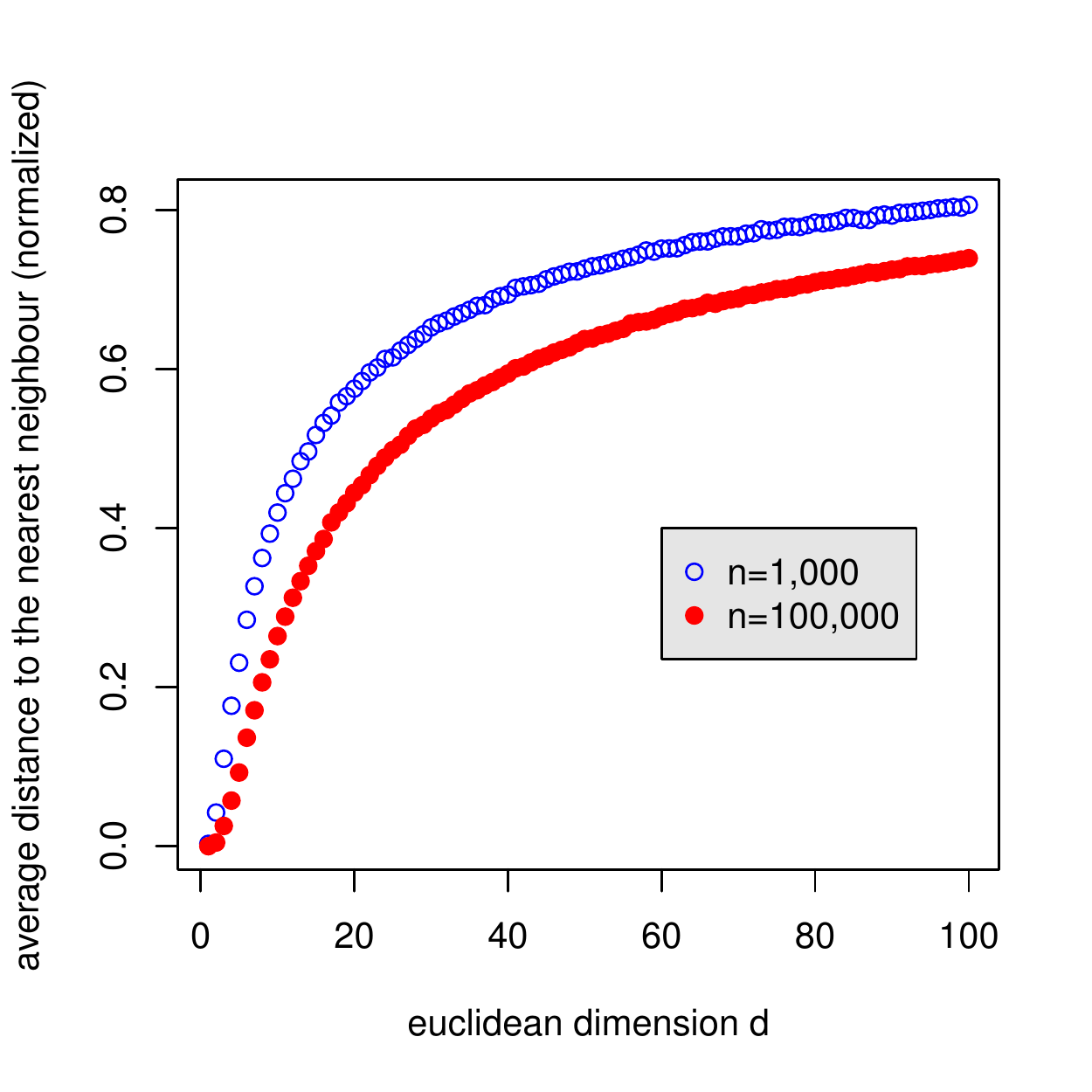}}
\caption{\label{fig:nn_dist}
A dist\^ancia m\'edia normalizada at\'e o vizinho mais pr\'oximo dentro da amostra com $n$ pontos tirada aleatoriamente da distribui\c c\~ao gaussiana em $\R^d$.}
\end{figure}

O fen\^omeno est\'a tamb\'em ilustrado na Figura \ref{fig:nn_dist} com o n\'umero de pontos constante ($n=10^3$ e $n=10^5$), e as dist\^ancias normalizadas de modo que o {\em tamanho carater\'\i stico} do espa\c co gaussiano $(\R^n,\gamma^n)$,
\begin{equation}
{\mathrm{CharSize}}\,(X) =\E_{\mu\otimes\mu}(d(x,y)),\end{equation}
seja igual a $1$.

As duas propriedades combinadas implicam que, como $d\to\infty$, cada vez menos pontos podem ser descartados durante a execu\c c\~ao de uma consulta de intervalo, e o desempenho de um esquema de indexa\c c\~ao degrada rapidamente.

\begin{figure}[ht]
\centering
  \scalebox{0.3}{\includegraphics{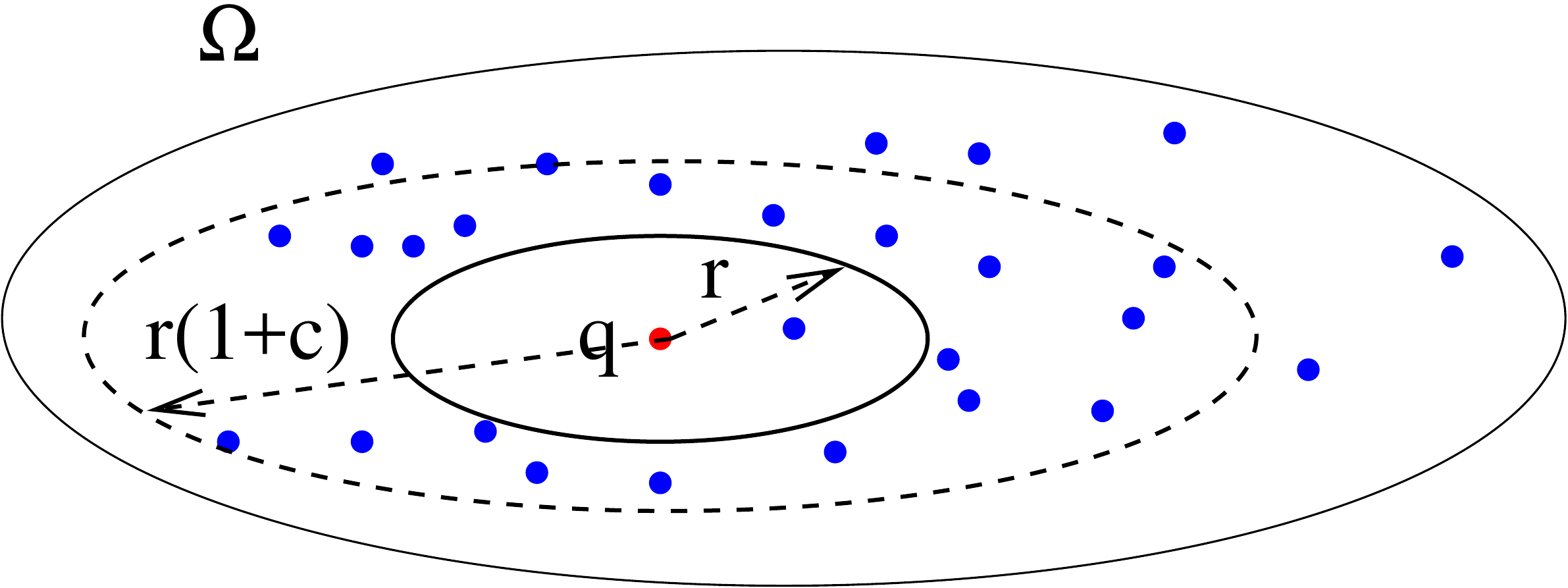}} 
  \caption{Instabilidade da busca do vizinho mais pr\'oximo.}
  \label{fig:unstable}
\end{figure}

No entanto, j\'a apenas o paradoxo de espa\c co vazio em si mesmo tem consequ\^encias imediatas para o classificador $k$-NN. Seja $x\in\Omega$ um ponto qualquer. Denotemos $r_{\mbox{\tiny NN}}(x)$ a dist\^ancia de $x$ para o seu vizinho mais pr\'oximo na amostra $\sigma$. Na consequ\^encia do paradoxo de espa\c co vazio, uma grande quantidade de pontos de $\sigma$ est\~ao quase a mesma dist\^ancia de $x$ que o seu vizinho mais pr\'oximo. Mais formalmente, seja $c>0$, e digamos, seguindo \cite{BGRS}, que a consulta de vizinho mais pr\'oximo de $x$ \'e {\em $c$-inst\'avel} 
\index{instabilidade da busca}
se a bola de raio $(1+c)r_{\mbox{\tiny NN}}(x)$ centrada em $x$ cont\'em pelo menos metade dos pontos de $\sigma$. (Figura \ref{fig:unstable}.)

Usando a concentra\c c\~ao da medida, n\~ao \'e dif\'\i cil de mostrar que, pelo $c>0$ fixo, no limite $d\to\infty$ a maioria das buscas ser\~ao $c$-inst\'aveis. 

\begin{figure}[ht]
\begin{center}
  \scalebox{0.35}[0.35]{\includegraphics{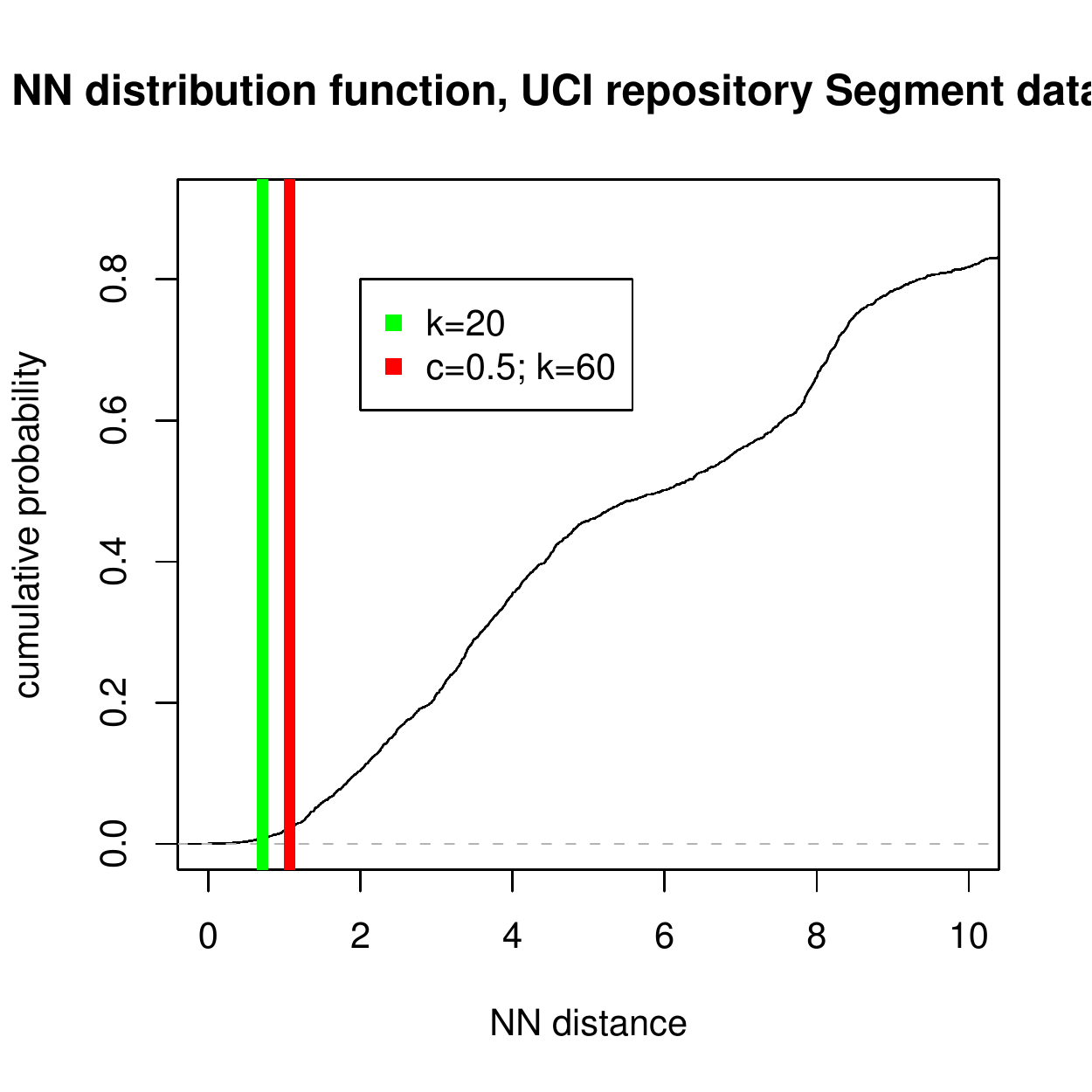}} 
  \hskip .2cm
  \scalebox{0.35}[0.35]{\includegraphics{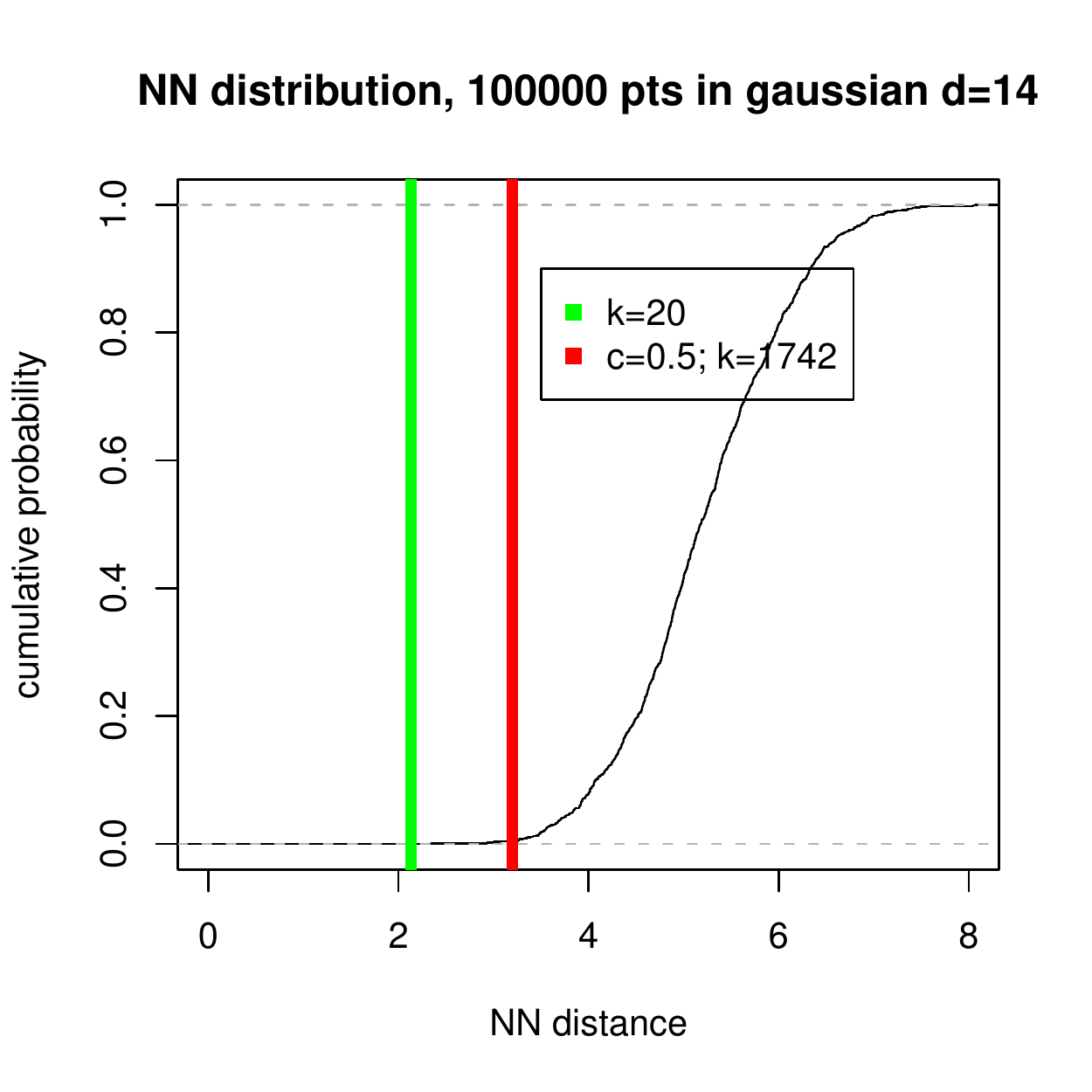}}
  \caption{A fra\c c\~ao m\'edia dos pontos de dados nas bolas de raio $(1+c)r_{\mbox{\tiny $k$-NN}}$.}
  \label{fig:instability}
\end{center}
\end{figure}

Nas dimens\~oes baixas, o fen\^omeno est\'a fraco (Figura \ref{fig:instability}, a esquerda, o conjunto de dados {\em Segment} da {\em UCI data repository}\footnote{\href{http://archive.ics.uci.edu/ml/index.php}
{http://archive.ics.uci.edu/ml/}}), mas nas dimens\~oes m\'edias, \'e j\'a pronunciado (Figura \ref{fig:instability}, a direita, o subconjunto aleat\'orio da distribui\c c\~ao gaussiana em $\R^{14}$).  Aqui, $k = 20$ e $c = 0.5$.
A linha esquerda vertical corresponde ao valor m\'edio de raio da bola que cont\'em $k$ vizinhos mais pr\'oximos, $r_{\mbox{\tiny $k$-NN}} $, e a segunda linha corresponde a $(1+c) r_ {\mbox {\tiny $k$-NN}}$. Para o conjunto {\em Segment,} a segunda bola cont\'em em m\'edia $60$ pontos. Para o gaussiano, o valor correspondente j\'a \'e de $1,742$ pontos.

O fen\^omeno da instabilidade significa uma perda da import\^ancia do fato de ser o vizinho mais pr\'oximo. Por exemplo, no caso de um erro quase inevit\'avel de recupera\c c\~ao do vizinho mais pr\'oximo exato, o r\'otulo do vizinho substitu\'\i do ser\'a mais ou menos aleat\'orio. Al\'em disso, a converg\^encia de raio $r_{\mbox{\tiny $k$-NN}}$ para zero, como no lema de Cover--Hart, ser\'a muito devagar.
O desempenho do classificador $k$-NN (e de qualquer outro algoritmo baseado sobre vizinhos mais pr\'oximos) degrada especialmente em dimens\~oes altas, mas tamb\'em em dimens\~oes m\'edias, mesmo se n\~ao t\~ao notoriamente. 

Apoiemos alguns dos argumentos emp\'\i ricos acima com resultados concretos.

\subsection{Paradoxo de espa\c co vazio e instabilidade da busca}
Relembramos que a nota\c c\~ao assint\'otica $f(n)=O(g(n))$ significa que existe uma constante $C>0$ e um n\'umero natural $C>0$ tais que $\abs{f(n)}\leq C\abs{g(n)}$ para todos $n\geq N$. Neste caso tamb\'em escrevemos $g(n)=\Omega(f(n))$. 
O fato $f=O(g)$ e $g=O(f)$ \'e expresso pela nota\c c\~ao $f=\Theta(g)$.
A nota\c c\~ao $f(n)=\omega(g(n))$ significa que $f(n)=\Omega(g(n))$ e $f(n)\neq O(g(n))$, ou seja, $f$ cresce estritamente mais r\'apido que $g$.
\index{On@$O(n)$}
\index{Omegan@$\Omega(n)$}
\index{omegan@$\omega(n)$}
\index{Thetan@$\Theta(n)$}

A fim de executar a an\'alise assint\'otica, vamos trabalhar com uma {\em sequ\^encia infinita} de dom\'\i nios, $\Omega_d$, bem como de conjuntos de dados $X_d\subseteq \Omega_d$, $d=1,2,\ldots$. Isso equivale a tirar um ponto \'unico (caminho amostral), $\bar x$, no produto infinito
\[\Omega_1^{n_1}\times\Omega_2^{d_2}\times\ldots\times\Omega_d^{n_d}\times\ldots,\]
em rela\c c\~ao ao produto infinito de medidas de probabilidade:
\[\bar X\sim \mu_1^{\otimes n_1}\otimes\mu_2^{\otimes d_2}\otimes\ldots\otimes\mu_d^{\otimes n_d}\otimes\ldots.\]
A {\em confian\c ca} neste contexto significa a medida de produto acima. 
Especificamente, a afirma\c c\~ao $Q(d,\bar x)$, parametrizado pela dimens\~ao $d$ e aceitando como uma vari\'avel o caminho amostral, {\em ocorre assintoticamente com alta confian\c ca} se para todo $\delta>0$ existe $D$ tal que
\[P[Q(d,\bar X)\mbox{ \'e verdadeiro }]>1-\delta\]
quando $d\geq D$. 

Ao mesmo tempo, a fim de manter a nota\c c\~ao simples, vamos suprimir o \'\i ndice $d$ e falar apenas de um dom\'\i nio \'unico $\Omega$ e um conjunto de dados $X\subseteq\Omega$. 

A fim de provar formalmente um resultado geral, aceitemos as seguintes hip\'oteses.

\subsubsection{Dom\'\i nio como um espa\c co m\'etrico com medida} 
O dom\'\i nio m\'etrico $(\Omega,d)$ \'e munido de uma medida de probabilidade $\mu$, e dados est\~ao tirados de $\Omega$ de maneira i.i.d., seguinte a lei $\mu$.

\subsubsection{Normaliza\c c\~ao da dist\^ancia} 
A dist\^ancia $d$ sobre o dom\'\i nio \'e normalizada de modo que o {\em tamanho carater\'\i stico} de $\Omega$ \'e constante:
\[{\mathrm{CharSize}}(\Omega) = \E_{\mu\otimes\mu}(d)=\Theta(1).\]
\index{tamanho carater\'\i stico}

\subsubsection{Dimens\~ao intr\'\i nseca crescente} O dom\'\i nio $\Omega=\Omega_d$
tem ``dimens\~ao intr\'\i nseca $\Omega(d)$'' no sentido que a fun\c c\~ao de concentra\c c\~ao de $(\Omega,d,\mu)$ admite uma cota superior gaussiana
\[\alpha_{\Omega}(\ve) = \exp(-\Omega(\ve^2d)).\]

\subsubsection{Tamanho de conjunto de dados} O n\'umero de pontos, $n=\sharp X_d$, cresce mais r\'apido do que alguma fun\c c\~ao polinomial em $d$, \'e mais devagar do que alguma fun\c c\~ao exponential em $d$:
\begin{eqnarray}
n=d^{\omega(1)},~~
d=\omega(\log n).
\end{eqnarray}

\begin{observacao}
A \'ultima hip\'otese \'e padr\~ao na an\'alise assint\'otica de esquemas de indexa\c c\~ao para pesquisa de semelhan\c ca, veja \citep*{indyk}. Um exemplo da tal taxa de crescimento \'e $n=2^{\sqrt d}$.
\end{observacao}

\begin{teorema} 
\label{th:empty}
Sob as hip\'oteses acima, para cada $\ve>0$, assintoticamente com alta confian\c ca todos pontos $q\in\Omega$ exceto um conjunto de medida demasiado pequena $\exp(-\Omega(\ve^2 d))$, satisfazem
\[\left\vert r^X_{\mbox{\tiny NN}}(q)-{\mathrm{CharSize}}(\Omega)\right\vert<\ve.\]
\end{teorema}

\begin{corolario}
Sob as hip\'oteses acima, para cada $\ve>0$ e $c>0$, assintoticamente com alta confian\c ca, para todos pontos $q\in\Omega$ exceto um conjunto de medida demasiado pequena $\exp(-\Omega(\ve^2 d))$, a busca de intervalo de raio $\ve$ em ponto $q$ \'e $c$-inst\'avel.
\label{c:c-instavel}
\end{corolario}

Como um outro subproduto da t\'ecnica, obtemos:

\begin{proposicao}
\label{p:simplex}
Sob as mesmas hip\'oteses, para todo $\ve>0$ as dist\^ancias dois a dois entre pontos de $X$ s\~ao todos no intervalo
${\mathrm{CharSize}}\,(\Omega)\pm \ve$, assintoticamente com alta confian\c ca.
\end{proposicao}

Sem perda de generalidade, normalizemos o tamanho carater\'\i stico de $\Omega$ para $1$.
Para cada $q\in\Omega$, a fun\c c\~ao dist\^ancia, $d(q,-)$, \'e $1$-Lipschitz cont\'\i nua, logo concentrada em torno do seu valor mediano, $R(q)$. Para evitar incerteza, escolhamos o {\em m\'\i nimo} valor mediano, que sempre existe (exerc\'\i cio).

\begin{exercicio}
Mostrar que a fun\c c\~ao $R\colon\Omega\to\R$, $q\mapsto R(q)$ \'e Lipschitz cont\'\i nua tamb\'em, de constante $1$.
\end{exercicio}

Por conseguinte, $R(q)$ concentra-se em torno do seu valor mediano, que denotemos por $R_M$. 

\begin{exercicio}
Deduza que para todos pontos $q\in\Omega$ fora de um conjunto de medida demasiado pequena, $\exp(-\Omega(\ve^2 d))$, a casca esf\'erica,
\[\{x\in\Omega\colon R_M-\ve \leq d(q,x)\leq R_M+\ve\},\]
tem medida $1-\exp(-\Omega(\ve^2 d))$. 
\end{exercicio}

A probabilidade para pelo menos um ponto de $X$ n\~ao pertencer \`a casca esf\'erica como acima \'e limitada por
\[n \exp(-\Omega(\ve^2 d)) = \exp(-\Omega(\ve^2 d))
\]
pois $n$ cresce subexponencialmente em $d$. Para concluir a prova do teorema \ref{th:empty}, precisamos do exerc\'\i cio seguinte.

\begin{exercicio}
Mostre que, com a confian\c ca $1-\exp(-\Omega(\ve^2 d))$,
\[1-R_M <\ve.\]
\end{exercicio}

Consideremos o conjunto $A$ de todos pares $(q,\bar x)$ que satisfazem a conclus\~ao do teorema \ref{th:empty}. N\'os estabelecemos que para todos $q$ fora de um conjunto de medida $\exp(-\Omega(\ve^2 d))$, a ``faixa vertical'' sobre $q$, 
\[\{\bar x\colon (q,\bar x)\in A\},\]
tem medida $1-\exp(-\Omega(\ve^2 d))$. O teorema de Fubini \ref{t:fubini} permite concluir que a medida de produto de $A$ \'e $1-\exp(-\Omega(\ve^2 d))$. Por conseguinte, para todos caminhos amostrais $\bar x$ fora de um conjunto de medida $\exp(-\Omega(\ve^2 d))$, a ``faixa horizontal'', ou seja, o conjunto de $q\in\Omega$ tais que $(q,\bar x)\in A$, tem medida $1-\exp(-\Omega(\ve^2 d))$. Isso \'e exatamente a afirma\c c\~ao do teorema \ref{th:empty}.

Segue-se que, qualquer que seja $c>0$, para todos os pontos $q$ fora de um conjunto da medida demasiado pequena, assintoticamente, a medida da bola de raio $(1+c)r^X_{\mbox{\tiny NN}}(q)$ com centro em $q$ converge para $1$, como $1-\exp(-\Omega(c^2 d))$. Isto estabelece a instabilidade de busca e o corol\'ario \ref{c:c-instavel}.

Para mostrar proposi\c c\~ao \ref{p:simplex}, fixemos $\e>0$. Para dois elementos aleat\'orios de $\Omega$, $X_1$ e $X_2$, temos com confian\c ca $1-n\exp(-O(d)\e^2)$ a propriedade $\abs{d(X_1,X_2)-R(X_1)}<\e/2$ (condicionando sobre $X_1$).
Dado uma amostra aleat\'oria $X$ com $n$ pontos, conclu\'\i mos que,
com confian\c ca $1-n^2\exp(-O(d)\e^2)$, $\abs{d(x,y)-R(x)}<\e/2$ para todos $x,y\in X$. Ao mesmo tempo, com a confian\c ca $1-n\exp(-O(d)\e^2)$, 
temos $\abs{R(x)-1}<\e/2$ para todos $x\in X$. Como $n$ \'e subexponencial em $d$, o resultado segue-se. 

\subsection{Fun\c c\~oes de dimens\~ao intr\'\i nseca}
Formular precisamente o que \'e a ``dimens\~ao in\-tr\'in\-seca de dados'' resta um desafio te\'orico na ci\^encia de dados, tendo um significado pr\'atico bastante \'obvio. Eis alguns objetivos que uma tal no\c c\~ao deveria cumprir.

\begin{enumerate}
\item Queremos que um valor elevado de dimens\~ao intr\'\i nseca seja indicativo da presen\c ca da maldi\c c\~ao da dimensionalidade.

\item O conceito n\~ao deve fazer qualquer distin\c c\~ao entre objetos cont\'\i nuos e discretos, e a dimens\~ao intr\'\i nseca de uma amostra discreta aleat\'oria deve estar pr\'oxima da do dom\'\i nio subjacente munido da lei de distribui\c c\~ao. 

\item A dimens\~ao intr\'\i nseca deve concordar com a nossa intui\c c\~ao geom\'etrica e devolver valores usuais para objetos familiares, tais como esferas euclidianas ou cubos de Hamming.

\item N\'os queremos o conceito ser insens\'\i vel ao ru\'\i do aleat\'orio da amplitude moderada.

\item Por fim, para ser \'util, a dimens\~ao intr\'\i nseca deve ser computacionalmente vi\'avel. 
\end{enumerate}

Vamos discutir uma abordagem, concentrando sobre os objetivos 1--3, que v\~ao ser formalizados e transformados em tr\^es axiomas. 

Para formalizar o objetivo 2, precisamos de uma no\c c\~ao de proximidade entre espa\c cos m\'etricos com medida, que concorde bem com o fen\^omeno de concentra\c c\~ao.
Sejam $X=(X,d_X,\mu_X)$ e $Y=(Y,d_Y,\mu_Y)$ dois espa\c cos m\'etricos (separ\'aveis e completos) munidos de medidas borelianas de probabilidade. A ideia da dist\^ancia de Gromov \citep*{Gr} \'e que $X$ e $Y$ ser\~ao pr\'oximos se toda fun\c c\~ao $1$-Lipschitz cont\'\i nua sobre $X$ pode ser emparelhada com uma tal fun\c c\~ao sobre $Y$, e vice-versa. Relembramos (teorema \ref{t:parametrizacao}) que todo espa\c co probabil\'\i stico padr\~ao, $\Omega$, pode ser {\em parametrizado} pelo intervalo unit\'ario, $[0,1]$: existe uma aplica\c c\~ao boreliana $\phi\colon [0,1]\to\Omega$ tal que a medida $\mu$ \'e a imagem direta da medida de Lebesgue (uniforme), $\lambda$, sobre o intervalo:
\[\mu=\phi_{\ast}(\lambda),\]
ou seja, para todo subconjunto boreliano $A\subseteq X$,
\[\mu_X(A)=\lambda(\phi^{-1}(A)).\]
Introduzimos a dist\^ancia ${\mathrm{me}}_1$ sobre o espa\c co de fun\c c\~oes borelianas sobre $[0,1]$ como segue:
\[{\mathrm{me}}_1(f,g)=\inf\left\{\e>0\colon \lambda\{t\in [0,1]\colon \abs{f(t)-g(t)}> \e\}<\e\right\}.\]

\begin{exercicio}
Mostrar que ${\mathrm{me}}_1$ \'e uma pseudom\'etrica, determinando a converg\^encia em medida. (Releia a discuss\~ao no in\'\i cio da subse\c c\~ao \ref{s:cobertura}).
\end{exercicio}

Agora definamos a dist\^ancia de Gromov,
\index{dist\^ancia! de Gromov}
$d_{conc}(X,Y)$, como o \'\i nfimo de todos $\e>0$ admitindo algumas parametriza\c c\~oes $\phi_X$ e $\phi_Y$ de $X$ e de $Y$ respetivamente, tendo a propriedade seguinte. Para cada $f\in {\mathrm{Lip}}_1(X)$ existe $g\in {\mathrm{Lip}}_1(Y)$ tal que
\begin{equation}
\label{eq:me1}
{\mathrm{me}}_1(f\circ \phi_X,g\circ \phi_Y)<\e,
\end{equation}
e vice versa: para cada $g\in {\mathrm{Lip}}_1(Y)$ existe $f\in {\mathrm{Lip}}_1(X)$ satisfazendo eq. (\ref{eq:me1}).

\begin{figure}[htp]
\centering
\scalebox{1}{\includegraphics{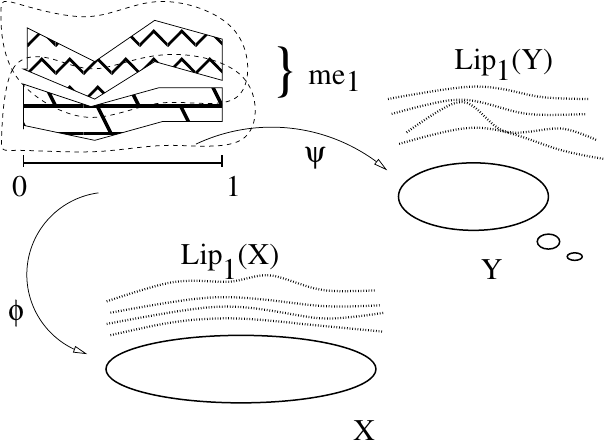}}
\caption{A no\c c\~ao da dist\^ancia de Gromov}
\label{fig:gromdist}
\end{figure}

\begin{exercicio}
Verificar que a dist\^ancia $d_{conc}$ \'e uma pseudom\'etrica.
\end{exercicio}

\begin{exercicio}
Verificar que se $d_{conc}(X,Y)=0$, ent\~ao os espa\c cos m\'etricos com medida $X=(X,d_X,\mu_X)$ e $Y=(Y,d_Y,\mu_Y)$ s\~ao isomorfos, a saber: existe uma isometria $f\colon\supp\mu_X\to\supp\mu_Y$ tal que $f_{\ast}(\mu_X)=\mu_Y$. (Este exerc\'\i cio \'e dif\'\i cil, veja Th. 5.16 em \citep*{shioya}).
\end{exercicio}

\begin{proposicao}
Seja $X$ um espa\c co m\'etrico com medida. Ent\~ao,
\[\left(d_{conc}(X,\{\ast\})\leq\e/2\right)\Rightarrow \left(\alpha(\e)\leq\e/2\right)\Rightarrow \left(d_{conc}(X,\{\ast\})\leq \e\right).\]
\end{proposicao}

\begin{proof} Suponha que $d_{conc}(X,\{\ast\})\leq \e/2<1/2$ e seja $A\subseteq X$ qualquer, com $\mu(A)\geq 1/2$. A fun\c c\~ao dist\^ancia $d_A(x)=d(A,x)=\inf\{d(a,x)\colon a\in A\}$ \'e $1$-Lipschitz cont\'\i nua, logo ela difere de uma fun\c c\~ao constante, $c$, pelo menos de $\e/2$ sobre um conjunto de medida $>1-\e/2$. \'E claro que $c\leq\e/2$, logo $d_A$ apenas pode tomar valor $>\e$ sobre um conjunto da medida $\leq\e/2$, significando $\alpha(\e)\leq\e/2$. 
Inversamente, se $d_{conc}(X,\{\ast\})\geq \e$, existe uma fun\c c\~ao $1$-Lipschitz cont\'\i nua, $f$, sobre $X$, que difere do seu valor mediano $M=M_f$ pelo menos por $\e$ sobre um conjunto de medida $\geq \e$. Por conseguinte, existem dois conjuntos, $A$ e $B$, tais que $\mu(A)\geq 1/2$, $\mu(B)\geq \e/2$, e para todos $a\in A$, $b\in B$ temos $\abs{f(a)-f(b)}\geq\e$, ou seja, $\alpha(\e)\geq \e/2$.
\end{proof}

\begin{figure}[ht]
\centering
\scalebox{.6}{\includegraphics{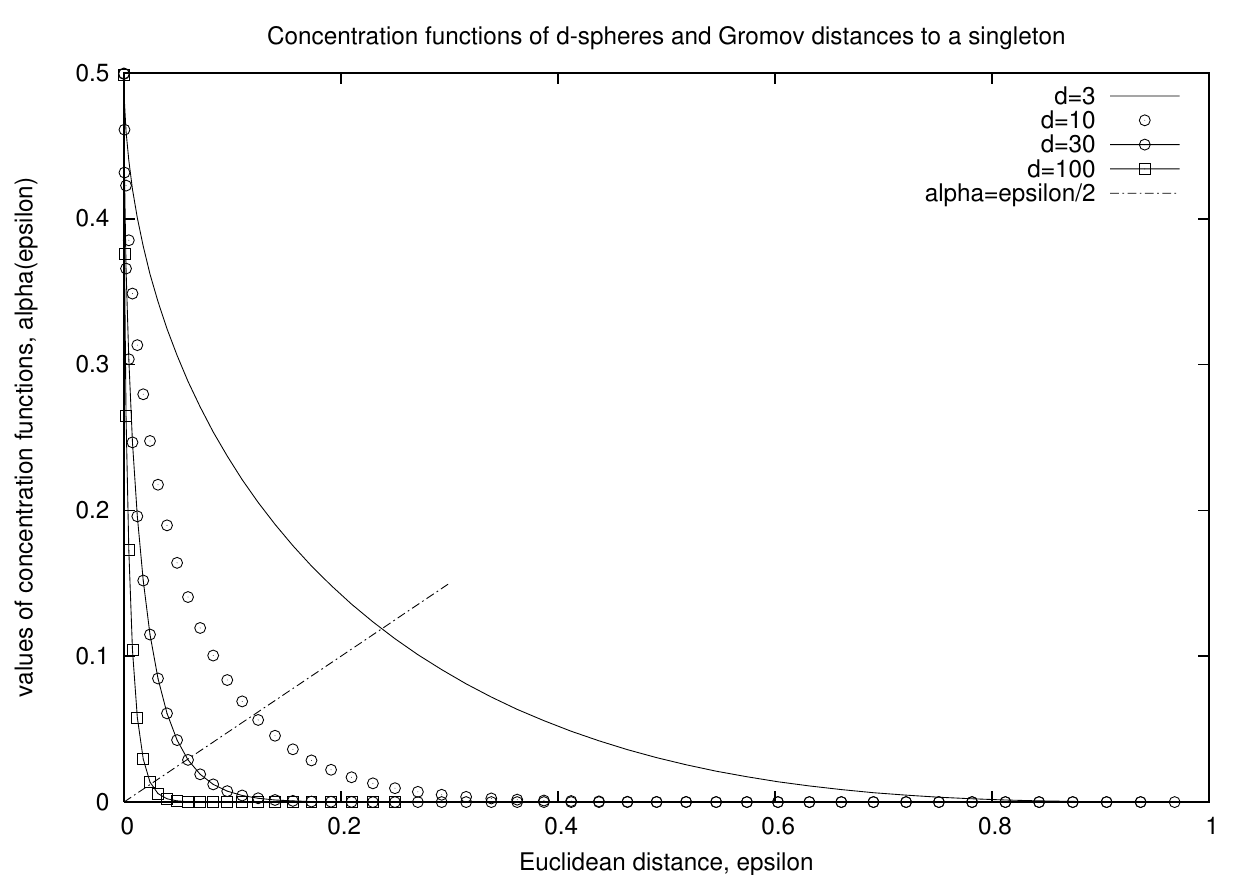}}
\caption{Fun\c c\~oes de concentra\c c\~ao das esferas $\s^n$, $n=3,10,30,100$, e a reta $\alpha=\e/2$.}
\label{fig:esferas+reta}
\end{figure}

\begin{exercicio}
Mostre que, para as esferas euclidianas, a dist\^ancia de Gromov at\'e um conjunto unit\'ario \'e exatamente a solu\c c\~ao da equa\c c\~ao $\alpha({\s^n},\e)=\e/2$, Fig. \ref{fig:esferas+reta}.
\end{exercicio}

\begin{exercicio} 
Deduza que uma fam\'\i lia $(X_n)$ de espa\c cos m\'etricos com medida \'e uma fam\'\i lia de L\'evy se e somente se ela converge para o espa\c co unit\'ario $\{\ast\}$ em rela\c c\~ao \`a m\'etrica de Gromov: $d_{conc}(X_n,\{\ast\})\to 0$ quando $n\to\infty$.
\end{exercicio}

\begin{observacao}
A explica\c c\~ao intuitiva \'e a seguinte:
quando uma sequ\^encia de conjuntos de dados mais se aproximar do conjunto unit\'ario na dist\^ancia de Gromov, mais se assemelhar a um ``buraco negro'' do ponto de vista da ci\^encia de dados, porque as ``carater\'\i sticas'' (fun\c c\~oes Lipschitz cont\'\i nuas) concentram-se e tornam-se cada vez menos discriminantes.
\end{observacao}

Agora seja $\partial$ uma fun\c c\~ao assinando a cada espa\c co m\'etrico com medida $(X,d,\mu)$ ou um n\'umero real n\~ao negativo ou o s\'\i mbolo $+\infty$. Digamos que $\partial$ \'e uma {\em fun\c c\~ao de dimens\~ao intr\'\i nseca} se ela satisfaz os tr\^es axiomas.
\index{fun\c c\~ao! de dimens\~ao intr\'\i nseca}

\subsubsection{\label{ax:conc}Axioma de concentra\c c\~ao} Dada uma fam\'\i lia $(X_n)$ de espa\c cos m\'etricos com medida,
$\partial(X_n)\uparrow\infty$ se e somente se $(X_n)$ \'e uma fam\'\i lia de L\'evy.

Isto axioma formaliza o requerimento que a dimens\~ao intr\'\i nseca \'e alta se e somente se o conjunto de dados sofre a maldi\c c\~ao de dimensionalidade.

\subsubsection{Axioma de depend\^encia suave de dados}  Se $d_{conc}(X_n,X)\to 0$, ent\~ao $\partial(X_n)\to \partial(X)$.

Este axioma garante, por exemplo, que a dimens\~ao intr\'\i nseca de uma amostra aleat\'oria \'e pr\'oxima \`a do dom\'\i nio subjacente.

\subsubsection{Axioma de normaliza\c c\~ao} $\partial(\s^n)=\Theta(n)$.

Relembremos que $f(n)=\Theta(g(n))$ se existem constantes $0<c<C$ e $N$ com $c\abs{f(n)}\leq \abs{g(n)}\leq C\abs{f(n)}$ para todos $n\geq N$. Neste caso, dizemos que $f$ e $g$ assintoticamente t\^em a mesma ordem de grandeza.

Este axioma serve para calibrar propriamente os valores da dimens\~ao intr\'\i nseca.

\begin{observacao}
Ao inv\'es das esferas, podem ser usadas os hipercubos normalizados, $\I^n$, os cubos de Hamming normalizados, os espa\c cos euclidianos munidas das medidas gaussianas, etc. --- pode-se mostrar que as defini\c c\~oes resultantes sejam equivalentes.
\end{observacao}

Os axiomas levam imediatamente a uma conclus\~ao paradoxal. Como, por exemplo, as esferas euclidianas $\s^n$ de raio um munidas da medida de probabilidade invariante pelas rota\c c\~oes formam uma fam\'\i lia de L\'evy normal (Ap\^endice \ref{a:esfera}), eles convergem para o espa\c co trivial $\{\ast\}$ na dist\^ancia de Gromov, e os axiomas 1 e 2 (ou 2 e 3) implicam que
\[\partial(\{\ast\})=+\infty.\]

O converso \'e tamb\'em verdadeiro. 

\begin{teorema} Seja $\partial$ uma fun\c c\~ao de dimens\~ao intr\'\i nseca. Ent\~ao $\partial(X)=+\infty$ se e somente se $X$ \'e isomorfo a um espa\c co trivial, $X\simeq\{\ast\}$, ou seja, a medida $\mu_X$ \'e uma medida de Dirac, suportada em um ponto.
\end{teorema}

\begin{proof}
Se $\partial(X)=+\infty$, ent\~ao a sequ\^encia constante $X_n=X$ \'e uma fam\'\i lia de L\'evy. Por conseguinte, $\alpha_X(\ve)=0$ para todos $\ve>0$, o que \'e apenas poss\'\i vel quando $\mu$ \'e uma medida de Dirac (exerc\'\i cio).
\end{proof}

Assim, o \'unico objeto de dimens\~ao infinita em uma teoria \'e um espa\c co trivial, um conjunto unit\'ario! 
Este paradoxo parece ser inevit\'avel se algu\'em quer uma no\c c\~ao de dimens\~ao intr\'\i nseca capaz de detectar a maldi\c c\~ao da dimensionalidade. No entanto, isto n\~ao parece levar a quaisquer problemas ou inconvenientes.

Talvez ainda mais surpreendente \'e o fato de que uma fun\c c\~ao de dimens\~ao satisfazendo os requisitos acima realmente existe.

\begin{exemplo}
Para um espa\c co m\'etrico com medida $(X,d,\mu)$, definamos
\begin{equation}
\label{eq:concdim}
\partial_{conc}(X)=\frac{1}{\left[2\int_0^1 \alpha_X(\e)~d\e\right]^2}.
\end{equation}
\end{exemplo}

\begin{exercicio}
Verifique o axioma 1. 
\par
[ {\em Sugest\~ao:} usar o teorema de converg\^encia dominada de Lebesgue. ]
\end{exercicio}

\begin{exercicio}
Mostrar que se $d_{conc}(X,Y)\leq \gamma$, ent\~ao para cada $\ve>0$, \[\abs{\alpha(X,\ve)-\alpha(Y,\ve)}\leq 2\gamma.\] 
[ {\em Sugest\~ao:} mostrar e usar o fato que $\alpha(X,\ve)$ \'e o supremo de todos os valores de forma $\mu_X\{x\in X\colon f(x)\geq M_f+\ve\}$, onde $f$ \'e uma fun\c c\~ao $1$-Lipschitz cont\'\i nua. ] Deduzir o axioma 2.
\end{exercicio}

\begin{exercicio}
Verificar o axioma 3. 
\par
[ {\em Sugest\~ao:} usar v\'arias propriedades geom\'etricas das esferas estabelecidas por Paul L\'evy, do Ap\^endice \ref{a:esfera}. ]
\end{exercicio}

Eis uma ilustra\c c\~ao do axioma de normaliza\c c\~ao.

\begin{figure}[htp]
\centering
\scalebox{.5}{\includegraphics{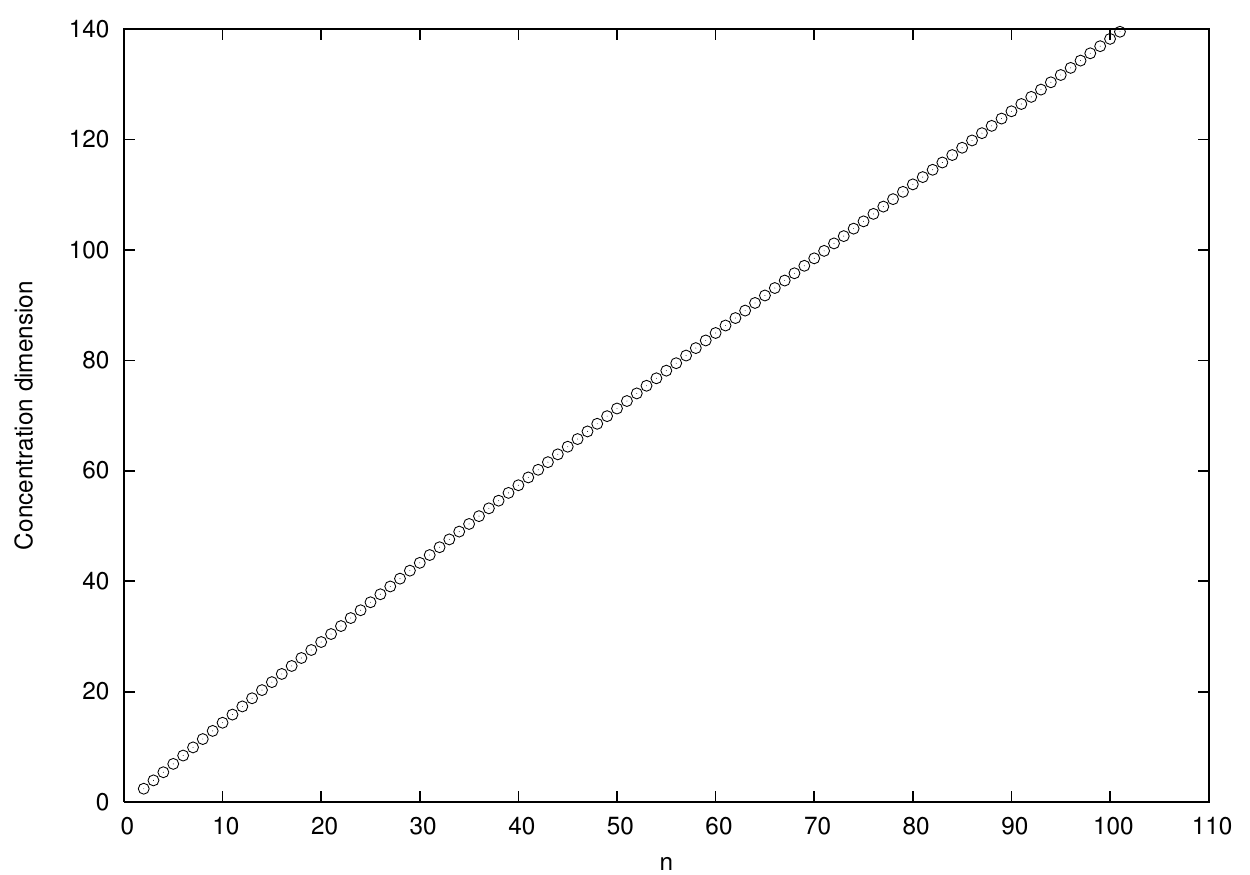}}
\caption{Dimens\~ao $\partial_{conc}$ de $n$-esferas para $2\leq n\leq 101$.}
\label{fig:concD100}
\end{figure}

\begin{exercicio}
Seja $(\Omega,d,\mu)$ um espa\c co m\'etrico com medida de probabilidade.
Mostra que, quando $n\to\infty$, temos as converg\^encias quase certas:
\begin{enumerate}
\item Para cada $\ve>0$, $\alpha((\varsigma_n,d\vert_{\varsigma_n},\mu_n),\ve)\to \alpha (\Omega,\ve)$,
\item $d_{conc}((\varsigma_n,d\vert_{\varsigma_n},\mu_n),\Omega)\to 0$,
\item se $\partial$ \'e uma fun\c c\~ao de dimensionalidade intr\'\i nseca, $\partial (\varsigma_n,d\vert_{\varsigma_n},\mu_n)\to \partial \Omega$.
\end{enumerate}
[ {\em Sugest\~ao:} use o lema de Cover--Hart. ]
\end{exercicio}

\begin{observacao}
A vers\~ao de dimens\~ao intr\'\i nseca apresentada acima, como sugerida em \citep*{pestov:07}, \citep*{pestov08c}, foi desenvolvida recentemente em um cen\'ario mais refinado e realista em \citep*{HSS}.
\end{observacao}

\subsection{Maldi\c c\~ao de dimensionalidade para alguns esquemas de indexa\c c\~ao} 
Obtenhamos alguns resultados mostrando a maldi\c c\~ao de dimensionalidade para esquemas de indexa\c c\~ao discutidos na subse\c c\~ao \ref{ss:mdck}. O modelo consiste de:

\begin{enumerate}
\item Uma fam\'\i lia $\mathscr F$ de fun\c c\~oes reais $1$-Lipschitz cont\'\i nuas sobre o dom\'\i nio m\'etrico $\Omega$, definidas parcialmente ou totalmente. As fun\c c\~oes de tal fam\'\i lia s\~ao aquelas fun\c c\~oes $1$-Lipschitz cont\'\i nuas que podem ser computados da maneira eficaz.
\smallskip
\item Dado um conjunto de dados $X\subseteq\Omega$, constr\'oi-se uma {\em esquema de indexa\c c\~ao} para $X$. O esquema consiste de uma subfam\'\i lia ${\mathscr F}_X\subseteq \mathscr F$ de fun\c c\~oes, submetidas \`a pr\'e-computa\c c\~ao, permitindo a execu\c c\~ao do seguinte algoritmo.
Dado uma subsequ\^encia finita de fun\c c\~oes $f_1,f_2,\ldots,f_k\in {\mathscr F}_X$ e a correspondente sequ\^encia de valores $f_i(q)$, o algoritmo devolve ou uma nova fun\c c\~ao $f_{k+1}\in {\mathscr F}_X$ para calcular $f_{k+1}(q)$, ou uma lista de pontos de $X$ a examinar.  
\smallskip
\item A lista de pontos devolvidos no final do c\'alculo deve incluir todos os pontos $x\in X$ tais que, para todos $f_1,\ldots,f_k\in{\mathscr F}$ indicadas pelo algoritmo, $\abs{f_i(q)-f_i(x)}\leq \ve$, onde $\ve>0$ \'e o raio de busca de intervalo inicial.
\smallskip
\item Para todos os pontos $x$ na lista, a condi\c c\~ao $d(x,q)<\ve$ \'e verificada, e os pontos satisfazendo-a s\~ao s\~ao devolvidos pelo algoritmo.
\end{enumerate}
\index{esquema! de indexa\c c\~ao}

Este tipo de esquema n\~ao \'e exclusivo, mas sim \'e muito comum. Para uma discuss\~ao mais detalhada com exemplos, consulte \citep*{PeSt06}, e para v\'arias implementa\c c\~oes na engenharia de dados, consulte \citep{chavez:01},\citep*{samet},\citep*{santini},\citep*{Yan},\citep{zezula:06}.

Precisamos de alguns resultados adicionais sobre a dimens\~ao VC e a concentra\c c\~ao de medida.

\begin{exercicio}
Seja $\mathscr C$ uma classe de conceitos, e $k$ um n\'umero natural. Ent\~ao a dimens\~ao da classe 
\[{\mathscr C}^{k{\mbox{\tiny -}}\cap}=\{\cap {\mathscr A}\colon {\mathscr A}\subseteq {\mathscr C},~
\sharp {\mathscr A}\leq k\},\]
que consiste de todas as interse\c c\~oes de $\leq k$ conceitos de $\mathscr C$, satisfaz
\[\VC\left({\mathscr C}^{k{\mbox{\tiny -}}\cap}\right)
 \leq 2k\VC(\mathscr C)\log(k\VC(\mathscr C)).\]
[ {\em Sugest\~ao:} comece com o lema de Sauer--Shelah, e depois imite o jeito na p\'agina \pageref{p:trick} baseado sobre lema \ref{l:trivial}. ]
\label{ex:vcdimk-intersect}
\end{exercicio}

\begin{proposicao}
Seja $C$ um subconjunto boreliano n\~ao negligenci\'avel de um espa\c co m\'etrico com medida de probabilidade, $(\Omega,d,\mu)$.
Denote $\alpha_C$ a fun\c c\~ao de con\-cen\-tra\-\c c\~ao de $C$ em rela\c c\~ao a m\'etrica induzida $d\vert C$ e a medida de probabilidade induzida $\mu/\mu(C)$. 
Ent\~ao para todos $\ve>0$
\[\alpha_C(\ve) \leq \frac{\alpha_{\Omega}(\ve/2)}{\mu(C)}.\]
\label{e:subspace}
\end{proposicao}

\begin{proof}
Sejam $\ve>0$ e $\delta<\alpha_C(\ve)$ quaisquer. Existem subconjuntos
$D,E\subseteq C$ a dist\^ancia $\geq\ve$ um de outro, satisfazendo $\mu(D)\geq\mu(C)/2$ e $\mu(E)\geq\delta\mu(C)$. Em particular, a medida de qualquer deles \'e pelo menos $\delta\mu(C)$. As $\ve/2$-vizinhan\c cas de $D$ e $E$ em $\Omega$ n\~ao se encontram pela desigualdade triangular. Por conseguinte, uma deles tem a medida $\leq 1/2$, e o seu complemento em $\Omega$, denote-o $F$, tem a propriedade $\mu(F)\geq 1/2$. Ao mesmo tempo, 
$\mu(F_{\ve/2})\leq 1-\delta\mu(C)$, porque $F_{\ve/2}$ n\~ao encontra um de dois conjuntos, $D$ ou $E$. Conclu\'\i mos:
$\alpha_{\Omega}(\ve/2)\geq \delta\mu(C)$,
e, formando o supremo sobre todos $\delta<\alpha_C(\ve)$,
\[\alpha_{\Omega}(\ve/2)\geq \alpha_C(\ve)\mu(C).\]
Em outras palavras, $\alpha_C(\ve)\leq\alpha_{\Omega}(\ve/2)/\mu(C)$, como desejado.
\end{proof}

\begin{corolario}
Sejam $C$ um subconjunto boreliano n\~ao negligenci\'avel de um espa\c co m\'etrico com medida de probabilidade, $(\Omega,d,\mu)$, e $f\colon C\to\R$ uma fun\c c\~ao $1$-Lipschitz cont\'\i nua. Seja $M$ um valor mediano de $f$ sobre $C$. Ent\~ao, para cada $\ve>0$,
\[\mu\{x\in C\colon \abs{f(x)-M}>\ve\}\leq 2\alpha(\Omega,\ve/2).
\]
\end{corolario}

\begin{proof}
\begin{align*}
\frac{1}{\mu(C)}\mu\{x\in C\colon \abs{f(x)-M}>\ve\} &=
\mu_C\{x\in C\colon \abs{f(x)-M}>\ve\}\\
 &\leq 2\alpha(C,\ve) \\
&\leq \frac{2}{\mu(C)}\alpha(\Omega,\ve/2).
\end{align*}
\end{proof}

Seja $\mathscr F$ uma classe de fun\c c\~oes reais sobre $\Omega$ (possivelmente parcialmente definidas). Denote ${\mathscr F}_{\geq}$ a fam\'\i lia de todos os subconjuntos 
\[\{\omega\in{\mathrm{dom}\, f}\colon f(\omega)\geq a\},~~a\in\R.\]

O resultado seguinte \'e uma varia\c c\~ao sobre os resultados de \citep*{pestov2012,pestov:13}.

\begin{teorema}
Suponha que o dom\'\i nio $\Omega$, munido de uma m\'etrica $d$ e uma medida de probabilidade $\mu$, tem o tamanho carater\'\i stico ${\mathbb{E}}_{\mu^2}d(X,Y)=1$ e a fun\c c\~ao de concentra\c c\~ao subgaussiana com exponente $\Theta(d)$:
\begin{equation}
\alpha_{\Omega}(\ve)=\exp\left(-\Theta(\ve^2 d) \right).
\label{eq:dim}
\end{equation} 
Seja $\mathscr F$ a fam\'\i lia de todas as fun\c c\~oes $1$-Lipschitz cont\'\i nuas sobre $\Omega$ usadas para escolher as fun\c c\~oes de decis\~ao para um esquema particular. Suponha que $\VC({\mathscr F}_{\geq})=\mbox{poly}\,(d)$ \'e polinomial em $d$.
Seja $X$ uma amostra aleat\'oria i.i.d. com $n$ pontos de $\Omega$ seguindo a lei $\mu$, onde $n$ \'e superpolinomial e subexponencial em $d$, $d=n^{o(1)}$ e $d=\omega(\log n)$. Ent\~ao cada esquema de indexa\c c\~ao para busca de vizinho mais pr\'oximo em $X$ de tipo descrito acima, usando as fun\c c\~oes de decis\~ao de uma subfam\'\i lia ${\mathscr F}_X\subseteq \mathscr F$ de tamanho $\mbox{poly}\,(n)$, tem o tempo esperado m\'edio da execu\c c\~ao superpolinomial em $d$, $d^{\omega(1)}$.
\label{t:maldicaoespacosmm}
\end{teorema}

Fixemos um valor $0<\ve_0<1$ estritamente positivo qualquer, por exemplo, $\ve_0=1/2$. 
Para cada $f\in {\mathscr F}$ definamos os conjuntos
\begin{align*}
\mbox{Out}_f&=\left\{x\in \mbox{dom}\,f\colon \abs{f(x)-M_f}>\frac{\ve}{4}\right\},\\
\mbox{In}_f &= \mbox{dom}\,f\setminus \mbox{Out}_f,
\end{align*}
onde $M_f$ \'e o valor mediano de $f$ sobre o seu dom\'\i nio de defini\c c\~ao.

\begin{exercicio}
Deduza que a dimens\~ao de Vapnik--Chervonenkis da classe ${\mathscr D}=\{\mbox{In}_f\colon f\in {\mathscr F}\}$ \'e polinomial em $d$.
\end{exercicio}

Seja $k$ o comprimento esperado da sequ\^encia $f_1,\ldots,f_i$ de fun\c c\~oes geradas pelo algoritmo de busca sobre todos os conjuntos de dados $X$ e todos os centros de busca $q$ de raio $\ve$. Se $k$ \'e superpolinomial em $d$, n\~ao tem nada a mostrar. Sen\~ao, segue-se do exerc\'\i cio \ref{ex:vcdimk-intersect} que a dimens\~ao VC da classe ${\mathscr D}^{k\mbox{\tiny -}\cap}$ de todas as interse\c c\~oes de $\leq k$ elementos de $\mathscr D$ \'e $\mbox{poly}\,d$ tamb\'em. 
Segundo o teorema de Glivenko--Cantelli, quando $d,n\to\infty$, com confian\c ca $1-\exp\left(-\Theta(-\ve^2 d)\right)$, a medida emp\'\i rica de todos os elementos de  ${\mathscr D}^{k\mbox{\tiny -}\cap}$ difere da medida $\mu$ por menos de $\ve/4$. 

Agora seja $X$ um conjunto de dados qualquer fixo tendo a propriedade acima. Um esquema de indexa\c c\~ao est\'a escolhida neste momento. Segundo a nossa hip\'otese, a subclasse ${\mathscr F}_X\subseteq {\mathscr F}$ satisfaz $\sharp{\mathscr F}_X =\mbox{poly}\,(n)$. Um centro de busca $q$, um elemento aleat\'orio de $\Omega$, satisfaz
\begin{align*}
P\left[q\in \bigcup_{f\in {\mathscr F}_X} \mbox{Out}_f\right]
&\leq \sharp{\mathscr F}_X \cdot\alpha(\Omega,\ve/8) \\
&= \exp\left(-\Theta(-\ve^2 d)\right).
\end{align*}
Em outras palavras, com confian\c ca $1-\exp\left(-\Theta(-\ve^2 d)\right)$, qualquer seja $f\in {\mathscr F}_X$, se $q$ pertence a $\mbox{dom}\,f$, ent\~ao $q\in \mbox{In}_f$. Al\'em disso, com alta confian\c ca, a vizinho mais pr\'oximo de $q$ dentro de $X$ vai ser a dist\^ancia $\geq \e_0$.
Fixemos um $q$ qualquer com estas duas propriedades.

A s\'erie de buscas de intervalo para determinar o vizinho mais pr\'oximo vai necessariamente incluir a busca de intervalo com o centro $q$ e um raio $\geq \e_0/4$. Executemos uma tal busca. Sejam $f_i$, $i=1,\ldots, k$ fun\c c\~oes elementos de ${\mathscr F}_X$ escolhidos pelo algoritmo para calcular os valores em $q$. Como $q\in\mbox{dom}\,f_i$ para todos $i$, temos 
\[\abs{f_i(q)-M_f}<\frac{\ve}4,~i=1,2,\ldots,k.\]
Os pontos de $X$ que n\~ao pertencem ao conjunto $\cup\{ \mbox{Out}_{f_i}\colon i=1,2,\ldots,k\}$ t\^em que ser devolvidos. Temos:
\begin{align*}
\sharp\{x\in X\colon x\notin \cup_{i=1}^k \mbox{Out}_{f_i}\} &= n\mu_{\sharp}\left[\Omega\setminus\cup_{i=1}^k \mbox{Out}_{f_i}\right] \\
&\geq n\left(\mu\left[\Omega\setminus\cup_{i=1}^k \mbox{Out}_{f_i}\right]-\ve/2 \right)\\
&\geq \frac n2 \\
&= d^{\omega(1)}.
\end{align*}
\qed

O resultado seguinte (que mencionemos sem demonstra\c c\~ao) sugere que, para todos os fins pr\'aticos, a hip\'otese $\VC({\mathscr F}_{\geq})=\mbox{poly}\,(d)$ \'e razo\'avel. 

\begin{teorema}[\citet*{GJ}, Theorem 2.3]  
Seja 
\[{\mathscr F}=\{x\mapsto f(\theta,x)\colon\theta\in \R^s\}\]
uma classe parametrizada de fun\c c\~oes bin\'arias. Suponha que, para todo input $x\in\R^n$, existe um algoritmo calculando $f(\theta,x)$, e o c\'alculo precisa ao m\'aximo $t$ opera\c c\~oes dos seguintes tipos:
\begin{itemize}
\item os opera\c c\~oes aritm\'eticas $+,-,\times$ e $/$ sobre reais,
\item saltos condicionados sobre $>$, $\geq$, $<$, $\leq$, $=$, e $\neq$ compara\c c\~oes de reais, e
\item sa\'\i da $0$ ou $1$.
\end{itemize}
Ent\~ao $\VC({\mathscr F})\leq 4s(t+2)$. \qed
\label{th:gj}
\end{teorema}

Ali\'as, a dimens\~ao $s$ do espa\c co de par\^ametros $\R^s$ n\~ao pode ser superpolinomial em $d$, pois neste caso j\'a lendo o par\^ametro vai demorar o tempo $d^{\omega(1)}$.

\begin{exercicio}
Tente mostrar o teorema de Goldberg e Jerrum usando as t\'ecnicas desenvolvidas na prova do teorema \ref{th:vc}.
\end{exercicio}

\section{Lema de Johnson--Lindenstrauss\label{s:ljl}}

\subsection{Formula\c c\~ao do resultado e discuss\~ao preliminar}

\begin{teorema}[Lema de Johnson--Lindenstrauss]
Seja $X$ um subconjunto com $n$ elementos num espa\c co de Hilbert $\mathcal H$, e seja $0<\e\leq 1$ qualquer. Existe um operador linear $T\colon {\mathcal H}\to \ell^2(k)$, onde
\[k=O(\e^{-2}\log n),\]
tal que
\[(1-\e)\norm{x-y}< \norm{T(x)-T(y)} <(1+\e)\norm{x-y}\]
para todos $x,y\in X$. 
\index{lema! de Johnson--Lindenstrauss}
\end{teorema}

Vamos deduzir valores concretos das constantes na nossa demonstra\c c\~ao. A saber, o nosso argumento em particular vai mostrar que se $n\geq 96$, ent\~ao a dimens\~ao $k$ \'e no m\'aximo
\[k \leq \left\lceil\frac{65\log n}{\e^2}\right\rceil.\]
(Certamente, as cotas n\~ao s\~ao \'otimas e podem ser melhoradas).

Obviamente, pode-se supor que $\dim {\mathcal H}\leq n$, substituindo $\mathcal H$ pelo espa\c co linear gerado por $X$. No entanto, o espa\c co $T(\mathcal H)$ tem dimens\~ao logar\'\i tmica em $n$. O resultado, originalmente motivado pelas necessidades da an\'alise funcional, tornou-se uma ferramenta importante na inform\'atica te\'orica bem como pr\'atica. O lema \'e sujeito a muita pesquisa contempor\^anea.

Chama a aten\c c\~ao a consequ\^encia seguinte do resultado, bastante surpreendente. Se um espa\c co de Hilbert $\mathcal H$ contiver um sistema ortogonal de $n$ vetores, ent\~ao, obviamente, a dimens\~ao de $\mathcal H$ seria maior ou igual \`a $n$. Agora, chamamos um sistema de vetores $\e$-{\em quase ortogonal} se o \^angulo entre quaisquer dois vetores distintos pertence ao intervalo $\pi/2\pm\e$. O lema de Johnson--Lindernstrauss, aplicada aos vetores b\'asicos $e_1,e_2,\ldots,e_n$, implica a exist\^encia de sistemas $\e$-quase ortogonais de tamanho {\em exponencial} em dimens\~ao do espa\c co de Hilbert, para um $\e>0$ fixo.

Agora fazemos algumas prepara\c c\~oes para a demonstra\c c\~ao. Substituamos $\mathcal H$ por $\ell^2(n)$. Vamos reduzir o problema para um de geometria da esfera euclidiana. Com esta finalidade, formemos o conjunto
\begin{equation}
\label{eq:y}
Y = \left\{\frac{x-y}{\norm{x-y}}\colon x,y\in X, x\neq y\right\}.\end{equation}
Temos: $Y\subseteq \s^{n-1}$ e $\abs Y = n(n-1)/2$. \'E claro que basta mostrar a exist\^encia de um operador linear $T$ tendo a propriedade seguinte:
\begin{equation}
\label{eq:1}
\forall y\in Y,~~1-\e < \norm{Ty} <1+\e.\end{equation}
Em outras palavras, $T$ envia $Y$ dentro a casca esf\'erica de espessura $2\e$.
De fato, basta mostrar que existe uma constante $M>0$ tal que
\begin{equation}
\label{eq:M}
\forall y\in Y,~~(1-\e)M < \norm{Ty} <(1+\e)M.\end{equation}
Neste caso, podemos substituir $T$ pelo operador linear $M^{-1}T$ e obter (\ref{eq:1}).

Denotemos por $p_k$ a proje\c c\~ao ortogonal de $\mathcal H=\ell^2(n)$ sob o subespa\c co gerado pelas $k$ primeiras coordenadas. Como $p_k$ \'e um operador linear de norma um, a fun\c c\~ao $f=\norm\cdot\circ p_k$ \'e $1$-Lipschitz cont\'\i nua.
A prova cl\'assica do lema de Johnson--Lindenstrauss \'e baseada sobre a concentra\c c\~ao dessa fun\c c\~ao, restrita sobre a esfera, em torno do seu valor mediano: 
\begin{equation}
\label{eq:f}
\s^{n-1}\ni x\mapsto f(x) = \norm{p_k(x)} = \left(\sum_{i=1}^k x_i^2\right)^{1/2} \in\R.\end{equation}
Aqui, a medida na esfera unit\'aria \'e a medida de Haar, ou seja, a \'unica medida boreliana de probabilidade invariante pelas rota\c c\~oes da esfera, veja se\c c\~ao (\ref{s:haar}). A dist\^ancia na esfera \'e euclidiana. 

Vamos estimar o valor mediano $M_f$, fazendo isso em poucas etapas.
Primeiramente, note que a esperan\c ca do quadrado de $f$ pode ser calculada exatamente:
\begin{align*}
\E(f^2) &= \E\left( \sum_{i=1}^k x_i^2\right) \\
&= \sum_{i=1}^k \E(x_i^2) \\
&= \frac kn,
\end{align*}
porque para $n=k$ temos
\[\E(f^2) = 1 = \sum_{i=1}^n \E(x_i^2) = n \E(x_i^2)\]
e $\E(x_i^2)$ s\~ao todas iguais para $i=1,2,\ldots,n$ por considera\c c\~ao de simetria.

Como $f\geq 0$, temos $\left(M_f\right)^2=M_{f^2}$. Ent\~ao, bastaria estimar o valor mediano de $f^2$. Isso parece ser poss\'\i vel gra\c cas \`a observa\c c\~ao feita no exerc\'\i cio \ref{e:medianomedio}: sobre as estruturas de alta dimens\~ao, o valor mediano e o valor m\'edio s\~ao pr\'oximos. Infelizmente, a diferen\c ca entre dois valores dada pelo exerc\'\i cio \'e da ordem de grandeza maior do que a esperan\c ca! 
Esse argumento sim se aplica \`a fun\c c\~ao $f$, e n\~ao a $f^2$.
E n\~ao se pode limitar $\E(f)$ por baixo atrav\'es de $\E(f^2)$ de modo \'util. Segundo a desigualdade de Jensen, $\abs{\E(f)}\leq\left(\E(f^2)\right)^{1/2}$, mas este \'e o sentido errado. A desigualdade de H\"older implica que $\E f^2 \leq \E f \cdot \sup f = \E f$, mas esta estimativa \'e fraca demais, como ser\'a visto.

A solu\c c\~ao ser\'a fornecida pelo fato que sobre as estruturas de alta dimens\~ao, a vari\^ancia de uma fun\c c\~ao Lipschitz cont\'\i nua,
\[\var f = \E(f-\E f)^2 = \E(f^2) - \E(f)^2,\]
\index{vari\^ancia}
\'e pequena. Ent\~ao, o nosso plano de trabalho \'e o seguinte: de $\E(f^2)$ para $\E(f)$, e depois para $M_f$. 

\subsection{Mais resultados sobre a concentra\c c\~ao de medida}
Comecemos com a prova do exerc\'\i cio \ref{e:medianomedio}.

\begin{lema}
Seja $(X,d,\mu)$ um espa\c co m\'etrico separ\'avel munido de uma medida boreliana de probabilidade, e seja $f\colon X\to\R$ uma fun\c c\~ao $1$-Lipschitz cont\'\i nua. Ent\~ao,
\begin{equation}
\abs{\E f-M_f} \leq \int_0^\infty \alpha_X(\e)\,d\e.
\end{equation}
\label{l:mm}
\end{lema}

\begin{proof}
Denotemos 
\[A_\pm = \{x\in X\colon f(x)\gtreqqless M\}.\]
Agora,
\begin{align*}
\abs{\E f-M_f} &= \left\vert \int_{A_+} (f-M_f)\,d\mu -\int_{A_-} (M_f-f)\,d\mu \right\vert \\
&\leq  \max\left\{\left\vert \int_{A_+} (f-M_f)\,d\mu\right\vert, \left\vert\int_{A_-} (M_f-f)\,d\mu \right\vert \right\},
\end{align*}
porque ambas integrais s\~ao positivas. Basta estimar uma delas, por exemplo $A_+$. Como $M_f$ \'e um valor mediano de $f$, o valor da medida de $A_\pm$ \'e pelo menos $1/2$. Como a fun\c c\~ao $f$ \'e $1$-Lipschitz cont\'\i nua, temos
\[\{x\colon f(x)\geq M_f+\e\}\subseteq \{x\colon x\notin (A_-)_\e\},\]
e por conseguinte
\begin{equation}
\mu\{x\colon f(x)\geq M_f+\e\} \leq \alpha_X(\e).
\end{equation}
Para calcular a integral, dividimos o volume entre o gr\'afico de $f$ e o n\'\i vel $M_f$ acima do conjunto $A_+$ em faixas horizontais infinitesimais (Fig. \ref{fig:faixas}).

\begin{figure}[htp]
\centerline{\includegraphics[width=6cm]{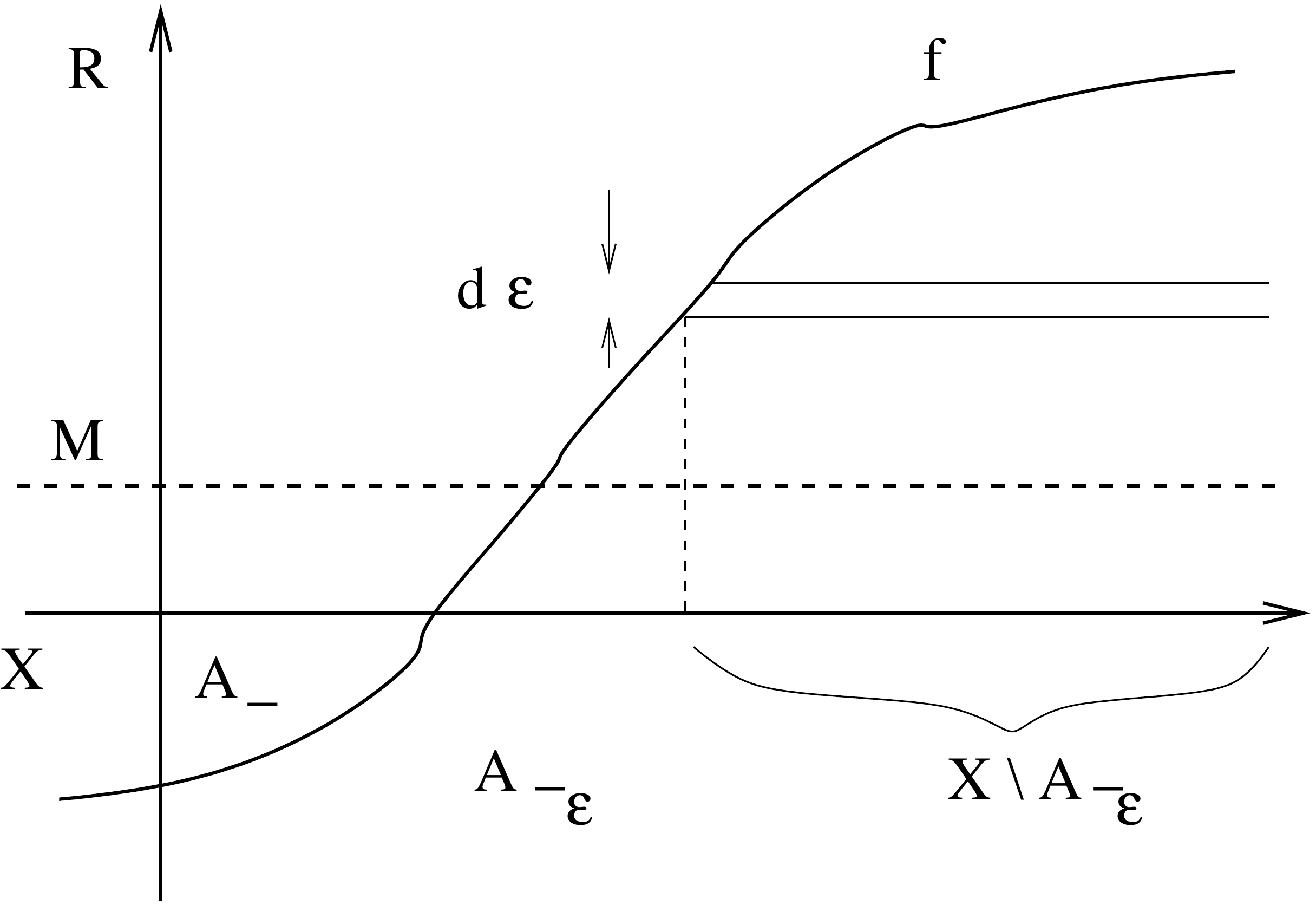}}
\caption{Estimando $\int_{A_+}(f-M_f)\,d\mu$.}
\label{fig:faixas}
\end{figure}
Segue-se que o volume da cada faixa \'e aproximada (menos as quantidades de alta ordem) por $\alpha(\e)\,d\e$. 
\end{proof}

\begin{corolario}
Suponha que a fun\c c\~ao de concentra\c c\~ao admita uma cota superior gaussiana,
\begin{equation}
\label{eq:gaussian}
\alpha_X(\e)\leq C_1 \exp(-C_2\e^2n).\end{equation}
Ent\~ao para toda fun\c c\~ao $f$ $1$-Lipschitz cont\'\i nua sobre $X$ temos
\begin{equation}
\abs{\E f-M_f} \leq  \frac{C_1\sqrt \pi}{2\sqrt{C_2 n}}.
\end{equation}
\label{c:paracotagaussiana}
\end{corolario}

\begin{proof}
\begin{align*}
\int_0^\infty \exp(-C_2\e^2n)\,d\e &=
 \frac{1}{\sqrt{C_2 n}}\int_0^\infty \exp(-C_2\e^2n)\,d(\sqrt{C_2 n}\e) \\
&= \frac{\sqrt \pi}{2\sqrt{C_2 n}}.
\end{align*}
\end{proof}

\begin{corolario}
Se $f$ \'e uma fun\c c\~ao real $1$-Lipschitz cont\'\i nua sobre a esfera euclidiana $\s^{n-1}$, ent\~ao
\begin{equation}
\abs{\E f-M_f} \leq  \frac{5\sqrt{2\pi}}{2\sqrt{n}}.
\end{equation}
\end{corolario}

De mesma maneira, pode-se estimar a vari\^ancia de uma fun\c c\~ao $1$-Lipschitz cont\'\i nua sobre um espa\c co de alta dimens\~ao.
 
\begin{lema}
Seja $(X,d,\mu)$ um espa\c co m\'etrico separ\'avel munido de uma medida boreliana de probabilidade, e seja $f\colon X\to\R$ uma fun\c c\~ao $1$-Lipschitz cont\'\i nua. Ent\~ao,
\begin{equation}
\var f \leq 2\int_0^\infty\alpha_X (\sqrt \e)\,d\e 
\end{equation}
\end{lema}
 
 \begin{proof}
 \begin{align*}
 \var f &=
 \E((f-\E f)^2) \\
 &= \E(f-M_f+M_f-\E f)^2 \\
&= \E((f-M_f)^2) + 2\E(f-M_f)(M_f-\E f) + (M_f-\E f)^2 \\
 &=  \E((f-M_f)^2)- (M_f-\E f)^2 \\
&\leq  \E((f-M_f)^2).
 \end{align*}
Dividamos de novo a imagem em faixas infinitesimais de altura $d \e$ para notar que
\[\mu\{x\in X\colon (f(x)-M_f)^2 \geq \e\}\leq \mu\{x\in X\colon\abs{f(x)-M_f}\geq\sqrt\e\}\leq 2\alpha_X (\sqrt \e).\]
Isso nos d\'a a estimativa
\[\E((f-M_f)^2)\leq 2\int_0^\infty\alpha_X (\sqrt \e)\,d\e.\]
\end{proof}

 \begin{corolario}
Temos, para uma fun\c c\~ao $1$-Lipschitz cont\'\i nua $f$ qualquer sobre a esfera $\s^{n-1}$:
 \[\var f \leq \frac{20}{n}.\]
\end{corolario}
 
\begin{proof}
\begin{align*}
\int_0^\infty\alpha_X (\sqrt \e)\,d\e &\leq 
 5\int_0^\infty \exp(-\e n/2)\,d\e \\
 &= \frac {10}n\int_0^\infty \exp(-\e n/2)\,d(\e n/2) \\
 &= \frac {10}n.
\end{align*}
\end{proof}

\subsection{Estimativa do valor mediano de $f$}
Reunindo nossas observa\c c\~oes sobre a fun\c c\~ao $f$ defininda na eq. (\ref{eq:f}), obtemos:
\[\E (f^2) =\frac kn,~~\abs{\E f-M_f} \leq  \frac{5\sqrt{2\pi}}{2\sqrt{n}},~~
\var f \leq \frac{20}{n}.\]
A \'ultima significa:
\[\left\vert \E(f^2) - (\E f)^2 \right\vert\leq \frac{20}{n}.\]
Por conseguinte,
\[(\E f)^2 \geq \frac{k-20}{n},~~\E f \geq \frac{\sqrt{k-20}}{\sqrt n}\]
e 
\[M_f \geq \frac{\sqrt{k-20}- (5/2)\sqrt{2\pi}}{\sqrt n}.\]

Assumindo que $k$ seja bastante grande de modo que 
\[\sqrt{k-20}- (5/2)\sqrt{2\pi} >\frac{\sqrt k}4\]
(este ser\'a o caso a partir de $k\geq 96$), chegamos a uma cota inferior razo\'avel:
\begin{equation}
\label{eq:mf}
M_f > \frac{\sqrt k}{4\sqrt{n}}.\end{equation}

\subsection{Prova do lema de Johnson--Lindenstrauss}

Seja $f$ a fun\c c\~ao definida na eq. (\ref{eq:f}), e $u\in O(n)$ uma transforma\c c\~ao ortogonal. 

\begin{exercicio} 
Verificar que o valor $^uf(x)$ \'e igual \`a norma da imagem de $x$ pela proje\c c\~ao ortogonal sobre o espa\c co de dimens\~ao $k$ de $\ell^2(n)$ gerado pelas imagens $u(e_1)$, $u(e_2)$, $\ldots$, $u(e_k)$. Em outras palavras,
\[^uf(x) = \norm{u^{\ast}p_ku(x)}.\]
\label{ex:u*pu}
\end{exercicio}

\'E claro que todo subespa\c co de dimens\~ao $k$ pode ser obtido desse modo.

Vamos aplicar o lema \ref{l:tool2} \`a fun\c c\~ao $^uf$, ao conjunto finito $Y\subseteq\s^{n-1}$ definido na eq. (\ref{eq:y}), assim que ao valor
\[\e^\prime = \e M_f.\]
A condi\c c\~ao (\ref{eq:card}) sobre a cardinalidade de $F=Y$ no lema 
torna-se assim:
\[\frac{n(n-1)}2 = \abs Y < \frac 14 \exp(\e^2 M_f^2 n /2),
\]
e, tendo em vista eq. (\ref{eq:mf}), a cota ser\'a verificada uma vez temos
\begin{align*}
\frac{n^2}2 &<
 \frac 14 \exp\left(\e^2 \left(\frac{\sqrt k}{4\sqrt{n}}\right)^2 n /2\right) \\
&= \frac 14 \exp(\e^2 k/32).
\end{align*}
Aplicando o logaritmo natural, resolvemos a desigualdade para obter
\[2\log n < \frac{\e^2 k}{32}-\log 2,\]
ou
\[k> \frac{64\log n}{\e^2} +32\log 2.\]
Isso \'e satisfeito, por exemplo, quando 
\[k> \frac{65\log n}{\e^2}.\]
Obtemos a eq. (\ref{eq:M}) com $T=u^{-1}p_ku$, m\'odulo a observa\c c\~ao no exerc\'\i cio \ref{ex:u*pu}.
\hfill \qed

\begin{exercicio}
Implemente o algoritmo das proje\c c\~oes aleat\'orias em R e aplique ao conjunto de dados {\em Phoneme} (p\'agina \pageref{pagina:phoneme}), combinando com o classificador $k$-NN, ou melhor, um ensemble de classificadores $k$-NN combinados com v\'arias proje\c c\~oes.
(Se voc\^e conseguiu melhorar o desempenho do classificador, voc\^e fez melhor do que eu ou meus alunos).
\end{exercicio}

\subsection{Discuss\~ao: proje\c c\~oes aleat\'orias e aprendizagem}

\subsubsection{}
O lema de Johnson--Lindenstrauss foi mostrada em artigo \citep*{JL}. O resultado tornou-se desde ent\~ao uma ferramenta maior na an\'alise funcional assim que na inform\'atica, onde ele \'e usado para projetar os algoritmos aleatorizados, incluindo os da redu\c c\~ao de dimensionalidade. A pesquisa moderna \'e amplamente dedicada aos meios de escolher o operador aleat\'orio $T$ de uma maneira computacionalmente eficaz.

\subsubsection{}
De fato, $T$ n\~ao precisa ser uma proje\c c\~ao (normalizada) aleat\'oria, como na prova acima. Por exemplo, uma matriz aleat\'oria do tamanho $n\times k$ com coeficientes gaussianos bastaria tamb\'em. Ainda mais, os coeficientes podem ser zeros e uns, e al\'em disso, a matriz pode ser esparsa (conter mais zeros do que uns). Para leituras futuras, eu recomendo \citep*{matousek} e \citep*{vempala}, bem como \citep*{naor}.

\subsubsection{\label{ss:limitacoes}} 
O exemplo seguinte mostra que a \'unica proje\c c\~ao aleat\'oria seguida pelo classificador qualquer n\~ao vai dar certo.
No espa\c co euclidiano $\R^N$ de uma alta dimens\~ao $N\gg 1$ sejam $\mu_0$ e $\mu_1$ duas medidas invariantes pelas rota\c c\~oes, da massa total $1/2$ cada uma, suportadas nas esferas conc\^entricas $(1/2)\s^{N-1}$ e $\s^{N-1}$, respetivamente, em torno de zero. Criemos o problema de aprendizagem, cujo lei subjacente \'e a medida de probabilidade $\mu=\mu_1+\mu_2$ e a fun\c c\~ao de regress\~ao \'e igual a $\eta=\chi_{s^{N-1}}$. 

Seja $\pi_k$ uma proje\c c\~ao de $\R^N$ sobre um subespa\c co de uma menor dimens\~ao $k\ll N$.
Segundo um resultado de Poincar\'e, as imagens diretas $\pi_{\ast}\mu_i$, $i=0,1$, depois da renormaliza\c c\~ao do argumento por $\sqrt{N-k}$ (o que n\~ao vai afeitar o desempenho do classificador), s\~ao aproximadamente gaussianas, centradas e com vari\^ncia $1/2$ e $1$, respetivamente. A regra composta ${\mathcal L}\circ \pi_k$ vai possuir aproximadamente o mesmo erro de aprendizagem que a regra $\mathcal L$ aprendendo o problema cuja lei tem a densidade 
\[\frac 12\left[ \frac 1{\sqrt 2\pi}e^{-\norm{x}^2/2}+\frac 1 {\sqrt \pi}e^{-\norm{x}^2} \right]\]
e a fun\c c\~ao de regress\~ao 
\[\eta(x) = \frac{\frac 1{\sqrt 2\pi}e^{-\norm{x}^2/2}}{\frac 1{\sqrt 2\pi}e^{-\norm{x}^2/2}+\frac 1 {\sqrt \pi}e^{-\norm{x}^2}}, \]
a norma sendo a de $\ell^2(k)$. \'E claro que o erro de Bayes vai ser estritamente positivo, e por conseguinte o conceito $C=\s^{N-1}$ n\~ao pode ser aprendido com qualquer regra composta $\mathcal L\circ \pi_k$.

\subsubsection{}
Ao mesmo tempo, o teorema de Cram\'er--Wold \ref{t:cramer-wold}, que vamos mostrar no pr\'oximo cap\'\i tulo, diz que uma medida boreliana de probabilidade sobre $\R^d$ \'e unicamente definida pelas suas imagens diretas sob todas as proje\c c\~oes unidimensionais. Por conseguinte, qualquer problema de aprendizagem no espa\c co $\R^d$, ou at\'e mesmo no espa\c co de Hilbert $\ell^2$, \'e unicamente definido, e pode ser aprendido, pelas suas imagens sob as proje\c c\~oes unidimensionais. Pode ser que aprendizagem ensemble de um conjunto de regras $k$-NN combinadas com $m$ proje\c c\~oes aleat\'orias unidimensionais \'e universalmente consistente, quando $m$ \'e de tamanho razo\'avel, por exemplo $m=O(\log n)$?

\subsubsection{}
A ideia de usar aprendizagem ensemble de classificadores $k$-NN combinados com proje\c c\~oes aleat\'orias unidimensionais foi sugerida em \citep*{FJS}. 

O algoritmo sugerido no artigo depende de tr\^es par\^ametros: $k$ (o n\'umero de vizinhos mais pr\'oximos), $r>1$, e $N$. Dado uma amostra rotulada, $\sigma$, no espa\c co euclidiano $\R^d$, e um ponto $x\in\R^d$, sugere-se escolher $\lfloor rk\rfloor$ vizinhos mais pr\'oximos de $x$ em $\sigma$, depois aplicar a eles, assim que a $x$, $N$ proje\c c\~oes aleat\'orias unidimensionais (seguindo a medida de Haar sobre a esfera), $\pi_1,\pi_2,\ldots,\pi_N$, depois executar o algoritmo $k$-NN para todas as imagens, e escolher o r\'otulo de $x$ pelo voto majorit\'ario. 
O resultado principal do artigo (teorema 3) afirma que o algoritmo \'e universalmente consistente quando $k/n\to\infty$. Por\'em, o teorema n\~ao cont\'em algumas hip\'oteses sobre os valores dos par\^ametros $r$ e $N$, assim que seu comportamento assint\'otico. A \'unica condi\c c\~ao mencionada \'e que $k,n\to\infty$, $k/n\to 0$. Por isso, n\~ao parece poss\'\i vel verificar a validade do resultado. 
Al\'em disso, o algoritmo n\~ao resolve uma das dificuldades principais em altas dimens\~oes, que \'e a busca de vizinhos mais pr\'oximos, e assim, em nossa opini\~ao, vai um pouco contra a l\'ogica. 

\subsubsection{} 
Para um panor\^amico do estado da arte na \'area de redu\c c\~ao m\'etrica aleatorizada de dimensionalidade, veja \citep*{naor}. 
Combinar esta teoria avan\c cada com a aprendizagem supervisionada pode ser uma dire\c c\~ao promissora.

\section{Redu\c c\~ao boreliana}

\subsection{}
Geralmente, as aplica\c c\~oes $f$ que realizam a redu\c c\~ao de dimensionalidade em v\'arias \'areas da inform\'atica s\~ao cont\'\i nuas, por exemplo Lipschitz cont\'\i nuas, como no m\'etodo de proje\c c\~oes aleat\'orias. No contexto de aprendizagem de m\'aquina, esta \'e uma condi\c c\~ao bastante restritiva. O problema pode ser intuitivamente descrito como segue. 

Sejam $\Omega$ e $\Upsilon$ dois dom\'\i nios, e seja $f\colon\Omega\to\Upsilon$ uma fun\c c\~ao cont\'\i nua, realizando a redu\c c\~ao de dimensionalidade. Dado uma amostra rotulada $\sigma$ em $\Omega$ e um ponto $x\in\Omega$, apliquemos uma regra de aprendizagem $\mathcal L$ ao dom\'\i nio $\Upsilon$ \`a imagens $f(\sigma)$ da amostra, afim de determinar o r\'otulo do ponto $f(x)$, que ser\'a devolvido pela regra composta ${\mathcal L}\circ f$ como r\'otulo de $x$. Agora suponha que a dimens\~ao topol\'ogica do dom\'\i nio $\Omega$ \'e mais alta do que a de $\Upsilon$ (a situa\c c\~ao comum). 

\begin{figure}[ht]
\centering
\scalebox{.25}{\includegraphics{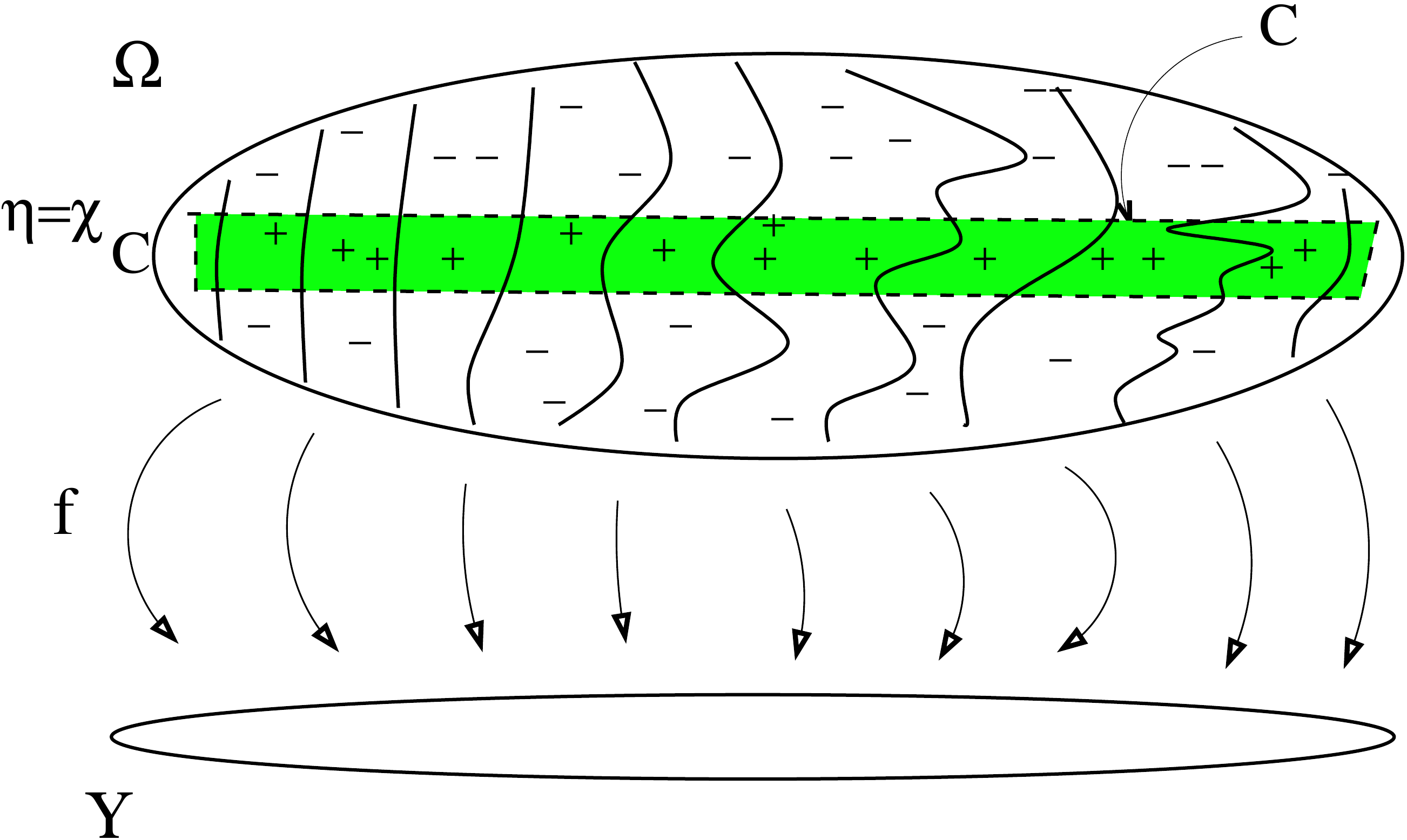}}
\caption{Fibra\c c\~ao produzida por uma fun\c c\~ao cont\'\i nua.}
\label{fig:distances}
\end{figure}

Pelas raz\~oes topol\'ogicas, as fibras $f^{-1}(y)$ tipicamente t\^em a dimens\~ao topol\'ogica estritamente positiva, $\dim\Omega - \dim\Upsilon$. Em particular, eles est\~ao altamente n\~ao triviais. A regra $\mathcal L$ n\~ao faz distin\c c\~ao entre os pontos $x\in \Omega$ tendo a mesma imagem, $f(x)$, devolvendo o mesmo r\'otulo para todos. Deste modo, se, por exemplo, o conceito $C\subseteq\Omega$ a ser aprendido passa a ser transversal \`a folia\c c\~ao definida por $f$, o resultado vai ser p\'essimo.

O erro da regra ${\mathcal L}\circ f$ vai ser pr\'oximo ao erro de Bayes se e somente se a fun\c c\~ao de regress\~ao toma valores aproximadamente constantes ao longo das fibras $f^{-1}(y)$. De modo equivalente, precisamos das garantias que as imagens diretas $f_{\ast}(\mu_0)$ e $f_{\ast}(\mu_1)$ das distribui\c c\~oes de pontos rotulados $0$ e $1$ respectivamente sejam t\~ao bem separadas em $\Upsilon$ que em $\Omega$. A este respeito, o exemplo na subse\c c\~ao \ref{ss:limitacoes} mostra que h\'a restri\c c\~oes sobre a validade de resultados deste g\^enero.

\subsection{}
No entanto, analisando de perto o modelo te\'orico da aprendizagem estat\'\i stica, pode notar-se que a no\c c\~ao da consist\^encia universal \'e na verdade insens\'\i vel a estrutura m\'etrica ou mesmo topol\'ogica no dom\'\i nio, enquanto a estrutura {\em boreliana} permanece intacta. Isto permite, atrav\'es de uma inje\c c\~ao boreliana, reduzir os dados para um caso de baixa dimens\~ao, at\'e mesmo unidimensional, ap\'os o qual o algoritmo de aprendizagem composto continua a ser universalmente consistente.

Relembramos que uma aplica\c c\~ao $f\colon \Omega\to W$ entre dois espa\c cos borelianos \'e dita {\em isomorfismo boreliano} se $f$ \'e bijetiva, e $f$ e $f^{-1}$ s\~ao borelianas. Isso significa que $f$ estabelece uma bije\c c\~ao entre a estrutura boreliana $\mathcal B_\Omega$ e $\mathcal B_W$. 

Cada aplica\c c\~ao cont\'\i nua \'e boreliana, e
cada homeomorfismo (isso \'e, uma bije\c c\~ao cont\'\i nua, com o inverso cont\'\i nuo) \'e um isomorfismo boreliano. Mas existem  muito mais aplica\c c\~oes borelianas do que aplica\c c\~oes cont\'\i nuas, e muito mais isomorfismos borelianos do que homeomorfismos. 

\begin{exercicio}
Mostre que n\~ao existe alguma inje\c c\~ao cont\'\i nua do quadrado $[0,1]^2$ dentro do intervalo $[0,1]$. 
\par
[ {\em Sugest\~ao:} uma inje\c c\~ao cont\'\i nua sobre um espa\c co compacto \'e um homeomorfismo... ]
\end{exercicio}

Ao mesmo tempo, existe uma inje\c c\~ao boreliana do quadrado no intervalo. 
Ela pode ser obtido usando o ``entrela\c camento'' dos d\'\i gitos nas expans\~oes bin\'arias de $x$ e de $y$ num par $(x, y) \in [0,1]^2$ (sujeito as precau\c  c\~oes habituais sobre as sequ\^encias infinitas de uns):
\begin{equation}
 \label{eq:borelmap}
  [0,1]^2\ni(0.a_1a_2\ldots, 0.b_1b_2\ldots)\mapsto (0.a_1b_1a_2b_2\ldots)\in [0,1].\end{equation}

\begin{figure}[ht]
\begin{center}
  \scalebox{0.25}[0.25]{\includegraphics{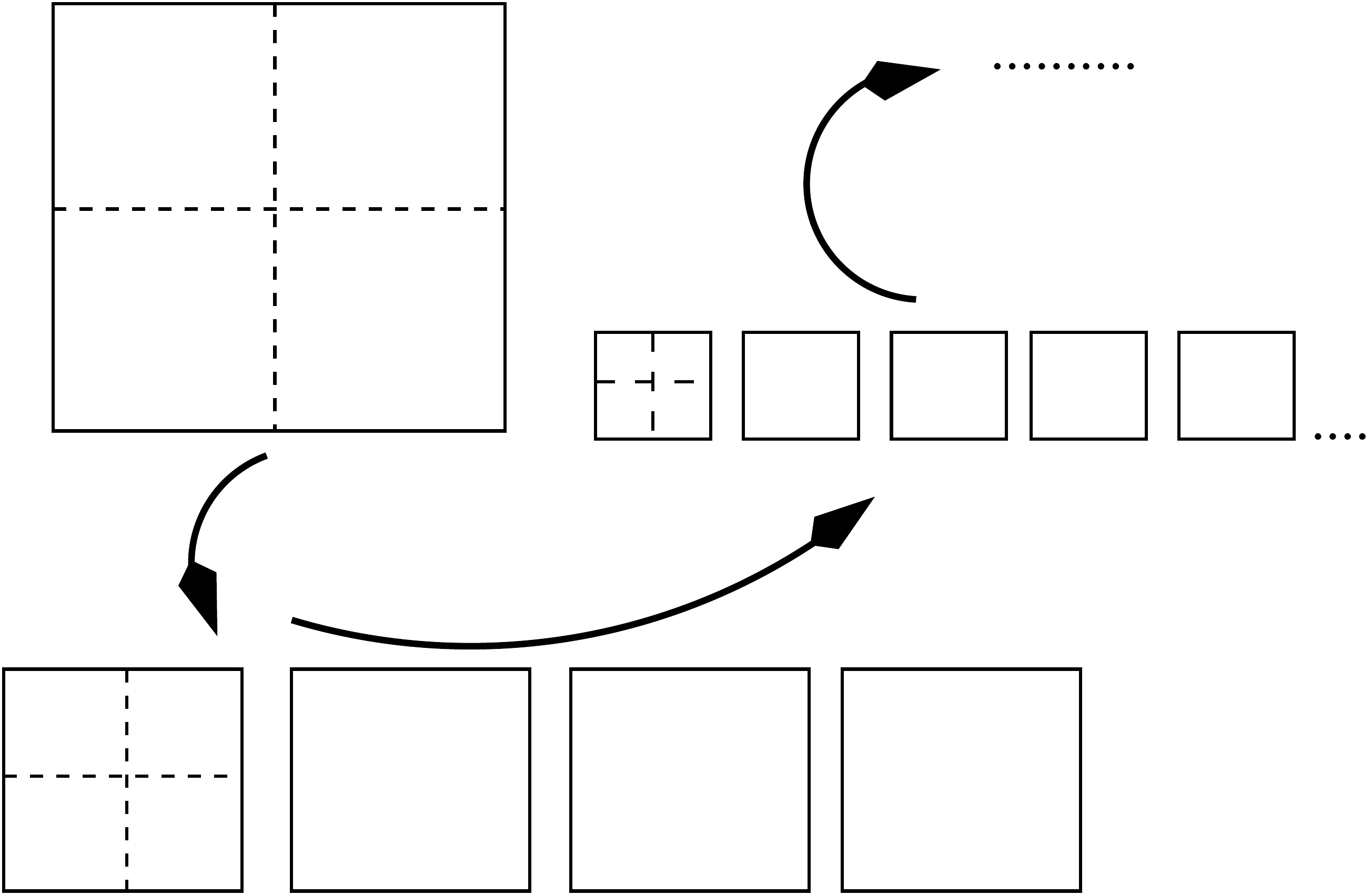}} 
  \caption{Construindo uma inje\c c\~ao boreliana do quadrado para o intervalo.}
  \label{fig:borel_isomorphism}
\end{center}
\end{figure}

Para uma representa\c c\~ao geom\'etrica desta inje\c c\~ao, veja a figura \ref{fig:borel_isomorphism}.

\begin{exercicio}
Mostrar que a aplica\c c\~ao acima \'e injetora e boreliana. 
\par
[ {\em Sugest\~ao:} basta mostrar que as imagens rec\'\i procas dos intervalos abertos formando uma base da topologia do intervalo s\~ao conjuntos borelianos. ]
\end{exercicio}

A aplica\c c\~ao $f$ acima n\~ao \'e sobrejetora, por exemplo o ponto $0.10101010\ldots$ n\~ao \'e na imagem de $f$. Mas ela pode ser modificada a fim de obter um isomorfismo boreliano entre $[0,1]^2$ e $[0,1]$. 
No lugar da base $2$, pode ser uma base qualquer.
Ademais, esta constru\c c\~ao pode ser generalizada para mostrar que n\~ao h\'a muita diversidade entre os espa\c cos borelianos padr\~ao. Eis um resultado cl\'assico (teorema \ref{t:isomorfismo1}).

\begin{teorema}
Dois espa\c cos borelianos padr\~ao s\~ao isomorfos se e somente se eles tem a mesma cardinalidade. 
\qed
\end{teorema}

\begin{corolario}
Sejam $\Omega$ e $W$ dois espa\c cos m\'etricos separ\'aveis e completos, de cardinalidade $\mathfrak c = 2^{\aleph_0}$ cada um. (Por exemplo, isso \'e o caso se eles n\~ao cont\'em os pontos isolados).  Ent\~ao os espa\c cos borelianos correspondentes s\~ao isomorfos.
\qed
\end{corolario}

Este ser\'a o caso da maioria dos dom\'\i nios de interesse na teoria. Por exemplo, o conjunto de Cantor, o intervalo unit\'ario, o espa\c co euclidiano $\R^d$, o espa\c co de Hilbert separ\'avel de dimens\~ao infinita $\ell^2$, e na verdade todos espa\c cos de Fr\'echet (espa\c cos lineares topol\'ogicos localmente convexos, metriz\'aveis e completos) separ\'aveis n\~ao triviais s\~ao todos isomorfos entre eles como espa\c cos borelianos. A estrutura de Borel de todos estes espa\c cos \'e a mesma do {\em espa\c co de Borel padr\~ao} com cardinalidade de cont\'inuo. 

\subsection{}
Agora, sejam $\Omega$ e $\Upsilon$ dois espa\c cos borelianos padr\~ao, e seja $f\colon\Omega\to W$ uma inje\c c\~ao boreliana. Esta $f$ pode ser canonicamente estendida a uma inje\c c\~ao boreliana de $\Omega\times \{0,1\}$ em $\Upsilon\times\{0,1\}$:
\[f(x,\e)=(f(x),\e),\]
onde $\e\in \{0,1\}$. 
Usemos a mesma letra $f$ pela prolonga\c c\~ao. 

Suponha agora que $\mathcal L$ seja um classificador qualquer no dom\'\i nio $\Upsilon$. Definiremos um novo classificador, $\mathcal L^f$, no dom\'\i nio $\Omega$, como a composi\c c\~ao de $\mathcal L$ com a inje\c c\~ao boreliana $f$:
\[{\mathcal L}^f_n(\sigma)(x)={\mathcal L}_n(f(\sigma))(f(x)).\]
(Como na Figura \ref{fig:reducao}).

Seja $\mu$ uma medida de probabilidade qualquer sobre $\Omega\times\{0,1\}$.
Definamos a imagem direta $f_{\ast}\mu$ da medida $\mu$ ao longo de $f$: qualquer seja um conjunto boreliano $B\subseteq \Upsilon\times\{0,1\}$, posemos
\[(f_{\ast}\mu)(B)= \mu(f^{-1}(B)).\]
\'E uma medida de probabilidade boreliana sobre $\Upsilon\times\{0,1\}$. 

\begin{exercicio}
Mostre que se
\[X_1,X_2,\ldots,X_n,\ldots\]
\'e uma sequ\^encia de vari\'aveis aleat\'orias i.i.d. com valores em $\Omega\times\{0,1\}$ segundo a lei $\mu$, ent\~ao 
\[f(X_1),f(X_2),\ldots,f(X_n),\ldots\]
\'e uma sequ\^encia das vari\'aveis aleat\'orias i.i.d. com valores em $\Upsilon\times\{0,1\}$ seguindo a lei $f_{\ast}(\mu)$. 
\end{exercicio}

Definamos $\iota$ a fun\c c\~ao de regress\~ao da medida $f_{\ast}(\mu)$ sobre $\Upsilon\times \{0,1\}$.

\begin{exercicio}
Verifique que
\[\eta = \iota\circ f.\]
\end{exercicio}

Por conseguinte, o classificador de Bayes (``o melhor classificador poss\'\i vel''), $T^{\Omega}_{bayes}$, para $\Omega$, e o classificador de Bayes, $T^{\Upsilon}_{bayes}$, para $W$,  satisfazem:
\[\forall x\in\Omega,~~T^{\Omega}_{bayes}(x) = T^{\Upsilon}_{bayes}(f(x)).\]

\begin{figure}[ht]
\begin{center}
\scalebox{0.25}[0.25]{\includegraphics{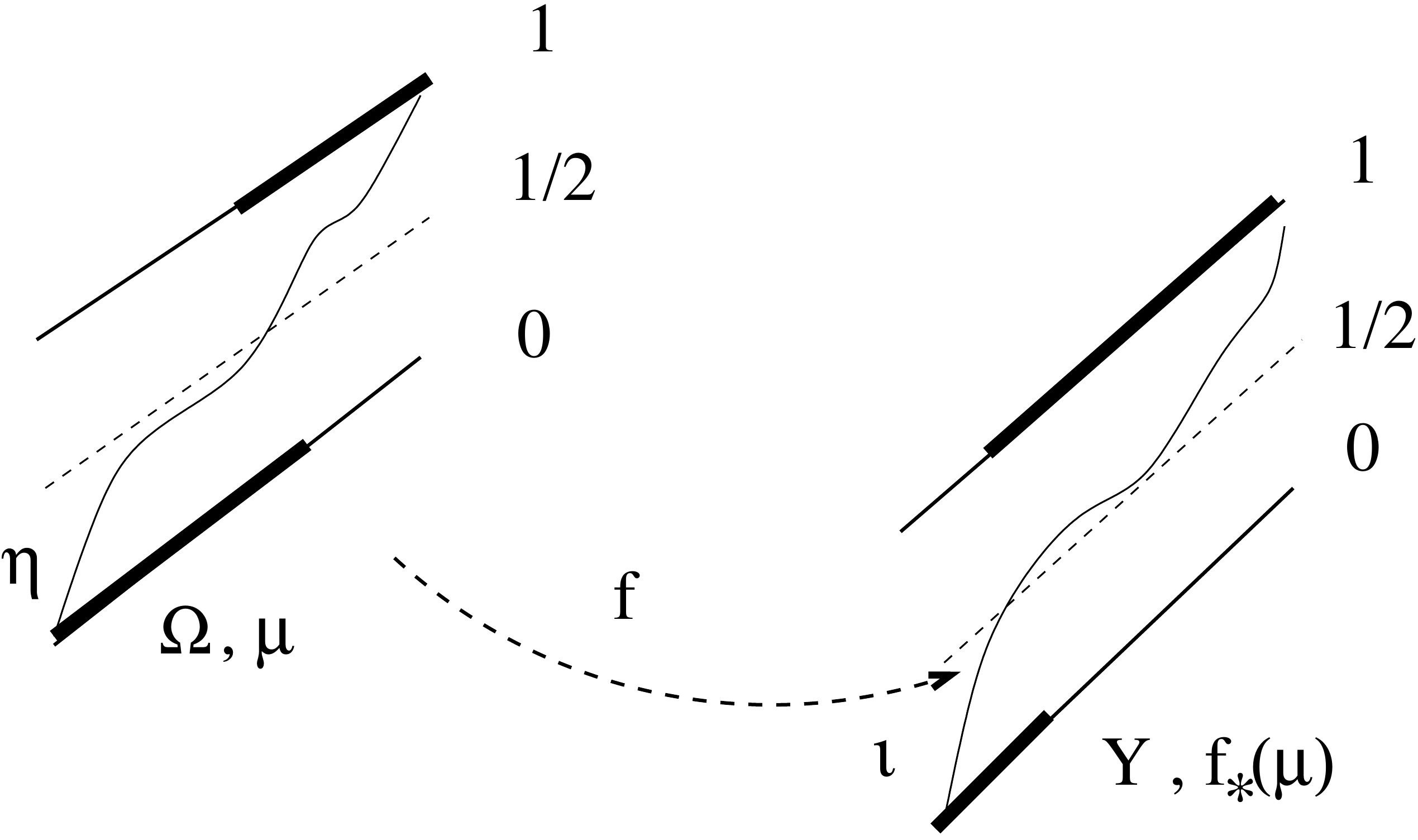}} 
\caption{Gr\'aficos de classificadores de Bayes em dois dom\'\i nios.}
\label{fig:prova}
\end{center}
\end{figure}

\begin{exercicio}
Mostre que, para $\mu^{\otimes n}$-quase todo $\sigma$,
\[\mbox{erro}_{\mu,\eta}{\mathcal L}^f_n(\sigma) = \mbox{erro}_{f_{\ast}(\mu),\theta}{\mathcal L}_n(f(\sigma)).\]
\end{exercicio}

Segue-se imediatamente que qualquer inje\c c\~ao boreliana fornece uma redu\c c\~ao de dimensionalidade que conserva a consist\^encia universal de regras de aprendizagem. O resultado seguinte oferece uma nova perspectiva da redu\c c\~ao de dimensionalidade em teoria da aprendizagem autom\'atica estat\'\i stica.
 
\begin{teorema} 
Sejam $\Omega$ e $\Upsilon$ dois dom\'\i nios (espa\c cos borelianos padr\~ao), e seja $f\colon \Omega\to \Upsilon$ uma inje\c c\~ao boreliana. Seja $\mathcal L$ \'e um classificador universalmente consistente em $\Upsilon$. Ent\~ao o classificador ${\mathcal L}^f$, obtido pela redu\c c\~ao de dimensionalidade $f$ de $\Omega$ para $\Upsilon$, seguida da aplica\c c\~ao do classificador $\mathcal L$, \'e universalmente consistente em $\Omega$ tamb\'em. Se $\mathcal L$ \'e universalmente fortemente consistente em $\Upsilon$ (a saber, o erro de aprendizagem converge para o erro de Bayes quase certamente), ent\~ao ${\mathcal L}^f$ \'e universalmente fortemente consistente em $\Omega$.
\qed
\index{redu\c c\~ao! de dimensionalidade! boreliana}
\end{teorema}

Em particular, pelo menos teoricamente, sempre existe a redu\c c\~ao com uma inje\c c\~ao boreliana do problema de aprendizagem em qualquer dom\'\i nio para o no intervalo, $[0,1]$, ou at\'e o no conjunto de Cantor, com a dimens\~ao topol\'ogica $d=0$. 

\begin{observacao}
Inje\c c\~oes borelianas para redu\c c\~ao de dimensionalidade foram sugeridas  em \citep*{pestov2013}, e aplicadas com sucesso na competi\c c\~ao CDMC'2013 pela equipe consistente do autor destas notas e de tr\^es alunos dele: Ga\"el Giordano, Hubert Duan, e Stan Hatko. O erro de classifica\c c\~ao pelo problema de dete\c c\~ao de intrusos numa rede (uma amostra rotulada em $\R^7$) foi reduzido de $0.3$ at\'e $0.1$ por cento. Veja \citep*{DGHP} para uma descri\c c\~ao de algoritmos mais detalhada, assim como \citep*{hatko:project}.

As nossas experi\^encias at\'e agora mostram que os melhores resultados s\~ao obtidos quando a dimens\~ao \'e reduzida por um fator constante (por exemplo, entre $4$ e $7$), que depende do conjunto de dados.
\end{observacao}

\begin{exercicio}
Escrever o c\'odigo em R para redu\c c\~ao de dimensionalidade usando as inje\c c\~oes borelianas, e combin\'a-lo com o classificador $k$-NN para melhorar o erro de classifica\c c\~ao no problema do reconhecimento de voz (o conjunto de dados {\em Phoneme}). Tentar as bases diferentes de expans\~ao dos n\'umeros.
\end{exercicio}

\begin{observacao}
A teoria est\'a no seu in\'\i cio. O que \'e necess\'ario do ponto de vista te\'orico, \'e desenvolver alguma no\c c\~ao de capacidade, talvez no esp\'\i rito de dimens\~ao de Vapnik--Chervonenkis ou m\'edias de Rademacher, para uma fam\'\i lia $\mathscr F$ de inje\c c\~oes de Borel $f$ entre dois dom\'\i nios, $\Omega$ e $W$. Quando esta no\c c\~ao for finita, isto garantiria que o classificador ${\mathcal L}^f$ seja universalmente consistente, quando $f$ \'e selecionado dentro de $\mathscr F$ dependendo da amostra $\sigma$, a fim de optimizar o desempenho emp\'\i rico de ${\mathcal L}^f$. Em outras palavras, $f$ \'e um elemento aleat\'orio de $\mathscr F$.

A perspetiva mais promissora ser\'a de combinar a redu\c c\~ao boreliana com aprendizagem ensemble.
\end{observacao}

%
%

\chapter{Aproxima\c c\~ao universal\label{ch:aproximacao}}

O perceptron produz uma cole\c c\~ao de conceitos realmente simples, eles s\~ao semiespa\c cos, e suas habilidades de classifica\c c\~ao s\~ao claramente muito limitadas. O que ocorre se permitirmos redes de arquitetura mais complexa? Qu\~ao complicada \'e poss\'\i vel que seja uma classe de fun\c c\~oes gerada por uma rede neural?  Podemos esperar gerar todos os conceitos?
A resposta \'e positiva: que se juntarmos apenas alguns unidades de computa\c c\~ao, podemos aproximar cada conceito (ou at\'e cada fun\c c\~ao) que queremos. 

O primeiro resultado deste cap\'\i tulo \'e o teorema cl\'assico de Kolmogorov obtido em resposta a um dos problemas na famosa lista de David Hilbert, logo antes da emerg\^encia da aprendizagem autom\'atica. O resultado pode ser interpretado assim: para cada $d$ existe uma rede fixa de unidades de computa\c c\~ao que, para v\'arios valores de par\^ametros, gera todas as fun\c c\~oes cont\'\i nuas de $d$ vari\'aveis. 

No entanto, o significado deste resultado \'e sobretudo metaf\'\i sico. O resultado ``p\'e no ch\~ao'' nesta dire\c c\~ao \' o Teorema de Aproxima\c c\~ao Universal de Cybenko, afirmando, em particular, que j\'a uma rede de perceptrons sem realimenta\c c\~ao que tem apenas uma camada interior pode aproximar qualquer conceito boreliano se o n\'umero de unidades \'e bastante grande (aqui, ao contr\'ario ao teorema de Kolmogorov, o tamanho da rede depende da fun\c c\~ao e de grau de aproxima\c c\~ao desejado). 

Finalmente, discutimos o princ\'\i pio de minimiza\c c\~ao de risco estrutural que combina algumas vantagens da teoria de Vapnik--Chervonenkis com as de consist\^encia universal.

\section{Problema 13 de Hilbert e teorema da superposi\c c\~ao de Kolmogorov}

\subsection{}
Em sua palestra plen\'aria no 1900 Congresso Internacional de Matem\'aticos (ICM), realizado em Paris, David Hilbert formulou uma lista de 23 problemas abertos que tiveram, ao todo, um profundo impacto sobre os desenvolvimentos matem\'aticos no s\'eculo XX. (Um livro altamente recomendado \'e \citep*{yandell}). O d\'ecimo terceiro da lista foi o problema seguinte.

\begin{quote}
{\small
Nomography deals with the problem: to solve equations by means of drawings of families of curves depending on an arbitrary parameter. It is seen at once that every root of an equation whose coefficients depend upon only two parameters, that is, every function of two independent variables, can be represented in manifold ways according to the principle lying at the foundation of nomography. Further, a large class of functions of three or more variables can evidently be represented by this principle alone without the use of variable elements, namely all those which can be generated by forming first a function of two arguments, then equating each of these arguments to a function of two arguments, next replacing each of those arguments in their turn by a function of two arguments, and so on, regarding as admissible any finite number of insertions of functions of two arguments. So, for example, every rational function of any number of arguments belongs to this class of functions constructed by nomographic tables; for it can be generated by the processes of addition, subtraction, multiplication and division and each of these processes produces a function of only two arguments. One sees easily that the roots of all equations which are solvable by radicals in the natural realm of rationality belong to this class of functions; for here the extraction of roots is adjoined to the four arithmetical operations and this, indeed, presents a function of one argument only. Likewise the general equations of the 5-th and 6-th degrees are solvable by suitable nomographic tables; for, by means of Tschirnhausen transformations, which require only extraction of roots, they can be reduced to a form where the coefficients depend upon two parameters only.
\\
Now it is probable that the root of the equation of the seventh degree is a function of its coefficients which does not belong to this class of functions capable of nomographic construction, i. e., that it cannot be constructed by a finite number of insertions of functions of two arguments. In order to prove this, the proof would be necessary that the equation of the seventh degree $f^7 + xf^3 + yf^2 + zf + 1 = 0$ is not solvable with the help of any continuous functions of only two arguments. I may be allowed to add that I have satisfied myself by a rigorous process that there exist analytical functions of three arguments $x, y, z$ which cannot be obtained by a finite chain of functions of only two arguments.
\\
By employing auxiliary movable elements, nomography succeeds in constructing functions of more than two arguments, as d'Ocagne has recently proved in the case of the equation of the 7-th degree.}
\end{quote}

A ess\^encia do problema 13 era ent\~ao comumente interpretada como a conjectura de que nem todas as fun\c c\~oes cont\'\i nuas em tr\^es vari\'aveis podem ser expressas por meio de composi\c c\~oes finitas de fun\c c\~oes cont\'\i nuas em duas vari\'aveis.

Em 1957, Vladimir Arnold, ent\~ao um aluno de gradua\c c\~ao de 3o ano e depois um dos mais influentes matem\'aticos russos, mostrou que, na verdade, a conjectura de Hilbert era falsa: cada fun\c c\~ao cont\'\i nua de tr\^es vari\'aveis pode ser expressa usando um n\'umero finito de substitui\c c\~oes de fun\c c\~oes cont\'\i nuas de duas vari\'aveis.

Desenvolvendo a prova de Arnold ainda mais, seu orientador de pesquisa Andre\u\i\ Kolmogorov estabeleceu \citep*{kolmogorov} o resultado seguinte, que se tornou cl\'assico.

\begin{teorema}[Teorema da Superposi\c c\~ao de Kolmogorov]
Para todo $n\geq 2$ existem fun\c c\~oes cont\'\i nuas $\psi^{pq}$, $p=1,2,\ldots,n$, $q=1,2,\ldots,2n+1$, definidas sobre $[0,1]$, tais que cada fun\c c\~ao cont\'\i nua $f\colon [0,1]^n\to\R$ pode ser representada assim:
\begin{equation}
f(x_1,x_2,\ldots,x_n)=\sum_{q=1}^{2n+1}g_q\left[\sum_{p=1}^n\psi^{pq}(x_p)\right],
\end{equation}
onde $g_q$, $q=1,2,\ldots,2n+1$ s\~ao fun\c c\~oes reais cont\'\i nuas, que dependem de $f$.
\index{teorema! da superposi\c c\~ao! de Kolmogorov}
\end{teorema}

Desde ent\~ao, o resultado tem atra\'\i do uma aten\c c\~ao consider\'avel e foi melhorado muitas vezes. Em particular, aqui est\'a uma forma dada por G. Lorentz \citep*{lorentz}.

\begin{teorema}
Seja $f\colon [0,1]^d\to\R$ uma fun\c c\~ao cont\'\i nua, e $\e>0$ um n\'umero racional. Ent\~ao
\begin{equation}
f=\sum_{i=1}^{2d+1} g(z_k),
\end{equation}
onde $g\colon\R\to\R$ \'e uma fun\c c\~ao cont\'\i nua apropriada (dependendo de $f$ e de $\e$), e para todo $k$,
\begin{equation}
z_k =\sum_{j=1}^d \lambda^k\psi(x_j+\e k)+k.
\end{equation}
Aqui, $\lambda\in\R$, $\psi$ \'e uma fun\c c\~ao real Lipschitz cont\'\i nua  crescente, e $\lambda,\psi$ n\~ao dependem de $f$ e $\e$. \qed
\index{teorema! da superposi\c c\~ao! de Kolmogorov--Lorentz}
\end{teorema}

O teorema acima pode ser interpretado para dizer que uma rede neural artificial relativamente simples (a ``rede neural de Kolmogorov--Lorentz'') pode gerar uma fun\c c\~ao cont\'\i nua qualquer em $d$ vari\'aveis:

\begin{figure}[ht]
 \begin{center}
\scalebox{0.45}{\includegraphics{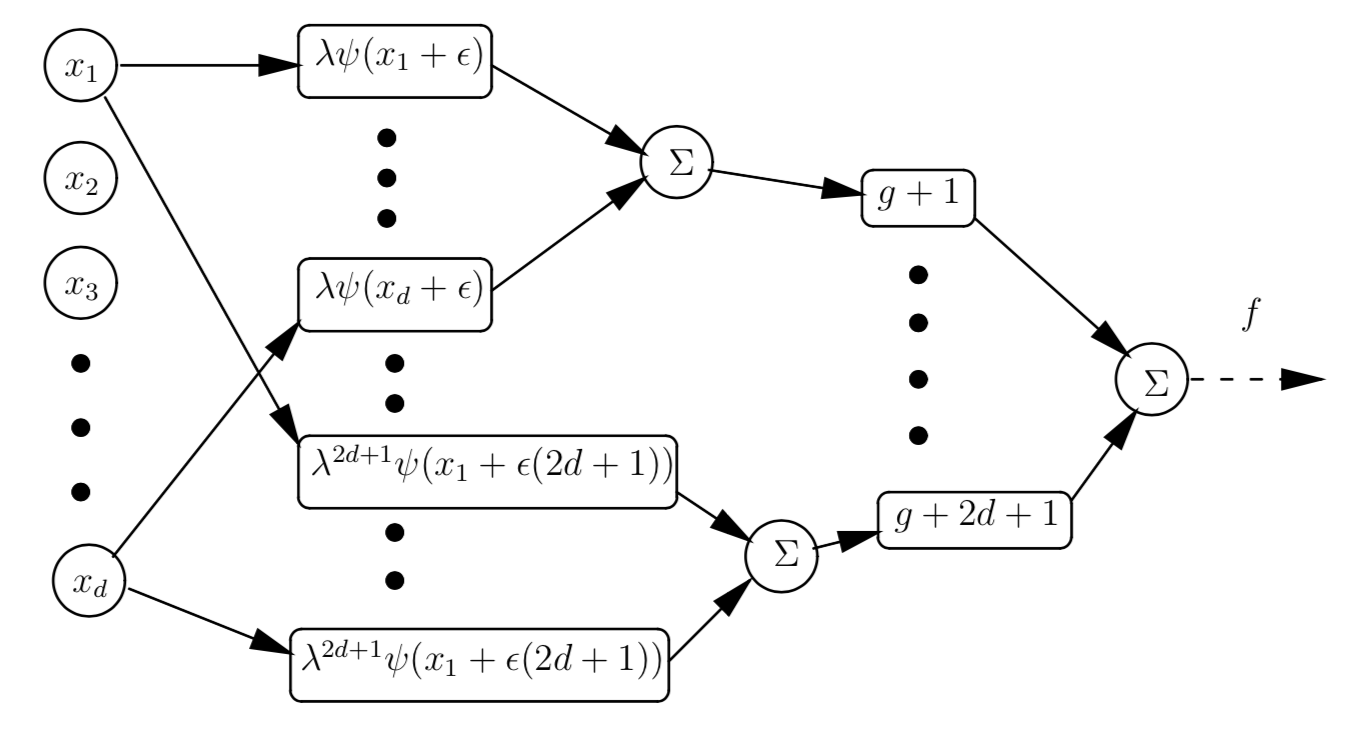}} 
\caption{A ``rede neural de Kolmogorov--Lorentz''.}
\end{center}
\end{figure}

O \'unico ``par\^ametro'' da rede a ser treinado \'e a fun\c c\~ao $g$ de uma vari\'avel. Em seu tempo a ideia acima atraiu o interesse de tais pesquisadores na comunidade da aprendizagem da m\'aquina como Robert Hecht-Nielsen \citep*{hecht-nielsen}. Houve tamb\'em publica\c c\~oes
argumentando que a rede de Kolmogorov--Lorentz n\~ao pode possivelmente ter algum valor pr\'atico \citep*{girosi}, e pelo contr\'ario, alguns resultados mis recentes \citep*{kurkova,igelnik,braun-griebel} demonstram que as fun\c c\~oes $\psi$, $\lambda$ e $g$ podem ser recuperadas de forma algor\'\i tmica eficiente. 

Apesar de tudo isso, a observa\c c\~ao acima tem principalmente um significado filos\'ofico: \'e tranquilizador estar ciente do fato que as redes neurais artificiais, em teoria pelo menos, s\~ao onipotentes. O teorema de Kolmogorov \'e assim mencionado em muitos fontes que tratam de redes neurais.

\subsection{}
Na sua revista \citep*{lorentz} do problema 13 de Hilbert, G.G. Lorentz oferece a perspetiva seguinte do teorema de Kolmogorov.

\begin{quote}
{\small
Let us take a good Calculus text and examine what genuine functions of two or three variables can we find. The function $x+y$ will be found, but the product $xy$ is already nothing new, it reduces to the sum and functions of one variable, $xy=e^{\log x+\log y}$, similarly for $x+y+z=(x+y)+z$. So perhaps there do not exist any other functions?\\  The astonishing result of Kolmogorov (1957) confirms this.}
\end{quote}

Nesta se\c c\~ao vamos apresentar a prova do Teorema de Superposi\c c\~ao de Kolmogorov devida a Jean-Pierre Kahane \citep*{kahane}, que \'e provavelmente a mais curta prova conhecida. Embora possa ser a menos construtiva, por\'em todas as ideias originais s\~ao conservadas e podem ser vistas claramente, e as tecnicalidades s\~ao reduzidas ao m\'\i nimo dos m\'\i nimos. 

\subsection{Estendendo as fun\c c\~oes Lipschitz cont\'\i nuas}

\begin{lemma} Seja $f$ uma fun\c c\~ao real Lipschitz cont\'\i nua de constante $L\geq 0$, definida sobre um subespa\c co $A$ de um espa\c co m\'etrico $(X,d)$. Ent\~ao $f$ admite uma extens\~ao, $\tilde f$, sobre todo $X$, que \'e Lipschitz cont\'\i nua da mesma constante $L$.
\label{l:extensaoL}
\end{lemma}

\begin{figure}[ht]
\begin{center}
\scalebox{0.65}[0.65]{\includegraphics{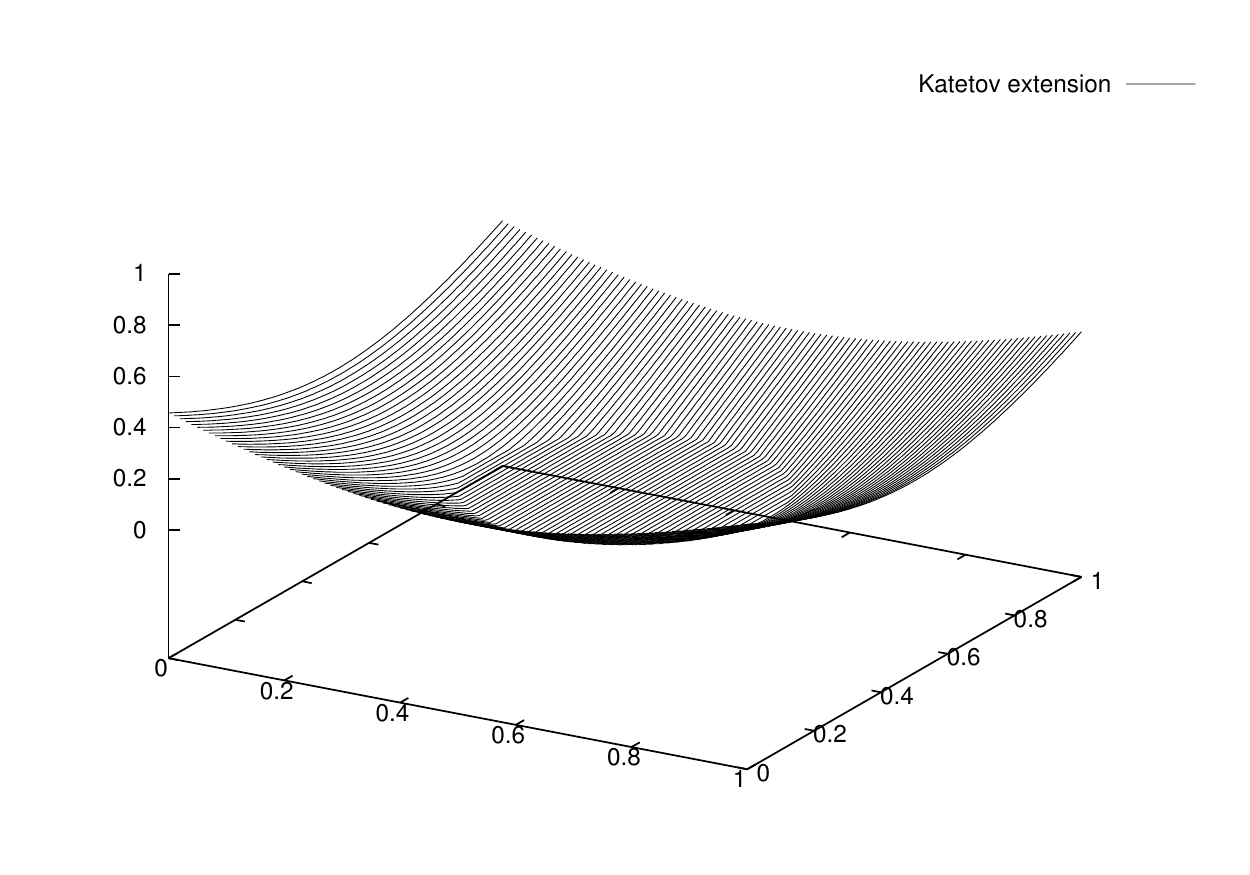}} 
\caption{Extens\~ao $\tilde f$ de uma fun\c c\~ao identicamente zero sobre o disco de raio $1/4$ em torno do meio do quadrado unit\'ario.}
\label{fig:katetext}
\end{center}
\end{figure}

\begin{proof}
Sem perda de generalidade e substituindo $f$ por $f/L$ se for necess\'ario, podemos supor que $f$ \'e $1$-Lipschitz cont\'\i nua. 
Al\'em disso, suponhamos que $A$ n\~ao \'e vazio.
Para todo $x\in X$, definamos
\begin{equation}
\tilde f(x) =\inf\left\{f(a)+d(a,x)\colon a\in A\right\}.
\label{eq:katetov}
\end{equation}

O valor $\tilde f(x)$ \'e bem-definido: escolhendo um $a_0\in A$ qualquer, a desigualdade triangular e a condi\c c\~ao $L=1$ implicam que 
\[\forall a\in A,~~f(a)+d(a,x)\geq f(a_0)-d(x,a_0).\]
A segunda observa\c c\~ao \'e que se $x\in A$, ent\~ao para todo $a\in A$, pela mesma raz\~ao que acima,
\[f(a)+d(a,x)\geq f(x),\]
ou seja, o \'\i nfimo na Eq. (\ref{eq:katetov}) \'e atingido em $a=x$, \'e por isso a fun\c c\~ao $\tilde f$ \'e uma extens\~ao de $f$. 

Basta mostrar que $\tilde f$ \'e $1$-Lipschitz cont\'\i nua. Sejam $x,z\in \Omega$ e $\e>0$ quaisquer. Escolha $y_1,y_2\in Y$ de modo que o valor de $f(x)$, respetivamente $f(z)$, seja a dist\^ncia $<\e$ do valor $f(y_1)+d(x,y_1)$, respetivamente $f(y_2)+d(z,y_2)$. Ent\~ao $\abs{f(x)-f(z)}$ \'e a menor de $2\e$ de 
$\abs{f(y_1)+d(x,y_1)-f(y_2)-d(z,y_2)}$, e al\'em disso,
\[f(y_1)+d(x,y_1)\leq f(y_2)+d(x,y_2)+\e,\]
embora, como \'e \'obvio,
\[d(x,y_2)\leq d(x,z)+d(z,y_2).\]
Por conseguinte,
\[f(y_1)+d(x,y_1)\leq f(y_2)+d(x,z)+d(z,y_2)+\e,\]
ou seja
\[f(y_1)+d(x,y_1)-f(y_2)-d(z,y_2)\leq d(x,z)+\e.\]
Isso implica
\[f(x)-f(z)\leq d(x,z)+3\e\]
e como $\e>0$ foi qualquer, enviando $\e\downarrow 0$,
\[f(x)-f(z)\leq d(x,z).\]
Um argumento sim\'etrico mostra que
\[f(z)-f(x)\leq d(x,z).\]
\end{proof}

\begin{observacao} 
A fun\c c\~ao $\tilde f$ se chama {\em extens\~ao de Kat\'etov}.
\'E claro que a extens\~ao $\tilde f$ \'e, em geral, longe de ser \'unica. A fun\c c\~ao estendida, $\tilde f$, \'e a {\em maior} de todas as extens\~oes Lipschitz cont\'\i nuas de $f$ tendo a constante Lipschitz determinada. (Exerc\'\i cio.) Considere tamb\'em Fig. \ref{fig:katetext}, onde essa extens\~ao \'e grafada para a fun\c c\~ao $f$ tomando identicamente o valor zero no c\'\i rculo $(x-0.5)^2 + (y-0.5)^2 = 1/16$. Os valores da fun\c c\~ao $\tilde f$ est\~ao subindo na taxa mais r\'apida permitida pela constante de Lipschitz $1$.

Pode-se, por exemplo, considerar em vez disso a {\em menor} extens\~ao poss\'\i vel, dada por
\[\bar f(x) =\sup\left\{f(a)-d(a,x)\colon a\in A\right\}.\]
\end{observacao}

\subsection{A prova de Kahane do teorema da superposi\c c\~ao de Kolmogorov}
\subsubsection{Formula\c c\~ao do teorema dada por Kahane}
Denotemos $\Phi$ o conjunto de todas as fun\c c\~oes cont\'\i nuas n\~ao decrescentes do intervalo fechado $\I=[0,1]$ nele mesmo, visto como um subconjunto de $C[0,1]$.

\begin{exercicio} Mostrar que $\Phi$ \'e fechado em $C[0,1]$, ent\~ao \'e um espa\c co m\'etrico completo relativo \`a m\'etrica uniforme.
\end{exercicio}

\begin{exercicio}
Mostrar que uma {\em fun\c c\~ao gen\'erica} $\varphi\in\Phi$ (veja subse\c c\~ao \ref{ss:genericidade}) \'e {\em estritamente crescente}. 
\par
[ {\em Sugest\~ao:} basta verificar que se $x,y\in\Q\cap [0,1]$ e $x<y$, ent\~ao a propriedade $\phi(x)<\phi(y)$ \'e gen\'erica. ]
\end{exercicio}

Para todo inteiro $k$, a pot\^encia $k$-\'esima $\Phi^k$ \'e um espa\c co m\'etrico completo quando munido da dist\^ancia 
\[d((\varphi_1,\ldots,\varphi_n),(\psi_1,\psi_2,\ldots,\psi_n))=\max_{i=1}^n
\norm{\varphi_i-\psi_i},\]
e assim pode-se falar de $k$-tuplos gen\'ericos de elementos de $\Phi$.

\begin{teorema}[Teorema da superposi\c c\~ao de Kolmogorov, vers\~ao de Kahane]
Sejam $\lambda_1,\lambda_2$, $\ldots,\lambda_n$ reais estritamente positivos, dois a dois distintos, de soma um. Ent\~ao, para uma $(2n+1)$-tupla gen\'erica $(\varphi_1,\varphi_2,\ldots,\varphi_{2n+1})\in \Psi^{2n+1}$, toda fun\c c\~ao cont\'\i nua real $f_0$ sobre o cubo unit\'ario $\I^n$ pode ser representada como
\begin{equation}
f_0(x_1,\ldots,x_n)=\sum_{q=1}^{2n+1} g\left( \sum_{p=1}^n\lambda_p\varphi_q(x_p)\right),
\label{eq:kolmogorov}
\end{equation}
onde $g$ \'e uma fun\c c\~ao cont\'\i nua sobre o intervalo $\I$ (que depende de $f$).
\index{teorema! da superposi\c c\~ao! vers\~ao de Kahane}
\end{teorema}

\subsubsection{}
Seja $\e>0$ um n\'umero bastante pequeno, a ser especificado mais tarde. Para cada $f\in C[0,1]^n$ n\~ao identicamente igual a zero, definamos o conjunto $\Omega(f)$ de todas as $(2n+1)$-tuplas $(\varphi_1,\varphi_2,\ldots,\varphi_{2n+1})$ de fun\c c\~oes de $\Phi$ tais que existe um $h\in C[0,1]$ tendo as propriedades:
\begin{align}
(i) & \norm h\leq \norm f,\nonumber \\
(ii) & \left\Vert f(x_1,x_2,\ldots,x_n) -\sum_{q=1}^{2n+1}h\left(\sum_{p=1}^n
\lambda_p\varphi_q (x_p)\right)\right\Vert < (1-\e)\norm f.
\label{eq:e}
\end{align}
(As normas s\~ao normas uniformes em $C[0,1]$ e em $C(\I^n)$, respectivamente). 

\subsubsection{}
\'E claro que $\Omega(f)$ \'e aberto em $\Phi^{2n+1}$. De fato, a norma da express\~ao na parte esquerda em $(ii)$ \'e estritamente menor que $(1-\e)\norm f$; denote a diferen\c ca entre dois valores da norma por $\gamma$. Se o valor de $h$ em todas $2n+1$ express\~oes dentro dos par\^enteses muda por menos de $\gamma/(2n+1)$, a norma da diferen\c ca muda por $<\gamma$, e a desigualdade vai ser v\'alida. Como $h$ \'e cont\'\i nua sobre o interval fechado unit\'ario, ela \'e uniformemente cont\'\i nua. Pode-se escolher um $\gamma_1>0$ com a propriedade que se dois valores do argumento de $h$ diferem por menos de $\gamma_1$, ent\~ao os valores correspondentes de $h$ diferem por $<\gamma/(2n+1)$. Concentremo-nos sobre a soma de $n$ termos dentro do argumento de $h$. Se todo termo muda por $<\gamma_1/n$, ent\~ao o argumento de $h$ muda por $<\gamma_1/n$. Todo $\lambda_p$ \'e limitado por um, logo basta verificar que os valores de cada $\varphi_q$ mudam por $<\gamma_1/n$. Mas isso \'e o caso se for permitido que a $(2n+1)$-tupla $(\varphi_q)$ varie dentro da bola aberta de raio $\gamma_1/n$ em torno do valor original. 

\subsubsection{}
Pretendamos mostrar que o conjunto $\Omega(f)$ \'e denso em $\Phi^{2n+1}$. Suponha, por enquanto, que isso foi estabelecido. A prova do teorema segue assim.

Como o espa\c co $C[0,1]^n$ \'e separ\'avel (proposi\c c\~ao \ref{p:c(x)separavel}), ele cont\'em um subconjunto enumer\'avel denso, $F$. Pode-se ainda supor que $0\notin F$. De acordo com o teorema da categoria de Baire \ref{t:baireformaequiv}, o conjunto $\Omega = \cap_{f\in f}\Omega (f)$ \'e um conjunto $G_\delta $ denso em $\Phi^{2n+1}$. Logo, um elemento gen\'erico $(\varphi_1,\varphi_2,\ldots,\varphi_{2n+1})$ de $\Phi^{2n+1}$ pertence a $\Omega$. 

\begin{exercicio} 
Mostrar que sob nossas hip\'oteses, dado $f_0\in C(\I^n)$, existe $f\in F$ com as propriedades $\norm f\leq\norm{f_0}$ e $\norm{f-f_0}<(\e/2)\norm{f_0}$. 
\end{exercicio}

Escolha uma fun\c c\~ao $f\in F$ como acima. Seja $(\varphi_1,\varphi_2,\ldots,\varphi_{2n+1})$ uma tupla qualquer de elementos de $\Omega$.
Agora escolhamos uma fun\c c\~ao $h$ verificando as condi\c c\~oes (\ref{eq:e}). N\'os iremos expressar este fato simplesmente escrevendo $h=\gamma(f_0)$. Define tamb\'em $\gamma(0)=0$. 

Agora definamos recursivamente para $j=0,1,\ldots$, as fun\c c\~oes $h_j=\gamma(f_j)$, onde
\[f_{j+1}(x_1,x_2,\ldots,x_n)=f_j(x_1,x_2,\ldots,x_n)-\sum_{q=1}^{2n+1}h_j
\left(\sum_{p=1}^n
\lambda_p\varphi_q (x_p)\right).\]

\begin{exercicio}
Verifique que a s\'erie de fun\c c\~oes $\sum_{j=0}^\infty h_j$ converge (at\'e mesmo absolutamente) no espa\c co normado $C[0,1]$. 
\end{exercicio}

\begin{exercicio}
Verifique que a soma da s\'erie acima, $h=\sum_{j=0}^\infty h_j$ (que \'e tamb\'em uma fun\c c\~ao cont\'\i nua), satisfaz a conclus\~ao do teorema de Kolmogorov (\ref{eq:kolmogorov}). \par
[ {\em Sugest\~ao:} use as somas telesc\'opicas. ]
\end{exercicio}

Agora concentremos nossos esfor\c cos para estabelecer que $\Omega(f)$ \'e denso em $\Phi^{2n+1}$.

\subsubsection{} Seja $G$ um subconjunto aberto n\~ao vazio qualquer de $\Phi^{2n+1}$, e seja $\delta=\delta(G,F,\e)$ um valor positivo bastante pequeno, a ser especificado quase no final da prova.

Para todo $q=1,2,\ldots,2n+1$ considere a sequ\^encia de intervalos fechados dois a dois disjuntos $\I_q(j)$, $j\in\Z$, de comprimento $2n\delta$ cada um, separados entre si por sucessivas lacunas de comprimento $\delta$. Eis a f\'ormula:
\[\I_q=\I_q(j)=[q\delta+(2n+1)j\delta,~q\delta+(2n+1)j\delta+2n\delta].\]
Observe que, como o valor de $q$ aumenta por um, todos os intervalos $\I_q$ mover para a direita por $\delta$ unidades, de modo que depois de $2n+1$ passos o arranjo de intervalos $\I_q$ novamente parece como no in\'\i cio.

\begin{figure}[ht]
\begin{center}
\scalebox{0.25}{\includegraphics{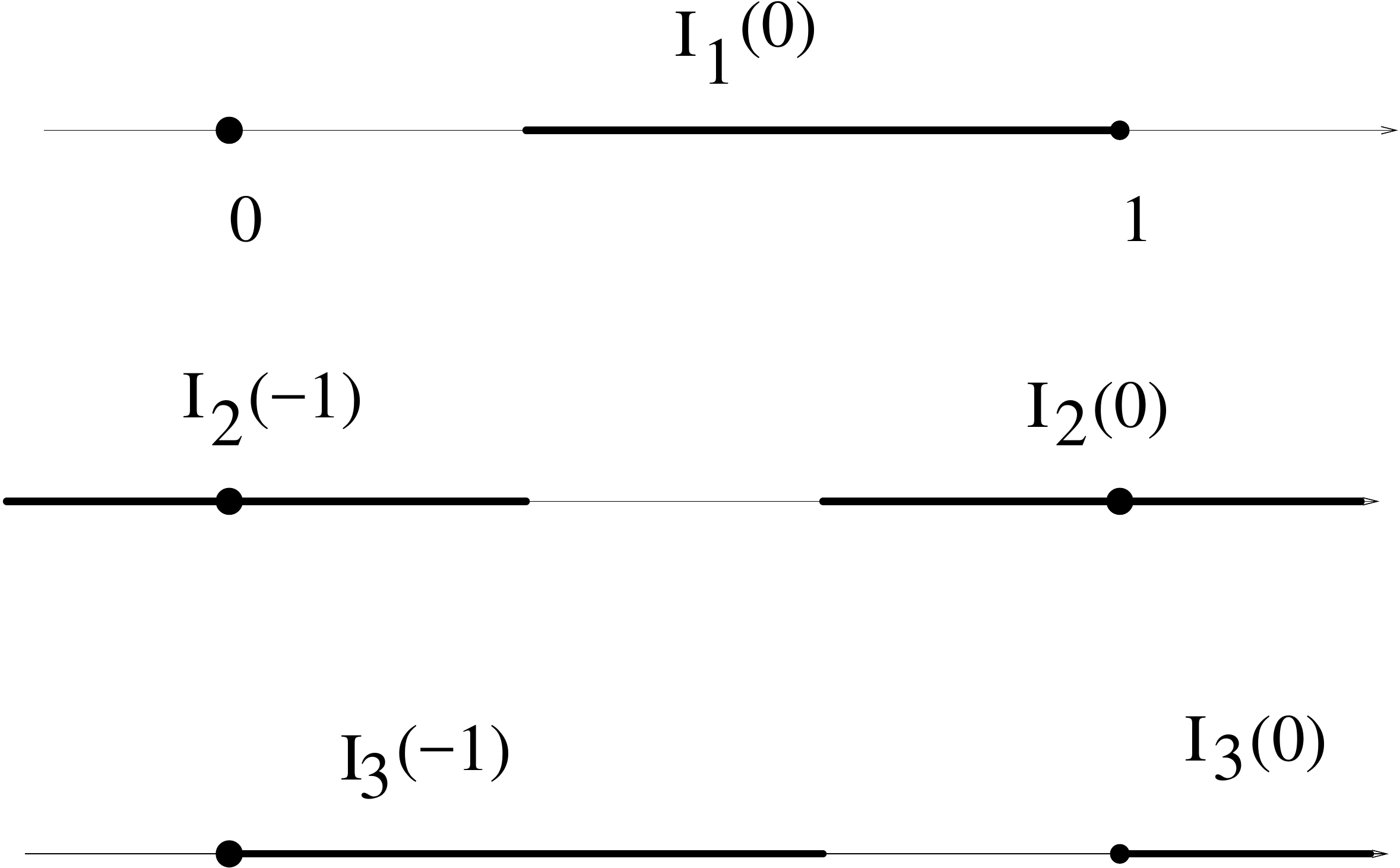}} 
\caption{Intervalos $\I_q(j)$ no caso mais simples onde $\delta=1/3$ e $n=1$, de modo que $2n+1=3$.}
\label{fig:katetext2}
\end{center}
\end{figure}

Por conseguinte, todo ponto $x\in [0,1]$ pertence \`a uni\~ao de todos os intervalos $\I_q$ para todos os valores de $q$ exceto no m\'aximo um.

Denote por $P_q$ o cubo da forma
\[P_q=P_q(j_1,j_2,\ldots,j_n)=\I_q(j_1)\times
\I_q(j_2)\times\ldots\times\I_q(j_n).\]
Se $x\in\I^n$, ent\~ao toda coordenada $x_i$ pertence a um dos intervalos $\I_q$ para todos $q$ exceto no m\'aximo um valor. Por conseguinte, $x$ pertence \`a uni\~ao de todos os cubos $P_q$ para todos os valores $q=1,2,\ldots,2n+1$ exceto no m\'aximo $n$ valores (o n\'umero de coordenadas do cubo). Ent\~ao, $x$ pertence \`a uni\~ao de cubos $P_q$ para pelo menos $n+1$ valores diferentes de $q$. 

\begin{observacao}
Este \'e um ponto importante que cada elemento do cubo $\I^n$ pertence aos cubos pequenos ``mais frequentemente do que n\~ao'', e este ponto sutil vai cumprir um papel crucial na estimativa final da prova. Isso explica o n\'umero de fun\c c\~oes, $2n + 1$, na afirma\c c\~ao do teorema. Foi mostrado (veja por exemplo \citep*{sternfeld}), que o n\'umero $2n + 1$ das fun\c c\~oes $\varphi_q$ \'e de fato o m\'\i nimo poss\'\i vel e n\~ao pode ser melhorado.
\end{observacao}

\subsubsection{} Agora denote por $\Delta$ a cole\c c\~ao de todas as $(2n+1)$-tuplas de fun\c c\~oes em $\Phi$ tais que $\varphi_q$ \'e constante sobre os intervalos $\I_q$ e afim entre eles. Escolha $\delta$ de modo que
\begin{enumerate}
\item a oscila\c c\~ao de $f$ sobre todo cubo $P_q$ \'e limitada por $\e\norm f$ (isso \'e certamente poss\'\i vel, pois $f$ \'e uniformemente cont\'\i nua sobre o cubo $\I^n$), e
\item $G\cap \Delta=\emptyset$. 
Observe que, como o tamanho dos intervalos de $\I_q$ se aproxima de zero, as fun\c c\~oes constantes sobre estes intervalos e afins entre eles se tornarem cada vez mais flex\'\i veis e podem aproximar as fun\c c\~oes cont\'\i nuas dadas cada vez melhor. Se n\'os escolhermos qualquer $(2n + 1)$-tupla de fun\c c\~oes $\varphi_1, \ldots, \varphi_{2n + 1}$ de $G$, e um $\gamma > 0$ de tal modo que o $\gamma $-bola em torno deste $ (2n + 1) $-tupla est\'a contida em $G$, ent\~ao, escolhendo o tamanho dos intervalos de $\I_q$ bastante pequeno, podemos garantir que existem $(2n + 1)$-tuplas de fun\c c\~oes de $\Delta$ que est\~ao $\gamma$-pr\'oximas de $\varphi_1, \ldots, \varphi_{2n + 1}$.
\end{enumerate}

\begin{exercicio}
Mostre que se pode escolher de modo recursivo uma sequ\^encia de n\'umeros suficientemente pequenos $\mu_{q,j}$ de modo que, adicionando $\mu_i$ ao valor de todo $\varphi_q$ sobre o $j$-\'esimo cubo, obtemos que as fun\c c\~oes
\[\chi_q(x_1,x_2,\ldots,x_n)=\sum_{p=1}^n\lambda_p \phi_1(x_p),~~q=1,2,3,\ldots,2n+1\]
assumem valores dois a dois diferentes sobre todos os cubos diferentes, $P_q(j_1,\ldots,j_n)$ e $P_{q'}(j_1',j_2',\ldots,j_n')$. Em outras palavras, a aplica\c c\~ao
\[(q,j_1,j_2,\ldots,j_n)\mapsto \chi_q(P_q(j_1,\ldots,j_n))\]
\'e injetora.
\par
[ {\em Sugest\~ao:} escolher todos os $\mu_{q,j}$ linearmente independentes sobre o corpo gerado por todos os $\lambda_i$ e, em seguida, usar o fato de que $\lambda_i$ s\~ao todos dois a dois distintos. ]
\end{exercicio}

\subsection{} Denote $M(P_q)$ o valor m\'edio de $f$ sobre o cubo $P_q=P_q(j_1,\ldots,j_n)$, ou seja,
\[M(P_q) =\int_{P_q}f(x)~dx.\]
Para todos os cubos, defina
\[h(\chi_q(P_q))=2\e M(P_q).\]
Lembre-se de que o valor da fun\c c\~ao $\chi_q$ \'e constante sobre cada cubo $P_q$, e n\~ao h\'a dois valores iguais entre si. Claramente, o valor m\'aximo de $h$ nos cubos \'e limitado por $2\e\norm f$.

Estenda esse $h$ sobre $\R$ de modo que a norma de $h$ \'e limitada pela mesma quantidade $2\e\norm f$:
\begin{equation}
\norm h\leq 2\e\norm f.
\label{eq:coarse}
\end{equation}
Pode-se, por exemplo, usar o lema \ref{l:extensaoL} e cortar os grandes valores da fun\c c\~ao resultante usando $\max\{-,2\e\norm f \}$ e $\min\{-,2\e\norm f\}$).

Para todo valor $x\in P_q$, pode-se escrever, segundo a defini\c c\~ao de $h$,
\begin{equation}
\label{eq:fine}
h(\chi_q(P_q))=2\e f(x)+\rho,\end{equation}
onde $\abs\rho\leq 2\e^2\norm f$. 

Como todo $x\in\I^n$ pertence, ao m\'\i nimo, aos $n+1$ cubos que correspondem aos valores distintos de $q$, podemos, quando
\[\e<\frac{n+1}2,\]
usar a estimativa fina (\ref{eq:fine}) para esses $n+1$ valores de $q$, bem como a estimativa grosseira (\ref{eq:coarse}) para os $\leq n$ cubos restantes, a fim de obter o seguinte,
\begin{align*}
\left\vert f(x)-\sum_{q=1}^{2n+1} h(\chi_q(x))\right\vert &\leq
(1-2(n+1)\e)\abs{f(x)} +2(n+1)\e^2\norm f+2n\e\norm f \\
&\leq (1-2\e+2(n+1)\e^2)\norm f \\
&\leq  (1-\e)\norm f,
\end{align*}
como desejado. N\'os mostramos que $\Omega (f)$ tem uma interse\c c\~ao n\~ao vazia com $G$, terminando a prova.
 \qed

\begin{observacao}
Observe novamente que a estimativa na \'ultima cadeia de desigualdades est\'a v\'alida pela mais magra das margens: ela depende do fato que $n+1>n$!
\end{observacao}

\section{Teorema da aproxima\c c\~ao universal de Cybenko}

\subsection{}
A rede de Kolmogorov--Lorentz ainda est\'a distante demais daquelas que ocorrem na pr\'atica. O que pode ser dito sobre \`a capacidade de aproxima\c c\~ao das redes mais realistas, por exemplo, de uma rede bin\'aria sem realimenta\c c\~ao feita de perceptrons, de tipo estudado na subse\c c\~ao \ref{ss:redes}? Teorema \ref{th:vc} afirma que uma tal rede de arquitetura fixa tem dimens\~ao VC finita, logo ela apenas pode aproximar uma classe pr\'opria de conceitos. E se permitirmos que o n\'umero de unidades cres\c ca? Acontece que neste caso, at\'e mesmo as redes da arquitetura a mais simples podem aproximar cada conceito arbitrariamente bem.

\begin{teorema}
Seja $\mathcal N$ uma rede neural sem re\-a\-li\-men\-ta\-\c c\~ao com valor limiar linear, cujas unidades de computa\c c\~ao s\~ao perceptrons. Suponha que $\mathcal N$ s\'o tem uma camada de entrada com $k$ unidades, uma camada escondida com $m$ unidades, e uma unidade de sa\'\i da. Ent\~ao, quando $m$ \'e bastante grande, $\mathcal N$ pode aproximar qualquer conceito boreliano $C\subseteq\R^k$. Em outras palavras, dado uma medida de probabilidade $\mu$ sobre $\R^k$, um conjunto boreliano $C\subseteq\R^k$ e $\ve>0$, existem $m$ e uma escolha de $(k+1)m+(m+1)$ par\^ametros da rede gerando um conceito $C^\prime$ com $\mu(C^\prime\Delta C)<\ve$.
\label{t:aproximacao_universal_conceitos}
\end{teorema}

Para mostrar o resultado, \'e necess\'ario coloc\'a-lo numa perspetiva ligeiramente mais geral, onde a unidade de sa\'\i da \'e linear, e deste modo, a rede produz fun\c c\~oes ao inv\'es de conceitos.

\begin{teorema}
Seja $\mathcal N$ uma rede neural sem re\-a\-li\-men\-ta\-\c c\~ao tendo uma camada de entrada com $k$ unidades, uma camada escondida com $m$ unidades, e uma unidade de sa\'\i da. Suponha que todas as unidades tenham fun\c c\~oes de modelagem afins, e as da camada escondida s\~ao perceptrons, enquanto a unidade de sa\'\i da \'e linear. Ent\~ao, para cada medida de probabilidade $\mu$ sobre $\R^k$, quando $m$ \'e bastante grande, $\mathcal N$ pode aproximar qualquer fun\c c\~ao $g\in L^1(\R^k,\mu)$. Em outras palavras, dado uma medida de probabilidade $\mu$ sobre $\R^k$, uma fun\c c\~ao $g\in L^1(\R^k\mu)$ e $\ve>0$, existem $m$ e uma escolha de $(k+1)m+m$ par\^ametros da rede gerando uma fun\c c\~ao $g^\prime$ com $\norm{g^\prime-g}_{L^1(\mu)}<\ve$.
\label{t:aproximacao_universal_eta}
\index{teorema! de Cybenko}
\end{teorema}

Forne\c camos algumas clarifica\c c\~oes. A fun\c c\~ao gerada pela unidade computacional da camada $1$ \'e da forma
\[u(x) = \eta\left(\sum_{i=1}^k a_i x_i -\theta\right),\]
onde $a_i,\theta\in\ R$ s\~ao os par\^ametros (pesos e o valor limiar) e $\eta$ a fun\c c\~ao de Heaviside. Deste modo, a fun\c c\~ao gerada pela rede \'e da forma
\begin{equation}
\sum_{j=1}^m\beta_j \eta\left(\sum_{i=1}^k a_{ij} x_i -\theta_j\right).
\label{eq:formadafuncao}
\end{equation}
O teorema diz que um espa\c co vetorial gerado pelas composi\c c\~oes da fun\c c\~ao de Heaviside $\eta$ com todas as fun\c c\~oes afins $z\colon \R^k\to \R$,
\[z(x) = \langle a,x\rangle -\theta,\]
\'e denso em $L^1(\R^k,\mu)$, qualquer seja $\mu\in P(\R^k)$. Note que todas as composi\c c\~oes $\eta\circ z$ s\~ao fun\c c\~oes borelianas e limitadas, logo pertencem ao espa\c co $L^1(\mu)$.

\begin{exercicio}
Deduza teorema \ref{t:aproximacao_universal_conceitos} do teorema \ref{t:aproximacao_universal_eta}.
\end{exercicio}

A demonstra\c c\~ao do teorema \ref{t:aproximacao_universal_eta} \'e baseada sobre o resultado seguinte.

\begin{teorema}[Cram\'er--Wold]
Uma medida boreliana de probabilidade sobre $\R^d$ \'e unicamente definida pelas suas imagens diretas sob todas as proje\c c\~oes unidimensionais.
\label{t:cramer-wold}
\index{teorema! de Cram\'er--Wold}
\end{teorema}

Em outras palavras, se $\mu_1$ e $\mu_2$ s\~ao duas medidas de probabilidade sobre $\R^d$ tais que, qualquer seja uma proje\c c\~ao ortogonal $\pi\colon\R^d\to\R$, temos $\pi_{\ast}(\mu_1)=\pi_{\ast}(\mu_2)$. Ent\~ao, $\mu_1=\mu_2$.

Para mostrar o resultado, relembra que a {\em fun\c c\~ao carater\'\i stica}, $\phi_{\mu}$, de uma medida de probabilidade, $\mu$, sobre $\R^d$ \'e uma fun\c c\~ao de $\R^d$ para $\C$, dada por
\begin{align*}
\R^d\ni t\mapsto \phi_{\mu}(t)&=\E_{\mu}(\exp{i\langle t,X\rangle})\\
&= \int_{\R^d}\exp{i\langle t,x\rangle}\,d\mu(x)\in\C.
\end{align*}
\index{fun\c c\~ao! carater\'\i stica}
A fun\c c\~ao carater\'\i stica permite de reconstruir a medida $\mu$ unicamente (teorema \ref{t:carfundeterminam}).

\begin{exercicio}
Seja $V$ um subespa\c co linear do espa\c co vetorial $\ell^2(d)$, com a proje\c c\~ao ortogonal correspondente $\pi\colon \ell^2(d)\to L$. Seja $\mu$ uma medida de probabilidade sobre $\ell^2(d)$. Verifique que para todo $t\in V$,
\[\phi_{\pi_{\ast}(\mu)}(t) = \phi_{\mu}(t).\]
[ {\em Sugest\~ao:} aplicando uma transforma\c c\~ao unit\'aria, pode-se supor que $V$ \'e gerado pelos primeiros $\dim V$ vetores de base... ]
\end{exercicio}

Teorema \ref{t:cramer-wold} de Cram\'er--Wold segue imediatamente modulo teorema \ref{t:carfundeterminam} e o fato que os subespa\c cos de dimens\~ao um cobram $\R^d$. 

\begin{observacao}
O argumento estabelece, de fato, um resultado mais geral (teorema de R\'enyi): cada medida de probabilidade sobre $R^d$ \'e determinada unicamente pelas suas imagens sob proje\c c\~oes sobre uma fam\'\i lia de subespa\c cos lineares cuja uni\~ao \'e $\R^d$.
\end{observacao}

\begin{observacao}
Geometricamente, o teorema de Cram\'er--Wold diz que uma medida de probabilidade sobre $\R^d$ \'e definida unicamente pelos valores sobre semiespa\c cos.
\end{observacao}

\begin{exercicio}
Deduza que a conclus\~ao do teorema de Cram\'er--Wold resta v\'alida para quaisquer medidas borelianas {\em finitas} sobre $\R^d$.
\end{exercicio}

\begin{proof}[Prova do teorema \ref{t:aproximacao_universal_eta}]
Seja $\mu$ uma medida de probabilidade boreliana sobre $\R^k$. Denotemos $V$ um subespa\c co linear fechado de $L^1(\mu)$ gerado pelas composi\c c\~oes de $\eta$ com todas as fun\c c\~oes afins $\R^k\to \R$. Seja $\phi$ um funcional linear limitado sobre $L^1(\mu)$ qualquer, que se anula sobre $V$. Vamos mostrar que $\phi$ \'e necessariamente trivial, $\phi\equiv 0$, o que implica, gra\c cas ao teorema de Hahn--Banach \ref{t:hb}, que $V=L^1(\mu)$.

Cada funcional linear limitado sobre $L^1(\mu)$ pode ser identificado com uma fun\c c\~ao $g\in L^{\infty}(\mu)$, essencialmente limitada em rela\c c\~ao com a medida $\mu$, pela f\'ormula:
\begin{equation}
\phi(h) = \int_{\R^k} g(x)h(x)\,d\mu(x).
\label{eq:defdephi}
\end{equation}
(Teorema \ref{t:dualdeL1}). Segundo a escolha de $\phi$, para toda fun\c c\~ao afim $z\colon\R^k\to\R$, temos
\begin{align}
\int \int_{\R^k} g(x)\eta(z(x))\,d\mu(x) &= \phi(\eta\circ z)\label{eq:=0} \\
&= 0. \nonumber
\end{align}

Consideremos a parte positiva e a parte negativa de $g$:
\[g_+(x)=\max\{g(x),0\},~~g_-(x)=-\min\{g(x),0\},\]
de modo que 
\[g=g_+ -g_-.\]
Dado qualquer subconjunto boreliano, $B\subseteq\R^k$, definamos
\[\nu_{\pm}(B) = \int_{B} g_{\pm}(x)\,d\mu(x).\]
Estas $\nu_{\pm}$ s\~ao medidas finitas, satisfazendo $\nu_{\pm}(\R^k)\leq \mbox{ess sup}\, g$. 
Vamos mostrar que $\nu_+=\nu_-$, o que terminar\'a a demonstra\c c\~ao, pois para todas fun\c c\~oes borelianas $h$,
\begin{equation}
\phi(h) = \int_{\R^k} h(x)\,d\nu_+(x) - \int_{\R^k} h(x)\,d\nu_-(x).
\label{eq:defdephi2}
\end{equation}
Segue-se da eq. (\ref{eq:=0}) que, qualquer seja fun\c c\~ao afim $z\colon\R^k\to\R$,
\begin{equation}
\int_{\R^k} \eta(z(x))\,d\nu_+(x) = \int_{\R^k} \eta(z(x))\,d\nu_-(x).
\end{equation}
Segundo o teorema de Cram\'er--Wold \ref{t:cramer-wold}, uma medida de probabilidade (logo, uma medida finita) sobre $\R^d$ \'e unicamente definida pelas imagens diretas sob todas as proje\c c\~oes unidimensionais. Basta mostrar que, dado um funcional linear $\psi\colon\R^k\to\R$, as medidas $\psi_{\ast}(\nu_+)$ e $\psi_{\ast}(\nu_-)$ s\~ao id\^enticas. Por sua vez, isto vai seguir se mostrarmos que as fun\c c\~oes de distribui\c c\~ao de $\psi_{\ast}(\nu_{\pm})$,
\[F_{\pm}(t) = \psi_{\ast}(\nu_{\pm})(-\infty, t],\]
s\~ao iguais para todos valores $t$. Seja $t$ qualquer. Definamos a fun\c c\~ao afim $z\colon\R^k\to\R$ por $z(x) = -\psi(x)+ t$. Agora, 
\[\psi(x)\leq t \iff z(x)\geq 0 \iff (\eta\circ z)(x)=1,\]
de onde conclu\'\i mos, usando eq. (\ref{eq:defdephi2}):
\begin{align}
F_{+}(t)-F_{-}(t) &= \nu_{+}(\psi^{-1}(-\infty, t))-\nu_{-}(\psi^{-1}(-\infty, t)) \label{eq:poslednee} \\
&= \int (\eta\circ z)\,d\nu_+ - \int (\eta\circ z)\,d\nu_- \nonumber \\
&= \int_{\R^k} g(x) (\eta\circ z) d\mu \nonumber\\
&= \phi(\eta\circ z)\nonumber \\
&=0.\nonumber
\end{align}
\end{proof}

Com pouco esfor\c co adicional, podemos substituir a fun\c c\~ao de Heaviside $\eta$ por qualquer fun\c c\~ao sigmoide $f$ como fun\c c\~ao de ativa\c c\~ao. Digamos que uma fun\c c\~ao $f\colon\R\to \R$ \'e {\em sigmoide} se
\[\lim f(x) =\begin{cases} 0,&\mbox{ quando }x\to-\infty,\\
1,&\mbox{ quando }x\to+\infty.
\end{cases}
\]
(Aqui, omitamos a condi\c c\~ao para $\eta$ ser mon\'otona).

\begin{exercicio}
Seja $f\colon\R\to\R$ uma fun\c c\~ao boreliana, limitada e sigmoide. Mostre que existe uma sequ\^encia $(z_n)$ de fun\c c\~oes afins, $z_n\colon\R\to\R$, tais que, para cada medida de probabilidade $\mu$ sobre $\R$, $\norm{f\circ z_n - \eta}_{L^1(\mu)}\to 0$. 
\par
[ {\em Sugest\~ao:} dado $\ve>0$, pode-se escolher uma fun\c c\~ao afim $z$ de modo que, se $x\notin (0,\ve)$, temos $\abs{f(z(x))-\eta(x)}<\ve$. Escolha recursivamente uma tal sequ\^encia $z_n$, verifique a converg\^encia simples, e use o teorema de Lebesgue de converg\^encia dominada. ]
\end{exercicio}

\begin{exercicio}
Adaptar a prova do teorema \ref{t:aproximacao_universal_eta} acima (nomeadamente, a parte em torno da eq. (\ref{eq:poslednee}) exige um pouco de trabalho) para mostrar que ao inv\'es da fun\c c\~ao de Heaviside $\eta$, pode-se substituir qualquer fun\c c\~ao limitada, boreliana, e sigmoide, $f$.
\end{exercicio}

Chegamos ao seguinte resultado.

\begin{teorema}[Teorema da Aproxima\c c\~ao Universal \citep*{cybenko}]
Seja $f\colon\R\to\R$ uma fun\c c\~ao boreliana e limitada sigmoide qualquer.
Seja $\mathcal N$ uma rede neural sem re\-a\-li\-men\-ta\-\c c\~ao tendo uma camada de entrada com $k$ unidades, uma camada escondida com $m$ unidades, e uma unidade de sa\'\i da. Suponha que todas as unidades tenham fun\c c\~oes de modelagem afins, e as da camada escondida tenham a fun\c c\~ao de ativa\c c\~ao $\eta$, enquanto a unidade de sa\'\i da \'e linear. Ent\~ao, para cada medida de probabilidade $\mu$ sobre $\R^k$, quando $m$ \'e bastante grande, $\mathcal N$ pode aproximar qualquer fun\c c\~ao $g\in L^1(\R^k,\mu)$. Em outras palavras, dado uma medida de probabilidade $\mu$ sobre $\R^k$, uma fun\c c\~ao $g\in L^1(\R^k\mu)$ e $\ve>0$, existem $m$ e uma escolha de $(k+1)m+m$ par\^ametros da rede gerando uma fun\c c\~ao $g^\prime$ com $\norm{g^\prime-g}_{L^1(\mu)}<\ve$.
\label{t:aproximacao_universal_cybenko}
\index{teorema! de Cybenko}
\end{teorema}

Neste caso, a fun\c c\~ao gerada pela unidade computacional da camada $1$ \'e da forma
\[u(x) = f\left(\sum_{i=1}^k a_i x_i -\theta\right),\]
enquanto a fun\c c\~ao gerada pela rede \'e
\begin{equation}
\sum_{j=1}^m\beta_j f\left(\sum_{i=1}^k a_{ij} x_i -\theta_j\right).
\label{eq:formadafuncao2}
\end{equation}
O teorema diz que um espa\c co vetorial gerado pelas composi\c c\~oes da fun\c c\~ao de ativa\c c\~ao sigmoide $f$ fixa com todas as fun\c c\~oes afins $z\colon \R^k\to \R$,
\[z(x) = \langle a,x\rangle -\theta,\]
\'e denso em $L^1(\R^k,\mu)$.

\subsection{Discuss\~ao}
\subsubsection{}
N\'os escolhemos de formular os resultados para norma $L^1(\mu)$, por causa de sua relev\^ancia para a nossa abordagem, mas de fato eles valem para qualquer $L^p(\mu)$, $1\leq p<\infty$, sem esfor\c co extra. (Exerc\'\i cio).

\subsubsection{}
O leitor pode querer modificar os argumentos acima para provar uma vers\~ao do teorema \ref{t:aproximacao_universal_cybenko} para fun\c c\~oes cont\'\i nuas, que Cybenko \citep*{cybenko} considerou como mais importante.

\begin{teorema}[George Cybenko, 1989]
Seja $f\colon\R\to\R$ uma fun\c c\~ao cont\'\i nua e limitada sigmoide qualquer.
Seja $\mathcal N$ uma rede neural sem re\-a\-li\-men\-ta\-\c c\~ao tendo uma camada de entrada com $k$ unidades, uma camada escondida com $m$ unidades, e uma unidade de sa\'\i da. Suponha que todas as unidades tenham fun\c c\~oes de modelagem afins, e as da camada escondida tenham a fun\c c\~ao de ativa\c c\~ao $\eta$, enquanto a unidade de sa\'\i da \'e linear. Ent\~ao, quando $m$ \'e bastante grande, $\mathcal N$ pode aproximar qualquer fun\c c\~ao cont\'\i nua $g\colon\R^k\to\R$ na topologia de converg\^encia uniforme sobre compactos. Em outras palavras, dado uma fun\c c\~ao $g\colon\R^k\to\R$ cont\'\i nua e limitada, um compacto $K\subseteq\R^k$, e $\ve>0$, existem $m$ e uma escolha de $(k+1)m+m$ par\^ametros da rede gerando uma fun\c c\~ao $g^\prime$ com $\max_{x\in K}\abs{g(x)-g^\prime(x)}<\ve$.
\label{t:aproximacao_universal_cybenko_cont}
\end{teorema}

A prova deve usar uma forma mais forte do teorema de representa\c c\~ao de Riesz do que n\'os estabelecemos no teorema \ref{t:riesz}: para qualquer espa\c co compacto metriz\'avel $K$, a f\'ormula 
\[\phi(f) =\int_X f(x)\,d\mu(x)\]
estabelece uma correspond\^encia bijetora entre os funcionais lineares limitados, $\phi: C(K)\to\R$, e {\em medidas finitas com sinal,} que tomam valores reais quaisquer, positivos assim como negativos.

\subsubsection{}
Refinando ainda mais o mesmo c\'\i rculo de ideias (teoria da transformada de Fourier, que est\'a escondida em nossa prova ao disfarce da fun\c c\~ao carater\'\i stica de uma medida), foi mostrado por Hornik \citep*{hornik} que, bastante surpreendentemente, a fun\c c\~ao de ativa\c c\~ao pode ser qualquer fun\c c\~ao n\~ao essencialmente constante em rela\c c\~ao \`a medida de Lebesgue, $\lambda$. (Uma fun\c c\~ao $f\colon \R\to \R$ \'e essencialmente constante se $f(x)=f(y)$ para quase todos $x,y\in\R$, ou seja, existe $c\in\R$ t\~ao que $\lambda\{x\in\R\colon f(x)\neq c\}=0$.)

\begin{teorema}[Kurt Hornik, 1991]
Seja $f\colon\R\to\R$ uma fun\c c\~ao boreliana qualquer, limitada \'e n\~ao essencialmente constante, em rela\c c\~ao \`a medida de Lebesgue.
Seja $\mathcal N$ uma rede neural sem re\-a\-li\-men\-ta\-\c c\~ao tendo uma camada de entrada com $k$ unidades, uma camada escondida com $m$ unidades, e uma unidade de sa\'\i da. Suponha que todas as unidades tenham fun\c c\~oes de modelagem afins, e as da camada escondida tenham a fun\c c\~ao de ativa\c c\~ao $\eta$, enquanto a unidade de sa\'\i da \'e linear. Ent\~ao, para cada medida de probabilidade $\mu$ sobre $\R^k$, quando $m$ \'e bastante grande, $\mathcal N$ pode aproximar qualquer fun\c c\~ao $g\in L^1(\R^k,\mu)$. Em outras palavras, dado uma medida de probabilidade $\mu$ sobre $\R^k$, uma fun\c c\~ao $g\in L^1(\R^k\mu)$ e $\ve>0$, existem $m$ e uma escolha de $(k+1)m+m$ par\^ametros da rede gerando uma fun\c c\~ao $g^\prime$ com $\norm{g^\prime-g}_{L^1(\mu)}<\ve$.
\label{t:aproximacao_universal_hornik}
\end{teorema}

A prova do teorema de Hornik necessita o uso de um teorema de Wiener na An\'alise Harm\^onica, oferecendo uma classifica\c c\~ao completa de todos os subespa\c cos lineares fechados de $L^1(\R,\lambda)$, invariantes pelas transla\c c\~oes. No mesmo artigo \citep*{hornik} Hornik mostrou tamb\'em uma vers\~ao cont\'\i nua do resultado, parecida ao teorema \ref{t:aproximacao_universal_cybenko_cont}, onde a fun\c c\~ao de ativa\c c\~ao $f\colon\R\to\R$ pode ser qualquer fun\c c\~ao cont\'\i nua e limitada n\~ao constante.

\begin{observacao}
Mencionemos uma pergunta de Hornik \citep*{hornik} que pode ser ainda em aberto. Quais s\~ao as condi\c c\~oes (necess\'arias e suficientes) sobre uma fun\c c\~ao de ativa\c c\~ao $f\colon\R\to\R$ (n\~ao necessariamente boreliana) para garantirem que a classe de fun\c c\~oes $g^\prime$ geradas pela rede como acima aproxime todas as fun\c c\~oes cont\'\i nuas $g\colon\R^k\to\R$ na topologia de converg\^encia uniforme sobre compactos? (Esta topologia tem sentido para espa\c co linear de todas as fun\c c\~oes de $\R^k$ para $\R$ limitadas sobre cada compacto).

\'E claro que algumas condi\c c\~oes sejam necess\'arias, pois se, por exemplo, $f$ \'e a fun\c c\~ao indicadora de um ponto, a conclus\~ao \'e falsa.
\end{observacao}

\section{Minimiza\c c\~ao de risco estrutural}

At\'e agora, estudamos dois conceitos de aprendizagem: a dentro da classe, e a universalmente consistente. Ambos t\^em vantagens e desvantagens. Aprendizagem dentro da classe permite estimar o erro de generaliza\c c\~ao, por\'em n\~ao pode aproximar qualquer conceito. De outro lado, regras de aprendizagem universalmente consistentes podem aprender qualquer conceito, mas a taxa de aprendizagem pode ser arbitrariamente devagar. 

N\'os discutiremos agora um conceito de aprendizagem que pertence a \citep*{vapnik} e que combina alguns bons aspectos das duas abordagens: o princ\'\i pio de minimiza\c c\~ao de risco estrutural. 

Senti-me tentado de desenvolver uma apresenta\c c\~ao diferente, e quaisquer erros e mal-entendidos surgiram, s\~ao certamente os meus pr\'oprios.

\subsection{Reexaminando o erro emp\'\i rico de aprendizagem}

Dada uma classe de conceitos $\mathscr C$ e uma amostra rotulada $(\sigma,\e)$, o princ\'\i pio de minimiza\c c\~ao do erro emp\'\i rico exige minimizar sobre $C\in {\mathscr C}$ o erro emp\'\i rico de aprendizagem do conceito desconhecido, $D$,
\[\mbox{erro}_{\mu_{\sigma},D}(C) = \mu_{\sigma}(C\Delta D).\]
Note que a express\~ao pode ser reescrita evitando $D$, pois somente usamos a rotulagem $D\upharpoonright\sigma$ produzida sobre $\sigma$ por $D$, or, de modo equivalente, a subamostra $\sigma_+\sqsubset \sigma$ formada pelos pontos $x_i$ com o rotulo correspondente $\e_i=1$:
\[\mbox{erro}_{\mu_{\sigma},\sigma_+}(C) = \mu_\sigma(\sigma_+\Delta C).\]
Reinterpretemos o erro emp\'\i rico $\mbox{erro}_{\mu_{\sigma},\sigma_+}(C)$ no contexto do problema de aprendizagem $(\mu,\eta)$ no dom\'\i nio $\Omega$, onde $\mu\in P(\Omega)$ e $\eta\colon\Omega\to [0,1]$ \'e uma fun\c c\~ao de regress\~ao. 

Recordemos o teorema \ref{t:rademacher}: dado uma classe de conceitos ${\mathscr C}$ qualquer, com confian\c ca $1-\delta$,
\[\sup_{C\in{\mathscr C}}\left\vert \mu_\sigma(C)-\mu(C)\right\vert\leq 2R_n({\mathscr C}) + \sqrt{\frac{\ln (2/\delta)}{2n}}.\]

Nos lemas seguintes, a amostra rotulada $(\sigma,\sigma_+)$ segue a lei conjunta dada por $(\mu,\eta)$.

\begin{lema}
Seja $\mathscr C$ uma classe de conceitos.
Para todos $C\in {\mathscr C}$, com confian\c ca $1-\delta$,
\begin{align*}
\mu_{\sigma}(\sigma_+\cap C)&\overset\ve \approx \mu_1(C) \\
& = \int_C\eta\,d\mu,
\end{align*}
onde
\begin{equation}
\ve = 2R_n({\mathscr C}) + \sqrt{\frac{\ln (2/\delta)}{2n}}.
\label{eq:valordeeps}
\end{equation}
\end{lema}

\begin{proof}
Basta aplicar o teorema \ref{t:rademacher} \`a medida emp\'\i rica $\mu_{(\sigma,\e)}$ no dom\'\i nio rotulado $\Omega\times\{0,1\}$, substituindo $C\times \{1\}$ em vez de $C$. A complexidade de Rademacher da classe de conjuntos $C\times \{1\}$, $C\in {\mathscr C}$, \'e limitada por cima pela complexidade da classe $\mathscr C$.
\end{proof}

Do modo parecido, usando o teorema \ref{t:rademacher} com a classe de conceitos $C\times\{0\}$, obtemos

\begin{lema}
Para todos $C\in {\mathscr C}$, com confian\c ca $1-\delta$,
\begin{align*}
\mu_{\sigma}(\sigma_-\cap C)&\overset\ve \approx \mu_0(C)  \\
& =\int_C(1-\eta)\,d\mu,
\end{align*}
onde $\ve$ \'e como na eq. (\ref{eq:valordeeps}).
\end{lema}

\begin{exercicio}
Mostre que a complexidade de Rademacher da classe ${\mathscr C}^c$, formada pelos conceitos complementares $\Omega\setminus C$, $C\in {\mathscr C}$, \'e igual a complexidade de Rademacher de $\mathscr C$.
\end{exercicio}

\begin{lema}
Para todos $C\in {\mathscr C}$, com confian\c ca $1-2\delta$,
\begin{align*}
\mbox{erro}_{\mu_{\sigma},\sigma_+}(C) &\overset\ve \approx \int_{\Omega}\abs{\chi_C-\eta}\,d\mu \\
&=\mbox{erro}_{\mu,\eta}(C),
\end{align*}
onde
\[\ve =4R_n({\mathscr C}) + 2\sqrt{\frac{\ln (2/\delta)}{2n}}.
\]
\label{l:erroeerro}
\end{lema}

\begin{proof}
\begin{align*}
\int_{\Omega}\abs{\chi_C-\eta}\,d\mu 
&
= \int_C (1-\eta)\,d\mu + \int_{\Omega\setminus C}\eta\,d\mu \\
& \overset{2\ve}\approx \mu_{\sigma}(\sigma_-\cap C) +
\mu_{\sigma}(\sigma_+\cap (\Omega\setminus C)) \\
& = \mu_{\sigma}\left[(C\setminus\sigma_+)\cup (\sigma_+\setminus C) \right]
\\
&= \mu_{\sigma}(C\Delta\sigma_+) \\
&= \mbox{erro}_{\mu_{\sigma},\sigma_+}(C).
\end{align*}
\end{proof}

Por conseguinte, mesmo quando apenas ambicionarmos a minimizar o erro de aprendizagem dentro de um problema determin\'\i stico (sob a hip\'otese $\eta=\chi_C$), na realidade, estamos minimizando o erro de aprendizagem no problema com ru\'\i do aleat\'orio tamb\'em. 

\subsection{Classes filtradas}

Seja ${\mathscr C}$ uma classe de conceitos num dom\'\i nio $\Omega$, munido de uma {\em filtra\c c\~ao,} ou seja, representado como uni\~ao de uma cadeia crescente de subclasses,
\[{\mathscr C} = \bigcup_{m=1}^{\infty} {\mathscr C}_m,\]
\[{\mathscr C}_1\subseteq {\mathscr C}_2\subseteq\ldots\subseteq {\mathscr C}_m\subseteq \ldots.\]
Para $C\in {\mathscr C}$ escrevamos 
\[\mbox{gr}\,(C) = \min\{m\in\N\colon C\in {\mathscr C}_m\}.\]
Suponhamos tamb\'em que
\begin{enumerate}
\item Dimens\~ao de Vapnik--Chervonenkis de cada classe $\VC({\mathscr C}_m) = d_m <\infty$ \'e finita,
\item ${\mathscr C}$ tem a propriedade de aproxima\c c\~ao universal: para cada medida de probabilidade $\mu$ sobre $\Omega$, cada conceito boreliano $D\subseteq\Omega$, e cada $\ve>0$, existe $C\in {\mathscr C}$ tal que $\mu(C\Delta D)<\ve$.
\end{enumerate}

\par
Vamos chamar uma tal classe {\em classe filtrada}.
\index{classe! filtrada}

\begin{observacao}
Segue-se que a sequ\^encia $(d_m)$ \'e mon\'otona e converge para infinito, $d_m\uparrow+\infty$.
\end{observacao}

\begin{exemplo} Seja ${\mathscr C}_m$ a classe de conceitos gerada pela rede neural sem reali\-men\-ta\-\c c\~ao, com valor limiar linear, tendo uma camada interior com $m$ unidades. Teoremas \ref{th:vc} e \ref{t:aproximacao_universal_cybenko} servem a verificar que a classe de conceitos filtrada ${\mathscr C} = \cup_{m=1}^{\infty} {\mathscr C}_m$ satisfaz as duas condi\c c\~oes. 
\label{ex:filtrado1}
\end{exemplo}

\begin{exemplo}
Definamos ${\mathscr C}_m$ a classe de todos os conceitos da forma $\eta\circ p_m$, onde $\eta$ \'e a fun\c c\~ao de Heaviside e $p_m$ \'e um polin\^omio qualquer de grau $\leq m$. As duas condi\c c\~oes acima seguem do teorema \ref{t:cf} (imediato) e do teorema de Stone--Weierstrass \ref{stone-weierstrass} (isto exige um pouco de trabalho, exerc\'\i cio).
\label{ex:filtrado2}
\end{exemplo}

\begin{exemplo}
Seja $\Omega=\{0,1\}^{\N}$ um espa\c co de Cantor (as vezes chamado o ``cubo de Cantor''), ou seja, um espa\c co compacto metriz\'avel totalmente desconexo sem pontos isolados, cuja topologia \'e gerada pela m\'etrica 
\[d(x,y)= 2^{-\min\{i\colon x_i\neq y_i\}}.\]
Dado $m\in\N$, definamos ${\mathscr C}_m$ a classe de todos os conjuntos cil\'\i ndricos com a base nas primeiras $m$ coordenadas:
\[{\mathscr C}_m = \{\pi_m^{-1}(A)\colon A\subseteq \{0,1\}^m\},\]
once $\pi_m$ \'e a proje\c c\~ao can\^onica de $\{0,1\}^{\N}$ sobre $\{0,1\}^m$. Tais conjuntos s\~ao abertos e fechados.
A classe ${\mathscr C}_m$ \'e finita, com $2^m$ elementos, e $\VC({\mathscr C}_m)=m$. A propriedade de aproxima\c c\~ao universal da classe ${\mathscr C} = \cup_{m=1}^{\infty} {\mathscr C}_m$ segue das duas afirma\c c\~oes seguintes, deixadas como exerc\'\i cios.
\begin{enumerate}
\item Dado um conjunto boreliano $C\subseteq \{0,1\}^{\N}$, uma medida de probabilidade $\mu$, e $\ve>0$, existe um conjunto aberto e fechado $V\subset \{0,1\}^{\N}$ tal que $\mu(C\Delta D)<\ve$. (Usar teorema \ref{t:regularidadeKU} sobre regularidade de $\mu$, mais um pouco da topologia geral).
\item Cada conjunto aberto e fechado $V\subset \{0,1\}^{\N}$ \'e necessariamente da forma $\pi_m^{-1}(A)$ para $m$ e $A\subseteq \{0,1\}^m$ apropriados. (Use compacidade). Assim, $\mathscr C$ \'e exatamente a classe de todos subconjuntos abertos e fechados do cubo de Cantor.
\end{enumerate}
\label{ex:filtrado3}
\end{exemplo}

\begin{exercicio}
Mostre que para cada classe de conceitos $\mathscr C$,
\[\hat R_n({\mathscr C})\leq \sqrt{2d \frac{\log n}{n}},\]
onde $d=\VC(\mathscr C)$.
\par
[ {\em Sugest\~ao:} combinar as estimativas na prova da implica\c c\~ao (2) $\Rightarrow$ (1), p\'agina \pageref{p:impl1implies2}, a desigualdade na subse\c c\~ao \ref{ss:cond3}, e o lema de Sauer--Shelah. ]
\label{ex:rademacherviad}
\end{exercicio}

\begin{lema}
Sejam $\mathscr C$ uma classe filtrada. Dada uma sequ\^encia $(\delta_m)$, $\delta_m>0$, a um n\'umero $\delta>0$ tais que $\delta=\sum_{m=1}^{\infty}\delta_m$, temos para cada problema de aprendizagem $(\mu,\eta)$ em $\Omega$, com confian\c ca $1-2\delta$,
\begin{align*}
\forall C\in {\mathscr C},~\left\vert\mbox{erro}_{\mu_{\sigma},\sigma_+}(C)- \mbox{erro}_{\mu,\eta}(C)\right\vert &\leq 4 R_n({\mathscr C}_{\mathrm{gr}\,(C)},\mu) + 2\sqrt{\frac{\ln (2/\delta_{\mathrm{gr}\,(C)})}{2n}} \\
& \leq 4 \sqrt{2d_{\mathrm{gr}\,(C)} \frac{\log n}{n}}  + 2\sqrt{\frac{\ln (2/\delta_{\mathrm{gr}\,(C)})}{2n}}.
\end{align*}
\label{l:riscoestrutural}
\end{lema}

\begin{proof}
Para todo $m$, pelo lema \ref{l:erroeerro} temos com confian\c ca $1-2\delta_m$,
\[\forall C\in {\mathscr C}_m,~\left\vert\mbox{erro}_{\mu_{\sigma},\sigma_+}(C)- \mbox{erro}_{\mu,\eta}(C)\right\vert\leq 4 R_n({\mathscr C}_{m},\mu) + 2\sqrt{\frac{\ln (2/\delta_{m})}{2n}}, \]
e a primeira desigualdade segue com risco $2\delta=\sum_{i=1}^{\infty}2\delta_m$ pela cota da uni\~ao. A segunda desigualdade segue do exerc\'\i cio \ref{ex:rademacherviad}.
\end{proof}

\begin{definicao}
Seja $\mathscr C$ uma classe filtrada, $(\mu,\eta)$ um problema de aprendizagem, e $(\delta_m)_{m=1}^{\infty}$ uma sequ\^encia som\'avel de reais, $\delta_m>0$. O {\em risco estrutural} \'e a fun\c c\~ao
\[\mbox{erro}_{\mu_{\sigma},\sigma_+}(C) + 4 \sqrt{2d_{\mathrm{gr}\,(C)} \frac{\log n}{n}}  + 2\sqrt{\frac{\ln (2/\delta_{\mathrm{gr}\,(C)})}{2n}}.\]
A regra de aprendizagem $\mathcal L$ segue o {\em princ\'\i pio de minimiza\c c\~ao do risco estrutural} se para todo $n$, ${\mathcal L}_n$ escolha a hip\'otese $H$ minimizando a express\~ao acima.
\index{princ\'\i pio! de minimiza\c c\~ao! de risco estrutural}
\end{definicao}

\begin{observacao}
\'E claro que a escolha de uma tal hip\'otese $H$ \'e sempre poss\'ivel. Mostrar que tal escolha \'e sempre poss\'\i vel em uma forma Borel mensur\'avel exige trabalho adicional, no esp\'\i rito da subse\c c\~ao \ref{ss:univsep}. Por exemplo, isto \'e sempre poss\'\i vel se a classe $\mathscr C$ \'e universalmente separ\'avel (exerc\'\i cio).
\end{observacao}

\begin{exercicio}
Mostrar que as classes filtradas nos exemplos \ref{ex:filtrado1}, \ref{ex:filtrado2}, e \ref{ex:filtrado3} s\~ao universalmente separ\'aveis.
\end{exercicio}

\begin{teorema}
Cada regra de aprendizagem seguindo o princ\'\i pio de minimiza\c c\~ao do risco estrutural numa classe filtrada \'e universalmente consistente.
\end{teorema}

\begin{proof}
Dada uma classe filtrada ${\mathscr C}$, escolhamos uma sequ\^encia $(m_n)$ que cresce, $m_n\uparrow+\infty$, mas bastante devagar, da maneira que a sequ\^encia
\[4 \sqrt{2d_{m_n} \frac{\log n}{n}}  + 2\sqrt{\frac{\ln (2/\delta_{m_n})}{2n}}\]
converge para zero monotonicamente. Lema \ref{l:riscoestrutural} implica que, com confian\c ca $1-2\sum_{i=m_n}^{\infty}\delta_i$, 
\[
\forall C\in {\mathscr C},~\left\vert\mbox{erro}_{\mu_{\sigma},\sigma_+}(C)- \mbox{erro}_{\mu,\eta}(C)\right\vert \leq 4 \sqrt{2d_{m_n\wedge \mathrm{gr}\,(C)} \frac{\log n}{n}}  + 2\sqrt{\frac{\ln (2/\delta_{m_n\wedge \mathrm{gr}\,(C)})}{2n}}.\]
Seja $\ve>0$ qualquer. Existe $M$ t\~ao grande que a classe ${\mathscr C}_{M}$ cont\'em um conceito $C$ com $\norm{\chi_C-T_{bayes}}_1<\ve$. Este $C$ pertence a todas as classes ${\mathscr C}_{m}$, $m\geq M$. Segundo lema \ref{l:erroeerro}, com confian\c ca $1-2\sum_{i=m_n}^{\infty}\delta_i$, a regra ${\mathcal L}_n$ vai escolher uma hip\'otese $H$ satisfazendo 
\[\left\vert\mbox{erro}_{\mu_{\sigma},\sigma_+}(H)- \mbox{erro}_{\mu,\eta}(H)\right\vert \leq 4 \sqrt{2d_{m_n\wedge \mathrm{gr}\,(H)} \frac{\log n}{n}}  + 2\sqrt{\frac{\ln (2/\delta_{m_n\wedge \mathrm{gr}\,(H)})}{2n}}.\]
Combinando duas desigualdades, conclu\'\i mos: com confian\c ca $1-4\sum_{i=m_n}^{\infty}\delta_i$,
\begin{align*}
\mbox{erro}_{\mu,\eta}{\mathcal L}_n(\sigma,\sigma_+) &\leq \ell^{\ast}(\mu,\eta)+\ve +8 \sqrt{2d_{m_n\wedge \mathrm{gr}\,(H)} \frac{\log n}{n}}  + 4\sqrt{\frac{\ln (2/\delta_{m_n\wedge \mathrm{gr}\,(H)})}{2n}}.
\end{align*}

Quando $n\to\infty$, as duas ra\'\i zes e o risco $\sum_{i=m_n}^{\infty}\delta_i$ convergem para zero, e $\ve>0$ foi qualquer. Deduzimos a converg\^encia em probabilidade 
\[\mbox{erro}_{\mu,\eta}{\mathcal L}_n(\sigma,\sigma_+)\to \ell^{\ast}(\mu,\eta).\]
\end{proof}

%
%

\chapter{Compress\~ao amostral\label{ch:compressao}}

\section{Compress\~ao, descompress\~ao, e aprendizagem}

\subsection{}
O assunto do \'ultimo cap\'\i tulo \'e esquemas de compress\~ao amostral.
Denotemos por $[\Omega]^d$ o conjunto de todos os subconjuntos com $d$ elementos, $[\Omega]^{\leq d}$ o conjunto de todos os subconjuntos com $\leq d$ elementos, e $[\Omega]^{\leq \infty}$o conjunto de todos os subconjuntos finitos. 
Na sua vers\~ao mais b\'asica e pura, uma {\em aplica\c c\~ao de descompress\~ao amostral de tamanho $d$} \'e  qualquer aplica\c c\~ao
\[{\mathcal H}\colon [\Omega]^{\leq d}\to 2^{\Omega},\]
\index{aplica\c c\~ao! de descompress\~ao amostral}
associando a cada subconjunto com $\leq d$ elementos uma hip\'otese, ou seja, um subconjunto de $\Omega$. A aplica\c c\~ao de descompress\~ao pode ser tamb\'em codificada assim:
\begin{equation}
{\mathcal H}\colon [\Omega]^{\leq d}\times\Omega\to \{0,1\}.
\label{eq:descompressao2}
\end{equation}

Deste modo, os subconjuntos finitos de $\Omega$, $\sigma=\{x_1,x_2,\ldots,x_n\}$, est\~ao vistos como ``amostras'' num sentido um pouco diferente: n\'os descartamos a ordem nas amostras, bem como repeti\c c\~oes de pontos. 

\begin{exemplo}
Seja $\Omega=\R$ e $d=2$. A cada conjunto $\sigma$ com $\leq 2$ pontos associemos o menor intervalo fechado, $I(\sigma)$, que cont\'em $\sigma$, ou seja, a envolt\'oria fechada convexa de $\sigma$ na reta.
A mesma constru\c c\~ao se aplica a qualquer conjunto $\Omega$ totalmente ordenado.
\label{ex:esqtotord}
\end{exemplo}

\begin{exemplo}
Seja $\Omega=\R^2$ e $d=4$. A cada conjunto $\sigma$ com $\leq 4$ pontos associemos o menor ret\^angulo fechado, $\Pi(\sigma)$, cujos lados s\~ao paralelos aos eixos coordenados e que cont\'em $\sigma$.
\label{ex:esqplano}
\end{exemplo}

Dado uma aplica\c c\~ao de descompress\~ao amostral, $\mathcal H$, definamos a classe ${\mathscr C}_{\mathcal H}$, que consiste de todos os subconjuntos $C\subseteq \Omega$ satisfazendo
\begin{equation}
\forall \sigma\in [\Omega]^{\leq \infty},~~\exists \tau\subseteq\sigma,~\tau\in [\Omega]^{\leq d},~~{\mathcal H}(\tau)\cap\sigma = C\cap\sigma.
\label{eq:condicaoC_H}
\end{equation}
Em outras palavras, a rotulagem induzida por $C$ sobre qualquer amostra finita, $\sigma$, pode ser tamb\'em induzida pela descompress\~ao de uma subamostra apropriada de $\sigma$ de tamanho $\leq d$.
\index{CH@${\mathscr C}_{\mathcal H}$}

\begin{exercicio} 
Seja $\mathcal H$ uma aplica\c c\~ao de descompress\~ao amostral de tamanho $d$. Mostrar que a classe ${\mathscr C}_{\mathcal H}$ tem dimens\~ao de Vapnik--Chervonenkis $\leq d$. 
\par 
[ {\em Sugest\~ao:} contar as rotulagens diferentes sobre $\sigma$. ]
\label{ex:descompressaoVC}
\end{exercicio}

Um problema maior em aberto \'e o seguinte. Seja $\mathscr C$ uma classe de conceitos. Existe uma aplica\c c\~ao de descompress\~ao de tamanho $d=\VC(\mathscr C)$ tal que ${\mathscr C}\subseteq{\mathscr C}_{\mathcal H}$?
\label{p:perguntaemaberto}

\begin{observacao}
Os exemplos \ref{ex:esqtotord} e \ref{ex:esqplano} fornecem uma boa ilustra\c c\~ao simples.
\end{observacao}

Para esquemas de compress\~ao, a ordem dos elementos numa amostra, bem como as repeti\c c\~oes, n\~ao importam. Por isso, uma {\em amostra} significa um subconjunto finito do dom\'\i nio, $\Omega$, enquanto uma {\em amostra rotulada} vai significar um subconjunto finito $\sigma$ do dom\'\i nio $\Omega$, munido de uma rotulagem, ou seja, um subconjunto $\tau\subseteq\sigma$ de pontos rotulados $1$, no pressuposto de que os pontos de $\sigma\setminus\tau$ s\~ao rotulados $0$. Denotemos por $\ls\Omega d$ o conjunto de todas as amostras rotuladas de tamanho $\leq d$, e da mesma maneira $\ls\Omega\infty$.

Dado uma aplica\c c\~ao de descompress\~ao amostral, $\mathcal H$, na defini\c c\~ao da classe ${\mathscr C}_{\mathcal H}$ fica impl\'\i cita uma {\em aplica\c c\~ao de compress\~ao amostral de tamanho $d$,}

\begin{equation}
\kappa\colon \ls\Omega\infty\to [\Omega]^{\leq d},
\label{eq:aplcompressao}
\end{equation}
\index{aplica\c c\~ao! de compress\~ao amostral}
tendo as propriedades 
\begin{equation}
\kappa(\sigma,\tau)\subseteq \sigma
\label{eq:kappasigmatau}
\end{equation}
e, para cada $C\in {\mathscr C}_{\mathcal H}$,
\begin{equation}
{\mathcal H}(\kappa(\sigma,C\cap\sigma))\cap\sigma = C\cap\sigma.
\label{eq:propkappa}
\end{equation}

\begin{definicao}
Um {\em esquema de compress\~ao amostral de tamanho $d$} para uma classe de conceitos $\mathscr C$ \'e um par $({\mathcal H},\kappa)$, onde ${\mathcal H}$ \'e uma aplica\c c\~ao de descompress\~ao de tamanho $d$ tal que ${\mathscr C}\subseteq {\mathscr C}_{\mathcal H}$, e $\kappa$ e uma aplica\c c\~ao de compress\~ao de tamanho $d$ correspondente.
\index{esquema! de compress\~ao amostral}
\end{definicao}

\begin{observacao}
De modo equivalente, um esquema de compress\~ao amostral de tamanho $d$ pode ser definido como um par $({\mathcal H},\kappa)$ de aplica\c c\~oes
\[{\mathcal H}\colon [\Omega]^{\leq d}\to 2^{\Omega},\]
\[\kappa\colon \ls\Omega\infty\to [\Omega]^{\leq d},\]
tendo as propriedades nas eq. (\ref{eq:kappasigmatau}) e (\ref{eq:propkappa}).
\end{observacao}

Em particular, ${\mathcal H}\circ\kappa$ \'e uma regra de aprendizagem consistente com a classe $\mathscr C$, o que conduz-nos ao assunto de aprendizagem. Por esta raz\~ao, estamos mais interessados no caso onde a aplica\c c\~ao de descompress\~ao na eq. (\ref{eq:descompressao2}) \'e boreliana. 
O t\'opico de mensurabilidade de esquemas de compress\~ao \'e praticamente inexplorado (no entanto, veja a disserta\c c\~ao \citep*{kalajdzievski}).

\begin{exercicio}
Seja $\Omega$ um espa\c co boreliano padr\~ao e $k\in\N$.
Mostrar que o espa\c co $[\Omega]^k$ tem uma estrutura boreliana padr\~ao induzida do produto $\Omega^k$. 
\par
[ {\em Sugest\~ao:} identificar $\Omega$ com o intervalo $\I$. ] Deduzir que $[\Omega]^{\leq k}$ e $[\Omega]^{\leq \infty}$ tamb\'em admitem estruturas borelianas padr\~ao naturais.
\end{exercicio}

A pergunta seguinte est\'a mais forte, e vermos que uma resposta positiva implicaria a resposta positiva na pergunta na p\'agina \pageref{p:perguntaemaberto}. Seja $\mathscr C$ uma classe de conceitos. Suponha que $\mathscr C$ seja munido de uma estrutura boreliana padr\~ao tal que $\{(x,C)\colon x\in C\in {\mathscr C}\}$ \'e um subconjunto boreliano de $\Omega\times {\mathscr C}$. Existe uma aplica\c c\~ao de descompress\~ao boreliana de tamanho $d=\VC(\mathscr C)$ tal que ${\mathscr C}\subseteq{\mathscr C}_{\mathcal H}$? Por exemplo, se $\mathscr C$ \'e universalmente separ\'avel?

\begin{exercicio}
Mostre que a classe ${\mathscr C}_{\mathcal H}$ gerado por uma fun\c c\~ao de descompress\~ao $\mathcal H$ \'e fechado na topologia de converg\^encia simples. Deduza que se $\mathscr C^\prime$ \'e universalmente denso em $\mathscr C$, ent\~ao cada esquema de compress\~ao amostral para $\mathscr C^\prime$ \'e automaticamente um esquema de compress\~ao amostral para $\mathscr C$ tamb\'em.
\end{exercicio}

Aqui \'e a liga\c c\~ao com a aprendizagem supervisionada.
Definamos a aplica\c c\~ao can\^onica $\iota$ que esquece a ordem e as repeti\c c\~oes numa amostra rotulada,
\begin{align*}
\iota\colon \bigcup_{n=1}^{\infty} \Omega^n\times \{0,1\}^n  &\to [\Omega]^{\leq \infty}\times [\Omega]^{\leq \infty},
\\
\sigma=(x_1,x_2,\ldots,x_n,\e_1,\e_2,\ldots,\e_n)&\mapsto (\{x_1,x_2,\ldots,x_n\}, \{x_i\colon \e_i=1\}).
\end{align*}
Agora, cada esquema de compress\~ao amostral para uma classe $\mathscr C$ defina uma regra de aprendizagem como segue: dado uma amostra rotulada $(\sigma,\e)$,
\[{\mathcal L}(\sigma,\e) = {\mathcal H}(\kappa(\iota(\sigma,\e))).\]
Segue-se das defini\c c\~oes que a regra \'e consistente com a classe $\mathscr C$, logo, PAC aprende $\mathscr C$ (teorema \ref{t:maintVC} e exerc\'\i cio \ref{ex:descompressaoVC}). 

Existe uma prova direta e muito simples, sem usar as classes de Glivenko--Cantelli. O resultado seguinte foi um dos principais no trabalho original \citep*{LW}, onde a teoria de esquemas de compress\~ao amostral foi lan\c cada. Curiosamente, o manuscrito nunca foi publicado, por\'em resta um das refer\^encias mais importantes nesta \'area. 

\begin{teorema}[Littlestone--Warmuth]
Seja $({\mathcal H},\kappa)$ um esquema de compress\~ao amostral de tamanho $d$ para uma classe $\mathscr C$ de conceitos borelianos, tal que a aplica\c c\~ao ${\mathcal H}\colon [\Omega]^{\leq d}\times\Omega\to \{0,1\}$ \'e boreliana. Ent\~ao a regra ${\mathcal L} = {\mathcal H}\circ\kappa\circ\iota$ PAC aprende a classe $\mathscr C$ com erro $\ve>0$ e confian\c ca
\[\geq 1 - \sum_{j=0}^d{n\choose j}(1-\ve)^{n-j}> 1 - \left(\frac{en}{d}\right)^d(1-\ve)^{n-d},\]
onde $n$ \'e o tamanho da amostra. 
\label{t:littlestone-warmuth}
\index{teorema! de Littlestone--Warmuth}
\end{teorema}

No seguinte, identifiquemos o espa\c co boreliano padr\~ao $\Omega$ com o intervalo $\I=[0,1]$. Fixemos o tamanho da amostra, $n$. Seja $J$ um subconjunto  de $[n]=\{1,2,3,\ldots,n\}$. Dado uma amostra $\sigma\in[\I]^d$, escrevamos $\sigma=\{x_1,x_2,\ldots,x_n\}$, onde $x_1<x_2<\ldots<x_n$, e formemos uma amostra $\sigma_{J}$ com $d$ elementos pela regra
\[x_i\in\sigma_{\tau}\iff i\in J.\]
A aplica\c c\~ao 
\[[\I]^{n}\ni\sigma\mapsto \sigma_J\in [\I]^{\sharp J}\]
\'e cont\'\i nua com rela\c c\~ao \`a topologia de produto, em particular boreliana.

\begin{exercicio}
Verificar que na presen\c ca de uma lei $\mu$ os elementos aleat\'orios $\sigma_J$ e $\sigma_{[n]\setminus J}$ s\~ao independentes. 
\end{exercicio}

Cada escolha de um conjunto de \'\i ndices $J\in[[n]]^{\leq d}$ de tamanho $j\leq d$ define a {\em hip\'otese aleat\'oria} ${\mathcal H}(\sigma_J)$. (A no\c c\~ao tem sentido porque a aplica\c c\~ao $\mathcal H$ \'e boreliana). Segue-se que o elemento aleat\'orio ${\mathcal H}(\sigma_J)$ \'e independente de $\sigma_{[n]\setminus J}$. 

Agora seja $C\in {\mathscr C}$ qualquer, induzindo uma rotulagem $C\upharpoonright\sigma$ sobre $\sigma$. Usamos o truque de condicionamento.
Para $J$ e $\sigma_J$ fixos, se $\mu({\mathcal H}(\sigma_J)\Delta C)>\ve$, ent\~ao a probabilidade que $\mu_{\sigma_{[n]\setminus J}}({\mathcal H}(\sigma_J)\Delta C)=0$ \'e menos de $(1-\ve)^{n-j}$. Isso significa que a probabilidade do evento $\sigma_J = \kappa(C\upharpoonright\sigma)$ \'e tamb\'em $<(1-\ve)^{n-j}$. A cota de uni\~ao termina a prova da primeira desigualdade. A segunda segue do lema \ref{l:<}. \qed

\begin{exercicio}
Estimar acima a complexidade amostral da regra ${\mathcal L} = {\mathcal H}\circ\kappa\circ\iota$ no teorema \ref{t:littlestone-warmuth}. Por exemplo, tentar deduzir a estimativa de \citep*{MY} (Th. 1.2):
\[s(\ve,\delta) \leq \frac{8}{\ve}\left(d\log\left(\frac 2\ve\right)+\log\left(\frac 1{\delta}\right)\right).\]
\label{ex:estimativaMY}
\end{exercicio}

\subsection{Dimens\~ao VC um} 
Mostremos que toda classe de dimens\~ao VC um admite um esquema de compress\~ao amostral de tamanho um, com boas propriedades de mensurabilidade.

\begin{exercicio}
Mostrar que a diferen\c ca sim\'etrica 
\[A\Delta B = (A\setminus B)\cup (B\setminus A)\]
\'e associativa,
\[(A\Delta B)\Delta C = A\Delta (B\Delta C)\]
e distributiva sobre interse\c c\~ao:
\[A\cap (B\,\triangle \,C)=(A\cap B)\,\triangle \,(A\cap C).\]
\end{exercicio}

\begin{lema}
Seja $\mathscr C$ uma classe de conceitos, e $D\subseteq\Omega$ um conceito qualquer. A dimens\~ao de Vapnik--Chervonenkis da classe
\[{\mathscr C}\Delta D = \{C\Delta D\colon C\in {\mathscr C}\}\]
\'e igual \`a de $\mathscr C$.
\end{lema}

\begin{proof}
Sejam $A$ um subconjunto fragmentado por $\mathscr C$, e $B\subseteq A$. Existe $C\in {\mathscr C}$ tal que $C\cap A = B\Delta (D\cap A)$. Agora temos:
\begin{align*}
A\cap(C\Delta D) &= (A\cap C)\Delta (A\cap D) \\
&= (B\Delta (D\cap A))\Delta (D\cap A) \\
&= B\Delta\emptyset \\
&= B.
\end{align*}
Segue-se que $\VC(\mathscr C)\leq\VC({\mathscr C}\Delta D)$. Como $({\mathscr C}\Delta D)\Delta D = {\mathscr C}$, obtemos a igualdade.
\end{proof}

\begin{lema}
Seja $\mathcal H$ uma aplica\c c\~ao de descompress\~ao de tamanho $d$, e $D\subseteq\Omega$ um conceito qualquer. A formula
\[[\Omega]^{\leq d}\ni\tau\mapsto {\mathcal H}(\tau)\Delta D\subseteq \Omega\]
defina uma aplica\c c\~ao de descompress\~ao de tamanho $d$ cuja classe de conceitos associada \'e
\begin{equation}
{\mathscr C}_{{\mathcal H}\Delta D}={\mathscr C}_{\mathcal H}\Delta D.
\label{eq:chdeltad}
\end{equation}
\label{eq:esquemadeltad}
\end{lema}

\begin{proof}
Seja $\kappa$ uma aplica\c c\~ao de compress\~ao correspondente a $\mathcal H$. Definamos a aplica\c c\~ao
\begin{align*}
{^D}\kappa\colon [\Omega]^{<\infty}\times [\Omega]^{<\infty}
 &\to [\Omega]^{\leq d},\\
{^D}\kappa(\sigma,\tau) &= \kappa(\sigma,\tau\Delta (D\cap\sigma)).
\end{align*}
Para qualquer elemento $C\Delta D\in {\mathscr C}_{\mathcal H}\Delta D$ e qualquer $\sigma\in [\Omega]^{<\infty}$, temos
\begin{align*}
\sigma\cap 
\left[{\mathcal H}\left({^D}\kappa(\sigma,\sigma\cap (C\Delta D)\right)\Delta D\right] &= 
\sigma\cap 
\left[
{\mathcal H}\left(\kappa(\sigma,(\sigma\cap C)\Delta (\sigma\cap D)\Delta (D\cap\sigma)\right)\Delta D\right]\\
&= \sigma\cap 
\left[{\mathcal H}\left(\kappa(\sigma,\sigma\cap C\right)\Delta D\right] \\
&= \left[\sigma\cap 
{\mathcal H}\left(\kappa(\sigma,\sigma\cap C\right)\right]\Delta (\sigma\cap D)
\\
&= (\sigma\cap C)\Delta (\sigma\cap D) \\
&= \sigma\cap (C\Delta D).
\end{align*}
Isto estabelece a inclus\~ao $\supseteq$ na eq. (\ref{eq:chdeltad}). Mas a situa\c c\~ao \'e sim\'etrica: $({\mathcal H}\Delta D)\Delta D={\mathcal H}$.
\end{proof}

\begin{lema}
Seja $\mathscr C$ uma classe de conceitos tal que $\emptyset\in {\mathscr C}$ e $\VC(\mathscr C)=1$. Formemos a classe ${\mathscr C}^{\cap}$ que consiste de todas as interse\c c\~oes de subfam\'\i lias n\~ao vazias de $\mathscr C$:
\[{\mathscr C}^{\cap} = \{\cap{\mathscr C}^\prime: \emptyset\neq {\mathscr C}^\prime\subseteq {\mathscr C}\}.\]
Ent\~ao,
\[\VC({\mathscr C}^{\cap})=1.\]
\label{l:c^cap}
\end{lema}

\begin{proof}
Suponha que existe um conjunto, $\{a,b\}$, com dois pontos distintos fragmentado pela classe ${\mathscr C}^{\cap}$. Em particular, existem $C^\prime_1,C^\prime_2,C^\prime_3\in {\mathscr C}^{\cap}$ tais que $C^\prime_1\cap \{a,b\}=\{a\}$, $C^\prime_2\cap \{a,b\}=\{b\}$, e $C^\prime_3\supseteq \{a,b\}$. A defini\c c\~ao da classe ${\mathscr C}^{\cap}$ implica imediatamente que existem $C_1,C_2,C_3\in {\mathscr C}$ tais que $C_1\cap \{a,b\}=\{a\}$, $C_2\cap \{a,b\}=\{b\}$, e $C_3\supseteq \{a,b\}$. Como $\emptyset\in {\mathscr C}$, conclu\'\i mos: $\mathscr C$ fragmenta $\{a,b\}$, o que \'e imposs\'\i vel.
\end{proof}

\begin{teorema}
Cada classe $\mathscr C$ de dimens\~ao $\VC(\mathscr C)=1$ admite um esquema de compress\~ao amostral de tamanho $1$.
\label{t:vcdim1admiteesquema}
\end{teorema}

\begin{proof}
Segundo lema \ref{eq:esquemadeltad}, podemos supor sem perda de generalidade que $\emptyset\in\mathscr C$; se n\~ao for o caso, substituamos a classe ${\mathscr C}\Delta D$ para $D\in{\mathscr C}$ qualquer. 
Segundo lema \ref{l:c^cap}, a classe maior ${\mathscr C}^{\cap}\supseteq {\mathscr C}$ tem dimens\~ao VC um e \'e fechado pelas interse\c c\~oes.

A esquema de descompress\~ao $\mathcal H$ vai associar a cada ponto $x$ a interse\c c\~ao de todos os conceitos que cont\'em $x$:
\[{\mathcal H}_x = \cap\{C\colon C\in {\mathscr C}^{\cap}\}.\]
Seja $C\in {\mathscr C}^{\cap}$ qualquer. A rela\c c\~ao entre elementos de $C$,
\begin{align*}
x\prec y &\iff {\mathcal H}_x\subseteq {\mathcal H}_y \\
&\iff x\in {\mathcal H}_y
\end{align*}
\'e uma pr\'e-ordem. Seja $\sigma$ uma amostra finita qualquer. Escolhamos um elemento maximal, $x\in C\cap\sigma$, com rela\c c\~ao \`a pr\'e-ordem $\prec$. Supondo que $C_x\cap\sigma\neq C\cap\sigma$, existe $y\in C\cap\sigma$ que \'e incompar\'avel com $x$. Mas neste caso o conjunto $\{x,y\}$ \'e fragmentado por ${\mathscr C}^{\cap}$:
\[\{x,y\}\subseteq C,~C_x\cap \{x,y\}=\{x\},~C_y\cap\{x,y\}=\{y\},~\emptyset\cap \{x,y\}=\emptyset,\]
uma contradi\c c\~ao. 
\end{proof}

\begin{observacao}
Parece que o resultado deve ser creditado a \citep*{FW}, veja p. 295. Ele foi redescoberto algumas vezes, inclusive pelo autor e Damjan Kalajdzievski em 2011 (nunca publicado).
\end{observacao}

\begin{observacao}
Seja $\mathscr C$ uma classe universalmente separ\'avel de dimens\~ao de Vapnik--Chervonenkis um.
Verifique que a constru\c c\~ao acima resulta em um esquema de compress\~ao boreliano. Neste caso, a solu\c c\~ao do problema \'e, ent\~ao, completamente satisfat\'oria.
\end{observacao}

\subsection{Teorema de compacidade: redu\c c\~ao aos dom\'\i nios finitos}

O resultado seguinte mostra que, se esquecermos da mensurabilidade, o problema de exist\^encia dos esquemas de compress\~ao pode ser reduzida ao problema finito. Para uma classe $\mathscr C$ de conceitos de um dom\'\i nio $\Omega$ e um subdom\'\i nio $\Upsilon\subseteq\Omega$, denotemos por ${\mathscr C}\upharpoonright\Upsilon$ a classe de todos os conceitos $C\cap\Upsilon$, $C\in {\mathscr C}$.

\begin{teorema}[Teorema de Compacidade \citep*{B-DL}]
Seja $\mathscr C$ uma classe de conceitos num dom\'\i nio $\Omega$ e $d\in\N$. Suponha que para cada subdom\'\i nio finito $F\subseteq\Omega$, a classe ${\mathscr C}\upharpoonright F$ admite um esquema de compress\~ao amostral de tamanho $d$ no dom\'\i nio $F$. Ent\~ao, $\mathscr C$ admite um esquema de compress\~ao amostral de tamanho $d$.
\label{t:compacidadeesquemas}
\index{teorema! de compacidade}
\end{teorema}

\begin{proof} Para cada $F\subseteq \Omega$ finito, escolhamos uma aplica\c c\~ao de descompress\~ao de tamanho $d$ para dom\'\i nio $F$, 
\[{\mathcal H}_F\colon [F]^{\leq d}\times F\to \{0,1\}.\]
Suponhamos que $\Omega$ \'e infinito (caso contr\'ario, h\'a nada a mostrar). 
Neste caso, a fam\'\i lia de todos os intervalos superiores semiinfinitos,
\[I_F = \{G\in [\Omega]^{<\infty}\colon F\subseteq G\},\]
\'e facilmente verificada para ser centrada. Segundo proposi\c c\~ao \ref{p:centradomaximal}, existe um ultrafiltro (fam\'\i lia centrada maximal), $\mathcal U$, sobre o conjunto $[\Omega]^{<\infty}$ que cont\'em todos os intervalos $I_F$, $F\in [\Omega]^{<\infty}$.

Agora sejam $\sigma\in [\Omega]^{d}$ uma amostra finita qualquer com $\leq d$ elementos em $\Omega$, e $x\in\Omega$. O conjunto $[\Omega]^{<\infty}$ admite uma parti\c c\~ao em tr\^es subconjuntos:

\begin{align*}
A_1 &= \{F\in [\Omega]^{<\infty}\colon {\mathcal H}_F(\sigma\cap F)(x) =1\},
\\
A_0 &= \{F\in [\Omega]^{<\infty}\colon {\mathcal H}_F(\sigma\cap F)(x) =0\},
\\
A_{?} & = \{F\in [\Omega]^{<\infty}\colon x\notin F\}.
\end{align*}
Segundo a escolha do ultrafiltro $\mathcal U$, ele cont\'em o intervalo $I_{\{x\}}=\Omega\setminus A_{?}$, e segundo teorema \ref{t:AAc}, $A_0\cup A_1\in {\mathcal U}$. Corol\'ario \ref{c:gammaxi} implica que exatamente uma das duas possibilidades tem lugar: seja $A_0\in {\mathcal U}$, seja $A_1\in {\mathcal U}$.

Definamos o valor da aplica\c c\~ao de descompress\~ao
\[{\mathcal H}\colon [\Omega]^{\leq d}\times\Omega\to \{0,1\}\]
assim:
\[{\mathcal H}(\sigma)(x) = \begin{cases} 1,&\mbox{ se }
\{F\in [\Omega]^{<\infty}\colon {\mathcal H}_F(\sigma\cap F)(x) =1\}\in {\mathcal U},\\
0,&\mbox{ se }
\{F\in [\Omega]^{<\infty}\colon {\mathcal H}_F(\sigma\cap F)(x) =1\}\in {\mathcal U}.
\end{cases}
\]
Resta mostrar que todo $C\in {\mathscr C}$ verifica eq. (\ref{eq:condicaoC_H}). Seja $\sigma\in [\Omega]^{<\infty}$ qualquer, e $x\in\sigma$. Para todo $F\in I_\sigma$, a hip\'otese implica que pode escolher $\tau_F\subseteq\sigma$, $\sharp\tau_F\leq d$, tal que
\[{\mathcal H}_F(\tau_F)(x)=\chi_C(x).\]
Dado $\tau\in [\sigma]^{\leq d}$, definamos a fam\'\i lia
\[X_{\tau} = \{F\in I_\sigma\colon \tau_F = \tau\}.\]
Os conjuntos $X_{\tau}$, $\tau\in [\sigma]^{\leq d}$, junto com o conjunto $\Omega\setminus I_\sigma$, formam uma cobertura finita de $[\Omega]^{<\infty}$, e o mesmo argumento que acima permite concluir que existe o \'unico $\tau$ tal que $X_{\tau}\in {\mathcal U}$. Segue-se da nossa defini\c c\~ao de $\mathcal H$ que
\[{\mathcal H}(\tau)(x)=\chi_C.\]
\end{proof}

\begin{observacao}
Seja $\mathscr C$ uma classe universalmente separ\'avel, e suponha que para cada subdom\'\i nio finito $F\subseteq\Omega$, a classe ${\mathscr C}\upharpoonright F$ admite um esquema de compress\~ao amostral de tamanho $d$ em $F$. Segue-se que $\mathscr C$ admite um esquema de compress\~ao {\em boreliana} de tamanho $d$? A resposta \'e positiva para classes $d$-{\em m\'aximas} \citep*{kalajdzievski}.
\end{observacao}

\section{Classes extremais}

A exist\^encia de esquemas de compress\~ao de tamanho $d$ foi estabelecida em \citep*{KW} para classes de conceitos ditas $d$-m\'aximas. Uma classe $\mathscr C$ \'e $d${\em -m\'axima} se para cada amostra $\sigma\in [\Omega]^{<\infty}$ a cota no lema de Sauer--Shelah est\'a atingida:
\[\sharp{\mathscr C}\upharpoonright\sigma = \sum_{j=0}^d{n\choose j}.\]
Em particular, $\mathscr C$ tem dimens\~ao de Vapnik--Chervonenkis $d$. 

Como a prova \'e um pouco t\'ecnica, vamos apresentar ao inv\'es disso um resultado bem mais transparente: cada classe $d$-m\'axima admite um esquema de compress\~ao {\em rotulado} de tamanho $d$. Um esquema de compress\~ao rotulado tem, como o dom\'\i nio da aplica\c c\~ao de descompress\~ao $\mathcal H$ (e o codom\'\i nio da aplica\c c\~ao de compress\~ao $\kappa$) o conjunto de todas as amostras {\em rotuladas} de tamanho $\leq d$. Deste modo, a compress\~ao torna-se mais f\'acil.

O resultado para esquemas rotulados, mostrado originalmente em \citep*{FW}, foi recentemente generalizado por \citep*{MW} para classes de conceitos mais gerais, chamadas {\em classes extremais}. Neste contexto de geometria combinat\'oria, a prova torna-se muito elegante, e forma o assunto desta se\c c\~ao.

\subsection{Classes m\'aximas}

\begin{definicao}
Seja $\mathscr C$ uma classe de conceitos de dimens\~ao $VC$ $\leq d$. Diz-se que a classe $\mathscr C$ \'e 
\begin{itemize}
\item
$d$-{\em maximal} se ela n\~ao est\'a contida numa classe estritamente maior de dimens\~ao VC $d$, e
\item $d$-{\em m\'axima} se para cada amostra $\sigma\in [\Omega]^{<\infty}$ a cota no lema de Sauer--Shelah est\'a atingida:
\[\sharp{\mathscr C}\upharpoonright\sigma = \sum_{j=0}^d{n\choose j}.\]
\end{itemize}
\index{classe! m\'axima}
\index{classe! maximal}
\end{definicao}

Nenhuma propriedade implica a outra.

\begin{exemplo} (Emprestado de \citep*{FW}).
Definamos $\Omega = \{1, 2, 3, 4\}$, e
\[{\mathscr C} = \{\{1\},\{2\},\{1,2\},\{1,3\},\{2,3\},\{1,4\},\{2,4\},\{3,4\},
\{1,2,3\},\{1,2,4\}\}.\]
\'E f\'acil verificar que a classe $\mathscr C$ \'e $2$-maximal, por\'em n\~ao \'e $2$-m\'axima, pois $\sharp{\mathscr C}=11$, enquanto $\sum_{j=0}^2{4\choose j}=12$. 
\end{exemplo}

Segue-se que nem cada classe de dimens\~ao VC $d$ \'e contida numa classe $d$-{\em m\'axima}. 

\begin{exercicio} 
Use lema de Zorn (\ref{l:zorn}) para mostrar que cada classe de dimens\~ao VC $d$ \'e contida numa classe $d$-{\em maximal.} 
\end{exercicio}

\begin{exemplo}
A classe que consiste de todos os intervalos fechados em $\R$ \'e $2$-m\'aximo, mas n\~ao \'e $2$-maximal.
\end{exemplo}

Portanto, em dom\'\i nios {\em finitos} cada classe $d$-m\'axima \'e $d$-maximal.

\begin{exercicio}
Seja $\Omega$ um dom\'\i nio finito com $n$ elementos, e seja $\mathscr C$ uma classe de dimens\~ao VC $d$. Verifique que $\mathscr C$ \'e $d$-m\'axima se e somente se
\[\sharp{\mathscr C} = \sum_{j=0}^d {n\choose j}.\]
[ {\em Sugest\~ao:} para sufici\^encia, usar indu\c c\~ao em $n$ junto com as propriedades de base dos coeficientes binomiais. ]
Em particular, segue-se do lema de Sauer--Shelah que uma classe $d$-m\'axima num dom\'\i nio finito \'e $d$-maximal.
\end{exercicio}

\begin{definicao}
Seja $\mathcal H\colon [\Omega]^{\leq d}\to 2^{\Omega}$ uma aplica\c c\~ao de descompress\~ao. Diz-se que $\mathcal H$ {\em evite embates} se, dado $\tau_1,\tau_2\in [\Omega]^{\leq d}$, $\tau_1\neq\tau_2$, as hip\'oteses ${\mathcal H}(\tau_1)$ e ${\mathcal H}(\tau_2)$ s\~ao distintas j\'a na uni\~ao $\tau_1\cup\tau_2$:
\[{\mathcal H}(\tau_1)\cap (\tau_1\cup\tau_2) \neq {\mathcal H}(\tau_2)\cap (\tau_1\cup\tau_2).
\]
\end{definicao}

A prova do teorema de \citep*{KW} sobre a exist\^encia de um esquema de compress\~ao para classes $d$-m\'aximas \'e baseada sobre a seguinte observa\c c\~ao de base.

\begin{exercicio}
Seja $\mathcal H\colon [\Omega]^{\leq d}\to {\mathscr C}$ uma aplica\c c\~ao qualquer, onde $\mathscr C$ \'e uma classe $d$-m\'aximo. Se $\mathcal H$ evite embates, mostre que ${\mathscr C}\subseteq {\mathscr C}_{\mathcal H}$.
\end{exercicio}

No entanto, deixemos o resultado mencionado acima, e provemos um resultado menos tecnicamente desafiador, por\'em muito instrutivo.

\subsection{Esquemas de compress\~ao rotulados}

Relembremos que $\ls\Omega d$ denota o conjunto de todas as amostras rotuladas de tamanho $\leq d$, e de mesmo jeito \'e definido $\ls\Omega\infty$.

\begin{definicao}
Um esquema de compress\~ao amostral rotulado \'e um par $({\mathcal H},\kappa)$, onde 
\[{\mathcal H}\colon \ls\Omega d\to 2^{\Omega}\]
\'e a aplica\c c\~ao de descompress\~ao, e 
\[\kappa\colon \ls\Omega\infty\to \ls\Omega d,\]
a aplica\c c\~ao de compress\~ao, que tem as propriedades: $\kappa(\sigma,\tau)$ \'e uma subamostra de $\sigma$ cujo r\'otulo \'e induzido por $\tau$, e
\begin{equation}
\forall C\in {\mathscr C},~\forall\sigma\in [\Omega]^{\infty},~{\mathcal H}(\kappa(\sigma,C\cap\sigma))\cap \sigma = C\cap\sigma.
\label{eq:forallsigmaconsistente}
\end{equation}
\index{esquema! de compress\~ao amostral! rotulado}
\end{definicao}

\begin{observacao}
Da maneira equivalente, a partir de uma aplica\c c\~ao $\mathcal H$ como acima, pode-se definir a classe ${\mathscr C}_{\mathcal H}$, como na eq. (\ref{eq:condicaoC_H}), e depois terminar a defini\c c\~ao requerendo que $\kappa(\sigma,\tau)$ seja uma subamostra rotulada de $(\sigma,\tau)$, e ${\mathscr C}\subseteq {\mathscr C}_{\mathcal H}$.
\end{observacao}

\begin{observacao}
Cada esquema de compress\~ao amostral $({\mathcal H},\kappa)$ n\~ao rotulada defina um esquema de descompress\~ao amostral rotulada, cuja aplica\c c\~ao de descompress\~ao \'e a composi\c c\~ao de $\mathcal C$ com a aplica\c c\~ao esquecendo a rotulagem:
\[\ls\Omega\infty \ni (\sigma,\tau)\mapsto \sigma\in[\Omega]^{<\infty}.\]
A aplica\c c\~ao de compress\~ao simplesmente adiciona o r\'otulo induzido de $\sigma$:
\[\ls\Omega\infty \ni (\sigma,\tau)\mapsto (\kappa(\sigma,\tau),\tau\cap \kappa(\sigma,\tau)).\]
\end{observacao}

\begin{observacao} 
Agora os resultados sobre os esquemas n\~ao rotulados se generalizam para esquemas rotulados. Alguns deles (como teorema \ref{t:vcdim1admiteesquema}) valem trivialmente. O leitor pode ser interessado de mostrar os resultados an\'alogos ao exerc\'\i cio \ref{ex:descompressaoVC}, teorema \ref{t:littlestone-warmuth}, e Teorema de Compacidade \ref{t:compacidadeesquemas}.
\end{observacao}

\begin{observacao}
O exemplo motivador de um esquema de compress\~ao rotulada \'e o famoso classificador de M\'aquina de Vetores de Suporte (Support Vector Machine, SVM). Deste ponto de vista, ele est\'a discutido em \citep*{vLBS}.
\end{observacao}

\subsection{Cubos de conceitos}

Teorema de Compacidade \ref{t:compacidadeesquemas} permite apenas trabalhar com dom\'\i nios finitos, e isto vai ser a nossa hip\'otese at\'e o final da se\c c\~ao.

\begin{definicao}
Seja $\mathscr C$ uma classe de conceitos num dom\'\i nio $\Omega$. Um {\em cubo} em $\mathscr C$ com o conjunto de coordenadas $D\subseteq\Omega$ \'e uma classe $\mathscr B\subseteq\mathscr C$ tal que:
\begin{enumerate}
\item $\mathscr B$ fragmenta $D$, e
\item $\mathscr B\vert_{\Omega\setminus D}$ consiste de \'unico conceito, que vamos denotar $\mbox{tag}(\mathscr B)$.
\end{enumerate}
\index{cubo! de conceitos}
\end{definicao}

Em outras palavras, se pensamos do cubo $\mathscr B$ como um subconjunto do cubo de Hamming $\{0,1\}^{\Omega}$, ele pode ser escrito assim:
\[\mathscr B = \{0,1\}^D\times \{\chi_{\mbox{tag}(\mathscr B)}\}.\]
As restri\c c\~oes dos elementos do cubo sobre $D$ podem tomar quaisquer valores, mas sobre $\Omega\setminus D$, \'e sempre apenas uma fun\c c\~ao bin\'aria fixa. Deste modo, $\mathscr B$ pode ser identificado com o cubo de Hamming, $\{0,1\}^D$: a aplica\c c\~ao
\[{\mathscr B}\ni C\mapsto [D\ni x\mapsto \chi_C(x)]\in \{0,1\}^D\]
\'e bijetora.

\begin{definicao}
Um conjunto $D\subseteq\Omega$ \'e {\em fortemente fragmentado} por uma classe $\mathscr C$ se existe um cubo ${\mathscr B}\subseteq {\mathscr C}$ tendo $D$ como conjunto de coordenadas.
\index{conjunto! fortemente fragmentado}
\end{definicao}

Obviamente, um conjunto fortemente fragmentado \'e fragmentado por $\mathscr C$.

\begin{exercicio}
Construir um exemplo simples (com o dom\'\i nio $\Omega$ de dois pontos) de um conjunto fragmentado mas n\~ao fortemente fragmentado por uma classe.
\end{exercicio}

\begin{teorema}
Seja $\mathscr C$ uma classe de conceitos num dom\'\i nio finito, $\Omega$.
Denotemos por $\mbox{sh}({\mathscr C})$ a fam\'\i lia de subconjuntos de $\Omega$ fragmentados por $\mathscr C$, e $\mbox{ssh}({\mathscr C})$ a fam\'\i lia de subconjuntos fortemente fragmentados por $\mathscr C$. Ent\~ao,
\[\sharp \mbox{ssh}({\mathscr C})\leq \sharp {\mathscr C} \leq \sharp \mbox{sh}({\mathscr C}).\]
\label{t:sandwich}
\end{teorema}

A segunda desigualdade \'e o teorema de Pajor \ref{t:pajor}. Cada das desigualdades pode ser de fato deduzida da outra (aplicada \`a classe complementar).

\begin{exercicio}
Mostre que para um conjunto $A\subseteq\Omega$ uma e uma s\'o condi\c c\~ao \'e verdadeira:
\begin{itemize}
\item $\mathscr C$ fragmenta $A$,
\item $2^\Omega\setminus {\mathscr C}$ fortemente fragmenta $\Omega\setminus A$.
\end{itemize}
\label{ex:cocubo}
\end{exercicio}

\begin{exercicio}
Deduza do exerc\'\i cio \ref{ex:cocubo} o seguinte:
a primeira desigualdade (chamada em \citep*{BoR} a desigualdade de Sauer reversa) vale para uma classe $\mathscr C$ se e somente se a segunda desigualdade vale para a classe complementar $2^{\Omega}\setminus {\mathscr C}$.
De mesmo, para igualdades.
\label{ex:dualidadesauer}
\end{exercicio}

Isto termina a prova do teorema \ref{t:sandwich}.

\begin{definicao}
Uma classe $\mathscr C$ \'e dita {\em extremal} se todo subconjunto fragmentado por $\mathscr C$ \'e fortemente fragmentado, ou seja, se as duas desigualdades no teorema \ref{t:sandwich} s\~ao igualdades.
\index{classe! extremal}
\end{definicao}

Mostremos que toda classe $d$-m\'aximo \'e extremal. Com este prop\'osito, verifiquemos que as duas igualdades s\~ao de fato equivalentes j\'a quando aplicadas \`a mesma classe (e n\~ao apenas \`a complementar). Vamos usar a t\'ecnica da prova original \citep*{sauer} do lema de Sauer--Shelah \ref{l:sauer-shelah}, um pouco mais complicada do que a prova do teorema de Pajor \ref{t:pajor}.

\subsection{Compress\~ao para baixo\label{ss:prabaixo}} 

Para cada fam\'\i lia $\mathscr C$ de subconjuntos de $\sigma$ e cada $x\in\sigma$, definamos a fam\'\i lia $T_x({\mathscr C})$ como segue. O operador $T_x$ remove o ponto $x$ de todo conjunto $A\in{\mathscr C}$, exceto quando $A\setminus\{x\}$ j\'a pertence a $\mathscr C$. Neste caso, $T_x$ devolve $A$. Formalmente,
\begin{equation}
T_x({\mathscr C})=\left\{A\setminus\{x\}\colon A\in{\mathcal H} \right\}\cup\left\{A\colon A\in {\mathscr C}\mbox{ e } A\setminus\{x\}\in {\mathscr C}\right\}.
\end{equation}
\index{compress\~ao! para baixo}

\begin{lema}
O operador $T_x$ \'e injetor sobre $\mathscr C$.
\label{cl:card}
\end{lema}

\begin{proof}
As imagens de $A$ e de $B$ s\~ao iguais apenas se um deles \'e obtido do outro deletando $x$, por exemplo, $B=A\setminus \{x\}$. Por\'em, neste caso $T_x(A)=A$ e $T_x(B)=B$.
\end{proof}
 
Definamos o {\em peso} da fam\'\i lia $\mathscr C$ como a soma de cardinalidades de todos os conjuntos:
\[w({\mathscr C}) = \sum_{C\in\mathscr C}\sharp C.\]

O seguinte \'e \'obvio.

\begin{lema} 
\begin{equation}
\label{eq:sumtotal}
w(\mathscr C) \leq w(T_x(\mathscr C)). 
\end{equation}
S\'o h\'a igualdade no caso $T_x({\mathscr C})={\mathscr C}$. 
\end{lema}

\begin{lema}
$\mbox{sh}(T_x({\mathscr C}))\subseteq \mbox{sh}({\mathscr C})$. Em particular,
$\VC(T_x({\mathscr C}))\leq\VC({\mathscr C})$.
\label{cl:then}
\end{lema}

\begin{proof}
Suponha $A\subseteq\sigma$ \'e fragmentado por $T_x({\mathscr C})$. Mostremos que $A$ \'e fragmentado por $\mathscr C$. Seja $B\subseteq A$. Procuremos $C\in {\mathscr C}$ tal que $A\cap C=B$. 
\par
{\em Caso 1.} $x\notin A$.
\\[1mm]
Existe $C\in T_x({\mathscr C}^\ast)$ com $A\cap C=B$, e pela defini\c c\~ao de $T_x({\mathscr C}^\ast)$, ou $C\cup\{x\}\in \mathscr C$ ou $C\in\mathscr C$. A interse\c c\~ao de qualquer conjunto com $A$ \'e igual a $B$.
\par
{\em Caso 2.} $x\in B$.
\\[1mm]
Fixe $C\in T_x({\mathscr C}^\ast)$ com $A\cap C=B$. Neste caso, $x\in C$, significando que $C\in {\mathscr C}$.
\par
{\em Caso 3.} $x\in A\setminus B$.
\\[1mm]
Como $B\cup\{x\}\subseteq A$ e $A$ \'e fragmentado por $T_x({\mathscr C}^\ast)$, esta fam\'\i lia cont\'em $C$ tal que $A\cap C = B\cup\{x\}$. Este $C$ cont\'em $x$, o que implica que $C$ e $C\setminus\{x\}$ pertencem a $\mathscr C$. Como $(C\setminus\{x\})\cap A = B$, conclu\'\i mos. 
\end{proof}

O comportamento de conjuntos fortemente fragmentados \'e dual.

\begin{lema}
$\mbox{ssh}(T_x({\mathscr C}))\supseteq \mbox{ssh}({\mathscr C})$.
\label{cl:then2}
\end{lema}

\begin{proof}
Seja $A\subseteq\sigma$ um conjunto fortemente fragmentado por $\mathscr C$. De modo equivalente, $\sigma\setminus A$ n\~ao \'e fragmentado por $2^\sigma\setminus{\mathscr C}$ (exerc\'\i cio \ref{ex:cocubo}): existe $B\subset\sigma$, $B\cap A=\emptyset$, tal que para qualquer $C\in {\mathscr C}$, $C\setminus A\neq B$.
Suponha, para obter uma contradi\c c\~ao, que $\sigma\setminus A$ \'e fragmentado pela classe $2^\sigma\setminus T_x{\mathscr C}$. Em particular, existe $C\in {\mathscr C}$ tal que $T_x(C)\setminus A = B$. Cado conceito $C$ com esta propriedade deve satisfazer as propriedades seguintes:
\begin{enumerate}
\item $x\notin B$, 
\item $x\notin A$,
\item $x\in C$,
\item $C\setminus \{x\}\notin {\mathscr C}$.
\end{enumerate}
Ao mesmo tempo, existe $C^\prime\in {\mathscr C}$ tal que $T_x(C^\prime)\setminus A = B\cup\{x\}$. Isso implica que $C^\prime\setminus\{x\}\in {\mathscr C}$, e $C^\prime\setminus\{x\}\setminus A=B$, a contradi\c c\~ao.
\end{proof}

Agora denotemos por $T=T_{x_n}T_{x_{n-1}}\ldots T_{x_2}T_{x_1}$, onde $\sigma=\{x_1,x_2,\ldots,x_n\}$. Apliquemos $T$ recursivamente \`a fam\'\i lia de conjuntos $\mathscr C$ at\'e tal itera\c c\~ao $N$ que o peso de $\mathscr C$ se estabilize. 

\begin{lema} 
J\'a uma itera\c c\~ao basta, ou seja, a fam\'\i lia $T({\mathscr C})$ atinge o peso minimal. 
\end{lema}

\begin{proof}
Chamemos uma fam\'\i lia $\mathscr C$ {\em $x$-est\'avel} para $x\in\Omega$ se $T_x({\mathscr C})={\mathscr C}$. De modo equivalente, como $\mathscr C$ \'e finita, para todo $C\in{\mathscr C}$ temos $T_x(C)=C$. Por exemplo, cada fam\'\i lia da forma $T_x({\mathscr C})$ \'e $x$-est\'avel.
Mostremos que se $\mathscr C$ \'e $x$-est\'avel, ent\~ao para cada $y\in\Omega$ a fam\'\i lia $T_y({\mathscr C})$ \'e $x$-est\'avel tamb\'em, o que termina a prova pela indu\c c\~ao finita.

Seja $C\in{\mathscr C}$. Basta mostrar que a imagem de $C$ por $T_y$ tem a propriedade $T_y(C)\setminus\{x\}\in T_y({\mathscr C})$. \'E verdadeiro de modo trivial se $x\notin C$. Suponhamos que $x\in C$. A $x$-estabilidade de $\mathscr C$ implica $C\setminus\{x\}\in {\mathscr C}$. 
Tem dois casos a estudar.

{\em Caso 1.} $T_y(C)=C$. Isso significa $C\setminus\{y\}\in {\mathscr C}$, logo o conjunto $C\setminus\{y\}$ \'e fixo pelo operador $T_x$. Conclu\'\i mos: $C\setminus\{x,y\}\in {\mathscr C}$, e por conseguinte $T_y(C\setminus\{x\})=C\setminus\{x\}$, como desejado.

{\em Caso 2.} $T_y(C) = C\setminus\{y\}$. Se $T_y(C\setminus\{x\}) = C\setminus\{x,y\}$, ent\~ao $T_y(C)\setminus\{x\} = C\setminus\{x,y\}\in T_y({\mathscr C})$ e tudo est\'a bem. Se, ao contr\'ario, $T_y(C\setminus\{x\})=C\setminus\{x\}$, ent\~ao $C\setminus\{x,y\}\in {\mathscr C}$, logo $C\setminus\{x,y\}\in T_y({\mathscr C})$, e chegamos \`a mesma conclus\~ao.
\end{proof}

\begin{observacao}
Exemplos simples (com $\sigma$ tendo apenas tr\^es pontos distintos) mostram que a fam\'\i lia $T({\mathscr C})$ \'e altamente n\~ao \'unica: ela depende da ordem de atua\c c\~ao dos operadores $T_x$, $x\in\sigma$. Dado uma permuta\c c\~ao $\tau\in S_n$ de $[n]$, escrevemos
\[T_{\tau} = 
T_{x_{\tau(n)}}T_{x_{\tau(n-1)}}\ldots T_{x_{\tau(2)}}T_{x_{\tau(1)}}.\]
\end{observacao}

Como o operador $T_x$, $x\in\sigma$, n\~ao muda mais a fam\'\i lia $T_{\tau}({\mathscr C})$, conclu\'\i mos: para cada $x\in\sigma$ e $A\in T_{\tau}({\mathscr C})$, 
\[A\setminus\{x\}\in T_{\tau}({\mathscr C}).\]
Por conseguinte, se $C\in T_{\tau}({\mathscr C})$ e $B\subseteq A$, ent\~ao $B\in T_{\tau}({\mathscr C})$. A fam\'\i lia $T_{\tau}({\mathscr C})$ \'e {\em fechada para baixo.} Em particular, ela n\~ao cont\'em algum conjunto $A$ de cardinalidade $d+1$ ou mais, e deste modo, \'e contida na bola de Hamming fechada de raio $d$. (Isto traz imediatamente a estimativa do lema de Sauer--Shelah).

\begin{lema}
Seja $\mathscr C$ uma classe de conceitos no dom\'\i nio $\sigma=\{x_1,\ldots,x_n\}$.
Sejam $I=\{x_{i_1},x_{i_2},\ldots,x_{i_k}\}\subseteq\sigma$, e $\sigma\setminus I = \{x_{j_1},\ldots,x_{j_{n-k}}\}$. Ent\~ao 
\begin{enumerate}
\item $I$ \'e fragmentado por $\mathscr C$ se e somente se $I\in T_{(i_1,i_2,\ldots,i_k,j_1,j_2,\ldots,j_{n-k})}(\mathscr C)$,
\item $I$ \'e fortemente fragmentado por $\mathscr C$ sse $I\in T_{(j_1,j_2,\ldots,j_{n-k},i_1,i_2,\ldots,i_k)}(\mathscr C)$.
\end{enumerate}
\label{l:Ishssh}
\end{lema}

\begin{proof}
(1)
O conjunto $I$ sendo fragmentado por $\mathscr C$ significa que para cada $A\subseteq I$ existe conjunto $C_A\in {\mathscr C}$ tal que $C_A\cap I = A$. A aplica\c c\~ao de operadores $T_{x}$ com $x$ fora de $I$ n\~ao afeita as interse\c c\~oes com $I$, e denotando $T_J=T_{x_{j_1}}\ldots T_{x_{j_{n-k}}}$, ainda temos $T_J(C_A)\cap I=A$. Como a classe $T_J(\mathscr C)$ \'e fechado pelas remo\c c\~oes de pontos $x\in\sigma\setminus I$, para cada $A\subseteq I$, existe $C^\prime_A\in {\mathscr C}$ tal que $T_J(C^\prime_A)=A$. Em outras palavras, os elementos de  $T_J(\mathscr C)$ que s\~ao subconjuntos de $I$ formam uma fam\'\i lia fechada por baixo, e por conseguinte eles pertencem \`a classe $T_{IJ}(\mathscr C)$ tamb\'em. Isto estabelece a necessidade ($\Rightarrow$). A sufici\^encia ($\Leftarrow$) segue de fato que $T_{IJ}(\mathscr C)$ \'e fechado por baixo, e a inclus\~ao $\mbox{sh}(T_{IJ}(\mathscr C))\subseteq \mbox{sh}(\mathscr C)$ (lema \ref{cl:then}).

(2) Agora suponha que $I$ \'e fortemente fragmentado por $\mathscr C$. Existe um conjunto $C\subseteq J=\sigma\setminus I$ tal que, qual quer seja $A\subseteq I$, o conjunto $C\cap A$ pertence a $\mathscr C$. \'E claro que os operadores $T_x$, $x\in I$ n\~ao mudam os elementos do cubo $\{C\}\times 2^I$. Depois da aplica\c c\~ao de operadores $T_x$, $x\in J$, obtemos uma fam\'\i lia de conjuntos $T_{JI}(\mathscr C)$, fechada por baixo, que ainda fragmenta $I$, logo $I\in T_{JI}(\mathscr C)$. A necessidade \'e mostrada. Para sufici\^encia, suponha que $I\in T_{JI}(\mathscr C)$. Ent\~ao, $T_{JI}(\mathscr C)$ fragmenta $I$, e por isso j\'a $T_{I}(\mathscr C)$ fragmenta $I$, logo $T_{I}(\mathscr C)={\mathscr C}$. Em particular, existe $C^\prime\in {\mathscr C}$ tal que $I\subseteq C^\prime$. Definamos $C=C^\prime\setminus I$. Como a fam\'\i lia $\mathscr C$ \'e fechada sob operadores $T_x$, $x\in I$, conclu\'\i mos que, qualquer seja $A\subseteq I$, o conjunto $C\cup A$ pertence a $\mathscr C$.
\end{proof}

\begin{teorema}[\citet*{BoR}]
Para uma classe de conceitos $\mathscr C$ num dom\'\i nio finito $\sigma$, as condi\c c\~oes seguintes s\~ao equivalentes.
\begin{enumerate}
\item\label{condicao:1} $\mathscr C$ \'e extremal, ou seja, $\sharp \mbox{ssh}({\mathscr C})= \sharp ({\mathscr C})=\sharp \mbox{sh}({\mathscr C})$.
\item\label{condicao:2} $\sharp {\mathscr C} = \sharp \mbox{sh}({\mathscr C})$.
\item\label{condicao:3} $\sharp \mbox{ssh}(2^\sigma\setminus{\mathscr C})=\sharp (2^\sigma\setminus{\mathscr C})$.
\item \label{condicao:4}$\mathscr C$ tem a \'unica compress\~ao para baixo. Em outras palavras, quaisquer sejam permuta\c c\~oes $\tau,\tau^\prime\in S_n$, $T_{\tau}({\mathscr C})=T_{\tau^\prime}({\mathscr C})$.
\item \label{condicao:5}$\sharp \mbox{ssh}({\mathscr C})= \sharp ({\mathscr C})$.
\end{enumerate}
\label{t:bollobas-radcliffe}
\index{teorema! de Bollob\'as--Radcliffe}
\end{teorema}

\begin{proof}
(\ref{condicao:1}) $\Rightarrow$ (\ref{condicao:2}) \'e trivial, e a equival\^encia (\ref{condicao:2})$\iff$(\ref{condicao:3}) j\'a foi estabelecida (exerc\'\i cio \ref{ex:cocubo}). 

Assuma (\ref{condicao:2}), ou seja, $\sharp{\mathscr C}=\sharp\mbox{sh}({\mathscr C})$. Para qualquer permuta\c c\~ao $\tau\in S_n$, $\mbox{sh}(T_{\tau}({\mathscr C}))=T_{\tau}({\mathscr C})$ e $T_{\tau}$ \'e injetora. Temos:
\[\sharp\mbox{sh}(T_{\tau}({\mathscr C})) = \sharp T_{\tau}({\mathscr C})
=\sharp{\mathscr C}=\sharp\mbox{sh}({\mathscr C}).\]
Lema \ref{cl:then} implica que $\mbox{sh}(T_{\tau}({\mathscr C}))\subseteq \mbox{sh}({\mathscr C})$, e por conseguinte
\[T_{\tau}({\mathscr C})=\mbox{sh}(T_{\tau}({\mathscr C})) = \mbox{sh}({\mathscr C})\]
qualquer seja permuta\c c\~ao $\tau\in S_n$, estabelecendo (\ref{condicao:4}).

Assumamos a condi\c c\~ao (\ref{condicao:4}). Usando o lema \ref{l:Ishssh}, conclu\'\i mos que um conjunto $I$ \'e fragmentado por $\mathscr C$ se e somente se $I$ \'e fortemente fragmentado por $\mathscr C$, o que estabelece (\ref{condicao:1}) e em particular (\ref{condicao:5}). Neste momento, as condi\c c\~oes (\ref{condicao:1})--(\ref{condicao:4}) s\~ao dois a dois equivalentes.

Finalmente, assumamos a validade de (\ref{condicao:5}). Usando a implica\c c\~ao j\'a mostrada, (\ref{condicao:3})$\Rightarrow$(\ref{condicao:1}), aplicada \`a classe complementar $2^{\sigma}\setminus{\mathscr C}$, conclu\'\i mos que esta classe \'e extremal, ou seja, $\sharp\mbox{ssh}(2^{\sigma}\setminus{\mathscr C})=\sharp(2^{\sigma}\setminus{\mathscr C})$. Exerc\'\i cio \ref{ex:dualidadesauer} implica condi\c c\~ao (\ref{condicao:2}) para $\mathscr C$.
\end{proof}

\begin{corolario}
Cada classe $d$-m\'axima \'e extremal.
\end{corolario}

\begin{proof}
Para uma classe $d$-m\'axima, $\sharp{\mathscr C}= \sum_{j=0}^d{n\choose j}$, o que \'e exatamente o n\'umero de subconjuntos com $\leq d$ elementos de $\sigma$, ou seja, subconjuntos fragmentados. Temos a igualdade $\sharp {\mathscr C} = \sharp \mbox{sh}({\mathscr C})$, e conclu\'\i mos pelo teorema \ref{t:bollobas-radcliffe}.
\end{proof}

\begin{exemplo}
Nem cada classe extremal de dimens\~ao VC $d$ \'e $d$-m\'axima. O exemplo mais simples \'e o de qualquer bola de Hamming em torno de zero que \'e estritamente intermedi\'aria entre a bola aberta e a bola fechada de raio $d$. Outra exemplo \'e qualquer cubo $\mathscr B$ cujo conjunto de coordenadas \'e um subconjunto pr\'oprio do dom\'\i nio, $D\subsetneq\Omega$.
\end{exemplo}

\begin{lema}
Seja $\mathscr C$ uma classe extremal, e seja $\sigma\subseteq\Omega$. Ent\~ao a classe $\mathscr C\upharpoonright\sigma$ \'e extremal.
\end{lema}

\begin{proof}
Um subconjunto $A\subseteq\sigma$ fragmentado por ${\mathscr C}\upharpoonright\sigma$ \'e fragmentado por $\mathscr C$, logo fortemente fragmentado por $\mathscr C$. Existe um cubo $\{C\}\times \{0,1\}^A\subseteq {\mathscr C}$, once $C\subseteq \Omega\setminus A$. Por conseguinte, o cubo $\{C\cap\sigma\}\times \{0,1\}^A$ \'e uma subfam\'\i lia de ${\mathscr C}\upharpoonright\sigma$, ou seja, $A$ \'e fortemente fragmentado por ${\mathscr C}\upharpoonright\sigma$.
\end{proof}

\begin{definicao}
Dado uma classe $\mathscr C$ e um subconjunto $D$ do dom\'\i nio $\Omega$, definamos a {\em redu\c c\~ao} de $\mathscr C$ modulo $D$ como uma classe no dom\'\i nio $\Omega\setminus D$ que consiste de todos os conceitos da forma $\mbox{tag}(\mathscr B)$ onde $\mathscr B\subseteq \mathscr C$ \'e um cubo com conjunto de coordenadas exatamente $D$. Nota\c c\~ao: ${\mathscr C}^{\backslash D}$.
\index{redu\c c\~ao! de uma classe}
\index{CD@${\mathscr C}^{\backslash D}$}
\end{definicao}

\begin{lema}
A redu\c c\~ao de uma classe extremal \'e uma classe extremal.
\label{l:reducaoextremal}
\end{lema}

\begin{proof}
Seja $I\subseteq \Omega\setminus D$ um subconjunto fragmentado por ${\mathscr C}^{\backslash D}$. Segue-se que $D\cup I$ \'e fragmentado por ${\mathscr C}$, logo fortemente fragmentado. Conclu\'\i mos da defini\c c\~ao de ${\mathscr C}^{\backslash D}$ que esta classe fortemente fragmenta $I$ em $\Omega\setminus D$.
\end{proof}

\begin{lema}
Seja $\mathscr B$ um cubo e $\mathscr C$ uma classe extremal, ambos no dom\'\i nio $\Omega$. Ent\~ao $\mathscr B\cap\mathscr C$ \'e uma classe extremal.
\end{lema}

\begin{proof}
Seja $D$ o conjunto de coordenadas do cubo $\mathscr B$. Podemos supor que $\mathscr B\cap\mathscr C$ n\~ao \'e vazia.
Suponha que $I\subseteq\Omega$ \'e fragmentado por $\mathscr B\cap\mathscr C$. Ent\~ao, em particular, $I\subseteq D$. Como $I$ \'e fragmentado por $\mathscr C$, ele \'e fortemente fragmentado, e segundo lema \ref{l:Ishssh}, $I$ pertence a $T_{\Omega\setminus D,D\setminus I,I}(\mathscr C)$. \'E claro que $T_{D\setminus I,I}(\mathscr C\cap\mathscr B)\cap D=T_{D\setminus I,I}(\mathscr C)\cap D$, logo este conjunto cont\'em $I$ (\ref{l:Ishssh},(2)). Como $\mathscr C\cap \mathscr B\setminus D$ consiste de \'unico conceito $\mbox{tag}(\mathscr B)$, e $T_{\Omega\setminus D,D\setminus I,I}(\mathscr C\cap\mathscr B)$ \'e fechado por baixo, conclu\'\i mos que $I\in T_{\Omega\setminus D,D\setminus I,I}(\mathscr C\cap\mathscr B)$ tamb\'em.
\end{proof}

\begin{definicao}
Cada classe de conceitos $\mathscr C$ suporta uma estrutura natural de grafo, nomeadamente, dois conceitos $C$ e $C^\prime$ s\~ao adjacentes se e somente se eles s\'o diferem em um ponto: $\sharp (C\Delta D^\prime)=1$. Em outras palavras, quando pensamos de $\mathscr C$ como um subconjunto do cubo de Hamming $\{0,1\}^\Omega$, a dist\^ancia de Hamming (n\~ao normalizada) entre $C$ e $C^\prime$ \'e exatamente $1$. Este grafo chama-se {\em grafo de $1$-inclus\~ao} ({\em $1$-inclusion graph}).
\index{grafo! de $1$-inclus\~ao}
\end{definicao}

\begin{definicao}
Um grafo (simples) \'e dito {\em conexo} se para cada par $x,y$ de v\'ertices, existe um {\em caminho} que junta $x$ e $y$, ou seja, uma sequ\^encia finita 
\[x=x_0,x_1,x_2,\ldots,x_n=y,\]
onde dois pontos consequentes s\~ao adjacentes. O m\'\i nimo $n$ com esta propriedade (o comprimento do caminho) \'e dito a {\em dist\^ancia de caminho} entre $x$ e $y$.
\index{grafo! conexo}
\index{dist\^ancia! de caminho}
\end{definicao}

\begin{exercicio}
Verifique que a dist\^ancia de caminho \'e uma m\'etrica.
\end{exercicio}

Eis um resultado importante sobre a geometria de classes extremais.

\begin{teorema}
Seja $\mathscr C$ uma classe extremal. A dist\^ancia de caminho entre dois conceitos \'e igual \`a dist\^ancia de Hamming.
Em particular, a grafo de $1$-inclus\~ao de $\mathscr C$ \'e conexo.
\label{t:conexidade}
\end{teorema}

\begin{proof}
Suponhamos que a afirma\c c\~ao \'e falsa, e escolhamos dois conceitos, $C$ e $C^\prime$, cuja dist\^ancia de caminho n\~ao \'e igual a dist\^ancia de Hamming, e tais que a dist\^ancia de Hamming \'e menor entre todos os pares com esta propriedade. Seja $D$ o conjunto de todos $x$ onde $\chi_C(x)\neq \chi_{C^\prime}(x)$, e seja $\mathscr B$ o menor cubo que cont\'em $C$ e $C^\prime$, ou seja, o cubo dom conjunto de coordenadas $D$ e $\mbox{tag}({\mathscr B})=C\setminus D = C^\prime\setminus D$. Segue-se da nossa hip\'otese que $\sharp D\geq 2$. No cubo $\mathscr B\cong \{0,1\}^D$ existe um elemento $C^{\prime\prime}\neq C,C^\prime$ com a propriedade 
\[d_{Hamming}(C,C^{\prime\prime})+d_{Hamming}(C^\prime,C^{\prime\prime})
=d_{Hamming}(C,C^{\prime}),\]
e a escolha de $C,C^\prime$ garante que 
\begin{align*}
d_{caminho}(C,C^{\prime})&\leq 
d_{caminho}(C,C^{\prime\prime})+d_{caminho}(C^\prime,C^{\prime\prime})\\
&= d_{Hamming}(C,C^{\prime\prime})+d_{Hamming}(C^\prime,C^{\prime\prime})\\
&= d_{Hamming}(C,C^{\prime}),
\end{align*}
uma contradi\c c\~ao.
\end{proof}

Digamos que um cubo $\mathscr B\subseteq\mathscr C$ \'e {\em maximal} se ele n\~ao est\'a contido num cube estritamente maior dentro de $\mathscr C$.

\begin{lema}
Sejam ${\mathscr B}_1$, ${\mathscr B}_2$ dois cubos dentro de uma classe extremal, $\mathscr C$, tendo os conjuntos de coordenadas $D_1$ e $D_2$. Se ${\mathscr B}_1$ \'e maximal e $D_1\subseteq D_2$, ent\~ao ${\mathscr B}_1={\mathscr B}_2$.
\label{l:cubomaximal}
\end{lema}

\begin{proof}
Suponha que $D_2\setminus D_1\neq\emptyset$ e ${\mathscr B}_1\neq {\mathscr B}_2$. A redu\c c\~ao ${\mathscr B}^{\backslash D}_1$ \'e um conjunto unit\'ario com o \'unico elemento $\mbox{tag}\,{\mathscr B}_1$, e a redu\c c\~ao ${\mathscr B}^{\backslash D}_2$ \'e um cubo com o conjunto de coordenadas $D_2\setminus D_1$, que n\~ao cont\'em $\mbox{tag}\,{\mathscr B}_1$ por causa da maximalidade de ${\mathscr B}_1$. 
Como o grafo de $1$-inclus\~ao de ${\mathscr C}^{\backslash D}$ \'e conexo (lema \ref{l:reducaoextremal} e teorema \ref{t:conexidade}), existe um caminho que junta $\mbox{tag}\,{\mathscr B}_1$ com algum elemento do cubo ${\mathscr B}^{\backslash D}_2$. O elemento do caminho adjacente a $\mbox{tag}\,{\mathscr B}_1$ difere de $\mbox{tag}\,{\mathscr B}_1$ em um ponto $x\in \Omega\setminus D$, e est\'a da forma $\mbox{tag}\,{\mathscr B}_3$, onde ${\mathscr B}_3$ tem $D$ como o conjunto de coordenadas. Segue-se das defini\c c\~oes que ${\mathscr B}_1\cup {\mathscr B}_3$ \'e um cubo com conjunto de coordenadas $D\cup \{x\}$, contradizendo a maximalidade de ${\mathscr B}_1$.
\end{proof}

Agora podemos mostrar o resultado central da se\c c\~ao.

\begin{teorema}[\citet*{MW}]
Cada classe extremal de dimens\~ao de Vapnik--Chervonenkis $d$ admite um esquema de compress\~ao amostral rotulado de tamanho $d$. 
\index{teorema! de Moran--Warmuth}
\end{teorema}

\begin{proof}
Teorema de Compacidade \ref{t:compacidadeesquemas} reduz o teorema ao caso de dom\'\i nios finitos. O esquema de compress\~ao tem a descri\c c\~ao seguinte.

\begin{enumerate}
\item A descompress\~ao ${\mathcal H}(\tau,\gamma)$ de uma amostra rotulada $(\tau,\gamma)$ \'e qualquer conceito $C$ consistente com a rotulagem (isto \'e, $C\cap\tau =\gamma$) que pertence a um cubo $\mathscr B\subseteq\mathscr C$ com o conjunto de coordenadas $\tau$.
\item 
A compress\~ao $\kappa(\sigma,\tau)$ de uma amostra rotulada $(\sigma,\tau)$ \'e da forma $(\delta,\tau\cap\delta)$, onde $\delta$ \'e o conjunto de coordenadas de um cubo maximal em $\mathcal C\cap\sigma$, que cont\'em $\tau$.
\end{enumerate}

Se a amostra $(\delta,\alpha)$ pertence \`a imagem da aplica\c c\~ao de compress\~ao, $\kappa$, ent\~ao $\delta$ \'e fortemente fragmentada pela classe  $\mathscr C\upharpoonright \sigma$, logo fragmentado por $\mathscr C$, logo fortemente fragmentado por $\mathscr C$. Existe um cubo $\mathscr B$ em $\mathscr C$ com o conjunto de coordenadas $\delta$, em particular existe um conceito $C\in {\mathscr B}$ que induz sobre $\delta$ a rotulagem $\alpha$. Ent\~ao, a hip\'otese ${\mathcal H}(\tau,\gamma)$ para tais amostras \'e bem definida.

Resta s\'o verificar a condi\c c\~ao na eq. (\ref{eq:forallsigmaconsistente}). Seja $(\sigma,C\cap\sigma)$ uma amostra rotulada por um elemento da classe $\mathscr C$, seja $\mathscr B$ um cubo maximal da classe ${\mathscr C}\upharpoonright\sigma$ que cont\'em $C\cap\sigma$, com o conjunto de coordenadas $\delta$. Seja ${\mathscr B}^\prime$ um cubo da classe $\mathscr C$ com coordenadas $\delta$, e seja $C^\prime\in {\mathscr B}^\prime$ um conceito que induz sobre $\delta$ a rotulagem $C\cap\sigma=C\cap\delta$. Basta verificar que $C^\prime$ \'e consistente com $C$ tamb\'em sobre o resto de $\sigma$, ou seja, que $C^\prime\cap (\sigma\setminus\delta)=C\cap (\sigma\setminus\delta)$.

O cubo ${\mathscr B}^\prime\upharpoonright\sigma$ tem $\delta$ como coordenadas, e como o cubo $\mathscr B$ \'e maximal, lema \ref{l:cubomaximal} implica que ${\mathscr B}={\mathscr B}^\prime\upharpoonright\sigma$. Como $\delta$ \'e o conjunto de coordenadas de $\mathscr B$, a restri\c c\~ao ${\mathscr B}\upharpoonright\sigma\setminus\delta$ consiste apenas de um conceito, $\mbox{tag}\,({\mathscr B})$. O mesmo vale para a restri\c c\~ao ${\mathscr B}^\prime\upharpoonright\sigma\setminus\delta$, e o resultado segue.
\end{proof}

\section{Teorema de Moran--Yehudayoff}

O mais forte resultado geral envolvendo a exist\^encia dos esquemas de compress\~ao at\'e o momento \'e um avan\c co not\'avel recente por \citep*{MY}, que mostravam que cada classe de conceitos $\mathscr C$ de dimens\~ao $\VC({\mathscr C})=d$ admite um esquema de compress\~ao de tamanho exponencial em $d$. Aqui a no\c c\~ao de um esquema \'e relaxado ainda mais: n\~ao somente o esquema \'e rotulado, mas al\'em disso o argumento do esquema permite um n\'umero exponencial em $d$ de bits de ``informa\c c\~ao adicional''. O tamanho do esquema no teorema de Moran--Yehudayoff n\~ao depende apenas da dimens\~ao de Vapnik--Chervonenkis, mas tamb\'em da dimens\~ao de Vapnik--Chervonenkis {\em dual,} que, por seu turno, admite uma cota exponencial em $d$.

\subsection{Dimens\~ao de Vapnik--Chervonenkis dual}
Seja $(\Omega,{\mathscr C})$ uma classe de conceitos, isto \'e, $\Omega$ \'e um conjunto e $\mathscr C$ \'e uma fam\'\i lia de subconjuntos de $\Omega$. A {\em classe de conceitos dual} a $(\Omega,{\mathscr C})$ tem $\mathscr C$ como o dom\'\i nio e $\Omega$ como a classe de conceitos, onde cada $x\in\Omega$ determina uma fun\c c\~ao bin\'aria, $\hat x$, sobre $\mathscr C$ como segue:
\[\hat x (C) = \begin{cases} 1,&\mbox{ se }x\in C,\\
0,&\mbox{ sen\~ao.}
\end{cases}
\]
\index{classe! dual}
A dimens\~ao VC da classe dual \'e chamada {\em dimens\~ao VC dual} de $\mathscr C$. Notemos: 
\[\VC^{\ast}(\mathscr C) = \VC({\mathscr C},\Omega).\]
\index{dimens\~ao! de Vapnik--Chervonenkis! dual}

\begin{teorema}
Suponha que $\VC((\Omega,{\mathscr C})=d$ seja finita. Ent\~ao, a dimens\~ao VC dual de $\mathscr C$ \'e estritamente limitada por $2^{d+1}-1$,
\[\VC^\ast({\mathscr C}) < 2^{d+1},\]
\'e a desigualdade \'e exata.
\end{teorema}

\begin{proof}
Suponha que $\VC(\Omega,\mathscr C)\geq 2^{d+1}$. Ent\~ao existe uma subfam\'\i lia finita $\mathscr A\subseteq \mathscr C$ de cardinalidade $\abs{\mathscr A} = 2^{d+1}$ fragmentada por $\Omega$. O que significa: para todo $\mathscr B \subseteq \mathscr A$ existe $x=x_{\mathscr B}\in \Omega$ que pertence aos todos $B\in\mathscr B$ e n\~ao pertence a nenhum $A\in\mathscr A\setminus \mathscr B$. 

Vamos agora selecionar um subconjunto de $\Omega$ com $d+1$ elementos fragmentado por $\mathscr A$, acabando o argumento por contraposi\c c\~ao. 
Indexamos os elementos de $\mathscr A$ com elementos do cubo de Hamming $\{0,1\}^{d+1}$. Para cada $i=1,2,\ldots,d+1$, escolhemos um ponto $x_i\in\Omega$ que pertence a um conjunto $A_\sigma$ indexado com a palavra $\sigma=\sigma_1\sigma_2\ldots\sigma_{d+1}$ se e somente se $\sigma_i=1$. Isso \'e poss\'\i vel porque $\mathscr A$ \'e fragmentado por $\Omega$.

Se agora $I$ \'e um subconjunto de $\{1,2,\ldots,d+1\}$, seja $\sigma$ um elemento do cubo de Hamming que tem $1$ nas todas as posi\c c\~oes que pertencem a $I$, e $0$ em outras partes. O conjunto correspondente $A_\sigma$ contem $x_i$ se e somente se $i\in I$. 

Resta mostrar que o valor $2^{d+1}-1$ pode ser atingido. Seja $\Omega=\{0,1\}^k$, e seja $\mathscr C$ uma classe que consiste de todos os subconjuntos cil\'\i ndricos do cubo de Hamming da forma
\[\pi_i^{-1}(1),~~i=1,2,\ldots,k.\]
Em outras palavras, 
\[{\mathscr C} =\{C_i\colon i=1,2,\ldots,k\},\]
onde $C_i$ consiste de todas as palavras cuja $i$-\'esima coordenada \'e um.

\'E claro que $\mathscr C$ \'e fragmentado por $\Omega$, logo a dimens\~ao VC dual desta classe \'e exatamente $k$. No mesmo tempo, a dimens\~ao VC de $\mathscr C$ \'e menor ou igual a $\lfloor\log k\rfloor$, pelas considera\c c\~oes da cardinalidade.

Defina $k=2^{d+1}-1$. A constru\c c\~ao acima resulta em uma classe de dimens\~ao VC dual $2^{d+1}-1$ e de dimens\~ao VC $\leq d$. A dimens\~ao VC de $\mathscr C$ n\~ao pode ser estritamente menor que $d$ por causa da desigualdade j\'a estabelecida. Portanto, $\VC(\mathscr C)=d$.
\end{proof}

\begin{corolario}
Se $d=\VC(\mathscr C)$ e $d^\ast=\VC^\ast(\mathscr C)$,
\[\log_2d - 1 < d^\ast < 2^{d+1}.\]
Em particular, $d^\ast = O(\exp d) \cap \Omega (\log d)$. \qed
\index{VCstar@$\VC^\ast$}
\end{corolario}

\subsection{Complexidade amostral de classes de Glivenko--Cantelli}
Dado uma classe de Glivenko--Cantelli de dimens\~ao VC $d$,
precisamos obter algumas estimativas para a complexidade amostral de $\mathscr C$, ou seja, um valor de $n$ garantindo que o erro uniforme da estimativa emp\'\i rica de medidas de todos os elementos de $\mathscr C$ seja menor que $\ve>0$ com confian\c ca $1-\delta$. A complexidade amostral \'e de ordem $O(d/\ve^2)$, e para deduzi-la, vamos usar o resultado t\'ecnico importante.

\subsubsection{Lema de separa\c c\~ao de Dudley}

\begin{lema}[Lema de Separa\c c\~ao (Dudley)]
\label{l:separacao}
\index{lema! de Dudley de separa\c c\~ao}
Seja $K$ um subconjunto $\e$-separado do cubo de Hamming $\Sigma^n$ (com a dist\^ancia normalizada). Suponha 
\[\VC(K)\leq d.\] 
Ent\~ao,
\begin{equation}
\abs K\leq \left(2e^2 \log\left(\frac{2e}{\e}\right)\right)^d\left(\frac 1{\e}\right)^d.
\end{equation}
\end{lema}

\begin{observacao}
Note que a cota sobre o tamanho de $K$ n\~ao depende da dimens\~ao $n$ do cubo de Hamming en quest\~ao.
\end{observacao}

\begin{lema}
Seja $K$ um subconjunto $\e$-separado de $\Sigma^n$. Ent\~ao existe um subconjunto de coordenadas
\[I\subseteq \{1,2,\ldots,n\}\]
com
\begin{equation}
\label{eq:I}
\abs I\leq \frac{2\log \abs K}{\e},\end{equation}
tendo a propriedade que cadas duas palavras distintas $\eta,\tau\in K$ diferem entre elas sobre pelo menos uma coordenada de $I$. Em outras palavras, a aplica\c c\~ao de restri\c c\~ao
\[K\ni\sigma\mapsto\sigma\upharpoonright I\in \{0,1\}^I\]
\'e injetora.
\label{l:I}
\end{lema}

\begin{proof}
Se $\abs K\leq 2$, a afirma\c c\~ao \'e \'obvia, e por isto assumamos que $\abs K\geq 3$, e por conseguinte
\begin{equation}
\label{eq:log}
\log\abs K\geq 1.
\end{equation}
Formemos o subconjunto $V\subseteq \{-1,0,1\}^n$ por
\[V =\{\eta-\tau\colon \eta,\tau\in K,~\eta\neq\tau\}.\]
(Aqui a subtra\c c\~ao \'e feita em $\R^n$, nem modulo $2$.) 

Claro, $\abs V\leq \abs K^2$.
Como para cada $\eta\neq\tau$ temos $\bar d(\eta,\tau)\geq \e$, cada $v\in V$ difere de zero sobre pelo menos $n\e$ coordenadas.

Sejam $X_i$, $i=1,2,\ldots,t$ vari\'aveis aleat\'orias i.i.d. tomando valores em $[n]=\{1,2,\ldots,n\}$ e distribu\'\i das uniformemente: para todos $i=1,2,\ldots,t$ e $j=1,2,\ldots,n$,
\[Pr\{X_i=j\}=\frac 1n.\]
Para todo $v\in V$, a probabilidade do que uma coordenada aleat\'oria n\~ao \'e nula \'e pelo menos $\e$, logo
\[Pr\left\{\forall i=1,2,\ldots,t,~v(X_i)=0\right\} =\prod_{i=1}^t Pr\{v(X_i)=0\}<(1-\e)^t.\]
Segundo a cota de uni\~ao,
\begin{align*}
Pr\left\{\exists v\in V,~\forall i=1,2,\ldots,t,~v(X_i)=0\right\} &\leq 
\sum_{v\in V} Pr\left\{\forall i=1,2,\ldots,t,~v(X_i)=0\right\} \\
&\leq  \abs K^2 (1-\e)^t.
\end{align*}
Por conseguinte, para o evento complementar,
\[Pr\left\{\forall v\in V,~\exists i=1,2,\ldots,t,~ v(X_i)\neq 0\right\}\geq 1- \abs K^2 (1-\e)^t.\]
Desde que 
\begin{equation}
1- \abs K^2 (1-\e)^t>0,
\label{eq:des}
\end{equation}
existe um subconjunto $I\subseteq [n]$ com $t$ elementos e a propriedade desejada. A eq. (\ref{eq:des}) \'e equivalente a
\[(1-\e)^t<\frac 1{\abs K^2},\]
ou seja, o tamanho de $I$ que garante a conclus\~ao do lema, \'e
\begin{equation}
t > - \frac{2\log\abs K}{\log(1-\ve)}.
\label{eq:con}
\end{equation}
Para todos $\ve$ estritamente entre $0$ e $t$ temos, usando a s\'erie de Taylor,
\begin{equation}
\label{eq:taylor}
-\log(1-\e)=\e+\frac{\e^2}2+\frac{\e^3}3+\ldots+\frac{\e^n}n+\ldots.\end{equation}
Em particular,
\[-\log(1-\e)>\e\]
logo
\[- \frac{2\log\abs K}{\log(1-\e)}<\frac{2\log\abs K}{\e}.\]
Tomando em considera\c c\~ao (\ref{eq:taylor}) e (\ref{eq:log}),
\begin{align*}
\frac{2\log\abs K}{\e}-\left(- \frac{2\log\abs K}{\log(1-\e)}\right) &= 
2\log\abs K \frac {\log(1-\e)+\e}{\e\log(1-\e)} \\
&= 2\log\abs K \frac{\frac{1}2+\frac{\e}3+\ldots+\frac{\e^{n-2}}n+\ldots}
{1+\frac{\e}2+\frac{\e^2}3+\ldots+\frac{\e^{n-1}}n+\ldots} \\
&\geq  \frac{1+\frac{2\e}3+\ldots+\frac{2\e^{n-2}}n+\ldots}
{1+\frac{\e}2+\frac{\e^2}3+\ldots+\frac{\e^{n-2}}{n-1}+\ldots} \\
&> 1.
\end{align*}
Por conseguinte, existe pelo menos um n\'umero inteiro $t$ estritamente entre os dois n\'umeros:
\[- \frac{2\log\abs K}{\log(1-\e)}<t<\frac{2\log\abs K}{\e}.\]
Este $t$ satisfaz (\ref{eq:con}), bem como (\ref{eq:I}), e n\'os conclu\'\i mos.
\end{proof}

\begin{exercicio}
\label{ex:exxe}
Mostre que
\[e^x>x^e\]
para todos $x>e$.
\end{exercicio}

\begin{lema}
\label{l:abba}
Suponhamos que $\beta>e^2$ e $\alpha\geq 1$. 
\par
(1) Se $\alpha\geq e\beta\log\beta$, ent\~ao $\alpha>\beta\log\alpha$. De modo equivalente, 
\par (2) se $\alpha\leq \beta\log\alpha$, ent\~ao $\alpha<e\beta\log\beta$.
\end{lema}

\begin{proof} 
Estabele\c camos a primeira parte do lema, de onde a segunda parte segue pela contraposi\c c\~ao. Defina a fun\c c\~ao $\phi(\alpha)=\alpha -\beta\log\alpha$, e seja $\alpha_0=e\beta\log\beta$. Como $\beta>e^2$, temos $\alpha_0>e\beta$. Al\'em disso,
\begin{align*}
\phi(\alpha_0) 
&= e\beta\log\beta -\beta (1+\log\beta+\log\log\beta) \\
&> \beta (\log\beta-1-\log\log\beta) \\
&> 0,
\end{align*}
sob a condi\c c\~ao que 
\[\log\beta>1+\log\log\beta,\]
isto \'e,
\[\beta>e\log\beta,\]
ou
\[e^\beta >\beta^e,\]
depois de exponenciar duas vezes. A \'ultima desigualdade \'e v\'alida se  $\beta>e^2$ (exerc\'\i cio \ref{ex:exxe}), de onde conclu\'\i mos: $\phi(\alpha_0)>0$ quando $\beta>e^2$. Como $\phi^\prime(\alpha)=1-\beta/\alpha>0$ para $\alpha>\alpha_0$, temos $\phi(\alpha)>0$ para todos $\alpha>\alpha_0$. 
\end{proof}

\begin{proof}[Prova do Lema de Separa\c c\~ao \ref{l:separacao}]
Escolhamos um subconjunto $I$ como no lema \ref{l:I}. Obviamente, 
\[\VC(K\upharpoonright I)\leq d,\]
e pelo lema de Sauer--Shelah,
\begin{align*}
\abs K &= \abs{K\upharpoonright I} \\
&\leq  \left(\frac{e \abs I}{d} \right)^d \\
&\leq  \left(\frac{e 2\log \abs K}{\e d} \right)^d,
\end{align*}
ou seja,
\[\abs K^{1/d} \leq \frac{2e\log \abs K}{\e d}=\frac{2e}{\e}\log\left(\abs K^{1/d} \right). \]
Aplicando lema \ref{l:abba}, parte (2), com valores
\[\alpha = \abs K^{1/d}\mbox{ (o que \'e $\geq 1$)}\]
e 
\[\beta = \frac{2e}{\e}\mbox{ (o que \'e $> e^2$ se $\e<2/e$)},\]
conclu\'\i mos
\[\abs K^{1/d}<e \frac{2e}{\e}\log\left(\frac{2e}{\e}\right)= 
\frac{2e^2}{\e}\log\left(\frac{2e}{\e}\right),\]
or, de modo equivalente,
\[\abs K < \left(2e^2 \log\left(\frac{2e}{\e}\right)\right)^d\left(\frac 1{\e}\right)^d,
\]
como desejado. 
S\'o basta notar que se $\e\geq 2/e$, ent\~ao $K$ cont\'em ao m\'aximo dois pontos, e a conclus\~ao \'e verdadeira tamb\'em.
\end{proof}

Relembramos que $\mu_n$ \'e a medida emp\'\i rica suportada pela amostra aleat\'oria $X_1,X_2,\ldots,X_n$ (ent\~ao, $\mu_n$ \'e uma medida aleat\'oria). A dist\^ancia $L^1(\mu_n)$ entre duas fun\c c\~oes reais  $f$ e $g$ \'e dada por
\[\norm{f-g}_{L^1(\mu_n)}=\frac 1n \sum_{i=1}^n \abs{f(X_i)-g(X_i)},\]
\'e ela \'e uma dist\^ancia aleat\'oria. Esta dist\^ancia n\~ao \'e, geralmente, uma m\'etrica, mas uma pseudom\'etrica. 

\begin{exercicio} 
O que \'e $\E_{\mu}\norm{f-g}_{L^1(\mu_n)}$?
\end{exercicio}

Seja $\sigma=\{x_1,x_2,\ldots,x_n\}$ uma inst\^ancia de amostra aleat\'oria, e suponhamos que todos os pontos $x_i$ s\~ao dois a dois distintos. Ent\~ao a dist\^ancia $L^1(\mu_n)$ sobre fun\c c\~oes bin\'arias com valores em $\{0,1\}$ \'e simplesmente a dist\^ancia de Hamming normalizada entre as restri\c c\~oes das fun\c c\~oes sobre a amostra:
\[\norm{\chi_C-\chi_D}_{L^1(\mu_n)} = \bar d(\chi_C\upharpoonright\sigma,\chi_D\upharpoonright\sigma).\]
Para uma classe de conceitos $\mathscr C$ sobre um dom\'\i nio $\Omega$, nos denotamos $N(\e,{\mathscr C},L^1(\mu_n))$ ou, mais precisamente, $N(\e,{\mathscr C},L^1(\mu_\sigma))$, o n\'umero de cobertura de $\mathscr C$ em rela\c c\~ao \`a medida emp\'\i rica suportada na amostra $\sigma$. Essencialmente, isso \'e o n\'umero de cobertura da imagem de $\mathscr C$ no cubo de Hamming correspondente $\{0,1\}^\sigma$. O seguinte \'e uma consequ\^encia imediata do lema de Dudley. 

\begin{teorema}[Dudley]
Seja $\VC(\mathscr C)\leq d$. Ent\~ao para todo $\ve>0$
\begin{equation}
N(\e,{\mathscr C},L^1(\mu_n)) \leq \left(2e^2 \log\left(\frac{2e}{\e}\right)\right)^d\left(\frac 1{\e}\right)^d.
\end{equation}
\qed
\index{teorema! de Dudley}
\end{teorema}

\begin{exercicio}
Seja $\mathscr C$ uma classe de conceitos de dimens\~ao VC finita $d$. Mostre que as cotas no teorema de Dudley valem para dist\^ancia $L^1(\mu)$ em rela\c c\~ao a {\em qualquer} medida de probabilidade boreliana $\mu$ sobre $\Omega$. 
\par
[ {\em Sugest\~ao:} assuma existe um subconjunto finito $\ve$-separado em $\mathscr C$ em rela\c c\~ao a dist\^ancia $L^1(\mu)$, cuja cardinalidade \'e estritamente maior do que a cota no teorema, e obtenha a contradi\c c\~ao usando amostragem aleat\'oria junto com o fato que $\mathscr C$ \'e uma classe de Glivenko--Cantelli. ]
\end{exercicio}

\subsubsection{Complexidade amostral de classes de Glivenko--Cantelli}

Segundo teorema \ref{t:rademacher}, dada uma classe de conceitos ${\mathscr C}$ qualquer, com confian\c ca $1-\delta$,
\begin{equation}
\sup_{C\in{\mathscr C}}\left\vert \mu_\sigma(C)-\mu(C)\right\vert\leq 2R_n({\mathscr C}) + \sqrt{\frac{\ln (2/\delta)}{2n}}.
\label{eq:teoremarademahera}
\end{equation}

A prova da implica\c c\~ao (2) $\Rightarrow$ (1) no teorema \ref{t:vct} (p\'agina \pageref{p:impl1implies2}) cont\'em a desigualdade seguinte: para qualquer $\ve>0$,
\begin{equation}
\hat R_n({\mathscr C})(\sigma) \leq  \sqrt{\frac{2\log N(\e,{\mathscr C},L^1(\mu_\sigma))}{n}}+\ve.
\label{eq:upperboundforRn}
\end{equation}

Combinando a desigualdade acima e o teorema de Dudley para o valor $\ve/8$ ao inv\'es de $\ve$, obtemos:

\begin{align*}
\hat R_n({\mathscr C})(\sigma) &= 
\sqrt{\frac{2\log N(\e/8,{\mathscr C},L^1(\mu_\sigma))}{n}}+\frac{\ve}{8}
\\
&\leq \sqrt{\frac{2\log \left(2e^2 \log\left(\frac{16e}{\e}\right)\right)^d\left(\frac 8{\e}\right)^d}{n}}+\frac{\ve}{8}\\
&\leq \frac{\ve}{8} + \frac{\ve}{8} = \frac{\ve}{4},
\end{align*}
quando 
\[n\geq \frac{128}{\ve^2}\left(d\log\frac 8\ve +d\log\log\frac{16e}{\ve}+\log 2 + 2 \right).\]
Para o segundo termo na eq. (\ref{eq:teoremarademahera}), temos
\[\sqrt{\frac{\ln (2/\delta)}{2n}}<\frac{\ve}{4}\]
quando 
\[n> \frac{8}{\ve^2}\log\frac{2}{\delta}.\]
Deduzamos o resultado seguinte.

\begin{teorema}[Vapnik--Chervonenkis]
Seja $\mathscr C$ uma classe satisfazendo $\VC({\mathscr C})=d<\infty$. Dado $\ve,\delta>0$, temos com confian\c ca $1-\delta$ para cada medida de probabilidade $\mu$ no dom\'\i nio $\Omega$,
\[
\sup_{C\in{\mathscr C}}
\left\vert \mu_\sigma(C)-\mu(C)\right\vert\leq \ve
\]
quando
\begin{align*}
n&\geq \frac{128}{\ve^2}\max\left\{ d\log\frac 8\ve +d\log\log\frac{16e}{\ve}+\log 2 + 2, 16\log\frac{2}{\delta}
\right\} \\
& = O\left(\frac{d}{\ve^2}\log\frac 1{\delta}\right).
\end{align*}
Em outras palavras, a complexidade amostral de $\mathscr C$ satisfaz
\begin{equation}
s({\mathscr C},\ve,\delta)\leq \frac{128}{\ve^2}\max\left\{ d\log\frac 8\ve +d\log\log\frac{16e}{\ve}+\log 2 + 2, 16\log\frac{2}{\delta}
\right\}.
\label{eq:complexidadeamostralGCd}
\end{equation}
\label{vcgccomplexidade}
\index{complexidade! amostral}
\end{teorema}

\begin{observacao}
Como sempre, as constantes podem ser melhoradas consideravelmente.
\end{observacao}

Tendo em vista corol\'ario \ref{c:consistente}, conclu\'\i mos:

\begin{corolario}
Seja $\mathscr C$ uma classe de conceitos de Glivenko--Cantelli. Ent\~ao $\mathscr C$ \'e consistentemente PAC aprendiz\'avel com a complexidade amostral $s({\mathscr C},\ve,\delta)$ da eq. (\ref{eq:complexidadeamostralGCd}). A saber, se $\mathcal L$ \'e uma regra de aprendizagem consistente com a classe $\mathscr C$, ent\~ao para todos $\ve,\delta>0$ e cada $n\geq s({\mathscr C},\ve,\delta)$, dado um conceito $C\in {\mathscr C}$, temos para toda medida de probabilidade $\mu$ sobre $\Omega$ que, com confian\c ca $1-\delta$,
\[\mbox{erro}_{\mu,C}{\mathcal L}_n(C\upharpoonright\sigma) \leq\ve.\]
\label{c:consistentecomplexidadeVCd}
\end{corolario}

Eis um corol\'ario imediato para classes duais. 

\begin{corolario}
Seja $\mathscr C$ uma classe de dimens\~ao VC dual $\VC^\ast({\mathscr C})=d^\ast<\infty$ num dom\'\i nio $\Omega$. 
Dado $\ve,\delta>0$, para $n$ bastante grande da ordem $O(\frac{d^\ast}{\ve^2}\log\frac 1{\delta})$ e cada medida de probabilidade $\mu$ sobre $\Omega$, temos que com confian\c ca $1-\delta$
\[\forall x\in\Omega,~\mu\{C\in {\mathscr C}\colon x\in C\}\overset\ve\approx \frac{\sharp\{i=1,2,\ldots,n\colon x\in C_i\}}{n},
\]
onde $C_1,C_2,\ldots,C_n$ \'e uma amostra i.i.d. de conceitos, seguindo a distribui\c c\~ao $\mu$.
\label{c:leinaclasse}
\end{corolario}

\begin{observacao}
Todos os tr\^es resultados exigem hip\'oteses adicionais de mensurabilidade. Por exemplo, teorema \ref{vcgccomplexidade} e corol\'ario \ref{c:consistentecomplexidadeVCd} s\~ao v\'alidos sob a hip\'otese usual que $\mathscr C$ \'e universalmente separ\'avel bastaria em ambos casos. A hip\'otese mais geral, que se aplica a todas as tr\^es situa\c c\~oes e que n\~ao estudamos neste texto, seria a seguinte: a classe $\mathscr C$ admite uma estrutura boreliana tal que o conjunto $\{(x,C)\colon x\in\Omega,~C\in {\mathscr C}\}$ \'e um subconjunto boreliano do produto $\Omega\times {\mathscr C}$. Todavia, s\'o vamos usar o corol\'ario \ref{c:leinaclasse} para classes $\mathscr C$ finitas, naquele caso todas as condi\c c\~oes da mensurabilidade est\~ao verificadas.
\end{observacao}

\subsection{Esquemas de compress\~ao com informa\c c\~ao adicional\label{ss:adicional}}

A seguinte defini\c c\~ao j\'a aparece na obra original \citep*{LW}.
 
\begin{definicao}
Um esquema de compress\~ao amostral rotulado com informa\c c\~ao adicional $I$, onde $I$ \'e um conjunto finito, \'e um par $({\mathcal H},\kappa)$, onde 
\[{\mathcal H}\colon \ls\Omega d\times I\to 2^{\Omega}\]
\'e a aplica\c c\~ao de descompress\~ao, e 
\[\kappa\colon \ls\Omega\infty\to \ls\Omega d\times I,\]
a aplica\c c\~ao de compress\~ao, satisfazendo as propriedades usuais: $\kappa(\sigma,\tau)$ \'e uma subamostra de $\sigma$ cujo r\'otulo \'e induzido por $\tau$, e
\begin{equation}
\forall C\in {\mathscr C},~\forall\sigma\in [\Omega]^{\infty},~{\mathcal H}(\kappa(\sigma,C\cap\sigma))\cap \sigma = C\cap\sigma.
\label{eq:forallsigmaconsist}
\end{equation}
\index{esquema! de compress\~ao amostral! com informa\c c\~ao adicional}
\end{definicao}

\begin{observacao}
Da maneira equivalente, a defini\c c\~ao pode ser modificada usando a classe ${\mathscr C}_{\mathcal H}$.
\end{observacao}

Temos a vers\~ao seguinte do teorema \ref{t:littlestone-warmuth}.

\begin{teorema}[\citet*{LW}]
Seja $({\mathcal H},\kappa)$ um esquema de compress\~ao amostral rotulado com informa\c c\~ao adicional para uma classe $\mathscr C$ de conceitos borelianos, tal que a aplica\c c\~ao ${\mathcal H}\colon [\Omega]^{\leq d}\times I\times\Omega\to \{0,1\}$ \'e boreliana. Ent\~ao a regra ${\mathcal L} = {\mathcal H}\circ\kappa\circ\iota$ PAC aprende a classe $\mathscr C$ com erro $\ve>0$ e confian\c ca
\[\geq 1 - \sharp I\sum_{j=0}^d{n\choose j}(1-\ve)^{n-j},\]
onde $n$ \'e o tamanho da amostra. 
\label{t:littlestone-warmuth3}
\qed
\index{teorema! de Littlestone--Warmuth}
\end{teorema}

Daqui, pode-se deduzir a mesma estimativa da complexidade amostral que no exerc\'\i cio \ref{ex:estimativaMY}, onde $d$ \'e substitu\'\i do por $k=d+\log\sharp I$:
\[s(\ve,\delta) \leq \frac{8}{\ve}\left(k\log\left(\frac 2\ve\right)+\log\left(\frac 1{\delta}\right)\right).\]

\begin{exercicio}
Tentar deduzir a estimativa acima.
\end{exercicio}

O resultado justifica a defini\c c\~ao seguinte.

\begin{definicao}
O {\em tamanho} de um esquema rotulado com informa\c c\~ao adicional \'e definido como
\[d+\log_2\sharp I.\]
\end{definicao}

\begin{teorema}[\citet*{MY}] Cada classe $\mathscr C$ com $\VC({\mathscr C})=d$ e $\VC^\ast({\mathscr C})=d^\ast$ admite um esquema de compress\~ao rotulado com informa\c c\~ao adicional de tamanho $O(d\cdot d^\ast)$, em particular, $\exp(O(d))$.
\label{t:moran-yehudayoff}
\index{teorema! de Moran--Yehudayoff}
\end{teorema}

A prova ocupa o resto do cap\'\i tulo e conclui a parte principal das nossas notas.

\subsubsection{}
Fixemos qualquer regra de aprendizagem 
\[\mathcal L\colon\cup_{n=1}^{\infty}\Omega^n\times\{0,1\}^n\to {\mathscr C}\]
consistente com a classe $\mathscr C$, com valores em $\mathscr C$. 
Podemos supor que $\mathcal L$ \'e invariante pelas permuta\c c\~oes das amostras (n\~ao depende da ordem dos elementos $x_1,\ldots,x_n\in\sigma$), bem como n\~ao depende das repeti\c c\~oes dos elementos: se $\sigma=(x_1,\ldots,x_n)$ e $\sigma^\prime=(y_1,\ldots,y_m)$ s\~ao tais que $\{x_1,\ldots,x_n\}= \{y_1,\ldots,y_m\}$, ent\~ao para cada $C\in {\mathscr C}$, temos ${\mathcal L}(\sigma,C\upharpoonright\sigma)={\mathcal L}(\sigma^\prime,C\upharpoonright\sigma^\prime)$.

Seja $s=s({\mathscr C},1/3,1/3)$ o valor da complexidade amostral de aprendizagem da classe $\mathscr C$ para $\ve=\delta=1/3$. Segue-se do corol\'ario \ref{c:leinaclasse} que $s=O(d)$.

\subsubsection{Aplica\c c\~ao de compress\~ao} 
Seja $(\sigma,\tau)\in\ls\Omega\infty$ uma amostra rotulada com um conceito da classe ${\mathscr C}$, a saber, $\sigma\in[\Omega]^{n}$, $n\in\N$, e existe $C\in{\mathscr C}$ com $\tau=C\cap\sigma$. Procuremos a imagem de $(\sigma,\tau)$ pela aplica\c c\~ao de compress\~ao $\kappa$. Note que $C$ \'e desconhecido.

Formemos a classe (finita), $\mathscr C^\prime$, de todas as hip\'oteses da forma ${\mathcal L}(\varsigma,\tau\cap\varsigma)\in {\mathscr C}$, onde $\varsigma\in [\sigma]^{\leq s}$. (Formalmente, $(\varsigma,\tau)$ n\~ao pertence ao dom\'\i nio de $\mathcal L$, no entanto, podemos fixar uma ordem boreliana sobre $\Omega$, por exemplo, identific\'a-lo com $[0,1]$, tornando todas as amostras ordenadas, ou seja, elementos de $\Omega^n$). Eis o cora\c c\~ao da prova e um novo discernimento importante.

\begin{lema}
Existem $T=O(d^\ast)$ elementos $\varsigma_1,\ldots,\varsigma_T\in[\sigma]^{\leq s}$ tais que, para cada $x\in\sigma$, o rotulo de $x$ \'e adivinhado corretamente pela maioria das hip\'oteses ${\mathcal L}(\varsigma_i,\tau\cap\varsigma_i)$:
\begin{equation}
\forall x\in\sigma,~
\sharp\{i=1,2,\ldots,T\colon x\in \tau\Delta {\mathcal L}(\varsigma_i,\tau\cap\varsigma_i)\}<T/2.
\end{equation}
\label{l:myprincipal}
\end{lema}

\begin{proof}
Definamos uma matriz bin\'aria (com coeficientes em $\{0,1\}$), de tamanho $[n]^{\leq s}\times n$, cujas linhas correspondem \`as amostras $\varsigma\in [\sigma]^{\leq s}$, e colunas, aos elementos $x\in\sigma$. O coeficiente $(\varsigma,x)$ \'e igual a $0$ se e somente se 
\[ x\in \tau\Delta {\mathcal L}(\varsigma_i,\tau\cap\varsigma_i),\]
ou seja, o coeficiente \'e igual a $1$ se e somente se os r\'otulos gerados para $x$ por $\tau$ e pela hip\'otese ${\mathcal L}(\varsigma_i,\tau\cap\varsigma_i)$ s\~ao iguais.

Segundo corol\'ario \ref{c:leinaclasse} e a escolha de $s$, dada uma medida de probabilidade $\mu$ sobre a amostra $\sigma$, com confian\c ca $2/3$, o erro de aprendizagem da hip\'otese ${\mathcal L}(\tilde\varsigma,\tau\cap\tilde\varsigma)$, $\tilde\varsigma\in\sigma^s$, \'e menor que $1/3$:
\[\mu^{\otimes s}\{\tilde\varsigma\in\sigma^s\colon \mu\left({\mathcal L}(\tilde\varsigma,\tau\cap\tilde\varsigma)\Delta\tau\right)<1/3\}\geq 2/3.\]
Em particular, existe $\tilde\varsigma\in\sigma^s$ tal que o erro de aprendizagem de ${\mathcal L}(\tilde\varsigma,\tau\cap\tilde\varsigma)$ \'e menor que $1/3$. Segundo a nossa escolha de $\mathcal L$, a amostra $\varsigma\in [\sigma]^{\leq s}$ obtida de $\tilde\varsigma$ pela elimina\c c\~ao de repeti\c c\~oes e a ordem tem a mesma propriedade:
\begin{equation}
\mu\left({\mathcal L}(\varsigma,\tau\cap\varsigma)\Delta\tau\right)<1/3.
\label{eq:varsigmaexpressao}
\end{equation}

O teorema Minimax de von Neumann \ref{t:minimax} diz o seguinte. Pensemos de uma medida de probabilidade sobre $[n]$ como um $n$-vetor, $p$, com coordenadas (pesos) n\~ao negativas, cuja soma \'e igual a $1$. 
Seja $M$ uma matriz com coeficientes reais, de formato $m\times n$. Ent\~ao,
\[\max_{p\in P[m]}\min_{q\in P[n]}p^t M q=\min_{q\in P[n]}\max_{p\in P[m]} p^t Mq.\]
(Veja tamb\'em uma reformula\c c\~ao mais pertinente na eq. (\ref{eq:minimaxconcretapure})).
Para $\varsigma\in [\sigma]^{\leq s}$, denotemos por $\delta_{\varsigma}$ a medida de Dirac suportada em ponto $\varsigma$, ou seja, o vetor de base cuja coordenada $\varsigma$-\'esima \'e $1$, e o resto s\~ao zeros. Agora o erro na eq. (\ref{eq:varsigmaexpressao}) pode ser escrito assim:
\[\mu\left({\mathcal L}(\varsigma,\tau\cap\varsigma)\Delta\tau\right)=\delta_{\varsigma}^t M \mu.\]
Por conseguinte, o que acabamos de mostrar, \'e que 
\[\max_{\varsigma}\min_{\mu\in P(\sigma)}\delta_{\varsigma}^t M \mu \geq 2/3.\]
Segue-se que a express\~ao a direita no teorema Minimax seja $\geq 2/3$ tamb\'em. Por conseguinte, para cada elemento $x\in\sigma$, existe uma distribui\c c\~ao $p$ sobre o conjunto de hip\'oteses $\mathscr C^\prime$ (ou: o conjunto de $\varsigma$), tal que 
\[p\{\varsigma\colon x\in \tau\Delta{\mathcal L}(\varsigma,\tau\cap\varsigma)\}\geq 2/3.\]
Deduzimos do corol\'ario \ref{c:leinaclasse} que existe um conjunto finito de $T=O(d^\ast)$ subamostras $\varsigma_i$, $i=1,2,\ldots,T$, tal que para cada $x\in\sigma$,
\[p\{\varsigma\colon x\in \tau\Delta{\mathcal L}(\varsigma,\tau\cap\varsigma)\}
\overset{1/8}\approx \frac 1T\sharp\{i=1,2,\ldots,T\colon x\in \tau\Delta{\mathcal L}(\varsigma,\tau\cap\varsigma)\},\]
e por conseguinte
\begin{align*}
\frac 1T\sharp\{i=1,2,\ldots,T\colon x\in \tau\Delta{\mathcal L}(\varsigma,\tau\cap\varsigma)\} &\geq p\{\varsigma\colon x\in \tau\Delta{\mathcal L}(\varsigma,\tau\cap\varsigma)\} - \frac 18 \\
&> \frac 12,
\end{align*}
estabelecendo o lema.
\end{proof}

Agora a imagem $\kappa(\sigma,\tau)$ \'e definido como o par consistindo da subamostra 
\[\alpha=\bigcup_{t=1}^T\varsigma_t\]
de $\sigma$, rotulada com $\tau$, assim como a palavra bin\'aria $I\in \{0,1\}^{2^T}$, para identificar todas as subamostras de $\alpha$ que correspondem \`as amostras $\varsigma_t$. Pense dos elementos de $I$ como rotulagens de $\alpha$ com subconjuntos de $[T]$, onde cada ponto $x$ \'e marcado com o conjunto $\{t\in T\colon x\in \varsigma_t\}$. 

\begin{exercicio} 
Mostrar que o n\'umero de tais rotulagens \'e $O(d^\ast\cdot d)$.
\end{exercicio}

\subsubsection{Aplica\c c\~ao de descompress\~ao}
Dado um par que consiste de uma amostra rotulada $(\alpha,\tau)$ junto com a informa\c c\~ao adicional $i$, recupere as amostras rotuladas $(\varsigma_t,\tau_t)$, $t\in [T]$. O rotulo de um ponto $x\in\Omega$ pela hip\'otese ${\mathcal H}((\alpha,\tau),i)$ \'e definido pelo voto majorit\'ario entre os r\'otulos 
\[{\mathcal L}(\varsigma_t,\tau_t)(x),~t\in [T].\]

Como $\alpha$ \'e uma subamostra rotulada de $\sigma$, apenas temos que verificar a condi\c c\~ao (\ref{eq:forallsigmaconsist}). Sejam $C\in{\mathscr C}$ e $\sigma\in[\Omega]^{<\infty}$ quaisquer. Consideremos a hip\'otese ${\mathcal H}(\kappa(\sigma,C\cap\sigma))$. Para cada $x$, o r\'otulo \'e determinado pelo voto majorit\'ario entre as hip\'oteses ${\mathcal L}((\varsigma_t,\tau_t)$, e lema \ref{l:myprincipal} garante que este r\'otulo \'e correto.


\appendix
%
%

\chapter{Vari\'aveis aleat\'orias: a primeira passagem\label{a:variaveis}}

Ambicionamos dar sentido \`a frase seguinte: {\em os dados s\~ao modelados por uma sequ\^en\-cia $X_1,X_2,\ldots, X_n,\ldots \in\Omega$ de elementos aleat\'orios independentes e identicamente distribu\'\i dos de um espa\c co boreliano padr\~ao, $\Omega$.} Essa \'e uma tarefa que ser\'a plenamente concretizada apenas no Ap\^endice \ref{a:integral}. O presente Ap\^endice tem um car\'ater bastante informal e principalmente motivador.

\subsection{De vari\'aveis usuais para vari\'aveis aleat\'orias}
Dado um dom\'\i nio $\Omega$, os dados s\~ao modelados pelas {\em vari\'aveis aleat\'orias} com valores em $\Omega$, ou {\em elementos aleat\'orios} de $\Omega$. Recordemos primeiramente a no\c c\~ao bem conhecida de uma {\em vari\'avel}, muito comum na matem\'atica pura (geometria, \'algebra, an\'alise...)
Eis alguns contextos t\'\i picos onde as vari\'aveis fazem a sua apari\c c\~ao.
\vskip .2mm

(1) Determinar os valores de $x$ por quais 
\[5x^2 -x + 3 =0.\]
\vskip .2mm

(2) Suponha que $t\in [0,1]$. Ent\~ao ....
\vskip .2mm

(3) Sejam $x,y,z\in\R$ quaisquer. Suponha que $x<y$. Ent\~ao $x+z>y+z$.
\vskip .2mm

(4) Seja $z$ um n\'umero complexo qualquer. O valor absoluto de $z$ ....
\vskip .2mm

Uma vari\'avel \'e um elemento qualquer (desconhecido) de um conjunto ($\R$ nos casos (1) e (3), $[0,1]$ no (2), $\C$ no (4), etc.). 
As vari\'aveis na teoria de probabilidade s\~ao de uma natureza ligeiramente diferente. Elas s\~ao denotadas habitualmente pelas letras {\em mai\'usculas,} $X,Y,Z,\ldots$, a fim de distinguir das vari\'aveis ``usuais''. Se $X$ \'e uma vari\'avel aleat\'oria (abreviamos: v.a.) real, isso significa duas coisas. Primeiramente, como no caso de uma vari\'avel usual, 

- $X$ \'e um n\'umero real cujo valor exato \'e desconhecido: $X\in\R$.

Mas h\'a mais informa\c c\~oes adicionais dispon\'\i veis. Mesmo com o valor exato de $X$ sendo desconhecido, se sabe  

- a probabilidade de $X$ pertencer \`a cada parte $A$ de $\R$.

Em outras palavras, se $A\subseteq\R$ \'e uma parte de $\R$, ent\~ao existe um n\'umero real entre $0$ e $1$ que fornece a probabilidade do evento $X\in A$. Este n\'umero \'e denotado por
\[P[X\in A],\]
e o conjunto de valores $P[X\in A]$ para todos $A$ se chamam a {\em lei de probabilidade,} ou simplesmente a {\em lei} de $X$. Ent\~ao, uma vari\'avel aleat\'oria \'e uma vari\'avel ``usual'' munida de uma lei. Por exemplo, se $a,b\in\R$, $a\leq b$ quaisquer, ent\~ao se sabe a probabilidade
\[P[a<X<b]\]
de que o valor de $X$ esteja entre $a$ e $b$. A lei de uma vari\'avel aleat\'oria se denota por uma letra grega, por exemplo, $\mu$ ou $\nu$. \'E uma aplica\c c\~ao associando a cada parte $A$ de $\R$ um n\'umero real,
\[\R\supseteq A\mapsto \mu(A) = P[X\in A]\in [0,1].\]

Eis alguns exemplos.

1.
Uma vari\'avel aleat\'oria de Bernoulli toma dois valores: $0$ e $1$, cada uma com a probabilidade $1/2$:
\[P[X=0]=\frac 12 = P[X=1].\]
Para calcular a lei de $X$, seja $A\subseteq\R$ um conjunto qualquer. Obviamente, se $A$ cont\'em ambos $0$ e $1$, ent\~ao a probabilidade que $X\in A$ \'e igual a $1$, \'e um evento certo. Se $A$ n\~ao cont\'em nem $0$ nem $1$, ent\~ao o evento $X\in A$ \'e improv\'avel, a sua probabilidade \'e $0$. Afinal, se $A$ cont\'em exatamente um dos pontos $\{0,1\}$, ent\~ao a probabilidade do evento $X\in A$ \'e $1/2$:
\[P[X\in A] = \left\{\begin{array}{cl} 1,&\mbox{ se }0,1\in A,\\
  \frac 12,&\mbox{ se }0\in A\mbox{ e }1\notin A,\\
  \frac 12,&\mbox{ se }0\notin A\mbox{ e }1\in A,\\
  0,&\mbox{ se }0\notin A,~1\notin A.
  \end{array}\right.
\]

Uma vari\'avel de Bernoulli modela uma jogada \'unica de uma moeda justa, onde a probabilidade de dar ``coroa'' (o valor $1$) \'e $1/2$, a mesma que a probabilidade de dar ``cara'' (o valor $0$).

De maneira mais geral, se a moeda n\~ao \'e justa, ent\~ao a probabilidade de dar ``coroa'' pode ser um valor qualquer $p\in [0,1]$,
\[P[X=1]=p,\]
\'e a probabilidade de dar ``cara'' \'e
\[P[X=0]=1-p=q.\]

\begin{figure}[ht]
\begin{center}
  \scalebox{0.25}[0.25]{\includegraphics{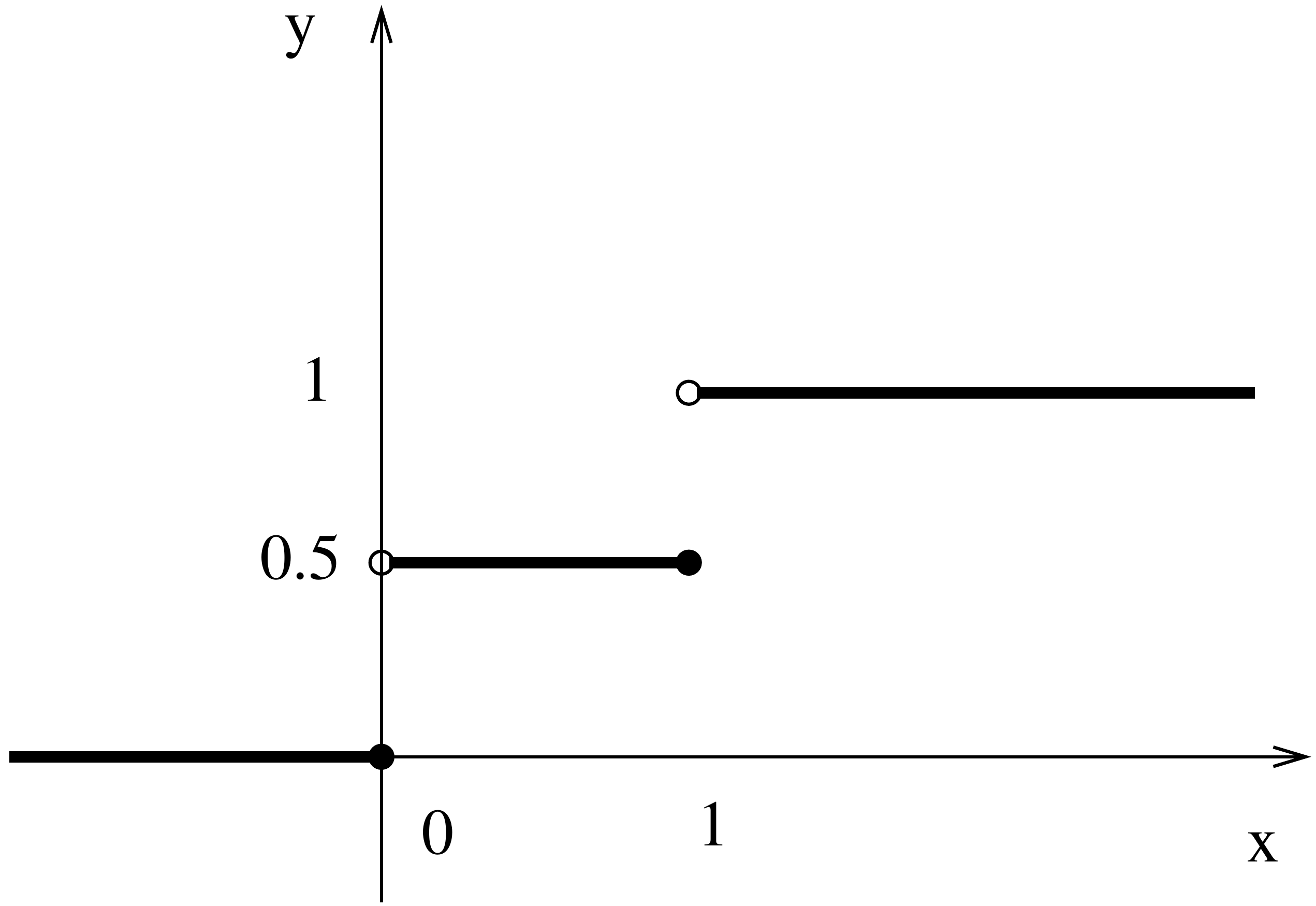}} 
\end{center}
\caption{\label{fig:phi_bernoulli}
Fun\c c\~ao de distribui\c c\~ao de uma vari\'avel aleat\'oria de Bernoulli.}
\end{figure}

A lei de probabilidade de uma vari\'avel aleat\'oria real, $X$, \'e completamente determinada pela sua {\em fun\c c\~ao de distribui\c c\~ao}, $\Phi$. \'E uma fun\c c\~ao real dada por
\[\Phi(t) = P[X<t].\]
\'E f\'acil calcular a fun\c c\~ao de distribui\c c\~ao de uma v.a. de Bernoulli, veja Figura \ref{fig:phi_bernoulli}.

Se o conjunto dos valores da fun\c c\~ao de distribui\c c\~ao de uma vari\'avel aleat\'oria $X$ \'e enumer\'avel, ent\~ao $X$ \'e dita {\em discreta.} Por exemplo, a vari\'avel aleat\'oria de Bernoulli \'e discreta.

\begin{figure}[ht]
\begin{center}
  \scalebox{0.25}[0.25]{\includegraphics{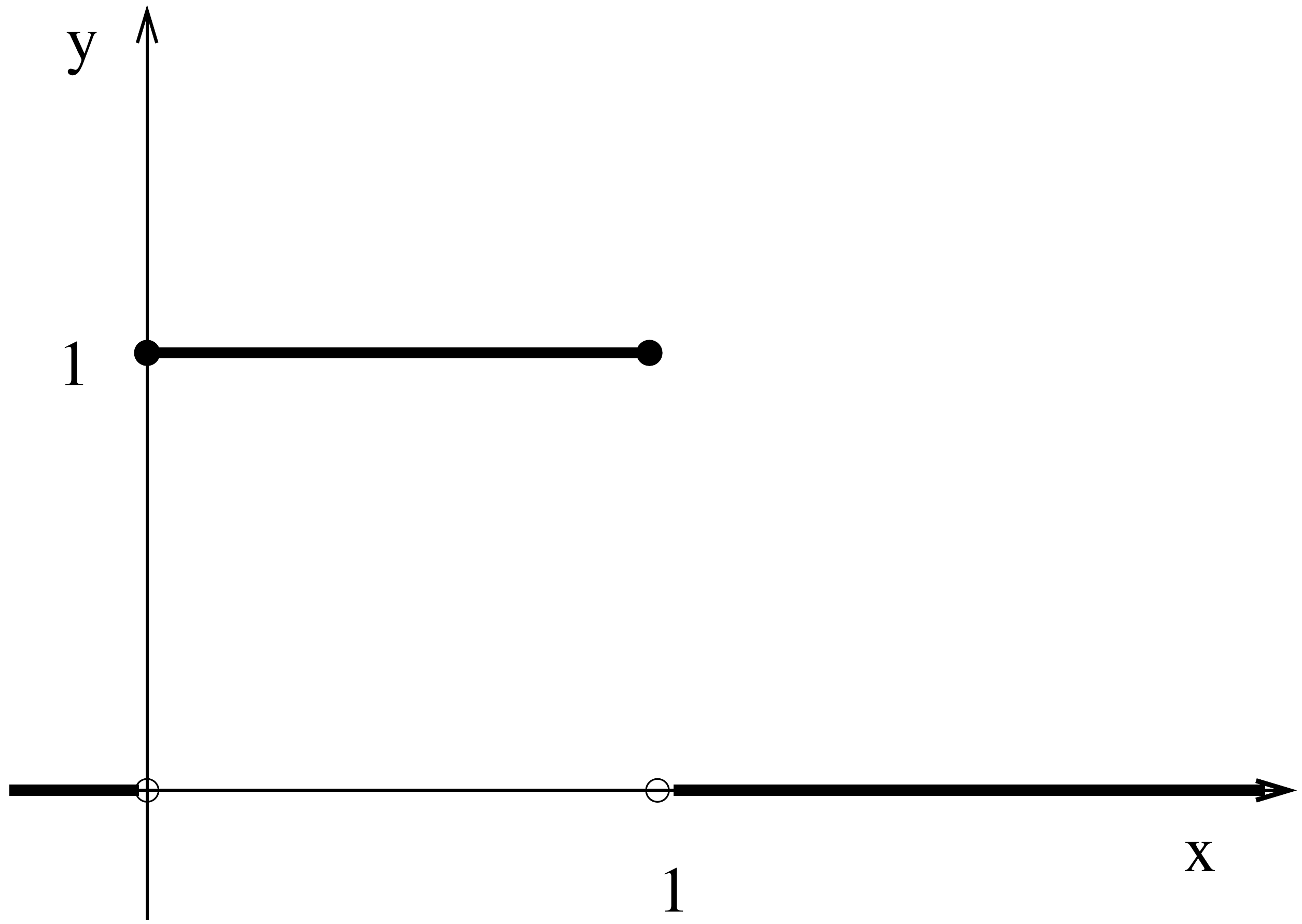}} 
\end{center}
\caption{\label{fig:indicatrice}
Grafo da fun\c c\~ao indicadora do intervalo $[0,1]$.}
\end{figure}

2. Uma vari\'avel aleat\'oria de lei {\em uniforme} com valores no intervalo $[0,1]$ \'e dada pela f\'ormula seguinte: quaisquer sejam $a,b\in\R$, $a<b$, 
\begin{equation}
  \label{eq:densite}
  P[X\in (a,b)] =\int_a^b \chi_{[0,1]}(t)\,dt.\end{equation}
Aqui, $\chi_{[0,1]}$ nota a {\em fun\c c\~ao indicadora} do intervalo $[0,1]$ (Figura \ref{fig:indicatrice}):
\[\chi_{[0,1]}(t) =\left\{\begin{array}{cl} 1,&\mbox{ se }x\in [0,1],\\
  0,&\mbox{ caso contr\'ario.}
\end{array}\right.\]

Por exemplo,
\[P[0\leq X \leq 1] =\int_0^1 \chi_{[0,1]}(t)\,dt =1,\]
e
\[P\left[-\frac 12\leq X \leq \frac 12\right] =\int_{-\frac 12}^{\frac 12} \chi_{[0,1]}(t)\,dt =\frac 12.\]

Se um intervalo $(a,b)$ est\'a contido em $[0,1]$, ent\~ao
\begin{eqnarray*}
  P[X\in (a,b)] &=& \int_a^b \chi_{[0,1]}(t)\,dt\\
  &=& \int_a^b 1\cdot dt \\
  &=& b-a.
  \end{eqnarray*}
Em outras palavras, neste caso a probabilidade de que $X$ perten\c ca ao intervalo $(a,b)$ \'e igual ao comprimento do intervalo. 

Se a lei de uma vari\'avel aleat\'oria \'e dada pela integral, como na f\'ormula (\ref{eq:densite}), ent\~ao a fun\c c\~ao sob a integral \'e chamada a {\em densidade} de $X$. A densidade de uma v.a. uniforme \'e a fun\c c\~ao indicadora:
\[\chi_{[0,1]}(t).\]

\begin{exercicio}
  Mostrar que a lei de Bernoulli n\~ao possui densidade.
\end{exercicio}

\begin{figure}[ht]
\begin{center}
  \scalebox{0.25}[0.25]{\includegraphics{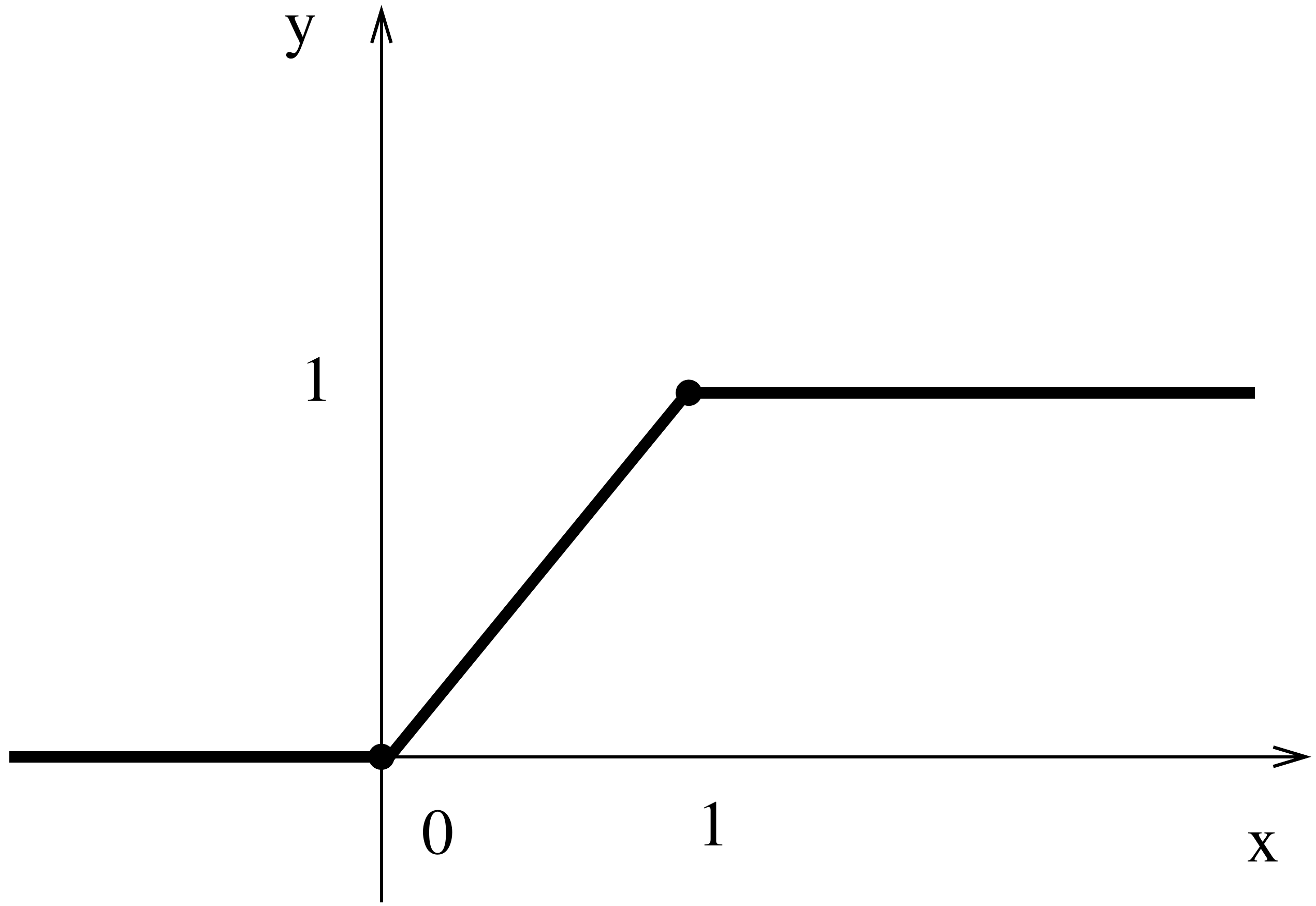}} 
\end{center}
\caption{\label{fig:phi_uniforme}
A fun\c c\~ao de distribui\c c\~ao de uma vari\'avel aleat\'oria uniforme.}
\end{figure}

A fun\c c\~ao de distribui\c c\~ao de uma v.a. uniforme \'e calculada facilmente (Figura \ref{fig:phi_uniforme}).

Uma vari\'avel aleat\'oria real $X$ \'e dita {\em cont\'\i nua} se valores da sua fun\c c\~ao de distribui\c c\~ao preenchem o intervalo $[0,1]$. A v.a. uniforme \'e obviamente cont\'\i nua. 

\begin{exercicio}
Seja $X$ uma v.a. que possui densidade. Mostrar que $X$ \'e cont\'\i nua.
\end{exercicio}

\begin{exercicio}[$\ast$]
Construir um exemplo de v.a. que \'e cont\'\i nua e n\~ao possui densidade.
\par 
[ {\em Sugest\~ao:} deve-se usar o conjunto de Cantor \ref{ex:conjuntodecantor}. ]
\end{exercicio}

\begin{exercicio}
  Construir um exemplo de v.a. nem discreta nem cont\'\i nua.
\end{exercicio}

Uma vari\'avel aleat\'oria real \'e {\em gaussiana} (ou: segue a lei {\em normal centrada e reduzida}), se $X$ possui densidade dada por
\[\frac{1}{\sqrt{2\pi}} e^{-t^2/2}.\]
Em outras palavras, quais quer sejam $a,b\in\R$, 
\[P[a<X<b] = \frac {1} {\sqrt{2\pi}}\int_a^b e^{-t^2/2}dt.\]

\begin{figure}[h]
\centering
  \scalebox{0.55}[0.55]{\includegraphics{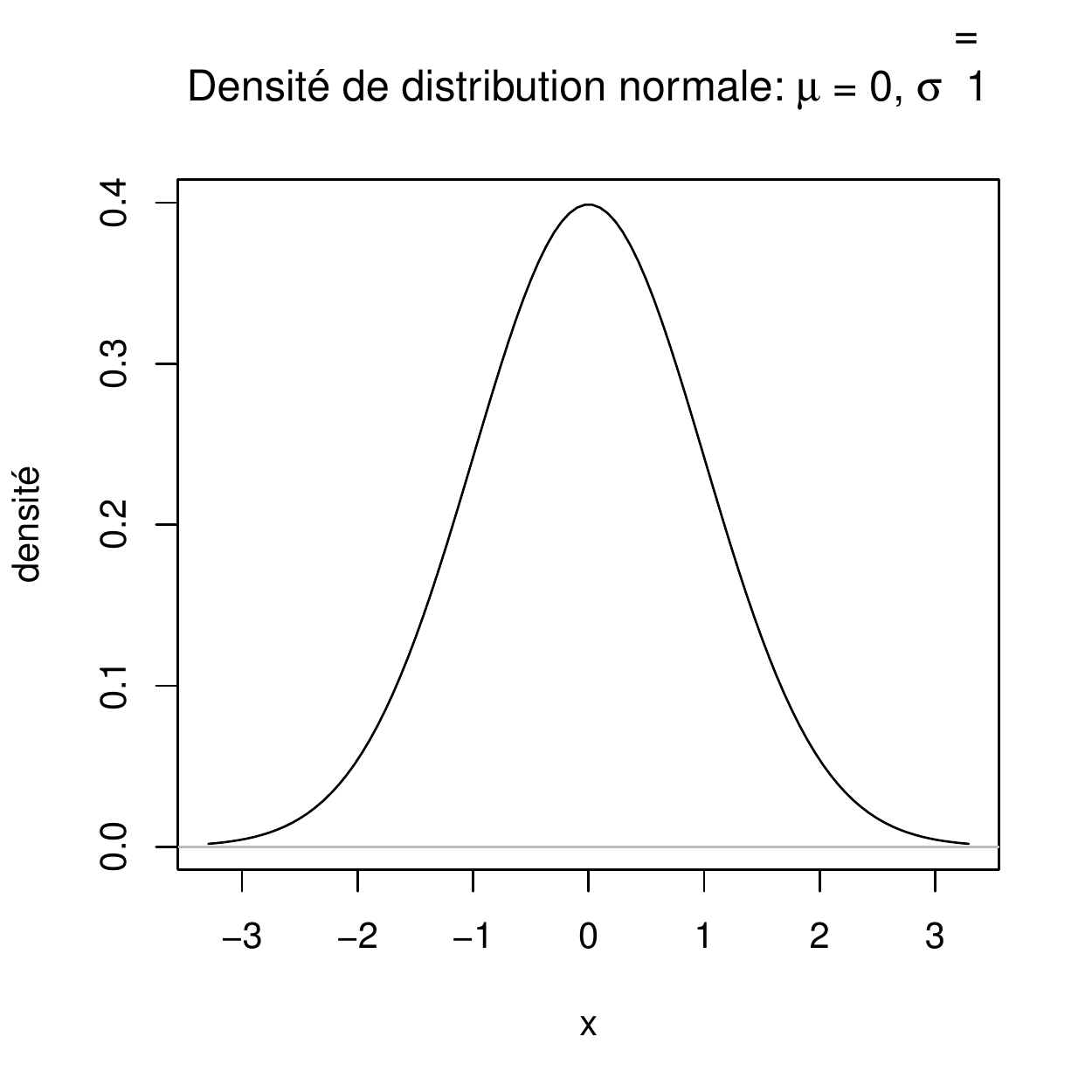}} 
\caption{\label{fig:normale}
  A fun\c c\~ao de densidade da lei gaussiana.}
\end{figure}  

A lei {\em semicircular} \'e dada pela fun\c c\~ao de densidade 
  \[f(t)=\left\{\begin{array}{cl}\frac{2}{\pi}\sqrt{1-t^2},&\mbox{ se }\abs t\leq 1,\\
    0,&\mbox{ se n\~ao.}
  \end{array}\right..\] 

Estritamente falando, o grafo da densidade n\~ao \'e um semic\'\i rculo, mas, melhor, uma semielipse -- o fator normalizador $2/\pi\approx 0.637$ \'e necess\'ario para que a probabilidade de um evento certo seja igual a $1$.

\begin{figure}[ht]
\begin{center}
  \scalebox{0.25}[0.25]{\includegraphics{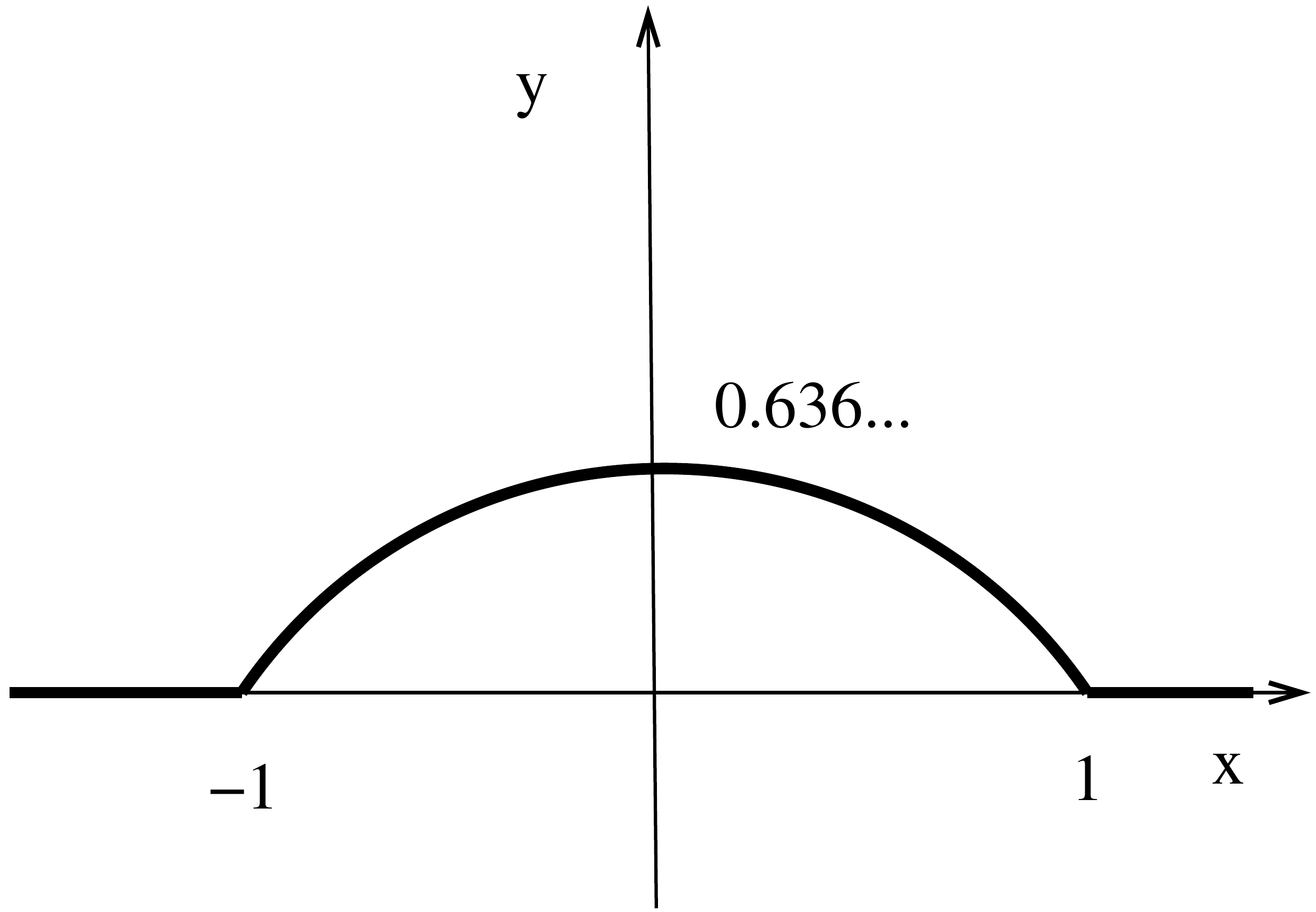}} 
\end{center}
\caption{\label{fig:phi_demicirc}
A densidade da lei semicircular.}
\end{figure}

A no\c c\~ao de uma vari\'avel aleat\'oria n\~ao \'e apenas o \'unico conceito mais fundamental da teoria de probabilidade, mas \'e, sem d\'uvida, uma das mais importantes no\c c\~oes em todas ci\^encias matem\'aticas. Alguns matem\'aticos eminentes argumentam que, eventualmente, os fundamentos da matem\'atica devem ser alterados de modo que as vari\'aveis aleat\'orias sejam t\~ao b\'asicos quanto os conjuntos \citep*{mumford}.

\subsection{\'Algebra de conjuntos borelianos}
At\'e agora, s\'o vimos as vari\'aveis aleat\'orias reais, com valores em $\R$. Mas elas podem assumir valores em dom\'\i nios $\Omega$ mais gerais. 

Seja $\Omega$ um dom\'\i nio geral.
Quais s\~ao as propriedades desejadas da lei, $\mu$, de uma vari\'avel aleat\'oria $X$ com valores em $\Omega$? Claro, os valores da lei pertencem no intervalo $[0,1]$, e a probabilidade que $X\in\Omega$ deve ser $1$:

\begin{enumerate}
\item[(P1)] $P[X\in\Omega]=\mu(\Omega)=1$.
\end{enumerate}

A probabilidade de $x$ pertencer a uni\~ao de uma fam\'\i lia disjunta dos conjuntos  $A_i$, $i\in I$ deve ser igual \`a soma das probabilidades de que $x\in A_i$ para todos $i$:
\[Pr\left[X\in \bigcup A_i\right] = \sum Pr[X\in A_i].\]

Qual \'e o tamanho de tais fam\'\i lias que devemos considerar? 
Se n\'os restringimos a propriedade \`as uni\~oes finitas, a no\c c\~ao de probabilidade resultante \'e muito geral e fraca demais. Neste caso, trata-se das {\em medidas de probabilidade finitamente aditivas}. Elas s\~ao bastante importantes em v\'arias partes da matem\'atica, mas n\~ao s\~ao suficientes para abranger as necessidades da teoria de probabilidade. V\'arios teoremas fundamentais s\~ao falsos neste contexto.

Se, pelo contr\'ario, permitimos as uni\~oes de fam\'\i lias {\em quaisquer}, a no\c c\~ao de probabilidade que obtemos \'e demasiado restritiva.

\begin{exercicio}
Suponha que a lei de uma vari\'avel aleat\'oria $X$ com valores em $\Omega$ lei possui a propriedade: qualquer seja uma fam\'\i lia dos conjuntos $A_i\subseteq \Omega$, $i\in I$ dois a dois disjuntos, $A_i\cap A_j=\emptyset$ para todos $i,j$, $i\neq j$, temos
\[P\left[X\in\cup_{i\in  I}A_i\right] =\sum_{i\in I}P[X\in A_i].\]
Mostrar que isso \'e o caso se e somente se $X$ \'e puramente at\^omica, ou seja, existe uma sequ\^encia (finita ou infinita) de pontos dois-a-dois distintos $a_j\in\Omega$, $j=1,2,3,\ldots$ ({\em \'atomos} de medida $\mu$), com $\mu\{a_j\}=p_j>0$, tal que o valor da medida de um subconjunto $A\subseteq\Omega$ \'e dado por
\[\mu(A)=\sum\{p_j\colon a_j\in A\}=\sum_{j=1}^\infty \chi_A(a_j)p_j=\int_\Omega\chi_Ad\mu.\]
Em particular,
\[\sum_{j=1}^\infty p_j=1.\]
\end{exercicio}

\begin{exercicio}
Mostrar que uma vari\'avel aleat\'oria real \'e discreta se e somente se ela \'e puramente at\^omica. 
\end{exercicio}

A escolha mais natural e frut\'\i fera (ditada pela pr\'atica da pesquisa matem\'atica) \'e a das fam\'\i lias {\em enumer\'aveis}.

\begin{enumerate}
\item[(P2)]
  Se $A_i$, $i=1,2,3,\ldots$ s\~ao disjuntos dois-a-dois, ent\~ao
$\mu(\cup_{i=1}^\infty A_i) = \sum_{i=1}^\infty\mu(A_i)$.
\end{enumerate}

Como um corol\'ario imediato, obtemos, no caso onde $A_1=A$ e $A_2=A^c=X\setminus A$:

\begin{enumerate}
\item[(P2$^\prime$)]
  Se $A\subseteq\Omega$, ent\~ao $P(A^c)=1-P(A)$.
\end{enumerate}

Se $\mu$ \'e a lei de uma vari\'avel puramente at\^omica, ent\~ao o valor 
\[\mu(A) = P[X\in A]\]
\'e bem definido qualquer seja um subconjunto $A\subseteq\Omega$ do dom\'\i nio. Podemos esperar o mesmo para cada vari\'avel aleat\'oria? A resposta \'e negativa. 

A {\em Hip\'otese do Cont\'\i nuo} (Ap\^endice \ref{a:conjuntos}) diz que cada subconjunto infinito de $\R$ ou tem a cardinalidade de $\N$, ou tem a cardinalidade de $\R$. 
Pode-se mostrar \citep*{BK} que, assumindo a validade da Hip\'otese do Cont\'\i nuo, as \'unicas medidas definidas sobre a fam\'\i lia de todos os subconjuntos de $\R$ \'e satisfazendo as propriedades (P1) e (P2) acima s\~ao puramente at\^omicas. 
Como n\~ao queremos que a teoria dependa de hip\'oteses conjunt\'\i sticas adicionais, somos for\c cados a restringir a cole\c c\~ao $\mathscr B$ dos subconjuntos $A\subseteq\Omega$, para as quais o valor $P[X\in A]$ \'e bem definido. Tentemos definir esta fam\'\i lia usando algum senso comum.

O axioma (P1) implica que $\Omega$ sempre pertence \`a fam\'\i lia $\mathscr B$. Segundo o axioma $(P2)$, se uma sequ\^encia de conjuntos pertence a $\mathscr B$,
\[A_1,A_2,\ldots \in {\mathcal B},\]
ent\~ao a sua uni\~ao pertence a $\mathscr B$ tamb\'em:
\[\bigcup_{i}A_i\in{\mathscr B}.\]
Tendo em conta o axioma (P2$^\prime$), conclu\'\i mos que, se $A\in {\mathscr B}$, ent\~ao $A^c\in {\mathscr B}$. Segue-se que a fam\'\i lia $\mathscr B$ deve conter $\Omega$, os complementares de todos os seus membros, e as uni\~oes de subfam\'\i lias enumer\'aveis. Uma fam\'\i lia com estas propriedades \'e dita uma {\em sigma-\'algebra} de subconjuntos de $\Omega$.

Se $\Omega$ \'e um espa\c co m\'etrico, \'e razo\'avel exigir que a lei seja bem definida para todas as bolas abertas:
\[B_r(x) = \{y\in\Omega\colon d(x,y)<r\}.\]
Isso \'e necess\'ario, por exemplo, para conhecer a probabilidade do evento
\[[d(X,x)<r].\]
Mais geralmente, exigimos que a lei seja bem definida para todos os conjuntos {\em abertos,} ou seja, os conjuntos que s\~ao uni\~oes das bolas:
\[V\mbox{ \'e aberto }\iff \forall x\in V,~\exists\e>0,~B_\e(x)\subseteq V.\]

\begin{exercicio}
Seja $\Omega$ um espa\c co m\'etrico. Mostrar que existe a menor sigma-\'algebra que contem todos os conjuntos abertos de $\Omega$.
\end{exercicio} 

\begin{definicao}
Seja $\Omega$ um espa\c co m\'etrico qualquer. 
A {\em menor} fam\'\i lia $\mathscr B$ que cont\'em todos os conjuntos abertos e \'e fechada com rela\c c\~ao aos complementares e uni\~oes de subfam\'\i lias enumer\'aveis, se chama a fam\'\i lia de subconjuntos {\em borelianos} de $\Omega$. 
\end{definicao}

Um espa\c co m\'etrico $\Omega$ \'e dito {\em separ\'avel} se existe um subconjunto enumer\'avel $A$ cujo fecho \'e $\Omega$:
\[\bar A = \Omega.\]

\begin{exercicio} 
Seja $\Omega$ um espa\c co m\'etrico separ\'avel. Mostrar que a fam\'\i lia de conjuntos borelianos \'e a menor sigma-\'algebra que contem todas as bolas abertas.
\end{exercicio} 

\'E claro que todo subconjunto fechado de $\Omega$ \'e boreliano tamb\'em. De fato, os conjuntos borelianos s\~ao muito mais n\'umerosos do que os abertos ou fechados.

\begin{exercicio}
Mostrar exemplos de subconjuntos borelianos de $[0,1]$ que n\~ao s\~ao nem abertos nem fechados.
\end{exercicio}

\begin{definicao} 
Um conjunto $\Omega$ munido de uma sigma-\'algebra $\mathscr B$ de subconjuntos \'e dito um {\em espa\c co boreliano} se existe uma m\'etrica $d$ sobre $\Omega$ tal que $\mathscr B$ \'e exatamente a fam\'\i lia de todos os conjuntos borelianos do espa\c co m\'etrico $(\Omega,d)$.
\index{espa\c co! boreliano}
\end{definicao}

\begin{exercicio}
Seja $\Upsilon$ um subconjunto de um espa\c co boreliano $(\Omega,{\mathscr B})$. Mostrar que $\Upsilon$ se torna um espa\c co boreliano se munido da sigma-\'algebra 
\[{\mathscr B}\vert_{\Upsilon} = \{B\cap \Upsilon\colon B\in {\mathscr B}\}.\]
[ {\em Sugest\~ao:} escolha uma m\'etrica $d$ sobre $\Omega$ que gera a estrutura boreliana $\mathscr B$, e verifique que a estrutura ${\mathscr B}\vert_{\Upsilon} $ \'e gerada pela restri\c c\~ao $d\vert_{\Upsilon}$... ]

O espa\c co boreliano $(\Upsilon,{\mathscr B}\vert_{\Upsilon})$ se chama um {\em subespa\c co boreliano} de $\Omega$.
\label{ex:subespacoboreliano}
\end{exercicio}

\subsection{Medidas de probabilidade borelianas}

\begin{definicao}
Uma fun\c c\~ao $\mu$ na classe $\mathscr B$ dos conjuntos borelianos de $\Omega$ com valores em $[0,1]$ que satisfaz (P1) e (P2) \'e uma {\em medida de probabilidade boreliana}. 
\index{medida! de probabilidade! boreliana}
\end{definicao}

\begin{definicao}
Um espa\c co boreliano $\Omega$ munido de uma medida boreliana de probabilidade $\mu$ \'e dito {\em espa\c co probabil\'\i stico}.
\index{espa\c co! probabil\'\i stico}
\end{definicao}

\begin{exemplo}
Se $\Omega=\Sigma^n$ \'e o cubo de Hamming, ent\~ao a estrutura boreliana do cubo $\Sigma^n$ consiste de todos os seus subconjuntos, e a medida de contagem normalizada $\mu_{\sharp}$ \'e uma medida de probabilidade boreliana. 
\end{exemplo}

\begin{exemplo}
Mais geralmente, seja $\Omega$ um dom\'\i nio qualquer, seja 
\[x_1,x_2,\ldots,x_n,\ldots\]
uma sequ\^encia (finita ou infinita) dos elementos de $\Omega$, e seja
\[p_1,p_2,\ldots,p_n,\ldots\]
uma sequ\^encia do mesmo tamanho dos valores positivos, $p_i>0$, com a propriedade
\[\sum_{i} p_i=1.\]
Ent\~ao a regra
\[\mu(A)=\sum \{p_i\colon x_i\in A\}\]
define uma medida de probabilidade boreliana sobre $\Omega$. De fato, esta medida \'e definida para todo subconjunto $A\subseteq\Omega$, boreliano ou n\~ao. Os conjuntos unit\'arios $\{x_i\}$ tem a medida positiva:
\[\mu\{x_i\}=p_i.\]
\label{ex:discreta}
\end{exemplo}

\begin{definicao}
Uma medida $\mu$ sobre $\Omega$ \'e {\em discreta,} ou {\em puramente at\^omica,} se existe um subconjunto enumer\'avel de $\Omega$ cuja medida \'e um.
\index{medida! puramente at\^omica}
\end{definicao}

\begin{exercicio} Mostrar que
cada medida discreta tem a mesma forma que a medida no exemplo \ref{ex:discreta}.
\end{exercicio}

Afim de dar exemplos matematicamente rigorosos de medidas de probabilidade n\~ao-discretas, tais como a distribui\c c\~ao uniforme sobre o intervalo $[0,1]$, j\'a precisamos mais trabalho t\'ecnico. Vamos fazer isso no ap\^endice \ref{a:medidas}.

\subsection{Fun\c c\~oes borelianas}
Cada medida de probabilidade sobre $\Omega$ \'e a lei de uma vari\'avel aleat\'oria com valores em $\Omega$. Vamos mostrar agora que pode-se compor vari\'aveis aleat\'orias e fun\c c\~oes para obter novas vari\'aveis aleat\'orias.

Sejam $\Omega$ e $W$ dois espa\c cos m\'etricos, e $f\colon \Omega\to W$ uma fun\c c\~ao. Seja $X$ uma vari\'avel aleat\'oria com valores em $\Omega$. Ent\~ao $f(X)$ \'e uma vari\'avel aleat\'oria com valores em $W$. A lei, $\nu$, de $f(X)$ \'e a {\em imagem direta} da lei $\mu$ de $X$ por $f$: se $B\subseteq W$, ent\~ao
\[\nu(B) =\mu(f^{-1}(B)).\]
Isso significa:
\[P[f(X)\in B] = P[X\in f^{-1}(B)].\]
A lei $\nu$ \'e as vezes denotada
\[\nu = f_{\ast}(\mu).\]
\index{imagem direta de medida}
\index{fastmu@$f_{\ast}(\mu)$}

\begin{figure}[ht]
\begin{center}
  \scalebox{0.25}[0.25]{\includegraphics{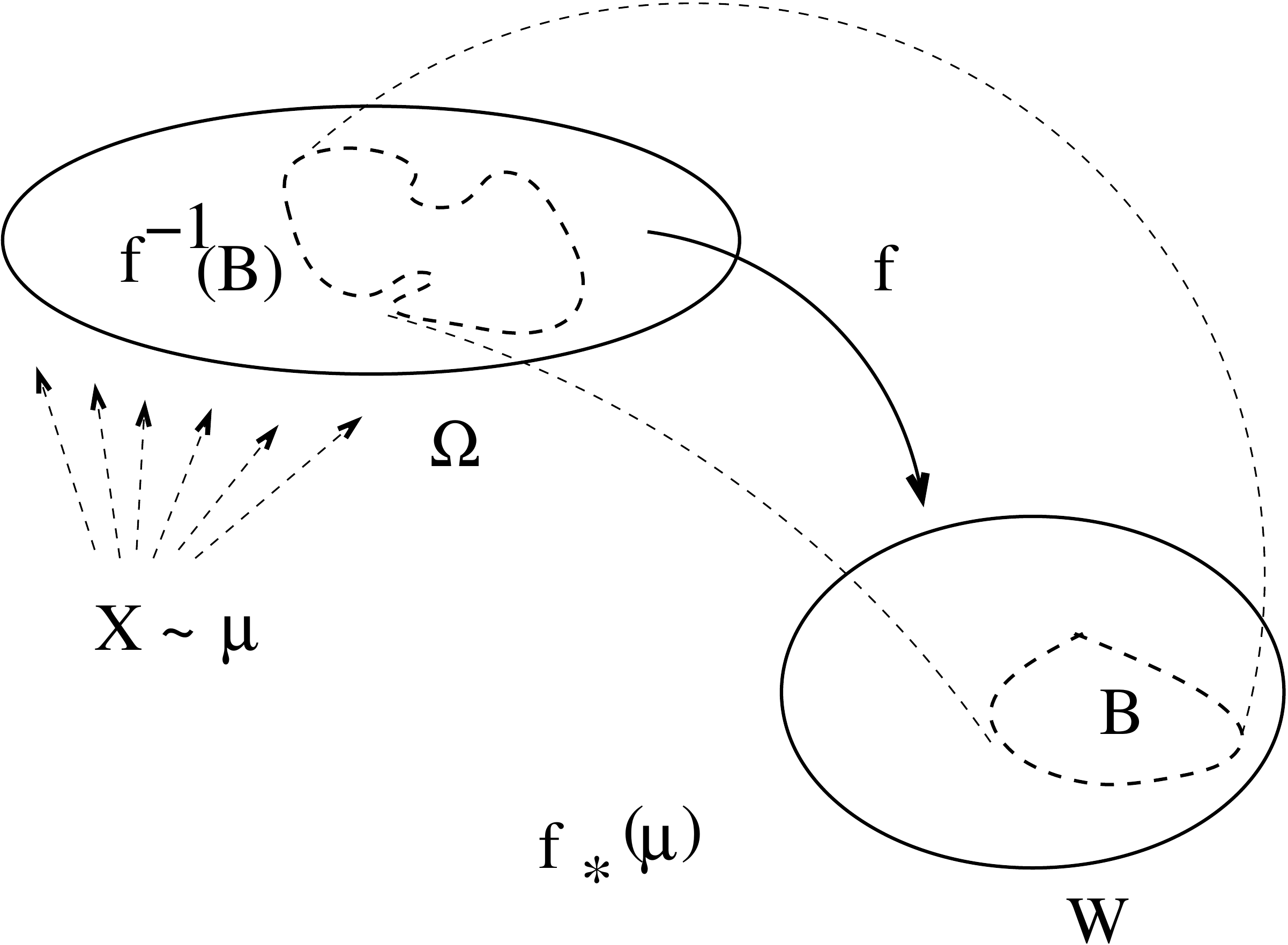}} 
\end{center}
\caption{\label{fig:imagem_direita}
Imagem direta de uma lei.}
\end{figure} 

Para esta defini\c c\~ao funcionar, a \'unica condi\c c\~ao necess\'aria sobre $f$ \'e que a imagem rec\'\i proca de cada subconjunto boreliano $B\subseteq W$ por $f$ seja um conjunto boreliano. Uma tal fun\c c\~ao se chama {\em fun\c c\~ao boreliana}, ou {\em Borel mensur\'avel.}
Pode se verificar que $f\colon\Omega\to W$ \'e boreliana se e somente se a imagem rec\'\i proca de cada subconjunto {\em aberto} de $W$ \'e boreliana. Em particular, cada fun\c c\~ao cont\'\i nua \'e boreliana, mas as fun\c c\~oes borelianas s\~ao muito mais n\'umerosas.
\index{fun\c c\~ao! boreliana}

\begin{exercicio}
  Construir uma fun\c c\~ao boreliana descont\'\i nua.
\end{exercicio}

\begin{exercicio}
Sejam $\Omega$ e $W$ dois espa\c cos m\'etricos, $f\colon \Omega\to W$ uma fun\c c\~ao boreliana, e $\mu$ uma medida de probabilidade boreliana sobre $\Omega$. Mostrar que a imagem direta $f_{\ast}(\mu)$ \'e uma medida de probabilidade boreliana sobre $W$.
\end{exercicio}

\'E neste sentido que cada fun\c c\~ao $f$ sobre um espa\c co de probabilidade $({\mathfrak X},\nu)$ com valores num espa\c co boreliano $\Omega$ pode ser tratada como uma vari\'avel aleat\'oria. A saber, $\nu$ \'e a lei de uma vari\'avel aleat\'oria, digamos $X$, com valores em $\mathscr X$. Ent\~ao, $f(X)$ \'e uma vari\'avel aleat\'oria com valores em $\Omega$, tendo a lei $f_{\ast}(\nu)$. Foi neste sentido que falamos da esperan\c ca de uma fun\c c\~ao real $f$ sobre o cubo de Hamming munido da medida de contagem normalizada.

\begin{definicao}
Uma {\em realiza\c c\~ao} de vari\'avel aleat\'oria $X$ com valores em $\Omega$ e a lei $\mu$ \'e uma fun\c c\~ao boreliana $f$ de um espa\c co de probabilidade $({\mathscr X},\nu)$ com valores em $\Omega$ tal que a imagem direta $f_\ast(\nu)$ \'e igual a $\mu$. 
\index{realiza\c c\~ao de uma vari\'avel aleat\'oria}
\end{definicao}

Cada vari\'avel aleat\'oria pode ser representada desta forma. Por exemplo, se $X\in\Omega$ \'e uma v.a. com a lei $\mu$, podemos definir ${\mathscr X}=\Omega$, $\nu=\mu$, e $f(x)=x$.

\subsection{Espa\c cos borelianos padr\~ao\label{ss:espborpad}}

A aprendizagem estat\'\i stica -- pelo menos, dentro do paradigma atual -- ou n\~ao \'e fact\'\i vel nos dom\'\i nios n\~ao separ\'aveis, ou se reduz a um subdom\'\i nio separ\'avel. Por isso, vamos supor que $\Omega$ seja separ\'avel. Uma outra restri\c c\~ao natural \'e que o dom\'\i nio seja metriz\'avel com uma m\'etrica completa. Isto leva-nos ao conceito seguinte.

\begin{definition}
Um {\em espa\c co boreliano padr\~ao} \'e um par $(\Omega,{\mathscr B})$ que consiste de um conjunto $\Omega$ e uma sigma-\'algebra $\mathscr B$, que \'e a estrutura boreliana gerada por uma m\'etrica completa e separ\'avel sobre $\Omega$.
\index{espa\c co! boreliano! padr\~ao}
\end{definition}

Relembramos que um espa\c co m\'etrico $(X,d)$ \'e {\em separ\'avel} se existe um subconjunto enumer\'avel $Y$ denso em $X$:
\[\bar Y =X.\]

\begin{exercicio}
Mostrar que um subespa\c co de um espa\c co m\'etrico separ\'avel \'e separ\'avel.
\par
[ Para uma prova, veja teorema \ref{t:subespsep}. ]
\end{exercicio}

Acontece que os espa\c cos borelianos padr\~ao distintos s\~ao muito poucos e admitem uma classifica\c c\~ao completa simples.

\begin{definicao}
Um {\em isomorfismo} entre dois espa\c cos borelianos $(\Omega_i,{\mathscr B}_i)$, $i=1,2$ \'e uma bije\c c\~ao $f\colon\Omega_1\to\Omega_2$ conservando os conjuntos borelianos: qual quer seja $B\subseteq\Omega_1$, temos
\[B\in {\mathscr B}_1\iff f(B)\in {\mathscr B}_2.\]
\label{d:isomorfismoboreliano}
\index{isomorfismo! boreliano}
\end{definicao}

\begin{teorema}
Dois espa\c cos borelianos padr\~ao s\~ao isomorfos se e somente se eles tem a mesma cardinalidade. 
\label{t:isomorfismo}
\end{teorema}

A prova do teorema ocupa o Ap\^endice \ref{apendice:padrao}. Em particular, vamos ver que cada espa\c co boreliano enumer\'avel \'e isomorfo ao espa\c co discreto da mesma cardinalidade cuja estrutura boreliana consiste de todos os subconjuntos; enquanto cada espa\c co boreliano padr\~ao n\~ao enumer\'avel \'e isomorfo a $\R$ com a sua estrutura boreliana can\^onica. 

\subsection{Independ\^encia\label{s:independencia}}

\begin{definicao}
Se $\Omega$ \'e um espa\c co boreliano padr\~ao e $\mu$ \'e uma medida de probabilidade boreliana (ou seja, definida para todos elementos da estrutura boreliana $\mathscr B$ de $\Omega$), ent\~ao $(\Omega,\mu)$ \'e dito {\em espa\c co probabil\'\i stico padr\~ao}.
\index{espa\c co! probabil\'\i stico! padr\~ao}
\end{definicao}

Se temos mais de uma vari\'avel aleat\'oria,
\[X_1,X_2,\ldots,X_n,\ldots,\]
tomando os valores, respectivamente, nos espa\c cos $\Omega_1,\Omega_2$, $\ldots$, $\Omega_n,\ldots$, ent\~ao elas podem ser combinadas numa \'unica vari\'avel aleat\'oria, tomando os valores no produto dos espa\c cos $\Omega_i$:
\[X=(X_1,X_2,\ldots,X_n,\ldots)\in \Omega_1\times\Omega_2\times\ldots\times\Omega_n.\]
A lei $\mu$ da vari\'avel $X$ \'e chamada a {\em lei conjunta} das vari\'aveis $X_1,X_2,\ldots,X_n,\ldots$. 

As vari\'aveis aleat\'orias $X_1,X_2,\ldots,X_n,\ldots$ s\~ao ditas {\em independentes} se, cada vez que $A_i$ \'e um subconjunto boreliano de $\Omega_i$, $i=1,2,\ldots$, temos
\begin{eqnarray*}
Pr[X_1\in A_1,X_2\in A_2,\ldots,X_n\in A_n,\ldots] = \\
Pr[X_1\in A_1]\times Pr[X_2\in A_2]\times\ldots\times Pr[X_n\in A_n]\times\ldots.\end{eqnarray*}

Nota\c c\~ao:
\[\mu=\otimes_{i=1}^\infty \mu_i.\]
Esta $\mu$ \'e tamb\'em chamada a medida produto das medidas de probabilidade $\mu_1,\mu_2,\ldots$.

Por exemplo, sejam $X$ e $Y$ duas v.a., cada uma de lei uniforme sobre o intervalo $[0,1]$. Se $X$ e $Y$ s\~ao independentes, isso significa que a vari\'avel aleat\'oria $Z=(X,Y)$ com valores no quadrado $[0,1]^2$ tem lei, $\mu$, que \'e uniforme no quadrado: quaisquer sejam $a,b,c,d$, $a\leq b$, $c\leq d$, temos
\[\mu([a,b]\times [c,d]) = (b-a)(d-c).\]

Por outro lado, se por exemplo $Y=X$, ent\~ao a lei da vari\'avel $Z=(X,Y)$ \'e concentrada na diagonal do quadrado: se $A_1,A_2\subseteq [0,1]$ s\~ao disjuntos, ent\~ao, obviamente,
\[P[X\in A_1,Y\in A_2]=0,\]
de onde \'e f\'acil de concluir que
\[\mu(\Delta)=1,\]
onde
\[\Delta = \{(x,x)\colon x\in [0,1]\}.\]

Se $X_1,X_2,\ldots,X_n$ \'e uma sequ\^encia das vari\'aveis aleat\'orias independentes distribu\'\i das segundo a lei gaussiana em $\R$, ent\~ao sua lei conjunta \'e a lei gaussiana $n$-dimensional em $\R^d$, determinada pela densidade
\[\frac{1}{(2\pi)^{n/2}} e^{-(t_1^2+t_2^2+\ldots+t_n^2)/2}.\]
Isso significa que, qualquer que seja um subconjunto boreliano $A\subseteq \R^d$,
\[P[X\in A] = \frac{1}{(2\pi)^{n/2}}\int_A e^{-(t_1^2+t_2^2+\ldots+t_n^2)/2}dt_1\ldots dt_n.\]
\index{medida! gaussiana}

%
%

\chapter{Elementos da teoria de conjuntos\label{a:conjuntos}}

\section{Cardinalidade e ordens} 

\subsection{Cardinalidade}
\begin{definicao}
Sejam $X$ e $Y$ dois conjuntos quaisquer. Eles s\~ao {\em equipotentes,} ou {\em t\^em a mesma cardinalidade,} se existe uma aplica\c c\~ao bijetora entre eles:
\[f\colon X\to Y,\]
ou seja, $f$ \'e injetora ($\forall x,y\in X$, $x\neq y\rightarrow f(x)\neq f(y)$) e sobrejetora ($\forall y\in Y$, $\exists x\in X$ tal que $f(x)=y$). 
Nota\c c\~ao:
\[\abs X=\abs Y.\]
Para conjuntos finitos, num contexto combinat\'orio usamos tamb\'em a nota\c c\~ao
\[\sharp X =\sharp Y.\]
\end{definicao}

\begin{observacao}
A rela\c c\~ao da equipot\^encia \'e uma rela\c c\~ao da equival\^encia entre conjuntos. Ela \'e
\begin{enumerate}
\item reflexiva: $\abs X=\abs X$,
\item sim\'etrica: se $\abs X=\abs Y$, ent\~ao $\abs Y=\abs X$, e
\item transitiva: se $\abs X=\abs Y$ e $\abs Y=\abs Z$, ent\~ao $\abs X=\abs Z$.
\end{enumerate}
Todas as tr\^es propriedades s\~ao f\'aceis a verificar.
\end{observacao}

\begin{definicao}
Seja $X$ um conjunto.
A classe de equival\^encia de $X$ relativo \`a rela\c c\~ao de equipot\^encia \'e denotado $\abs X$. Ela \'e chamado a {\em cardinalidade} de $X$.

En particular, a cardinalidade do conjunto $\N$ dos n\'umeros naturais \'e denotado $\aleph_0$ (l\^e: \'alefe zero):
\[\abs\N = \aleph_0.\]
A cardinalidade do conjunto $\R$ dos n\'umeros reais \'e denotado ou $\mathfrak c$ ou $2^{\aleph_0}$.
\index{cardinalidade}
\index{c@$\mathfrak c$}
\index{aleph0@$\aleph_0$}
\end{definicao}

\begin{definicao}
Seja $X$ e $Y$ dois conjuntos. Digamos que a cardinalidade de $X$ \'e maior ou igual \`a cardinalidade de $Y$,
\[\abs X\geq \abs Y,\]
se existe uma inje\c c\~ao de $Y$ em $X$.
\end{definicao}

\begin{definicao}
Digamos que a cardinalidade de $X$ \'e estritamente maior que a cardinalidade de $Y$,
\[\abs X > \abs Y,\]
se $\abs X\geq \abs Y$ e n\~ao \'e verdadeiro que $\abs X=\abs Y$.
\end{definicao}

\begin{teorema}[Cantor--Bernstein] Sejam $X$ e $Y$ dois conjuntos tais que $\abs X\leq \abs Y$ e $\abs Y\leq \abs X$. Ent\~ao
\[\abs X=\abs Y.\]
\index{teorema! de Cantor--Bernstein}
\end{teorema}

\begin{proof}
Definamos $X_1=X\setminus g(Y)$, $Y_1=Y\setminus f(X)$, e depois, da maneira recursiva,
\[X_{n+1}=g(Y_n),\]
\[Y_{n+1}=f(X_n),\]
para todos $n\in\N_+=\{1,2,\ldots\}$.

\begin{figure}[ht]
\begin{center}
\scalebox{0.3}[0.3]{\includegraphics{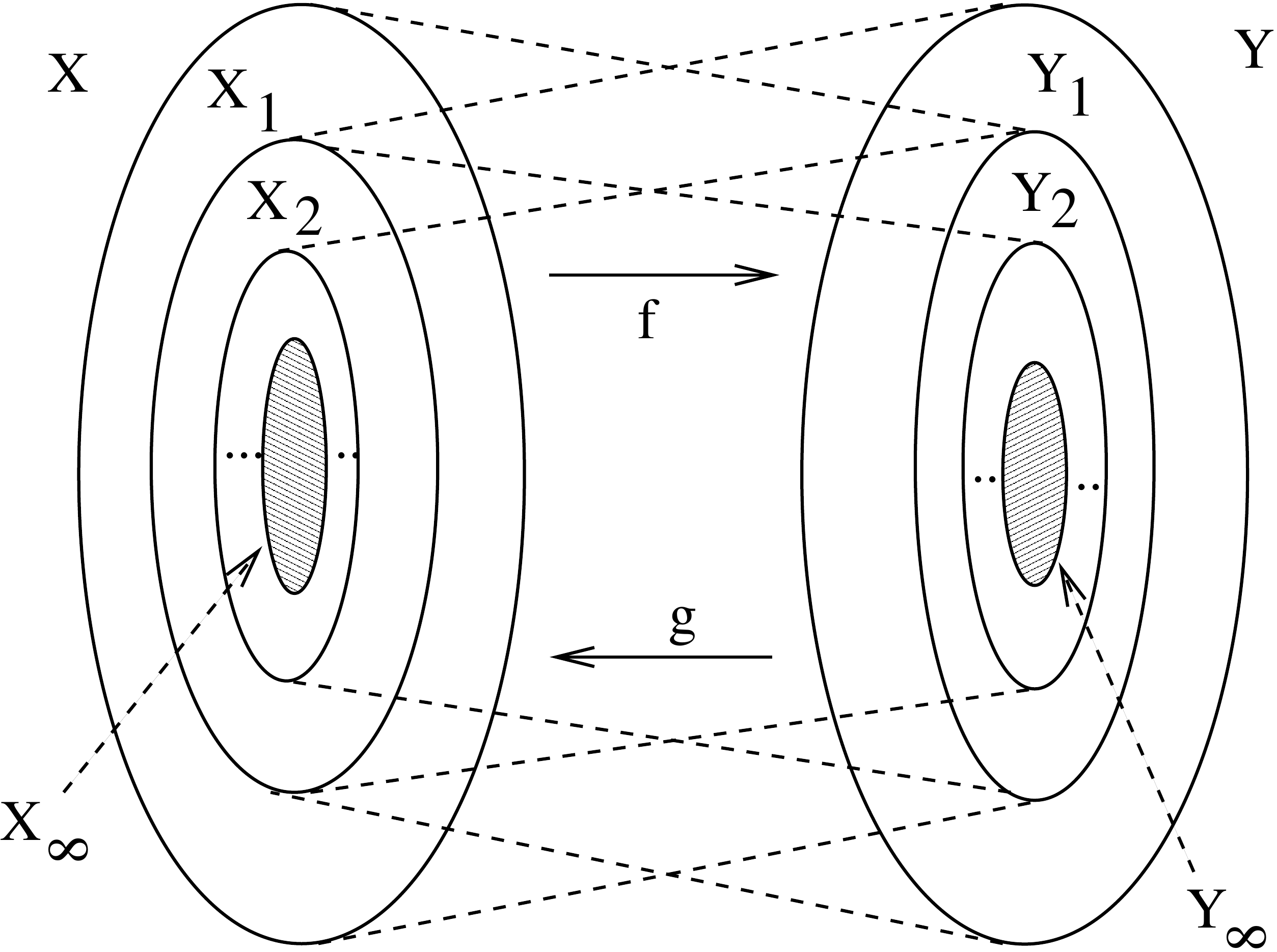}} 
\caption{Prova do teorema de Cantor--Bernstein.}
\label{fig:cb}
\end{center}
\end{figure}

Os conjuntos $X_n$, $n=1,2,3,\ldots$ s\~ao dois a dois disjuntos, bem como os conjuntos $Y_n$. Definamos
\[X_{\infty} = X\setminus \cup_{n=1}^\infty X_n,\]
\[Y_{\infty} = Y\setminus \cup_{n=1}^\infty Y_n.\]

Para cada $x\in X$, seja
\begin{equation}
\label{eq:h}
h(x) =\begin{cases}
f(x),&\mbox{ si }x\in X_{2n+1},~n=1,2,3,\ldots,\\
g^{-1}(x),&\mbox{ si }x\in X_{2n},~n=1,2,3,\ldots,\\
f(x),&\mbox{ si }x\in X_{\infty}.
\end{cases}
\end{equation}
Agora n\~ao \'e dif\i cil verificar que $h\colon X\to Y$ \'e uma fun\c c\~ao bijetora.
\end{proof}

\begin{exercicio}
Verifique que $h\colon X\to Y$ na eq. (\ref{eq:h}) \'e bijetora.
\par
({\em Sugest\~ao:} mostre que a aplica\c c\~ao $h^{\ddag}\colon Y\to X$ definida por
\[h^{\ddag}(y) = \begin{cases}
g(y),&\mbox{ si }y\in Y_{2n+1},~n=1,2,3,\ldots,\\
f^{-1}(y),&\mbox{ si }y\in Y_{2n},~n=1,2,3,\ldots,\\
f^{-1}(x),&\mbox{ si }y\in Y_{\infty}.
\end{cases}
\]
\'e a aplica\c c\~ao inversa de $h$, em outras palavras, $h\circ h^\ddag = {\mathrm{Id}}_Y$ e $h^\ddag\circ h = {\mathrm{Id}}_X$.)
\end{exercicio}

\begin{teorema}
\label{th:comparabilidade}
Sejam $X$ e $Y$ dois conjuntos quaisquer. Ent\~ao, ou $\abs X\leq \abs Y$, ou $\abs Y\leq \abs X$.
\end{teorema}

Deixemos a prova do teorema \ref{th:comparabilidade} como exerc\'\i cio \ref{ex:comparabilidade} (uma aplica\c c\~ao do lema de Zorn, a ser discutido mais tarde).

\begin{corolario}
Sejam $X$ e $Y$ dois conjuntos quaisquer. Ent\~ao uma e uma s\'o possibilidade entre as seguintes tem lugar:
\begin{itemize}
\item $\abs X <\abs Y$,
\item $\abs X=\abs Y$, ou
\item $\abs X>\abs Y$.
\end{itemize}
\end{corolario}

\begin{definicao}
Seja $X$ um conjunto. Denotaremos $2^X$ o conjunto de todas as partes de $X$:
\[2^X = \{Y\colon Y\subseteq X\}.\]
\end{definicao}

\begin{teorema}
Seja $X$ um conjunto qualquer. Ent\~ao,
\[\abs X<\abs{2^X}.\]
\end{teorema}

\begin{proof}
Para come\c car, a aplica\c c\~ao $H\colon X\to 2^X$ dada por
\[H(x) =\{x\}\]
\'e obviamente injetora, de onde conclu\'\i mos:
\[\abs X\leq \abs{2^X}.\]
Resat mostrar que
\[\abs X\neq\abs{2^X}.\]
Suponha, para uma contradi\c c\~ao, que os dois cardinais sejam iguais:
\begin{equation}
\label{eq:hypo}
\abs X = \abs{2^X}.
\end{equation}
Isto significa que existe uma aplica\c c\~ao bijetora
\[F\colon X\bij 2^X.\]
Definamos
\[A = \{x\in X\colon x\notin F(x)\}.\]
Como $F$ \'e sobrejetora, existe $x_0\in X$ tal que
\[F(x_0)=A.\]
Perguntamos: $x_0\in A$? Segundo a defini\c c\~ao do conjunto $A$, temos $x_0\in A$ se e somente se $x_0\notin A$. \'E obviamente uma afirma\c c\~ao autocontradit\'oria. Deduzimos: a hip\'otese inicial (\ref{eq:hypo}) \'e falsa.
\end{proof}

\begin{exemplo}
\label{ex:dva}
Mostremos que $\abs {2^{\N}}=\abs \R$.

Associamos a cada subconjunto $A$ de $\N$ a sua fun\c c\~ao carater\'\i stica, ou seja, uma sequ\^encia dos zeros e uns da maneira que se $n\in A$, ent\~ao na $n$-\'esima posi\c c\~ao temos $1$, e se $n\notin A$, temos $0$. Por exemplo, ao subconjunto dos n\'umeros naturais pares associamos a sequ\^encia 
\[0,1,0,1,0,1,\ldots\]
Desta maneira, pensaremos de $2^\N$ como o conjunto de todas as sequ\^encias bin\'arias.

A aplica\c c\~ao que associa \`a uma sequ\^encia
\[\e_1\e_2\e_3\ldots,~~\e_i\in\{0,1\},\]
o \'unico n\'umero real 
\[\sum_{n=1}^{\infty} 10^{-n}\e_n\]
cuja expans\~ao decimal \'e esta sequ\^encia, \'e injetora. Logo,
\[\abs{2^\N}\leq \abs\R.\]

De outro lado, definamos uma inje\c c\~ao de $\R$ em $2^{\N}$ como a composi\c c\~ao de uma bije\c c\~ao qualquer entre $\R$ e o intervalo $(0,1)$ e a aplica\c c\~ao que envia $x\in (0,1)$ na sequ\^encia dos d\'\i gitos da sua expan\c c\~ao bin\'aria,
\[x=\sum_{n=1}^\infty 2^{-n}\e_n\mapsto (\e_1,\e_2,\e_3,\ldots),~~\e_n\in\{0,1\}.\]
Esta inje\c c\~ao $\R\to 2^\N$ estabelece que
\[\abs{\R}\leq \abs{2^\N}.\]
Conclu\'\i mos usando o teorema de Cantor--Bernstein.
\tri
\end{exemplo}

\begin{corolario}
$\aleph_0<{\mathfrak c}$.
\end{corolario}

A {\em hip\'otese do cont\'\i nuo} (Continuum Hypothesis, CH) afirma que n\~ao existem cardinalidades estritamente intermedi\'arias entre $\aleph_0$ e ${\mathfrak c}$. Vamos discutir a CH em mais detalhas no final do Ap\^endice.

\subsection{Ordens e permuta\c c\~oes}

\begin{definicao}
Sejam $X$ um conjunto e $R$ uma rela\c c\~ao bin\'aria sobre $X$: $R\subseteq X^2$. Diz-se que $R$ \'e uma {\em rela\c c\~ao de ordem parcial}, ou simplesmente um {\em ordem parcial,} se ela verifica as condi\c c\~oes seguintes:
\begin{enumerate}
\item\label{ax:refl} $R$ \'e reflexiva: $xRx$ para todos $x\in X$.
\item $R$ \'e antissim\'etrica: se $xRy$ e $yRx$, ent\~ao $x=y$.
\item\label{ax:trans} $R$ \'e transitiva: se $xRy$ e $yRz$, ent\~ao $xRz$.
\end{enumerate}
Habitualmente, uma rela\c c\~ao de ordem parcial \'e denotada pelo s\'\i mbolo $\leq$. Um par $(X,\leq)$ que consiste de um conjunto $X$ munido de uma ordem parcial $\leq$ \'e dito {\em conjunto parcialmente ordenado}.
\label{d:ordemparcial}
\end{definicao}

\begin{definicao}
Uma ordem parcial $\leq$ sobre um conjunto $X$ \'e dita {\em ordem total} se  todos elementos de $X$ s\~ao compar\'aveis dois a dois entre eles:
\[\forall x,y\in X,~~\mbox{ ou bem }x\leq y\mbox{, ou bem }y\leq x.\]
Nesto caso, diz-se que o conjunto $(X,\leq)$ \'e {\em totalmente ordenado,} ou {\em linearmente ordenado.}
\label{d:ordemtotal}
\index{ordem! parcial}
\index{ordem! total}
\end{definicao}

\begin{definicao}
Uma rela\c c\~ao \'e dita {\em pr\'e-ordem} ({\em parcial}) se ela satisfaz os axiomas (\ref{ax:refl}) e (\ref{ax:trans}) da defini\c c\~ao \ref{d:ordemparcial}. Usamos as pr\'e-ordens algumas vezes no texto principal. Usualmente, {\em pr\'e-ordem} significa uma pr\'e-ordem total, ou seja, todos elementos de $X$ s\~ao dois a dois compar\'aveis.
\index{pr\'e-ordem}
\end{definicao}

\begin{exemplo}
A reta $\R$, munida da ordem usual, \'e totalmente ordenada.
\end{exemplo}

\begin{exemplo}
Seja $X$ um conjunto qualquer. A fam\'\i lia $2^X$ das todas as partes de $X$ \'e parcialmente ordenada pela rela\c c\~ao da inclus\~ao:
\[A\leq B\iff A\subseteq B.\]
Se $\abs{X}\geq 2$, esta ordem n\~ao \'e total: existem sempre pelo menos dois subconjuntos $A,B\subseteq X$ que n\~ao s\~ao compar\'aveis entre eles:
\[A\not\subseteq B\mbox{ e }B\not\subseteq A.\]
\end{exemplo}

\begin{definicao}
\label{tos}
Um conjunto totalmente ordenado $(X,\leq)$ \'e dito ser {\it bem ordenado} se cada subconjunto n\~ao vazio, $A\subseteq X$, tem um elemento m\'\i nimo: existe
$a\in A$ tal que $a\leq y$ para todos $y\in A$.
\index{ordem! boa}
\end{definicao}

\begin{exemplo} O conjunto $\N$ de n\'umeros naturais \'e bem ordenado. Se $A$ \'e um subconjunto n\~ao vazio de n\'umeros naturais, escolha $n\in A$ qualquer; o conjunto de todos
$a\in A$ com $a\leq n$ \'e finito e n\~ao vazio, e um simples argumento indutivo mostra que ele cont\'em o elemento menor, $m$. Este $m$ \'e tamb\'em o elemento menor de $A$.
\end{exemplo}

\begin{exemplo} O conjunto $\Z$ de n\'umeros inteiros, munido da ordem usual, n\~ao \'e bem ordenado, porque o pr\'oprio $\Z$ n\~ao tem um elemento m\'\i nimo.
\end{exemplo}

\begin{exemplo} 
O conjunto vazio $\emptyset$ munido da sua \'unica ordem (uma rela\c c\~ao vazia) \'e bem ordenado.
\end{exemplo}

\begin{definicao} Seja $E$ um conjunto qualquer (finito ou infinito). Uma {\em permuta\c c\~ao de} $E$ \'e uma bije\c c\~ao de $E$ sobre si mesmo:
\[\sigma\colon E\bij E.\]
O conjunto de todas as permuta\c c\~oes de $E$ \'e denotado $S_E$. Se $E$ \'e um conjunto finito da cardinalidade $n$, as permuta\c c\~oes de $E$ s\~ao ditas as {\em do posto} $n$.
\index{permuta\c c\~oes}
\end{definicao}

\begin{definicao} Seja $n$ um n\'umero inteiro natural. Na tradi\c c\~ao combinat\'oria, nota-se
\[[n] = \{1,2,3,\ldots,n\}.\]
\end{definicao}

\begin{definicao}
O conjunto de permuta\c c\~oes de $[n]$ se nota $S_n$.
\index{Sn@$S_n$}
\end{definicao}

Dado uma permuta\c c\~ao $\sigma$ de $[n]$, obtemos uma ordem $\preceq$ sobre $[n]$, definindo 
\[i\preceq j\iff \sigma(i)\leq \sigma(j).\]
\'E f\'acil ver que $\preceq$ \'e uma ordem total. O resultado seguinte mostra que toda ordem total sobre $[n]$ vem deste modo da ordem usual, permutando os elementos de $[n]$ entre eles.

\begin{teorema}
Dada uma ordem total qualquer, $\preceq$, sobre $[n]$, existe uma e uma s\'o  permuta\c c\~ao $\sigma\in S_n$ tal que, quaisquer sejam $i,j\in [n]$, temos
\begin{equation}
i\preceq j\iff \sigma(i)\leq \sigma(j),
\label{eq:prec}
\end{equation}
onde $\leq$ \'e a ordem usual sobre os n\'umeros inteiros.
\end{teorema}

\begin{proof} 
A propriedade na eq. (\ref{eq:prec}) pode ser reescrita da maneira equivalente como segue: quaisquer sejam $i,j\in [n]$,
\begin{equation}
\sigma^{-1}(i)\preceq \sigma^{-1}(j)\iff i\leq j.
\label{eq:prec2}
\end{equation}
Como $1$ \'e o menor elemento de $[n]$ na ordem usual, $\sigma^{-1}(1)$ dever ser o menor elemento de $[n]$ em rela\c c\~ao \`a ordem $\preceq$. Ent\~ao, n\~ao h\'a escolha para $\sigma^{-1}(1)$: deve ser $\sigma^{-1}(1)=\min\{[n],\preceq\}$. 
Como cada conjunto totalmente ordenado finito possui o m\'\i nimo, definamos
\[\sigma^{-1}(1) = \min\{[n],\preceq\}.\]
O n\'umero $2$ \'e o m\'\i nimo de $\{2,3,\ldots,n\}$, e usando eq. (\ref{eq:prec2}), conclu\'\i mos que $\sigma^{-1}$ deve ser o menor elemento do conjunto ordenado $([n],\preceq)$ menos $\sigma^{-1}(1)$. Definamos
\[\sigma^{-1}(2) = \min\{[n]\setminus\{\sigma^{-1}(1)\},\preceq\}.\]
De modo recursivo, se as imagens de  $1,2,\ldots,k$ por $\sigma^{-1}$ tinham sido escolhidas, definamos 
\[\sigma^{-1}(k+1) = \min\{[n]\setminus\{\sigma^{-1}(1),
\sigma^{-1}(2),\ldots,\sigma^{-1}(k)\},\preceq\}.\]
\'E claro que a fun\c c\~ao $\sigma^{-1}\colon [n]\to[n]$ escolhida deste modo \'e injetora, porque a cada passo a imagem de $k+1$ por $\sigma^{-1}$ foi diferente dos valores j\'a escolhidos para $1,2,3,\ldots,k$. Como $[n]$ \'e um  conjunto finito, $\sigma^{-1}$ \'e bijetora, \'e possui a aplica\c c\~ao inversa, que denotemos por $\sigma$. 

Esta $\sigma$ \'e uma permuta\c c\~ao de $[n]$, que, segundo a sua defini\c c\~ao, verifica a propriedade (\ref{eq:prec2}), logo a propriedade (\ref{eq:prec}) tamb\'em. 

Para mostrar a unicidade de $\sigma$, notemos que a escolha de cada valor por $\sigma^{-1}$ possuindo a propriedade (\ref{eq:prec2}) foi \'unica.
\end{proof}

\begin{definicao}
Dado um n\'umero natural, $n$, o {\em fatorial} de $n$, denotado $n!$, \'e um n\'umero natural definido recursivamente como segue:
\begin{enumerate}
\item $1!=1$, et
\item $(n+1)!=(n+1)n!$.
\end{enumerate}
Ent\~ao, pode-se escrever:
\[n! = n(n-1)(n-2)\ldots3\cdot 2\cdot 1 =\prod_{i=1}^n i.\]
\end{definicao}

\begin{exercicio}
$\sharp S_n=n!$.
\end{exercicio}

 \begin{observacao}
O problema seguinte, chamado o {\em problema do caixeiro-viajante}
(em ingl\^es: {\em travelling salesman problem,} ou {\em TSP}), \'e provavelmente o mais importante problema em aberto da inform\'atica te\'orica moderna.  
 
Dado $n$ pontos (as ``cidades'') e as dist\^ancias dois a dois entre eles, o problema \'e, calcular, em tempo polinomial em $n$, a rota mais curta que visita todas as cidades uma \'unica vez.

\'E claro que o problema pode ser resolvido enumerando todos os $(n-1)!$ caminhos poss\'iveis \'e comparando-os dois a dois para escolher o mais curto. Mas este algoritmo vai exigir o tempo (n\'umero de opera\c c\~oes) proporcional a $(n-1)!$, que \'e uma quantidade enorme. Segundo a {\em f\'ormula de Stirling,}
\[n! \approx \sqrt{2\pi n}\left(\frac n e\right)^n.\]
Por exemplo,
\[100! \approx 9.3326\cdot 10^{157}.\]
O algoritmo mais eficaz conhecido exige o tempo da ordem de grandeza $2^nn^2$, o que \'e exponencial em $n$.
 \end{observacao}

\section{Sobre o sistema axiom\'atico ZFC}

\subsection{Sistema ZF}
O que isso significa, provar um teorema? Na compreens\~ao moderna, significa:
deduzi-lo a partir de um conjunto de axiomas, usando uma cole\c c\~ao de ferramentas l\'ogicas exatamente especificadas.

Os axiomas que t\^em mais ou menos toda a matem\'atica atual baseada
sobre eles, s\~ao os chamados axiomas ZFC da teoria do conjuntos axiom\'atica.
Este axiom\'atica remonta ao trabalho de Cantor, que foi o primeiro a propor
uma cole\c c\~ao de axiomas que regem a teoria dos conjuntos. Mais tarde, sua axiom\'atica foi corrigida e atualizada por Zermelo, Fraenkel, e Skolem. A origem da sigla ZFC \'e:

\begin{itemize}
\item {\bf Z} para Ernst Zermelo,
\item {\bf F} para Adolf Abraham Fraenkel,
\item {\bf C} para Axioma de Escolha (Axiom of Choice).
\end{itemize}

Para dar uma ideia de como esses axiomas se parecem, deixe-me declarar alguns axiomas da lista original de 7 axiomas da teoria do conjuntos propostos pelo pr\'oprio Georg Cantor. 

\paragraph{\bf Axioma do Conjunto Vazio} Existe um conjunto, denotado $\emptyset$, que cont\'em nenhum elemento:
\[\forall x,~~ \neg(x\in\emptyset).\]
Aqui o s\'\i mbolo $\neg$ significa nega\c c\~ao l\'ogica. A nota\c c\~ao mais familiar para $\neg(x\in \emptyset)$ seria
$x\notin\emptyset$. 
\paragraph{\bf Axioma da Extens\~ao}
Dois conjuntos, $x$ e $y$, s\~ao iguais se e somente se eles consistem dos mesmos elementos:
\[\forall x\forall y~~(x=u)\Leftrightarrow \forall z~
(z\in x)\leftrightarrow (z\in y)\]

\paragraph{\bf Axioma do Par} \
Este axioma permite, a partir de dois conjuntos $x,y$, criar um novo conjunto, $\{x,y\}$, que cont\'em exatamente dois elementos, $x$ e $y$:
\[\forall x\forall y\exists z~~\forall t~(t\in z)\leftrightarrow
(t=x)\vee (t=y)\]

\paragraph{\bf Axioma da Pot\^encia} 
Dado um conjunto $X$, pode-se formar um novo conjunto cujos elementos s\~ao todos os subconjuntos de $X$. Este conjunto \'e chamado {\em conjunto pot\^encia} de $X$ e denotado $2^X$ ou $2^X$. Em nota\c c\~ao l\'ogica,
\[\forall x\exists y\forall z~~(z\in y)\leftrightarrow (z\subseteq x).\]
Aqui $y\subseteq x$ \'e uma substitui\c c\~ao para
\[\forall z~~(z\in y)\rightarrow (z\in x).\]

Um dos axiomas originais de Cantor tinha postulado que se pode formar novos conjuntos por suas propriedades. Isso \'e realmente o que estamos fazendo todos os dias quando fazemos
matem\'atica, e assim um axioma deste tipo \'e vital.

\paragraph{\bf Axioma de Compreens\~ao}
Seja $\mathcal P$ uma propriedade de conjuntos que pode se exprimir em uma linguagem formal apropriada. Ent\~ao a cole\c c\~ao de todos os conjuntos $x$ que possuem a propriedade 
$\mathcal P$,
\[\{x\colon {\mathcal P}(x)\},\]
\'e um conjunto. 

No entanto, este axioma leva a problemas, que foi notado pela primeira vez em 1902 por Bertrand Russell. 
Esta contradi\c c\~ao, conhecida como a {\em antinomia de Russell,} causou alguns problemas para as funda\c c\~oes axiom\'aticas nascentes da teoria de conjuntos.
Em ess\^encia, \'e simplesmente uma reformula\c c\~ao do bem conhecido paradoxo do barbeiro.

\paragraph{\bf Antinomia de Russell}
Seja $\mathcal P$ a propriedade de um conjunto $x$ seguinte: 
\[x\notin x.\]
O Axioma de Compreens\~ao permite formar o conjunto
\[A = \{x\colon x\notin x\}.\]
Pergunta: $A$ \'e um elemento de si mesmo?
\'E claro que ambas poss\'\i veis respostas levam a uma contradi\c c\~ao.

A antinomia levou \`a necessidade de formular a Axioma de Compreens\~ao em uma
maneira mais exata. Al\'em disso, alguns outros axiomas foram adicionados. A primeira vers\~ao do 
axiomas que estava livre da antinomia de Russell e alguns outros paradoxos foi
proposta por Zermelo em 1908. Foram introduzidas novas melhorias por
Fraenkel em 1919 e Skolem em 1922. Observe que Cantor j\'a morreu em 1918. 

Os nove axiomas atuais incluem os axiomas que vimos acima, 
incluindo uma vers\~ao corrigida do Axioma da Compreens\~ao
(que n\~ao discutiremos), bem como, por exemplo, o seguinte.

\paragraph{\bf Axioma da Regularidade} 
Cada conjunto n\~ao vazio $x$ cont\'em um elemento que, visto como um conjunto, \'e disjunto de $x$:
\[\forall x,~ (\neg(x=\emptyset)\rightarrow \exists y~(y\in x \wedge
y\cap x =\emptyset).\]

Pode-se mostrar que o Axioma da Regularidade \'e equivalente \`a inexist\^encia de cadeias infinitas da forma
\[x_1\ni x_2\ni x_2\ni\ldots\ni x_n\ni\ldots\]
Em particular, nenhum conjunto pode ser o seu pr\'oprio elemento:
\[\forall x,~~x\notin x.\]

\paragraph{\bf Axioma do Infinito} 
Existe um conjunto infinito. Eis uma forma econ\^omica para exprimir isso:
\[\exists x~~(\emptyset\in x\wedge\forall y~(y\in x\rightarrow
\{y\}\in x).\]
Em outras palavras, existe um conjunto $X$ tendo a propriedade:
\[\emptyset\in X,\{\emptyset\}\in X,\{\{\emptyset\}\}\in X,\ldots\]

\subsection{Axioma da Escolha}

Axioma da escolha (AC) \'e, provavelmente, o mais controverso de todos os axiomas do sistema ZFC, e definitivamente o mais conhecido de todos: mesmo se um matem\'atico for duramente pressionado para declarar exatamente os 8 axiomas restantes, ningu\'em vai ter qualquer problema afirmando o axioma da escolha.

AC afirma uma coisa muito simples: dada uma fam\'\i lia de conjuntos n\~ao-vazios e dois a dois disjuntos, $A_i$, indexada com um conjunto de \'\i ndices, $I$, pode-se formar um novo conjunto, $B$, que intercepta cada conjunto em nossa fam\'\i lia em exatamente um ponto.

\paragraph{\bf Axioma da Escolha}
{\em
Seja $A_i$, $i\in I$ uma fam\'\i lia de conjuntos n\~ao vazios, tal que $A_i\cap A_j =\emptyset$ quando $i\neq j$. Existe um conjunto $B$ tal que para cada $i\in I$, a interse\c c\~ao $B\cap A_i$ \'e um conjunto unit\'ario.}

\label{page:axiomadaescolha}
\paragraph{\textbf{Axioma da Escolha}} (Forma B)
{\em 
Seja $\gamma$ uma fam\'\i lia qualquer dos conjuntos n\~ao vazios. Ent\~ao existe existe uma fun\c c\~ao $f\colon\gamma\to\cup\gamma$ (uma {\em fun\c c\~ao de escolha}) tal que, para todos $A\in\gamma$,}
\[f(A)\in A.\]
\index{Axioma da Escolha (AC)}

\begin{exercicio}
Verifique que as duas formas do Axioma da Escolha s\~ao equivalentes.
\end{exercicio}

\begin{exercicio}
Verifique o Axioma da Escolha no caso onde $\gamma$ \'e enumer\'avel, usando a indu\c c\~ao matem\'atica.
\end{exercicio}

O axioma da escolha parece t\~ao natural que ele estava sendo usado implicitamente no s\'eculo XIX, sem ser declarado explicitamente. No entanto, para conjuntos infinitos quaisquer AC n\~ao, de fato, siga do resto dos axiomas da ZF!
 
A primeira formula\c c\~ao de AC possivelmente pertence a Peano. Em 1904 o Axioma da Escolha foi ``legitimada'' por Zermelo, que provou o princ\'\i pio da boa ordem (teorema \ref{t:well-ordering}) usando Axioma da Escolha. 

Assim, o sistema axiom\'atico ZFC, em certo sentido,
serve como base da matem\'atica moderna. Qu\~ao s\'olido \'e essa base? 
Por exemplo, como podemos acertar que o sistema ZFC n\~ao leve a contradi\c c\~oes, alguns resultados tipo $0 = 1$? Em outras palavras, o
sistema ZFC \'e {\em consistente}?

Paradoxalmente, n\~ao s\'o n\~ao sabemos a resposta a esta pergunta, mas
mesmo assumindo que ZFC \'e consistente, nunca seremos capazes de provar este fato! \'E uma consequ\^encia do famoso {\em teorema de Incompletude,} uma
das contribui\c c\~oes fundamentais para a matem\'atica do s\'eculo XX feitas
por Kurt G\"odel. 
O teorema da incompletude diz que, dado um sistema axiom\'atico, a afirma\c c\~ao ``este sistema \'e consistente'' (isto \'e, n\~ao leva a
contradi\c c\~oes) n\~ao pode ser provada {\em dentro do pr\'oprio sistema.}

E pode o axioma da escolha causar problemas por conta pr\'opria? Surpreendentemente --- e este foi outra grande contribui\c c\~ao de G\"odel --- apesar de sua natureza controversa, a AC n\~ao acrescenta problemas adicionais aos fundamentos da matem\'atica. \'E {\em relativamente consistente:}

\paragraph{\bf Teorema de G\"odel da Consist\^encia Relativa da AC} (1935)
Se ZF \'e consistente, ent\~ao ZFC \'e consistente.

Em outras palavras, se n\~ao houver contradi\c c\~ao em ZF, ent\~ao ZF mais o Axioma da Escolha tamb\'em levar\'a a nenhuma contradi\c c\~ao. Se houver problemas com fundamentos da matem\'atica, ent\~ao \'e um problema com ZF, e o Axioma da Escolha n\~ao tem nada a ver com isso.

\subsection{Hip\'otese do Cont\'\i nuo (CH)\label{s:CH}}

Aqui est\'a a famosa conjectura de Georg Cantor, um dos problemas mais famosos e influentes em toda a hist\'oria da matem\'atica:

\paragraph{\bf Hip\'otese do Cont\'\i nuo (Continuum Hypothesis, CH)} (Georg Cantor, 1878). 
{\em N\~ao existem n\'umeros cardinais estritamente entre $\aleph_0$ e $\mathfrak c$. A saber, se $A$ \'e um subconjunto infinito da reta, $\R$, ent\~ao ou $\abs A = \aleph_0$ ou $\abs A = {\mathfrak c}$.}
\index{Hip\'otese do Cont\'\i nuo (CH)}

David Hilbert considerou o problema da validade do CH para ser de tal grande import\^ancia que ele incluiu-o como problema $1$ em sua lista de problemas matem\'aticos em seu discurso ao segundo Congresso Internacional de
Matem\'aticos (ICM) em Paris em 1900.

Em 1937 G\"odel provou talvez o seu maior resultado. 

\paragraph{\bf Consist\^encia da Hip\'otese do Cont\'\i nuo.} (Kurt G\"odel) {\em
A Hip\'otese do Cont\'\i nuo \'e consistente com ZFC. Em outras palavras, se ZF \'e consistente, ent\~ao a lista dos axiomas de ZFC mais CH \'e consistente 
(n\~ao leva a contradi\c c\~oes).}

Este resultado excluiu a possibilidade de que a Hip\'otese do Cont\'\i nuo pudesse ser refutada. \'E geralmente considerado como ``a primeira metade da 
solu\c c\~ao para a Hip\'otese do Cont\'\i nuo''.

A fim de ver a ``segunda metade'', os matem\'aticos tiveram que esperar at\'e 1963.

\paragraph{\bf Independ\^encia da Hip\'otese do Cont\'\i nuo.} (Paul Cohen)
{\em 
A Hip\'otese do Cont\'\i nuo n\~ao pode ser provada a partir de axiomas ZFC.
Em outras palavras, supondo que ZFC \'e consistente, se um acrescenta \`a lista de axiomas de ZFC a nega\c c\~ao da
Hip\'otese do Cont\'\i nuo, ent\~ao a lista resultante de axiomas,
 \[ZFC + \neg (CH), \]
tamb\'em \'e consistente.} 

Assim, a Hip\'otese do Cont\'\i nuo n\~ao pode ser provada nem
refutada no sistema axiom\'atico ZFC. \'E {\em independente}
de ZFC.

\section{Lema de Zorn}

\subsection{Formula\c c\~ao}

\begin{definicao}
Um elemento $x$ de um conjunto parcialmente ordenado $X$ \'e dito {\em maximal} se n\~ao existem nenhumas elementos estritamente maiores do que $x$:
\[\forall y\in X,~~x\leq y\Rightarrow x=y.\]
\end{definicao}

\begin{exemplo}
O elemento $1$ \'e maximal no segmento $[0,1]$ munido da ordem usual.
\end{exemplo}

\begin{exemplo}
\label{ex:b}
Seja $V$ um espa\c co vetorial sobre um corpo $\K$ qualquer. Fazemos $\mathfrak B$ a fam\'\i lia dos todos os subconjuntos linearmente independentes de $V$:
\begin{eqnarray*}
X\in{\mathscr B}&\iff& X\subseteq V\mbox{ e }\forall n\in\N_+,~\forall \lambda_i\in\R,~\forall x_i\in X,\\
&& ~i=1,2,\ldots,n,~~x_i\neq x_j\mbox{ para }i\neq j,\\
&& \mbox{ se }\sum_{i=1}^n\lambda_ix_i=0\mbox{, ent\~ao }\lambda_1=\lambda_2=\ldots=\lambda_n=0.
\end{eqnarray*}
O conjunto $\mathfrak B$, munido da rela\c c\~ao da rela\c c\~ao de inclus\~ao, \'e parcialmente ordenado. Um elemento $X\in {\mathfrak X}$ \'e maximal se e somente se $X$ \'e uma base (isso \'e, $X$ n\~ao \'e contido em nenhum conjunto linearmente independente estritamente maior).
\end{exemplo}

\begin{observacao}
A no\c c\~ao de um elemento maximal \'e diferente desta de um elemento m\'aximo. No exemplo precedente, n\~ao existe nenhuma elemento m\'aximo.
\end{observacao}

\begin{definicao}
Um conjunto parcialmente ordenado e n\~ao vazio $\mathfrak X$ \'e chamado {\em indutivo} se ele possui a propriedade seguinte. Seja ${\mathfrak C}\subseteq {\mathfrak X}$ um subconjunto de $\mathfrak X$ tal que a restri\c c\~ao da ordem sobre ${\mathfrak C}$ \'e uma ordem total. Ent\~ao ${\mathfrak C}$ \'e majorado em $\mathfrak X$:
\[\exists x\in X~~\forall c\in {\mathfrak C},~~c\leq x.\]
\index{conjunto! indutivo}
\end{definicao}

O resultado seguinte encontra-se entre aos instrumentos mais potentes e frequentemente utilizados da matem\'atica moderna. A demonstra\c c\~ao \'e deduzida do Axioma de Escolha (e \'e, de fato, uma forma equivalente dele). 

\begin{teorema}[Lema de Zorn] Seja $\mathfrak X$ conjunto n\~ao vazio parcialmente ordenado indutivo. Ent\~ao $\mathfrak X$ cont\'em pelo menos um elemento maximal. \qed
\label{l:zorn}
\index{lema! de Zorn}
\end{teorema}

\subsection{Dedu\c c\~ao do lema de Zorn do Axioma da Escolha}

A nossa prova \'e uma das mais curtas em exist\^encia, seguindo \citep*{weston}.

Seja $\mathscr X$ um conjunto parcialmente ordenado n\~ao vazio e indutivo. Mostremos a exist\^encia de um elemento maximal em $\mathscr X$.

\subsubsection{}
Para cada cadeia $\mathscr C\subseteq \mathscr X$, denotemos por $\widehat{\mathcal C}$ o conjunto de todos os majorantes de ${\mathcal C}$ dentro ${\mathscr X}$ que n\~ao pertencem a ${\mathcal C}$:
\[\widehat{\mathcal C} = \{x\in{\mathscr X}\setminus{\mathcal C}\colon \forall y\in{\mathcal C}~~y\leq x\}.\]
Por exemplo, $\widehat\emptyset={\mathscr X}$ (onde $\emptyset$ \'e uma cadeia vazia). Com certeza, n\~ao pode se excluir a possibilidade do que $\widehat{\mathcal C}$ seja vazio para algumas cadeias $\mathscr C$.

\subsubsection{}
Denotemos
\[\gamma = \{\widehat{\mathcal C} \colon {\mathcal C}\mbox{ \'e uma cadeia dentro }{\mathscr X}\mbox{ t.q. }\widehat{\mathcal C}\neq\emptyset\}.\]
Segundo AC, existe uma fun\c c\~ao de escolha
\[f\colon\gamma\to\bigcup\gamma,\]
tal que
\[\forall \widehat{\mathcal C}\in\gamma,~~f(\widehat{\mathcal C})\in\widehat{\mathcal C}.\]

\subsubsection{}
Notemos que, se ${\mathcal C}$ \'e uma cadeia e ${\mathscr K}\subseteq{\mathcal C}$, ent\~ao $\mathscr K$ \'e uma cadeia tamb\'em.
Digamos que uma cadeia ${\mathcal C}$ \'e uma {\em $f$-cadeia} se ela possui a propriedade seguinte. Qualquer seja ${\mathscr K}\subseteq {\mathcal C}$, se $\widehat{\mathscr K}\cap{\mathcal C}\neq\emptyset$, ent\~ao $f(\widehat{\mathscr K})\in {\mathcal C}$ e $f(\widehat{\mathscr K})$ \'e o menor majorante de ${\mathscr K}$ dentro ${\mathcal C}\setminus{\mathscr K}$.

Por exemplo, pois $\widehat\emptyset\cap{\mathcal C}={\mathcal C}$, toda $f$-cadeia cont\'em o elemento $f(\widehat \emptyset)$, que \'e o menor elemento de ${\mathcal C}$. Existe sempre pelo menos uma $f$-cadeia n\~ao-vazia: o conjunto unit\'ario $\{f(\widehat \emptyset)\}$.

\subsubsection{\label{cup}}
Se ${\mathcal C}$ \'e uma $f$-cadeia tal que $\widehat{\mathcal C}\neq\emptyset$, ent\~ao ${\mathcal C}\cup\{f(\widehat{\mathcal C})\}$ \'e uma $f$-cadeia. 

$\triangleleft$
Denotemos $x=f(\widehat{\mathcal C})$.
Seja ${\mathscr K}\subseteq{\mathcal C}$ um subconjunto tal que $\widehat{\mathscr K}\cap({\mathcal C}\cup\{x\})\neq\emptyset$. H\'a dois casos a considerar. 
Se ${\mathscr K}\cap{\mathcal C}\neq\emptyset$, ent\~ao, como ${\mathcal C}$ \'e uma $f$-cadeia, temos $f(\widehat{\mathscr K})\in {\mathcal C}\subseteq {\mathcal C}\cup\{x\}$ e $f(\widehat{\mathscr K})$ \'e o menor majorante de ${\mathscr K}$ dentro ${\mathcal C}\setminus{\mathscr K}$, logo dentro ${\mathcal C}\cup\{x\}$ tamb\'em. Se ${\mathscr K}\cap{\mathcal C} =\emptyset$, ent\~ao $\widehat{\mathscr K}\cap({\mathcal C}\cup\{x\})=\{x\}$, de onde  conclu\'\i mos: $\widehat{\mathscr K}=\widehat{\mathcal C}$. Por conseguinte, $f(\widehat{\mathscr K})=f(\widehat{\mathcal C})=x$, e a condi\c c\~ao de ser uma $f$-cadeia \'e satisfeita.
\hfill
$\triangleright$

\subsubsection{\label{soit}}
Sejam ${\mathcal C}$ e ${\mathcal C}^{\prime}$ duas cadeias quaisquer. Ent\~ao, ou ${\mathcal C}^{\prime}\subseteq{\mathcal C}$, ou ${\mathcal C}\subseteq{\mathcal C}^{\prime}$. No primeiro caso, ${\mathcal C}^{\prime}$ \'e um segmento inicial de ${\mathcal C}$, ou seja, se $y,z\in{\mathcal C}$, $z\leq y$, e $y\in{\mathcal C}^{\prime}$, ent\~ao $z\in{\mathcal C}^{\prime}$.

$\triangleleft$ Suponha que ${\mathcal C}\setminus{\mathcal C}^{\prime}\neq\emptyset$, e seja $x\in{\mathcal C}\setminus{\mathcal C}^{\prime}$ qualquer. Definamos
\[{\mathscr K}=\{y\in{\mathcal C}\cap{\mathcal C}^{\prime}\colon y\leq x\}\equiv \{y\in{\mathcal C}\cap{\mathcal C}^{\prime}\colon y < x\}.\]
Como $x\in\widehat{\mathscr K}\cap{\mathcal C}$, temos ${\mathscr K}\cap{\mathcal C}\neq\emptyset$ e $f(\hat{\mathscr K})\in{\mathcal C}$ e por conseguinte $f(\widehat{\mathscr K})\leq x$. Como $\widehat K{\mathscr K}\cap{\mathscr K}=\emptyset$, temos $f(\widehat{\mathscr K})\notin{\mathscr K}$, logo
$f({\mathscr K})\notin{\mathcal C}^{\prime}$. Isso implica que necessariamente $\widehat{\mathscr K}\cap{\mathcal C}^{\prime}=\emptyset$. Como nenhum elemento de ${\mathcal C}^{\prime}$ majora ${\mathscr K}$, e ${\mathcal C}^{\prime}$ \'e uma cadeia, conclu\'\i mos que ${\mathscr K}$ \'e {\em cofinal} dentro ${\mathcal C}^{\prime}$, a saber,
\[\forall y\in{\mathcal C}^{\prime}~~\exists z\in{\mathscr K}~~ y\leq z,\]
et por conseguinte,
\[\forall y\in{\mathcal C}^{\prime}~~y\leq x.\]
Conclu\'\i mos: ${\mathcal C}^{\prime}\subseteq{\mathcal C}$. Al\'em disso, o argumento implica que $y\in{\mathcal C}^{\prime}$ e $z\leq y$, ent\~ao $z$ n\~ao pode pertencer a ${\mathcal C}\setminus{\mathcal C}^{\prime}$. Por conseguinte, ${\mathcal C}^{\prime}$ \'e um segmento inicial de ${\mathcal C}$. 
\hfill
$\triangleright$

\subsubsection{\label{tot}}
Em particular, a fam\'\i lia de todas as $f$-cadeias \'e totalmente ordenada: em outras palavras, todas as $f$-cadeias s\~ao compar\'aveis duas a duas. Logo, a uni\~ao
\[{\mathcal C}_0 =\bigcup\{{\mathcal C}\colon {\mathcal C}\mbox{ est une $f$-cadeia}\}\]
\'e uma cadeia.

\subsubsection{\label{seg}}
Cada $f$-cadeia ${\mathcal C}$ \'e uma segmento inicial da cadeia ${\mathcal C}_0$.

$\triangleleft$
Sejam $x\in{\mathcal C}$ e $y\in{\mathcal C}_0$ tais que $y\leq x$. Existe uma $f$-cadeia ${\mathcal C}^{\prime}$ tal que $y\in{\mathcal C}^{\prime}$. Segundo \ref{soit}, uma das $f$-cadeias ${\mathcal C}$ e ${\mathcal C}^{\prime}$ cont\'em a outra. Se ${\mathcal C}^{\prime}\subseteq{\mathcal C}$, ent\~ao $y\in{\mathcal C}^{\prime}\subseteq {\mathcal C}$ e n\~ao tem nada a mostrar. Se ${\mathcal C}\subseteq{\mathcal C}^{\prime}$, ent\~ao, mais uma vez segundo \ref{soit}, ${\mathcal C}$ forma um segmento inicial de ${\mathcal C}^{\prime}$, de onde conclu\'\i mos: $y\in{\mathcal C}$.
\hfill
$\triangleright$

\subsubsection{}
A cadeia ${\mathcal C}_0$ \'e uma $f$-cadeia. 

$\triangleleft$
Seja ${\mathscr K}\subseteq{\mathcal C}_0$ uma cadeia tal que $\widehat{\mathscr K}\cap{\mathcal C}_0\neq\emptyset$. Ent\~ao, existe uma $f$-cadeia ${\mathcal C}$ que tem uma interse\c c\~ao n\~ao-vazia com $\widehat{\mathscr K}$. Deduzimos duas coisas: primeiramente, ${\mathscr K}\subseteq{\mathcal C}$ (gra\c cas a \ref{seg}), e al\'em disso, 
$\widehat{\mathscr K}\cap{\mathcal C}\neq\emptyset$. Pois ${\mathcal C}$ \'e uma $f$-cadeia, temos
$f(\widehat{\mathscr K})\in{\mathcal C}\subseteq{\mathcal C}_0$, e $f(\widehat{\mathscr K})$ \'e o elemento menor de $\widehat{\mathscr K}\cap{\mathcal C}$. Como ${\mathcal C}$ \'e um segmento inicial de $Ch_0$, conclu\'\i mos que $\widehat{\mathscr K}\cap Ch$ \'e um segmento inicial de $\widehat{\mathscr K}\cap{\mathcal C}_0$, e por conseguinte
$f(\widehat{\mathscr K})$ \'e o menor elemento de $\widehat{\mathscr K}\cap{\mathcal C}$.
\hfill
$\triangleright$

\subsubsection{\label{empty}} $\widehat{{\mathcal C}_0}=\emptyset$.

$\triangleleft$
Redu\c c\~ao para o absurdo. Suponha que $\widehat{{\mathcal C}_0}\neq\emptyset$, e notemos $x=f(\widehat{{\mathcal C}_0})$. Segundo \ref{cup}, ${\mathcal C}_0\cup\{x\}$ \'e uma $f$-cadeia. Pois ${\mathcal C}$ \'e a uni\~ao de todas as $f$-cadeias, temos ${\mathcal C}\cup\{x\}\subseteq{\mathcal C}$, ou seja, $x\in{\mathcal C}$. Ao mesmo tempo, $x\in\widehat{{\mathcal C}_0}$, qual conjunto \'e disjunto de ${\mathcal C}_0$.
\hfill
$\triangleright$

\subsubsection{} Porque ${\mathscr X}$ \'e indutivo, existe um majorante, $x$, para a cadeia ${\mathcal C}_0$. De acordo com \ref{empty}, $x\in{\mathcal C}_0$. Cada $y$ tal que $y>x$ seria um majorante para ${\mathcal C}_0$ que n\~ao pertence a ${\mathcal C}_0$, o que \'e imposs\'\i vel segundo \ref{empty}. Isso significa que $x$ \'e um elemento maximal de ${\mathscr X}$. \qed

\begin{exercicio}
Deduza o Axioma da Escolha do lema de Zorn.
\par
[ {\em Sugest\~ao:} considere o conjunto parcialmente ordenado ${\mathscr X}$  que consiste de todas as fun\c c\~oes de escolha parcialmente definidas sobre  $\gamma$. ]
\end{exercicio}

\subsection{Ilustra\c c\~ao: exist\^encia de uma base num espa\c co vetorial}

\begin{exemplo}
\label{e:indutivo}
O conjunto $\mathfrak B$ dos subespa\c cos vetoriais do exemplo \ref{ex:b} \'e indutivo. 

Eis a verifica\c c\~ao. Seja ${\mathfrak C}$ uma fam\'\i lia dos elementos de  $\mathfrak B$ totalmente ordenada pela rela\c c\~ao de inclus\~ao:
\[\forall A,B\in {\mathfrak C},~~A\subseteq B\mbox{ ou }B\subseteq A.\]
Portanto, $A\cup B$ pertence a ${\mathfrak C}$ para todos $A,B\in {\mathfrak C}$. Pela indu\c c\~ao finita, qualquer sejam $n\in\N$ e $A_1,A_2,\ldots,A_n\in {\mathfrak C}$, temos $\cup_{i=1}^n A_i\in {\mathfrak C}$. 

Fazemos $X=\cup {\mathfrak C}=\cup\{A\colon A\in {\mathfrak C}\}$. Vamos mostrar que $X\in {\mathfrak B}$, isso \'e, $X$ \'e linearmente independente. Sejam  $n\in\N$, $\lambda_i\in\K$, $x_i\in X$, $i=1,2,\ldots,n$, tais que $x_i\neq x_j$ para $i\neq j$. Suponhamos que $\sum_{i=1}^n\lambda_ix_i=0$. Existe $A_i\in {\mathfrak C}$ tais que $x_i\in A_i$, $i=1,2,\ldots,n$. O conjunto $A=\cup_{i=1}^n A_i$ pertencem a ${\mathfrak C}$, e por conseguinte $A$ \'e linearmente independente. Deduzimos: $\lambda_i=0$ para todos $i=1,2,\ldots,n$. 
\end{exemplo}

Eis uma aplica\c c\~ao do lema de Zorn.

\begin{teorema}
Cada espa\c co vetorial possui uma base.
\end{teorema}

\begin{proof}
Seja $V$ um espa\c co vetorial qualquer. A fam\'\i lia $\mathfrak B$ definida como o exemplo \ref{ex:b} \'e indutiva (exemplo \ref{e:indutivo}). De acordo com o lema de Zorn, ele possui pelo menos um elemento maximal, $X$. Esto $X$ \'e uma base de $V$.
\end{proof}

\begin{exercicio}
Mostrar teorema \ref{th:comparabilidade}. ({\em Sugest\~ao:} aplique o lema de Zorn \`a fam\'\i lia de todos as inje\c c\~oes parciais de $X$ para $Y$, bem como de $Y$ para $X$.)
\label{ex:comparabilidade}
\end{exercicio}

\subsection{Princ\'\i pio da boa ordem\label{ss:boaordem}}
Mesmo se $\Z$ n\~ao \'e bem ordenado com sua ordem natural (usual), podemos
d\^e-lhe uma boa ordem.
Para obter essa ordem, $\prec$, organize os inteiros da seguinte maneira: 
primeiro, coloque todos os n\'umeros naturais
em sua ordem habitual, e depois disso, colocar todos os inteiros negativos em um
ordem inversa:
\[0\prec1\prec2\prec3\prec\ldots\prec10^{10}\prec\ldots 
\prec-1\prec-2\prec-3\prec\ldots\prec-10^{10}\prec\ldots\]

\begin{exercicio}
Mostre que a ordem $\prec$ como acima \'e uma boa ordem, isto \'e, $(\Z,\prec)$ \'e bem-ordenado.
\end{exercicio}

Acontece que se pode construir uma boa ordem sobre qualquer conjunto, mesmo tendo uma propriedade mais forte.

\begin{definicao}
Seja $(X,\prec)$ um conjunto totalmente ordenado. Um {\em segmento inicial} de $X$ \'e qualquer subconjunto $Y$ munido da ordem induzida de $X$ e tal que, se $y\in Y$ e $z\prec y$, ent\~ao $z\in Y$.
\end{definicao}

\begin{definicao}
Uma boa ordem $\prec$ sobre um conjunto $X$ chama-se {\em minimal} se cada pr\'oprio segmento inicial de $(X,\prec)$ tem a cardinalidade menor do que $X$.
\end{definicao}

\begin{teorema}[Princ\'\i pio da Boa Ordem; Zermelo, 1904]
Dado um conjunto qualquer, $X$, existe uma boa ordem minimal $\prec$ sobre $X$. 
\qed
\label{t:well-ordering}
\index{princ\'\i pio! da boa ordem}
\end{teorema}

\begin{exercicio}
Seja $X$ um conjunto bem-ordenado. Mostrar que n\~ao existe nenhuma bije\c c\~ao que preserva ordem entre $X$ e um pr\'oprio segmento inicial de $X$.
\end{exercicio}

\begin{exercicio}
Mostre a exist\^encia de uma boa ordem sobre qualquer conjunto usando o Lema de Zorn. 
\par
[ {\em Sugest\~ao:} considere a fam\'\i lia de todos os boas ordens (minimais) sobre subconjuntos de $X$, definindo uma ordem parcial entre eles de uma maneira cuidada, para usar o exerc\'\i cio acima. ]
\end{exercicio}

\begin{exercicio}
Deduza que todo conjunto admite uma boa ordem minimal.
\end{exercicio}

\subsection{Ultrafiltros}

\begin{definicao}
Um {\em filtro} sobre um conjunto $X$ qualquer \'e uma fam\'\i lia $\mathcal F$ (n\~ao vazia) dos subconjuntos n\~ao vazios de $X$ tal que
\begin{enumerate}
\item se $A,B\in {\mathcal F}$, ent\~ao $A\cap B\in {\mathcal F}$, e
\item se $A\in {\mathcal F}$ e $A\subseteq B\subseteq X$, ent\~ao $B\in {\mathcal F}$.
\end{enumerate}
\end{definicao}

\begin{exemplo}
Sejam $X$ um espa\c co m\'etrico (ou, mais geralmente, espa\c co topol\'ogico), $x$ um ponto de $X$. A fam\'\i lia ${\mathcal N}(x)$ de todas as vizinhan\c cas de $x$ em $X$ \'e um filtro.
\end{exemplo}

\begin{exemplo}
O {\em filtro de Fr\'echet} sobre um conjunto infinito $X$ qualquer consiste de todos os subconjuntos de $X$ cofinitos. 
\end{exemplo}

\begin{definicao}
Seja $X$ um conjunto n\~ao vazio qualquer. Uma fam\'\i lia $\mathcal F$ de subconjuntos de $X$ \'e chamada {\em centrada} se a interse\c c\~ao de cada subfam\'\i lia finita de $\mathcal F$ \'e n\~ao vazia:
\[\forall n,~~\forall F_1,F_2,\ldots,F_n\in{\mathcal F},~~\bigcap_{i=1}^nF_i\neq\emptyset.\]
\index{fam\'\i lia! centrada}
\end{definicao}

\begin{exemplo}
Cada filtro \'e uma fam\'\i lia centrada.
\end{exemplo}

\'A f\'acil verificar que a cole\c c\~ao das todas as fam\'\i lias centradas sobre um conjunto n\~ao vazio $X$, ordenada pela inclus\~ao, \'e indutiva.
Como corol\'ario de lema de Zorn, obtemos:

\begin{proposicao}
Cada sistema centrado sobre um conjunto $X$ qualquer \'e contido num sistema centrado maximal.
\label{p:centradomaximal}
\end{proposicao}

\begin{definicao}
Uma sistema centrado maximal \'e dito {\em ultrafiltro}.
\index{ultrafiltro}
\end{definicao}

Eis um crit\'erio maior.

\begin{teorema}
\label{t:AAc}
Seja $\mathcal F$ um sistema centrado sobre um conjunto n\~ao vazio $X$. Ent\~ao $\mathcal F$ \'e um ultrafiltro se e somente se, qualquer seja $A\subseteq X$, temos $A\in {\mathcal F}$ ou $A^c\in {\mathcal F}$.
\end{teorema}

\begin{proof}
$\Rightarrow$: suponhamos que $\mathcal F$ \'e um ultrafiltro, e seja $A\subseteq X$ um subconjunto qualquer. Se $A$ encontra a interse\c c\~ao de cada subcole\c c\~ao finita de $\mathcal F$, ent\~ao o sistema ${\mathcal F}\cup \{A\}$ \'e centrada, e como $\mathcal F$ \'e maximal, temos $A\in {\mathcal F}$. De mesmo modo, se $A^c$ encontra a interse\c c\~ao de cada subcole\c c\~ao finita de $\mathcal F$, temos $A^c\in {\mathcal F}$. O caso onde nem $A$, nem $A^c$ encontram a interse\c c\~ao de cada subcole\c c\~ao finita de $\mathcal F$ \'e imposs\'\i vel: suponha a exist\^encia dos $A_1,A_2\ldots,A_n,A_{n+1},\ldots,A_{n+m}$ tais que
\[A\cap A_1\cap \ldots \cap A_n=\emptyset,~~
A^c\cap A_{n+1},\ldots,A_{n+m} =\emptyset,\]
logo
\[\bigcap_{i=1}^n A_i\subseteq A^c,~~\bigcap_{i=n+1}^{n+m} A_i\subseteq A,\]
e finalmente
\[\bigcap_{i=1}^{n+m}A_i\subseteq A\cap A^c=\emptyset,\]
a contradi\c c\~ao com a hip\'otese que $\mathcal F$ \'e centrada.

$\Leftarrow$: Seja $\mathcal F$ uma sistema centrada com a propriedade que para cada $A\subseteq X$, ou $A\in{\mathcal F}$, ou $A^c\in {\mathcal F}$. Seja $\mathcal F^\prime$ uma sistema centrada tal que ${\mathcal F}\subseteq {\mathcal F}^\prime$. Se a inclus\~ao \'e pr\'opria, obtemos uma contradi\c c\~ao imediata: existe $A\in {\mathcal F}^\prime\setminus {\mathcal F}$, e como $A\notin {\mathcal F}$, temos $A^c\in {\mathcal F}$. Os conjuntos $A$ e $A^c$ ambos pertencem ao sistema centrada ${\mathcal F}^\prime$, que \'e imposs\'\i vel.
\end{proof}

\begin{observacao}
Evidentemente, ``ou'' no teorema \ref{t:AAc} \'e exclusivo.
\end{observacao}

\begin{corolario}
Seja ${\mathcal U}$ um ultrafiltro sobre um conjunto $X$. Sejam $A,B\subseteq X$ tais que $A\cup B=X$. Ent\~ao ou $A\in{\mathcal U}$, ou $B\in{\mathcal U}$.
\end{corolario}

\begin{proof}
Suponha que $A\notin{\mathcal U}$ e $B\notin {\mathcal U}$. Segundo o teorema \ref{t:AAc}, temos
$A^c\in{\mathcal U}$ e $B^c\in{\mathcal U}$, e por conseguinte
\[\emptyset=A^c\cap B^c\in{\mathcal U},\]
uma contradi\c c\~ao.
\end{proof}

Pela indu\c c\~ao matem\'atica finita \'obvia, obtemos:

\begin{corolario}
\label{c:gammaxi}
Sejam $\gamma$ uma cobertura finita de um conjunto $X$ e ${\mathcal U}$ um ultrafiltro sobre $X$. Ent\~ao existe $A\in\gamma$ tal que $A\in{\mathcal U}$. Em outras palavras,
\[\gamma\cap{\mathcal U}\neq\emptyset.\]
\end{corolario}

\begin{corolario}
Cada ultrafiltro \'e um filtro.
\label{c:cueuf}
\end{corolario}

\begin{proof}
Seja ${\mathcal U}$ um ultrafiltro sobre um conjunto $X$. Sejam $A,B\in{\mathcal U}$. Se $A\cap B\notin{\mathcal U}$, ent\~ao 
\[A^c\cup B^c=(A\cap B)^c\in {\mathcal U},\]
e obtemos uma contradi\c c\~ao com o fato que ${\mathcal U}$ \'e uma fam\'\i lia centrada, porque
\[A\cap B\cap (A^c\cup B^c)=\emptyset.\]
Conclu\'\i mos: $A\cap B\in{\mathcal U}$.

De maneira semelhante, sejam $A\in{\mathcal U}$ e $A\subseteq B\subseteq X$. Suponhamos que $B\notin {\mathcal U}$. Logo $B^c\in{\mathcal U}$, e $A\cap B^c=\emptyset$, a contradi\c c\~ao.
\end{proof}

%
%

\chapter{Espa\c cos m\'etricos. Teorema da categoria de Baire\label{ch:metricos}}

\section{N\'umeros reais\label{a:R}}

\subsection{Corpos ordenados}

\begin{definicao}
Seja $F$ um conjunto munido de duas opera\c c\~oes bin\'arias, $+$ e $\cdot$, assim como dois elementos distinguidos, $0$ e $1$.  O sistema $\langle F, +, \cdot, 0,1\rangle$ \'e dito um {\em corpo} se ele satisfaz os axiomas seguintes.

\begin{enumerate}
\item[(F1)] $\forall a,b,c\in F, ~a+(b+c) = (a+b)+c$ (associatividade de adi\c c\~ao).
\item[(F2)] $\forall a,b\in F,~a+b=b+a$ (comutatividade de adi\c c\~ao).
\smallskip\item[(F3)] $\forall a\in F,~ a+0=a$.
\smallskip\item[(F4)] $\forall a\in F,~\exists -a~: a+(-a)=0$.
\smallskip\item[(F5)] $\forall a,b,c\in F,~a(bc)=(ab)c$ (associatividade de multiplica\c c\~ao)
\smallskip\item[(F6)] $\forall a,b\in F,~ab=ba$ (comutatividade de multiplica\c c\~ao)
\smallskip\item[(F7)] $\forall a\in F,~a\cdot 1 = a$.
\smallskip\item[(F8)] $\forall a\neq 0,~\exists a^{-1} : a\cdot a^{-1}=1$.
\smallskip\item[(F9)] $\forall a,b,c\in F,~ a(b+c)=ab+ac$ (distributividade)  
\end{enumerate}
\end{definicao}

\begin{exemplo}
O sistema de n\'umeros reais $\R$ \'e um corpo.
\end{exemplo}

\begin{exemplo}
A totalidade de n\'umeros racionais, ou seja, os reais que podem ser representados como um r\'acio de inteiros, 
$p/q$, $p,q\in\Z$, $q\neq 0$, forma um corpo sobre as opera\c c\~oes usuais, o corpo de n\'umeros racionais, denotado $\Q$.
\end{exemplo}

\begin{exemplo} Para todo n\'umero primo $p$, pode-se formar um corpo, denotado $\Z_p$ e chamado o {\em corpo de res\'\i duos m\'odulo $p$}, ou simplesmente um {\em corpo finito com $p$ elementos.} Os elementos de $\Z_p$ s\~ao denotados  $0,1,2,\dots, p-1$, e a adi\c c\~ao e multiplica\c c\~ao s\~ao executadas m\'dulo $p$. 

Relembramos que dois inteiros, $m$ e $n$, s\~ao {\em iguais m\'odulo $p$} (nota\c c\~ao: $m\equiv n$ ($\mod p$)), se $m-n$ \'e um m\'ultiplo de $p$.
\end{exemplo}

\begin{exemplo} 
O corpo de n\'umeros complexos, $\C$, consiste de todos os polin\^omios em uma vari\'avel $\im$ da forma $z=a+b\im$, com coeficientes reais, e $\im$ \'e a {\em unidade imagin\'aria}, tendo a propriedade $\im^2=-1$. O coeficiente $a$ \'e dito a {\em parte real} de $z$, e $b$ \'e a {\em parte imagin\'aria} de $z$. 

Adi\c c\~ao e multiplica\c c\~ao dos n\'umeros complexos s\~ao executadas como as dos polin\^omios, e a express\~ao $\im^2$ est\'a depois substitu\'\i da por $-1$. Para executar a divis\~ao, definamos o {\em conjugado} de $z$,
$\bar z = a-b\im$. Temos:
\[z\bar z = (a+b\im)(a-b\im)=a^2+b^2=\abs{z}^2,\]
onde $\abs z$ \`a a norma euclidiana de $z$. Agora
\[\frac{z}{w} = \frac{z\bar w}{w\bar w} = \frac{1}{\abs{w}^2}z\bar w.\]
\end{exemplo}

\begin{exemplo}
Dado um corpo $F$ qualquer, formemos um novo corpo, denotado $F(t)$ \'e chamado uma {\em extens\~ao transcendental simples} de $F$, ou um {\em corpo de fun\c c\~oes racionais sobre} $F$. O corpo $F(t)$ consiste de todas as express\~oes da forma 
\[{p(t)\over q(t)}\equiv
{a_0+a_1t+a_2t^2+\dots+a_mt^m\over b_0+b_1t+b_2t^2+\dots+b_nt^n},\]
onde $p(t)$ e $q(t)$ s\~ao (relativamente primos) polin\^omios em uma vari\'avel $t$ com coeficientes de $F$. Em outras palavras, elementos de $F(t)$ s\~ao {\em fun\c c\~oes racionais} sobre $F$. As opera\c c\~oes s\~ao executadas de modo usual. 
\end{exemplo}

\begin{definicao}
Seja $F=\langle F,+,\cdot,0,1\rangle$ um corpo, e seja $\leq $ uma rela\c c\~ao de ordem total sobre $F$ (defini\c c\~ao \ref{d:ordemtotal}). O sistema  $F=\langle F,+,\cdot,0,1,\leq \rangle$ \'e chamado um {\em corpo ordenado} se ele satisfaz
\smallskip
\item[(OF1)] $\forall a,b,c\in F$, se $a\leq b$ ent\~ao $a+c\leq b+c$.
\smallskip
\item[(OF2)] $\forall a,b,c\in F$, se $a\leq b$ e $0\leq c$ ent\~ao $ac\leq bc$.
\end{definicao}

\begin{exercicio}
Deduza dos axiomas de um corpo ordenado que, se $a>0$ e $b>0$, ent\~ao $a+b>0$.
\end{exercicio} 

\begin{exemplo}
$\R$ \'e um corpo ordenado.
\end{exemplo}

\begin{exemplo} 
O corpo de racionais, $\Q$, \'e um {\em subcorpo ordenado} de $\R$.
\end{exemplo}

\begin{exercicio}
Mostre que o corpo de n\'umeros complexos, $\C$, n\~ao admite uma estrutura de corpo ordenado.
\end{exercicio}

\begin{observacao} 
Um corpo $F$ \'e {\em formalmente real} se existe uma ordem $\leq $ sobre $F$ tornando-o um corpo ordenado. O corpo $\C$ n\~ao \'e formalmente real, enquanto o corpo $\R$ \'e. Pode-se mostrar que um corpo $F$ \'e formalmente real se e somente se $-1$ n\~ao pode ser representado como uma soma finita de quadrados de elementos de $F$ (teorema de Artin-Schreier, resolvendo um dos problemas de Hilbert, o d\'ecimo s\'etimo).
\end{observacao}

\begin{exercicio}
Um corpo finito $F$ qualquer, inclusive $\Z_p$, n\~ao \'e formalmente real.
\end{exercicio}

\begin{exemplo} 
O corpo $F(t)$ de fun\c c\~oes racionais com coeficientes em um corpo formalmente real, $F$, \'e formalmente real. Mostremos uma tal constru\c c\~ao para o corpo $\Q(t)$. 

Definamos o sinal de um polin\^omio, 
\[p\equiv p(t)\equiv a_0+a_1t+a_2t^2+\dots+a_nt^n,\]
como o sinal de $p$ \'e igual ao sinal do primeiro coeficiente n\~ao nulo. Por exemplo, $t>0$, $-2+t+t^{11}<0$. O sinal de $x=p/q$ \'e equal ao produto dos sinais de $p$ e $q$. Por exemplo,
\[{3+t\over -11-t+t^2+t^{25}}<0,\]
pois $3+t>0$ e $-11-t+t^2+t^{25}<0$.

Finalmente, $x<y$ se e somente se $y-x>0$. 
\end{exemplo}

\begin{exercicio}
Mostrar que a ordem linear definido sobre o corpo $\Q(t)$ como acima satisfaz os axiomas de um corpo ordenado.
\end{exercicio}

Quando munido da ordem acima, o corpo $\Q(t)$ \'e as vezes denotado 
$\Q(\alpha)$, onde $\alpha=t$, mas este s\'\i mbolo enfatiza o fato de que o campo est\'a ordenado em um modo particular.
O elemento $\alpha$ \'e positivo ($a_0=0,a_1=1>0$). e ao mesmo tempo, para todo n\'umero racional estritamente positivo $\e>0$ temos 
\[\alpha<\e,\]
pois $\alpha-\e<0$, com $a_0=-\e<0$). De fato, $\abs\alpha$ \'e menor do que qualquer racional estritamente positivo. Tais elementos s\~ao chamados {\em infinitesimais.} 

O elemento $\alpha^{-1}$ \'e {\em infinitamente grande:} qualquer que seja um n\'umero racional $r$, temos $\alpha^{-1}>r$;
\[\alpha^{-1}-r={1\over \alpha}-{r\over 1}={1-r\alpha\over \alpha}>0.\]

\subsection{Axioma de Dedekind}

\begin{definicao}
Um conjunto $X$ totalmente ordenado satisfaz a {\em propriedade de completude de Dedekind,} ou \'e {\em completo no sentido de Dedekind,} se todo subconjunto de $X$ n\~ao vazio e limitado por cima tem um supremo.
\par
Em outras palavras, se $\emptyset\neq A\subseteq X$ e existe $x\in X$ tal que
\[\forall a\in A,~a\leq x,\]
ent\~ao existe $y=\sup A$: 
\[\forall a\in A,~a\leq y,\mbox{ e se } \forall a\in A,~a\leq z,\mbox{ ent\~ao }y\leq z.\]
\index{completude de Dedekind}
\end{definicao}

\begin{exemplo}
O corpo $\Q(\alpha)$ n\~ao \'e completo no sentido de Dedekind. O conjunto $\N$ de n\'umeros naturais \'e um subconjunto limitado por cima, por exemplo,
\[\forall n\in\N,~~n<\frac 1{\alpha}.\]
Suponha agora que existe o supremo de $\N$, $s=\sup\N$. Em particular, $s$ \'e uma cota superior para $\N$, a saber,
\[\forall n\in\N,~~n\leq s,\]
logo
\[\forall n\in\N,~~n+1\leq s,\]
ou, de modo equivalente,
\[\forall n\in\N,~~n\leq s-1 <s,\]
contradizendo a defini\c c\~ao do supremo.
\label{ex:qofalpha}
\end{exemplo}

\begin{exemplo}
O corpo $\Q$ n\~ao \'e completo no sentido de Dedekind. O conjunto $A$ de todos os elementos da forma
\[0,~0.1,~0.101,~0.101001,~0.1010010001,~0.101001000100001,~\dots\]
\'e limitado por cima, por\'em n\~ao tem o supremo em $\Q$. \'E f\'acil a verificar que se o supremo existisse, sua expans\~ao digital seria da forma
\[r=0.101001000100001\dots,\]
um n\'umero irracional.
\end{exemplo}

\begin{definicao}
Um {\it isomorfismo} entre dois corpos ordenados, $K$ e $F$, \'e uma bije\c c\~ao $f\colon K\to F$ tal que $f$ conserva adi\c c\~ao, ou seja,
para cada $x,y\in K$ temos $f(x+y)=f(x)+f(y)$, multiplica\c c\~ao, ou seja, para  $f(xy)=f(x)f(y)$, e a ordem, ou seja,
se $x<y$, ent\~ao $f(x)<f(y)$.
\end{definicao}
 
\begin{definicao}
Dois corpos ordenados, $F$ e $K$, s\~ao {\em isomorfos} se existe um isomorfismo
$f\colon F\to K$.
\end{definicao}

\begin{teorema} 
A menos de isomorfismo, existe um e apenas um corpo ordenado completo no sentido de Dedekind. Ele \'e chamado o corpo dos reais e denotado $\R$.
\label{t:existenciadeR}
\index{R@$\R$}
\end{teorema}

Idealmente, ter\'\i amos que come\c car com uma teoria de n\'umeros naturais, definidos pelos axiomas de Peano:

\begin{definicao}
O sistema de n\'umeros naturais, $\N$, \'e definido pelos {\em axiomas de Peano} seguintes.
\begin{enumerate}
\item Existe um elemento de $\N$ denotado $0$.
\item Cada elemento $n\in\N$, possui o {\em sucessor,} $Sn\in\N$.
\item Se $m\neq n$, ent\~ao $Sm\neq Sn$.
\item O elemento $0$ n\~ao \'e o sucessor de algum elemento.
\item ({\em Axioma de recorr\^encia}).
Seja $P$ um subconjunto de $\N$ tal que $0\in\N$ e para cada $n\in P$, $Sn\in P$. Ent\~ao $P=\N$.
\end{enumerate}
\end{definicao}

No entanto, s\'o vamos esbo\c car uma constru\c c\~ao de $\R$ a partir de n\'umeros racionais. 
Definamos
\[\Q_{>0} = \{r\in Q\colon r >0\},\]
o conjunto de n\'umeros racionais estritamente positivos. 

\begin{definicao}
\label{d:r+}
O conjunto $\R_{>0}$ consiste de todos os subconjuntos $A$ de $\Q_{>0}$ tais que
\begin{enumerate}
\item Quaisquer que sejam $a\in A$ e $x\in X$ com $x\leq a$, temos $x\in A$ ($A$ \'e um {\em intervalo initial} de $X$),
\item $\emptyset \neq A \neq X$ ($A$ \'e um subconjunto {\em pr\'oprio} de $X$).
\item $A$ n\~ao possui o m\'aximo ($A$ \'e um intervalo aberto).
\end{enumerate}
\end{definicao}

\begin{exercicio}
Verigique que $\R_{>0}$ \'e totalmente ordenado pela rela\c c\~ao de inclus\~ao:
\[A\leq A^\prime\iff A\subseteq A^\prime. \]
\end{exercicio}

\begin{exercicio}
\label{l:+ver}
Sejam $A,B\in\R_{>0}$. Mostre que a soma conjunt\'\i stica
\[A+B = \{a+b\colon a\in A,b\in B\}\]
pertence a $\R_{>0}$.
\end{exercicio}

\begin{exercicio}
Verifique que a a soma conjunt\'\i stica define uma opera\c c\~ao de adi\c c\~ao comutativa e associativa.
\end{exercicio}

\begin{exercicio}
Se $A\leq B$, ent\~ao $A+C \leq B+C$.
\end{exercicio}

\begin{exercicio}
Mostre que, se $A\leq B$, pode se definir a subtra\c c\~ao
\[B-A = \Q_{>0}\cap\{b-a\colon b\in B, a\in A^c\},\]
de modo que
\[(B-A)+A = B.\]
\end{exercicio}

\begin{definicao}
Definamos o produto conjunt\'\i stico dentro de $\R_{>0}$ por
\[AB = \{ab\colon a\in A, b\in B\}.\]
\end{definicao}

\begin{exercicio} 
Verifique que o produto $AB$ pertence a $\R_{>0}$ (satisfaz as propriedades na defini\c c\~ao \ref{d:r+}). Depois, verifique que a multiplica\c c\~ao \'e comutativa e associativa, bem como distributiva em rela\c c\~ao \`a adi\c c\~ao:
\begin{align*}
A(B+C) &= \{a(b+c)\colon a\in A,b\in B,c\in C\} \\
&= \{ab+ac\colon  a\in A,b\in B,c\in C\} \\
&= AB+AC.
\end{align*}
Finalmente, se $A\leq B$ e $C\in\R_{>0}$, ent\~ao
\[AC\leq BC.\]
\end{exercicio}

O conjunto $(0,1)$ pertence a $\R_{>0}$ e joga o papel da identidade multiplicativa.

\begin{exercicio}
Qualquer que seja $A\in\R_{>0}$, 
\[(0,1)\cdot A = A.\]
\end{exercicio}

A constru\c c\~ao do inverso multiplicativo \'e um pouco mais complicada. Dado $A\in\R_{>0}$, formamos o conjunto de inversos do conjunto complimentar, 
\[\left( A^c\right)^{-1} = \{b^{-1}\colon b\in A^c\},\]
e removamos o m\'\i nimo deste conjunto (que sempre existe). Denotemos o conjunto resultante $A^\S$. 

\begin{exercicio}
$A\cdot A^\S = (0,1)$.
\end{exercicio}

Agora definamos o sistema $\R_{<0}$ de n\'umeros estritamente negativos como uma c\'opia formal de $\R_{>0}$,
\[\R_{<0} = \{-x\colon x\in\R_{>0}\}.\]
Munamos $\R_{>0}$ de uma ordem,
\[-r < -t \iff t < r,\]
uma adi\c c\~ao,
\[(-r) + (-t) = -(r+t),\]
assim como uma {\em multiplica\c c\~ao externa} que toma seus valores dentro de $\R_{>0}$:
\[(-r)\cdot (-t) = rt.\]

Finalmente, definamos o sistema $\R$ como a uni\~ao disjunta de $\R_{>0}$, $\R_{<0}$, e o zero:
\[\R = \R_{<0}\cup \{0\}\cup \R_{>0}.\]
A ordem total \'e definido sobre $\R$ por
\[-r < 0 < t.\]
A multiplica\c c\~ao \'e estendida sobre $\R$ pela regra
\[r\cdot (-t)  =- rt.\]
A adi\c c\~ao ser\'a definida assim. Se $r<t$, ent\~ao
\[(-r) + t = t-r\]
(a subtra\c c\~ao sendo bem definida), e se $r>t$, 
\[ (-r) + t = -(r-t).\]
Finalmente,
\[1 = (0,1)\in\R_{>0}.\]

A verifica\c c\~ao do que o sistema $\R=\langle\R,+,\times,0,1,\leq \rangle$ forme um corpo ordenado n\~ao \'e dif\'\i cil, \'e uma cadeia dos exerc\'\i cios puramente t\'ecnicos. Verifiquemos o seguinte.

\begin{teorema}
O sistema $\R=\langle\R,+,\times,0,1,\leq \rangle$ satisfaz o axioma de Dedekind.
\label{t:rsatisfazdedekind}
\end{teorema}

Seja $X$ um subconjunto n\~ao vazio \'e limitado por cima: existe $b\in\R$ tal que para todos $x\in X$, temos $x\leq b$.
Escolhamos um elemento $x_0\in X$ qualquer. Denotemos
\[\hat X = \{x-x_0+1\colon x\in X\}.\]
O conjunto $\hat X$ cont\'em o elemento $x_0 -x_0+1 = 1$,
de onde conclu\'\i mos
\[\hat X\cap \R_{>0}\neq\emptyset.\]
Al\'em disso, $\hat X$ \'e majorado por $b^\prime = b-x_0+1$.
Basta mostrar que $\hat X$ possui um supremo, $s$, pois neste caso $s+x_0-1$ 
ser\'a o supremo de $X$.

Referindo \`a nossa defini\c c\~ao de $\R_{>0}$, definamos o conjunto 
\[S = \bigcup_{A\in\hat X}A.\]
Este $S$ \'e um subconjunto de $\Q_{>0}$.
Como $\hat X\neq \emptyset$, conclu\'\i mos que $S\neq\emptyset$. 
O conjunto $\hat X$ est majorado por $b^\prime$. Na interpreta\c c\~ao conjunt\'\i stica dos reais,
\begin{equation}
\label{eq:bprime}
A\subseteq b^\prime.
\end{equation}
Por conseguinte, $\emptyset\neq(b^\prime)^c \subseteq A^c$,
de onde conclu\'\i mos que $A^c\neq\emptyset$.

Sejam $s\in S$ e $x\in\Q_{>0}$ quaisquer tais que $x\leq s$.
Existe $A\in\hat X$ com $s\in A$.
Este $A$ \'e um intervalo inicial de $\Q_{>0}$, e por conseguinte
\[x\in A\subseteq S.\]
Agora suponha que $S$ cont\'em o elemento maximal, $d\in\Q_{>0}$:
\[d\in S\mbox{ e }\forall s\in S,~~s\leq d.\]
Ent\~ao, existe $A\in\hat X$ tal que $d\in A$,
e como $A\subseteq S$, conclu\'\i mos que $d$ \'e o elemento maximal de $A$, o que \'e imposs\'\i vel.

As propriedades acima significam que $S$ \'e um n\'umero real.

Qualquer que seja $A\in \hat X$,
\[A\subseteq \cup\{B\colon B\in\hat X\}=S,\]
o que significa
\[A\leq S,\]
ou seja, $S$ \'e um majorante de $\hat X$. Agora seja $C$ um majorante qualquer de $\hat X$. Para todos $A\in\hat X$, $A\leq C$,
ou seja,
$A\subseteq C$.
Conclu\'\i mos:
\[S =\cup\{A\colon A\in\hat X\}\subseteq C,\]
o que \'e interpretado como $S\leq C$.
A demonstra\c c\~ao do teorema \ref{t:rsatisfazdedekind} \'e terminada.

A unicidade de $\R$ no teorema \ref{t:existenciadeR} a menos de um isomorfismo mais ou menos reproduz o argumento acima, a partir do fato seguinte.

\begin{exercicio}
Mostre que cada corpo ordenado cont\'em o corpo de racionais, $\Q$, como um subcorpo ordenado.
\end{exercicio}

\subsection{Axioma de Arquimedes e propriedade de Cantor}

\begin{definicao}
Um corpo ordenado $K$ satisfaz o {\em axioma de Arquimedes,} ou \'e {\em arquimediano,} se, quaisquer que sejam $x$ e $y$ estritamente positivos, existe um n\'umero natural $n$ tal que $nx>y$, ou seja,
\[\underbrace{x+x+x+\dots+x}_{n~\text{\rm vezes}}>y.\]
Se $K$ n\~ao satisfaz o axioma de Arquimedes, ele \'e dito {\em n\~ao arquimediano}.
\end{definicao}

\begin{teorema}
Um corpo completo no sentido de Dedekind \'e arquimediano.
\label{3.9}
\end{teorema}

\begin{exercicio}
Mostre o teorema pela contraposi\c c\~ao, usando a mesma ideia que no exemplo \ref{ex:qofalpha}.
\end{exercicio}

\begin{exemplo}
O corpo $\Q(\alpha)$ \'e n\~ao arquimediano: para cada $n\in\N$, $n\alpha<1$.
\end{exemplo}

\begin{exemplo}
Como o exemplo de $\Q$ mostra, o axioma de Arquimedes n\~ao implica o axioma de Dedekind.
\end{exemplo}

\begin{teorema}
O corpo de n\'umeros reais possui a propriedade de interse\c c\~ao de Cantor: cada sequ\^encia encaixada de intervalos fechados finitos tem a interse\c c\~ao n\~ao vazia. Se o comprimento dos intervalos converge para zero, a interse\c c\~ao consiste de um ponto. 
\label{th:cantor}
\index{teorema! de interser\c c\~ao de Cantor}
\end{teorema}

\begin{proof}
Seja $[a_n,b_n]$, $n\in\N_+$ uma sequ\^encia encaixada,
\[a_1\leq a_2\leq \ldots \leq a_n\ldots \leq \ldots b_n\ldots b_2\leq b_1.\]
O conjunto $A=\{a_n:n\in\N\}$ \'e limitado por cima, logo possui o supremo$c=\sup A$. Em particular, $c\leq b_n$ para todos $n\in\N$. Isso significa
\[\forall n\in\N,~~a_n\leq c \leq b_n,\]
como desejado. Supondo que $b_n-a_n\to 0$, se $d\in \cap_{n\in\N}[a_n,b_n]$, ent\~ao
\[\abs{c-d}\leq b_n-a_n
\]
para todos $n$, logo $c=d$.
\end{proof}

\begin{exemplo}
O corpo ordenado $\Q$ n\~ao satisfaz a propriedade de interse\c c\~ao de Cantor, como testemunhado pela sequ\^encia de intervalos
\[[0,2],~[0.1, 0.2], ~[0.101,0.102],~[0.101001,0.101002],[0.1010010001,0.1010010002],\dots\]
\end{exemplo}

\begin{exemplo} 
O corpo ordenado $\Q(\alpha)$ n\~ao satisfaz a propriedade de interse\c c\~ao de Cantor tamb\'em:  a interse\c c\~ao da sequ\^encia encaixada
\[\left[n\alpha, \frac 1 n\right],~~n=1,2,\dots\]
\'e vazia.
\end{exemplo}

\begin{observacao}
A propriedade de interse\c c\~ao de Cantor \'e mais fraca do que o axioma de Dedekind. Pode-se construir alguns corpos ordenados n\~ao-arquimedianos onde cada sequ\^encia encaixada de intervalos fechados tem uma interse\c c\~ao n\~ao vazia (tais corpos s\~ao ditos {\em esfericamente completos}), e ademais, nenhuma sequ\^encia enumer\'avel converge para zero.
\end{observacao}

\begin{exercicio}
Mostre que o axioma de Dedekind \'e equivalente \`a propriedade de interse\c c\~ao de Cantor junto com o axioma de Arquimedes.
\end{exercicio}

\section{M\'etricas, topologias, fun\c c\~oes}

\subsection{Espa\c cos m\'etricos\label{ss:espmet}}

\begin{definicao}
Seja $X$ um conjunto n\~ao vazio. Uma {\em m\'etrica} sobre $X$ \'e uma fun\c c\~ao 
\[d\colon X\times X\to\R_+\]
que verifica os axiomas seguintes:
\begin{enumerate}
\item \label{axone}
$\forall x,y\in X$, $d(x,y)=0$ se e somente se $x=y$.
\item (Simetria) $\forall x,y\in X$, $d(x,y)=d(y,x)$.
\item (Desigualdade triangular) $\forall x,y,z\in X$, $d(x,z)\leq d(x,y)+d(y,z)$.
\end{enumerate}
\label{def:distance}
\end{definicao}

\begin{exercicio}
\label{r:positif}
Mostre que a positividade de valores de uma m\'etrica \'e uma consequ\^encia imediata dos tr\^es axiomas. 
\end{exercicio}

\begin{definicao}
Um conjunto $X$ n\~ao vazio qualquer, munido duma dist\^ancia $d$, \'e chamado um {\em espa\c co m\'etrico.} Em outros termos, um espa\c co m\'etrico \'e um par $(X,d)$, onde $d$ \'e uma m\'etrica sobre o conjunto n\~ao vazio $X$.

Habitualmente, um espa\c co m\'etrico \'e notado pela \'unica letra $X$.
\index{espa\c co! m\'etrico}
\par
Os elementos de um espa\c co m\'etrico s\~ao chamados os {\em pontos}, a fim de sublinhar a natura geom\'etrica dos objetos desse tipo. 
\end{definicao}

\begin{exemplo}
A dist\^ancia dita {\em usual} sobre $\R$ \'e dada pela formula
\[d(x,y)=\abs{x-y}.\]
\end{exemplo}

\begin{exemplo}
Seja $X$ um conjunto n\~ao vazio. A {\em m\'etrica discreta}, ou {\em m\'etrica zero-um,} sobre $X$ \'e dada pela regra 
\[d_{discr}(x,y)=\begin{cases} 1,&\mbox{ se }x\neq y,\\
0,&\mbox{ se }x=y.
\end{cases}\]
\end{exemplo}

\begin{definicao}
Sejam $X=(X,d)$ um espa\c co m\'etrico, $Y\subseteq X$ um subconjunto de $X$ n\~ao vazio. A {\em restri\c c\~ao} de $d$ sobre $Y$, notada $d\vert_Y$, \'e, com efeito, a restri\c c\~ao de $d$ sobre o produto cartesiano $Y\times Y$, ou seja,
\[\forall x,y\in Y,~~d\vert_Y(x,y)=d(x,y).\]
\'E \'obvio que $d\vert_Y$ \'e uma m\'etrica sobre $Y$, chamada a {\em m\'etrica} (ou {\em dist\^ancia}) {\em induzida} sobre $Y$ (de $X$).
O espa\c co m\'etrico $(Y,d\vert_Y)$ \'e dito {\em um subespa\c co m\'etrico} de $X$. Cada subconjunto n\~ao vazio de um espa\c co m\'etrico $X$ define um subespa\c co m\'etrico da maneira \'unica. 
\end{definicao}

\begin{exemplo}
O espa\c co
\[\alpha\N=\left\{\frac 1n\colon n=1,2,3,\ldots\right\}\cup \{0\}\]
\'e munido da dist\^ancia induzida da reta $\R$, ou seja, a dist\^ancia entre dois pontos $x,y\in\alpha\N$ \'e a mesma que no $\R$. Isso espa\c co consiste dos membros duma sequ\^encia convergente com o seu limite. Ele \'e chamado {\em a compactifica\c c\~ao de Alexandroff}
(ou {\em a compactifica\c c\~ao por um ponto}) {\em de um conjunto enumer\'avel}.
\end{exemplo}

\begin{exemplo}
O intervalo fechado
\[[0,1]=\{x\in\R\colon 0\leq x\leq 1\},\]
munida da dist\^ancia usual (induzida da reta $\R$), encontra-se entre aos exemplos mais importantes dos espa\c cos m\'etricos.
\end{exemplo}

\begin{exemplo} 
O espa\c co m\'etrico dos n\'umeros racionais, 
\[\Q = \left\{\frac mn\colon m,n\in\Z,~n\neq 0\right\}\]
munido da fun\c c\~ao dist\^ancia usual, \'e um subespa\c co m\'etrico da reta.
\end{exemplo}

\begin{exemplo}
O espa\c co vetorial real da dimens\~ao $n$,
\[\R^n=\underbrace{\R\times\R\times\ldots\times\R}_{\mbox{$n$ vezes}},\]
munido da {\em dist\^ancia euclideana usual:} 
\[d(x,y)=\sqrt{\sum_{i=1}^n \abs{x_i-y_i}^2}.\]
\end{exemplo}

Os axiomas 1 e 2 s\~ao \'obvios, enquanto o axioma 3 exige uma demonstra\c c\~ao. 

\begin{definicao}
Recordemos que o {\em produto escalar padr\~ao} de dois vetores no $\R^n$ \'e dado pela formula
\[(x,y) = \sum_{i=1}^nx_iy_i.\]
\end{definicao}

O produto escalar tem as propriedades seguintes verificadas facilmente. 

\begin{proposicao}
\begin{enumerate}
\item $(x,x)\geq 0$ e $(x,x)=0$ se e somente se $x=0$.
\item $(x,y)=(y,x)$.
\item $(x+y,z)=(x,z)+(y,z)$.
\item $(\lambda x,y)=\lambda (x,y)$.
\end{enumerate}
\end{proposicao}

Quaisquer que sejam $x,y\in \R^n$, temos
\[d(x,y)=\sqrt{(x-y,x-y)}.\]
A desigualdade triangular pela dist\^ancia euclideana \'e uma consequ\^encia imediata do teorema seguinte. 

\begin{teorema}[Desigualdade de Minkowski]
Para todos $x,y\in\R^n$ temos
\[\sqrt{(a+b,a+b)}\leq \sqrt{(a,a)}+\sqrt{(b,b)}.\]
\label{th:minkowski}
\end{teorema}

A fim de mostrar a desigualdade de Minkowski, precisamos de um outro resultado.

\begin{teorema}[Desigualdade de Cauchy--Schwarz]
Para todos $x,y\in\R^n$ temos
\[\abs{(x,y)}^2\leq (x,x)(y,y).\]
\end{teorema}

\begin{proof}
Sejam $x,y\in\R^n$ dois vetores quaisquer. Para todos 
$t\in\R$ temos
\[0\leq (tx+y, tx+y) = t^2 (x,x) +2t(x,y) + (y,y).\]
Por conseguinte, o polin\^omio do grau $2$ em $t$, 
\[t^2 (x,x) +2t(x,y) + (y,y),\]
tem um raiz no m\'aximo. Isto quer dizer que o discriminante do polin\^omio \'e negativo,
\[[2(x,y)]^2 - 4(x,x)(y,y) \leq 0,\]
ou seja
\[\vert (x,y)\vert^2 = (x,y)^2 \leq (x,x)(y,y).\]
\end{proof}

\begin{proof}[Demonstra\c c\~ao do teorema \ref{th:minkowski}]
\begin{align*}
(x+y,x+y) &= (x,x)+(x,y)+(y,x)+(y,y) \\
&=
 (x,x)+(x,y)+(x,y)+(y,y) \\ &=
(x,x)+2(x,y) +(y,y)\\ & \leq
(x,x)+2\vert (x,y)\vert + (y,y)\\ \mbox{\small (ap\'os Cauchy--Schwarz)}&\leq
 (x,x)+2\sqrt{(x,x)}\sqrt{(y,y)}+(y,y) \\
&= [\sqrt{(x,x)}+\sqrt{(y,y)}]^2,
\end{align*}
o que implica o resultado desejado.
\end{proof}

\begin{exemplo}
\label{e:ellinfty}
Notaremos $\ell^\infty$ o conjunto de todas as sequ\^encias limitadas dos n\'umeros reais, $x=(x_n)_{n\in\N_+},~x_n\in\R$. 
(Recordemos que uma sequ\^encia infinita 
$(x_n)$ \'e dita {\it limitada} se existe um n\'umero real $L\geq 0$ tal que
para todos $n=1,2,3,\dots$ temos $\vert x_n\vert\leq L$).

O valor da dist\^ancia 
\[d_{\infty}(x,y)=\sup_{n\in\N_+}\vert x_n-y_n\vert\]
entre duas sequ\^encias $x,y\in \ell^\infty$ \'e bem definido porque $x$ e $y$ s\~ao limitadas, bem como a sua diferen\c ca $x-y$. A verifica\c c\~ao dos axiomas de uma m\'etrica para $d_{\infty}$ ser\'a deixada como exerc\'\i cio. Esta m\'etrica $d_{\infty}$ \'e chamada a m\'etrica do tipo $\ell^\infty$, ou a m\'etrica uniforme.
\end{exemplo}

\begin{exemplo}
O
{\em espa\c co de Baire},
notado $\Z^\omega$, consiste de todas as sequ\^encias infinitas dos n\'umeros inteiros
$x=(x_1,x_2,\dots,x_n,\dots)$, onde $x_i\in\Z$, $i=1,2,\ldots$. O valor da dist\^ancia entre as duas sequ\^encias,
\[x=(x_1,x_2,\dots,x_n,\dots)\]
e
\[y=(y_1,y_2,\dots,y_n,\dots),\]
\'e dado pela express\~ao seguinte:
\[d (x,y)=\begin{cases} 0, ~\text{se $x=y$}, \\
2^{-\min\{ i\colon~ x_i\neq y_i\}} ~\text{se $x\neq y$}.
\end{cases}\]
Por exemplo, a dist\^ancia entre os elementos
\[(1,2,3,4,5,\dots, n,\dots)\]
e
\[(1,2,3,0,0,0,0,\dots,0,\dots)\]
\'e igual a $1/8$.

Os axiomas 1 e 2 s\~ao praticamente evidentes. A fim de verificar a desigualdade triangular, sejam $x,y,z$ tr\^es pontos em $\Z^\omega$. O nosso objetivo \'e mostrar que
\begin{equation}
\label{eq:baire}
d (x,z)\leq  d (x,y)+ d (y,z).
\end{equation}
Se pelo menos dois entre os elementos $x,y,z$ s\~ao iguais, ent\~ao a desigualdade (\ref{eq:baire}) \'e verificada da maneira trivial. Por conseguinte, pode-se supor que todas sequ\^encias $x,y,z$ s\~ao distintas.

Denotemos $i$ o menor n\'umero inteiro tal que (i) para todos $j<i$ temos $x_j=y_j=z_j$, e (ii) os inteiros $x_i,y_i,z_i$ n\~ao s\~ao todos iguais. 
Um tal $i$ existe ap\'os a hip\'otese. Agora \'e \'obvio o que $d (x,z)\leq 2^{-i}$. Logo, o n\'umero \`a esquerda da equa\c c\~ao (\ref{eq:baire}) \'e menor ou igual a $2^{-i}$. Ao mesmo tempo, pelo menos um entre os n\'umeros $d(x,y)$ e $ d (y,z)$ \`a direita \'e igual a $2^{-i}$. (Se n\~ao, ent\~ao $x_i=y_i$ e $y_i=z_i$, por conseguinte  
$x_i=y_i=z_i$, o que contradiz a escolha de $i$).    
\label{ex:espacodebaire}
\end{exemplo}

\subsection{Espa\c cos ultram\'etricos}

\begin{observacao}
\label{obs:ultra}
Com efeito, o argumento acima mostra que a m\'etrica sobre o espa\c co de Baire satisfaz a {\em desigualdade triangular forte}:
\begin{equation}
\label{eq:ultra}
d(x,z)\leq \max\{d(x,y),d(y,z)\}.
\end{equation}
\end{observacao}

\begin{definicao}
Uma m\'etrica \'e chamada {\em ultram\'etrica} se ela satisfaz a desigualdade (\ref{eq:ultra}). Um conjunto munido de uma ultram\'etrica \'e dito {\em espa\c co ultram\'etrico}.
\end{definicao}

\begin{exemplo}
O espa\c co m\'etrico cuja m\'etrica \'e a m\'etrica zero-um \'e ultram\'etrico.
\end{exemplo}

\begin{exemplo}
Segundo observa\c c\~ao \ref{obs:ultra}, o espa\c co de Baire \'e um espa\c co ultram\'etrico.
\end{exemplo}

\subsection{Conjuntos abertos}

\begin{definicao}
Sejam $(X,d)$ um espa\c co m\'etrico, $x$ um ponto no $X$, e $\epsilon>0$. A {\em bola aberta de centro $x$ e raio $\epsilon$} \'e o conjunto
\[B_\epsilon (x) := \{y\in X:  d  (x,y) < \epsilon\}\subseteq X.\]
\end{definicao} 

\begin{exemplo}
Num espa\c co m\'etrico $(X,d)$ munido da dist\^ancia zero-um, cada bola aberta \'e igual ou ao espa\c co inteiro ou a um conjunto unit\'ario.
\end{exemplo}

\begin{definicao}
Seja $X$ um espa\c co m\'etrico. Um subconjunto $V$ de $X$ \'e dito {\em aberto} no $X$ se para cada ponto $x$ de $V$ existe um valor $\e>0$ tal que a bola $B_\e(x)$ \'e contida no $V$. 
\end{definicao} 

\begin{exercicio}
Cada bola aberta \'e um conjunto aberto. 
\end{exercicio}

\begin{proposicao} Seja $(X,d )$ um espa\c co m\'etrico. Ent\~ao:
\begin{enumerate}
\item O conjunto vazio $\emptyset$ e o espa\c co $X$ s\~ao abertos no $(X,d)$.
\smallskip
\item A interse\c c\~ao de dois abertos \'e aberto. Por conseguinte, se
$V_1, ... ,V_k$ s\~ao abertos, ent\~ao $\cap_{i=1}^k V_i$ \'e aberto.
\smallskip
\item A uni\~ao $V=\cup_{\alpha\in A}V_\alpha$ duma fam\'\i lia dos abertos \'e aberta.
\end{enumerate}
\label{p:topologiametrica}
\end{proposicao}

\begin{observacao}
A fam\'\i lia $\mathscr T$ de todos os abertos de um espa\c co m\'etrico $X$ \'e chamada a {\em topologia} de $X$. As tr\^es propriedades acima s\~ao escolhidas como os axiomas de um {\em espa\c co topol\'ogico,} que \'e uma no\c c\~ao mais geral do que a no\c c\~ao de espa\c co m\'etrico.
\end{observacao}

\begin{definicao} Sejam $X=(X,d)$ um espa\c co m\'etrico, $A$ um subconjunto de $X$, e $x$ um ponto no $A$. Suponhamos que existe um $\e>0$ tal que 
$x\in B_\epsilon (x) \subseteq A$. Ent\~ao $x$ \'e chamado um {\em ponto interior} de $A$, enquanto $A$ \'e chamado uma {\em vizinhan\c ca} de $x$. 
\end{definicao}

\begin{observacao}
A vizinhan\c ca de um ponto n\~ao \'e necessariamente um conjunto aberto em $X$. 
\end{observacao}

\begin{definicao}
Uma parte $A$ de um conjunto $X$ \'e chamada de {\em cofinita} se o seu complemento $A^c=X\setminus A$ \'e finito.
\end{definicao}

\begin{proposicao}
Cada parte cofinita de um espa\c co m\'etrico qualquer \'e aberta. 
\label{p:cofinito}
\end{proposicao}

\begin{proof}
Exerc\'\i cio.
\end{proof}

\begin{proposicao}
Cada bola aberta num espa\c co ultram\'etrico \'e um conjunto fechado.
\end{proposicao}

\begin{proof}
Sejam $X$ um espa\c co ultram\'etrico, $x\in X$, e $\e>0$. Se $y\notin B_\e(x)$, escolhamos $\delta>0$ qualquer tal que $\delta<\e$. Suponhamos que existe $z\in B_\e(x)\cap B_\delta(y)$. Ent\~ao temos
\[d(x,y)\leq \max\{d(x,y),d(y,z)\}\leq \max\{d(x,y),\delta\}<\e,\]
logo $y\in B_\e(x)$, a contradi\c c\~ao. Conclu\'\i mos: $B_\e(x)$ e $B_\delta(y)$ s\~ao disjuntas, isto \'e, o complemento da bola $B_\e(x)$ \'e uma vizinhan\c ca de $y$.
\end{proof}

\begin{exercicio}
Verdadeiro ou falso: cada bola fechada num espa\c co ultram\'etrico \'e um conjunto aberto?
\end{exercicio}

\begin{definicao} Seja $X$ um espa\c co m\'etrico. Um ponto $x\in X$ \'e dito {\em isolado} se o conjunto unit\'ario $\{x\}$ \'e aberto no $X$. 
\end{definicao}

\begin{exemplo} Num espa\c co m\'etrico $X$ munido da dist\^ancia zero-um cada ponto \'e isolado. 
\end{exemplo}

\begin{exemplo} Cada elemento com exce\c c\~ao de $0$ \'e isolado no espa\c co m\'etrico $\alpha\N$. 
\end{exemplo}

\begin{exercicio} Mostre que um subconjunto, $A$, de $\alpha\N$ \'e aberto se \'e somente se ou $A$ n\~ao contem zero, ou $A$ \'e cofinito. (As duas propriedades n\~ao s\~ao mutualmente exclusivas).  \ref{p:cofinito}.
\end{exercicio}

\subsection{Interior}

\begin{definicao}
Sejam $(X,d)$ um espa\c co m\'etrico e $A$ um subconjunto de $X$. O {\em interior} de $A$, notado $\Int_XA$ ou simplesmente $\Int A$, \'e um subconjunto de $X$ que consiste de todos os pontos interiores de $A$:
\[\Int_XA = \{x\in A\colon \exists \e>0~~B_\e(x)\subseteq A\}.\]
\`As vezes utiliza-se o s\'\i mbolo $\overset{\circ}{A}$.
\end{definicao}

\begin{observacao}
Seja $X$ um espa\c co m\'etrico. Uma parte $A$ de $X$ \'e aberta se e somente se $A=\Int_XA$. \'E uma consequ\^encia imediata das defini\c c\~oes.
\end{observacao}

\begin{exemplo}
O interior de cada espa\c co m\'etrico $X$ em si mesmo \'e $X$:
\[\Int_XX=X.\]
Por exemplo,
\[\Int_\Q\Q =\Q.\]
\end{exemplo}

\begin{exemplo}
O interior do mesmo conjunto $\Q$ formado no espa\c co m\'etrico $\R$ \'e vazio:
\[\Int_\R\Q=\emptyset.\]
Em outras palavras, nenhum intervalo aberto n\~ao vazio, $(a,b)$, dos n\'umeros reais n\~ao est\'a contido em $\Q$: existe pelo menos um ponto irracional no intervalo $(a,b)$. 
\end{exemplo}

\subsection{M\'etricas equivalentes}

\begin{definicao}
 Duas m\'etricas $d$ e $d^\prime$ sobre um conjunto $X$ s\~ao {\em equivalentes} se elas geram a mesma topologia: um subconjunto $A$ de $X$ \'e aberto no espa\c co m\'etrico $(X,d)$ se \'e somente se ele \'e aberto no espa\c co $(X,d^\prime)$.
 \end{definicao}
 
\begin{exemplo}
A m\'etrica zero-um sobre $\R$ n\~ao \'e equivalente \`a m\'etrica usual, por que o conjunto unit\'ario $\{0\}$ \'e aberto no espa\c co $(\R,d_{0-1})$, mas n\~ao \'e aberto no espa\c co $\R$ relativo \`a m\'etrica usual. 
 \end{exemplo}
 
\begin{exercicio}
 A m\'etrica usual (induzida do $\R$) sobre o espa\c co $\alpha\N$ n\~ao \'e uma ultram\'etrica:
 \[d(0,1) = 1 >\frac 12 = \max\{d(0,1/2),d(1/2,1)\}.\]
 Verifique que a regra seguinte define uma ultram\'etrica equivalente: $d(x,x)=0$ para cada $x$, e se $x\neq y$, ent\~ao $(d(x,y)=\max\{x,y\}$.
\end{exercicio}
 
\begin{exercicio}
Seja $(X,d)$ um espa\c co m\'etrico qualquer. Mostre que a f\'ormula
\[\tilde d(x,y) = \min\{d(x,y),1\}\]
define uma m\'etrica equivalente sobre $X$.
\label{ex:metricalimitadaequivalente}
\end{exercicio}

\subsection{Conjuntos fechados}

\begin{definicao}
Digamos que uma parte $F$ de um espa\c co m\'etrico $X$ \'e {\em fechada} em $X$ (ou: $F$ \'e um {\em fechado}), se o complemento $F^c= X\setminus F$ \'e aberto.
\end{definicao}

\begin{exemplo} 
Cada conjunto unit\'ario $\{x\}$, $x\in X$ \'e fechado num espa\c co m\'etrico $X$ qualquer. 
\end{exemplo}

\begin{exemplo} 
No espa\c co m\'etrico $\R$ dos n\'umeros reais munido da distancia usual $d(x,y)=\abs{x-y}$, todo intervalo fechado
\[[a,b] = \{x\in\R\colon a\leq x\leq b\}\]
\'e um subconjunto fechado de $\R$. 
\end{exemplo}

\begin{observacao}
Uma parte $A$ de um espa\c co m\'etrico pode ser ao mesmo tempo aberta \'e fechada, as dual propriedades n\~ao s\~ao mutualmente exclusivas. 
\end{observacao}

\begin{exercicio}
Sejam $X$ um espa\c co m\'etrico qualquer, $Y\subseteq X$ um subespa\c co m\'etrico de $X$. Se $V\subseteq X$ \'e aberto (fechado) em $X$, ent\~ao a interse\c c\~ao $V\cap Y$ \'e aberta (respectivamente, fechada) em $Y$. 
\label{ex:rel}
\end{exercicio}

\begin{exemplo}
Sejam $a,b\in\R\setminus\Q$, $a<b$. No espa\c co $\Q$ dos n\'umeros racionais, denotemos
\[A = \{x\in\Q\colon a<x<b\}.\]
Ent\~ao a parte $A$ \'e ao mesmo tempo aberta e fechada em $\Q$. Ela \'e aberta por que 
\[A = \Q\cap (a,b),\]
gra\c cas ao Exerc\'\i cio \ref{ex:rel}. Da mesma maneira, como
\[A = \Q\cap [a,b],\]
conclu\'\i mos que $A$ \'e fechado em $\Q$. 
\end{exemplo}

\begin{definicao} 
Seja $X=(X,d)$ um espa\c co m\'etrico. A {\em bola fechada} de
raio $r>0$ em torno de $x\in X$ \'e o conjunto definido por
\[\bar B_\e(x)=\{y\in X\colon d(x,y)\leq r\}.\]
\end{definicao}

\begin{exercicio} Cada bola fechada \'e um conjunto fechado.
\end{exercicio}

\begin{teorema}
Seja $X=(X,d)$ um espa\c co m\'etrico qualquer. 

\begin{enumerate}
\item $X$ \'e o conjunto vazio $\emptyset$ s\~ao fechados.
\item A uni\~ao de dois conjuntos fechados \'e fechado. 
\item A interse\c c\~ao de uma fam\'\i lia n\~o vazia qualquer de fechados \'e um fechado. 
\end{enumerate}
\end{teorema}

\begin{definicao} A {\em esfera} de raio $r>0$ centrada em $x\in X$ \'e o conjunto
\[S_r(x) =\{y\in X\colon d(x,y)=r\}.\]
\end{definicao}

\begin{exemplo} 
Cada esfera \'e um conjunto fechado, sendo a interse\c c\~ao de dois fechados:
\[S_r(x) = \bar B_r(x)\cap B_r(x)^c.\]
\end{exemplo}

\subsection{Ader\^encia}

\begin{definicao} Uma sequ\^encia $(x_n)$ de pontos de um espa\c co m\'etrico $(X,d)$ \'e dita {\em convergente} {\em para o limite} $x\in X$ se 
\[\forall \e>0~~\exists n_0~~\forall n\geq n_0~~d(x_n,x)<\e.\]
Em outras palavras, cada bola aberta em torno de $x$ contem todos os membros da sequ\^encia excepto um n\'umero finito. Simbolicamente: 
\[\lim_{n\to\infty} x_n = x,\]
ou 
\[x_n\overset{n\to\infty}{\longrightarrow} x.\]
\end{definicao}

\begin{definicao}
Sejam $X$ um espa\c co m\'etrico, $A$ um subconjunto de $X$. Um ponto $x$ de $X$ \'e chamado um ponto {\em aderente} de $A$ (em $X$) se existe uma sequ\^encia  $(a_n)$ dos pontos de $A$ que converge para $x$:
\[\lim_{n\to\infty} a_n = x.\]
\end{definicao}

\begin{exemplo}
\label{ex:supremo}
Seja $A$ um subconjunto n\~ao vazio qualquer da reta $\R$. Suponhamos que $A$ seja limitado superiormente. Ent\~ao o supremo $b=\sup A$ de $A$ e um ponto aderente a $A$. De mesmo para o \'\i nfimo $\inf A$.
\end{exemplo}

\begin{proposicao}
Um ponto $x$ de um espa\c co m\'etrico qualquer $X$ \'e um ponto aderente de uma parte $A$ de $X$ se \'e somente se toda vizinhan\c ca de $x$ tem pontos em comum com $A$.
\end{proposicao}

\begin{proof}
Exerc\'\i cio.
\end{proof}

\begin{definicao} A {\em ader\^encia}, ou o {\em fecho,} de uma parte $A$ de um espa\c co m\'etrico $X$ \'e o conjunto de todos os pontos aderentes de $A$. A ader\^encia \'e denotada por $\bar A$ ou, se quer-se sublinhar o espa\c co \^ambito, por $\bar A^X$. (A nota\c c\~ao alternativa e bastante popular: ${\mathrm{cl}}\,A$, ou ${\mathrm{cl}}_X(A)$, da palavra ingl\^esa {\tt ``closure''} para a ader\^encia.)
\end{definicao}

\begin{teorema} 
O ``operador da ader\^encia'' possui as propriedades seguintes:
\begin{enumerate}
\item $\bar\emptyset =\emptyset$.
\\[.5mm]
\item $A\subseteq\bar A$.
\\[.5mm]
\item se $A\subseteq B$, ent\~ao $\bar A\subseteq \bar B$.
\\[.5mm]
\item $\bar{\bar A}=\bar A$.
\end{enumerate}
\label{th:4}
\end{teorema}

\begin{proof} 
Exerc\'\i cio.
\end{proof}

\begin{observacao}
As propriedades (1)-(4) podem ser usados como axiomas de um espa\c co topol\'ogico (axiomas de Kuratowski).
\end{observacao}

\begin{teorema} Uma parte $A$ de um espa\c co m\'etrico \'e fechada se e somente se $\bar A = A$.
\label{th:ssi}
\end{teorema}

\begin{proof} $\Rightarrow$: se a parte $A$ \'e fechada, ent\~ao o complemento $A^c$ \'e aberto e para cada $x\in A^c$ existe $\e>0$ com a propriedade $B_\e(x)\subseteq A^c$. Logo, a bola $B_\e(x)$ n\~ao cont\'em nenhum ponto de $A$, e por conseguinte $x$ n\~ao \'e aderente a $A$. Conclu\'\i mos: todos os pontos aderentes a $A$ j\`a pertencem a $A$. De acordo com a afirma\c c\~ao (2) do teorema \ref{th:4}, $\bar A = A$.

$\Leftarrow$: Suponhamos que $\bar A = A$. Seja $x\in A^c$ um ponto qualquer. Ent\~ao $x$ n\~ao \'e aderente a $A$, \'e pode-se escolher um n\'umero real $\e>0$ tal que $B_\e(x)\cap A=\emptyset$. Segue-se que o complemento $A^c$ \'e aberto.
\end{proof}

\begin{corolario}
A ader\^encia $\bar A$ de qualquer conjunto $A$ \'e um conjunto fechado.
\label{c:ferme}
\end{corolario}

\begin{proof}
De acordo com o teorema  \ref{th:4}(3), temos $\bar{\bar A}=\bar A$.
\end{proof}

\begin{corolario} 
A ader\^encia de um conjunto $A$ \'e o menor fechado contendo $A$. 
\end{corolario}

\begin{proof}
Se $A$ \'e um subconjunto de fechado qualquer, $F$, ent\~ao 
\[\bar A\subseteq \bar F = F\]
(teorema \ref{th:ssi}). 

Ao mesmo tempo, $\bar A$ \'e fechado (corol\'ario \ref{c:ferme}). Conclu\'\i mos: $\bar A$ \'e um fechado que contem $A$ e que \'e contido em cada fechado contendo $A$. 
\end{proof}

\subsection{Partes densas}

\begin{definicao}
Uma parte $A$ de um espa\c co m\'etrico qualquer $X$ \'e dita {\em densa}, se
\[\bar A= X.\]
\end{definicao}

\begin{observacao}
As condi\c c\~oes seguintes s\~ao (evidentemente) equivalentes:
\begin{enumerate}
\item $A$ \'e denso em $X$.
\item Cada ponto de $X$ \'e aderente a $A$.
\item Cada aberto n\~ao vazio de $X$ encontra $A$, ou seja, contem algum ponto de $A$.
\item Para todos $x\in X$ e todos $\e>0$, a bola $B_\e(x)$ contem algum ponto de $A$.
\end{enumerate}
\end{observacao}

\begin{exemplo} O conjunto $\Q$ dos n\'umeros racionais \'e denso na reta $\R$.
Este fato \'e conhecido como {\em teorema da densidade}. Sejam  $a,b\in\R$, $a<b$. De acordo com a propriedade de Arquimedes, existe um $n\in\N_+$ tal que 
\[\frac 1n <b-a.\]
Por conseguinte, entre os n\'umeros racionais
\[\frac k n, ~~k\in\Z,\]
ao menos um \'e contido no intervalo $(a,b)$. Se n\~ao, podem escolher dois n\'umeros consecutivos, $\frac kn$ e $\frac{k+1}n$, tais que 
\[\frac k n \leq a,\mbox{ enquanto }\frac{k+1}n\geq b,\]
que n\~ao \'e poss\'\i vel gra\c cas a escolha de $n$: a quantidade $1/n$ \'e demasiado pequena para ``atrav\'essar a racha (o intervalo $[a,b]$) de um salto''! 
\end{exemplo}

\subsection{Espa\c cos separ\'aveis}
\begin{definicao}
Um espa\c co m\'etrico $X$ \'e dito {\em separ\'avel} se existe um subconjunto $A\subseteq X$ enumer\'avel e denso:
\[A=\{a_n\colon n=1,2,\ldots\},~~\bar A=X.\]
\index{espa\c co! m\'etrico! separ\'avel}
\end{definicao}

\begin{exemplo}
Cada espa\c co m\'etrico enumer\'avel tal que $\alpha\N$ ou $\Q$ \'e obviamente separ\'avel.
\end{exemplo}

\begin{exemplo}
A reta $\R$ \'e separ\'avel: o conjunto enumer\'avel $\Q$ dos n\'umeros racionais \'e denso em $\R$.  
\end{exemplo}

\begin{exemplo}
O espa\c co euclidiano $\R^n$ \'e separ\'avel: por exemplo, o conjunto $\Q^n$ que consiste de todos os vetores cujas coordenadas s\~ao racionais \'e enumer\'avel e denso. 
\end{exemplo}

\begin{exemplo}
\label{e:01sep}
Um espa\c co m\'etrico munido com a fun\c c\~ao dist\^ancia zero-um \'e separ\'avel se e somente se $X$ \'e enumer\'avel. (Exerc\'\i cio.)
\end{exemplo}

\begin{teorema}
Cada subespa\c co m\'etrico de um espa\c co m\'etrico separ\'avel \'e separ\'avel.
\label{t:subespsep}
\end{teorema}

\begin{proof}
Seja $X$ um espa\c co separ\'avel e seja $Y\subseteq X$ um subespa\c co. Escolhamos uma parte enumer\'avel e denso em $X$,
\[A=\{a_n\colon n=1,2,3,\ldots\}.\]
Seja $y_0\in Y$ um elemento qualquer. Para cada $m=1,2,3,\ldots$, escolhamos um ponto $b_{n,m}$, onde
\[\begin{cases}
b_{n,m}\in B_{1/m}(a_n)\cap Y,& \mbox{ se a bola $B_{1/m}(a_n)$ encontra $Y$,}
\\
b_{n,m}=y_0, & \mbox{ se n\~ao.}
\end{cases}
\]
O conjunto
\[B=\{b_{n,m}\colon n,m=1,2,3,\ldots\}\]
\'e contido em $Y$. Ele \'e evidentemente enumer\'avel. Basta verificar que $B$ \'e denso em $Y$. Sejam $y\in Y$ e $\e>0$. Existe $m$ tal que 
\[\frac 1m<\frac{\e}2.\]
Como $A$ \'e denso em $X$, existe $n$ tal que $a_n\in B_{1/m}(y)$. Por conseguinte, a bola $B_{1/m}(a_n)$ encontra $Y$ (ela cont\'em $y$, por exemplo). Conclu\'\i mos: o ponto $b_{n,m}\in B$ \'e contido em $B_{1/m}(a_n)$, e portanto
\[d(y,b_{n,m})\leq d(y,a_n)+d(a_n,b_{n,m})<\frac 1m+\frac 1m =\frac 2m <\e.\]
Isto significa que a interse\c c\~ao $B_\e(y)\cap B$ \'e n\~ao vazia.
\end{proof}

\begin{observacao}
O resultado acima (teorema \ref{t:subespsep}) n\~ao \'e verdadeiro para os espa\c cos topol\'ogicos mais gerais.
\end{observacao}

\begin{proposicao}
\label{p:pp}
O espa\c co m\'etrico $\ell^\infty$ cont\'em um subespa\c co m\'etrico $A$ de  cardinalidade do cont\'\i nuo e tal que a m\'etrica induzida sobre $A$ \'e uma m\'etrica zero-um.
\end{proposicao}

\begin{proof}
Cada sequ\^encia bin\'aria
\[\e_1\e_2\e_3\ldots,~~\e_i\in\{0,1\},\]
\'e obviamente limitada, logo pertence a $\ell^\infty$. Denotemos $A$ o conjunto de todas tais sequ\^encias. Sejam $a,b\in A$, $a\neq b$. Existe $n$ tal que
$a_n\neq b_n$, de onde
\[\abs{a_n-b_n}=1,\]
e por conseguinte
\[d_{\infty}(a,b)=\sup_{i\in\N}\abs{a_i-b_i}=1.\]
De acordo com Exemplo \ref{ex:dva}, a cardinalidade de $A$ \'e igual \`a cardinalidade de $\R$. 
\end{proof}

\begin{corollary}
O espa\c co m\'etrico $\ell^\infty$ n\~ao \'e separ\'avel.
\end{corollary}

\begin{proof}
Combinemos exemplo \ref{e:01sep} com teorema \ref{t:subespsep}.
\end{proof}

\begin{exercicio} 
Mostrar que cada sequ\^encia num espa\c co m\'etrico possui no m\'aximo um limite.
\end{exercicio}

\begin{teorema}
Seja $X$ um espa\c co m\'etrico separ\'avel. Ent\~ao
\[\abs X\leq {2^{\aleph_0}}.\]
\end{teorema}

\begin{proof}
Escolhamos um subespa\c co $Y$ denso e enumer\'avel. Denotemos $\mathscr C$ o conjunto de todas as sequ\^encias dos pontos de $Y$ que s\~ao convergentes em $X$. A aplica\c c\~ao que associa \`a sequ\^encia $(y_n)$ o seu \'unico limite \'e uma sobreje\c c\~ao de $\mathscr C$ sobre $X$. 
O conjunto $Y^{\N}$ de todas as sequ\^encias dos pontos de $Y$ (convergentes e n\~ao) cont\'em $\mathscr C$ como um subconjunto. Ent\~ao,
\[\abs X\leq \abs X\leq \abs{Y^\N}.\]
Basta mostrar que $\abs{Y^\N}\leq\abs{\R}$. Como o conjunto $Y$ \'e enumer\'avel, ele pode ser identificado com um subconjunto de $\N$, e a nossa tarefa \'e de mostrar que
\[\abs{\N^\N}\leq \abs\R.\]
Basta produzir uma inje\c c\~ao de $\N^\N$ em $2^{\N\times \N}$. Associemos a cada se sequ\^encia $(n_1,n_2,\ldots,n_k,\ldots)$ de n\'umeros naturais o seu gr\'afico: 
\[A = \{(n,a_n)\colon n\in\N\}.\]
\'E evidente que a correspond\^encia acima \'e injetora: cara fun\c c\~ao \'e unicamente determinada pelo seu gr\'afico.
\end{proof}

\subsection{Imagens diretas e imagens rec\'\i procas}
\begin{observacao} 
Sejam $f\colon X\to Y$ uma aplica\c c\~ao entre dois conjuntos quaisquer, $A\subseteq X$ e $B\subseteq Y$ subconjuntos quaisquer. A {\em imagem} de $A$ por $f$ \'e o conjunto
\[f(A) = \{f(a)\colon a\in A\}\subseteq Y,\]
e a {\em imagem inversa}, ou {\em rec\'\i proca,} de $B$ por $f$ \'e o conjunto
\[f^{-1}(B)=\{x\in X\colon f(x)\in B\}\subseteq X.\]
\end{observacao}

\begin{observacao}
A imagem inversa $f^{-1}(B)$ \'e o maior subconjunto $A$ de $X$ que possui a propriedade seguinte:
\[f(A)\subseteq B.\]
\end{observacao}

\subsection{Aplica\c c\~oes cont\'\i nuas em um ponto}

\begin{definicao}
Seja $f$ uma aplica\c c\~ao entre dois espa\c cos m\'etricos, $X$ e $Y$, e seja $x\in X$. Digamos que $f$ \'e {\em cont\'\i nua em} $x$ se
\[\forall\e>0,~\exists\delta>0,~~\forall y\in X,~~(d_X(x,y)<\delta)\Rightarrow (d_Y(f(x),f(y))<\e,\]
ou, da maneira equivalente e um pouco mais econ\^omica, 
\[\forall\e>0,~\exists\delta>0,~~f(B_\delta(x))\subseteq B_{\epsilon}(f(x)).\]
\end{definicao}

\begin{proposicao}[Criterio sequencial da continuidade] 
Uma aplica\c c\~ao $f\colon X\to Y$ \'e cont\'\i nua em um ponto $x\in X$ se e somente se para cada sequ\^encia $(x_n)$ de pontos de $X$ que converge para $x$, a sequ\^encia das imagens $(f(x_n))$ converge para $f(y)$.
\end{proposicao}

\begin{proof} $\Rightarrow$: suponhamos que $f$ \'e cont\'\i nua no ponto $x$, e seja $(x_n)$ uma sequ\^encia tal que $x_n\to x$ quando $n\to\infty$.
Seja $\e>0$ qualquer. Escolhamos $\delta>0$ da maneira que
\[f(B_\delta(x))\subseteq B_{\epsilon}(f(x)).\]
Como $(x_n)$ tenha $x$ com o limite, existe um n\'umero natural, $N$, tal que
\[\forall n\geq N,~~x_n\in B_\delta(x).\]
Conclu\'\i mos:
\[f(x_n)\to f(x)\mbox{ quand }n\to\infty.\]

$\Leftarrow$: por contraposi\c c\~ao. Suponhamos que $f$ \'e descont\'\i nua no ponto $x$. Logo,
\[\exists\epsilon>0,~~\forall\delta>0,~~\exists y\in B_\delta(x),~~f(y)\notin B_\epsilon(f(x)).\]
Fixemos um tal $\e>0$. Para cada $n=1,2,3\ldots$ encontremos um ponto $x_n\in X$ tal que
\[d_X(x_n,x)<\frac 1n\mbox{ e }f(x_n)\notin B_\epsilon(f(x)).\]
A sequ\^encia $(x_n)$ converge para $x$, enquanto a sequ\^encia $(f(x_n))$ n\~ao converge para $f(x)$.
\end{proof} 

\subsection{Aplica\c c\~oes cont\'\i nuas}

\begin{definicao} 
Uma aplica\c c\~ao $f$ entre dois espa\c cos m\'etricos $X$ e $Y$ \'e dita {\em cont\'\i nua} se $f$ \'e cont\'\i nua em cada ponto $x\in X$. Em outras palavras,
\[\forall x\in X,~~\forall\e>0,~\exists\delta>0,~~\forall y\in X,~~(d_X(x,y)<\delta)\Rightarrow (d_Y(f(x),f(y))<\e.\]
\end{definicao}

\begin{teorema}
Seja $f\colon X\to Y$ uma aplica\c c\~ao entre dois espa\c cos m\'etricos quaisquer. As condi\c c\~oes seguintes s\~ao equivalentes:
\begin{enumerate}
\item\label{equiv:1} $f$ \'e cont\'\i nua.
\item\label{equiv:2} Para cada $x\in X$ e cada vizinhan\c ca $V$ de $f(x)$, a imagem inversa $f^{-1}(V)$ \'e uma vizinhan\c ca de $x$. 
\item\label{equiv:3} A imagem inversa, $f^{-1}(V)$, de cada aberto, $V$, de $Y$ \'e aberta em $X$.
\item\label{equiv:4} A imagem inversa, $f^{-1}(F)$, de cada fechado, $F$, de $Y$ \'e fechada em $X$.
\item\label{equiv:5} Para cada parte $A\subseteq X$, temos
\[f(\overline{A})\subseteq\overline{f(A)}.\]
\end{enumerate}
\label{th:ouv}
\end{teorema}

\begin{proof}[Demonstra\c c\~ao do Teorema \ref{th:ouv} ``em sentido anti-hor\'ario'']
\par
(\ref{equiv:1}) $\Rightarrow$ (\ref{equiv:5}): Seja $x\in\bar A$. Ent\~ao $x$ \'e um ponto de ader\^encia de $A$, logo existe uma sequ\^encia $(a_n)$ dos pontos de $A$ convergente para $x$. A fun\c c\~ao $f$ \'e cont\'\i nua em ponto $x$, e
devido ao crit\'erio sequencial da continuidade, temos 
\[f(a_n)\to f(x).\]
Conclu\'\i mos: $f(x)$ pertence \`a ader\^encia de $f(A)$ pois $f(a_n)\in f(A)$.
\par
(\ref{equiv:5}) $\Rightarrow$ (\ref{equiv:4}): seja $G$ um subconjunto fechado de $Y$. Seja $B=f^{-1}(G)$. Ent\~ao $B$ \'e o maior subconjunto de $X$ tal que $f(B)\subseteq G$ (Observa\c c\~ao 3.3.2). Segundo (4), temos
\[f(\bar B)\subseteq \overline{f(B)}\subseteq \bar G = G,\]
logo $\bar B\subseteq B$,
e $B=f^{-1}(G)$ \'e fechado em $X$.
\par
(\ref{equiv:4}) $\Rightarrow$ (\ref{equiv:3}): se $V$ \'e um subconjunto aberto de $Y$, ent\~ao
\[f^{-1}(V)=f^{-1}(Y\setminus V) = X\setminus f^{-1}(Y\setminus V),\]
e o conjunto $f^{-1}(Y\setminus V)$ \'e fechado em $X$ gra\c cas \`a hip\'otese (4).
\par
(\ref{equiv:3}) $\Rightarrow$ (\ref{equiv:2}): Seja $V$ uma vizinhan\c ca de $f(x)$ em $Y$. Escolhamos $\e>0$ tal que $B_\e(f(x))\subseteq V$. O conjunto $f^{-1}(B_\e(f(x)))$ \'e aberto em $X$ e cont\'em $x$, logo \'e uma vizinhan\c ca de $x$ em $X$.
\par
(\ref{equiv:2}) $\Rightarrow$ (\ref{equiv:1}): A bola $B_\e(f(x))$ \'e uma vizinhan\c ca de $f(x)$ em $Y$, logo $f^{-1}(B_\e(f(x)))$ \'e uma vizinhan\c ca de $x$ em $X$. Existe $\delta>0$ com $B_\delta(x)\subseteq f^{-1}(B_\e(f(x)))$. Conclu\'\i mos:
\[f(B_\delta(x))\subseteq B_\e(f(x)),\]
o que implica a continuidade de $f$ em $x$.
\end{proof}

\subsection{Aplica\c c\~oes Lipschitz cont\'\i nuas}

\begin{definicao}
Uma fun\c c\~ao $f\colon X\to Y$ entre dois espa\c cos m\'etricos \'e dita {\em Lipschitz cont\'\i nua} se existe uma constante $L\geq 0$ ({\em constante de Lipschitz de} $f$) tal que
\[\forall x,y\in X~~ d_Y(f(x),f(y))\leq L\cdot d_X(x,y).\]
\index{fun\c c\~ao! Lipschitz cont\'\i nua}
\end{definicao}

\begin{observacao}
Cada aplica\c c\~ao Lipschitz cont\'\i nua \'e cont\'\i nua. 
(Exerc\'\i cio.)
\end{observacao}

\begin{exemplo}
A fun\c c\~ao cont\'\i nua
\[f(x)=x^2\]
de $\R$ a $\R$ n\~ao \'e Lipschitz cont\'\i nua.
Seja $L\geq 0$ qualquer. Seja $x>L$ e $y=x+1$. Temos
\[\abs{f(x)-f(y)} = y^2-x^2 = (x+1)^2-x^2 = 2x+1 > L.\]
Conclu\'\i mos: nenhum valor $L\in\R$ \'e uma constante de Lipschitz para $f$. 
\end{exemplo}

\begin{exemplo}
Sejam $X$ um espa\c co m\'etrico e $x_0\in X$. A {\em fun\c c\~ao dist\^ancia} de $x_0$, dada por
\[d_{x_0}(y) = d(x_0,y),\]
\'e Lipschitz cont\'\i nua, com a constante de Lipschitz $1$. (Exerc\'\i cio.)
\end{exemplo}

\begin{exercicio}
Mais geralmente, sejam $X$ um espa\c co m\'etrico e $A\subseteq X$ um subconjunto n\~ao vazio qualquer. A {\em fun\c c\~ao dist\^ancia} de $A$, notada $d_A$, \'e dada pela f\'ormula seguinte:
\[d_A(x) = \inf_{a\in A}d(x,a).\]
Mostrar que a fun\c c\~ao $d_A\colon X\to \R$ \'e Lipschitz cont\'\i nua, com a constante de Lipschitz $1$. (Para solu\c c\~ao, veja a prova do lema \ref{l:distanciadeA}.)
\end{exercicio}

\subsection{Topologia produto\label{ss:topologiadeproduto}}

\begin{definicao}
Dada uma fam\'\i lia de conjuntos $X_i$, $i\in I$, definamos o produto cartesiano como o conjunto de todas as fun\c c\~oes 
\begin{align*}
x\colon I&\to \cup_{i\in I}X_i,\\
i & \mapsto x_i
\end{align*}
tais que
\[\forall i\in I,~~x_i\in X_i.\]
Nota\c c\~ao do produto cartesiano: $\prod_{i\in I}X_i$.
\end{definicao}

\begin{definicao}
Dado $i\in I$, definamos a {\em $i$-\'esima proje\c c\~ao} do produto $\prod_{i\in I}X_i$ sobre $X_i$ pela formula
\[\pi_i(x) = x_i.\]
\end{definicao}

\begin{definicao}
Se $X_i$, $i\in I$ \'e uma fam\'\i lia enumer\'avel (finita ou infinita) de espa\c cos m\'etricos, $\abs I\leq\aleph_0$, digamos que uma m\'etrica $d$ sobre o produto cartesiano $\prod_{i\in I}X_i$ {\em induz a topologia produto} se qualquer sequ\^encia $(x_n)$ de elementos do produto converge para um elemento $x$ se e somente se, para todo $i\in I$,
\[\pi_i(x_n)\to \pi(x).\]
\label{d:topologiadeproduto}
\index{topologia! produto}
\end{definicao}

\begin{exercicio}
Mostre que a condi\c c\~ao acima \'e equivalente \`a condi\c c\~ao seguinte: 
a fam\'\i lia de conjuntos abertos do espa\c co m\'etrico $\left(\prod_{i\in I}X_i,d\right)$ \'e a menor topologia (fam\'\i lia de conjuntos satisfazendo as propriedades na proposi\c c\~ao \ref{p:topologiametrica}) tal que todas as proje\c c\~oes $\pi_i$, $i\in I$ s\~ao cont\'\i nuas.

Em outras palavras, a topologia produto \'e a menor topologia que cont\'em todos os conjuntos da forma
\[\pi_i^{-1}(V),~~i\in I,~~\mbox{$V$ \'e aberto em $X_i$.}\]
\label{ex:topprodmenorquepi}
\end{exercicio}

\begin{exercicio}
Sejam $(X_i,d_i)$, $i\in I$ espa\c cos m\'etricos, $\abs I\leq\aleph_0$. Mostrar que a m\'etrica no produto dada por
\[d(x,y) = \sum_{i\in I}\frac{1}{2^n}\min\{d_i(x_i,y_i),1\}\]
gera a topologia produto sobre $\prod_{i\in I}X_i$.
\end{exercicio}

\begin{exercicio}
Sejam $X_i$, $i\in \N$ espa\c cos m\'etricos finitos quaisquer (logo, discretos). Mostra que a m\'etrica no produto dada por
\[d(x,y) = \begin{cases}0,&\mbox{ se }x=y,\\
2^{-\min\{i\colon x_i\neq y_i\}},&\mbox{ se }x\neq y,
\end{cases}\]
gera a topologia produto sobre $\prod_{i\in \N}X_i$.
\end{exercicio}

\subsection{Imers\~oes isom\'etricas}
\begin{definicao}
Uma aplica\c c\~ao $i\colon X\to Y$ entre dois espa\c cos m\'etricos \'e dita {\em imers\~ao isom\'etrica} quando para quaisquer $x,y\in X$ tivermos
\[d_Y(i(x),i(y)) = d_X(x,y).\]
\end{definicao}

\begin{observacao}
Sejam $X$ um espa\c co m\'etrico qualquer e $Y\subseteq X$ um subespa\c co m\'etrico dele. Ent\~ao, como no exemplo precedente, a aplica\c c\~ao id\^entica
\[Y\ni y\mapsto i(y)=y\in X\]
\'e uma imers\~ao isom\'etrica de $Y$ em $X$. 
\end{observacao}

\subsection{Imers\~ao de Kuratowski\label{ss:imersaokuratowski}}

\begin{lema}
\label{l:distancia}
Se $X$ \'e um espa\c co m\'etrico qualquer e $x,y\in X$ dois pontos de $X$, ent\~ao
\[\sup_{z\in X}\abs{d_x(z)-d_y(z)}=d(x,y),\]
onde $d_x$ \'e a fun\c c\~ao dist\^ancia de $x$.
\end{lema}

\begin{proof}
A desigualdade $\leq$ \'e uma consequ\^encia da desigualdade triangular: para cada $z\in X$, temos
\[\abs{d_x(z)-d_y(z)}=\max\{d_x(z)-d_y(z),d_y(z)-d_x(z) \}\leq d(x,y).\]
Para mostrar a desigualdade $\geq$, notemos que a igualdade \'e atingida no ponto $z=x$:
\[\abs{d_x(x)-d_y(x)}=\abs{0-d(y,x)} = d(x,y).\]
\end{proof}

\begin{exemplo}
Seja $\Gamma$ um conjunto qualquer. Generalizemos o exemplo \ref{e:ellinfty}, e definamos o espa\c co m\'etrico $\ell^{\infty}(\Gamma)$ como o conjunto de todas as aplica\c c\~oes limitadas 
\[x\colon \Gamma\ni\gamma\mapsto x_{\gamma}\in\R,\]
munido da m\'etrica
\[d_{\infty}(x,y)=\sup_{\gamma\in\Gamma} \abs{x_{\gamma}-y_{\gamma}}.\]
Por exemplo, se $\Gamma=\N_+$, ent\~ao $\ell^{\infty}(\Gamma)=\ell^{\infty}$. 
\end{exemplo}

\begin{observacao}
Se $X=(X,d)$ \'e um espa\c co m\'etrico, vamos denotar o espa\c co $\ell^{\infty}(\Gamma)$ onde $\Gamma=X$ por $\ell^{\infty}(\abs X)$, para sublinhar que o conjunto de coordenadas do espa\c co \'e o conjunto subjacente do espa\c co $X$. 
\end{observacao}

\begin{definicao}
Seja $A$ um subconjunto de um espa\c co m\'etrico $X$ qualquer. 
O {\em di\^ametro} de $A$ \'e dado pela formula
\[{\mathrm{diam}}\,(A) =
\inf \{r\in [0,+\infty]\colon \forall a,b\in A,~~d(a,b)\leq r\}.
\]
\end{definicao}

\begin{exercicio} 
Mostrar que para cada subconjunto $A$ {\em n\~ao vazio} temos
\[{\mathrm{diam}}\,(A) = \sup_{a,b\in A}d(a,b).\]
\end{exercicio}

\begin{exercicio}
\label{exer:limitado}
Mostrar que para um espa\c co m\'etrico $X$ as defini\c c\~oes seguintes s\~ao equivalentes:
\begin{enumerate}
\item $X$ \'e limitado, ou seja, o di\^ametro de $X$ \'e finito.
\item Para cada $x\in X$, a fun\c c\~ao dist\^ancia $d_x\colon X\to\R$ \'e limitada.
\item Existe $x\in X$ tal que a fun\c c\~ao dist\^ancia $d_x\colon X\to\R$ \'e limitada.
\end{enumerate}
\end{exercicio}

\begin{definicao}
\label{d:limitado}
Um subconjunto $A$ de um espa\c co m\'etrico $X$ \'e dito {\em limitado} se $A$ satisfaz uma das condi\c c\~oes equivalentes de exerc\'\i cio \ref{exer:limitado}.
\end{definicao}

\begin{proposicao}
Seja $X$ um espa\c co m\'etrico limitado. Ent\~ao a f\'ormula seguinte define uma imers\~ao isom\'etrica de $X$ em $\ell^{\infty}(\abs X)$:
\[X\ni x\overset{i}\mapsto i(x) = d_x\in \ell^{\infty}(\abs X).\]
(Essa imers\~ao \'e chamada a {\em imers\~ao de Kuratowski}.)
\end{proposicao}

\begin{proof}
Gra\c cas ao exerc\'\i cio \ref{exer:limitado}, cada fun\c c\~ao $d_x$ pertence a $\ell^{\infty}(\abs X)$. E gra\c cas ao lema \ref{l:distancia}, temos, para quaisquer que sejam $x,y\in X$,
\begin{align*}
d_{\infty}(i(x),i(y)) &=
 \sup_{z\in X}\abs{d_x(z)-d_y(z)}\\
&= d_X(x,y).
\end{align*}
\end{proof}

\begin{observacao}
Em caso geral, a fun\c c\~ao $d_x$ n\~ao \'e limitada, e a constru\c c\~ao acima deve ser modificada.
\end{observacao}

\begin{teorema}
\label{th:geral}
Seja $X$ um espa\c co m\'etrico qualquer. Escolhamos um ponto $x_0\in X$.
A f\'ormula seguinte define uma imers\~ao isom\'etrica de $X$ em $\ell^{\infty}(\abs X)$:
\[X\ni x\overset{i}\mapsto i(x) = d_x-d_{x_0}\in \ell^{\infty}(\abs X).\]
(Esta imers\~ao \'e tamb\'em uma vers\~ao da imers\~ao de Kuratowski.)
\end{teorema}

\begin{proof}
Seja $x\in X$. A fun\c c\~ao $i(x) = d_x-d_{x_0}$ \'e limitada, gra\c cas \`a desigualdade triangular: qualquer que seja $z\in M$, temos
\[\abs{d_x(z)-d_{x_0}(z)}\leq d(x,x_0).\]
Agora, sejam $x,y\in X$ quaisquer.
Temos:
\begin{align*}
d_{\infty}(i(x),i(y)) &= \sup_{z\in X} \abs{(d_x(z)-d_{x_0}(z))-(d_y(z)-d_{x_0}(z))} \\
&= \sup_{z\in X}\abs{d_x(z)-d_y(z)} \\
&= d(x,y)
\end{align*}
(lema \ref{l:distancia}). 
\end{proof}

\begin{lema}
\label{l:sup}
Sejam $f$ uma fun\c c\~ao real limitada sobre um espa\c co m\'etrico $X$ e $Y$ um subespa\c co denso de $X$. Ent\~ao temos
\[\sup_{x\in X} f(x) = \sup_{y\in Y}f(y).\]
\end{lema}

\begin{proof}
A desigualdade $\geq$ \'e \'obvia. Denotemos
$b = \sup_{x\in X} f(x)$.
Seja $\e>0$. Existe $x\in X$ tal que 
$f(x)>b-\frac{\e}2$.
A fun\c c\~ao $f$ \'e cont\'\i nua em $x$, logo existe $\delta>0$ tal que, se $d(x,y)<\delta$, temos $\abs{f(x)-f(y)}<\e/2$. 
Como $Y$ \'e denso em $X$, existe $y\in Y\cap B_{\delta}(x)$. Por conseguinte,
\[f(y)>f(x)-\frac{\e}2 > b-\e,\]
o que implica a desigualdade $\leq$.
\end{proof}

Isto nos leva a mais uma vers\~ao da imers\~ao de Kuratowski.

\begin{teorema}
Sejam $X$ um espa\c co m\'etrico e $Y$ um subespa\c co denso de $X$. Escolhamos um ponto $x_0\in X$ qualquer.
A f\'ormula seguinte define uma imers\~ao isom\'etrica de $X$ em $\ell^{\infty}(\abs Y)$:
\[X\ni x\overset{i}\mapsto i(x) = \left(d_x-d_{x_0}\right)\vert_Y\in \ell^{\infty}(\abs Y).\]
\label{t:kuratowskiY}
\end{teorema}

\section{Espa\c cos completos, compactos, e pr\'e-compactos}

\subsection{Sequ\^encias de Cauchy}

\begin{definicao}
Uma sequ\^encia de elementos de um espa\c co m\'etrico $X$ \'e chamada {\em sequ\^encia de Cauchy} se para cada $\e>0$ existe um $N$ tal que
\[d(x_m,x_n)<\e\]
quaisquer que sejam $m,n\geq N$.
\end{definicao}

\begin{observacao} 
Cada sequ\^encia convergente \'e uma sequ\^encia de Cauchy.
\end{observacao}

A rec\'iproca \'e falsa em geral. 

\begin{exemplo} 
Eis um exemplo de uma sequ\^encia de Cauchy no espa\c co m\'etrico $\Q$ que n\~ao converge neste espa\c co:
\[0,~0.1,~0.101,~0.101001,~0.1010010001, \dots .\]
\end{exemplo}

\begin{lema}
Uma sequ\^encia $(x_n)$ \'e uma sequ\^encia de Cauchy se e somente se 
\[\forall \e>0~~\exists N~~\forall n\geq N~~x_n\in B_\e(x_N).\]
\label{l:defeqseqcauchy}
\end{lema}

\begin{proof}
Exerc\'\i cio.
\end{proof}

\subsection{Espa\c cos m\'etricos completos}

\begin{definicao}
Um espa\c co m\'etrico $X$ \'e dito {\em completo} quando cada sequ\^encia de Cauchy tiver um limite (no mesmo espa\c co $X$).
\index{espa\c co~ m\'etrico! completo}
\end{definicao}

\begin{exemplo} 
O espa\c co m\'etrico $(X,d)$ munido de uma dist\^ancia zero-um $d=d_{0-1}$ \'e completo. Toda sequ\^encia de Cauchy $(x_n)$ em $X$ \'e {\em eventualmente  constante,} ou seja, existem $x\in X$ e $N\in\N$ tais que para cada $n\geq N$ temos $x_n=c$.
\end{exemplo}

\begin{lema}
Seja $X$ em espa\c co m\'etrico. As condi\c c\~oes seguintes s\~ao equivalentes:
\begin{enumerate}
\item 
$X$ \'e completo. 
\item 
A interse\c c\~ao da cada sequ\^encia das bolas fechadas encaixados de $X$, cuja o raio converge para zero, \'e n\~ao vazia.
\end{enumerate}
\label{l:encaixados}
\end{lema} 

\begin{proof}
(1) $\Rightarrow$ (2): esta dire\c c\~ao \'e praticamente evidente por que os centros das bolas formam uma sequ\^encia de Cauchy. Mais exatamente, seja $(\bar B_{\e_n}(x_n))_{n=1}^\infty$ uma sequ\^encia qualquer das bolas fechadas encaixados em $X$ cujas raios convergem para zero. Dado $\e>0$, existe $N$ tal que $\e_n\leq \e/2$ quando $n\geq N$. Se $m,n\geq N$, ent\~ao $x_m,x_k\in \bar B_{\e_N}(x_N)$, e assim ter\'\i amos
\[d(x_m,x_n)\leq d(x_m,x_N)+d(x_k,x_N)<2\e_n\leq \e.\]
A sequ\^encia dos centros \'e uma sequ\^encia de Cauchy, e como $X$ \'e completo, existe um limite
\[x=\lim_{n\to\infty}x_n.\]
Esto $x$ \'e adherente a cada bola fechada $\bar B_{\e_n}(x_n)$, logo $x\in \bar B_{\e_n}(x_n)$ e por conseguinte,
\[x\in\cap_{n=1}^\infty\bar B_{\e_n}(x_n).\]
(2) $\Rightarrow$ (1): 
seja $(x_n)$ uma sequ\^encia de Cauchy em $X$. Para todo $k=1,2,3,\ldots$ escolhamos $N=N(k)$ de modo que
\[\forall m,n\geq N(k),~~d(x_m,x_n)<2^{-k-1}.\]
Se $k\leq m$, ent\~ao para cada $y\in\bar B_{2^{-m}}(x_{N(m)})$ temos
\[d(x_k,y)\leq d(x_k,x_m)+d(x_m,y)\leq 2^{-k-1}+2^{-m}\leq 
2^{-k-1}+2^{-k-1} = 2^{-k},\]
logo $y\in \bar B_{2^k}(x_{N(k)})$. Por conseguinte, 
\[\bar B_{2^{-m}}(x_{N(m)})\subseteq \bar B_{2^{-k}}(x_{N(k)}).\]
Conclu\'\i mos: as bolas $\bar B_{2^{-k}}(x_{N(k)})$ formam uma sequ\^encia das bolas fechadas encaixados, cujas raios convergem para zero. De acordo com a hip\'otese, a interse\c c\~ao das bolas \'e n\~ao vazia, e como os raios convergem para zero, a interse\c c\~ao consiste de um ponto s\'o, digamos $x$:
\[\cap_{k=1}^\infty \bar B_{2^{-k}}(x_{N(k)}) =\{x\}.\]
Para cada $k$ temos
\[d(x,x_{N(k)})\leq 2^{-k-1}.\]
Concluimos que $x$ \'e o limite da sequ\^encia $(x_{N(k)})_{k=1}^\infty$. 
\end{proof}

\begin{exemplo} Para o espa\c co $\R$ munido da dist\^ancia usual a propriedade (2) do Lema \ref{l:encaixados} \'e exatamente a propriedade dos intervalos fechados encaixados (Teorema de Cantor). Conclu\'\i mos que $\R$ \'e completo como um espa\c co m\'etrico. 
\end{exemplo}

\begin{exemplo}
O crit\'erio acima (lema \ref{l:encaixados}) acima mostra mais ou menos imediatamente a completude do espa\c co normado $\ell^\infty (\Gamma) $ de todas as fun\c c\~oes reais limitadas sobre um conjunto $\Gamma$ qualquer, munido da norma uniforme
\[\norm f=\sup_{\gamma\in \Gamma}\abs{f(\gamma)}\]
e a dist\^ancia correspondente,
\[d(f,g)=\norm{f-g}.\]
(Exerc\'\i cio.)
\label{t:gamma}
\end{exemplo}

\begin{teorema}
Seja $Y$ um subespa\c co m\'etrico de um espa\c co m\'etrico $X$. Se $Y$ \'e completo, ent\~ao ele \'e fechado em $X$.
\end{teorema}

\begin{proof}
Seja $x$ um ponto aderente qualquer de $Y$ em $X$:
\[x\in\bar Y^X.\]
Existe uma sequ\^encia $(y_n)$ de pontos de $Y$ convergente para $x$ em $X$. Por conseguinte, $(y_n)$ \'e uma sequ\^encia de Cauchy em $X$, logo ela resta uma sequ\^encia de Cauchy no subespa\c co $Y$, e possui o limite, $y$, em $Y$, porque $Y$ \'e completo. Certamente, $y_n\to y$ em $X$ tamb\'em. A unicidade do limite implica que $x=y$, e portanto
\[x=y\in Y.\]
\end{proof}

\begin{teorema}
\label{t:fechado}
Cada subespa\c co fechado de um espa\c co m\'etrico completo \'e completo. 
\end{teorema}

\begin{proof}
Exerc\'\i cio.
\end{proof}

\begin{exemplo}
Um exemplo importante \'e o espa\c co $C[0,1]$ de todas as fun\c c\~oes cont\'\i nuas reais sobre o intervalo fechado unit\'ario, munido da norma uniforme
\[\norm f=\max_{x\in [0,1]}\abs{f(x)}.\]
A maneira mais f\'acil de verificar a completude de $C[0,1]$ \'e perceber que este espa\c co \'e fechado no espa\c co m\'etrico ambiente $\ell^\infty ([0,1])$. (Exerc\'\i cio. Veja uma prova um pouco mais tarde, teorema \ref{th:complet}.)
\end{exemplo}

\begin{exercicio}
Deduzir do lema \ref{l:encaixados} o resultado seguinte. Um espa\c co m\'etrico $X$ \'e completo se e somente se cada sequ\^encia de {\em conjuntos fechados} encaixados, cujos di\^ametros convergem para zero, tem uma interse\c c\~ao n\~ao vazia.
\label{ex:fechadosencaixados}
\end{exercicio}

\subsection{Completamento de um espa\c co m\'etrico: exist\^encia}

\begin{definicao}
Seja $X$ um espa\c co m\'etrico qualquer. O {\em completamento} de $X$ \'e um par $(\hat X,i)$, que consiste de um espa\c co m\'etrico completo $\hat X$ e uma imers\~ao isom\'etrica $i\colon X\hookrightarrow\hat X$, onde $i(X)$ \'e denso em $\hat X$.
\end{definicao}

\begin{exemplo}
O completamento do espa\c co $\Q$ munido da dist\^ancia usual \'e (isom\'etrico ao) espa\c co m\'etrico $\R$. 
\end{exemplo} 

\begin{teorema}
Cada espa\c co m\'etrico, $X$, possui um completamento.
\label{t:complecao}
\end{teorema}

\begin{proof}
De acordo com Teorema \ref{th:geral}, existe uma imers\~ao isom\'etrica, $i$, de $X$ num espa\c co m\'etrico $\ell^{\infty}(\abs X)$. Definamos
\[\hat X = \overline{i(X)}.\]
O espa\c co $\ell^{\infty}(\abs X)$ \'e completo (exerc\'\i cio \ref{t:gamma}), e como o subespa\c co $\hat X$ \'e fechado em $\ell^{\infty}(X)$, ele \'e completo tamb\'em (teorema \ref{t:fechado}). A imagem $i(X)$ \'e densa em $\hat X$.
\end{proof}

\subsection{Espa\c cos isom\'etricos}

\begin{definicao}
Uma {\em isometria} entre dois espa\c cos m\'etricos $X$ e $Y$ e uma imers\~ao isom\'etrica sobrejetiva do $Y$ em $X$. 
\end{definicao}

\begin{exemplo}
A fun\c c\~ao 
\[\R\ni x\mapsto i(x)=x+1\in\R\]
\'e uma isometria da reta com si pr\'opria.
\end{exemplo}

\begin{definicao}
Dois espa\c cos m\'etricos $X$ e $Y$ s\~ao chamados {\em isom\'etricos} se existe uma isometria
\[i\colon X\to Y.\]
Nota\c c\~ao: $X\cong Y$. 
\end{definicao}

\begin{exercicio}
A rela\c c\~ao da isometria entre espa\c cos m\'etricos \'e uma rela\c c\~ao da equival\^encia, ou seja,
\begin{enumerate}
\item Reflexividade: cada espa\c co $X$ \'e isom\'etrico a si mesmo: $X\cong X$.
\item Simetria: se $X\cong Y$, ent\~ao $Y\cong X$.
\item Transitividade: se $X\cong Y$ e $Y\cong Z$, ent\~ao $X\cong Z$.
\end{enumerate}
\end{exercicio}

\subsection{Completamento de um espa\c co m\'etrico: unicidade}

\begin{teorema}
O completamento $\hat X$ de um espa\c co m\'etrico $X$ \'e \'unico a menos de isometria. Sejam $(\hat X,i)$ e $(\tilde X,j)$ dois completamentos de $X$. Ent\~ao existe uma isometria $\kappa\colon \hat X\to \tilde X$ tal que \[\kappa\circ i=j.\]
\label{t:unica}
\end{teorema}

\begin{proof}
Escolhamos um ponto $x_0\in X$ qualquer.
Segundo Teorema \ref{t:kuratowskiY}, a f\'ormula
\[\hat X\ni x\overset{i^\prime}\mapsto i^\prime(x) = \left(d_{i(x)}-d_{i(x_0)}\right)\circ i\in \ell^{\infty}(\abs X)\]
define uma imers\~ao isom\'etrica
\[i^{\prime}\colon \hat X\hookrightarrow \ell^{\infty}(\abs X),\]
e a f\'ormula semelhante
\[\tilde X\ni x\overset{j^\prime}\mapsto j^\prime(x) = \left(d_{j(x)}-d_{j(x_0)}\right)\circ j\in \ell^{\infty}(\abs X)\]
define uma imers\~ao isom\'etrica
\[j^{\prime}\colon \tilde X\hookrightarrow \ell^{\infty}(\abs X).\]
Temos
\[i^\prime\circ i = j^\prime\circ j,\]
e por essa raz\~ao as imagens de $\hat X$ para $i^\prime$ e de $\tilde X$ para $j^\prime$ s\~ao iguais (cada uma delas \'e o fecho em $\ell^\infty(\abs X)$ do conjunto $i^\prime(i(X)) = j^\prime(j(X))$). 

As aplica\c c\~oes 
\[i^\prime\colon \hat X\to i^\prime(\hat X)\mbox{ e }
j^\prime\colon \tilde X\to j^\prime(\tilde X)\]
s\~ao isometrias sobre as suas imagens, logo possuem as aplica\c c\~oes rec\'\i procas. Definamos
\[\kappa = \left(j^\prime\right)^{-1}\circ i^\prime.\]
\'E uma isometria entre $\hat X$ e $\tilde X$. Al\'em disso, temos
\begin{align*}
\kappa\circ i &=
 \left(j^\prime\right)^{-1}\circ i^\prime\circ i \\
&= \left(j^\prime\right)^{-1}\circ j^\prime\circ j \\
&= j.
\end{align*}
\end{proof}

\subsection{Teorema de Weierstrass sobre a converg\^encia uniforme}

\begin{definicao}
Seja $X$ um espa\c co m\'etrico qualquer. Denotemos $CB(X)$ o conjunto de todas as fun\c c\~oes reais limitados e cont\'\i nuas sobre $X$, munido da dist\^ancia induzida de $\ell^\infty(X)$.
\end{definicao}

\begin{teorema} 
Seja $X$ um espa\c co m\'etrico qualquer. O espa\c co m\'etrico $CB(X)$ \'e completo.
\label{th:complet}
\end{teorema}

\begin{proof}
Mostremos que $CB(X)$ \'e fechado em $\ell^\infty(X)$, ou seja, que o complemento de $CB(X)$ em $\ell^{\infty}(\Gamma)$ \'e aberto. Seja $f\in CB(X)^c=\ell^\infty(X)\setminus CB(X)$ uma fun\c c\~ao qualquer. Como $f$ \'e descont\'\i nua, existem  $x_0\in CB(X)$ e $\e_0>0$ tais que
\[\forall\delta>0~~\exists y\in K~~d(x_0,y)<\delta~~\wedge~~
\abs{f(x_0)-f(y)}\geq\e_0.\]
Mostremos que a bola aberta $B_{\e_0/3}(g)$ n\~ao encontra $CB(X)$. Sejam
$g\in B_{\e_0/3}(x_0)$ e $\delta>0$ quaisquer. Escolhamos $y\in K$ como acima. Agora estabele\c camos a desigualdade 
\begin{equation}
\label{eq:claim}
\abs{g(x_0)-g(y)}\geq\frac{\e_0}3,\end{equation}
de onde a descontinuidade de $g$ seguir. 

Para obter uma contradi\c c\~ao, suponhamos que (\ref{eq:claim}) n\~ao \'e verdadeiro. Temos
\[\abs{g(x_0)-g(y)}<\frac{\e_0}3,\]
e, seguindo a desigualdade triangular,
\begin{align*}
\abs{f(x_0)-f(y)}&= 
\abs{f(x_0)-g(x_0)+g(x_0)-g(y)+g(y)-f(y)}
\\ &\leq
\abs{f(x_0)-g(x_0)}+\abs{g(x_0)-g(y)}+\abs{g(y)-f(y)}\\
&< \frac{\e_0}3+\frac{\e_0}3+\frac{\e_0}3\\
&= \e,
\end{align*}
uma contradi\c c\~ao.
\end{proof}

\begin{definicao}
Seja $(f_n)$ uma sequ\^encia das fun\c c\~oes de um conjunto $X$ para $\R$. Diz-se que a sequ\^encia $(f_n)$ {\em converge uniformemente} para uma fun\c c\~ao  $f\colon X\to\R$ se
\[d_{\infty}(f_n,f)\to o\mbox{ quando }n\to\infty,\]
onde $d_{\infty}$ \'e a dist\^ancia no espa\c co $\ell^\infty(X)$.
Nota\c c\~ao:
\[f_n\rightrightarrows f.\]
\end{definicao}

Eis uma reformula\c c\~ao equivalente do teorema \ref{th:complet}.

\begin{teorema}[Teorema de Weierstrass sobre a converg\^encia uniforme]
O limite uniforme de uma sequ\^encia de fun\c c\~oes cont\'
\i nuas \'e cont\'\i nua. \qed
\end{teorema}

\subsection{Espa\c cos compactos}

\begin{definicao}
Um subconjunto $K$ de um espa\c co m\'etrico $X$ \'e dito {\em compacto} se cada sequ\^encia dos pontos de $K$ cont\'em uma subsequ\^encia que converge e tem o seu limite em $K$. 
\index{espa\c co! m\'etrico! compacto}
\end{definicao}

\begin{exemplo} $\R$ n\~ao \'e compacto: a sequ\^encia
\[1,2,3,4,\ldots, n,\ldots\]
n\~ao cont\'em nenhuma subsequ\^encia convergente.
\end{exemplo}

\begin{exemplo} 
Cada espa\c co m\'etrico finito, $X$, \'e compacto. 
\end{exemplo}

\begin{exercicio}
Mostre que o produto de uma fam\'\i lia enumer\'avel de espa\c cos m\'etricos compactos, munido de uma m\'etrica que gera a topologia de produto, \'e compacto.
\label{ex:produtocompacto}
\end{exercicio}

\begin{teorema}
Seja $K$ um subconjunto compacto de um espa\c co m\'etrico $X$ qualquer. Ent\~ao $K$ \'e fechado em $X$.
\label{th:ferme}
\end{teorema}

\begin{proof}
Seja $x\in X$ um ponto aderente a $K$. Existe uma sequ\^encia $(x_n)$ dos elementos de $K$ que converge para $x$:
\[x_n\to x.\]
Como $K$ \'e compacto, a sequ\^encia $(x_n)$ tem uma subsequ\^encia convergente $(x_{n_k})_{k=1}^\infty$ cujo limite \'e contido em $K$:
\[\exists \varkappa\in K~~x_{n_k}\to\varkappa\mbox{ quando }k\to\infty.\]
Ao mesmo tempo, a subsequ\^encia certamente converge para $x$, todo como a sequ\^encia original:
\[x_{n_k}\to x\mbox{ quando }k\to\infty.\]
Conclu\'\i mos:
$x=\varkappa\in K$.
\end{proof}

\begin{teorema} 
Cada espa\c co m\'etrico compacto \'e completo.
\label{t:compactocompleto}
\end{teorema}

\begin{proof}
Seja $(x_n)$ uma sequ\^encia de Cauchy no espa\c co m\'etrico compacto $K$. Existem uma subsequ\^encia $(x_{n_k})$ convergente para um limite $x\in K$. Verifica-se facilmente que com efeito
\[\lim_{n\to\infty}x_n=x.\]
\end{proof}

\begin{teorema} Cada subconjunto fechado de um espa\c co m\'etrico compacto \'e compacto.
\label{th:compact}
\end{teorema}

\begin{proof}
Exerc\'\i cio.
\end{proof}

\subsection{Teoremas de Weierstrass}

\begin{teorema} A imagem de um conjunto compacto por uma aplica\c c\~ao cont\'\i nua \'e compacta. 
\label{th:image}
\end{teorema}

\begin{proof}
Sejam $X$ e $Y$ dois espa\c cos m\'etricos, $K\subseteq X$ um subconjunto compacto, e $f\colon X\to Y$ uma aplica\c c\~ao cont\'\i nua.
Seja $(y_n)$ uma sequ\^encia qualquer dos pontos de $f(K)$. Para cada $n$ escolhamos $x_n\in K$ tal que $f(x_n)=y_n$. Como $K$ \'e compacto, existe uma subsequ\^encia $(x_{n_k})_{k=1}^\infty$ convergente em $K$ para um limite $\varkappa\in K$. Como a aplica\c c\~ao $f$ \'e cont\'\i nua, conclu\'\i mos
\[y_n=f(x_n)\to f(\varkappa)\in f(K)\mbox{ quando }n\to\infty.\]
\end{proof}

Eis um corol\'ario importante que, pelas raz\~oes hist\'oricas, habitualmente \'e cortado em duas partes de maneira um pouco grotesca. 

\begin{teorema}[Primeiro teorema de Weierstrass]
Cada fun\c c\~ao real cont\'\i nua sobre um espa\c co m\'etrico compacto \'e limitada...
\end{teorema}

\begin{teorema}[Segundo teorema de Weierstrass]
...e atinge os seus limites.
\end{teorema}

\begin{proof}
Seja $K$ um espa\c co m\'etrico compacto (em especial, n\~ao vazio), e seja $f\colon K\to\R$ uma fun\c c\~ao cont\'\i nua. A imagem $f(K)$ \'e compacto gra\c ca ao teorema \ref{th:image}, logo limitado em $\R$ (teorema de Heine-Borel \ref{t:heine-borel}). Como $f(K)$ n\~ao \'e vazio, o supremo $b=\sup f(K)$ existe e pertence \`a $\R$. Evidentemente, $b$ \'e um ponto aderente de $f(K)$. Como $f(K)$ \'e fechado em $\R$, temos $b\in f(K)$, de onde conclu\'\i mos.
\end{proof}

\subsection{Aplica\c c\~oes fechadas}

\begin{definicao}
Uma aplica\c c\~ao cont\'\i nua $f\colon X\to Y$ entre dois espa\c cos m\'etricos quaisquer $X$ e $Y$ \'e dita {\em fechada} se para cada parte fechada $G$ de $X$ a sua imagem, $f(G)$, por $f$ \'e fechada em $Y$.
\end{definicao}

\begin{exemplo}
Sejam $X=\R^2$ munido da dist\^ancia usual, $Y=\R$, $G\subseteq \R^2$ o gr\'afico da fun\c c\~ao $y=1/x$, ou seja,
\[G=\left\{\left(x,\frac 1x\right)\colon x\in\R\setminus\{0\}\right\},\]
e $p\colon\R^2\to \R$ a proje\c c\~ao:
\[p(x,y)=x.\]
Verifica-se facilmente que a aplica\c c\~ao $p$ \'e cont\'\i nua e que o conjunto $G$ \'e fechado em $\R^2$, mas ao mesmo tempo a imagem n\~ao \'e:
\[p(G)=\R\setminus\{0\}\]
\end{exemplo}

\begin{proposicao}
Cada aplica\c c\~ao cont\'\i nua de um espa\c co m\'etrico compacto $K$ para um espa\c co m\'etrico $Y$ qualquer \'e fechada. 
\label{p:fermee}
\end{proposicao}

\begin{proof}
Seja $F\subseteq K$ uma parte fechada qualquer. Logo $F$ \'e compacto, e a sua imagem $f(F)$ pela aplica\c c\~ao cont\'\i nua $f$ e compacto em $Y$. Cada subconjunto compacto de um espa\c co m\'etrico \'e fechado. 
\end{proof}

\begin{corolario}
\label{c:homeo}
Seja $f\colon K\to Y$ uma aplica\c c\~ao cont\'\i nua e bijetiva entre dois espa\c cos m\'etricos, onde $K$ \'e compacto. Ent\~ao a aplica\c c\~ao inversa $f^{-1}$ \'e cont\'\i nua. (Como diz-se, $f$ \'e um {\em homeomorfismo.})
\end{corolario}

\begin{proof}
A fim de mostrar que a aplica\c c\~ao $g=f^{-1}\colon Y\to K$ \'e cont\'\i nua, seja $G\subseteq K$ uma parte fechada qualquer. Temos
\[g^{-1}(G) = f(G),\]
e $f(G)$ \'e fechado em $Y$ porque a aplica\c c\~ao $f$ \'e fechada segundo a proposi\c c\~ao \ref{p:fermee}. Conclu\'\i mos: a imagem inversa de cada parte fechada de $K$ por $g$ \'e fechada em $Y$, e portanto $g=f^{-1}$ \'e cont\'\i nua.
\end{proof}

\subsection{Espa\c cos pr\'e-compactos\label{ss:espacospre-compactos}}

\begin{definicao}
Um subconjunto $A$ de um espa\c co m\'etrico $X$ \'e dito {\em pr\'e-compacto,} ou {\em totalmente limitado,} se para cada valor $r>0$ existe uma cobertura finita de $A$ pelas bolas abertas de raio $r$:
\[\exists n,~\exists x_1,x_2,\ldots,x_n\in X,\mbox{ tais que }A\subseteq \cup_{i=1}^n B_r(x_i).\]
Em outras palavras, para cada $r>0$ existe uma cole\c c\~ao finita $x_1,x_2,\ldots,x_n$ dos pontos de $X$ tal que cada $a\in A$ \'e a dist\^ancia  $<r$ de algum deles:
\[\forall a\in A,~~\exists i=1,2,3,\ldots,n\mbox{ tal que } d(a,x_i)<r.\]
\index{espa\c co~ m\'etrico! pr\'e-compacto}
\end{definicao}

\begin{exemplo}
\label{ex:01}
Um subconjunto $A$ de $\R$ \'e pr\'e-compacto se e somente se $A$ \'e limitado (contido num intervalo finito).
\end{exemplo}

\begin{definicao}
Um espa\c co m\'etrico $X$ \'e dito {\em pr\'e-compacto} se $X$ \'e um subconjunto pr\'e-compacto de si mesmo.
\end{definicao}

Com efeito, a no\c c\~ao da pr\'e-compacidade de um subespa\c co m\'etrico $A$ \'e ``absoluta'', ou seja, n\~ao depende de um espa\c co ambiental, $X$.

\begin{proposicao} 
Seja $A$ um subconjunto pr\'e-compacto n\~ao vazio de um espa\c co m\'etrico $X$, ou seja, para cada $r>0$ existe uma cobertura finita de $A$ pelas bolas abertas de raio $r$ centradas em $X$. Ent\~ao, $A$ \'e pr\'e-compacto em si mesmo, ou seja, para cada $r>0$ existe uma cobertura finita de $A$ pelas bolas abertas de raio $r$ centradas em $A$.
\label{p:absoluta} 
\end{proposicao}

\begin{proof} 
Seja $r>0$. Podemos escolher uma cole\c c\~ao finita dos pontos $x_1,x_2,\ldots,x_n\in X$ tal que para cada $a\in A$ existe $i$ com a propriedade
\[d(x_i,a)<r/2.\]
Sem perda da generalidade, podemos supor que cada bola aberta $B_{r/2}(x_i)$  encontra $A$. Para cada $i$, escolhamos um ponto
\[a_i\in B_{r/2}(x_i)\cap A.\]
Seja $a\in A$ um ponto qualquer. Existe $i$ tal que $d(a,x_i)<r/2$. Temos finalmente, gra\c ca \`a desigualdade triangular:
\[d(a,a_i)\leq d(a,x_i)+d(x_i,a_i)<\frac r2+\frac r2=r.\]
Portanto, $A$ \'e contido na uni\~ao das bolas abertas $B_r(a_i)$, $i=1,2,\ldots,n$, onde as bolas podem ser consideradas em $A$ em vez de em $X$. 
\end{proof}

\begin{corolario} 
Um subespa\c co m\'etrico de um espa\c co m\'etrico pr\'e-compacto \'e pr\'e-compacto. \qed
\label{c:imm}
\end{corolario}

\subsection{Compacidade e pr\'e-compacidade}

\begin{teorema} 
Cada conjunto compacto \'e pr\'e-compacto.
\label{th:comp}
\end{teorema}

\begin{proof} Vamos mostrar a contraposi\c c\~ao: se $X$ n\~ao \'e pr\'e-compacto, ent\~ao $X$ n\~ao \'e compacto. 

Suponhamos que $X$ \'e um subconjunto n\~ao pr\'e-compacto de um espa\c co m\'etrico. A nega\c c\~ao da pr\'e-compacidade significa que existe $r>0$ com a propriedade que para cada cole\c c\~ao finita $x_1,x_2,\ldots,x_n\in X$, existe $x_{n+1}\in X$ tal que
\[d(x_i,x_{n+1})\geq r,~~i=1,2,\ldots,n.\]
Utilizando esta propriedade, escolhamos pela recorr\^encia uma sequ\^encia infinita dos pontos de $X$, 
\[x_1,x_2,\ldots,x_n,\ldots,\]
de modo que 
\[\forall i,j,~~(i\neq j)\Rightarrow d(x_i,x_j)\geq r.\]
Evidentemente, cada subsequ\^encia de $(x_n)$ tem a mesma propriedade, logo n\~ao \'e convergente. Conclu\'\i mos: o conjunto $X$ n\~ao \'e compacto. 
\end{proof}

\begin{observacao}
Existem espa\c cos pr\'e-compactos n\~ao compactos, por exemplo, o intervalo $(0,1)$.
\end{observacao}

\begin{teorema}
Um espa\c co m\'etrico $X$ \'e compacto se e somente se $X$ \'e pr\'e-compacto \'e completo. 
\label{t:criterio}
\end{teorema}

\begin{proof}
A necessidade ($\Rightarrow$) foi mostrada nos teoremas \ref{th:comp} e \ref{t:compactocompleto}.

A fim de verificar a sufici\^encia ($\Leftarrow$), seja $X$ um espa\c co pr\'e-compacto e completo, e seja $(x_n)$ uma sequ\^encia qualquer dos pontos de $X$. Cobremos $X$ com uma fam\'\i lia finita das bolas abertas de raio um. Como esta cole\c c\~ao \'e finita, existe pelo menos uma bola entre elas, $B_1(a_1)$, que cont\'em uma infinidade dos membros da sequ\^encia $(x_n)$. 

De acordo com o corol\'ario \ref{c:imm}, a bola $B_1(a_1)$ \'e pr\'e-compacta, logo ela pode ser coberta por uma fam\'\i lia finita das bolas abertas de raio $1/2$ (formadas no subespa\c co m\'etrico $B_1(a_1)$). Entre elas, pelo menos uma bola cont\'em uma infinidade dos membros da sequ\^encia $(x_n)$. 

Continuando pela recorr\^encia, escolhamos uma sequ\^encia infinita das bolas abertas $B_1(a_1)$, $B_{1/2}(a_2)$, $\ldots$, $B_{1/n}(a_n)$, 
$\ldots$, cada uma das quais cont\'em uma infinidade dos membros da sequ\^encia $(x_n)$. 

Finalmente, escolhamos uma sequ\^encia dos n\'umeros inteiros
\[n_1<n_2<\ldots<n_k<\ldots\]
de tal modo que cada elemento $x_{n_k}$, $k=1,2,3,\ldots$ pertence \`a bola $B_{1/k}(a_k)$. Esta subsequ\^encia \'e uma sequ\^encia de Cauchy: cada vez que $i,j>k$, temos 
\[d(x_{n_i},x_{n_j})<2/k.\]
Como o espa\c co m\'etrico $X$ \'e completo, a sequ\^encia extrata $(x_{n_k})_{k=1}^\infty$ converge em $X$. Conclu\'\i mos: $X$ \'e compacto.
\end{proof}

\begin{observacao}
\label{o:precauchy}
Com efeito, temos mostrado que se $X$ \'e um espa\c co pr\'e-compacto, ent\~ao cada sequ\^encia $(x_n)$ dos pontos de $X$ cont\'em uma subsequ\^encia de Cauchy. Prova-se facilmente que esta propriedade carateriza os espa\c cos pr\'e-compactos (exerc\'\i cio).
\end{observacao}

\begin{observacao} 
O crit\'erio da compacidade acima (teorema \ref{t:criterio}) \'e muito conveniente porque, geralmente, \'e mais f\'acil de verificar a pr\'e-compacidade de um espa\c co m\'etrico que a compacidade dele. 
\end{observacao}

\begin{exercicio}
Deduza o resultado cl\'assico seguinte.
\end{exercicio}

\begin{teorema}[Teorema de Heine-Borel]
\label{t:heine-borel}
Uma parte $K$ de $\R$ \'e compacta se e somente se $K$ \'e fechada e limitada. 
\end{teorema}

\begin{teorema}
Seja $X$ um subespa\c co denso de um espa\c co m\'etrico $Y$. Se $X$ \'e pr\'e-compacto, ent\~ao $Y$ \'e pr\'e-compacto.
\end{teorema}

\begin{proof}
Seja $r>0$ um n\'umero real qualquer. Escolhamos uma cole\c c\~ao finita  $x_1,x_2,\ldots,x_n\in X$ da tal maneira que 
\[X\subseteq \cup_{i=1}^n B_{r/2}(x_i).\]
(Aqu\'\i\ $B_{r/2}(x_i)$ pode denotar as bolas abertas seja em $X$, seja em $Y$, isso n\~ao importe).

Vamos verificar agora que 
\[Y\subseteq \cup_{i=1}^n B_{r}(x_i).\]
Seja $y\in Y$. Como $X$ \'e denso em $Y$, existe $x\in X\cap B_{r/2}(y)$. O ponto $x$ \'e contido na bola $B_{r/2}(x_i)$ por um $i$. Conclu\'\i mos, utilizando a desigualdade de tri\^angulo:
\[d(y,x_i)\leq d(y,x)+d(x,x_i)<\frac r 2+\frac r2 = r.\]
(Aqu\'\i\ $B_r(x_i)$ denota uma bola aberta em $Y$.)
\end{proof}

\begin{observacao}
Se $X$ \'e denso em $Y$, a condi\c c\~ao $X\subseteq\cup_{i=1}^n B_r(x_i)$ n\~ao implica, em geral, que $Y\subseteq\cup_{i=1}^n B_r(x_i)$. Queiram olhar o exemplo seguinte:$Y=(0,1)$, $X=(0,1)\setminus\{1/2\}$, $x=1/4$, $y=3/4$, $r=1/4$. \qed
\end{observacao}

\begin{corolario} 
Um espa\c co m\'etrico $X$ \'e pr\'e-compacto se e somente se o seu completamento $\hat X$ \'e compacto. \qed
\label{c:complprec}
\end{corolario}

\subsection{Aplica\c c\~oes uniformemente cont\'\i nuas}

\begin{observacao}
Lembremos que a imagem de um espa\c co compacto por uma fun\c c\~ao cont\'\i nua \'e compacto. \'E {\em falso} para os espa\c cos pr\'e-compactos. 

Por exemplo, a fun\c c\~ao cont\'\i nua
\[\left(-\frac{\pi}2,\frac{\pi}2\right)\ni x\mapsto \tan x\in \R\]
\'e sobrejetiva, o intervalo limitado \'e pr\'e-compacto, e a imagem n\~ao \'e pr\'e-compacta.
\end{observacao}

\begin{definicao}
Uma fun\c c\~ao $f\colon X\to Y$ entre dois espa\c cos m\'etricos \'e dita {\em uniformemente cont\'\i nua} se
\[\forall \e>0~~\exists\delta>0~~\forall x,y\in X~~(d_X(x,y)<\delta) \Rightarrow
(d_Y(f(x),f(y))<\e).\]
\label{d:fucms}
\index{fun\c c\~ao! uniformemente cont\'\i nua}
\end{definicao}

O valor de $\delta=\delta(\e)$ pode ser escolhido ``uniformemente'' em rela\c c\~ao a $x$. 
Certamente, cada fun\c c\~ao uniformemente cont\'\i nua \'e cont\'\i nua. 

\begin{exemplo}
A fun\c c\~ao
\[f(x)=x^2\]
de $\R$ para $\R$ \'e cont\'\i nua mas n\~ao uniformemente cont\'\i nua, porque, dados $\e>0$ e $x\in\R$, o valor de $\delta>0$ correspondente \'e da ordem da grandeza
\[\delta\sim \frac{\e}{2\abs x}.\]
Por conseguinte, $\delta$ depende de $x$ de maneira essencial. 
\end{exemplo}

\begin{teorema}
\label{t:compactocontinua}
Cada fun\c c\~ao cont\'\i nua sobre um espa\c co m\'etrico compacto \'e uniformemente cont\'\i nua.
\end{teorema}

\begin{proof}
Vamos mostrar a contraposi\c c\~ao: seja $f\colon X\to Y$ uma fun\c c\~ao cont\'\i nua entre dois espa\c cos m\'etricos. Suponhamos que $f$ n\~ao \'e uniformemente cont\'\i nua. Existe $\e_0>0$ tal que para cada $n=1,2,3,\ldots$ pode-se encontrar os pontos $x_n,y_n\in X$ que satisfazem
\begin{equation}
\label{eq:font}
d_X(x_n,y_n)<\frac 1n\end{equation}
e
\begin{equation}
\label{eq:et}
d_Y(f(x_n),f(y_n))\geq\e_0.\end{equation}
A sequ\^encia $(x_n)$ cont\'em uma subsequ\^encia convergente 
\[x_{n_i}\to x\mbox{ quando }i\to\infty.\]
Gra\c ca \`a hip\'otese (\ref{eq:font}), 
\[d_X(x,y_{n_i})\leq d_X(x,x_{n_i})+d_X(x_{n_i},y_{n_i})\to 0\mbox{ quando }i\to\infty.\]
Por conseguinte,
\[y_{n_i}\to x.\]
Por causa de (\ref{eq:et}), n\~ao temos
\[f(x_{n_i})\to f(x) \leftarrow f(y_{n_i}),\]
o que significa que $f$ n\~ao \'e cont\'\i nua.
\end{proof}

\begin{exercicio}
Mostre que a imagem de um espa\c co m\'etrico pr\'e-compacto para uma aplica\c c\~ao uniformemente cont\'\i nua \'e pr\'e-compacto.
\end{exercicio}

\section{Teorema da categoria de Baire}

\subsection{Conjuntos nunca densos}

Eis uma forte nega\c c\~ao da propriedade de um subconjunto de ser denso num espa\c co m\'etrico. 

\begin{definicao} 
Um subconjunto $A$ de um espa\c co m\'etrico $X$ \'e dito {\em nunca denso}\footnote{{\em Nowhere dense} (ingl\^es), {\em rare} (franc\^es).} se cada bola aberta, $B_\e(x)$, de $X$ conta uma bola $B_\delta(y)$ disjunta de $A$:
\[\forall x\in A,~~\forall \e>0~~\exists y\in X,~~\exists \delta>0,\mbox{ tais que }B_\delta(y)\subseteq B_\e(x)\mbox{ e }B_\delta(y)\cap A=\emptyset.\]
\end{definicao}

\begin{observacao}
Um subconjunto de um conjunto nunca denso \'e nunca denso.
\label{r:sous}
\end{observacao}

\begin{observacao}
$A$ \'e nunca denso se e somente se $\bar A$ \'e nunca denso.
\\[3mm]
A implica\c c\~ao $\Leftarrow$ segue-se da observa\c c\~ao \ref{r:sous}. A fim de mostrar a implica\c c\~ao $\Rightarrow$, suponhamos que $A$ \'e um subconjunto nunca denso de $X$. Sejam $x\in X$ e $\e>0$ quaisquer. Como $A$ \'e nunca denso, existem $y\in X$ e $\delta>0$ tais que $B_\delta(y)\subseteq B_\e(x)$ e $B_\delta(y)\cap A=\emptyset$. Mas toda bola aberta disjunta de um conjunto, $A$, \'e disjunta da ader\^encia $\bar A$.
\label{r:barre}
\end{observacao}

\begin{observacao} 
\label{r:ferme}
Um conjunto fechado, $F$, \'e nunca denso se e somente se $\Int F=\emptyset$.
\\[3mm]
A implica\c c\~ao ``somente se'' ($\Rightarrow$) \'e evidentemente verdadeira para todos subconjuntos, fechados ou n\~ao. A fim de verificar a implica\c c\~ao ``se'' ($\Leftarrow$), seja $F$ um conjunto fechado de interior vazio, e sejam $x\in X$ e $\e>0$ quaisquer. A bola $B_\e(x)$ n\~ao \'e contida em $F$ de acordo com a hip\'otese, por que $\Int F=\emptyset$. Logo a diferen\c ca $B_\e(x)\setminus F$ \'e n\~ao vazia. De outro lado, esta diferen\c ca \'e aberta como a interse\c c\~ao de dois abertos: 
\[B_\e(x)\setminus F=B_\e(x)\cap F^c.\]
Como todo aberto n\~ao vazio, $B_\e(x)\setminus F$ cont\'em uma bola aberta,  $B_\delta(y)$. Ela \'e contida na bola original, $B_\e(x)$, e disjunta de $F$.
\end{observacao}

As observa\c c\~oes \ref{r:barre} e \ref{r:ferme} implicam imediatamente:

\begin{proposicao}
Um subconjunto $A$ de um espa\c co m\'etrico $X$ \'e nunca denso em $X$ se e somente se a ader\^encia de $A$ possui o interior vazio. \qed
\end{proposicao}

\begin{exemplo} 
Cada conjunto unit\'ario $\{x\}$, onde $x$ n\~ao \'e isolado, \'e nunca denso.
\end{exemplo}

\begin{exemplo} 
A reta $y=0$ (o eixo coordenado $Ox$) \'e nunca denso no plano euclideano $\R^2$.
\end{exemplo}

\begin{exemplo} 
O conjunto $\Q$ n\~ao \'e nunca denso em $\R$, porque $\Q$ \'e denso. 
\end{exemplo}

\begin{exemplo} $\Z$ \'e nunca denso em $\R$ mas n\~ao \'e nunca denso em si mesmo.
\end{exemplo}

\begin{lema}
A uni\~ao de uma fam\'\i lia finita dos conjuntos nunca densos \'e nunca densa. 
\end{lema}

\begin{proof} 
Basta mostrar o resultado para dois conjuntos nunca densos, $A$ e $B$. Seja $B_\e(x)$ uma bola aberta qualquer. Ela cont\'em uma bola aberta $B_\delta(y)$ desjunta de $A$, e por sua vez, a bola $B_\delta(y)$ cont\'em uma bola aberta ainda menor, $B_\gamma(z)$, desjunta de $B$. Evidentemente, a bola $B_\gamma(z)$ \'e desjunta de $A\cup B$.
\end{proof}

\subsection{Conjuntos magros}

\begin{definicao} 
Uma parte $A$ de um espa\c co m\'etrico $X$ \'e dita um conjunto {\em magro,}\footnote{Como em franc\^es, {\em maigre}.}
ou um conjunto {\em da primeira categoria}\footnote{{\em Set of first category} (ingl\^es).} se ela \'e igual \`a uni\~ao de uma fam\'\i lia enumer\'avel de subconjuntos nunca densos:
\[A = \cup_{i=1}^\infty A_i,\]
onde $A_i$ s\~ao nunca densos em $X$.
\end{definicao}

\begin{observacao} Evidentemente, cada subconjunto de um conjunto magro \'e magro. Por conseguinte, um subconjunto magro pode ser definido como um conjunto {\em contido} na uni\~ao de uma fam\'\i lia enumer\'avel de subconjuntos nunca densos (ou: nunca densos e fechados).
\end{observacao}

\begin{exemplo} 
Cada conjunto nunca denso \'e magro.
\end{exemplo}

\begin{exemplo} $\Q$ \'e magro em $\R$, assim como em si mesmo.
\end{exemplo}

\begin{exemplo} $\Z$ \'e magro em $\R$, mas n\~ao \'e magro em si mesmo: todos os conjuntos unit\'arios s\~ao abertos em $\Z$, logo n\~ao nunca densos.
\end{exemplo}

\subsection{Teorema da categoria de Baire}

O teorema a seguir encontra-se entre as ferramentas mais \'uteis.

\begin{teorema}[Teorema da categoria de Baire]
Cada subconjunto magro de um espa\c co m\'etrico completo tem o interior vazio. 
\index{teorema! da categoria de Baire}
\end{teorema}

\begin{proof} Basta mostrar que a uni\~ao de uma sequ\^encia de subconjuntos fechados e nunca densos, $(F_n)$, tem o interior vazio. Seja $B_\e(x)$ uma bola aberta qualquer em $X$. Vamos provar que
\[B_\e(x)\not\subseteq\cup_{n=1}^\infty F_n.\]
Definamos $x_1=x$, $\e_1=\e$. Escolhamos uma bola fechada contida em $B_\e(x)$,  por exemplo $\bar B_{\e/2}(x)$. Como $F_1$ \'e nunca denso, existem $x_2\in X$ e $\e_2>0$ tais que
\[B_{\e_2}(x_2)\subseteq B_{\e/2}(x)\mbox{ e } B_{\e_2}(x_2)\cap F_1=\emptyset.\]
Reduzamos o raio a fim de obter uma bola fechada contida em $B_{\e_2}(x_2)$:
\[\bar B_{\e_2/2}(x_2)\subseteq B_{\e_2}(x_2).\]
Escolhamos $x_3$ e $\e_3>0$ de tal modo que
\[B_{\e_3}\subseteq \bar B_{\e_2/2}(x_2)\mbox{ e }B_{\e_3}\cap F_3=\emptyset.\]
Continuemos da maneira recorrente a fim de obter uma sequ\^encia das bolas fechadas encaixadas, $B_1,B_2,\ldots$, cujas raios aproximam-se de zero e que possuem a propriedade seguinte: cada bola $B_n$ \'e desjunta dos conjuntos nunca densos $F_1,F_2,\ldots, F_n$.
 
De acordo com lema \ref{l:encaixados}, a interse\c c\~ao $\cap B_n$ \'e n\~ao vazia. \'E \'obvio que o ponto comum de todas as bolas pertence \`a bola original, e ao mesmo tempo pertence ao nenhum conjunto $F_n$, $n=1,2,3,\ldots$. 
\end{proof}
 
\begin{exemplo} A conclus\~ao do teorema \'e falsa sem a hip\'otese que $X$ seja completo. Consideremos o espa\c co $\Q$ munido da dist\^ancia usual. Os conjuntos unit\'arios $\{q\}$, $q\in\Q$ s\~ao fechados, e o interior deles \'e vazio. Apesar disso,
\[\Q=\cup_{q\in\Q}\{q\}\]
\'e a uni\~ao de uma fam\'\i lia {\em enumer\'avel} dos subconjuntos nunca densos. 
 \end{exemplo}
 
 \begin{definicao}
Um subconjunto de um espa\c co m\'etrico $X$ \'e dito um {\em subconjunto do tipo $G_\delta$} (ou simplesmente {\em um $G_\delta$}), se $G$ \'e igual \`a interse\c c\~ao de uma fam\'\i lia enumer\'avel de abertos de $X$: 
\[G=\cap_{i=1}^\infty U_i,\mbox{ onde $U_i$ s\~ao abertos.}\]
\end{definicao}
 
\begin{exemplo} Cada aberto, $U$, em um espa\c co m\'etrico qualquer \'e um $G_\delta$:
\[U=\cap\{U\},\]
ou, se n\~ao convencer ainda,
\[U=\cap_{i=1}^\infty U_n,\mbox{ ou $U_n=U$ para todos $n$.}\]
\end{exemplo}
 
\begin{exemplo} O conjunto dos n\'umeros irracionais \'e um $G_\delta$ na reta $\R$:
\[\R\setminus\Q =\cap_{q\in\Q}(\R\setminus\{q\}).\]
\end{exemplo}
 
\begin{exemplo} Cada conjunto unit\'ario $\{x\}$ em um espa\c co m\'etrico qualquer $X$ \'e um $G_\delta$: verifica-se simplesmente que 
\[\{x\}=\cap_{n=1}^\infty B_{1/n}(x).\]
\end{exemplo}

\begin{exercicio}
Mais geralmente, cada subconjunto fechado de um espa\c co m\'etrico \'e um $G_\delta$. 
\end{exercicio}

Existem n\'umerosos corol\'arios e formas equivalentes do teorema de Baire. 
 
\begin{lema} Se $U$ \'e uma parte aberta densa num espa\c co m\'etrico $X$, ent\~ao o complemento $U^c$ \'e nunca denso em $X$. 
\label{l:l}
\end{lema}

\begin{proof} 
Como o complemento $U^c$ \'e fechado, basta mostrar que o seu interior \'e vazio. Se n\~ao, ent\~ao existem $x\in X$ e $\e>0$ tais que $B_\e(x)\subseteq \Int U^c$. Em particular, $B_\e(x)$ \'e desjunta de $U$, uma contradi\c c\~ao por que $U$ \'e denso em $X$.
\end{proof}

Deduz-se imediatamente:

\begin{teorema}[Forma equivalente do teorema de Baire]
Sejam $X$ um espa\c co m\'etrico completo e $(U_n)$ uma fam\'\i lia enumer\'avel de partes abertas de $X$ que s\~ao densas em $X$. Ent\~ao a interse\c c\~ao 
\[\cap_{n=1}^\infty U_n\]
\'e densa em $X$.
\index{teorema! da categoria de Baire}
\label{t:baireformaequiv}
\end{teorema}

\begin{proof} O complemento
\[\left(\cap_{n=1}^\infty U_n\right)^c = \cup_{n=1}^\infty U_n^c\]
\'e a uni\~ao de uma fam\'\i lia enumer\'avel dos conjuntos nunca densos, de acordo com o lema \ref{l:l}. Logo, o complemento acima \'e magro em $X$, e pelo teorema de Baire, o seu interior \'e vazio. Em outras palavras, esta uni\~ao n\~ao cont\'em nenhuma bola aberta. Portanto, o seu complemento, $\cap_{n=1}^\infty U_n$, encontro cada bola aberta.
\end{proof} 

\subsection{Genericidade\label{ss:genericidade}}

\begin{definicao}
Um subconjunto $G$ de um espa\c co m\'etrico $X$ \'e dito {\em comagro,} ou {\em gen\'erico,} se $A$ cont\'em um subconjunto $B$ do tipo $G_\delta$ que \'e denso em $X$.
\end{definicao}

Eis uma justifica\c c\~ao do nome ``comagro''.

\begin{lema} 
\label{l:maigre}
Uma parte $A$ de um espa\c co m\'etrico $X$ \'e comagro se e somente se o complemento $A^c$ \'e magro.
\end{lema} 

\begin{proof}
$\Rightarrow$: Seja $A$ uma parte comagra. Escolhamos um $G_\delta$ denso, $G$, contido em $A$. Ent\~ao
\[G=\cap_{n=1}^\infty U_n,\]
onde todos os $U_n$ s\~ao abertos e evidentemente densos em $X$ (por que $G\subseteq U_n$ e j\'a $G$ \'e denso). Temos
\[A^c\subseteq G^c=\cup_{n=1}^\infty U_n^c,\]
e de acordo com o lema \ref{l:l}, os subconjuntos $U_n^c$ s\~ao nunca densos. Conclu\'\i mos: $A^c$ \'e magro.
\par
$\Leftarrow$: a prova \'e semelhante.
\end{proof}

\begin{exemplo}
O conjunto dos n\'umeros racionais n\~ao \'e um $G_\delta$ em $\R$. Suponhamos o contrario: que $\Q$ \'e um $G_\delta$. Ent\~ao $\Q$ \'e comagro e o seu complemento, o conjunto $\R\setminus\Q$ dos n\'umeros irracionais, \'e magro. Como $\Q$ \'e si mesmo magro em $\R$, a uni\~ao $\R=\Q\cup(\R\setminus\Q)$ \'e magro assim, logo o interior em $\R$ \'e vazio, que \'e absurdo.  
\end{exemplo}

Eis uma outra reformula\c c\~ao do teorema de Baire. 
 
\begin{teorema}[Mais uma forma equivalente do teorema de Baire]
A interse\c c\~ao de uma fam\'\i lia enumer\'avel de partes gen\'ericas de um espa\c co m\'etrico completo \'e uma parte gen\'erica. 
\index{teorema! da categoria de Baire}
\end{teorema}

\begin{proof} Seja $A_n$, $n=1,2,\ldots$ uma fam\'\i lia enumer\'avel de subconjuntos gen\'ericos de $X$. Os complementos deles, $A_1^c$, $A_2^c$, $\ldots$, $A_n^c$, $\ldots$ s\~ao magros, de acordo com o lema \ref{l:maigre}. Evidentemente, a uni\~ao de uma fam\'\i lia enumer\'avel das partes magros, 
\[\cup_{n=1}^\infty A_n^c,\]
\'e magra. Logo, o complemento desta uni\~ao,
\[\left( \cup_{n=1}^\infty A_n^c\right)^c = \cap_{n=1}^\infty A_n,\]
\'e comagro ($=$ gen\'erico). 
\end{proof}
 
\begin{observacao} 
Na matem\'atica, uma {\em propriedade} dos elementos de um conjunto $X$ \'e mais ou menos sin\^onimo com uma {\em parte} de $X$, mesmo se habitualmente uma ``propriedade'' significa uma parte que pode ser definida por uma formula l\'ogica. 
\end{observacao}
 
\begin{corolario}
Seja
\[{\mathscr P}_1,{\mathscr P}_2,\ldots,{\mathscr P}_n,\ldots\]
uma sequ\^encia de propriedades de elementos de um espa\c co m\'etrico completo, $X$. Suponhamos que todas as propriedades ${\mathscr P}_n$ s\~ao gen\'ericas. Ent\~ao, a conjun\c c\~ao infinita delas, 
\[{\mathscr P}_1\wedge {\mathscr P}_2\wedge\ldots\wedge{\mathscr P}_n\wedge\ldots\]
\'e uma propriedade gen\'erica. Em particular, existe pelo menos um elemento $x\in X$ que verifica todas as propriedades ${\mathscr P}_n$.
\end{corolario}

Este resultante potente se aplica frequentemente \`as propriedades diversas de fun\c c\~oes, por example umas fun\c c\~oes cont\'\i nuas, diferenci\'aveis, e cetera. Neste caso, o espa\c co m\'etrico em quest\~ao consiste de fun\c c\~oes.

\begin{exemplo}
A fam\'\i lia das todas as fun\c c\~oes cont\'\i nuas $f\colon [0,1]\to\R$ que s\~	as n\~ao diferenci\'aveis em todos os pontos $t\in [0,1]$ do intervalo \'e comagro no espa\c co m\'etrico $C[0,1]$ das todas as fun\c c\~oes cont\'\i nuas reais sobre o intervalo, munido da dist\^ancia uniforme:
\[d(f,g)=\sup_{t\in [0,1]} \abs{f(t)-g(t)}.\]
Em outras palavras, uma fun\c c\~ao cont\'\i nua gen\'erica sobre o intervalo fechado \'e n\~ao diferenci\'avel em todos os pontos do intervalo. A demonstra\c c\~ao \'e n\~ao completamente \'obvia, mas eu acho que ela pode ser reconstru\'\i da direitamente, sem utilizar os livros.
\end{exemplo}

%
%

\chapter{Espa\c cos borelianos padr\~ao. Teorema de isomorfismo\label{apendice:padrao}}

Recordemos a defini\c c\~ao seguinte da subse\c c\~ao \ref{ss:espborpad}.

\begin{definition}
Um {\em espa\c co boreliano padr\~ao} \'e um par $(\Omega,{\mathscr B})$ que consiste de um conjunto $\Omega$ e uma sigma-\'algebra $\mathscr B$, que \'e uma estrutura boreliana gerada por uma m\'etrica completa e separ\'avel sobre $\Omega$.
\index{espa\c co! boreliano! padr\~ao}
\end{definition}

Neste ap\^endice, vamos mostrar o teorema seguinte (teorema \ref{t:isomorfismo} do Ap\^endice \ref{a:variaveis}).

\begin{teorema}
Dois espa\c cos borelianos padr\~ao s\~ao isomorfos se e somente se eles tem a mesma cardinalidade. 
\label{t:isomorfismo1}
\index{teorema! de isomorfismo boreliano}
\end{teorema}

Necessidade \'e trivial. Para mostrar a sufici\^encia, notemos que se $\Omega$ \'e enumer\'avel, ent\~ao a estrutura boreliana correspondente consiste de todos os subconjuntos de $\Omega$:
\[{\mathscr B}=2^{\Omega}.\]
De fato, todo conjunto unit\'ario de um espa\c co m\'etrico \'e fechado, logo boreliano, e num espa\c co enumer\'avel cada subconjunto \'e a uni\~ao de uma fam\'\i lia enumer\'avel de conjuntos unit\'arios. Por isso, $\Omega$ \'e Borel isomorfo ao (\'unico) espa\c co m\'etrico da mesma cardinalidade munido da dist\^ancia zero-um.

Eis um exerc\'\i cio para ilustrar a no\c c\~ao de um isomorfismo boreliano (defini\c c\~ao \ref{d:isomorfismoboreliano}).

\begin{exercicio}
Seja $\Omega$ um espa\c co boreliano n\~ao enumer\'avel, e seja $A\subseteq\Omega$ um subconjunto enumer\'avel. Mostrar que
 os espa\c cos borelianos $\Omega$ e $\Omega\setminus A$ s\~ao isomorfos (compare o exerc\'\i cio \ref{ex:subespacoboreliano}).
\par
[ {\em Sugest\~ao:} escolha um subconjunto enumer\'avel infinito qualquer $Z\subseteq\Omega\setminus A$, assim que uma bije\c c\~ao qualquer $\phi\colon A\cup Z\to Z$. Agora verifica que a bije\c c\~ao entre $\Omega$ e $\Omega\setminus A$, dada por
\[\Omega\ni x\mapsto\begin{cases} x,&\mbox{ se }x\in \Omega\setminus (A\cup Z),\\
\phi(x),&\mbox{ se }x\in A\cup Z,\end{cases}\]
\'e um isomorfismo boreliano. ]
\label{ex:omegaminusa}
\end{exercicio}

Relembremos que o {\em espa\c co de Baire} (exemplo \ref{ex:espacodebaire}) \'e o conjunto $\N^{\N}$ de todas as sequ\^encias de n\'umeros naturais, munido da m\'etrica
\[d(x,y)=\begin{cases} 0,&\mbox{ se }x=y,\\
2^{-\min\{i\colon x_i\neq y_i\}},&\mbox{ caso contr\'ario.}
\end{cases}\]
(Assim, a topologia do espa\c co de Baire \'e a topologia produto, no sentido da subse\c c\~ao \ref{ss:topologiadeproduto}, sobre a pot\^encia infinita enumer\'avel de $\Z$ munido da topologia discreta.)

\begin{exercicio}
Mostrar que o espa\c co de Baire \'e completo e separ\'avel.
\end{exercicio}

\begin{exercicio} ($\ast$) Mostrar que, como espa\c co topol\'ogico, o espa\c co de Baire \'e homeomorfo ao espa\c co $\R\setminus\Q$ de n\'umeros irracionais.
\end{exercicio}

Agora basta mostrar o resultado seguinte.

\begin{teorema}
Cada espa\c co boreliano padr\~ao n\~ao enumer\'avel \'e isomorfo ao espa\c co de Baire $\N^{\N}$ com a sua estrutura boreliana can\^onica.
\label{t:isomorfoaobaire}
\end{teorema}

Segue-se que, como espa\c cos borelianos, $\R$, $\R^d$, $\ell^2$, o espa\c co de Cantor $C$, todos os espa\c cos de Banach separ\'aveis n\~ao triviais, etc., s\~ao dois a dois isomorfos. 
A prova do teorema \ref{t:isomorfoaobaire} ocupa o resto do ap\^endice.

\begin{definicao} Um subconjunto $G$ de um espa\c co m\'etrico $(X,d)$ \'e dito um {\em subconjunto $G_\delta$} se existe uma fam\'\i lia enumer\'avel de conjuntos abertos $U_n$, $n\in\N$, tal que
\[G =\cap_{i\in\N} U_n.\]
\end{definicao}

Por exemplo, todo subconjunto aberto \'e $G_\delta$.

\begin{exercicio} Mostrar que todo subconjunto fechado de um espa\c co m\'etrico \'e um conjunto $G_\delta$.
\end{exercicio}

\begin{teorema}
Seja $(X,d)$ um espa\c co m\'etrico, e seja $Y$ um subespa\c co que admite uma m\'etrica $\rho$ completa e equivalente a $d\vert_Y$ (isto \'e, $\rho$ e $d$ induzem a mesma topologia sobre $Y$). Ent\~ao $Y$ \'e um subconjunto $G_\delta$ em $X$.
\label{t:completogdelta}
\end{teorema}

\begin{exercicio}
Mostre que um subconjunto $Y$ de um espa\c co m\'etrico $X$ \'e um $G_\delta$ em $X$ se e somente se $Y$ \'e um $G_\delta$ na sua ader\^encia $\bar Y^X$.
\end{exercicio}

Portanto, na prova do teorema \ref{t:completogdelta} suponhamos sem perda de generalidade que $Y$ \'e denso em $X$. 
Seja $x\in X\setminus Y$ qualquer. Para cada $\ve>0$, a bola aberta $B_{\ve}(x)$ encontra $Y$, ent\~ao o conjunto $\bar B_{\ve}(x)\cap Y$ \'e fechado \'e n\~ao vazio. Se $\ve_n\downarrow 0$, a sequ\^encia de fechados $\bar B_{\ve_n}(x)\cap Y$ em $Y$ \'e encaixada e tem a interse\c c\~ao vazia. Exerc\'\i cio \ref{ex:fechadosencaixados} implica que os di\^ametros destes conjuntos em rela\c c\~ao \`a m\'etrica $\rho$ n\~ao convergem para zero. Denotemos
\begin{align*}
D(x) & = \inf_{\ve>0}\mbox{diam}_{\rho} \bar B_{\ve_n}(x)\cap Y \\
& = \lim_{\ve\downarrow 0}\mbox{diam}_{\rho} \bar B_{\ve_n}(x)\cap Y,
\end{align*}
onde os di\^ametros s\~ao calculados em $(Y,\rho)$. Para cada $\delta>0$, denotemos
\[A_{\delta} = \{x\in X\colon  D(x)\geq\delta\}.\]

\begin{exercicio}
Mostre que
\[X\setminus Y = \bigcup_{\delta>0} A_{\delta}.\]
\end{exercicio}

\begin{exercicio}
Mostre que $A_{\delta}$ \'e fechado em $X$.
\end{exercicio}

Para estabelecer o teorema \ref{t:completogdelta}, \'e bastante observar que se $\delta\leq\delta^\prime$, ent\~ao $A_{\delta}\supseteq A_{\delta^\prime}$, e combinar os exerc\'\i cios acima. \hfill \qed

\begin{teorema} Seja $(X,d)$ um espa\c co m\'etrico completo, e seja $Y$ um subconjunto $G_\delta$ de $X$. Ent\~ao $Y$ admite uma m\'etrica equivalente completa, $\rho$. Al\'em disso, $\rho$ pode ser escolhida da maneira que $\rho\geq d\vert_Y$, ou seja, quais quer sejam $x,y\in Y$,
\[d(x,y)\leq \rho(x,y).\]
\label{t:completeequivegdelta}
\end{teorema} 

Come\c camos com um passo intermedi\'ario, o caso onde $Y$ \'e aberto em $X$.

\begin{exercicio}
Seja $f$ uma fun\c c\~ao real sobre um conjunto $X$. Mostre que a f\'ormula
\[d(x,y)=\abs{f(x)-f(y)}\]
defina uma pseudom\'etrica sobre $X$ (ou seja, uma fun\c c\~ao real de dois argumentos que satisfaz os axiomas 2 e 3 de uma m\'etrica).
\end{exercicio}

\begin{exercicio}
Seja $U$ um subconjunto aberto de um espa\c co m\'etrico completo $X=(X,d)$. Mostre que a m\'etrica $\rho_U$ sobre $U$, definida pela f\'ormula
\[\rho_U(x,y) = d(x,y)  + \min\left\{1,\left\vert \frac{1}{d(x,X\setminus U)}-\frac{1}{d(y,X\setminus U)}\right\vert\right\},
\]
\'e equivalente a $d$ e completa.
\end{exercicio}

Agora escolhamos uma sequ\^encia de abertos $(U_n)$ tal que $\cap_nU_n=Y$. Para cada $n$ escolhemos uma m\'etrica equivalente e completa sobre $U_n$, limitada por $1$. 

\begin{exercicio}
Mostre que espa\c co m\'etrico $\prod_{n=1}^{\infty} U_n$, munido da m\'etrica
\[d(x,y)=\sum_{n=1}^{\infty} 2^{-n}d_n(x_n,y_n),\]
\'e completo.
\end{exercicio}

\begin{exercicio} Mostre que a aplica\c c\~ao diagonal
\[Y\ni y\mapsto \Delta(y) = (y,y,y,\ldots,y,\ldots)\in \prod_{n} U_n,\]
\'e um homeomorfismo de $Y$ com um subespa\c co do produto.
\end{exercicio}

\begin{exercicio}
Mostre que a imagem de $Y$ \'e fechado dentro de produto.
\par
[ {\em Sugest\~ao:} se $x=(x_n)$ n\~ao pertence \`a imagem, ent\~ao a sequ\^encia $(x_n)$ n\~ao \'e constante.... ]
\end{exercicio}

\begin{exercicio}
Deduza o teorema \ref{t:completeequivegdelta}.
\end{exercicio}

\begin{exercicio}
Seja $\Omega$ um espa\c co m\'etrico separ\'avel. Denotemos por $A$ o conjunto de $x\in\Omega$ com a propriedade que existe $r>0$ tal que a bola $B_r(x)$ \'e enumer\'avel. Mostrar que $A$ \'e enumer\'avel. 
\label{ex:formemosA}
\end{exercicio}
 
\begin{lema}
Seja $X$ um espa\c co m\'etrico separ\'avel e completo e n\~ao enumer\'avel. Ent\~ao, dado $\e>0$, existe um subconjunto enumer\'avel $A\subseteq X$ e uma parti\c c\~ao enumer\'avel infinita de $X\setminus A$, 
\[X\setminus A = \bigcup_{i\in\N}G_i,~~G_i\cap G_j=\emptyset\mbox{ quando }i\neq j,\]
onde para todo $i$, $G_i$ \'e um conjunto $G_\delta$ n\~ao enumer\'avel de di\^ametro $<\e$. 
\label{l:particao}
\end{lema}

\begin{proof}
Formemos o conjunto enumer\'avel $A_1$ como no exerc\'\i cio \ref{ex:formemosA}, e definamos $X^\prime = X\setminus A_1$.
Seja $0<\gamma<\e/2$.
Escolha uma sequ\^encia densa em $X^\prime$, $(x_n)$. Temos $\cup_{i=1}^{\infty}B_{\gamma}(x_n)=X$. Escolha recursivamente uma subsequ\^encia, eliminando um ponto $x_n$ se e somente se $x_n\in\cup_{i=1}^{n-1}\bar B_\gamma(x_i)$, onde as bolas s\~ao formadas em $X^\prime$.
Para todo $n$, defina $X_n=\bar B_\gamma(x_n)\setminus \cup_{i=1}^n \bar B_\gamma(x_i)$. Pode ocorrer que esta fam\'\i lia seja infinita ou finita, mas ela consiste de conjuntos de di\^ametro $\leq 2\gamma<\e$, que cobrem $X^\prime$. 

Agora dividimos $X_1=\bar B_\gamma(x_1)$, em ``an\'eis'', a saber: existe uma sequ\^encia infinita $\gamma_m\downarrow 0$, $\gamma_1=\gamma$, tal que todo conjunto $Y_m=\bar B_{\gamma_m}(x_1)\setminus \bar B_{\gamma_{m+1}}(x_1)$ n\~ao \'e enumer\'avel. 

Denotemos por $G_{n}, {n\in\N}$ a fam\'\i lia infinita que consiste de todos os conjuntos $Y_m$, $m=1,2,\ldots$, e $X_n$, $n=2,3,\ldots$. Para cada um deles, o conjunto $A_n\subseteq G^\prime_n$ de pontos $x\in G^\prime_n$ que admitem uma bola $B_r(x)\cap G^\prime_n$ enumer\'avel \'e enumer\'avel. Denotaremos $A_2=\cup_{n\in\N}A_n$, $A=A_1\cup A_2$, e $G_n=G^\prime_n\setminus A$. Se alguns conjuntos $G_n$ foram vazios, enlevemos-os. Os restantes, junto com $A$, satisfazem as conclus\~oes do lema.
\end{proof}

Vamos terminar a prova do teorema \ref{t:isomorfoaobaire}.

Denotemos por $\N^{<\omega}$ o conjunto de todos as sequ\^encias finitas dos n\'umeros naturais. 
Recursivamente no comprimento $k=\ell(\bar n)$ de uma sequ\^encia $\bar n\in \N^{<\omega}$ definamos para todo $\bar n$ um subconjunto $G_{\bar n}\subseteq \Omega$, uma m\'etrica $\rho_{\bar n}$ sobre $G_{\bar n}$, e um conjunto enumer\'avel $A_{\bar n}\subseteq \Omega$ tais que:
\begin{itemize}
\item $G_{\emptyset}=\Omega$, $\rho_{\emptyset}=d$,
\item $G_{\bar n}$ \'e um conjunto $G_\delta$ em $\Omega$,
\item $G_{\bar n}$ n\~ao \'e enumer\'avel,
\item se o comprimento de $\bar n$ \'e igual a $k$, ent\~ao para todo $m\in\N$, ${\mathrm{diam}}_{\rho_{\bar n}}(G_{\bar nm})<1/k$,
\item para todo $\bar n\in \N^{<\omega}$,
os conjuntos da forma $G_{\bar n m}$, onde $m\in\N$, formam uma parti\c c\~ao de $G_{\bar n}\setminus A_{\bar n}$,
\item a m\'etrica $\rho_{\bar n}$ sobre $G_{\bar n}$ \'e completa (e automaticamente separ\'avel),
\item se $\bar n\in \N^{<\omega}$ e $m\in\N$, ent\~ao $\rho_{\bar n m}\geq \rho_{\bar n}\vert_{G_{\bar n m}}$.
\end{itemize}

Denotemos $A=\cup A_{\bar n}$.
Dado um elemento qualquer do espa\c co de Baire, 
\[x=(n_1,n_2,\ldots)\in\N^{\N},\]
a sequ\^encia encaixada de conjuntos $G_{(n_1,n_2,\ldots,n_k)}\subseteq\Omega$, $k=1,2,\ldots$, que correspondem aos prefixes de $x$, tem a propriedade ${\mathrm{diam}}_d(G_{(n_1,n_2,\ldots,n_k)})\to 0$. Segue-se que a sequ\^encia de fechos destes conjuntos tem um e apenas um ponto em comum, que denotemos por $f(x)$ (exerc\'\i cio \ref{ex:fechadosencaixados}).  Deste modo, a fun\c c\~ao $f\colon\N^{\N}\to\Omega$ \'e bem definida. De fato, $f(x)\in\Omega\setminus A$: se $\bar k$ \'e um prefixo de $x$ qualquer, ent\~ao os di\^ametros de $G_{\bar n}$ convergem para zero relativo a m\'etrica completa $\rho_{\bar k}$. Por isso, $f(x)\in G_{\bar k}$. Segue-se que $f(x)\notin A$. 

A fun\c c\~ao $f$ \'e sobrejetora: se $y\in\Omega\setminus A$, ent\~ao para cada $k$ existe a \'unica sequ\^encia finita $(n_1,\ldots,n_k)\in \N^{<\omega}$, tal que $y\in G_{(n_1,\ldots,n_k)}$, e al\'em disso, se $k<\ell$, ent\~ao a sequ\^encia do comprimento $\ell$ come\c ca com os elementos $n_1,\ldots,n_k$. Deste modo, obtemos uma sequ\^encia infinita $n_1,n_2,\ldots,n_k,\ldots,n_\ell,\ldots$, cuja imagem por $f$ \'e igual a $y$. 

A fun\c c\~ao $f$ \'e injetora: se dois elementos, $x$ e $y$, de $\N^{\N}$ s\~ao diferentes, existe $k$ tal que os $k$-prefixos de sequ\^encias, $\bar n$ e $\bar m$, n\~ao s\~ao iguais. Ent\~ao, $f(x)$ e $f(y)$ pertencem aos dois espa\c cos m\'etricos completos disjuntos, $(G_{\bar n},\rho_{\bar n})$ e $(G_{\bar m},\rho_{\bar m})$. Gra\c cas \`as propriedades das m\'etricas escolhidas, conclu\'\i mos que $f(x)\in G_{\bar n}$ e $f(y)\in G_{\bar m}$, logo $f(x)\neq f(y)$. 

A fun\c c\~ao $f$ \'e boreliana. Isso \'e uma consequ\^encia imediata dos resultados seguintes.

\begin{exercicio}
Sejam $(\Omega_i,{\mathscr B}_i)$, $i=1,2$, dois espa\c cos borelianos, e $f\colon \Omega_1\to\Omega_2$ uma fun\c c\~ao. Suponha que ${\mathscr B}_2$ seja gerada pela fam\'\i lia $\mathcal D$ de subconjuntos de $\Omega_2$, ou seja, ${\mathscr B}_2$ \'e a menor sigma-\'algebra que contem a fam\'\i lia $\mathcal D$. Suponha que para todo $D\in {\mathscr D}$ o conjunto $f^{-1}(D)$ \'e boreliano. Ent\~ao $f$ \'e uma fun\c c\~ao boreliana.
\end{exercicio}

\begin{exercicio}
Mostrar que cada bola aberta de $\Omega$ \'e a uni\~ao de uma (necessariamente enumer\'avel) fam\'\i lia de conjuntos da forma $G_{\bar n}$, $\bar n\in\N^{<\omega}$. Deduzir que a fam\'\i lia de tais conjuntos gera a estrutura boreliana de $\Omega$.
\end{exercicio} 

Como $f^{-1}(G_{\bar n})$ \'e o conjunto aberto de todas sequ\^encias que tem o prefixo $\bar n$, segue-se que $f$ \'e boreliana.

Finalmente, $f^{-1}$ \'e boreliana, usando o mesmo resultado e a observa\c c\~ao
que a imagem por $f$ de conjunto aberto de todas as sequ\^encias que tem o prefixo $\bar n$ do comprimento $k$ (ou seja, uma bola aberta de raio $2^{-k+1}$) \'e igual ao conjunto $G_{\bar n}\setminus A$. 

Temos um isomorfismo boreliano $f\colon\N^{\N}\to\Omega\setminus A$, onde $A$ \'e enumer\'avel. O exerc\'\i cio \ref{ex:omegaminusa} diz que $\Omega\setminus A$ \'e isomorfo a $\Omega$, o que termina a demonstra\c c\~ao. \hfill\qed

O livro \citep*{kechris} \'e uma refer\^encia padr\~ao da teoria de conjuntos descritiva.

%
%

\chapter{Medidas de probabilidade. Teorema de Carath\'eodory
\label{a:medidas}}

Neste Ap\^endice vamos desenvolver alguns ferramentas t\'ecnicas necess\'arias para mostrar a exist\^encia de medidas n\~ao-at\^omicas (difusas) e algumas de suas propriedades, assim como definir o produto de uma fam\'\i lia de medidas, uma no\c c\~ao indispens\'avel para estudar a independ\^encia de vari\'aveis aleat\'orias. 

\section{Teorema de extens\~ao de medida}

Relembramos a defini\c c\~ao seguinte.

\begin{definicao}
Seja $\Omega$ um conjunto n\~ao vazio. Uma {\em sigma-\'algebra} de subconjuntos de $\Omega$ \'e uma fam\'\i lia de subconjuntos de $\Omega$,
\[{\mathcal F}\subseteq 2^{\Omega},\]
tal que
\begin{enumerate}
\item $\Omega\in{\mathcal F}$,
\item se $A\in\F$, ent\~ao $A^c\in\F$ (onde $A^c=\Omega\setminus A$).
\item se $A_1,\cdots,A_n,\cdots\in\F$ s\~ao elementos de $\F$, ent\~ao $\cup_{i=1}^{\infty} A_i\in\F$.
\end{enumerate}
\index{sigma-\'algebra}
\end{definicao}

\begin{observacao} Segue-se que uma sigma-\'algebra contem as interse\c c\~oes enumer\'aveis de seus membros tamb\'em, assim como o conjunto vazio. 
\end{observacao}

Todavia, vamos come\c car com uma no\c c\~ao auxiliar de um campo de conjuntos. Os campos s\~ao mais f\'acil a manipular do que as sigma-\'algebras.

\begin{definicao}
\label{cylin}
Seja $\Omega$ um conjunto. Um {\em campo} ({\em field}) de subconjuntos de $\Omega$ \'e uma fam\'\i lia de subconjuntos de $\Omega$,
\[{\mathcal F}_0\subseteq 2^{\Omega},\]
tal que
\begin{enumerate}
\item $\Omega\in{\mathcal F}_0$,
\item se $A\in\F_0$, ent\~ao $A^c\in\F_0$,
\item se $n\in\N$ e $A_1,\cdots,A_n\in\F_0$ s\~ao dois a dois disjuntos, 
ent\~ao $\cup_{i=1}^{n} A_i\in\F_0$.
\end{enumerate}
\index{campo de conjuntos}
\end{definicao}

\begin{exemplo}
\label{ex:semi}
A fam\'\i lia $\F_0$ de todos os subconjuntos $A$ do intervalo semiaberto 
$(0,1]$ que podem ser representados como uni\~oes finitas de intervalos semiabertos,
\[A= (a_1,b_1]\cup(a_2,b_2]\cup\cdots \cup (a_k,b_k],~~ k\in\N,\]
forma um campo. 
\end{exemplo}

\begin{definicao} Un subconjunto $A$ de $\{0,1\}^\N$ chama-se um {\em conjunto cil\'\i ndrico} se ele \'e definido pelo um conjunto finito de coordenadas, $I$. 
Precisamente, existe um conjunto finito $I\subset\N$ e um conjunto $B\subseteq
\{0,1\}^I$ tais que
\begin{equation}
\label{repre}
A = \pi_I^{-1}(B).
\end{equation}
Aqui, 
\[\pi_I\colon \{0,1\}^\N\to \{0,1\}^I\]
\'e a proje\c c\~ao usual, definida por 
\[\{0,1\}^\N\ni\omega \mapsto [I\ni i\mapsto \pi_I(\omega)_i:=
\omega_i\in\{0,1\}]\in \{0,1\}^I.\]
\index{conjunto! cil\'\i ndrico}
\end{definicao}

\begin{exemplo}
O conjunto
\[A=\{\omega\in \{0,1\}^\N \mid \omega_1=0\}\]
\'e cil\'\i ndrico. Neste caso, podemos tomar $I=\{1\}$ e
\[B=\{0\}\subseteq \{0,1\}=\{0,1\}^I.\]
\end{exemplo}

\begin{exercicio}
A fam\'\i lia de todos os conjuntos cil\'\i ndricos de $\{0,1\}^\N$ \'e um corpo de conjuntos. 
\end{exercicio}

\begin{definicao} Seja $\F_0$ um corpo de subconjuntos de um conjunto $\Omega$. Digamos que a fun\c c\~ao 
\[\mu\colon \F_0\to\R_+\]
\'e uma medida de probabilidade sobre o corpo $\F_0$ se ela satisfaz as condi\c c\~oes seguintes.
\begin{enumerate}
\item $\mu(\Omega)=1$,
\item se $A_1,\cdots,A_n,\cdots\in\F_0$ s\~ao elementos de $\F_0$ dois a dois disjuntos, tais que sua uni\~ao $\cup_{i=1}^{\infty} A_i$ pertence a $\F_0$, 
ent\~ao $\mu\left(\cup_{i=1}^{\infty} A_i\right)=
\sum_{i=1}^{\infty} \mu(A_i)$. 
\end{enumerate}

\label{d:mpcorpo}
\end{definicao}

Tais medidas s\~ao f\'aceis a construir \`a m\~ao.

\begin{exemplo}
Sobre o corpo $\F_0$ de subconjuntos de $(0,1]$ do exemplo 
\ref{ex:semi}, definamos a medida de probabilidade $P$ por
\begin{align*}
\mu(A) &= \mu\left((a_1,b_1]\cup(a_2,b_2]\cup\cdots \cup (a_k,b_k]\right) \\
&:= l(a_1,b_1]+l(a_2,b_2]+\cdots +l (a_k,b_k] \\
&=  \sum_{i=1}^k (b_i-a_i),
\end{align*}
onde os subintervalos $(a_i,b_i]$ s\~ao dois a dois disjuntos. Todo elemento de $\F_0$ pode ser representado desta maneira ({\em exerc\'\i cio}).
\label{ex:semiintervalos} 
\end{exemplo}

\begin{exercicio}
Verificar que a defini\c c\~ao da medida no exemplo \ref{ex:semiintervalos} \'e consistente, ou seja, independente da representa\c c\~ao de um conjunto $A$ como acima, e que trata-se de uma medida de probabilidade no sentido da defini\c c\~ao \ref{d:mpcorpo}.
\end{exercicio}

\begin{exemplo}
Dado um conjunto cil\'\i ndrico $A$ (defini\c c\~ao \ref{cylin}), definamos a sua medida por
\[\mu(A) = \frac{\sharp(B)}{2^{\abs I}},\]
onde $A$ \'e representado como em Eq. (\ref{repre}). 

Em outras palavras, $\mu(A)$ \'e igual \`a medida de contagem normalizada do conjunto $B$ no cubo de Hamming $\{0,1\}^I$.

Precisamos verificar que a defini\c c\~ao n\~ao depende da escolha de um conjunto de \'\i ndices $I$ e um subconjunto $B\subseteq\{0,1\}^I$. Com esta finalidade, notemos que entre todos os $I$ que permitem uma representa\c c\~ao de $A$ como em Eq. (\ref{repre}), ent\~ao existe o menor conjunto de \'\i ndices, $I_0$. O resto \'e mais ou menos evidente. ({\em Exerc\'\i cio}). 

\'E menos \'obvio que a medida $\mu$ \'e sigma-aditiva. Precisamos demostrar que, cada vez que um conjunto cil\'\i ndrico $A$ for representado como a uni\~ao de uma sequ\^encia de cilindros disjuntos, 
$A_1,A_2,\cdots, A_n,\cdots$, temos necessariamente
\[\mu(A) = \sum_i \mu(A_i).\]
Se a sequ\^encia terminar, isso \'e f\'acil: neste caso basta mostrar a igualdade para $i=2$. Sejam $A_1$, $A_2$ dois conjuntos cil\'\i ndricos disjuntos, suportados sobre os conjuntos de \'\i ndices $I_1$ e $I_2$ respectivamente. Neste caso, o conjunto $A_1\cup A_2$ \'e um cilindro suportado pelo conjunto de \'\i ndices $I=I_1\cup I_2$. De fato, temos conjuntos
$B_1,B_2\subseteq \{0,1\}^I$ tais que
\[A_1=\pi_I^{-1}(A_1),~~ A_2=\pi_I^{-1}(A_2),\]
e por conseguinte $A_1\cup A_2 = \pi_I^{-1}(B_1\cup B_2)$. Agora temos
\begin{eqnarray*}
\mu(A_1\cup A_2) &=& \frac{\sharp(B_1\cup B_2)}{2^{\abs I}} \\
&=& \frac{\sharp(B_1)}{2^{\abs I}}+\frac{\sharp( B_2)}{2^{\abs I}}\\
&=& \mu(A_1)+\mu(A_2),
\end{eqnarray*}
como evidentemente $B_1$ e $B_2$ s\~ao disjuntos. 

\'E o caso dif\'\i cil onde todos os cilindros $A_1,A_2,\cdots, A_n,\cdots$ s\~ao dois a dois disjuntos e n\~ao vazios? O fato \'e que este caso nunca ocorre. Com efeito, isso \'e uma consequ\^encia da propriedade fundamental da compacidade do espa\c co $\{0,1\}^\N$ munido da topologia padr\~ao de produto, ou seja, a topologia do espa\c co (ou: cubo) de Cantor. 
\label{ex:verrrrificacao}
\end{exemplo} 

Para mostrar este fato, vamos reexprimir-o numa forma equivalente.

\begin{lema} 
\label{l:computacaocilindricos} Se
\[B_1\supseteq B_2\supseteq\cdots\supseteq B_n\supseteq\cdots\]
\'e uma sequ\^encia encaixada de subconjuntos cil\'\i ndricos n\~ao vazias de
$\{0,1\}^\N$, ent\~ao
\[\cap_{i=1}^\infty B_i\neq\emptyset.\]
\end{lema}

Para notar que isso \'e exatamente o que precisamos, dado uma fam\'\i lia $A_1,A_2,\cdots$ de conjuntos cil\'\i ndricos dois a dois disjuntos e n\~ao vazios, cuja uni\~ao, $A$, \'e um conjunto cil\'\i ndrico, renomeie $B_i=A\setminus\left(\cup_{i=1}^n A_i\right)$ e aplica o lema \ref{l:computacaocilindricos}.

\begin{proof}[Prova do lema \ref{l:computacaocilindricos}] Vamos escolher recursivamente um elemento comum
$\omega$ contido em todos os conjuntos $B_i$. Tem um elemento $\omega_1\in\{0,1\}$ com a propriedade seguinte: 
cada $A_i$ contem uma palavra que come\c ca com $\omega_1$. (Exerc\'\i cio de 30 segundos). 
Defina $\omega_1=\e_1$. Agora, existe um elemento $\omega_2$, igual a $0$ ou a $1$, com a propriedade: cada 
$A_i$, $i\geq 2$, contem uma palavra cuja primeira coordenada \'e igual a  $\omega_1$ e a segunda, a $\omega_2$. (Porque se n\~ao, ent\~ao...?) 

Continuando a repetir este procedimento, obtemos uma sequ\^encia 
$\omega_1,\omega_2,\omega_3,\cdots$ que pertence \`a interse\c c\~ao de todos os $A_i's$. 
\end{proof}

Os campos em si s\~ao pouco importantes, mas eles geram as sigma-\'algebras.

\begin{lema} Cada campo (e de fato, cada fam\'\i lia de subconjuntos de $\Omega$), $\F_0$, \'e contida em uma \'unica menor sigma-\'algebra. Em outras palavras, existe uma sigma-\'algebra $\F$
tal que 
\[\F_0\subseteq \F\subseteq 2^{\Omega},\]
e se $\F_1$ \'e uma sigma-\'algebra qualquer que contem $\F_0$,
ent\~ao
\[\F\subseteq\F_1.\]
\end{lema}

\begin{proof} A fam\'\i lia, dizemos $\Phi$, de todas as sigma-\'algebras $\Xi$ de subconjuntos de $\Omega$ com a propriedade 
$\F_0\subseteq\Xi$ n\~ao \'e vazia (ela contem sempre o conjunto $2^{\Omega}$ de partes de $\Omega$), e por conseguinte a interse\c c\~ao, 
\[\F:=\cap\Phi =\cap_{\Xi\in\Phi}\Xi,\]
\'e uma fam\'\i lia bem definida de subconjuntos de $\Omega$. \'E f\'acil a verificar que a interse\c c\~ao de uma fam\'\i lia de sigma-\'algebras \'e uma sigma-\'algebra. E pela defini\c c\~ao, $\F$ \'e contido em toda sigma-\'algebra que contem $\F_0$.
\end{proof}

\begin{exemplo}
Os subconjuntos borelianos de $\left\{ \begin{matrix} (0,1] \\ \{0,1\}^{\N} \end{matrix}
\right\}$ s\~ao exatamente os elementos da menor sigma-\'algebra que contem 
 $\left\{ \begin{matrix} \mbox{ todo intervalo semiaberto } \\
 \mbox{ todo conjunto cil\'\i ndrico } \end{matrix}
\right\}$, respetivamente. \qed
\end{exemplo}

Aqui um resultado extremamente poderoso e \'util para construir as medidas.

\begin{teorema}[Teorema de Carath\'eodory] Seja $\F_0$ um corpo de subconjuntos de um conjunto $\Omega$, e seja
$\mu\colon\F_0\to\R_+$ uma fun\c c\~ao tendo as propriedades:
\begin{enumerate}
\item $\mu(\Omega)=1$.
\item Se $A\in\F_0$, ent\~ao $\mu(A^c)=1-\mu(A)$.
\item Se $A_1,A_2,\cdots,A_n,\cdots
\in\F_0$ s\~ao dois a dois disjuntos e tais que $\cup_{i=1}^\infty A_i\in\F_0$,
ent\~ao $\mu(\cup_{i=1}^\infty A_i)=\sum_{i=1}^\infty \mu(A_i)$.
\end{enumerate}
Ent\~ao existem uma sigma-\'algebra $\F$ que contem $\F_0$, e uma medida de probabilidade $\tilde \mu$ definida sobre $\F$, de modo que a restri\c c\~ao de  $\tilde \mu$ a $\F_0$ \'e igual a $\mu$:
\[\forall A\in\F_0,~~ \tilde \mu(A) = \mu(A).\]
Al\'em disso, a medida de probabilidade $\tilde \mu$ sobre $\F$ com esta propriedade \'e \'unica.
\label{t:extensao}
\index{teorema! de Carath\'eodory}
\end{teorema}

O resto da sec\c c\~ao \'e dedicado \`a demonstra\c c\~ao do teorema. 

\subsection{Medida exterior e conjuntos mensur\'aveis}
\begin{definicao}
Seja $A\subseteq\Omega$ um subconjunto qualquer. A {\it medida exterior} de
$A$ \'e o valor
\[\mu^\ast(A)=\inf \sum_{i=1}^\infty \mu(A_i),\]
onde o \'\i nfimo \'e tomado sobre todas as coberturas enumer\'aveis $\{ A_i\colon i\in\N\}$ do conjunto $A$ com elementos do campo $\F_0$:
\[A\subseteq \cup_{i\in\N} A_i, ~~ A_i\in\F_0.\]
\end{definicao}

O \'\i nfimo ao lado direita \'e bem definido pois todo subconjunto $A\subseteq\Omega$ admite uma cobertura enumer\'avel pelos elementos de $\F_0$, por exemplo, $A\subseteq \cup\{\Omega\}$. Al\'em disso, \'e claro que a soma ao lado direita \'e sempre menor ou igual a $1$.

\begin{proposicao}
A medida exterior tem as propriedades seguintes.
\begin{enumerate}
\item $\mu^\ast(\emptyset)=0$.
\item (Monotonicidade.) Se $A\subseteq B$, ent\~ao $\mu^\ast(A)\leq \mu^\ast(B)$.
\item (Subaditividade.) Se $A_i, i=1,2,\dots$ \'e uma fam\'\i lia enumer\'avel de subconjuntos de $\Omega$, ent\~ao
\[\mu^\ast\left( \cup_{i}A_i\right)\leq \sum_i \mu^\ast(A_i).\]
\end{enumerate} 
\label{threep}
\end{proposicao}

\begin{proof}
As duas primeiras propriedades s\~ao mais ou menos \'obvias. A terceira \'e uma consequ\^encia do fato que a soma de uma s\'erie absolutamente convergente n\~ao \'e afeita pelos rearranjos dos termos. Seja $\e>0$ qualquer. Para cada
$n\in\N_+$, escolhemos os elementos $B_{ni}\in\F_0$ da maneira que
\[A_n\subseteq\bigcup_{i=1}^\infty B_{ni}\mbox{ e }
\sum_{i=1}^\infty \mu(B_{ni})\leq \mu^\ast(A_n)+\frac{\e}{2^n}.\] 
A fam\'\i lia 
$(B_{ni})_{n,i=1}^\infty$ de elementos do campo $\F_0$ \'e enumer\'avel e satisfaz
\[\sum_{n,i=1}^\infty \mu(B_{ni}) = \sum_{n=1}^\infty\sum_{i=1}^\infty
 \mu(B_{ni}) \leq \sum_{n=1}^\infty \mu^\ast(B_{ni})+\e.\]
Tomando o \'\i nfimo sobre todos $\e>0$, obtemos a propriedade desejada. 
\end{proof}

Em geral, a medida exterior at\'e n\~ao \'e finitamente aditiva (pelo menos, assumindo o axioma de escolha). Vamos definir uma subclasse de conjuntos sobre qual $\mu^\ast$ \'e sigma-aditiva. Estes conjuntos formam uma sigma-\'algebra que contem $\F_0$, e s\~ao chamados {\em conjuntos mensur\'aveis}.

\begin{definicao}
Um subconjunto $A\subseteq\Omega$ diz-se {\it $\mu$-mensur\'avel,} ou simplesmente
{\it mensur\'avel,} se para todo subconjunto $B\subseteq\Omega$ 
\[\mu^\ast(B) = \mu^\ast(B\cap A) + \mu^\ast(B\setminus A).\]
\index{conjunto! mu-mensuravel@$\mu$-mensur\'avel}
\end{definicao}

\begin{observacao}
Como a medida exterior \'e subaditiva
(proposi\c c\~ao \ref{threep}), sempre temos a desigualdade
\[\mu^\ast(B) \leq \mu^\ast(B\cap A) + \mu^\ast(B\setminus A),\]
quaisquer que sejam $A$ e $B$. Ent\~ao, um subconjunto $A\subseteq\Omega$ \'e mensur\'avel se e somente se, qualquer que seja subconjunto $B\subseteq\Omega$, temos a desigualdade
\[\mu^\ast(B) \geq \mu^\ast(B\cap A) + \mu^\ast(B\setminus A).\]

\label{oneside}
\end{observacao}

Denotaremos $\mathcal M$ a fam\'\i lia de todos os subconjuntos $\mu$-mensur\'aveis de $\Omega$.

Substituindo $B\leftarrow A\cup B$ na defini\c c\~ao de um conjunto mensur\'avel, obtemos:

\begin{proposicao}
A medida exterior \'e finitamente aditiva sobre os conjuntos mensur\'aveis: se 
$A$ e $B$ s\~ao disjuntos e mensur\'aveis, ent\~ao $\mu^\ast(A\cup B)=\mu^\ast(A)+
\mu^\ast(B)$. 
\qed
\end{proposicao}

Esta propriedade estende-se indutivamente \`as uni\~oes finitas.  

\begin{proposicao} Todo elemento do campo original $\F_0$ \'e um conjunto mensur\'avel, ou seja,
\[\F_0\subseteq{\mathcal M}.\]
\end{proposicao}

\begin{proof} Seja $A\in\F_0$, e seja $B\subseteq\Omega$ qualquer.
Dado $\e>0$ qualquer, escolhemos uma sequ\^encia enumer\'avel $(A_i)_{i=1}^\infty$ de elementos de $\F_0$ tal que
\[B\subseteq\bigcup_{i=1}^\infty A_i\]
e
\[\sum_{i=1}^\infty \mu(A_i)\leq
\mu^\ast(B) +\e.\]
Definamos $A_i'=A_i\cap A$ e $A_i''=A_i\setminus A$. Os conjuntos
$A_i',A_i''$ pertencem ao corpo $\F_0$, e 
\[B\cap A\subseteq\cup_{i=1}^\infty
A_i',~~ B\setminus A \subseteq \cup_{i=1}^\infty
A_i'.\]
Por conseguinte, 
\[\mu^\ast(B\cap A)\leq \sum_{i=1}^\infty
\mu(A_i')\mbox{ e }\mu^\ast(B\setminus A)\leq  \sum_{i=1}^\infty
\mu(A_i'').\]
Notemos que para todo $i$,
\[\mu(A_i) = \mu(A_i') + \mu(A_i'').\]
Deduzimos das desigualdades acima:
\begin{align*}
\mu^\ast(B) &\geq  \sum_{i=1}^\infty \mu(A_i)-\e\\
&= \sum_{i=1}^\infty \mu(A_i') + \sum_{i=1}^\infty \mu(A_i'')-\e \\
&\geq  \mu^\ast(B\cap A) + \mu^\ast(B\setminus A)-\e.
\end{align*}
Enviando $\e\to 0$, obtemos
\[\mu^\ast(B) \geq \mu^\ast(B\cap A) + \mu^\ast(B\setminus A),\]
o que significa, considerando a observa\c c\~ao \ref{oneside}, que $A$ \'e mensur\'avel.
\end{proof}

\begin{proposicao} A restri\c c\~ao do medida exterior $\mu^\ast$ sobre o campo $\F_0$ \'e igual \`a medida $\mu$:
\[\forall A\subseteq\F_0,~~ \mu^\ast(A) = \mu(A).\]
\end{proposicao}

\begin{proof} Tomando a cobertura de $A$ com o \'unico elemento $A\in\F_0$, conclu\'\i mos
\[\mu^\ast(A)\leq \mu(A).\]
Seja $\e>0$ qualquer. Existe uma cobertura enumer\'avel de $A$ pelos elementos $A_1,A_2,\cdots\in\F_0$ tal que
\begin{equation}
\label{ff}
\sum_{i=1}^\infty \mu(A_i)\leq \mu^\ast(A)+\e.
\end{equation}
Os conjuntos
\[A_i'=\left(A_i\setminus\cup_{j=1}^{i-1}A_j\right)\cap A\]
pertencem ao campo $\F_0$, s\~ao dois a dois disjuntos, e a sua uni\~ao \'e igual a $A$, que \'e um elemento de $\F_0$. Segundo as propriedades de $P$, 
\[\mu(A) =\sum_{i=1}^\infty \mu(A_i').\]
Como $A_i'\subseteq A_i$, temos
\[\sum_{i=1}^\infty \mu(A_i')\leq\sum_{i=1}^\infty \mu(A_i),\]
e gra\c cas \`a Eq. (\ref{ff}),
\[\mu(A)\leq \mu^\ast(A)+\e.\]
Como isso \'e verdadeiro para todos $\e>0$, conclu\'\i mos, enviando $\e\downarrow 0$, que
\[\mu(A)\leq \mu^\ast(A).\]
\end{proof}

\begin{teorema}
\label{mathcal}
A fam\'\i lia $\mathcal M$ de todos os conjuntos mensur\'aveis forma uma sigma-\'algebra.
\end{teorema}

\begin{proof}
Somente a sigma-aditividade precisa uma prova, pois as primeiras duas propriedades s\~ao mais ou menos evidentes.

Vamos mostrar inicialmente que as uni\~oes finitas de conjuntos mensur\'aveis s\~ao mensur\'aveis. Basta mostrar para dois conjuntos mensur\'aveis, $A_1$ e $A_2$. 
Com esta finalidade, seja $B\subseteq\Omega$ qualquer. Temos, como $A_1$ \'e mensur\'avel,
\[\mu^\ast(B)=\mu^\ast(B\cap A_1)+\mu^\ast(B\setminus A_1),\]
e como $A_2$ \'e mensur\'avel,
\[\mu^\ast(B\setminus A_1) = \mu^\ast((B\setminus A_1)\cap A_2) + 
\mu^\ast(B\setminus A_1\setminus A_2).\]
Por conseguinte,
\begin{align}
\mu^\ast(B)&=\mu^\ast(B\cap A_1) + 
\mu^\ast(B\cap (A_2\setminus A_1))+\mu^\ast(B\setminus (A_1\cup A_2))\nonumber \\
& =  \mu^\ast(B\cap (A_1 \cup A_2)) + \mu^\ast(B\setminus (A_1\cup A_2)),
\end{align}
o que estabelece a mensurabilidade de $A_1\cup A_2$. 

Em consequ\^encia disso, as interse\c c\~oes finitas e as diferen\c cas de conjuntos mensur\'aveis s\~ao mensur\'aveis. Conclu\'\i mos: $\mathcal M$ \'e um campo.

Seja $(A_i)_{i=1}^\infty$ uma sequ\^encia infinita de subconjuntos mensur\'aveis de $\Omega$. Sem perda de generalidade e substituindo para $A_i$ o conjunto mensur\'avel
$A_i\setminus(\cup_{j=1}^i A_j)$ se for necess\'ario, podemos supor que os conjuntos $A_i$ s\~ao dois a dois disjuntos. Denotemos $A=\cup_{i=1}^\infty A_i$. Vamos mostrar que $A$ \'e mensur\'avel. 

Seja $B\subseteq\Omega$ qualquer. Temos, para todo $i$, 
\[\mu^\ast(B) = \mu^\ast(\cup_{j=1}^i A_j\cap B) + 
\mu^\ast(B\setminus\cup_{j=1}^i A_j).\]
Como a medida exterior \'e mon\'otona, conclu\'\i mos
\[\mu^\ast(B) \geq \mu^\ast(\cup_{j=1}^i A_j\cap B) 
+ \mu^\ast(B\setminus A).\]
Como os conjuntos s\~ao dois a dois disjuntos e mensur\'aveis, podemos reescrever a \'ultima express\~ao assim: 
\[\mu^\ast(B) \geq \sum_{j=1}^i \mu^\ast(A_j\cap B) 
+ \mu^\ast(B\setminus A).\]
No limite  $i\to\infty$ temos
\begin{equation}
\mu^\ast(B) \geq \sum_{j=1}^\infty \mu^\ast(A_j\cap B) 
+ \mu^\ast(B\setminus A).
\label{valuable}
\end{equation}
Como $\mu^\ast(A\cap B)\leq  \sum_{j=1}^\infty \mu^\ast(A_j\cap B)$
(subaditividade de $\mu^\ast$), temos
\[\mu^\ast(B) \geq \mu^\ast(A\cap B)
+ \mu^\ast(B\setminus A).\]
A outra desigualdade segue-se da subaditividade da medida exterior: 
\[\mu^\ast(B) \leq \mu^\ast(A\cap B)
+ \mu^\ast(B\setminus A).\]
Combinamos as duas para concluir:
\[\mu^\ast(B) = \mu^\ast(A\cap B) + \mu^\ast(B\setminus A).\]
Isso estabelece que $A=\cup_{i=1}^\infty A_i$ \'e mensur\'avel.
\end{proof}

De fato, na prova mostramos algo mais. Substituindo na Eq. (\ref{valuable})
$B=A$, obtemos
\[\mu^\ast(A) \geq \sum_{j=1}^\infty \mu^\ast(A_j),\]
a junto com a subaditividade, esta desigualdade implica 
\begin{equation}
\label{equality}
\mu^\ast(A) = \sum_{j=1}^\infty \mu^\ast(A_j).
\end{equation}

A restri\c c\~ao, $\tilde \mu$, da medida exterior $\mu^\ast$ sobre a sigma-\'algebra $\mathcal M$ de conjuntos mensur\'aveis \'e uma medida de probabilidade. A restri\c c\~ao de $\tilde \mu$ sobre $\F_0$ \'e igual a $\mu$.

\subsection{Unicidade de extens\~ao}
Seja $\mu_1$ uma medida de probabilidade qualquer, definida sobre a sigma-\'algebra $\mathcal M$ e tal que a restri\c c\~ao de $\mu_1$ sobre $\F_0$ \'e igual a $\mu$:
\[\forall A\in\F_0,~~ \mu(A) = \mu_1(A).\]

\begin{lema} Para todo $A\in\mathcal M$, $\mu_1(A)\leq \tilde \mu(A)$.
\label{l:leq}
\end{lema}

\begin{proof} Seja $\e>0$ qualquer. Segundo a defini\c c\~ao de uma medida exterior, existe uma cobertura de $A$ por elementos de $\F_0$,
\[A\subseteq \cup_{i=1}^\infty A_i,~~ A_i\in\F_0,\]
tal que
\[\sum_{i=1}^\infty \mu(A_i)\leq\tilde \mu(A)+\e.\]
Usando a monotonicidade e $\sigma$-subaditividade da medida 
$\mu_1$, obtemos
\begin{align*}
\mu_1(A) &\leq
\mu_1\left(\cup_{i=1}^\infty A_i\right)\\
&\leq \sum_{i=1}^\infty \mu_1(A_i) \\
&= \sum_{i=1}^\infty \mu(A_i)\\
&\leq\tilde \mu(A)+\e.
\end{align*}
Enviando $\e\downarrow 0$, conclu\'\i mos
\[\mu_1(A)\leq \tilde \mu(A).\]
\end{proof} 

Seja $A\in {\mathcal M}$ qualquer. Lema \ref{l:leq} diz que 
\[\mu_1(A)\leq \tilde \mu(A),\]
e no mesmo tempo, 
\[\mu_1(A^c)\leq \tilde \mu(A^c).\]
Por conseguinte,
\begin{eqnarray*}
\mu_1(A) &=& 1-\mu_1(A^c) \\
&\geq & 1-\tilde \mu(A^c) \\
&=& \tilde \mu(A),
\end{eqnarray*}
o que significa 
\[\mu_1(A) = \tilde \mu(A).\]
Conclu\'\i mos: a extens\~ao da medida \'e \'unica. Isso termina a demonstra\c c\~ao do teorema \ref{t:extensao}.

Com efeito, nos demonstramos mais. No argumento acima, basta supor que a medida de probabilidade $\mu_1$ seja definida sobre uma sigma-\'algebra, dizemos $\F_1$, presa entre $\F_0$ e $\mathcal M$,
\[\F\subseteq\F_1\subseteq{\mathcal M}.\]
O que nos demonstramos, \'e o resultado seguinte, uma forma um pouco mais exata do teorema de extens\~ao. 

\begin{teorema}[Teorema de Carath\'eodory] 
 Seja $\F_0$ um campo de conjuntos de um conjunto $\Omega$, e seja
$\mu\colon\F_0\to\R_+$ uma fun\c c\~ao com as propriedades
\begin{enumerate}
\item $\mu(\Omega)=1$.
\item se $A_1,A_2,\cdots,A_n,\cdots
\in\F_0$ s\~ao dois a dois disjuntos e $\cup_{i=1}^\infty A_i\in\F_0$,
ent\~ao $\mu(\cup_{i=1}^\infty A_i)=\sum_{i=1}^\infty \mu(A_i)$.
\end{enumerate}
Denotaremos $\tilde \mu$ a medida de probabilidade sobre a sigma-\'algebra
$\mathcal M$ de todos os conjuntos $\mu$-mensur\'aveis de $\Omega$. Seja $\F_1$ uma sigma-\'algebra qualquer de subconjuntos de $\Omega$ que satisfaz
\[\F_0\subseteq\F_1\subseteq{\mathcal M},\]
e seja $\mu_1$ uma medida de probabilidade sobre $\F_1$, cuja restri\c c\~ao sobre 
$\F_0$ \'e igual a $\mu$. Ent\~ao $\mu_1$ \'e igual \`a restri\c c\~ao de $\tilde \mu$ sobre $\F_1$.
\qed
\index{teorema! de Carath\'eodory}
\end{teorema}

Este resultado diz que usar a nota\c c\~ao $\tilde \mu$ n\~ao tem sentido, pois usando a letra $\mu$ para a medida estendida vai nunca resultar numa ambiguidade.

\subsection{Dois exemplos}
\begin{exemplo}
O espa\c co \'e o espa\c co de Cantor $\Omega=\{0,1\}^\N$ de sequ\^encias bin\'arias infinitas. Aqui, 
o campo $\F_0$ \'e formado por todos conjuntos cil\'\i ndricos de $\Omega$, e os elementos da sigma-\'algebra correspondente s\~ao os conjuntos borelianos. Os elementos da sigma-\'algebra $\mathcal M$ s\~ao ditos conjuntos {\em Haar mensur\'aveis.} A medida de probabilidade resultante \'e chamada {\em medida de Haar.} 
\index{medida! de Haar}
\end{exemplo}

\begin{observacao}
A maneira de lig\'a-la com a medida de Haar constru\'\i da na esfera na se\c c\~ao \ref{s:haarnaesfera}, \'e notar que esta medida vai resultar da mesma constru\c c\~ao se $\{0,1\}^{\N}$ \'e visto como um grupo compacto sob adi\c c\~ao m\'odulo $1$ munida da m\'etrica $\sum_{i=1}^{\infty}2^{-i}\abs{\sigma_i-\tau_i}$ (compare tamb\'em as observa\c c\~oes \ref{obs:haar} e \ref{obs:unicidade}). \'E um exerc\'\i cio interessante.
\end{observacao}

\begin{exemplo} O espa\c co \'e o intervalo $(0,1]$. Aqui,
$\F_0$ consiste de todos os uni\~oes finitos de intervalos semiabertos. A sigma-\'algebra gerada por $\F_0$ consiste de conjuntos borelianos do intervalo. Os elementos da fam\'\i lia $\mathcal M$ s\~ao conjuntos
{\it mensur\'aveis no sentido de Lebesgue.} A medida de probabilidade estendida \'e frequentemente denotada $\lambda$ ou as vezes $\lambda^{(1)}$, \'e dita a {\em medida de Lebesgue}, ou {\em medida uniforme} sobre o intervalo. 
\index{medida! de Lebesgue}
\end{exemplo}

\subsection{$\F$ contra $\mathcal M$} 
Nossa pr\'oxima tarefa \'e mostrar que, de certa forma, a sigma-\'algebra gerada por $\F_0$ j\'a basta para dar vaz\~ao \`as necessidades da probabilidade e teoria de medida. 

Estamos sempre no \^ambito delineado em
\ref{t:extensao}. 
Seja $A\in{\mathcal M}$ um conjunto mensur\'avel qualquer de
$\Omega$. 
Para todo $n\in\N_+$, podemos cobrir $A$ com uma fam\'\i lia enumer\'avel de elementos de $\F_0$,
\[A\subseteq \cup_{i=1}^\infty A_i,~~ A_i\in\F_0\]
de modo que
\[\sum_{i=1}^\infty \mu(A_i)\leq \mu(A)+\frac 1n.\]
(Lembra que $\mu(A)$ \'e igual \`a medida exterior $\mu^\ast(A)$ para todo conjunto mensur\'avel, $A$.) Se denota-se
\[B_n=\cup_{i=1}^\infty A_i,\]
ent\~ao
\[A\subseteq B_n\]
e 
\[\mu(A)\leq \mu(B_n)\leq \mu(A)+\frac 1n.\]
Al\'em disso, $B_n\in\F$. Agora definiremos
\[B=\cap_{i=1}^\infty B_n.\]
Claro, $B\in\F$. Tamb\'em,
\[A\subseteq B.\]
Qual quer seja $n$,
\[\mu(B)\leq \mu(B_n)\leq \mu(A)+\frac 1n.\]
De fato, temos
\[\mu(B)\leq \mu(A),\]
e
\[\mu(B)=\mu(A).\]
Ademais,
\[A\Delta B = (A\setminus B)\cup (B\setminus A) = B\setminus A,\]
e como
\[\mu(A) = \mu(B) = \mu(A) + \mu(B\setminus A),\]
conclu\'\i mos:
\[\mu(A\Delta B)= \mu(B\setminus A)=0.\]

\begin{definicao} Um conjunto mensur\'avel $N$ de $\Omega$ \'e dito um {\em conjunto nulo,} ou {\em conjunto negligenci\'avel,} se $\mu(N)=0$.

\index{conjunto! negligenci\'avel}
\end{definicao}

Demonstremos o seguinte.

\begin{teorema}
Sobre as hip\'oteses do teorema \ref{t:extensao}, seja $A$ um subconjunto mensur\'avel de $\Omega$. Ent\~ao existe um subconjunto $B\in\F$ tal que a diferen\c ca sim\'etrica $A\Delta B$ \'e negligenci\'avel.

Mais exatamente, cada conjunto mensur\'avel $A\subseteq\Omega$ pode ser representado ou como 
\[A = B\setminus N,\]
onde $B\in\F$ e $N$ \'e negligenci\'avel, ou como 
\[A = B\cup N,\]
onde de novo $B\in\F$ e $N$ \'e negligenci\'avel.
\qed
\label{t:perturb}
\end{teorema}

(A \'ultima representa\c c\~ao \'e obtida pela aplica\c c\~ao do resultado ao complemento $A^c$.)

\begin{exercicio}
Seja $N$ um conjunto negligenci\'avel. Mostre que qualquer subconjunto $A\subseteq N$ \'e mensur\'avel (logo, certamente, negligenci\'avel).
\end{exercicio}

\begin{observacao}
Tipicamente, h\'a muito mais conjuntos mensur\'aveis do que elementos de $\F$ (por exemplo, conjuntos borelianos). Por exemplo, no caso do intervalo unit\'ario 
$(0,1]$ assim como do espa\c co de Cantor $\{0,1\}^\N$, pode-se mostrar que a cardinalidade da fam\'\i lia de todos os conjuntos borelianos (elementos de $\F$) \'o a do cont\'\i nuo,
${\mathfrak c}=2^{\aleph_0}$, enquanto a cardinalidade da fam\'\i lia de todos os conjuntos mensur\'aveis \'e igual a
\[2^{\mathfrak c} = 2^{2^{\aleph_0}}.\]
(Deixemos ambas afirma\c c\~oes como exerc\'\i cios para os leitores interessados).
Em qualquer caso, segundo o teorema \ref{t:perturb}, um conjunto mensur\'avel s\'i difere de um elemento apropriado de $\F$ por uma parte negligenci\'avel, que n\~ao importa. \qed
\end{observacao}

\begin{exercicio}
Deduzir, sobre as hip\'oteses do teorema \ref{t:extensao}, o resultado seguinte. Seja $f$ uma fun\c c\~ao ${\mathcal M}$-mensur\'avel com valores reais, ou seja, dado $a,b\in\R$, $a<b$, a imagem rec\'\i proca $f^{-1}(a,b)$ pertence a ${\mathcal M}$. Ent\~ao existe uma fun\c c\~ao $g$ $\F$-mensur\'avel (para todos $a,b$, $g^{-1}(a,b)$ pertence a $\F$) tal que $f$ e $g$ s\'o diferem sobre um conjunto $\mu$-negligenci\'avel:
\[\mu\{x\in\Omega\colon f(x)\neq g(x)\}=0.\]
[ {\em Sugest\~ao:} aplicar o teorema \ref{t:perturb} recursivamente a cada conjunto da forma $f^{-1}(a,b)$ com $a,b$ racionais... ]
\label{ex:fmudag}
\end{exercicio}

\subsection{Completamento de um espa\c co probabil\'\i stico}

Seja $(\Omega,\F, \mu)$ um espa\c co probabil\'\i stico qualquer. Ent\~ao, a sigma-\'algebra $\F$ \'e, em particular, um campo, \'e o procedimento acima pode se aplicar ao espa\c co $(\Omega,\F, \mu)$ tamb\'em. A sigma-\'algebra $\F$ \'e estendida at\'e uma sigma-\'algebra $\mathcal M$ de todos os conjuntos $\mu$-mensur\'aveis, e a medida de probabilidade $\mu$ se estende para uma medida de probabilidade (notada normalmente pela mesma letra) sobre $\mathcal M$.

\begin{definicao}
O espa\c co de probabilidade $(\Omega,{\mathcal M},\mu)$, chama-se o {\em completamento} de $(\Omega,\F, \mu)$. 
\end{definicao}

\begin{definicao} 
Um espa\c co de probabilidade $(\Omega,\F, P)$ \'e dito {\em completo} se ele coincide com o seu pr\'oprio completamento, ou seja, se todo subconjunto $\mu$-mensur\'avel de $\Omega$ \'e j\'a contido em $\F$. 
\end{definicao}

Ao cada espa\c co probabil\'\i stico $(\Omega,{\mathcal M},\mu)$ podemos associar um espa\c co pseudom\'etrico, cujo conjunto subjacente \'e ${\mathcal M}$ e a pseudom\'etrica \'e dada por a express\~ao $\mu(A\Delta B)$. 

\begin{exercicio}
Verifique que a express\~ao acima \'e uma pseudom\'etrica (satisfaz axiomas 2 e 3 de uma m\'etrica).
\end{exercicio}

A cada espa\c co pseudom\'etrico pode-se associar um espa\c co m\'etrico quociente, como descrito no exerc\'\i cio \ref{ex:espacometricoquociente}.

\begin{exercicio}
Verificar que os espa\c cos m\'etricos quocientes associados a um espa\c co probabil\'\i stico e ao seu completamento s\~ao isometricamente isomorfos, e ambos s\~ao completos. 
\end{exercicio}

Por isso, o nome ``completamento de um espa\c co probabil\'\i stico'' n\~ao \'e consistente com a estrutura (pseudo)m\'etrica sobre a sigma-\'algebra (que \'e importante na parte principal das nossas notas). Trata-se de completamento num sentido diferente.

Pode-se mostrar que um espa\c co $(\Omega,\F,\mu)$ \'e completo --- ou seja, seu completamento n\~ao adiciona novos conjuntos mensur\'aveis --- se e somente se cada subconjunto de um conjunto negligenci\'avel $A\in\F$, $\mu(A)=0$, j\'a pertence a $\F$. No entanto, temos que parar em algum momento. Deixemos a prova deste crit\'erio como um (fact\'\i vel) exerc\'\i cio.

\subsection{Produtos de espa\c cos probabil\'\i sticos}

Seja $\Omega_\alpha$, $\alpha\in A$, uma fam\'\i lia de espa\c cos probabil\'\i sticos,
onde $A$ \'e um conjunto de \'\i ndices e para todo $\alpha\in A$
\[\Omega_\alpha=(\Omega_\alpha,\F_\alpha,P_\alpha).\]
Denotemos por
\[\Omega=\prod_{\alpha\in A}\Omega_\alpha\]
o produto cartesiano da fam\'\i lia de conjuntos $\Omega_\alpha$, $\alpha\in A$.
Em outras palavras, elementos de $\Omega$ s\~ao aplica\c c\~oes $x\colon A\to
\cup_{\alpha\in A}\Omega_\alpha$ tendo a propriedade que para todos
$\alpha\in A$
\[x_\alpha= x(\alpha)\in\Omega_\alpha.\]
Axioma da Escolha garante que $\Omega\neq\emptyset$ (pois os espa\c cos $\Omega_{\alpha}$ nunca s\~ao vazios).

Para um subconjunto $I\subseteq A$, definamos a {\em proje\c c\~ao can\^onica} $\pi^A_I$, ou simplesmente $\pi_I$, 
\[\pi_I\colon \Omega\ni x\mapsto \pi_I(x)\in \Omega_I,\]
onde 
\[\Omega_I=\prod_{\beta\in I}\Omega_\beta\]
e para cada $\beta\in I$
\[\pi_I(x)_\beta = x_\beta.\]
Por exemplo, se $I=\{\beta\}$ \'e um conjunto unit\'ario, obtemos a proje\c c\~ao de $\beta$-\'esima coordenada, $\pi_\beta$, de $\Omega$
sobre o fator $\Omega_\beta$, dada por
$\pi_\beta(x)=x_\beta$.

Seja $\Omega_1,\Omega_2,\ldots,\Omega_n$ uma subcole\c c\~ao finita qualquer de espa\c cos probabil\'\i sticos, tendo as sigma-\'algebras de conjuntos mensur\'aveis
$\F_1,\F_2,\ldots,\F_n$, respetivamente. Chamemos um subconjunto $A\subseteq\prod_{i=1}^n\Omega_i$ {\it retangular} se
$A$ \'e da forma
\[A=A_1\times A_2\times\ldots\times A_n,\]
para apropriados $A_i\in\F_i$, $i=1,2,3,\ldots$. Chamemos um subconjunto
$B\subseteq\prod_{i=1}^n\Omega_i$ {\em elementar} se $B$ \'e a uni\~ao de um n\'umero finito de subconjuntos retangulares dois a dois disjuntos.
A observa\c c\~ao seguinte \'e elementar, embora talvez um pouco tediosa para verificar em todos os detalhes.

\begin{exercicio}
Mostre que a cole\c c\~ao de todos os subconjuntos elementares de um produto finito $\prod_{i=1}^n\Omega_i$ \'e um campo de conjuntos. \qed
\end{exercicio}

Denotemos por $\F_0$ a fam\'\i lia de todos os subconjuntos {\em cil\'\i ndricos}, $B$, do produto cartesiano
$\Omega=\prod_{\alpha\in A}\Omega_\alpha$, ou seja, conjuntos da forma
\[B=\pi_I^{-1}(C),\]
onde $I\subseteq A$ \'e {\it finito,} e $C\subseteq \prod_{\beta\in I}
\Omega_\beta$ \'e um subconjunto elementar. O seguinte \'e f\'acil a mostrar.

\begin{lema} 
A fam\'\i lia $\F_0$ \'e um campo de subconjuntos de $\Omega$.
\qed
\end{lema}

Agora denotemos por $\F$ a sigma-\'algebra gerada por $\F_0$. De modo equivalente, $\F$ \'e a menor sigma-\'algebra que cont\'em todos os conjuntos da forma $\pi_\alpha^{-1}(A)$, $A\in\F_\alpha$, $\alpha\in A$, ou tamb\'em como a menor sigma-\'algebra que torna todas as proje\c c\~oes $\pi_{\alpha}$, $\alpha\in A$, mensur\'aveis.

Nossa tarefa ser\'a de munir $\Omega$ com uma medida de probabilidade definida sobre a sigma-\'algebra $\F$. Isto \'e feito em duas etapas. 

\begin{lema} 
Seja $A\in\F_0$. Represente $A$ sob a forma
\[\pi_I^{-1}\left(\bigcup_{i=1}^k \prod_{j=1}^{k}
A_{ij}, \right),\]
onde $I\subseteq A$ \'e finito, $\abs I=k$, a uni\~ao \'e a de conjuntos dois a dois disjuntos, e $A_{ij}\in \F_j$. Ent\~ao o n\'umero 
\[P(A):= \sum_{i=1}^k P_1(A_{i1})\cdot P_2(A_{i2})\cdot\ldots\cdot
P_k(A_{ik})\]
n\~ao depende da escolha de uma representa\c c\~ao para $A$ como acima, e determina uma fun\c c\~ao sobre o campo $\F_0$ tendo as propriedades
\begin{enumerate}
\item $P(\Omega)=1$.
\item Se $A_1,A_2,\cdots,A_n,\cdots
\in\F_0$ s\~ao dois a dois disjuntos e tais que $\cup_{i=1}^\infty A_i\in\F_0$,
ent\~ao $P(\cup_{i=1}^\infty A_i)=\sum_{i=1}^\infty P(A_i)$.
\end{enumerate}
\label{l:f_0produtogeral}
\end{lema}

Deixemos a primeira afirma\c c\~ao como exerc\'\i cio. Al\'em disso, como n\'os apenas vamos usar produtos enumer\'aveis, vamos fazer a verifica\c c\~ao da segunda afirma\c c\~ao s\'o para eles, porque para os produtos n\~ao enumer\'aveis a sigma-\'algebra resultante n\~ao \'e uma estrutura boreliana padr\~ao, e n\'os n\~ao temos ferramentas para lidar com isso.

\begin{exercicio}
Suponha que $\abs A\leq \aleph_0$, e que todos os espa\c cos $(\Omega_\alpha,\F_\alpha)$ s\~ao espa\c cos borelianos padr\~ao. Verificar que o produto $(\Omega,\F)$ \'e um espa\c co boreliano padr\~ao. Mais exatamente, sejam $d_{\alpha}$ m\'etricas separ\'aveis e completas sobre $\Omega_\alpha$ gerando as estruturas $\F_\alpha$. Ent\~ao, $\F$ \'e gerada por qualquer m\'etrica sobre o produto que gera a topologia produto correspondente.
\end{exercicio}

\begin{definicao}
Um espa\c co m\'etrico $X$ tem {\em dimens\~ao zero} se os conjuntos abertos e fechados formam uma {\em base,} ou seja: dado $x\in X$ e $\ve>0$, existe um conjunto aberto e fechado, $V$, tal que
\[x\in V\subseteq B_{\ve}(x).\]
\end{definicao}

\begin{exemplo}
Todo espa\c co m\'etrico finito tem dimens\~ao zero.
\end{exemplo}

\begin{exercicio}
Verifique que o espa\c co de Cantor $\{0,1\}^\N$ tem dimens\~ao zero.
\end{exercicio}

\begin{exercicio}
Mostre que o espa\c co $\Q$ de racionais munido da dist\^ancia usual tem dimens\~ao zero.
\end{exercicio}

\begin{exercicio}
Mostre que qualquer espa\c co m\'etrico enumer\'avel tem dimens\~ao zero.
\par
[ {\em Sugest\~ao:} estudar as fun\c c\~oes cont\'\i nuas $d(x,-)$ da dist\^ancia e suas imagens em $\R$... ]
\end{exercicio}

Agora usamos o teorema de isomorfismo boreliano \ref{t:isomorfismo1} para realizar cada $\Omega_\alpha$ como um espa\c co m\'etrico compacto de dimens\~ao zero, a saber:
\begin{itemize}
\item Se $\Omega$ \'e finito, ser\'a o espa\c co m\'etrico finito correspondente com qualquer m\'etrica (por exemplo, zero-um).
\item Se $\Omega$ \'e infinito e enumer\'avel, ser\'a o espa\c co m\'etrico compacto $\alpha\N = \{0\}\cup\{1/n\colon n\in\N_+\}$.
\item Se $\Omega$ tem cardinalidade $\mathfrak c$, ser\'a o espa\c co de Cantor $\{0,1\}^{\N}$ munido de qualquer m\'etrica gerando a topologia produto, como, por exemplo, $d(x,y)=2^{-\min\{i\colon x_i\neq y_i\}}$.
\end{itemize}

\begin{exercicio}
Mostrar que o produto de uma fam\'\i lia enumer\'avel de espa\c cos m\'etricos de dimens\~ao zero, munido de uma m\'etrica que gera a topologia produto, tem dimens\~ao zero.
\label{ex:produtodimzero}
\end{exercicio}

Agora o produto $\Omega=\prod_{\alpha\in A}\Omega_{\alpha}$ \'e si mesmo um espa\c co m\'etrico compacto (exerc\'\i cio \ref{ex:produtocompacto}) de dimens\~ao zero.

\begin{exercicio}
Mostre que a fam\'\i lia $\F_{00}$ de todos os conjuntos cil\'\i ndricos abertos e fechados forma um corpo em $\Omega$ e gera a sigma-\'algebra $\F$.
\end{exercicio}

\begin{exercicio}
Use os mesmos argumentos topol\'ogicos que no caso particular $\Omega=\{0,1\}^{\N_+}$ tratado no exemplo \ref{ex:verrrrificacao} e no lema \ref{l:computacaocilindricos} para mostrar que a fun\c c\~ao $\mu$ definida sobre $\F_{00}$ como no lema \ref{l:f_0produtogeral} satisfaz a conclus\~ao do lema.
\end{exercicio}

Agora apliquemos o teorema de extens\~ao de medidas \ref{t:extensao} para concluir que $\mu$ estende-se at\'e a \'unica medida de probabilidade definida sobre a sigma-\'algebra $\F$. Esta medida chama-se a {\em medida produto} \'e denotada 
\[\mu=\otimes_{\alpha \in A} \mu_\alpha.\]
O espa\c co de probabilidade padr\~ao $(\Omega,\F,\otimes_{\alpha \in A} P_\alpha)$ se chama o {\it produto} de espa\c cos de probabilidade padr\~ao
$\Omega_\alpha$, $\alpha\in A$. No caso onde $\Omega_\alpha$ s\~ao dois a dois isomorfos, $\Omega_\alpha = \Omega$, trata-se de uma 
{\it pot\^encia} de um espa\c co de probabilidade,
$\Omega^A$, e a medida produto \'e denotada $\mu^{\otimes A}$.

\begin{exemplo}
O espa\c co de Cantor munido da medida de Haar, $\Omega=\{0,1\}^{\N_+}$, \'e a pot\^encia infinita enumer\'avel do espa\c co de Bernoulli, $\{0,1\}$, munido da medida de contagem normalizada. 
\end{exemplo}

\begin{exemplo} 
Se $\Omega$ \'e o intervalo unit\'ario $\I=[0,1]$ munido da medida de Lebesgue
$\lambda$, ent\~ao a pot\^encia finita $\I^n$, $n\in\N$, \'e o cubo de dimens\~ao $n$, e a medida produto $\lambda^{\otimes n}$ \'e a medida de Lebesgue de dimens\~ao $n$, denotada $\lambda^{(n)}$.
\end{exemplo}

\subsection{Realiza\c c\~oes de vari\'aveis aleat\'orias e independ\^encia}

J\'a discutimos brevemente a no\c c\~ao de independ\^encia de uma fam\'\i lia de vari\'aveis aleat\'orias na se\c c\~ao \ref{s:independencia}. Relembramos a defini\c c\~ao.

\begin{definicao}
Uma fam\'\i lia $(X_i)_{i\in I}$ de vari\'aveis aleat\'orias com valores respetivos em espa\c cos borelianos padr\~ao $\Omega_i$, $i\in I$ \'e {\em independente} se para cada cole\c c\~ao finita de \'\i ndices, $i_1,i_2,\ldots,i_k$, \'e cada cole\c c\~ao de conjuntos borelianos (ou mensur\'aveis), $A_1\in\Omega_{i_1},A_2\in\Omega_{i_2},\ldots,A_k\in\Omega_{i_k}$, temos
\begin{align*}
P(X\in A_1,X_{i_2}\in A_2,\ldots,& X_{i_k}\in A_k)
=\\
& P\left(X_{i_1}\in A_1)\right)\cdot\left(X_{i_2}\in A_2)\right)\cdot
\ldots\cdot P\left( X_{i_k}\in A_k)\right).
\end{align*}
\index{independ\^encia de vari\'aveis aleat\'orias}
\end{definicao}

Em outras palavras, a lei da vari\'avel aleat\'oria 
\[(X_i)_{i\in I}\in \prod_{i\in I}\Omega_i\]
\'e o produto $\otimes_{i\in I}\mu_i$ das leis $\mu_i$ de $X_i$.

A abordagem comum para tratar as vari\'aveis aleat\'orias da maneira matematicamente rigorosa e de falar das suas {\em realiza\c c\~oes}.

\begin{definicao}
Seja $X$ uma vari\'avel aleat\'oria tomando valores em um espa\c co boreliano padr\~ao $\Omega$, com a lei $\mu$. Uma {\em realiza\c c\~ao} de $X$ sobre um espa\c co probabil\'\i stico $(\mathfrak X,\nu)$ \'e qualquer fun\c c\~ao mensur\'avel $f\colon {\mathfrak X}\to\Omega$, tendo a propriedade
\[f_{\ast}(\nu)=\mu.\]
\index{realiza\c c\~ao de uma vari\'avel aleat\'oria}
\end{definicao}

\begin{exemplo}
Toda vari\'avel aleat\'oria $X\in\Omega$ possui a realiza\c c\~ao can\^onica que \'e uma aplica\c c\~ao de identidade:
\[\mbox{id}_\Omega\colon \Omega\ni x\mapsto x\in\Omega.\]
\end{exemplo}

\begin{observacao}
Temos que uma fam\'\i lia de v.a. $X_i\in\Omega$, $i\in I$ \'e independente se e somente se as vari\'aveis $X_i$ podem ser realizadas simultaneamente como proje\c c\~oes coordenadas 
\[\pi_i\colon \left(\prod_{i\in I}\Omega_i,\otimes_{i\in I}\mu_i\right)\to\Omega_i.\]
\end{observacao}

Frequentemente, nenhuma distin\c c\~ao \'e feita entre uma vari\'avel e sua realiza\c c\~ao, e a mesma letra $X$ est\'a usada para denotar a \'ultima.

A seguinte consequ\^encia imediata da exist\^encia da medida produto fornece uma fonte rica de vari\'aveis aleat\'orias independentes.

\begin{teorema}
Sejam $\mu_n$ medidas de probabilidade borelianas em espa\c cos borelianos padr\~ao $\Omega_n$, $n=1,2,3,\ldots$. Ent\~ao existe uma sequ\^encia independente de v.a. $X_1,X_2,X_3,\ldots$ tais que a lei de cada $X_n$ \'e $\mu_n$.
\qed
\end{teorema}

\begin{proof} 
Cada v.a. $X_n$ \'e realizada como a $n$-\'esima proje\c c\~ao coordenada, $\pi_n$, sobre o espa\c co probabil\'istico padr\~ao  $(\prod_{n=1}^{\infty}\Omega_n,\otimes_{n=1}^{\infty}\mu_n)$.
\end{proof}

\begin{observacao}
Uma das perguntas mais intrigantes para o autor --- e possivelmente, absurda --- \'e a seguinte. Pode-se dar o significado \`a frase ``os dados s\~ao modelados com uma sequ\^encia de vari\'aveis aleat\'orias livres no sentido de Voiculescu''?

As vari\'aveis aleat\'orias reais limitadas podem ser realizadas como fun\c c\~oes mensur\'aveis sobre um espa\c co probabil\'\i stico padr\~ao $(\Omega,\mu)$, ou seja, elementos da \'algebra comutativa de fun\c c\~oes $L^{\infty}(\mu)$.
As vari\'aveis aleat\'orias livres n\~ao s\~ao vari\'aveis aleat\'orias, mas sim, elementos de uma \'algebra de operadores {\em n\~ao comutativa} \citep*{VDN}.
\end{observacao}

\section{Parametriza\c c\~ao, regularidade, converg\^encia}

\subsection{Parametriza\c c\~ao de espa\c cos probabil\'\i sticos padr\~ao}

\begin{teorema}
Seja $(\Omega,\mu)$ um espa\c co probabil\'\i stico padr\~ao. Ent\~ao existe uma aplica\c c\~ao boreliana $f\colon [0,1]\to\Omega$ tal que a medida $\mu$ \'e a imagem direta da medida de Lebesgue (uniforme), $\lambda$, sobre o intervalo:
\[\mu=f_{\ast}(\lambda).\]
\label{t:parametrizacao}
\end{teorema}

\begin{observacao} 
A parametriza\c c\~ao \'e nunca \'unica: por exemplo, o intervalo admite um grupo $\mathrm{Aut}(\lambda)$ de automorfismos borelianos conservando a medida de Lebesgue, qual grupo \'e muito rico e interessante.
\end{observacao}

\begin{exemplo}
O espa\c co de Bernoulli $\Omega=\{0,1\}$ com $\mu(1)=p$, $\mu(0)=q$, $p+q=1$, pode ser parametrizado assim:
\[f=\chi_{[0,p]}.\]
\end{exemplo}

\begin{exercicio}
Mostre o teorema \ref{t:parametrizacao} no caso onde $\mu$ \'e puramente at\^omica. Deduza que ser\'a bastante mostrar o teorema para medidas $\mu$ n\~ao at\^omicas.
\label{ex:naoatomicabasta}
\end{exercicio}

\begin{lema}
Seja $f\colon A\to [0,1]$ uma fun\c c\~ao mon\'otona (ou seja, n\~ao crescente ou n\~ao decrescente), onde $A\subseteq [0,1]$ \'e um subconjunto boreliano. Mostre que $f$ \'e boreliana. Se $A=[0,1]$, ent\~ao a imagem de $f$ \'e um espa\c co boreliano padr\~ao.
\label{l:monotonaeisomorfismo}
\end{lema}

\begin{proof}
A imagem inversa de cada intervalo por $f$ \'e a interse\c c\~ao de um intervalo (aberto, semiaberto, ou fechado) com $A$, logo, um conjunto boreliano. Agora suponha que o dom\'\i nio de $f$ seja $[0,1]$. A fun\c c\~ao $f$ possui, ao m\'aximo, uma fam\'\i lia enumer\'avel de descontinuidades de salto, e a cada tal descontinuidade corresponde um intervalo removido do intervalo $[f(0),f(1)]$. Um intervalo (seja aberto, semiaberto, ou fechado) \'e um conjunto $F_{\sigma}$, ou seja, a uni\~ao de uma fam\'\i lia enumer\'avel de subconjuntos fechados de $[0,1]$. A sua uni\~ao \'e um $F_{\sigma}$ tamb\'em, portanto o complemento desta uni\~ao, que \'e exatamente a imagem de $f$, \'e um $G_\delta$. A topologia de um subconjunto $G_\delta$ de um espa\c co m\'etrico completo \'e gerada por uma m\'etrica completa (que e, claro, separ\'avel), segundo o teorema \ref{t:completeequivegdelta}, logo o mesmo se aplica \`a estrutura boreliana de $f([0,1])$.
\end{proof}

Seja $\Omega$ um espa\c co boreliano padr\~ao munido de uma medida de probabilidade $\mu$ n\~ao at\^omica. Segue-se que $\Omega$ tem cardinalidade $\mathfrak c$. Segundo teorema \ref{t:isomorfismo1}, existe um isomorfismo boreliano $\phi\colon \Omega\to [0,1]$. A imagem direta da medida $\mu$ por $\phi$ \'e uma medida de probabilidade n\~ao at\^omica, $\phi_{\ast}(\mu)$, sobre o intervalo $[0,1]$. A fun\c c\~ao de distribui\c c\~ao de $\phi_{\ast}(\mu)$, dada por 
\[\Phi(t) = \phi_{\ast}(\mu) [0,t) = \mu(\phi^{-1}[0,1)),\]
\'e cont\'\i nua e n\~ao decrescente, $\Phi(0)=0$, $\Phi(1)=1$.

\begin{exercicio}
Mostrar que a fun\c c\~ao 
\begin{align*} 
\Phi^{\leftarrow}\colon [0,1]&\to [0,1],\\
\Phi^{\leftarrow}(x) &= \min\{t\colon \Phi(t)\geq x\}
\end{align*}
\'e (estritamente) crescente (logo boreliana), e tem a propriedade
\[\Phi^{\leftarrow}_{\ast}(\lambda)= \phi_{\ast}(\mu).\]
[ {\em Sugest\~ao:} basta mostrar que para todo $t\in [0,1]$,
\[\Phi^{\leftarrow}_{\ast}(\lambda)[0,t)= \phi_{\ast}(\mu)[0,t).~~]\]
\label{ex:Phileftarrow}
\end{exercicio}

Por conseguinte, a aplica\c c\~ao boreliana $\phi^{-1}\circ\Phi^{\leftarrow}\colon [0,1]\to\Omega$ \'e uma parametriza\c c\~ao desejada do espa\c co probabil\'\i stico $(\Omega,\mu)$, no caso $\mu$ \'e n\~ao at\^omica. O teorema \ref{t:parametrizacao} segue-se deste resultado junto com exerc\'\i cio \ref{ex:naoatomicabasta}.

\begin{teorema}
Cada espa\c co boreliano padr\~ao, $\Omega$, munido de uma medida de probabilidade n\~ao at\^omica, $\mu$, \'e isomorfo ao intervalo fechado $[0,1]$ munido da medida de Lebesgue $\lambda$.
\label{t:unicidadedemedidanaoatomica}
\end{teorema}

\begin{observacao}
Aqui a palavra ``isomorfismo'' significa um isomorfismo entre dois espa\c cos probabil\'\i sticos padr\~ao, $f\colon(\Omega,\mu)\to(\Upsilon,\nu)$, ou seja, um isomorfismo boreliano $f$ entre subespa\c cos borelianos padr\~ao $\Omega^\prime$ e $\Upsilon^\prime$, respetivamente, cada um tendo a medida plena, tal que $f_{\ast}(\mu)=\nu$ e $(f^{-1})_{\ast}(\nu)=\mu$. (Apenas uma de duas condi\c c\~oes j\'a basta).
\end{observacao}

\begin{exercicio}
Mostre o teorema \ref{t:unicidadedemedidanaoatomica}.
\par
[ {\em Sugest\~ao:} mostre que a fun\c c\~ao $\Phi^{\leftarrow}$ no exerc\'\i cio \ref{ex:Phileftarrow}, \'e bijetora, logo (usando o lema \ref{l:monotonaeisomorfismo}) um isomorfismo boreliano entre $[0,1]$ e um subespa\c co boreliano padr\~ao do intervalo de medida um. ]
\end{exercicio}

\subsection{Regularidade de medidas\label{s:regularidade}}

\begin{teorema}
Toda medida boreliana de probabilidade, $\mu$, sobre um espa\c co m\'etrico $\Omega$ qualquer (n\~ao necessariamente completo ou separ\'avel) satisfaz a condi\c c\~ao de regularidade seguinte. Seja $A$ um subconjunto mensur\'avel de $\Omega$. Para todo $\ve>0$ existem um conjunto fechado, $F$, e um conjunto aberto, $U$, de $\Omega$ tais que
\[F\subseteq A\subseteq U\]
e
\[\mu(U\setminus F)<\ve.\]
\label{t:regularidadeFU}
\end{teorema}

\begin{proof}
Consideremos a fam\'\i lia $\mathcal A$ de subconjuntos $A$ de $\Omega$ que tem a propriedade na conclus\~ao do teorema: dado $\ve>0$, existem um conjunto fechado, $F$, e um conjunto aberto, $U$, de $\Omega$ com $F\subseteq A\subseteq U$ e $\mu(U\setminus F)<\ve$. \'E f\'acil verificar que $\mathcal A$ forma uma sigma-\'algebra (exerc\'\i cio), e cont\'em todos subconjuntos fechados (usa o fato que $\mu(F_\ve)\downarrow \mu(F)$ quando $\ve\downarrow 0$). Por conseguinte, $\mathcal A$ cont\'em todos os conjuntos borelianos.

Se $A$ \'e mensur\'avel, usamos o teorema \ref{t:perturb} para achar conjuntos borelianos $B,B^\prime$ tais que $B\subseteq A\subseteq B^\prime$ e $\mu(B^\prime\setminus B)=0$. Agora o resultado para conjuntos borelianos permite concluir.
\end{proof}

Para alguns matem\'aticos, a palavra ``regularidade'' significa uma condi\c c\~ao mais forte, como no resultado seguinte.

\begin{teorema}
Toda medida boreliana de probabilidade, $\mu$, sobre um espa\c co m\'etrico $\Omega$ completo e separ\'avel (ou seja, polon\^es), satisfaz a condi\c c\~ao de regularidade seguinte. Seja $A$ um subconjunto mensur\'avel de $\Omega$. Para todo $\ve>0$ existem um conjunto compacto, $K$, e um conjunto aberto, $U$, de $\Omega$ tais que
\[K\subseteq A\subseteq U\]
e
\[\mu(U\setminus K)<\ve.\]
\label{t:regularidadeKU}
\end{teorema}

Isso \'e uma consequ\^encia do resultado seguinte.

\begin{teorema}
Seja $\mu$ uma medida boreliana de probabilidade sobre um espa\c co m\'etrico $\Omega$ completo e separ\'avel. Para cada $\ve>0$, existe um subconjunto compacto $K\subseteq\Omega$ com $\mu(K)>1-\ve$.
\label{t:suporteK}
\end{teorema}

\begin{exercicio}
Mostrar que cada espa\c co m\'etrico separ\'avel, $\Omega$, tem a {\em propriedade de Lindel\"of:} toda cobertura aberta de $\Omega$ admite uma subcobertura enumer\'avel. ({\em Advert\^encia:} n\~ao \'e bastante de escolher uma subcobertura enumer\'avel de um conjunto enumer\'avel denso, ela pode n\~ao cobrir o espa\c co inteiro...)
\end{exercicio}

\begin{proof}[Prova do teorema \ref{t:suporteK}]
Seja $k\in\N_+$ qualquer.
Como $\Omega$ \'e separ\'avel, existe uma cobertura enumer\'avel de $\Omega$ com as bolas de raio $2^{-k}$:
\[\bigcup_{i=1}^{\infty}B_{2^{-k}}(x_i)=\Omega.\]
Por causa da sigma-aditividade de $\mu$,
\[\mu\left( \bigcup_{i=1}^{N}\bar B_{2^{-k}}(x_i)\right) \uparrow 1\mbox{ quando }N\to\infty,\]
e pode se escolher um $N_k$ tal que
\[\mu\left( \bigcup_{i=1}^{N_k}\bar B_{2^{-k}}(x_i)\right) >1-\frac{\ve}{2^k}.\]
O conjunto
\[K=\bigcap_{k=1}^{\infty} \bigcup_{i=1}^{N_k}\bar B_{2^{-k}}(x_i)\]
\'e pr\'e-compacto (qualquer que seja $k$, ele \'e coberto por uma fam\'\i lia finita de bolas fechadas de raio $2^{-k}$, logo por uma fam\'\i lia finita de bolas abertas de raio ligeiramente maior, digamos $2^{-k+1}$). Ao mesmo tempo, $K$ \'e fechado num espa\c co m\'etrico completo, logo compacto. Finalmente,
\[\mu(\Omega\setminus K)\geq 1 - \sum_{k=1}^{\infty}\mu\left(\Omega\setminus \bigcup_{i=1}^{N_k}\bar B_{2^{-k}}(x_i) \right) >1-\sum_{k=1}^{\infty}\frac{\ve}{2^k} = 1-\ve.
\]
\end{proof}

\begin{proof}[Demonstra\c c\~ao do teorema \ref{t:regularidadeKU}]
Sejam $F$ e $U$ como no teorema \ref{t:regularidadeFU}, e $K$ como no teorema \ref{t:suporteK}. O subconjunto $K\cap F$ de $A$ \'e compacto, e $\mu(U\setminus K\cap F)\leq 2\ve$.
\end{proof}

\begin{teorema}[Teorema de Luzin]
Seja $f$ uma fun\c c\~ao real mensur\'avel sobre um espa\c co m\'etrico completo e separ\'avel $\Omega$ munido de uma medida de probabilidade boreliana, $\mu$. Dado $\ve>0$, existe um subconjunto compacto $K\subseteq\Omega$ com $\mu(\Omega\setminus K)<\ve$ tal que a restri\c c\~ao $f\vert_{K}$ \'e cont\'\i nua.
\label{t:luzin}
\index{teorema! de Luzin}
\end{teorema}

\begin{proof} Escolhamos uma sequ\^encia som\'avel $(\ve_n)$, $\ve_n>0$, com $\sum_{n=1}^{\infty}\ve_n=\ve$.
Enumeremos a fam\'\i lia de todos os intervalos abertos com pontos extremos racionais: $(a_n,b_n)$, $n\in\N$. Para cada $n$, o conjunto $f^{-1}(a_n,b_n)$ \'e mensur\'avel, e segundo o teorema \ref{t:regularidadeKU},
existem os conjuntos compactos $K_n\subseteq f^{-1}(a_n,b_n)$ e $K^{\prime}_n$, $K^{\prime}_n\cap f^{-1}(a_n,b_n)=\emptyset$, e tais que sua uni\~ao, $\tilde K_n= K_n\cup K^\prime_n$, satisfaz $\mu(\tilde K_n)>\ve_n$. 
Definamos o compacto 
\[K=\bigcap_{n=1}^{\infty} \tilde K_n.\]
Segue-se que $\mu(K)>1-\sum_{n=1}^{\infty}\ve_n=1-\ve$. 
Para cada $n$, o conjunto $f^{-1}(a_n,b_n)\cap K$ \'e aberto em $K$ porque o seu complemento,
\[K\setminus f^{-1}(a_n,b_n) = K^{\prime}_n\cap K,\]
\'e fechado em $K$. 
\end{proof}

(A prova curta acima pertence a Erik Talvila and Peter Loeb).

\begin{observacao} 
Diz-se que uma fun\c c\~ao mensur\'avel \'e ``quase cont\'\i nua''. Para os espa\c cos com medida mais gerais (n\~ao separ\'aveis), a defini\c c\~ao acima \'e usada como a defini\c c\~ao de uma fun\c c\~ao mensur\'avel (fun\c c\~oes mensur\'aveis no sentido de Bourbaki).
\end{observacao}

\begin{observacao}
\'E claro que o resultado generaliza-se para as fun\c c\~oes com valores num espa\c co m\'etrico separ\'avel qualquer, ou mesmo num espa\c co topol\'ogico que tem uma base enumer\'avel de conjuntos abertos.
\end{observacao}

\subsection{Converg\^encia\label{s:quasecerta}}

\begin{definicao} Uma sequ\^encia $X_i,i=1,2,\cdots$ de vari\'aveis aleat\'orias com valores em um espa\c co m\'etrico $(\Omega,d)$ converge {\em em probabilidade} para uma v.a. $X$ se 
\begin{equation}
\forall\e>0,~~P\left(d(X_n,X)\geq\e\right)
\to 0\mbox{ quando }n\to\infty.
\label{sequence}
\end{equation}
Nota\c c\~ao:
\[X_n\overset{{p}}\longrightarrow X.\]

\index{converg\^encia! em probabilidade}
\end{definicao}

\begin{definicao}
\label{convas}
Uma sequ\^encia $X_i,i=1,2,\cdots$ de vari\'aveis aleat\'orias com valores em um espa\c co m\'etrico $(\Omega,d)$ converge {\em quase certamente} para uma v.a. $X$ se 
\begin{equation}
\label{as}
P\left(\lim_{n\to\infty} X_n=X\right)=1.
\end{equation}
Nota\c c\~ao:
\[X_n\overset{{a.s.}}\longrightarrow X.\]

\index{converg\^encia! quase certa}
\end{definicao}

\begin{observacao}
Como interpretar a probabilidade do evento na \'ultima defini\c c\~ao? Se realizarmos todas as vari\'aveis $X_n,X$ sobre um espa\c co probabil\'\i stico qualquer, $(\mathfrak X,\nu)$, ent\~ao trata-se da medida $\nu$ do conjunto
\[\{\omega\in\mathfrak X\colon \lim_{n\to\infty} X_n(\omega)
=X(\omega)\}.\]
Sen\~ao, trata-se da medida do conjunto de todas as sequ\^encias $(x,x_1,x_2,\ldots,x_n,\ldots)$ satisfazendo
\[\lim_{n\to\infty}x_n=x\]
em rela\c c\~ao \`a medida da lei conjunta de v.a. $X,X_1,X_2,\ldots,$ sobre o espa\c co $\Omega^{\infty}\times\Omega$. Este espa\c co probabil\'\i stico realiza todas as vari\'aveis aleat\'orias implicitamente, e cada outra realiza\c c\~ao delas vai fatorizar atrav\'es deste espa\c co.
\end{observacao}

\begin{exercicio}
Mostre que o conjunto acima \'e boreliano.
\end{exercicio}

\begin{definicao}
Uma sequ\^encia $(X_n)$ de vari\'aveis aleat\'orias converge para uma vari\'avel aleat\'oria $X$ quase certamente se
\[P[X_n\to X]=1.\]
\end{definicao}

\begin{proposicao}
A converg\^encia quase certa implica a converg\^encia em probabilidade. Dizendo de outra forma, seja $X_i,i=1,2,\cdots$ \'e uma sequ\^encia de vari\'aveis aleat\'orias com valores em um espa\c co m\'etrico $(\Omega,d)$, e seja $X$ \'e uma v.a. com valores em $\Omega$. Ent\~ao a condi\c c\~ao
\[X_n\overset{{a.s.}}\longrightarrow X\]
implica
\[X_n\overset{{p}}\longrightarrow X.\]
\end{proposicao}

\begin{proof} 
Realizemos todas as v.a. $X_n,X$ sobre um espa\c co probabil\'\i stico qualquer, $(\mathfrak X,\nu)$.
Fixemos $\e>0$. Para cada $N\in\N$, denotemos
\[A_{N,\e}=\{\omega\in\mathfrak X\colon d(X_n(\omega),X(\omega))<\e
\mbox{ para todos } n\geq N\}.\]
A hip\'otese que $X_n$ converge para $X$ quase certamente implica
\[P\left(\cup_{N=1}^\infty A_{N,\e} \right) =1,\]
e como os conjuntos $A_{N,\e}$ para $\e$ fixo e $N$ indo para infinito formam uma sequ\^encia encaixada crescente,
\[P(A_{N,\e})\uparrow 1.\]
Como isto \'e verdadeiro para cada $\e>0$, conclu\'\i mos que $X_n$ convergem em probabilidade para $X$.
\end{proof}

O converso n\~ao \'e verdadeiro.

\begin{observacao}
Um {\em evento} \'e uma vari\'avel aleat\'oria bin\'aria. Neste sentido, podemos falar da independ\^encia de uma fam\'\i lia de eventos. Tipicamente, um evento \'e da forma $[X\in A]$, onde $X$ \'e uma v.a. e $A\subseteq\Omega$ um subconjunto boreliano. A fun\c c\~ao indicadora $\chi_A$ uma realiza\c c\~ao deste evento. Neste sentido, pode-se falar da independ\^encia de uma fam\'\i lia $(A_i)$ de subconjuntos mensur\'aveis de um espa\c co probabil\'\i stico padr\~ao, $({\mathfrak X},\nu)$. Aplicando a defini\c c\~ao da independ\^encia, conclu\'\i mos que isso significa o seguinte: para toda cole\c c\~ao finita de \'\i ndices, $i_1,i_2,\ldots,i_k$,
\[\mu(A^\ast_{i_1}\cap A^\ast_{i_2}\cap\ldots \cap A^\ast_{i_k})
= \prod_{j=1}^k \mu(A^\ast_{i_j}),\]
onde $A^\ast_i$ significa seja $A_i$, seja o seu complemento.
\end{observacao}

\begin{exercicio}
Mostrar o {\em 2o lema de Borel--Cantelli.} 
Seja $(B_n)$ uma sequ\^encia de eventos {\em independentes} tais que a soma de probabilidades deles \'e infinita:
\[\sum_{i=1}^{\infty} P(B_n)<+\infty.\]
Ent\~ao, quase certamente, um n\'umero infinito destes eventos ocorre:
\[P\left[\forall n~\exists m\geq n~B_m\right]=1.\]
 Na nota\c c\~ao conjunt\'\i stica: se $B_1,B_2,\ldots$ s\~ao independentes e $\sum_{i=1}^{\infty}\mu(B_n)=+\infty$, ent\~ao
\[\mu \left(\bigcap_{n=1}^{\infty}\bigcup_{i=n}^{\infty} B_n\right) =1.\]
[ {\em Sugest\~ao:} a probabilidade que s\'o $n$ deles ocorrem \'e majorada pela probabilidade que todos eventos $\bar B_i$ complementares a $B_i$ ocorrem para $i\geq N$, e agora pode-se usar a independ\^encia de $\bar B_i$ e a desigualdade \'util $1-x\leq e^{-x}$ para $x\geq 0$... ]
\index{lema! de Borel--Cantelli, segunda}
\end{exercicio}

\begin{exercicio}
Construir uma sequ\^encia de subconjuntos borelianos $B_n$ do intervalo $[0,1]$ munido da medida de Lebesgue $\lambda$ tais que $\lambda(B_n)= 1/n$ e $B_n$ s\~ao independentes. Concluir que as fun\c c\~oes indicadores $\chi_{B_n}$ convergem para zero em medida (e na dist\^ancia $L^1$), mas para cada $x$, $\limsup_{n\to\infty}\chi_{B_n}(x)=1$, ent\~ao a sequ\^encia diverge quase em todo ponto.
\end{exercicio}

Portanto, tem algumas situa\c c\~oes onde a converg\^encia em probabilidade, ou em esperan\c ca, implica a converg\^encia quase certa. Na aprendizagem estat\'\i stica, uma destas situa\c c\~oes \'e particularmente \'util: isso ocorre quando as vari\'aveis aleat\'orias s\~ao altamente concentradas.

\begin{teorema}[Lei Forte dos Grandes N\'umeros]
Sejam $(\Omega,\mu)$ um espa\c co probabil\'\i stico padr\~ao, e $U$ um subconjunto mensur\'avel. Quase certamente, para um caminho amostral aleat\'orio $\varsigma\sim \mu^{\infty}$, a medida emp\'\i rica de $U$ com rela\c c\~ao ao segmento inicial de $\varsigma$ converge para o valor da medida de $U$:
\[\mu_{\varsigma_n}(U)\to \mu(U).\]
\label{l:munmu}
\index{lei dos grandes n\'umeros! forte}
\end{teorema}

\begin{proof}
Como $\varsigma_n$ \'e um elemento aleat\'orio de $\Omega^n$ com a distribui\c c\~ao $\mu^{\otimes n}$, a lei de grandes n\'umeros implica 
\begin{eqnarray*}
P[\mu_{\varsigma_n}(U)\to \mu(U)] &\geq &
P\left[\exists N~\forall m~\left\vert\mu_{\varsigma_{N+m}}(U)- \mu(U)\right\vert\leq \frac 1m\right] \\
&\geq & \sup_{N} 
P\left[\forall m~\left\vert\mu_{\varsigma_{N+m}}(U)- \mu(U)\right\vert\leq \frac 1m\right] \\
&\geq & \sup_{N} \left(1-2\sum_{m=1}^{\infty}e^{-2\ve^2(N+m)} \right)\\
&=& \sup_{N} \left(1-2\frac{e^{-2\ve^2N}}{1-e^{-2\ve^2}} \right)\\
&=& 1.
\end{eqnarray*}
(Acima, $P$ \'e a medida $\mu$). 
\end{proof}

\begin{exercicio}
Mostrar o {\em 1o lema de Borel--Cantelli.} 
Seja $(B_n)$ a sequ\^encia de eventos (n\~ao necessariamente independentes) tais que a soma de probabilidades deles \'e finita:
\[\sum_{i=1}^{\infty} P(B_n)<+\infty.\]
Ent\~ao, a probabilidade da ocorr\^encia de um n\'umero infinito destes eventos \'e zero:
\[P\left[\forall n~\exists m\geq n~B_m\right]=0.\]
 Na nota\c c\~ao conjunt\'\i stica: se $\sum_{i=1}^{\infty}\mu(B_n)<+\infty$, ent\~ao
\[\mu \left(\bigcap_{n=1}^{\infty}\bigcup_{i=n}^{\infty} B_n\right) =0.\]
[ {\em Sugest\~ao:} o que pode se dizer da s\'erie $\sum_{i=n}^{\infty}\mu(B_n)$? ]
\label{ex:1lemaborel-cantelli}
\index{lema! de Borel--Cantelli, primeira}
\end{exercicio}

O resultado seguinte \'e uma varia\c c\~ao sobre a prova do teorema \ref{l:munmu}.

\begin{lema}
Seja $(\Omega,\mu)$ um espa\c co probabil\'\i stico padr\~ao, e seja $(f_n)$ uma sequ\^encia de fun\c c\~oes reais mensur\'aveis sobre $\Omega$ satisfazendo uma concentra\c c\~ao gaussiana uniforme: para todo $\ve>0$ e todo $n$,
\[\mu\{x\in\Omega\colon \abs{f_n-\E_\mu f_n}>\ve\}\leq Ce^{-c\ve^2n}.\]
Se $f_n\to f$ em probabilidade, ent\~ao $f_n\to f$ quase em toda parte (e $f$ \'e uma fun\c c\~ao constante quase em toda parte).
\label{l:prob+conc->qs}
\end{lema}

\begin{proof}
Exerc\'\i cio.
\end{proof}

%
%

\chapter{Teoremas de Hahn--Banach e de Stone--Weierstrass\label{ch:norma}}

\section{Espa\c cos normados e espa\c cos de Banach\label{s:normados}}

\subsection{Espa\c cos normados}
\begin{definicao}
Seja $E$ um espa\c co vetorial (real ou complexo).
Uma {\em norma} sobre $E$ \'e uma fun\c c\~ao real
\[N\colon E\to\R\]
que verifica as propriedades seguintes: 
\begin{enumerate}
\item $N(x)=0$ se e somente se $x=0$.
\item $N(\lambda x)=\abs\lambda N(x)$ para cada $x\in E$ e cada $\lambda\in\R$ (ou $\lambda\in\C$).
\item $N(x+y)\leq N(x)+N(y)$.
\end{enumerate}
\label{d:normaN}
\index{espa\c co! normado}
\end{definicao}

\begin{definicao}
Um espa\c co vetorial $E$ munido de uma norma \'e dito {\em espa\c co normado.}
\end{definicao}

\begin{observacao}
Geralmente, o valor de norma de um elemento $x\in E$ \'e denotado 
$\norm x$ ao inv\'es de $N(x)$.
\end{observacao}

\begin{exemplo}
Seja $\Gamma$ um conjunto qualquer. 
O espa\c co vetorial $\ell^\infty(\Gamma)$ 
\'e um espa\c co normado. 
\end{exemplo}

\begin{exemplo}
A reta $\R$ com a sua norma usual
\[\norm x =\abs x\]
\'e um espa\c co normado da dimens\~ao $1$. 
\end{exemplo}

\begin{exemplo}
A norma euclideana padr\~ao sobre o espa\c co vetorial $\R^n$, $n\in\N$, \'e dada por
\[\norm x_2 =\sqrt{\sum_{i=1}^n \abs{x_n}^2}.\]
Com efeito, temos j\'a mostrados que $\norm\cdot_2$ \'e uma norma: a propriedade (3) \'e exatamente a desigualdade de Minkowski.

O espa\c co $\R^n$ munido da norma euclideana $\norm\cdot_2$ acima \'e denotado $\ell^2(n)$.
\end{exemplo}

\begin{exemplo}
O espa\c co $\ell^2$ consiste de todas as sequ\^encias de escalares (reais ou complexas) som\'aveis com quadrado:
\[\sum_{i=1}^{\infty}\abs{x_i}<\infty,\]
munido da norma
\[\norm{x}_2 = \sqrt{\sum_{i=1}^{\infty}\abs{x_i}}.\]
Mostre que $\ell^2$ \'e um espa\c co normado, desenvolvendo uma vers\~ao da desigualdade de Minkowski e imitando a prova no caso de dimens\~ao finita. A norma acima \'e gerada pelo produto escalar
\[\langle x,y\rangle = \sum_{i=1}^{\infty} x_iy_i.\]
\par
Um pouco mais geralmente, pode-se definir o espa\c co $\ell^2(\Gamma)$ a partir de qualquer conjunto $\Gamma$. Por exemplo,  $\ell^2(n)$ \'e $\ell^2([n])$, enquanto $\ell^2$ \'e $\ell^2(\N_+)$. Os vetores $e_\gamma$, $\gamma\in\Gamma$ formam uma base ortonormal can\^onica do espa\c co, onde
\[(e_{\gamma})_\delta = \begin{cases} 1,&\mbox{ se }\gamma=\delta,\\
0,&\mbox{ caso contr\'ario.}
\end{cases}
\]
\label{ex:ell2}
\end{exemplo}

\begin{observacao}
Seja $E=(E,\norm\cdot)$ um espa\c co normado. Verifica-se facilmente que a formula 
\[d(x,y)=\norm{x-y}\]
define uma m\'etrica sobre $E$.
\end{observacao}

\begin{exercicio}
\label{ex:continuas}
Seja $E$ um espa\c co m\'etrico qualquer.
Verificar que a adi\c c\~ao,
\[E\times E\ni (x,y)\mapsto x+y\in E,\]
todo como a multiplica\c c\~ao escalar,
\[\K\times E\ni (\lambda, x)\mapsto \lambda x\in E,\]
s\~ao aplica\c c\~oes cont\'\i nuas.
\end{exercicio}

\subsection{Espa\c cos de Banach}

\begin{definicao}
Um espa\c co normado $E$ \'e dito {\em espa\c co de Banach} se $E$ \'e completo como um espa\c co m\'etrico. 
\index{espa\c co! de Banach}
\end{definicao}

\begin{exemplo}
  $\R$ \'e um espa\c co de Banach de dimen\c c\~ao real $1$.
\end{exemplo}

\begin{exercicio}
Mostrar que os espa\c cos $\ell^2(n)$ e $\ell^2$ s\~ao espa\c cos de Banach.
\end{exercicio}

\begin{exemplo} 
$\ell^{\infty}(\Gamma)$ \'e um espa\c co de Banach.
\end{exemplo}

\begin{observacao}
  Pode-se mostrar que cada espa\c co normado de dimens\~ao finita \'e um espa\c co de Banach.
\end{observacao}

\begin{exemplo} 
O espa\c co $C(X)$, onde $X$ \'e um compacto, \'e um espa\c co de Banach.
\end{exemplo}

\begin{exemplo}
Denotaremos $c_{00}$ o conjunto de todas as sequ\^encias $(x_n)$ dos n\'umeros reais cuja todos os membros, excepto um n\'umero finito, s\~ao iguais ao zero:
\[\{n\colon x_n\neq 0\}\mbox{ \'e finito.}\]
O subespa\c co vetorial $c_{00}$ do espa\c co de Banach $\ell^\infty$, munido da norma (e dist\^ancia) induzida, n\~ao \'e completo. 
\end{exemplo}

\subsection{Aplica\c c\~oes lineares}

\begin{teorema}
Seja $f\colon E\to F$ uma aplica\c c\~ao linear entre dois espa\c cos normados. As condi\c c\~oes seguintes s\~ao equivalentes.
\begin{enumerate}
\item\label{lincont:1} $f$ \'e cont\'\i nua.
\item \label{lincont:2} $f$ \'e cont\'\i nua em zero.
\item \label{lincont:3} $f$ \'e limitada sobre a bola fechada $\bar B_1(0)$.
\item \label{lincont:4} Existe $K\geq 0$ tal que para todos $x\in E$ temos
\[\norm{f(x)}_F\leq K\cdot\norm{x}_E.\]
\item \label{lincont:5} $f$ \'e Lipschitz cont\'\i nua.
\item \label{lincont:6} $f$ \'e uniformemente cont\'\i nua.
\end{enumerate}
\label{t:aplicacoeslimitadas}
\end{teorema}

\begin{proof}
As implica\c c\~oes (\ref{lincont:1})$\Rightarrow$(\ref{lincont:2}), (\ref{lincont:5})$\Rightarrow$(\ref{lincont:6}), e (\ref{lincont:6})$\Rightarrow$(\ref{lincont:1}) s\~ao triviais.

(\ref{lincont:2})$\Rightarrow$(\ref{lincont:3}): Suponhamos que $f$ \'e cont\'\i nua em $0$. Existe $\delta>0$ tal que, para todos $x\in E$,
\[(\norm x_E<\delta)\Rightarrow (\norm{f(x)}_F<1).\]
Seja $y\in \bar B_\e(0)$. Ent\~ao $\norm y\leq 1$, e pois
\[\left\Vert \frac 2{\delta}y\right\Vert \leq  \frac 2{\delta}<\delta,\]
temos
\[\left\Vert f\left( \frac 2{\delta}y\right)\right\Vert <1,\]
de onde
\begin{align*}
\norm{f(y)} &=\left\Vert f\left(\frac{\delta}2\cdot\frac 2{\delta}y\right)\right\Vert \\
&= \frac{\delta}2\left\Vert f\left(\frac 2{\delta}y\right)\right\Vert \\
&<\frac\delta 2.
\end{align*}

(\ref{lincont:3})$\Rightarrow$(\ref{lincont:4}): Suponha que existe $K\geq 0$ tal que para todos $y\in\bar B_1(0)$ temos $\norm{f(y)}\leq K$. Seja $x\in E$ qualquer. Temos:
\[\left\Vert \frac x{\norm x}\right\Vert = \frac 1{\norm x}\norm x = 1,\]
ou seja,
\[\frac x{\norm x}\in\bar B_1(0),\]
logo
\begin{align*}
\norm x &= \left\Vert \norm x\frac{f(x)}{\norm x}\right\Vert \\
&= \norm x\cdot \left\Vert f\left(\frac x{\norm x}\right)\right\Vert \\
&\leq  \norm x\cdot K.
\end{align*}

(\ref{lincont:4})$\Rightarrow$(\ref{lincont:5}): Suponhamos que existe $K\geq 0$ tal que para todos $x\in E$,
\[\norm{f(x)}_F\leq K\cdot\norm{x}_E.\]
Se $x,y\in E$, ent\~ao
\begin{align*}
d_F(f(x),f(y)) &= 
\norm{f(x)-f(y)} \\ &=  \norm{f(x-y)} \\
&\leq  K\norm{x-y} \\
&= Kd_E(x,y).
\end{align*}
\end{proof}

\begin{corolario}
Uma aplica\c c\~ao linear $f$ entre dois espa\c cos normados \'e cont\'\i nua se e somente se a restri\c c\~ao de $f$ sobre a bola unit\'aria $B=B=\bar B_1(0)$ de $E$ \'e cont\'\i nua.
\label{c:continuassenabola}
\end{corolario}

\begin{definition} 
Digamos que uma aplica\c c\~ao linear $f\colon E\to F$ entre dois espa\c cos normados \'e {\em limitada} se ela satisfaz a condi\c c\~ao equivalente (\ref{lincont:4}) no teorema \ref{t:aplicacoeslimitadas}.
O menor valor $K\geq 0$ como em (\ref{lincont:4}) \'e chamado a {\em norma} da aplica\c c\~ao $f$ (ou: {\em norma operadora}) e notado $\norm f$. Dito de outro modo,
\begin{equation}
\norm f =\inf\{K\geq\colon  f(B_1(0))\subseteq B_K(0)\}.
\end{equation}
\label{d:normaoperadora}
\end{definition}

\begin{exercicio}
Sejam $E$ e $F$ dois espa\c cos normados. Mostre que a norma operadora \'e uma norma sobre o espa\c co linear $B(E,F)$ de todas as aplica\c c\~oes lineares limitadas de $E$ para $F$, munido com as opera\c c\~oes naturais.
\label{ex:b(ef)}
\end{exercicio}

\begin{exercicio}
Mostre que, se $F$ \'e um espa\c co de Banach, ent\~ao o espa\c co $B(E,F)$ \'e um espa\c co de Banach tamb\'em.
\end{exercicio}

Relembramos que uma aplica\c c\~ao linear com valores no corpo de escalares \'e dita {\em funcional linear}.

\begin{exemplo}
O funcional linear $\delta_0\colon C[0,1]\to\R$,
definido por
\[\delta_0(f) = f(0)\]
(e chamado a {\em fun\c c\~ao delta de Dirac}), \'e limitado, com $\norm{\delta_0}=1$.
\end{exemplo}

\begin{exemplo}
O funcional linear $I\colon C[0,1]\to\R$, 
definido por
\[I(f)=\int_0^1f(x)~dx,\]
\'e limitado, com $\norm I=1$. 
\end{exemplo}

\begin{exercicio}
Mostre que cada aplica\c c\~ao linear $f$ do espa\c co $\ell^\infty(n)$ com valores num espa\c co normado $F$ qualquer \'e limitado. Combine esta oberva\c c\~ao com proposi\c c\~ao \ref{p:fermee} e corol\'ario \ref{c:continuassenabola} para mostrar que cada aplica\c c\~ao linear de um espa\c co normado de dimens\~ao finita para um espa\c co normado $F$ qualquer \'e limitada.
\end{exercicio}

Os exemplos conduzem-nos de maneira natural \`a pergunta seguinte. Seja $E$ um espa\c co normado de dimens\~ao infinita qualquer. Existem sempre os funcionais lineares limitados sobre $E$? A resposta positiva \'e dada pelo importante {\em teorema de Hahn-Banach}, uma das pedras angulares da An\'alise Funcional.

\section{Teorema de Hahn--Banach}

\subsection{Formula\c c\~ao do teorema de Hahn--Banach}

\begin{teorema}[Teorema de Hahn--Banach]
\label{t:hb}
Sejam $E$ um espa\c co normado e $F$ um subespa\c co vetorial de $E$. Seja $\phi\colon F\to\R$ um funcional linear limitado qualquer. Ent\~ao existe um funcional linear limitado $\tilde\phi\colon E\to\R$ que estende $\phi$,
\[\tilde\phi\vert_F=\phi,\]
e tem a mesma norma,
\[\norm{\tilde\phi}=\norm{\phi}.\]
\index{teorema! de Hahn--Banach}
\end{teorema}

Como um corol\'ario imediato, funcionais lineares limitados s\~ao n\'umerosos sobre cada espa\c co normado.

\begin{corolario}
Sejam $E$ um espa\c co normado qualquer e $x\in E$ um vetor, $x\neq 0$. Ent\~ao existe um funcional linear limitado $\phi$ sobre $E$ tal que
\begin{itemize}
\item $\norm\phi=1$, e
\item $\phi(x)=\norm x$. 
\end{itemize}
(Diz-se que $\phi$ {\em separe $x$ de zero}.)
\label{c:separacaodepontos}
\end{corolario}

\begin{proof}
Formemos 
\[\R x =\{\lambda x\colon \lambda\in\R\}\]
um subespa\c co linear de $E$ de dimens\~ao um. Seja $\psi\colon \R x\to \R$ um funcional dado pela formula seguinte:
\[\psi(\lambda x )=\lambda\norm x.\]
Este funcional \'e visivelmente linear. Assim, temos 
\[\abs{\psi(\lambda x)} =\abs\lambda\cdot\norm x =\norm{\lambda x},\]
logo $\psi$ \'e limitado, da norma um.

De acordo com o teorema de Hahn-Banach \ref{t:hb}, existe uma extens\~ao,  $\phi=\tilde\psi$, de $\psi$ sobre $E$ da norma um. Em particular,
\[\phi(x)=\psi(x)=\norm x.\]
\end{proof}

\subsection{Demonstra\c c\~ao no caso da codimens\~ao um}
Suponha que $\dim E/F=1$,
em outras palavras, existe um vetor $x\in E\setminus F$ tal que o conjunto $F\cup\{x\}$ gera $E$ com um espa\c co vetorial.

\begin{lema}
\label{l:codim}
Sejam $G$ um subespa\c co vetorial de um espa\c co normado $E$, $g$ um funcional linear limitado sobre $F$, e $x\in E$. Ent\~ao existe um funcional linear limitado $g_1$ sobre o espa\c co vetorial
\[G+\R x = \{y+tx\colon y\in G,t\in\R\},\]
gerado pela uni\~ao $G\cup\{x\}$, cuja restri\c c\~ao sobre $G$ \'e igual a $g$ e tal que $\norm{g_1}=\norm g$.
\end{lema}

\begin{proof}
Para cada $y\in G$ definamos
\[\varphi(y) =\norm g\cdot \norm{x+y}-g(y),\]
\[\psi(y) =-\norm g\cdot \norm{x+y}-g(y).\]
Qualquer que sejam $y,z\in G$, temos
\begin{align*}
\varphi(y)-\psi(z) &= \norm g\left(\norm{x+y}+\norm{x+z}\right)-g(y-z) \\
&= \norm g\left(\norm{x+y}+\norm{-x-z}\right)-g(y-z) \\
&\geq \norm g\left(\norm{x+y-x-z}\right)-g(y-z) \\
&= \norm g\cdot\norm{y-z}-g(y-z)\\
&\geq 0,
\end{align*}
de acordo com a defini\c c\~ao da norma $\norm g$. Por conseguinte, 
\[\inf_{y\in G}\varphi(y)\geq\sup_{z\in G}\phi(z).\]
Escolhamos $a\in\R$ de tal modo que
\[\forall y\in G~~\varphi(y)\geq a \geq\psi(y).\]
Este $a$ ser\'a a imagem de $x$ por $g_1$. Definamos para todos $t\in\R$ e $y\in G$
\[g_1(y+tx) = g(y)+ta.\]
Evidentemente, $g_1$ \'e uma forma linear sobre $G+\R x$. \'E \'obvio que $g_1\vert_G=g$. A fim de verificar que a norma do funcional $g_1$ n\~ao excede a de $g$, observemos que para cada $t\neq 0$,
\begin{align*}
\abs{g(y)+ta} &=\abs t\cdot \left\vert f\left(\frac yt\right) +a\right\vert \\
&\leq  \abs t\cdot \norm g\cdot\left\Vert \frac yt+a\right\Vert \\
&= \norm g\cdot \norm{y+tx}.
\end{align*}
\end{proof}

\subsection{Demonstra\c c\~ao no caso geral}

Denotemos $\mathfrak X$ a cole\c c\~ao de todos os pares $(V,\psi)$, onde
\begin{itemize}
\item $V$ \'e um subespa\c co vetorial de $E$ tal que $F\subseteq V\subseteq E$,
\item $\psi\colon V\to\R$ \'e um funcional linear limitado tal que
\begin{itemize}
\item $\norm{\psi}\leq\norm{\phi}$,
\item $\psi\vert_F = \phi$.
\end{itemize}
\end{itemize}

Digamos que $(V,\psi)\leq(Z,\zeta)$ se e somente se
\begin{itemize}
\item $V\subseteq Z$, e
\item $\zeta\vert_V=\psi$.
\end{itemize}

A rela\c c\~ao $\leq$ acima \'e uma rela\c c\~ao de ordem parcial sobre $\mathfrak X$. A fam\'\i lia $\mathfrak X$ n\~ao \'e vazia: ela cont\'em o par $(F,\phi)$. 

Verifiquemos que o conjunto $\mathfrak X$ \'e intuitivo. Seja $\mathfrak C$ uma parte de $\mathfrak X$ (n\~ao vazia) totalmente ordenada qualquer. 
Definamos
\[V = \cup\{W\colon \exists \psi,~~(W,\psi)\in {\mathfrak X}\}.\]
\'E f\'acil a ver que o subconjunto $V$ de $E$ \'e um subespa\c co vetorial.
Sejam $x,y\in V$.  Existem dois pares $(V,\psi),(Z,\zeta)\in {\mathfrak C}$ tais que $x\in V$ e $Y\in Z$. Porque $\mathfrak C$ \'e totalmente ordenado, um dos dois \'e verdadeiro: ou $(V,\psi)\leq (Z,\zeta)$, ou $(Z,\zeta)\leq (V,\psi)$. Suponhamos sem perda de generalidade o primeiro. Logo $x,y\in Z$ e, como $Z$ \'e um subespa\c co vetorial de $E$, temos para todos $\lambda,\mu\in\R$
\[\lambda x+\mu y\in Z\subseteq V.\]
Porque $V\supset F\ni 0$, conclu\'\i mos.

Para cada $x\in V$ definamos
\[\upsilon(x) = \psi(x),\]
onde $\psi$ \'e um funcional linear tal que existe $W$ com as propriedades $x\in W$ e $(W,\psi)\in {\mathfrak C}$. 
Esta defini\c c\~ao n\~ao depende da escolha do par $(W,\psi)$. Seja $(Z,\xi)\in{\mathfrak C}$ tal que $x\in W$. Porque $(W,\psi)$ e $(Z,\xi)$ s\~ao compar\'aveis, podemos supor que $(W,\psi)\leq (Z,\xi)$. Ent\~ao $\xi\vert_W=\psi$, e porque $x\in W$,
\[\xi(x)=\psi(x).\]

Sejam $x,y\in V$, $\lambda,\mu\in\R$. Existem $(W,\psi)$ e $(Z,\xi)$ em $\mathfrak X$ tais que $x\in W$ e $y\in Z$. Como $(W,\psi)$ e $(Z,\xi)$ s\~ao compar\'aveis, podemos supor sem perda de generalidade que $(W,\psi)\leq (Z,\xi)$. Isso significa que $x,y\in Z$ e, porque $Z$ \'e um subespa\c co vetorial de $E$ e $\xi$ \'e linear sobre $Z$, temos
\[\upsilon(\lambda x+\mu y)=\xi(\lambda x+\mu y)=\lambda\xi(x)+\mu\xi(y)=
\lambda\upsilon(x)+\mu\upsilon(y).\]
Conclu\'\i mos: o funcional $\upsilon$ sobre $V$ \'e linear. 

Se $x\in F$, logo, de acordo com a defini\c c\~ao de $V$ temos $x\in V$, e de acordo com a defini\c c\~ao do funcional $\upsilon$, temos
\[\upsilon(x)=\phi(x),\]
onde conclui-se:
\[\upsilon\vert_F=\phi.\]

Finalmente, seja $x\in V$ qualquer. Existe um par $(W,\psi)\in {\mathfrak X}$ tal que $x\in W$. Portanto,
\[\abs{\upsilon(x)}=\abs{\psi(x)}\leq \norm\psi\cdot\norm x\leq \norm\phi\cdot\norm x.\]
Conclu\'\i mos: o funcional $\upsilon$ \'e limitado, e a sua norma satisfaz:
\[\norm{\upsilon}\leq\norm{\phi}.\]

O par $(V,\upsilon)$ pertence a $\mathfrak X$, e evidentemente, para todos $(W,\psi)\in {\mathfrak C}$, temos
\[(W,\psi)\leq (V,\upsilon).\]
O subconjunto totalmente ordenado $\mathfrak C$ \'e portanto majorado. 

Gra\c cas ao lema de Zorn, existe um elemento maximal, $(W,\psi)$, em $\mathfrak X$. Suponha por absurdo, que $W\neq E$. Ent\~ao existe $x\in E\setminus W$. O espa\c co vetorial $W+\R x$ \'e estritamente maior que $W$, e segundo o lema \ref{l:codim} existe um funcional linear $\tilde\psi$ sobre $W+\R x$ que estende $\psi$ e cuja norma n\~ao excede a de $\psi$ (portanto, a de $\phi$). Por conseguinte, o par $(W+\R x,\tilde\psi)$ \'e contida em $\mathfrak X$, e ele \'e estritamente maior do que $(W,\psi)$, uma contradi\c c\~ao. Ent\~ao, $W=E$, e o funcional linear 
$\tilde\phi=\psi$ \'e uma extens\~ao desejada de $\phi$ sobre $E$. \qed

\section{Teorema de Stone--Weierstrass}

\subsection{Formula\c c\~ao do teorema}

\begin{teorema}[Teorema de Stone--Weierstrass]
  Seja $X$ um espa\c co m\'etrico compacto qualquer, e seja $A$ um conjunto de fun\c c\~oes cont\'\i nuas sobre $X$ com valores em $\R$ ou $\C$ tal que
  \begin{enumerate}
  \item $A$ cont\'em todas as fun\c c\~oes constantes,
  \item $A$ \'e fechado pela adi\c c\~ao: se $f,g\in A$, ent\~ao $f+g\in A$,
  \item $A$ \'e fechado pela multiplica\c c\~ao: se $f,g\in A$, ent\~ao $fg\in A$,
\item $A$ \'e fechado pela conjuga\c c\~ao complexa: se $f\in A$, ent\~ao $\bar f\in A$,
  \item $A$ separa os pontos de $X$: se $x,y\in X$ e $x\neq y$, ent\~ao existe $f\in A$ tal que $f(x)\neq f(y)$.
  \end{enumerate}
  Ent\~ao, $A$ \'e uniformemente denso em $C(X)$:
  \[\bar A=C(X).\]
  Em outras palavras, cada fun\c c\~ao cont\'\i nua $f\colon X\to \R$ pose ser aproximada uniformemente sobre $X$ por uma sequ\^encia de elementos de $A$:
  \[\exists g_n\in A~~ g_n\rightrightarrows f.\]
\label{stone-weierstrass}
\index{teorema! de Stone--Weierstrass}
\end{teorema}

\begin{observacao}
  No teorema, $X$ pode ser um espa\c co {\em topol\'ogico} compacto qualquer: a exist\^encia de uma m\'etrica sobre $X$ n\~ao joga papel nenhum ma demonstra\c c\~ao.
\end{observacao}

\begin{observacao}
As condi\c c\~oes (1)-(3) significam que $A$ \'e uma {\em sub\'algebra (real)} de $C(X)$. O teorema de Stone-Weierstrass pode ser reformulada como segue: a \'unica sub-\'algebra fechada de $C(X)$ que separa os pontos de $C(X)$ \'e $C(X)$ pr\'opria.
\end{observacao}

\begin{observacao}
Quando $X=[0,1]$ e $A$ consiste de todas as fun\c c\~oes polinomiais, ent\~ao as condi\c c\~oes (1)-(4) se verificam, e a conclus\~ao do teorema \'e conhecido como o {\em teorema de Weierstrass}. 
\end{observacao}

As vers\~oes reais e complexas do teorema s\~ao equivalentes.

\begin{exercicio}
Mostra que a vers\~ao complexa implica a vers\~ao real.
\par
[ {\em Sugest\~ao:} note que se $A\subseteq C(X,\R)$ satisfaz as hip\'oteses do teorema, ent\~ao a \'algebra $A+\im A\subseteq C(X,\C)$ satisfaz as condi\c c\~oes tamb\'em, e se agora
$f_n\in C(X,\C)$ e $f_n\to f$, $f\in C(X,\R)$, ent\~ao $\mbox{Re}\,f_n\in C(X,\R)$ e $\mbox{Re}\,f_n\to f$. ]
]
\end{exercicio}

\begin{exercicio}
Mostra que a vers\~ao real implica a vers\~ao complexa.
\par
[ {\em Sugest\~ao:} primeiramente mostre que, qualquer que seja fun\c c\~ao cont\'\i nua $f\colon X\to\C$, $f\in A$, a sua parte real $\mbox{Re}\,f$ e a parte complexa $\mbox{Im}\,f$ pertencem \`a \'algebra $A$. Depois mostre que as fun\c c\~oes reais em $A$ satisfazem as condi\c c\~oes do teorema. ]
\end{exercicio}

Por conseguinte, basta mostrar apenas a vers\~ao real do teorema.
Vamos seguir a prova incomum de \citep*{BroD}.

\subsection{Alguns resultados t\'ecnicos}

\begin{exercicio} Mostrar a {\em desigualdade de Bernoulli:} se $b\geq -1$ e $m\geq 1$, ent\~ao
  \[(1+b)^m\geq 1+mb.\]
[ {\em Sugest\~ao:} use a indu\c c\~ao em $m$. ]
\end{exercicio}

\begin{lema}
Sob as hip\'oteses do teorema de Stone-Weierstrass, seja $x\in X$ e seja $U$ uma vizinhan\c ca de $x$ quaisquer. Existe uma vizinhan\c ca $V$ de $x$ tal que $V\subseteq U$ e qualquer que seja $\e>0$, existe $f\in A$ com as propriedades
  \begin{itemize}
  \item $f(X)\subseteq [0,1]$,
  \item $f(V)\subseteq [0,\e)$,
  \item $f(X\setminus U)\subseteq (1-\e,1]$.
  \end{itemize}
\end{lema}

\begin{proof}
Sem perda de generalidade e substituindo $\Int U$ por $U$ se necess\'ario, podemos supor que $U$ \'e aberto. Por conseguinte, $X\setminus U$ \'e fechado em $X$, logo compacto.
Para cada ponto $y\in X$, existe uma fun\c c\~ao $f_y\in A$ tal que 
  \[f_y(x)\neq f_y(y).\]
A fun\c c\~ao
  \[h_y = \frac{(f_y - f_y(x))^2}{\norm{(f_y - f_y(x))^2}}_\infty\]
pertence \`a \'algebra $A$, toma suas valores no intervalo $[0,1]$, e satisfaz
  \[h_y(x)=0,~~h_y(y)>0.\]
Os conjuntos abertos
  \[V_y = \{z\in X\colon f_y(z)>0\},~~y\in X\setminus U\]
  cobrem o espa\c co compacto $X\setminus U$. Ent\~ao, existe uma subcobertura finita de $X\setminus U$:
  \[V_{y_1}\cup V_{y_2}\cup\ldots \cup V_{y_m}\supseteq X\setminus U,\]
onde $y_1,y_2,\ldots,y_m\in Y\setminus U$.
A fun\c c\~ao
  \[h =\frac 1m\sum_{i=1}^m h_{y_i}\]
  pertence a $A$, toma os valores no intervalo $[0,1]$, e satisfaz
$h(x) =0$.
  Se $z\in X\setminus U$ qualquer, ent\~ao existe $i=1,2,\ldots,m$ tal que $z\in V_{y_i}$, e conclu\'\i mos:
  \[h(z) \geq \frac 1m h_{y_i}(z)>0.\]
  Devido ao segundo teorema de Weierstrass, o \'\i nfimo de $f$ sobre o espa\c co compacto $X\setminus U$ \'e o m\'\i nimo, atingido num ponto $z_0$. Conclu\'\i mos:
  \[\forall z\in X\setminus U,~~h(z)\geq h(z_0)>0.\]
  Escolhamos um valor irracional $\delta>0$ tal que $\delta\leq h(z_0)$. Temos:
  \[\forall z\in X\setminus U,~~h(z)\geq \delta>0,~~\delta\in\R\setminus\Q.\]
  Em particular, $\delta<1$. 
  O conjunto aberto
  \[V=\left\{z\in X\colon h(z)<\frac{\delta}2\right\}\]
  satisfaz $x\in V\subseteq U$. 
  Definamos
  \[k=\left\lceil\frac 1\delta\right\rceil.\]
Como $\delta$ \'e irracional, $k>1/\delta$, e como $\delta<1$, temos $k\geq 2$. Formemos uma sequ\^encia dos elementos de $A$:
  \[q_n(z) = \left(1-h^n(z)\right)^{k^n},~n=1,2,3,\ldots.\]
  \'E claro que os valores de $q_n$ pertencem ao intervalo $[0,1]$, e $q_n(x)=1$.
  Analisemos o comportamento das fun\c c\~oes $q_n$ sobre os conjuntos $V$ e $X\setminus U$, separadamente. 
  
  (a) Seja $z\in V$ qualquer. Usando a desigualdade de Bernoulli com $b=h^n(z)$ e $m=k^n$, obtemos
  \begin{align*}
    1 &\geq 
    q_n(z) \\
    &= \left(1-h^n(z)\right)^{k^n}\\
    &\geq 1 - k^nh^n(z) \\
    &> 1 -(kh(z))^n.
    \end{align*}
Como $k-1<1/\delta$, 
  \[k <\frac 1\delta +1 =\frac{\delta+1}{\delta} <\frac 2\delta,\]
  e por conseguinte, 
  \[kh(z)<\frac {2k}\delta <\frac{\delta}2\cdot \frac 2\delta=1.\]
  Temos,
  uniformemente sobre $V$,
  \[\abs{q_n(z)-1} \leq \left(\frac {2k}\delta\right)^n\to 0\mbox{ quando }n\to\infty.\]
  Conclu\'\i mos:
  \[q_n\overset{V}{\rightrightarrows} 1,\]
  uniformemente sobre $V$. Dado $\e>0$, se $n$ \'e bastante grande,
  \[\forall z\in V,~~\abs{q_n(z)-1}<\e.\]
  
  (b) Seja $z\in X\setminus U$ qualquer. 
  Usando a desigualdade de Bernoulli na outra dire\c c\~ao, obtemos
  \begin{align*}
    0&\leq     q_n(x) \\
    &= \left(1-h^n(z)\right)^{k^n}\\ 
    &= \frac{1}{k^nh^n(z)}\cdot k^nh^n(z) \cdot\left(1-h^n(z)\right)^{k^n}\\ 
    &= \frac{1}{k^nh^n(z)}(1+k^nh^n(z))\left(1-h^n(z)\right)^{k^n}\\ 
    &\leq 
    \frac{1}{k^nh^n(z)}(1+k^nh^n(z))(1-k^nh^n(z)) \\
    &= 
    \frac{1}{k^nh^n(z)}(1-k^{2n}h^{2n}(z)) \\
    &\leq  \frac{1}{k^nh^n(z)} \\
    &< \frac{1}{(k\delta)^n}.
    \end{align*}
  Como $k>1/\delta$, temos
  \[\frac{1}{k\delta}<1,\]
de onde
  \[q_n\overset{X\setminus U}{\rightrightarrows} 0,\]
  uniformemente sobre $X\setminus U$.
  Dedo $\e>0$, se $n$ \'e bastante grande, 
  \[0\leq q_n(z)<\e\]
  para todos $z\in X\setminus U$.
  
Agora, se $\e>0$ \'e qualquer, para $n$ suficientemente grande, a fun\c c\~ao
  \[f = 1-q_n\]
  possui todas as propriedades desejadas la conclus\~ao do lema.
\end{proof}

\begin{lema} Sobre as hip\'oteses do teorema de Stone-Weierstrass, sejam $F$ e $G$ dois subconjuntos de $X$ fechados e disjuntos. Seja $\e>0$ qualquer. Existe $f\in A$ tal que
  \begin{itemize}
    \item $f(X)\subseteq [0,1]$,
    \item $f(F)\subseteq [0,\e)$,
    \item $f(G)\subseteq (1-\e,1]$.
  \end{itemize}
\label{l:12.2.4}
\end{lema}

\begin{proof}
  Para cada $x\in F$, usando o lema precedente, escolhamos uma vizinhan\c ca $V_x$ de $x$ com $U= X\setminus F$. A cobertura aberta do subconjunto compacto $G$ de $X$ pelos conjuntos $V_x,~~x\in F$,
  admite ume subcobertura finita 
  \[V_{x_i},~~i=1,2,\ldots,m.\]
  Para cada $i=1,2,\ldots,m$, escolhamos a fun\c c\~ao $f_i\in A$ com valores no intervalo $[0,1]$ e tal que $f_i(V_{x_i})\subseteq [0,\e/m)$, $f_i(G)\subseteq (1-\e/m,1]$.
  A fun\c c\~ao
  \[f(z) =\prod_{i=1}^m f_i(z)\]
  pertence a $A$ e toma seus valores no intervalo $[0,1]$. Seja $z\in F$. Existe $i=1,2,\ldots,m$ tal que $z\in V_{x_i}$, e por conseguinte
  \[f_i(z)<\e/m.\]
  Conclu\'\i mos:
  \[f(z) = f_i(z)\times \prod_{j\neq i}^m f_j(z)\leq \frac{\e}m\leq \e.\]
  Seja $z\in G$. Para cada $i=1,2,\ldots,m$, temos
  \[f_i(z) > 1-\frac{\e}m,\]
  de onde, usando a desigualdade de Bernoulli pela \'ultima vez, deduzimos
  \[f(z) =\prod_{i=1}^m f_i(z) > \left(1-\frac{\e}{m}\right)^m \geq 1 - m\cdot \frac{\e}{m} = 1-\e.\]
\end{proof}

\subsection{Demonstra\c c\~ao do teorema de Stone-Weierstrass}
\'E claro que bastaria aproximar uma fun\c c\~ao cont\'\i nua com valores em $[0,1]$. 
Dada uma fun\c c\~ao $f\in C(X)$ tal que $f(X)\subseteq [0,1]$, e um n\'umero natural $n\in\N_+$, para cada $i=0,1,\ldots,n-1$ definamos conjuntos
\[F_i=\left\{x\in X\colon f(x)\leq \frac in\right\},\]
\[G_i = \left\{x\in X\colon f(x)\geq \frac {i+1}n\right\}.\]
Para cada $i$, os conjuntos $F_i$ e $G_i$ s\~ao fechados e disjuntos em $X$. Segundo lema \ref{l:12.2.4}, existe uma fun\c c\~ao $g_i\colon X\to [0,1]$, $g_i\in A$, tal que 
\[g_i(F_i)\subseteq [0,1/n)\mbox{ e }g_i(G_i)\subseteq (1-1/n,1].\]
Ponhamos
\[g=g_{(n)}=\frac 1n\sum_{i=0}^{n-1} g_i.\]
Obviamente, $g\in A$ e $g(X)\subseteq [0,1]$.
A nota\c c\~ao $g_{(n)}$ sublinha o fato que $g$ depende de $n$. Deste modo, temos uma sequ\^encia $g_{(n)}\in A$, $n=1,2,3,\ldots$. 

Seja $x\in X$ qualquer. Existe $i=0,1,\ldots,n-1$ tal que
\[\frac in\leq f(x) \leq \frac{i+1}n.\]
Deduzimos que 
\[x\in F_j,~~j=i+1,i+2,\ldots,n-1,\]
\[x\in G_j,~~j=0,1,\ldots,i-1.\]
Se $j\geq i+1$, ent\~ao $x\in F_j$, de onde $0\leq g_j(x) <1/n$. Se $j\leq i-1$, ent\~ao $x\in G_j$, de onde $1-1/n<g_j(x)\leq 1$. O valor $g_i(x)$ pode ser qualquer, entre $0$ e $1$. Somando $n$ desigualdades e dividindo a soma por $n$, conclu\'\i mos que a diferen\c ca entre $g(x)$ e $i/n$ \'e menor que $2/n$, e por conseguinte a diferen\c ca entre $g(x)$ e $f(x)$ \'e menor que $3/n$. Quando $n\to\infty$, temos
\[\norm{f(x)-g_{(n)}(x)}_{\infty}< \frac 3n\to 0,\]
ou seja,
\[g_{(n)}\rightrightarrows f.\]

%
%

\chapter{Lema de Kronecker\label{a:kronecker}}

Neste ap\^endice, vamos estudar os subgrupos aditivos fechados de $\R^n$. Comecemos acertando que apenas existe uma ``topologia natural'' sobre um espa\c co vetorial de dimens\~ao finita.

\begin{exercicio}
Verifique que a bola unit\'aria fechada do espa\c co $\ell^1(n)$ \'e igual \`a envolt\'oria convexa do conjunto $\pm e_i$, $i=1,2,\ldots,n$, onde $(e_i)$ s\~ao vetores de base:
\[B_{\ell^1(n)} = \left\{\sum_{i=1}^n \lambda_i e_i\colon \lambda_i\in [-1,1],~\sum\abs{\lambda_i}\leq 1\right\}.\]
Deduza que a norma de qualquer funcional linear $\psi$ sobre $\ell^1(n)$ \'e igual a $\max_{i=1}^n\abs{\psi(e_i)}$.
\label{ex:envoltoria}
\end{exercicio}

\begin{exercicio}
Verifique que a bola unit\'aria de $\ell^{\infty}(n)$ \'e igual \`a envolt\'oria convexa do conjunto de todas as combina\c c\~oes lineares da forma $\sum_{i=1}^n \ve_i e_i$, onde $\ve_i=\pm 1$.
\label{ex:boladeallinfty}
\end{exercicio}

\begin{teorema}
As topologias seguintes sobre $\R^n$ s\~ao duas a duas iguais.
\begin{enumerate}
\item\label{topp:1} A topologia produto, ou seja, uma topologia metriz\'avel tal que $x_m\to x$ se e somente a $i$-\'esima coordenada de $x_m$ converge para a de $x$, $i=1,2,\ldots,n$.
\item \label{topp:2} A topologia gerada por todos os funcionais lineares, ou seja, a menor topologia sobre $\R^n$ tal que cada funcional linear $\phi\colon\R^n\to\R$ \'e cont\'\i nuo.
\item\label{topp:5}  A topologia gerada pela norma $\ell^{\infty}(n)$.
\item\label{topp:3}  A topologia gerada por qualquer norma sobre $\R^n$.

\item\label{topp:4}  A topologia gerada pela norma $\ell^1(n)$.
\end{enumerate}
\label{t:topp}
\end{teorema}

\begin{proof}
Vamos mostrar, de modo circular, que cada topologia \'e contida na seguinte. 

(\ref{topp:1})$\subseteq$(\ref{topp:2}) segue do fato que cada proje\c c\~ao coordenada, $\pi_i$, \'e um funcional linear. 

(\ref{topp:2})$\subseteq$(\ref{topp:5}): temos que mostrar que um funcional linear qualquer, $\phi$, sobre $\ell^{\infty}(n)$ e limitado. Usando o exerc\'\i cio \ref{ex:boladeallinfty}, conclu\'imos:
\begin{align*}
\norm{\phi} &= \max\{\sum_{i=1}^n \ve_i \phi(e_i)\colon \ve_i=\pm 1\}\\
&\leq \sum_{i=1}^n \left\vert\phi(e_i)\right\vert.
\end{align*}

(\ref{topp:5})$\subseteq$(\ref{topp:3}): Seja $\norm\cdot$ uma norma qualquer sobre $\R^n$. Mostremos que existe $L\geq 0$ tal que a bola $\ell^{\infty}(n)$ de raio $L$ cont\'em a bola unit\'aria $B_{\norm\cdot}$. Como a topologia $\ell^\infty(n)$ \'e a de produto, a bola $B_{\norm\cdot}$ \'e compacta. Existe $\delta>0$ t\~ao pequeno que os vetores $\sum_{i=1}^n \ve_i e_i$, $\e_i=\pm 1$ todos pertencem \`a bola $B_{\norm\cdot}$. Por conseguinte, o operator identidade $\mbox{id}_{\R^n}$ de $\ell^\infty(n)$ para $(\R^n,\norm\cdot)$ \'e limitado, logo cont\'\i nuo, e a norma $\norm\cdot$ \'e uma aplica\c c\~ao cont\'\i nua sobre $\ell^{\infty}(n)$ tamb\'em.
Para cada $N\in\N$ seja $F_N$ o conjunto de todos $x\in B_{\infty}$ tais que $\norm x\geq N$. A continuidade da norma em $\ell^{\infty}(n)$ implica que $F_N$ s\~ao compactos. A sequ\^encia $(F_N)$ sendo encaixada, se todos $F_N\neq\emptyset$, existe $x\in B_{\infty}\cap\cap_NF_N$. Tal $x$ deve satisfazer $\norm x=+\infty$, o que \'e imposs\'ivel. 

(\ref{topp:3})$\subseteq$(\ref{topp:4}): Seja $\norm\cdot$ uma norma qualquer sobre $\R^n$. Definamos o operador linear e bijetor $T\colon \ell^1(n)\to \R^n$ por
 \[T(e_i)=\norm{e_i}^{-1} e_i.\]
Em virtude do exerc\'\i cio \ref{ex:envoltoria}, $\norm T=\max\norm{e_i}^{-1}=1$, logo $T$ \'e limitado e cont\'\i nuo, e segue-se que a topologia gerada pelas bolas abertas na norma $\ell^1(n)$ \'e mais fina do que a topologia gerada pela norma $\norm\cdot$.

(\ref{topp:4})$\subseteq$(\ref{topp:1})
Sejam $(x_m)$ uma sequ\^encia de elementos de $\R^n$ e $x\in\R^n$ tais que que para cada $i=1,2,\ldots,n$, $\pi_i(x_m)\to\pi_i(x)$. Temos:
\begin{align*} \norm{x_i-x}_1 &= \sum_{i=1}^n \abs{x_i-x} \\
&\to 0,
\end{align*}
o que significa que a topologia gerada pela norma $\ell^1$ \'e mais fraca do que a topologia produto sobre $\R^n$.
\end{proof}

\begin{definicao}
Relembremos que um {\em grupo} \'e um conjunto $G$ munido de uma opera\c c\~ao bin\'aria, chamado o produto, e um elemento fixo, $e$, o {\em elemento neutro,} tais que o produto \'e associativo, para todo $x$, $xe=ex=x$, e todo elemento $x$ admite um inverso, $x^{-1}$, de modo que $xx^{-1}=x^{-1}x=e$.
\index{grupo}
\end{definicao}

Um grupo cuja opera\c c\~ao \'e comutativa, $xy=yx$, \'e chamado {\em abeliano,} e o produto ser\'a frequentemente (mas n\~ao sempre) denotado $+$ (nota\c c\~ao aditiva).

\begin{exemplo}
Cada espa\c co linear \'e um grupo abeliano, onde adi\c c\~ao \'e a opera\c c\~ao bin\'aria, zero \'e o elemento neutro, e $-x$ \'e o inverso de $x$. Em particular, $\R^n$ \'e um grupo, e $\Z^n$ \'e um subgrupo.
\end{exemplo}

\begin{exemplo}
O grupo $\T=U(1)=\{z\in\C\colon \abs z =1\}$ \'e um subgrupo multiplicativo do grupo $\C^\times$ de elementos n\~ao zero de $\C$, munido da multiplica\c c\~ao dos n\'umeros complexos.
\end{exemplo}

\begin{exemplo} O grupo sim\'etrico $S_n$, que consiste de permuta\c c\~oes do conjunto $[n]$ munidos da opera\c c\~ao de composi\c c\~ao de fun\c c\~oes, n\~ao \'e abeliano.
\index{Sn@$S_n$}
\end{exemplo}

Dado um grupo $G$ e um subconjunto $X\subseteq G$, existe o menor subgrupo de $G$ que cont\'em $X$. 

\begin{exemplo}
O subgrupo de $\R^n$ gerado pelos vetores b\'asicos, $e_1,e_2,\ldots,e_n$, \'e exatamente o grupo $\Z^n$.
\end{exemplo}

\begin{exemplo}
Seja $x\in\R$, $x\neq 0$. O subgrupo gerado por $x$ \'e igual a 
\[x\Z = \{xn\colon n\in\Z\}.\]
\end{exemplo}

\begin{definicao}
Um grupo $G$ munido de uma topologia \'e dito {\em grupo topol\'ogico} se as opera\c c\~oes s\~ao cont\'\i nuas:
\begin{align*}
m\colon G\times G &\ni (g,h)\mapsto gh\in G,\\
i\colon G&\ni g\mapsto g^{-}\in G.
\end{align*}
\end{definicao}

\begin{exercicio}
Mostre que $\R^n$ e $\T^n$ (com a topologia usual) s\~ao grupos topol\'ogicos.
\end{exercicio}

\begin{exercicio}
Seja $G$ um grupo topol\'ogico e $H$ um subgrupo. Mostre que a ader\^encia $\bar H^G$ \'e um subgrupo.
\par
[ {\em Sugest\~ao:} suponha que $G$ \'e metriz\'avel, e deste modo basta trabalhar com as sequ\^encias. Mostre que se $g_n\to g$ e $h_n\to h$, ent\~ao $g_nh_n\to gh$ e $g_n^{-1}\to g^{-1}$. ]
\end{exercicio}

\begin{teorema}
Cada subgrupo fechado e pr\'oprio de $\R$ \'e da forma $x\Z$, $x\neq 0$.
\label{t:discretesbgrpR}
\end{teorema}

\begin{proof}
Seja $G$ um subgrupo fechado de $\R$. Se $G$ cont\'em uma sequ\^encia n\~ao trivial convergente para zero, $x_n\to 0$, ent\~ao $G$ cont\'em todos subgrupos $x_n\Z$, cuja uni\~ao \'e densa em $\R$, e como $G$ \'e fechado, $G=\R$. 
Suponha agora que $G\neq\{0\}$ e $G$ n\~ao cont\'em sequ\^encias convergentes para zero. Neste caso, o elemento seguinte \'e bem definido:
\[x=\inf\{g\in G\colon g>0\},\]
e como $G$ \'e fechado, $x\in G$, \'e um m\'\i nimo. Afirmamos que $x\Z=G$. \'E claro que $x\Z\subseteq G$. Se existisse $y\in G\setminus x\Z$, o elemento $z=x\lfloor y/x\rfloor$ satisfaria $z\in x\Z$, $y-z\in G$, e $0<y-z<x$ ($G$ \'e um grupo ordenado), o que \'e imposs\'\i vel.
\end{proof}

\begin{lema}
Seja $x\in\R^n$. O subgrupo de $\R^n$ gerado por $\Z^n\cup\{x\}$ \'e discreto se e somente se $x\in \Q^n$.
\label{l:Zcupx}
\end{lema}

\begin{proof} 
Suponha que o grupo $G$ gerado por $\Z^n\cup\{x\}$ \'e discreto, ou seja, n\~ao cont\'em sequ\^encias convergentes n\~ao triviais. Seja $g\in G$ qualquer, $g=(g_1,g_2,\ldots, g_n)$. Para todo $m\in\Z$, consideremos o vetor 
\[\tilde g_m=mg - \sum_{i=1}^n\lfloor mg_i\rfloor e_i = (mg_1-\lfloor mg_1\rfloor,
mg_2-\lfloor mg_2\rfloor, \ldots, mg_n-\lfloor mg_n\rfloor ).\]
Ele pertence a $G$, sendo a diferen\c ca de dois elementos de $G$, e tamb\'em pertence \`a bola $B_{\infty}$. Como a interse\c c\~ao $G\cap B_{\infty}$ \'e finita por causa da compacidade da bola, existem dois inteiros distintos, $m$ e $k$, tais que $\tilde g_m =\tilde g_k$. Por conseguinte,
\[(m-k)g_i = \lfloor mg_i\rfloor-\lfloor kg_i\rfloor,~i=1,2,\ldots,n,\]
o que significa que $g_i\in\Q$.

Suponha agora que $x\in\Q^n$. Se $q$ \'e um denominator comum de $x_i$, ent\~ao o grupo gerado por $\Z^n\cup\{x\}$  \'e um subgrupo do grupo discreto $\frac 1q\Z^n$.
\end{proof}

\begin{lema}
Seja $G$ um subgrupo fechado e n\~ao discreto de $\R^n$. Ent\~ao, $G$ cont\'em uma reta passando por $0$, ou seja, um subgrupo fechado da forma $\R x$, $x\in\R^n$, $x\neq 0$.
\label{l:retapassando}
\end{lema}

\begin{proof}
Existe uma sequ\^encia convergente e n\~ao trivial, $x_n\to x$, $x_n,x\in G$, $x_n\neq x$. Substituindo $x_n = x_n-x\in G$, suponhamos que $x_n\to 0$. Como $\norm{x_n}_{\infty}\to 0$, temos $\norm{x_n}_{\infty}^{-1}\to\infty$. A sequ\^encia $\norm{x_n}_{\infty}^{-1}x_n$ pertence \`a esfera unit\'aria $S_{\infty}$ em rela\c c\~ao \`a norma $\ell^{\infty}(n)$, logo converge para um elemento $x$ da esfera. O mesmo se aplica \`a sequ\^encia $\lfloor\norm{x_n}_{\infty}^{-1}\rfloor x_n$, pois a diferen\c ca entre as duas sequ\^encias converge para zero. Como os elementos da \'ultima sequ\^encia pertencem a $G$, conclu\'\i mos que $x\in G$ tamb\'em. Para todo $\lambda\in\R$, os elementos da sequ\^encia 
\[\left\lfloor \frac{\lambda}{\norm{x_n}_{\infty}}\right\rfloor \cdot x_n\]
pertencem a $G$, e \'e f\'acil a ver que a sequ\^encia acima converge para $\lambda x$.
\end{proof}

A reta, $\R$, pode ser vista como um espa\c co vetorial sobre o corpo $\Q$.

\begin{exercicio}
Mostre que $\R$, visto como um espa\c co vetorial sobre o corpo $\Q$, tem dimens\~ao do cont\'\i nuo, $\mathfrak c$.
\end{exercicio}

\begin{definicao}
Um conjunto de reais \'e dito {\em racionalmente independente} se ele \'e um conjunto linearmente independente em $\R$ como um espa\c co linear sobre $\Q$.
\index{reais racionalmente independentes}
\end{definicao}

\begin{lema}
Sejam $1\leq m\leq n$, e $x=(x_1,x_2,\ldots,x_n)\in\R^n$ um vetor tal que os reais
\[x_1,x_2,\ldots,x_n,1\]
s\~ao racionalmente independentes. Seja $A$ uma matriz $m\times n$ com coeficientes racionais, de posto $m$. Ent\~ao o vetor $y= Ax$, $y=(y_1,y_2,\ldots,y_m)$, tem a propriedade que os reais
\[y_1,y_2,\ldots,y_m,1\]
s\~ao racionalmente independentes.
\label{l:imagemderacionalmenteindep}
\end{lema}

\begin{proof}
Os reais $y_1,y_2,\ldots,y_m,1$ s\~ao racionalmente independentes se e apenas se, quaisquer que sejam coeficientes racionais $\lambda_1,\ldots,\lambda_m$, n\~ao todos iguais a zero, 
\[\sum_{i=1}^m\lambda_i y_i\notin\Q.\]
Temos
\begin{align*}
\sum_{i=1}^m\lambda_i y_i &= \sum_{i=1}^m\lambda_i \sum_{j=1}^n a_{ij}x_j \\
&= \sum_{j=1}^n \left(\sum_{i=1}^m \lambda_{i} a_{ij} \right)x_i,
\end{align*}
e como a matriz $A$ tem posto $m$ sobre o corpo $\Q$, as $m$ linhas da matriz s\~ao linearmente independentes sobre $\Q$, ou seja, $\sum_{i=1}^m \lambda_i a_{ij}$ n\~ao se anulam simultaneamente. Gra\c cas \`a hip\'otese sobre $x$, conclu\'\i mos.
\end{proof}

\begin{lema}
Seja $x=(x_1,x_2,\ldots,x_n)$ um vetor em $\R^n$, tal que os reais
\[x_1,x_2,\ldots,x_n,1\]
s\~ao racionalmente independentes, ou seja, linearmente independentes como vetores sobre o corpo $\Q$. Ent\~ao o \'unico subgrupo fechado de $\R^n$ que cont\'em $x$ e $\Z^n$ \'e $\R^n$ pr\'oprio.
\label{l:pre-kronecker}
\end{lema}

\begin{proof} 
Denotemos $V$ a uni\~ao de todas as retas contidas em $G$ e passando por zero. \'E claro que $V$ \'e um subespa\c co vetorial de $\R^{n}$. Escolhamos algum subespa\c co linear $V^\prime$ complementar a $V$, ou seja, $V^\prime + V = \R^{n}$. Denotemos $G^\prime = G\cap V^\prime$. Segue-se que $G= V+G^\prime$.  Segundo o lema \ref{l:retapassando}, o grupo $G^\prime$ \'e discreto.

Denotemos $\pi$ a aplica\c c\~ao linear quociente de $\R^{n}$ sobre o espa\c co linear quociente, $\R^{n}/V$. Entre os vetores $\pi(e_i)$, $i=1,2,\ldots,n$, escolhamos uma base do espa\c co quociente.
Identificamos $\R^{n+1}/V$ com o espa\c co $\R^m$, onde $m=n-\dim V$.
O operador $\pi$, restrito sobre $V^\prime$, \'e um isomorfismo de $V^\prime$ com $\R^m$, e existem $m$ vetores de base em $\R^{n}$ cujas imagens s\~ao vetores de base padr\~ao em $\R^m$. 
Ademais, a imagem de $G$ por $T\circ \pi$ \'e igual a imagem de $G^\prime$, logo \'e um subgrupo discreto. Em virtude do lema \ref{l:Zcupx}, a imagens de todos elementos de $\Z^n$ tem coordenadas racionais, logo a matriz do operador quociente tem coeficientes racionais. Lema \ref{l:imagemderacionalmenteindep} diz que as coordenadas de $\pi(x)$ em $\R^m$, mais $1$, formam uma fam\'\i lia de reais racionalmente independente. Lema \ref{l:Zcupx} 
implica que isso \'e apenas poss\'\i vel se $m=0$.
\end{proof}

\begin{definicao}
Um {\em homomorfismo} entre dois grupos, $f\colon G\to H$, \'e uma aplica\c c\~ao que conserva as opera\c c\~oes: $f(gh)=f(g)f(h)$ e $f(e_G)=e_H$.
O {\em n\'ucleo} de um homomorfismo \'e a imagem rec\'\i proca de identidade:
\[N=f^{-1}(e).\]
\'E um {\em subgrupo normal,} ou seja, um subgrupo $N$ tendo a propriedade $x^{-1}Nx=N$ qualquer que seja $x\in G$.
\end{definicao}

\begin{exemplo}
Um homomorfismo cont\'\i nuo $f\colon \R\to\T$ \'e dado por
\[f(t) = \exp(2\pi\im t).\]
Este homomorfismo estende-se at\'e um homomorfismo cont\'\i nuo entre $\R^n$ e o toro $\R^n$:
\begin{equation}
f(t_1,\ldots,t_n) = (\exp(2\pi\im t_1), \ldots, \exp(2\pi\im t_n)).
\label{eq:homomorfismotoro}
\end{equation}
O n\'ucleo deste homomorfismo \'e o subgrupo $\Z^n$. A toro pode ser visto como o {\em grupo quociente} $\R^n/\Z^n$, que consiste de todas as classes de equival\^encia $x+\Z^n$ m\'odulo o subgrupo $\Z^n$, munidas da adi\c c\~ao de conjuntos: a f\'ormula
\[(x+\Z^n)+(y+\Z^n)=(x+y)+\Z^n\]
\'e bem definida. Deste ponto de vista, identificamos
\[x+\Z^n \leftrightarrow f(x).\]
\label{ex:toro}
\end{exemplo}

\begin{teorema}[Lema de Kronecker]
Seja $x\in\R^n$ um vetor tal que \[x_1,x_2,\ldots,x_n,1\]
s\~ao racionalmente independentes. Ent\~ao o subgrupo gerado pela imagem de $x$ sob o homomorfismo (\ref{eq:homomorfismotoro}) \'e denso no toro $\T^n$.
\label{t:kronecker}
\index{lema! de Kronecker}
\end{teorema}

\begin{proof} Denotemos $G$ o subgrupo do toro gerado pelo elemento $f(x)$. A imagem inversa $f^{-1}G$ \'e um subgrupo fechado de $\R^n$ que cont\'em $x$ e $\Z^n$. Agora apliquemos o lema \ref{l:pre-kronecker}, junto com o fato que $f\left[f^{-1}(G)\right]=G$ pois $f$ \'e sobrejetora.
\end{proof}

A prova neste ap\^endice \'e baseada sobre a de Bourbaki \citep*{bourbaki}, Chap. VII, \S 1. Quero agradecer Vladimir V. Uspenskij por ter trazido a prova de Bourbaki \`a minha aten\c c\~ao. Todos os erros que eu possivelmente tenho introduzidos tentando encurtar a prova ainda mais s\~ao meus pr\'oprios.

%
%

\chapter{Integra\c c\~ao e desintegra\c c\~ao\label{a:integral}}

\section{Integral de Lebesgue, espa\c co $L^1(\mu)$, esperan\c ca\label{s:integrall1mu}}

\subsection{Integral de Lebesgue}
Sejam $\mu$ uma medida de probabilidade sobre um espa\c co boreliano padr\~ao $\Omega$, e $f\colon\Omega\to\R$ uma fun\c c\~ao boreliana e n\~ao negativa. A integral de $f$ no sentido de Lebesgue \'e o valor
\begin{eqnarray}
\int_{\Omega}f(x)\,d\mu(x) = \int_{0}^{+\infty}\mu\{x\in\Omega\colon f(x)\geq y\}\,dy,
\label{eq:integraldeLebesgue}
\end{eqnarray}
\index{integral de Lebesgue}
onde a integral \`a direita \'e a integral definida de uma fun\c c\~ao real (no sentido de Riemann). Isto corresponde ao dividir a \'area sob o gr\'afico de $f$ em faixas horizontais, ao inv\'es de verticais, como na defini\c c\~ao da integral de Riemann.

\begin{observacao}
Uma fun\c c\~ao mensur\'avel \'e dita {\em fun\c c\~ao $L^1(\mu)$} se a integral acima da fun\c c\~ao positiva $\abs f$ \'e finita.
\end{observacao}

\begin{exercicio}
Observe que a fun\c c\~ao 
\[\R\ni y\mapsto \mu\{x\in\Omega\colon f(x)\geq y\}\in \R\]
\'e mon\'otona, logo a integral no sentido de Riemann \'e bem definida, com um valor em $\R_+\cup\{+\infty\}$.
\end{exercicio}

\begin{exercicio}
Deduza que o conjunto
\[S_f=\{(x,y)\in \Omega\times\R\colon f(x)\leq y\}\]
(a \'area sob o gr\'afico de $f$) \'e boreliano, e verifique que
\[\int_{\Omega}f(x)\,d\mu(x) = (\mu\otimes\lambda) (S_f),\]
onde $\lambda$ \'e a medida de Lebesgue unidimensional.
\end{exercicio}

\begin{lema}
Seja $(\Omega,\mu)$ um espa\c co probabil\'\i stico padr\~ao, e seja $f\colon\Omega\to\R$ uma fun\c c\~ao boreliana. Ent\~ao a transforma\c c\~ao
\[
\mbox{id}_{\Omega}\times T_f\colon\Omega\times\R\ni (x,y)\mapsto (x,y+ f(x))\in\Omega\times\R
\]
\'e um automorfismo boreliano do produto $\Omega\times\R$ que conserva a medida $\mu\otimes\lambda$ (um ``automorfismo vertical'').
\label{l:automorfismovertical}
\end{lema}

\begin{proof}
\'E claro que a transforma\c c\~ao acima \'e bijetora, com a inversa que corresponde \`a fun\c c\~ao $-f$. Portanto,
para verificar a primeira afirma\c c\~ao, basta mostrar que a imagem de um subconjunto retangular aberto qualquer, $U\times O$, seja boreliano. 
Usemos o teorema de Luzin \ref{t:luzin} para escolher, para cada $\ve>0$, um compacto $K_{\ve}\subseteq\Omega$ tal que $\mu(K_{\ve})>1-\ve$ \'e $f\vert_{K_{\ve}}$ \'e cont\'\i nua. Temos
\begin{align*}
(\mbox{id}_{\Omega}\times T_f)(U\times O)\cap (K_{\ve}\times\R) & =
(\mbox{id}_{\Omega}\times T_{f\vert_{K_{\ve}}})(U\cap K_{\ve})\times O,
\end{align*}
um subconjunto aberto de $K_{\ve}\times\R$, logo boreliano em $\Omega\times\R$, e quando $\ve_n\downarrow 0$,
\[(\mbox{id}_{\Omega}\times T_f)(U\times O) = \bigcup_{n=0}^{\infty} (\mbox{id}_{\Omega}\times T_f)(U\times O)\cap (K_{\ve_n}\times\R),\]
de onde conclu\'\i mos.

Para mostrar a segunda afirma\c c\~ao, notemos que o resultado para a parte at\^omica de $\Omega$ \'e evidente, e suponhamos que $\mu$ seja n\~ao at\^omica.
Identifiquemos $\Omega$ atrav\'es de um isomorfismo de espa\c cos probabil\'\i sticos com o intervalo $[0,1]$ (teorema \ref{t:unicidadedemedidanaoatomica}).
Estendamos $f\vert_{K_{\ve}}$ at\'e uma fun\c c\~ao cont\'\i nua $\tilde f$ sobre $[0,1]$ (veja o argumento na p\'agina \pageref{extensaodefuncoescontinuas}). Agora \'e bastante simples fazer a conclus\~ao desejada, aproximando a imagem de $U\times O$ com uni\~oes de subconjuntos retangulares cuja base converge para zero.
\end{proof}

\begin{exercicio}
Conclua que a fun\c c\~ao dist\^ancia $L^1(\mu)$,
\[d(f,g)_1=\int_{\Omega}\abs{f(x)-g(x)}\,d\mu(x),\]
\'e uma pseudom\'etrica sobre o conjunto de todas as fun\c c\~oes $L^1(\mu)$ positivas.
\par
[ {\em Sugest\~ao:} use um automorfismo vertical do lema \ref{l:automorfismovertical} para concluir que a dist\^ancia entre duas fun\c c\~oes, $f$ e $g$, \'e igual ao valor da medida $\mu\otimes\lambda$ da diferen\c ca sim\'etrica $S_f\Delta S_g$. ]
\label{ex:pseudometricaL1}
\end{exercicio}

\begin{exercicio}
Mostre que a integral de Lebesgue \'e aditivo para fun\c c\~oes positivas:
\[\int_{\Omega}(f+g)\,d\mu = \int_{\Omega}f\,d\mu+\int_{\Omega}g\,d\mu.\]
[ {\em Sugest\~ao:} mais uma vez, use um automorfismo vertical para mostrar que area sob o gr\'afico de $f+g$ \'e igual a soma das \'areas sob os gr\'aficos de $f$ e de $g$. ]
\end{exercicio}

Se agora a fun\c c\~ao mensur\'avel $f\colon\Omega\to\R$ \'e qualquer, representemos
\[f=f_+-f_-,\]
onde 
\[f_+=\max\{f,0\},~~f_-=\max\{-f,0\}\]
s\~ao positivas, e escrevamos
\[\int_{\Omega}f(x)\,d\mu(x) = \int_{\Omega}f_+(x)\,d\mu(x) -\int_{\Omega}f_-(x)\,d\mu(x).\]
A fun\c c\~ao \'e $L^1$ se e somente se ambas integrais s\~ao finitas. Neste caso, o valor da integral de $f$ \'e bem definido e finito. 

\begin{observacao}
As defini\c c\~oes acima t\^em sentido para todas as fun\c c\~oes $\mu$-men\-su\-r\'aveis. No entanto, como toda fun\c c\~ao $\mu$-men\-sur\'avel apenas difere de uma fun\c c\~ao boreliana $f^\prime$ sobre um conjunto negligenci\'avel (exerc\'\i cio \ref{ex:fmudag}), basta observar que
\[\int_{\Omega}f\,d\mu=\int_{\Omega}f^\prime\,d\mu,\]
e sempre trabalhar s\'o com as fun\c c\~oes borelianas.
\end{observacao}

\begin{exercicio}
Estenda a afirma\c c\~ao do exerc\'\i cio \ref{ex:pseudometricaL1} para todas as fun\c c\~oes $L^1(\mu)$.
\end{exercicio}

\subsection{$L^1(\mu)$\label{ss:l1mu}}

\begin{exercicio}
Mostre que as fun\c c\~oes $L^1(\mu)$ formam um espa\c co linear (denotado $L^1(\mu)$), e que
a integral $\int_{\Omega}d\mu$ \'e um funcional linear sobre o espa\c co $L^1(\mu)$.
\end{exercicio}

\begin{exercicio}
Mostre que a fun\c c\~ao
\[\norm{f}_1=\int_{\Omega}\abs{f(x)}\,d\mu(x)\]
\'e uma pr\'e-norma sobre o espa\c co $L^1(\mu)$, ou seja, satisfaz os axiomas 2 e 3 de uma norma (defini\c c\~ao \ref{d:normaN}).
\end{exercicio}

\begin{exercicio}
Mostre que o n\'ucleo da pr\'e-norma $L^1(\mu)$,
\[N=\{x\in L^1(\mu)\colon \norm{x}_1=0\},\]
\'e um subespa\c co linear, e que a f\'ormula
\[\norm{x+N}_1 = \norm{x}_1\]
defina uma norma sobre o espa\c co vetorial quociente $L^1(\mu)/N$.
\end{exercicio}

Este espa\c co \'e tamb\'em denotado pelo mesmo s\'\i mbolo $L^1(\mu)$. Seus elementos s\~ao classes de equival\^encia de fun\c c\~oes. Cada classe cont\'em pelo menos um representante que \'e uma fun\c c\~ao boreliana.
\index{L1mu@$L^1(\mu)$}

\begin{exercicio}
Mostre que as condi\c c\~oes seguintes s\~ao equivalentes.
\begin{enumerate}
\item $f$ e $g$ pertencem \`a mesma classe da equival\^encia modulo $N$.
\item $f$ e $g$ s\'o diferem sobre um conjunto negligenci\'avel.
\end{enumerate}
\end{exercicio}

\begin{exercicio}
Mostre que, se $\mu$ n\~ao \'e puramente at\^omica, o n\'ucleo $N$ \'e n\~ao trivial.
\end{exercicio}

\begin{exercicio}
Sejam $f$ e $g$ duas fun\c c\~oes mensur\'aveis sobre um espa\c co probabil\'\i stico padr\~ao $\Omega$. Suponha que $\norm{f-g}_1<\ve$.
Observe que o conjunto de pontos $x$ onde $\abs{f(x)-g(x)}\geq\ve$ tem medida $\leq\ve$. Agora escolha sobre $\Omega$ uma m\'etrica completa e separ\'avel gerando a estrutura boreliana, e
deduza a seguinte: se $(f_n)$ \'e uma sequ\^encia de Cauchy no espa\c co $L^1(\mu)$, ent\~ao para cada $\ve>0$ existe um aberto $U\subseteq\Omega$ com $\ve(U)<\ve$ e uma subsequ\^encia $(f_{n_i})$ que \'e Cauchy em rela\c c\~ao \`a m\'etrica uniforme sobre $\Omega\setminus U$. Conclua que o espa\c co normado $L^1(\mu)$ \'e um espa\c co de Banach.
\par
[ {\em Sugest\~ao:} a escolha de $U$ e de uma subsequ\^encia s\~ao recursivas, usando uma sequ\^encia som\'avel dos reais positivos $(\ve_n)$ e a defini\c c\~ao equivalente de uma sequ\^encia de Cauchy do lema \ref{l:defeqseqcauchy}. ]
\end{exercicio}

\subsection{Esperan\c ca}
Agora, dada uma vari\'avel aleat\'oria real $X$, tomando valores positivos, podemos definir a esperan\c ca de $X$ no mesmo esp\'\i rito que a integral na eq. (\ref{eq:integraldeLebesgue}):
\[\E X = \int_0^{\infty}(1-\Phi_X(t))\,dt,\]
onde 
\[\Phi(t) = P[X<t]\]
\'e a fun\c c\~ao de distribui\c c\~ao de $X$. A defini\c c\~ao se estende  sobre v.a. reais quaisquer pela formula:
\[\E X = \E X_+ - \E X_-,\]
onde a nota\c c\~ao \'e clara.
Na pr\'atica, a carateriza\c c\~ao seguinte ser\'a mais \'util.

\begin{exercicio}
Seja $X$ uma vari\'avel aleat\'oria real qualquer. Mostre que, qualquer que seja uma realiza\c c\~ao $f\colon ({\mathfrak X},\nu)\to\R$ de $X$, temos
\[\E X = \int_{\mathfrak X}f(x)\,d\nu(x).\]
\label{ex:mudancavariavel}
\index{esperan\c ca}
\index{EX@$\E X$}
\end{exercicio}

\begin{observacao}
Em particular, 
\[\E X =\int_{\R}x\,d\mu(x),\]
onde $\mu$ \'e a lei de $X$.
\end{observacao}

\subsection{Converg\^encia dominada}

\begin{teorema}[Teorema de Lebesgue de converg\^encia dominada]
Seja $(f_n)$ uma sequ\^encia de fun\c c\~oes mensur\'aveis sobre um espa\c co boreliano padr\~ao $\Omega$ munido de uma medida de probabilidade, $\mu$. Suponhamos que $f_n$ convergem quase em toda parte para uma fun\c c\~ao $f$:
\[\mu\{x\in\Omega\colon f_n(x)\to f(x)\}=1,\]
e que existe uma fun\c c\~ao $L^1(\mu)$, $g$,
\[\int_\Omega\abs{g}\,d\mu<+\infty,\]
que limita por cima as fun\c c\~oes da sequ\^encia:
\[\forall n,~\abs{f_n}\leq \abs{g}.\]
Ent\~ao $f$ \'e mensur\'avel, e
\[\int_\Omega \abs{f_n-f}\,d\mu\to 0\mbox{ quando }n\to\infty,\]
ou seja, $f_n\to f$ na pseudom\'etrica $L^1(\mu)$.
\label{t:dominada}
\index{teorema! de converg\^encia dominada}
\end{teorema}

\begin{exercicio}
Seja $(f_n)$ uma sequ\^encia de fun\c c\~oes borelianas que converge simplesmente para uma fun\c c\~ao $f$ sobre um dom\'\i nio boreliano padr\~ao:
\[\forall x\in\Omega,~f_n(x)\to f(x)\mbox{ quando }n\to\infty.\]
Mostrar que $f$ \'e boreliana.
\label{ex:limitborelianas}
\end{exercicio}

\begin{observacao}
Todo subconjunto de um conjunto negligenci\'avel \'e negligen\-ci\'a\-vel, logo mensur\'avel. Por conseguinte, se uma fun\c c\~ao $f$ \'e boreliana sobre um subconjunto de um espa\c co probabil\'\i stico $\Omega$ que tem medida um, segue-se que $f$ \'e mensur\'avel sobre $\Omega$.
\label{o:mesuravel}
\end{observacao}

\begin{proof}[Prova do teorema \ref{t:dominada}]
Denotemos $Y$ um conjunto mensur\'avel da medida um sobre qual $f_n\to f$ simplesmente. 
Gra\c cas ao exerc\'\i cio \ref{ex:fmudag}, podemos supor sem perda de generalidade que as fun\c c\~oes $f_n\vert_Y$ s\~ao borelianas sobre $Y$. (O conjunto $Y$ tem uma estrutura boreliana --- n\~ao necessariamente padr\~ao --- induzida de $\Omega$).
Segundo o exerc\'\i cio \ref{ex:limitborelianas}, a fun\c c\~ao limite $f$ \'e boreliana sobre $Y$, logo mensur\'avel sobre $\Omega$ (oberva\c c\~ao \ref{o:mesuravel}). Seja $\ve>0$ qualquer fixo.
Para cada $n$, o conjunto
\[B_n=\{x\in Y\colon \forall n^\prime\geq n,~ \abs{f_n(x)-f(x)}<\ve\}\]
\'e boreliano (exerc\'\i cio), $B_n\subseteq B_{n+1}$, e $\cup_{n}B_n$ tem medida $1$. Segue-se que, quando $n\to\infty$, 
\[\mu(B_n)\to 1.\]
 Temos
\begin{eqnarray*}
\int_\Omega \abs{f_n-f}\,d\mu &=& \int_{B_n}+\int_{\Omega\setminus B_n} \\
&<& \ve + \int_{\Omega\setminus B_n} 2\abs{g}\,d\mu \\
&\to & \ve\mbox{ quando }n\to\infty,
\end{eqnarray*}
pois $\mu(\Omega\setminus B_n)\to 0$ quando $n\to\infty$, e $g$ \'e integr\'avel. 
\end{proof}

\subsection{Fun\c c\~ao carater\'\i stica de uma medida de probabilidade}

A {\em fun\c c\~ao carater\'\i stica}, $\phi_{\mu}$, de uma medida de probabilidade, $\mu$, sobre $\R^d$ \'e uma fun\c c\~ao de $\R^d$ para $\C$, dada por
\begin{align*}
\R^d\ni t\mapsto \phi_{\mu}(t)&=\E_{\mu}(\exp{i\langle t,X\rangle})\\
&= \int_{\R^d}\exp{i\langle t,x\rangle}\,d\mu(x)\in\C,
\end{align*}
onde $X\in\R^d$ segue a lei $\mu$.
\index{fun\c c\~ao! carater\'\i stica}
\'E claro que $\phi_{\mu}$ \'e sempre bem definida.

\begin{exercicio}
Mostrar as propriedades seguintes da fun\c c\~ao carater\'\i stica $\phi_{\mu}$.
\begin{enumerate}
\item $\phi_{\mu}$ \'e uniformemente cont\'\i nua.
\item $\phi_{\mu}(0)=1$ and $\abs{\phi_{\mu}}\leq 1$.
\end{enumerate}
\end{exercicio}

A fun\c c\~ao carater\'\i stica permite de reconstruir a medida $\mu$ unicamente. 

\begin{exercicio} 
Sejam $\mu_1,\mu_2$ duas medidas de probabilidade sobre $\R^d$. Mostrar que existe uma fun\c c\~ao cont\'\i nua $f\colon\R^d\to\R$, que se anula fora de um compacto, e tal que $\int f\,d\mu_1\neq\int f\,d\mu_2$. (As fun\c c\~oes de suporte compacto separam as medidas). 
\par
[ {\em Sugest\~ao:} existe um compacto $K\subseteq\R^d$ tal que $\mu_1(K)\neq\mu_2(K)$. Agora pode-se escolher um $\ve>0$ tal que $\mu_1(K_\ve)\neq\mu_2(K_\ve)$, onde $K_\ve$ denota a $\ve$-vizinhan\c ca de $K$....etc. ] Compare o exerc\'\i cio \ref{ex:funcoesseparammedidas}.
\label{ex:Ccsepara}
\end{exercicio}

\begin{teorema}
Sejam $\mu_1$ e $\mu_2$ duas medidas de probabilidade borelianas distintas sobre $\R^d$. Ent\~ao, $\phi_{\mu_1}\neq\phi_{\mu_2}$.
\label{t:carfundeterminam}
\end{teorema}

Suponha que $\phi_{\mu_1}=\phi_{\mu_2}$.
Segundo exerc\'\i cio \ref{ex:Ccsepara}, basta mostrar que para todas fun\c c\~oes $f\colon\R^d\to\R$ de suporte compacto temos $\int f\,d\mu_1 =\int f\,d\mu_2$. Fixemos uma tal fun\c c\~ao $f$ e um $\ve>0$ qualquer. 
Existe um real $L>0$ t\~ao grande que $\mbox{supp}\,f\subseteq [-L/2,L/2]^d$ e $\mu_i[-L/2,L/2]^d>1-\ve$, $i=1,2$. Formemos o toro (exemplo \ref{ex:toro})
\[\T^d=\R^d/L\Z^d\]
e definamos a fun\c c\~ao $\tilde f\colon T\to\R$ por
\[\tilde f(x+L\Z^d) = f(x),\]
para todos $x\in [-L/2,L/2]^d$. \'E claro que $\tilde f$ \'e bem definida \'e cont\'\i nua (exerc\'\i cio). 

Para cada $m\in\Z^d$, defina
\begin{align*}
g_m\colon\R^d&\to\C,\\
x&\mapsto \exp(i\langle \pi m/L, x\rangle)
\end{align*}
Seja $\mathcal F$ o espa\c co linear gerado pelas fun\c c\~oes $g_m$, $m\in\Z^d$. Este espa\c co \'e fechado pela multiplica\c c\~ao (exerc\'\i cio), logo, forma uma \'algebra. Tudo como as fun\c c\~oes geradoras, $g_m$, todos os elementos $g\in\mathcal F$ s\~ao peri\'odicos:
\[g(x+2Ln)=g(x),~~n\in\Z^d,~x\in\R^d.\]
Por isso, a fun\c c\~ao
\begin{align*}
\tilde g\colon \T^d\to \C,\\
\tilde g(x+2L\Z^d)=g(x)
\end{align*}
\'e bem definida, cont\'\i nua, e limitada. 

\begin{exercicio}
Verificar que o conjunto de fun\c c\~oes com valores complexes
\[\tilde{\mathcal F} = \{\tilde g\colon g\in \mathcal F\}\]
forma uma \'algebra, fechada pelas conjuga\c c\~oes.
\end{exercicio}

Como o toro $\T^d$ \'e compacto, o teorema de Stone--Weierstrass \ref{stone-weierstrass} implica que $\tilde{\mathcal F}$ \'e densa em \'algebra $C(T^d)$. Em particular, existe $\tilde g\in \tilde{\mathcal F}$ tal que $\norm{\tilde g - \tilde f}_{\infty}<\ve$ (a norma uniforme). Isso implica, segundo a defini\c c\~ao das duas fun\c c\~oes,
\[\norm{(f-g)\chi_{[-L,L]^d}}_{\infty}<\ve,\]
assim como
\begin{align*}
\norm{(f-g)\chi_{\R^d\setminus [-L,L]^d}}_{\infty} &\leq \norm{g}_{\infty} \\
&=\norm{\tilde g}_{\infty} \\
&\leq \norm{\tilde f}_{\infty}+\ve \\
&= \norm{f}_{\infty}+\ve.
\end{align*}

Para todo $m\in\Z^d$,
\begin{eqnarray*}
\int g_m\,d\mu_1 &= \phi_{\mu_1}(\pi m/L) \\
&= \phi_{\mu_2}(\pi m/L) \\
&=\int g_m\,d\mu_2,
\end{eqnarray*}
logo, para cada $g\in {\mathcal F}$,
\[\int g\,d\mu_1 = \int g\,d\mu_2.\]
Usando a \'ultima igualdade, conclu\'\i mos:
\begin{align*}
\left\vert \int f\,d\mu_1 -  \int f\,d\mu_2\right\vert & = 
\left\vert \int (f-g)\,d\mu_1 -  \int (f-g)\,d\mu_2\right\vert \\
&\leq \left\vert \int \abs{f-g}\,d\mu_1 +  \int \abs{f-g}\,d\mu_2\right\vert\\
&\leq \ve\left(2\norm{f}_{\infty}+2\ve+\mu_1(\R^d) +\mu_2(\R^d) \right).
\end{align*}
Como $\ve>0$ foi qualquer, isto termina a demonstra\c c\~ao do teorema \ref{t:carfundeterminam}. (A demonstra\c c\~ao foi emprestada do livro \citep*{klenke}, teorema 15.8).

\section{Teoremas de Riesz e de Banach--Alaoglu}

\subsection{Teorema de representa\c c\~ao de Riesz}
Relembramos que, para um espa\c co compacto $X$, o espa\c co de Banach $C(X)$ consiste de todas as fun\c c\~oes cont\'\i nuas sobre $X$, munidas da norma uniforme.
Um funcional linear $\phi$ sobre um espa\c co linear que consiste de fun\c c\~oes reais \'e dito {\em positivo} se $\phi(f)\geq 0$ cada vez que $f\geq 0$. \'E claro que, dada uma medida boreliana de probabilidade, $\mu$, sobre um espa\c co compacto m\'etrico $X$, o funcional
\[C(X)\ni f\mapsto \int_Xf(x)\,d\mu(x)\in\R\]
\'e linear, positivo, e envia a fun\c c\~ao $1$ para $1$. Verifica-se que cada funcional linear sobre $C(X)$ com estas propriedades \'e desta forma.

\begin{teorema}[Teorema de Representa\c c\~ao de Riesz]
Seja $X$ um espa\c co m\'etrico compacto. Ent\~ao a f\'ormula
\[\mu\mapsto [C(X)\ni f\mapsto \int_Xf(x)\,d\mu(x)\in\R]
\]
estabelece uma correspond\^encia bijetora entre as medidas de probabilidade borelianas sobre $X$ e os funcionais lineares positivos sobre o espa\c co de Banach $C(X)$ enviando $1$ para $1$.
\label{t:riesz}
\index{teorema! de Riesz}
\end{teorema}

\begin{exercicio}
Mostre que cada funcional linear e positivo $\phi\colon C(X)\to\R$ enviando $1$ para $1$ \'e limitado, de norma $1$.
\par
[ {\em Sugest\~ao:} dado uma $f\in C(X)$ com $\norm f_{\infty}\leq 1$, estudar as imagens por $\phi$ das fun\c c\~oes $f_+$, $1-f_+$, etc. ]
\label{ex:positivo1to1norma1}
\end{exercicio}

\begin{exercicio}
Seja $X$ um espa\c co compacto, e $F\subseteq K$ um subconjunto fechado. Mostre que para cada $\ve>0$, existe uma fun\c c\~ao cont\'\i nua $f_{\ve}\colon X\to [0,1]$ tal que $f_{\ve}\vert_F\equiv 1$ e $f_{\ve}$ se anula fora da $\ve$-vizinhan\c ca de $F$. Conclua que
\[\int_{X}f_{\ve}(x)\,d\mu(x)\to \mu(F)\mbox{ quando }\ve\to 0.\]
\label{ex:funcoesseparammedidas}
\end{exercicio}

\begin{exercicio} 
Deduza que se $\mu_1\neq\mu_2$, ent\~ao existe uma fun\c c\~ao cont\'\i nua $f\in C(X)$ tal que 
\[\int_Xf(x)\,d\mu_1(x)\neq \int_Xf(x)\,d\mu_2(x).\]
\end{exercicio}

Deste modo, apenas resta a mostrar que cada funcional linear positivo $\phi$, com $\phi(1)=1$, seja da forma
\[\phi(f) =\int_Xf(x)\,d\mu(x)\]
para uma medida de probabilidade apropriada sobre $X$. Com esta finalidade, precisamos mais dois resultados (teoremas \ref{t:HBpositivo} e \ref{compactoimagemdecantor}) que s\~ao importantes em si mesmos.

\begin{definicao}
Seja $V$ um espa\c co linear real.
Um {\em cone} (ou: {\em cone positivo}) em $V$ \'e um subconjunto $P$ tendo as propriedades seguintes:
\begin{itemize}
\item Se $x,y\in P$, ent\~ao $x+y\in P$.
\item Se $x\in P$ e $\lambda\geq\R$, ent\~ao $\lambda x\in P$.
\item Se $x\in P$ e $-x\in P$, ent\~ao $x=0$.
\end{itemize}
\end{definicao}

\begin{exemplo}
O nosso exemplo motivador \'e o espa\c co $V=C(X)$ de fun\c c\~oes cont\'\i nuas reais sobre um espa\c co compacto $X$, munido do cone $P=\{f\colon f\geq 0\}$.
\end{exemplo}

\begin{exercicio}
Dado um cone $P\subseteq V$, verifique que a rela\c c\~ao 
\[x\leq y\iff y-x\in P\]
\'e uma ordem parcial sobre $V$. (Portanto, 
o par $(V,P)$ se chama um {\em espa\c co linear parcialmente ordenado}). Deste modo, o cone $P$ consiste exatamente de todos os elementos (n\~ao estritamente) positivos:
\[P=\{x\in V\colon x\geq 0\}.\]
\end{exercicio}

\begin{definicao}
Um funcional linear $\phi\colon V\to \R$ sobre um espa\c co linear parcialmente ordenado $(V,P)$ \'e {\em positivo} se $f(x)\geq 0$ quando $x\in P$ (ou seja, $x\geq 0$).
\end{definicao}

\begin{observacao}
Se $X$ \'e um espa\c co linear parcialmente ordenado tendo o cone $P$, e $Y$ \'e um subespa\c co linear de $X$, ent\~ao $Y$ \'e parcialmente ordenado tamb\'em, com o cone $P\cap Y$.
\end{observacao}

\begin{teorema}[Teorema de Hahn--Banach para funcionais positivos]
Seja $V$ um espa\c co linear parcialmente ordenado, e seja $E$ um subespa\c co linear
de $V$. Ent\~ao cada funcional linear positivo $\phi$ sobre $E$ se estende at\'e um funcional linear positivo sobre $V$.
\label{t:HBpositivo}
\index{teorema! de Hahn--Banach}
\end{teorema}

A prova segue as mesmas linhas que a prova do teorema de Hahn--Banach para funcionais limitados (\ref{t:hb}). Sentimos que n\~ao h\'a alguma necessidade de replicar o argumento completo. Apenas um an\'alogo positivo do lema \ref{l:codim} pode colocar ligeiras dificuldades, e por esta raz\~ao, discutamos a prova aqui.

\begin{exercicio}
Mostrar que $\sup\emptyset=-\infty$ e $\inf\emptyset =+\infty$.
\label{ex:surinfinfty}
\end{exercicio}

\begin{lema}
\label{l:codimpositivo}
Sejam $G$ um subespa\c co vetorial de um espa\c co parcialmente ordenado $E=(E,P)$, $g$ um funcional linear positivo sobre $F$, e $x\in E$. Ent\~ao existe um funcional linear positivo $g_1$ sobre o espa\c co vetorial
\[G+\R x = \{y+tx\colon y\in G,t\in\R\},\]
gerado pela uni\~ao $G\cup\{x\}$, cuja restri\c c\~ao sobre $G$ \'e igual a $g$.
\end{lema}

\begin{proof}
A ideia da escolha \'e tamb\'em parecida \`a no lema \ref{l:codim}, s\'o que ao inv\'es da norma temos que usar a positividade para escolhermos corretamente a imagem, $a\in\R$, de $x$ por $g_1$. Seja $t\geq 0$ qualquer, $y\in G$. A condi\c c\~ao 
\[y+tx\geq 0\]
\'e equivalente a $x\geq t^{-1}y$, e deve implicar 
\[g_1(y+tx) = g(y)+ ta \geq 0,\]
ou seja $a\geq g(t^{-1}y)$. Da mesma maneira, a condi\c c\~ao $y-tx\geq 0$ \'e equivalente a $x\leq t^{-1}y$, e deve implicar $a\leq g(t^{-1}y)$. Conclu\'\i mos: a extens\~ao desejada existe se e somente se existe $a\in\R$ tal que, quaisquer que sejam $y^{\prime}, y^{\prime\prime}\in G$ tais que $y^\prime\leq x\leq y^{\prime\prime}$, temos
\[g(y^{\prime})\leq a \leq g(y^{\prime\prime}).\]
Como $y^{\prime}\leq y^{\prime\prime}$, temos que 
\[g(y^{\prime})\leq g(y^{\prime\prime}).\]
a por conseguinte,
\[\sup_{y^\prime\leq x}g(y^{\prime})\leq\inf_{y^{\prime\prime}\geq x} g(y^{\prime\prime}),\]
se permitirmos que o supremo e o \'\i nfimo assumam seus valores na reta estendida $\R\cup\{\pm\infty\}$, como no exerc\'\i cio \ref{ex:surinfinfty}.
Basta escolher $a$ de modo que 
\[\sup_{y^\prime\leq x}g(y^{\prime})\leq a \leq\inf_{y^{\prime\prime}\geq x} g(y^{\prime\prime}).\] 
\end{proof}

Agora, um resultado bastante importante da topologia geral.

\begin{teorema}
Seja $X$ um espa\c co m\'etrico compacto. Existe uma aplica\c c\~ao cont\'\i nua e sobrejetora $g\colon \{0,1\}^{\N}\to X$ do espa\c co de Cantor sobre $X$.
\label{compactoimagemdecantor}
\end{teorema}

\begin{proof}
Usando a pr\'e-compacidade de $X$, para cada $n$ construamos uma $2^{-n}$-rede finita $F_n$ para $X$. Escolhamos uma sequ\^encia $d_n\geq 1$ tendo a propriedade $2^{d_{n}}\geq \sharp F_n$. Agora construamos recursivamente uma sequ\^encia de aplica\c c\~oes sobrejetoras 
\[g_n\colon \{0,1\}^{\sum_{i\leq n}d_{i}} \to F_n,\]
$n\geq 1$, satisfazendo a propriedade: se $\sigma\in \{0,1\}^{\sum_{i\leq n}d_{i}}$, onde $n\geq 2$, ent\~ao
\[d_X(g_n(\sigma),g_{n-1}(\sigma\upharpoonright[\sum_{i\leq n-1}d_{i}])<2^{-n+1}.
\]
Dado um $\sigma\in \{0,1\}^{\N}$, a sequ\^encia de pontos 
\[g_n(\sigma\upharpoonright [\sum_{i\leq n}d_{i}])\in X\]
\'e Cauchy, e portanto possui um e apenas um limite em $X$, que denotemos por $g(\sigma)$. Verifica-se facilmente que a aplica\c c\~ao $g\colon \{0,1\}^{\N}\to X$ \'e sobrejetora, usando o fato que cada ponto $x\in X$ \'e o limite de uma sequ\^encia da forma $(x_n)$, onde $x_n\in F_n$ e $d(x,x_n)<2^{-n}$. A imagem por $g$ de um conjunto aberto cil\'\i ndrico da forma 
\[\pi^{-1}_{[\sum_{i\leq n}d_{i}]}(\tau)\]
tem di\^ametro $\leq \sum_{i=n-1}^{\infty} 2^{-i}= 2^{-n+2}$, o que estabelece a continuidade de $g$.
\end{proof}

\begin{exercicio}
Seja $g\colon X\to Y$ uma aplica\c c\~ao cont\'\i nua entre dois espa\c cos compactos. Ent\~ao, a f\'ormula
\[g^{\ast}(f)(x) = f(g(x))\]
estabelece um operador linear limitado e positivo $g^\ast\colon C(Y)\to C(X)$, enviando $1$ para $1$. \'E o {\em operador dual} a $g$.
\label{ex:operadordual}
\end{exercicio}

Agora seja $\phi$ um funcional linear positivo qualquer sobre $C(X)$, enviando $1$ para $1$. Escolhemos uma aplica\c c\~ao sobrejetora $g\colon \{0,1\}^{\N}\to X$. Formemos o operador dual $g^\ast\colon C(X)\to C(\{0,1\}^{\N})$. Como $g$ \'e sobrejetora, o operador $g^{\ast}$ \'e uma imers\~ao isom\'etrica (exerc\'\i cio). O funcional $\phi$, definido sobre um subespa\c co fechado do espa\c co de Banach $C(\{0,1\}^{\N})$, se estende at\'e um funcional linear positivo $\tilde\phi$ sobre $C(\{0,1\}^{\N})$, gra\c cas ao teorema de Hahn--Banach na vers\~ao parcialmente ordenada \ref{t:HBpositivo}. \'E claro que $\tilde\phi(1)=\phi(1)=1$, logo $\tilde\phi$ \'e limitado, da norma $1$ (exerc\'\i cio \ref{ex:positivo1to1norma1}).

Os conjuntos cil\'\i ndricos de $\{0,1\}^{\N}$ s\~ao exatamente os conjuntos abertos e fechados (exerc\'\i cio). Para cada aberto e fechado, $V$, a fun\c c\~ao $\chi_V$ \'e cont\'\i nua. Definamos
\[\nu(V) = \tilde\phi(\chi_V).\]

\begin{exercicio}
Verifique que a medida $\nu$ sobre o campo de abertos e fechados do espa\c co $\{0,1\}^{\N}$ \'e finitamente aditiva, logo sigma-aditiva pelo mesmo argumento que no exemplo \ref{ex:verrrrificacao}.
\end{exercicio}

Agora $\nu$ estende-se at\'e uma medida de probabilidade boreliana sobre $\{0,1\}^{\N}$ (teorema de Carath\'eodory \ref{t:extensao}). As combina\c c\~oes lineares finitas de fun\c c\~oes indicadoras de conjuntos abertos e fechados satisfazem as hip\'oteses do teorema de Stone--Weierstrass \ref{stone-weierstrass} (exerc\'\i cio), logo s\~ao densas em $C(\{0,1\}^{\N})$. 

\begin{exercicio}
Mostrar este fato diretamente, sem usar o teorema de Stone--Weierstrass.
\end{exercicio} 

Segue-se que para cada $f\in C(\{0,1\}^{\N})$,
\[\tilde\phi(f) = \int_{\{0,1\}^{\N}}f(x)\,d\nu(x).\]
Denotemos 
\[\mu = g_{\ast}(\nu)\]
a imagem direta da medida $\nu$ pela aplica\c c\~ao cont\'\i nua $g\colon \{0,1\}^{\N}\to X$. Qualquer que seja $f\in C(X)$, temos
\begin{align*}
\phi(f) &=\tilde\phi(g^{\ast}(f))\\
& = \int_{\{0,1\}^{\N}}f(g(x))\,d\nu(x) \\
&= \int_X f(x)\,d\mu,
\end{align*}
estabelecendo o teorema de Riesz \ref{t:riesz}.

\subsection{Teorema de Banach--Alaoglu}

\begin{definicao}
Seja $E$ um espa\c co normado qualquer. O espa\c co dual $E^\ast$ consiste de todos os funcionais lineares cont\'\i nuas sobre $E$, munido das opera\c c\~oes naturais da adi\c c\~ao e da multiplica\c c\~ao escalar, assim que da norma operadora (defini\c c\~ao \ref{d:normaoperadora}).
\end{definicao}

\begin{exercicio}
Mostrar que $E^\ast$ \'e um espa\c co de Banach (ou mostre o exerc\'\i cio \ref{ex:b(ef)}, mais geral).
\end{exercicio}

Gra\c cas ao teorema de Hahn--Banach, o espa\c co dual n\~ao \'e trivial. Suas elementos {\em separam pontos} de $E$: se $x,y\in E$, $x\neq y$, ent\~ao existe $\phi\in E^\ast$ tal que $\phi(x)\neq \phi(y)$ (corol\'ario \ref{c:separacaodepontos}).

Denotemos por $B_{E^\ast}=\{\phi\in E^\ast\colon\norm{\phi}\leq 1\}$ a bola unit\'aria {\em fechada} do espa\c co dual de $E$.

\begin{teorema}[Teorema de Banach--Alaoglu, caso separ\'avel]
Seja $E$ um espa\c co normado separ\'avel. Existe uma m\'etrica sobre a bola unit\'aria $B_{E^\ast}$ do espa\c co dual $E^\ast$ tal que a bola \'e compacta, e a restri\c c\~ao de cada funcional $\phi\in E^\ast$ sobre $B_{E^\ast}$ \'e cont\'\i nua.
\label{t:banach-alaoglu}
\index{teorema! de Banach--Alaoglu}
\end{teorema}

Esta topologia se chama a {\em topologia fraca$^\ast$}.
\index{topologia! fracaest@fraca$^\ast$}

\begin{observacao}
A topologia fraca$^\ast$ \'e bem definida e o teorema de Banach--Alaoglu vale para um espa\c co normado qualquer, s\'o que para espa\c cos n\~ao separ\'aveis a prova j\'a exige o Axioma de Escolha (o teorema de Tychonoff sobre a compacidade do produto de uma fam\'\i lia qualquer dos espa\c cos topol\'ogicos compactos). Como n\'os apenas precisamos uma vers\~ao separ\'avel, d\^emos uma vers\~ao restrita do resultado.
\end{observacao}

Transformamos $[0,1]^{\N}$ num espa\c co m\'etrico como segue:
\begin{equation}
d(x,y) = \sum_{k\in\N_+} 2^{-k}\abs{x_k-y_k}.
\label{eq:metricanocubo}
\end{equation}

\begin{exercicio}
Mostrar que $[-1,1]^{\N}$, munido da m\'etrica acima, \'e um espa\c co compacto.
\par
[ {\em Sugest\~ao:} mostre que a m\'etrica acima gera a topologia produto no sentido da defini\c c\~ao \ref{d:topologiadeproduto}, e use o exerc\'\i cio \ref{ex:produtocompacto}. ]
\end{exercicio}

Seja $\bar x=(x_k)$ uma sequ\^encia de elementos da bola unit\'aria $B$ de $E$ cujos elementos s\~ao densos na bola. 
Definamos uma aplica\c c\~ao $i_{\bar x}$ da bola $B_{E^\ast}$ para o cubo $[-1,1]^{\N}$ como segue:
\[B_{E^\ast}\ni\phi\mapsto i_{\bar x}(\phi),\mbox{ onde }\forall k,~~
i(\phi)_{k} = \phi(x_k)\in [-1,1].\]

\begin{exercicio}
Mostre que a aplica\c c\~ao $i_{\bar x}$ \'e injetora e cont\'\i nua.
\end{exercicio}

\begin{lema}
A imagem da bola $B_{E^\ast}$ por $i_{\bar x}$ \'e fechada no cubo $[-1,1]^{\N}$, logo compacta.
\label{l:imagemdepx}
\end{lema}

Seja $(\phi_n)$ uma sequ\^encia de elementos da bola cujas imagens $i_{\bar x}(\phi_n)$ formam uma sequ\^encia de Cauchy em rela\c c\~ao \`a m\'etrica (\ref{eq:metricanocubo}). 
A propriedade de Cauchy \'e equivalente \`a condi\c c\~ao seguinte: para cada $\ve>0$ e $M$, existe $N$ e $\delta>0$ tais que, se $n\geq N$, ent\~ao 
\[\abs{\phi_n(x_k)-\phi_N(x_k)}<\ve\mbox{ quando }k=1,2,\ldots,M.\]
Em particular, para cada $k\in\N$, a sequ\^encia $(\phi_n(x_k))$ \'e Cauchy em $\R$, logo possui um limite, que denotemos por $\phi(x_k)$.

\begin{exercicio}
Verifique que a aplica\c c\~ao com o dom\'\i nio $\{x_k\}_{k\in\N}$
\[x_k\mapsto \phi(x_k)\in\R\]
\'e Lipschitz cont\'\i nua de constante $1$.
\end{exercicio}

\begin{exercicio}
Seja $f\colon X\to Y$ uma aplica\c c\~ao uniformemente cont\'\i nua entre dois espa\c cos m\'etricos. Mostre que $f$ admite uma e uma s\'o prorroga\c c\~ao uniformemente cont\'\i nua sobre os completamentos: $\hat f\colon \hat X\to \hat Y$. Se $f$ for Lipschitz cont\'\i nua, ent\~ao $\hat f$ \'e Lipschitz cont\'\i nua, com a mesma constante.
\end{exercicio}

Estendamos $\phi$ at\'e uma fun\c c\~ao (que denotemos pela mesma letra) de $B_{E^\ast}$ para $\R$.

\begin{exercicio}
Sejam $x,y\in B_{E^\ast}$ and $\lambda\in\R$ tais que $x+y\in B_{E^\ast}$ e $\lambda x\in B_{E^\ast}$. Mostre que $\phi(x+y)=\phi(x)+\phi(y)$ e $\phi(\lambda(x)=\lambda\phi(x)$.
\end{exercicio}

\begin{exercicio}
Mostre que $\phi$ admite uma extens\~ao \'unica at\'e um funcional linear da norma $\leq 1$, definido sobre $E$.
\end{exercicio}

Isto conclui a prova do lema \ref{l:imagemdepx}.

Agora coloquemos sobre $B_{E^\ast}$ a topologia induzida do cubo $[-1,1]^{\N}$ pela imers\~ao da bola como um subespa\c co compacto com ajuda da aplica\c c\~ao $i_{\bar x}$. Em outras palavras, $B_{E^\ast}$ \'e munido da m\'etrica
\[d_{\bar x}(\phi,\psi) = d(i_{\bar x}(\phi), i_{\bar x}(\psi)),\]
onde $d$ \'e como na eq. (\ref{eq:metricanocubo}).

\begin{exercicio}
Deduza dos exerc\'\i cios \ref{ex:produtocompacto} e \ref{ex:topprodmenorquepi} que a topologia induzida pela m\'etrica $d_{\bar x}$ sobre a bola $B_{E^\ast}$ (atrav\'es da imers\~ao $i_{\bar x}$) \'e a menor topologia tal que todas as aplica\c c\~oes $\phi(x_n) = \pi_n(i_{\bar x}(\phi)$, $n\in\N$ s\~ao cont\'\i nuas. 
\label{ex:descricaotopboladual}
\end{exercicio}

\begin{lema}
Em rela\c c\~ao \`a m\'etrica $d_{\bar x}$, para cada $x\in B_E$, a aplica\c c\~ao 
\begin{equation}
B_{E^\ast}\ni \phi\mapsto \phi(x)\in\R
\label{eq:weakstarappl}
\end{equation}
\'e cont\'\i nua.
\label{l:emrelacaoecontinua}
\end{lema}

\begin{proof}
Seja $x\in B_E$ qualquer. Formemos uma nova sequ\^encia, incluindo $x$ como o primeiro elemento: $\tilde x = (x,x_1,x_2,\ldots)$. A topologia induzida sobre a bola pela m\'etrica $d_{\tilde x}$ \'e tamb\'em compacta, e mais fina do que a topologia induzida pela m\'etrica $d_{\bar x}$ (exerc\'\i cio \ref{ex:descricaotopboladual}). Corol\'ario \ref{c:homeo} aplicada \`a aplica\c c\~ao identidade $\mbox{id}_{B_{E^\ast}}$ implica que as duas topologias s\~ao iguais, logo a aplica\c c\~ao $\phi\mapsto \phi(x)$ \'e cont\'\i nua em rela\c c\~ao \`as ambas.
\end{proof}

\begin{exercicio}
Conclua do exerc\'\i cio \ref{ex:descricaotopboladual} e do lema \ref{l:emrelacaoecontinua}, que, qualquer que seja sequ\^encia $\bar x$ densa na bola $B_E$, a topologia gerada pela m\'etrica $d_{\bar x}$ sobre a bola dual $B_{E^\ast}$ \'e a menor topologia tal que para cada $x\in B$ a aplica\c c\~ao $\phi\mapsto\phi(x)$ \'e cont\'\i nua.
\end{exercicio}

Segue-se que $\bar x$ e $\bar y$ s\~ao duas sequ\^encias densas na bola $B_E$ quaisquer, ent\~ao as m\'etricas $d_{\bar x}$ e $d_{\bar y}$ s\~ao equivalentes sobre a bola $B_{E^\ast}$, o que conclui a prova do teorema \ref{t:banach-alaoglu}.

\begin{exercicio}
Mostrar que a topologia normada sobre $B_{E^\ast}$ \'e mais forte do que \`a topologia fraca$^\ast$, e que as duas coincidem se e somente se $\dim E<\infty$.
\end{exercicio}

\subsection{Topologia sobre o espa\c co de medidas de probabilidade}
Denotemos $P(X)$ o conjunto de todas as medidas de probabilidade sobre um espa\c co compacto $X$. 
Segundo o teorema de Riesz, $P(X)$ pode se identificar com o conjunto de funcionais lineares e positivos sobre $C(X)$, enviando $1$ para $1$. Deste modo, $P(X)$ \'e um subconjunto da bola dual $B_{C(X)^\ast}$. 
\index{PX@$P(X)$}

\begin{proposicao}
Seja $X$ um espa\c co m\'etrico compacto. Ent\~ao o espa\c co de Banach $C(X)$ \'e separ\'avel.
\label{p:c(x)separavel}
\end{proposicao}

Isso segue imediatamente da combina\c c\~ao de dois resultados estabelecidos na prova do teorema de Riesz: o espa\c co $C(X)$ \'e um subespa\c co normado do espa\c co $C(\{0,1\}^{\N})$, e o \'ultimo espa\c co \'e separ\'avel. No mesmo tempo, o resultado \'e muito mais simples, e n\~ao exige algumas ferramentas avan\c cadas. Eis uma prova direta deste fato, que vamos fazer s\'o para $X=[0,1]$.

\begin{exercicio}
Denotemos por $A$ o subconjunto do espa\c co $\ell^\infty[0,1]$ que consiste de todas as fun\c c\~oes $f\colon [0,1]\to \Q$ tais que existe $n\in\N_+$ com a propriedade que $f$ \'e constante sobre cada intervalo semiaberto 
\[\left[\frac in,\frac{i+1}n\right),~~i=0,1,2,\ldots,n-1.\]
Mostrar que $A$ \'e enumer\'avel.
\end{exercicio}

\begin{exercicio}
Sejam $f\in C[0,1]$ e $\e>0$ quaisquer. Escolhamos $\delta>0$ tal que
\[\forall x,y\in [0,1],~~\abs{x-y}<\delta \Rightarrow \abs{f(x)-f(y)}<\e.\]
Existe $n$ tal que 
\[\frac 1n<\delta\]
(a propriedade de Arquimedes). 
Definiremos a fun\c c\~ao $g\in A$ pelos condi\c c\~oes seguintes: $g$ \'e constante sobre cada intervalo $[i/n,(i+1)/n)$, $i=0,1,\ldots,n-1$, e
\[\forall i=1,2,3,\ldots,n,~~ g\left(\frac in\right)\in \Q\mbox{ e }\left\vert g\left(\frac in\right) - f\left(\frac in\right)\right\vert < \e.\]
Mostrar que $d_{\infty}(d,f)<2\e$.
\end{exercicio}

Segue-se que o subespa\c co $\bar A$ (a ader\^encia em $\ell^\infty[0,1]$) \'e separ\'avel e cont\'em $C[0,1]$, logo $C[0,1]$ \'e separ\'avel (teorema \ref{t:subespsep}).

\begin{exercicio}
Generalizar o argumento no caso do espa\c co m\'etrico $C(X)$, onde $X$ \'e um espa\c co m\'etrico compacto qualquer.
\end{exercicio}

\begin{lema}
O espa\c co $P(X)$ \'e fechado na bola dual $B_{C(X)^\ast}$ em rela\c c\~ao \`a topologia fraca$^\ast$, logo compacto.
\end{lema}

\begin{proof}
Seja $\phi\in B_{C(X)^\ast}\setminus P(X)$. Indiquemos uma vizinhan\c ca $V$ de $\phi$ na topologia fraca$^\ast$ disjunta de $P(X)$. Se $\phi(1)=a\neq 1$, ent\~ao definamos $\e=\abs{a-1}$ e
\[V = \{\psi\in B_{C(X)^\ast}\colon \psi(1)\in (a-\e,a+\e)\}.\]
Se $\phi$ n\~ao \'e positivo, existe $f\in B_E$ tal que $f\geq 0$ e $\phi(f)=b<0$. Definamos 
\[V = \{\psi\in B_{C(X)^\ast}\colon \psi(f)\in (2b,0)\}.\]
\end{proof}

Chegamos ao resultado seguinte.

\begin{teorema} 
O espa\c co $P(X)$ de medidas borelianas de probabilidade sobre um espa\c co compacto metriz\'avel $X$ \'e munido de uma topologia compacta metriz\'avel, que \'e a menor topologia tal que todas as aplica\c c\~oes
\[\mu\mapsto \int_Xf(x)\,d\mu(x)\in\R\]
s\~ao cont\'\i nuas, quando $f\in C(X)$.
\label{t:topologiaPX}
\end{teorema}

Esta topologia tem o mesmo nome: topologia fraca$^\ast$ sobre $P(X)$.

\section{Converg\^encia de martingales e derivada de Radon--Nikod\'ym}

\subsection{Teorema de converg\^encia de martingales}
N\'os j\'a vimos martingales sobre espa\c cos probabil\'\i sticos finitos na subse\c c\~ao \ref{ss:martingales}. Agora consideremos martingales um pouco mais complicados: ainda definidas pelas parti\c c\~oes finitas, embora formando uma sequ\^encia infinita, sobre um espa\c co probabil\'\i stico padr\~ao.

\begin{definicao}
Seja $(X,\mu)$ um espa\c co probabil\'\i stico padr\~ao, e seja 
\[\Omega_0\succ \Omega_1\succ\dots\succ \Omega_{n}\succ\Omega_{n+1}\succ
\dots\]
uma sequ\^encia refinadora de parti\c c\~oes finitas mensur\'aveis de $\Omega$, onde $\Omega_0=\{\Omega\}$ \'e uma parti\c c\~ao trivial. Um martingale (formado em rela\c c\~ao \`a sequ\^encia acima) \'e uma cole\c c\~ao de fun\c c\~oes mensur\'aveis reais $(f_0,f_1,\dots)$ sobre $X$, que pertencem a $L^1(\mu)$ e satisfazem, para todo $i=1,2,\dots$,
\[E_{\mu}(f_i\mid\Omega_{i-1})=f_{i-1}.\]
\end{definicao}

A esperan\c ca condicional \'e entendida no sentido da defini\c c\~ao \ref{d:esperancacondicional}.

\begin{definicao}
Digamos que uma fun\c c\~ao mensur\'avel real $f$ sobre um espa\c co probabil\'\i stico padr\~ao $(\Omega,\mu)$ \'e {\em essencialmente limitada} se existe uma fun\c c\~ao limitada, $g$, que difere de $f$ sobre um conjunto negligenci\'avel. O {\em supremo essencial} de $f$ \'e o valor
\[
\mbox{ess}\sup f = \inf\{\sup{f\vert A}\colon \mu(A)=1\}.
\]
\end{definicao}

\begin{exercicio}
Verifique que as fun\c c\~oes mensur\'aveis e essencialmente limitadas formam um subespa\c co linear de $\R^{\Omega}$, e
a f\'ormula
\[\norm{f}_{\infty} = \mbox{ess}\sup \abs{f}\]
defina uma pre-norma sobre este espa\c co.
\end{exercicio}

O espa\c co acima \'e notado $L^{\infty}(\Omega,\mu)$, ou simplesmente $L^{\infty}(\mu)$.
O espa\c co normado associado ao espa\c co acima \'e denotado pelo mesmo s\'\i mbolo.

\begin{exercicio}
Mostre que o espa\c co normado $L^{\infty}(\mu)$ \'e um espa\c co de Banach.
\par
[ {\em Sugest\~ao:} use o teorema \ref{th:complet}. ]
\end{exercicio}

O resultado seguinte forma uma ferramenta t\'ecnica maior. 

\begin{teorema}[Primeiro teorema de Doob de converg\^encia de martingale]
Seja $(f_n)$ um martingale, tal que $\norm{f_n}_{1}<\infty$ uniformemente em $n$. Ent\~ao $f_n$ converge em quase todo ponto para um limite que pertence a $L^1(\mu)$.
\label{t:doobverdadeiro}
\index{teorema! de converg\^encia de martingales}
\end{teorema}

Vamos ``quase'' mostrar o resultado, e a \'unica coisa que nos faltar\'a, ser\'a o lema de Fatou, que garante que o limite simples da sequ\^encia $(f_n)$ perten\c ca a $L^1(\mu)$. Como queremos evitar uma prova extra bastante t\'ecnica, formulemos uma vers\~ao do teorema de Doob mais fraca poss\'\i vel do que atende \`as nossas necessidades.

\begin{teorema}
Seja $(f_n)$ um martingale, tal que $\norm{f_n}_{\infty}<\infty$ uniformemente em $n$. Ent\~ao $f_n$ converge em $L^1(\mu)$ para um limite.
\label{t:doob}
\index{teorema! de converg\^encia de martingales}
\end{teorema}

\begin{exercicio}
Mostre que sem a hip\'otese sobre as normas $L^{\infty}$ de $f_n$, a conclus\~ao do teorema \ref{t:doob} n\~ao \'e verdadeira.
\end{exercicio}

\begin{exercicio}
Mostre que, se apenas exigirem que as normas $L^1(\mu)$ de $f_n$ sejam uniformemente limitadas, pode ocorrer que $\E_{\mu}f_n=1$ para todos $n$, portanto $f_n\to 0$ q.c.
\end{exercicio}

\begin{observacao}
O segundo teorema de Doob d\'a condi\c c\~oes necess\'arias e suficientes para $(f_n)$ convergirem em $L^1(\mu)$.
No entanto, s\'o vamos precisar do resultado mais fraco \ref{t:doob}.
\end{observacao}

\begin{definicao}
Sejam $(x_n)$ uma sequ\^encia de reais, finita ou infinita, e $a<b$. Uma {\em travessia para cima} ({\em upcrossing}) do intervalo $(a,b)$ pela sequ\^encia $(x_n)$ \'e qualquer par de elementos $x_m,x_n$, onde $m<n$, $x_m\leq a$, $x_n\geq b$, e para qualquer $k$ com $m<k<n$, temos $x_k\in (a,b)$. 
\end{definicao}

\begin{definicao}
O n\'umero total de travessias para cima do intervalo $(a,b)$ pela sequ\^encia $(x_n)$ \'e denotado $U[(x_n),(a,b)]$.
\end{definicao}

De mesmo modo, as {\em travessias para baixo} ({\em downcrossings}) s\~ao definidas, e o seu n\'umero \'e denotado $D((x_n),(a,b))$. \'E claro que a diferen\c ca entre os n\'umeros $U[(x_n),(a,b)]$ e $D[(x_n),(a,b)]$ pode ser $\pm 1$ ao m\'aximo.

\begin{exercicio}
Mostre que uma sequ\^encia limitada $(x_n)$ de reais converge para um limite se e somente se, para todos $a<b$ (equivalente: para todos $a<b$ racionais), o n\'umero de travessias para cima do intervalo $(a,b)$ pela sequ\^encia \'e finito.
\label{ex:updowncrossings}
\end{exercicio}

Agora o teorema \ref{t:doob} segue do resultado seguinte, aplicado ao conjunto de intervalos com pontos extremos racionais, e depois combinado com o teorema de converg\^encia dominada \ref{t:dominada}. Relembramos que $f_-$ denota a parte negativa de uma fun\c c\~ao:
\[f_-=\max\{-f,0\}.\]

\begin{lema}[Lema de Doob de travessias para cima]
Suponha que $L^1(f_n)<\infty$ uniformemente em $n$. Ent\~ao, para todos $a<b$ e $N\in\N$,
\[(b-a)\E_{x\sim\mu}U[(f_i(x))_{i=1}^N,(a,b)] \leq \E_{\mu}(f_N-a)_-.
\]
\label{l:doobtravessias}
\index{lema! de Doob}
\end{lema}

Definamos recursivamente as fun\c c\~oes $t_k$, $k\in\N$, e $s_k$, $k\in\N_+$, sobre $X$ com valores em $\N$ (ou: realiza\c c\~oes de vari\'aveis aleat\'orias)  como segue:
\begin{align*}
t_0 &=0,\\
s_k &= \min\{n > t_{k-1}\colon f_n\leq a\}, \\
t_k &= \min\{n>s_{k}\colon f_n\geq b\}.
\end{align*}
Deste modo,
\[0=t_0<s_1<t_1<s_2<\ldots\]

\begin{exercicio}
Verifique que
\[U[(f_i(x))_{i=1}^N,(a,b)] = \max\{k\colon t_k(x)\leq N\}.\]
\end{exercicio}

\begin{definicao}
Uma fun\c c\~ao $N\colon X\to \N$ \'e chamada {\em tempo de parada} se para todo $n$, o evento
\[\{x\in X\colon N(x)=n\}\]
pertence \`a \'algebra de conjuntos gerada pela parti\c c\~ao $\Omega_n$.
\end{definicao}

\begin{exercicio}
Mostre que $t_k$ e $s_k$ s\~ao tempos de parada.
\end{exercicio}

Para dois fun\c c\~oes $f$ e $g$, usemos a nota\c c\~ao 
\[(f\wedge g)(x)=\min\{f(x),g(f)\}.\]

\begin{exercicio}
Mostre que, se $(f_n)$ \'e um martingale e $N$ \'e um tempo de parada, ent\~ao a sequ\^encia $(f_{n\wedge N})$ \'e um martingale. A fun\c c\~ao $f_{n\wedge N}$ deve ser entendida como
\[f_{n\wedge N}(x) = f_{\min\{n,N(x)\}}.\]
[ {\em Sugest\~ao:} para um elemento $A\in\Omega_{n-1}$ e os elementos $A_i\in\Omega_n$, $\cup A_i=A$, considere v\'arias possibilidades para o valor de $n\wedge N$ sobre $A$ e $A_i$. ]
\end{exercicio}

\begin{teorema}
Seja $(f_n)$ um martingale e $N$ um tempo de parada. Ent\~ao,
\[\E f_{N\wedge n} = \E f_0.\]
\label{t:nwedgeN}
\end{teorema}

Para mostrar o resultado, precisamos da no\c c\~ao seguinte.

\begin{definicao}
Uma sequ\^encia $h=(h_n)$ de fun\c c\~oes sobre $X$ chama-se {\em previs\'\i vel} se cada $h_n$ \'e mensur\'avel em rela\c c\~ao \`a \'algebra gerada por $\Omega_{n-1}$, ou seja, \'e constante sobre os elementos de $\Omega_{n-1}$.
\end{definicao}

Dado um martingale $f=(f_n)$ e uma sequ\^encia previs\'\i vel $(h_n)$, definamos a {\em transfor\-ma\-\c c\~ao martingale,}
\[(h\cdot f)_n = f_0+\sum_{i=1}^n h_jd_j,\]
onde $d_j=f_j-f_{j-1}$ s\~ao diferen\c cas de martingale (defini\c c\~ao \ref{d:diferencasdemartingale}). Por exemplo, se $h_n=1$, obtemos o martingale original.

\begin{exercicio}
Mostre que a transforma\c c\~ao martingale $((h\cdot f)_n)_{n=0}^{\infty}$ \'e um martingale.
\par
[ {\em Sugest\~ao:} use proposi\c c\~oes \ref{p:diffprop} e \ref{properties}.(\ref{factorout}). ]
\end{exercicio}

\begin{exercicio}
Mostre que para cada $n$, 
\[\E(h\cdot f)_n = \E f_0.\]
\label{ex:expectacaodatransformacao}
\end{exercicio}

Modulo o exerc\'\i cio \ref{ex:expectacaodatransformacao}, a prova do teorema \ref{t:nwedgeN} ser\'a terminada pela observa\c c\~ao seguinte.

\begin{lema}
Seja $(f_n)$ um martingale e $N$ um tempo de parada. Ent\~ao,
\[(f_{n\wedge N}) = (z\cdot f)_n,\]
onde $(z_n)$ \'e uma sequ\^encia previs\'\i vel, dada por
\[z(x) = \begin{cases} 1,&\mbox{ se }N\geq n,\\
0,&\mbox{ se } N<n
\end{cases}
\]
\end{lema}

\begin{proof}
O evento $z_n=1$ \'e igual ao evento $N\geq n$, e ele \'e $\Omega_{n-1}$-mensur\'avel porque o evento complementar $N\leq n-1$ \'e (defini\c c\~ao de tempo de parada). Logo, $(z_n)$ \'e uma sequ\^encia previs\'\i vel. Agora,
\begin{align*}
(z\cdot f)_n &= f_0 + \sum_{i=1}^n z_n(f_n-f_{n-1})\\
&= f_0 + \sum_{i=1}^{n\wedge N} (f_n-f_{n-1})\\
&= f_{n\wedge N},
\end{align*}
pois a soma \'e telesc\'opica.
\end{proof}

\begin{proof}[Prova do lema \ref{l:doobtravessias}]
Denotemos $K(x)=U[(f_i(x))_{i=1}^N,(a,b)]$. Qualquer que seja $n$, temos:

\begin{align*}
0 & = \sum_{j=1}^{\infty} (\E f_0 - \E f_0) \\
&= 
\E\sum_{j=1}^{\infty} (f_{t_j\wedge n} - f_{s_j\wedge n}) 
\\
&= \E\sum_{j=1}^{K(x)}(f_{t_j} - f_{s_j}) + \E(f_n-f_{s_K(x)})\\
&\geq \E K(x) (b-a) - \E (f_n-a)_-,
\end{align*}
e o resultado segue.
\end{proof}

\begin{exercicio}
Deduza o teorema \ref{t:doob} do lema \ref{l:doobtravessias} e o exerc\'\i cio \ref{ex:updowncrossings}.
\end{exercicio}

\subsection{Derivada de Radon--Nikod\'ym\label{ss:RN}}

\begin{definicao}
Sejam $\mu$ e $\nu$ duas medidas sobre um espa\c co boreliano padr\~ao, $\Omega$. Diz-se que $\mu$ \'e absolutamente cont\'\i nua em rela\c c\~ao a $\nu$, nota\c c\~ao
\[\mu\ll\nu,\]
se cada conjunto $\nu$-negligenci\'avel \'e $\mu$-negligenci\'avel:
\[\forall B\in {\mathscr B}_{\Omega},~\nu(B)=0 \Rightarrow \mu(B)=0.\]
\end{definicao}

\begin{exercicio}
Sejam $\mu$ uma medida finita boreliana sobre um espa\c co boreliano padr\~ao, $\Omega$, e $f\in L^1(\mu)$. Mostre que a f\'ormula
\[(f\mu)(B) = \int_{\Omega}\chi_B d\mu\]
defina uma medida finita boreliana, $f\mu$, sobre $\Omega$.
\end{exercicio}

\begin{exemplo}
$f\mu\ll\mu$.
\end{exemplo}

O teorema de Radon--Nikod\'ym diz que isto \'e o \'unico exemplo poss\'\i vel.

\begin{teorema}[Teorema de Radon--Nikod\'ym]
Sejam $\mu$ e $\nu$ duas medidas finitas sobre um espa\c co boreliano padr\~ao, $\Omega$, tais que
\[\mu\ll\nu.\]
Ent\~ao, existe uma fun\c c\~ao $f\in L^1(\nu)$ tal que
\[\mu=f\nu.\]
\label{t:RN}
\index{teorema! de Radon--Nikod\'ym}
\end{teorema}

\begin{definicao}
A fun\c c\~ao $f$ no teorema \ref{t:RN} chama-se a {\em derivada de Radon--Nikod\'ym}. Nota\c c\~ao:
\[f=\frac{d\mu}{d\nu}.\]
\end{definicao}

\begin{exercicio}
Mostre o teorema de Radon--Nikod\'ym no caso onde $\nu$ \'e puramente at\^omica. Deduza que basta mostrar o teorema sob a hip\'otese que $\nu$ seja uma medida n\~ao at\^omica.
\label{ex:bastaconsiderarnaoatomicas}
\end{exercicio}

Na parte principal, somente usamos o teorema de Radon--Nikod\'ym para mostrar a exist\^encia da fun\c c\~ao de regress\~ao (p. \pageref{page:usingRN}), e neste caso, a medida $\mu$ satisfaz uma forte restri\c c\~ao adicional: para cada conjunto $A$ mensur\'avel, $\mu(A)\leq\nu(A)$. A prova torna se consideravelmente mais simples. Vamos apenas mostrar este resultado:

\begin{teorema}[Teorema de Radon--Nikod\'ym, caso especial]
Sejam $\mu$ e $\nu$ duas medidas finitas sobre um espa\c co boreliano padr\~ao, $\Omega$, tais que para cada $A\in{\mathscr B}_{\Omega}$,
\[\mu(A)\leq\nu(A).\]
Ent\~ao, existe uma fun\c c\~ao boreliana $f$ com valores em $[0,1]$ tal que
\[\mu=f\nu.\]
\label{t:RNcasoespecial}
\end{teorema}

Gra\c cas ao exerc\'\i cio \ref{ex:bastaconsiderarnaoatomicas}, podemos supor que $\nu$ \'e uma medida n\~ao at\^omica. Aplicando o teorema \ref{t:unicidadedemedidanaoatomica}, suponhamos sem perda de generalidade que $\Omega=\{0,1\}^{\N}$ \'e o cubo de Cantor, e $\nu$ \'e a medida de Haar. 

Para cada $n$, seja $\Omega_n$ a parti\c c\~ao de $\{0,1\}^{\N}$ em $2^n$ subconjuntos cil\'\i ndricos da forma $\pi_n^{-1}(\sigma)$, $\sigma\in \{0,1\}^n$. Definamos $f_n\colon \{0,1\}^{\N}\to\R$ por
\[f_n(x) = \frac{\mu\left(\pi_n^{-1}(\pi_n(x))\right)}{\nu(\pi_n^{-1}(\pi_n(x)))} =2^n\mu\left(\pi_n^{-1}(\pi_n(x))\right).\]

\begin{exercicio}
Verifique que $(f_n)$ \'e um martingale em rela\c c\~ao \`a sequ\^encia de parti\c c\~oes $(\Omega_n)$.
\end{exercicio}

A hip\'otese adicional sobre $\mu$ implica que $\norm{f_n}_{\infty}\leq 1$, e segundo o teorema de converg\^encia de martingale \ref{t:doob}, $f_n\to f$ no espa\c co $L^1(\mu)$.

\begin{exercicio}
Mostre que para todo conjunto cil\'\i ndrico $C=\pi_n^{-1}(\sigma)$ temos
\[f(C) = f_n(C) =2^n\mu\left(\pi_n^{-1}(\pi_n(x))\right).\]
Com ajuda do teorema de extens\~ao de medidas, deduza que
\[\mu=f\nu.\]
\end{exercicio}

\subsection{Espa\c co dual de $L^1(\mu)$}

Todo elemento $f\in L^{\infty}(\mu)$ defina um funcional linear sobre $L^1(\mu)$, como segue:
\[L^1(\mu)\ni g\mapsto \hat f(g) = \int_{\Omega} f(x)g(x)\,d\mu(x)\in\R.\]

\begin{exercicio}
Verifique que a funcional acima \'e limitado, e a sua norma satisfaz
\[\norm{\hat f}=\norm{f}_{\infty}.\]
Deste modo, a correspond\^encia
\begin{equation}
L^{\infty}(\mu)\ni f\mapsto \hat f\in L^1(\mu)^{\ast}
\label{eq:correspondenciaL1*Linfty}
\end{equation}
\'e ima imers\~ao isom\'etrica linear.
\end{exercicio}

De fato, verifica-se que todos os funcionais limitados sobre $L^1(\mu)$ s\~ao daquela forma.

\begin{teorema}
O espa\c co de Banach $L^1(\mu)^{\ast}$ dual ao espa\c co $L^1(\mu)$ \'e isometricamente isomorfo ao espa\c co $L^{\infty}(\mu)$. O isomorfismo \'e estabelecido pela eq. (\ref{eq:correspondenciaL1*Linfty}).
\label{t:dualdeL1}
\end{teorema}

Apenas resta mostrar que para cada funcional linear limitado $\phi\colon L^1(\mu)\to\R$ existe uma fun\c c\~ao $f\in L^{\infty}(\mu)$ tal que $\phi=\hat f$. No caso de um espa\c co probabil\'\i stico finito, isso segue das considera\c c\~oes de dimens\~ao.

\begin{exercicio} 
Mostre o teorema \ref{t:dualdeL1} no caso onde $\Omega$ \'e finito.
\end{exercicio}

A prova do caso geral \'e muito semelhante \`a prova do teorema \ref{t:RNcasoespecial}. Sem perda de generalidade, pode-se supor que $\Omega$ \'e o cubo de Cantor, $\{0,1\}^{\N}$, munido de uma medida de probabilidade, $\mu$. Seja $\phi\in L^1(\mu)^{\ast}$. Mais uma vez, denotemos por $\Omega_n$ a parti\c c\~ao de $\{0,1\}^{\N}$ em $2^n$ subconjuntos cil\'\i ndricos $\pi_n^{-1}(\sigma)$, $\sigma\in \{0,1\}^n$. 
Para cada $n$, definamos $f_n\colon \{0,1\}^{\N}\to\R$ por
\[f_n(x)=\frac{\phi(\chi_{\pi_n^{-1}(\pi_n(x))})}{\mu(\pi_n^{-1}(\pi_n(x)))}
= \phi\left(\frac{\chi_{\pi_n^{-1}(\pi_n(x))}}
{\norm{\chi_{\pi_n^{-1}(\pi_n(x))}}_1}\right).\]

\begin{exercicio}
Verifique que $(f_n)$ \'e um martingale, e $f_n$ s\~ao uniformemente limitados em $L^{\infty}(\mu)$ por $\norm\phi$.
\end{exercicio}

O teorema de converg\^encia de martingale \ref{t:doob} garante que $f_n\to f$ no espa\c co $L^1(\mu)$.

\begin{exercicio}
Verifique que para todo conjunto cil\'\i ndrico da forma $C=\pi_n^{-1}(\sigma)$,
\[\int_Cf(x)\,d\mu(x)= \phi(C).\]
Deduza que para toda combina\c c\~ao linear finita,
\begin{equation}
g=\sum_{i=1}^k \lambda_k \chi_{C_k},
\label{eq:comblinfinchiC}
\end{equation}
onde $C_k$ s\~ao conjuntos cil\'\i ndricos como acima, temos
\[\int_{\{0,1\}^{\N}} g(x)f(x)\,d\mu(x)=\phi(g).\]
\end{exercicio}

\begin{exercicio}
Mostre que as fun\c c\~oes como na eq. (\ref{eq:comblinfinchiC}) ({\em fun\c c\~oes simples}) s\~ao densas em $L^1(\mu)$.
\par
[ {\em Sugest\~ao:} use a regularidade de $\mu$ para aproximar cada conjunto da forma $\{x\in\Omega\colon t_i<f(x)<t_{i+1}\}$ com um conjunto aberto e fechado, e depois use o fato que abertos e fechados em $\{0,1\}^{\N}$ s\~ao exatamente conjuntos cil\'\i ndricos. ]
\par
Em particular, o espa\c co $L^1(\mu)$, onde $\mu$ \'e uma medida de probabilidade sobre um espa\c co boreliano padr\~ao, \'e separ\'avel.
\end{exercicio}

\begin{exercicio}
Complete a prova do teorema \ref{t:dualdeL1}.
\end{exercicio}

\section{Teorema da desintegra\c c\~ao de Rokhlin\label{s:rokhlin}}

\subsection{\label{ss:rokhlin}}
N\'os precisamos a vers\~ao seguinte do teorema de Rokhlin.

\begin{teorema}[Teorema da desintegra\c c\~ao de Rokhlin]
Sejam $f\colon\Omega\to\Upsilon$ uma fun\c c\~ao boreliana entre dois espa\c cos borelianos padr\~ao, e $\mu$ uma medida de probabilidade boreliana sobre $\Omega$. Para quase todo $x\in\Upsilon$, existe uma medida de probabilidade $\mu_x$ sobre a fibra $f^{-1}(x)\subseteq\Omega$, tal que, qualquer que seja fun\c c\~ao $g\in L^1(\mu)$,
\begin{equation}
\int_{\Omega}g(x)\,d\mu(x) = \int_{\Upsilon} df_{\ast}(\mu)\int_{f^{-1}(x)}g(y)\,d\mu_x(y).
\label{eq:rokhlin}
\end{equation}
Em particular, para cada conjunto mensur\'avel $A\subseteq\Omega$,
\[\mu(A) = \int_{\Upsilon} \mu_x(A\cap f^{-1}(x))\,df_{\ast}(\mu).\]
\index{teorema! de desintegra\c c\~ao de medidas}
\end{teorema}

\begin{observacao}
Na fala probabil\'\i stica, as medidas $\mu_x$ s\~ao {\em probabilidades condicionais dado $f(X)=x$}, onde $X$ \'e um elemento aleat\'orio de $\Omega$ seguindo a lei $\mu$. Nota\c c\~ao:
\[P[ \cdot \mid f(X)=x].\]
Analogamente, a integral $\int d\mu_x$ \'e a {\em esperan\c ca condicional dado $f(X)=x$.}
A igualdade (\ref{eq:rokhlin}) serve da base do {\em jeito de condicionamento}.
\index{jeito! de condicionamento}
\label{o:truquedecond}
\end{observacao}

\begin{exercicio}
Mostre o teorema de Rokhlin no caso onde a medida $f_{\ast}(\mu)$ \'e puramente at\^omica em $\Upsilon$. \par
[ {\em Sugest\~ao:} se $x$ \'e um \'atomo, a medida $\mu_x$ \'e simplesmente a normaliza\c c\~ao da restri\c c\~ao de $\mu$ sobre $f^{-}(x)$,
\[\mu_x = \frac{\mu\vert_{f^{-1}(x)}}{\mu(f^{-1}(x))}.~~ ]\]
Deduza que basta mostrar o teorema de Rokhlin apenas para medidas $\mu$ cuja imagem direta $f_{\ast}(\mu)$ \'e n\~ao-at\^omica.
\label{ex:bastanaoatomizade}
\end{exercicio}

\subsection{} Para come\c car, mostremos o teorema de Rokhlin no caso especial onde $\Omega$ e $\Upsilon$ s\~ao espa\c cos compactos m\'etricos, $f\colon\Omega\to\Upsilon$ \'e uma fun\c c\~ao cont\'\i nua, e $\Upsilon=\{0,1\}^{\N}$ \'e homeomorfo ao cubo de Cantor. Com o exerc\'\i cio \ref{ex:bastanaoatomizade} em mente, suponhamos que $f_{\ast}(\mu)$ \'e uma medida n\~ao-at\^omica.

Diz-se que um espa\c co m\'etrico tem {\em dimens\~ao zero} se para cada ponto $x$ e cada vizinhan\c ca $V$ de $x$, existe um conjunto aberto e fechado, $U$, tal que $x\in U\subseteq V$.

\begin{exercicio}
Mostre que o cubo de Cantor $\{0,1\}^{\N}$ \'e um espa\c co de dimens\~ao zero.
\end{exercicio}

\begin{exercicio}
Seja $K$ um espa\c co m\'etrico compacto, de dimens\~ao zero, e sem pontos isolados. Mostre que $K$ \'e homeomorfo ao cubo de Cantor $\{0,1\}^{\N}$. 
\par
[ {\em Sugest\~ao:} construa uma sequ\^encia de parti\c c\~oes de $K$ com abertos e fechados, once a cada passo todo elemento da parti\c c\~ao \'e dividido em dois.... ]
\end{exercicio}

\begin{exercicio}
Mostre que o suporte, $\mbox{supp}\,\mu$, de uma medida boreliana n\~ao-at\^omica $\mu$ sobre um espa\c co m\'etrico qualquer n\~ao tem pontos isolados. (Como diz-se, $\mbox{supp}\,\mu$ \'e um conjunto {\em perfeito}).
\end{exercicio}

Os exerc\'\i cios acima permitem assumir, 
sem perda de generalidade, que a medida $f(\mu)$ tem $\Upsilon=\{0,1\}^{\N}$ como o suporte.
Como nas se\c c\~oes precedentes, formemos uma sequ\^encia refinadora de parti\c c\~oes finitas do cubo de Cantor, $(\Omega_N)$, onde $\Omega_n$ consiste de conjuntos abertos e fechadas cil\'\i ndricos da forma $\pi_n^{-1}(\sigma)$, $\sigma\in \{0,1\}^n$. Dado um elemento $C\in \Omega_n$, definamos uma medida de probabilidade $\mu_C$ sobre $\Omega$ como a restri\c c\~ao normalizada de $\mu$ sobre o conjunto aberto e fechado $f^{-1}(C)$:
\[\mu_C=\frac{\mu\vert_{f^{-1}(C)}}{\mu(f^{-1}(C))}.\]
Como $\mu(f^{-1}(C))=f_{\ast}(\mu)(C)\neq 0$, a defini\c c\~ao \'e correta.

Definamos, para cada $n$, uma aplica\c c\~ao $\mu_n\colon \Upsilon\to P(\Omega)$ como segue:
\[\mu_n(x) = \mu_C,\mbox{ para o \'unico }C\in\Omega_n\mbox{ que cont\'em }x.\]
A sequ\^encia $(\mu_n)$ forma um martingale com valores no espa\c co $P(\Omega)$ (relativo a sua estrutura afim), que converge para uma aplica\c c\~ao $\nu\colon\Upsilon\to P(\Omega)$, cujos valores, $\nu(x)$, ser\~ao exatamente as nossas medidas $\mu_x$. Ao inv\'es de desenvolvermos uma teoria de martingales com valores nos espa\c cos localmente convexos, reduzamos tudo ao caso de martingais reais.

Seja $g\in C(\Omega)$ uma fun\c c\~ao cont\'\i nua qualquer sobre o espa\c co compacto $\Omega$. Para cada $n$, definamos a fun\c c\~ao 
\[g_n\colon \Upsilon\to\R,\]
igual a esperan\c ca condicional de $g$ em rela\c c\~ao \`a parti\c c\~ao $f^{-1}(\Omega_n)$, pela condi\c c\~ao
\[\tilde g_n(x) = \int_Cg(x)\,d\mu_C,\mbox{ onde }
x\in C\in \Omega_n.\]

\begin{exercicio}
Verificar que a sequ\^encia $[(\tilde g_n)]$ \'e um martingale sobre $\Upsilon$ em rela\c c\~ao \`a sequ\^encia de parti\c c\~oes $(\Omega_n)$.
\end{exercicio}

Como $(g_n)$ s\~ao uniformemente limitados por $\norm{g}_{\infty}$, o martingale tem limite em $L^1(f_{\ast}(\mu))$, que denotemos por $\tilde g$. Tamb\'em temos a converg\^encia em $f_{\ast}(\mu)$-quase todo $x$, $\tilde g_n(x)\to \tilde g(x)$.

Fixemos uma sequ\^encia $(g_k)_{k=1}^{\infty}$ densa na bola unit\'aria de $C(\Omega)$. Segue-se que que, para $f_{\ast}(\mu)$-quase todo $x\in\Upsilon$ e todos $k$, $\tilde{(g_k)}_n(x)\to \tilde{g_k}(x)$, e $\left\Vert\tilde{(g_k)}_n-\tilde{g_k}\right\Vert_1\to 0$. Denotemos por $\mathfrak X$ o conjunto de $x\in\Upsilon$ onde a converg\^encia pontual tem lugar para todos $k$.

\begin{exercicio}
Mostre que, para $x\in {\mathfrak X}$, a aplica\c c\~ao
\[g_k\mapsto \tilde{g_k}(x)\]
estende-se at\'e um funcional linear positivo sobre $C(\Omega)$ enviando $1$ para $1$. Logo, quase certamente existe a \'unica medida de probabilidade $\mu_x$ sobre $\Omega$ tal que
\[\tilde{g_k}(x) = \int_{\Omega}g_k\,d\mu_x.\]
\end{exercicio}

\begin{exercicio}
Verifique que, para $x\in {\mathfrak X}$, se $g,h\in C(\Omega)$, $V\ni x$ e uma vizinhan\c ca de $x$ em $\Upsilon$ s\~ao tais que $g\vert_{f^{-1}(V)}= h\vert_{f^{-1}(V)}$, ent\~ao 
\[\tilde g(x) =\tilde h(x).\]
Deduza que, para quase todo $x$,
\[\mbox{supp}\,\mu_x\subseteq f^{-1}(x).\]
[ {\em Sugest\~ao:} $f$ \'e cont\'\i nua. ]
\end{exercicio}

\begin{exercicio}
Seja $x\in\Upsilon$ e $(C_n)$ uma sequ\^encia de abertos e fechados cil\'\i ndricos da forma $C_n=\pi_n^{-1}(\pi_n(x))$. Mostre que, para quase todo $x$,
\[\mu_{C_n}\to \mu_x\mbox{ quando }n\to\infty\]
na topologia fraca$^\ast$ em $P(\Omega)$.
\end{exercicio}

\begin{exercicio}
Combine o exerc\'\i cio precedente com a observa\c c\~ao que para cada $g\in C(\Omega)$ e todo $k$,
\[\int_{\Upsilon} \tilde{g_k}\,df_{\ast}(\mu)=\int_{\Omega} g\,d\mu\]
para deduzir a afirma\c c\~ao do teorema de Rokhlin (eq. \ref{eq:rokhlin}) sob as nossas hip\'oteses adicionais. (Primeiramente, para fun\c c\~oes cont\'\i nuas, e no conseguinte, para fun\c c\~oes $L^1(\mu)$).
\end{exercicio}

\subsection{} 
Agora, o caso geral. Comecemos com uma varia\c c\~ao do teorema de Luzin \ref{t:luzin}. 

\begin{exercicio}
Mostre o seguinte.
Seja $f$ uma fun\c c\~ao mensur\'avel sobre um espa\c co m\'etrico completo e separ\'avel, $\Omega$, munido de uma medida de probabilidade boreliana, $\mu$, com valores no cubo de Cantor $\{0,1\}^{\N}$. Dado $\ve>0$, existe um subconjunto compacto $K\subseteq\Omega$ com $\mu(\Omega\setminus K)<\ve$ tal que a restri\c c\~ao $f\vert_{K}$ \'e cont\'\i nua.
\label{l:luzin-2}
\end{exercicio}

Agora identifiquemos $\Upsilon$, como espa\c co boreliano, com $\{0,1\}^{\N}$, e $\Omega$, com um espa\c co compacto m\'etrico qualquer. Para cada $\ve>0$, escolhemos um compacto $K\subseteq \Omega$ com $\mu(K)>1-\ve$ e $f\vert_K$ cont\'\i nua. Definamos a medida de probabilidade normalizada
\[\mu^{\ve} = \frac{\mu\vert_{K_{\ve}}}{\mu(K_{\ve})}.\]
Escolhemos uma fam\'\i lia de medidas de probabilidade condicionais, $(\mu^{\ve}_x)$, $x\in\Upsilon$, desintegrando a medida $\mu^{\ve}$. Deixemos como um exerc\'\i cio, de verificar que, se $\ve_n\downarrow 0$ bastante r\'apido, ent\~ao o lema de Borel--Cantelli \ref{ex:1lemaborel-cantelli} garante que, para quase todo $x\in\Upsilon$, a sequ\^encia de medidas $(\mu^{\ve_n}_x)_{n=1}^{\infty}$ converge na topologia fraca$^{\ast}$ para uma medida de probabilidade $\mu_x$, e a fam\'\i lia $(\mu_x)$ satisfaz todas as conclus\~oes do teorema de Rokhlin.

\subsection{Teorema de Fubini}
Um exemplo cl\'assico da desintegra\c c\~ao de medidas \'e o teorema de Fubini.

\begin{teorema}
Sejam $(\Upsilon,\mu)$ e $(\Omega,\nu)$ dois espa\c cos probabil\'\i sticos padr\~ao. Ent\~ao, a desintegra\c c\~ao da medida de produto $\mu\otimes\nu$ sobre $\Upsilon\times\Omega$ tem forma $\mu_x=\nu$ para quase todos $x\in\Upsilon$. Em outras palavras, para toda fun\c c\~ao $f\in L^1(\mu\otimes\nu)$,
\[\int_{\Upsilon\times\Omega} f\,d(\mu\otimes\nu) = \int_{\Upsilon} d\mu(x)\int_{\Omega} f(x,y)\,d\nu(y).\]
\label{t:fubini}
\index{teorema! de Fubini}
\end{teorema} 

\begin{exercicio} 
Mostre o teorema, primeiramente para $f$ cont\'\i nua (na topologia de produto, onde $\Omega$ e $\Upsilon$ s\~ao munidas de estruturas de espa\c cos m\'etricos apropriados), usando as fun\c c\~oes $\tilde f_n$ na prova do teorema de Rokhlin, e depois para $f\in L^1(\mu)$.
\end{exercicio}

%
%

\chapter{Desigualdade de concentra\c c\~ao de Paul L\'evy \label{a:esfera}}

\section{Fen\^omeno de concentra\c c\~ao}

Consideremos um dom\'\i nio, $\Omega$, potencialmente de ``alta dimens\~ao'', como a esfera euclideana:
\[
  \s^d=\{x\in\R^{d+1}\mid \abs x=1\}.
\]

Suponhamos que o \'unico meio de estudar o objeto em quest\~ao seja por uma s\'erie dos experimentos aleat\'orios, do seguinte modo. Cada experimento produz um ponto $x\in \Omega$ tirado de maneira aleat\'oria (cuja distribui\c c\~ao \'e conforme a medida natural de $\Omega$, como o volume), e cada vez podemos registrar os valores $f(x)$ de fun\c c\~oes (quantidades observ\'aveis)
\[
f\colon X\to\R
\]
para $x$. Quanta informa\c c\~ao sobre a geometria de $\Omega$ podemos obter desta maneira? 

Por exemplo, o que podemos deduzir sobre o {\em di\^ametro} de $\Omega$? O di\^ametro de $\Omega$ \'e a quantidade

\[
\diam \Omega:=\sup\{d(x,y)\mid x,y\in \Omega\},
\]
onde $d(x,y)$ denota a dist\^ancia entre $x$ e $y$. Nesta situa\c c\~ao, como as observ\'aveis $f\colon \Omega\to\R$, \'e l\'ogico considerar as fun\c c\~oes {\em Lipschitz cont\'\i nuas} com a constante de Lipschitz $1$, isso \'e, as fun\c c\~oes que n\~ao aumentam a dist\^ancia:

\[
\forall x,y\in \Omega~~\abs{f(x)-f(y)}\leq d(x,y).
\]
Eis uma fonte das tais fun\c c\~oes.

\begin{exercicio}
\label{distanc} 
Seja $x_0\in \Omega$ um ponto de $X$ qualquer.
Mostrar que a fun\c c\~ao dist\^ancia definida por
\[x\mapsto\mathrm{dist}(x_0,x)\]
\'e Lipschitz cont\'\i nua com a constante $1$.
\end{exercicio}

Por conseguinte, obtemos o resultado seguinte.

\begin{exercicio}
  Mostre que
  \begin{eqnarray*}
    \diam \Omega &=& \sup\left\{\abs{f(x)-f(y)} \colon x,y\in \Omega,\right. \\
    && \left.f\colon \Omega\to\R
    \mbox{ \'e $1$-Lipschitz cont\'\i nua com }L=1 \right\}.\end{eqnarray*}
\label{brute}
\end{exercicio}

\begin{exercicio}
Verifique que, se $\Omega$ \'e compacto, ent\~ao o valor do di\^ametro \'e realizado pela uma fun\c c\~ao $1$-Lipschitz cont\'\i nua $f$ apropriada:
\[\diam\Omega = \max\{d(f(x),f(y))\colon x,y\in\Omega\}.\]
\end{exercicio}

Se n\'os pud\'essemos medir os valores de {\em todos as observ\'aveis} para {\em todos os pares de pontos de $X$} e ent\~ao escolher o supremo, saber\'\i amos o di\^ametro $\diam X$. Mas isto \'e imposs\'\i vel. Podemos escolher {\em uma} observ\'avel $f$, e gerar ent\~ao a sequ\^encia mais ou menos longa, mas {\em finita}, de pontos aleat\'orios,
\[x_1,x_2,\ldots,x_N,\]
registrando cada vez o valor $f(x_i)$, $i=1,2,3,\ldots$.

Depois que produzimos uma s\'erie de n\'umeros reais
\[f(x_1),f(x_2),\ldots,f(x_N),\]
calcularemos a diferen\c ca m\'axima
\[D_N=\max_{i,j=1}^N\abs{f(x_i)-f(x_j)}.\]
\'E imediato que
\[D_N\leq D_{N+1}\leq D,\]
de modo que os valores $D_N$ ``melhoram'' cada vez.

Pararemos o experimento quando a probabilidade de melhorar o valor precedente, $D_N$, se torna demasiado pequena. Mais precisamente, seja $\kappa>0$ (um valor limiar) um n\'umero fixo muito pequeno, tal como $\kappa=10^{-10}$ (sugerido por Gromov).

N\'os pararemos depois que o n\'umero $D=D_N$ satisfaz
\[\mu\{x\mid \abs{f(x)-M}<D\}>1-\kappa\},\]
onde $\mu$ \'e a medida ``natural'' sobre $X$. O valor $D=D_N$ obtido desta maneira chama-se {\em o di\^ametro observ\'avel} de $X$. Mais precisamente, o di\^ametro observ\'avel $\obs_\kappa X$ \'e definido por

\begin{eqnarray*}
\obs_\kappa X&=&\inf\{D>0 \colon \mbox{ para cada observ\'avel $f$ sobre $X$, 
}  \\
& & \mu\{x\in X 
\mid \abs{f(x)-M}\geq D \}\leq \kappa\}.
\end{eqnarray*}
\index{di\^ametro! observ\'avel}

Ilustremos o conceito para as esferas euclideanas $\s^n$. 
Neste experimento, substituamos a reta $\R$ com uma ``tela'' $\R^2$, e usemos como a observ\'avel a proje\c c\~ao coordenada $\R^{d+1}\to\R^2$,
\[(x_1,x_2,\dots,x_{d+1})\to (x_1,x_2).\]
O n\'umero dos pontos tirados ser\'a fixo a $N=1000$. 

\begin{figure}
\begin{center}
\scalebox{0.25}[0.307]{\includegraphics{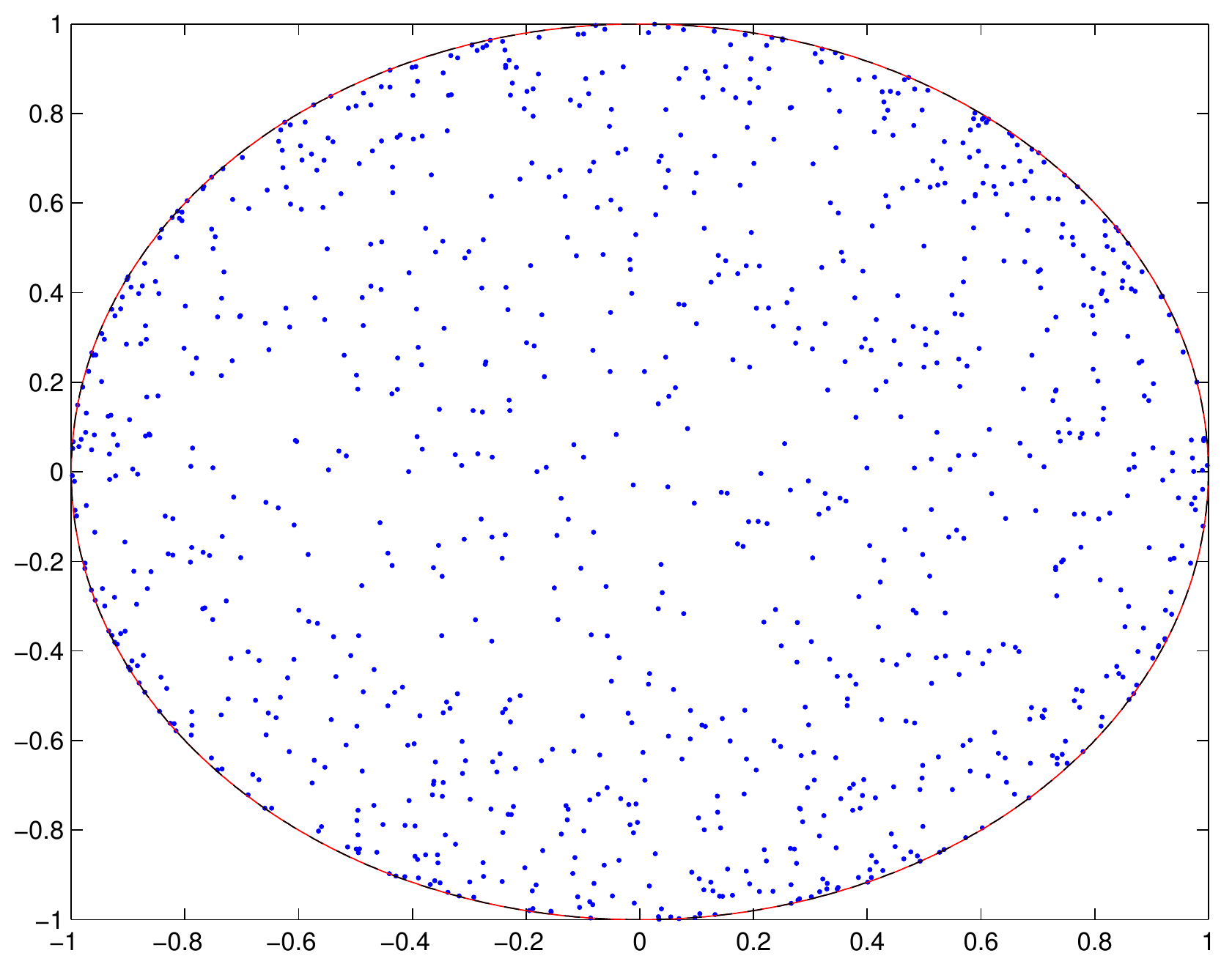}}
\hskip .2cm
\scalebox{0.25}[0.307]{\includegraphics{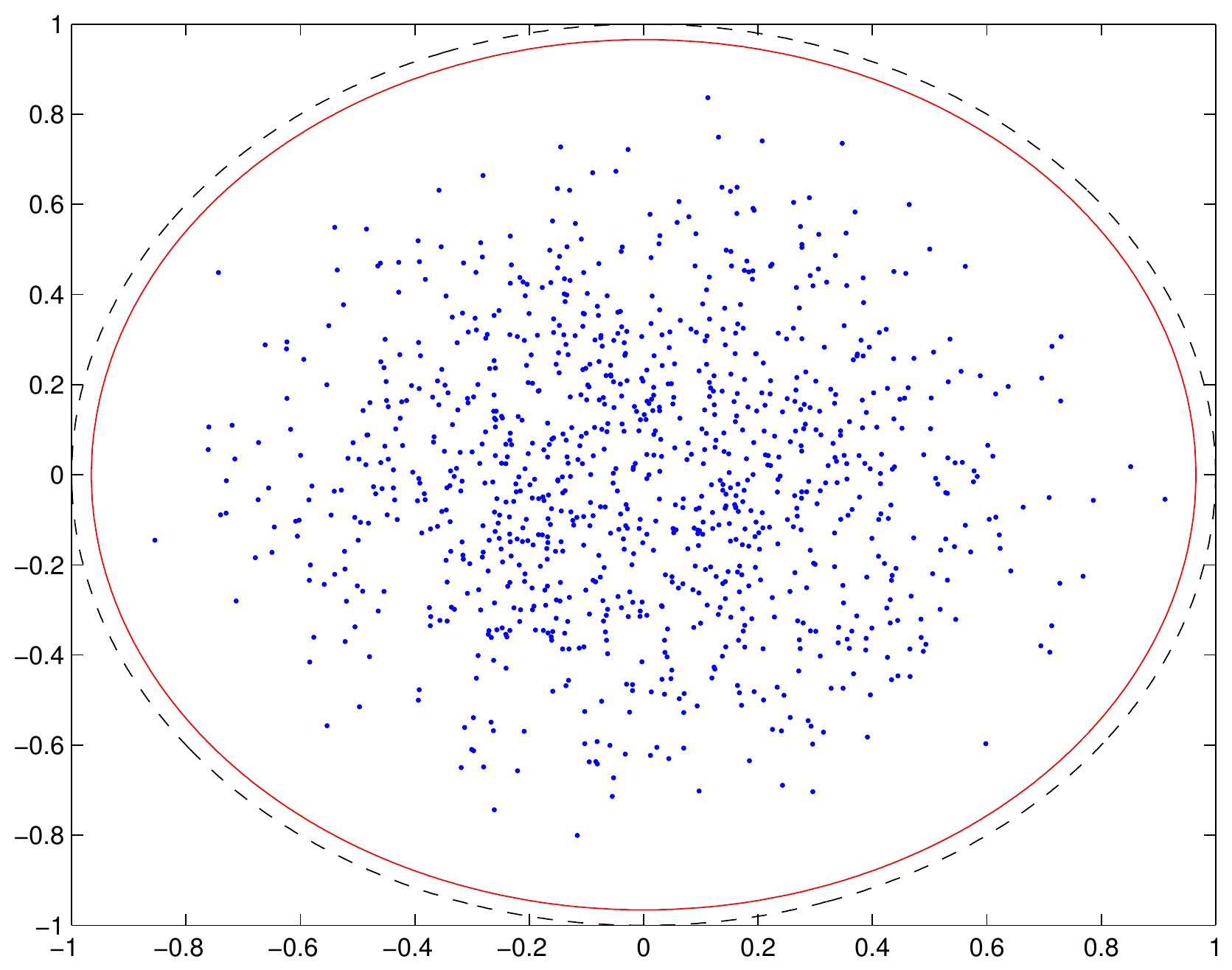}} 
\caption
{$\s^2$ e $\s^{10}$}
\label{210}
\end{center}
\end{figure}

\begin{figure}
\begin{center}
\scalebox{0.25}[0.307]{\includegraphics{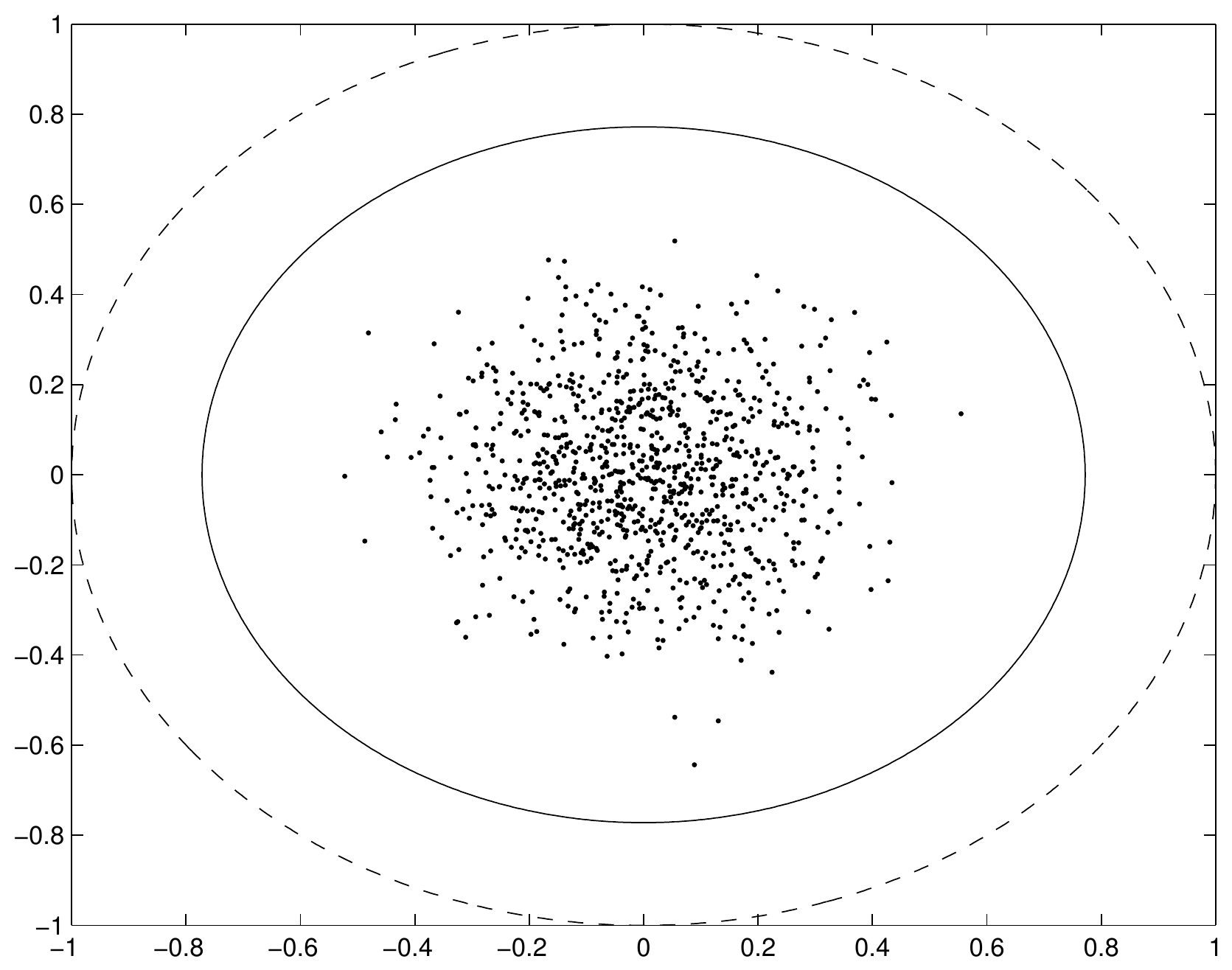}}
\hskip .2cm
\scalebox{0.25}[0.307]{\includegraphics{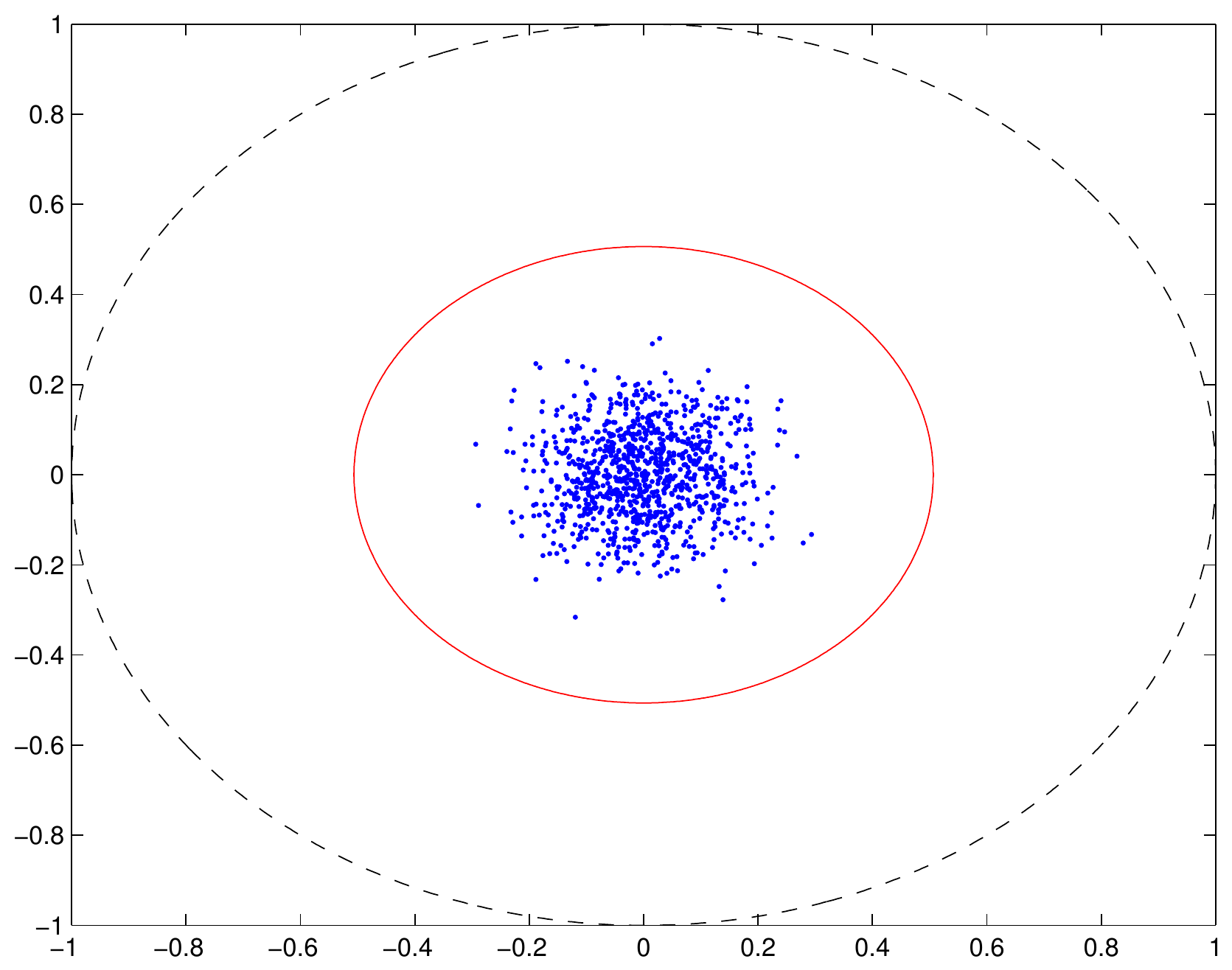}}
\caption{ $\s^{30}$ e $\s^{100}$}
\label{30100}
\end{center}
\end{figure}

\begin{figure}
\begin{center}
\scalebox{0.25}[0.307]{\includegraphics{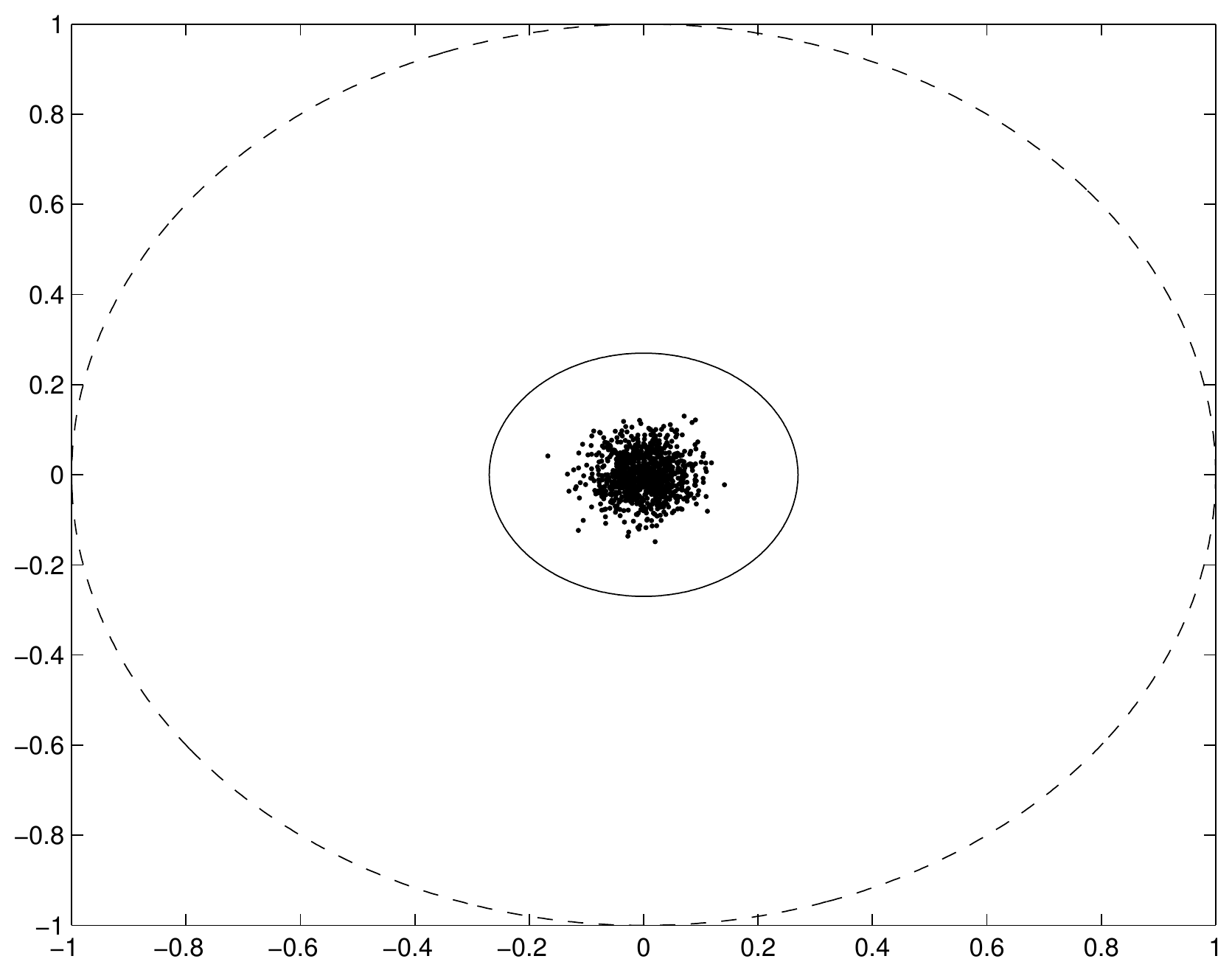}}
\hskip .2cm
\scalebox{0.25}[0.307]{\includegraphics{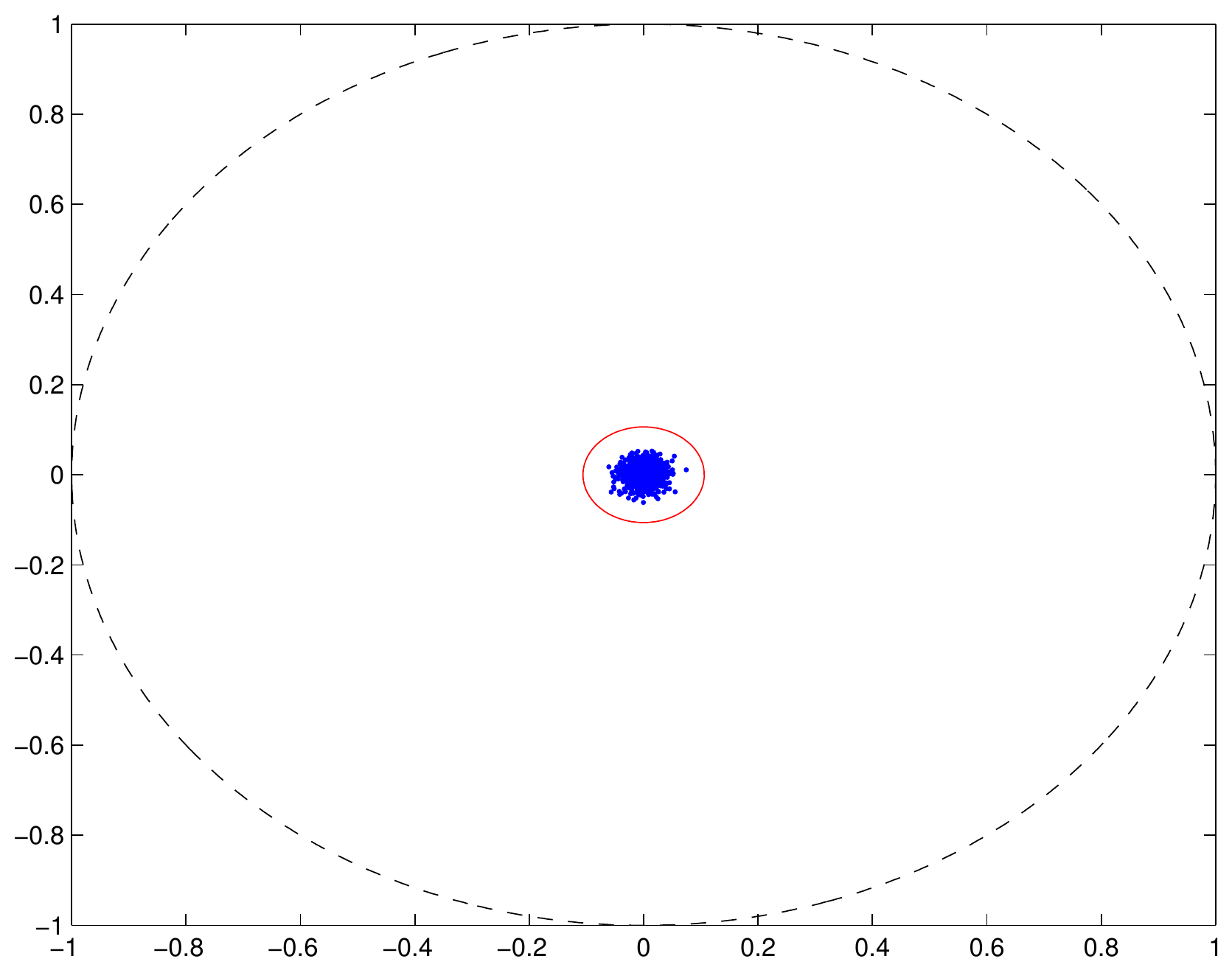}}
\caption{$\s^{500}$ e $\s^{2500}$}
\label{500}
\end{center}
\end{figure}

As Figuras \ref{210}, \ref{30100} e \ref{500} mostram os resultados de experimentos. 
A linha pontilhada representa a proje\c c\~ao da esfera de raio um (a saber, o c\'\i rculo de raio um), enquanto a linha s\'olida mostra um c\'\i rculo de tal raio que a probabilidade de um evento de que a proje\c c\~ao de um ponto aleat\'orio na esfera esteja fora deste c\'\i rculo \'e menos que $\kappa = 10^{-10}$. Em outras palavras, o di\^ametro do c\'\i rculo solido \'e o di\^ametro observ\'avel da esfera $\s^d$.

\'E poss\'\i vel provar que o di\^ametro observ\'avel da esfera satisfaz
\[\obs_\kappa(\s^d) = \Theta\left(\frac 1{\sqrt d}\right)\]
para cada valor limiar $\kappa>0$. Em outras palavras, assintoticamente, o di\^ametro observ\'avel da esfera $\s^n$ \'e de ordem $1/\sqrt d$. 

\begin{figure}[ht]
\scalebox{0.3}[0.3]{\includegraphics{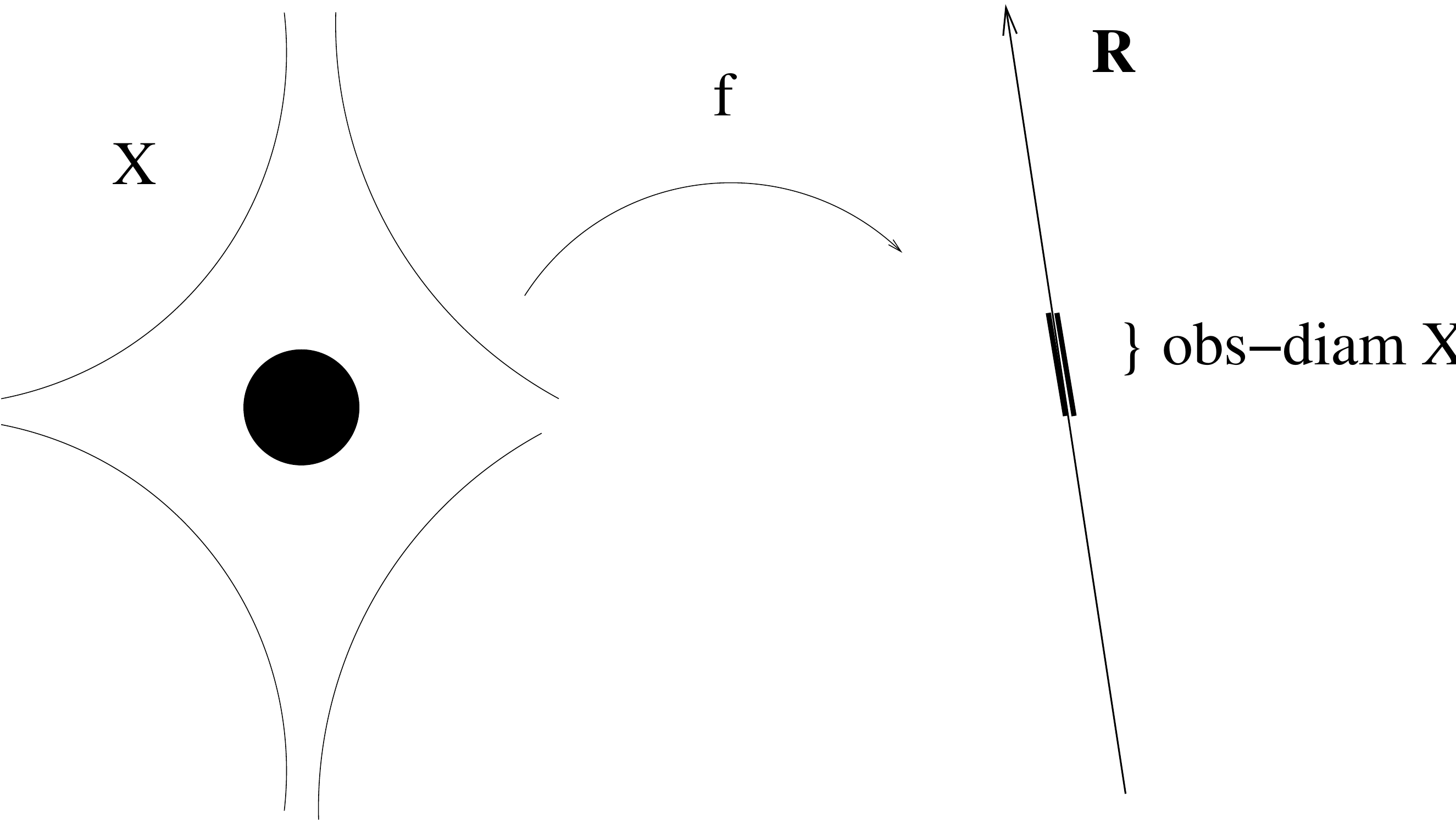}} \\
\caption{O di\^ametro observ\'avel de um espa\c co da alta dimens\~ao.}
\label{diam-obs}
\end{figure}

Come o di\^ametro atual da esfera $\s^d$ \'e $2$, uma esfera da alta dimens\~ao  aparece como um ``cometa'' formado de um ``n\'ucleo'' muito pequeno e um ``envolt\'orio gasoso'' de grande tamanho e de densidade baixa. (Figura \ref{diam-obs}).

Esta observa\c c\~ao \'e t\'\i pica de outros objetos geom\'etricos da alta dimens\~ao.
Por exemplo, \'e poss\'\i vel mostrar que o di\^ametro observ\'avel do cubo unit\'ario,
\[
  \I^d=\{x\in\R^{d}\mid \forall i=1,\ldots,d,~0\leq\abs{x_i}\leq 1\},
\]
satisfaz
\[\obs_\kappa(\I^d) = \Theta\left(1\right).\]
Ou seja, assintoticamente $\obs_\kappa(\I^d)$ \'e constante. Ao mesmo tempo,
\[\diam(\I^d)=\sqrt d.\]

\begin{figure}
\begin{center}
\scalebox{0.25}[0.3]{\includegraphics{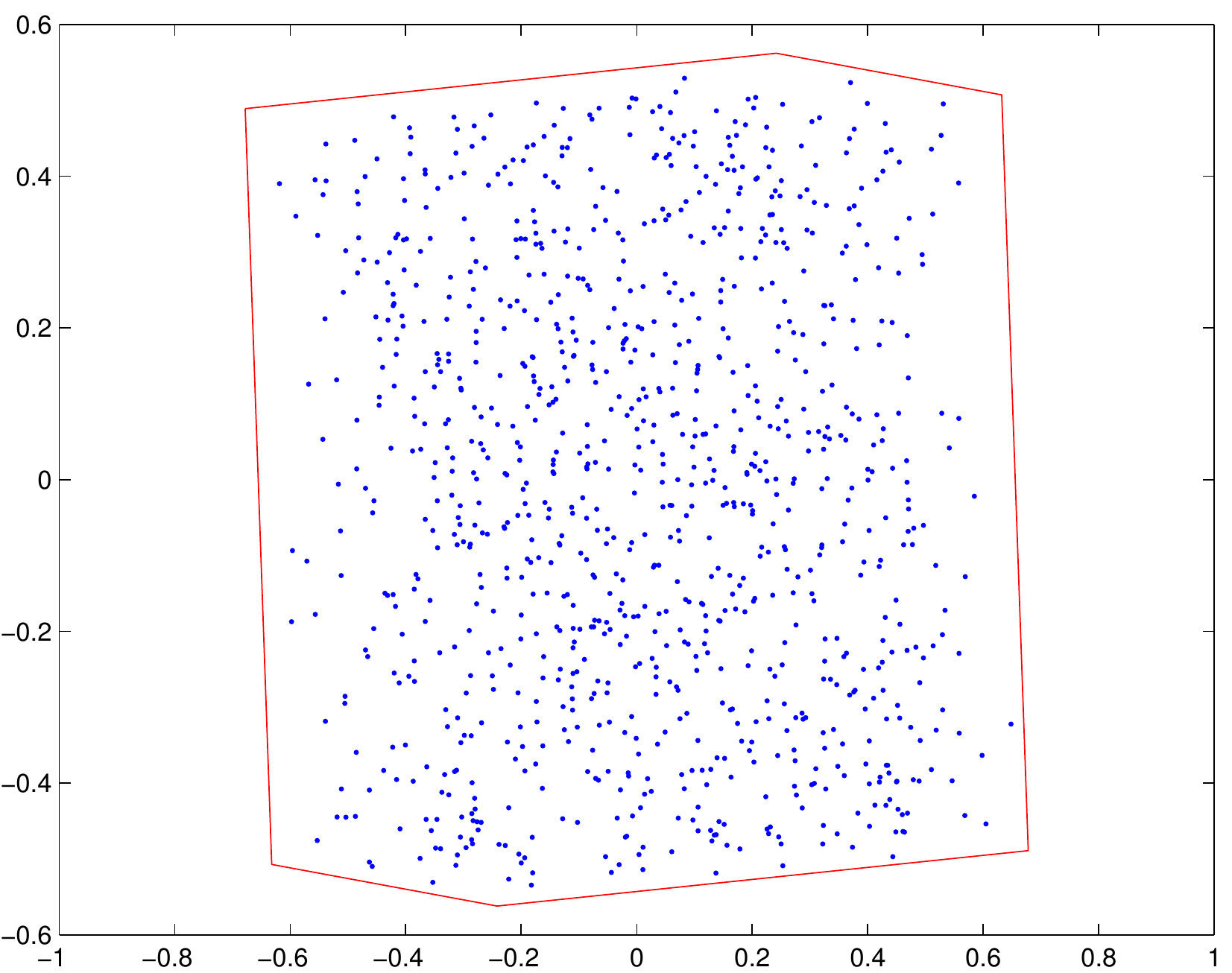}}
\scalebox{0.25}[0.3]{\includegraphics{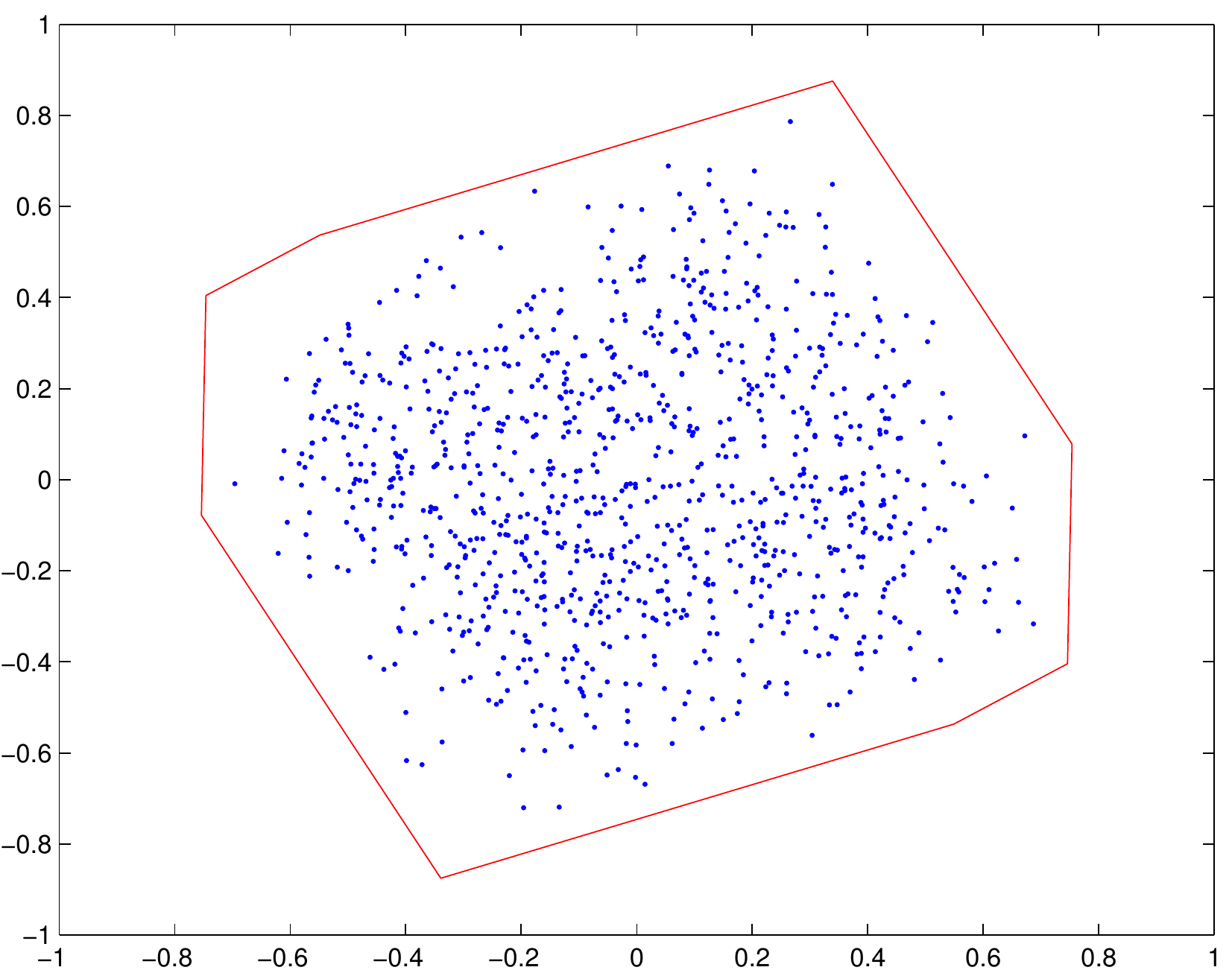}}
\caption{Proje\c c\~oes do cubo $\I^d$ e dos $1,000$ pontos aleat\'orios no cubo sobre um plano aleat\'orio, $d=3,4$.}
\label{cubo34}
\end{center}
\end{figure}

\begin{figure}
\begin{center}
\scalebox{0.25}[0.3]{\includegraphics{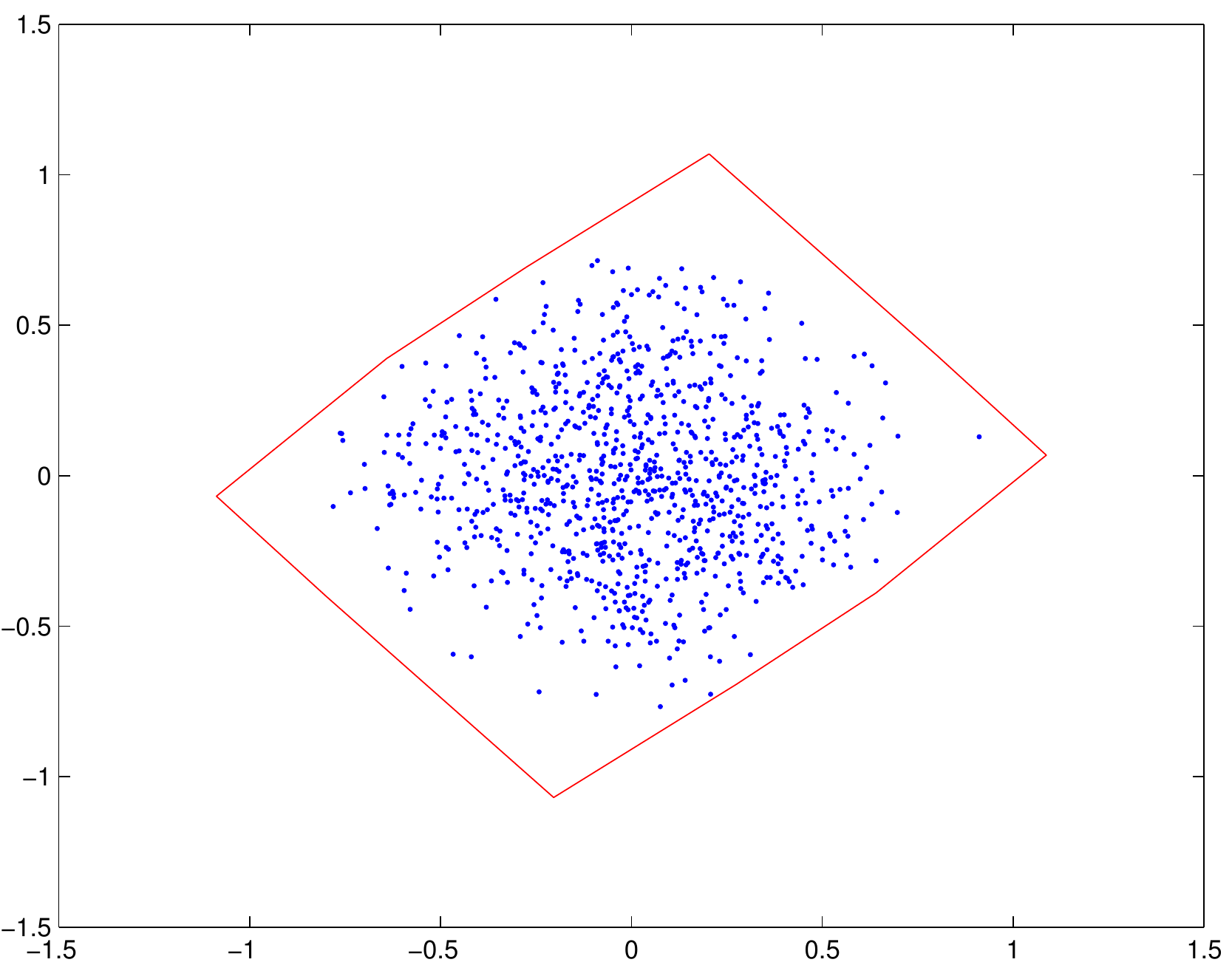}}
\scalebox{0.25}[0.3]{\includegraphics{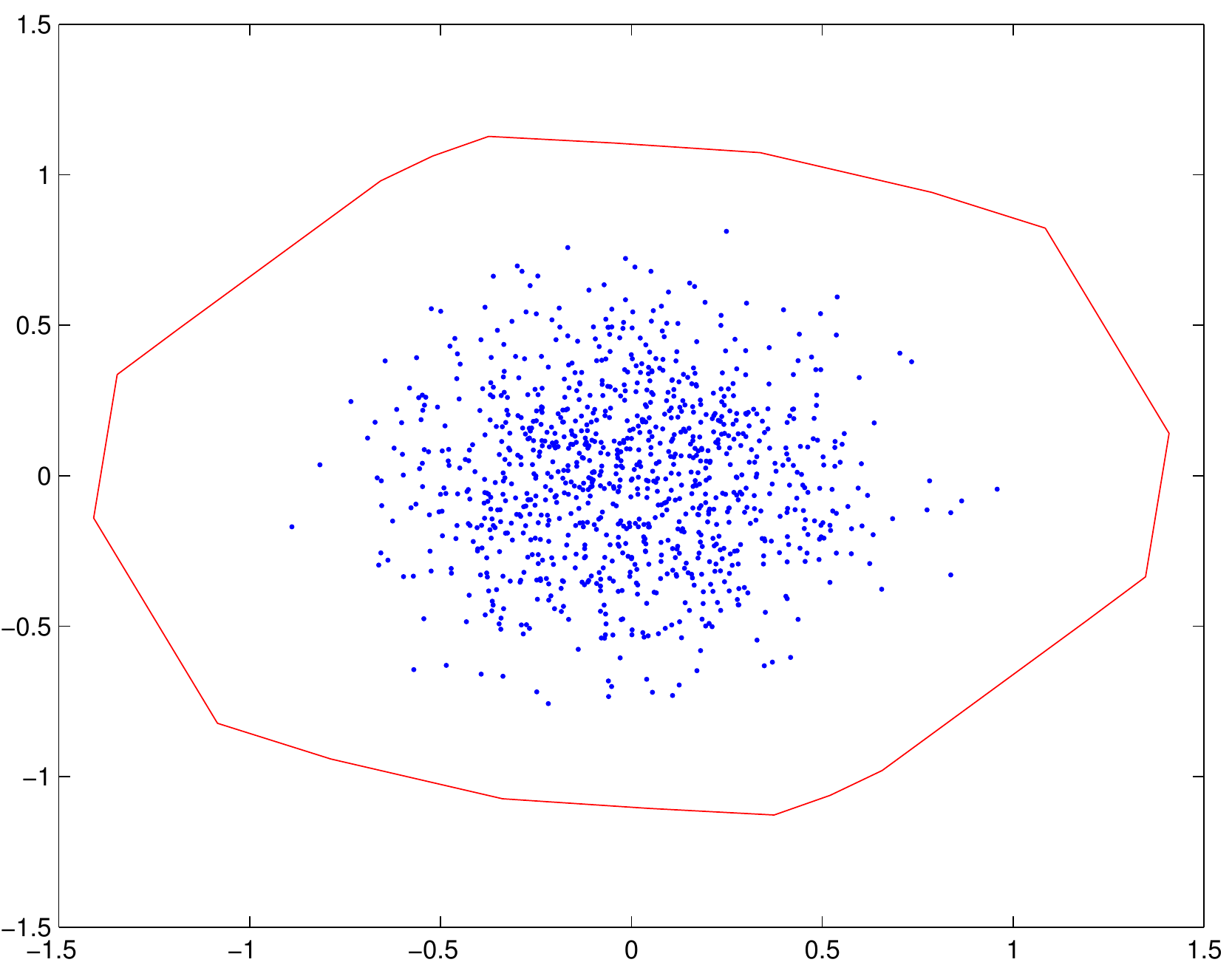}}
\end{center}
\caption{O mesmo, $d=5,10$.}
\label{cubo510}
\end{figure}

\begin{figure}
\begin{center}
\scalebox{0.25}[0.3]{\includegraphics{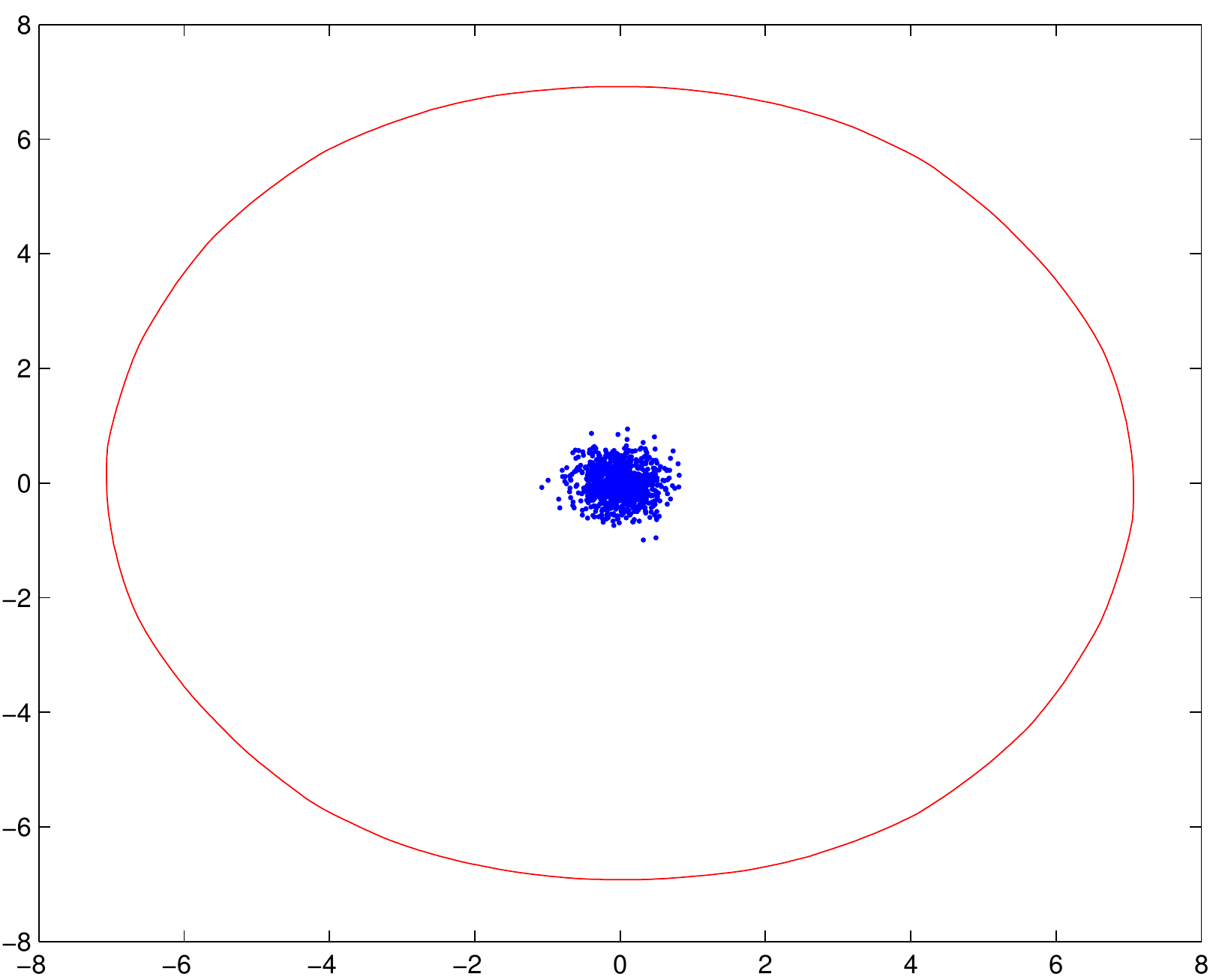}}
\scalebox{0.25}[0.3]{\includegraphics{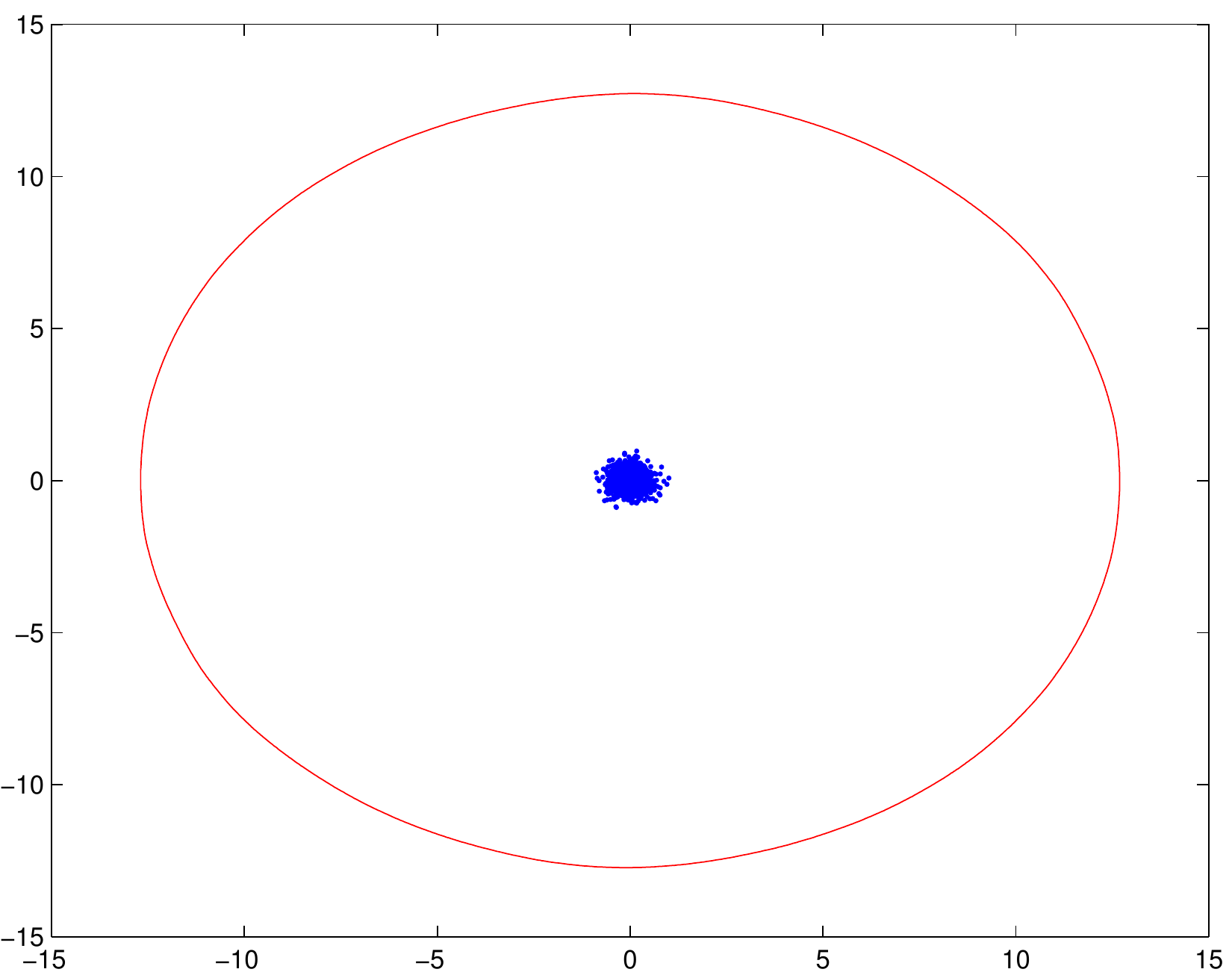}}
\caption{O mesmo, $d=300,1000$.}
\label{cubo500}
\end{center}
\end{figure}

Com efeito, nas altas dimens\~oes a proje\c c\~ao ortogonal do cubo $\I^d$ com $N=1000$ pontos aleat\'orios sobre um plano aleat\'orio assemelha-se fortemente \`a proje\c c\~ao da esfera. Veja as Figuras \ref{cubo34}, \ref{cubo510}, e \ref{cubo500}.

\begin{figure}
\begin{center}
\scalebox{0.25}[0.3]{\includegraphics{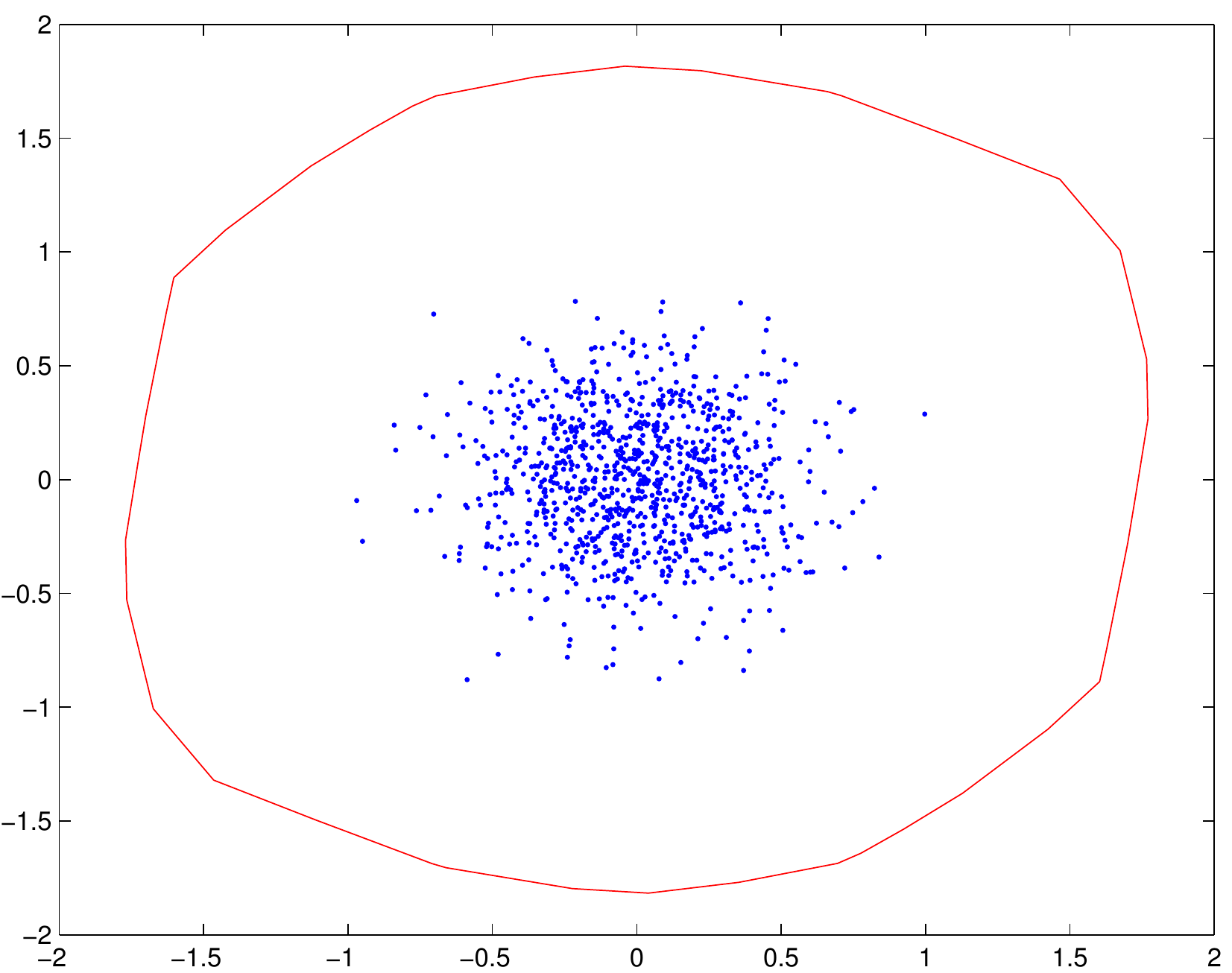}}
\scalebox{0.25}[0.3]{\includegraphics{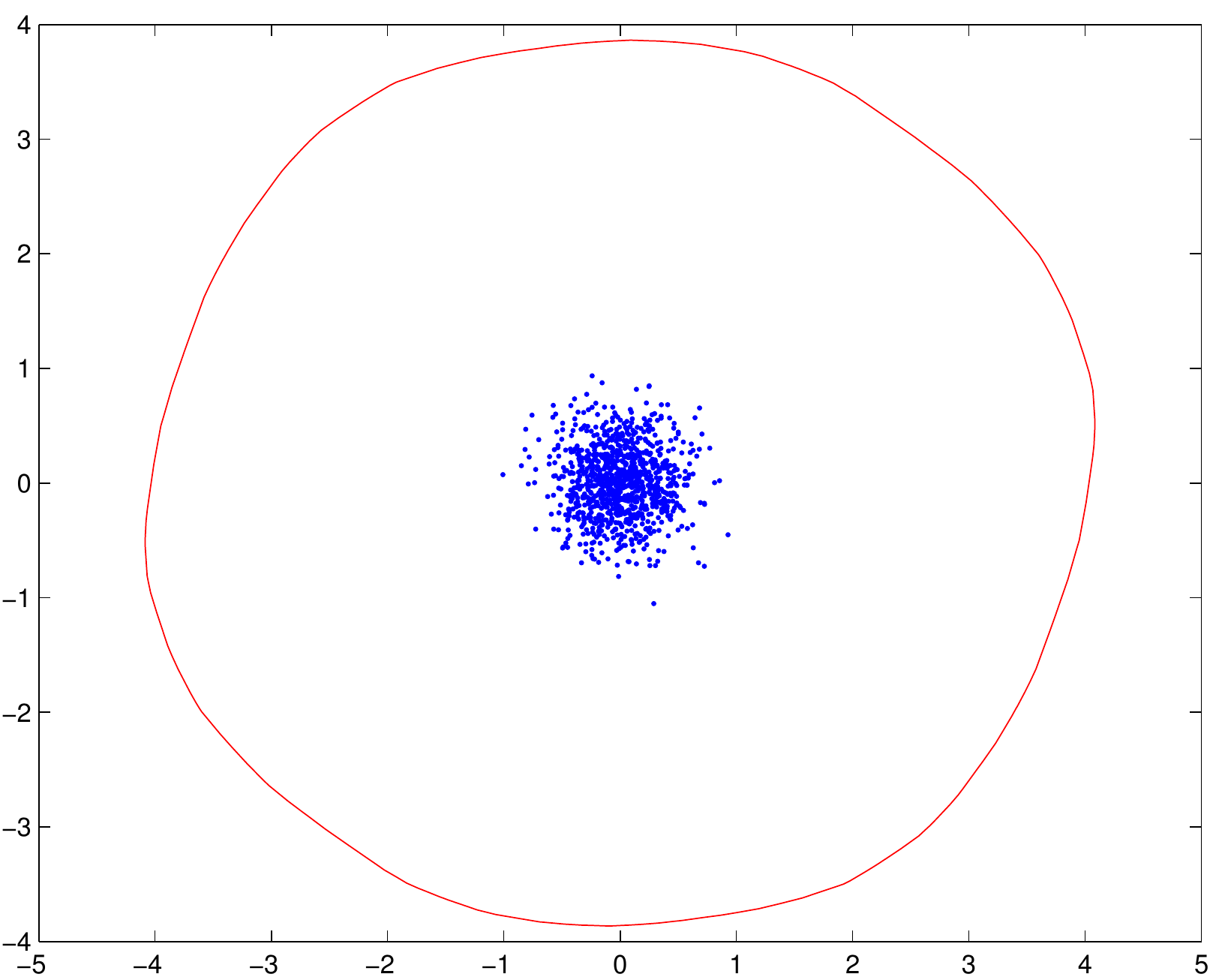}}
\caption{O mesmo, $d=20,100$.}
\end{center}
\end{figure}

A depend\^encia do di\^ametro observ\'avel no valor limiar $\kappa$ n\~ao \'e muito sens\'\i vel (somente logar\'\i tmica).

O {\em fen\^omeno de concentra\c c\~ao de medida sobre as estruturas de alta dimens\~ao} pode ser exprimido da seguinte maneira informal: 
\begin{quote}
{\em
O di\^ametro observ\'avel de um objeto geom\'etrico de alta dimens\~ao \'e tipicamente demasiado pequeno comparado ao di\^ametro atual:}
\[\obs(X) \ll\diam(X).\]
\end{quote}
\index{fen\^omeno de concentra\c c\~ao de medida}

\begin{figure}
\begin{center}
\scalebox{0.25}[0.3]{\includegraphics{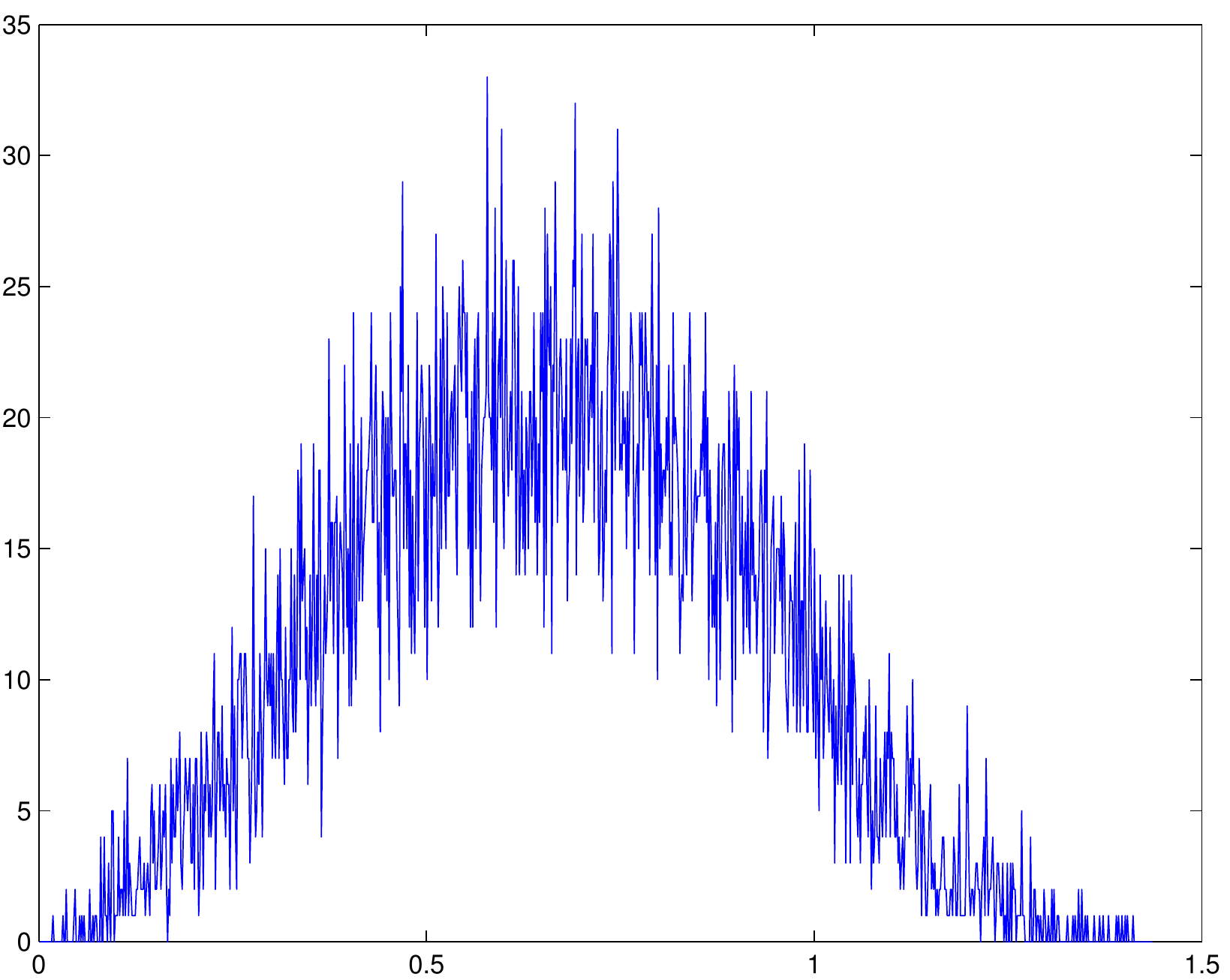}}
\scalebox{0.25}[0.3]{\includegraphics{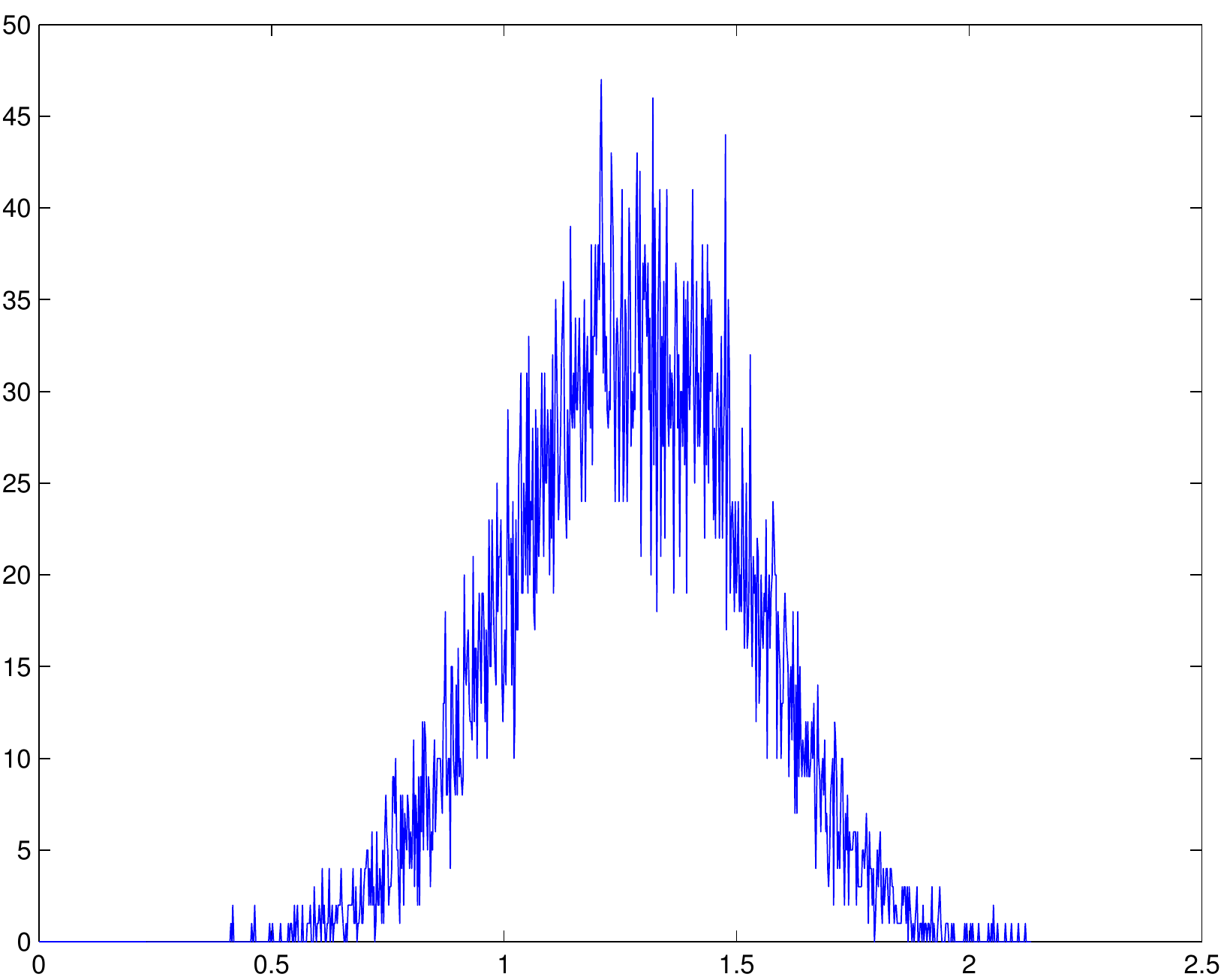}}
\caption{Distribui\c c\~ao das dist\^ancias entre $10,000$ pontos aleat\'orios do cubo $\I^d$, $d=3,10$.}
\label{tc310}
\end{center}
\end{figure}

\begin{figure}
\begin{center}
\scalebox{0.25}[0.3]{\includegraphics{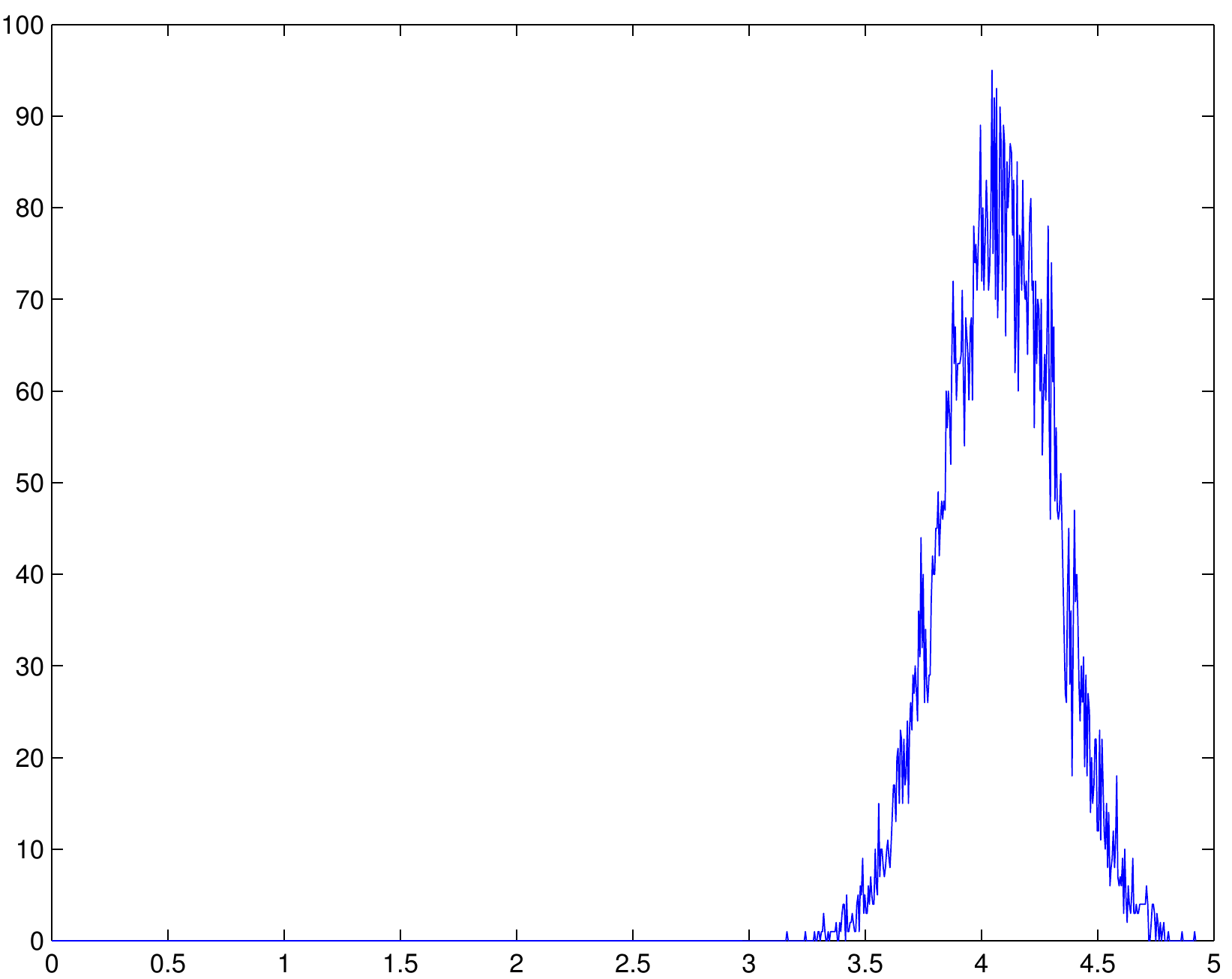}}
\scalebox{0.25}[0.3]{\includegraphics{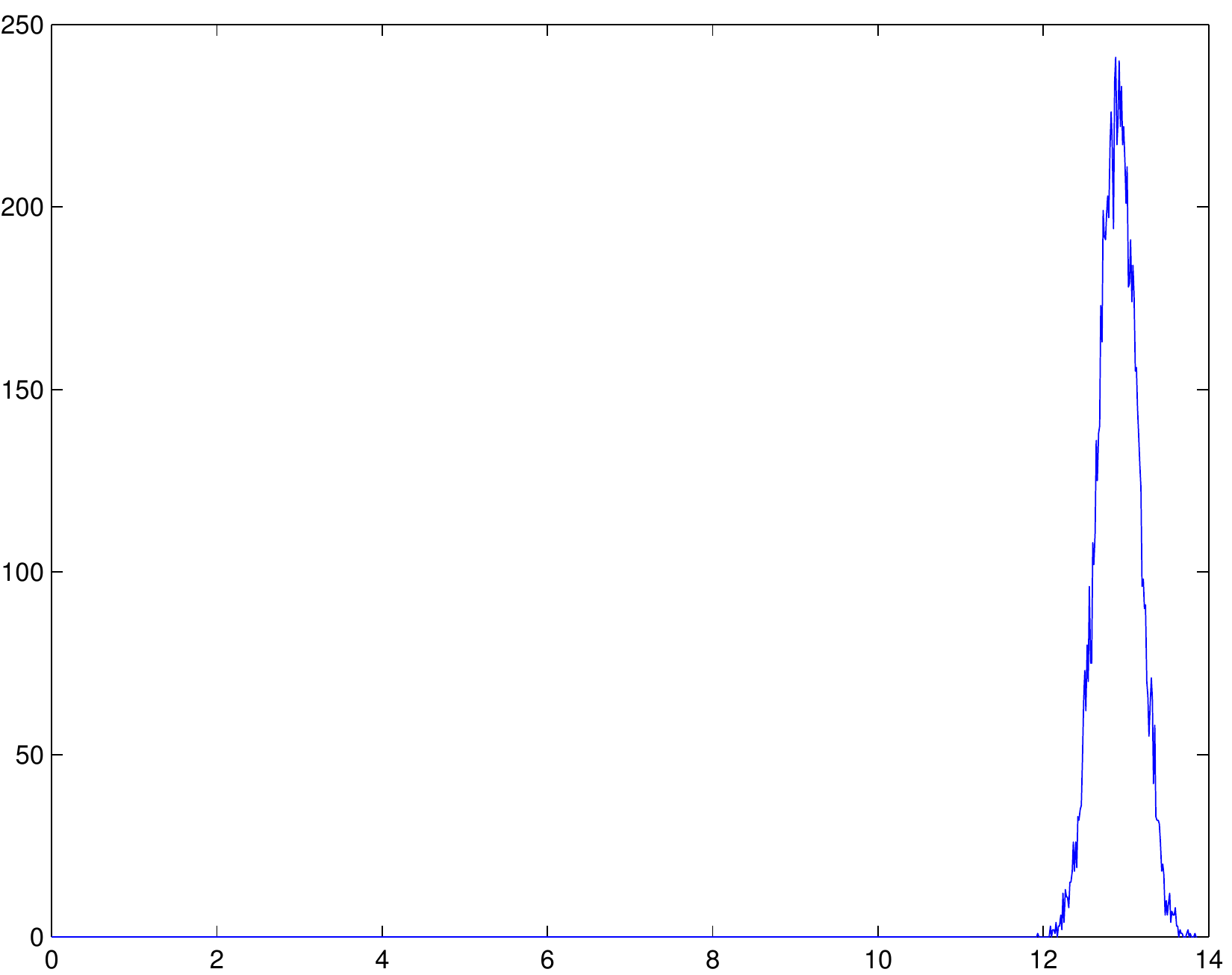}}
\caption{O mesmo, $d=100,1000$.}
\label{tc100}
\end{center}
\end{figure}

A formula\c c\~ao mais precisa usa a no\c c\~ao do {\em tamanho carater\'\i stico} de $X$ em vez do di\^ametro. Sobre um espa\c co de grande dimens\~ao, os valores da dist\^ancia $d(x,y)$ tipicamente concentram em torno da esperan\c ca da dist\^ancia, ou do {\em tamanho carater\'\i stico} de $X$,
\[{\mathrm{charSize}}\,(X)=\E_{\mu\otimes \mu}(d(x,y)).\]
\index{tamanho carater\'\i stico}
Veja as Figuras \ref{tc310} e \ref{tc100} para o cubo $\I^d$.

Por exemplo, o tamanho carater\'\i stico da esfera \'e, assintoticamente, $\Theta(1)$:
\[{\mathrm{charSize}}\,(\s^n)\to \sqrt 2\mbox{ quando }n\to\infty.\]

O fen\^omeno de concentra\c c\~ao de medida na forma mais exata diz o que
\[\mbox{Di\^ametro observ\'avel} \ll \mbox{tamanho carater\'\i stico}.\]

O fen\^omeno \'e o assunto de estudo de uma disciplina matem\'atica relativamente nova: a {\em an\'alise geom\'etrica assint\'otica.}
Esta introspe\c c\~ao na geometria dos objetos de dimens\~ao alta \'e a mais importante, e tem muitas aplica\c c\~oes e consequ\^encias amplas em ci\^encias matem\'aticas.

Eis uma reformula\c c\~ao heur\'\i stica equivalente (embora n\~ao evidente) do fen\^omeno:

\begin{quote}
{\em
Tipicamente,
num espa\c co $\Omega$ de grande dimens\~ao, para cada subconjunto $A\subseteq \Omega$ que cont\'em pelo menos a metade dos pontos, a maior parte dos pontos de $\Omega$ est\~ao pr\'oximos ao $A$.}
\end{quote}

N\'os j\'a formalizamos a no\c c\~ao de uma ``estrutura'' no texto principal, como um espa\c co m\'etrico $\Omega=(\Omega,\rho,\mu)$ munido de uma medida de probabilidade boreliana. 
Concentra\c c\~ao no cubo de Hamming $\{0,1\}^n$ forma o assunto do cap\'\i tulo \ref{ch:rotulagens}. Neste ap\^endice vamos estabelecer a concentra\c c\~ao de medida na esfera euclideana unit\'aria $\s^d$.

Denotemos
\[A_\e=\{x\in\Omega\colon \exists a\in A~~\rho(x,a)<\e\}\]
a $\e$-vizinhan\c ca do subconjunto $A$ de $\Omega$.

\begin{definicao}
  Seja $(\Omega_d,\rho_d,\mu_d)$, $d=1,2,3,\ldots$ uma fam\'\i lia de espa\c cos m\'etricos munidos de uma medida de probabilidade boreliana (espa\c cos m\'etricos com medida). Esta fam\'\i lia \'e {\em uma fam\'\i lia de L\'evy} se, para cada fam\'\i lia $A_d$, $d=1,2,\ldots$, de subconjuntos boreleanos de $\Omega_d$, tais que
  \[\liminf\mu_d(A_d)>0,\]
e por cado $\e>0$, temos
\[\mu_d((A_d)_\e)\to 1.\]
\index{fam\'\i lia! de L\'evy}
\end{definicao}

As fam\'\i lias ``naturais'' dos espa\c cos m\'etricos com medida s\~ao tipicamente as fam\'\i lias de L\'evy, depois de normalizarmos o tamanho carater\'\i stico. 
Um instrumento conveniente para quantificar o fen\^omeno da concentra\c c\~ao  \'e a {\em fun\c c\~ao de concentra\c c\~ao.} 

\begin{definicao}
 Seja $(\Omega,d,\mu)$ um espa\c co m\'etrico com medida. A {\em fun\c c\~ao de concentra\c c\~ao} de $\Omega$, notada $\alpha_\Omega(\e)$, \'e definida pelas condi\c c\~oes seguintes:

\[\alpha(\e)=\left\{
  \begin{array}{ll} \frac 12, & \mbox{se $\e=0$,} \\
1-\min\left\{\mu_\sharp\left(A_\e\right) \colon
A\subseteq\Sigma^n, ~~ \mu_\sharp(A)\geq\frac 12\right\}, &
\mbox{se $\e>0$.}
\end{array}\right.
\]
\label{def:concfn1}
\index{fun\c c\~ao! de concentra\c c\~ao}
\end{definicao}

\begin{teorema}
  Uma fam\'\i lia $(\Omega_d,\rho_d,\mu_d)$ \'e uma fam\'\i lia de L\'evy se e apenas se as fun\c c\~oes de concentra\c c\~ao tendem a zero:
  \[\alpha(\Omega_d,\e)\to 0\mbox{ para cada }\e>0.\]
\qed
\end{teorema}

\begin{definicao}
  Uma fam\'\i lia de L\'evy $(\Omega_d,d_d,\mu_d)$ \'e chamada uma {\em fam\'\i lia de L\'evy normal} se existem $C_1,C_2>0$ tais que
  \[\alpha(\Omega_d,\e)\leq C_1e^{-C_2\e^2d}.\]
\index{fam\'\i lia! de L\'evy! normal}
\end{definicao}

Aqui est\'a a liga\c c\~ao com o di\^ametro observ\'avel:
sobre uma estrutura de grande dimens\~ao, toda fun\c c\~ao Lipschitz cont\'\i nua \'e quasi constante em toda parte exceto sobre um conjunto da medida muito pequena. 

Relembramos que um n\'umero real $M=M_f$ \'e dito um {\em valor mediano} de uma fun\c c\~ao boreliana $f$, sobre um espa\c co com medida de probabilidade $(\Omega,\mu)$ se
\[\mu\{x\in \Omega\colon f(x)\geq M\}\geq\frac 12
\mbox{ e } 
\mu\{x\in \Omega\colon f(x)\leq M\}\geq\frac 12.\]
\index{valor! mediano}
Um valor mediano $M=M_f$ existe sempre, mas geralmente, n\~ao \'e \'unico.

\begin{exercicio}
  Seja $f$ uma fun\c c\~ao Lipschitz cont\'\i nua com a constante de Lipschitz $L\geq 0$ sobre um espa\c co m\'etrico com medida, $(\Omega,\rho,\mu)$. Provar que
  \[\mu\{\abs{f(x)-M}>\e\}\leq 2\alpha_\Omega\left(\frac{\e}{L}\right).\]
  Mais geralmente, se $f$ \'e uniformemente cont\'\i nua de tal modo que
\[\forall x,y\in X,~~d(x,y)<\delta\Rightarrow \abs{fx-fy}<\e,\]
ent\~ao
\[\mu\{\abs{f(x)-M}>\e\}\leq 2\alpha_X(\delta).\]
\tri
\end{exercicio}

A fun\c c\~ao dist\^ancia $d(-,p)$ de um ponto $p$ fixo qualquer \'e Lipschitz cont\'\i nua (com a constante $1$), e nos dom\'\i nios de alta dimens\~ao uma tal fun\c c\~ao concentra-se em torno do valor mediano.

Entre os livros tratando o fen\^omeno de concentra\c c\~ao de medida, \citep*{MS} \'e o mais acess\'\i vel, \citep*{L} o mais abrangente e \citep*{Gr} cont\'em uma riqueza de ideias. Eu n\~ao tive a chance de obter uma c\'opia do livro recente \citep*{AAGM}, que parece ser excelente e pode tornar-se a fonte principal de refer\^encia. Tamb\'em, eu ainda n\~ao estudei o livro \citep*{vershynin} dedicado \`as aplica\c c\~oes do fen\^omeno de concentra\c c\~ao na ci\^encia de dados, inclusive t\'opicos na aprendizagem estat\'\i stica.

\section{Medida de Haar sobre a esfera euclideana\label{s:haarnaesfera}}
\subsection{Duas m\'etricas}

A esfera euclideana de raio um 
\[
\s^n=\{x\in\ell^{2}(n+1)\mid \abs x=1\}
\]
admite duas m\'etricas padr\~ao. Elas s\~ao: a dist\^ancia euclideana induzida do espa\c co $\ell^2(n+1)$,
\[d_{eucl}(x,y)=\norm{x-y}_2,\]
assim como a dist\^ancia geod\'esica, ou seja, o comprimento do menor arco do c\'\i rculo m\'aximo entre $x$ e $y$, ou, de modo equivalente, o valor do \^angulo entre dois vetores (em radianos):
\[d_{geo}(x,y)=\angle(x,y).\]
As duas dist\^ancias s\~ao equivalentes: para todos $x,y\in\s^n$, temos
\[d_{eucl}(x,y)\leq d_{geo}(x,y)\leq \frac\pi 2 d_{eucl}(x,y),\]
e no caso onde $d_{geo}(x,y)\leq \pi/2$, temos
\begin{equation}
d_{eucl}(x,y)\leq d_{geo}(x,y)\leq \frac\pi {2\sqrt 2} d_{eucl}(x,y).\end{equation}

At\'e melhor, as duas dist\^ancias s\~ao ligadas por uma transforma\c c\~ao m\'etrica:
\begin{equation}
d_{eucl}(x,y) = 2\sin\left(\frac{d_{geo}(x,y)}{2}\right).
\label{eq:metrictransform}
\end{equation}

O grupo ortogonal
\[O(n+1)=\{u\in M_{n+1}(\R)\colon u^t u=uu^t = 1\}\]
\index{grupo! ortogonal}
age sobre a esfera por isometrias:
\[\s^n\ni x\mapsto ux\in \s^n,~~u\in O(n+1).\]
Aqui $ux$ \'e o produto usual de uma matriz ortogonal $u$ do comprimento 
$(n+1)\times (n+1)$ e um vetor-coluna $x$ do comprimento $(n+1)\times 1$. De fato, isometrias com rela\c c\~ao a uma dist\^ancia s\~ao isometrias com rela\c c\~ao a outra.

\subsection{Medidas de probabilidade sobre a esfera\label{s:haar}}
Como antes, denotemos por $P(X)$ a fam\'\i lia de todas as medidas de probabilidade borelianas sobre um espa\c co boreliano padr\~ao $X$. O teorema de representa\c c\~ao de Riesz \ref{t:riesz} diz que as medidas de probabilidade borelianas sobre um espa\c co compacto $X$ admitem uma correspond\^encia bijetora com os funcionais lineares e positivos sobre $C(X)$ enviando $1$ para $1$.

No teorema de Riesz, em vez de fun\c c\~oes cont\'\i nuas, basta considerar as fun\c c\~oes Lipschitz cont\'\i nuas. 
Por exemplo, teorema de Stone--Weierstrass \ref{stone-weierstrass} implica que as fun\c c\~oes Lipschitz cont\'\i nuas s\~ao densas em $C(X)$.

Para $X$ compacto metriz\'avel, $P(X)$ torna-se em um espa\c co compacto metriz\'avel (teorema \ref{t:topologiaPX}). A topologia sobre $P(X)$ \'e completamente determinada pelas sequ\^encias convergentes, e uma sequ\^encia $(\mu_n)$ de medidas de probabilidade converge para uma medida de probabilidade $\mu$ se e somente se, para toda fun\c c\~ao $f\colon X\to\R$, temos
\begin{equation}
\label{eq:seq}
\int f\,d\mu_n \to \int f\,d\mu.
\end{equation}

Uma medida de probabilidade, $\mu$, sobre $\s^n$ \'e {\em invariante por rota\c c\~oes} se para todo subconjunto boreliano $A\subseteq\s^n$ e toda $u\in O(n+1)$, temos
\[\mu(uA)=\mu(A).\]
De modo equivalente, isso \'e o caso se para todas fun\c c\~oes cont\'\i nuas (ou Lipschitz cont\'\i nuas, ou mesmo $1$-Lipschitz cont\'\i nuas) $f\colon \s^n\to\R$ e todas $u\in O(n+1)$, temos
\[\int \,^uf\,d\mu = \int f\,d\mu,\]
onde 
\[^uf(x) = f(u^{-1}x)\]
\'e a transla\c c\~ao a esquerda de $f$ por $u$. 

Finalmente, pode-se definir uma transla\c c\~ao a esquerda da medida $\mu$ por $u$, como segue:
\[u\ast \mu (A) = \mu(u^{-1}A).\]
A invari\^ancia de $\mu$ significa que para cada $u$,
\[u\ast\mu=\mu.\]

O nosso resultado principal \'e o seguinte. 

\begin{teorema}
Existe uma e uma s\'o medida de probabilidade boreliana $\nu=\nu_n$ sobre $\s^n$ que \'e invariante por isometrias:
\[\nu_n(A)=\nu_n(uA)\mbox{ para toda }u\in O(n+1).\]
\index{medida! de Haar}
\end{teorema}

Esta medida \'e chamada a {\em medida de Haar}. A prova que vamos fazer estende-se facilmente sobre todos grupos compactos e seus espa\c cos homog\^eneos. Ela pode ser estendido sobre todos os grupos localmente compactos, mas a extens\~ao exige a Axioma de Escolha.

\subsection{Teorema de Hall--Rado}
Seja $C$ um subconjunto finito de v\'ertices de um grafo. Denotemos por $\partial C$ a cole\c c\~ao de todos os v\'ertices adjacentes a um ou mais v\'ertices de $C$. 

\begin{teorema}[Teorema de emparelhamento de Hall--Rado]
Seja $\Gamma=(V,E)=(A,B,E)$ um grafo finito bipartido, onde $V$ denota v\'ertices, $E$ denota arestas, $V=A\sqcup B$. As condi\c c\~oes seguintes s\~ao equivalentes.
\begin{enumerate}
\item Existe uma aplica\c c\~ao injetora $i\colon A\to B$ com $(a,i(a))\in E$ para todo $a\in A$ (um {\em emparelhamento} para $A$).
\item Para todo $C\subseteq A$, temos
\[\abs{\partial C}\geq \abs C.\]
\end{enumerate}
\index{teorema! de Hall-Rado}
\end{teorema}

\begin{proof}
A implica\c c\~ao (1) $\Rightarrow$ (2) \'e \'obvia porque $i(C)\subseteq\partial C$. A implica\c c\~ao contr\'aria (2) $\Rightarrow$ (1) tem pelo menos $7$ diferentes provas conhecidas. Vamos apresentar uma delas, usando indu\c c\~ao matem\'atica sobre o tamanho $n$ de $A$. 

Para $n=1$ a afirma\c c\~ao \'e evidente. Suponha que ela foi mostrada para todos valores $1\leq k\leq n$, e seja $\abs A=n+1$. Vamos estudar dois casos separadamente.

(a) Tem muita redund\^ancia no grafo, nomeadamente, para todo subconjunto pr\'oprio $C\subsetneqq A$ temos $\abs{\partial C}>\abs C$. Se for o caso, escolha um v\'ertice $a_0\in A$ assim como um v\'ertice adjacente $b_0\in B$. Denote $\tilde\Gamma = (\tilde V,\tilde E)$ o grafo obtido de $\Gamma$ apagando os v\'ertices $a_0,b_0$ assim como todas as arestas provenientes de $a_0$ ou chegando em $b_0$. Se $C\subseteq \tilde V$, ent\~ao $C\subseteq A$ e por que no m\'aximo um elemento da fronteira de $C$ foi perdido (nomeadamente, $b_0$), temos
$\abs{\partial_{\tilde \Gamma} C} \geq \abs{\partial_{\Gamma}}C -1 \geq \abs C$. Logo, o grafo $\tilde\Gamma$ satisfaz a hip\'otese e como $\tilde A = A\setminus\{a_0\}$ s\'o tem $n$ v\'ertices, existe um emparelhamento $i$ para $\tilde A$ cuja imagem exclui $b_0$. Estendendo $i$ sobre $A$ por $i(a_0)=b_0$, obtemos o emparelhamento desejado para $A$.

(b) N\~ao tem redund\^ancia como acima, logo existe um conjunto $C\subsetneqq A$, $C\neq \emptyset$ tal que $\abs{\partial C} = \abs C$. Neste caso, todo v\'ertice $c\in C$ \`a adjacente \`a um v\'ertice de $B$ s\'o, logo pode-se definir um emparelhamento parcial $i_C$ de $C$ de modo \'unico poss\'\i vel. Definamos $D = A\setminus C$, e seja $\tilde\Gamma$ um grafo bipartido obtido de $\Gamma$ apagando $C$ e $i_C(C)$, assim como todas as arestas entre estes dois conjuntos. O grafo $\Gamma$ satisfaz a hip\'otese: de fato, supondo que para $E\subsetneqq D$ tem $\abs{\partial E} <\abs E$, obtemos uma contradi\c c\~ao, porque $\partial(D\cup C) = \partial D\cup\partial C$, onde a uni\~ao \'e disjunta, de modo que a cardinalidade dela \'e $<\abs D + \abs C$. Como $\abs D\leq n$, existe um emparelhamento parcial $i_D$ segundo a hip\'otese indutiva. Coalescendo $i_C$ e $i_D$, obtemos o emparelhamento desejado $i$ para $A$.
\end{proof}

\subsection{Constru\c c\~ao da medida de Haar}
Seja $\e>0$ qualquer. A esfera $\s^n$, sendo compacta, admite uma $\e$-rede finita. Seja $\mathcal N={\mathcal N}(\e)$ uma tal $\e$-rede do menor tamanho poss\'\i vel, com $N(\e)=N(\e,\s^n)$ elementos. (A dist\^ancia sobre a esfera pode ser euclideana ou geod\'esica).
Denotemos por $\mu_{{\mathcal N}(\e)}$ a medida emp\'\i rica suportada sobre os elementos da rede $\mathcal N(\e)$, e seja
\[\int f\,d\mu_{{\mathcal N}(\e)} =\frac{1}{N(\e,\s^n)}\sum_{x\in {\mathcal N}(\e)}f(x)\]
a integral correspondente (o valor m\'edio de $f$ sobre elementos de $\mathcal N(\e)$).

A sequ\^encia de medidas $(\mu_{{\mathcal N}(1/k)})$, $k\in\N_+$ admite uma subsequ\^encia convergente, pois o espa\c co $P(\s^n)$ \'e compacto e metriz\'avel. Escolha uma sequ\^encia de valores de $\e$, $(\e_k)$, e uma medida $\mu$, tais que
\[\mu_{{\mathcal N}(\e_k)} \to \mu.\]

\'E a nossa medida de Haar, faltando s\'o verificar a sua invari\^ancia e a unicidade.

\subsection{Invari\^ancia de $\mu$ pelas isometrias}

\begin{lema} Seja $\mathcal N$, $\mathcal N^\prime$ duas $\e$-redes quaisquer do tamanho m\'\i nimo $N(\e)$, e seja $f\colon\s^n\to\R$ uma fun\c c\~ao $1$-Lipschitz cont\'\i nua. Ent\~ao,
\[\left\vert \int f\,d\mu_{\mathcal N} - \int f\,d\mu_{\mathcal N^\prime}\right\vert < 2\e.
\]
\label{l:nosso}
\end{lema}

\begin{proof}
Formemos um grafo bipartido com o conjunto de v\'ertices ${\mathcal N}\sqcup {\mathcal N}^\prime$, onde $x\in {\mathcal N}$ e $y\in {\mathcal N}^\prime$ s\~ao adjacentes se e somente se as bolas abertas de raio $\e$ em torno de $x$ e de $y$ se encontram. Vamos verificar a condi\c c\~ao (ii) do teorema de Hall--Rado. Seja $C\subseteq {\mathcal N}$ qualquer. 

Segundo a defini\c c\~ao de uma $\e$-rede, todo elemento de uma bola $B_\e(x)$, $x\in C$, e contido numa bola $B_\e(y)$, onde $y\in{\mathcal N}^\prime$. Isso implica que $y$ \'e adjacente a $x$, logo pertence a $\partial C$. Conclu\'\i mos: 
\[\bigcup_{x\in C} B_\e(x)\subseteq \bigcup_{y\in\partial C} B_\e(y).\]
Por conseguinte, se eliminarmos da rede $\mathcal N$ o conjunto $C$ e substituirmos $\partial C$, obtemos uma $\e$-rede de novo. Como o tamanho de uma $\e$-rede n\~ao pode ser menor do tamanho de $\mathcal N$, conclu\'\i mos: $\abs{\partial C}\geq \abs C$. 

Segundo o teorema de Hall--Rado, existe uma inje\c c\~ao $i\colon {\mathcal N}\to {\mathcal N}^\prime$ (logo, uma bije\c c\~ao) tal que 
\[d(x,i(x))<2\e\]
para todo $x\in {\mathcal N}$. Porque $f$ \'e $1$-Lipschitz cont\'\i nua, temos
\[\abs{f(x)-f(i(x))}<2\e,\]
o que traduz-se na propriedade parecida para os valores m\'edios.
\end{proof}

Sejam agora $u\in O(n+1)$ qualquer, e $f$ uma fun\c c\~ao $1$-Lipschitz cont\'\i nua sobre a esfera. Seja $\delta>0$. Escolha $N$ t\~ao grande que para todo $k\geq N$,  
\[\left\vert \int f\,d\mu - \int f\,d \mu_{\e_k}\right\vert <\delta
\mbox{ and }\left\vert \int \,^uf\,d\mu - \int \,^uf\,d \mu_{\e_k}\right\vert <\delta.\]
A medida $u\ast \mu_{\e_k}$ \'e uma medida uniforme suportada sobre os elementos de $u^{-1}{\mathcal N}({\e_k})$, uma $\e_k$-rede do tamanho m\'\i nimo. Segundo lema \ref{l:nosso}, 
\[\left\vert \int f\,d\mu_{\e_k} - \int \,^uf\,d \mu_{\e_k}\right\vert <2\e_k.\]
Conclu\'\i mos:
\[\left\vert \int f\,d\mu - \int \,^uf\,d \mu\right\vert <2(\delta+\e_k)\to 0,\]
e o resultado segue-se. \qed

\begin{observacao}
O mesmo argumento estabelece a exist\^encia de uma medida de probabilidade boreliana sobre cada grupo compacto metriz\'avel, munido de uma m\'etrica  \'e invariante pelas transla\c c\~oes a esquerda e a direita. Particularmente, isto se aplica ao grupo ortogonal $O(n+1)$.
Como uma m\'etrica bi-invariante gerando a topologia, pode-se usar, por exemplo, a dist\^ancia de Hilbert-Schmidt, ou seja, a dist\^ancia entre as matrizes vistas como vetores num espa\c co euclideano: 
\[\norm{u-v}_2 = {\mathrm{tr}}\,((u-v)^t(u-v)).\]
Mais uma vez, esta medida chama-se a medida de Haar sobre $O(n+1)$.
\label{obs:haar}
\end{observacao}

\begin{exercicio}
Verifique que a dist\^ancia de Hilbert--Schmidt \'e bi-invariante sobre o grupo $O(n+1)$.
\end{exercicio}

\subsection{Unicidade de $\mu$}
Com a finalidade de estabelecer a unicidade da medida $\mu$, vamos utilizar uma medida de Haar sobre $O(n+1)$, denotada $\nu$. 
Dado uma $f\in C(\s^n)$, temos, usando o teorema de Fubini \ref{t:fubini} e a invari\^ancia das medidas,
\begin{eqnarray*}
\mu(f) &=& 
\int_{\s^n} f(x)\,d\mu(x) \\
&=& \int_{O(n+1)}\underbrace{\int_{\s^n} f(ux)\,d\mu(x)}_{\mbox{\tiny constante em $u$}}\,d\nu(u)\\
&=&
\int_{\s^n}\int_{O(n+1)} f(ux)\,d\nu(u)\,d\mu(x).
\end{eqnarray*}
Agora vamos mostrar que, de fato, a integral interior
\[\int_{O(n+1)} f(ux)\,d\nu(u)\] 
n\~ao depende de $x$. Para todo $y\in\s^n$, tem $v\in O(n+1)$ tal que $v^{-1}x=y$, o que implica 
\begin{eqnarray*}
\int_{O(n+1)} f(ux)\,d\nu(u) &=& \int_{O(n+1)} \,^vf(ux)\,d\nu(u) \\
&=&
\int_{O(n+1)} f(v^{-1}uv y)\,d\nu(u) \\
&=& \int_{O(n+1)} f(wy)\,d\nu(w)\end{eqnarray*}
(depois a mudan\c ca de vari\'avel $w=v^{-1}uv$, compare exerc\'\i cio \ref{ex:mudancavariavel}).

Denotemos o valor dessa integral por $\bar\nu(f)$. Deste modo,
\[\mu(f) =\bar\nu(f).\]
Se agora $\mu^\prime$ \'e uma outra medida de probabilidade invariante sobre a esfera, temos pelo mesmo argumento
\[\mu^\prime(f) =\bar\nu(f) = \mu(f).\]
\qed

\begin{observacao}
O argumento id\^entico estabelece a unicidade de uma medida de Haar sobre um espa\c co homog\^eneo qualquer de um grupo compacto, inclusive sobre o grupo $O(n)$ pr\'oprio.
\label{obs:unicidade}
\end{observacao}

\section{Concentra\c c\~ao de medida na esfera}

\subsection{Medida de calotas esf\'ericas}

Denotemos por
\[C_r(x)=\{y\in\s^n\colon d_{geo}(x,y)\leq r\}\]
a calota fechada esf\'erica de raio geod\'esico $r$. Por exemplo, as calotas de raio $\pi/2$ s\~ao hemisf\'erios.

\begin{exercicio}
Verificar que
\[\cos t \leq e^{-t^2/2}\]
quando $0\leq t\leq \pi/2$. \tri
\end{exercicio}

\begin{lema}
A medida de Haar normalizada, $\mu_{n+1}$, da calota esf\'erica da esfera $\s^{n+1}$ de raio geod\'esico $\frac\pi 2-\e$, onde $\e>0$, satisfaz
\begin{equation}
\label{eq:fracpi8}
\mu_{n+1}(C_\e)\leq \sqrt{\frac{\pi}{8}}e^{-n\e^2/2}.\end{equation}
\label{l:fracpi8}
\end{lema}

\begin{proof}
Denotemos, para simplificar, $C=C_{\pi/2-\e}(x_0)$. Para todo $\theta>0$, o conjunto de pontos \`a dist\^ancia $\pi/2-\theta$ de $x_0$ forma uma esfera de dimens\~ao $n$ de raio $\cos\theta$, cujo volume \'e proporcional a $\cos^n\theta$. Por conseguinte,
\begin{eqnarray*}
\mu(C) &=& \frac{\int_{\e}^{\pi/2}\cos^n\theta d\theta}{\int_{-\pi/2}^{\pi/2}\cos^n\theta d\theta}.
\end{eqnarray*}
Denotemos
\[I_n=\int_{0}^{\pi/2}\cos^n\theta d\theta.\]
Integrando por partes, 
\begin{eqnarray*}
I_k &=& \int_0^{\pi/2}\cos^k\theta\,d\theta \\
&=& \int_0^{\pi/2}\cos^{k-1}\theta\,d(\sin\theta) \\
&=&\cos^{k-1}\theta\sin\theta \vert_0^{\pi/2} - \int_0^{\pi/2}(k-1)\cos^{k-2}\theta(-\sin\theta)\sin\theta\,d\theta \\
&=& (k-1)\int_0^{\pi/2}\cos^{k-2}\theta\sin^2\theta\,d\theta \\
&=& (k-1)\int_0^{\pi/2}\cos^{k-2}\theta\,d\theta - (k-1)\int_0^{\pi/2}\cos^k\theta\,d\theta \\
&=& (k-1) I_{k-2} - (k-1)I_k,
\end{eqnarray*}
obtemos a rela\c c\~ao recursiva
\[I_k=\frac{k-1}{k}I_{k-2},\]
qual, depois a multiplica\c c\~ao por $\sqrt k$ e usando o fato \'obvio $k-1\geq \sqrt{k(k-2)}$, leva a
\[\sqrt k I_k\geq \sqrt{k-2}I_{k-2}.\]
Conclu\'\i mos:
\[\sqrt n I_n\geq \min\{I_1,\sqrt 2 I_2\}=\min\{1,\sqrt 2 \pi/4\}=1.\]
Segue-se que
\begin{equation}
\frac{1}{2 I_n}\leq \frac {\sqrt n}2.
\label{eq:I_n}
\end{equation}

Usando a mudan\c ca de vari\'aveis $\theta = \tau/\sqrt n$, seguida por mais uma mudan\c ca de vari\'aveis, $t=\tau - \e\sqrt n$, conseguimos
\begin{eqnarray*}
2I_n\mu(C) &=& \int_{\e}^{\pi/2}\cos^n\theta d\theta \\
&=&
\frac 1{\sqrt n} \int_{\e\sqrt n}^{(\pi/2)\sqrt n}\cos^n\left(\frac\tau{\sqrt n}\right)\,d\tau \\
&\leq &\frac 1{\sqrt n} \int_{\e\sqrt n}^{(\pi/2)\sqrt n}e^{-\tau^2/2}\,d\tau \\
&\leq & \frac 1{\sqrt n}e^{-\e^2n/2}\int_0^{(\pi/2-\e)\sqrt n}e^{-t^2/2}\,dt \\
&\leq & \frac 1{\sqrt n}e^{-\e^2n/2}\int_0^{\infty}e^{-t^2/2}\,dt \\
&=& \frac{e^{-\e^2n/2}\sqrt{\pi/2}}{\sqrt n},
\end{eqnarray*}
e a eq. (\ref{eq:I_n}) leva a
\begin{eqnarray*}
\mu(C) &\leq& \frac{e^{-\e^2n/2}\sqrt{\pi/2}}{\sqrt n} \times \frac {\sqrt n}2
\\
&=& \sqrt{\frac{\pi}{8}}e^{-n\e^2/2}.
\end{eqnarray*}
\end{proof}

\subsection{Dist\^ancia de Hausdorff}

Seja $X=(X,d)$ um espa\c co m\'etrico de di\^ametro finito. Para todos subconjuntos $A,B\subseteq X$, defina-se a {\em dist\^ancia de Hausdorff} entre eles, como segue:
\[d_H (A,B)=\inf\{\e>0\colon A\subseteq B_\e \wedge B\subseteq A_\e\}.\]
\index{dist\^ancia! de Hausdorff}

\begin{exercicio}
Verificar que a dist\^ancia de Hausdorff sobre o espa\c co de todos os subconjuntos de $X$ \'e n\~ao negativa, sim\'etrica, e satisfaz a desigualdade triangular. (Ou seja, $d_H$ \'e uma {\em pseudom\'etrica}).
\tri
\end{exercicio}

\begin{exercicio}
Mostre que, se $F,G$ s\~ao subconjuntos fechados num espa\c co m\'etrico $X$ tais que $d_H(F,G)=0$, ent\~ao $F=G$. 
\end{exercicio}

Por conseguinte, a fam\'\i lia ${\mathfrak F}(X)$ de todos os subconjuntos n\~ao v\'arios e fechados de $X$, munida da dist\^ancia de Hausdorff, \'e um espa\c co m\'etrico. 

Relembremos que um espa\c co m\'etrico $X$ \'e {\em totalmente limitado,} ou {\em pr\'e-compacto,} se para todo $\e>0$ existe uma $\e$-rede finita para $X$. Um espa\c co m\'etrico $X$ \'e compacto se e somente se ele \'e pr\'e-compacto e completo. 

\begin{lema}
Se $X$ \'e um espa\c co m\'etrico totalmente limitado, ent\~ao ${\mathfrak F}(X)$ \'e totalmente limitado.
\end{lema}

\begin{proof}
Seja $\e>0$ qualquer. Escolha uma $\e$-rede para $X$, $\mathcal N$. \'E f\'acil de verificar que a fam\'\i lia finita $2^{\mathcal N}$ de todos os subconjuntos n\~ao vazios de $\mathcal N$ forma uma $\e$-rede para o espa\c co ${\mathfrak F}(X)$.
\end{proof} 

\begin{lema}
Se $X$ \'e um espa\c co m\'etrico compacto, ent\~ao ${\mathfrak F}(X)$ \'e compacto.
\end{lema}

\begin{proof}
S\'o temos que mostrar que o espa\c co m\'etrico ${\mathfrak F}(X)$ \'e completo. 
Seja $(F_n)$ uma sequ\^encia de Cauchy em ${\mathfrak F}(X)$. Sem perda de generalidade e passando a uma subsequ\^encia se for necess\'ario, pode-se supor que para todos $n\leq m$,
\[F_m\subseteq (F_n)_{2^{-n}}.\]
Seja $\overline{(F_n)_{2^{-n-1}}}$ a ader\^encia da $2^{-n-1}$-vizinhan\c ca de $F_n$, para todo $n$. \'E claro que a sequ\^encia seguinte de conjuntos fechados \'e encaixada:
\[\overline{(F_1)_{1/4}}\supseteq \overline{(F_2)_{1/8}}\supseteq \ldots
\supseteq \overline{(F_n)_{2^{-n-1}}}\supseteq \ldots\]
A interse\c c\~ao
\[F=\cap_{n=1}^\infty \overline{(F_n)_{2^{-n-1}}}\]
n\~ao \'e vazia em virtude da compacidade de $X$, e agora verifica-se facilmente que a sequ\^encia $(F_n)$ converge para $F$.
\end{proof}

\begin{definicao}
Uma fun\c c\~ao $f\colon X\to\R$ sobre um espa\c co m\'etrico (ou topol\'ogico) $X$ \'e dita {\em semicont\'\i nua superiormente} em um ponto $x_0\in X$ se para todo $\e>0$ tem uma vizinhan\c ca $V$ de $x_0$ tal que
\[\forall y\in V,~~f(y)\leq f(x_0)+\e.\]
A fun\c c\~ao $f$ \'e semicont\'\i nua superiormente se ela \'e semicont\'\i nua superiormente em todo ponto.

De modo equivalente, $f$ \'e semicont\'\i nua superiormente se a imagem inversa de todo intervalo aberto e semiinfinito
$(-\infty,b)$ \'e aberto em $X$. Mais uma defini\c c\~ao equivalente: a imagem inversa de todo intervalo fechado e semiinfinito $[a,+\infty)$ \'e fechado em $X$.
\tri
\index{fun\c c\~ao! semicont\'\i nua superiormente}
\end{definicao}

Os dois resultados seguintes s\~ao verdadeiros se a esfera $\s^n$ for substitu\'\i da por um espa\c co m\'etrico qualquer do di\^ametro limitado, munido de uma medida boreliana de probabilidade.

\begin{exercicio}
Mostre que a fun\c c\~ao
\[A\mapsto \mu(A)\]
sobre o espa\c co ${\mathfrak F}(\s^n)$ munido da dist\^ancia de Hausdorff e a medida de Haar \'e semicont\'\i nua superiormente, mas n\~ao \'e cont\'\i nua. \tri
\label{ex:continua}
\end{exercicio}

\begin{exercicio}
\label{ex:bB}
Seja $A$ um subconjunto fechado de $\s^n$. Definamos a fam\'\i lia
${\mathcal G}\subseteq {\mathfrak F}(\s^n)$ como a cole\c c\~ao de todos subconjuntos fechados $B$ da esfera $\s^n$ tendo duas propriedades:
\begin{enumerate}
\item $\nu(B)=\nu(A)$,
\item para todo $\e>0$, $\nu(B_\e)\leq \nu(A_\e)$.
\end{enumerate}
Mostre que o conjunto $\mathcal G$ \'e fechado em
${\mathfrak F}(\s^n)$ com rela\c c\~ao \`a m\'etrica de Hausdorff.
\tri
\end{exercicio}

O seguinte \'e a ``metade superior'' do teorema de Weierstrass sobre as fun\c c\~oes cont\'\i nuas sobre os conjuntos compactos.

\begin{exercicio}
Mostre que toda fun\c c\~ao semicont\'\i nua superiormente sobre um espa\c co compacto tem o m\'aximo.
\par
[ {\em Sugest\~ao:} use a contrapositiva. ]
\label{ex:maximo}
\end{exercicio}

\subsection{Simetriza\c c\~ao por dois pontos}

Seja $\vec x\in\s^n\subseteq \ell^2(n+1)$ um vetor de comprimento um. Denotemos
\[H=H(\vec x)=\{y\in \ell^2(n+1)\colon \langle y,\vec x\rangle =0\}\]
o hiperplano ortogonal \`a dire\c c\~ao $\vec x$, 
\[\s_0=\s_0(\vec x)=\{y\in\s^n\colon \langle y,\vec x\rangle =0\}=\s^n\cap H,\]
a esfera equatorial correspondente, 
\[\s_+ = \s_+(\vec x) = \{y\in\s^n\colon \langle y,\vec x\rangle >0\},\]
o hemisf\'erio aberto superior, e
\[\s_- = \s_-(\vec x) = \{y\in\s^n\colon \langle y,\vec x\rangle <0\},\]
o hemisf\'erio aberto inferior (com rela\c c\~ao ao vetor de dire\c c\~ao $\vec x$). 

Denotemos por $\sigma=\sigma(\vec x)$ a involu\c c\~ao (reflex\~ao) com rela\c c\~ao a $H$:
\[\sigma(y) = y - 2\langle y,\vec x\rangle \vec x.\]
Temos claramente
\[\sigma^2={\mathrm{Id}}.\]
Sendo uma transforma\c c\~ao ortogonal, $\sigma$ conserva a dist\^ancia (euclideana assim como geod\'esica), assim como a medida de Haar sobre a esfera $\s^n$. Em particular, $\sigma$ estabelece um isomorfismo entre os hemisf\'erios $\s_+$ e $\s_-$ vistos como espa\c cos m\'etricos com medida.

\begin{exercicio}
\label{ex:ineq}
Mostre que, dado $y,z\in \s_+$, temos
\[d_{geo}(y,z)\leq d_{geo}(y,\sigma(z)).\]
[ {\em Sugest\~ao:} gra\c cas \`a Eq. (\ref{eq:metrictransform}), a desigualdade \'e verdadeira para a dist\^ancia euclideana se e somente se ele vale para a dist\^ancia geod\'esica, e no \'ultimo caso basta mostrar o resultado para $y = e_1$, o primeiro vetor da base de coordenadas padr\~ao. ]
\end{exercicio}

Para um subconjunto $A\subseteq\s^n$, denotemos
\[A_{\pm/0}=A\cap \s_{\pm/0}.\]

A {\em simetriza\c c\~ao de $A$ com dois pontos} \'e o conjunto
\begin{eqnarray*}
A^{\dag} &=& A_+\cup A_0\cup \sigma(A_-\setminus \sigma(A_+))\cup \left(A_-\setminus \sigma(A_+)\right)
\\
&=& A_+\cup A_0\cup \sigma(A_-)\cup \left(A_-\setminus \sigma(A_+)\right).\end{eqnarray*}
\index{simetriza\c c\~ao! por dois pontos}

Em outras palavras, os pontos de $A_+$ sempre ficam onde est\~ao, mas os pontos de $A_-$ s\~ao ``empurrados'' na dire\c c\~ao de $A_+$. Um ponto $y\in A_-$ \'e permitido ficar onde ele est\'a s\'o no caso onde o ponto $\sigma(y)$ j\'a est\'a ocupado por um elemento de $A_+$. Se $\sigma(y)\notin A_+$, ent\~ao $y$ \'e retirado de $A_-$ e $\sigma(y)$ \'e adicionado a $A_+$. A express\~ao ``simetriza\c c\~ao por dois pontos'' n\~ao foi muito bem escolhida. \'E um ``deslocamento'' na dire\c c\~ao $\vec x$.

\'E imediato que a simetriza\c c\~ao de um conjunto boreliano \'e boreliano, e que
\[\mu(A^{\dag}) = \mu(A).\]
Ademais, temos, usando exerc\'\i cio \ref{ex:ineq}:

\begin{exercicio} Seja $A$ um subconjunto boreliano da esfera, e seja $\e>0$. Ent\~ao
\[\left(A^{\dag}\right)_{\e}\subseteq \left(A_{\e}\right)^{\dag}.\]
(A inclus\~ao n\~ao \'e necessariamente estrita).
\tri
\end{exercicio}

Eis um corol\'ario importante:

\begin{equation}
\mu\left((A^{\dag})_{\e}\right)\leq \mu\left(\left(A_{\e}\right)^{\dag}\right)=  \mu\left(A_{\e}\right).\end{equation}

\subsection{A desigualdade isoperim\'etrica de L\'evy}

\begin{teorema}[Paul L\'evy, 1922]
Seja $A$ um subconjunto boreliano da esfera $\s^n$ e seja $C$ uma calota esf\'erica em $\s^n$ que tem a mesma medida de Haar:
\[\nu(C)=\nu(A).\]
Ent\~ao temos para todo $\e>0$
\begin{equation}
\nu(C_\e)\leq \nu(A_\e).
\end{equation} 
\label{t:levy}
\index{desigualdade! isoperim\'etrica! de L\'evy}
\end{teorema}

\begin{proof}
Primeiramente, vamos mostrar que em vez de um conjunto boreliano $A$, basta mostrar o resultado para um conjunto fechado, nomeadamente, para a ader\^encia $\bar A$.

\begin{exercicio}
Seja $A$ um subconjunto de um espa\c co m\'etrico $X$. Mostre que para todo $\e>0$,
\[A_\e = (\overline A)_\e.\]
Por conseguinte, para toda medida de probabilidade boreliana $\mu$ sobre $X$, temos
\[\mu(A_\e) = \mu((\overline A)_\e).\]
\tri
\label{ex:fechado}
\end{exercicio}

Como $\mu(\bar A)\geq \mu(A)$, assumindo a conclus\~ao do teorema \'e verdadeira para uma calota $C^\prime$ da medida $\mu(\bar A)$, conclu\'\i mos
\[\mu(C_\e)\leq \mu(C^\prime_\e)\leq \mu((\bar A)_\e) =\mu(A_\e),\]
como desejado.

Denotemos por $\mathcal G$, como no exerc\'\i cio \ref{ex:bB}, a fam\'\i lia de todos subconjuntos fechados de $\s^n$ tendo as propriedades
\begin{enumerate}
\item $\mu(B)=\mu(A)$,
\item para todo $\e>0$, $\mu(B_\e)\leq \mu(A_\e)$.
\end{enumerate}
Este $\mathcal G$ \'e um subconjunto fechado e n\~ao vazio de ${\mathfrak F}(\s^n)$. Escolha uma calota esf\'erica $C$ em torno de um elemento $x_0\in \s^n$ tal que
\[\mu(C)=\mu(A).\]
A nossa tarefa \'e de mostrar que $C\in {\mathcal G}$. Com ajuda do exerc\'\i cio \ref{ex:maximo} assim como uma modifica\c c\~ao \'obvia do exerc\'\i cio \ref{ex:continua}, a fun\c c\~ao $B\mapsto \mu(B\cap C)$ tem o m\'aximo sobre $\mathcal G$, digamos atingido no ponto $B^\prime\in {\mathcal G}$.
Como $B^\prime$ \'e fechado, basta mostrar que 
\[\overset{o}{C}\subseteq B^\prime,\]
onde $\overset{o}{C}$ denota o interior de $C$ na esfera $\s^n$.

Suponha o contr\'ario: $\overset{o}{C}\subsetneqq B$. Ent\~ao, 
$\mu(\overset{o}{C}\setminus B)=\mu(B\setminus \overset{o}{C})>0$. O primeiro conjunto sendo aberto, existem $z\in \overset{o}{C}\setminus B$ e $r>0$ tais que $U_r(z)\subseteq \overset{o}{C}\setminus B$. Aqui $U_r(z)$ denota uma calota esf\'erica aberta de raio $r$ em torno de $z$. 
Como o conjunto $B\setminus \overset{o}{C}$ \'e compacto, ele pode ser coberto com um n\'umero finito de bolas abertas $U_r(y)$, $y\in B\setminus C$. Pelo menos uma delas satisfaz $\mu(U_r(y)\cap B)>0$. Escolha um tal elemento $y$. 

Definamos
\[\vec x = \frac{z-y}{\norm{z-y}}.\]
Como $\norm{x_0-z}<r_C<\norm{x_0-y}$, um c\'alculo simples mostra que 
\[\langle x_0, z-y\rangle = \langle x_0,z\rangle - \langle x_0,y\rangle >0,\]
e por conseguinte $x_0\in S_+(\vec x)$. 

Como $\sigma_{\vec x}(y)=z$ e a involu\c c\~ao $\sigma_{\vec x}$ \'e isom\'etrica, a imagem de $U_r(y)$ por $\sigma_{\vec x}$ \'e igual a $U_r(z)$. Este \'ultimo conjunto n\~ao contem pontos de $B$, logo a simetriza\c c\~ao $B\mapsto B^\dag$ na dire\c c\~ao $\vec x$ vai transferir todos os pontos do conjunto $U_r(y)\cap B$ dentro a calota $U_r(z)$. 

Se $a\in C_-$, ent\~ao $\sigma(a)\in S_+$ e por conseguinte $d(x_0,\sigma(a))\leq d(x_0,a)\leq r(C)$ (exerc\'\i cio \ref{ex:ineq}), logo $\sigma(a)\in C_+$. Conclu\'\i mos: $\sigma(C_-)\subseteq C_+$.
Isso significa que 
para todo $x\in B\cap\overset{o}{C}$, a imagem de $x$ pela simetriza\c c\~ao ou muda para
$B^\dag \cap\overset{o}{C}$, ou fica onde ele est\'a. Conclu\'\i mos:
$\mu(B^\dag\cap C)\geq \mu(B\cap C)+\mu(U_r(z)\cap B)>\mu(B\cap C)$, e no mesmo tempo $B^\dag\in {\mathcal G}$, contradizendo a escolha de $B$.
\end{proof}

\subsection{Concentra\c c\~ao de medida na esfera euclideana}

\begin{teorema}
Seja $\s^{n+1}$ a esfera euclideana munida da dist\^ancia geod\'esica e a medida de Haar. A fun\c c\~ao de concentra\c c\~ao dessa esfera satisfaz
\begin{equation}
\alpha(\s^{n+1},\e)\leq \sqrt{\frac\pi 8} e^{-n\e^2/2}.
\label{eq:alfaesfera}
\end{equation}
\label{t:alfaesfera}
\end{teorema}

\begin{proof}
\'E uma consequ\^encia imediata da desigualdade isoperim\'etrica de L\'evy (teorema \ref{t:levy}), combinada com o lema \ref{l:fracpi8}.
\end{proof}

\begin{figure}[htp]
\centerline{\includegraphics[width=6cm]{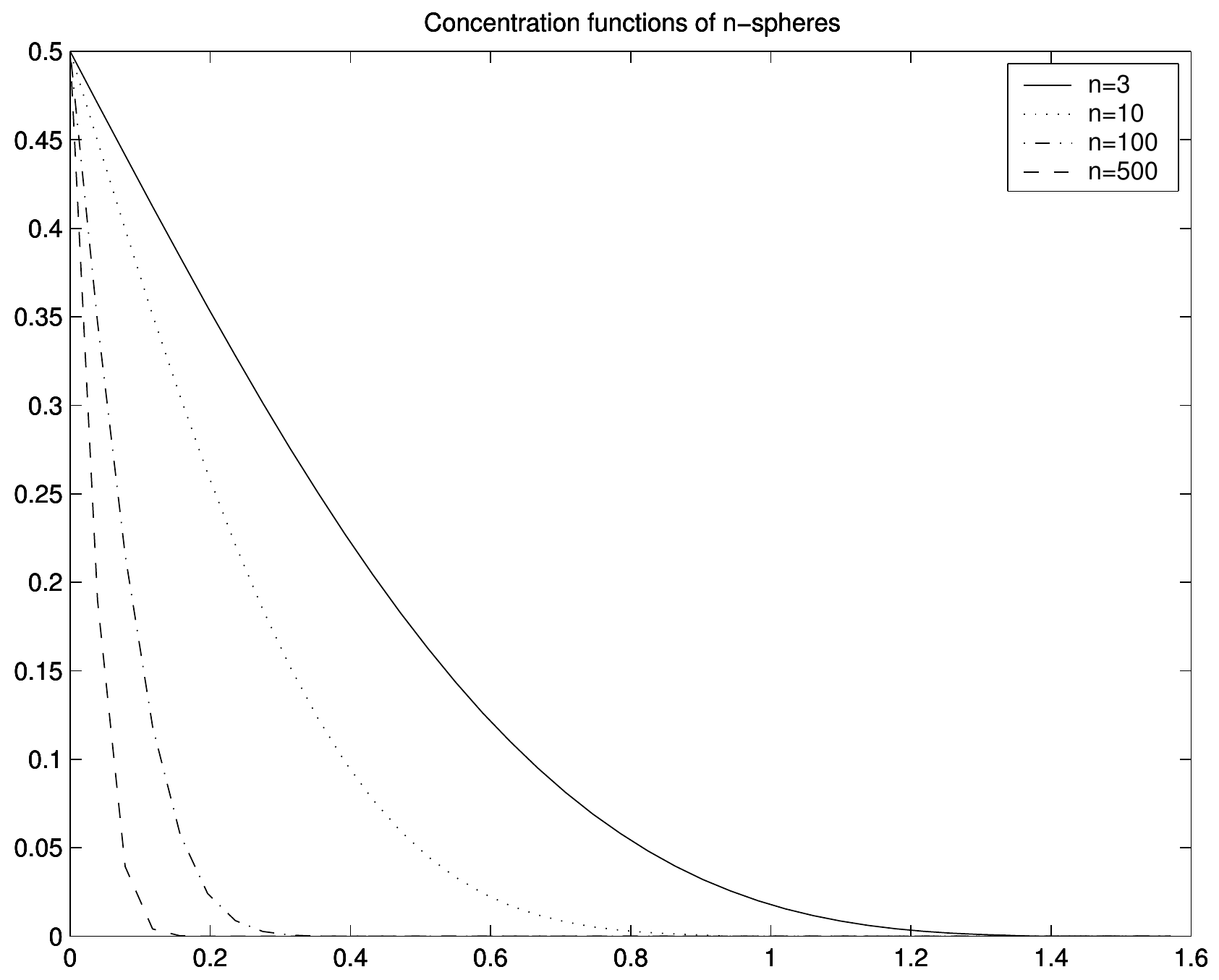}}
\caption{Fun\c c\~oes de concentra\c c\~ao de esferas de dimens\~oes $n=3,10,100,500$.}
\label{fig:sph500}
\end{figure}

\begin{corolario}
Seja 
\[f\colon \s^{n+1}\to\R\]
uma fun\c c\~ao $1$-Lipschitz cont\'\i nua (relativo \`a dist\^ancia geod\'esica). Para todo $\e>0$, temos
\[\mu\{x\in\s^{n+1}\colon \abs{f(x)- M_f}>\e\}<\sqrt{\frac\pi 2} e^{-n\e^2/2}.\]
\qed
\end{corolario}

\begin{figure}[htp]
\centerline{\includegraphics[width=6cm]{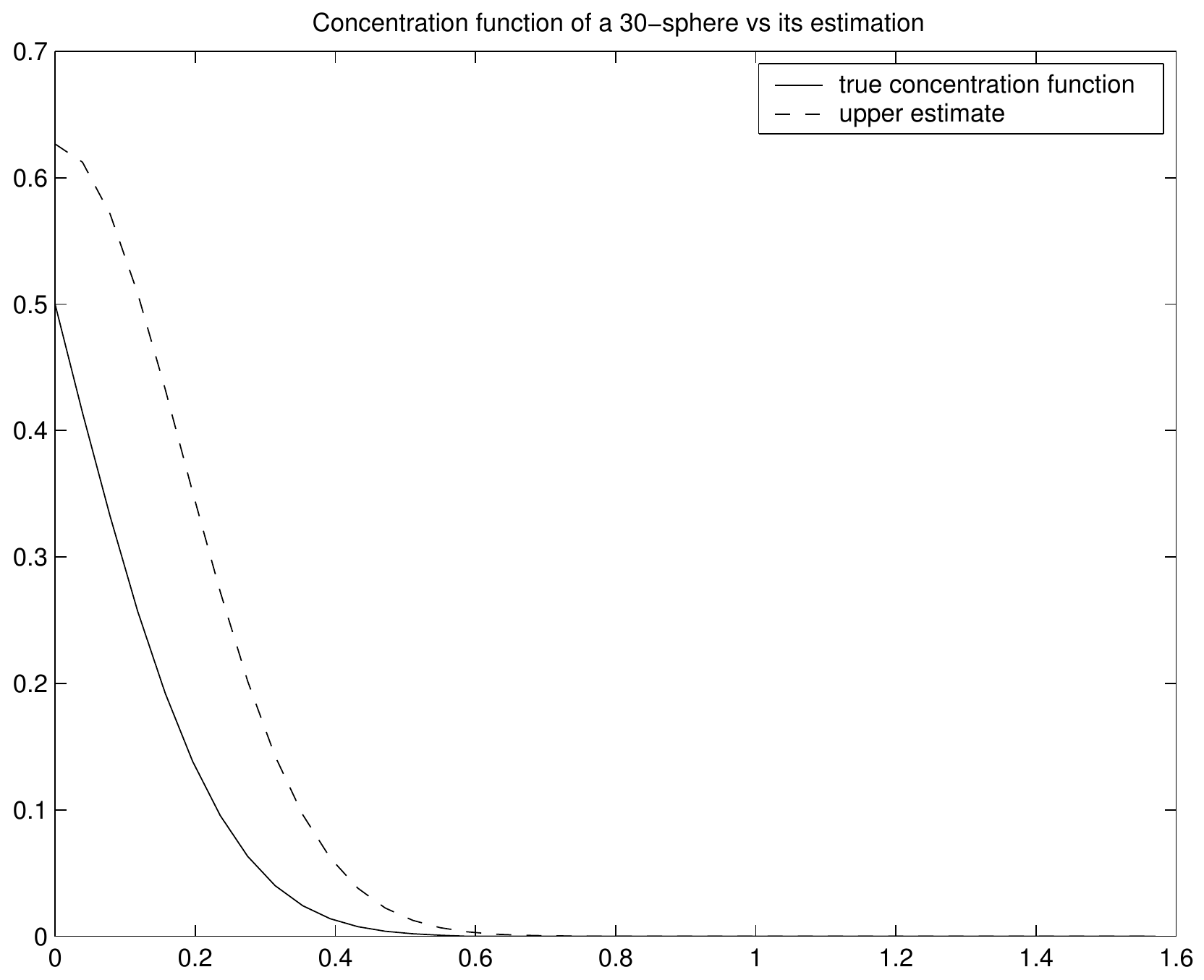}}
\caption{A fun\c c\~ao de concentra\c c\~ao da esfera de dimens\~ao $30$ contra a sua cota superior gaussiana.}
\label{fig:sph30}
\end{figure}

\begin{observacao}
Como $d_{eucl}\leq d_{geo}$, a fun\c c\~ao de concentra\c c\~ao da esfera  $\s^{n+1}$ munida da dist\^ancia euclideana admite a mesma cota superior (\ref{eq:alfaesfera}). \tri
\end{observacao}

Na se\c c\~ao \ref{s:ljl} vamos trabalhar com a esfera $\s^{n-1}$ do espa\c co euclideano $\ell^2(n)$, e por isso \'e mais c\^omodo extrair do teorema \ref{t:alfaesfera} o seguinte.

\begin{corolario}
Relativo \`a dist\^ancia euclideana,
\[\alpha(\s^{n-1},\e) \leq 5 \exp(-\e^2 n/2),
\]
e na regi\~ao $\e\leq 1$, temos uma estimativa melhor:
\[\alpha(\s^{n-1},\e) \leq 2 \exp(-\e^2 n/2).
\]
\end{corolario}

\begin{proof}
\begin{eqnarray*}
\alpha_{n-1}(\e) &\leq& \sqrt{\frac{\pi}8}\exp(-\e^2 (n-2)/2) \\
&=& \sqrt{\frac{\pi}8}\cdot \exp(\e^2)\exp(-\e^2 n/2) \\
&<& 5 \exp(-\e^2 n/2),
\end{eqnarray*}
porque se $\e\leq \sqrt 2$, ent\~ao
\[\sqrt{\frac{\pi}8}\cdot \exp(\e^2) \leq \sqrt{\frac{\pi}8}\cdot e^2 \approx 4.630404 <5,\] 
e se $\e\geq \sqrt 2$, ent\~ao a desigualdade \'e verdadeira pelas raz\~oes triviais:
 $\alpha_{n-1}(\e)=0$. (A dist\^ancia \'e euclideana!)
O caso $\e\leq 1$ \'e parecido.
\end{proof}

\subsection{Estabilidade por oscila\c c\~oes}

\begin{lema}
\label{l:A}
Para todo subconjunto boreliano, $A$, da esfera $\s^{n-1}$ e $\xi\in\s^{n-1}$ qualquer, temos
\begin{equation}
\label{eq:A}
\mu(A) = \nu\{u\in O(n)\colon u\xi\in A\},\end{equation}
onde $\nu$ \'e a medida de Haar sobre o grupo ortogonal $O(n)$.
\end{lema}

\begin{proof}
A express\~ao no lado direito da f\'ormula (\ref{eq:A}) defina uma medida de probabilidade sobre $\s^{n-1}$, digamos $\tilde\mu$, nomeadamente, a imagem direita de $\nu$ pela aplica\c c\~ao de \'orbita. Essa medida \'e invariante por isometrias:
\begin{eqnarray*}
\tilde\mu(vA) &=& 
\nu\{u\in O(n)\colon u\xi\in vA\}\\ &=&
\nu\{u\in O(n)\colon v^{-1}u\xi\in A\}\\
&=& \nu\left( v^{-1}\cdot \{u\in O(n)\colon u\xi\in vA\}\right) \\
&=& \nu\{u\in O(n)\colon u\xi\in vA\}\\
&=& \tilde\mu(A).
\end{eqnarray*}
Por causa da unicidade da medida de Haar sobre a esfera, $\tilde\mu$ deve ser igual a $\mu$.
\end{proof}

O resultado seguinte \'e uma ferramenta importante da an\'alise funcional geom\'etrica.

\begin{lema}
\label{l:tool}
Seja $f\colon\s^{n-1}\to\R$ uma fun\c c\~ao real $1$-Lipschitz cont\'\i nua sobre a esfera $\s^{n-1}$, e seja $0<\e\leq 1$. Qual quer seja um subconjunto finito
$F\subseteq\s^{n-1}$ da cardinalidade
\[\abs F < \frac 14 \exp(\e^2 n /2),\]
existe uma transforma\c c\~ao ortogonal $u\in O(n)$ tal que a fun\c c\~ao $f$ tem pequena oscila\c c\~ao sobre a transla\c c\~ao $uF$ de $F$:
\[\forall y\in uF,~~\abs{f(y)-M_f}<\e.\]
\index{estabilidade por oscila\c c\~oes}
\end{lema}

\begin{proof}
Para todo $x\in F$, definamos
\[U_x = \{u\in O(n)\colon \abs{f(u x)-M_f}<\e\}.\]
Como 
\[\mu\{x\in\s^{n-1}\colon \abs{f(x)-M_f}<\e\}\geq 1-2\alpha_{\s^{n-1}}(\e) > 1 - 4\exp(-\e^2 n/2),\]
o lema \ref{l:A} implica que
\[\nu(U_x)\geq 1 - 4\exp(-\e^2 n/2).\]
Logo, a interse\c c\~ao de todos os conjuntos $U_x$, $x\in F$ n\~ao \'e vazia por causa da nossa hip\'otese sobre a cardinalidade de $F$. Escolha 
\[u\in\bigcap_{x\in F} U_x.\]
Esse $u$ tem a propriedade desejada.
\end{proof}

Reformulamos o lema. Para uma fun\c c\~ao $f\colon \s^{n-1}\to\R$ e uma transforma\c c\~ao ortogonal $u\in O(n)$ defina a fun\c c\~ao $^uf$, como antes, por
\[^uf(x) = f(u^{-1}x).\]
\'E \'obvio que se $f$ for $1$-Lipschitz cont\'\i nua, ent\~ao $^uf$ \'e tal, e al\'em disso,
\[M_{^uf} = M_f.\]
Eis uma forma equivalente do lema \ref{l:tool}.

\begin{lema} 
\label{l:tool2}
Seja $f\colon\s^{n-1}\to\R$ uma fun\c c\~ao real $1$-Lipschitz cont\'\i nua sobre a esfera $\s^{n-1}$, e seja $\e>0$. Qual quer seja um subconjunto finito
$F\subseteq\s^{n-1}$ com
\begin{equation}
\label{eq:card}
\abs F < \frac 14 \exp(\e^2 n /2),\end{equation}
existe uma transforma\c c\~ao ortogonal $u\in O(n)$ tendo a propriedade que para todo $x\in F$,
\[\abs{^uf(x) - M_{^uf}}<\e.\]
\qed
\end{lema}

\subsection{Coment\'arios hist\'oricos e sugest\~oes de leitura}

Tem provavelmente uma d\'uzia de v\'arias provas do teorema \ref{t:alfaesfera} sobre a concentra\c c\~ao de medida sobre as esferas. Historicamente, a primeira prova foi devida a de Paul L\'evy \citep*{levy}, mesmo se ela s\'o tornou-se matematicamente exata no trabalho de Gromov \citep*{GrApp}. Em particular, o lema \ref{l:fracpi8} foi diretamente emprestado do livro de L\'evy ({\em loc. cit.}), veja tamb\'em \citep*{MS}, pp. 5-6.

A prova com simetriza\c c\~ao, provavelmente a mais simples de todas, foi descoberto por Figiel, Lindenstrauss, e Milman \citep*{FLM}. Ela foi simplificada por Benyamini \citep*{benyamini}. A nossa presenta\c c\~ao segue a de Schechtman \citep*{schechtman}, mas \'e simplificada mesmo mais, tendo eliminado a necessidade de usar o teorema de densidade de Lebesgue. Desta forma, o argumento torna-se elementar.

Para outras provas da desigualdade de L\'evy (ou seja, a cota superior gaussiana no teorema \ref{t:alfaesfera}), usando os v\'arios m\'etodos da an\'alise e geometria, veja por exemplo \citep*{GrApp}, \citep*{L}, \citep*{matousekBook}, \citep*{pisier}, \citep*{BL}.

O lema \ref{l:tool}, descoberto e explorado por Vitali Milman nos anos 1960 \citep*{M67}, est\'a ligando a an\'alise funcional geom\'etrica, teoria combinat\'oria, e teoria de grupos de transfor\-ma\-\c c\~oes. Esta liga\c c\~ao \'e o assunto principal do livro \citep*{Pes06}.

%
%

\chapter{Teorema Minimax de von Neumann\label{a:minimax}}

Pensemos de uma medida de probabilidade sobre o espa\c co boreliano finito $[n]$ como um $n$-vetor, $p$, com coordenadas (pesos) n\~ao negativas, $p_i$, cuja soma \'e igual a $1$.  

\begin{teorema}[Teorema Minimax ``concreto'' de von Neumann (1928)]
\label{t:minimax}
Seja $M$ uma matriz com coeficientes reais, de formato $m\times n$. Ent\~ao,
\[\max_{p\in P[m]}\min_{q\in P[n]}p^t M q=\min_{q\in P[n]}\max_{p\in P[m]} p^t Mq.\]
\index{teorema! Minimax}
\end{teorema}

Primeiramente, notemos que a desigualdade $\leq$ n\~ao \'e espec\'\i fica ao teorema e seque do resultado geral seguinte.

\begin{teorema}[Desigualdade min-max]
Seja $f\colon X\times Y\to \R$ uma fun\c c\~ao qualquer. Ent\~ao
\[\sup_{x\in X}\inf_{y\in Y} f(x,y) \leq \inf_{y\in Y}\sup_{x\in X} f(x,y).\]
\index{desigualdade! min-max}
\end{teorema}

\begin{proof}
A desigualdade simplesmente afirma que cada n\'umero real (ou s\'\i mbolo $\pm\infty$) da forma $\inf_{y\in Y} f(x,y)$, $x\in X$, \'e menor ou igual a cada elemento de $\R\cup\{\pm\infty\}$ da forma $\sup_{x\in X} f(x,y)$, $y\in Y$. Sejam $x\in X$, $y\in Y$ quaisquer. Temos:
\[\inf_{y^\prime\in Y} f(x,y^\prime)\leq f(x,y)\leq \sup_{x^\prime\in X} f(x^\prime,y),\]
estabelecendo a afirma\c c\~ao.
\end{proof}

Resta-nos apenas verificar a desigualdade $\geq$.
Note que o conjunto $P[n]$ pode se identificar com a envolt\'oria convexa do conjunto dos vetores de base, $e_1,e_2,\ldots,e_n$ (o {\em simplex} padr\~ao), denotado $\Delta^n$.

\begin{exercicio}
Seja $f$ uma fun\c c\~ao cont\'\i nua e {\em convexa} sobre $\Delta^n$, ou seja, quaisquer que sejam $x,y\in\Delta^n$ e $t,s\in [0,1]$,
\[f(tx+sy)\leq tf(x)+sf(y).\]
Mostre que o m\'aximo de $f$ sobre $\Delta^n$ \'e necessariamente atingido em um v\'ertice $e_i$, $i=1,2,\ldots,n$. Do mesmo jeito, o m\'\i nimo de uma fun\c c\~ao {\em c\^oncava} \'e atingido em um v\'ertice.
\end{exercicio}

Segue-se do exerc\'\i cio que o m\'\i nimo $\min_{q\in P[n]}p^t M q$ da fun\c c\~ao linear $q\mapsto p^t M q$ sobre $P(m)$ seja atingido sobre um vetor de base:
\begin{align*}
\min_{q\in P[n]}p^t M q &= \min_{j\in [m]} p^t M e_j\\
&=  \min_{j\in [m]} p^t M_{\ast,j},
\end{align*}
onde $M_{\ast,j}$ significa o $j$-\'esima coluna de $M$. 

Similarmente, o {\em m\'aximo} $\max_{p\in P[m]} p^t Mq$ da fun\c c\~ao linear (logo c\^oncava) $p\mapsto p^t Mq$ \'e atingido sobre um vetor de base tamb\'em:
\begin{align*}
\max_{p\in P[m]} p^t Mq &=
\max_{i\in [m]} e_i^t Mq \\
&= \max_{i\in [m]} M_{i,\ast} q,
\end{align*}
onde $M_{i,\ast}$ \'e a $i$-\'esima linha de $M$.

Agora o teorema de Minimax ``concreto'' pode ser reformulado assim: 
\begin{equation}
\max_{p\in P[m]}\min_{j\in [n]}p^t M_{\ast,j}=\min_{q\in P[n]}\max_{i\in [m]} M_{i,\ast} q.
\label{eq:minimaxconcretapure}
\end{equation}
(Note que \'e exatamente esta forma que usamos na prova do teorema de Moran--Yehudayoff \ref{t:moran-yehudayoff}).

\begin{exercicio}
Sejam $C$ um subconjunto compacto e convexo do espa\c co euclideano $\ell^2(n)$, $x\notin C$. Mostre que $C$ possui o ponto $c\in C$ mais pr\'oximo de $x$:
\[\forall y\in C,~y\neq c\Rightarrow \norm{x-c} < \norm{x-y}.\]
\end{exercicio}

\begin{lema}[Alternativa para matrizes]
Seja $M$ uma matriz com coeficientes reais, de formato $m\times n$. Ent\~ao, vale uma e apenas uma das duas condi\c c\~oes seguintes:
\begin{enumerate}
\item \label{alternativa1} O vetor $0\in\R^m$ pertence \`a envolt\'oria convexa de $n$ colunas de $M$ e $m$ vetores de base de $\R^m$.
\item \label{alternativa2} Existe uma medida de probabilidade $p\in P[m]$ com todos pesos estritamente positivos tal que $\langle p, A_{\ast,j}\rangle >0$ para todos $j=1,2,\ldots,n$.
\end{enumerate}
\label{l:alternativa}
\end{lema}

\begin{proof}
Denote
\[C = \mbox{conv}\,(M_{\ast,1},\ldots,M_{\ast,n},e_1,\ldots,e_m)\]
e suponha que $0\notin C$. Seja $c\in C$ o ponto mais pr\'oximo de $x$ em $C$. O funcional linear $\langle -,c\rangle$ toma o valor $c^2>0$ em $c$, e para cada $y\in C$, $y\neq c$, temos $\langle c,y\rangle > c^2$. (Caso contr\'ario, um subintervalo n\~ao trivial do intervalo $[c,y]=\{tc+(1-t)y\colon t\in [0,1]\}\subseteq C$ pertence \`a esfera $S_{c^2}(0)$ pelas raz\~oes puramente planim\'etricas, o que seria uma contradi\c c\~ao com a escolha de $c$). Em particular, para todo $y\in C$, $\langle y,c\rangle >0$. Substituindo $y=e_i\in C$, conclu\'\i mos que $c_i >0$ para todos $i$. Agora basta normalizar $c$, definindo a medida de probabilidade $p=c/\sum_i c_i$.
\end{proof}

\begin{observacao}
O funcional $\langle c,-\rangle$ que aparece na prova acima fazendo uma separa\c c\~ao entre $0$ e $C$ \'e um caso particular da vers\~ao seguinte (n\~ao imediatamente \'obvia) do teorema de Hahn--Banach \ref{t:hb}: seja $C$ um subconjunto fechado e convexo de um espa\c co de Banach $E$, e seja $x\notin C$. Ent\~ao existe um funcional linear limitado $\phi\in E^\ast$ tal que
\[\phi(x)>\sup_{c\in C}\phi(c).\]
\end{observacao}

Suponha que a possibilidade (\ref{alternativa1}) ocorre no lema \ref{l:alternativa}: existem coeficientes $\lambda_j\geq 0$, $j=1,\ldots,m+n$, tais que $\sum_{j=1}^{m+n}\lambda_j=1$ e
\[0=\sum_{j=1}^n \lambda_jA_{\ast,j}+\sum_{k=1}^m \lambda_{n+k}e_k.\]
\'E claro que $\sum_{j=1}^n \lambda_j>0$. Por conseguinte, 
\[q_j=\left(\sum_{k=1}^n \lambda_k\right)^{-1}\lambda_j,~j\in [n]\]
define uma medida de probabilidade, $q$, sobre $[n]$. Qualquer que seja $i\in [m]$, temos
\begin{align*}
\langle A_{i,\ast},q\rangle &= - \left(\sum_{k=1}^n \lambda_k\right)^{-1} \sum_{k=1}^m \lambda_{n+k}e_k \\
&\leq 0.
\end{align*}
Conclu\'\i mos: 
\[\min_{q\in P[n]}\max_{i\in [m]} M_{i,\ast} q\leq 0.\]

Agora suponha que a possibilidade (\ref{alternativa2}) ocorre. Neste caso, existe $p\in P[m]$ tal que temos para todos $j\in [n]$
\[\langle p, A_{\ast,j}\rangle >0.\]
Por conseguinte,
\[\max_{p\in P[m]}\min_{j\in [n]}p^t M_{\ast,j}>0.\]
Denotemos para simplicidade
\begin{align*}
v_{(I)} &= \max_{p\in P[m]}\min_{j\in [n]}p^t M_{\ast,j} = \max_{p\in P[m]}\min_{q\in P[n]}p^t M q,\\
v_{(II)} &= \min_{q\in P[n]}\max_{i\in [m]} M_{i,\ast} q=\min_{q\in P[n]}\max_{p\in P[m]} p^t Mq.
\end{align*}
O nosso argumento acima exclui a possibilidade do que
\[v_{(I)}  \leq 0 < v_{(II)}.\]
Denotemos $\mathbf{J}=\mathbf{J}_{m\times n}$ uma matriz com todos coeficientes iguais a $1$, e seja $t$ um real qualquer. Formemos uma matriz do mesmo tamanho $m\times n$,
\[B = M+t\mathbf{J}.\]
\begin{exercicio}
Verifique que, quaisquer que sejam $p\in P[m]$, $q\in P[n]$,
\[p^t \mathbf{J} q = 1.\]
Por conseguinte,
\[p^t B q = p^t M q + t.\]
\end{exercicio}
N\'os impossibilitamos o caso onde $v_{(I)} +t \leq 0 < v_{(II)}+t$,
ou seja,
\[v_{(I)}  \leq -t < v_{(II)},\]
qualquer que seja um n\'umero real $t$. Conclu\'\i mos:  $v_{(I)}  \geq v_{(II)}$.

A apresenta\c c\~ao foi inspirada pela nota \citep*{scarvalone}.

\backmatter

\bibliographystyle{plainnat}
\bibliography{etaas_biblio}

\begin{thebibliography}{137}
\providecommand{\natexlab}[1]{#1}
\providecommand{\url}[1]{\texttt{#1}}
\expandafter\ifx\csname urlstyle\endcsname\relax
  \providecommand{\doi}[1]{doi: #1}\else
  \providecommand{\doi}{doi: \begingroup \urlstyle{rm}\Url}\fi

\bibitem[Alon and Spencer(1992)]{alon_spencer}
Noga Alon and Joel~H. Spencer.
\newblock \emph{The probabilistic method}.
\newblock Wiley-Interscience Series in Discrete Mathematics and Optimization.
  John Wiley \& Sons, Inc., New York, 1992.

\bibitem[Alon et~al.(1997)Alon, Ben-David, Cesa-Bianchi, and Haussler]{ABDCBH}
Noga Alon, Shai Ben-David, Nicol\`o Cesa-Bianchi, and David Haussler.
\newblock Scale-sensitive dimensions, uniform convergence, and learnability.
\newblock \emph{J. ACM}, 44\penalty0 (4):\penalty0 615--631, 1997.

\bibitem[Anthony and Bartlett(1999)]{AB}
Martin Anthony and Peter~L. Bartlett.
\newblock \emph{Neural network learning: theoretical foundations}.
\newblock Cambridge University Press, Cambridge, 1999.

\bibitem[Artstein-Avidan et~al.(2015)Artstein-Avidan, Giannopoulos, and
  Milman]{AAGM}
Shiri Artstein-Avidan, Apostolos Giannopoulos, and Vitali~D. Milman.
\newblock \emph{Asymptotic geometric analysis. {P}art {I}}, volume 202 of
  \emph{Mathematical Surveys and Monographs}.
\newblock American Mathematical Society, Providence, RI, 2015.

\bibitem[Assouad and Quentin~de Gromard(2006)]{assouad_gromard}
Patrice Assouad and Thierry Quentin~de Gromard.
\newblock Recouvrements, derivation des mesures et dimensions.
\newblock \emph{Rev. Mat. Iberoamericana}, 22\penalty0 (3):\penalty0 893--953,
  2006.

\bibitem[Banach and Kuratowski(1929)]{BK}
Stefan Banach and Kazimierz Kuratowski.
\newblock Sur une g\'en\'eralisation du probleme de la mesure.
\newblock \emph{Fundamenta Mathematicae}, 14:\penalty0 127--131, 1929.

\bibitem[Barkol and Rabani(2002)]{BR}
Omer Barkol and Yuval Rabani.
\newblock Tighter lower bounds for nearest neighbor search and related problems
  in the cell probe model.
\newblock \emph{J. Comput. Syst. Sci.}, 64\penalty0 (4):\penalty0 873--896,
  2002.

\bibitem[Bartlett and Mendelson(2002)]{bartlett_mendelson}
Peter~L. Bartlett and Shahar Mendelson.
\newblock Rademacher and gaussian complexities: Risk bounds and structural
  results.
\newblock \emph{Journal of Machine Learning Research}, 3:\penalty0 463--482,
  2002.

\bibitem[Ben-David and Litman(1998)]{B-DL}
Shai Ben-David and Ami Litman.
\newblock Combinatorial variability of {V}apnik-{C}hervonenkis classes with
  applications to sample compression schemes.
\newblock \emph{Discrete Applied Mathematics}, 86\penalty0 (1):\penalty0 3--25,
  1998.

\bibitem[Benedek and Itai(1991)]{BI}
Gyora~M. Benedek and Alon Itai.
\newblock Learnability with respect to fixed distributions.
\newblock \emph{Theor. Comput. Sci.}, 86\penalty0 (2):\penalty0 377--390, 1991.

\bibitem[Benyamini(1984)]{benyamini}
Yoav Benyamini.
\newblock Two-point symmetrization, the isoperimetric inequality on the sphere
  and some applications.
\newblock In \emph{Texas functional analysis seminar 1983--1984 ({A}ustin,
  {T}ex.)}, Longhorn Notes, pages 53--76. Univ. Texas Press, Austin, TX, 1984.

\bibitem[Benyamini and Lindenstrauss(2000)]{BL}
Yoav Benyamini and Joram Lindenstrauss.
\newblock \emph{Geometric nonlinear functional analysis. {V}ol. 1}, volume~48
  of \emph{American Mathematical Society Colloquium Publications}.
\newblock American Mathematical Society, Providence, RI, 2000.

\bibitem[Beyer et~al.(1999)Beyer, Goldstein, Ramakrishnan, and Shaft]{BGRS}
Kevin~S. Beyer, Jonathan Goldstein, Raghu Ramakrishnan, and Uri Shaft.
\newblock When is ''nearest neighbor'' meaningful?
\newblock In \emph{Database Theory - {ICDT} '99, 7th International Conference,
  Jerusalem, Israel, January 10-12, 1999, Proceedings.}, pages 217--235, 1999.

\bibitem[Biau et~al.(2008)Biau, Devroye, and Lugosi]{BDL}
G\'erard Biau, Luc Devroye, and G\'abor Lugosi.
\newblock Consistency of random forests and other averaging classifiers.
\newblock \emph{Journal of Machine Learning Research}, 9:\penalty0 2015--2033,
  2008.

\bibitem[Blum et~al.(2016)Blum, Hopcroft, and Kannan]{BHK}
Avrim Blum, John Hopcroft, and Ravindran Kannan.
\newblock
  \emph{\href{https://www.cs.cornell.edu/jeh/book2016June9.pdf}{Foundations of
  Data Science}}.
\newblock 2016.
\newblock Book in preparation.

\bibitem[Blumer et~al.(1989)Blumer, Ehrenfeucht, Haussler, and Warmuth]{BEHW}
Anselm Blumer, Andrzej Ehrenfeucht, David Haussler, and Manfred~K. Warmuth.
\newblock Learnability and the {V}apnik-{C}hervonenkis dimension.
\newblock \emph{J. ACM}, 36\penalty0 (4):\penalty0 929--965, 1989.

\bibitem[Bollob\'as and Radcliffe(1995)]{BoR}
B\'ela Bollob\'as and A.~J. Radcliffe.
\newblock Defect {S}auer results.
\newblock \emph{J. Comb. Theory, Ser. A}, 72\penalty0 (2):\penalty0 189--208,
  1995.

\bibitem[Bourbaki(1974)]{bourbaki}
Nicolas Bourbaki.
\newblock \emph{\'El\'ements de math\'ematique. Topologie g\'en\'erale,
  Chapitres 5 \`a 10, Deuxi\`eme edition}.
\newblock Hermann, Paris, 1974.

\bibitem[Braun and Griebel(2009)]{braun-griebel}
J\"{u}rgen Braun and Michael Griebel.
\newblock On a constructive proof of {K}olmogorov's superposition theorem.
\newblock \emph{Constr. Approx.}, 30\penalty0 (3):\penalty0 653--675, 2009.

\bibitem[Bro~Miltersen(2000)]{miltersen}
Peter Bro~Miltersen.
\newblock Cell probe complexity - a survey.
\newblock In \emph{Proceedings of the 19th Conference on the Foundations of
  Software Technology and Theoretical Computer Science, 1999. Advances in Data
  Structures Workshop}, April 2000.

\bibitem[Brosowski and Deutsch(1981)]{BroD}
Bruno Brosowski and Frank Deutsch.
\newblock An elementary proof of the {S}tone-{W}eierstrass theorem.
\newblock \emph{Proc. Amer. Math. Soc.}, 81\penalty0 (1):\penalty0 89--92,
  1981.

\bibitem[C\'{e}rou and Guyader(2006)]{CG}
Fr\'{e}d\'{e}ric C\'{e}rou and Arnaud Guyader.
\newblock Nearest neighbor classification in infinite dimension.
\newblock \emph{ESAIM Probab. Stat.}, 10:\penalty0 340--355, 2006.

\bibitem[Chavez et~al.(2001)Chavez, Navarro, Baeza-Yates, and
  Marroquin]{chavez:01}
Edgar Chavez, Gonzalo Navarro, Ricardo~A. Baeza-Yates, and Jose~L. Marroquin.
\newblock Searching in metric spaces.
\newblock \emph{ACM Computing Surveys}, 33\penalty0 (3):\penalty0 273--321,
  2001.

\bibitem[Cover and Hart(1967)]{cover_hart}
Thomas~M. Cover and Peter~E. Hart.
\newblock Nearest neighbor pattern classification.
\newblock \emph{IEEE Transactions on Information Theory}, 13\penalty0
  (1):\penalty0 21--27, 1967.

\bibitem[Crawley(2012)]{C}
Michael~J. Crawley.
\newblock \emph{\href{https://www.xarg.org/ref/a/0470973927/}{The R Book}}.
\newblock Wiley, dec 2012.

\bibitem[Cucker and Smale(2002)]{cucker_smale}
Felipe Cucker and Steve Smale.
\newblock On the mathematical foundations of learning.
\newblock \emph{Bull. Amer. Math. Soc. (N.S.)}, 39\penalty0 (1):\penalty0
  1--49, 2002.

\bibitem[Cucker and Zhou(2007)]{cucker_zhou}
Felipe Cucker and Ding-Xuan Zhou.
\newblock \emph{Learning theory: an approximation theory viewpoint}, volume~24
  of \emph{Cambridge Monographs on Applied and Computational Mathematics}.
\newblock Cambridge University Press, Cambridge, 2007.

\bibitem[Cybenko(1989)]{cybenko}
George Cybenko.
\newblock Approximation by superpositions of a sigmoidal function.
\newblock \emph{{Mathematics of Control, Signals, and Systems}}, 2\penalty0
  (4):\penalty0 303--314, 1989.

\bibitem[Davies(1971)]{davis}
Roy~O. Davies.
\newblock Measures not approximable or not specifiable by means of balls.
\newblock \emph{Mathematika}, 18\penalty0 (02):\penalty0 157, December 1971.

\bibitem[Despres(2014)]{despres}
Christian J.~J. Despres.
\newblock The {V}apnik-{C}hervonenkis dimension of cubes in $\mathbb{R}^d$,
  2014.
\newblock Fall 2014 Honours research project, University of Ottawa, 4 pp.
  \href{https://arxiv.org/abs/1412.6612}{arXiv:1412.6612}.

\bibitem[Devroye(1981)]{devroye_1981}
Luc Devroye.
\newblock On the almost everywhere convergence of nonparametric regression
  function estimates.
\newblock \emph{Ann. Statist.}, 9\penalty0 (6):\penalty0 1310--1319, 1981.

\bibitem[Devroye and Gy\"{o}rfi(1985)]{devroye_gyorfi}
Luc Devroye and L\'{a}szl\'{o} Gy\"{o}rfi.
\newblock \emph{Nonparametric density estimation. {T}he $L_1$ view}.
\newblock Wiley Series in Probability and Mathematical Statistics: Tracts on
  Probability and Statistics. John Wiley \& Sons, Inc., New York, 1985.

\bibitem[Devroye et~al.(1994)Devroye, Gy\"{o}rfi, Krzy\.{z}ak, and
  Lugosi]{DevroyeGK}
Luc Devroye, L\'{a}szl\'{o} Gy\"{o}rfi, Adam Krzy\.{z}ak, and G\'{a}bor Lugosi.
\newblock On the strong universal consistency of nearest neighbor regression
  function estimates.
\newblock \emph{Ann. Statist.}, 22\penalty0 (3):\penalty0 1371--1385, 1994.

\bibitem[Devroye et~al.(1996)Devroye, Gy\"{o}rfi, and Lugosi]{DGL}
Luc Devroye, L\'{a}szl\'{o} Gy\"{o}rfi, and G\'{a}bor Lugosi.
\newblock \emph{A probabilistic theory of pattern recognition}, volume~31 of
  \emph{Applications of Mathematics (New York)}.
\newblock Springer-Verlag, New York, 1996.

\bibitem[Duan(2014)]{duan}
Hubert~Haoyang Duan.
\newblock Applying supervised learning algorithms and a new feature selection
  method to predict coronary artery disease, 2014.
\newblock M.Sc. thesis, University of Ottawa, 2014, 102 pp.
  \href{https://arxiv.org/abs/1402.0459}{arXiv:1402.0459 [cs.LG]}.

\bibitem[Duan et~al.()Duan, Giordano, Hatko, and Pestov]{DGHP}
Hubert~Haoyang Duan, Ga\"el Giordano, Stan Hatko, and Vladimir Pestov.
\newblock \href{http://www.csmining.org/cdmc2013/tl_files/results/319.pdf}{2013
  Cybersecurity Data Mining Competition Methods overview}.
\newblock 8 pp.

\bibitem[Dudley(2014)]{dudley}
R.~M. Dudley.
\newblock \emph{Uniform central limit theorems}, volume 142 of \emph{Cambridge
  Studies in Advanced Mathematics}.
\newblock Cambridge University Press, New York, second edition, 2014.

\bibitem[Durst and Dudley(1980)]{DD}
Mark Durst and Richard~M. Dudley.
\newblock Empirical processes, {V}apnik-{C}hervonenkis classes and {P}oisson
  processes.
\newblock \emph{Probab. Math. Statist.}, 1\penalty0 (2):\penalty0 109--115
  (1981), 1980.

\bibitem[Engelking(1989)]{engelking}
Ryszard Engelking.
\newblock \emph{General topology}, volume~6 of \emph{Sigma Series in Pure
  Mathematics}.
\newblock Heldermann Verlag, Berlin, second edition, 1989.

\bibitem[Figiel et~al.(1977)Figiel, Lindenstrauss, and Milman]{FLM}
T.~Figiel, J.~Lindenstrauss, and V.~D. Milman.
\newblock The dimension of almost spherical sections of convex bodies.
\newblock \emph{Acta Math.}, 139\penalty0 (1-2):\penalty0 53--94, 1977.

\bibitem[Floyd and Warmuth(1995)]{FW}
Sally Floyd and Manfred~K. Warmuth.
\newblock Sample compression, learnability, and the {V}apnik-{C}hervonenkis
  dimension.
\newblock \emph{Machine Learning}, 21\penalty0 (3):\penalty0 269--304, 1995.

\bibitem[Fraiman et~al.(2010)Fraiman, Justel, and Svarc]{FJS}
Ricardo Fraiman, Ana Justel, and Marcela Svarc.
\newblock Pattern recognition via projection-based {$k\rm NN$} rules.
\newblock \emph{Comput. Statist. Data Anal.}, 54\penalty0 (5):\penalty0
  1390--1403, 2010.

\bibitem[Frankl and F\"{u}redi(1981)]{FF}
P.~Frankl and Z.~F\"{u}redi.
\newblock A short proof for a theorem of {H}arper about {H}amming-spheres.
\newblock \emph{Discrete Math.}, 34\penalty0 (3):\penalty0 311--313, 1981.

\bibitem[Freiling(1986)]{freiling}
Chris Freiling.
\newblock Axioms of symmetry: throwing darts at the real number line.
\newblock \emph{J. Symbolic Logic}, 51\penalty0 (1):\penalty0 190--200, 1986.

\bibitem[Girosi and Poggio(1989)]{girosi}
Federico Girosi and Tomaso~A. Poggio.
\newblock Representation properties of networks: Kolmogorov's theorem is
  irrelevant.
\newblock \emph{Neural Computation}, 1\penalty0 (4):\penalty0 465--469, 1989.

\bibitem[Goldberg and Jerrum(1995)]{GJ}
P.W. Goldberg and M.R. Jerrum.
\newblock Bounding the {V}apnik--{C}hervonenkis dimension of concept classes
  parametrized by real numbers.
\newblock \emph{Machine Learning}, 18:\penalty0 131--148, 1995.

\bibitem[Gromov(1986)]{GrApp}
M.~Gromov.
\newblock Isoperimetric inequalities in {R}iemannian manifolds. {A}ppendix {I}
  in \cite{MS}, 1986.

\bibitem[Gromov(1998)]{gromov_trends}
Mikhael Gromov.
\newblock Possible trends in mathematics in the coming decades.
\newblock \emph{Notices Amer. Math. Soc.}, 45\penalty0 (7):\penalty0 846--847,
  1998.

\bibitem[Gromov(1999)]{Gr}
Misha Gromov.
\newblock \emph{Metric structures for {R}iemannian and non-{R}iemannian
  spaces}, volume 152 of \emph{Progress in Mathematics}.
\newblock Birkh\"{a}user Boston, Inc., Boston, MA, 1999.
\newblock Based on the 1981 French original [ MR0682063 (85e:53051)], With
  appendices by M. Katz, P. Pansu and S. Semmes, Translated from the French by
  Sean Michael Bates.

\bibitem[Guyon et~al.(2006)Guyon, Nikravesh, Gunn, and Zadeh]{guyon}
Isabelle Guyon, Masoud Nikravesh, Steve Gunn, and Lotfi~A. Zadeh, editors.
\newblock \emph{Feature Extraction, Foundations and Applications}.
\newblock Springer Berlin Heidelberg, 2006.

\bibitem[Hanika et~al.(2018)Hanika, Schneider, and Stumme]{HSS}
Tom Hanika, Friedrich~Martin Schneider, and Gerd Stumme.
\newblock Intrinsic dimension of geometric data sets, 2018.
\newblock 28 pp. \href{https://arxiv.org/abs/1801.07985}{arXiv:1801.07985}.

\bibitem[Hatko(2013)]{hatko:project}
Stan Hatko.
\newblock Borel isomorphic dimensionality reduction of data and supervised
  learning, 2013.
\newblock Honours research project, University of Ottawa,
  \href{https://arxiv.org/abs/1307.8333}{arXiv:1307.8333}, 45 pp.

\bibitem[Hatko(2015)]{hatko:thesis}
Stan Hatko.
\newblock $k$-{N}earest neighbour classification of datasets with a family of
  distances, 2015.
\newblock M.Sc. thesis, University of Ottawa,
  \href{https://arxiv.org/abs/1512.00001}{arXiv:1512.00001}.

\bibitem[Hecht-Nielsen(1987)]{hecht-nielsen}
Robert Hecht-Nielsen.
\newblock {K}olmogorov's mapping neural network existence theorem.
\newblock In \emph{{P}roceedings of the IEEE First International Conference on
  Neural Networks {\rm ({S}an {D}iego, {CA})}}, volume III, pages 11--13.
  Piscataway, NJ: IEEE, 1987.

\bibitem[Hefez and Villela(2017)]{HV}
Abramo Hefez and Maria L\'ucia~T. Villela.
\newblock \emph{C\'odigos Corretores de Erros}.
\newblock IMPA, Rio de Janeiro, 2017.
\newblock 2a edi\c c\~ao, 206 p\'aginas.

\bibitem[Hofmann and Morris(2013)]{hofmann_morris}
Karl~H. Hofmann and Sidney~A. Morris.
\newblock \emph{The structure of compact groups}, volume~25 of \emph{De Gruyter
  Studies in Mathematics}.
\newblock De Gruyter, Berlin, 2013.
\newblock Third edition.

\bibitem[Hornik(1991)]{hornik}
Kurt Hornik.
\newblock Approximation capabilities of multilayer feedforward networks.
\newblock \emph{Neural Networks}, 4\penalty0 (2):\penalty0 251--257, 1991.

\bibitem[Igelnik(2003)]{igelnik}
Boris Igelnik.
\newblock Use of the {K}olmogorov's superposition theorem and cubic splines for
  efficient neural-network modeling.
\newblock In \emph{Knowledge-Based Intelligent Information and Engineering
  Systems, 7th International Conference, {KES} 2003, Oxford, UK, September 3-5,
  2003, Proceedings, Part {I}}, pages 184--190, 2003.

\bibitem[Indyk(2004)]{indyk}
Piotr Indyk.
\newblock Nearest neighbors in high-dimensional spaces.
\newblock In \emph{Handbook of Discrete and Computational Geometry, Second
  Edition.}, pages 877--892. 2004.

\bibitem[Johnson and Lindenstrauss(1984)]{JL}
William~B. Johnson and Joram Lindenstrauss.
\newblock Extensions of {L}ipschitz mappings into a {H}ilbert space.
\newblock In \emph{Conference in modern analysis and probability ({N}ew
  {H}aven, {C}onn., 1982)}, volume~26 of \emph{Contemp. Math.}, pages 189--206.
  Amer. Math. Soc., Providence, RI, 1984.

\bibitem[Kahane(1975)]{kahane}
Jean-Pierre Kahane.
\newblock Sur le th\'{e}or\`eme de superposition de {K}olmogorov.
\newblock \emph{J. Approximation Theory}, 13:\penalty0 229--234, 1975.
\newblock Collection of articles dedicated to G. G. Lorentz on the occasion of
  his sixty-fifth birthday, III.

\bibitem[Kalajdzievski(2012)]{kalajdzievski}
Damjan Kalajdzievski.
\newblock Measurability aspects of the compactness theorem for sample
  compression schemes, 2012.
\newblock M.Sc. thesis, University of Ottawa, 64 pp., 2012.

\bibitem[Kearns and Vazirani(1994)]{kearns_vazirani}
Michael~J. Kearns and Umesh Vazirani.
\newblock \emph{An Introduction to Computational Learning Theory}.
\newblock The MIT Press, 1994.

\bibitem[Kechris(1995)]{kechris}
Alexander~S. Kechris.
\newblock \emph{Classical descriptive set theory}, volume 156 of \emph{Graduate
  Texts in Mathematics}.
\newblock Springer-Verlag, New York, 1995.

\bibitem[Klenke(2014)]{klenke}
Achim Klenke.
\newblock \emph{Probability theory. {A} comprehensive course}.
\newblock Universitext. Springer, London, second edition, 2014.

\bibitem[Kolmogorov(1963)]{kolmogorov}
A.~N. Kolmogorov.
\newblock On the representation of continuous functions of many variables by
  superposition of continuous functions of one variable and addition.
\newblock \emph{Amer. Math. Soc. Transl. (2)}, 28:\penalty0 55--59, 1963.

\bibitem[Koltchinskii and Panchenko(2000)]{koltchinskii_panchenko}
Vladimir Koltchinskii and Dmitriy Panchenko.
\newblock Rademacher processes and bounding the risk of function learning.
\newblock In \emph{High dimensional probability, {II} ({S}eattle, {WA}, 1999)},
  volume~47 of \emph{Progr. Probab.}, pages 443--457. Birkh\"{a}user Boston,
  Boston, MA, 2000.

\bibitem[Kumari(2018)]{kumari}
Sushma Kumari.
\newblock Topics in random matrices and statistical machine learning, 2018.
\newblock Ph.D. thesis, Kyoto University, 125 pages,
  \href{https://arxiv.org/abs/1807.09419}{arXiv:1807.09419}.

\bibitem[Kurkov{\'{a}}(1991)]{kurkova}
Vera Kurkov{\'{a}}.
\newblock Kolmogorov's theorem is relevant.
\newblock \emph{Neural Computation}, 3\penalty0 (4):\penalty0 617--622, 1991.

\bibitem[Kushilevitz et~al.(2000)Kushilevitz, Ostrovsky, and Rabani]{KOR}
Eyal Kushilevitz, Rafail Ostrovsky, and Yuval Rabani.
\newblock Efficient search for approximate nearest neighbor in high dimensional
  spaces.
\newblock \emph{{SIAM} J. Comput.}, 30\penalty0 (2):\penalty0 457--474, 2000.

\bibitem[Kuzmin and Warmuth(2007)]{KW}
Dima Kuzmin and Manfred~K. Warmuth.
\newblock Unlabeled compression schemes for maximum classes.
\newblock \emph{Journal of Machine Learning Research}, 8:\penalty0 2047--2081,
  2007.

\bibitem[Ledoux(2001)]{L}
Michel Ledoux.
\newblock \emph{The concentration of measure phenomenon}, volume~89 of
  \emph{Mathematical Surveys and Monographs}.
\newblock American Mathematical Society, Providence, RI, 2001.

\bibitem[L\'evy(1922)]{levy}
Paul L\'evy.
\newblock \emph{Le\c cons d'analyse fonctionnelle}.
\newblock Gauthier-Villars, Paris, 1922.

\bibitem[Littlestone and Warmuth(1986)]{LW}
Nick Littlestone and Manfred Warmuth.
\newblock
  \href{https://users.soe.ucsc.edu/~manfred/pubs/lrnk-olivier.pdf}{Relating
  Data Compression and Learnability}.
\newblock Unpublished manuscript, June 1986.

\bibitem[Loeb(1975)]{loeb}
Peter~A. Loeb.
\newblock Conversion from nonstandard to standard measure spaces and
  applications in probability theory.
\newblock \emph{Trans. Amer. Math. Soc.}, 211:\penalty0 113--122, 1975.

\bibitem[Lorentz(1976)]{lorentz}
G.~G. Lorentz.
\newblock The 13th problem of {H}ilbert.
\newblock In \emph{Mathematical developments arising from {H}ilbert problems
  ({P}roc. {S}ympos. {P}ure {M}ath., {V}ol. {XXVIII}, {N}orthern {I}llinois
  {U}niv., {D}e {K}alb, {I}ll., 1974)}, pages 419--430. Amer. Math. Soc.,
  Providence, R.I., 1976.

\bibitem[Maindonald(2008)]{M}
J.H. Maindonald.
\newblock Using {R} for data analysis and graphics. {I}ntroduction, code and
  commentary, 2008.
\newblock \href{http://cran.r-project.org/doc/contrib/usingR.pdf}
  {http://cran.r-project.org/doc/contrib/usingR.pdf}.

\bibitem[Mansuy(2009)]{mansuy}
Roger Mansuy.
\newblock \href{http://www.jehps.net/juin2009/Mansuy.pdf} {The Origins of the
  Word ``Martingale''}.
\newblock \emph{Journal \'Electronique d'Histoire des Probabilit\'es e de la
  Statistique}, 5\penalty0 (1), 6 2009.
\newblock 10 pp.

\bibitem[Matou\v{s}ek(2002)]{matousekBook}
Ji\v{r}\'{\i} Matou\v{s}ek.
\newblock \emph{Lectures on discrete geometry}, volume 212 of \emph{Graduate
  Texts in Mathematics}.
\newblock Springer-Verlag, New York, 2002.

\bibitem[Matou\v{s}ek(2008)]{matousek}
Ji\v{r}\'{\i} Matou\v{s}ek.
\newblock On variants of the {J}ohnson-{L}indenstrauss lemma.
\newblock \emph{Random Structures Algorithms}, 33\penalty0 (2):\penalty0
  142--156, 2008.

\bibitem[Mendelson(2002)]{mendelson}
Shahar Mendelson.
\newblock A few notes on statistical learning theory.
\newblock In Shahar Mendelson and Alexander~J. Smola, editors, \emph{Machine
  Learning Summer School}, volume 2600 of \emph{Lecture Notes in Computer
  Science}, pages 1--40. Springer, 2002.

\bibitem[Milman(1967)]{M67}
V.~D. Milman.
\newblock The infinite dimensional geometry of the unit sphere of a {B}anach
  space.
\newblock \emph{Dokl. Akad. Nauk SSSR}, 177:\penalty0 514--517, 1967.

\bibitem[Milman and Schechtman(1986)]{MS}
Vitali~D. Milman and Gideon Schechtman.
\newblock \emph{Asymptotic theory of finite-dimensional normed spaces}, volume
  1200 of \emph{Lecture Notes in Mathematics}.
\newblock Springer-Verlag, Berlin, 1986.
\newblock With an appendix by M. Gromov.

\bibitem[Moran and Warmuth(2016)]{MW}
Shay Moran and Manfred~K. Warmuth.
\newblock Labeled compression schemes for extremal classes.
\newblock In \emph{Algorithmic Learning Theory - 27th International Conference,
  {ALT} 2016, Bari, Italy, October 19-21, 2016, Proceedings}, pages 34--49,
  2016.

\bibitem[Moran and Yehudayoff(2016)]{MY}
Shay Moran and Amir Yehudayoff.
\newblock Sample compression schemes for {VC} classes.
\newblock \emph{J. {ACM}}, 63\penalty0 (3):\penalty0 21:1--21:10, 2016.

\bibitem[Mumford(2000)]{mumford}
David Mumford.
\newblock The dawning of the age of stochasticity.
\newblock In \emph{Mathematics: frontiers and perspectives}, pages 197--218.
  Amer. Math. Soc., Providence, RI, 2000.

\bibitem[Naor(2018)]{naor}
Assaf Naor.
\newblock Metric dimension reduction: A snapshot of the ribe program, 2018.
\newblock \href{https://arxiv.org/abs/1809.02376}{arXiv:1809.02376} {A}lso
  Naor's 2018 plenary ICM lecture,
  \href{https://www.youtube.com/watch?v=MjHCB3lEBcU}
  {https://www.youtube.com/watch?v=MjHCB3lEBcU}.

\bibitem[Ostrand(1965)]{ostrand}
Phillip~A. Ostrand.
\newblock A conjecture of {J}. {N}agata on dimension and metrization.
\newblock \emph{Bull. Amer. Math. Soc.}, 71:\penalty0 623--625, 1965.

\bibitem[Owen(2010)]{O}
W.J. Owen.
\newblock The {R} guide.
\newblock \href{http://cran.r-project.org/doc/contrib/Owen-TheRGuide.pdf}
  {http://cran.r-project.org/doc/contrib/Owen-TheRGuide.pdf}, 2010.

\bibitem[Pajor(1985)]{pajor}
Alain Pajor.
\newblock \emph{Sous-espaces {$l^n_1$} des espaces de {B}anach}, volume~16 of
  \emph{Travaux en Cours}.
\newblock Hermann, Paris, 1985.

\bibitem[Patrascu and Thorup(2009)]{PT}
Mihai Patrascu and Mikkel Thorup.
\newblock Higher lower bounds for near-neighbor and further rich problems.
\newblock \emph{{SIAM} J. Comput.}, 39\penalty0 (2):\penalty0 730--741, 2009.

\bibitem[Pestov(2006)]{Pes06}
Vladimir Pestov.
\newblock \emph{Dynamics of infinite-dimensional groups}, volume~40 of
  \emph{University Lecture Series}.
\newblock American Mathematical Society, Providence, RI, 2006.
\newblock The Ramsey-Dvoretzky-Milman phenomenon, Revised edition of {\it
  Dynamics of infinite-dimensional groups and Ramsey-type phenomena} [IMPA, Rio
  de Janeiro, 2005].

\bibitem[Pestov(2007)]{pestov:07}
Vladimir Pestov.
\newblock Intrinsic dimension of a dataset: what properties does one expect?
\newblock In \emph{Proceedings of the International Joint Conference on Neural
  Networks, {IJCNN} 2007, Celebrating 20 years of neural networks, Orlando,
  Florida, USA, August 12-17, 2007}, pages 2959--2964, 2007.

\bibitem[Pestov(2008)]{pestov08c}
Vladimir Pestov.
\newblock An axiomatic approach to intrinsic dimension of a dataset.
\newblock \emph{Neural Networks}, 21\penalty0 (2-3):\penalty0 204--213, 2008.

\bibitem[Pestov(2010)]{pestov2010sbrn}
Vladimir Pestov.
\newblock A note on sample complexity of learning binary output neural networks
  under fixed input distributions.
\newblock In Teresa~Bernarda Ludermir, Karla Figueiredo, and Carlos~E. Thomaz,
  editors, \emph{SBRN}, pages 7--12. IEEE Computer Society, 2010.

\bibitem[Pestov(2011)]{pestov2010}
Vladimir Pestov.
\newblock {PAC} learnability versus {VC} dimension: {A} footnote to a basic
  result of statistical learning.
\newblock In \emph{The 2011 International Joint Conference on Neural Networks,
  {IJCNN} 2011, San Jose, California, USA, July 31 - August 5, 2011}, pages
  1141--1145, 2011.

\bibitem[Pestov(2012)]{pestov2012}
Vladimir Pestov.
\newblock Indexability, concentration, and {VC} theory.
\newblock \emph{J. Discrete Algorithms}, 13:\penalty0 2--18, 2012.

\bibitem[Pestov(2013{\natexlab{a}})]{pestov2013}
Vladimir Pestov.
\newblock Is the {$k$}-{NN} classifier in high dimensions affected by the curse
  of dimensionality?
\newblock \emph{Comput. Math. Appl.}, 65\penalty0 (10):\penalty0 1427--1437,
  2013{\natexlab{a}}.

\bibitem[Pestov(2013{\natexlab{b}})]{pestov:13}
Vladimir Pestov.
\newblock Lower bounds on performance of metric tree indexing schemes for exact
  similarity search in high dimensions.
\newblock \emph{Algorithmica}, 66\penalty0 (2):\penalty0 310--328,
  2013{\natexlab{b}}.

\bibitem[Pestov(2013{\natexlab{c}})]{pestov:13tcs}
Vladimir Pestov.
\newblock P{AC} learnability under non-atomic measures: a problem by
  {V}idyasagar.
\newblock \emph{Theoret. Comput. Sci.}, 473:\penalty0 29--45,
  2013{\natexlab{c}}.

\bibitem[Pestov and Stojmirovi\'{c}(2006)]{PeSt06}
Vladimir Pestov and Aleksandar Stojmirovi\'{c}.
\newblock Indexing schemes for similarity search: an illustrated paradigm.
\newblock \emph{Fund. Inform.}, 70\penalty0 (4):\penalty0 367--385, 2006.

\bibitem[Pisier(1989)]{pisier}
Gilles Pisier.
\newblock \emph{The volume of convex bodies and {B}anach space geometry},
  volume~94 of \emph{Cambridge Tracts in Mathematics}.
\newblock Cambridge University Press, Cambridge, 1989.

\bibitem[Preiss(1979)]{preiss1}
D.~Preiss.
\newblock Invalid {V}itali theorems.
\newblock In \emph{Proc. Seventh Winter School in Abstract Analysis (Prague,
  1979)}, pages 58--60. Institute of Mathematics of the Academy of Sciences of
  the Czech Republic, 1979.

\bibitem[Preiss(1981)]{preiss}
D.~Preiss.
\newblock Gaussian measures and the density theorem.
\newblock \emph{Comment. Math. Univ. Carolin.}, 22\penalty0 (1):\penalty0
  181--193, 1981.

\bibitem[Preiss(1983)]{preiss83}
D.~Preiss.
\newblock Dimension of metrics and differentiation of measures.
\newblock In \emph{General topology and its relations to modern analysis and
  algebra, {V} ({P}rague, 1981)}, volume~3 of \emph{Sigma Ser. Pure Math.},
  pages 565--568. Heldermann, Berlin, 1983.

\bibitem[Samet(2006)]{samet}
Hanan Samet.
\newblock \emph{Foundations of Multidimensional and Metric Data Structures (The
  Morgan Kaufmann Series in Computer Graphics)}.
\newblock Morgan Kaufmann, aug 2006.

\bibitem[Santini(2001)]{santini}
Simone Santini.
\newblock \emph{Exploratory Image Databases: Content-Based Retrieval
  (Communications, Networking and Multimedia)}.
\newblock Academic Press, sep 2001.

\bibitem[Sauer(1972)]{sauer}
N.~Sauer.
\newblock On the density of families of sets.
\newblock \emph{J. Combinatorial Theory Ser. A}, 13:\penalty0 145--147, 1972.

\bibitem[Scarvalone(2008)]{scarvalone}
Mira Scarvalone.
\newblock
  \href{http://www.math.uchicago.edu/~may/VIGRE/VIGRE2008/REUPapers/Scarvalone.pdf}
  {Game theory and the Minimax theorem}.
\newblock University of Chicago VIGRE undergraduate project, 5 pp., 2008.

\bibitem[Schechtman(2003)]{schechtman}
Gideon Schechtman.
\newblock Concentration results and applications.
\newblock In \emph{Handbook of the geometry of {B}anach spaces, {V}ol. 2},
  pages 1603--1634. North-Holland, Amsterdam, 2003.

\bibitem[Shalev-Shwartz and Ben-David(2014)]{shalev-shwartz_ben-david}
Shai Shalev-Shwartz and Shai Ben-David.
\newblock \emph{Understanding Machine Learning: From Theory to Algorithms}.
\newblock Cambridge University Press, 2014.

\bibitem[Shannon(1948)]{shannon}
C.~E. Shannon.
\newblock A mathematical theory of communication.
\newblock \emph{Bell System Tech. J.}, 27:\penalty0 379--423, 623--656, 1948.

\bibitem[Shelah(1972)]{shelah}
Saharon Shelah.
\newblock A combinatorial problem; stability and order for models and theories
  in infinitary languages.
\newblock \emph{Pacific J. Math.}, 41:\penalty0 247--261, 1972.

\bibitem[Shioya(2016)]{shioya}
Takashi Shioya.
\newblock \emph{Metric measure geometry}, volume~25 of \emph{IRMA Lectures in
  Mathematics and Theoretical Physics}.
\newblock EMS Publishing House, Z\"{u}rich, 2016.
\newblock Gromov's theory of convergence and concentration of metrics and
  measures.

\bibitem[Song et~al.(2008)Song, Takakura, and Kwon]{STK}
Jungsuk Song, Hiroki Takakura, and Yongjin Kwon.
\newblock A generalized feature extraction scheme to detect 0-day attacks via
  {IDS} alerts.
\newblock In \emph{2008 International Symposium on Applications and the
  Internet}. {IEEE}, July 2008.

\bibitem[Sontag(1992)]{sontag}
Eduardo~D. Sontag.
\newblock Feedforward nets for interpolation and classification.
\newblock \emph{J. Comput. Syst. Sci.}, 45\penalty0 (1):\penalty0 20--48, 1992.

\bibitem[Sternfeld(1985)]{sternfeld}
Y.~Sternfeld.
\newblock Dimension, superposition of functions and separation of points, in
  compact metric spaces.
\newblock \emph{Israel J. Math.}, 50\penalty0 (1-2):\penalty0 13--53, 1985.

\bibitem[Stone(1977)]{stone:77}
Charles~J. Stone.
\newblock Consistent nonparametric regression.
\newblock \emph{Ann. Statist.}, 5\penalty0 (4):\penalty0 595--645, 1977.

\bibitem[Talagrand(1984)]{talagrand84}
Michel Talagrand.
\newblock Solution du probl\`eme de {G}livenko-{C}antelli.
\newblock \emph{C. R. Acad. Sci. Paris S\'{e}r. I Math.}, 298\penalty0
  (9):\penalty0 213--216, 1984.

\bibitem[Talagrand(1996)]{talagrand96}
Michel Talagrand.
\newblock The {G}livenko-{C}antelli problem, ten years later.
\newblock \emph{J. Theoret. Probab.}, 9\penalty0 (2):\penalty0 371--384, 1996.

\bibitem[Torgo(2010)]{torgo}
Luis Torgo.
\newblock \emph{Data Mining with R: Learning with Case Studies (Chapman \&
  Hall/CRC Data Mining and Knowledge Discovery Series)}.
\newblock Chapman and Hall/CRC, 2010.

\bibitem[Valiant(1984)]{valiant84learnable}
Les Valiant.
\newblock {A Theory of the Learnable}.
\newblock \emph{Communications of the ACM}, 27\penalty0 (11):\penalty0
  1134--1142, 1984.

\bibitem[Vapnik and \v{C}ervonenkis(1971)]{VC:71}
V.~N. Vapnik and A.~Ja. \v{C}ervonenkis.
\newblock The uniform convergence of frequencies of the appearance of events to
  their probabilities.
\newblock \emph{Teor. Verojatnost. i Primenen.}, 16:\penalty0 264--279, 1971.

\bibitem[Vapnik(1998)]{vapnik}
Vladimir~N. Vapnik.
\newblock \emph{Statistical learning theory}.
\newblock Adaptive and Learning Systems for Signal Processing, Communications,
  and Control. John Wiley \& Sons, Inc., New York, 1998.
\newblock A Wiley-Interscience Publication.

\bibitem[Vapnik(2000)]{vapnik2}
Vladimir~N. Vapnik.
\newblock \emph{The nature of statistical learning theory}.
\newblock Statistics for Engineering and Information Science. Springer-Verlag,
  New York, second edition, 2000.

\bibitem[Vempala(2004)]{vempala}
Santosh~S. Vempala.
\newblock \emph{The random projection method}, volume~65 of \emph{DIMACS Series
  in Discrete Mathematics and Theoretical Computer Science}.
\newblock American Mathematical Society, Providence, RI, 2004.

\bibitem[Vershynin(2018)]{vershynin}
Roman Vershynin.
\newblock \emph{High-dimensional probability}, volume~47 of \emph{Cambridge
  Series in Statistical and Probabilistic Mathematics}.
\newblock Cambridge University Press, Cambridge, 2018.
\newblock An introduction with applications in data science, With a foreword by
  Sara van de Geer.

\bibitem[Vidyasagar(2003)]{vidyasagar}
M.~Vidyasagar.
\newblock \emph{Learning and generalization. With applications to neural
  networks}.
\newblock Communications and Control Engineering Series. Springer-Verlag
  London, Ltd., London, second edition, 2003.

\bibitem[Voiculescu et~al.(1992)Voiculescu, Dykema, and Nica]{VDN}
D.~V. Voiculescu, K.~J. Dykema, and A.~Nica.
\newblock \emph{Free random variables}, volume~1 of \emph{CRM Monograph
  Series}.
\newblock American Mathematical Society, Providence, RI, 1992.

\bibitem[von Luxburg et~al.(2003/04)von Luxburg, Bousquet, and
  Sch\"{o}lkopf]{vLBS}
Ulrike von Luxburg, Olivier Bousquet, and Bernhard Sch\"{o}lkopf.
\newblock A compression approach to support vector model selection.
\newblock \emph{J. Mach. Learn. Res.}, 5:\penalty0 293--323, 2003/04.

\bibitem[Weinberger and Saul(2009)]{weinberger_saul:09}
Kilian~Q. Weinberger and Lawrence~K. Saul.
\newblock Distance metric learning for large margin nearest neighbor
  classification.
\newblock \emph{Journal of Machine Learning Research}, 10:\penalty0 207--244,
  2009.

\bibitem[Weston(1957)]{weston}
J.~D. Weston.
\newblock A short proof of {Z}orn's lemma.
\newblock \emph{Arch. Math. (Basel)}, 8:\penalty0 279, 1957.

\bibitem[Yandell(2002)]{yandell}
Benjamin~H. Yandell.
\newblock \emph{The honors class}.
\newblock A K Peters, Ltd., Natick, MA, 2002.
\newblock Hilbert's problems and their solvers.

\bibitem[Yianilos(1993)]{Yan}
Peter~N. Yianilos.
\newblock Data structures and algorithms for nearest neighbor search in general
  metric spaces.
\newblock In \emph{Proceedings of the {F}ourth {A}nnual {ACM}-{SIAM}
  {S}ymposium on {D}iscrete {A}lgorithms ({A}ustin, {TX}, 1993)}, pages
  311--321. ACM, New York, 1993.

\bibitem[Zezula et~al.(2006)Zezula, Amato, Dohnal, and Batko]{zezula:06}
Pavel Zezula, Giuseppe Amato, Vlastislav Dohnal, and Michal Batko.
\newblock \emph{Similarity Search: The Metric Space Approach}.
\newblock Springer Science $+$ Business Media, New York, 2006.

\bibitem[Zhao(1987)]{zhao:87}
L.~C. Zhao.
\newblock Exponential bounds of mean error for the nearest neighbor estimates
  of regression functions.
\newblock \emph{J. Multivariate Anal.}, 21\penalty0 (1):\penalty0 168--178,
  1987.

\bibitem[Zhou(2012)]{zhou_ensemble}
Zhi-Hua Zhou.
\newblock \emph{Ensemble Methods: Foundations and Algorithms}.
\newblock Chapman \& Hall/Crc Machine Learning \& Pattern Recognition. Chapman
  and Hall/CRC, 2012.

\end{thebibliography}


\printindex

\end{document}